\documentclass[11pt,oneside,openany]{amsbook}
\pdftrailerid{}
\usepackage[T1]{fontenc}
\usepackage{lmodern}
\usepackage[letterpaper,margin=1in]{geometry}
\usepackage{amsmath,amssymb,amsthm,mathtools,bm,mathrsfs}
\usepackage{booktabs,array,longtable,tabularx,graphicx,microtype,enumitem}
\usepackage[dvipsnames]{xcolor}
\usepackage{etoolbox,placeins}
\makeatletter
\def\@captionfont{\normalfont\normalsize}
\patchcmd{\LT@makecaption}{\reset@font}{\reset@font\normalsize}{}{\PackageError{monograph}{Longtable caption font hook failed}{}}

\newcommand{\keepheadingwithtext}{\if@nobreak\else
  \par
  \begingroup
  \@tempdima=8\baselineskip
  \vskip 0pt plus\@tempdima
  \penalty-100
  \vskip 0pt plus-\@tempdima
  \vskip\@tempdima
  \penalty9999
  \vskip-\@tempdima
  \vskip0pt
  \endgroup
  \fi}
\AtBeginDocument{\pretocmd{\@startsection}{\keepheadingwithtext}{}{\PackageError{monograph}{Section hook failed}{}}}
\newcommand{\appendixcaptionlist}[1]{\begingroup
  \makeatletter
  \setTrue{#1}\parskip\z@skip
  \def\l@figure{\@tocline{0}{3pt plus2pt}{0pt}{3pc}{}}\let\l@table\l@figure
  \@input{\jobname.#1}\if@filesw
    \@xp\newwrite\csname tf@#1\endcsname
    \immediate\@xp\openout\csname tf@#1\endcsname\jobname.#1\relax
  \fi
  \global\@nobreakfalse
  \endgroup}
\makeatother
\usepackage{tikz}
\usetikzlibrary{positioning,arrows.meta}
\usepackage[hyphens]{url}
\usepackage[colorlinks=true,allcolors=MidnightBlue,bookmarksdepth=2]{hyperref}
\makeatletter
\patchcmd{\NR@part}{\protect\enspace\protect\noindent}
  {\texorpdfstring{\protect\enspace\protect\noindent}{\space}}{}{\PackageError{monograph}{Numbered part bookmark hook failed}{}}
\patchcmd{\NR@spart}{\protect\noindent}{\texorpdfstring{\protect\noindent}{}}{}{\PackageError{monograph}{Unnumbered part bookmark hook failed}{}}
\makeatother

\hypersetup{pdftitle={Training and Inference Dynamics of PLDR-LLMs: Row-Map Collapse, Renormalization, and Predictive Reduction},pdfauthor={Burc Gokden},pdfsubject={Single-pass training, row geometry, conditional laws and inference}}
\newtheorem{theorem}{Theorem}[chapter]
\newtheorem{proposition}[theorem]{Proposition}
\newtheorem{lemma}[theorem]{Lemma}
\newtheorem{corollary}[theorem]{Corollary}

\theoremstyle{definition}
\newtheorem{definition}[theorem]{Definition}

\theoremstyle{remark}
\newtheorem{remark}[theorem]{Remark}
\numberwithin{equation}{chapter}
\numberwithin{figure}{chapter}
\numberwithin{table}{chapter}

\newcommand{\R}{\mathbb R}
\newcommand{\N}{\mathbb N}
\newcommand{\E}{\mathbb E}
\newcommand{\Prob}{\mathbb P}
\newcommand{\D}{\mathcal D}
\newcommand{\RG}{\mathcal R}
\newcommand{\Cov}{\operatorname{Cov}}
\newcommand{\Var}{\operatorname{Var}}
\newcommand{\KL}{\operatorname{KL}}
\newcommand{\tr}{\operatorname{tr}}
\newcommand{\diag}{\operatorname{diag}}
\newcommand{\softmax}{\operatorname{softmax}}
\newcommand{\clip}{\operatorname{clip}}
\newcommand{\LN}{\operatorname{LN}}
\newcommand{\rank}{\operatorname{rank}}
\newcommand{\range}{\operatorname{range}}
\newcommand{\norm}[1]{\left\lVert #1\right\rVert}
\newcommand{\abs}[1]{\left\lvert #1\right\rvert}
\newcommand{\ip}[2]{\left\langle #1,#2\right\rangle}
\newcommand{\op}{\mathrm{op}}
\newcommand{\Fro}{\mathrm F}
\newcommand{\rowq}{\mathsf Q}
\newcommand{\edge}{\mathfrak a}
\newcommand{\idEdge}{\mathfrak e}
\newcommand{\G}{G_{\mathrm{LM}}}
\newcommand{\Alm}{A_{\mathrm{LM}}}
\newcommand{\Hess}{\mathcal H}
\newcommand{\codepath}[1]{\path{#1}}
\newcommand{\hashpair}[2]{\texttt{\scriptsize #1}\allowbreak\texttt{\scriptsize #2}}

\newcommand{\RowSourceObservableStateDimension}{61,868,742}

\newcommand{\RowSourceObservableHeldoutPasses}{30}
\newcommand{\RowSourceObservableHeldoutTotal}{54}
\newcommand{\RowSourceObservableAttenuating}{7}

\newcommand{\RowSourceObservablePlaceboFailures}{64}
\newcommand{\RowSourceObservableMaximumGain}{200.8}

\newcommand{\RowSourceBalanceHeldoutTotal}{54}
\newcommand{\RowSourceBalanceHeldoutEvaluable}{54}
\newcommand{\RowSourceBalanceCancellationDominant}{6}
\newcommand{\RowSourceBalanceRouteChanged}{17}
\newcommand{\RowSourceBalanceStateDominant}{42}
\newcommand{\RowSourceBalanceObservationDominant}{12}
\newcommand{\RowSourceBalanceCampaignOutcome}{\textsc{refuted}}
\newcommand{\RowSourceBalanceGateTheoryPasses}{9}

\newcommand{\RowSourceRetainedDenseEdges}{1{,}056}
\newcommand{\RowSourceRetainedDenseMapEdges}{304{,}128}
\newcommand{\RowSourceRetainedDenseReopeningEdges}{131{,}748}
\newcommand{\RowSourceRetainedDenseReopeningPercent}{43.32\%}

\newcommand{\RowSourceRetainedEndpointIncrements}{864}
\newcommand{\RowSourceRetainedEndpointReopenings}{474}
\newcommand{\RowSourceRetainedEndpointReopeningPercent}{54.86\%}
\newcommand{\RowSourceRetainedEndpointIdentityResidual}{\ensuremath{1.5\times 10^{-12}}}
\newcommand{\RowSourceRetainedNonevaluableRecords}{46}
\newcommand{\RowSourceRetainedCancellationTotal}{6}
\newcommand{\RowSourceRetainedCancellationLargestPath}{5}
\newcommand{\RowSourceRetainedCancellationLayerTwo}{4}

\newcommand{\RowSourcePrecisionReplayArmMapCount}{5184}
\newcommand{\RowSourcePrecisionReplayGateOnlyExcluded}{616}
\newcommand{\RowSourcePrecisionReplayLockedEvaluable}{1112}
\newcommand{\RowSourcePrecisionReplayPrimaryEvaluable}{1728}
\newcommand{\RowSourcePrecisionReplayFailAbsMin}{2.208\times10^{-8}}
\newcommand{\RowSourcePrecisionReplayFailAbsMax}{1.488\times10^{-6}}
\newcommand{\RowSourcePrecisionReplayNativeClosure}{8.882\times10^{-16}}
\newcommand{\RowSourcePrecisionReplayDefectMatch}{8.882\times10^{-16}}

\newcommand{\RowSourcePrecisionReplayWorkResidual}{4.062\times10^{-13}}
\newcommand{\RowSourcePrecisionReplayOldWorkToleranceMinimum}{2.000\times10^{-5}}

\newcommand{\RowSourceReplicationEffectFloor}{1.000\times10^{-12}}
\newcommand{\RowSourceReplicationCampaignOutcome}{\textsc{refuted}}
\newcommand{\RowSourceReplicationHeldoutEvaluable}{1632}
\newcommand{\RowSourceReplicationHeldoutPlanned}{1728}
\newcommand{\RowSourceReplicationHeldoutPositive}{662}
\newcommand{\RowSourceReplicationHeldoutNegative}{751}
\newcommand{\RowSourceReplicationNaturalReopening}{761}
\newcommand{\RowSourceReplicationControlReopening}{797}

\newcommand{\RowSourceReplicationPeakGiB}{5.35}

\newcommand{\RowSourceMechanismEvaluable}{2304}

\newcommand{\RowSourceMechanismContrastResidual}{1.480\times10^{-13}}
\newcommand{\RowSourceMechanismFourSourceResidual}{7.598\times10^{-16}}
\newcommand{\RowSourceMechanismStateSignAgreement}{96.8\%}
\newcommand{\RowSourceMechanismStateDominant}{2115}

\newcommand{\RowSourceMechanismCancellationPercentile}{4.545}
\newcommand{\RowSourceMechanismTerminalDefectDominant}{356}

\newcommand{\RowSourceRadialArmEligible}{4{,}608}
\newcommand{\RowSourceRadialArmContract}{2{,}452}
\newcommand{\RowSourceRadialArmReopen}{2{,}156}
\newcommand{\RowSourceRadialFailureCount}{2{,}015}
\newcommand{\RowSourceTangentExcessCount}{141}
\newcommand{\RowSourceRadialPairedEligible}{2{,}304}
\newcommand{\RowSourceRadialSignAgreement}{2{,}236}
\newcommand{\RowSourceRadialSignComparisons}{2{,}304}
\newcommand{\RowSourceRadialSignAgreementPercent}{97.05\%}
\newcommand{\RowSourceRadialLinearDominant}{2{,}115}
\newcommand{\RowSourceRadialChargeDominant}{0}
\newcommand{\RowSourceTangentInteractionDominant}{24}
\newcommand{\RowSourceTangentChargeDominant}{165}
\newcommand{\RowSourceRadialGainResidual}{\ensuremath{2.62\times 10^{-14}}}
\newcommand{\RowSourceRadialPairedResidual}{\ensuremath{1.67\times 10^{-15}}}

\newcommand{\RowSourceRadialProductResidual}{\ensuremath{2.17\times 10^{-15}}}

\newcommand{\RowSourceRadialHoldoutOutcome}{insufficient}
\newcommand{\RowSourceRadialHoldoutPoweredCells}{17}
\newcommand{\RowSourceRadialHoldoutPlannedCells}{18}
\newcommand{\RowSourceRadialHoldoutEligible}{1{,}632}
\newcommand{\RowSourceRadialHoldoutAgreements}{1{,}579}
\newcommand{\RowSourceRadialHoldoutAgreementPercent}{96.75\%}
\newcommand{\RowSourceRadialHoldoutGainResidual}{\ensuremath{2.53\times 10^{-14}}}
\newcommand{\RowSourceRadialHoldoutPairedResidual}{\ensuremath{1.28\times 10^{-15}}}

\newcommand{\RowSourceObserverExactZeroMaps}{0}
\newcommand{\RowSourceObserverAllSourceMaps}{5,184}
\newcommand{\RowSourceObserverTargetMinimumEnergy}{3.846675\times10^{-15}}
\newcommand{\RowSourceObserverTargetMaximumEnergy}{6.287699\times10^{-15}}
\newcommand{\RowSourceObserverTargetFloorFactor}{159.04}
\newcommand{\RowSourceObserverTargetMaximumCoordinate}{1.839362\times10^{-8}}
\newcommand{\RowSourceObserverRetainedCensoredMaps}{288}
\newcommand{\RowSourceObserverRetainedResolvedMaps}{2,304}
\newcommand{\RowSourceObserverHoldoutCensoredMaps}{288}
\newcommand{\RowSourceObserverHoldoutResolvedMaps}{2,304}
\newcommand{\RowSourceBlockWindowCount}{18}
\newcommand{\RowSourceBlockThirtyTwoComparisons}{143,424}
\newcommand{\RowSourceBlockThirtyTwoContracting}{85,061}
\newcommand{\RowSourceBlockThirtyTwoExpanding}{58,363}

\newcommand{\RowSourceBlockThirtyTwoResolvedComparisons}{143,136}
\newcommand{\RowSourceBlockThirtyTwoResolvedExpanding}{58,166}

\newcommand{\ModelSourceNestedChainError}{4.274\times10^{-15}}
\newcommand{\ModelSourceNestedDirectError}{8.882\times10^{-16}}

\newcommand{\ModelSourceContextSmallChainError}{3.414\times10^{-15}}

\newcommand{\ModelSourceContextSmallForwards}{96}
\newcommand{\ModelSourceContextSmallQualification}{2}
\newcommand{\ModelSourceContextSmallWorkerSeconds}{597.6}
\newcommand{\ModelSourceContextSmallLogitBytes}{98,304,000}

\newcommand{\ModelSourceContextLargeChainError}{2.956\times10^{-15}}
\newcommand{\ModelSourceContextLargeDirectError}{2.220\times10^{-16}}
\newcommand{\ModelSourceContextLargeCells}{20}
\newcommand{\ModelSourceContextLargeChains}{6,144}
\newcommand{\ModelSourceContextLargeForwards}{192}
\newcommand{\ModelSourceContextLargeQualification}{2}
\newcommand{\ModelSourceContextLargeWorkerSeconds}{613.1}
\newcommand{\ModelSourceContextLargeLogitBytes}{196,608,000}
\newcommand{\ModelSourceContextLargeRetained}{96.35\text{--}97.14}
\newcommand{\ModelSourceContextLargeRMS}{0.175\text{--}0.193}
\newcommand{\ModelSourceContextLargeNativeVariance}{0.07739\text{--}0.09877}
\newcommand{\ModelSourceContextLargeResidualVariance}{0.00238\text{--}0.00353}
\newcommand{\ModelSourceCategoricalDictionaryBytes}{512,000}

\newcommand{\ModelSourceContextRiskRetained}{96.07\text{--}96.82}
\newcommand{\ModelSourceContextRiskRMS}{0.182\text{--}0.197}
\newcommand{\ModelSourceContextRiskWeightedMass}{4.86\text{--}11.10}
\newcommand{\ModelSourceContextRiskBound}{0.530\text{--}0.621}
\newcommand{\ModelSourceContextRiskWorkerSeconds}{527.6}
\newcommand{\ModelSourceContextRiskForwards}{384}
\newcommand{\ModelSourceContextRiskQualification}{2}
\newcommand{\ModelSourceContextRiskChains}{12,288}
\newcommand{\ModelSourceContextRiskDirectError}{4.441\times10^{-16}}
\newcommand{\ModelSourceContextRiskChainError}{4.441\times10^{-15}}
\newcommand{\ModelSourceContextRiskLogitBytes}{393,216,000}

\newcommand{\ModelSourceHOne}{0.5017}
\newcommand{\ModelSourceHLoOne}{0.4897}
\newcommand{\ModelSourceHHiOne}{0.5074}
\newcommand{\ModelSourceResponseOne}{0.365}
\newcommand{\ModelSourceDiagonalOne}{33.85}
\newcommand{\ModelSourceOffDiagonalOne}{0.351}
\newcommand{\ModelSourceSegmentOne}{1.611}
\newcommand{\ModelSourceSegmentLoOne}{1.575}
\newcommand{\ModelSourceSegmentHiOne}{1.655}
\newcommand{\ModelSourcePermutedOne}{0.991}
\newcommand{\ModelSourceFrozenKLOne}{5.45\times10^{-7}}

\newcommand{\ModelSourceNativeNLLOne}{3.593}
\newcommand{\ModelSourceADMaxOne}{2.25\times10^{-3}}
\newcommand{\ModelSourceHTwo}{0.4968}
\newcommand{\ModelSourceHLoTwo}{0.4898}
\newcommand{\ModelSourceHHiTwo}{0.5073}
\newcommand{\ModelSourceResponseTwo}{0.370}
\newcommand{\ModelSourceDiagonalTwo}{32.66}
\newcommand{\ModelSourceOffDiagonalTwo}{0.297}
\newcommand{\ModelSourceSegmentTwo}{1.704}
\newcommand{\ModelSourceSegmentLoTwo}{1.624}
\newcommand{\ModelSourceSegmentHiTwo}{1.756}
\newcommand{\ModelSourcePermutedTwo}{0.992}
\newcommand{\ModelSourceFrozenKLTwo}{3.21\times10^{-12}}

\newcommand{\ModelSourceNativeNLLTwo}{3.524}
\newcommand{\ModelSourceADMaxTwo}{1.35\times10^{-3}}

\newcommand{\ModelSourceCriticalityScanRuns}{140}
\newcommand{\ModelSourceCriticalityFineRuns}{48}

\newcommand{\ModelSourceCriticalityWiderRuns}{4}

\newcommand{\ModelSourceCriticalityReplayParents}{4}
\newcommand{\ModelSourceCriticalityReplayBranches}{8}
\newcommand{\ModelSourceCriticalityInitializationModels}{60}
\newcommand{\ModelSourceCriticalityPrecisionCases}{4}
\newcommand{\ModelSourceCriticalityPrecisionContexts}{30}
\newcommand{\ModelSourceCriticalityMaximumMemory}{16.32}
\newcommand{\ModelSourceCriticalityInitialEntropyMinimum}{0.0288}
\newcommand{\ModelSourceCriticalityInitialEntropyMaximum}{0.0297}
\newcommand{\ModelSourceCriticalityInitialRowMinimum}{$2.64\times10^{-5}$}
\newcommand{\ModelSourceCriticalityInitialRowMaximum}{$3.06\times10^{-5}$}
\newcommand{\ModelSourceCriticalityIntermediateRowSmall}{$1.04\times10^{-2}$}
\newcommand{\ModelSourceCriticalityIntermediateRowLarge}{$5.53\times10^{-3}$}
\newcommand{\ModelSourceCriticalityIntermediateEnhancementSmall}{1.49}
\newcommand{\ModelSourceCriticalityIntermediateEnhancementLarge}{8.98}
\newcommand{\ModelSourceCriticalitySaturatedComparisonPercent}{97.7}
\newcommand{\ModelSourceCriticalitySaturatedComparisonContextKL}{0.534}
\newcommand{\ModelSourceCriticalityCoarseTailRow}{$1.66\times10^{-2}$}
\newcommand{\ModelSourceCriticalityCoarseTailOmitted}{$1.94\times10^{-9}$}
\newcommand{\ModelSourceCriticalityCoarseTailMean}{0.0303}
\newcommand{\ModelSourceCriticalityCoarseOtherMean}{$4.26\times10^{-7}$}
\newcommand{\ModelSourceCriticalityPulseEntropyOrdinaryMinimum}{2.34}
\newcommand{\ModelSourceCriticalityPulseEntropyOrdinaryMaximum}{2.98}
\newcommand{\ModelSourceCriticalityPulseEntropyFrozenMinimum}{2.22}
\newcommand{\ModelSourceCriticalityPulseEntropyFrozenMaximum}{2.66}
\newcommand{\ModelSourceCriticalityPulseRowOrdinaryMinimum}{10.64}
\newcommand{\ModelSourceCriticalityPulseRowOrdinaryMaximum}{60.37}
\newcommand{\ModelSourceCriticalityPulseRowFrozenMinimum}{5.87}
\newcommand{\ModelSourceCriticalityPulseRowFrozenMaximum}{17.22}
\newcommand{\ModelSourceCriticalityPulseAmplitudeMinimum}{1.00}
\newcommand{\ModelSourceCriticalityPulseAmplitudeMaximum}{1.12}
\newcommand{\ModelSourceCriticalityPeakMeanVarianceMinimum}{0.0264}
\newcommand{\ModelSourceCriticalityPeakMeanVarianceMaximum}{0.0456}
\newcommand{\ModelSourceCriticalityHorizonRowSmallStart}{0.748}

\newcommand{\ModelSourceCriticalityHorizonRowSmallEnd}{0.139}

\newcommand{\ModelSourceCriticalityHorizonRowLargeStart}{0.906}
\newcommand{\ModelSourceCriticalityHorizonChiLargeStart}{$5.53\times10^{-3}$}
\newcommand{\ModelSourceCriticalityHorizonNLLLargeStart}{7.442}
\newcommand{\ModelSourceCriticalityHorizonRowLargeEnd}{0.314}
\newcommand{\ModelSourceCriticalityHorizonChiLargeEnd}{$0.559$}
\newcommand{\ModelSourceCriticalityHorizonNLLLargeEnd}{7.315}
\newcommand{\ModelSourceCriticalityHorizonWidthRatioStart}{0.53}
\newcommand{\ModelSourceCriticalityHorizonWidthRatioEnd}{9.27}
\newcommand{\ModelSourceCriticalityPrecisionPrimalError}{$2.39\times10^{-11}$}
\newcommand{\ModelSourceCriticalityPrecisionNativeDeviceKL}{$4.95\times10^{-4}$}
\newcommand{\ModelSourceCriticalityPrecisionArithmeticKL}{0.0318}
\newcommand{\ModelSourceCriticalityKernelSamples}{512}
\newcommand{\ModelSourceCriticalityKernelReplicas}{4}

\newcommand{\ModelSourceCriticalityKernelQuadratureError}{$1.93\times10^{-11}$}
\newcommand{\ModelSourceCriticalityKernelLogitPrediction}{$7.97\times10^{-3}$}
\newcommand{\ModelSourceCriticalityKernelEntropySmall}{0.1267}
\newcommand{\ModelSourceCriticalityKernelEntropyLarge}{0.0358}
\newcommand{\ModelSourceCriticalityKernelRowSmall}{$3.71\times10^{-3}$}
\newcommand{\ModelSourceCriticalityKernelRowLarge}{$1.07\times10^{-3}$}
\newcommand{\ModelSourceCriticalityKernelCovarianceSmall}{0.1078}
\newcommand{\ModelSourceCriticalityKernelCovarianceLarge}{0.0303}
\newcommand{\ModelSourceCriticalityKernelMonteEntropy}{$3.56\times10^{-3}$}
\newcommand{\ModelSourceCriticalityKernelMonteRow}{$1.06\times10^{-4}$}
\newcommand{\ModelSourceCriticalityWiderEntropyPowerPercent}{20.7}
\newcommand{\ModelSourceCriticalityWiderEntropyRegularPercent}{2.9}
\newcommand{\ModelSourceCriticalityWiderRowPowerPercent}{15.1}
\newcommand{\ModelSourceCriticalityWiderRowRegularPercent}{46.0}

\newcommand{\ModelSourceDynamicsPeakSmallStart}{$0.160$}
\newcommand{\ModelSourceDynamicsPeakSmallEnd}{$1.60\times10^{-6}$}
\newcommand{\ModelSourceDynamicsPeakMiddleStart}{$0.211$}
\newcommand{\ModelSourceDynamicsPeakMiddleEnd}{$1.78\times10^{-5}$}
\newcommand{\ModelSourceDynamicsPeakLargeStart}{$0.638$}
\newcommand{\ModelSourceDynamicsPeakLargeEnd}{$5.81\times10^{-7}$}
\newcommand{\ModelSourceDynamicsAdamPrimalError}{$4.44\times10^{-16}$}
\newcommand{\ModelSourceDynamicsAdamMomentError}{$6.94\times10^{-18}$}
\newcommand{\ModelSourceDynamicsResolvedTangentCases}{24}

\newcommand{\ModelSourceObservationHorizonNllBefore}{7.300}
\newcommand{\ModelSourceObservationHorizonNllAfter}{7.287}
\newcommand{\ModelSourceObservationHorizonContextBefore}{0.501}
\newcommand{\ModelSourceObservationHorizonContextAfter}{0.512}
\newcommand{\ModelSourceObservationHorizonBefore}{0.5336}
\newcommand{\ModelSourceObservationHorizonAfter}{0.1375}
\newcommand{\ModelSourceObservationHorizonDifference}{-0.3961}
\newcommand{\ModelSourceObservationHorizonSE}{0.0917}
\newcommand{\ModelSourceObservationHorizonLow}{-0.5267}
\newcommand{\ModelSourceObservationHorizonHigh}{-0.2134}
\newcommand{\ModelSourceObservationAdditionalDifference}{-0.3651}
\newcommand{\ModelSourceObservationAdditionalLow}{-0.4562}
\newcommand{\ModelSourceObservationAdditionalHigh}{-0.2010}
\newcommand{\ModelSourceObservationHorizonMeanBefore}{0.2330}
\newcommand{\ModelSourceObservationHorizonMeanAfter}{0.0367}
\newcommand{\ModelSourceObservationMeanDecreases}{16}
\newcommand{\ModelSourceObservationStableCells}{8}

\newcommand{\ModelSourceScalingTrainingArtifacts}{96}
\newcommand{\ModelSourceScalingAddedUpdates}{2,154,496}
\newcommand{\ModelSourceScalingMaximumHorizon}{131,072}

\newcommand{\ModelSourceCacheWorkerSeconds}{534.4}
\newcommand{\ModelSourceCacheRMS}{0.042--0.124}
\newcommand{\ModelSourceCacheLatency}{62.96--65.12}
\newcommand{\ModelSourceCacheMaxKL}{0.000712}

\newcommand{\ModelSourceCacheRiskRMS}{0.0396--0.1582}
\newcommand{\ModelSourceCacheRiskMaxKL}{0.001497}
\newcommand{\ModelSourceCacheRiskMaxContext}{0.8514}
\newcommand{\ModelSourceCacheRiskWorkers}{954.5}
\newcommand{\ModelSourceCacheRiskPeakGiB}{1.154}
\newcommand{\ModelSourceCacheRiskRetained}{0.97748--1.00984}

\title[Training and Inference Dynamics of PLDR-LLMs]{Training and Inference Dynamics of PLDR-LLMs\\[6pt]\large Row-Map Collapse, Renormalization, and Predictive Reduction}
\author{Burc Gokden}
\address{Fromthesky Research Labs LLC, Oregon, USA}
\email{burc@fromtheskyresearchlabs.com}
\date{September 18, 2026}
\dedicatory{\normalfont September 18, 2026}
\begin{document}
\frontmatter
\maketitle
\chapter*{Abstract}
I develop a unified account of training and inference in Power Law Decoder
Representation language models. The starting observable is the absolute
energy of the row-centered learned map. Exact finite work identities resolve
its changes into parameter sources, signed interactions, and numerical
observation defects. Radial and tangential coordinates give a positive-energy
gain, while a canonical affine completion retains restarts at the exact
row-constant face. Chronological blocking is associative and preserves the
loss of homogeneous dependence on incoming scalar energy at a face hit.
This scalar representation does not erase the complete optimizer state. The augmented AdamW state supplies
the corresponding dynamical description, with explicit forcing, observation
nonlinearity, and sufficient attraction conditions.

Predictive renormalization acts on the complete conditional training law.
A smaller autonomous state requires closure; finite approximate reductions
carry successor and emission errors. I center this law on a single pass over
distinct corpus target blocks, retaining optimizer memory, remaining data,
schedule, and numerical policy. Finite-population covariance, matched
physical clocks, generated-matrix fluxes, and signed temporal energy connect
row dynamics to model-wide observations. Absolute row collapse, a small
normalized row fraction, downstream operator stabilization, and predictive
accuracy are distinguished mathematically.

The executed evidence resolves observer and optimizer dependence, rejects
the tested autonomous row-state candidates, and supports finite conditional
prediction and state-specific operator reduction. Independent single-pass
families show moving finite fluctuation regions without establishing a
thermodynamic critical class. Conditional sign symmetry, head limits,
covariance flows, and readout error budgets identify the additional
assumptions needed to transfer a scaling law into inference. The resulting
theory combines exact path identities, conditional dynamical statements,
and finite empirical findings without identifying them with one another.
All mathematical proofs are given in the text; selected formal checks and
a compact numerical evidence collection accompany the monograph.

\setcounter{tocdepth}{1}
\begingroup
\makeatletter
\pretocmd{\l@chapter}{\par\begingroup
  \@tempdima=\pagegoal\advance\@tempdima-\pagetotal
  \ifdim\@tempdima<4\baselineskip\newpage\fi
  \endgroup}{}{\PackageError{monograph}{Contents chapter reservation failed}{}}
\makeatother
\tableofcontents
\endgroup
\chapter*{Guide to the monograph}
\label{model:app:supplement-guide}
The organizing question is what must be retained when a microscopic
training computation is replaced by a coarser description. The answer
depends on the task. A realized row-energy path admits a very small exact
coordinate. Predicting its successor generally requires more information.
Preserving a predictive distribution requires an emission as well as a
state transition. Preserving fluctuation exponents requires errors that
decrease on the fluctuation scale. A two-row worked example in
Section~\ref{sec:worked-row-prediction} follows this chain from endpoint
energy through the closure condition to a conditional predictive-error bound.

Part I develops the architecture, row quotient, finite source work, and
complete optimizer dynamics. Part II organizes those objects under
chronological blocking and gives the closure tests and conditional
orbit classifications. Part III makes the consuming corpus an explicit
state variable and develops the connection to shared model-wide fields.
Part IV studies predictive distributions, cached operators, conditional
memory, and adaptation. Part V gives conditional scaling laws, the
empirical size and time comparisons, and the requirements for inference
visibility. Part VI contains the appendices, with statistical units, complete outcome
grids, reproduction boundaries, and selected formal correspondence.

The arguments use finite-dimensional linear algebra, difference equations,
probability kernels, conditional covariance, and elementary scaling theory.
The word \emph{exact} applies to a stated mathematical identity or specified
execution comparison. It does not assert that floating-point arithmetic
recovers a real-arithmetic zero. The word \emph{conditional} records
hypotheses about a law, a retained state, or a limiting family; it does not
mean that the hypotheses have been established experimentally.

The primary training law is single-pass RefinedWeb. Earlier captured
trajectories and deliberately repeated-corpus controls retain their own
data laws and serve only the conclusions their designs support. Released
pretrained models provide observations of trained inference; their complete
ordered pretraining histories are not reconstructed here. No new training
experiment is needed for the synthesis presented in this monograph.

The companion code and reported evidence are available at
\begin{center}
\url{https://github.com/burcgokden/PLDR-LLM-Training-Dynamics}\\[4pt]
\url{https://huggingface.co/datasets/fromthesky/pldr-llm-training-dynamics-data}.
\end{center}
The code repository contains selected Lean developments and scientific
programs for the theory and experiments, organized by the monograph's subjects.
The data repository contains numerical records, complete reported outcome
grids, data dictionaries and portable provenance. Its coverage index links
reported displays and claims to those records and states the limits on
recovering raw observations. Each companion has its own file manifest and
integrity checks.

\noindent\textbf{Zero notation.}
The symbol $0$ denotes a scalar zero, including scalar components, norms,
energies, and individual entries of a numerical array. We write $\bm0$
for a zero vector, $\bm0_{m\times n}$ for an $m$-by-$n$ zero matrix,
and $\bm0_{\mathrm{mat}}$ for a zero matrix whose shape is fixed by its
operands. A superscript $\mathsf T$ indicates a zero row vector.
The symbols $0_{\mathrm{op}}$ and $0_{\mathrm{fun}}$ denote a zero linear
operator and an identically zero scalar function, respectively.
Vector and matrix inequalities using $\le,\ge,<,>$ are entrywise;
$\preceq,\succeq$ denote the positive-semidefinite order on symmetric
matrices, and $\prec,\succ$ its positive-definite version. Thus an
entrywise positive PLGA base and a positive-definite covariance impose
different conditions. In a vector or matrix limit, the zero has the same
type as the quantity converging to it; a limit of its norm remains scalar.

\section*{Physical terminology and renormalization objects}
\label{sec:physical-terminology}
A learned ``metric'' names a matrix stage; symmetry and positive
definiteness require separate proof. ``Curvature'' names an architectural
operator or specified derivative, without implying a Riemannian curvature
tensor. Row energy is the operational statistic
$\|C_rA\|_F^2$. The power potential is an elementwise feature map; a
source-tilted log moment-generating function is a statistical potential.
Neither usage by itself establishes equilibrium thermodynamics.

Four renormalization objects are distinguished. Chronological affine
blocking composes edges of a realized path
(Section~\ref{rg:sec:ordered-affine-blocking}); complete-state kernel powers
compose transition laws (Chapter~\ref{ch:closed-states}); fluctuation-law
normalization specifies a limiting family
(Chapter~\ref{ch:conditional-criticality}); and experimental scale maps
compare declared widths and clocks (Chapter~\ref{ch:physical-clocks}).
Composition alone does not establish a physical scale limit.
The local context specifies the object, scale and conditioning whenever
the construction changes.

\section*{Cross-part observable index}
\label{sec:observable-index}
The table maps local notation without identifying different physical objects.
In particular, $Q(A)$ below is centered energy, whereas the finite-step
charge $\mathcal Q_t$ in Equation~\eqref{eq:overview-work} is an increment norm.

\begingroup
\setlength{\LTpre}{4pt}\setlength{\LTpost}{4pt}
\begin{longtable}{@{}p{.22\textwidth}p{.40\textwidth}p{.30\textwidth}@{}}
\toprule Observable and domain & Centering and normalization & Statistical unit and interpretation\\\midrule\endhead
Absolute row energy $Q(A)$; any finite matrix &
$C_r=I_r-r^{-1}{\bf1}{\bf1}^{\mathsf T}$,
$Q(A)=\|C_rA\|_F^2$. Part I writes $E_t$ for the named physical row map. &
One map at one state and context. Row centering removes the common row;
any further plotted scale must be stated.\\[2pt]
Normalized row fraction $u(A)$; $\|A\|_F>0$ &
$u(A)=Q(A)/\|A\|_F^2$. The total-energy symbol $E$ in matrix-flux formulas
is the denominator, not the Part I centered energy. &
One map; dimensionless contrast share. Transfer to absolute collapse
requires the amplitude bounds of Proposition~\ref{prop:absolute-relative}.\\[2pt]
Operator context scatter &
$G-\overline G$ is centered across contexts at a fixed state. Cache studies
use per-layer Frobenius norm divided by $\sqrt{Nd^2}$ and state their
layer and initialization averages. & Contexts within a checkpoint/panel.
Context centering differs from row centering; the relative centered-RMS
denominator follows the specified predictive experiment.\\[2pt]
Predictive fidelity and target risk & KL from the native predictive law
under the declared proper-prefix convention; external-target NLL is a
separate loss. & Paired prefix/context evaluations. Average fidelity,
target performance and uniform-context accuracy are different claims.\\[2pt]
Collective fluctuation $\Var(U_N)$ and $\chi_N$ & Ensemble centering of the
intensive observation; $\chi_N=N\Var(U_N)$ uses head count $N$. &
Whole initialization/path under the specified corpus and panel law.
Heads, contexts and times remain nested or paired.\\
\bottomrule
\end{longtable}
\endgroup
Figure groups identify the plotted observable and averaging unit locally.
Arithmetic zeros and observation floors retain their stated numerical scope.

\chapter*{Principal results and evidence}
\label{sec:claim-index}
The table follows the dependency chain from finite mechanics to predictive
reduction. A path identity is exact on its stated domain; a dynamical theorem
uses its written assumptions; an empirical conclusion has the sampling unit
and scope shown here. Appendix~\ref{app:formal} gives selected independent
formal checks, and Appendix~\ref{app:reproducibility} locates the evidence.

\begin{longtable}{@{}p{.25\textwidth}p{.34\textwidth}p{.33\textwidth}@{}}
\toprule Result and location & Evidence and unit & Boundary\\\midrule\endhead
Finite row work and PLGA defect
(Sections~\ref{row:sec:direct-energy}, \ref{row:sec:plga-defect}) &
Three trajectories, nine trajectory-layer units, and 864 maps per observation
time in the principal long study. Finite work and interventions resolve
source interactions. & Maps and contexts are nested observations. A native
zero is not an all-future real-arithmetic collapse certificate.\\[4pt]
Chronological row blocking
(Section~\ref{rg:sec:ordered-affine-blocking}) &
Exact affine composition, including the realized path $1\to0\to1$. &
Loss of homogeneous scalar-energy dependence does not erase optimizer or
corpus memory. Predictive closure is a separate condition.\\[4pt]
Transported forcing
(Corollary~\ref{row:cor:linear-stochastic-forcing}) &
Deterministic propagators and square-integrable martingale differences. &
Mean collapse concerns the summed mean; mean-square collapse also requires
the exact transported variance to vanish. Envelopes are sufficient.\\[4pt]
Differential row transport
(Corollary~\ref{model:cor:row-path-error}) &
Actual finite matrix increments and a stated generator remainder. &
The summed budget is sufficient at the specified scale; radial rescaling
shows that it is not necessary.\\[4pt]
Row-RG decision procedure
(Section~\ref{rg:sec:executed-results}) &
Four trajectories of 196,608 updates; 12 trajectory-layer cells. &
Ten complete cells fail the nominal precision gate; two are face-interrupted.
Native batch dependence is not justified by Gaussian calibration.\\[4pt]
Single-pass width refinement
(Section~\ref{model:sec:critical-independent-results}) &
Six initialization identities per width/gain cell, with a fixed shared
generator and panel. The 216 refinement paths use a disjoint second
partition of one master corpus. &
Moving finite fluctuation regions and unresolved systematic narrowing do not
identify a thermodynamic critical class.\\[4pt]
Reduced forecasts and conditional memory
(Sections~\ref{model:sec:law-closure-results}, \ref{model:sec:memory-results}) &
Two outer initialization identities produce eight incoming states.
At each state, 16 development, eight calibration and 16 validation
branches share a fixed risk panel; coverage is 117/128. &
Tube widths are $0.0431$--$0.3472$ nats; states derive from two outer
initializations. Coverage conditions on the protocol and calibration law.\\[4pt]
State-specific cache reduction
(Section~\ref{model:sec:cache-state-results}) &
Six paired initialization identities at each of two widths give twelve
initial states and 24 trained endpoints. A disjoint 16/64-document
calibration/assessment panel supports 18 passing recalibrated cells. &
Exceptions reach six of 64 contexts. Initial-cache transfer fails in all six
positive-control cells. This is fixed-panel fidelity, not latency or
uniform generated-history accuracy.\\[4pt]
Signed head limit
(Theorem~\ref{model:thm:native-sign-gaussian-mixture}) &
Complete-law head-sign invariance, fixed observation dimension, a uniform
head bound, and joint convergence of invariant observation and covariance. &
The limit is $(H,C,C^{1/2}W)$ with $W$ independent of $(H,C)$; $C$ may be
random or singular. A finite sign check does not prove convergence.\\[4pt]
Physical readout
(Proposition~\ref{model:prop:readout-refinement}) &
Condition on checkpoint, fitted map and configuration law. For positive
reference second moment $\mu_L$, require
$\delta_L=o(\sqrt{\mu_L})$. &
Transfers an existing second-moment slope. Connected variance,
generated laws and fourth moments have separate error conditions.\\
\bottomrule
\end{longtable}

\section*{Assumptions and conditioning at a glance}
\label{sec:assumption-guide}
A fixed corpus and an ensemble of newly drawn corpora define different laws.
Likewise, contexts at one checkpoint do not provide independent trained
initializations. The guide records what is retained before any averaging;
Appendix~\ref{app:reproducibility} identifies the concrete panels and overlaps.
Single-pass finite-corpus training is the primary law. Any asymptotic claim
must specify its family of widths, training ages, corpus sizes and observation
scales; corpus exhaustion is not removed by changing the clock.

\begin{longtable}{@{}p{.20\textwidth}p{.43\textwidth}p{.29\textwidth}@{}}
\toprule Claim family & Retained state or conditioning & Horizon and conclusion\\\midrule\endhead
Finite row work and blocking & Realized row maps, source increments,
observation rule and chronological edge list
(Sections~\ref{row:sec:direct-energy}, \ref{rg:sec:ordered-affine-blocking}). &
Finite path identity; closure and all-future attraction require more.\\[4pt]
AdamW and consuming corpus & Weights, both moments, counters, schedule,
remaining source positions, selection history and program randomness
(Sections~\ref{model:sec:training}, \ref{model:sec:data-resource}). & Complete successor law or a separately
specified conditional reduction; not a stationary source.\\[4pt]
Conditional prediction tubes & Incoming complete state, development fit,
remaining-source law and fixed risk panel; calibration and future-path
randomness are separate (Section~\ref{model:sec:law-closure-results}). &
Finite future-path coverage averaged over calibration and one future path;
no simultaneous or outer-model coverage.\\[4pt]
Cache risk & Checkpoint, proper-prefix convention, calibration panel or its
sampling law, assessment panel, cache rule, width and control
(Section~\ref{model:sec:cache-state-results}). & Fixed-panel aggregate
fidelity; context exceptions and state-transfer limits remain.\\[4pt]
Signed operator limit & Complete-law sign invariance at $T_N$, fixed
observation dimension, uniform head bound, joint convergence of invariant
observation and covariance
(Theorem~\ref{model:thm:native-sign-gaussian-mixture}). & Conditional mixture
along the declared family, or the specified convergent subsequence.\\[4pt]
Physical readout & Checkpoint, fitted readout, configuration/source law,
size convention, positive reference fluctuation scale and error norm
(Section~\ref{model:sec:readout-budget}). & Transport of an existing signal;
centered and higher-moment conclusions need their own budgets.\\
\bottomrule
\end{longtable}

\section*{Organization of the theory}
Part I develops quotient geometry, observer-resolved work, radial-face
coordinates, complete AdamW dynamics, and the PLGA defect. Part II develops
affine blocking, gauge normalization, orbit criteria, and closure and
observer tests. Parts III--V connect these results to complete conditional
laws, consuming-corpus training, collective and optimizer clocks, predictive
memory, cache risk, and conditional size/time limits. The interfaces between
these descriptions state explicitly which observations and hypotheses
support each conclusion.

\mainmatter
\part{Architecture, observations, and exact row dynamics}
\chapter{A unified theory of training and inference}
This chapter introduces the problem of relating learned row geometry to
training dynamics and predictive inference. It sets out the chain of
arguments, the distinctions between exact identities and conditional limits,
and the role of the finite experimental evidence.

\label{ch:overview}
\label{row:sec:introduction}
\label{rg:sec:introduction}
\section{The central question}
PLDR-LLMs generate the operators that couple queries and keys inside
attention. Their learned row maps can become highly concentrated while
the model retains a nontrivial common component and useful predictive
variation. I study how that geometry is produced by training, how its
finite evolution can be composed across time, and when a reduced
description remains predictive after the full model is frozen.

The architecture originates in power-law graph attention and its language
model implementations \cite{gokden2019,gokden2021,gokden2024,gokden2025,gokden2026foundations}.
The mathematical problem is more specific than associating a power-law
nonlinearity with a critical phenomenon. One must identify an observable,
its conditioning law, a scale transformation, and the assumptions that
make a limiting statement meaningful. A learned power exponent is an
ordinary parameter. A critical exponent is a property of a specified
family of probability laws under a specified scale map.

Three levels of description structure the analysis. At the first level,
finite endpoint algebra gives identities that hold on every admissible
realized path. At the second, an augmented state and a source law define
the predictive process. At the third, a chosen emission and its
observation rule determine what survives in inference. These levels
are compatible, but each passage between them introduces a question
that an endpoint identity alone cannot answer.

\section{The chain of conclusions}
Finite row mechanics supplies exact work and mixed gate/shape identities.
Chronological blocking then describes a realized energy path, while the
complete-state kernel retains the variables needed to specify training.
Closure and emission-error tests determine whether those descriptions can
be reduced. Frozen-state continuation tubes and recalibrated caches provide
finite predictive successes at specified states and error budgets
(Sections~\ref{model:sec:law-closure-results} and
\ref{model:sec:cache-state-results}). Conditional scaling theorems describe
the additional assumptions needed for a limiting law
(Section~\ref{model:sec:theory-synthesis}). None of these passages substitutes for the next.

The architecture-specific content includes the PLGA response-plus-defect
decomposition and the compensating query/operator head-sign action,
including the AdamW first-moment transformation
(Section~\ref{model:sec:singlepass-critical-observations}). Signed cross-head
cancellation does not remove invariant cross-head dependence. In inference,
context scatter, estimation of a cached operator, and displacement between
training states are separate contributions to risk. These distinctions
connect the row mechanics to what a reduced predictive computation can
actually preserve.

\section{The row quotient and its finite work}
For a matrix of generated rows $X_t\in\R^{r\times d}$, write
\[
 C_r=I_r-r^{-1}\mathbf1\mathbf1^\top,\qquad
 Z_t=C_rX_t,\qquad E_t=\|Z_t\|_F^2.
\]
The quotient removes a common row. For fixed finite $r$, vanishing
$E_t$ is equivalent to vanishing maximum pairwise row distance.
It neither forces the common row to vanish nor fixes a downstream
operator. At the final affine normalization,
$Z_t=Y_{c,t}\operatorname{diag}(\gamma_t)$ and
$E_t=\sum_k\gamma_{k,t}^2\|Y_{c,:k,t}\|^2$.
Collapse therefore concerns the mixed gate and shape coordinates;
separate collapse of every gate is unnecessary. The quotient equivalence is
proved in Lemma~\ref{row:lem:row-quotient-geometry}, and
Section~\ref{row:sec:gate-shape-geometry} derives the mixed factorization.

Let $D_t=Z_{t+1}-Z_t$. Expansion of a squared norm gives
\begin{equation}
 E_{t+1}-E_t
 =2\langle Z_t,D_t\rangle_F+\|D_t\|_F^2
 =-\mathcal W_t+\mathcal Q_t.
 \label{eq:overview-work}
\end{equation}
The signed work $\mathcal W_t=-2\langle Z_t,D_t\rangle_F$ must
exceed the nonnegative finite-step charge $\mathcal Q_t$ for contraction.
An inward differential direction can fail this test at finite amplitude.
For a source split $D_t=\sum_aD_t^{(a)}$, the charge contains every
pairwise source inner product. Discarding cross terms changes the
identity and can change its interpretation. Section~\ref{row:sec:finite-work}
derives Equation~\eqref{eq:overview-work} and the complete source ledger.

On a positive-energy edge, define
$\alpha_t=-\langle Z_t,D_t\rangle_F/E_t$ and
$T_t=D_t+\alpha_t Z_t$. Orthogonality yields
\[
 \frac{E_{t+1}}{E_t}=(1-\alpha_t)^2+\frac{\|T_t\|_F^2}{E_t}.
\]
At the exact face $E_t=0$, the same division is unavailable and
$E_{t+1}=\|D_t\|_F^2$. The complete source increment determines
whether the face is preserved. Small native energy and exact real
row equality have different meanings; an observer must preserve that
distinction when reporting gains or restarts.
Sections~\ref{row:sec:radial-normal-form} and \ref{row:sec:face-restarts}
derive the positive-energy gain and face completion;
Section~\ref{row:sec:observer-resolution} specifies the numerical observer.

\section{A path coordinate and a predictive state}
The canonical edge $(q_t,\zeta_t)$ has $q_t=E_{t+1}/E_t$ and
$\zeta_t=0$ on a positive source, and $q_t=0$, $\zeta_t=E_{t+1}$
on the face. It acts by $e\mapsto q_te+\zeta_t$.
Chronological composition is
\begin{equation}
 (q_2,\zeta_2)\circ(q_1,\zeta_1)
 =(q_2q_1,q_2\zeta_1+\zeta_2).
 \label{eq:overview-compose}
\end{equation}
The product-convolution formula retains every source after transport
by later gains. A block that crosses a face can lose dependence on its
incoming energy even when both block endpoints are positive. Recomputing
a ratio from those endpoints loses this information. This concerns the
homogeneous scalar-energy coefficient only; parameters, optimizer moments,
common rows, and corpus state can still influence later sources.
Section~\ref{rg:sec:ordered-affine-blocking} derives the ordered composition
in Equation~\eqref{eq:overview-compose} and its product-convolution formula.

The gain uses an observed successor. It is therefore a path coordinate,
not a causal model for an unseen update. To predict, let $S_t$ include
parameters, AdamW moments, schedule and bias-correction clocks, remaining
corpus, and the stochastic or numerical program state. Its transition
kernel $K_t$ can be composed chronologically. A time-homogeneous notation
$K^b$ is justified by augmenting deterministic clocks and resource state;
it does not make a consuming corpus stationary.
Section~\ref{model:sec:training} constructs this augmented training law, and
Section~\ref{model:sec:data-resource} specifies its consuming-corpus kernel.

An observable $\pi(S)$ admits a state-independent reduced kernel only
when the pushed-forward successor law is the same for all incoming
states in each relevant fiber. Section~\ref{rg:sec:kernel-lumpability}
states and proves this closure condition. The matched-fiber interventions
in Section~\ref{rg:sec:matched-fiber-results} show that energy and raw
optimizer coordinates combined with energy do not supply that closure
on their tested domains. Section~\ref{rg:sec:staged-closure-results}
tests and rejects the specified richer row states. Those findings
leave the exact path identity intact and identify information that a
predictive model must retain or approximate.

\begin{figure}[htbp]
\centering
\begin{tikzpicture}[node distance=14mm,>=Stealth,
 every node/.style={align=center,font=\small},
 box/.style={draw,rounded corners,text width=5.6cm,minimum height=1.1cm}]
\node[box] (state) {Complete incoming state $S_t$\\parameters, memory, corpus, clocks};
\node[box,right=14mm of state] (next) {Successor state $S_{t+b}$\\chronological kernel composition};
\node[box,below=of state] (obs) {Retained observation $\pi(S_t)$\\row, collective, or predictive coordinate};
\node[box,below=of next] (out) {Retained successor and emission\\conditional law and error budget};
\draw[->] (state)--node[above]{$K_{t:t+b}$}(next);
\draw[->] (state)--(obs);
\draw[->] (next)--(out);
\draw[->,dashed] (obs)--(out);
\end{tikzpicture}
\caption{The complete law always has a chronological composition.
The lower predictive map requires additional hypotheses. Realized
energy edges describe an observed path through this diagram.}
\label{fig:law-diagram}
\end{figure}
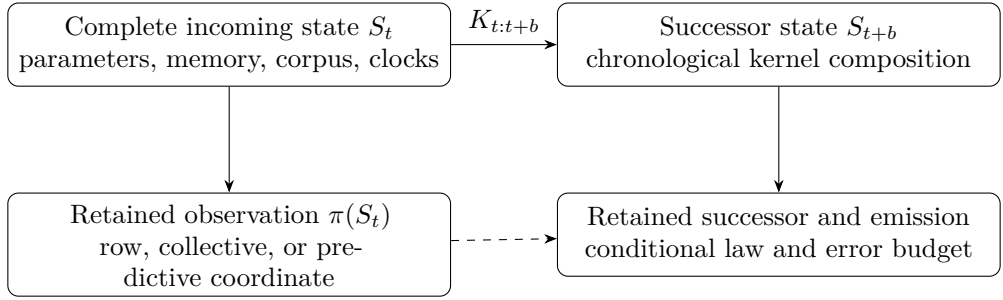

\section{The single-pass resource and the choice of clock}
A finite corpus is consumed. Distinct token blocks mean distinct source
positions under the stated tokenizer and block construction, not distinct
token values or necessarily independent documents. A uniform random
ordering without replacement has a negative finite-population covariance
between different draws. The remaining population and the consumed
fraction therefore belong to the law. Adaptive gradients add state
dependence to this resource effect; a covariance identity for a frozen
population is not a diffusion limit for adaptive training.
Section~\ref{model:sec:data-resource} derives the resource identities;
Section~\ref{model:sec:consuming-bridge} transports the frozen-population
covariance through a specified linear response.

The optimizer also has a clock. Keeping $\beta_1,\beta_2$ fixed preserves
memory in update counts. A physical-step refinement $h_N\to0$ with
$\beta_i(N)=\exp(-\gamma_i h_N)$ describes another specified family.
Matching memory, physical duration, and consumed fraction permits
meaningful finite comparisons, but does not force generated row matrices
to move at a width-independent speed. The exact finite row flux keeps
normalization changes, quadratic increments, and cross-time interactions.
It is the appropriate starting point before taking a differential limit.
Sections~\ref{model:sec:matched-memory-bound} and
\ref{model:sec:physical-clock} derive the memory and clock conditions;
Section~\ref{model:sec:collective-clock} gives the exact finite row flux.

Repeated-corpus studies are useful controlled changes of this source law.
They can reveal fitting and resource-reuse effects. Their fitted laws and
effective exponents do not automatically transfer to the single-pass
family. Likewise, a fixed checkpoint observed on many contexts and an
ensemble of independently trained checkpoints represent different random
units. Section~\ref{model:sec:onepass-results} reports the source and
schedule comparisons, and Section~\ref{model:sec:statistics} specifies
the statistical units.

\section{From geometric concentration to inference}
There are two geometric measurements used throughout the book. The
absolute row energy $\|C_rX\|_F^2$ resolves a physical row difference.
The normalized fraction $\|C_rA\|_F^2/\|A\|_F^2$ measures the share of
matrix energy in row contrast. They refer to the same quotient operation
but may concern different stages of the generator. Even for the same
matrix, they require an amplitude bound to be interchangeable.
Section~\ref{sec:bridge-geometry} supplies the precise comparison.

PLGA can create row variation from a constant-row input. Its
response-plus-defect formula therefore retains both propagated input
contrast and an intrinsic constant-input defect. A row theorem reaches
the predictive output only after these terms are controlled over the
relevant inputs and transported through the decoder. A fixed finite
probe registry does not provide a uniform cover of every inference prefix.
Section~\ref{row:sec:plga-defect} derives the response-plus-defect formula;
Section~\ref{sec:bridge-geometry} gives the conditional decoder error budget.

Conversely, predictive reduction can succeed without exact row collapse.
An operator cache may approximate the relevant context-conditioned mean
while its risk retains context scatter and training-state displacement.
Nested vocabulary projections preserve a specified resolution of a
probability vector. Their error bounds concern that emission and
conditioning law. They do not establish an economical autonomous
training algorithm. The cache-risk decomposition is derived in
Section~\ref{model:sec:cache-risk-law}, and the nested predictive
error bounds in Section~\ref{model:sec:nested-resolution}.

\section{What the finite evidence supports}
The experiments provide several complementary conclusions. The row
measurements in Section~\ref{row:sec:executed-evidence} support the finite
work and source accounting while resolving arithmetic floors, direction
dependence, and reopenings. The long row study does not identify a
directional phase or universality class
(Section~\ref{rg:sec:temporal-flow-results}). Matched interventions reject
the proposed reduced closure states
(Sections~\ref{rg:sec:matched-fiber-results} and
\ref{rg:sec:staged-closure-results}).

Native single-pass families exhibit finite training-selected row, common,
and predictive fluctuations (Sections~\ref{model:sec:critical-onepass-results}
and \ref{model:sec:critical-independent-results}). Their moving fluctuation
regions do not show systematic narrowing across the measured widths.
A distinct 48-trajectory family matches memory, time, and consumption
(Section~\ref{model:sec:matched-clock-results}). Recorded one-step and
multistep continuations verify finite matrix transport and chronological
blocking, including signed temporal interference
(Sections~\ref{model:sec:collective-flux-results} and
\ref{model:sec:row-path-results}).

At frozen inference, disjoint assessment panels support the declared
aggregate accuracy targets for state-specific calibration
(Sections~\ref{model:sec:operator-cache-results},
\ref{model:sec:cache-risk-results}, and \ref{model:sec:cache-state-results}).
The same results expose context dependence, state transfer error, and
acquisition cost. The appropriate positive statement is finite conditional predictive
reduction. A thermodynamic critical limit, self-organized critical
selection, and economical autonomous reduced training remain unresolved.
Conditional theorems below state sufficient premises for these stronger
conclusions and do not treat a finite measurement as an all-future premise.

\section{Hypotheses behind the limiting summaries}
Complete conditional head-sign invariance, a fixed observation dimension,
a uniform head bound, and joint convergence of $(H_N,C_N)$ give
\[
 (H_N,C_N,Z_N)\Longrightarrow(H,C,C^{1/2}W)
\]
in Theorem~\ref{model:thm:native-sign-gaussian-mixture}
(Section~\ref{model:sec:native-sign-limit}), with $W$ standard
Gaussian and independent of $(H,C)$. The covariance may be random or
singular. Sign-orbit averaging does not make trained heads independent,
and a finite symmetry check does not establish covariance convergence.
The statement applies along the chosen size/time family, or only along
a subsequence when that is the available convergence premise.

For physical readout, condition on the checkpoint, fitted map and
configuration law. With $\mu_L=\E\|v_L\|^2>0$ and
$\delta_L=\|\widehat v_L-v_L\|_{L^2}$,
Proposition~\ref{model:prop:readout-refinement} in
Section~\ref{model:sec:readout-budget} gives
\[
 \left|\frac{\E\|\widehat v_L\|^2}{\mu_L}-1\right|
 \le 2r_L+r_L^2,\qquad r_L=\frac{\delta_L}{\sqrt{\mu_L}}.
\]
Thus $\delta_L=o(\sqrt{\mu_L})$ transfers an existing second-moment slope;
it does not establish that slope. Connected variance requires a centered
error budget, generated-law transfer its total-variation or prefix-KL
condition, and fourth-moment or Binder ratios stronger moment control.
An absolute accuracy threshold alone does not suffice as the reference
fluctuations shrink. Imposed Ising/Potts source laws supply calibration,
not evidence of native self-organization. Their experimental design and
results are in Sections~\ref{model:sec:physical-results} and
\ref{model:sec:readout-results}.

\section{Notation shared across the descriptions}
\begin{longtable}{@{}p{.22\textwidth}p{.72\textwidth}@{}}
\toprule Coordinate & Meaning and scope\\\midrule\endhead
$C_r$, $\rowq$, $\Pi$ & Orthogonal projection that removes the common row;
the dimension is specified locally.\\
$X$, $A$, $\Alm$, $\G$ & The physical residual row map, a generic generated
matrix, the positive PLGA base, and the signed attention operator.
They are distinct stages; a bound must state which one it concerns.\\
$Z$, $E$ & In the row chapters, $Z=C_rX$ and $E=\|Z\|_F^2$.
In normalized matrix-flux formulas, $E=\|A\|_F^2$ is explicitly the
total energy. The fraction is always displayed as a ratio.\\
$D_t$ & A finite matrix increment when used in work or flux identities;
the data distribution is denoted $\mathcal D$.\\
$\mathcal W$, $\mathcal Q$ & Signed inward work and squared finite-step charge;
$Q_{\rm rope}$ is instead the rotary query matrix.\\
$(q,\zeta)$ & Canonical row-energy gain and face restart. The consumed
corpus fraction, also conventionally written $q_N$ in resource formulas,
is always indexed by the size family and defined there.\\
$S_t$, $s_t$, $\theta_t,m_t,v_t$ & Complete training state and its
parameter and optimizer coordinates.\\
$t$, $h_Nt$, $N$ & Update count, physical time for a declared step
$h_N$, and the size coordinate of a specified family.\\
$\mu$, $K$, $\psi$ & A conditioned law, transition kernel, and observation
map; their domains are fixed where each is introduced.\\
$\chi_N=N\Var(U_N)$ & A count susceptibility for the intensive observation
$U_N$ under its stated ensemble; it is not a geometric-length convention.\\
\bottomrule
\end{longtable}
Other symbols have local definitions. This convention permits the row
mechanics and model-wide laws to share their natural coordinates while
making their interfaces explicit.

\section{Physical terminology}
\begin{longtable}{@{}p{.20\textwidth}p{.74\textwidth}@{}}
\toprule Term & Mathematical use in this monograph\\\midrule\endhead
Metric and curvature & Learned arrays or generator coordinates in the
PLDR program. Symmetry, positive definiteness, and differential-geometric
curvature identities are used only where explicitly imposed.\\
Potential & A specified scalar observation of the generator or model;
it is not automatically a mechanical potential or Lyapunov function.\\
Energy & A declared squared norm, normalized fraction, or named potential
observation. The local definition fixes its units and random unit.\\
Work & The signed finite secant contraction in
Equation~\eqref{eq:overview-work}; its balance includes the quadratic
finite-step charge and implies no mechanical conservation law.\\
Renormalization & A specified composition, elimination, or pushforward
of a retained observation or law. Equilibrium and universality require
additional hypotheses.\\
Power and critical exponent & A learned power is a model parameter;
a critical exponent describes a specified limiting family.\\
\bottomrule
\end{longtable}

\FloatBarrier
\par\medskip\noindent
Chapter~\ref{ch:forward-program} specifies the forward computation and
conditioning law used throughout the book. These definitions identify the
model, observations and sources to which the geometric analysis applies.

\chapter{The forward program and its conditional observations}
\label{ch:forward-program}
This chapter fixes the PLDR forward program, its layer and head variables,
and the conditional experiment that gives training observations their
meaning. It also places the architectural mechanisms and reduction questions
in relation to optimizer dynamics and collapse theories.

\section{The complete PLDR forward state}
\label{model:sec:model}

Fix a vocabulary of size $V$, a context length $S$, $L$ decoder layers, $h$
heads per layer, and head dimension $d$. The index $\ell\in\{1,\ldots,L\}$
denotes the decoder layer, and $a\in\{1,\ldots,h\}$ denotes the attention
head within that layer. A sample is $(x,y)\sim\D$, with
$x\in\{1,\ldots,V\}^{S}$ and $y$ an external next token. Only $x$ enters
the model. Let $X_0$ be the scaled, normalized embedding array and
$X_\ell\in\R^{S\times hd}$ the decoder boundary state. In the native
implementation used here,
\begin{alignat}{2}
 Q_{\ell a}&=\operatorname{RoPE}(X_{\ell-1}W^Q_{\ell a}+b^Q_{\ell a}),
 &\quad K_{\ell a}&=\operatorname{RoPE}(X_{\ell-1}W^K_{\ell a}+b^K_{\ell a}),\\
 V_{\ell a}&=X_{\ell-1}W^V_{\ell a}+b^V_{\ell a},\notag\\
 D_{\ell a}&=Q_{\ell a}^{\mathsf T}Q_{\ell a},
 &\quad A_{\ell a}&=\Phi_\ell\bigl(\LN(D_{\ell a})\bigr),\label{model:eq:gram}\\
 \Alm^{(\ell a)}&=f(W_{\ell a}A_{\ell a}+b_{\ell a})+\epsilon_A,
 &\quad P^{\mathrm{pot}}_{\ell a}&=\left(\Alm^{(\ell a)}\right)^{\odot P_{\ell a}},\\
 \G^{(\ell a)}&=a_{\ell a}P^{\mathrm{pot}}_{\ell a}+b^a_{\ell a},
 &\quad T_{\ell a}&=\softmax_{\mathrm{causal}}
 \left(Q_{\ell a}\G^{(\ell a)}K_{\ell a}^{\mathsf T}/\sqrt d\right),\label{model:eq:plga}\\
 O_{\ell a}&=T_{\ell a}V_{\ell a}.
\end{alignat}
Here $f(t)=t^2\sigma(t)$ is iSwiGLU, $\epsilon_A=10^{-9}$, the powers are
entrywise, and both products $WA$ and $aP^{\mathrm{pot}}$ are ordinary
matrix products acting on the left. The generator $\Phi_\ell$ has hidden width $A_{\rm dff}=170$ and eight
residual units, each with two SwiGLU maps and a final LayerNorm.
The value projection uses learned weights $W^V_{\ell a}$ and bias
$b^V_{\ell a}$, with the bias broadcast over token positions. The concatenated
head outputs pass through a learned output projection to give $O_\ell$.
The complete decoder is
\begin{alignat}{2}
 U_\ell&=\LN_{\ell,1}(X_{\ell-1}+O_\ell),&\quad X_\ell&=\LN_{\ell,2}\bigl(U_\ell+\operatorname{FFN}_\ell(U_\ell)\bigr)
       =F_\ell(X_{\ell-1}),\label{model:eq:decoder}\\
 z_\theta(x)&=W_{\mathrm{out}}X_L[S]+b_{\mathrm{out}},&\quad p_\theta(\cdot\mid x)&=\softmax z_\theta(x).
\end{alignat}
Matrix orientation at the vocabulary projection is conventional. The
actual implementation acts on the last array axis. All LayerNorm
denominators include $\epsilon_{\mathrm{LN}}=10^{-6}$.

The optimizer partition used in the experiments assigns the shared residual metric
units, their internal normalizations, and the learned PLGA parameters
$W,b,P,a,b^a$ to the generator group. The body group contains the
remaining parameters, including the initial Gram normalization in
Equation~\eqref{model:eq:gram}, the query, key and value projections, the
embedding, decoder feed-forward and normalization layers, and the
vocabulary projection. Both groups can therefore change the generated
operator. This parameter partition is distinct from projecting a
metric matrix into common-row and row-contrast coordinates.
Section~\ref{model:sec:potential-intervention-law} develops the corresponding
training interventions, and Section~\ref{model:sec:potential-factorial-results}
reports their executed comparisons.

The names ``metric'' and ``energy-curvature'' designate architecture
outputs. Entrywise positivity of $\Alm$ does not make it a symmetric
positive-definite metric, and $\G$ need not be symmetric. Its eigenvalues
are therefore not automatically masses or inverse susceptibilities of a
statistical field theory.

The learned entries of $P_{\ell a}$ are ordinary model parameters;
they are distinct from critical exponents of a size-indexed probability
law. At a fixed probe input, write $M_t=\Alm(t)>\bm0_{d\times d}$,
$V_t=M_t^{\odot P_t}$ and $\Delta Z_t=Z_{t+1}-Z_t$. Then
\begin{equation}
 \Delta\log V_t=(\Delta P_t)\odot\log M_t
       +P_{t+1}\odot\Delta\log M_t.
 \label{model:eq:potential-increment}
\end{equation}
This exact finite identity follows by subtracting
$P_t\odot\log M_t$ from $P_{t+1}\odot\log M_{t+1}$ and adding the
intermediate product $P_{t+1}\odot\log M_t$. Hence a burst in the
power parameters can be attenuated where $M_t$ is near one, while
changes of the power base can produce potential activity even when
$P_t$ barely moves. Changing the probe input adds source variation to
this observation. Potential activity, its projection into $\G$ and
its predictive emission are separate coordinates of the complete law;
none alone identifies a critical scaling class.
Section~\ref{model:sec:potential-dynamics} extends
Equation~\eqref{model:eq:potential-increment} to source and temporal budgets;
Section~\ref{model:sec:potential-results} reports the potential measurements.

Equation~\eqref{model:eq:gram} uses every supplied query. Score masking does not
remove that global dependence. Consequently our predictive observable is
the final supplied position, whose target $y$ is excluded from the input.
The default entropy, head output and hidden-state projections are
also observed at the final supplied position. The explicit prefix
experiments in Section~\ref{model:sec:inference-results} compare this law
with earlier positions in a parallel pass. Those historical rows are
not independently evaluated shorter-prefix conditionals.

Let $\Xi_\theta(x,y)$ denote the flattened collection of all forward
tensors, the final logits or probabilities, and any specified external
loss. It can be as large as the full computational trace. For a measurable
observation $\psi$ define
\begin{equation}
 \Phi_\theta=\psi(\Xi_\theta),\qquad
 m_{\theta,\D}=\E_\D\Phi_\theta,\qquad
 \varphi_\theta=\Phi_\theta-m_{\theta,\D},\qquad
 \mu_{\theta,\D}=(\varphi_\theta)_\#\D.\label{model:eq:law}
\end{equation}
The identity observation gives the complete trace law. A finite
measurement such as the 174-vector specified in
Section~\ref{model:sec:joint-observation-methods} gives its pushforward. Either state is
a \emph{joint} law: replacing it by a product of marginal head laws is an
additional approximation.

\begin{proposition}[Finite-call regularity]\label{model:prop:regularity}
For fixed finite weights, fixed $S,V,L,h,d$, positive normalization and
power-base offsets, and a fixed mask, the real-arithmetic forward map is
smooth in its continuous inputs and in small continuous source
perturbations. Any specified finite collection of forward observables,
including the external next-token negative log probability, is bounded
over the finite set of token samples. Its log moment generating function
\begin{equation}
 W_{\theta,\D}(j)=\log\E_\D e^{j^{\mathsf T}\varphi_\theta}\label{model:eq:cgf}
\end{equation}
is finite and real analytic for every real $j$.
\end{proposition}
\begin{proof}
Affine maps, sigmoid and SwiGLU are smooth. A LayerNorm denominator is a
square root of a strictly positive number. The elementwise power can be
written $u^p=\exp(p\log u)$ with $u\geq\epsilon_A>0$. Masked softmax is a
smooth ordinary softmax on each fixed allowed support, and its allowed
probabilities are strictly positive. Composition establishes smoothness.
There are finitely many token samples of fixed length, so any finite-valued
observable has a finite maximum and minimum on that set. Its moment
generating function is a finite positively weighted sum of exponentials;
it is strictly positive on real sources. Taking its logarithm therefore
preserves local real analyticity everywhere on $\R^p$.
\end{proof}

This proposition concerns the mathematical map. Floating-point execution
is an observation of that map with rounding, and can suppress small
variations. The precision replays in
Section~\ref{row:sec:precision-endpoints} and finite-response controls in
Section~\ref{model:sec:numerical-response-results} quantify this distinction. Exact equality of real tensors is never inferred from their small
native-precision variance.

\section{The conditioned single-pass experiment}
\label{model:sec:conditioned-experiment}
Fix a document law $\mathcal D$, tokenizer, block/target rule, architecture,
initialization law, optimizer, schedule and arithmetic policy. A realized
corpus $\mathcal C_M$ and an ordered single pass are additional conditioning
choices. Distinct source target positions define nonrepetition; identical
words in different documents do not constitute repeated exposure.
The complete state is
\[
 S_t=(\theta_t,m_t,v_t,\sigma_t,R_t,\xi_t),
\]
with parameters, Adam moments, schedule and bias clocks, remaining block
identities, and sampler or stochastic-program state. The native update
removes each used block from $R_t$. Conditioning on the complete ordering
makes the next batch deterministic; marginalizing its unrevealed part
gives the without-replacement kernel in Equation~\eqref{model:eq:single-pass-kernel}.
A remaining count alone generally does not identify that kernel.

The observation law fixes a proper prefix $X$ and external next token $Y$.
Only $X$ enters $p_{\theta_t}(\cdot\mid X)$; $Y$ supplies target risk after
the prediction is formed. Because the global query Gram uses every supplied
token, a causal attention mask does not justify including $Y$ in that call.
Training emission is the trained-weight marginal of the complete state
followed by this declared observation. Freezing weights stops optimizer
and resource evolution while retaining input dependence.

The common object on which the maps act is the joint law
\begin{equation}
 \mathbf P_N^{\mathcal D}(ds_{0:T},dx,du)
 =\rho_{N,0}^{\mathcal D}(ds_0)
  \prod_{t=0}^{T-1}K_{N,t}^{\mathcal D}(s_t,ds_{t+1})\,
  \mu_{N,T}(dx\mid s_T)\,
  \mathcal E_{N,T}(du\mid s_T,x^-).
 \label{model:eq:complete-conditioned-law}
\end{equation}
Here $N$ indexes the declared model-size family; in the native width
families it is the number of heads per layer, $N=h$. The index $t$ counts
optimizer updates, $T$ is the training horizon, and
$s_{0:T}=(s_0,\ldots,s_T)$ is a path of realizations of the complete state
$S_t$ defined above. The initial-state law is $\rho_{N,0}^{\mathcal D}$,
and $K_{N,t}^{\mathcal D}(s_t,ds_{t+1})$ is the conditional law of the
next complete state given $s_t$; the product orders these transitions
chronologically along the path. The evaluation kernel $\mu_{N,T}(dx\mid s_T)$
selects $x=(x^-,y_{\rm target})$, a proper prefix and its external target,
conditional on the terminal state. The emission kernel
$\mathcal E_{N,T}(du\mid s_T,x^-)$ gives the conditional law of the
declared model output $u$, which can include operators and the entire
predictive distribution. The target is absent from its arguments.
The measure arguments $ds_{0:T},dx,du$ specify integration over the
state path, evaluation example and emission under the joint probability
law $\mathbf P_N^{\mathcal D}$.

The index $\mathcal D$ abbreviates
the specified document law and protocol; any additional conditioning
on a realized corpus, order or generator seed is retained in every
factor. Marginalizing a factor changes the experiment and its covariance
sectors. Aligned temporal blocking integrates intermediate $s_t$;
decoder elimination integrates internal graph variables in $\mathcal E$;
fixed observation maps push forward $u$. Thus compatible maps share one
law rather than separate, implicitly matched ensembles.
Section~\ref{model:sec:training} develops these update and emission
kernels and their chronological composition;
Section~\ref{model:sec:data-resource} gives the complete single-pass
kernel and its corpus-conditioning conventions.

The central experiments condition on a fixed shared generator initialization
and one recorded ordering in each of two disjoint partitions of a master
RefinedWeb corpus. Covariances use complete model initialization identities
at each fixed context, then average over contexts. Heads, decoders,
checkpoints on one path and paired contexts do not increase the number of
training replicas. Averaging independently sampled corpora would add the
between-corpus mean covariance of Proposition~\ref{model:prop:corpus-conditioning};
the two disjoint partitions do not measure that outer population sector.
Sections~\ref{model:sec:critical-onepass-results} and
\ref{model:sec:critical-independent-results} give the two families' designs
and results, including their pairing and replication counts.

A thermodynamic sequence must specify $N$, initialization normalization,
horizon $T_N$, batch $B_N$, source size $M_N$, consumed fraction
$q_N=B_NT_N/M_N$, both learning-rate groups and both Adam memory clocks.
Fixed Adam coefficients keep memory fixed in updates; coefficients
$\beta_i(N)=e^{-\gamma_i h_N}$ keep finite memory in the rescaled step
$h_N\to0$. These are different families. A consuming finite source has no
nonterminal stationary law without refill. Section~\ref{model:sec:data-resource}
proves the resource and coupling statements, and
Section~\ref{model:sec:theory-synthesis} states the limiting hypotheses.
The completed measurements use their recorded finite family and are not
relabelled as a different joint limit.

\section{Relation to optimizer dynamics and collapse theories}
\label{row:sec:related-work}

PLDR-LLM and PLGA introduce a learned power-law attention geometry and a row
map that can replace part of its generating network at inference
\cite{gokden2021,gokden2024,gokden2025,gokden2026foundations}.  The present
problem is the training-time selection of a row-constant quotient.  It is not
an architectural expressivity theorem and it is not implied by the inference
replacement identity.

Layer normalization subtracts a feature mean and rescales the centered
component \cite{ba2016layernorm}.  Its role in attention expressivity has
also been studied directly \cite{brody2023layernorm}.  We retain the positive
regularizer used by the implementation, which makes all displayed
normalization denominators positive.  The final affine gate multiplies
normalized features after this operation.  This order produces the exact
gate-shape factorization in Section~\ref{row:sec:row-geometry}, but the
factorization is kinematic and does not determine training dynamics.

Rank collapse in pure self-attention concerns token-representation rank as
network depth grows \cite{dong2021rankcollapse,noci2022signal}.  Attention
masks and LayerNorm change that behavior \cite{wu2024masks}, and transformer
oversmoothing depends on learned spectra \cite{dovonon2024oversmoothing}.
Attention entropy collapse instead concerns concentration of softmax
probabilities and training instability \cite{zhai2023entropy}.  PLDR row-map
collapse differs in three respects.  Its row index belongs to a learned
\(64\times64\) geometry inside one attention head, the common row is not
required to vanish, and the limiting variable is optimizer time rather than
network depth.

Adam and AdamW retain coordinatewise moment memory and adaptive
preconditioning \cite{kingma2015adam,loshchilov2019adamw}.  Decoupled weight
decay acts in addition to the adaptive write.  Even low-dimensional Adam
systems can cycle \cite{bock2023cycles}.  Replacing the optimizer by gradient
flow, freezing its denominator, or deleting moment coordinates can therefore
change the normal successor.  We derive the smooth-cell derivative in the
implementation order and explicitly separate the full state from its
final-gate subblock (Section~\ref{row:sec:smooth-adamw}).

The ordered estimates in Section~\ref{row:sec:complete-state} are discrete
variation-of-constants formulas for nonautonomous recurrences
\cite{elaydi2005}.  Chronological products and changing metrics are standard
tools for switched and nonnormal systems \cite{jungers2009}.  The
model-specific work is to identify the row-collapse stratum, construct the
implemented successor, retain its invariance defect, and connect its normal
coordinate to a physical row map.  A generic Duhamel identity alone does not
supply those ingredients.

The radial and tangential energy split used here is the orthogonal projection
of an executed finite secant onto its current observable state.  It is not an
eigendecomposition of the optimizer Jacobian, a stable-manifold assertion,
or a continuous-time Lyapunov derivative.  Its gain identity is exact on each
positive-energy edge.  The zero-face affine coordinate is likewise an exact
scalar variation-of-constants representation, distinguished from an
inequality-based attraction estimate by its canonical restart term.
Sections~\ref{row:sec:radial-normal-form} and \ref{row:sec:face-restarts}
give these two exact constructions.

Positive variation and bounded variation provide useful scalar path
envelopes \cite{folland1999}.  Our exact block excursion criterion is
stronger as a characterization because it is both necessary and sufficient
for the declared orbit.  Positive variation remains a conservative
sufficient budget.  This distinction matters for paths that oscillate
inside a shrinking envelope. Theorem~\ref{row:thm:block-orbit-criterion}
in Section~\ref{row:sec:row-geometry} proves the exact excursion criterion.

Endpoint divided differences and interval bounds have separate roles.
Divided differences give exact finite identities on realized branches.
Outward interval arithmetic can enclose floating-point evaluation or local
derivatives \cite{moore2009}.  Neither form of arithmetic establishes that a
training orbit enters or remains in an invariant tube.  We therefore keep
algebraic identity checks, numerical-resolution checks, and dynamical tests
as separate evidence classes.

\subsection{Probabilistic reduction and the architecture-specific contribution}
Probabilistic renormalization and composition are established tools
\cite{jona2001}. The contribution here is the native state retained under
elimination, its row-face bookkeeping, and explicit tests of predictive
closure. The without-replacement covariance identity is likewise standard;
its role is to expose corpus resource depletion and the assumptions needed
to transport it through adaptive training
(Section~\ref{model:sec:data-resource}). The executed predictive-closure
tests are in Sections~\ref{rg:sec:matched-fiber-results} and
\ref{rg:sec:staged-closure-results}.

Attention Gaussian-process limits depend on the architecture and width
scaling \cite{hron2020}. The fresh-projection initialization construction
in Section~\ref{model:sec:initialization-limit} is separate from the trained
sign-orbit mixture in Theorem~\ref{model:thm:native-sign-gaussian-mixture}.
Neither conclusion asserts that trained heads are independent.
Exchangeability-based rank calibration is also established
\cite{shafer2008}; here its application specifies frozen-state restoration,
continuation scores, calibration and validation units, and measured widths.
Its coverage averages over the calibration draw and does not hold for every
realized tube at every state. Section~\ref{model:sec:law-closure-results}
gives the continuation design, rank calibration, and validation outcomes.

Operator caching originates in the PLDR architecture work
\cite{gokden2025}. The present cache analysis concerns state transport,
recalibration, signed risk accounting, and fresh-context fidelity
(Section~\ref{model:sec:cache-risk-law} for the risk law and
Sections~\ref{model:sec:cache-risk-results} and
\ref{model:sec:cache-state-results} for the experiments). The
broader criticality interpretation in \cite{gokden2026foundations} motivates
tests but is not supplied by a learned power parameter. A diverging
correlation length, autonomous critical attractor, and universal measure
of reasoning are not established by the finite evidence here.

\FloatBarrier
\par\medskip\noindent
With the forward state and conditioning specified,
Chapter~\ref{ch:row-geometry} isolates the row-constant geometry and explains
how to observe its absolute energy with certified numerical bounds.

\chapter{Row geometry, absolute energy, and numerical observation}
\label{ch:row-geometry}
This chapter defines row collapse as a geometric property of the
implemented map. It develops the row quotient, the mixed gate and shape
characterization, and numerical enclosures that distinguish exact collapse
from a small or unresolved measured energy.

\section{Implemented row map and exact collapse geometry}
\label{row:sec:row-geometry}

The implemented geometry, the collapse criterion along a trajectory, and the
exact energy strata are established here. Section~\ref{rg:sec:row-geometry}
then specifies what finite-precision observations can certify.

\subsection{Implemented map}

Fix one registered context, decoder layer, and attention head.  The rotary
query matrix \(Q_{\rm rope}\in\R^{t\times d}\) supplies the unscaled Gram
input
\begin{equation}
 U=Q_{\rm rope}^{\top}Q_{\rm rope}.
 \label{row:eq:implemented-gram}
\end{equation}
The row program treats the \(r=d\) rows of \(U\) separately, with shared
learned weights; the reference row experiments use \(d=64\).  Represent each row by a column vector \(x\in\R^d\) during the rowwise calculation, and put
\begin{gather}
 P_d=I_d-d^{-1}\mathbf1\mathbf1^\top, \quad c(x)=P_dx,
 \label{row:eq:regularized-layernorm}\\
 s(x)=\left(\epsilon_{\rm LN}+d^{-1}\norm{c(x)}^2\right)^{1/2}, \quad \nu(x)=c(x)/s(x),\notag
\end{gather}
with \(\epsilon_{\rm LN}=10^{-6}\).  Affine LayerNorm is
\(
 \mathcal N_{\gamma,\beta}(x)=\gamma\odot\nu(x)+\beta.
\)

One gated block has the implemented order
\begin{equation}
 G(z;\psi)=W_3\!\left[
 \operatorname{silu}(W_1z+b_1)\odot(W_2z+b_2)
 \right]+b_3.
 \label{row:eq:implemented-gated-block}
\end{equation}
Residual unit \(j\) applies two such blocks followed by affine LayerNorm:
\begin{align}
 x_0&=\mathcal N_{\gamma_0,\beta_0}(u),\nonumber\\
 w_j&=x_{j-1}+G_{j,2}(G_{j,1}(x_{j-1})),\nonumber\\
 x_j&=\mathcal N_{\gamma_j,\beta_j}(w_j),
 \qquad 1\le j\le8.
 \label{row:eq:implemented-row-program}
\end{align}
The physical row map is the matrix \(X\) whose rows are the corresponding
\(x_8\).  The final gate \(\gamma_8\) has one \(d\)-vector per decoder
layer and is shared across that layer's heads.  Upstream normalized
shapes remain context and head specific.

The regularizer makes \(s(x)>0\) for every input.  It also gives
\begin{equation}
 \norm{\nu(x)}^2
 =\frac{\norm{c(x)}^2}
 {\epsilon_{\rm LN}+d^{-1}\norm{c(x)}^2}<d.
 \label{row:eq:layernorm-range}
\end{equation}
This is a global boundedness fact, not a mechanism for row collapse.

\subsection{Physical row quotient}

For \(X\in\R^{r\times d}\), define
\begin{gather}
 C_r=I_r-r^{-1}\mathbf1\mathbf1^\top, \quad J(X)=\norm{C_rX}_{\Fro},
 \label{row:eq:physical-quotient-observables}\\
 E(X)=J(X)^2, \quad D(X)=\max_{i<j}\norm{X_{i:}-X_{j:}}_2.\notag
\end{gather}
Write \(Z=\rowq X:=C_rX\) for the centered physical row map.
The geometric statements below hold for arbitrary finite \(r\) and \(d\).

\begin{lemma}[Physical quotient and row diameter]
\label{row:lem:row-quotient-geometry}
\label{rg:lem:quotient-diameter}
For every \(r\ge2\),
\begin{equation}
 \sum_{i<j}\norm{X_{i:}-X_{j:}}_2^2=rE(X),
 \qquad
 D(X)^2\le2E(X)\le(r-1)D(X)^2.
 \label{row:eq:row-quotient-equivalence}
\end{equation}
Consequently, for fixed finite \(r\), \(J(X_t)\to0\), \(E(X_t)\to0\),
and \(D(X_t)\to0\) are equivalent.
\end{lemma}

\begin{proof}
Let \(\bar x=r^{-1}\sum_iX_{i:}\).  Expanding all unordered pairs gives
\[
 \sum_{i<j}\norm{X_{i:}-X_{j:}}^2
 =r\sum_i\norm{X_{i:}}^2-\norm{\textstyle\sum_iX_{i:}}^2
 =r\sum_i\norm{X_{i:}-\bar x}^2=rE(X).
\]
If \(i,j\) attain the diameter, then
\[
 D(X)^2\le
 2\norm{X_{i:}-\bar x}^2+2\norm{X_{j:}-\bar x}^2
 \le2E(X).
\]
Conversely, the pairwise sum contains \(r(r-1)/2\) terms, each at most
\(D(X)^2\).  Dividing its upper bound by \(r\) gives
\(E(X)\le(r-1)D(X)^2/2\).  The convergence equivalences follow by
squeezing between fixed positive constants.
\end{proof}

The quotient removes only common-row translations:
\begin{equation}
 C_r(X+\mathbf1a^\top)=C_rX.
 \label{row:eq:common-row-unrestricted}
\end{equation}
The learned common row can therefore retain nontrivial geometry at collapse.

\subsection{Layer-resolved gate-shape characterization}
\label{row:sec:gate-shape-geometry}

Let \(Y\in\R^{r\times d}\) contain the normalized rows immediately before
the final affine LayerNorm, let \(Y_c=C_rY\), and suppress the layer index on
the shared gate.  The bias is row constant, so
\begin{equation}
 X=Y\diag(\gamma)+\mathbf1\beta^\top,\qquad
 C_rX=Y_c\diag(\gamma).
 \label{row:eq:final-layernorm-factorization}
\end{equation}
Define the coordinate shape energy
\(
 \upsilon_k=\norm{Y_{c,:k}}_2^2
\)
and physical coordinate energy \(e_k=\gamma_k^2\upsilon_k\).

\begin{theorem}[Layer-resolved mixed gate-shape collapse]
\label{row:thm:layer-resolved-gate-shape}
\label{rg:thm:mixed-gate-shape}
For every implemented row map,
\begin{equation}
 E(X)=\sum_{k=1}^{d}e_k
 =\sum_{k=1}^{d}\gamma_k^2\upsilon_k.
 \label{row:eq:layer-resolved-energy-factorization}
\end{equation}
For a sequence with fixed finite \(d\), physical row-map collapse occurs if
and only if
\begin{equation}
 \abs{\gamma_{k,t}}\sqrt{\upsilon_{k,t}}\longrightarrow0
 \quad\text{for every }k.
 \label{row:eq:mixed-coordinate-criterion}
\end{equation}
Neither \(\gamma_{k,t}\) nor \(\upsilon_{k,t}\) must converge separately.
\end{theorem}

\begin{proof}
Centering Equation~\eqref{row:eq:final-layernorm-factorization} removes the
bias.  The squared Frobenius norm is the sum of the squared Euclidean norms
of the columns, and column \(k\) is \(\gamma_kY_{c,:k}\).  This proves
Equation~\eqref{row:eq:layer-resolved-energy-factorization}.  Its \(d\) summands
are nonnegative.  A finite sum of nonnegative sequences tends to zero if and
only if every summand tends to zero.  Taking nonnegative square roots gives
Equation~\eqref{row:eq:mixed-coordinate-criterion};
Lemma~\ref{row:lem:row-quotient-geometry} converts energy convergence to physical
row-diameter convergence.
\end{proof}

On an edge where all three quantities are positive, the coordinate identity
has the exact logarithmic form
\begin{equation}
 \log\frac{e_{k,t+1}}{e_{k,t}}
 =2\log\frac{\abs{\gamma_{k,t+1}}}{\abs{\gamma_{k,t}}}
 +\log\frac{\upsilon_{k,t+1}}{\upsilon_{k,t}}.
 \label{row:eq:coordinate-log-cocycle}
\end{equation}
At a zero, the absolute energies remain primary and the log ratio is not
defined.  This zero policy is essential near the row-constant face.

An effective squared gate can be defined without dividing coordinatewise:
\begin{equation}
 a_{m,t}=\frac{E_{m,t}}{S_{m,t}},\qquad
 S_{m,t}=\sum_k\upsilon_{m,k,t},
 \label{row:eq:effective-squared-gate}
\end{equation}
whenever \(S_{m,t}>0\).  This is an occupancy-weighted descriptive factor.
It does not isolate a gate intervention because the occupancy weights also
move.

\subsection{Exact orbit criterion}

Kinematic endpoint classification does not control reopenings between
endpoints.  Let \(E_t\ge0\) be one map's energy and let
\(0\le\tau_0<\tau_1<\cdots\) be a fixed, strictly increasing, unbounded block
sequence.  Define
\begin{equation}
 M_j=\max_{\tau_j\le t\le\tau_{j+1}}E_t,\qquad
 A_j=M_j-E_{\tau_j},\qquad
 B_j=\sum_{t=\tau_j}^{\tau_{j+1}-1}(E_{t+1}-E_t)_+.
 \label{row:eq:block-excursion-definitions}
\end{equation}
Here \(A_j\) is exact excursion above the left anchor and \(B_j\) is positive
variation.

\begin{theorem}[Necessary and sufficient block orbit criterion]
\label{row:thm:block-orbit-criterion}
For every such partition,
\begin{equation}
 E_t\longrightarrow0
 \quad\Longleftrightarrow\quad
 E_{\tau_j}\longrightarrow0
 \ \text{and}\ A_j\longrightarrow0.
 \label{row:eq:block-orbit-iff}
\end{equation}
Moreover \(0\le A_j\le B_j\).  Hence vanishing anchor energy and vanishing
positive variation are sufficient but positive variation is not necessary.
\end{theorem}

\begin{proof}
The block is finite, so its maximum is attained and
\(M_j=E_{\tau_j}+A_j\).  If \(E_t\to0\), both the subsequence
\(E_{\tau_j}\) and the maximum over every sufficiently late finite block
tend to zero.  Since \(0\le A_j\le M_j\), the right side follows.

Conversely, suppose the two right-side sequences vanish.  Given
\(\varepsilon>0\), choose \(j_0\) such that
\(E_{\tau_j}+A_j<\varepsilon\) for \(j\ge j_0\).  Every
\(t\ge\tau_{j_0}\) lies in some block with index \(j\ge j_0\), and
\[
 0\le E_t\le M_j=E_{\tau_j}+A_j<\varepsilon.
\]
Thus \(E_t\to0\).  Finally, telescoping inside a block gives
\[
 E_t-E_{\tau_j}
 =\sum_{s=\tau_j}^{t-1}(E_{s+1}-E_s)
 \le\sum_{s=\tau_j}^{t-1}(E_{s+1}-E_s)_+\le B_j.
\]
Taking the maximum positive excursion proves \(A_j\le B_j\).
\end{proof}

For a finite registry \(\mathcal M\) of contexts, layers, and heads, define
the stacked energy
\begin{equation}
 E_t^{\mathcal M}=\sum_{m\in\mathcal M}E_{m,t}.
 \label{rg:eq:stacked-energy}
\end{equation}
Because the registry is finite and every summand is nonnegative,
\(E_t^{\mathcal M}\to0\) is equivalent to collapse of every registered map.
Nothing in this finite statement gives a uniform result over unseen contexts.
Such an extension requires a separate physical cover, introduced in
Section~\ref{row:sec:complete-state}.

\subsection{Exact face and energy strata}

An empirical observer must not turn a small positive row map into an exact
face hit.  Let \(Z\) be any centered row map in the finite registry, write
\(E(Z)=\norm{Z}_{\Fro}^2\), and fix an effect floor
\(\epsilon_\ast>0\).  Define
\begin{equation}
 \mathcal F_0=\{Z:E(Z)=0\},\qquad
 \mathcal C_{\epsilon_\ast}=\{Z:0<E(Z)\le\epsilon_\ast\},\qquad
 \mathcal R_{\epsilon_\ast}=\{Z:E(Z)>\epsilon_\ast\}.
 \label{row:eq:observer-strata}
\end{equation}
The first set is the exact row-constant face in centered coordinates.  The
second contains floor-censored positive states.  The third contains resolved
positive states, subject to the remaining technical checks of a protocol.

\begin{theorem}[Exact observer-stratum partition]
\label{row:thm:observer-stratum-partition}
The three sets in Equation~\eqref{row:eq:observer-strata} are pairwise disjoint
and exhaust the centered row-map space.  Moreover,
\(\mathcal F_0=\{\bm0_{r\times d}\}\).  In particular, membership in
\(\mathcal C_{\epsilon_\ast}\) implies \(Z\ne\bm0_{r\times d}\) and cannot activate a
theorem whose premise is \(E(Z)=0\).
\end{theorem}

\begin{proof}
Energy is a finite sum of nonnegative real squares.  It is zero if and only
if every coordinate of \(Z\) is zero, so \(\mathcal F_0=\{\bm0_{r\times d}\}\).
For every nonnegative real number \(e\), exactly one of
\[
 e=0,\qquad 0<e\le\epsilon_\ast,\qquad e>\epsilon_\ast
\]
holds.  Apply this trichotomy to \(e=E(Z)\).  Exhaustion and pairwise
disjointness follow.  The middle case includes the strict inequality
\(E(Z)>0\), hence excludes the exact face.
\end{proof}

\section{Certified numerical observation of row collapse}
\label{rg:sec:row-geometry}

The centered map $Z=\rowq X$ and its energy $E(X)$ are defined in
Section~\ref{row:sec:row-geometry}. Lemma~\ref{row:lem:row-quotient-geometry}
identifies energy decay with row-diameter decay, and
Theorem~\ref{row:thm:layer-resolved-gate-shape} gives the exact gate-shape
criterion. The question here is what a computed energy establishes about
that exact geometry. The energy and observer information supply the
chronological construction in Section~\ref{rg:sec:exact-rg}.

\subsection{Certified energy enclosures}

The exact face and positive-energy strata are defined in
Equation~\eqref{row:eq:observer-strata} and characterized by
Theorem~\ref{row:thm:observer-stratum-partition}.
A finite-precision zero is a statement about a represented computation.
An analytical face claim additionally requires an exact argument or an
error enclosure that certifies zero energy.

\begin{theorem}[Certified quotient-energy enclosure]
\label{rg:thm:observer-energy-enclosure}
Let \(E\ge0\) be the exact quotient energy, let \(\widehat E\in\R\) be a
computed value, and suppose a certified absolute error \(\delta\ge0\)
satisfies \(\abs{E-\widehat E}\le\delta\). Define
\begin{equation}
 I(\widehat E,\delta)=[L,U]
 =\bigl[\max\{0,\widehat E-\delta\},\widehat E+\delta\bigr].
 \label{rg:eq:observer-energy-interval}
\end{equation}
Then \(E\in[L,U]\). If \(U=0\), the state is a certified analytical face;
if \(L>0\), it is certified positive; and if \(L=0<U\), it is an
observer-near-face state whose branch is unresolved by this enclosure.
For a sequence of valid enclosures \([L_t,U_t]\), the condition
\(U_t\to0\) implies \(E_t\to0\). A fixed tolerance statement
\(U_t\le\epsilon_*\) does not imply convergence as \(t\to\infty\).
\end{theorem}

\begin{proof}
The error bound gives
\(\widehat E-\delta\le E\le\widehat E+\delta\). Combining the lower
inequality with \(E\ge0\) proves membership in
Equation~\eqref{rg:eq:observer-energy-interval}. If \(U=0\), membership and
nonnegativity force \(E=0\). If \(L>0\), membership forces \(E>0\). The
remaining nondegenerate case contains zero and positive values, so neither
branch follows from the interval alone. Finally, \(0\le E_t\le U_t\), and
the squeeze theorem proves the convergence assertion. The constant sequence
\(E_t=\epsilon_*/2\) proves that a fixed tolerance is insufficient.
\end{proof}

A numerical assignment to the strata in Equation~\eqref{row:eq:observer-strata}
requires a valid enclosure or an independent exact argument. Bitwise row
equality is reported separately. A positive subfloor state remains on the
positive canonical branch, while an observer-near-face state does not support
either a gain ratio or a restart claim without additional information.
Section~\ref{rg:sec:interval-affine-observer} derives the interval blocking
rule, and Section~\ref{rg:sec:observer-results} reports the stage-resolved
replays that distinguish represented zeros from analytical face claims.

\subsection{Arithmetic resolution and the native estimand}
\label{row:sec:observer-resolution}

Equation~\eqref{row:eq:observer-strata} is an observer interface, not a new
dynamical law.  Mathematical radial coordinates are defined whenever
\(E(Z)>0\).  A protocol may withhold relative inference throughout
\(\mathcal C_{\epsilon_\ast}\) because division by a small stored energy is
poorly conditioned.  Absolute energy and the unnormalized work-charge ledger
remain defined there. Section~\ref{row:sec:radial-normal-form} defines the
radial coordinates, while Section~\ref{row:sec:finite-work} supplies the
branch-free work-charge ledger.

The exact identities above are real-arithmetic statements.  Near collapse,
relative floating-point diagnostics can be undefined or misleading.  Given
a float64 reference \(E^{64}\) and float32 replay \(E^{32}\), the experiment
declares the absolute arithmetic floor
\begin{equation}
 \delta_{\rm arith}=8\abs{E^{64}-E^{32}}.
 \label{row:eq:arithmetic-resolution-floor}
\end{equation}
A value is called resolved only when
\(\abs{E^{64}}>\delta_{\rm arith}\).  Absolute energy and absolute
factorization residual remain reportable below that floor.  Relative gains
and zeros from the lower-precision path do not.
This replay discrepancy measures disagreement between two arithmetic paths.
It does not by itself provide the certified bound on exact energy assumed in
Theorem~\ref{rg:thm:observer-energy-enclosure}; the two paths can share an
error. Such a bound requires a separate error analysis.
Section~\ref{row:sec:precision-endpoints} reports the float32/float64 replay
and its coordinate-zero census; Section~\ref{row:sec:observer-block-analysis}
reports the stored-energy strata and finite-block measurements.

For a causal endpoint comparison, the natively executed centered rows and
their native energies are the primary estimand.  Reconstructing the same
endpoint as \(Y_c\diag(\gamma)\) is an exact real-arithmetic identity but a
finite-precision diagnostic, because the two evaluation paths can accumulate
rounding in a different order.  Section~\ref{row:sec:direct-energy} therefore
retains their implementation defect explicitly and gives the native
four-source ledger.  A factorization discrepancy cannot by itself invalidate
a native causal contrast when the native work, effect, execution, and replay
checks close.

\FloatBarrier
\par\medskip\noindent
Chapter~\ref{ch:finite-work} turns this geometric observable into an exact
finite-step balance. It identifies the parameter sources, interactions and
face restarts that change the energy defined here.

\chapter{Finite work, parameter sources, and face restarts}
\label{ch:finite-work}
This chapter derives the exact work ledger for finite changes of the
row-centered map. It resolves gate, shape and optimizer sources, retains
their interactions, and treats both contraction at positive energy and
reopening from the row-constant face.

\section{Exact direct observable-energy dynamics}
\label{row:sec:direct-energy}

This section derives an exact finite-step balance for physical row energy
directly from the observed row maps.  The balance exposes every source of
energy change without a Taylor truncation.  It neither closes the optimizer
state nor assumes that the row-constant face is invariant.  The complementary
complete-state cocycle in Section~\ref{row:sec:complete-state} resolves how
hidden optimizer coordinates transport perturbations.

Fix a finite registry \(\mathcal M\) of contexts, layers, and heads.  Stack
the natively evaluated centered row maps into the finite-dimensional
Frobenius space
\[
 \mathcal Y=\bigoplus_{m\in\mathcal M}\R^{r_m\times d_m},
 \qquad Z_t=Z_t^{\rm nat}=(C_{r_m}X_{m,t})_{m\in\mathcal M},
 \qquad E_t=\norm{Z_t}_{\Fro}^{2}.
\]
The inner product below is the sum of the Frobenius inner products over the
registry.

\subsection{The exact physical work ledger}
\label{row:sec:finite-work}

Let \(D_t=Z_{t+1}-Z_t\) be the executed physical increment.  Define the
dissipative work and finite-step charge by
\begin{equation}
 \mathcal W_t=-2\ip{Z_t}{D_t},\qquad
 \mathcal Q_t=\norm{D_t}_{\Fro}^{2}.
 \label{row:eq:direct-work-charge-definition}
\end{equation}
Positive \(\mathcal W_t\) means that the linear part of the increment points
toward the row-constant face.  The charge is always nonnegative.

\begin{theorem}[Exact physical work-charge identity]
\label{row:thm:exact-direct-work-charge}
For every implemented update, with no small-step assumption,
\begin{equation}
 E_{t+1}-E_t=-\mathcal W_t+\mathcal Q_t.
 \label{row:eq:exact-direct-work-charge}
\end{equation}
If the physical increment has any finite source decomposition
\(D_t=\sum_{a\in\mathcal A}D_t^{(a)}\), then
\begin{alignat}{2}
 \mathcal W_t
   &=\sum_a\mathcal W_t^{(a)},
 &\quad \mathcal W_t^{(a)}&=-2\ip{Z_t}{D_t^{(a)}},
 \label{row:eq:direct-source-work}\\
 \mathcal Q_t
   &=\sum_a\norm{D_t^{(a)}}_{\Fro}^{2}
     +2\sum_{a<b}\ip{D_t^{(a)}}{D_t^{(b)}}.
 \label{row:eq:direct-source-gram}
\end{alignat}
Thus a source ledger is complete only when it retains the cross-Gram terms.
\end{theorem}

\begin{proof}
Since \(Z_{t+1}=Z_t+D_t\), expansion of the squared norm gives
\[
 \norm{Z_t+D_t}_{\Fro}^{2}-\norm{Z_t}_{\Fro}^{2}
 =2\ip{Z_t}{D_t}+\norm{D_t}_{\Fro}^{2}
 =-\mathcal W_t+\mathcal Q_t.
\]
Linearity of the inner product in its second argument proves
Equation~\eqref{row:eq:direct-source-work}.  Expanding
\(\norm{\sum_aD_t^{(a)}}^2\) gives the diagonal terms and twice every
unordered cross term, which proves Equation~\eqref{row:eq:direct-source-gram}.
\end{proof}

The identity is stronger than a differential approximation and weaker than
an attraction theorem.  A negative energy increment can result from direct
dissipative work, destructive interaction among source increments, or both.
Conversely, \(\mathcal W_t>0\) does not imply \(E_{t+1}<E_t\) when the
finite-step charge is larger.

\begin{theorem}[Branch-free work-charge classification]
\label{row:thm:observer-safe-work-charge}
Let \(Z\) and \(D\) be arbitrary finite Frobenius vectors, and set
\[
 E=\norm{Z}_{\Fro}^{2},\qquad
 E^+=\norm{Z+D}_{\Fro}^{2},\qquad
 \mathcal W=-2\ip{Z}{D},\qquad
 \mathcal Q=\norm{D}_{\Fro}^{2}.
\]
Then all four quantities are defined without dividing by \(E\), and
\begin{alignat}{3}
 E^+<E&\quad\Longleftrightarrow\quad \mathcal W>\mathcal Q,
 &\quad E^+=E&\quad\Longleftrightarrow\quad \mathcal W=\mathcal Q,
 &\quad E^+>E&\quad\Longleftrightarrow\quad \mathcal W<\mathcal Q.
 \label{row:eq:branch-free-work-charge-classification}
\end{alignat}
At the exact row-constant face \(E=0\), one has \(Z=\bm0_{r\times d}\),
\(\mathcal W=0\), and \(E^+=\mathcal Q\).  Consequently, strict
contraction from the face is impossible, and the face is preserved exactly
when \(D=\bm0_{r\times d}\).
\end{theorem}

\begin{proof}
Direct expansion gives
\[
 E^+-E=2\ip{Z}{D}+\norm{D}_{\Fro}^{2}
       =-\mathcal W+\mathcal Q.
\]
The three equivalences follow by comparing this real number with zero.
If \(E=\norm{Z}_{\Fro}^{2}=0\), positive definiteness of the finite
Frobenius norm gives \(Z=\bm0_{r\times d}\).  Substitution gives \(\mathcal W=0\) and
\(E^+=\norm{D}_{\Fro}^{2}=\mathcal Q\).  Finally,
\(\mathcal Q=0\) holds exactly when \(D=\bm0_{r\times d}\).
\end{proof}

This classification is the native estimand at every observer stratum in
Theorem~\ref{row:thm:observer-stratum-partition}.  Normalized radial coordinates
are mathematically defined whenever the stored energy is strictly positive,
but they need not be reported when the observer cannot resolve that energy
from its numerical floor.

\begin{theorem}[Interval-certified work-charge decision]
\label{row:thm:interval-certified-work-charge}
Suppose computed values \(\widehat{\mathcal W}\) and
\(\widehat{\mathcal Q}\) have verified absolute error bounds
\[
 |\widehat{\mathcal W}-\mathcal W|\le\delta_{\mathcal W},
 \qquad
 |\widehat{\mathcal Q}-\mathcal Q|\le\delta_{\mathcal Q}.
\]
Then
\begin{alignat}{2}
 \widehat{\mathcal W}-\delta_{\mathcal W}
   &>\widehat{\mathcal Q}+\delta_{\mathcal Q}
   &\quad &\Longrightarrow E^+<E,
 \label{row:eq:interval-certified-contraction}\\
 \widehat{\mathcal Q}-\delta_{\mathcal Q}
   &>\widehat{\mathcal W}+\delta_{\mathcal W}
   &\quad &\Longrightarrow E^+>E.
 \label{row:eq:interval-certified-reopening}
\end{alignat}
If the two intervals overlap, these bounds alone do not certify a sign.
Likewise, fix the observer floor \(\epsilon_\ast>0\).  For any verified
energy enclosure \(0\le E_-\le E\le E_+\), the following implications hold:
\begin{align*}
 E_+=0 &\Longrightarrow E=0,\\
 E_->0 &\Longrightarrow E>0,\\
 0<E_-\ \text{and}\ E_+\le\epsilon_\ast
   &\Longrightarrow 0<E\le\epsilon_\ast,\\
 E_->\epsilon_\ast &\Longrightarrow E>\epsilon_\ast.
\end{align*}
Thus a positive lower endpoint certifies strict positivity, while resolved
membership requires the stronger bound \(E_->\epsilon_\ast\).  An enclosure
that intersects more than one observer stratum does not certify membership in
either adjacent stratum from these bounds alone.

For approximate endpoints \(\widehat Z,\widehat D\), valid deterministic
error bounds are
\begin{align}
 |\mathcal W(Z,D)-\mathcal W(\widehat Z,\widehat D)|
 &\le 2\bigl(
 \norm{Z-\widehat Z}_{\Fro}\norm{D}_{\Fro}
 +\norm{\widehat Z}_{\Fro}\norm{D-\widehat D}_{\Fro}\bigr),
 \label{row:eq:work-stability-bound}\\
 |\mathcal Q(D)-\mathcal Q(\widehat D)|
 &\le
 \bigl(\norm{D}_{\Fro}+\norm{\widehat D}_{\Fro}\bigr)
 \norm{D-\widehat D}_{\Fro}.
 \label{row:eq:charge-stability-bound}
\end{align}
\end{theorem}

\begin{proof}
The assumed errors place \(\mathcal W\) and \(\mathcal Q\) in their
respective closed intervals.  The first strict interval separation implies
\(\mathcal W>\mathcal Q\), and the second implies the reverse inequality.
Theorem~\ref{row:thm:observer-safe-work-charge} gives the two energy decisions.
Overlapping work-charge intervals contain values in either order, so no
order follows from those enclosures alone.  For the energy enclosure,
nonnegativity and \(E\le E_+=0\) force \(E=0\).  If \(E_->0\), then
\(E\ge E_->0\).  Adding \(E_+\le\epsilon_\ast\) gives
\(0<E\le\epsilon_\ast\), while \(E_->\epsilon_\ast\) gives
\(E>\epsilon_\ast\).  These are precisely the three observer-stratum
conditions.  If an enclosure intersects a stratum boundary, it contains
admissible energies on both sides of that boundary, so the enclosure alone
cannot select one adjacent stratum.

For the work bound, add and subtract
\(-2\ip{\widehat Z}{D}\), use bilinearity, and apply Cauchy--Schwarz:
\[
 |\mathcal W(Z,D)-\mathcal W(\widehat Z,\widehat D)|
 \le2|\ip{Z-\widehat Z}{D}|
    +2|\ip{\widehat Z}{D-\widehat D}|.
\]
For the charge bound, use
\(
 |\norm{D}_{\Fro}^{2}-\norm{\widehat D}_{\Fro}^{2}|
 =|\norm D_{\Fro}-\norm{\widehat D}_{\Fro}|
  (\norm D_{\Fro}+\norm{\widehat D}_{\Fro})
\)
and the reverse triangle inequality.
\end{proof}

\subsection{Exact gate, shape, interaction, and arithmetic-defect secants}

Write the centered pre-affine shape as \(U_t=C_rY_t\), the final LayerNorm
gate as \(\Gamma_t=\diag(\gamma_t)\), and the factorized evaluation as
\(\widetilde Z_t=U_t\Gamma_t\).  In exact real arithmetic,
\(Z_t=\widetilde Z_t\).  For a finite-precision native evaluation, define the
implementation defect
\begin{equation}
 \delta_t=Z_t^{\rm nat}-\widetilde Z_t.
 \label{row:eq:implementation-factorization-defect}
\end{equation}
The defect is an arithmetic source, not an additional architectural
mechanism.  Define
\begin{align}
 D_t^{\rm shape}&=(U_{t+1}-U_t)\Gamma_t,\nonumber\\
 D_t^{\rm gate}&=U_t(\Gamma_{t+1}-\Gamma_t),\nonumber\\
 D_t^{\rm int}&=(U_{t+1}-U_t)(\Gamma_{t+1}-\Gamma_t),\nonumber\\
 D_t^{\rm def}&=\delta_{t+1}-\delta_t.
 \label{row:eq:gate-shape-secant-components}
\end{align}

\begin{proposition}[Exact gate-shape secant]
\label{row:prop:exact-gate-shape-secant}
For every pair of endpoints, the factorized and native increments obey
\begin{equation}
 \widetilde Z_{t+1}-\widetilde Z_t
 =D_t^{\rm shape}+D_t^{\rm gate}+D_t^{\rm int},
 \qquad
 D_t=D_t^{\rm shape}+D_t^{\rm gate}+D_t^{\rm int}+D_t^{\rm def}.
 \label{row:eq:exact-gate-shape-secant}
\end{equation}
Let
\(\mathcal A_4=\{\mathrm{shape},\mathrm{gate},\mathrm{int},\mathrm{def}\}\).
Consequently, the complete native energy ledger is
\begin{equation}
 E_{t+1}-E_t
 =2\sum_{a\in\mathcal A_4}\ip{Z_t}{D_t^{(a)}}
  +\sum_{a,b\in\mathcal A_4}\ip{D_t^{(a)}}{D_t^{(b)}}.
 \label{row:eq:exact-gate-shape-energy-ledger}
\end{equation}
These formulas remain defined when a gate, shape coordinate, or physical
energy is zero.
\end{proposition}

\begin{proof}
Add and subtract \(U_{t+1}\Gamma_t\) and then expand both factors:
\begin{align*}
 U_{t+1}\Gamma_{t+1}-U_t\Gamma_t
 &=(U_{t+1}-U_t)\Gamma_t
   +U_{t+1}(\Gamma_{t+1}-\Gamma_t)\\
 &=(U_{t+1}-U_t)\Gamma_t
   +U_t(\Gamma_{t+1}-\Gamma_t)
   +(U_{t+1}-U_t)(\Gamma_{t+1}-\Gamma_t).
\end{align*}
This proves the first identity in
Equation~\eqref{row:eq:exact-gate-shape-secant}.  Subtracting the two endpoint
instances of Equation~\eqref{row:eq:implementation-factorization-defect} gives
the native identity.  Substitution into
Theorem~\ref{row:thm:exact-direct-work-charge}, followed by expansion of the full
four-source squared norm, proves
Equation~\eqref{row:eq:exact-gate-shape-energy-ledger}.  No division is used.
\end{proof}

The interaction term is bilinear in the two endpoint changes.  Omitting it
turns an identity into a first-order approximation.  Likewise, reporting
four component norms without their Gram matrix cannot determine the native
charge.  In real arithmetic \(D_t^{\rm def}=\bm0_{r\times d}\), and the ledger reduces to
the three architectural secants.

\begin{proposition}[Precision-resolved collapse criterion]
\label{row:prop:precision-resolved-collapse}
For any endpoint sequence,
\begin{equation}
 Z_t^{\rm nat}\longrightarrow\bm0_{r\times d}
 \quad\Longleftrightarrow\quad
 \widetilde Z_t+\delta_t\longrightarrow\bm0_{r\times d}.
 \label{row:eq:precision-resolved-collapse}
\end{equation}
Moreover,
\begin{equation}
 \bigl|\norm{\widetilde Z_t}_{\Fro}-\norm{\delta_t}_{\Fro}\bigr|
 \le \norm{Z_t^{\rm nat}}_{\Fro}
 \le \norm{\widetilde Z_t}_{\Fro}+\norm{\delta_t}_{\Fro}.
 \label{row:eq:precision-resolved-triangle}
\end{equation}
Thus factorized collapse together with \(\delta_t\to\bm0_{r\times d}\) is sufficient for
native collapse.  Neither condition is necessary on its own, because the two
terms may cancel.
\end{proposition}

\begin{proof}
Equation~\eqref{row:eq:precision-resolved-collapse} is the definition
\(Z_t^{\rm nat}=\widetilde Z_t+\delta_t\).  The upper inequality in
Equation~\eqref{row:eq:precision-resolved-triangle} is the triangle inequality.
The two reverse triangle inequalities
\(\norm{\widetilde Z_t}\le\norm{Z_t^{\rm nat}}+\norm{\delta_t}\) and
\(\norm{\delta_t}\le\norm{Z_t^{\rm nat}}+\norm{\widetilde Z_t}\) give the
lower bound.  If both summands tend to zero, the upper bound proves native
collapse.  Conversely, choosing any nonzero sequence \(V_t\) and setting
\(\widetilde Z_t=V_t\), \(\delta_t=-V_t\) shows why separate convergence is
not necessary.
\end{proof}

\subsection{Endpoint-exact parameter sources under AdamW}

Let \(\Psi_{m,t}(\theta)\) denote the centered implemented row map for the
fixed registered input used at update \(t\).  The AdamW position increment
from Equations~\eqref{row:eq:adamw-moment-one}--
\eqref{row:eq:adamw-position} is exactly
\begin{equation}
 \Delta\theta_t
 =-\eta_t\Lambda\theta_t
 -\eta_t\frac{m_{t+1}/a_{1,t}}
 {\sqrt{v_{t+1}/a_{2,t}}+\epsilon_{\rm A}}
 =:\Delta\theta_t^{\rm wd}+\Delta\theta_t^{\rm loss}.
 \label{row:eq:adamw-exact-position-sources}
\end{equation}
Moment lag, clipping, bias correction, and the live denominator are already
inside the second endpoint displacement.  They are not frozen in the
following secant.

\begin{theorem}[Integrated implemented-parameter secant]
\label{row:thm:integrated-parameter-secant}
Suppose \(\Psi_{m,t}\) is continuously differentiable on each member of a
finite partition of the segment
\(\theta_t(\tau)=\theta_t+\tau\Delta\theta_t\), and is continuous on the
whole segment.  Then
\begin{equation}
 D_{m,t}=\Psi_{m,t}(\theta_{t+1})-\Psi_{m,t}(\theta_t)
 =\int_0^1D\Psi_{m,t}(\theta_t(\tau))[\Delta\theta_t]\,d\tau.
 \label{row:eq:integrated-parameter-secant}
\end{equation}
For any exact finite partition
\(\Delta\theta_t=\sum_a\Delta\theta_t^{(a)}\), including the two AdamW
sources in Equation~\eqref{row:eq:adamw-exact-position-sources},
\begin{equation}
 D_{m,t}=\sum_a D_{m,t}^{(a)},\qquad
 D_{m,t}^{(a)}=
 \int_0^1D\Psi_{m,t}(\theta_t(\tau))
 [\Delta\theta_t^{(a)}]\,d\tau.
 \label{row:eq:integrated-parameter-source-ledger}
\end{equation}
The same statement holds after partitioning parameters by decoder layer or
registered module.
\end{theorem}

\begin{proof}
On each smooth subinterval, the fundamental theorem of calculus applied to
\(\tau\mapsto\Psi_{m,t}(\theta_t(\tau))\) gives the integral of
\(D\Psi_{m,t}(\theta_t(\tau))[\Delta\theta_t]\).  Summing adjacent
subintervals telescopes because \(\Psi_{m,t}\) is continuous at their
boundaries, proving Equation~\eqref{row:eq:integrated-parameter-secant}.
The derivative is linear in its direction argument.  Substitute
\(\Delta\theta_t=\sum_a\Delta\theta_t^{(a)}\), distribute the derivative,
and use linearity of the finite integral to obtain
Equation~\eqref{row:eq:integrated-parameter-source-ledger}.
\end{proof}

This theorem is an endpoint identity, not a claim that the training
successor follows the artificial parameter segment.  Its purpose is to
assign an executed physical increment exactly to declared parameter
displacements.  A causal intervention requires an additional paired
successor, because a descriptive source secant alone does not identify what
would have happened had that source been removed.

\subsection{Exact paired-intervention contrast}

Consider natural and controlled successors from the same native source
\(Z_t\), with increments \(D_t^{N}\) and \(D_t^{C}\).  Put
\begin{equation}
 H_t=D_t^{C}-D_t^{N},\qquad
 I_t=\norm{Z_t+D_t^{C}}_{\Fro}^{2}
     -\norm{Z_t+D_t^{N}}_{\Fro}^{2}.
 \label{row:eq:direct-work-intervention-contrast}
\end{equation}

\begin{theorem}[Exact paired-contrast mechanism]
\label{row:thm:exact-paired-energy-contrast}
Every paired successor satisfies
\begin{equation}
 I_t
 =2\ip{Z_t}{H_t}
  +2\ip{D_t^{N}}{H_t}
  +\norm{H_t}_{\Fro}^{2}.
 \label{row:eq:exact-paired-energy-contrast}
\end{equation}
The three terms are respectively the state projection, natural-increment
interaction, and nonnegative intervention charge.  Therefore the sign of a
fixed intervention is conditional on the current physical state and natural
increment.  It is not determined by the source label or by
\(\norm{H_t}\) alone.
\end{theorem}

\begin{proof}
By definition, \(D_t^{C}=D_t^{N}+H_t\).  Hence
\begin{align*}
 I_t
 &=\norm{Z_t+D_t^{N}+H_t}_{\Fro}^{2}
   -\norm{Z_t+D_t^{N}}_{\Fro}^{2}\\
 &=2\ip{Z_t+D_t^{N}}{H_t}+\norm{H_t}_{\Fro}^{2}\\
 &=2\ip{Z_t}{H_t}+2\ip{D_t^{N}}{H_t}
   +\norm{H_t}_{\Fro}^{2}.
\end{align*}
This is Equation~\eqref{row:eq:exact-paired-energy-contrast}.  Since either inner
product can change sign as \(Z_t\) or \(D_t^N\) changes, no state-independent
sign follows without additional inequalities controlling both terms.
\end{proof}

If the native natural and controlled increments are each resolved into the
four sources of Proposition~\ref{row:prop:exact-gate-shape-secant}, then
\(H_t=\sum_{a\in\mathcal A_4}(D_t^{C,(a)}-D_t^{N,(a)})\).  Substitution in
Equation~\eqref{row:eq:exact-paired-energy-contrast} gives a source-complete
paired ledger.  Its linear and quadratic cross terms must be retained when
assigning a sign.

\subsection{Exact radial--tangential normal form}
\label{row:sec:radial-normal-form}

The work-charge ledger becomes a complete one-step classification after
separating the realized increment into its component parallel to the current
observable and its orthogonal residual.  This separation is intrinsic to the
physical row quotient and does not assume an optimizer model.  On an edge
with \(E_t>0\), define
\begin{equation}
 \alpha_t=-\frac{\ip{Z_t}{D_t}}{E_t},\qquad
 \mathcal T_t=D_t+\alpha_tZ_t,\qquad
 \tau_t^2=\frac{\norm{\mathcal T_t}_{\Fro}^2}{E_t}.
 \label{row:eq:radial-tangential-definition}
\end{equation}
Thus positive \(\alpha_t\) is inward radial motion, while
\(\mathcal T_t\) is the tangential finite increment after radial removal.
The notation is edgewise: it describes the exact executed secant, not a
linearized tangent vector.

\begin{theorem}[Exact radial--tangential gain and closing criterion]
\label{row:thm:exact-radial-tangential-gain}
For every edge with \(E_t>0\),
\begin{alignat}{2}
 \ip{Z_t}{\mathcal T_t}&=0,
 &\quad D_t&=-\alpha_tZ_t+\mathcal T_t,
 \label{row:eq:radial-tangential-split}\\
 g_t^2:=\frac{E_{t+1}}{E_t}
 &=(1-\alpha_t)^2+\tau_t^2,
 &\quad \frac{E_{t+1}-E_t}{E_t}
 &=-2\alpha_t+\alpha_t^2+\tau_t^2.
 \label{row:eq:exact-radial-tangential-gain}
\end{alignat}
In particular,
\begin{equation}
 E_{t+1}<E_t
 \quad\Longleftrightarrow\quad
 0<\alpha_t<2
 \quad\hbox{and}\quad
 \tau_t^2<\alpha_t(2-\alpha_t).
 \label{row:eq:radial-tangential-closing-criterion}
\end{equation}
Thus an edge fails to contract either because the radial coefficient is
noninward or overshoots the interval \((0,2)\), or because tangential charge
exceeds the remaining radial margin.  These alternatives are exhaustive.
\end{theorem}

\begin{proof}
Using Equation~\eqref{row:eq:radial-tangential-definition},
\[
 \ip{Z_t}{\mathcal T_t}
 =\ip{Z_t}{D_t}+\alpha_t\norm{Z_t}_{\Fro}^2=0.
\]
The definition of \(\mathcal T_t\) also rearranges to the split in
Equation~\eqref{row:eq:radial-tangential-split}.  Hence
\(Z_{t+1}=(1-\alpha_t)Z_t+\mathcal T_t\).  Orthogonality and the Pythagorean
identity give
\[
 E_{t+1}=(1-\alpha_t)^2E_t+\norm{\mathcal T_t}_{\Fro}^2.
\]
This orthogonal split is unique.  Indeed, if
\(D_t=-aZ_t+T\) with \(\ip{Z_t}{T}=0\), taking the inner product with
\(Z_t\) gives \(a=-\ip{Z_t}{D_t}/E_t=\alpha_t\), and then
\(T=D_t+\alpha_tZ_t=\mathcal T_t\).
Division by the positive number \(E_t\) proves the first gain formula;
subtracting one proves the second.  Finally,
\[
 g_t^2<1
 \quad\Longleftrightarrow\quad
 \tau_t^2<1-(1-\alpha_t)^2=\alpha_t(2-\alpha_t).
\]
Because \(\tau_t^2\ge0\), the strict inequality forces
\(\alpha_t(2-\alpha_t)>0\), which is equivalent to
\(0<\alpha_t<2\).  The converse follows by reversing the same steps.
\end{proof}

The radial coefficient is a normalized version of the exact work:
\(\mathcal W_t=2\alpha_tE_t\), while
\(\mathcal Q_t/E_t=\alpha_t^2+\tau_t^2\).  Theorem~\ref{row:thm:exact-radial-tangential-gain}
therefore resolves the work-charge inequality into the unique radial part
and an orthogonal nonnegative charge.

\begin{proposition}[Complete source-resolved radial ledger]
\label{row:prop:source-resolved-radial-ledger}
Suppose \(E_t>0\) and \(D_t=\sum_{a\in\mathcal A}D_t^{(a)}\) is any finite
exact source decomposition.  Define
\begin{equation}
 \alpha_t^{(a)}=-\frac{\ip{Z_t}{D_t^{(a)}}}{E_t},\qquad
 \mathcal T_t^{(a)}=D_t^{(a)}+\alpha_t^{(a)}Z_t.
 \label{row:eq:source-radial-definition}
\end{equation}
Then every \(\mathcal T_t^{(a)}\) is orthogonal to \(Z_t\), and
\begin{alignat}{2}
 \alpha_t&=\sum_a\alpha_t^{(a)},
 &\quad \mathcal T_t&=\sum_a\mathcal T_t^{(a)},
 \label{row:eq:source-radial-recomposition}\\
 \tau_t^2&=\frac{1}{E_t}\sum_{a,b}
   \ip{\mathcal T_t^{(a)}}{\mathcal T_t^{(b)}},
 &\quad \frac{E_{t+1}-E_t}{E_t}
 &=-2\sum_a\alpha_t^{(a)}+
   \left(\sum_a\alpha_t^{(a)}\right)^2+
   \frac{1}{E_t}\sum_{a,b}
   \ip{\mathcal T_t^{(a)}}{\mathcal T_t^{(b)}}.
 \label{row:eq:source-radial-gram}
\end{alignat}
The ordered double sum retains all tangential diagonal and cross-Gram terms.
\end{proposition}

\begin{proof}
The orthogonality calculation in the preceding proof applies to each source.
Linearity of the inner product gives
\[
 \alpha_t=-\frac{\ip{Z_t}{\sum_aD_t^{(a)}}}{E_t}
 =\sum_a\alpha_t^{(a)}.
\]
Substitution into the definitions gives
\(
 \mathcal T_t=\sum_aD_t^{(a)}+(\sum_a\alpha_t^{(a)})Z_t
 =\sum_a\mathcal T_t^{(a)}
\).
Expanding the squared norm of this last sum proves the formula for
\(\tau_t^2\).  The final identity follows from
Equation~\eqref{row:eq:exact-radial-tangential-gain}.
\end{proof}

For the four native secants of
Proposition~\ref{row:prop:exact-gate-shape-secant},
Proposition~\ref{row:prop:source-resolved-radial-ledger} produces four radial
coefficients and the complete \(4\times4\) tangential Gram matrix.  It
therefore distinguishes inward source work from radial overshoot,
tangential self-charge, and tangential source cancellation without dropping
the arithmetic-defect endpoint.

\begin{theorem}[Exact radial--tangential paired contrast]
\label{row:thm:radial-tangential-paired-contrast}
Let natural and controlled successors share a nonzero source state.  Resolve
\begin{equation}
 D_t^N=-\alpha_t^N Z_t+\mathcal T_t^N,\qquad
 H_t=D_t^C-D_t^N=-\beta_tZ_t+\mathcal S_t,
 \label{row:eq:paired-radial-splits}
\end{equation}
where both residuals are orthogonal to \(Z_t\).  Then
\begin{equation}
 \frac{I_t}{E_t}
 =\underbrace{-2(1-\alpha_t^N)\beta_t}_{\text{radial linear}}
  +\underbrace{\beta_t^2}_{\text{radial charge}}
  +\underbrace{\frac{2\ip{\mathcal T_t^N}{\mathcal S_t}}{E_t}}
    _{\text{tangential interaction}}
  +\underbrace{\frac{\norm{\mathcal S_t}_{\Fro}^2}{E_t}}
    _{\text{tangential charge}}.
 \label{row:eq:radial-tangential-paired-contrast}
\end{equation}
The four terms are exact.  In particular, the radial contribution can be a
useful state-conditional predictor, but its sign is decisive only when it
dominates the two tangential terms.
\end{theorem}

\begin{proof}
The natural endpoint is
\((1-\alpha_t^N)Z_t+\mathcal T_t^N\), and the controlled endpoint is
\((1-\alpha_t^N-\beta_t)Z_t+\mathcal T_t^N+\mathcal S_t\).
Orthogonality removes every cross term containing \(Z_t\).  Subtracting the
two squared norms and dividing by \(E_t>0\) gives
\[
 (1-\alpha_t^N-\beta_t)^2-(1-\alpha_t^N)^2
 +\frac{\norm{\mathcal T_t^N+\mathcal S_t}_{\Fro}^2
       -\norm{\mathcal T_t^N}_{\Fro}^2}{E_t}.
\]
Expanding the two differences gives
Equation~\eqref{row:eq:radial-tangential-paired-contrast}.
\end{proof}

\begin{corollary}[Exact positive-excursion product]
\label{row:cor:exact-positive-excursion-product}
If \(E_j>0\) for every \(k\le j<n\), then
\begin{equation}
 E_n=E_k\prod_{j=k}^{n-1}
 \left((1-\alpha_j)^2+\tau_j^2\right).
 \label{row:eq:exact-positive-excursion-product}
\end{equation}
Consequently, collapse along an infinite positive excursion is equivalent to
vanishing of this exact chronological product.
\end{corollary}

\begin{proof}
Theorem~\ref{row:thm:exact-radial-tangential-gain} gives
\(E_{j+1}=((1-\alpha_j)^2+\tau_j^2)E_j\) on every stated edge.  Iteration
telescopes to the displayed product.  Since \(E_k>0\), its endpoint tends to
zero exactly when the product does.
\end{proof}

\subsection{A direct nonautonomous attraction envelope}

The exact work ledger yields a scalar route to collapse that does not require
a contracting complete-state norm.  Let \(\varrho_t,b_t\ge0\) and suppose
\begin{equation}
 \mathcal W_t-\mathcal Q_t
 \ge (1-\varrho_t)E_t-b_t.
 \label{row:eq:direct-dissipation-margin}
\end{equation}
Equivalently, \(E_{t+1}\le\varrho_tE_t+b_t\).  For \(n\ge k\), define the
ordered scalar product
\begin{equation}
 P_E(n,k)=\prod_{j=k}^{n-1}\varrho_j,\qquad P_E(k,k)=1.
 \label{row:eq:direct-scalar-propagator}
\end{equation}

\begin{theorem}[Direct nonautonomous energy envelope]
\label{row:thm:direct-energy-envelope}
Under Equation~\eqref{row:eq:direct-dissipation-margin},
\begin{equation}
 E_n\le P_E(n,0)E_0+
 \sum_{j=0}^{n-1}P_E(n,j+1)b_j.
 \label{row:eq:direct-energy-envelope}
\end{equation}
If both terms on the right tend to zero, then every row map in the finite
registry collapses.  No pointwise condition \(\varrho_t<1\) is required.
\end{theorem}

\begin{proof}
The formula is immediate at \(n=0\).  Assume it holds at \(n\).  Multiply
the bound by \(\varrho_n\ge0\), add \(b_n\), and use
\(P_E(n+1,k)=\varrho_nP_E(n,k)\) for \(k\le n\) and
\(P_E(n+1,n+1)=1\).  This proves the formula at \(n+1\) by induction.  If its
right side tends to zero, nonnegativity gives \(E_n\to0\).  Each registered
energy is bounded by the stacked energy, and
Lemma~\ref{row:lem:row-quotient-geometry} converts energy convergence to
row-diameter convergence.
\end{proof}

Chronological contraction is therefore needed only at the level at which a
certificate is asserted.  The complete state may be expansive while the
observable work margin closes, or the direct margin may fail while hidden
state cancellation closes the exact observable cocycle.  Neither route
implies the other without comparison hypotheses.

\subsection{Reopenings and the row-constant face}
\label{row:sec:face-restarts}

Define the absolute positive reopening
\begin{equation}
 \mathfrak r_t=(E_{t+1}-E_t)_+=(-\mathcal W_t+\mathcal Q_t)_+.
 \label{row:eq:absolute-reopening}
\end{equation}
It is well defined at \(E_t=0\), unlike a relative gain.

\begin{theorem}[Summable reopening closure]
\label{row:thm:summable-reopening-closure}
Suppose \(E_t\ge0\), arbitrarily late energies are arbitrarily small, and
\(\sum_{t=0}^{\infty}\mathfrak r_t<\infty\).  Explicitly, assume that for every
\(\varepsilon>0\) and every \(N\), some \(t\ge N\) satisfies
\(E_t<\varepsilon\).  Then \(E_t\to0\).
\end{theorem}

\begin{proof}
Given \(\varepsilon>0\), choose \(N\) so that
\(\sum_{t=N}^{\infty}\mathfrak r_t<\varepsilon/2\).  Choose \(t_0\ge N\) with
\(E_{t_0}<\varepsilon/2\).  For every \(n\ge t_0\), telescoping and the
definition of positive part give
\[
 E_n-E_{t_0}
 =\sum_{t=t_0}^{n-1}(E_{t+1}-E_t)
 \le\sum_{t=t_0}^{n-1}\mathfrak r_t<\varepsilon/2.
\]
Thus \(0\le E_n<\varepsilon\) for all \(n\ge t_0\), proving the limit.
\end{proof}

\begin{proposition}[Exact face-preservation criterion]
\label{row:prop:direct-face-preservation}
At a row-constant source \(Z_t=\bm0_{r\times d}\),
\begin{equation}
 E_{t+1}=\norm{D_t}_{\Fro}^{2}.
 \label{row:eq:face-endpoint-charge}
\end{equation}
Hence the physical face is preserved by that update if and only if
\(D_t=\bm0_{r\times d}\).  For any exact source ledger, this is equivalent to
\(\sum_aD_t^{(a)}=\bm0_{r\times d}\).  In particular, the gate-shape ledger requires
\(D_t^{\rm shape}+D_t^{\rm gate}+D_t^{\rm int}+D_t^{\rm def}=\bm0_{r\times d}\) for a
native finite-precision endpoint.
\end{proposition}

\begin{proof}
Substitute \(Z_t=\bm0_{r\times d}\) into
\(Z_{t+1}=Z_t+D_t\) and square the Frobenius norm.  A finite-dimensional
squared norm is zero exactly when its vector is zero.  The source statements
follow from their exact sum identities.
\end{proof}

\begin{theorem}[Exact radial-face affine cocycle]
\label{row:thm:exact-radial-face-cocycle}
For every nonnegative energy sequence, define
\begin{equation}
 q_t=\begin{cases}
 E_{t+1}/E_t,&E_t>0,\\
 0,&E_t=0,
 \end{cases}
 \qquad
 \zeta_t=\begin{cases}
 0,&E_t>0,\\
 E_{t+1},&E_t=0.
 \end{cases}
 \label{row:eq:radial-face-coefficients}
\end{equation}
Then \(q_t,\zeta_t\ge0\), every edge satisfies the exact affine identity
\begin{equation}
 E_{t+1}=q_tE_t+\zeta_t,
 \label{row:eq:exact-radial-face-step}
\end{equation}
and, with
\(P_q(n,k)=\prod_{j=k}^{n-1}q_j\) and \(P_q(k,k)=1\),
\begin{equation}
 E_n=P_q(n,0)E_0+
 \sum_{j=0}^{n-1}P_q(n,j+1)\zeta_j.
 \label{row:eq:exact-radial-face-cocycle}
\end{equation}
On a positive-energy edge,
\(q_t=(1-\alpha_t)^2+\tau_t^2\) and \(\zeta_t=0\).  On a zero-face edge,
\(q_t=0\) and
\(\zeta_t=E_{t+1}=\norm{D_t}_{\Fro}^2\).  Thus
Equation~\eqref{row:eq:exact-radial-face-cocycle} is defined across every face
hit and restart without assigning a relative gain at zero.
\end{theorem}

\begin{proof}
If \(E_t>0\), substitution into
Equation~\eqref{row:eq:radial-face-coefficients} gives
\(q_tE_t+\zeta_t=E_{t+1}\).  If \(E_t=0\), it gives the same equality as
\(0+E_{t+1}=E_{t+1}\).  Nonnegativity follows from nonnegativity of the
energies.  Equation~\eqref{row:eq:exact-radial-tangential-gain} identifies
\(q_t\) on the positive branch, and
Proposition~\ref{row:prop:direct-face-preservation} identifies \(\zeta_t\) on
the zero branch.

It remains to iterate Equation~\eqref{row:eq:exact-radial-face-step}.  The
formula is immediate at \(n=0\).  If it holds at \(n\), multiply by \(q_n\),
add \(\zeta_n\), and use
\(P_q(n+1,k)=q_nP_q(n,k)\) for \(k\le n\) and
\(P_q(n+1,n+1)=1\).  This proves the formula for \(n+1\), hence for every
finite \(n\).
\end{proof}

\begin{theorem}[Intermittent block radial-face closure]
\label{row:thm:intermittent-block-radial-face-closure}
Let \(0=\sigma_0<\sigma_1<\cdots\) be an unbounded sequence of integer
anchors.  For the canonical coefficients in
Equation~\eqref{row:eq:radial-face-coefficients}, define the exact block
quantities
\begin{align}
 \rho_j
 &=P_q(\sigma_{j+1},\sigma_j),\nonumber\\
 b_j
 &=\sum_{\ell=\sigma_j}^{\sigma_{j+1}-1}
   P_q(\sigma_{j+1},\ell+1)\zeta_\ell,\nonumber\\
 A_j
 &=\max_{\sigma_j\le t\le\sigma_{j+1}}
   (E_t-E_{\sigma_j})_+.
 \label{row:eq:block-radial-face-quantities}
\end{align}
Then every block obeys the exact anchor recurrence
\begin{equation}
 E_{\sigma_{j+1}}=\rho_jE_{\sigma_j}+b_j.
 \label{row:eq:block-radial-face-recurrence}
\end{equation}
Suppose that for some \(J\) and \(0\le\rho<1\),
\[
 \rho_j\le\rho\quad\text{for all }j\ge J,\qquad
 b_j\longrightarrow0,\qquad
 A_j\longrightarrow0.
\]
Then \(E_t\to0\).  Consequently, every row map in the finite registry
collapses.  The hypotheses permit expanding individual steps, zero-face
restarts, and transient excursions inside each block.
\end{theorem}

\begin{proof}
Apply the affine cocycle of
Theorem~\ref{row:thm:exact-radial-face-cocycle} on the shifted interval
\([\sigma_j,\sigma_{j+1}]\).  The propagated anchor term is
\(\rho_jE_{\sigma_j}\), and the transported restart terms sum to \(b_j\),
which proves Equation~\eqref{row:eq:block-radial-face-recurrence}.

Write \(x_j=E_{\sigma_j}\).  Nonnegativity and the eventual bound on
\(\rho_j\) give, for \(j\ge J\),
\[
 x_j\le \rho^{j-J}x_J+
 \sum_{k=J}^{j-1}\rho^{j-1-k}b_k.
\]
To prove that the right side vanishes, fix \(\varepsilon>0\).  Choose
\(K\ge J\) so that
\(b_k\le(1-\rho)\varepsilon/3\) for all \(k\ge K\).  The contribution from
these tail terms is at most \(\varepsilon/3\).  The finite expression
\[
 \rho^{j-J}x_J+
 \sum_{k=J}^{K-1}\rho^{j-1-k}b_k
\]
tends to zero, so it is below \(2\varepsilon/3\) for all sufficiently large
\(j\).  Hence \(x_j\to0\).

For \(\sigma_j\le t\le\sigma_{j+1}\), the definition of \(A_j\) gives
\[
 0\le E_t\le E_{\sigma_j}+A_j=x_j+A_j.
\]
Both terms tend to zero, and the anchors are unbounded, so \(E_t\to0\) for
the full sequence.  Finally, every registered energy is bounded by the
stacked energy, and Lemma~\ref{row:lem:row-quotient-geometry} converts its
vanishing to row-diameter collapse.
\end{proof}

The block premise is sufficient, not necessary.  The executed dense windows
do not satisfy its fixed-window empirical specialization: each of the
\(\RowSourceBlockWindowCount\) windows contains expanding endpoint pairs at every
measured horizon from \(1\) through \(32\).  At horizon \(32\), only
\(\RowSourceBlockThirtyTwoContracting\) of \(\RowSourceBlockThirtyTwoComparisons\) comparisons
contract, while \(\RowSourceBlockThirtyTwoExpanding\) expand.  Thus those records
falsify a simple fixed-window uniform-contraction description.  Failure of
that sufficient premise falsifies neither the exact cocycle nor row-map
collapse.  Collapse may still follow through nonuniform ordered-product
decay, a different closing block sequence, summable reopening, or exact
observable cancellation, provided the corresponding stated conditions hold.
Section~\ref{row:sec:observer-block-analysis} gives the dense-window methods
and results; Table~\ref{row:tab:block-gain-census} contains the horizon counts.

At the face, every linear source work is zero.  Face preservation is
therefore decided by the complete increment and its quadratic charge, not by
the sign of a first-order work term.  This observation links the direct
ledger to the gate-zero criterion in
Proposition~\ref{row:prop:gate-zero-invariance}: a zero gate is preserved only
when the implemented optimizer produces no effective physical increment,
possibly after exact cancellation with a zero-shape coordinate.

\subsection{Comprehensive direct collapse theory}

\begin{theorem}[Comprehensive exact orbit characterization]
\label{row:thm:comprehensive-direct-collapse}
For a fixed finite registry, the following statements are equivalent:
\begin{enumerate}[label=(\roman*)]
\item every registered row diameter tends to zero;
\item \(E_n\to0\);
\item in the canonical radial-face cocycle,
\begin{equation}
 P_q(n,0)E_0\longrightarrow0,
 \qquad
 \sum_{j=0}^{n-1}P_q(n,j+1)\zeta_j\longrightarrow0;
 \label{row:eq:radial-face-collapse-characterization}
\end{equation}
\item the cumulative work-charge ledger satisfies
\begin{equation}
 E_0+\sum_{t=0}^{n-1}(-\mathcal W_t+\mathcal Q_t)
 \longrightarrow0.
 \label{row:eq:cumulative-collapse-characterization}
\end{equation}
\end{enumerate}
Thus collapse is exactly simultaneous extinction of the propagated initial
energy and all transported zero-face restarts.  This characterization does
not assert that AdamW makes either contribution vanish.
\end{theorem}

\begin{proof}
The equivalence of the first two assertions follows from
Lemma~\ref{row:lem:row-quotient-geometry} for each map and finiteness of the
registry.  By Theorem~\ref{row:thm:exact-radial-face-cocycle}, the two terms in
(iii) are nonnegative and their sum equals \(E_n\) for every \(n\).
Therefore their separate convergence to zero is equivalent to (ii).
Summing Equation~\eqref{row:eq:exact-direct-work-charge} from \(t=0\) to
\(n-1\) shows that the expression in (iv) is also exactly \(E_n\), proving
the remaining equivalence.
\end{proof}

\begin{proposition}[Cumulative ledger telescope]
\label{row:prop:cumulative-direct-ledger}
For every finite \(n\),
\begin{equation}
 E_0+\sum_{t=0}^{n-1}(-\mathcal W_t+\mathcal Q_t)=E_n.
 \label{row:eq:cumulative-direct-ledger}
\end{equation}
Consequently, the cumulative expression tends to zero if and only if
\(E_n\to0\).  This is an exact restatement of energy convergence, not an
independent dynamical criterion.
\end{proposition}

\begin{proof}
Sum Equation~\eqref{row:eq:exact-direct-work-charge} from \(t=0\) to \(n-1\).
The left side telescopes to \(E_n-E_0\).  Rearrangement proves the displayed
identity, and equality for every \(n\) proves the convergence equivalence.
\end{proof}

\begin{theorem}[Sufficient direct closures]
\label{row:thm:sufficient-direct-closures}
Any of the following hypotheses is sufficient for the equivalent collapse
conclusions of Theorem~\ref{row:thm:comprehensive-direct-collapse}:
\begin{enumerate}[label=(\alph*)]
\item the direct nonautonomous envelope of
Theorem~\ref{row:thm:direct-energy-envelope} has vanishing homogeneous and
transported-forcing terms;
\item arbitrarily late energies are arbitrarily small and the absolute
reopenings in Equation~\eqref{row:eq:absolute-reopening} are summable;
\item the anchor contraction, transported-restart, and within-block excursion
conditions of Theorem~\ref{row:thm:intermittent-block-radial-face-closure} hold.
\end{enumerate}
\end{theorem}

\begin{proof}
Condition \(a\) gives \(E_n\to0\) by
Theorem~\ref{row:thm:direct-energy-envelope}.  Condition \(b\) gives the same limit
by Theorem~\ref{row:thm:summable-reopening-closure}, and condition \(c\) gives it
by Theorem~\ref{row:thm:intermittent-block-radial-face-closure}.  Apply
Theorem~\ref{row:thm:comprehensive-direct-collapse} in each case.
\end{proof}

The characterization, telescope, and sufficient closures have deliberately
different logical roles.  AdamW need not supply either sufficient closure
generically.  Establishing one requires an all-future analytical bound.  The
complete-state theory in Section~\ref{row:sec:complete-state} gives a
complementary route through optimizer dynamics, chart transport, forcing,
and observation geometry. The paired-source and radial tests of this
chapter's finite identities are reported in
Sections~\ref{row:sec:direct-work-evidence} and
\ref{row:sec:radial-tangential-evidence}.

\FloatBarrier
\par\medskip\noindent
The endpoint balances identify what changed along an observed path.
Chapter~\ref{ch:complete-adamw} supplies the complete optimizer state and
forced dynamics needed to study why such changes occur and when attraction
can follow.

\chapter{Complete-state AdamW dynamics}
\label{ch:complete-adamw}
This chapter develops the augmented AdamW dynamics behind the row-energy
ledger. It treats optimizer memory, transverse coordinates, forcing and
observation nonlinearity, and states sufficient attraction conditions
together with obstructions and continuous or stochastic specializations.

\section{Complete-state forced AdamW cocycle}
\label{row:sec:complete-state}

\subsection{Why the optimizer state is part of the mechanism}

Let \(\theta_t\in\R^p\) denote trainable parameters and let
\(m_t,v_t\in\R^p\) denote AdamW first and second moments.  The complete
continuous state for a selected layer contains every coordinate of
\((\theta_t,m_t,v_t)\) whose perturbation can return to that layer's row-map
normal.  Coordinates proved structurally inactive may be removed; coordinates
removed only because their realized gradient happened to be small may not.

The environment \(\omega_t\) records the minibatch, data cursor, scheduler
clock, clipping cell, decay mask, random state, and intervention branch.  It
is not differentiated.  Conditional on this environment, one update is a
deterministic map
\begin{equation}
 s_{t+1}=\mathcal F_t(s_t),\qquad s_t=(\theta_t,m_t,v_t).
 \label{row:eq:implemented-complete-successor}
\end{equation}
The sequence is nonautonomous because \(\mathcal F_t\) changes with
\(\omega_t\) and the optimizer clock.

\subsection{Implemented smooth-cell derivative}
\label{row:sec:smooth-adamw}

Fix one clipping cell.  Write \(g(\theta)\) for the raw minibatch gradient,
\(c=\clip(g(\theta);-L,L)\), \(C\) for the diagonal clipping derivative,
and \(K=Dg(\theta)\).  Let \(\Lambda=\lambda_{\rm wd}\diag(d)\), where
\(d\in\{0,1\}^p\) is the implemented decay mask.  The update order is
\begin{align}
 m^+&=\beta_1m+(1-\beta_1)c,\label{row:eq:adamw-moment-one}\\
 v^+&=\beta_2v+(1-\beta_2)c^{\odot2},\label{row:eq:adamw-moment-two}\\
 \widehat m^+&=m^+/a_1,\qquad
 \widehat v^+=v^+/a_2,
 \quad a_i=1-\beta_i^\tau,\label{row:eq:adamw-bias}\\
 \theta^+&=(I-\eta\Lambda)\theta
 -\eta\widehat m^+\oslash
   (\sqrt{\widehat v^+}+\epsilon_{\rm A}).
 \label{row:eq:adamw-position}
\end{align}
All roots and divisions are coordinatewise, and \(\tau\ge1\) is the live
optimizer step.  Define
\begin{align}
 K_c&=CK,\nonumber\\
 M_\theta&=(1-\beta_1)K_c,\qquad
 V_\theta=2(1-\beta_2)\diag(c)K_c,\label{row:eq:adamw-moment-position-blocks}\\
 D_m&=\diag\left(
  \frac{1}{a_1(\sqrt{\widehat v^+}+\epsilon_{\rm A})}
 \right),\nonumber\\
 D_v&=\diag\left(
 -\frac{\widehat m^+}
 {2a_2\sqrt{\widehat v^+}
  (\sqrt{\widehat v^+}+\epsilon_{\rm A})^2}
 \right).
 \label{row:eq:adamw-denominator-derivatives}
\end{align}

\begin{theorem}[Implemented lifted AdamW derivative]
\label{row:thm:implemented-adamw-lift}
Suppose every component of \(\widehat v^+\) is positive and no component of
\(g(\theta)\) lies on a clipping boundary.  On that smooth cell, the
Jacobian of
\((\theta,m,v)\mapsto(\theta^+,m^+,v^+)\) is
\begin{equation}
 \mathcal A_t=
 \begin{pmatrix}
 I-\eta\Lambda-\eta(D_mM_\theta+D_vV_\theta)
   &-\eta\beta_1D_m&-\eta\beta_2D_v\\
 M_\theta&\beta_1I&\bm0_{p\times p}\\
 V_\theta&\bm0_{p\times p}&\beta_2I
 \end{pmatrix}.
 \label{row:eq:implemented-adamw-matrix}
\end{equation}
A differentiable post-update intervention premultiplies the applicable
position block by its derivative. A state-independent additive post-update
perturbation changes the forcing but not the homogeneous derivative. If the
same additive forcing is applied to two affine responses, it cancels in their
difference.
\end{theorem}

\begin{proof}
On the fixed clipping cell,
\(\delta c=CK\,\delta\theta\).  Differentiating
Equations~\eqref{row:eq:adamw-moment-one} and
\eqref{row:eq:adamw-moment-two} gives
\[
 \delta m^+=M_\theta\delta\theta+\beta_1\delta m,
 \qquad
 \delta v^+=V_\theta\delta\theta+\beta_2\delta v.
\]
For
\(
 q(m^+,v^+)=\widehat m^+\oslash
 (\sqrt{\widehat v^+}+\epsilon_{\rm A}),
\)
coordinatewise differentiation gives
\(
 \delta q=D_m\delta m^++D_v\delta v^+.
\)
Substitute the two moment differentials into
\(
 \delta\theta^+=(I-\eta\Lambda)\delta\theta-\eta\delta q.
\)
Collecting the coefficients of \(\delta\theta,\delta m,\delta v\) yields
the first block row in Equation~\eqref{row:eq:implemented-adamw-matrix};
the moment equations give the remaining rows. The intervention statements
are the chain rule and differentiation of an additive constant. For the last
claim, \((Au+b+c)-(Av+b+c)=A(u-v)\). Thus the same additive forcing cancels
in the difference of two affine responses; this is a ring identity.
\end{proof}

The \(D_vV_\theta\) term and the third block column are the response of the
live denominator.  A frozen-preconditioner approximation deletes them and is
therefore a different successor.  Likewise, the loss Hessian \(K\) contains
the actual batch and all upstream model paths that affect the selected
parameters. Section~\ref{row:sec:gate-operators} checks the implemented
final-gate block and its signed responses;
Section~\ref{row:sec:complete-tangent-evidence} measures selected
complete-state tangent directions.

\subsection{The gate-zero stratum is a dynamical condition}

Consider one coordinate of the final LayerNorm gate.  At a source with
\(\gamma=0\), the decoupled decay term vanishes and the exact AdamW update is
\begin{align}
 m^+&=\beta_1m+(1-\beta_1)c,\nonumber\\
 v^+&=\beta_2v+(1-\beta_2)c^2,\nonumber\\
 \gamma^+&=-\eta
 \frac{m^+/a_1}{\sqrt{v^+/a_2}+\epsilon_{\rm A}},
 \label{row:eq:gate-zero-successor}
\end{align}
where \(c\) is the clipped gate gradient.  The denominator is strictly
positive.

\begin{proposition}[Gate-zero invariance criterion]
\label{row:prop:gate-zero-invariance}
Assume \(\eta>0\), \(a_1,a_2>0\), \(\epsilon_{\rm A}>0\), and
\(v^+\ge0\).  At a gate-zero source,
\begin{equation}
 \gamma^+=0
 \quad\Longleftrightarrow\quad
 \beta_1m+(1-\beta_1)c=0.
 \label{row:eq:gate-zero-general-criterion}
\end{equation}
If also \(m=0\) and \(0\le\beta_1<1\), then
\begin{equation}
 \gamma^+=0\quad\Longleftrightarrow\quad c=0.
 \label{row:eq:gate-zero-gradient-criterion}
\end{equation}
The physical row quotient after the step is zero if and only if
\(\lvert\gamma_k^+\rvert\sqrt{\upsilon_k^+}=0\) for every coordinate \(k\).
Thus invariance of the physical face can be weaker than invariance of the
entire gate-zero parameter face when a normalized-shape coordinate vanishes.
\end{proposition}

\begin{proof}
In Equation~\eqref{row:eq:gate-zero-successor}, the denominator, \(\eta\), and
\(a_1\) are nonzero.  Hence \(\gamma^+=0\) if and only if \(m^+=0\).
Substituting the first-moment update proves
Equation~\eqref{row:eq:gate-zero-general-criterion}.  When \(m=0\), this condition
is \((1-\beta_1)c=0\).  Since \(1-\beta_1>0\), it is equivalent to \(c=0\),
which proves Equation~\eqref{row:eq:gate-zero-gradient-criterion}.  The last
statement follows from the nonnegative coordinate factorization in
Theorem~\ref{row:thm:layer-resolved-gate-shape}.
\end{proof}

This proposition distinguishes a zero observation from an invariant
optimizer stratum.  Weight decay cannot reopen a zero gate, but a clipped
loss gradient generally can.  Retaining \(v\) changes the size of the
reopening through the denominator and does not change its zero criterion.
Section~\ref{row:sec:gate-zero-evidence} tests this condition by retaining the
incoming second moment and controlling only the gate-gradient write.
\subsection{Transverse charts and the invariance defect}
\label{row:sec:trajectory-chart}

Fix a realized complete-state trajectory
\(s_{t+1}=\mathcal F_t(s_t)\) and a declared chart itinerary. Realized
exogenous inputs are included in this pathwise specification of
\(\mathcal F_t\). For layer \(\ell\), let \(\mathscr S_{\ell,t}\) be the
selected row-constant stratum in the complete state space \(\mathcal X_t\).
On an admitted smooth stratum, suppose a full local chart
\[
 \Psi_t:V_t^{\rm tan}\times U_t\longrightarrow\mathcal V_t
 \subset\mathcal X_t,\qquad U_t\subset\R^{q_t},\quad\bm0\in U_t,
\]
is a diffeomorphism onto its image, with
\[
 \Psi_t(V_t^{\rm tan}\times\{\bm0\})
 =\mathscr S_{\ell,t}\cap\mathcal V_t.
\]
Here \(V_t^{\rm tan}\) is the tangential coordinate domain and \(q_t\)
is the local normal dimension. Chart existence is a local smooth-stratum
assumption, not a claim across every singular face. Suppose
\(s_t\in\mathcal V_t\), and set
\[
 (a_t,z_t)=\Psi_t^{-1}(s_t),\qquad
 \pi_t(\Psi_t(a,z))=z,\qquad \iota_t(z)=\Psi_t(a_t,z).
\]
The tangential coordinate \(a_t\) is selected from the realized state and
held fixed only while varying the argument \(z\) within this slice.
Thus both the right-inverse and trajectory-lift identities hold:
\begin{equation}
 \pi_t\circ\iota_t=\mathrm{id}_{U_t},\qquad
 z_t=\pi_t(s_t),\qquad s_t=\iota_t(z_t).
 \label{row:eq:trajectory-lift}
\end{equation}
Restrict the domains, when possible, so that
\(\mathcal F_t(\iota_t(U_t))\subset\mathcal V_{t+1}\).
All claims below are restricted to this admitted itinerary and domain.
Define the charted successor by
\begin{equation}
 T_t=\pi_{t+1}\circ\mathcal F_t\circ\iota_t.
 \label{row:eq:charted-successor}
\end{equation}
The lift, rather than the right-inverse identity alone, proves the realized
recurrence by direct substitution:
\begin{equation}
 z_{t+1}=\pi_{t+1}(s_{t+1})
 =\pi_{t+1}\mathcal F_t(s_t)
 =\pi_{t+1}\mathcal F_t(\iota_t(z_t))=T_t(z_t).
 \label{row:eq:realized-chart-step}
\end{equation}
The reference \(\iota_t(\bm0)\) lies on the selected stratum and shares
the realized tangential coordinate; it need not equal \(s_t\).
At a smooth reference define
\begin{equation}
 f_t=T_t(\bm0),\qquad A_t=DT_t(\bm0),\qquad
 R_t(z)=T_t(z)-f_t-A_tz.
 \label{row:eq:normal-components}
\end{equation}
Substitution gives the exact identity
\begin{equation}
 z_{t+1}=A_tz_t+f_t+R_t(z_t).
 \label{row:eq:exact-normal-cocycle}
\end{equation}
The force \(f_t\) is zero exactly when
\(\mathcal F_t(\iota_t(\bm0))\) lies in the next selected row-constant
stratum within its extraction chart. Tangential coordinates may change.
Batch imbalance, movement of the tangent base, data drift, branch changes,
and numerical evaluation can contribute; the force need not be small.

The chain rule, with its evaluation points explicit, gives
\begin{equation}
 A_t=\left.D\pi_{t+1}\right|_{\mathcal F_t(\iota_t(\bm0))}
     \left.D\mathcal F_t\right|_{\iota_t(\bm0)}
     \left.D\iota_t\right|_{\bm0}.
 \label{row:eq:full-normal-operator}
\end{equation}
The middle factor is the implemented complete-state derivative
\(\mathcal A_t\) in Equation~\eqref{row:eq:implemented-adamw-matrix},
evaluated at the reference. A final-gate block omits upstream parameter and
moment coordinates unless a proved invariant splitting removes them or a
Schur complement incorporates them with controlled memory.

This construction concerns one realized itinerary. The slices, forces,
and operators can depend on the complete tangential history. It does not
make \(z_t\) an autonomous or Markov state and does not establish fiberwise
predictive closure. Differentiation holds the selected slice fixed; it
does not differentiate through a rule selecting another history or slice.

\paragraph{Prescribed slices and their two defects.}
\label{row:par:fixed-slice-defects}
If a prescribed slice \(\bar\iota_t\) is used with \(z_t=\pi_t(s_t)\), set
\begin{align}
 \bar T_t&=\pi_{t+1}\mathcal F_t\bar\iota_t,\nonumber\\
 e_t&=\pi_{t+1}\mathcal F_t(s_t)-\bar T_t(z_t).
 \label{row:eq:fixed-slice-defect}
\end{align}
For \(\bar f_t=\bar T_t(\bm0)\), a declared compatible linear map
\(\bar A_t\), and \(\bar R_t(z)=\bar T_t(z)-\bar f_t-\bar A_tz\),
addition and subtraction give
\[
 z_{t+1}=\bar A_tz_t+\bar f_t+\bar R_t(z_t)+e_t.
\]
Its transport formula contains
\(\sum_{j<n}\bar\Phi(n,j+1)e_j\), where \(\bar\Phi\) is the chronological
product of the \(\bar A_t\). A complete-state physical observation
\(\mathcal H_t\) has an additional defect
\begin{equation}
 d_t=\mathcal H_t(s_t)-\mathcal H_t(\bar\iota_t(z_t)).
 \label{row:eq:fixed-slice-observation-defect}
\end{equation}
Writing \(\bar h_t=\mathcal H_t\circ\bar\iota_t\) and
\(\bar O_t=D\bar h_t(\bm0)\) when defined, its observable transport formula
therefore includes both \(d_n\) and
\(\sum_{j<n}\bar O_n\bar\Phi(n,j+1)e_j\).
The state defect alone cannot justify the observation formula.
If \(\pi_{t+1}\mathcal F_t\) is \(K_t\)-Lipschitz on the relevant pair,
\[
 \|e_t\|\le K_t\|s_t-\bar\iota_t(z_t)\|.
\]
An observation Lipschitz bound controls \(d_t\) in the same way. A tube
forcing budget must include \(\|e_t\|\) in its declared norm. Neither
defect inherits a quadratic-in-\(z_t\) bound from the remainder definition.

For a concrete necessity example on \(\R^2\), let
\(\mathcal F(x,y)=(x+y,y)\), \(\pi(x,y)=x\), and
\(\bar\iota(z)=(z,0)\). Although \(\pi\bar\iota=\mathrm{id}\), the
state \(s=(1,1)\) has \(\pi\mathcal F(s)=2\) whereas
\(\bar T(\pi(s))=1\). Its fixed-slice defect is \(e=1\).
The adapted slice \(\iota(z)=(z,1)\) gives \(T(z)=z+1\), \(f=1\),
and \(R=0_{\rm fun}\), recovering the realized successor. If the physical
observation is \(\mathcal H(x,y)=y\), the fixed slice also has \(d=1\).

\subsection{Exact observable transport and cancellation}
\label{row:sec:observable-transport}

Let \(\mathcal Y\) be the finite product of the centered physical row-map
spaces for a declared registry \(\mathcal M\), and let
\(\mathcal H_t:\mathcal X_t\to\mathcal Y\) be the stacked complete-state
row observation. On the compatible slice of
Section~\ref{row:sec:trajectory-chart}, define
\(h_t=\mathcal H_t\circ\iota_t\). The same lift gives
\begin{equation}
 y_t=\mathcal H_t(s_t)=h_t(z_t)\in\mathcal Y.
 \label{row:eq:realized-chart-observation}
\end{equation}

For the realized native endpoints, identify \(y_t=Z_t\) and
\(D_t=y_{t+1}-y_t\).  The radial coefficient \(\alpha_t\), tangential
residual \(\mathcal T_t\), positive-face gain \(q_t\), and zero-face restart
\(\zeta_t\) are therefore exact observations of the complete-state
successor.  They are not derivatives.  The matrices \(A_t\) and \(O_t\)
describe base-route first variation, whereas the executed endpoint also
contains \(f_t\), \(R_t(z_t)\), \(Q_t(z_t)\), and any route change.
Consequently, a local tangent gain does not determine the nonlinear realized
energy gain without bounds on those additional terms.  Conversely, the
radial normal form determines the observable edge exactly but does not by
itself identify which hidden optimizer coordinates produced it.  The two
descriptions are complementary rather than competing closure assumptions.
The normalized radial quantities are mathematically defined for every
strictly positive stored energy.  Observer policy is separate: a
floor-censored-positive source may withhold \(\alpha_t\), \(\tau_t\), and
\(q_t\) without converting that source into an exact face hit.  In that
stratum the unnormalized work \(\mathcal W_t\), charge \(\mathcal Q_t\), and
their verified intervals remain the branch-free observables.  Neither a
local tangent quantity nor observer censoring determines the nonlinear
endpoint gain.
Around the chart origin, define
\begin{equation}
 o_t=h_t(\bm0),\qquad O_t=Dh_t(\bm0),\qquad
 Q_t(z)=h_t(z)-o_t-O_tz.
 \label{row:eq:observable-components}
\end{equation}
For \(n\ge k\), write
\begin{equation}
 \Phi(n,k)=A_{n-1}A_{n-2}\cdots A_k,\qquad
 \Phi(k,k)=I_{q_k}.
 \label{row:eq:matrix-propagator}
\end{equation}
The products act between the corresponding chart spaces.

\begin{theorem}[Observable complete-state variation of constants]
\label{row:thm:observable-complete-state}
Suppose the admitted smooth-cell itinerary is trajectory-compatible as in
Section~\ref{row:sec:trajectory-chart}, with the realized recurrence
\eqref{row:eq:realized-chart-step} and physical observation
\eqref{row:eq:realized-chart-observation}. With
\(A_t:\R^{q_t}\to\R^{q_{t+1}}\) and the components above,
\begin{align}
 z_n={}&\Phi(n,0)z_0
 +\sum_{j=0}^{n-1}\Phi(n,j+1)f_j
 +\sum_{j=0}^{n-1}\Phi(n,j+1)R_j(z_j),
 \label{row:eq:state-variation-of-constants}\\
 y_n={}&o_n+O_n\Phi(n,0)z_0
 +\sum_{j=0}^{n-1}O_n\Phi(n,j+1)f_j \nonumber\\
 &+\sum_{j=0}^{n-1}O_n\Phi(n,j+1)R_j(z_j)
 +Q_n(z_n).
 \label{row:eq:observable-variation-of-constants}
\end{align}
Consequently, physical row-map collapse over the finite registry is
equivalent to convergence of the complete vector on the right side of
Equation~\eqref{row:eq:observable-variation-of-constants} to zero.

It is sufficient, but not necessary, that the norms of all five displayed
components tend to zero.  More generally,
\begin{align}
 \norm{y_n}
 \le{}&\norm{o_n}+\norm{O_n\Phi(n,0)z_0}
 +\sum_{j<n}\norm{O_n\Phi(n,j+1)f_j}\nonumber\\
 &+\sum_{j<n}\norm{O_n\Phi(n,j+1)R_j(z_j)}
 +\norm{Q_n(z_n)}.
 \label{row:eq:observable-component-envelope}
\end{align}
\end{theorem}

\begin{proof}
The trajectory lift proves Equation~\eqref{row:eq:realized-chart-step}.
Substituting the definition of \(R_t\) yields the exact recurrence
\eqref{row:eq:exact-normal-cocycle}. Equation~\eqref{row:eq:state-variation-of-constants}
now follows by induction and is immediate at \(n=0\).  If it holds at \(n\), substitute it into
\(z_{n+1}=A_nz_n+f_n+R_n(z_n)\).  Left multiplication by \(A_n\) changes
each old \(\Phi(n,j+1)\) into \(\Phi(n+1,j+1)\), while the new force and
remainder enter with \(\Phi(n+1,n+1)=I\).  This gives the formula at
\(n+1\).

The observation lift \eqref{row:eq:realized-chart-observation} and
Equation~\eqref{row:eq:observable-components} give
\(y_n=o_n+O_nz_n+Q_n(z_n)\).  Substitute the state identity and use the
linearity of \(O_n\) to obtain
Equation~\eqref{row:eq:observable-variation-of-constants}.  Equality makes
convergence of \(y_n\) equivalent to convergence of its full right side.
For a finite registry, convergence of the stacked centered maps is
equivalent to convergence of every physical quotient norm, and
Lemma~\ref{row:lem:row-quotient-geometry} converts this to row-diameter collapse.
The triangle inequality gives
Equation~\eqref{row:eq:observable-component-envelope}.  Vanishing of every term
on that upper bound is sufficient.  It is not necessary because two nonzero
vector terms can have equal and opposite limits.
\end{proof}

The cancellation is measurable rather than rhetorical.  If the five terms
on the right side of
Equation~\eqref{row:eq:observable-variation-of-constants} are denoted
\(c_{a,n}\), then
\begin{equation}
 \norm{y_n}^2
 =\sum_a\norm{c_{a,n}}^2
 +2\sum_{a<b}\ip{c_{a,n}}{c_{b,n}}.
 \label{row:eq:observable-cancellation-gram}
\end{equation}
Negative cross terms can offset expanding component norms.  Replacing this
identity by the triangle envelope removes that route and yields a stronger
certificate than physical collapse itself.

\begin{theorem}[Exact finite-increment observable balance]
\label{row:thm:finite-increment-observable-balance}
Let a realized base trajectory satisfy
\(s^0_{t+1}=\mathcal F_t(s^0_t)\).  Choose compatible linear maps \(A_t\)
along that base trajectory and define \(\Phi\) by
Equation~\eqref{row:eq:matrix-propagator}.  For a scale \(\delta>0\) and a
declared state direction \(d\), initialize two realized trajectories by
\(s^\pm_0=s^0_0\pm\delta d\), expose them to the same declared exogenous
environment, and set
\begin{equation}
 \Delta_t=\frac{s^+_t-s^-_t}{2},\qquad
 \rho_t=\Delta_{t+1}-A_t\Delta_t.
 \label{row:eq:finite-increment-state-residual}
\end{equation}
The signed trajectories may follow different clipping or boundary routes.
For an observation map \(\mathcal H_n\), choose a compatible linear map
\(O_n\) at the base state and define
\begin{alignat}{3}
 Y_n&=\frac{\mathcal H_n(s^+_n)-\mathcal H_n(s^-_n)}{2},
 &\quad u_n&=\Phi(n,0)\delta d,\nonumber\\
 L_n&=O_nu_n,
 &\quad N_n&=\sum_{j=0}^{n-1}O_n\Phi(n,j+1)\rho_j,
 &\quad Q_n&=Y_n-O_n\Delta_n.
 \label{row:eq:finite-increment-balance-terms}
\end{alignat}
Then, for every finite horizon for which these quantities are defined,
\begin{equation}
 \Delta_n=u_n+\sum_{j=0}^{n-1}\Phi(n,j+1)\rho_j,
 \qquad
 \boxed{Y_n=L_n+N_n+Q_n}.
 \label{row:eq:finite-increment-observable-balance}
\end{equation}
Thus the full-state correction \(N_n\) contains every departure of the
finite state difference from base-route linear transport, while \(Q_n\)
contains the remaining observation nonlinearity.
\end{theorem}

\begin{proof}
By construction, \(\Delta_0=\delta d=u_0\) and
\(\Delta_{t+1}=A_t\Delta_t+\rho_t\).  Suppose the first identity in
Equation~\eqref{row:eq:finite-increment-observable-balance} holds at time \(n\).
Multiplication by \(A_n\) changes \(\Phi(n,j+1)\) into
\(\Phi(n+1,j+1)\), and addition of \(\rho_n\) supplies the final term with
\(\Phi(n+1,n+1)=I\).  Induction proves the state identity at every finite
horizon.  Applying \(O_n\), using its linearity, and substituting
\(Q_n=Y_n-O_n\Delta_n\) gives
\[
 Y_n=O_nu_n+
 \sum_{j=0}^{n-1}O_n\Phi(n,j+1)\rho_j+Q_n
 =L_n+N_n+Q_n.
\]
No differentiability or common route for the signed trajectories was used.
\end{proof}

For a numerical effect floor \(\epsilon_*>0\), the finite-increment
cancellation index is
\begin{equation}
 \chi_n=\frac{\norm{L_n}+\norm{N_n}+\norm{Q_n}}
 {\max\{\norm{Y_n},\epsilon_*\}}.
 \label{row:eq:finite-increment-cancellation-index}
\end{equation}
Large \(\chi_n\), a small response-to-homogeneous ratio, and a negative
inner product \(\ip{L_n}{N_n+Q_n}\) diagnose destructive cancellation.
They are measurements, not consequences of
Theorem~\ref{row:thm:finite-increment-observable-balance}.  The theorem also
does not make \(L_n\), \(N_n\), or \(Q_n\) small and does not turn finite
horizons or sampled directions into an asymptotic or operator-norm result.
When \(A_t\) and \(O_n\) are derivatives on a smooth base route, the terms
have their usual tangent interpretation.  At a boundary they retain the
exact algebraic meaning for the declared compatible maps, but no derivative
claim follows. Section~\ref{row:sec:finite-balance-evidence} reports the
measured homogeneous, state-correction, and observation-correction terms,
including their direction-dependent cancellation.

\begin{proposition}[Stratified itinerary identity]
\label{row:prop:stratified-itinerary}
Suppose each realized update is assigned a route label \(\sigma_t\) encoding
its clipping cell, zero-second-moment face, and local chart. Suppose every
route map \(T_t^{\sigma_t}\) is defined on the admitted states and comes
from a compatible lift, so that \(z_{t+1}=T_t^{\sigma_t}(z_t)\), with the
physical observation identity \eqref{row:eq:realized-chart-observation}.  At a smooth
reference set \(A_t=DT_t^{\sigma_t}(\bm0)\).  At a nonsmooth reference choose
any declared compatible linear map \(A_t\).  In both cases define
\[
 f_t=T_t^{\sigma_t}(\bm0),\qquad
 R_t(z)=T_t^{\sigma_t}(z)-f_t-A_tz.
\]
Then Equations~\eqref{row:eq:state-variation-of-constants} and
\eqref{row:eq:observable-variation-of-constants} remain exact along the entire
itinerary.  A quadratic remainder bound may be invoked only on steps where
it has been established for the selected route; the algebraic identity does
not imply such a bound.
\end{proposition}

\begin{proof}
The compatible lift first gives \(z_{t+1}=T_t^{\sigma_t}(z_t)\).
The definition of \(R_t\) then gives
\(z_{t+1}=A_tz_t+f_t+R_t(z_t)\) at each admitted step, whether or
not \(A_t\) is a derivative.  The inductive proof of
Theorem~\ref{row:thm:observable-complete-state} uses only this identity and
compatibility of consecutive linear maps.  It therefore applies across
route changes and chart dimensions.  Differentiability enters only when
identifying \(A_t\) with a Jacobian and when deriving a higher-order bound
for \(R_t\), so no such regularity follows from the exact decomposition
alone.
\end{proof}

This separation is essential at exact-zero AdamW variances.  The square-root
derivative is not a two-sided smooth map there.  A tangent confined to the
zero-variance face may still have a valid directional propagation, but a
direction leaving the variance cone requires a different stratum analysis.
\subsection{Scheduled metrics and variable tubes}
\label{row:sec:scheduled-tubes}

Retain the trajectory-compatible itinerary of
Section~\ref{row:sec:trajectory-chart}, or the admitted route identities of
Proposition~\ref{row:prop:stratified-itinerary}.
Let \(\mathsf H_t\succ\bm0_{\mathrm{mat}}\) be a scheduled metric and write
\begin{equation}
 \norm{z}_t=(z^\top \mathsf H_tz)^{1/2},\qquad
 \mathfrak a_t=\norm{A_t}_{\mathsf H_{t+1}\leftarrow \mathsf H_t}.
 \label{row:eq:scheduled-normal-norm}
\end{equation}
The tube recursion below is meaningful in the scheduled norms themselves.
To transfer scheduled decay to a fixed Euclidean norm, only a uniform lower
comparison is required: for some \(h_->0\),
\begin{equation}
 h_-I\preceq \mathsf H_t\quad\text{for all }t.
 \label{row:eq:uniform-metric-conditioning}
\end{equation}

\begin{theorem}[Lower-metric fixed-norm transfer]
\label{row:thm:lower-metric-transfer}
If Equation~\eqref{row:eq:uniform-metric-conditioning} holds and
\(\norm{z_t}_t\to0\), then \(\norm{z_t}_2\to0\).  No upper bound on
\(\mathsf H_t\) is needed for this implication.
\end{theorem}

\begin{proof}
The matrix inequality gives
\[
 h_-\norm{z_t}_2^2\le z_t^\top \mathsf H_tz_t=\norm{z_t}_t^2.
\]
Since \(h_->0\),
\(
 0\le\norm{z_t}_2\le h_-^{-1/2}\norm{z_t}_t.
\)
The right side tends to zero, so the squeeze theorem proves the claim.
\end{proof}

An optional upper comparison \(\mathsf H_t\preceq h_+I\) is needed for the reverse
transfer from a fixed-norm radius to a scheduled radius, or for two-sided
uniform equivalence.  It is not a premise of physical decay when the
physical cover is already stated in the scheduled norm.
Assume that on a radius \(r_t>0\),
\begin{equation}
 \norm{R_t(z)}_{t+1}\le C_t\norm{z}_t^2
 \quad\text{whenever }\norm{z}_t\le r_t,
 \qquad C_t\ge0.
 \label{row:eq:quadratic-remainder-bound}
\end{equation}
Set
\begin{equation}
 F_t=\norm{f_t}_{t+1},\qquad
 \mathfrak q_t=\mathfrak a_t+C_tr_t.
 \label{row:eq:effective-tube-gain}
\end{equation}
The variable-radius forward condition is
\begin{equation}
 \mathfrak q_tr_t+F_t\le r_{t+1}.
 \label{row:eq:variable-radius-condition}
\end{equation}
It is checked before using the Taylor bound at the next step.

For \(n\ge k\), define the chronological scalar product
\begin{equation}
 P_{\rm tube}(n,k)=\prod_{i=k}^{n-1}\mathfrak q_i,
 \qquad P_{\rm tube}(k,k)=1.
 \label{row:eq:chronological-scalar-product}
\end{equation}

\begin{theorem}[Scheduled-norm tube collapse]
\label{row:thm:scheduled-tube-collapse}
Suppose a trajectory-compatible itinerary satisfies
Equations~\eqref{row:eq:exact-normal-cocycle} through
\eqref{row:eq:variable-radius-condition} hold for all \(t\ge0\), and
\(\norm{z_0}_0\le r_0\).  Then \(\norm{z_t}_t\le r_t\) for every \(t\) and
\begin{equation}
 \norm{z_n}_n\le
 P_{\rm tube}(n,0)\norm{z_0}_0+
 \sum_{j=0}^{n-1}P_{\rm tube}(n,j+1)F_j.
 \label{row:eq:complete-state-product-convolution}
\end{equation}
In particular, \(\norm{z_n}_n\to0\) if
\begin{equation}
 P_{\rm tube}(n,0)\norm{z_0}_0\longrightarrow0,
 \qquad
 \sum_{j=0}^{n-1}P_{\rm tube}(n,j+1)F_j\longrightarrow0.
 \label{row:eq:product-forcing-limit}
\end{equation}
This conclusion is entirely in the scheduled norms and requires no
comparison between \(\mathsf H_t\) and a fixed metric.
\end{theorem}

\begin{proof}
Put \(u_t=\norm{z_t}_t\).  Suppose \(u_t\le r_t\).  The exact recurrence,
the induced-norm bound, and
Equation~\eqref{row:eq:quadratic-remainder-bound} give
\[
 u_{t+1}\le\mathfrak a_tu_t+F_t+C_tu_t^2
 \le(\mathfrak a_t+C_tr_t)u_t+F_t=\mathfrak q_tu_t+F_t.
\]
Using \(u_t\le r_t\) once more and
Equation~\eqref{row:eq:variable-radius-condition} yields
\(u_{t+1}\le r_{t+1}\).  Induction from \(u_0\le r_0\) proves invariance of
the variable tube.  Iterating \(u_{t+1}\le \mathfrak q_tu_t+F_t\) gives
Equation~\eqref{row:eq:complete-state-product-convolution}.  Indeed,
multiplication of the horizon-\(n\) bound by \(\mathfrak q_n\) prepends \(\mathfrak q_n\) to
every chronological suffix, while the new force enters with empty suffix
product one.  The two limits in
Equation~\eqref{row:eq:product-forcing-limit} squeeze \(u_n\) to zero.
\end{proof}

\begin{corollary}[Scheduled-cover physical collapse]
\label{row:cor:scheduled-cover-collapse}
Assume \(\norm{z_n}_n\to0\).  Let \(\mathcal M\) be a finite registered set
of row maps whose physical quotient norms satisfy
\begin{equation}
 J_{m,n}\le b_{m,n}+g_{m,n}\norm{z_n}_n,
 \quad
 \max_{m\in\mathcal M}b_{m,n}\to0,
 \quad
 0\le g_{m,n}\le G<\infty.
 \label{row:eq:physical-cover}
\end{equation}
Here \(G\) is a single finite constant bounding all cover coefficients
\(g_{m,n}\), independently of the map index \(m\) and time \(n\).
Then \(\max_{m\in\mathcal M}J_{m,n}\to0\), and every registered physical
row diameter tends to zero.  If the cover is uniform over a compact or
otherwise controlled context class, the same conclusion holds uniformly
over that class.  No fixed-metric comparison is required.
\end{corollary}

\begin{proof}
The cover gives
\[
 0\le\max_mJ_{m,n}
 \le\max_mb_{m,n}+G\norm{z_n}_n\longrightarrow0.
\]
Lemma~\ref{row:lem:row-quotient-geometry} gives
\(D_{m,n}\le\sqrt2J_{m,n}\), proving row-diameter collapse.  The identical
supremum argument proves the uniform statement.
\end{proof}

\begin{corollary}[Fixed-norm complete-state transfer]
\label{row:cor:fixed-norm-collapse-transfer}
Assume the hypotheses and the two limits of
Theorem~\ref{row:thm:scheduled-tube-collapse}.  If
\(h_-I\preceq \mathsf H_t\) for one \(h_->0\) and all \(t\), then
\(\norm{z_t}_2\to0\).  If, in addition, the physical quotient norms obey
\begin{equation}
 J_{m,n}\le \widetilde b_{m,n}+
 \widetilde g_{m,n}\norm{z_n}_2,
 \quad
 \max_m\widetilde b_{m,n}\to0,
 \quad
 0\le\widetilde g_{m,n}\le\widetilde G<\infty,
 \label{row:eq:fixed-physical-cover}
\end{equation}
then every registered physical row diameter tends to zero.  No uniform upper
bound on \(\mathsf H_t\) is needed.
\end{corollary}

\begin{proof}
The scheduled-norm theorem gives \(\norm{z_t}_t\to0\), and
Theorem~\ref{row:thm:lower-metric-transfer} gives \(\norm{z_t}_2\to0\).  The
fixed-norm cover and Lemma~\ref{row:lem:row-quotient-geometry} then give the
physical conclusion by the same finite-maximum argument as in
Corollary~\ref{row:cor:scheduled-cover-collapse}.
\end{proof}

The theorem does not require \(\mathfrak q_t<1\) at every step.  Transient gains may
exceed one, and noncommuting matrix products may behave differently from
their individual factors.  The required objects are the chronological
product and the chronologically transported force.

\begin{theorem}[Finite-horizon cocycle envelope]
\label{row:thm:finite-cocycle-envelope}
Under the finite-horizon versions of the hypotheses above, for every declared
\(N\),
\begin{align}
 \norm{z_N}_N&\le
 P_{\rm tube}(N,0)\norm{z_0}_0+
 \sum_{j=0}^{N-1}P_{\rm tube}(N,j+1)F_j,\label{row:eq:finite-normal-envelope}\\
 \max_{m\in\mathcal M}J_{m,N}&\le
 \max_m b_{m,N}+G\left[
 P_{\rm tube}(N,0)\norm{z_0}_0+
 \sum_{j=0}^{N-1}P_{\rm tube}(N,j+1)F_j\right].
 \label{row:eq:finite-physical-envelope}
\end{align}
These are finite inequalities only.  No finite \(N\) implies either limit in
Equation~\eqref{row:eq:product-forcing-limit}.
\end{theorem}

\begin{proof}
The tube induction and scalar iteration in the preceding proof use only
steps \(0,\ldots,N-1\), which gives
Equation~\eqref{row:eq:finite-normal-envelope}.  Substitute that bound into the
physical cover and take the finite maximum to obtain
Equation~\eqref{row:eq:finite-physical-envelope}.  The final statement follows
because a finite list does not constrain a continuation of the sequence.
\end{proof}

\subsection{Continuous-time specialization}

The discrete optimizer is primary.  A continuous-time equation is useful only
when a separate scaling argument justifies it.

\begin{corollary}[Observable continuous-time specialization]
\label{row:cor:continuous-observable}
Let \(z:[0,T]\to\R^q\) be absolutely continuous and satisfy almost
everywhere
\begin{equation}
 \dot z(t)=B(t)z(t)+b(t)+r(t,z(t)).
 \label{row:eq:continuous-normal-flow}
\end{equation}
Assume the displayed terms are integrable and let \(U(t,s)\) be the
fundamental propagator of \(\dot u=B(t)u\), with \(U(s,s)=I\).  If
\[
 y(t)=o(t)+O(t)z(t)+Q(t,z(t)),
\]
then for \(0\le t\le T\),
\begin{align}
 z(t)={}&U(t,0)z(0)
 +\int_0^tU(t,s)b(s)\,ds
 +\int_0^tU(t,s)r(s,z(s))\,ds,
 \label{row:eq:continuous-state-duhamel}\\
 y(t)={}&o(t)+O(t)U(t,0)z(0)
 +\int_0^tO(t)U(t,s)b(s)\,ds\nonumber\\
 &+\int_0^tO(t)U(t,s)r(s,z(s))\,ds+Q(t,z(t)).
 \label{row:eq:continuous-observable-duhamel}
\end{align}
The continuous observable collapses exactly when the full vector on the
right side of Equation~\eqref{row:eq:continuous-observable-duhamel} tends to
zero along an unbounded continuation for which the formula remains valid.
\end{corollary}

\begin{proof}
Fix \(t\).  The backward propagator identity is
\(\partial_sU(t,s)=-U(t,s)B(s)\).  At almost every \(s\),
\[
 \frac{d}{ds}\bigl[U(t,s)z(s)\bigr]
 =-U(t,s)B(s)z(s)+U(t,s)\dot z(s)
 =U(t,s)\bigl[b(s)+r(s,z(s))\bigr].
\]
Integrating from \(0\) to \(t\), using \(U(t,t)=I\), and rearranging gives
Equation~\eqref{row:eq:continuous-state-duhamel}.  Substitution into the
observation identity and linearity of \(O(t)\) give
Equation~\eqref{row:eq:continuous-observable-duhamel}.  The final equivalence is
equality, just as in the discrete theorem.
\end{proof}

No conclusion about discrete AdamW follows from this corollary alone.
Clipping strata, bias-correction clocks, finite learning rates, and
coordinatewise second moments must survive any claimed discrete-to-continuous
limit. Sections~\ref{model:sec:physical-clock} and
\ref{model:sec:joint-scaling} state the clock, resource, and optimizer
scalings needed for such a passage.
\subsection{Stochastic forcing}

The realized-batch theorem is pathwise.  A complementary linear statement
separates systematic invariance bias from martingale noise using the
propagator in Equation~\eqref{row:eq:matrix-propagator}.
A history-dependent adapted chart need not satisfy the deterministic
transport hypotheses below. Conditioning on an entire realized trajectory
does not establish the martingale orthogonality used in this corollary.

\begin{corollary}[Transported stochastic forcing]
\label{row:cor:linear-stochastic-forcing}
In a finite-dimensional real Hilbert space, consider
\begin{equation}
 z_{t+1}=A_tz_t+b_t+\xi_t.
 \label{row:eq:linear-stochastic-cocycle}
\end{equation}
Suppose $A_t,b_t,z_0$ are deterministic. For a filtration
$(\mathcal F_t)$, let $\xi_t$ be $\mathcal F_{t+1}$-measurable,
square-integrable, with $\E[\xi_t\mid\mathcal F_t]=\bm0$.
With the propagator of Equation~\eqref{row:eq:matrix-propagator}, set
\begin{align}
 \mu_n=\E z_n&=\Phi(n,0)z_0+
       \sum_{j<n}\Phi(n,j+1)b_j,\label{row:eq:stochastic-mean}\\
 V_n&=\sum_{j<n}\E\norm{\Phi(n,j+1)\xi_j}^2.
 \label{row:eq:exact-transported-variance}
\end{align}
Then the exact second moment is
\begin{equation}
 \E\norm{z_n}^2=\norm{\mu_n}^2+V_n.
 \label{row:eq:stochastic-second-moment}
\end{equation}
Mean collapse, here defined as $\E z_n\to\bm0$, is equivalent to the
\emph{sum} in Equation~\eqref{row:eq:stochastic-mean} tending to zero.
Mean-square collapse is equivalent to $\mu_n\to\bm0$ and $V_n\to0$.
If $\E\norm{\xi_j}^2\leq\sigma_j^2$, then
\begin{equation}
 0\leq V_n=\E\norm{z_n-\mu_n}^2
 \leq B_n:=\sum_{j<n}\norm{\Phi(n,j+1)}_{\op}^2\sigma_j^2.
 \label{row:eq:transported-variance}
\end{equation}
Separate decay of $\Phi(n,0)z_0$ and
$\sum_{j<n}\Phi(n,j+1)b_j$ suffices for mean collapse. Those conditions
together with $B_n\to0$ suffice for mean-square collapse. Separate
component decay and decay of this envelope are not necessary.
\end{corollary}
\begin{proof}
Repeated substitution gives
$z_n=\mu_n+\sum_{j<n}U_{n,j}$, where
$U_{n,j}=\Phi(n,j+1)\xi_j$. Deterministic transport and centering give
$\E U_{n,j}=\bm0$, proving the mean formula. For $i<j$, $U_{n,i}$ is
$\mathcal F_j$-measurable. Cauchy--Schwarz makes the inner product
integrable, and conditional expectation gives
\[
 \E\ip{U_{n,i}}{U_{n,j}}
 =\E\ip{U_{n,i}}{\Phi(n,j+1)\E[\xi_j\mid\mathcal F_j]}=0.
\]
Expanding the finite squared sum therefore gives
$\E\norm{z_n-\mu_n}^2=V_n$. The deterministic mean has zero expected
inner product with the centered sum, proving
Equation~\eqref{row:eq:stochastic-second-moment}. Its two terms are
nonnegative, so their sum tends to zero exactly when each does.
Finally, $\norm{\Phi\xi}\leq\norm\Phi_{\op}\norm\xi$ proves the
variance envelope. The triangle inequality and nonnegative squeeze
principle prove the stated sufficient conditions.
\end{proof}

\paragraph{Cancellation and an unforced maximal-gain direction.}
For $A_t=1$, $z_0=1$, $b_0=-1$, later $b_t=0$, and all $\xi_t=0$,
the mean is zero for every $n\geq1$ although its homogeneous and forced
components are $1$ and $-1$. For the variance envelope, take
$A_t=\operatorname{diag}(1,0)$, $z_0=b_t=\bm0$, and
$\xi_0=\varepsilon e_2$, with an equiprobable sign $\varepsilon$;
all later noise is zero. Use $\sigma_0^2=1$ and later $\sigma_t^2=0$.
Then $z_n=\bm0$ and $V_n=0$ for $n\geq2$, whereas $B_n=1$.
The operator norm retains the unforced first direction.

The mean $\E z_n$, mean norm $\E\norm{z_n}$, and second moment
$\E\norm{z_n}^2$ are different observables. Mean-square collapse implies
mean-norm collapse by Cauchy--Schwarz and hence mean collapse; the mean
alone does not control fluctuations. Random state-dependent propagators
can depend on the noise they transport, so the orthogonality argument
above does not apply automatically. A nonlinear stochastic theorem must
control that dependence and its remainder inside the invariant tube.
Section~\ref{model:sec:nonstationary-response} retains the full chronological
covariance before a stationary approximation, and
Section~\ref{model:sec:consuming-bridge} gives the finite-population correction
for a specified linear response.

\subsection{Necessary obstructions}

\begin{theorem}[Dynamical and asymptotic collapse obstructions]
\label{row:thm:collapse-obstructions}
The following statements give independent obstructions to stronger collapse
claims.

\begin{enumerate}[label=(\roman*)]
\item On a trajectory-compatible itinerary, suppose \(z_t\to\bm0\), \(\sup_t\norm{A_t}<\infty\), and
  \(\norm{R_t(z_t)}\le C\norm{z_t}^2\) for a finite \(C\).  Then the exact
  recurrence forces \(f_t\to\bm0\).  Consequently, a persistent full-state
  invariance defect obstructs attraction to the chosen row-constant stratum.
\item If the metric lower bound in
  Equation~\eqref{row:eq:uniform-metric-conditioning} is removed, convergence in
  scheduled norm need not imply convergence in a fixed Euclidean norm.
\item For every finite observation horizon \(N\), there is a nonnegative
  energy path equal to zero on \(0,\ldots,N\) that does not converge to zero.
\end{enumerate}
\end{theorem}

\begin{proof}
For \(i\), rearrange the exact recurrence:
\[
 f_t=z_{t+1}-A_tz_t-R_t(z_t).
\]
Hence
\[
 \norm{f_t}\le\norm{z_{t+1}}
 +\norm{A_t}\norm{z_t}+C\norm{z_t}^2\longrightarrow0.
\]
The contrapositive gives the obstruction.

For (ii), take a fixed \(z_*\ne\bm0\), let \(H_t=\varepsilon_tI\) with
positive \(\varepsilon_t\to0\), and set \(z_t=z_*\).  Then
\(\norm{z_t}_{H_t}=\sqrt{\varepsilon_t}\norm{z_*}\to0\), although the
Euclidean norm remains constant.

For (iii), define \(E_t=0\) for \(t\le N\) and \(E_t=1\) for \(t>N\).
It agrees exactly with collapse on the observed prefix and then reopens
permanently, so \(E_t\not\to0\).
\end{proof}

\begin{theorem}[Linear observation-kernel obstruction]
\label{row:thm:linear-observation-kernel-obstruction}
Let \(Z\) and \(Y_t\) be normed vector spaces and let
\(O_t:Z\to Y_t\) be linear.  If a vector \(z_*\ne\bm0\) satisfies
\(O_tz_*=\bm0\) for every \(t\), then the constant path \(z_t=z_*\) has
\(O_tz_t\to\bm0\), while \(z_t\not\to\bm0\).  Consequently, collapse of the
linearized observation alone does not imply complete-state collapse.
\end{theorem}

\begin{proof}
For every \(t\), \(O_tz_t=O_tz_*=\bm0\), so the observed path is identically
zero and hence converges to zero.  The state path has constant norm
\(\norm{z_*}>0\), so it cannot converge to zero.
\end{proof}

\begin{proposition}[Nonlinear zero-fiber obstruction and derivative caveat]
\label{row:prop:nonlinear-zero-fiber-obstruction}
Let \(\mathcal O_t:Z\to Y_t\) be arbitrary observation maps with
\(\mathcal O_t(\bm0)=\bm0\).  If
\[
 z_*\in\bigcap_{t\ge0}\mathcal O_t^{-1}(\{\bm0\})
 \quad\text{for some }z_*\ne\bm0,
\]
then the constant path \(z_t=z_*\) has collapsed observations without state
collapse.  By contrast, when the maps are differentiable at zero, membership
of a nonzero vector in \(\bigcap_t\ker D\mathcal O_t(\bm0)\) is not sufficient
for this conclusion.
\end{proposition}

\begin{proof}
The zero-fiber hypothesis gives \(\mathcal O_t(z_t)=\bm0\) for every \(t\),
whereas the constant nonzero state cannot converge to zero.  To see why the
derivative condition is weaker, take \(Z=Y_t=\mathbb R\) and
\(\mathcal O_t(x)=x^2\) for every \(t\).  Then
\(D\mathcal O_t(0)=0_{\mathrm{op}}\), so every direction lies in every derivative kernel,
but \(\mathcal O_t(1)=1\).  The derivative kernel therefore need not lie in
the nonlinear zero fiber.
\end{proof}

These obstructions separate necessary facts from sufficient estimates.  A
bounded nonzero forcing envelope may yield a bounded tube, but an upper bound
alone cannot prove a positive lower floor because vector forcing can cancel.
The exact necessity statement is instead part \(i\): any genuinely convergent
full normal must have a vanishing realized invariance defect under the stated
boundedness conditions.

\FloatBarrier
\par\medskip\noindent
Chapter~\ref{ch:plga-emission} follows the geometric signal through PLGA
and into predictive emission. This transfer requires control of downstream
defects and amplitudes in addition to the optimizer dynamics studied here.

\chapter{From row geometry to PLGA and predictive emission}
\label{ch:plga-emission}
This chapter connects row geometry to downstream operators and predictive
error. It derives the PLGA response-plus-defect description, distinguishes
absolute row energy from normalized concentration, and assembles a finite
geometric-to-predictive error budget.

\section{PLGA transfer as a quotient-defect problem}
\label{row:sec:plga-defect}

The PLGA layer is downstream of the PLDR row map, but it need not preserve
the row-constant face.  For one head, write its implemented fixed-parameter
map as
\begin{align}
 B(X)&=\phi(WX+b)+\epsilon_{\rm P},\nonumber\\
 P(X)&=\exp\bigl(p\odot\log B(X)\bigr),\nonumber\\
 \mathcal P(X)&=aP(X)+b_a,
 \label{row:eq:implemented-plga-map}
\end{align}
where \(\phi(z)=z\operatorname{silu}(z)=z^2\operatorname{sigmoid}(z)\)
acts entrywise and \(\epsilon_{\rm P}=10^{-9}\).  Thus \(B(X)>\bm0_{r\times d}\)
entrywise, so arbitrary real learned exponents are defined.  The matrices
\(W,b,p,a,b_a\) retain their trained row and column structure.

Let \(\rowq(X)=C_rX\), and let
\(
 \bar X=(I-C_r)X=\mathbf1\bar x^\top
\)
be the constant-row projection.  Define
\begin{align}
 \delta(\bar X)&=\rowq\mathcal P(\bar X),
 \label{row:eq:plga-constant-input-defect}\\
 r(X;\bar X)&=\rowq\left[\mathcal P(X)-\mathcal P(\bar X)\right].
 \label{row:eq:plga-segment-response}
\end{align}
The first quantity is a structural invariance defect.  The second is the
response to the occupied quotient segment.

\begin{theorem}[PLGA quotient-defect decomposition]
\label{row:thm:plga-quotient-defect}
For every \(X\),
\begin{equation}
 \rowq\mathcal P(X)=r(X;\bar X)+\delta(\bar X).
 \label{row:eq:plga-defect-decomposition}
\end{equation}
If \(\mathcal P\) is differentiable on the segment
\(\bar X+s(X-\bar X)\), define
\begin{equation}
 \kappa(X)=\sup_{0\le s\le1}
 \norm{\rowq D\mathcal P_{\bar X+s(X-\bar X)}}_{\op}.
 \label{row:eq:plga-segment-derivative-bound}
\end{equation}
Then
\begin{equation}
 \norm{\rowq\mathcal P(X)}_{\Fro}
 \le \kappa(X)\norm{\rowq X}_{\Fro}
 +\norm{\delta(\bar X)}_{\Fro}.
 \label{row:eq:plga-quotient-defect-bound}
\end{equation}
At a row-constant input, the output quotient is exactly the defect.  Hence
the row-constant face is preserved at \(\bar X\) if and only if
\(\delta(\bar X)=\bm0_{r\times d}\).

For a sequence, upstream quotient collapse transfers downstream if
\(\sup_t\kappa(X_t)<\infty\) and
\(\norm{\delta(\bar X_t)}_{\Fro}\to0\).  Neither condition follows from
upstream collapse alone.
\end{theorem}

\begin{proof}
Add and subtract \(\rowq\mathcal P(\bar X)\) to obtain
Equation~\eqref{row:eq:plga-defect-decomposition}.  Since
\(X-\bar X=C_rX=\rowq X\), the fundamental theorem of calculus gives
\[
 r(X;\bar X)=\int_0^1
 \rowq D\mathcal P_{\bar X+s(X-\bar X)}[\rowq X] \,ds.
\]
Taking the Frobenius norm and applying the induced-norm bound gives
\(
 \norm{r(X;\bar X)}_{\Fro}
 \le\kappa(X)\norm{\rowq X}_{\Fro}.
\)
The triangle inequality proves
Equation~\eqref{row:eq:plga-quotient-defect-bound}.  If \(\rowq X=\bm0_{r\times d}\), then
\(X=\bar X\), the response is zero, and the output quotient equals the
defect.  This proves the invariance equivalence.  The sequential statement
follows by taking limits in the displayed upper bound.
\end{proof}

The direct segment response
\(\norm{r(X;\bar X)}\) and its realized secant ratio are descriptive finite
quantities.  A ratio formed by dividing by a very small input quotient is not
the independently controlled \(\kappa\) required by the theorem.  Near zero,
the absolute response and absolute defect remain primary.
Section~\ref{row:sec:plga-defect-evidence} measures these quantities at the
registered endpoints, and Section~\ref{rg:sec:plga-checkpoint-results}
reports the checkpoint sequence and constrained-head controls.

\subsection{Five sufficient structural conditions}

Let a matrix be row constant when all its rows agree.  Let
\(\mathbf1\) have the compatible length.

\begin{proposition}[Sufficient PLGA row preservation]
\label{row:prop:plga-row-preservation}
The implemented PLGA map sends every row-constant \(X\) to a row-constant
output if all five conditions hold:
\begin{enumerate}[label=(\alph*)]
\item \(W\mathbf1\in\operatorname{span}\{\mathbf1\}\);
\item \(b\) is row constant;
\item \(p\) is row constant;
\item \(a\mathbf1\in\operatorname{span}\{\mathbf1\}\);
\item \(b_a\) is row constant.
\end{enumerate}
These conditions are sufficient, not necessary.
\end{proposition}

\begin{proof}
Let \(X=\mathbf1x^\top\).  The first condition gives
\[
 WX=(W\mathbf1)x^\top=\mathbf1y^\top
\]
for some \(y\).  Adding the row-constant \(b\) preserves row constancy.
The entrywise map \(\phi(\cdot)+\epsilon_{\rm P}\) therefore produces a
positive row-constant matrix.  The third condition applies the same exponent row
to every identical base row, so \(P(X)=\mathbf1q^\top\) for some \(q\).
The fourth condition gives
\(
 aP(X)=(a\mathbf1)q^\top=\mathbf1w^\top.
\)
Adding the row-constant \(b_a\) proves that \(\mathcal P(X)\) is row
constant.  Special parameter cancellations can also make the defect zero,
so the conditions are not necessary.
\end{proof}

The five empirical residuals are the quotient norms of
\(W\mathbf1\), \(b\), \(p\), \(a\mathbf1\), and \(b_a\), respectively.
They diagnose these sufficient conditions.  The direct defect
\(\delta(\bar X)\) is the definitive test at the registered constant input.

Let \(r,c\ge1\), and let \(Z\in\R^{r\times c}\) contain the finite real
logits supplied to ordinary rowwise softmax \(\mathcal S\), restricted
to one common nonempty allowed support of size \(c\). Its Jacobian at
a probability vector \(q\) is \(\diag(q)-qq^\top\), whose Euclidean
operator norm is at most \(1/2\). A proved pre-softmax quotient bound
therefore gives a sufficient post-softmax bound. Pre-softmax row
invariance is not necessary, however, because softmax removes an
additive constant in each row. With
\[
 C_r=I_r-r^{-1}\mathbf1_r\mathbf1_r^\top,\qquad
 C_c=I_c-c^{-1}\mathbf1_c\mathbf1_c^\top,
\]
the exact criterion and upper bound are
\[
 \begin{aligned}
 C_r\mathcal S(Z)=\bm0_{r\times c}
 &\quad\Longleftrightarrow\quad C_rZC_c=\bm0_{r\times c},\\
 \norm{C_r\mathcal S(Z)}_F
 &\le\tfrac12\norm{C_rZC_c}_F
 \le\tfrac12\norm{C_rZ}_F.
 \end{aligned}
\]
For example,
\[
 Z=\begin{pmatrix}1&1\\2&2\end{pmatrix},\qquad
 \norm{C_2Z}_F=1,\qquad
 \mathcal S(Z)=\begin{pmatrix}1/2&1/2\\1/2&1/2\end{pmatrix}.
\]
Here \(C_2\mathcal S(Z)=\bm0_{2\times2}\), although the pre-softmax
row quotient is nonzero. Thus no corresponding positive lower bound
holds without additional assumptions. These statements concern the
logits supplied to softmax. Transferring a feature-space PLGA quotient
to token scores still requires the intervening query--operator--key
map. Rows with different causal masks require a separate comparison.

\section{PLGA and complete-state transfer}
\label{rg:sec:plga-transfer}

\subsection{Constant-input invariance defect}

The learned PLGA map is downstream of the PLDR row program.  Upstream row
collapse need not be inherited automatically.  For one head, write the
implemented fixed-parameter map schematically as
\begin{align}
 B(X)&=\phi(WX+b)+\epsilon_{\rm P},\nonumber\\
 P(X)&=\exp\!\bigl(p\odot\log B(X)\bigr),\nonumber\\
 \mathcal P(X)&=aP(X)+b_a,
 \label{rg:eq:implemented-plga}
\end{align}
where \(\phi(z)=z\operatorname{silu}(z)\) acts entrywise and
\(\epsilon_{\rm P}=10^{-9}\) in the implementation. Since
\(z\operatorname{silu}(z)=z^2\operatorname{sigmoid}(z)\ge0\), the base is
strictly positive. Arbitrary real learned powers are therefore well defined,
and the fixed-parameter map is continuously differentiable for finite inputs.

Let \(\mathcal C=\{\mathbf 1u^{\mathsf T}:u\text{ compatible}\}\) be the
row-constant face.  For fixed parameters, global face preservation means
\begin{equation}
 \rowq\mathcal P(\mathbf 1u^{\mathsf T})=\bm0_{r\times d}
 \quad\text{for every compatible }u.
 \label{rg:eq:plga-functional-criterion}
\end{equation}
This is the exact functional preservation criterion, not an independent
parameter characterization.

\begin{proposition}[PLGA structural sufficiency and nondegenerate converse]
\label{rg:prop:plga-face-preservation}
The five conditions
\begin{equation}
 \rowq W\mathbf 1=\bm0,\qquad
 \rowq b=\bm0_{r\times d},\qquad
 \rowq p=\bm0_{r\times d},\qquad
 \rowq a\mathbf 1=\bm0,\qquad
 \rowq b_a=\bm0_{r\times d}
 \label{rg:eq:plga-five-conditions}
\end{equation}
are jointly sufficient. Under the first three, write
\(W\mathbf1=\kappa_W\mathbf1\),
\(p=\mathbf1\pi^{\mathsf T}\), and
\(P(\mathbf1u^{\mathsf T})=\mathbf1v(u)^{\mathsf T}\).
Here \(\kappa_W\) is the common row sum of \(W\), and \(\pi\) is the
column vector of exponents whose transpose is the common row of \(p\).
For the implemented nonlinearity, \(v\) is constant as a function of \(u\)
if and only if \(\kappa_W=0\) or \(p=\bm0_{r\times d}\). If \(v\) is nonconstant and
Equation~\eqref{rg:eq:plga-functional-criterion} holds, then the last two
conditions are also necessary.

Neither the first three conditions nor the last two are necessary without
additional nondegeneracy. In particular, \(a=\bm0_{r\times r}\) and \(\rowq b_a=\bm0_{r\times d}\)
annihilate every intermediate defect, regardless of \(W,b,p\). Also, if
\(\kappa_W=0\) or \(p=\bm0_{r\times d}\), then \(v(u)\equiv v_0\). Choose any nonzero
centered vector \(g\), set
\(a=r^{-1}g\mathbf1^{\mathsf T}\), and set
\(b_a=-gv_0^{\mathsf T}\). This nonempty cancellation family preserves the
face while violating both final conditions.
\end{proposition}

\begin{proof}
A matrix lies in \(\mathcal C\) exactly when its row quotient is zero, which
gives Equation~\eqref{rg:eq:plga-functional-criterion}. For sufficiency, the
first condition gives \(W\mathbf1=\kappa_W\mathbf1\), while the next two make
\(b\) and \(p\) row constant. Thus every entrywise operation in \(B\) and
\(P\) preserves row constancy and
\(P(\mathbf1u^{\mathsf T})=\mathbf1v(u)^{\mathsf T}\). The final two
conditions give \(a\mathbf1=\kappa\mathbf1\) and
\(b_a=\mathbf1\gamma^{\mathsf T}\), hence
\[
 \mathcal P(\mathbf1u^{\mathsf T})
 =\mathbf1\bigl(\kappa v(u)+\gamma\bigr)^{\mathsf T}\in\mathcal C.
\]

Under the first three conditions, write
\(b=\mathbf1\beta^{\mathsf T}\) and
\(p=\mathbf1\pi^{\mathsf T}\).  Coordinatewise,
\[
 v_j(u)=
 \left[\phi(\kappa_Wu_j+\beta_j)+\epsilon_{\rm P}\right]^{\pi_j}.
\]
If \(\kappa_W=0\), this is independent of \(u\), and if \(p=\bm0_{r\times d}\), it is
identically one. Conversely, if \(\kappa_W\ne0\) and \(p\ne\bm0_{r\times d}\), choose
\(j\) with \(\pi_j\ne0\). The affine argument spans \(\R\);
\(\phi(z)=z^2\operatorname{sigmoid}(z)\) is nonconstant;
\(\phi(z)+\epsilon_{\rm P}>0\); and raising positive values to the nonzero
power \(\pi_j\) is injective. Hence \(v_j\), and therefore \(v\), is
nonconstant. This proves the stated equivalence.

For the partial converse, put \(g=\rowq a\mathbf1\) and
\(D=\rowq b_a\). Face preservation is equivalent to
\begin{equation}
 gv(u)^{\mathsf T}+D=\bm0_{r\times d}\quad\text{for every }u.
 \label{rg:eq:plga-partial-converse}
\end{equation}
Choose \(u_1,u_2\) with \(v(u_1)\ne v(u_2)\) and subtract the two
identities. The outer product
\(g(v(u_1)-v(u_2))^{\mathsf T}\) is zero. Since its second factor is
nonzero, \(g=\bm0\); Equation~\eqref{rg:eq:plga-partial-converse} then gives
\(D=\bm0_{r\times d}\). These are the fourth and fifth conditions. Finally, direct
substitution verifies the annihilation family.  For the cancellation family,
\(\mathbf1^{\mathsf T}\mathbf1=r\) and \(\rowq g=g\), so
\(\rowq a\mathbf1=g\) and \(\rowq b_a=-gv_0^{\mathsf T}\).  Their sum in
Equation~\eqref{rg:eq:plga-partial-converse} is zero.  Since \(g\ne\bm0\) and
\(v_0\) has positive entries, both final structural residuals are nonzero.
This completes the proof.
\end{proof}

Let \(\bar X=(I-C_r)X\) be the row-constant projection and define
\begin{equation}
 \delta(\bar X)=\rowq\mathcal P(\bar X),
 \qquad
 r(X;\bar X)=\rowq\bigl(\mathcal P(X)-\mathcal P(\bar X)\bigr).
 \label{rg:eq:plga-response-defect}
\end{equation}
The first term is a structural invariance defect.  The second is the response
to the occupied quotient segment.

\begin{theorem}[PLGA response-plus-defect transfer]
\label{rg:thm:plga-rg-transfer}
For every input,
\begin{equation}
 \rowq\mathcal P(X)=r(X;\bar X)+\delta(\bar X).
 \label{rg:eq:plga-defect-decomposition}
\end{equation}
If \(\mathcal P\) is continuously differentiable on a neighborhood of the
segment \(\bar X+s(X-\bar X)\), \(0\le s\le1\), let
\(D\mathcal P_A\) be its derivative between matrix spaces equipped with the
Frobenius norm, and set
\begin{equation}
 \kappa(X)=\sup_{0\le s\le1}
 \norm{\rowq D\mathcal P_{\bar X+s(X-\bar X)}}_{\op}.
 \label{rg:eq:plga-kappa}
\end{equation}
Then
\begin{equation}
 \norm{\rowq\mathcal P(X)}_{\Fro}
 \le \kappa(X)\norm{\rowq X}_{\Fro}
   +\norm{\delta(\bar X)}_{\Fro}.
 \label{rg:eq:plga-norm-bound}
\end{equation}
At a row-constant input, the output quotient is exactly \(\delta(\bar X)\).
Hence the face is preserved at that input if and only if
\(\delta(\bar X)=\bm0_{r\times d}\).
\end{theorem}

\begin{proof}
Add and subtract \(\rowq\mathcal P(\bar X)\) to obtain
Equation~\eqref{rg:eq:plga-defect-decomposition}.  Since
\(X-\bar X=C_rX=\rowq X\), the fundamental theorem of calculus on the
segment gives
\[
 r(X;\bar X)=\int_0^1
 \rowq D\mathcal P_{\bar X+s(X-\bar X)}[\rowq X]\,ds.
\]
The Frobenius-induced operator-norm inequality and triangle inequality yield
Equation~\eqref{rg:eq:plga-norm-bound}.  If \(\rowq X=\bm0_{r\times d}\), then \(X=\bar X\)
and the response term is zero, proving the face statement.
\end{proof}

\subsection{Embedding the defect in the affine RG}

Put \(E_{\rm in}=\norm{\rowq X}_{\Fro}^2\),
\(E_{\rm out}=\norm{\rowq\mathcal P(X)}_{\Fro}^2\), and
\(d=\norm{\delta(\bar X)}_{\Fro}\).  For every \(\eta>0\), Young's
inequality gives
\begin{equation}
 (u+v)^2\le(1+\eta)u^2+(1+\eta^{-1})v^2.
 \label{rg:eq:young-two-term}
\end{equation}
Applying it to Equation~\eqref{rg:eq:plga-norm-bound} gives the positive affine
energy edge
\begin{equation}
 E_{\rm out}\le
 \underbrace{(1+\eta)\kappa(X)^2}_{q_{\rm PLGA}}E_{\rm in}
 +\underbrace{(1+\eta^{-1})d^2}_{\zeta_{\rm PLGA}}.
 \label{rg:eq:plga-affine-edge}
\end{equation}
Thus PLGA response is a gain coordinate and its constant-input defect is a
source coordinate.  A sequence of downstream maps composes by the same
ordered law as Equation~\eqref{rg:eq:edge-composition}.  Optimizing \(\eta\)
can tighten a finite edge, but no choice of \(\eta\) removes a nonzero defect.

\begin{corollary}[Sequential downstream collapse]
\label{rg:cor:plga-sequential-collapse}
For a sequence \(X_t\), suppose
\begin{equation}
 E_{\rm in,t}\to0,
 \qquad \sup_t\kappa(X_t)<\infty,
 \qquad \norm{\delta(\bar X_t)}_{\Fro}\to0.
 \label{rg:eq:plga-transfer-assumptions}
\end{equation}
Then \(E_{\rm out,t}\to0\).  Conversely, at any exact row-constant input, a
nonzero defect obstructs downstream row collapse at that endpoint.
\end{corollary}

\begin{proof}
Use Equation~\eqref{rg:eq:plga-norm-bound}.  Its first term tends to zero by the
uniform response bound, and its second tends to zero by assumption.  At an
exact row-constant input, Theorem~\ref{rg:thm:plga-rg-transfer} gives
\(\rowq\mathcal P(X)=\delta(\bar X)\), so a nonzero defect yields positive
output energy.
\end{proof}

The corollary is a sequential transfer theorem, not a neighborhood theorem.
A realized secant ratio can be small even when the independent supremum in
Equation~\eqref{rg:eq:plga-kappa} is large.  Near the face, absolute response and
absolute defect are therefore primary. Likewise, directional derivatives
sampled along the occupied segment are lower bounds on the full operator
supremum, not certificates of it.
Section~\ref{rg:sec:plga-checkpoint-results} measures
the realized secant, sampled directional response, and constant-input defect
separately.

\subsection{Complete optimizer state}

AdamW carries parameter, first-moment, and second-moment coordinates
\cite{kingma2015adam,loshchilov2019adamw}.  Let \(y_t\) denote a complete
transverse state.

Theorem~\ref{rg:thm:complete-state-kernel-rg} records standard kernel blocking
and the scale-indexed closure criterion on this state.  The inequality below
transports any proved complete-state Lyapunov comparison to the row-energy
observable without assuming energy lumpability.

\paragraph{Complete-state product-convolution application.}
The complete-state comparison mechanism in
Theorem~\ref{row:thm:observable-complete-state} is an application of
Theorem~\ref{rg:thm:affine-rg-semigroup}. Suppose a nonnegative comparison quantity \(V_t=V(y_t)\) satisfies
\begin{equation}
 V_{t+1}\le a_tV_t+b_t,\qquad a_t,b_t\ge0.
 \label{rg:eq:complete-state-affine}
\end{equation}
For integers \(n\ge k\), define
\begin{align}
 A(n,k)&=\prod_{j=k}^{n-1}a_j,\nonumber\\
 B(n,k)&=\sum_{j=k}^{n-1}
   \left(\prod_{s=j+1}^{n-1}a_s\right)b_j,
 \label{rg:eq:complete-state-block}
\end{align}
where an empty product is one and an empty sum is zero.  Then
\begin{equation}
 V_n\le A(n,k)V_k+B(n,k).
 \label{rg:eq:complete-state-bound}
\end{equation}
Thus \((A(n,k),B(n,k))\) is the same ordered positive-affine
product-convolution as the observable RG.  If
\(0\le E_n\le c_nV_n\), \(c_n\ge0\), \(\sup_n c_n<\infty\), and the
right-hand side of Equation~\eqref{rg:eq:complete-state-bound} tends to zero for fixed \(k\), then
\(E_n\to0\).

\begin{proof}[Derivation]
For \(n=k\), the conventions give \(A(k,k)=1\), \(B(k,k)=0\), and
Equation~\eqref{rg:eq:complete-state-bound} is equality.  Suppose it holds at
\(n\).  Then
\[
 \begin{split}
 V_{n+1}
 &\le a_nV_n+b_n\\
 &\le a_nA(n,k)V_k+a_nB(n,k)+b_n.
 \end{split}
\]
The first coefficient is
\(a_nA(n,k)=\prod_{j=k}^{n}a_j=A(n+1,k)\).  Distributing \(a_n\) through
the sum defining \(B(n,k)\) and adjoining the final term \(b_n\) gives
\[
 a_nB(n,k)+b_n
 =\sum_{j=k}^{n}
   \left(\prod_{s=j+1}^{n}a_s\right)b_j
 =B(n+1,k).
\]
This completes the induction.  The last assertion follows from
\[
 0\le E_n\le\left(\sup_m c_m\right)
 \bigl(A(n,k)V_k+B(n,k)\bigr)
\]
and the squeeze theorem.
\end{proof}

The converse is false without observability.  Hidden optimizer modes can
remain nonzero while \(E_t=0\), and a later update can project them back into
the row quotient.  In RG language, such a projection appears as a source
field.  A complete-state RG and an observable RG can therefore share the
affine algebra while having different gains, sources, fixed sets, and
universality classes.

\section{Absolute collapse, relative concentration, and predictive error}
\label{sec:bridge-geometry}
The quotient used in the row dynamics and the normalized collective used
in the model-wide laws describe related but inequivalent properties.
The following comparison fixes the amplitude hypothesis required at this
interface. It applies to the same specified matrix stage; transferring
between stages also requires the PLGA defect bound above.

\begin{proposition}[Absolute and normalized row energy]
\label{prop:absolute-relative}
Let $r\ge2$, $d\ge1$, and let $A_t\in\R^{r\times d}$ have total energy
$T_t=\|A_t\|_F^2>0$. Set $e_t=\|C_rA_t\|_F^2$ and $u_t=e_t/T_t$.
Then $0\le u_t\le1$ and $e_t=u_tT_t$.
If $\sup_tT_t<\infty$, $u_t\to0$ implies $e_t\to0$.
If $\inf_tT_t>0$, $e_t\to0$ implies $u_t\to0$.
Consequently the two limits are equivalent under both amplitude bounds.
Neither implication holds for arbitrary positive $T_t$.
\end{proposition}
\begin{proof}
Orthogonal projection is contractive, so $0\le e_t\le T_t$.
If $T_t\le M<\infty$, then $0\le e_t\le Mu_t$, proving the first
limit by squeezing. If $T_t\ge m>0$, then $0\le u_t\le e_t/m$,
proving the second.
For failure without an upper bound, take a fixed nonzero centered
matrix $Z$ and a nonzero common row $\mathbf1c^\top$, and put
$A_t=Z+t\mathbf1c^\top$, $t\ge1$. Orthogonality gives
$e_t=\|Z\|_F^2>0$ and
$T_t=\|Z\|_F^2+t^2r\|c\|^2$, so $u_t\to0$ without absolute
collapse. For failure without a lower bound, put $A_t=t^{-1}B$ for
any fixed $B$ with nonzero row contrast. Then $e_t\to0$ while
$u_t=\|C_rB\|_F^2/\|B\|_F^2>0$ remains constant.
\end{proof}

The same distinction governs finite-step transport. Normalizing the row
work ledger introduces a moving total-energy denominator. Its exact
increment is the finite flux of
Proposition~\ref{model:prop:collective-clock}, including the denominator
at the successor (Section~\ref{model:sec:collective-clock}). A derivative
evaluated at the incoming denominator requires an additional finite-increment
error estimate. Sections~\ref{model:sec:collective-flux-results} and
\ref{model:sec:row-path-results} measure the one-step and chronological
transport on recorded native continuations.

\begin{proposition}[A finite geometric-to-predictive error budget]
\label{prop:row-predictive-bridge}
Fix a finite prefix and two decoder computations
$x_\ell=F_\ell(x_{\ell-1})$ and
$\widetilde x_\ell=\widetilde F_\ell(\widetilde x_{\ell-1})$ with
the same input $x_0=\widetilde x_0$. Suppose $F_\ell$ is
$L_\ell$-Lipschitz on the relevant states and
\[
 \|F_\ell(\widetilde x_{\ell-1})-
       \widetilde F_\ell(\widetilde x_{\ell-1})\|
 \le \varepsilon_\ell.
\]
Suppose the common output map is $L_{\rm out}$-Lipschitz.
For the final logits $z,\widetilde z$, define
\[
 B=L_{\rm out}\sum_{\ell=1}^{L}
       \left(\prod_{j=\ell+1}^{L}L_j\right)\varepsilon_\ell.
\]
Then $\|z-\widetilde z\|_2\le B$, and for
$p=\softmax(z)$, $\widetilde p=\softmax(\widetilde z)$,
\begin{equation}
 \KL(p\|\widetilde p)\le\frac14 B^2.
 \label{eq:bridge-kl}
\end{equation}
In particular, if a proved row-to-decoder bound gives
$\varepsilon_\ell\le a_\ell\sqrt{e_\ell}+\delta_\ell$ on
all required hybrid inputs, it may be substituted in $B$.
Here $\delta_\ell$ retains any constant-input or replacement defect.
\end{proposition}
\begin{proof}
Writing $d_\ell=\|x_\ell-\widetilde x_\ell\|$, add and
subtract $F_\ell(\widetilde x_{\ell-1})$ to obtain
$d_\ell\le L_\ell d_{\ell-1}+\varepsilon_\ell$.
Induction from $d_0=0$ gives the displayed product-convolution
bound, and the output Lipschitz condition gives the logit bound.
For $f(z)=\log\sum_i e^{z_i}$,
$\nabla^2f(z)=\operatorname{diag}(p)-pp^\top$.
For a unit vector $v$, its quadratic form is the variance of the
values $v_i$ under $p$. Any variable in $[a,b]$ has variance at most
$(b-a)^2/4$: its second moment is bounded by
$(a+b)$ times its mean minus $ab$, and completion of the square
gives the claim. Since
$\max_i v_i-\min_i v_i\le\sqrt2\|v\|_2$, the Hessian
operator norm is at most $1/2$. The exact Bregman formula is
\[
 \KL(p\|\widetilde p)
 =f(\widetilde z)-f(z)-\nabla f(z)^\top(\widetilde z-z).
\]
Taylor's integral remainder is bounded by
$\int_0^1(1-s)\tfrac12\|\widetilde z-z\|_2^2\,ds
=\tfrac14\|\widetilde z-z\|_2^2$. This proves
\eqref{eq:bridge-kl}. Substitution of valid upper bounds for
the nonnegative $\varepsilon_\ell$ preserves the estimate.
\end{proof}

The proposition is a conditional bridge, not an inference certificate
for the measured models. Its domain includes the states created by the
reduced computation, not just the native endpoint probes. Large
downstream gains or a persistent defect can prevent a small row
contrast from giving a useful bound. The result also explains why a
directly measured cache risk can be informative when no uniform
row-to-prediction certificate is available.
Section~\ref{model:sec:cache-risk-law} derives that conditional risk, and
Sections~\ref{model:sec:operator-cache-results} and
\ref{model:sec:cache-risk-results} report the operator-substitution and
prefix-transfer experiments.

\subsection{A worked chain from row energy to predictive fidelity}
\label{sec:worked-row-prediction}
Consider the two-row, one-column matrix
\[
 X_t=\begin{pmatrix}\mu_t+a_t\\\mu_t-a_t\end{pmatrix},\qquad
 Z_t=C_2X_t=\begin{pmatrix}a_t\\-a_t\end{pmatrix},\qquad E_t=2a_t^2.
\]
Here \(a_t,\mu_t\) are scalar row coordinates, unrelated to the tangential
chart coordinate in Section~\ref{row:sec:trajectory-chart}. If the realized
contrast increment is \(b_t\), then
\[
 D_t=\begin{pmatrix}b_t\\-b_t\end{pmatrix},\qquad
 E_{t+1}-E_t=4a_tb_t+2b_t^2.
\]
For \(a_t=1\) and \(b_t=-1/2\), the energy falls from \(2\) to \(1/2\).
The work term is \(-2\), the finite-step charge is \(1/2\), the radial
coefficient is \(\alpha_t=1/2\), the tangential increment is
\(\bm0_{2\times1}\), and the exact gain is \(1/4\). These conclusions
follow from the observed endpoints and the finite work identity.

Prediction requires the hidden increment. The same incoming matrix with
\(b_t=1/2\) instead gives successor energy \(9/2\). Thus energy alone
does not select a successor. An autonomous predictive law requires agreement
of the projected successor laws on every admitted incoming fiber, or an
explicitly qualified approximate closure; see
Section~\ref{rg:sec:kernel-lumpability}. Complete optimizer and remaining-corpus state
can determine the increment without making energy itself closed.

For inference at a fixed admitted proper prefix, put
\(X_t^{\rm c}=\mathbf1\mu_t\). Then
\(\|X_t-X_t^{\rm c}\|_F=\sqrt{E_t}\) exactly. Let \(G\) be the downstream
operator generator and suppose, on the required inputs,
\[
 \|G(X_t)-G(X_t^{\rm c})\|_F
 \le K\|X_t-X_t^{\rm c}\|_F.
\]
For a proposed replacement \(\widehat G\), define the common-input and
replacement defect \(\delta_t=\|G(X_t^{\rm c})-\widehat G\|_F\).
The triangle inequality proves
\begin{equation}
 \|G(X_t)-\widehat G\|_F\le K\sqrt{E_t}+\delta_t.
 \label{eq:worked-operator-budget}
\end{equation}
A common-row input may produce a nonconstant downstream operator or a common
component that varies across contexts. Small row energy does not bound
\(\delta_t\) and does not by itself justify a shared cache.

If the remaining decoder has justified logit sensitivity \(L\) on all
necessary native and hybrid inputs, then
\(\|\ell-\widehat\ell\|_2\le L(K\sqrt{E_t}+\delta_t)\).
Proposition~\ref{prop:row-predictive-bridge} yields
\begin{equation}
 \KL\bigl(\softmax(\ell)\|\softmax(\widehat\ell)\bigr)
 \le\frac{L^2}{4}(K\sqrt{E_t}+\delta_t)^2.
 \label{eq:worked-predictive-budget}
\end{equation}
For illustration only, \(K=L=1\), \(\delta_t=0\), and \(E_t=1/2\)
give the bound \(1/8\). These constants are not estimates for a trained
PLDR model. Several replaced layers require the proposition's full
product-convolution budget.

This conditional fidelity bound differs from measured cache risk and
external-target negative log likelihood (NLL). Improved target NLL does not
imply native-model fidelity; small native KL does not determine the sign
of the target-NLL change. The paired initializations and specified contexts
in Section~\ref{model:sec:cache-risk-results} test declared cache targets.
They do not establish the illustrative uniform sensitivities used here.

\FloatBarrier
\par\medskip\noindent
Chapter~\ref{ch:row-evidence} examines these mechanisms in recorded
experiments. Its finite interventions and numerical controls test the work,
optimizer and PLGA descriptions developed in the preceding chapters.

\chapter{Observed row mechanisms and finite interventions}
\label{ch:row-evidence}
This chapter presents the observations and interventions that resolve
finite row mechanisms. The results cover arithmetic qualification, gate and
shape changes, optimizer-state response, PLGA defects and the exact
finite-increment balances, with the statistical unit stated for each study.

\section{Executed evidence}
\label{row:sec:executed-evidence}

\subsection{Logical role and analysis unit}

The primary statistical unit is a trajectory-layer pair.  The three paths
are denoted A, B, and C, with seeds \(10444\), \(11444\), and \(12444\).
Within each path and layer, \(24\) registered contexts and four heads give
\(96\) map observations.  They are shared-weight subsamples, not \(96\)
independent replicas.  The complete population has nine trajectory-layer
units and \(864\) physical maps at each registered timepoint.

The fixed 24-context registry is reused across times and interventions.
Each source path has its own nonrepeating RefinedWeb chunk order;
these measurements condition on the recorded corpus and checkpoints.

The experiment measures finite objects from the theory and retains adverse
outcomes.  It does not infer an infinite-time limit from a finite trajectory.
An exact identity is treated as an implementation check.  A derivative is
treated as a local object only after signed probes test it.  A component
effect is interpreted only after same-device no-op and numerical-placebo
controls.  A downstream transfer claim is evaluated through its absolute
constant-input defect.

The trained model has three decoder layers, four heads, head width \(64\),
and eight row-program residual units.  Training uses float32 AdamW with
\(\beta_1=0.9\), \(\beta_2=0.95\),
\(\epsilon_{\rm A}=10^{-5}\), weight decay \(0.1\), and value clipping at
\(1\).  The learning rate warms for \(250\) updates to
\(7.5\times10^{-4}\) and is then constant.  Batch size is \(32\), context
length is \(256\), and each path has its own nonrepeating RefinedWeb chunk
order.  The mechanism campaign reuses the content-bound checkpoints and data
orders of three completed \(65{,}536\)-update paths.

\subsection{Frozen measurement graph}

The measurement design contains \(36\) GPU observation records and one
numerical reduction:

\begin{enumerate}[label=(\arabic*)]
\item six float32 and float64 precision replays at updates \(16{,}384\) and
  \(65{,}536\);
\item nine layer-resolved final-gate operators at update \(4096\);
\item nine same-device one-step component records at update \(4096\);
\item nine sign-paired, source-written perturbation validations at update
  \(4096\); and
\item three direct PLGA quotient-defect records at update \(65{,}536\).
\end{enumerate}

Every record binds the source checkpoint digest, real next batch, fixed map
registry, trajectory, layer, device, and output schema.  Scientific outcomes
do not control which observations are included.

\clearpage
\subsection{Float64 endpoints and mixed closure}
\label{row:sec:precision-endpoints}

The displayed $E$ is absolute centered row energy. Layer summaries aggregate
24 fixed contexts and four shared-weight heads; they are not normalized row
fractions or additional independent training realizations. Their endpoint
threshold has this observable and arithmetic scope.

\begin{table*}[t]
\centering
\caption{Float64 layer-resolved endpoint energy and late gains from updates $16{,}384$ to $65{,}536$.  Each entry is the median over the $24\times4=96$ registered maps in one trajectory-layer unit.}
\label{row:tab:layer-endpoint-gains}
\begin{tabular}{ccrrrrr}
\toprule
Path & Layer & $E_{65536}$ & $\Delta\log_{10}E$ & $\Delta\log_{10}S$ & $\Delta\log_{10}a$ & small \\
\midrule
A & 0 & $1.726\mathord{\times}10^{-14}$ & -16.077 & -12.575 & -3.672 & yes \\
A & 1 & $2.090\mathord{\times}10^{-16}$ & -15.636 & -12.607 & -3.051 & yes \\
A & 2 & $61.6185$ & -0.667 & -0.276 & -0.341 & no \\
B & 0 & $1.839\mathord{\times}10^{-11}$ & -9.102 & -7.468 & -1.663 & yes \\
B & 1 & $9.169\mathord{\times}10^{-5}$ & 0.336 & 2.228 & -1.884 & yes \\
B & 2 & $9.838\mathord{\times}10^{-12}$ & -13.457 & -9.599 & -3.871 & yes \\
C & 0 & $1.548\mathord{\times}10^{-15}$ & -14.599 & -13.740 & -0.802 & yes \\
C & 1 & $522.9156$ & 0.136 & 0.102 & 0.023 & no \\
C & 2 & $21.8474$ & -1.250 & 0.071 & -1.302 & no \\
\bottomrule
\end{tabular}
\end{table*}

\begin{table*}[t]
\centering
\caption{Endpoint arithmetic resolution.  Relative error compares the float32 and float64 physical energies.  A false coordinate zero is exactly zero after the float32 forward path and positive in the float64 replay.}
\label{row:tab:arithmetic-resolution}
\begin{tabular}{ccrrr}
\toprule
Path & Layer & median relative error & maximum relative error & false coordinate zeros \\
\midrule
A & 0 & $0.0844$ & $0.2263$ & 812 \\
A & 1 & $2.6579$ & $12.3469$ & 4016 \\
A & 2 & $4.992\mathord{\times}10^{-8}$ & $1.648\mathord{\times}10^{-7}$ & 0 \\
B & 0 & $3.419\mathord{\times}10^{-4}$ & $0.0018$ & 0 \\
B & 1 & $2.216\mathord{\times}10^{-6}$ & $2.208\mathord{\times}10^{-5}$ & 0 \\
B & 2 & $0.0027$ & $0.0129$ & 0 \\
C & 0 & $2.5406$ & $5.3073$ & 1857 \\
C & 1 & $2.665\mathord{\times}10^{-8}$ & $9.699\mathord{\times}10^{-8}$ & 0 \\
C & 2 & $5.447\mathord{\times}10^{-8}$ & $4.466\mathord{\times}10^{-7}$ & 0 \\
\midrule
\multicolumn{4}{r}{Total false coordinate zeros} & 6685 \\ 
\bottomrule
\end{tabular}
\end{table*}

\begin{table*}[t]
\centering
\caption{Lifted final-gate AdamW derivative at update $4096$.  The state is $(\gamma,m,v)\in\mathbb R^{192}$.  Norms use the declared live-second-moment scaling.  Probe error is the maximum relative residual at the smaller signed amplitude.}
\label{row:tab:gate-operator-diagnostics}
\begin{tabular}{ccrrrrr}
\toprule
Path & Layer & $\|A_t\|$ & $\rho(A_t)$ & nonnormality & $\|f_t\|$ & probe error \\
\midrule
A & 0 & $8.5824$ & $1.0090$ & $8.5057$ & $1.058\mathord{\times}10^{4}$ & $8.265\mathord{\times}10^{-4}$ \\
A & 1 & $78.7713$ & $1.2094$ & $65.1313$ & $6.930\mathord{\times}10^{4}$ & $7.496\mathord{\times}10^{-5}$ \\
A & 2 & $147.8915$ & $1.2220$ & $121.0226$ & $2.091\mathord{\times}10^{5}$ & $0.0026$ \\
B & 0 & $50.1910$ & $1.1244$ & $44.6363$ & $2657.7585$ & $5.949\mathord{\times}10^{-4}$ \\
B & 1 & $35.0231$ & $1.0818$ & $32.3755$ & $6.050\mathord{\times}10^{5}$ & $1.010\mathord{\times}10^{-4}$ \\
B & 2 & $446.5383$ & $1.4713$ & $303.5056$ & $1.088\mathord{\times}10^{5}$ & $0.0095$ \\
C & 0 & $54.0178$ & $1.1734$ & $46.0338$ & $1195.0262$ & $7.282\mathord{\times}10^{-4}$ \\
C & 1 & $791.8063$ & $1.2797$ & $618.7497$ & $3.704\mathord{\times}10^{5}$ & $0.0070$ \\
C & 2 & $213.4014$ & $1.1601$ & $183.9468$ & $2.846\mathord{\times}10^{6}$ & $0.0016$ \\
\bottomrule
\end{tabular}
\end{table*}

\begin{table*}[t]
\centering
\caption{Same-device one-step effects at update $4096$.  Responses are root-mean-square physical quotient norms over the 96 maps in one trajectory-layer unit.  The final column is the unopened half-amplitude prediction residual; the fixed pass threshold is $0.25$.}
\label{row:tab:one-step-mechanism-tests}
\resizebox{\textwidth}{!}{\begin{tabular}{ccrrrrrr}
\toprule
Path & Layer & one ULP & gate removed & upstream removed & joint removed & ULP/min & prediction error \\
\midrule
A & 0 & $1.030\mathord{\times}10^{-7}$ & $3.987\mathord{\times}10^{-4}$ & $0.0050$ & $0.0051$ & $2.583\mathord{\times}10^{-4}$ & $0.1808$ \\
A & 1 & $4.479\mathord{\times}10^{-7}$ & $2.746\mathord{\times}10^{-4}$ & $0.0467$ & $0.0469$ & $0.0016$ & $0.3137$ \\
A & 2 & $3.363\mathord{\times}10^{-7}$ & $0.0037$ & $1.3486$ & $1.3497$ & $9.116\mathord{\times}10^{-5}$ & $1.0814$ \\
B & 0 & $5.962\mathord{\times}10^{-6}$ & $0.0103$ & $0.0684$ & $0.0699$ & $5.772\mathord{\times}10^{-4}$ & $0.6989$ \\
B & 1 & $1.295\mathord{\times}10^{-5}$ & $0.0016$ & $0.4249$ & $0.4252$ & $0.0082$ & $0.2305$ \\
B & 2 & $4.232\mathord{\times}10^{-7}$ & $0.0111$ & $0.9780$ & $0.9786$ & $3.803\mathord{\times}10^{-5}$ & $0.6855$ \\
C & 0 & $4.946\mathord{\times}10^{-6}$ & $0.0038$ & $0.0506$ & $0.0514$ & $0.0013$ & $0.5929$ \\
C & 1 & $8.788\mathord{\times}10^{-6}$ & $0.0061$ & $0.4186$ & $0.4191$ & $0.0014$ & $0.7901$ \\
C & 2 & $5.638\mathord{\times}10^{-6}$ & $0.0040$ & $1.6175$ & $1.6176$ & $0.0014$ & $0.4372$ \\
\bottomrule
\end{tabular}}
\end{table*}

\begin{table*}[t]
\centering
\caption{Direct PLGA row-quotient decomposition at update $65{,}536$.  The constant-input defect is measured by replacing each input by its common row while retaining the learned PLGA parameters.}
\label{row:tab:plga-row-defect}
\begin{tabular}{ccrrrr}
\toprule
Path & Layer & input quotient & segment response & constant-input defect & output quotient \\
\midrule
A & 0 & $1.314\mathord{\times}10^{-7}$ & $2.124\mathord{\times}10^{-8}$ & $81.7727$ & $81.7727$ \\
A & 1 & $1.446\mathord{\times}10^{-8}$ & $2.509\mathord{\times}10^{-10}$ & $48.6810$ & $48.6810$ \\
A & 2 & $7.8496$ & $1.0076$ & $37.2410$ & $37.2438$ \\
B & 0 & $4.288\mathord{\times}10^{-6}$ & $6.011\mathord{\times}10^{-8}$ & $74.5849$ & $74.5849$ \\
B & 1 & $0.0096$ & $8.391\mathord{\times}10^{-4}$ & $76.1873$ & $76.1874$ \\
B & 2 & $3.137\mathord{\times}10^{-6}$ & $4.988\mathord{\times}10^{-7}$ & $57.4629$ & $57.4629$ \\
C & 0 & $3.891\mathord{\times}10^{-8}$ & $4.789\mathord{\times}10^{-9}$ & $76.5473$ & $76.5473$ \\
C & 1 & $22.8607$ & $3.3310$ & $15.8631$ & $13.8202$ \\
C & 2 & $4.6741$ & $0.4571$ & $6.2650$ & $6.2769$ \\
\bottomrule
\end{tabular}
\end{table*}

Table~\ref{row:tab:layer-endpoint-gains} uses float64 replay.  The fixed small
endpoint threshold is \(E<10^{-3}\).  Six of nine trajectory-layer medians
meet it: A0, A1, B0, B1, B2, and C0.  A2, C1, and C2 remain far from the
row-constant face.  This layer resolution is the central empirical fact;
pooling all layers would erase it.

The late gains use the \(16{,}384\) to \(65{,}536\) interval.  Five of the six
small endpoints have much larger normalized-shape closure than effective-gate
closure on the log scale.  B1 is different: its normalized shape reopens by
\(2.228\) decades while its effective gate closes by \(1.884\) decades, leaving
a net energy reopening of \(0.336\) decades at an endpoint that remains below
the fixed threshold.  C2 also shows gate closure against weak shape reopening,
but its final energy is still \(21.85\).  Thus gate closure can oppose shape
growth without being sufficient for physical collapse.

Table~\ref{row:tab:arithmetic-resolution} explains why the replay precision is
part of the estimand.  At A1 the median float32 relative energy error is
\(2.66\) and the maximum is \(12.35\).  At C0 the median is \(2.54\).  Across
the nine endpoints, \(6{,}685\) coordinate energies are exactly zero in
float32 and positive in float64.  The real-arithmetic factorization still
holds to relative residual below \(10^{-12}\) in every float64 record.  The
false zeros therefore diagnose observer resolution, not failure of
Theorem~\ref{row:thm:layer-resolved-gate-shape}.

\subsection{Lifted final-gate operators}
\label{row:sec:gate-operators}

For each trajectory and layer, the selected state is
\begin{equation}
 x=(\gamma,m,v)\in\R^{64}\times\R^{64}\times\R^{64}.
 \label{row:eq:empirical-gate-state}
\end{equation}
The producer evaluates the real next-batch loss in float64.  It constructs
the exact \(64\times64\) gate Hessian by weighted, basiswise
forward-over-reverse differentiation.  It substitutes that Hessian, the live
moments, clipping cell, bias corrections, denominator, learning rate, and
weight decay into
Equation~\eqref{row:eq:implemented-adamw-matrix}.  An independent autodiff
of the implemented AdamW composition checks the resulting
\(192\times192\) matrix.  Signed probes in gate, first-moment,
second-moment, and mixed directions use dimensionless amplitudes
\(2^{-10}\) and \(2^{-11}\).

All analytic-to-autodiff relative matrix residuals are below the frozen
\(2\times10^{-9}\) tolerance.  In every unit, the maximum nonlinear residual
at the smaller amplitude is below \(10^{-2}\), and it decreases by at least
the registered factor when the amplitude is halved.  The derivatives are
therefore numerically resolved at the tested scale.

Table~\ref{row:tab:gate-operator-diagnostics} gives the scientific result.  None
of the nine induced norms is below one; they range from \(8.58\) to
\(791.81\).  None of the spectral radii is below one; they range from
\(1.009\) to \(1.471\).  Nonnormality ratios range from \(8.51\) to \(618.75\).
The scaled gate-state origin defects range from \(1.20\times10^3\) to
\(2.85\times10^6\), and zero of nine is resolved as zero under the fixed
tolerance.

This rules out instantaneous contraction of the final-gate lifted substate
at update \(4096\).  It does not rule out the complete-state theorem.  The
measured block excludes upstream parameter and moment coordinates, and a
single operator does not determine a long chronological product.  Likewise,
the measured origin defect is a defect of the gate-state restriction, not a
certificate for the full charted \(f_t\).  The data show why the full
operator and forcing convolution cannot be replaced by a gate-only decay
factor. Section~\ref{row:sec:complete-tangent-evidence} reports the subsequent
measurements of complete-state directional transport.

\subsection{Same-device component interventions}

Each one-step record starts from the same update-\(4096\) checkpoint and
opens the same next batch.  The selected layer has eight registered arms:
native, no-op restamp, one-ULP gate placebo, gate update removed, upstream
row-program update removed, joint row-program update removed, gate decay
only, and gate adaptive write only.  All nonselected parameters retain the
native successor.  The no-op arm is exactly equal to native.

Table~\ref{row:tab:one-step-mechanism-tests} reports root-mean-square physical
response from native over the \(96\) maps in each unit.  The one-ULP response
is at most \(0.00822\) of the smallest of the three main removal responses,
well below the predeclared \(0.1\) gate.  Thus all nine component comparisons
are identifiable above same-device numerical sensitivity at one step.

In every unit, removing the upstream row-program write has a larger response
than removing the final-gate write.  The joint response is close to the
upstream response but is not assumed additive.  These are controlled
one-step counterfactual effects at the registered state.  They do not show
that the same component controls a long trajectory, nor do they supply the
chronological product and forcing limits in
Theorem~\ref{row:thm:scheduled-tube-collapse}.

An earlier \(64\)-update component diagnostic is retained as a negative
sensitivity result.  Moving one gate coordinate by one float32 ULP before the
branch produced layer energy ratios \([1.061,0.912,0.166]\), on the scale of
the long component arms.  Those long branches are therefore not used for a
mechanism claim.  The one-step replacement above was chosen because its
placebo is decisively smaller than every registered treatment response.

\subsection{Source-written same-device perturbations}

For every trajectory-layer unit, a deterministic direction spans all
selected upstream row-program parameters except the final gate.  The
direction is globally normalized relative to the parameter norm.  Native,
\(+h\), and \(-h\) successors at \(h=2^{-12}\) construct the central-secant
prediction
\begin{equation}
 \widehat X_{h/2}=X_0+\frac14(X_{+h}-X_{-h}).
 \label{row:eq:source-written-half-prediction}
\end{equation}
The complete predicted physical maps and their digest are written before the
\(+h/2\) validation begins.  All four successors run on the same device and
batch, and the record stores both pre-step and post-step distances.

The fixed pass rule is relative centered-response residual at most \(0.25\).
Only A0 and B1 pass, with residuals \(0.181\) and \(0.230\).  The other seven
range from \(0.314\) to \(1.081\).  Median one-step quotient gains range from
\(1.32\) to \(45.0\).  This is an adverse outcome for a universal first-order
upstream predictor at the declared amplitude.  It is compatible with strong
nonlinearity, cross-coordinate optimizer coupling, and movement across local
cells.  The failed predictor is not refit and is not used as evidence for the
asymptotic theorem.

A separate numerical-radius diagnostic compared one float32 GPU successor
with a float32 CPU successor.  Its median pre-step cross-device difference,
\(1.20\times10^{-5}\), already matched the post-step quantity, while the
underlying float32 state differed at about \(1.6\times10^{-9}\).  That record
is a portability and determinism measurement.  It is not a state-perturbation
radius.  The same-device sign-paired experiment above is the scientific
replacement.

\subsection{Direct PLGA quotient defect}
\label{row:sec:plga-defect-evidence}

For each registered endpoint input, the producer evaluates both
\(\mathcal P(X)\) and \(\mathcal P(\bar X)\) in float64 with identical
learned parameters.  It records the input quotient, occupied-segment
response, constant-input defect, output quotient, and signed slack in
Equation~\eqref{row:eq:plga-quotient-defect-bound}.  Every triangle slack is
nonnegative to numerical tolerance.

Table~\ref{row:tab:plga-row-defect} shows that the defect dominates near the
upstream row face.  For example, A0 has median input quotient
\(1.31\times10^{-7}\) but median PLGA defect \(81.77\); A1 has input quotient
\(1.45\times10^{-8}\) and defect \(48.68\).  None of the nine
trajectory-layer units preserves the row-constant face at its registered
constant inputs.  In addition, none of the \(180\) tested sufficient-condition
residuals, three paths times three layers times four heads times five
conditions, falls below \(10^{-10}\).

The result rejects automatic PLGA inheritance.  It does not assert a defect in
PLGA as an attention mechanism.  A learned non-row-constant output can be
functional.  The result says that upstream PLDR row collapse and downstream
PLGA row collapse are distinct mathematical statements, connected by the
constant-input defect in Theorem~\ref{row:thm:plga-quotient-defect}.

\subsection{Matrix-free complete-state tangent response}
\label{row:sec:complete-tangent-evidence}

A second executed graph measures the complete-state tangent object without
materializing a square Jacobian.  The differentiated state is the entire
implemented AdamW triple \((\theta,m,v)\), with
\RowSourceObservableStateDimension{} scalar coordinates.  Jacobian-vector and
vector-Jacobian products propagate three frozen directions through the real
successor: a row-observable adjoint direction, an optimizer-displacement
direction, and a balanced direction containing parameter and moment
coordinates.  Signed nonlinear responses are compared with the matrix-free
linear response.  Construction uses path A only.  Horizons, amplitudes,
tolerances, and the fallback rule are locked before paths B and C are opened.

The graph has one resource-qualification node, nine construction nodes, one
lock node, \(18\) held-out prediction nodes, \(18\) held-out validation nodes,
nine gate-zero stratum nodes, and one analysis node.  All \(57\) nodes
completed.  The qualification derivative has relative response residual
\(0.0372\), AdamW replay discrepancy \(5.11\times10^{-8}\), and peak reserved
memory below the declared cap.

\begingroup
\scriptsize
\setlength{\tabcolsep}{2pt}
\begin{table*}[t]
\centering
\caption{Construction calibration of the matrix-free complete-state AdamW cocycle. Qualification is directionwise and finite-horizon; it is not an operator-norm estimate.}
\label{row:tab:observable-cocycle-construction}
\begin{tabular}{crrrr}
\toprule
Layer & selected horizon & joint qualification & qualified cells & total cells \\
\midrule
0 & 1 & no & 5 & 45 \\
1 & 1 & no & 5 & 45 \\
2 & 1 & no & 4 & 45 \\
\bottomrule
\end{tabular}
\end{table*}

\begin{table*}[t]
\centering
\caption{Held-out signed-response tests of the observable complete-state cocycle. Gains are visible directional gains from the source row quotient to the locked endpoint.}
\label{row:tab:observable-cocycle-heldout}
\begin{tabular}{lrrrrrr}
\toprule
Direction & pass/total & median residual & maximum residual & minimum gain & median gain & attenuating \\
\midrule
\texttt{row\_adjoint} & 5/18 & $0.9939$ & $1.0546$ & $0.6933$ & $1.3401$ & 4 \\
\texttt{optimizer\_displacement} & 11/18 & $0.0568$ & $0.9897$ & $0.8715$ & $34.0398$ & 1 \\
\texttt{balanced\_full\_state} & 14/18 & $0.0593$ & $0.9896$ & $0.7222$ & $4.0044$ & 2 \\
\bottomrule
\end{tabular}
\end{table*}

\begin{table}[t]
\centering
\caption{One-update gate-zero stratum test at update $4096$. The endpoint column is the physical row-quotient norm.}
\label{row:tab:observable-cocycle-stratum}
\begin{tabular}{ccrrc}
\toprule
Path & Layer & source norm & endpoint norm & preserved \\
\midrule
A & 0 & $0$ & $0.0134$ & no \\
A & 1 & $0$ & $0.0057$ & no \\
A & 2 & $0$ & $0.0720$ & no \\
B & 0 & $0$ & $0.1544$ & no \\
B & 1 & $0$ & $0.0261$ & no \\
B & 2 & $0$ & $0.1550$ & no \\
C & 0 & $0$ & $0.1040$ & no \\
C & 1 & $0$ & $0.1157$ & no \\
C & 2 & $0$ & $0.0763$ & no \\
\bottomrule
\end{tabular}
\end{table}

\begin{table*}[t]
\centering
\caption{Horizon-resolved construction result at the smaller amplitude. The predictor becomes catastrophically large beyond one update. Placebo failures are retained in the fixed 27-cell population at each horizon.}
\label{row:tab:observable-cocycle-horizons}
\begin{tabular}{rrrrrr}
\toprule
Horizon & qualified & median prediction & prediction range & median observation & placebo failures \\
\midrule
1 & 14/27 & $0.0387$ & $1.659\mathord{\times}10^{-4}$--$173.4660$ & $0.0389$ & 1/27 \\
2 & 0/27 & $0.2139$ & $0.0068$--$9531.8791$ & $0.1143$ & 0/27 \\
4 & 0/27 & $86.6646$ & $0.1411$--$6.518\mathord{\times}10^{8}$ & $0.6900$ & 10/27 \\
8 & 0/27 & $3.359\mathord{\times}10^{5}$ & $728.9982$--$3.226\mathord{\times}10^{16}$ & $1.6459$ & 26/27 \\
16 & 0/27 & $1.495\mathord{\times}10^{13}$ & $2.828\mathord{\times}10^{10}$--$9.671\mathord{\times}10^{28}$ & $4.3268$ & 27/27 \\
\bottomrule
\end{tabular}
\end{table*}

\begingroup
\makeatletter
\patchcmd{\LT@makecaption}{\reset@font}{\reset@font\normalsize}{}{\PackageError{monograph}{Longtable caption hook failed}{}}
\makeatother
\begin{longtable}{ccrlll}
\caption{Every held-out trajectory-layer-source unit. Each direction cell gives response pass (P) or fail (F), followed by the sampled visible gain. The stored physical observation arrays are float32.}\label{row:tab:observable-cocycle-heldout-units}\\
\toprule
Path & Layer & Source & row adjoint & optimizer displacement & balanced state \\
\midrule
\endfirsthead
\toprule
Path & Layer & Source & row adjoint & optimizer displacement & balanced state \\
\midrule
\endhead
\bottomrule
\endfoot
B & 0 & 4096 & \texttt{F}; $1.1543$ & \texttt{P}; $48.3702$ & \texttt{P}; $2.2686$ \\
B & 0 & 8192 & \texttt{P}; $1.0149$ & \texttt{P}; $34.6184$ & \texttt{P}; $1.9807$ \\
B & 0 & 16384 & \texttt{P}; $0.8888$ & \texttt{F}; $0.8715$ & \texttt{P}; $1.0496$ \\
B & 1 & 4096 & \texttt{F}; $1.5260$ & \texttt{P}; $34.3469$ & \texttt{P}; $6.3224$ \\
B & 1 & 8192 & \texttt{P}; $1.0028$ & \texttt{F}; $37.4525$ & \texttt{F}; $0.7222$ \\
B & 1 & 16384 & \texttt{P}; $0.9543$ & \texttt{F}; $4.5681$ & \texttt{F}; $0.9703$ \\
B & 2 & 4096 & \texttt{F}; $3.7374$ & \texttt{P}; $37.2520$ & \texttt{P}; $29.8498$ \\
B & 2 & 8192 & \texttt{F}; $70.7240$ & \texttt{P}; $200.7824$ & \texttt{P}; $77.9948$ \\
B & 2 & 16384 & \texttt{F}; $1.1258$ & \texttt{P}; $6.3816$ & \texttt{P}; $1.2098$ \\
C & 0 & 4096 & \texttt{F}; $2.8988$ & \texttt{F}; $26.9645$ & \texttt{F}; $5.7403$ \\
C & 0 & 8192 & \texttt{P}; $1.0055$ & \texttt{P}; $19.4852$ & \texttt{P}; $1.7863$ \\
C & 0 & 16384 & \texttt{F}; $0.6933$ & \texttt{P}; $18.2310$ & \texttt{P}; $1.0566$ \\
C & 1 & 4096 & \texttt{F}; $19.7764$ & \texttt{P}; $170.9101$ & \texttt{P}; $32.4481$ \\
C & 1 & 8192 & \texttt{F}; $9.0324$ & \texttt{P}; $27.1183$ & \texttt{P}; $16.9324$ \\
C & 1 & 16384 & \texttt{F}; $7.1014$ & \texttt{P}; $20.2260$ & \texttt{P}; $2.0869$ \\
C & 2 & 4096 & \texttt{F}; $5.7506$ & \texttt{F}; $50.6419$ & \texttt{F}; $19.4864$ \\
C & 2 & 8192 & \texttt{F}; $0.8965$ & \texttt{F}; $53.2401$ & \texttt{P}; $32.1660$ \\
C & 2 & 16384 & \texttt{F}; $17.4484$ & \texttt{F}; $33.7328$ & \texttt{P}; $10.8831$ \\
\end{longtable}
\endgroup

\endgroup

The horizon-resolved construction record is adverse to a stable multistep
linear predictor.  At the smaller amplitude, \(14/27\) one-update cells
qualify, but no cell qualifies at horizons two, four, eight, or sixteen under
the joint derivative and numerical-placebo rules.  There are
\RowSourceObservablePlaceboFailures{} placebo failures in the fixed \(135\)-cell
population.  Median predicted response grows from \(0.0387\) at one update to
\(1.495\times10^{13}\) at sixteen updates, while the corresponding observed
median is \(4.3268\).  No horizon qualifies jointly in any layer, so the frozen
fallback selects one update in all three layers.

Every held-out unit is reported in
Table~\ref{row:tab:observable-cocycle-heldout-units}.  On paths B and C,
\RowSourceObservableHeldoutPasses{} of \RowSourceObservableHeldoutTotal{} signed-response tests
pass the fixed relative-residual threshold, and only
\RowSourceObservableAttenuating{} of \(54\) visible directional gains are below one.
The maximum visible gain is \RowSourceObservableMaximumGain{}.  Directional medians are
\(1.3401\), \(34.0398\), and \(4.0044\) for row adjoint,
optimizer displacement, and balanced full state.  Thus local linear fidelity
and visible contraction are separate properties, and a generic one-update
observable contraction law fails at the sampled states.

The stored physical observation arrays in this graph are float32.  Its pass
rates and visible gains therefore describe the implemented float32 observer,
not an exact real-arithmetic quotient.  The endpoint replay in
Table~\ref{row:tab:arithmetic-resolution} shows why this matters near the
arithmetic floor.  No failed cell is removed or refit, no sampled directional
gain is promoted to an operator norm, and no finite horizon is used as an
asymptotic certificate.

\subsection{Retained physical-energy and reopening replay}
\label{row:sec:retained-energy-replay}

A CPU-only replay inspected the immutable dense-window and finite-increment
archives without changing them.  The dense archive contains six registered
windows on each of paths A, B, and C.  Their five interior windows have 64
one-update edges and the terminal window has 32, for
\RowSourceRetainedDenseEdges{} captured update edges.  Across 288 maps this gives
\RowSourceRetainedDenseMapEdges{} scalar energy increments.  Exactly
\RowSourceRetainedDenseReopeningEdges{}, or \RowSourceRetainedDenseReopeningPercent{}, have
positive stored energy change.  Every one of the 288 registered map identities
reopens at least once somewhere in the captured windows.  Layerwise positive
fractions are \(0.4234\), \(0.4320\), and \(0.4443\).  Thus intermittent
reopening is not an exceptional feature of these finite windows.

The dense records retain \(E_t\), so their \(\Delta E_t\) and
\(\mathfrak r_t=(\Delta E_t)_+\) are directly evaluable.  They do not retain
the physical arrays \(Z_t\), so \(D_t\), signed work, and finite-step charge
are not evaluable from those records.  They retain final gates and squared
coordinate shape energies, not the signed centered normalized shapes.
Consequently the gate, shape, and interaction secants and their cross-Gram
matrix are also not evaluable.  The replay does not invert energies or other
aggregates to manufacture any missing array.

The finite-increment archive supplies a narrower full-map population.  Its
nine construction records each retain physical endpoints at horizons one and
two, giving \RowSourceRetainedEndpointIncrements{} successive map pairs.  The replay
reapplies the row-centering projector, forms \(D_t\), and evaluates both sides
of Equation~\eqref{row:eq:exact-direct-work-charge}.  A separately implemented
float64 calculation reproduces every increment under the fixed absolute
\(2\times10^{-10}\) and relative \(2\times10^{-5}\) tolerances.  The maximum
absolute work-charge residual is \RowSourceRetainedEndpointIdentityResidual{}.
There are \RowSourceRetainedEndpointReopenings{} positive increments, or
\RowSourceRetainedEndpointReopeningPercent{}.  The remaining
\RowSourceRetainedNonevaluableRecords{} files contain only one physical endpoint,
response secants without base endpoints, or row quotients.  They are reported
as ``not evaluable from retained records.''

\begin{table}[t]
\centering
\caption{CPU replay of retained one-update energy increments. Dense rows contain stored scalar energies; the full-map row contains successive centered physical maps and therefore admits the exact work-charge ledger.}
\label{row:tab:retained-energy-audit}
\begin{tabular}{lrrrr}
\toprule
Record & update edges & map increments & $\Delta E>0$ & fraction \\
\midrule
Dense path A & 352 & 101{,}376 & 46{,}218 & 45.59\% \\
Dense path B & 352 & 101{,}376 & 37{,}028 & 36.53\% \\
Dense path C & 352 & 101{,}376 & 48{,}502 & 47.84\% \\
Dense total & 1{,}056 & 304{,}128 & 131{,}748 & 43.32\% \\
\midrule
Full-map pairs & 9 & 864 & 474 & 54.86\% \\
\bottomrule
\end{tabular}

\end{table}

The same replay independently reproduces the finite cancellation
concentration: \RowSourceRetainedCancellationTotal{} cancellation-dominant held-out
units, of which \RowSourceRetainedCancellationLargestPath{} occur on one path and
\RowSourceRetainedCancellationLayerTwo{} occur in layer 2.  This is a descriptive
finite concentration, not a trajectory or layer population law.  The replay
establishes that finite reopenings occur and that the available full-map
increments obey the exact ledger.  It cannot establish the summability of
\(\mathfrak r_t\), a future work margin, or an asymptotic collapse theorem.

\subsection{Exact finite-increment observable balance}
\label{row:sec:finite-balance-evidence}

A third executed graph measures all terms in
Theorem~\ref{row:thm:finite-increment-observable-balance}.  For each path, layer,
source update, and direction, it propagates the base-route homogeneous tangent
\(L\), computes the signed finite state difference, forms the accumulated
state correction \(N\), and closes the observed response with \(Q\).  The
same batch schedule and device are used within each signed comparison.  Route
changes are retained and absorbed by the exact residual \(\rho_t\).

The construction path is A, source updates are \(4096\), \(8192\), and
\(16384\), and the horizon grid is \(1,2,4,8\).  Two frozen amplitudes are
used for each of the row-adjoint, optimizer-displacement, and balanced
full-state directions.  The longest jointly evaluable construction horizon is
selected independently by layer, with a one-update fallback.  The resulting
lock chooses horizons two, one, and one for layers zero, one, and two.  The
last choice is the declared fallback because layer two has no jointly
evaluable horizon.

The qualification node reproduces the implemented AdamW successor with
relative residual \(5.11\times10^{-8}\).  Its exact-balance reconstruction
relative residual is \(1.97\times10^{-11}\).  These residuals check the
implementation of the decomposition; they are not empirical confirmation of
the algebraic identity, which follows from
Theorem~\ref{row:thm:finite-increment-observable-balance}.

\begingroup
\scriptsize
\setlength{\tabcolsep}{2pt}
\begin{table*}[t]
\centering
\caption{Construction decomposition at the smaller signed amplitude. The three norms are the homogeneous tangent, accumulated state correction, and nonlinear observation correction.}
\label{row:tab:balance-construction}
\begin{tabular}{rrrrrrrr}
\toprule
Horizon & evaluable & $\operatorname{med}\|L\|$ & $\operatorname{med}\|N\|$ & $\operatorname{med}\|Q\|$ & $\operatorname{med}\|Y\|$ & $\operatorname{med}\chi$ & route changes \\
\midrule
1 & 27/27 & $0.0387$ & $0.0037$ & $0.0010$ & $0.0389$ & $1.1064$ & 0 \\
2 & 27/27 & $0.2139$ & $0.2318$ & $0.0036$ & $0.1143$ & $4.5211$ & 0 \\
4 & 20/27 & $86.6646$ & $85.9362$ & $0.0203$ & $0.6900$ & $130.7800$ & 9 \\
8 & 3/27 & $3.359\mathord{\times}10^{5}$ & $3.359\mathord{\times}10^{5}$ & $0.2429$ & $1.6459$ & $2.797\mathord{\times}10^{5}$ & 5 \\
\bottomrule
\end{tabular}
\end{table*}

\begin{table*}[t]
\centering
\caption{Held-out exact-balance outcomes by direction. Cancellation dominance is distinct from numerical evaluability.}
\label{row:tab:balance-heldout-direction}
\begin{tabular}{lrrrr}
\toprule
Direction & evaluable & cancellation-dominant & median $\chi$ & median $\|Y\|/\|L\|$ \\
\midrule
row adjoint & 18/18 & 6/18 & $13.8312$ & $0.1816$ \\
optimizer displacement & 18/18 & 0/18 & $1.3376$ & $0.9988$ \\
balanced full state & 18/18 & 0/18 & $1.1631$ & $0.9917$ \\
\bottomrule
\end{tabular}
\end{table*}

\begin{table*}[t]
\centering
\caption{Every held-out trajectory-layer-source unit. C denotes cancellation-dominant, N denotes evaluable but not cancellation-dominant, and X denotes numerically inevaluable. Values in parentheses are cancellation indices.}
\label{row:tab:balance-heldout-units}
\begin{tabular}{ccrrrr}
\toprule
Path & Layer & Source & row adjoint & optimizer displacement & balanced state \\
\midrule
B & 0 & 4096 & N (14) & N (2.9) & N (1.2) \\
B & 0 & 8192 & N (1.5) & N (4.9) & N (1.6) \\
B & 0 & 16384 & N (1.4) & N (1.5) & N (1.1) \\
B & 1 & 4096 & N (5.2) & N (1) & N (1) \\
B & 1 & 8192 & N (1) & N (1.4) & N (1.3) \\
B & 1 & 16384 & N (1) & N (1.1) & N (1.1) \\
B & 2 & 4096 & N (7.8) & N (1) & N (1.2) \\
B & 2 & 8192 & C (\ensuremath{1.5\times10^{2}}) & N (1) & N (1) \\
B & 2 & 16384 & N (2.6) & N (1) & N (1.1) \\
C & 0 & 4096 & C (68) & N (5.8) & N (2.3) \\
C & 0 & 8192 & N (1.3) & N (14) & N (6.6) \\
C & 0 & 16384 & N (14) & N (4) & N (1.1) \\
C & 1 & 4096 & C (30) & N (1.1) & N (1.2) \\
C & 1 & 8192 & N (17) & N (1) & N (1.1) \\
C & 1 & 16384 & N (14) & N (1) & N (1) \\
C & 2 & 4096 & C (39) & N (1.3) & N (1.6) \\
C & 2 & 8192 & C (39) & N (1.4) & N (1.3) \\
C & 2 & 16384 & C (81) & N (2) & N (1.1) \\
\bottomrule
\end{tabular}
\end{table*}

\begin{table*}[t]
\centering
\caption{Coordinatewise gate-zero formula and controlled-gradient test. The control changes only the selected clipped gate gradient before the moment write.}
\label{row:tab:gate-gradient-control}
\begin{tabular}{ccrrrr}
\toprule
Path & Layer & nonzero gradient coordinates & natural force & controlled force & bitwise pass \\
\midrule
A & 0 & 64 & $0.0134$ & 0 & yes \\
A & 1 & 64 & $0.0057$ & 0 & yes \\
A & 2 & 64 & $0.0720$ & 0 & yes \\
B & 0 & 64 & $0.1544$ & 0 & yes \\
B & 1 & 64 & $0.0261$ & 0 & yes \\
B & 2 & 64 & $0.1550$ & 0 & yes \\
C & 0 & 64 & $0.1040$ & 0 & yes \\
C & 1 & 64 & $0.1157$ & 0 & yes \\
C & 2 & 64 & $0.0763$ & 0 & yes \\
\bottomrule
\end{tabular}
\end{table*}

\endgroup

Table~\ref{row:tab:balance-construction} reveals the mechanism missed by the
linear predictor.  At horizon one, the median norms are
\((\norm L,\norm N,\norm Q,\norm Y)=(0.0387,0.0037,0.0010,0.0389)\), with
median \(\chi=1.1064\).  At horizon four, among the \(20/27\) evaluable cells,
\(\norm L\) and \(\norm N\) have medians \(86.6646\) and \(85.9362\), while
\(\norm Y\) has median \(0.6900\) and \(\chi\) has median \(130.7800\).
The unstable multistep tangent is therefore balanced primarily by the state
correction in part of the construction population.  This is an exact
finite-increment accounting statement, not evidence that the base tangent is
contracting. Section~\ref{model:sec:row-path-results} gives a separate
single-pass test of signed cross-time interference in finite matrix increments;
its observable and continuation design are specified there.

The frozen held-out definition calls a unit cancellation-dominant only when
the response exceeds the numerical floor,
\(\chi\ge10\), \(\norm Y/\norm L\le0.1\), and
\(\langle L,N+Q\rangle<0\).  All \RowSourceBalanceHeldoutEvaluable{} of
\RowSourceBalanceHeldoutTotal{} held-out units are evaluable, but only
\RowSourceBalanceCancellationDominant{} are cancellation-dominant.  Path B contributes
\(1/27\), and path C contributes \(5/27\).  Both are below the frozen
per-trajectory support fraction of two thirds, so the universal held-out
cancellation hypothesis is \RowSourceBalanceCampaignOutcome{}.

The failure is structured by direction.  All six cancellation-dominant units
are row-adjoint directions.  Five occur on path C and four occur in layer 2.
This finite concentration is descriptive and is not evidence for a
trajectory-level or layer-level population law.  Their direction-wide median is
\(\chi=13.8312\), compared with \(1.3376\) for optimizer displacement and
\(1.1631\) for balanced full state.  The corresponding median
\(\norm Y/\norm L\) values are \(0.1816\), \(0.9988\), and \(0.9917\).
Across all held-out units, the state correction is the larger correction in
\RowSourceBalanceStateDominant{} cases, the observation correction is larger in
\RowSourceBalanceObservationDominant{}, and \RowSourceBalanceRouteChanged{} signed comparisons
change route.  The exact balance survives every such route change, as the
stratified identity predicts, but strong destructive cancellation does not
transfer uniformly across directions or trajectories.

The supported theory is consequently direction selective.  Row-map collapse
can coexist with an expansive base-route tangent when nonlinear state
transport cancels the visible row-adjoint response.  Optimizer-displacement
and balanced directions in this experiment instead have observed responses
close to their homogeneous responses.  Neither pattern supplies the
chronological product and forcing limits required for asymptotic
complete-state collapse.

\subsection{Precision-resolved paired source experiments}
\label{row:sec:direct-work-evidence}

A paired experiment measures the one-step physical ledger in
Theorem~\ref{row:thm:exact-direct-work-charge} and removes one declared loss
source.  The natural arm executes the clipped AdamW successor.  In the
control, the incoming adjoint at the selected row-map output is replaced by
its row-constant orthogonal projection before upstream backpropagation.  The
forward values, source adjoint before projection, parameters, moments, batch,
clock, clipping rule, random state, device, upstream gradients, and unrelated
parameter gradients are held fixed or checked bitwise as appropriate.  Both
AdamW endpoints and both observations are reconstructed from the stored
source state.

The retained campaign uses source updates \(4096\), \(8192\), and \(16384\),
all three layers, the 96-map registry, and paths A, B, and C.  Its original
locked eligibility rule mixed native causal checks with a relative
factorization diagnostic.  A sealed precision reanalysis separates those
roles.  Of \(\RowSourcePrecisionReplayArmMapCount{}\) arm-map records, all
\(\RowSourcePrecisionReplayGateOnlyExcluded{}\) exclusions are gate-shape diagnostic
failures; none fails the native work identity, effect floor, execution, or
replay checks.  Their maximum-coordinate factorization residuals range from
\(\RowSourcePrecisionReplayFailAbsMin{}\) to \(\RowSourcePrecisionReplayFailAbsMax{}\).
Native-primary eligibility therefore increases from
\(\RowSourcePrecisionReplayLockedEvaluable{}\) to
\(\RowSourcePrecisionReplayPrimaryEvaluable{}\) held-out maps.

This separation does not change the locked conclusion, which remains
insufficient.  It is a post-outcome design sensitivity, not a confirmatory
reclassification.  The stored implementation defect matches the
native-minus-factorized increment to
\(\RowSourcePrecisionReplayDefectMatch{}\), and the four-source native secant closes to
\(\RowSourcePrecisionReplayNativeClosure{}\).  The largest native work-charge residual
is \(\RowSourcePrecisionReplayWorkResidual{}\), far below the minimum registered work
tolerance \(\RowSourcePrecisionReplayOldWorkToleranceMinimum{}\).  The work mask was
therefore nonbinding.  Table~\ref{row:tab:direct-work-mask-sensitivity} reports
the separated sensitivity by layer.

\begin{table}[t]
\centering
\small
\caption{Design sensitivity after separating native causal eligibility from factorization diagnostics. This post-outcome calculation is not a confirmatory reclassification.}
\label{row:tab:direct-work-mask-sensitivity}
\begin{tabular}{cp{7.2cm}ccc}
\toprule
Layer & Construction signs by source update & Locked sign & Path B support & Path C support \\
\midrule
0 & 96/96 positive; 96/96 positive; 96/96 positive & positive & 2/3 & 2/3 \\
1 & 96/96 positive; 96/96 negative; 96/96 positive & positive & 2/3 & 3/3 \\
2 & 84/96 positive; 81/96 negative; 56/96 negative & unresolved & 0/3 & 0/3 \\
\bottomrule
\end{tabular}
\end{table}

A new temporal-context replication tests the same intervention at updates
\(18{,}000\), \(32{,}768\), and \(65{,}536\).  The first two use disjoint
registered next batches.  At the terminal checkpoint, no historical
successor exists, so the record is explicitly a registered reserve-batch
counterfactual continuation.  It is never interpreted as the realized next
training step.  Path A constructs the layerwise sign lock; paths B and C
remain unopened until its digest is written.  The frozen graph contains two
device-qualification nodes, nine construction nodes, one lock, 18 held-out
nodes, and one analysis node.

Both devices reproduce the qualification record bitwise, with zero
cross-device energy discrepancy and a fixed effect floor of
\(\RowSourceReplicationEffectFloor{}\).  Every construction record passes the native
work-charge, four-source secant, factorization-defect, and extended energy
checks.  Layers 0 and 1 are unresolved because their signs reverse across
construction times.  Layer 2 locks positive, with qualifying positive
majorities at two construction times.

\begingroup
\scriptsize
\setlength{\tabcolsep}{2pt}
\begin{table}[t]
\centering
\small
\caption{Executed temporal-context replication by held-out trajectory and layer. The contrast is control minus natural native endpoint energy.}
\label{row:tab:direct-work-replication}
\begin{tabular}{ccccrrrc}
\toprule
Path & Layer & Locked sign & Eval. & $I>0$ & $I<0$ & Median $I$ & Status \\
\midrule
B & 0 & unresolved & 288/288 & 32 & 133 & \(-7.718\times10^{-14}\) & unresolved \\
B & 1 & unresolved & 288/288 & 1 & 287 & \(-2.155\times10^{-7}\) & unresolved \\
B & 2 & positive & 288/288 & 121 & 71 & \(7.838\times10^{-15}\) & refuted \\
C & 0 & unresolved & 192/288 & 192 & 0 & \(6.334\times10^{-6}\) & unresolved \\
C & 1 & unresolved & 288/288 & 150 & 138 & \(0.2282\) & unresolved \\
C & 2 & positive & 288/288 & 166 & 122 & \(0.1293\) & refuted \\
\bottomrule
\end{tabular}
\end{table}

\endgroup

Table~\ref{row:tab:direct-work-replication} reports all held-out
trajectory-layer cells.  Of \(\RowSourceReplicationHeldoutPlanned{}\) planned maps,
\(\RowSourceReplicationHeldoutEvaluable{}\) are evaluable:
\(\RowSourceReplicationHeldoutPositive{}\) are positive,
\(\RowSourceReplicationHeldoutNegative{}\) are negative, and \(219\) are neutral at
the fixed effect floor.  In locked layer 2, path B supports the positive sign
only at update \(18{,}000\), while path C supports it only at the terminal
counterfactual continuation.  Both paths have all \(288\) maps evaluable and
fail the required two-of-three temporal support rule.  The locked
layer-2 claim is therefore \RowSourceReplicationCampaignOutcome{}.  Natural and
controlled successors reopen in \(\RowSourceReplicationNaturalReopening{}\) and
\(\RowSourceReplicationControlReopening{}\) evaluable cases, respectively.  The
experiment rejects a persistent, state-independent sign for this source.

The exact identity in
Theorem~\ref{row:thm:exact-paired-energy-contrast} explains the temporal reversal.
Across \(\RowSourceMechanismEvaluable{}\) evaluable paired maps, its maximum absolute
residual is \(\RowSourceMechanismContrastResidual{}\), and the paired four-source
closure residual is at most \(\RowSourceMechanismFourSourceResidual{}\).  The state
projection \(2\langle Z,H\rangle\) is the largest-magnitude term in
\(\RowSourceMechanismStateDominant{}\) maps and agrees with the contrast sign in
\(\RowSourceMechanismStateSignAgreement{}\) of sign-eligible maps.  The 95th
percentile cancellation index is \(\RowSourceMechanismCancellationPercentile{}\), so the
natural-increment interaction and intervention charge still matter in the
tail.  The corrected summation-aware technical masks were live throughout
the replication but recorded zero failures, with closure maxima between
\(10^{-13}\) and \(10^{-16}\).  This live but unexercised mask behavior is
distinct from the terminal arithmetic-defect classification, where the
defect source is the largest paired-source norm in
\(\RowSourceMechanismTerminalDefectDominant{}\) maps.  Neither observation is assigned
a population interpretation.

\begin{table}[t]
\centering
\small
\caption{Exact paired-contrast decomposition in heldout layer 2. Entries are mapwise medians, so the three displayed median terms need not sum exactly to the median contrast.}
\label{row:tab:paired-contrast-mechanism}
\begin{tabular}{ccrcrrrr}
\toprule
Path & Step & Eval. & Sign & $I$ & $2\langle Z,H\rangle$ & $2\langle D^{N},H\rangle$ & $\lVert H\rVert^2$ \\
\midrule
B & 18000 & 96 & positive & \(1.388\) & \(1.257\) & \(-3.334\times10^{-2}\) & \(0.1337\) \\
B & 32768 & 96 & negative & \(-1.824\times10^{-3}\) & \(-1.813\times10^{-3}\) & \(1.065\times10^{-4}\) & \(1.340\times10^{-5}\) \\
B & 65536 & 96 & neutral & \(1.309\times10^{-15}\) & \(1.494\times10^{-15}\) & \(-6.751\times10^{-16}\) & \(7.307\times10^{-16}\) \\
C & 18000 & 96 & positive & \(2.649\times10^{-2}\) & \(2.478\times10^{-2}\) & \(-2.221\times10^{-3}\) & \(4.084\times10^{-3}\) \\
C & 32768 & 96 & negative & \(-0.7447\) & \(-3.967\) & \(-1.707\) & \(4.392\) \\
C & 65536 & 96 & positive & \(4.029\) & \(0.7570\) & \(-0.5757\) & \(5.204\) \\
\bottomrule
\end{tabular}
\end{table}

The sealed mechanism regeneration classifies both all-neutral
B/\(65{,}536\) units, in layers 0 and 2, as neutral and assigns neither a
majority sign.  Its scientific summary is byte-identical to the preceding
sealed summary.  The final control-sequence names and math-safe table cells
are emitted by the sealed analyzer itself, so the mechanism manuscript files
are imported byte for byte.  The optional consecutive-update stage was not
launched because it was not required by the locked decision cascade, and no
claim in this monograph depends on it.

The mechanism is consequently exact but state conditional.  A positive
contrast means that this control raises endpoint energy at that particular
state and natural increment.  It does not assign the centered adjoint an
intrinsic dissipative sign, and it does not supply an all-future work margin
or summable reopening budget.

\subsection{Exact radial and tangential edge analysis}
\label{row:sec:radial-tangential-evidence}

The radial reduction reuses the 27 content-bound paired records only after
the native-primary technical checks of the precision replay.  For each
natural and controlled arm map with source energy above the registered
observer floor, it computes
\(\alpha_t\), \(\mathcal T_t\), \(\tau_t^2\), and \(g_t^2\) directly from
the stored physical arrays.  Eligibility also requires the paired-source and
bitwise invariants, the native four-source closure, and the radial identity.
No missing physical array is reconstructed.

Of 5,184 planned arm maps, \(\RowSourceRadialArmEligible{}\) are eligible.
Table~\ref{row:tab:radial-tangential-retained} gives their exhaustive classes.
There are \(\RowSourceRadialArmContract{}\) strict contractions, no exact
preservations, and \(\RowSourceRadialArmReopen{}\) strict reopenings.  Among the
reopenings, \(\RowSourceRadialFailureCount{}\) have
\(\alpha_t\notin(0,2)\); the remaining \(\RowSourceTangentExcessCount{}\) have an
inward nonovershooting radial coefficient but violate
\(\tau_t^2<\alpha_t(2-\alpha_t)\).  The largest normalized-gain identity
residual is \(\RowSourceRadialGainResidual{}\).

\begin{table}[t]
\centering
\caption{Exact radial and tangential classification of retained physical row-map updates. For gain, the classes are contraction and reopening. For reopening, they are radial-interval failure and tangent excess. For the paired row, they are agreement and disagreement.}
\label{row:tab:radial-tangential-retained}
\begin{tabular}{lrrr}
\toprule
Check & eligible & first class & second class \\
\midrule
Arm-map gain & 4{,}608 & 2{,}452 & 2{,}156 \\
Radial interval failure & 2{,}156 & 2{,}015 & 141 \\
Paired sign comparison & 2{,}304 & 2{,}236 & 68 \\
\bottomrule
\end{tabular}

\end{table}

The same records resolve shape, gate, bilinear interaction, and arithmetic
defect into source radial coefficients and a full \(4\times4\) tangential
Gram matrix.  Radial coefficients and residual vectors recompose to the
native total to within \(2.67\times10^{-15}\) and
\(3.08\times10^{-15}\), respectively.  The largest difference between the
source-reduced energy and its stored native counterpart is
\(1.94\times10^{-12}\).  These are closure checks, not fitted residuals.

For each of \(\RowSourceRadialPairedEligible{}\) eligible natural-control pairs,
Equation~\eqref{row:eq:radial-tangential-paired-contrast} is evaluated as radial
linear work, radial charge, tangential interaction, and tangential charge.
Its maximum residual is \(\RowSourceRadialPairedResidual{}.\)  The combined radial
part agrees with the full raw contrast sign in \(\RowSourceRadialSignAgreement{}\) of
\(\RowSourceRadialSignComparisons{}\) maps, or \(\RowSourceRadialSignAgreementPercent{}\).
The largest-magnitude term is radial linear work in
\(\RowSourceRadialLinearDominant{}\) maps, tangential charge in
\(\RowSourceTangentChargeDominant{}\), tangential interaction in
\(\RowSourceTangentInteractionDominant{}\), and radial charge in
\(\RowSourceRadialChargeDominant{}\).  The 68 sign disagreements establish that
radial dominance is not an identity.

An independent reducer imports neither the primary analyzer nor its radial
identity module.  It recomputes 937 aggregate leaves with direct coordinate
formulas; all match under a frozen absolute tolerance \(10^{-11}\) plus
relative tolerance \(5\times10^{-12}\).  The largest absolute difference is
\(3.53\times10^{-12}\).  The retained dense scalar-energy logs supply 18
positive excursions, 5,184 map windows, and 304,128 positive source edges.
All finite products reconstruct, with maximum relative residual
\(\RowSourceRadialProductResidual{}\); the affine replay residual is at most
\(1.79\times10^{-15}\).  Those dense logs contain no physical component
arrays, so their \(\alpha_t\) and \(\tau_t\) values are unavailable and are
not inferred from scalar energy alone.  No edge in those 18 windows hits the
zero face.

\subsection{Fresh-context radial holdout}
\label{row:sec:radial-context-holdout}

A separate sealed two-GPU graph tests whether the radial contribution in
Equation~\eqref{row:eq:radial-tangential-paired-contrast} predicts the full
paired-contrast sign in fresh temporal contexts.  Trajectory A is used only
for construction.  Before either held-out path is analyzed, the graph freezes
the registry, terminal reserve batch, effect floor \(10^{-12}\), and decision
rule.  Trajectories B and C are held out at updates \(18{,}000\),
\(32{,}768\), and \(65{,}536\), across all three layers and 96 maps per cell.
The terminal update is a registered reserve-batch continuation and is not
represented as the historical next training step.

The frozen rule requires every one of \RowSourceRadialHoldoutPlannedCells{} held-out
cells to contain at least
64 radial-eligible maps, at least 90\% pooled raw sign agreement, and at least
70\% raw agreement in every adequately powered cell.  A cell below 64 maps
makes the outcome insufficient; an adequately powered cell below 70\% or a
pooled result below 90\% makes it refuted.  There is no pooled fallback for an
underpowered cell.

\begingroup
\scriptsize
\setlength{\tabcolsep}{3pt}
\begin{table}[t]
\centering
\caption{Fresh-context held-out radial-predictor result. Each cell contains 96 planned physical maps; the frozen power requirement is 64 radial-eligible maps.}
\label{row:tab:radial-context-holdout}
\begin{tabular}{lrrrrl}
\toprule
Path & update & layer & eligible & agreement & powered \\
\midrule
B & 18000 & 0 & 96 & 100.0\% & yes \\
B & 18000 & 1 & 96 & 100.0\% & yes \\
B & 18000 & 2 & 96 & 90.6\% & yes \\
B & 32768 & 0 & 96 & 100.0\% & yes \\
B & 32768 & 1 & 96 & 100.0\% & yes \\
B & 32768 & 2 & 96 & 100.0\% & yes \\
B & 65536 & 0 & 96 & 99.0\% & yes \\
B & 65536 & 1 & 96 & 97.9\% & yes \\
B & 65536 & 2 & 96 & 99.0\% & yes \\
C & 18000 & 0 & 96 & 100.0\% & yes \\
C & 18000 & 1 & 96 & 99.0\% & yes \\
C & 18000 & 2 & 96 & 100.0\% & yes \\
C & 32768 & 0 & 96 & 100.0\% & yes \\
C & 32768 & 1 & 96 & 100.0\% & yes \\
C & 32768 & 2 & 96 & 85.4\% & yes \\
C & 65536 & 0 & 0 & n/a & no \\
C & 65536 & 1 & 96 & 100.0\% & yes \\
C & 65536 & 2 & 96 & 74.0\% & yes \\
\bottomrule
\end{tabular}

\end{table}

\endgroup

Every scientific node passes the native technical and exact-identity checks.
\RowSourceRadialHoldoutPoweredCells{} cells are adequately powered.  They contribute
\(\RowSourceRadialHoldoutEligible{}\) radial-eligible maps, of which
\(\RowSourceRadialHoldoutAgreements{}\) have matching nonzero raw signs, giving
\(\RowSourceRadialHoldoutAgreementPercent{}\) pooled agreement.  No powered cell is
below 70\%; the lowest is trajectory C, update \(65{,}536\), layer 2, at
74.0\%.  Radial linear work is the largest-magnitude term in 1,473 held-out
maps, tangential charge in 138, tangential interaction in 21, and radial
charge in none.  The maximum gain and paired-identity residuals are
\(\RowSourceRadialHoldoutGainResidual{}\) and \(\RowSourceRadialHoldoutPairedResidual{}\).

Trajectory C, update \(65{,}536\), layer 0, has strictly positive source
energy in every planned map, but all 96 values lie below the frozen
\(10^{-12}\) observer floor.  They range from
\(\RowSourceObserverTargetMinimumEnergy\) to \(\RowSourceObserverTargetMaximumEnergy\), the
largest absolute centered coordinate is \(\RowSourceObserverTargetMaximumCoordinate\), and the
stored centered rows round-trip through float32.  The floor is
\(\RowSourceObserverTargetFloorFactor\) times the largest energy.  The exact-zero
count is \(\RowSourceObserverExactZeroMaps{}\).  Thus the normalized radial quantities exist mathematically,
but the registered policy makes every map ineligible.  The all-cells power
condition fails, so the sealed outcome remains
\RowSourceRadialHoldoutOutcome{}.  The cell does not exercise the exact-face restart
branch.

\subsection{Observer-stratum and finite-block analysis}
\label{row:sec:observer-block-analysis}

A sealed CPU analysis recomputes exact stored-energy classes for the 27
retained records and the 27 fresh-context records.  It also replays every
sealed dense scalar window at endpoint horizons \(1,2,4,8,16,32\).  A
separately implemented scalar checker recomputes all per-cell stratum rows,
endpoint-gain and maximum-relative-excursion quantiles, resolved-population
quantiles, censored counts, and per-window rows.  It agrees on 1,194 discrete
fields and 352 numerical fields, with zero maximum absolute difference.  Its
quantiles use an independent Hyndman--Fan type-7 interpolation covered by
boundary fixtures.

\begin{table}[t]
\centering
\small
\caption{Observer-stratum census of stored source energies. Mathematical radial coordinates require strict positivity; policy eligibility additionally requires energy above the frozen effect floor.}
\label{row:tab:observer-stratum-census}
\begin{tabular}{lrrrrrr}
\toprule
Collection & Maps & Exact zero & Censored & Resolved & Math. radial & Policy radial \\
\midrule
Retained & 2,592 & 0 & 288 & 2,304 & 2,592 & 2,304 \\
Fresh holdout & 2,592 & 0 & 288 & 2,304 & 2,592 & 2,304 \\
\bottomrule
\end{tabular}
\end{table}

\begin{table}[t]
\centering
\small
\caption{Sliding endpoint gains in the 18 sealed dense windows. The windows overlap, so counts are descriptive. The last column is the number of windows containing at least one expanding comparison.}
\label{row:tab:block-gain-census}
\begin{tabular}{rrrrrrr}
\toprule
$L$ & Comparisons & Contract & Fraction & Median & P95 & Expanding windows \\
\midrule
1 & 304,128 & 172,380 & 56.68\% & 0.99969 & 1.02309 & 18/18 \\
2 & 298,944 & 170,330 & 56.98\% & 0.99937 & 1.04044 & 18/18 \\
4 & 288,576 & 164,389 & 56.97\% & 0.99879 & 1.06704 & 18/18 \\
8 & 267,840 & 152,559 & 56.96\% & 0.99768 & 1.10758 & 18/18 \\
16 & 226,368 & 130,785 & 57.78\% & 0.99540 & 1.15127 & 18/18 \\
32 & 143,424 & 85,061 & 59.31\% & 0.99224 & 1.18524 & 18/18 \\
\bottomrule
\end{tabular}
\end{table}

Table~\ref{row:tab:observer-stratum-census} shows zero exact source-energy zeros
among \(\RowSourceObserverAllSourceMaps{}\) maps.  The retained collection has
\(\RowSourceObserverRetainedCensoredMaps{}\) floor-censored-positive maps and
\(\RowSourceObserverRetainedResolvedMaps{}\) resolved-positive maps.  The fresh
holdout collection has \(\RowSourceObserverHoldoutCensoredMaps{}\) and
\(\RowSourceObserverHoldoutResolvedMaps{}\), respectively.  Mathematical radial
coordinates therefore exist for every stored source map, whereas the
frozen policy reports them only for the resolved stratum.  Together with the
positive dense-window sources, this means that no measured edge in these
radial analyses exercises the exact-face branch.  The branch remains part of
the complete analytical theory because an optimizer orbit can hit or reopen
the face, not because these records observed such a hit.

The retained aggregates contain three \(0/96\) policy-eligible cells,
while the fresh-context design requires 64 eligible maps in every cell
and prohibits pooling. The construction sample contains two floor-limited
cells. The resulting design limitation is measurability under the fixed
observer floor: mathematical positivity alone does not meet the policy's
eligibility condition. All reported cells, including the underpowered cell,
remain in the outcome assessment.

Table~\ref{row:tab:block-gain-census} rejects a fixed-window
uniform-contraction account on the captured records.  Every one of the
\(\RowSourceBlockWindowCount\) windows contains an expansion at every registered
horizon.  At horizon 32 there are \(\RowSourceBlockThirtyTwoContracting\)
contracting and \(\RowSourceBlockThirtyTwoExpanding\) expanding comparisons among
\(\RowSourceBlockThirtyTwoComparisons\) endpoint pairs.  Restricting to the
\(\RowSourceBlockThirtyTwoResolvedComparisons\) pairs whose endpoints both exceed the
observer floor still leaves \(\RowSourceBlockThirtyTwoResolvedExpanding\) expansions.
The overlapping counts are descriptive, but a single counterexample is
already enough to falsify uniform contraction on this sample.  This finite
analysis does not test arbitrary anchor sequences, vanishing transported
restart mass, or vanishing within-block excursion, so it is not evidence for
the sufficient premises of
Theorem~\ref{row:thm:intermittent-block-radial-face-closure}.  Failure of that
premise is not a necessary-condition test and does not rule out collapse
through another exact route.

\subsection{Controlled gate-zero invariance test}
\label{row:sec:gate-zero-evidence}

The gate-zero source test sets the selected final gate and first moment to
zero, retains the live second moment, and otherwise retains the real upstream
state and next batch.  The natural arm uses the implemented clipped gate
gradient.  The control zeros only that selected gradient immediately before
the moment write; all nonselected gradients are required to remain bitwise
identical.

Table~\ref{row:tab:gate-gradient-control} reports a complete controlled result.
The coordinatewise AdamW formula matches the natural successor bitwise in
\RowSourceBalanceGateTheoryPasses{} of nine units.  Every natural arm has \(64\)
nonzero clipped gate-gradient coordinates and reopens the physical row face.
Every controlled arm has zero gate force and preserves the face bitwise, while
all other gradients match.  Thus the nine natural reopenings are attributable
to the next clipped gate-gradient write at these states, and the exact
invariance condition in Proposition~\ref{row:prop:gate-zero-invariance} is
verified by a surgical control rather than by observation alone.

\subsection{Numerical methods and computational scope}
The paired interventions preserve the declared incoming model state,
optimizer moments and next minibatch. Producing-device observations and
float64 diagnostic reductions have separate roles: same-device comparisons
identify finite intervention effects, while the numerical controls assess
sensitivity to precision and reduction order. The direct-work bounds include
primitive coordinate magnitudes and finite summation length.

The retained radial reduction is CPU-only. Independent floating-point
reductions use an absolute tolerance of \(10^{-11}\) plus a relative tolerance
of \(5\times10^{-12}\), and agree on all 937 aggregate leaves. The
fresh-context radial experiment uses disjoint context chunks for
construction, held-out paths and terminal reserve batches. Peak producer
reservation for the temporal-context and fresh-context experiments is
\(\RowSourceReplicationPeakGiB{}\) GiB. Sample counts, eligibility floors,
precision rules and complete outcomes are reported with each experiment.

\subsection{What the finite record establishes}

The executed record establishes the following finite claims.

\begin{enumerate}[label=(\roman*)]
\item Row-map collapse is layer selective: six of nine float64 endpoints are
small and three are not.
\item Gate and normalized shape form a mixed closure.  Shape supplies most
late closure in five small units, while gate closure offsets shape reopening
in B1.
\item Pointwise final-gate contraction is false at the measured states.  All
nine lifted suboperators are expansive and nonnormal, with nonzero restricted
origin defects.
\item Same-device component effects are identifiable at one update, but the
source-written upstream predictor fails in seven of nine units.  The full
nonlinear optimizer state cannot be discarded.
\item Matrix-free complete-state differentiation is feasible in selected
directions, but its multistep linear prediction is unstable and only \(7/54\)
visible held-out gains attenuate.
\item The exact finite-increment identity accounts for every held-out response.
Strong cancellation occurs in six row-adjoint units, but a generic
cross-direction cancellation law is refuted.  State correction, observation
correction, and route changes must remain explicit.
\item Native paired direct-work ledgers close to numerical precision.  A
precision-resolved replay shows that all \(616\) disputed exclusions were
factorization diagnostics, while the locked result remains insufficient.
\item A disjoint temporal-context replication locks a positive sign only in
layer 2 and refutes its persistence on both held-out trajectories.  The exact
paired contrast is state conditional: its state-projection term agrees with
the observed sign in \(96.8\%\) of sign-eligible maps, with interaction and
charge retained for the cancellation tail.
\item The exact radial criterion classifies all 4,608 eligible retained arm
maps: 2,452 contract, while 2,015 reopen by radial-interval failure and 141 by
excess tangential charge.  The positive-excursion and affine energy replays
close for all 18 retained scalar windows.
\item The paired radial contribution agrees with 2,236 of 2,304 retained raw
signs.  In fresh contexts it agrees with 1,579 of 1,632 policy-eligible signs
and no powered cell is below its floor, but the frozen all-cells claim remains
insufficient because one floor-censored-positive cell has no policy-eligible
radial map.  All 96 maps in that cell are mathematically radial eligible.
\item The observer census finds zero exact source-energy zeros among 5,184
retained and fresh-context maps.  Every dense window has expanding endpoint
pairs at every measured horizon through 32, so fixed-window uniform
contraction is false on the captured sample.
\item The gate-zero face reopens naturally in all nine units and is preserved
in all nine controlled-gradient arms, identifying the implemented invariance
force at those states.
\item PLGA does not automatically inherit the upstream row-constant face.  Its
constant-input defect is the dominant downstream term near collapsed inputs.
\end{enumerate}

These observations support the observer-resolved work-charge theory, the
exact positive-energy radial normal form, and the unexercised exact-face
branch together with the stratified complete-state cocycle.  They do not
supply an infinite-time premise.  No complete-state tube, closing block
sequence, contracting full normal product, summable reopening budget, or
vanishing invariance-force convolution is inferred from finite data.  The
analytical theory states sufficient asymptotic conditions separately and
retains observable cancellation as a distinct, direction-dependent route to
physical row-map collapse.

\FloatBarrier
\par\medskip\noindent
Chapter~\ref{ch:affine-blocking} begins the chronological reduction of
these paths. It packages finite energy changes into affine maps whose
composition retains restarts and the effects of intermediate face hits.

\part{Chronological renormalization and predictive closure}
\chapter{Positive affine blocking through the row-constant face}
\label{ch:affine-blocking}
This chapter constructs the positive affine representation of row-energy
transport and its ordered blocking rule. It extends the representation
through the zero-energy face, includes interval observations and finite
registries, and states the exact two-sector collapse criterion.

\section{Exact positive affine renormalization map}
\label{rg:sec:exact-rg}

The radial identity of Theorem~\ref{row:thm:exact-radial-tangential-gain} supplies the positive-energy edge. We now retain its chronological composition and complete it across exact face hits.

\subsection{Radial and tangential edge geometry}

Let \(Z_t=\rowq X_t\), \(E_t=\norm{Z_t}_{\Fro}^2\), and
\(D_t=Z_{t+1}-Z_t\).  When \(E_t>0\), define
\begin{equation}
 \alpha_t=-\frac{\ip{Z_t}{D_t}}{E_t},\qquad
 T_t=D_t+\alpha_tZ_t,
 \qquad
 \tau_t^2=\frac{\norm{T_t}_{\Fro}^2}{E_t}.
 \label{rg:eq:radial-definition}
\end{equation}
Positive \(\alpha_t\) is inward radial motion.  The residual \(T_t\) is the
finite tangential secant, not a differential approximation.

\begin{theorem}[Exact radial gain and closing criterion]
\label{rg:thm:radial-closing}
On every edge with \(E_t>0\),
\begin{alignat}{2}
 \ip{Z_t}{T_t}&=0,
 &\quad Z_{t+1}&=(1-\alpha_t)Z_t+T_t,
 \label{rg:eq:radial-orthogonal}\\
 q_t:=\frac{E_{t+1}}{E_t}
  &=(1-\alpha_t)^2+\tau_t^2,
 &\quad \frac{E_{t+1}-E_t}{E_t}
  &=-2\alpha_t+\alpha_t^2+\tau_t^2.
 \label{rg:eq:radial-gain}
\end{alignat}
Moreover,
\begin{equation}
 E_{t+1}<E_t
 \quad\Longleftrightarrow\quad
 0<\alpha_t<2
 \quad\text{and}\quad
 \tau_t^2<\alpha_t(2-\alpha_t).
 \label{rg:eq:radial-closing}
\end{equation}
\end{theorem}

\begin{proof}
Equation~\eqref{rg:eq:radial-definition} gives
\[
 \ip{Z_t}{T_t}=\ip{Z_t}{D_t}+\alpha_t\norm{Z_t}_{\Fro}^2=0.
\]
It also gives \(D_t=-\alpha_tZ_t+T_t\), hence the endpoint split in
Equation~\eqref{rg:eq:radial-orthogonal}.  Orthogonality and Pythagoras imply
\[
 E_{t+1}=(1-\alpha_t)^2E_t+\norm{T_t}_{\Fro}^2.
\]
Divide by \(E_t>0\) and subtract one to obtain
Equation~\eqref{rg:eq:radial-gain}.  Finally,
\[
 q_t<1
 \quad\Longleftrightarrow\quad
 \tau_t^2<1-(1-\alpha_t)^2=\alpha_t(2-\alpha_t).
\]
Since \(\tau_t^2\ge0\), this strict inequality requires
\(\alpha_t(2-\alpha_t)>0\), which is equivalent to
\(0<\alpha_t<2\).  Reversing the steps proves the converse.
\end{proof}

The decomposition is unique.  If \(D_t=-aZ_t+T\) and
\(\ip{Z_t}{T}=0\), taking the inner product with \(Z_t\) gives
\(a=\alpha_t\), after which \(T=T_t\).  Therefore the exact gain has only
two exhaustive failure channels: a noninward or overshooting radial step, or
tangential charge exceeding the radial margin.

\subsection{Canonical face completion}

At \(E_t=0\), division by \(E_t\) is undefined.  The endpoint energy is
instead
\begin{equation}
 E_{t+1}=\norm{D_t}_{\Fro}^2.
 \label{rg:eq:face-restart-energy}
\end{equation}
Thus the exact face is preserved if and only if the complete centered
increment is zero.  Cancellation among parameter sources is allowed, but
zero of one source is not sufficient.

\begin{definition}[Canonical positive affine edge]
\label{rg:def:canonical-edge}
For any nonnegative energy sequence, set
\begin{equation}
 q_t=\begin{cases}
 E_{t+1}/E_t,&E_t>0,\\
 0,&E_t=0,
 \end{cases}
 \qquad
 \zeta_t=\begin{cases}
 0,&E_t>0,\\
 E_{t+1},&E_t=0.
 \end{cases}
 \label{rg:eq:canonical-edge}
\end{equation}
The pair \(\edge_t=(q_t,\zeta_t)\in\R_+^2\) is the canonical edge.
\end{definition}

\begin{theorem}[Closure of the canonical realized cone]
\label{rg:thm:canonical-cone-closure}
Let
\begin{equation}
 \mathcal C=\{(q,\zeta)\in\R_+^2:q\zeta=0\},
 \qquad \mathcal A_+=\R_+^2.
 \label{rg:eq:canonical-cone}
\end{equation}
Every edge in Definition~\ref{rg:def:canonical-edge} lies in \(\mathcal C\).
If \(\edge_1,\edge_2\in\mathcal C\), then
\(\edge_2\circ\edge_1\in\mathcal C\). Consequently every finite
chronological block of realized edges is in \(\mathcal C\).
\end{theorem}

\begin{proof}
The branch definition makes either \(q_t\) or \(\zeta_t\) zero, so
\(q_t\zeta_t=0\). For two canonical edges, the product of the blocked
coordinates is
\[
 (q_2q_1)(q_2\zeta_1+\zeta_2)
 =q_2^2(q_1\zeta_1)+q_1(q_2\zeta_2)=0.
\]
Both blocked coordinates are nonnegative. Induction on block length proves
the final statement, with the empty block \((1,0)\) as base case.
\end{proof}

A block crossing an internal face need not equal the endpoint-only
recanonicalization from Definition~\ref{rg:def:canonical-edge}. For example,
the path \(1,0,1\) composes to \((0,1)\), whereas recanonicalizing only
the endpoints gives \((1,0)\). Both reproduce that single endpoint, but the
blocked edge retains the loss of homogeneous dependence on incoming
scalar energy in this realized affine representation. It does not erase
parameters, optimizer moments, the remaining corpus, or common-row
information. These can still determine the future source.

The edge acts on a scalar by
\(\edge_t\cdot e=q_te+\zeta_t\).  Definition~\ref{rg:def:canonical-edge}
immediately gives \(E_{t+1}=\edge_t\cdot E_t\).  On the positive branch,
Theorem~\ref{rg:thm:radial-closing} supplies
\(q_t=(1-\alpha_t)^2+\tau_t^2\).  On the face branch,
Equation~\eqref{rg:eq:face-restart-energy} supplies \(\zeta_t\).

The canonical edge is an exact description of a realized orbit.  It is not,
by itself, a causal predictor of an unseen update because its gain uses the
endpoint.  A predictive theorem may replace it by any proved comparison edge
\((\bar q_t,\bar\zeta_t)\) satisfying
\(E_{t+1}\le\bar q_tE_t+\bar\zeta_t\).  The same blocking algebra applies,
with equality replaced by an upper bound.
Section~\ref{rg:sec:kernel-lumpability} gives the separate state-space
criterion for an autonomous predictive law, and
Section~\ref{rg:sec:staged-closure-results} tests candidate reduced states.

\subsection{Ordered affine blocking}
\label{rg:sec:ordered-affine-blocking}

For two edges define
\begin{equation}
 (q_2,\zeta_2)\circ(q_1,\zeta_1)
  =(q_2q_1,q_2\zeta_1+\zeta_2),
 \qquad \idEdge=(1,0).
 \label{rg:eq:edge-composition}
\end{equation}
The right edge acts first.  Source transport makes this convention
essential.

\begin{theorem}[Positive affine RG semigroup]
\label{rg:thm:affine-rg-semigroup}
The operation in Equation~\eqref{rg:eq:edge-composition} is associative, has
identity \(\idEdge\), preserves \(\R_+^2\), and satisfies
\begin{equation}
 (\edge_2\circ\edge_1)\cdot e
   =\edge_2\cdot(\edge_1\cdot e).
 \label{rg:eq:edge-action-composition}
\end{equation}
For a chronological block \([k,n)\), define
\begin{alignat}{2}
 Q(n,k)&=\prod_{j=k}^{n-1}q_j,
 &\quad Q(k,k)&=1,
 \label{rg:eq:block-gain}\\
 Z(n,k)&=\sum_{j=k}^{n-1}
   \left(\prod_{s=j+1}^{n-1}q_s\right)\zeta_j,
 &\quad Z(k,k)&=0.
 \label{rg:eq:block-source}
\end{alignat}
Then the blocked edge is \((Q(n,k),Z(n,k))\), and
\begin{equation}
 E_n=Q(n,k)E_k+Z(n,k).
 \label{rg:eq:exact-block-cocycle}
\end{equation}
All formulas remain valid through exact face hits.
\end{theorem}

\begin{proof}
Direct calculation gives
\begin{align*}
 ((q_3,\zeta_3)\circ(q_2,\zeta_2))\circ(q_1,\zeta_1)
  &=(q_3q_2q_1,q_3q_2\zeta_1+q_3\zeta_2+\zeta_3),\\
 (q_3,\zeta_3)\circ((q_2,\zeta_2)\circ(q_1,\zeta_1))
  &=(q_3q_2q_1,q_3q_2\zeta_1+q_3\zeta_2+\zeta_3).
\end{align*}
This proves associativity.  The identity, positivity, and
Equation~\eqref{rg:eq:edge-action-composition} follow by substitution.

For the block formula, the claim is immediate when \(n=k\).  Assume it holds
at \(n\).  Then
\[
 E_{n+1}=q_nE_n+\zeta_n
 =q_nQ(n,k)E_k+q_nZ(n,k)+\zeta_n.
\]
The first coefficient is \(Q(n+1,k)\).  Distributing \(q_n\) into
Equation~\eqref{rg:eq:block-source} and appending \(\zeta_n\) gives
\(Z(n+1,k)\).  Induction proves the result.  At a face hit the canonical
gain is zero, so all earlier transported terms are correctly erased while
the restart enters as a new source.  No division occurs in the block formula.
\end{proof}

For an aligned block factor \(b\), the discrete RG transformation on an edge
sequence is
\begin{equation}
 (\RG_b\edge)_m
   =\edge_{(m+1)b-1}\circ\cdots\circ\edge_{mb}.
 \label{rg:eq:sequence-rg}
\end{equation}
Associativity proves the exact semigroup identity
\begin{equation}
 \RG_c\RG_b=\RG_{bc}
 \label{rg:eq:block-semigroup}
\end{equation}
on aligned sequences.  The fine edges need not be equal or commute.

\subsection{Interval-affine observer RG}
\label{rg:sec:interval-affine-observer}

Write a nonnegative interval as \(I=[\underline e,\overline e]\) with
\(0\le\underline e\le\overline e\). An interval edge is
\(\mathbf A=(Q,Z)\), where
\(Q=[\underline q,\overline q]\) and
\(Z=[\underline\zeta,\overline\zeta]\). Its action and chronological
composition are
\begin{align}
 \mathbf A\odot I
 &= [\underline q\,\underline e+\underline\zeta,
      \overline q\,\overline e+\overline\zeta],
 \label{rg:eq:interval-affine-action}\\
 (Q_2,Z_2)\odot(Q_1,Z_1)
 &=\bigl(
 [\underline q_2\underline q_1,\overline q_2\overline q_1],
 [\underline q_2\underline\zeta_1+\underline\zeta_2,
  \overline q_2\overline\zeta_1+\overline\zeta_2]
 \bigr).
 \label{rg:eq:interval-affine-composition}
\end{align}

\begin{theorem}[Nonnegative interval-affine RG]
\label{rg:thm:interval-affine-rg}
The operation in Equation~\eqref{rg:eq:interval-affine-composition} is
associative, has identity \(([1,1],[0,0])\), preserves valid nonnegative
intervals, and is monotone under componentwise interval inclusion. If
\(e\in I\), \(q\in Q\), and \(\zeta\in Z\), then
\(qe+\zeta\in\mathbf A\odot I\). If exact edges
\(\edge_i=(q_i,\zeta_i)\) lie in interval edges \(\mathbf A_i\), then the
exact chronological composition lies in
\(\mathbf A_2\odot\mathbf A_1\). Singleton intervals recover the exact
affine RG.

If \(E_t\in[L_t,U_t]\), \(E_{t+1}\in[L_{t+1},U_{t+1}]\), and \(L_t>0\),
then the exact positive-branch gain obeys
\begin{equation}
 \frac{L_{t+1}}{U_t}\le q_t\le\frac{U_{t+1}}{L_t}.
 \label{rg:eq:interval-gain-bound}
\end{equation}
For every admissible exact state with \(E_t>0\),
\begin{equation}
 q_t=0\quad\Longleftrightarrow\quad E_{t+1}=0.
 \label{rg:eq:positive-source-zero-gain}
\end{equation}
When \(L_t=0\), \(U_t>0\), and \(U_{t+1}>0\), no finite upper gain bound follows
from the two energy intervals alone. If \(U_{t+1}=0\),
Equation~\eqref{rg:eq:positive-source-zero-gain} applies to every admissible
positive source, while an admissible zero source belongs to the restart branch
and has no gain quotient.
\end{theorem}

\begin{proof}
All variables are nonnegative, so multiplication and addition are increasing
in every argument. This proves validity, inclusion monotonicity, the action
enclosure, and both endpoints in Equation~\eqref{rg:eq:interval-affine-composition}.
For three interval edges, both association orders have gain endpoints
\(\underline q_3\underline q_2\underline q_1\) and
\(\overline q_3\overline q_2\overline q_1\). Their source endpoints are,
respectively,
\[
 \underline q_3\underline q_2\underline\zeta_1
 +\underline q_3\underline\zeta_2+\underline\zeta_3,
 \qquad
 \overline q_3\overline q_2\overline\zeta_1
 +\overline q_3\overline\zeta_2+\overline\zeta_3,
\]
so composition is associative. Direct substitution proves the identity.
Replacing every interval by a singleton gives
Equation~\eqref{rg:eq:edge-composition}, which also proves enclosure of exact
compositions by induction.

For \(L_t>0\), positivity gives
\(q_t=E_{t+1}/E_t\). The smallest possible quotient is bounded by
\(L_{t+1}/U_t\), and the largest by \(U_{t+1}/L_t\), proving
Equation~\eqref{rg:eq:interval-gain-bound}. For any \(E_t>0\), division by the
positive denominator gives both directions of
Equation~\eqref{rg:eq:positive-source-zero-gain}: if \(q_t=0\), then
\(E_{t+1}=q_tE_t=0\), and if \(E_{t+1}=0\), then
\(q_t=E_{t+1}/E_t=0\). If \(L_t=0<U_t\) and
\(U_{t+1}>0\), choose admissible positive starting energies tending to zero
and a fixed positive successor. Their quotients diverge, so the intervals
alone supply no finite upper bound. If \(U_{t+1}=0\), interval validity and
nonnegativity force \(E_{t+1}=0\), and the equivalence gives zero gain exactly
when the source is positive. At a zero source, the canonical definition instead
sets \(q_t=0\) and records \(E_{t+1}\) as the restart \(\zeta_t\), without
forming a quotient.
\end{proof}

The theorem concerns exact real endpoint operations, as in classical interval
analysis \cite{moore1966,neumaier1990}. In floating-point use, each endpoint
operation is rounded outward and edges are folded in chronological order.
That implementation is enclosure-valid, but floating-point addition and
multiplication are not associative. Regrouping or parallel reduction can
therefore produce a different valid interval and must not be justified by the
exact associativity theorem. A long fold can widen, but it cannot turn an
uncertified near-face state into an exact source event. This is the observer
counterpart of the exact semigroup, not a replacement for it.
Section~\ref{rg:sec:observer-results} applies the observer distinction to
recorded near-face states and stage-resolved arithmetic replays.

\subsection{Endpoint-gauge covariance}

An energy normalization is part of an RG specification.  Let \(s_t>0\) and
write \(\widehat E_t=E_t/s_t\).  An edge transforms to
\begin{equation}
 \mathcal G_{s_t,s_{t+1}}(q_t,\zeta_t)
 =\left(q_t\frac{s_t}{s_{t+1}},
       \frac{\zeta_t}{s_{t+1}}\right).
 \label{rg:eq:gauge-edge}
\end{equation}

\begin{theorem}[Gauge covariance]
\label{rg:thm:gauge-covariance}
For positive \(s_0,s_1,s_2\),
\begin{equation}
 \mathcal G_{s_1,s_2}(\edge_2)\circ
 \mathcal G_{s_0,s_1}(\edge_1)
 =\mathcal G_{s_0,s_2}(\edge_2\circ\edge_1).
 \label{rg:eq:gauge-covariance}
\end{equation}
More generally, for \(n\ge1\), positive \(s_0,\ldots,s_n\), and edges
\(\edge_1,\ldots,\edge_n\),
\begin{equation}
 \mathcal G_{s_{n-1},s_n}(\edge_n)\circ\cdots\circ
 \mathcal G_{s_0,s_1}(\edge_1)
 =\mathcal G_{s_0,s_n}(\edge_n\circ\cdots\circ\edge_1).
 \label{rg:eq:gauge-finite-block}
\end{equation}
Thus changing units before or after any exact finite block gives the same
normalized edge.
\end{theorem}

\begin{proof}
For two edges, the gain on the left is
\[
 q_2\frac{s_1}{s_2}q_1\frac{s_0}{s_1}
 =q_2q_1\frac{s_0}{s_2}.
\]
Its source is
\[
 q_2\frac{s_1}{s_2}\frac{\zeta_1}{s_1}
 +\frac{\zeta_2}{s_2}
 =\frac{q_2\zeta_1+\zeta_2}{s_2}.
\]
These are exactly the two coordinates on the right of
Equation~\eqref{rg:eq:gauge-covariance}.

For Equation~\eqref{rg:eq:gauge-finite-block}, use induction on \(n\).  The
case \(n=1\) is an identity.  Assume the formula holds for \(n\).  By
associativity, substitute the induction hypothesis into the last \(n\)
normalized factors.  Applying Equation~\eqref{rg:eq:gauge-covariance} with
\(\edge_2=\edge_{n+1}\), \(\edge_1=\edge_n\circ\cdots\circ\edge_1\),
and intermediate gauge \(s_n\) gives
\[
 \mathcal G_{s_0,s_{n+1}}
   (\edge_{n+1}\circ\edge_n\circ\cdots\circ\edge_1),
\]
which is the required formula for \(n+1\).
\end{proof}

Gauge covariance does not make every physical claim gauge invariant.  The
statement \(E_t\to0\) is preserved only when the chosen gauges are bounded
above and away from zero, or when the transfer is otherwise controlled.
Normalization is therefore recorded, not silently absorbed.

\subsection{Finite-registry matrix extension}

Let \(\bm E_t\in\R_+^M\) collect a finite registry. A coupled comparison
edge is \((K_t,\bm h_t)\), where \(K_t\in\R^{M\times M}\) has nonnegative
entries and \(\bm h_t\in\R^M\) satisfies \(\bm h_t\ge\bm0\).
Here \(\bm0\in\R^M\) is the zero vector. Vector inequalities are
componentwise, so \(\bm h_t\ge\bm0\) means \(h_{t,i}\ge0\) for every
\(i=1,\ldots,M\). The one-step comparison bound is
\begin{equation}
 \bm E_{t+1}\le K_t\bm E_t+\bm h_t
 \label{rg:eq:matrix-comparison}
\end{equation}

\begin{proposition}[Matrix comparison RG]
\label{rg:prop:matrix-comparison-rg}
Define
\begin{equation}
 (K_2,\bm h_2)\circ(K_1,\bm h_1)
  =(K_2K_1,K_2\bm h_1+\bm h_2).
 \label{rg:eq:matrix-rg}
\end{equation}
Nonnegative matrix comparison edges form an associative affine semigroup with
identity \((I,\bm0)\). If two successive one-step inequalities have edges
\((K_1,\bm h_1)\) and \((K_2,\bm h_2)\), their composition is a valid
two-step comparison edge.
\end{proposition}

\begin{proof}
Since every entry of \(K_2\) is nonnegative, componentwise order is
preserved by multiplication with \(K_2\). Therefore
\[
 \bm E_2\le K_2\bm E_1+\bm h_2
 \le K_2(K_1\bm E_0+\bm h_1)+\bm h_2
 =(K_2K_1)\bm E_0+K_2\bm h_1+\bm h_2.
\]
The product and source remain nonnegative. For three edges, both association
orders have matrix coordinate \(K_3K_2K_1\) and source
\(K_3K_2\bm h_1+K_3\bm h_2+\bm h_3\), proving associativity. Direct
substitution proves the identity property.
\end{proof}

A diagonal \(K_t\) recovers independent scalar comparison edges.
Off-diagonal entries can encode a proved physical cover, shared optimizer
comparison, or layer-to-layer transfer. For a positive diagonal endpoint
gauge \(S_t\), direct substitution into
\(\widehat{\bm E}_t=S_t^{-1}\bm E_t\) gives
\begin{equation}
 \widehat K_{n,k}=S_n^{-1}K_{n,k}S_k,
 \qquad
 \widehat{\bm h}_{n,k}=S_n^{-1}\bm h_{n,k}.
 \label{rg:eq:matrix-gauge}
\end{equation}

\subsection{Exact two-sector collapse criterion}

The right side of Equation~\eqref{rg:eq:exact-block-cocycle} splits into
\begin{equation}
 H_n=Q(n,0)E_0,
 \qquad
 F_n=Z(n,0),
 \qquad E_n=H_n+F_n.
 \label{rg:eq:two-sectors}
\end{equation}
Both sectors are nonnegative.

\begin{theorem}[Canonical two-sector collapse]
\label{rg:thm:two-sector-collapse}
For every nonnegative energy sequence with canonical edges,
\begin{equation}
 E_n\longrightarrow0
 \quad\Longleftrightarrow\quad
 H_n\longrightarrow0\ \text{and}\ F_n\longrightarrow0.
 \label{rg:eq:two-sector-iff}
\end{equation}
For a fixed finite registry, these conditions are also equivalent to collapse
of every registered row diameter.
\end{theorem}

\begin{proof}
If both sectors vanish, their sum vanishes.  Conversely,
\(0\le H_n\le E_n\) and \(0\le F_n\le E_n\).  If \(E_n\to0\), squeezing
gives both sector limits.  The registry statement follows by applying the
same argument to the stacked nonnegative energy and then using
Lemma~\ref{rg:lem:quotient-diameter} for each of its finitely many summands.
\end{proof}

The theorem is a characterization, not an optimizer guarantee.  It says
exactly what a row-collapse theory must close: memory of the initial
transverse energy and every restart after transport by the later dynamics.

\FloatBarrier
\par\medskip\noindent
Chapter~\ref{ch:stratified-flow} uses this composition law to classify
longer orbits. The distinction between homogeneous transport and injected
energy becomes central when faces recur or comparisons vary with time.

\chapter{Stratified flow and orbitwise collapse criteria}
\label{ch:stratified-flow}
This chapter studies the flow generated by chronological affine blocking.
It separates positive-energy evolution, finitely many reopenings and
recurrent-face excursions, and identifies the hypotheses under which
comparison bounds and clean scaling descriptions apply.

\section{RG flow and stratified criticality}
\label{rg:sec:flow-criticality}

\subsection{Homogeneous flow in the comparison cone}

Suppose every fine comparison edge is the same pair
\((q,\zeta)\in\mathcal A_+\). Iterating the affine law gives, for every
integer \(b\ge1\),
\begin{equation}
 \RG_b(q,\zeta)=
 \left(q^b,\zeta\sum_{j=0}^{b-1}q^j\right)
 =\begin{cases}
 \left(q^b,\displaystyle\zeta\frac{1-q^b}{1-q}\right),&q\ne1,\\[7pt]
 (1,b\zeta),&q=1.
 \end{cases}
 \label{rg:eq:homogeneous-rg}
\end{equation}
The finite sum is the primary formula and is nonsingular at \(q=1\).
A homogeneous pair with \(q\zeta>0\) is not a canonical realized fine
edge. It is a forced model or comparison edge. The source-free line and the
memoryless line \(q=0\) are the two axes of the canonical edge cone. A
constant sequence of memoryless edges \((0,\zeta)\) with \(\zeta>0\) is not,
however, generated by its own affine orbit: after the first restart its next
starting energy is positive. Repeated homogeneous blocking of that pair is
therefore a comparison flow, not a realized fine-edge trajectory.

\begin{theorem}[Clean fixed set and phase flow]
\label{rg:thm:clean-fixed-points}
Fix an integer \(b>1\). The fixed set of \(\RG_b\) on
\(\mathcal A_+\) is
\begin{equation}
 \{(0,\zeta):\zeta\ge0\}\cup\{(1,0)\}.
 \label{rg:eq:clean-fixed-set}
\end{equation}
For \(0\le q<1\),
\begin{equation}
 \RG_b^{\,n}(q,\zeta)\longrightarrow
 \left(0,\frac{\zeta}{1-q}\right),
 \label{rg:eq:forced-fixed-line}
\end{equation}
For \(q=1\) and \(\zeta>0\), the source grows as \(b^n\zeta\); for
\(q>1\), the gain diverges. For the homogeneous fine recurrence on the
critical ray, \(E_{t+1}=E_t+\zeta\), one has
\(E_N=E_0+N\zeta\), which diverges for every \(\zeta>0\). Only
\((0,0)\) on the memoryless fixed line is the collapsed absorbing point.
Equivalently, for every integer \(b\ge2\),
\begin{equation}
 \RG_b(1,\zeta)=(1,b\zeta),\qquad
 \RG_b^{\,n}(1,\zeta)=(1,b^n\zeta).
 \label{rg:eq:critical-source-exact-scaling}
\end{equation}
Thus the critical source is an exact relevant field of scaling dimension
one.
\end{theorem}

\begin{proof}
The fixed-gain equation \(q^b=q\) is
\(q(q^{b-1}-1)=0\). For \(q\ge0\), its only solutions are zero and one.
At \(q=0\), the source sum in Equation~\eqref{rg:eq:homogeneous-rg} is one,
so every \((0,\zeta)\) is fixed. At \(q=1\), the source is multiplied by
\(b>1\), so fixedness requires \(\zeta=0\).

Repeated blocking by \(b\) is blocking once by \(b^n\). If \(q<1\),
then \(q^{b^n}\to0\) and
\[
 \zeta\sum_{j=0}^{b^n-1}q^j
 =\zeta\frac{1-q^{b^n}}{1-q}
 \longrightarrow\frac{\zeta}{1-q}.
\]
At \(q=1\), Equation~\eqref{rg:eq:homogeneous-rg} gives source
\(b^n\zeta\). Summing \(E_{t+1}-E_t=\zeta\) from \(t=0\) to \(N-1\)
gives \(E_N=E_0+N\zeta\). If \(q>1\), then
\(q^{b^n}\to\infty\). This proves every claim.
\end{proof}

Let \(b=e^\ell\). For \(q>0\), continuous source-free blocking is
\begin{equation}
 q(\ell)=q_0^{e^\ell},\qquad
 \beta_q(q)=\frac{dq}{d\ell}=q\log q.
 \label{rg:eq:gain-beta}
\end{equation}
For \(q\ne1\), the stationary comparison coordinate
\begin{equation}
 u=\frac{\zeta}{1-q}
 \label{rg:eq:stationary-coordinate}
\end{equation}
is invariant, and \(\zeta(\ell)=u(1-q(\ell))\). Hence
\begin{equation}
 \beta_\zeta(q,\zeta)
 =-\frac{\zeta q\log q}{1-q}.
 \label{rg:eq:source-beta}
\end{equation}
Its continuous limit at \(q=1\) is \(\beta_\zeta=\zeta\), consistent
with \(\zeta\mapsto e^\ell\zeta\).
\subsection{Relevant fields and clean exponents}

Write \(\tau=1-q\). For fixed integer \(b>1\), expansion at
\((\tau,\zeta)=(0,0)\) gives
\begin{align}
 \tau'&=b\tau-\frac{b(b-1)}2\tau^2+O(\tau^3),
 \label{rg:eq:tau-linearization}\\
 \zeta'&=b\zeta-\frac{b(b-1)}2\tau\zeta
   +O(\tau^2\zeta).
 \label{rg:eq:zeta-linearization}
\end{align}
Thus the two linearized scaling dimensions in \(\mathcal A_+\) are
\begin{equation}
 y_\tau=1,\qquad y_\zeta=1,\qquad
 \phi=y_\zeta/y_\tau=1.
 \label{rg:eq:scaling-dimensions}
\end{equation}
These dimensions are exact linearization coefficients of the homogeneous RG,
not fitted exponents from a finite training trajectory.
The source direction is transverse to the positive source-free branch of
the canonical cone; it remains physically relevant for face restarts and
comparison defects.

\begin{theorem}[Clean critical scaling]
\label{rg:thm:clean-critical-scaling}
For the source-free homogeneous model, define
\begin{equation}
 \xi_{\parallel}(q)=\frac1{\abs{\log q}},
 \qquad q>0,\quad q\ne1.
 \label{rg:eq:clean-correlation-scale}
\end{equation}
Then
\begin{equation}
 \xi_{\parallel}(q)\sim\abs{1-q}^{-1}
 \quad(q\to1),
 \label{rg:eq:clean-nu}
\end{equation}
so \(\nu_{\parallel}=1\). Here \(\nu_{\parallel}\) is the exponent induced
by the declared scale \(\xi_{\parallel}\), not a separately fitted empirical
exponent. For a subcritical homogeneous comparison model,
\begin{equation}
 E_*=\frac{\zeta}{1-q},\qquad
 \chi_\zeta=\frac{\partial E_*}{\partial\zeta}
 =\frac1{1-q},
 \label{rg:eq:source-susceptibility}
\end{equation}
and the source susceptibility exponent is one.
\end{theorem}

\begin{proof}
Taylor's formula gives
\(\log q=(q-1)+O((q-1)^2)\), so
\(\abs{\log q}/\abs{1-q}\to1\). Taking reciprocals proves
Equation~\eqref{rg:eq:clean-nu}. The fixed energy solves
\(E_*=qE_*+\zeta\), which gives the first part of
Equation~\eqref{rg:eq:source-susceptibility}; differentiation gives the second.
\end{proof}

There is no path-independent order-parameter exponent for \(E_*\).
If \(\zeta\) scales as \(\tau^p\), then \(E_*\) scales as
\(\tau^{p-1}\). Collapse along that approach requires \(p>1\), or
equivalently \(\zeta/\tau\to0\).

\subsection{Face-free and finitely reopened orbits}

On an interval where every energy is positive, canonical sources vanish and
\begin{equation}
 E_n=E_k\exp\!\left(\sum_{t=k}^{n-1}Y_t\right),
 \qquad Y_t=\log q_t.
 \label{rg:eq:log-gain-cocycle}
\end{equation}
On a strictly positive tail, products of scalar energy gains become sums
of log gains, yielding the cumulative-log criterion below.

\begin{lemma}[Exact cumulative-log criterion]
\label{rg:lem:cumulative-log-criterion}
Fix \(k\) and suppose \(E_t>0\) for every \(t\ge k\). Set
\[
 S_{n,k}=\sum_{t=k}^{n-1}Y_t .
\]
Then
\begin{equation}
 E_n\to0\quad\Longleftrightarrow\quad S_{n,k}\to-\infty,
 \qquad
 E_n\to\infty\quad\Longleftrightarrow\quad S_{n,k}\to+\infty.
 \label{rg:eq:cumulative-log-criterion}
\end{equation}
For every \(L>0\),
\(E_n\to L\) if and only if
\(S_{n,k}\to\log(L/E_k)\in\R\).
Thus the cumulative log gain, rather than only its Cesàro average, gives the
exact criterion on a positive tail.
\end{lemma}

\begin{proof}
Equation~\eqref{rg:eq:log-gain-cocycle} gives
\(\log(E_n/E_k)=S_{n,k}\). Because \(E_k>0\), continuity and strict
monotonicity of the exponential map give all three conclusions.
\end{proof}

\begin{proposition}[Regular nonautonomous approach to criticality]
\label{rg:prop:regular-nonautonomous-drift}
Suppose the positive-tail hypothesis of
Lemma~\ref{rg:lem:cumulative-log-criterion} holds and, for constants
\(c>0\) and \(\gamma\ge0\),
\begin{equation}
 Y_t=-c(t+1)^{-\gamma}+\epsilon_t,\qquad
 R_{n,k}=\sum_{t=k}^{n-1}\epsilon_t .
 \label{rg:eq:regular-drift-model}
\end{equation}
For fixed \(k\), the following statements hold.
\begin{enumerate}
 \item If \(0\le\gamma<1\) and
 \(R_{n,k}=o(n^{1-\gamma})\), then
 \begin{equation}
  \log E_n=-\frac{c}{1-\gamma}n^{1-\gamma}
  +o(n^{1-\gamma}),
  \label{rg:eq:stretched-exponential-critical-approach}
 \end{equation}
 so \(E_n\to0\) with stretched-exponential asymptotics.  The endpoint
 \(\gamma=0\) is ordinary exponential collapse.
 \item If \(\gamma=1\) and \(R_{n,k}=o(\log n)\), then
 \begin{equation}
  \log E_n=-c\log n+o(\log n),\qquad E_n=n^{-c+o(1)},
  \label{rg:eq:power-law-critical-approach}
 \end{equation}
 so \(E_n\to0\) algebraically.
 \item If \(\gamma>1\) and \(R_{n,k}\) converges to a finite limit, then
 \(E_n\) converges to a strictly positive finite limit.  A summable
 approach of the log gain to zero therefore does not force collapse.
\end{enumerate}
Moreover, for every integer \(b\ge1\), every \(m\ge1\), and the
deterministic leading field,
\begin{equation}
 0\le cb^{1-\gamma}m^{-\gamma}
       -c\sum_{s=1}^{b}(bm+s)^{-\gamma}
 \le \frac{c\gamma b(b+1)}2(bm)^{-\gamma-1}.
 \label{rg:eq:regular-drift-finite-block-bound}
\end{equation}
For fixed \(b\) and \(\gamma>0\), the upper bound is asymptotically sharp:
\begin{equation}
 cb^{1-\gamma}m^{-\gamma}
 -c\sum_{s=1}^{b}(bm+s)^{-\gamma}
 =\frac{c\gamma b(b+1)}2(bm)^{-\gamma-1}
 +O(m^{-\gamma-2}).
 \label{rg:eq:regular-drift-finite-block-remainder}
\end{equation}
Consequently
\begin{equation}
 c\sum_{s=1}^{b}(bm+s)^{-\gamma}
 =cb^{1-\gamma}m^{-\gamma}\bigl(1+O(m^{-1})\bigr)
 \qquad(m\to\infty).
 \label{rg:eq:regular-drift-block-scaling}
\end{equation}
Thus the asymptotic amplitude transforms as \(c_b=b^{1-\gamma}c\), with
scaling dimension \(y_c=1-\gamma\). This is not an exact finite-origin
identity. The field is relevant for \(\gamma<1\), marginal for \(\gamma=1\),
and irrelevant for \(\gamma>1\). At finite \(m\), especially when
\(\gamma\) is near one, an observed small change of amplitude cannot by
itself determine the sign of \(y_c\); the \(O(m^{-1})\) correction must be
resolved.
\end{proposition}

\begin{proof}
Taking logarithms in Equation~\eqref{rg:eq:log-gain-cocycle} gives
\[
 \log E_n=\log E_k-c\sum_{t=k}^{n-1}(t+1)^{-\gamma}+R_{n,k}.
\]
For \(0\le\gamma<1\), integral comparison yields
\[
 \sum_{t=k}^{n-1}(t+1)^{-\gamma}
 =\frac{n^{1-\gamma}}{1-\gamma}+o(n^{1-\gamma}).
\]
For \(\gamma=1\), the same comparison gives \(\log n+O(1)\).
These estimates and the assumed residual bounds prove the first two
claims.  If \(\gamma>1\), the deterministic series converges; convergence
of \(R_{n,k}\) makes the cumulative log gain converge as well.
Lemma~\ref{rg:lem:cumulative-log-criterion} then gives a positive finite
energy limit.

For \(f(x)=x^{-\gamma}\), the mean-value theorem and monotonicity of
\(\abs{f'(x)}=\gamma x^{-\gamma-1}\) give, for \(m\ge1\),
\[
 \left|(bm+s)^{-\gamma}-(bm)^{-\gamma}\right|
 \le\gamma s(bm)^{-\gamma-1}.
\]
Since \(f\) is nonincreasing, each summand is at most \(f(bm)\), which
proves the left inequality. Multiply the displayed bound by \(c\) and sum
\(s=1,\ldots,b\) to prove
Equation~\eqref{rg:eq:regular-drift-finite-block-bound}. Dividing its remainder
by \(cb^{1-\gamma}m^{-\gamma}\) proves
Equation~\eqref{rg:eq:regular-drift-block-scaling}. The relevance classification
is the sign of the asymptotic amplitude dimension \(1-\gamma\).

For fixed \(b\), Taylor's theorem gives, uniformly over
\(s=1,\ldots,b\),
\[
 (bm+s)^{-\gamma}
 =(bm)^{-\gamma}
 -\gamma s(bm)^{-\gamma-1}
 +O(m^{-\gamma-2}).
\]
Summing \(s=b(b+1)/2\) proves
Equation~\eqref{rg:eq:regular-drift-finite-block-remainder}.
\end{proof}

When it exists, define
\begin{equation}
 \lambda_k=\lim_{n\to\infty}
 \frac1{n-k}\sum_{t=k}^{n-1}Y_t.
 \label{rg:eq:lyapunov-rate}
\end{equation}

\begin{theorem}[Nonzero mean-rate specialization]
\label{rg:thm:positive-excursion-phase}
Assume \(E_t>0\) for all \(t\ge k\) and that \(\lambda_k\) exists.
If \(\lambda_k<0\), then \(E_n\to0\) exponentially. If
\(\lambda_k>0\), then \(E_n\to\infty\) exponentially. If
\(\lambda_k=0\), the rate alone does not determine the orbit.
\end{theorem}

\begin{proof}
Let \(S_{n,k}=\sum_{t=k}^{n-1}Y_t\). If \(\lambda_k<0\), then for all
large \(n\),
\(S_{n,k}/(n-k)<\lambda_k/2<0\), so
\(E_n\le E_ke^{(n-k)\lambda_k/2}\to0\). The positive case follows by
reversing the inequality. At zero rate, the positive sequences
\(E_n=1\), \(E_n=(n+1)^{-1}\), and \(E_n=n+1\) respectively remain
constant, collapse, and diverge, while
\(n^{-1}\log(E_n/E_0)\to0\) in every case.
\end{proof}

For a stationary ergodic positive-gain process with
\(\E\abs{Y_0}<\infty\), Birkhoff's theorem gives
\(\lambda_0=\E Y_0\) almost surely \cite{birkhoff1931}. Training data
must establish the stationarity and integrability premises before this
identification is used. The temporal-domain and finite-law tests of the
recorded PLDR trajectories are reported in
Section~\ref{rg:sec:temporal-flow-results}, with their design in
Section~\ref{rg:sec:long-design}.

If the orbit has only finitely many face hits, let \(k\) be the final one.
If it remains on the face forever, it is collapsed. Otherwise
\(E_{k+1}>0\), and Theorem~\ref{rg:thm:positive-excursion-phase} applies from
the restart at \(k+1\). The zero canonical gain erases the homogeneous
contribution of the earlier scalar energy to this realized affine recurrence.
Optimizer moments, the remaining corpus, and other hidden state coordinates
can still determine the restart source and subsequent gains. Indeed, for
fixed realized coefficients, chronological composition gives
\begin{equation}
 E_n=\left(\prod_{t=i}^{n-1}q_t\right)E_i
   +\sum_{j=i}^{n-1}\zeta_j\prod_{t=j+1}^{n-1}q_t.
 \label{rg:eq:scalar-face-memory}
\end{equation}
A zero gain inside the block makes the coefficient of \(E_i\) vanish.
It does not remove dependence of the coefficient list itself on the complete
state. Holding this list fixed differs from changing the training state and
regenerating the list.

\subsection{Recurrent faces and the exact excursion criterion}

Assume the orbit has infinitely many face hits. Let
\(\sigma_0<\sigma_1<\cdots\) enumerate successive times with
\(E_{\sigma_j}=0\). Define
\begin{equation}
 r_j=E_{\sigma_j+1},\qquad
 P_j=\max_{\sigma_j<t\le\sigma_{j+1}}E_t,\qquad
 M_j=\begin{cases}P_j/r_j,&r_j>0,\\0,&r_j=0.\end{cases}
 \label{rg:eq:excursion-variables}
\end{equation}
If two face times are consecutive, then \(r_j=P_j=0\). Otherwise
\(r_j>0\), \(M_j\ge1\), and \(P_j=r_jM_j\).

The face-time decomposition specializes the recurrent restart
criterion of Theorem~\ref{row:thm:intermittent-block-radial-face-closure}
by making the excursion peak the exact collapse variable.

\begin{theorem}[Recurrent-face collapse criterion]
\label{rg:thm:recurrent-face-collapse}
Under Equation~\eqref{rg:eq:excursion-variables},
\begin{equation}
 E_t\longrightarrow0
 \quad\Longleftrightarrow\quad
 P_j=r_jM_j\longrightarrow0.
 \label{rg:eq:recurrent-face-criterion}
\end{equation}
Consequently \(r_j\to0\) is necessary but is not sufficient.
\end{theorem}

\begin{proof}
If \(E_t\to0\), then for every \(\varepsilon>0\) there is \(N\) such
that \(E_t<\varepsilon\) for all \(t\ge N\). For every sufficiently
large \(j\), \(\sigma_j\ge N\), so the maximum over its excursion obeys
\(P_j\le\varepsilon\).

Conversely, suppose \(P_j\to0\). Given \(\varepsilon>0\), choose \(J\)
so that \(P_j<\varepsilon\) for \(j\ge J\). Every
\(t\ge\sigma_J\) is either a face time, where \(E_t=0\), or lies in one of
the successive excursion intervals, where \(E_t\le P_j<\varepsilon\).
Thus \(E_t\to0\). Finally \(0\le r_j\le P_j\), so collapse implies
\(r_j\to0\).

To prove insufficiency, define for \(m\ge0\)
\begin{equation}
 E_{3m}=0,\qquad E_{3m+1}=\frac1{m+1},\qquad E_{3m+2}=1.
 \label{rg:eq:canonical-obstruction-orbit}
\end{equation}
Its three successive canonical edges are
\begin{equation}
 \left(0,\frac1{m+1}\right),\qquad
 (m+1,0),\qquad(0,0).
 \label{rg:eq:canonical-obstruction-edges}
\end{equation}
They replay the orbit exactly. The restarts tend to zero, but
\(M_m=m+1\) and \(r_mM_m=1\), so the orbit does not collapse.
\end{proof}

The counterexample is canonical: every edge lies in the realized cone, so
the obstruction does not arise from allowing gain and source to be positive
on the same fine edge.

The competition can be expressed as a recurrent-face scaling field. Suppose
eventually \(0<r_j<1\), \(r_j\to0\), and define
\begin{equation}
 \rho_j=-\log r_j,\qquad a_j=\log M_j,\qquad
 \chi_{\rm exc}=\lim_{j\to\infty}\frac{a_j}{\rho_j},
 \label{rg:eq:excursion-exponent}
\end{equation}
when the limit exists.

\begin{proposition}[Restart-amplification critical exponent]
\label{rg:prop:restart-amplification-criticality}
Under these assumptions, \(\chi_{\rm exc}<1\) implies collapse,
whereas \(\chi_{\rm exc}>1\) implies \(P_j\to\infty\). At
\(\chi_{\rm exc}=1\), subleading terms decide the orbit.
\end{proposition}

\begin{proof}
The peak has logarithm
\[
 \log P_j=\log r_j+\log M_j=-\rho_j+a_j.
\]
Because \(r_j\to0\), \(\rho_j\to\infty\). If
\(\chi_{\rm exc}<1\), choose \(c\) strictly between the limit and one.
Eventually \(a_j\le c\rho_j\), hence
\(\log P_j\le-(1-c)\rho_j\to-\infty\). The recurrent-face criterion
gives collapse. If \(\chi_{\rm exc}>1\), choose \(c\) strictly between
one and the limit. Eventually
\(\log P_j\ge(c-1)\rho_j\to\infty\). At equality, choosing
\(a_j=\rho_j\), \(a_j=\rho_j-\sqrt{\rho_j}\), or
\(a_j=\rho_j+\sqrt{\rho_j}\) gives respectively constant, vanishing, or
diverging peaks while the ratio tends to one.
\end{proof}

\subsection{Uniform comparison closure}

A useful sufficient condition applies to comparison edges even when a
Lyapunov limit is unavailable.

For affine comparison blocks, uniform contraction and vanishing forcing
imply decay through the geometric-convolution estimate proved below.

\begin{theorem}[Uniform block closure]
\label{rg:thm:uniform-block-closure}
Let \(A_j,h_j\ge0\) satisfy
\begin{equation}
 A_{j+1}\le\rho A_j+h_j,\qquad
 0\le\rho<1,\qquad h_j\longrightarrow0.
 \label{rg:eq:uniform-block-recurrence}
\end{equation}
Then \(A_j\to0\).
\end{theorem}

\begin{proof}
Iteration from \(N\) gives
\begin{equation}
 A_{N+k}\le\rho^kA_N+
 \sum_{i=0}^{k-1}\rho^{k-1-i}h_{N+i}.
 \label{rg:eq:uniform-convolution}
\end{equation}
Given \(\varepsilon>0\), choose \(N\) so that
\(h_j<(1-\rho)\varepsilon/2\) for \(j\ge N\). The convolution is then
less than \(\varepsilon/2\). For sufficiently large \(k\),
\(\rho^kA_N<\varepsilon/2\), proving \(A_j\to0\).
\end{proof}
The next corollary passes from vanishing block maxima to collapse along
the full orbit.

\begin{corollary}[Block-cover closure]
\label{rg:cor:block-cover-closure}
Let \(0\le\sigma_0<\sigma_1<\cdots\) be an unbounded sequence of integer
anchors, let \(E_t\ge0\), and define
\[
 B_j=\max_{\sigma_j\le t<\sigma_{j+1}}E_t .
\]
If \(B_j\to0\), then \(E_t\to0\).
\end{corollary}

\begin{proof}
Given \(\varepsilon>0\), choose \(J\) so that \(B_j<\varepsilon\) for
all \(j\ge J\).  Every integer \(t\ge\sigma_J\) belongs to one of the
successive intervals \([\sigma_j,\sigma_{j+1})\) with \(j\ge J\), and hence
\(E_t\le B_j<\varepsilon\).
\end{proof}

\subsection{The stratified critical set}

The exact critical object depends on face history:

\begin{enumerate}[label=(\roman*)]
\item On every face-free positive tail, the exact additive control variable is
  cumulative log gain \(S_{n,k}\).  If \(\lambda_k\) exists and is nonzero,
  its sign selects exponential collapse or growth.  The rate-critical surface
  \(\lambda_k=0\) remains stratified by the sublinear behavior of \(S_{n,k}\),
  including the regimes in
  Proposition~\ref{rg:prop:regular-nonautonomous-drift}.
\item After finitely many restarts, the same cumulative criterion applies to
  the final positive tail; an eventually invariant face is already collapsed.
\item With recurrent positive restarts and an existing exponent in
  Equation~\eqref{rg:eq:excursion-exponent}, the restart-amplification boundary
  is \(\chi_{\rm exc}=1\). The exact criterion remains
  \(r_jM_j\to0\), including cases where no exponent exists.
\item In the full comparison cone, collapse requires both homogeneous memory
  and transported source to vanish as in
  Theorem~\ref{rg:thm:two-sector-collapse}; no scalar gain coordinate alone is
  complete.
\end{enumerate}

This stratification separates typical contraction, uniform contraction,
restart extinction, and actual orbit collapse. They are related but
none may be substituted for another without its stated hypotheses.
Section~\ref{rg:sec:temporal-flow-results} reports the finite flow decisions,
and Section~\ref{rg:sec:observer-results} explains why represented restarts
do not establish the analytical recurrent-face premises.

\FloatBarrier
\par\medskip\noindent
An exact description of an observed orbit does not by itself define a
closed predictive state. Chapter~\ref{ch:closed-states} addresses that
closure question and then states conditional universality results for the
resulting probability laws.

\chapter{Closed states, gauge fibers, and conditional universality}
\label{ch:closed-states}
This chapter separates pathwise blocking from autonomous state reduction.
It develops the closure condition on complete-state fibers, treats gauge
and optimizer coordinates, and distinguishes clean, disordered and affine
universality descriptions under their stated probabilistic assumptions.

\section{State-space closure and multiplicative universality classes}
\label{rg:sec:universality}

\subsection{Standard complete-state blocking and scale-indexed lumpability}
\label{rg:sec:kernel-lumpability}

Let \((\mathsf X,\mathcal X)\) be a standard Borel state space containing
all coordinates needed to determine the distribution of the next training
update, including parameters, optimizer moments, data-stream position, and
randomness state.  Let \(K\) be its time-homogeneous Markov transition
kernel.  For two kernels, with the right kernel acting first, define
\begin{equation}
 (K_2\circ K_1)(x,A)
 =\int_{\mathsf X}K_2(y,A)\,K_1(x,dy),
 \qquad I(x,A)=\mathbf1_A(x).
 \label{rg:eq:kernel-composition}
\end{equation}
Write \(K^0=I\), \(K^{n+1}=K\circ K^n\), and
\(\RG_bK=K^b\) for integer \(b\ge1\).

\begin{theorem}[Kernel blocking and scale-indexed observable closure]
\label{rg:thm:complete-state-kernel-rg}
The operation in Equation~\eqref{rg:eq:kernel-composition} is associative and
has identity \(I\).  Consequently,
\begin{equation}
 \RG_c(\RG_bK)=\RG_{bc}K
 \qquad\text{for all integers }b,c\ge1.
 \label{rg:eq:kernel-rg-semigroup}
\end{equation}

Let \((\mathsf E,\mathcal E)\) be a standard Borel observable space and let
\(e:\mathsf X\to\mathsf E\) be measurable.  For a fixed integer \(b\ge1\),
call \(K^b\) \emph{strongly lumpable through \(e\)} when there is a Markov
kernel \(\bar K_b\) on \(\mathsf E\) such that
\begin{equation}
 K^b(x,e^{-1}(B))=\bar K_b(e(x),B)
 \quad\text{for every }x\in\mathsf X,\ B\in\mathcal E.
 \label{rg:eq:energy-lumpability}
\end{equation}
An autonomous \(b\)-step observable transition kernel exists exactly under
this condition, with ``autonomous'' meaning that the same kernel represents
the observable transition for every complete-state initial probability law.
Equivalently, for every probability measure \(\mu\) on \(\mathsf X\),
\begin{equation}
 e_\#(\mu K^b)=(e_\#\mu)\bar K_b,
 \label{rg:eq:every-initial-law}
\end{equation}
where \(e_\#\) denotes pushforward. If the condition holds at scale \(b\),
then for every integer \(c\ge1\)
\begin{equation}
 K^{bc}(x,e^{-1}(B))=\bar K_b^{\,c}(e(x),B)
 \quad\text{for every }x\in\mathsf X,\ B\in\mathcal E.
 \label{rg:eq:lumped-blocking}
\end{equation}
Thus closure at scale \(b\) propagates to every multiple of \(b\), but it
need not descend to scale one.

For a deterministic update \(F\), so that
\(K(x,\cdot)=\delta_{F(x)}\), scale-\(b\) closure implies
\begin{equation}
 e(x)=e(x')\quad\Longrightarrow\quad
 e(F^b(x))=e(F^b(x')).
 \label{rg:eq:deterministic-scale-lumpability}
\end{equation}
Conversely, this condition gives scale-\(b\) closure whenever its induced
map \(g_b(e(x))=e(F^b(x))\) is measurable, as it always is for finite
discrete spaces.
In particular, row energy need not be Markov.  If two complete states have
the same energy but different pushforward next-energy laws, no one-step
energy kernel represents both.
\end{theorem}

\begin{proof}
For three kernels, both association orders evaluate to
\[
 \int_{\mathsf X}\!\int_{\mathsf X}
 K_3(z,A)\,K_2(y,dz)\,K_1(x,dy).
\]
The integrand is nonnegative, so the kernel form of Tonelli's theorem
identifies the two iterated integrals.  Direct integration against the Dirac
kernel \(I\) proves the two identity laws.  Associativity then gives
\((K^b)^c=K^{bc}\), which is
Equation~\eqref{rg:eq:kernel-rg-semigroup}.

First assume Equation~\eqref{rg:eq:energy-lumpability}. For every initial law
\(\mu\) and measurable \(B\), Tonelli's theorem gives
\[
 \bigl[e_\#(\mu K^b)\bigr](B)
 =\int_{\mathsf X}K^b(x,e^{-1}(B))\,\mu(dx)
 =\int_{\mathsf E}\bar K_b(u,B)\,(e_\#\mu)(du),
\]
which is Equation~\eqref{rg:eq:every-initial-law}. Conversely, if one observable
kernel satisfies Equation~\eqref{rg:eq:every-initial-law} for every \(\mu\), take
\(\mu=\delta_x\). This recovers Equation~\eqref{rg:eq:energy-lumpability} for
every \(x\) and \(B\), proving the claimed equivalence and making the
all-initial-law quantifier explicit.

Assume now that Equation~\eqref{rg:eq:energy-lumpability} holds. The case
\(c=1\) of Equation~\eqref{rg:eq:lumped-blocking} is immediate.  If it holds
for \(c\), associativity and the change-of-variables formula for the
pushforward under \(e\) give
\[
\begin{split}
 K^{b(c+1)}(x,e^{-1}(B))
 &=\int_{\mathsf X}K^{bc}(y,e^{-1}(B))\,K^b(x,dy)\\
 &=\int_{\mathsf X}\bar K_b^{\,c}(e(y),B)\,K^b(x,dy)\\
 &=\int_{\mathsf E}\bar K_b^{\,c}(u,B)\,
       \bar K_b(e(x),du)\\
 &=\bar K_b^{\,c+1}(e(x),B).
\end{split}
\]
Induction proves closure at every multiple of \(b\).

For deterministic \(F\),
\[
 K^b(x,e^{-1}(B))=\mathbf1_B(e(F^b(x))).
\]
If scale-\(b\) closure holds, equal values of \(e(x)\) give equal Dirac
laws on the right for every measurable \(B\), hence equal values of
\(e(F^b(x))\).  Conversely, when the induced \(g_b\) is measurable,
\(\bar K_b(u,\cdot)=\delta_{g_b(u)}\) verifies
Equation~\eqref{rg:eq:energy-lumpability} by substitution.

It remains to prove that closure need not descend.  Give
\(\mathsf X=\{0,1,2\}\) and \(\mathsf E=\{A,B\}\) their discrete
sigma-algebras.  Let \(F(0)=1\), \(F(1)=2\), \(F(2)=0\), and set
\(e(0)=e(1)=A\), \(e(2)=B\).  Since \(F^3\) is the identity, \(K^3\)
is lumpable with \(\bar K_3\) equal to the identity kernel on
\(\mathsf E\).  But \(e(0)=e(1)\), whereas
\(e(F(0))=A\) and \(e(F(1))=B\).  Thus \(K\) is not lumpable through
\(e\), completing the counterexample and the proof.
\end{proof}

Kernel powering, strong lumpability, and Markov functions are classical
constructions \cite{kemenysnell1960,burkerosenblatt1958,rogerspitman1981,kallenberg2021}.
The theorem states that standard construction here only to delimit the
closure condition.  No PLDR transition kernel \(K\) or lumped kernel
\(\bar K_b\) is estimated in this work.  A time-inhomogeneous schedule is
blocked by chronological composition of its kernels, or made homogeneous by
adjoining the clock and schedule state to \(\mathsf X\). For PLDR training,
the consuming-corpus augmentation is specified in
Section~\ref{model:sec:data-resource}. Energy forgets row orientation,
common-row coordinates, gradient information,
and AdamW moments.  Lumpability therefore cannot be assumed merely because
the scalar cocycle replays a realized trajectory.

State-abstraction work likewise distinguishes quotients that preserve a
transition model from coarser summaries useful only for selected objectives
\cite{liwalshlittman2006}. Projection formalisms give a complementary warning:
eliminating unresolved coordinates generally produces memory and fluctuating
forcing rather than an autonomous reduced evolution
\cite{zwanzig1961,mori1965}. Random affine recurrence theory begins with an
affine stochastic law and studies its stationary or tail behavior
\cite{kesten1973}. Our canonical coefficients instead factor realized PLDR
endpoints. They become a predictive reduced process only after the
scale-specific fiber condition above is established. Thus the classical tools
supply the comparison framework; the PLDR-specific content is the observable,
face handling, gauge-aware state tests, and executed closure gate. The
matched-fiber tests and staged candidate-state experiment are detailed in
Sections~\ref{rg:sec:matched-fiber-results} and
\ref{rg:sec:staged-closure-results}, respectively.

For a deterministic update, this observation gives a direct falsification
criterion. If \(e(x)=e(x')\), define the scale-\(b\) fiber defect
\[
 D_b(x,x')=\norm{e(F^b(x))-e(F^b(x'))}_\infty.
\]
Strong lumpability on a stated domain requires \(D_b(x,x')=0\) for every
same-fiber pair in that domain. A single reproducible positive defect rejects
closure there. It does not, by itself, decide closure on a smaller reachable
training manifold that excludes the intervention pair.

The conclusion is about information on a fiber, not uniqueness of a chosen
parameterization. Suppose an intervention holds every declared coordinate
fixed except $a$, preserves $e$, and gives $D_b>0$. Then the successor
observable is nonconstant along the tested $a$-direction inside an
$e$-fiber, so $e$ alone discards predictive information on that domain.
The coordinate $a$ is thereby a demonstrated fiber coordinate. This does
not prove that $(e,a)$ closes on all states, that every component of $a$ is
needed, or that raw $a$ is the unique or minimal sufficient augmentation.
A different statistic may separate the tested pair and close the dynamics.
Those stronger conclusions require their own fiber-wide tests.

\subsection{Gauge fibers and optimizer-coordinate closure}
\label{rg:sec:gauge-fibers}

Energy equality can arise either from a physical change of complete state or
from a parameter symmetry that leaves the row map invariant. These cases must
be separated. The following result supplies the null control for that
separation.

\begin{proposition}[Gauge-orbit null control]
\label{rg:prop:gauge-orbit-null}
Let \(F:\mathsf X\to\mathsf X\) be a deterministic update, let a family
\(G\) act on \(\mathsf X\), and let \(e:\mathsf X\to\mathsf E\). Suppose
that, for every \(g\in G\) and \(x\in\mathsf X\),
\begin{equation}
 F(gx)=gF(x),\qquad e(gx)=e(x).
 \label{rg:eq:gauge-equivariance}
\end{equation}
Then every finite block is gauge invariant:
\begin{equation}
 e(F^b(gx))=e(F^b(x))
 \qquad (b\in\N).
 \label{rg:eq:gauge-block-null}
\end{equation}
Consequently, a pair related by the consistent action \(x\mapsto gx\) is a
null control for a deterministic fiber experiment, not a closure witness.
\end{proposition}

\begin{proof}
We first prove \(F^b(gx)=gF^b(x)\) by induction on \(b\). At \(b=0\), both
sides equal \(gx\). If the identity holds at \(b\), then
\[
 F^{b+1}(gx)=F(F^b(gx))=F(gF^b(x))=gF(F^b(x))=gF^{b+1}(x),
\]
where the third equality is equivariance. Applying the invariant observable
\(e\) proves Equation~\eqref{rg:eq:gauge-block-null}. The last assertion follows
because every successor defect along such a pair must vanish.
\end{proof}

\begin{corollary}[Paired-orbit identity]
\label{rg:cor:paired-gauge-successor}
Write a complete optimizer state as \(x=(\theta,m,v)\), and let \(g\) be an
involution on the selected parameter and first-moment coordinates. Define
\[
 C_g(\theta,m,v)=(g\theta,gm,v),\qquad
 P_g(\theta,m,v)=(g\theta,m,v),\qquad
 M_g(\theta,m,v)=(\theta,gm,v).
\]
Suppose \(F C_g=C_g F\) and \(e C_g=e\). Then, for every \(b\in\N\),
\begin{equation}
 e\bigl(F^b(P_gx)\bigr)=e\bigl(F^b(M_gx)\bigr).
 \label{rg:eq:paired-gauge-successor}
\end{equation}
Thus the parameter-only branch and the moment-only branch form a paired
physical witness whenever the consistent branch is a valid symmetry control.
\end{corollary}

\begin{proof}
Because \(g^2\) is the identity on the selected coordinates,
\[
 C_g\circ M_g(\theta,m,v)
   =C_g(\theta,gm,v)=(g\theta,g^2m,v)=P_g(\theta,m,v).
\]
We next show directly that \(F^bC_g=C_gF^b\). The assertion is immediate for
\(b=0\). If it holds for \(b\), then
\[
 F^{b+1}C_g=F(F^bC_g)=F(C_gF^b)=C_gF^{b+1},
\]
where the last equality uses \(FC_g=C_gF\). Therefore
\[
 eF^bP_g=eF^bC_gM_g=eC_gF^bM_g=eF^bM_g,
\]
which is Equation~\eqref{rg:eq:paired-gauge-successor}.
\end{proof}

\begin{lemma}[Augmented-fiber witness]
\label{rg:lem:augmented-fiber-witness}
Let \(a:\mathsf X\to\mathsf A\) be any proposed deterministic state
observable, let \(h:\mathsf A\to\mathsf E\) be a readout, and fix \(b\ge1\).
If states \(x,y\) satisfy
\[
 a(x)=a(y),\qquad
 h(a(F^b(x)))\ne h(a(F^b(y))),
\]
then no deterministic scale-\(b\) map on \(a(\mathsf X)\) represents both
successors. In particular, \(a\) is not closed on any domain containing this
pair.
\end{lemma}

\begin{proof}
If a map \(H_b\) closed the proposed state, then
\(a(F^b(z))=H_b(a(z))\) for every state \(z\) in the domain. Hence
\[
 a(F^b(x))=H_b(a(x))=H_b(a(y))=a(F^b(y)).
\]
Applying \(h\) contradicts the assumed unequal readouts.
\end{proof}

For the implemented two-residual row program, let \(g_\ell\) act on
layer \(\ell\in\{0,1,2\}\). It negates both affine parameters of the initial
metric LayerNorm; every GLU input weight and every GLU output weight and
output bias, while leaving the two GLU input biases fixed; the first residual
LayerNorm bias; and the second residual LayerNorm weight. The first
LayerNorm output and the first residual output then change sign, while the
second residual output, hence the row map and its energy, are unchanged.
Each \(g_\ell\) is an involution. Actions on different layers have disjoint
coordinate support and commute, so the generated concrete sign-action group
is
\[
 \langle g_0,g_1,g_2\rangle\simeq(\mathbb Z/2\mathbb Z)^3.
\]
The full action is \(g=g_0g_1g_2\). In the executed checkpoint census, each
layer action covers 20 parameter tensors and 98,816 scalar elements, while
the full action covers 60 tensors and 296,448 elements. These counts describe
the measured implementation support; they make no minimality claim. This is
a parameter gauge, distinct from the endpoint change of energy units used for
the affine cocycle. Section~\ref{rg:sec:matched-fiber-results} gives the measured
support census, consistent-action null controls, and parameter-only tests.

A consistent optimizer action also negates the AdamW first moments attached
to the sign-changed parameters and leaves second moments fixed. Under an
exactly equivariant update, Proposition~\ref{rg:prop:gauge-orbit-null} requires
zero successor defect. A parameter-only action leaves the raw first and
second moments numerically unchanged. It therefore makes the proposed
augmented observation \(a=(e,m,v)\) identical at the source even though the
hidden parameter representative differs. Corollary~\ref{rg:cor:paired-gauge-successor}
shows that this parameter-only branch is exactly paired with the moment-only
branch in the mirrored gauge. Relative to \(x\), the latter applies the
full-support first-moment flip \(m\mapsto-m\), with elementwise dose
\(2|m|\). It is not an independent physical direction. If the paired
successors have unequal energy, Lemma~\ref{rg:lem:augmented-fiber-witness}
rejects closure of raw \((e,m,v)\) on that domain. This does not exclude a
gauge-covariant moment coordinate, a quotient of the optimizer state, or
another sufficient statistic.

There is a separate finite-observation fiber that is not a gauge orbit. Let
\(R_P(\theta)\) denote all row maps returned by a deterministic observer on a
fixed probe \(P\), let \([m,v]_G\) be a chosen gauge-canonical optimizer
coordinate, and let \(\phi\) contain the declared exogenous phase. Consider
the candidate
\[
 a_P(\theta,m,v,\phi)=\bigl(R_P(\theta),[m,v]_G,\phi\bigr).
\]
If a model coordinate is not read in evaluating \(R_P\), changing only that
coordinate produces an exact \(a_P\)-fiber. In particular, for an embedding
lookup, changing a row whose token is absent from every input in \(P\) cannot
change the probe computation, while the optimizer tensors and \(\phi\) can be
held fixed. If a later training batch reads that row and the row-map readout
after \(b\) updates differs, Lemma~\ref{rg:lem:augmented-fiber-witness} rules out
a deterministic scale-\(b\) map on \(a_P\) for every domain containing the
pair. The conclusion is structural: no statistic based only on a fixed finite
probe, optimizer state, and phase is globally sufficient on a domain that
allows dynamically relevant unobserved model coordinates. It does not decide
whether a restriction to an unmodified reachable orbit closes. The executed
probe-null embedding test in Section~\ref{rg:sec:staged-closure-results}
instantiates
this fiber exactly.

\subsection{The renormalized random variable}

Universality requires both a coarse-graining operation and a normalization.
On a positive excursion, the exact block gain is
\begin{equation}
 Q_b=\prod_{t=1}^{b}q_t,
 \qquad
 S_b=\log Q_b=\sum_{t=1}^{b}Y_t,
 \qquad Y_t=\log q_t.
 \label{rg:eq:log-block-variable}
\end{equation}
Conditional on an observer-certified positive excursion and a temporal law,
the natural probability-space RG acts on the law of \(Y_t\) by convolution
followed by centering and scale normalization.  This differs from both the
standard complete-state kernel blocking above and the physical energy gauge
of Section~\ref{rg:sec:exact-rg}.  Kernel powers block a specified transition
law, \(\mathcal T_b\) below compares fluctuation laws, and the endpoint gauge
changes energy units.

Here and below, \(f_b\asymp g_b\) means that their ratio is bounded above
and away from zero on the stated limit; \(f_b\sim g_b\) means the ratio
tends to one.

\begin{definition}[Multiplicative row-collapse universality]
\label{rg:def:multiplicative-universality}
Two row-collapse systems are in the same multiplicative universality class
only on a declared domain satisfying all of the following conditions:
\begin{enumerate}[label=(\roman*)]
\item their log gains are defined on observer-certified positive excursions;
\item the same aligned blocking rule is used, and direct blocks agree with
      iterated blocks in the exact cocycle;
\item either a stationary temporal domain or an explicit nonautonomous local
      or triangular-array limit is specified;
\item the dependence assumptions needed for the proposed limit law hold;
\item centering constants \(a_b\) and positive scales \(c_b\) are fixed
      independently of the sample being judged;
\item the laws of \((S_b-a_b)/c_b\) are tight and converge under the same
      block sequence to the same nondegenerate fixed law;
\item the relevant drift fields have the same scaling dimensions, and the
      limiting behavior persists across held-out seeds and the declared
      architecture family;
\item face restarts are absent, proved irrelevant under the full affine RG,
      or incorporated into a common joint marked-excursion fixed law.
\end{enumerate}
A gain-only class is not a class of the complete row-collapse process unless
the final source condition is closed.
\end{definition}

The definition concerns the quotient observable.  It does not assert that
the complete AdamW state, hidden activations, or PLGA output shares the same
class. Such a transfer needs strong lumpability, an explicit comparison map,
or a joint complete-state fixed-law analysis.

\subsection{Clean multiplicative class}

If \(Y_t\equiv\mu\), then \(S_b=b\mu\) and the law has no fluctuations.
The sign of \(\mu=\log q\) gives the clean phase, and the source-free
critical point is \(\mu=0\).  The physical e-fold scale is
\(\xi_{\rm typ}=1/\abs{\mu}\), giving the exponent one derived in
Theorem~\ref{rg:thm:clean-critical-scaling}.  This zero-variance class is not a
special sample from the normalized Gaussian class because the
square-root-variance normalization is singular at variance zero.

\subsection{Finite-variance temporal disorder}

Suppose first that \(Y_1,Y_2,\ldots\) are independent and identically
distributed with
\begin{equation}
 \E Y_1=\mu,
 \qquad
 \operatorname{Var}(Y_1)=\sigma^2\in(0,\infty).
 \label{rg:eq:finite-variance-assumption}
\end{equation}
Define the law transformation
\begin{equation}
 \mathcal T_b[\mathcal L(Y)]
 =\mathcal L\!\left(
   \frac{\sum_{t=1}^{b}(Y_t-\mu)}{\sigma\sqrt b}
  \right).
 \label{rg:eq:law-rg}
\end{equation}

\begin{theorem}[Gaussian log-gain fixed point]
\label{rg:thm:gaussian-log-gain-fixed-point}
Under Equation~\eqref{rg:eq:finite-variance-assumption}, the classical
Lindeberg--L\'evy central limit theorem \cite{billingsley1995} gives
\begin{equation}
 \mathcal T_b[\mathcal L(Y)]
 \Longrightarrow \mathcal N(0,1)
 \qquad\text{as }b\to\infty.
 \label{rg:eq:gaussian-fixed-point}
\end{equation}
The standard normal law is a fixed point of every \(\mathcal T_b\).  The
standardized drift \(h=\mu/\sigma\) transforms as
\begin{equation}
 h_b=\sqrt b\,h.
 \label{rg:eq:drift-scaling}
\end{equation}
Thus its scaling dimension is \(y_h=1/2\).  Define the
drift-to-fluctuation crossover length by the balance
\(b\abs{\mu}\asymp\sigma\sqrt b\).  It then obeys
\begin{equation}
 \xi_{\rm dis}\asymp\frac{\sigma^2}{\mu^2},
 \qquad \nu_{\rm dis}=2.
 \label{rg:eq:disorder-exponent}
\end{equation}
Here \(\nu_{\rm dis}\) is the derived exponent of that declared crossover
length, not a separately fitted empirical exponent.
\end{theorem}

\begin{proof}
Let \(X_t=(Y_t-\mu)/\sigma\).  The variables \(X_t\) are iid with zero
mean and unit variance.  The Lindeberg--L\'evy theorem therefore gives
\[
 \frac{1}{\sqrt b}\sum_{t=1}^{b}X_t
 \Longrightarrow\mathcal N(0,1),
\]
which is exactly Equation~\eqref{rg:eq:gaussian-fixed-point}.  If \(Y\) is
normal, closure of normal laws under independent sums shows that
Equation~\eqref{rg:eq:law-rg} returns the standard normal law for every \(b\).

Before centering, the block mean is \(b\mu\) and its fluctuation scale is
\(\sigma\sqrt b\).  Their ratio is \(\sqrt b\,\mu/\sigma\), proving
Equation~\eqref{rg:eq:drift-scaling}.  Drift and fluctuation are comparable when
\(b\abs{\mu}\asymp\sigma\sqrt b\), which rearranges to
Equation~\eqref{rg:eq:disorder-exponent}.
\end{proof}

There are two useful time scales near this disordered critical surface.
Under the stated iid assumptions, \(S_b/b\to\mu\) in probability, so fixed
interior quantiles of \(S_b/b\) converge to \(\mu\). The typical energy
e-fold scale is therefore asymptotically \(1/\abs{\mu}\). The larger scale
\(\sigma^2/\mu^2\) is where deterministic drift dominates a typical
square-root fluctuation. Their exponents one and two answer different
questions and should not be conflated.

For a stationary mixing sequence, the same fixed point follows whenever a
central limit theorem holds with positive long-run variance
\begin{equation}
 \sigma_{\rm eff}^2
 =\operatorname{Var}(Y_0)+2\sum_{k=1}^{\infty}
   \operatorname{Cov}(Y_0,Y_k).
 \label{rg:eq:long-run-variance}
\end{equation}
In that case \(\sigma\) is replaced by \(\sigma_{\rm eff}\).  Stationarity,
mixing, summability of correlations, and positivity of this variance are
assumptions to test.  They do not follow from a finite set of block
quantiles.
\subsection{Stationary and nonautonomous probability RG domains}

The convolution RG in Equation~\eqref{rg:eq:law-rg} is exact for independent
copies and specializes, by a central limit theorem, to stated stationary
dependence classes.  Its domain therefore requires a stable law under time
translation, a valid fluctuation normalization, and closure of the affine
source sector.

For changing training dynamics, write
\[
 Y_t=m_t+\varepsilon_t,\qquad
 M_{k,b}=\sum_{t=k}^{k+b-1}m_t,\qquad
 \Xi_{k,b}=\sum_{t=k}^{k+b-1}\varepsilon_t,
\]
so the exact block log gain is \(S_{k,b}=M_{k,b}+\Xi_{k,b}\).  A
nonautonomous probability-space RG must retain the block origin \(k\), the
cumulative drift \(M_{k,b}\), and location-dependent centering and scaling:
\begin{equation}
 \mathcal T_{k,b}
 =\mathcal L\!\left(\frac{S_{k,b}-a_{k,b}}{c_{k,b}}\right),
 \qquad c_{k,b}>0.
 \label{rg:eq:nonautonomous-law-rg}
\end{equation}
The regular family in Proposition~\ref{rg:prop:regular-nonautonomous-drift} is
asymptotically closed in its deterministic drift sector:
\(c\mapsto b^{1-\gamma}c\). Its centered fluctuation sector can still depend
on the block origin and requires a separate limiting-law hypothesis.
A common limiting law can be assigned only along a declared local or
triangular-array limit for which \(a_{k,b}\), \(c_{k,b}\), dependence, and
source behavior are controlled.  When no such stable domain is established,
the exact cumulative-log flow of
Lemma~\ref{rg:lem:cumulative-log-criterion} still applies, but a stationary
universality label does not.

\subsection{Other conditional subclasses}

The exact additive log coordinate also identifies when the Gaussian class is
not appropriate.

The centered fluctuation law is held fixed as the finite mean drift
\(\mu\) varies. Put \(\delta=\abs{\mu}\); comparison constants below
are uniform as \(\delta\downarrow0\). A normalization need not be a
pure power: a slowly varying factor can change a two-sided crossover bound.

\begin{proposition}[Conditional drift-to-fluctuation balance]
\label{rg:prop:universality-taxonomy}
Let \(0<H<1\) and \(c_b=b^H L(b)>0\), with
\begin{equation}
 \frac{\log L(b)}{\log b}\longrightarrow0
 \qquad(b\longrightarrow\infty).
 \label{rg:eq:subpower-normalization}
\end{equation}
Let \(\xi(\delta)\to\infty\) be a crossover scale satisfying
\begin{equation}
 \rho(\delta)=\frac{\delta\xi(\delta)}{c_{\xi(\delta)}},
 \qquad 0<r\le\rho(\delta)\le R<\infty
 \label{rg:eq:crossover-balance}
\end{equation}
for all sufficiently small \(\delta>0\). Then
\begin{equation}
 \frac{\xi(\delta)^{1-H}}{L(\xi(\delta))}\asymp\delta^{-1},
 \qquad
 \lim_{\delta\downarrow0}
 \frac{\log\xi(\delta)}{\log(1/\delta)}=\frac1{1-H}.
 \label{rg:eq:hurst-crossover}
\end{equation}
If, in addition, \(0<a\le L(b)\le A<\infty\) eventually, then
\begin{equation}
 (ar)^{1/(1-H)}\delta^{-1/(1-H)}
 \le\xi(\delta)\le
 (AR)^{1/(1-H)}\delta^{-1/(1-H)}.
 \label{rg:eq:bounded-factor-crossover}
\end{equation}
Thus a bounded factor gives the stronger two-sided comparison
\(\xi(\delta)\asymp\delta^{-1/(1-H)}\).
\end{proposition}

\begin{proof}
Equation~\eqref{rg:eq:crossover-balance} is equivalent to
\(\delta\xi^{1-H}=\rho L(\xi)\), which proves the first comparison.
Positivity permits taking logarithms, giving the exact identity
\begin{equation}
 (1-H)\log\xi
 =\log(1/\delta)+\log L(\xi)+\log\rho.
 \label{rg:eq:log-balance-identity}
\end{equation}
Divide by \(\log\xi>0\), eventually. The second term on the right
tends to zero by Equation~\eqref{rg:eq:subpower-normalization}; the
third tends to zero because \(\log\rho\) is bounded and
\(\xi\to\infty\). Consequently
\(\log(1/\delta)/\log\xi\to1-H>0\). Taking reciprocals proves
the logarithmic exponent in Equation~\eqref{rg:eq:hurst-crossover}.
With the additional bounds on \(L\), the balance gives
\(ar/\delta\le\xi^{1-H}\le AR/\delta\). Raising positive
quantities to \(1/(1-H)>0\) proves
Equation~\eqref{rg:eq:bounded-factor-crossover}.
\end{proof}

The subpower property in Equation~\eqref{rg:eq:subpower-normalization}
is the property of a positive measurable slowly varying normalization used
by this proof. The proposition is conditional on a diverging scale satisfying
the stated balance. It does not assert uniqueness of a finite crossover or
existence of a limiting fluctuation law. Integer block sizes are compatible
with the bounded-ratio formulation.

For iid centered increments in a nondegenerate \(\alpha\)-stable domain of
attraction, \(1<\alpha<2\), the normalization has the form
\(b^{1/\alpha}L(b)\) \cite{gnedenko1954}. Hence the logarithmic
crossover exponent is \(\alpha/(\alpha-1)\). Normal attraction with a
fixed nonzero scale permits asymptotically constant \(L\); an explicit
bound \(L\asymp1\) also suffices for the pure-power comparison.
Normal attraction and general stable attraction differ in this respect
\cite{chen2018stable}. A marginal tail alone does not establish a stable
limit for dependent optimizer increments. Likewise a correlated-sum class
requires a declared normalization \(b^H L(b)\) and the dependence
hypotheses for its limiting law. Finite variance alone supplies neither.

\paragraph{A logarithmic-tail example.}
Fix \(1<\alpha<2\). Let \(X\) be symmetric, with
\begin{equation}
 \Prob(\abs X>x)=x^{-\alpha}\log x,\qquad x\ge e,
 \label{rg:eq:logarithmic-tail}
\end{equation}
and the remaining mass at zero. More explicitly, its magnitude has density
\(x^{-\alpha-1}(\alpha\log x-1)\) on \((e,\infty)\), of total
mass \(e^{-\alpha}\); the atom at zero has mass \(1-e^{-\alpha}\).
Choose either sign equally for each nonzero magnitude. The density is
positive, and \(\int_e^\infty x^{-\alpha}\log x\,dx<\infty\),
so the first absolute moment is finite and \(\E X=0\). The balanced
regularly varying tails place iid copies in a stable domain of attraction.

Choose the diverging tail-quantile scale with
\(b c_b^{-\alpha}\log c_b\to1\). Then
\(\alpha\log c_b=\log b+\log\log c_b+o(1)\).
Dividing first by \(\log c_b\) gives
\(\log b/\log c_b\to\alpha\). Substitution back into the balance yields
\begin{equation}
 c_b\asymp(b\log b)^{1/\alpha},\qquad
 \frac{c_b}{b^{1/\alpha}}\longrightarrow\infty.
 \label{rg:eq:logarithmic-tail-normalization}
\end{equation}
Adding a drift \(\mu\) leaves the centered law fixed. Its crossover obeys
\(\xi^{\alpha-1}\delta^\alpha\asymp\log\xi\).
Equation~\eqref{rg:eq:hurst-crossover} gives
\(\log\xi\asymp\log(1/\delta)\); substitution now proves
\begin{equation}
 \xi(\delta)\asymp
 \delta^{-\alpha/(\alpha-1)}
 [\log(1/\delta)]^{1/(\alpha-1)}.
 \label{rg:eq:logarithmic-tail-crossover}
\end{equation}
The logarithmic multiplier is unbounded. Thus general stable attraction
has the stated logarithmic exponent but need not have a two-sided pure-power
crossover bound. At \(\alpha=3/2\) the power exponent is three and the
logarithmic power is two. This example is an analytic obstruction to the
stronger bound, not an empirical law fitted to training data.

If \(\alpha\le1\), a finite mean log gain need not exist, so the critical
coordinate \(\lambda=\E Y\) is unavailable. Quantile or stable-location
flows must then replace the mean-drift model. Nonstationary training retains
the block origin, cumulative drift, and conditional normalization of
Equation~\eqref{rg:eq:nonautonomous-law-rg}. A finite consuming corpus bounds
the accessible block length. An asymptotic crossover needs a separately
specified limiting family; a scale beyond the available corpus is not an
observed training regime. Section~\ref{model:sec:data-resource} derives
the finite-source restriction, and Section~\ref{model:sec:joint-scaling}
specifies the joint size, horizon, memory, and consumption limits.

\begin{table}[t]
\centering\small
\caption{Conditional positive-excursion taxonomy. Here \(\delta=\abs\mu\),
\(\nu_{\log}=\lim\log\xi/\log(1/\delta)\). Stable rows use
\(1<\alpha<2\), and \(L\) satisfies
Equation~\eqref{rg:eq:subpower-normalization}. The last column distinguishes
a two-sided comparison from a logarithmic exponent.}
\label{rg:tab:universality-classes}
\begin{tabularx}{\textwidth}{@{}>{\raggedright\arraybackslash}p{.23\textwidth}
>{\raggedright\arraybackslash}p{.20\textwidth}
>{\raggedright\arraybackslash}p{.19\textwidth}X@{}}
\toprule
Conditional class & Scale \(c_b\) & Drift factor \(b/c_b\) & Crossover \\
\midrule
Nondegenerate CLT & \(\sigma\sqrt b\), fixed \(\sigma>0\) &
 \(\sqrt b/\sigma\) & \(\xi\asymp\delta^{-2}\) \\[4pt]
Stable normal attraction or bounded factor & \(b^{1/\alpha}L(b)\),
 \(L\asymp1\) & \(b^{1-1/\alpha}/L(b)\) &
 \(\xi\asymp\delta^{-\alpha/(\alpha-1)}\) \\[4pt]
General stable attraction & \(b^{1/\alpha}L(b)\) &
 \(b^{1-1/\alpha}/L(b)\) & \(\nu_{\log}=\alpha/(\alpha-1)\);
 balance retains \(L\) \\[4pt]
Correlated sums with declared scale & \(b^H L(b)\), \(0<H<1\) &
 \(b^{1-H}/L(b)\) & \(\nu_{\log}=1/(1-H)\);
 pure-power bound if \(L\asymp1\) \\[4pt]
Deterministic clean evolution & No fluctuations & Not applicable &
 Energy e-fold length \(\xi_{\rm typ}=\delta^{-1}\) \\
\bottomrule
\end{tabularx}
\end{table}

\subsection{Reopening is a relevant affine field}

The preceding fixed laws concern positive excursions with \(\zeta_t=0\).
At critical clean gain, a homogeneous source transforms as
\(\zeta\mapsto b\zeta\), so it has scaling dimension one and is more relevant
than the standardized random drift, which has dimension one half.  A single
exact face restart ends the positive-excursion logarithmic description and
sets the homogeneous scalar coefficient to zero. The later sources and gains
can still depend on optimizer moments, corpus position, and hidden state.
Therefore a full universality claim requires one of the following:
\begin{enumerate}[label=(\alph*)]
\item exact face invariance after some scale;
\item a proof that the transported restart sector vanishes under blocking;
\item a joint affine fixed law for gain and source rather than a gain-only
  law.
\end{enumerate}
Without one of these closures, Gaussianization of log gains is informative
about positive excursions but not sufficient for row-map collapse.

\subsection{Recurrent-face affine classes}

When faces recur, the gain-only law is incomplete. Consecutive face hits have
\(r_j=P_j=M_j=0\) and are zero-peak events without logarithmic marks. For
every positive-restart excursion, a natural mark is
\begin{equation}
 \left(\sigma_{j+1}-\sigma_j,\rho_j,a_j\right)
 =\left(\sigma_{j+1}-\sigma_j,-\log r_j,\log M_j\right).
 \label{rg:eq:marked-excursion}
\end{equation}
Blocking consecutive excursions must retain both the timing mark and the peak
coordinate \(a_j-\rho_j=\log P_j\). A recurrent-face universality class
would therefore be a fixed law of this marked process under a stated
centering, scale normalization, and excursion-blocking rule. The scalar
critical ratio \(\chi_{\rm exc}\) from
Proposition~\ref{rg:prop:restart-amplification-criticality} identifies its
leading relevant direction when that ratio exists, but it does not identify a
limiting law. No recurrent-face fixed distribution is asserted without
assumptions on the restart and amplification process.

\subsection{What identifies an empirical class}

For PLDR trajectories, empirical identification therefore requires raw
consecutive energies; certified positive observer intervals; direct and
iterated aligned blocks; a stable temporal domain; explicit dependence tests;
prespecified centering and normalization; evidence of tightness and
convergence across increasing scales; held-out seeds and architectures; and
either source irrelevance or a fitted marked-reopening law. Failure of any
one gate prevents the class label. Persistent projection dependence,
nonstationarity, incompatible limiting laws, scale-dependent exponents, or a
nonvanishing source field refutes the corresponding claim on its measured
domain. Section~\ref{rg:sec:temporal-flow-results} reports the executed
temporal-domain and law tests; Section~\ref{rg:sec:observer-results} gives
their numerical branch limits. The controlled finite-variance examples in
Section~\ref{rg:sec:synthetic-qualification} qualify the reducer separately.

\FloatBarrier
\par\medskip\noindent
Chapter~\ref{ch:closure-evidence} tests the proposed reduced states and
observer conventions on executed computations. The tests determine which
finite predictive claims survive the closure requirements established here.

\chapter{Executed closure and observer tests}
\label{ch:closure-evidence}
This chapter reports computational tests of chronological renormalization,
numerical observers and predictive closure. Matched-fiber interventions and
staged state tests distinguish exact algebraic reconstruction from the
ability of a reduced state to predict an unseen successor.

\setcounter{topnumber}{3}
\section{Executed computational results}
\label{rg:sec:executed-results}

The registered long comparison comprises four source trajectories and twelve
trajectory-layer cells, conditional on their recorded training data and fixed
observation registry. Heads, contexts and time windows are nested measurements.
Ten complete positive-domain cells fail the nominal precision gate; two meet
face interruptions. Neither counting more maps nor pooling those regimes
supplies additional independent trajectories or selects a universality class.

The computations have six distinct roles. Synthetic experiments qualify the
algebra and finite-variance reducer under controlled assumptions. A
content-addressed retained summary describes captured PLDR trajectories
without treating overlapping comparisons as independent. The registered
four-trajectory run is the first full execution of the frozen cellwise
decision system and is also a design-precision record for its declared
nominal confidence-width target. Finally, exact-input and observer-regime analysis,
together with a stage-resolved CPU/two-GPU replay of the four protocol-defined saved steps for
each trajectory, determines what the recorded zeros and finite precision
support. A source-remeasured suite of matched optimizer-state interventions
tests whether the full row-map energy vector closes as a deterministic
predictive state. A gated four-trajectory stage then tests three successively
richer reduced states, held-out conditional equality, block scale 16, and
direct-versus-iterated continuation. No synthetic result is counted as PLDR
evidence.
Sections~\ref{rg:sec:synthetic-qualification}, \ref{rg:sec:long-design} and \ref{rg:sec:observer-results} give the qualification, registered design and observer analysis; Sections~\ref{rg:sec:matched-fiber-results} and \ref{rg:sec:staged-closure-results} detail the two closure tests.

\newcommand{\RGSourceRGSyntheticTrials}{20,000}
\newcommand{\RGSourceRGSyntheticLongEdges}{4,096}
\newcommand{\RGSourceRGAlgebraResidual}{\ensuremath{9.114\times10^{-15}}}
\newcommand{\RGSourceRGCriticalExponentFit}{1.00337}
\newcommand{\RGSourceRGBetaResidual}{\ensuremath{7.025\times10^{-7}}}
\newcommand{\RGSourceRGSyntheticReplicas}{30,000}
\newcommand{\RGSourceRGFinalKSDistance}{0.02834}
\newcommand{\RGSourceRGFinalCrossDistance}{0.03309}
\begin{table}[b]
\centering
\caption{Synthetic finite-variance log-gain blocking. The Kolmogorov distance is to the standard normal law; skewness and excess kurtosis are reported at block size 256.}
\label{rg:tab:synthetic-fixed-point}
\begin{tabular}{lrrrr}
\toprule
Increment law & $D_{\rm KS}(1)$ & $D_{\rm KS}(256)$ & skewness & excess kurtosis \\ 
\midrule
Gaussian & 0.0045 & 0.0077 & -0.0161 & -0.0343 \\
Laplace & 0.0651 & 0.0045 & 0.0147 & 0.0039 \\
Uniform & 0.0561 & 0.0074 & -0.0172 & -0.0058 \\
Rademacher & 0.3433 & 0.0283 & 0.0054 & -0.0266 \\
\bottomrule
\end{tabular}
\end{table}

\subsection{Deterministic and stochastic qualification}
\label{rg:sec:synthetic-qualification}

The affine implementation was exercised on \RGSourceRGSyntheticTrials{} random
triples of nonnegative edges. Sequential action, both association orders, and
endpoint gauge normalization were compared with direct blocking. A separate
path contained \RGSourceRGSyntheticLongEdges{} chronological edges with nonzero
sources. The largest relative discrepancy was
\(\RGSourceRGAlgebraResidual{}\), consistent with floating-point roundoff.

For clean gains \(q=1-\tau\), 17 logarithmically spaced values
\(10^{-5}\le\tau\le0.1\) gave
\begin{equation}
 \widehat\nu_{\parallel}=\RGSourceRGCriticalExponentFit{}.
 \label{rg:eq:synthetic-clean-exponent}
\end{equation}
Finite differences of the continuous block flow at six gains agreed with
\(q\log q\) to maximum relative discrepancy \(\RGSourceRGBetaResidual{}\).
These are implementation checks of exact formulas, not fitted PLDR
exponents.

The controlled law experiment used Gaussian, Laplace, uniform, and
Rademacher increments with common standard deviation \(0.04\).
For each law, \RGSourceRGSyntheticReplicas{} independent nested blocks were
normalized by \(\sigma\sqrt b\) through \(b=256\). At the final block,
the largest non-Gaussian Kolmogorov distance to the standard normal was
\(\RGSourceRGFinalKSDistance{}\), and the largest pairwise empirical Wasserstein
distance was \(\RGSourceRGFinalCrossDistance{}\). This qualifies the reducer on
known finite-variance independent inputs. It does not establish stationarity,
mixing, or a domain of attraction for PLDR training.

\subsection{Guarded reduction of retained trajectories}

\newcommand{\RGSourceRGObservedComparisons}{1,529,280}
\newcommand{\RGSourceRGObservedMaps}{5,184}
\newcommand{\RGSourceRGExactFaceMaps}{0}
\newcommand{\RGSourceRGFloorCensoredMaps}{576}
\newcommand{\RGSourceRGEffectFloor}{\ensuremath{1\times10^{-12}}}
\newcommand{\RGSourceRGMedianRate}{\ensuremath{-2.550163\times10^{-4}}}
\newcommand{\RGSourceRGMedianEFold}{3,921}
\newcommand{\RGSourceRGMedianFitRMS}{\ensuremath{3.032\times10^{-4}}}
\newcommand{\RGSourceRGMedianFitMax}{\ensuremath{5.317\times10^{-4}}}
\newcommand{\RGSourceRGMedianFitRSquared}{0.98683}
\newcommand{\RGSourceRGMedianPowerExponent}{0.938}
\newcommand{\RGSourceRGRateWeakening}{21.3\%}
\newcommand{\RGSourceRGStrongestRateWeakening}{22.5\%}
\newcommand{\RGSourceRGTenOrderCells}{4}
\newcommand{\RGSourceRGIncreasingCells}{0}
\newcommand{\RGSourceRGScaleOutcome}{\textsc{aggregate medians below one with heterogeneous cell flow}}
\begin{table}[t]
\centering\scriptsize
\caption{Descriptive PLDR block-gain census. Counts overlap and are not independent observations. The last column removes comparisons with a floor-censored endpoint.}
\label{rg:tab:retained-scale-flow}
\begin{tabular}{rrrrrrrrr}
\toprule
$b$ & count & median & p05 & p95 & max. & $Q_b<1$ & censored & resolved median \\
\midrule
1 & 304,128 & 0.99969059 & 0.96657 & 1.02309 & 16.6 & 56.68\% & 9,216 & 0.99968702 \\
2 & 298,944 & 0.99937153 & 0.93692 & 1.04044 & 86.2 & 56.98\% & 8,928 & 0.99937066 \\
4 & 288,576 & 0.99879113 & 0.88765 & 1.06704 & 358.9 & 56.97\% & 8,352 & 0.99880731 \\
8 & 267,840 & 0.99768043 & 0.80353 & 1.10758 & 5490.7 & 56.96\% & 7,200 & 0.99769767 \\
16 & 226,368 & 0.99539861 & 0.67510 & 1.15127 & 11290.6 & 57.78\% & 4,896 & 0.99548324 \\
32 & 143,424 & 0.99223896 & 0.51175 & 1.18524 & 5826.8 & 59.31\% & 288 & 0.99220674 \\
\bottomrule
\end{tabular}
\end{table}

\begin{table}[t]
\centering\small
\caption{Excursion and window heterogeneity in the descriptive PLDR summary. Expansion fractions and maxima are ranges across the 18 captured windows.}
\label{rg:tab:retained-excursions}
\begin{tabular}{rrrrrr}
\toprule
$b$ & excursion median & excursion p95 & excursion max. & window max. range & expanding fraction range \\
\midrule
1 & 0 & 0.02309 & 15.6 & 1.01--16.6 & 0.218--0.578 \\
2 & 0 & 0.04798 & 85.2 & 1.01--86.2 & 0.206--0.574 \\
4 & 0.0002288 & 0.09458 & 357.9 & 1.02--359 & 0.196--0.605 \\
8 & 0.001492 & 0.1632 & 5489.7 & 1.04--\ensuremath{5.49\times10^{3}} & 0.176--0.612 \\
16 & 0.004746 & 0.26 & 11289.6 & 1.06--\ensuremath{1.13\times10^{4}} & 0.111--0.612 \\
32 & 0.009031 & 0.3278 & 13909.7 & 1.05--\ensuremath{5.83\times10^{3}} & 0.018--0.663 \\
\bottomrule
\end{tabular}
\end{table}

\begin{table}[t]
\centering\scriptsize
\caption{Trajectory-layer energy envelopes at the first and last retained checkpoints, pooling retained and holdout registries only within each cell. The change interval is in base-10 logarithmic units.}
\label{rg:tab:retained-cell-flow}
\begin{tabular}{@{}rrllll@{}}
\toprule
trajectory & layer & first energy range & last energy range & log$_{10}$ change bounds & classification \\
\midrule
A & 0 & $[\ensuremath{83.3},\ensuremath{1.13\times10^{2}}]$ & $[\ensuremath{8.58\times10^{-15}},\ensuremath{2.25\times10^{-14}}]$ & $[-16.12,-15.57]$ & decreased envelope \\
A & 1 & $[\ensuremath{3.01\times10^{-4}},\ensuremath{5.09\times10^{-4}}]$ & $[\ensuremath{3.94\times10^{-16}},\ensuremath{1.53\times10^{-15}}]$ & $[-12.11,-11.29]$ & decreased envelope \\
A & 2 & $[\ensuremath{1.35\times10^{2}},\ensuremath{3.08\times10^{2}}]$ & $[\ensuremath{25.4},\ensuremath{79.7}]$ & $[-1.08,-0.23]$ & decreased envelope \\
B & 0 & $[\ensuremath{8.59\times10^{-3}},\ensuremath{1.45\times10^{-2}}]$ & $[\ensuremath{4.42\times10^{-12}},\ensuremath{2.81\times10^{-11}}]$ & $[-9.51,-8.48]$ & decreased envelope \\
B & 1 & $[\ensuremath{1.02\times10^{-3}},\ensuremath{1.46\times10^{-3}}]$ & $[\ensuremath{4.60\times10^{-5}},\ensuremath{2.59\times10^{-4}}]$ & $[-1.50,-0.60]$ & decreased envelope \\
B & 2 & $[\ensuremath{1.60\times10^{2}},\ensuremath{3.69\times10^{2}}]$ & $[\ensuremath{4.42\times10^{-12}},\ensuremath{1.84\times10^{-11}}]$ & $[-13.92,-12.94]$ & decreased envelope \\
C & 0 & $[\ensuremath{0.157},\ensuremath{0.159}]$ & $[\ensuremath{3.75\times10^{-15}},\ensuremath{6.31\times10^{-15}}]$ & $[-13.63,-13.40]$ & decreased envelope \\
C & 1 & $[\ensuremath{2.77\times10^{2}},\ensuremath{3.80\times10^{2}}]$ & $[\ensuremath{3.44\times10^{2}},\ensuremath{8.62\times10^{2}}]$ & $[-0.04,0.49]$ & overlapping envelopes \\
C & 2 & $[\ensuremath{2.79\times10^{2}},\ensuremath{5.33\times10^{2}}]$ & $[\ensuremath{16.3},\ensuremath{49.9}]$ & $[-1.51,-0.75]$ & decreased envelope \\
\bottomrule
\end{tabular}
\end{table}

The retained input is a content-addressed observer and block summary from
three PLDR trajectories. It contains 18 dense windows and horizons
\(b\in\{1,2,4,8,16,32\}\). The windows and endpoint comparisons overlap,
so the \RGSourceRGObservedComparisons{} comparisons are descriptive populations of
captured records. No independence-based standard error or sampling
\(p\)-value is assigned.

For each horizon, the endpoint gain is \(E_{t+b}/E_t\) on a mathematically
positive source. The aggregate median fit
\begin{equation}
 \log\operatorname{median}(Q_b)=\lambda_{\rm med}b+\varepsilon_b
 \label{rg:eq:median-flow-fit}
\end{equation}
gives
\begin{equation}
 \lambda_{\rm med}=\RGSourceRGMedianRate{}\ \text{per update},\qquad
 \xi_{\rm med}=\RGSourceRGMedianEFold{}\ \text{updates}.
 \label{rg:eq:retained-typical-rate}
\end{equation}
This is not a Lyapunov estimate or an exact semigroup test: a median of
products is not generally a product of medians, and the compact input does
not contain the complete aligned edge paths.

All six aggregate medians are below one, but the per-update rate weakens by
\RGSourceRGRateWeakening{} from \(b=1\) to \(b=32\), and by
\RGSourceRGStrongestRateWeakening{} from its strongest value at \(b=2\) to
\(b=32\). The through-origin residual signs are negative at the first five
scales and positive at the last; the fitted magnitude scales with exponent
\(\RGSourceRGMedianPowerExponent{}\), rather than the exactly homogeneous value
one. The fit has \(R^2=\RGSourceRGMedianFitRSquared{}\), root-mean-square residual
\(\RGSourceRGMedianFitRMS{}\), and maximum absolute residual \(\RGSourceRGMedianFitMax{}\).
Every 5th-to-95th percentile interval straddles one, and every captured
window contains expansion at every horizon.

The trajectory-layer energy envelopes make the mixture explicit. Eight of
nine cells have a final envelope strictly below their initial envelope; one
cell has overlapping envelopes; no cell is guaranteed by the recorded bounds
to increase. Exactly \(\RGSourceRGIncreasingCells{}\) cells are guaranteed to
increase, while \(\RGSourceRGTenOrderCells{}\) cells, not the pooled count, are
guaranteed to decrease by at least ten orders of magnitude. Thus the supported
statement is \RGSourceRGScaleOutcome{}, not uniform collapse of every cell.

The observer census contains \RGSourceRGObservedMaps{} maps, of which
\RGSourceRGExactFaceMaps{} are exact zeros and \RGSourceRGFloorCensoredMaps{} are positive
but at or below the registered effect floor \(\RGSourceRGEffectFloor{}\). The
retained record therefore exercises only positive canonical edges. Its
maximum relative excursion over a block \([t,t+b]\) is defined as
\(\max_{0\le s\le b}E_{t+s}/E_t-1\). It measures the largest intervening
rise above the source energy and is not the endpoint gain
\(E_{t+b}/E_t\). Its largest retained value is
\(1.39097\times10^4\), so a typical
contracting statistic coexists with large finite amplification.

\subsection{Registered four-trajectory PLDR experiment}

\newcommand{\RGSourceConfirmResultDigest}{\hashpair{f2e66e483d2b03ee8ea268709989379f}{ca2e6e3362c6e7c455b7a1f41d739f69}}
\newcommand{\RGSourceConfirmCodeCommit}{\hashpair{b205ec386437bb8bdf93}{82c0e35a14049b0f3025}}
\newcommand{\RGSourceConfirmTrajectories}{4}
\newcommand{\RGSourceConfirmUpdates}{196{,}608}
\newcommand{\RGSourceConfirmMapsPerTrajectory}{48}
\newcommand{\RGSourceConfirmCells}{12}
\newcommand{\RGSourceConfirmCompleteCells}{10}
\newcommand{\RGSourceConfirmNegativeDriftCells}{0}
\newcommand{\RGSourceConfirmPositiveDriftCells}{0}
\newcommand{\RGSourceConfirmUnresolvedCells}{0}
\newcommand{\RGSourceConfirmPrecisionTargetUnmetCells}{10}
\newcommand{\RGSourceConfirmFaceInterruptedCells}{2}
\newcommand{\RGSourceConfirmNonautonomousCells}{2}
\newcommand{\RGSourceConfirmPrecisionGateCells}{0}
\newcommand{\RGSourceConfirmStableCells}{8}
\newcommand{\RGSourceConfirmGaussianNonrejectedCells}{4}
\newcommand{\RGSourceConfirmScalingCells}{10}
\newcommand{\RGSourceConfirmFiniteScaleCells}{0}
\newcommand{\RGSourceConfirmLawRejections}{11}
\newcommand{\RGSourceConfirmLawTests}{30}
\newcommand{\RGSourceConfirmRegularEligibleCells}{2}
\newcommand{\RGSourceConfirmRegularPredictions}{0}
\newcommand{\RGSourceConfirmRecordedStates}{37{,}748{,}928}
\newcommand{\RGSourceConfirmFaceStates}{528{,}575}
\newcommand{\RGSourceConfirmFaceClosures}{52{,}111}
\newcommand{\RGSourceConfirmRestarts}{52{,}091}
\newcommand{\RGSourceConfirmRightCensoredExcursions}{0}
\newcommand{\RGSourceConfirmMaximumRestartAmplification}{6.872\times10^{10}}
\newcommand{\RGSourceConfirmMaximumExcursionPeak}{2.403\times10^{-13}}
\newcommand{\RGSourceConfirmLargeIncrementEvents}{1}
\newcommand{\RGSourceConfirmLargeIncrementClusters}{1}
\newcommand{\RGSourceConfirmMinimumEnergy}{0.000}
\newcommand{\RGSourceConfirmSemigroupResidual}{2.218\times10^{-16}}
\newcommand{\RGSourceConfirmCumulativeResidual}{1.039\times10^{-15}}
\newcommand{\RGSourceConfirmGateShapeResidual}{8.941\times10^{-8}}
\newcommand{\RGSourceConfirmGateShapeAbsoluteResidual}{1.728\times10^{-4}}
\newcommand{\RGSourceConfirmGateShapeSymmetricResidual}{1.000}
\newcommand{\RGSourceConfirmInitialPLGADefectMedian}{11.38}
\newcommand{\RGSourceConfirmFinalPLGADefectMedian}{49.87}
\newcommand{\RGSourceConfirmSampledOperatorMaximum}{5.661\times10^{2}}
\newcommand{\RGSourceConfirmControlDefectMaximum}{8.468\times10^{-15}}
\newcommand{\RGSourceConfirmDecompositionResidual}{1.389\times10^{-16}}

\begin{table}[t]
\centering
\caption{Registered cellwise flow decision. The projected simultaneous
nominal confidence half-width uses only design-segment variance and the planned
holdout size; it is a precision forecast, not a power calculation. The
reported holdout bound is
the simultaneous one-sided upper bound for a design-selected negative direction
and the lower bound for a selected positive direction.}
\label{rg:tab:confirmation-flow}
\scriptsize
\begin{tabular}{@{}ccrrrrcc@{}}
\toprule
Traj. & Layer & Design mean & Projected $h$ & Holdout mean & Dir. bound & Precision gate & Flow \\
\midrule
T0 & 0 & \multicolumn{6}{c}{face interrupted} \\
T0 & 1 & $-2.428\times10^{-4}$ & $6.240\times10^{-4}$ & $-2.035\times10^{-5}$ & $4.014\times10^{-4}$ & no & precision target unmet \\
T0 & 2 & $-2.407\times10^{-4}$ & $3.959\times10^{-4}$ & $6.520\times10^{-6}$ & $3.377\times10^{-4}$ & no & precision target unmet \\
T1 & 0 & $-2.649\times10^{-4}$ & $6.453\times10^{-4}$ & $8.431\times10^{-5}$ & $3.746\times10^{-4}$ & no & precision target unmet \\
T1 & 1 & $-7.760\times10^{-5}$ & $4.997\times10^{-4}$ & $-3.049\times10^{-5}$ & $4.188\times10^{-4}$ & no & precision target unmet \\
T1 & 2 & $-1.488\times10^{-4}$ & $2.872\times10^{-4}$ & $1.396\times10^{-5}$ & $3.936\times10^{-4}$ & no & precision target unmet \\
T2 & 0 & \multicolumn{6}{c}{face interrupted} \\
T2 & 1 & $-2.092\times10^{-4}$ & $4.963\times10^{-4}$ & $-2.829\times10^{-5}$ & $4.473\times10^{-4}$ & no & precision target unmet \\
T2 & 2 & $-3.854\times10^{-6}$ & $1.987\times10^{-4}$ & $-1.184\times10^{-5}$ & $1.474\times10^{-4}$ & no & precision target unmet \\
T3 & 0 & $-8.685\times10^{-5}$ & $2.479\times10^{-4}$ & $-2.242\times10^{-5}$ & $2.994\times10^{-4}$ & no & precision target unmet \\
T3 & 1 & $-1.532\times10^{-4}$ & $4.555\times10^{-4}$ & $1.856\times10^{-5}$ & $4.416\times10^{-4}$ & no & precision target unmet \\
T3 & 2 & $-2.950\times10^{-5}$ & $1.147\times10^{-4}$ & $6.430\times10^{-6}$ & $1.077\times10^{-4}$ & no & precision target unmet \\
\bottomrule
\end{tabular}
\end{table}

\begin{table}[t]
\centering
\caption{Cellwise energy transients, per-map holdout signs, and update-level
holdout excess kurtosis. Energies are medians over the 16 maps in a cell;
signs are negative/zero/positive counts of per-map mean log gain.}
\label{rg:tab:confirmation-transients}
\scriptsize
\begin{tabular}{@{}ccrrrrcc@{}}
\toprule
Traj. & Layer & $E_0$ & $E_{32{,}768}$ & $E_{65{,}536}$ & $E_{196{,}608}$ & Signs & Kurtosis \\
\midrule
T0 & 0 & $4.059\times10^{3}$ & $1.090$ & $3.623\times10^{-2}$ & $7.681\times10^{-3}$ & n/a & n/a \\
T0 & 1 & $4.059\times10^{3}$ & $0.5236$ & $1.948\times10^{-4}$ & $1.265\times10^{-5}$ & 16/0/0 & 5.90 \\
T0 & 2 & $4.062\times10^{3}$ & $0.2460$ & $7.736\times10^{-5}$ & $2.023\times10^{-4}$ & 3/0/13 & 4.50 \\
T1 & 0 & $4.053\times10^{3}$ & $2.645\times10^{-2}$ & $7.148\times10^{-6}$ & $0.4953$ & 0/0/16 & 151.45 \\
T1 & 1 & $4.040\times10^{3}$ & $7.678\times10^{-2}$ & $7.009\times10^{-3}$ & $1.116\times10^{-4}$ & 16/0/0 & 4.11 \\
T1 & 2 & $4.053\times10^{3}$ & $0.1413$ & $1.143\times10^{-3}$ & $6.331\times10^{-3}$ & 0/0/16 & 3.00 \\
T2 & 0 & $4.042\times10^{3}$ & $0.4061$ & $1.051\times10^{-2}$ & $0.000$ & n/a & n/a \\
T2 & 1 & $4.045\times10^{3}$ & $0.5496$ & $5.630\times10^{-4}$ & $1.299\times10^{-5}$ & 16/0/0 & 10.97 \\
T2 & 2 & $4.038\times10^{3}$ & $0.2684$ & $0.2024$ & $3.659\times10^{-2}$ & 16/0/0 & 1.16 \\
T3 & 0 & $4.057\times10^{3}$ & $1.042$ & $6.226\times10^{-2}$ & $3.761\times10^{-3}$ & 16/0/0 & 29.75 \\
T3 & 1 & $4.058\times10^{3}$ & $0.3951$ & $2.343\times10^{-3}$ & $3.074\times10^{-2}$ & 0/0/16 & 10.78 \\
T3 & 2 & $4.041\times10^{3}$ & $0.5060$ & $0.1390$ & $0.3556$ & 0/0/16 & 2.10 \\
\bottomrule
\end{tabular}
\end{table}

\begin{table}[t]
\centering
\caption{Registered layer-1 large-increment ledger.
An event has $|Y_t|\geq 1.0$;
events no more than 128 updates apart form a cluster.}
\label{rg:tab:confirmation-large-increments}
\scriptsize
\begin{tabular}{@{}crrrrrrr@{}}
\toprule
Traj. & Burn-in & Design & Holdout & Total & Clusters & Maps & Max. $|Y_t|$ \\
\midrule
T0 & 0 & 0 & 0 & 0 & 0 & 0 & n/a \\
T1 & 0 & 0 & 0 & 0 & 0 & 0 & n/a \\
T2 & 0 & 0 & 0 & 0 & 0 & 0 & n/a \\
T3 & 1 & 0 & 0 & 1 & 1 & 1 & $4.525$ \\
\bottomrule
\end{tabular}
\end{table}

\begin{table}[t]
\centering
\caption{Registered temporal-domain and finite-scale law diagnostics. KS is
number of Holm-rejected matched-normal tests among the three registered scales.}
\label{rg:tab:confirmation-law}
\scriptsize
\begin{tabular}{@{}ccrrcccc@{}}
\toprule
Traj. & Layer & Batch $\rho_1$ & Holdout var. exp. & Stable & KS rej. & FV scale & FV consistent \\
\midrule
T0 & 0 & \multicolumn{6}{c}{face interrupted} \\
T0 & 1 & -0.24 & 0.88 & yes & 1 & yes & no \\
T0 & 2 & -0.14 & 0.82 & no & 1 & yes & no \\
T1 & 0 & -0.14 & 0.99 & yes & 3 & yes & no \\
T1 & 1 & -0.25 & 0.91 & yes & 0 & yes & no \\
T1 & 2 & -0.22 & 0.93 & yes & 0 & yes & no \\
T2 & 0 & \multicolumn{6}{c}{face interrupted} \\
T2 & 1 & -0.16 & 0.86 & yes & 1 & yes & no \\
T2 & 2 & -0.25 & 0.81 & yes & 0 & yes & no \\
T3 & 0 & -0.09 & 0.96 & yes & 3 & yes & no \\
T3 & 1 & -0.23 & 0.91 & no & 2 & yes & no \\
T3 & 2 & -0.18 & 0.89 & yes & 0 & yes & no \\
\bottomrule
\end{tabular}
\end{table}

\begin{table}[t]
\centering
\caption{Post-run regular-drift diagnostic for cells passing the
uniform-negative design-window sign screen. Fit-quality and exponent
gates are evaluated before any held-out prediction is confirmatory.}
\label{rg:tab:confirmation-drift}
\scriptsize
\begin{tabular}{@{}cccrrrrrc@{}}
\toprule
Traj. & Layer & Sign screen & $\gamma$ & Design $R^2$ & Pred. sum & Obs. sum & Rel. error & Pass \\
\midrule
T0 & 0 & no & \multicolumn{6}{c}{not eligible} \\
T0 & 1 & no & \multicolumn{6}{c}{not eligible} \\
T0 & 2 & no & \multicolumn{6}{c}{not eligible} \\
T1 & 0 & no & \multicolumn{6}{c}{not eligible} \\
T1 & 1 & no & \multicolumn{6}{c}{not eligible} \\
T1 & 2 & yes & 0.33 & 0.03 & $-13.49$ & $1.830$ & 1.14 & no \\
T2 & 0 & no & \multicolumn{6}{c}{not eligible} \\
T2 & 1 & no & \multicolumn{6}{c}{not eligible} \\
T2 & 2 & no & \multicolumn{6}{c}{not eligible} \\
T3 & 0 & no & \multicolumn{6}{c}{not eligible} \\
T3 & 1 & yes & 3.20 & 1.00 & $-1.297$ & $2.433$ & 2.88 & no \\
T3 & 2 & no & \multicolumn{6}{c}{not eligible} \\
\bottomrule
\end{tabular}
\end{table}

\begin{table}[t]
\centering
\caption{Aligned affine block census in recorded arithmetic. Gain quantiles
use only blocks without a represented zero and are descriptive over maps and trajectories.}
\label{rg:tab:confirmation-blocks}
\small
\begin{tabular}{@{}rrrrrrr@{}}
\toprule
$b$ & Blocks & Face blocks & Source blocks & Median & 5th pct. & 95th pct. \\
\midrule
1 & 37{,}748{,}736 & 580{,}666 & 52{,}091 & 0.999786 & 0.976017 & 1.024575 \\
2 & 18{,}874{,}368 & 301{,}224 & 36{,}792 & 0.999570 & 0.954269 & 1.047969 \\
4 & 9{,}437{,}184 & 155{,}727 & 23{,}645 & 0.999133 & 0.914525 & 1.093410 \\
8 & 4{,}718{,}592 & 79{,}979 & 13{,}800 & 0.998258 & 0.847721 & 1.179741 \\
16 & 2{,}359{,}296 & 40{,}859 & 7{,}783 & 0.996438 & 0.750869 & 1.333957 \\
32 & 1{,}179{,}648 & 20{,}785 & 4{,}215 & 0.992620 & 0.628317 & 1.591156 \\
64 & 589{,}824 & 10{,}537 & 2{,}193 & 0.985561 & 0.499294 & 1.990601 \\
128 & 294{,}912 & 5{,}328 & 1{,}187 & 0.972214 & 0.374124 & 2.674250 \\
256 & 147{,}456 & 2{,}691 & 620 & 0.944733 & 0.264839 & 3.780648 \\
512 & 73{,}728 & 1{,}360 & 316 & 0.898441 & 0.182985 & 5.563342 \\
\bottomrule
\end{tabular}
\end{table}

\begin{table}[t]
\centering
\caption{Represented-zero and positive-restart census in recorded arithmetic.
Here $r$ is restart energy, $M$ is the maximum subsequent amplification
before the next represented zero, and $P=rM$ is that excursion's peak energy.}
\label{rg:tab:confirmation-faces}
\scriptsize
\begin{tabular}{@{}crrrrrrrr@{}}
\toprule
Traj. & Maps & Face states & Closures & Restarts & Censored & Median $r$ & Max. $M$ & Max. $P$ \\
\midrule
T0 & 4 & 151{,}287 & 14{,}788 & 14{,}784 & 0 & $1.962\times10^{-16}$ & $2.684\times10^{8}$ & $2.403\times10^{-13}$ \\
T1 & 0 & 0 & 0 & 0 & 0 & n/a & n/a & n/a \\
T2 & 16 & 377{,}288 & 37{,}323 & 37{,}307 & 0 & $1.242\times10^{-29}$ & $6.872\times10^{10}$ & $7.260\times10^{-24}$ \\
T3 & 0 & 0 & 0 & 0 & 0 & n/a & n/a & n/a \\
\bottomrule
\end{tabular}
\end{table}

\begin{table}[t]
\centering
\caption{Implemented PLGA diagnostics over all 192 registered maps at each
checkpoint. The sampled operator norm is a lower bound, not a certificate of
the segment supremum in Theorem~\ref{rg:thm:plga-rg-transfer}.}
\label{rg:tab:confirmation-plga}
\scriptsize
\begin{tabular}{@{}rrrrrrrr@{}}
\toprule
Step & Defect min. & Defect med. & Defect max. & Secant med. & $\widehat\kappa$ max. & Control max. & Face pass \\
\midrule
0 & 8.78 & 11.38 & 13.94 & 0.028 & $3.190\times10^{2}$ & $2.221\times10^{-15}$ & no \\
32{,}768 & 22.21 & 41.24 & 91.45 & 0.064 & $5.661\times10^{2}$ & $5.053\times10^{-15}$ & no \\
65{,}536 & 4.82 & 38.93 & 117.50 & 0.004 & $1.965\times10^{2}$ & $4.301\times10^{-15}$ & no \\
196{,}608 & $3.202\times10^{-5}$ & 49.87 & 124.92 & 0.004 & $85.96$ & $8.468\times10^{-15}$ & no \\
\bottomrule
\end{tabular}
\end{table}

\begin{table}[t]
\centering
\caption{Maximum relative residuals for the five jointly sufficient PLGA
face-preservation conditions. Each maximum is over all registered maps.}
\label{rg:tab:confirmation-plga-structural}
\scriptsize
\begin{tabular}{@{}rrrrrr@{}}
\toprule
Step & $W\mathbf1$ & Bias & Power & $a\mathbf1$ & Coupling bias \\
\midrule
0 & $1.000$ & $0.000$ & $0.9947$ & $1.000$ & $0.000$ \\
32{,}768 & $1.000$ & $0.9958$ & $0.6661$ & $1.000$ & $0.9970$ \\
65{,}536 & $1.000$ & $0.9961$ & $0.8729$ & $1.000$ & $0.9984$ \\
196{,}608 & $1.000$ & $0.9960$ & $0.8867$ & $1.000$ & $0.9980$ \\
\bottomrule
\end{tabular}
\end{table}

\begin{table}[t]
\centering
\caption{Training diagnostics and disjoint document ranges by trajectory.}
\label{rg:tab:confirmation-training}
\scriptsize
\begin{tabular}{@{}crlrrrrr@{}}
\toprule
Traj. & Seed & Documents & Tokens/upd. & Initial & Final & Min. & Median \\
\midrule
T0 & 202609300 & [7{,}000,25{,}467) & 260 & 5.850 & 2.208 & 0.511 & 1.789 \\
T1 & 202609301 & [36{,}000,53{,}826) & 260 & 6.111 & 1.773 & 0.275 & 1.736 \\
T2 & 202609302 & [66{,}000,86{,}879) & 260 & 5.934 & 1.668 & 0.702 & 1.781 \\
T3 & 202609303 & [95{,}000,113{,}881) & 260 & 5.956 & 2.392 & 0.470 & 1.795 \\
\bottomrule
\end{tabular}
\end{table}

\newcommand{\RGSourceObserverFloor}{$1.000\times10^{-12}$}
\newcommand{\RGSourceObserverRouteBound}{$1.728\times10^{-4}$}
\newcommand{\RGSourceObserverPositiveSubfloorStates}{1,037,258}
\newcommand{\RGSourceObserverExcursionPeakMaximum}{$2.403\times10^{-13}$}
\newcommand{\RGSourceObserverSingleStateFraction}{58.1\%}
\newcommand{\RGSourceObserverMinRequiredHoldout}{172,444}
\newcommand{\RGSourceObserverMaxRequiredHoldout}{5,458,022}
\newcommand{\RGSourceObserverCompletedRunGPUHours}{3.793}

\begin{table}[t]
\centering
\caption{Observer-band occupancy and design-variance precision projection. Fractions use the largest measured checkpoint route discrepancy; this is a sensitivity band, not a between-checkpoint certificate.}
\label{rg:tab:observer-bands-precision}
\scriptsize
\begin{tabular}{lrrrr}
\toprule
Cell & represented zero & positive below floor & holdout in route band & required holdout \\
\midrule
T0 L0 & 151,287 & 111,108 & 30.3\% & not defined \\
T0 L1 & 0 & 0 & 63.7\% & 5,103,778 \\
T0 L2 & 0 & 0 & 29.9\% & 2,054,881 \\
T1 L0 & 0 & 0 & 8.3\% & 5,458,022 \\
T1 L1 & 0 & 0 & 24.8\% & 3,272,307 \\
T1 L2 & 0 & 0 & 9.9\% & 1,081,001 \\
T2 L0 & 377,288 & 926,150 & 86.9\% & not defined \\
T2 L1 & 0 & 0 & 43.3\% & 3,227,846 \\
T2 L2 & 0 & 0 & 0.0\% & 517,335 \\
T3 L0 & 0 & 0 & 1.4\% & 805,449 \\
T3 L1 & 0 & 0 & 21.7\% & 2,720,036 \\
T3 L2 & 0 & 0 & 0.0\% & 172,444 \\
\bottomrule
\end{tabular}
\end{table}

\begin{table}[t]
\centering
\caption{All-layer large-increment census using the same absolute log-gain threshold. Only edges with positive represented endpoints enter the logarithm.}
\label{rg:tab:all-layer-large-increments}
\begin{tabular}{lrr}
\toprule
Layer & event count & maximum $|Y|$ \\
\midrule
0 & 119,953 & 24.953 \\
1 & 1 & 4.525 \\
2 & 2 & 1.088 \\
\bottomrule
\end{tabular}
\end{table}

\newcommand{\RGSourceStageCheckpointCount}{16}
\newcommand{\RGSourceStageDeviceCount}{3}
\newcommand{\RGSourceStageMapEvaluations}{2{,}304}
\newcommand{\RGSourceStageFinalZeros}{20}
\newcommand{\RGSourceStageInheritedZeros}{8}
\newcommand{\RGSourceStagePositiveRoundedZeros}{12}
\newcommand{\RGSourceStageUnresolvedZeros}{0}
\newcommand{\RGSourceStageFirstSOne}{4}
\newcommand{\RGSourceStageFirstSTwoOut}{4}
\newcommand{\RGSourceStageFirstSThreeOut}{12}
\newcommand{\RGSourceStageFirstCoalescencePositive}{20}
\newcommand{\RGSourceStageFirstCoalescenceUnresolved}{0}
\newcommand{\RGSourceStageNetworkSixtyFourZeroCount}{1}
\newcommand{\RGSourceStagePipelineSixtyFourZeroCount}{1}
\newcommand{\RGSourceStageNetworkSixtyFourOneUlpCount}{3}
\newcommand{\RGSourceStagePipelineSixtyFourOneUlpCount}{3}
\newcommand{\RGSourceStageNetworkSixtyFourPositiveCount}{16}
\newcommand{\RGSourceStagePipelineSixtyFourPositiveCount}{16}
\newcommand{\RGSourceStageNetworkSixtyFourMinimum}{0 (exact)}
\newcommand{\RGSourceStageNetworkSixtyFourPositiveMinimum}{$2.694\times10^{-48}$}
\newcommand{\RGSourceStageNetworkSixtyFourMaximum}{$1.175\times10^{-23}$}
\newcommand{\RGSourceStageNetworkSixtyFourRoundedMinimum}{$3.815\times10^{-44}$}
\newcommand{\RGSourceStagePipelineSixtyFourMinimum}{0 (exact)}
\newcommand{\RGSourceStagePipelineSixtyFourPositiveMinimum}{$2.694\times10^{-48}$}
\newcommand{\RGSourceStagePipelineSixtyFourMaximum}{$1.175\times10^{-23}$}
\newcommand{\RGSourceStagePipelineSixtyFourRoundedMinimum}{$3.815\times10^{-44}$}
\newcommand{\RGSourceStageDeviceDependentZeros}{0}
\newcommand{\RGSourceStageRecordDigest}{\hashpair{3ee01eee9b2bb7be150e39f0a61f49a8}{5931a3535cf0ec23a46acf9fe90f8448}}
\newcommand{\RGSourceStageSourceCommit}{\hashpair{0147b4144aee36431e1a}{83f1e6ae24a33a26bc59}}

\begin{table}[t]
\centering
\caption{Stage-resolved localization of producing-device float32 final row-equality events. Checkpoints with no final represented-zero map are omitted. A positive final input means that exact dyadic evaluation of the final LayerNorm from its represented input has a strictly positive output-energy lower endpoint.}
\label{rg:tab:stage-resolved-zero-provenance}
\begin{tabular}{lrrrrl}
\toprule
Trajectory & Update & Final zeros & Inherited zero & Positive input & First coalescence \\
\midrule
T0 & 196,608 & 4 & 4 & 0 & S1: 4 \\
T2 & 196,608 & 16 & 4 & 12 & S2out: 4, S3out: 12 \\
\bottomrule
\end{tabular}
\end{table}

\begin{table}[t]
\centering
\caption{Local diagnostics at the first float32 row-coalescing LayerNorm. The represented input energy is computed from the stored float32 array. The exact-output column gives the range of lower endpoints from exact dyadic evaluation of those arrays and represented gains. Gain and bias norms are constant within each row at the shown precision.}
\label{rg:tab:first-coalescence-local-diagnostics}
\scriptsize
\begin{tabular}{@{}llrcccc@{}}
\toprule
Output & Input & Maps & \shortstack{Input-energy\\range} & \shortstack{Exact-output lower\\endpoint range} & $\|\gamma\|_2$ & $\|\beta\|_2$ \\
\midrule
S1 & S0 & 4 & \shortstack{$9.856\times10^{-23}$\\to\\$1.250\times10^{-22}$} & \shortstack{$8.429\times10^{-24}$\\to\\$1.092\times10^{-23}$} & $8.517\times10^{-3}$ & $0.1219$ \\
S2out & S2in & 4 & \shortstack{$1.624\times10^{-28}$\\to\\$2.245\times10^{-28}$} & \shortstack{$7.623\times10^{-40}$\\to\\$1.091\times10^{-39}$} & $4.058\times10^{-8}$ & $2.232\times10^{-8}$ \\
S3out & S3in & 12 & \shortstack{$3.019\times10^{-29}$\\to\\$3.331\times10^{-24}$} & \shortstack{$2.030\times10^{-38}$\\to\\$2.499\times10^{-33}$} & $4.686\times10^{-8}$ & $3.396\times10^{-8}$ \\
\bottomrule
\end{tabular}
\end{table}

\begin{table}[t]
\centering
\caption{Exact represented-array classification for the float64 controls corresponding to producing-device float32 final zeros. Exact energies use the pairwise dyadic identity. The rounded column shows the centered-sum arithmetic floor.}
\label{rg:tab:stage-resolved-precision-controls}
\scriptsize
\begin{tabular}{lrrrrrr}
\toprule
Route & zero & one-ulp & other positive & exact min. & min. positive & rounded min. \\
\midrule
Whole network, float64 & 1 & 3 & 16 & 0 (exact) & $2.694\times10^{-48}$ & $3.815\times10^{-44}$ \\
Row-map pipeline, float64 from float32 S0 & 1 & 3 & 16 & 0 (exact) & $2.694\times10^{-48}$ & $3.815\times10^{-44}$ \\
\bottomrule
\end{tabular}
\end{table}

\newcommand{\RGSourceClosureTrajectoryCount}{2}
\newcommand{\RGSourceClosureScaleCount}{4}
\newcommand{\RGSourceClosureBranchCount}{16}
\newcommand{\RGSourceClosureCheckpointCount}{64}
\newcommand{\RGSourceClosureRejectedScaleCount}{4}
\newcommand{\RGSourceClosureRecordDigest}{\hashpair{864a68f6758320af6792ec7eb486912f}{e150190150be4db08f4c2fa2249c0ffa}}
\newcommand{\RGSourceClosureExecutionCommit}{\hashpair{8225edd891dd744591d8}{ee74b17761e823c2ea6a}}
\newcommand{\RGSourceClosureEnergyDecision}{rejected on the tested domain}
\newcommand{\RGSourceClosureRawMomentDecision}{rejected on the tested domain}
\newcommand{\RGSourceClosureAugmentationStatus}{not decided}

\begin{table}[t]
\centering
\caption{Independent source remeasurement. Changed maps count unequal stored float64 energy bit patterns relative to the recorder row. Source originals were measured on the producing GPU, the other GPU, and CPU. Rewritten copies were measured only on their producing GPU.}
\label{rg:tab:closure-source-gate}
\small
\begin{tabular}{llrrr}
\toprule
Trajectory & Route & Changed maps & $\|\Delta_0\|_\infty$ & Max. symmetric defect \\
\midrule
T0 & producing GPU & 0/48 & $0$ & $0$\\
T0 & other GPU & 0/48 & $0$ & $0$\\
T0 & CPU control & 48/48 & $1.974\times10^{-4}$ & $4.916\times10^{-8}$\\
T1 & producing GPU & 0/48 & $0$ & $0$\\
T1 & other GPU & 0/48 & $0$ & $0$\\
T1 & CPU control & 48/48 & $2.455\times10^{-4}$ & $6.104\times10^{-8}$\\
\bottomrule
\end{tabular}
\end{table}

\begin{table}[t]
\centering
\caption{Primary matched-fiber test. Only the AdamW first moments are zeroed. A scale is rejected exactly when at least one successor-energy bit pattern changes.}
\label{rg:tab:energy-closure-probe}
\small
\begin{tabular}{lrrrr}
\toprule
Trajectory & $b$ & Changed maps & $\|\Delta_b\|_\infty$ & Max. symmetric defect \\
\midrule
T0 & 1 & 48/48 & $29.07$ & $7.312\times10^{-3}$\\
T0 & 2 & 48/48 & $32.97$ & $8.269\times10^{-3}$\\
T0 & 4 & 48/48 & $52.25$ & $1.307\times10^{-2}$\\
T0 & 8 & 48/48 & $1.080\times10^{2}$ & $2.716\times10^{-2}$\\
T1 & 1 & 48/48 & $15.81$ & $3.954\times10^{-3}$\\
T1 & 2 & 48/48 & $46.07$ & $1.151\times10^{-2}$\\
T1 & 4 & 48/48 & $62.61$ & $1.562\times10^{-2}$\\
T1 & 8 & 48/48 & $93.88$ & $2.350\times10^{-2}$\\
\bottomrule
\end{tabular}
\end{table}

\begin{table*}[p]
\centering
\caption{Second-moment dose sweep at every measured scale. Counts aggregate two 48-map trajectories. These finite-horizon values are reported without a monotonicity assumption or fitted scaling law; a nonmonotone sequence is therefore evidence about sensitivity, not a beta function.}
\label{rg:tab:second-moment-dose}
\small
\begin{tabular}{lrrrr}
\toprule
Intervention & $b$ & Changed maps & $\|\Delta_b\|_\infty$ & Max. symmetric defect \\
\midrule
$0.999v$ & 1 & 96/96 & $1.271\times10^{-2}$ & $3.185\times10^{-6}$\\
$0.999v$ & 2 & 96/96 & $3.965$ & $9.868\times10^{-4}$\\
$0.999v$ & 4 & 96/96 & $21.91$ & $5.521\times10^{-3}$\\
$0.999v$ & 8 & 96/96 & $75.27$ & $1.901\times10^{-2}$\\
$0.5v$ & 1 & 96/96 & $7.513$ & $1.881\times10^{-3}$\\
$0.5v$ & 2 & 96/96 & $31.06$ & $7.759\times10^{-3}$\\
$0.5v$ & 4 & 96/96 & $49.48$ & $1.244\times10^{-2}$\\
$0.5v$ & 8 & 96/96 & $91.07$ & $2.296\times10^{-2}$\\
$2v$ & 1 & 96/96 & $6.759$ & $1.695\times10^{-3}$\\
$2v$ & 2 & 96/96 & $15.08$ & $3.773\times10^{-3}$\\
$2v$ & 4 & 96/96 & $43.24$ & $1.088\times10^{-2}$\\
$2v$ & 8 & 96/96 & $75.61$ & $1.895\times10^{-2}$\\
$v=\bm0$ & 1 & 96/96 & $1.387\times10^{2}$ & $3.514\times10^{-2}$\\
$v=\bm0$ & 2 & 96/96 & $1.123\times10^{2}$ & $2.828\times10^{-2}$\\
$v=\bm0$ & 4 & 96/96 & $96.67$ & $2.416\times10^{-2}$\\
$v=\bm0$ & 8 & 96/96 & $1.672\times10^{2}$ & $4.244\times10^{-2}$\\
\bottomrule
\end{tabular}
\end{table*}

\begin{table*}[t]
\centering
\caption{Sign-gauge controls at every scale. The consistent action co-transforms first moments and is the exact null control. The parameter-only action preserves the source energy and raw first and second moments, so changed successors reject closure of the raw $(e,m,v)$ augmentation on this domain.}
\label{rg:tab:gauge-closure-control}
\small
\begin{tabular}{lcclrr}
\toprule
Action & Source & Role & $b$ & Changed maps & Max. symmetric defect \\
\midrule
full $C_g(\theta,m,v)$ & exact & null control & 1 & 0/96 & $0$\\
full $C_g(\theta,m,v)$ & exact & null control & 2 & 0/96 & $0$\\
full $C_g(\theta,m,v)$ & exact & null control & 4 & 0/96 & $0$\\
full $C_g(\theta,m,v)$ & exact & null control & 8 & 0/96 & $0$\\
full $P_g(\theta,m,v)$ & exact & raw-moment closure & 1 & 96/96 & $6.885\times10^{-4}$\\
full $P_g(\theta,m,v)$ & exact & raw-moment closure & 2 & 96/96 & $2.503\times10^{-3}$\\
full $P_g(\theta,m,v)$ & exact & raw-moment closure & 4 & 96/96 & $9.415\times10^{-3}$\\
full $P_g(\theta,m,v)$ & exact & raw-moment closure & 8 & 96/96 & $1.926\times10^{-2}$\\
full $M_g(\theta,m,v)$ & exact & paired control & 1 & 96/96 & $6.885\times10^{-4}$\\
full $M_g(\theta,m,v)$ & exact & paired control & 2 & 96/96 & $2.503\times10^{-3}$\\
full $M_g(\theta,m,v)$ & exact & paired control & 4 & 96/96 & $9.415\times10^{-3}$\\
full $M_g(\theta,m,v)$ & exact & paired control & 8 & 96/96 & $1.926\times10^{-2}$\\
layer-0 $C_{g_0}(\theta,m,v)$ & exact & null control & 1 & 0/96 & $0$\\
layer-0 $C_{g_0}(\theta,m,v)$ & exact & null control & 2 & 0/96 & $0$\\
layer-0 $C_{g_0}(\theta,m,v)$ & exact & null control & 4 & 0/96 & $0$\\
layer-0 $C_{g_0}(\theta,m,v)$ & exact & null control & 8 & 0/96 & $0$\\
layer-0 $P_{g_0}(\theta,m,v)$ & exact & raw-moment closure & 1 & 96/96 & $2.931\times10^{-4}$\\
layer-0 $P_{g_0}(\theta,m,v)$ & exact & raw-moment closure & 2 & 96/96 & $2.375\times10^{-3}$\\
layer-0 $P_{g_0}(\theta,m,v)$ & exact & raw-moment closure & 4 & 96/96 & $9.737\times10^{-3}$\\
layer-0 $P_{g_0}(\theta,m,v)$ & exact & raw-moment closure & 8 & 96/96 & $1.567\times10^{-2}$\\
\bottomrule
\end{tabular}
\end{table*}

\begin{table*}[t]
\centering
\caption{Direct product-gauge comparisons. The full moment-only branch is compared with the full parameter-only branch as required by the paired-orbit identity. The layer-0 consistent and parameter-only actions are compared with baseline. Counts aggregate both 48-map trajectories.}
\label{rg:tab:gauge-product-decomposition}
\small
\begin{tabular}{lrrrr}
\toprule
Comparison & $b$ & Changed maps & $\|\Delta_b\|_\infty$ & Max. symmetric defect \\
\midrule
full $M_g$ versus $P_g$ & 1 & 0/96 & $0$ & $0$\\
full $M_g$ versus $P_g$ & 2 & 0/96 & $0$ & $0$\\
full $M_g$ versus $P_g$ & 4 & 0/96 & $0$ & $0$\\
full $M_g$ versus $P_g$ & 8 & 0/96 & $0$ & $0$\\
layer-0 $C_{g_0}$ versus baseline & 1 & 0/96 & $0$ & $0$\\
layer-0 $C_{g_0}$ versus baseline & 2 & 0/96 & $0$ & $0$\\
layer-0 $C_{g_0}$ versus baseline & 4 & 0/96 & $0$ & $0$\\
layer-0 $C_{g_0}$ versus baseline & 8 & 0/96 & $0$ & $0$\\
layer-0 $P_{g_0}$ versus baseline & 1 & 96/96 & $1.196$ & $2.931\times10^{-4}$\\
layer-0 $P_{g_0}$ versus baseline & 2 & 96/96 & $9.561$ & $2.375\times10^{-3}$\\
layer-0 $P_{g_0}$ versus baseline & 4 & 96/96 & $38.73$ & $9.737\times10^{-3}$\\
layer-0 $P_{g_0}$ versus baseline & 8 & 96/96 & $61.92$ & $1.567\times10^{-2}$\\
\bottomrule
\end{tabular}
\end{table*}

\begin{table*}[t]
\centering
\caption{First-moment, model-coordinate, and represented-resolution controls. The $b=1$ and $b=8$ endpoints summarize finite-horizon sensitivity only. A model-coordinate intervention is closure evidence only when its source gate is exact.}
\label{rg:tab:closure-sensitivity-controls}
\small
\begin{tabular}{lclrrrr}
\toprule
Intervention & Source & Role & $N_1$ & $\delta_1$ & $N_8$ & $\delta_8$ \\
\midrule
$0.999m$ & exact & energy closure & 96/96 & $6.887\times10^{-6}$ & 96/96 & $1.839\times10^{-2}$\\
$0.9m$ & exact & energy closure & 96/96 & $6.810\times10^{-4}$ & 96/96 & $2.596\times10^{-2}$\\
$m=\bm0$ & exact & energy closure & 96/96 & $7.312\times10^{-3}$ & 96/96 & $2.716\times10^{-2}$\\
one ulp in $m$ & exact & resolution & 0/96 & $0$ & 0/96 & $0$\\
LayerNorm bias $+0.01$ & 96/96 changed & sensitivity & 96/96 & $4.924\times10^{-3}$ & 96/96 & $2.366\times10^{-2}$\\
\bottomrule
\end{tabular}
\end{table*}

\newcommand{\RGSourceStageOneRecordDigest}{\hashpair{0232b9e0ff38f21a3e8d669267e9c1bd}{02541164a2a4832a29856e26c0a29567}}
\newcommand{\RGSourceStageOneGPUHours}{0.021013}
\newcommand{\RGSourceStageOneTrajectoryCount}{4}
\newcommand{\RGSourceStageOneCandidateCount}{3}
\newcommand{\RGSourceStageOnePromotedCount}{0}
\newcommand{\RGSourceStageOneStatus}{\texttt{COMPLETED\_STOPPED\_AT\_CLOSURE}}
\newcommand{\RGSourceStageOneGaugeStatus}{\texttt{NOT\_REJECTED}}
\newcommand{\RGSourceStageOneScaleList}{1,2,4,8,16}

\begin{table}[t]
\centering
\small
\caption{Executed Stage-1 candidate-state closure decisions. Each challenge first requires exact equality of the registered source candidate. Design and holdout decisions use disjoint source trajectories and the frozen scales $b\in\{1,2,4,8,16\}$.}
\label{rg:tab:staged-predictive-closure}
\begin{tabularx}{\textwidth}{@{}XXll@{}}
\toprule
Candidate state & Challenge & Design & Holdout \\
\midrule
energy vector & parameter-only gauge & \texttt{REJECTED} & \texttt{REJECTED} \\
energy + raw AdamW + phase & parameter-only gauge & \texttt{REJECTED} & \texttt{REJECTED} \\
row maps + canonical AdamW + phase & probe-null embedding & \texttt{REJECTED} & \texttt{REJECTED} \\
\bottomrule
\end{tabularx}

\end{table}

\begin{table}[t]
\centering
\caption{Exact 95\% Clopper--Pearson intervals for the executed protocol-qualification simulation. Counts are reconstructed exactly from the sealed record's integer trial totals and proportions.}
\label{rg:tab:qualification-calibration}
\small
\begin{tabularx}{\textwidth}{@{}Xrrr@{}}
\toprule
Qualification event & Count & Estimate & Exact 95\% interval \\
\midrule
Flow familywise false-positive event & 48/1,000 & 0.048000 & [0.035600, 0.063140] \\
All-cell stationary-domain acceptance & 972/1,000 & 0.972000 & [0.959785, 0.981315] \\
Gaussian familywise nonrejection & 978/1,000 & 0.978000 & [0.966880, 0.986163] \\
Finite-variance cell acceptance & 11,995/12,000 & 0.999583 & [0.999028, 0.999865] \\
Directional detection at the qualified alternative & 23,999/24,000 & 0.999958 & [0.999768, 0.999999] \\
\bottomrule
\end{tabularx}
\end{table}

\begin{table}[t]
\centering
\caption{Generated empirical falsification and status ledger. Source, seal, and replay checks are excluded because they validate execution rather than constitute scientific outcomes.}
\label{rg:tab:evidence-decision-ledger}
\begin{tabularx}{\textwidth}{@{}
 >{\raggedright\arraybackslash}p{0.28\textwidth}
 >{\raggedright\arraybackslash}p{0.20\textwidth}
 >{\raggedright\arraybackslash}X@{}}
\toprule
Claim & Status & Executed basis \\
\midrule
Retained aggregate typical flow & contracting but heterogeneous & 6 descriptive medians are below one; every captured window also contains expansion. \\
Registered cellwise flow & no directional classification & 0 negative, 0 positive, 0 unresolved, 10 precision-target-unmet, and 2 face-interrupted cells. \\
Stationary temporal domain & partial & 8 of 10 complete cells pass all temporal gates. \\
Gaussian finite-variance class & not identified & 4 cells have no matched-normal rejection and 10 pass the variance screen, but 0 pass the joint criterion. \\
Regular nonautonomous drift & premise not established & 2 cells pass the sign screen and 0 reach the registered held-out prediction. \\
Recurrent analytical source sector & not established & All 20 first-coalescing LayerNorm outputs have a strictly positive exact-dyadic lower endpoint. Whole-network exact represented arrays: 1 zero, 3 one-ulp, 16 other positive; the S0 replay gives 1, 3, and 16, respectively. \\
Autonomous row-energy-vector closure & rejected on tested domain & Independent producing-GPU and other-GPU source measurements are bitwise exact. First-moment fibers disagree at scales 1, 2, 4, 8. Second-moment fibers disagree at scales 1, 2, 4, 8. \\
Raw $(e,m,v)$ augmentation & rejected on tested domain & The parameter-only sign action preserves source energy and raw AdamW moments but changes successors; the consistent action is bitwise exact. The moment-only partner agrees with the paired orbit. \\
Unique or minimal state augmentation & not decided & The tested moments are fiber coordinates, while the gauge witness shows that their raw coordinates are not sufficient. \\
Prespecified reduced predictive states & rejected & All 3 candidate states fail at scales 1, 2, 4, 8, and 16 in both design and held-out trajectories; the complete-state quotient control remains exact. \\
Implemented PLGA face preservation & rejected on the measured domain & Direct defects exceed tolerance for all 192 trained maps at all four checkpoints; constrained controls remain at roundoff. \\
PLDR universality class & not identified & The registered prerequisites fail and the record covers one architecture at finite scales. \\
Asymptotic PLDR collapse & not decided & Finite trajectories do not close the all-future positive-tail or recurrent-excursion premise. \\
Cross-system universality & not tested & The executed record covers one architecture, four seeds, and no cell passing the joint universality criterion. \\
\bottomrule
\end{tabularx}
\end{table}

\subsubsection{Experimental design and statistical calibration}
\label{rg:sec:long-design}

The decision procedure was qualified under simulated null and alternative
models before the evidence-producing run.  The exact binomial counts and
two-sided 95\% Clopper--Pearson intervals were: flow familywise false
positives \(48/1000\), interval \([0.035600,0.063140]\); all-cell
stationary-domain acceptance \(972/1000\), interval
\([0.959785,0.981315]\); Gaussian familywise nonrejection \(978/1000\),
interval \([0.966880,0.986163]\); and finite-variance cell acceptance
\(11995/12000\), interval \([0.999028,0.999865]\).  At mean magnitude
\(2\times10^{-4}\) per update, directional detection occurred in
\(23999/24000\) alternative cells, interval
\([0.999768,0.999999]\).  Table~\ref{rg:tab:qualification-calibration} is
generated from the sealed qualification record.

The four trajectories use seeds 202609300--202609303 and a full-map
backward-error gate. The quotient-relative LayerNorm diagnostic is treated
separately because division by a nearly vanishing centered norm can amplify
roundoff. The 131,072-update holdout is the largest resource-capped horizon
in the fixed design. Its adequacy is assessed from design-segment variance
before the holdout directional test.

Four independently initialized compact PLDR models ran on two NVIDIA RTX
4090 GPUs. Each model had three decoder layers,
four heads, \(d_k=64\), \(d_{\rm model}=256\), four fixed probe contexts of
length 64, 3,067,571 parameters, and a 257-symbol byte vocabulary. Each
training update consumed 260 tokens. Training streams were disjoint, while
the fixed probe text was shared. AdamW used learning rate \(10^{-3}\), betas
\((0.9,0.95)\), epsilon \(10^{-5}\), weight decay \(0.1\), elementwise
gradient clipping at one, and no schedule. Table~\ref{rg:tab:confirmation-training}
records seeds, exact document intervals and training losses.

Each trajectory contained \(\RGSourceConfirmUpdates{}\) updates: 32,768 fixed burn-in,
32,768 design, and 131,072 locked holdout updates. The native float64 energy
of \(\RGSourceConfirmMapsPerTrajectory{}\) row maps was recorded at initialization
and after every update. For trajectory \(i\) and layer \(\ell\), the primary
series was the mean of the 16 per-map log gains,
\begin{equation}
 Y_{i\ell,t}
 =\frac1{16}\sum_{m\in(i,\ell)}
   \log\frac{E_{m,t+1}}{E_{m,t}},
 \label{rg:eq:registered-cell-series}
\end{equation}
whenever every involved endpoint was positive. Maps were therefore not
treated as independent temporal replicates.

Nonoverlapping 512-update means gave \(n_d=64\) design batches and
\(n_h=256\) holdout batches per complete cell. Let \(C=12\) be the
simultaneous cell count, \(\alpha_F=0.05\) the nominal familywise level,
\(s_d^2\) the design-batch sample variance, and \(\nu_d=n_d-1\). The
registered design-only variance upper bound under independent normal
design batch means is
\begin{equation}
 \widehat\sigma^2_{d,U}
 =\frac{\nu_d s_d^2}{\chi^2_{\alpha_F/C,\nu_d}}.
 \label{rg:eq:registered-design-variance-upper}
\end{equation}
With \(\nu_h=n_h-1\), the projected nominal simultaneous confidence half-width is
\begin{equation}
 h_{\rm proj}
 =t_{1-\alpha_F/C,\nu_h}
  \sqrt{\frac{\widehat\sigma^2_{d,U}}{n_h}}.
 \label{rg:eq:registered-projected-halfwidth}
\end{equation}
The frozen precision gate is
\begin{equation}
 h_{\rm proj}\le\delta,
 \qquad \delta=10^{-4}\ \text{per update}.
 \label{rg:eq:registered-precision-gate}
\end{equation}
This statistic forecasts a nominal simultaneous one-sided confidence width
at the planned holdout size. Transporting the design variance to holdout
requires variance stability or a justified variance-transport bound. The
projection is a planning statistic, not a guarantee of the realized width. It contains no type-II-error quantile and is not a
conventional power calculation. The separate qualification
simulation used an alternative of magnitude \(2\delta\) and produced the
directional detection count reported above. The design sign selected the
held-out direction. The corresponding Student \(t\) procedure has nominal
familywise calibration over all \(\RGSourceConfirmCells{}\) trajectory-layer cells
under the reference model below. A
stationary phase label further required split and trend statistics no larger
than 3.5 in magnitude, variance ratios in \([0.25,4]\), and the same bounds
across design and holdout.

\paragraph{Calibration model and native dependence.}
Let $\mathcal D$ contain the design information selecting signs
$a_c\in\{-1,1\}$. A sufficient model for exact Student calibration is
that, conditional on $\mathcal D$, each cell's $n_h$ holdout batch means
are independent normal observations with common mean $\mu_c$ and
positive variance $\sigma_{h,c}^2$. Then
\[
 L_c=a_c\bar H_c-t_{1-\alpha_F/C,n_h-1}\frac{S_{h,c}}{\sqrt{n_h}}
 \quad\hbox{satisfies}\quad
 \Prob(L_c>a_c\mu_c\mid\mathcal D)\leq\alpha_F/C.
\]
The union bound yields
$\Prob\{\exists c:L_c>a_c\mu_c\}\leq\alpha_F$, without independence
between cells. It requires valid marginal conditional pivots, including
any additional eligibility selection. Independently sampled normal
design batches justify the chi-square upper bound on the \emph{design}
variance. A joint design-and-holdout confidence claim must allocate error
across both events; two separate $0.05$ budgets do not make a joint
$0.05$ bound.

Nonoverlapping native batches need not be independent or normal.
Optimizer memory, changing parameters, corpus depletion, and shared
history can violate the reference model. A dependence justification for
these native pivots has not been established. Split, trend, and variance
screens and Gaussian qualification do not establish that missing premise.
The native procedure is therefore reported with nominal calibration.
Its precision failures diagnose the registered design's inadequacy;
they do not establish zero drift, marginality, or a critical phase.

The finite-scale law test used block sizes 128, 256, and 512, 4,999
matched-size Gaussian bootstrap replicates per test, and Holm correction over
all cells and scales. Every simulated sample was recentered and rescaled by
the same operations as its observed counterpart. Finite-variance scaling
required a holdout block-sum variance exponent in \([0.8,1.2]\) and a
design-to-holdout exponent shift no larger than 0.25. Thus nonrejection is
only finite calibration-model compatibility, never asymptotic class identification. The
separate nonautonomous prediction fitted
\(\overline Y(t)=-ct^{-\gamma}\) on four design windows and, when
\(R^2\ge0.8\) and \(0\le\gamma\le1\), required the held-out cumulative log
gain to agree within 25\%.

\subsubsection{Flow, temporal domain, and finite-scale laws}
\label{rg:sec:temporal-flow-results}

The run produced \(\RGSourceConfirmTrajectories{}\) trajectories,
\(\RGSourceConfirmRecordedStates{}\) recorded map states, and
\(\RGSourceConfirmCompleteCells{}\) cells whose recorded endpoints remained positive
out of \(\RGSourceConfirmCells{}\). The other
\(\RGSourceConfirmFaceInterruptedCells{}\) cells received the registered
\emph{face-interrupted} label after bitwise row equality and were excluded
from logarithmic inference. Under
Theorem~\ref{rg:thm:observer-energy-enclosure}, that is a numerical branch label,
not by itself an analytical-face certificate. Of the complete cells,
\(\RGSourceConfirmPrecisionGateCells{}\) met the design-only simultaneous
nominal confidence-half-width target. The registered flow census was
\(\RGSourceConfirmNegativeDriftCells{}\) negative-drift,
\(\RGSourceConfirmPositiveDriftCells{}\) positive-drift,
\(\RGSourceConfirmUnresolvedCells{}\) unresolved,
\(\RGSourceConfirmPrecisionTargetUnmetCells{}\) precision-target-unmet, and
\(\RGSourceConfirmFaceInterruptedCells{}\) face-interrupted cells
(Table~\ref{rg:tab:confirmation-flow}). An interval containing zero is not
interpreted as evidence for marginality when the precision target fails.

All ten complete-cell design means were negative, but their holdout means
split evenly between negative and positive signs. The projected simultaneous
nominal confidence half-widths ranged
from \(1.147\times10^{-4}\) to \(6.453\times10^{-4}\) per update, all above
the \(10^{-4}\) target. These classifications follow from the design-only
precision rule and do not use the holdout to excuse an unfavorable
direction. The negative design signs cannot be promoted to evidence of
subcritical flow.

Recomputing the design-variance requirement without changing the registered
decisions gives \RGSourceObserverMinRequiredHoldout{} to
\RGSourceObserverMaxRequiredHoldout{} holdout updates across the ten positive cells,
or 1.32 to 41.64 times the available horizon. Thus the available holdout
length does not meet the declared precision target in any complete cell.

Only \(\RGSourceConfirmStableCells{}\) complete cells passed every temporal-domain
gate, and \(\RGSourceConfirmNonautonomousCells{}\) received the explicit
nonautonomous phase label. At the law level,
\(\RGSourceConfirmLawRejections{}\) of \(\RGSourceConfirmLawTests{}\) Holm-adjusted
matched-normal tests rejected; \(\RGSourceConfirmGaussianNonrejectedCells{}\) cells
had no rejection at any registered law scale. Independently,
\(\RGSourceConfirmScalingCells{}\) cells met the finite-variance scaling condition.
After combining the precision gate, temporal stability, law nonrejection, and
variance scaling, \(\RGSourceConfirmFiniteScaleCells{}\) cells met the full registered
finite-scale criterion. This result does not identify a universality class.
The four cells without a matched-normal rejection were T1 layers 1 and 2,
T2 layer 2, and T3 layer 2. Because no complete cell met the precision target,
these marginal nonrejections cannot be interpreted as affirmative
finite-variance universality evidence.

The uniform-negative design-window screen selected
\(\RGSourceConfirmRegularEligibleCells{}\) cells, but neither satisfied all
prespecified fit gates. T1 layer 2 had \(R^2=0.03\), while T3 layer 1 had
\(\gamma=3.20\), outside \([0,1]\). Consequently zero cells reached the
held-out prediction stage and \(\RGSourceConfirmRegularPredictions{}\) registered
predictions passed. The predicted and observed sums printed in
Table~\ref{rg:tab:confirmation-drift} are post-run diagnostics for the two
screened cells, not failed confirmatory predictions. The experiment therefore
does not test, confirm, or falsify the regular-drift branch.

Table~\ref{rg:tab:confirmation-transients} shows why endpoint compression is
inadequate: it retains all four energy checkpoints, per-map holdout signs, and
update-level excess kurtosis for every cell. For example, the T1 layer-0
median fell to \(7.148\times10^{-6}\) at design end and rebounded to
\(0.4953\) finally, with holdout excess kurtosis 151.45. T3
layers 1 and 2 also rebounded after design end. The registered layer-1 ledger
contains
\(\RGSourceConfirmLargeIncrementEvents{}\) events in
\(\RGSourceConfirmLargeIncrementClusters{}\) temporal clusters
(Table~\ref{rg:tab:confirmation-large-increments}): T3 map
\(\mathtt{c2.L1.H3}\) jumped at burn-in update 3,572 from energy 0.673 to
62.09, with log gain 4.525. No registered layer-1 large increment occurred in
design or holdout.

\subsubsection{Observer regime and stage-resolved represented zeros}
\label{rg:sec:observer-results}

Among the \(\RGSourceConfirmRecordedStates{}\) recorded map states,
\(\RGSourceConfirmFaceStates{}\) were bitwise zeros and the recorded global minimum
was \(\RGSourceConfirmMinimumEnergy{}\). The represented ledger contains
\(\RGSourceConfirmFaceClosures{}\) positive-to-zero transitions,
\(\RGSourceConfirmRestarts{}\) zero-to-positive transitions, and
\(\RGSourceConfirmRightCensoredExcursions{}\) right-censored positive excursions.
These are reproducible statements about recorded arithmetic. They are not,
without an exact enclosure, counts of analytical face hits or canonical
sources.

The full trace contains \RGSourceObserverPositiveSubfloorStates{} strictly positive
states below the observer floor \RGSourceObserverFloor{}. Every one of the
\RGSourceConfirmRestarts{} represented excursions has peak at most
\RGSourceObserverExcursionPeakMaximum{}, equal to the independently reproduced
\(\RGSourceConfirmMaximumExcursionPeak{}\), and therefore remains below the floor.
The maximum represented restart amplification is
\(\RGSourceConfirmMaximumRestartAmplification{},\) while
\RGSourceObserverSingleStateFraction{} of excursions contain only one positive state.
This scale separation explains why the exact theory retains
\(P_j=r_jM_j\), but a fixed subfloor envelope does not prove \(P_j\to0\).
The sparse checkpoint route bound \RGSourceObserverRouteBound{} is used only for the
occupancy diagnostic in Table~\ref{rg:tab:observer-bands-precision}; it is not a
uniform bound between checkpoints.

The stage-resolved replay evaluated the complete frozen checkpoint set:
\RGSourceStageCheckpointCount{} files, comprising four protocol-defined saved steps
for each trajectory.  It used \RGSourceStageDeviceCount{} routes, namely CPU and two
GPUs, for \RGSourceStageMapEvaluations{} map-device instances. It retained stages S0,
S1, S2in, S2out, S3in, and S3out for float32, a whole-network float64 route,
and a float64 row-program route starting from the exact float32 S0 array. On
each checkpoint's producing
device, \RGSourceStageFinalZeros{} maps were row equal at float32 S3out. Of these,
\RGSourceStageInheritedZeros{} already had identical represented rows before the
final LayerNorm: \RGSourceStageFirstSOne{} first coalesced at S1 and
\RGSourceStageFirstSTwoOut{} first coalesced at S2out. The other
\RGSourceStagePositiveRoundedZeros{} represented-input-positive cases, equal to
\RGSourceStageFirstSThreeOut{} maps, first coalesced at S3out although an exact
dyadic evaluation from the represented S3in array gave a strictly positive
output-energy lower endpoint.
Table~\ref{rg:tab:stage-resolved-zero-provenance} therefore locates represented
coalescence, but does not identify an exact real face of the underlying
training computation.

The local enclosure was also applied at S1 and S2out. At the first
coalescing LayerNorm, all \RGSourceStageFirstCoalescencePositive{} cases had a
strictly positive exact-output lower endpoint and
\RGSourceStageFirstCoalescenceUnresolved{} were unresolved. Thus the eight later
identical S3in arrays are inherited products of an earlier float32
coalescence, not evidence that the first coalescing LayerNorm reached the
real face. Table~\ref{rg:tab:first-coalescence-local-diagnostics} gives the
represented input energies, exact-output lower endpoints, and affine gain
and bias norms for each first-stage group.

At the final LayerNorm, \RGSourceStageUnresolvedZeros{} cases remained unresolved.
Exact pairwise dyadic evaluation of the represented float64 output arrays
corrects the rounded centered-sum classification.  The whole-network route
contains \RGSourceStageNetworkSixtyFourZeroCount{} represented zero,
\RGSourceStageNetworkSixtyFourOneUlpCount{} one-ulp near-collapses, and
\RGSourceStageNetworkSixtyFourPositiveCount{} positive maps beyond one ulp. Its exact
energies range from \RGSourceStageNetworkSixtyFourMinimum{} to
\RGSourceStageNetworkSixtyFourMaximum{}, with smallest strictly positive value
\RGSourceStageNetworkSixtyFourPositiveMinimum{}, while its rounded centered-sum
minimum is \RGSourceStageNetworkSixtyFourRoundedMinimum{}.  The replay from float32
S0 gives \RGSourceStagePipelineSixtyFourZeroCount{},
\RGSourceStagePipelineSixtyFourOneUlpCount{}, and
\RGSourceStagePipelineSixtyFourPositiveCount{} in the same three classes. Its exact
range is \RGSourceStagePipelineSixtyFourMinimum{} to
\RGSourceStagePipelineSixtyFourMaximum{}, its smallest strictly positive value is
\RGSourceStagePipelineSixtyFourPositiveMinimum{}, and its rounded minimum is
\RGSourceStagePipelineSixtyFourRoundedMinimum{}. Here \emph{one ulp} means that the
largest corresponding-coordinate separation between any two represented
output rows is one local float64 spacing.

The exact calculation is exact for each already rounded float64 array, not
for the ideal real computation that produced it.  Both float64 routes
therefore remain sensitivity controls.  Across the complete frozen checkpoint
set, \RGSourceStageDeviceDependentZeros{} final float32 zero statuses depended on
device. The result is bound to source commit \RGSourceStageSourceCommit{} and record
digest \RGSourceStageRecordDigest{}.
Table~\ref{rg:tab:stage-resolved-precision-controls} reports both routes.
Tables~\ref{rg:tab:observer-bands-precision} and
\ref{rg:tab:all-layer-large-increments} report observer-band occupancy, the
precision projection, and the all-layer increment census. In particular, large
increments are concentrated in layer 0 rather than absent outside the
registered layer-1 diagnostic.

The represented block medians and source counts in
Tables~\ref{rg:tab:confirmation-blocks} and \ref{rg:tab:confirmation-faces} remain
valid summaries of the stored tensor. Their terms ``face'' and ``source'' are
schema labels for represented arithmetic and are not promoted here to
analytical branch assignments.

\subsubsection{Source-remeasured matched-fiber test}
\label{rg:sec:matched-fiber-results}

The closure experiment starts from update 16 of \RGSourceClosureTrajectoryCount{} independently seeded
retained closure-source trajectories, one produced on each GPU. Three objects
are kept distinct: the recorder row restored from each archived checkpoint, an
independent source remeasurement after reconstructing the model and probe, and
each first-successor vector. The source locator is relative to an explicit data
root. Each reconstructed source is measured on its producing GPU, the other
GPU, and CPU. On both trajectories the producing-GPU vector matches all 48
stored float64 energy bit patterns; the other GPU also matches 48 of 48. CPU
changes 48 of 48 patterns, with maximum absolute defects
\(1.974\times10^{-4}\) and \(2.455\times10^{-4}\), and maximum symmetric
relative defects \(4.916\times10^{-8}\) and \(6.104\times10^{-8}\).
These CPU differences are a quantitative route control. Only the exact
producing-GPU source measurement licenses the matched-fiber inference.
Original source checkpoints were evaluated on all three routes; rewritten
branch copies were remeasured only on their producing GPU.

For each trajectory, \RGSourceClosureBranchCount{} rewritten resume copies produce
\RGSourceClosureBranchCount{} continuations, hence 32 continuations in total. The
registry contains two unchanged baselines; first-moment multipliers \(0.999\),
\(0.9\), and \(0\); second-moment multipliers \(0.999\), \(0.5\), \(2\),
and \(0\); a one-ulp change to one first-moment element; a \(+0.01\)
final-LayerNorm-bias control; full consistent, parameter-only, and moment-only
sign actions; and layer-0 consistent and parameter-only sign actions. Every
continuation emits checkpoints at steps 0 and 24. The exact source census thus
contains \RGSourceClosureCheckpointCount{} bound checkpoint files and 1,585,341,312
bytes.

Each continuation starts from the same declared incoming state, apart
from its specified intervention, and runs for 24 updates. Random states,
data position and every undeclared model and optimizer value are preserved.
The two unchanged baselines reproduce bitwise at all measured scales, so
numerical nondeterminism does not explain an intervention defect.

The analysis separates structural validity, inferential eligibility, and
scientific outcome. All eight structural checks pass, including the exact
checkpoint census, measured gauge support, source remeasurement, branch
registry, checkpoint rewrites, baseline reproducibility, and predecessor-leaf
retention. The energy-vector and raw-moment tests are separately eligible.
Ordinary branch-level comparison flags remain reported per scale but are not
promoted to structural gates. Every inferential branch first compares its
independently remeasured source vector with the stored recorder row. The
optimizer interventions and the one-ulp resolution control pass this source
gate exactly. The bias branch changes all 96 source bit patterns and cannot
enter a closure decision. At each scale \(b\), the exact decision rule for an
admissible pair is
\begin{equation}
 \mathtt{closure\_rejected}(b)
 \quad\Longleftrightarrow\quad
 \mathtt{bitwise\_different\_map\_count}(b)>0.
 \label{rg:eq:closure-bitwise-rule}
\end{equation}
This deterministic rule compares stored float64 energy bit patterns computed
from native float32 row maps. It uses no tolerance and is not a statistical
significance test. This represented pipeline is also the experiment's
detection floor.

In the zero-first-moment branch, all 96 trajectory-map successor energies
differ from baseline at all four scales. The aggregate maximum symmetric
relative defects for \(b=1,2,4,8\) are, respectively,
\(7.312\times10^{-3}\), \(1.151\times10^{-2}\),
\(1.562\times10^{-2}\), and \(2.716\times10^{-2}\).
The maximum absolute fiber defect is 29.07 at \(b=1\) and 107.97 at
\(b=8\). At the endpoint scales, the corresponding first-moment dose
values are \(6.887\times10^{-6}\) and \(1.839\times10^{-2}\) for
\(0.999m\), and \(6.810\times10^{-4}\) and
\(2.596\times10^{-2}\) for \(0.9m\). These values establish
finite-horizon sensitivity, not an optimizer beta function.

All four second-moment doses also change all 96 successors at every scale.
For the \(v=\bm0\) branch the aggregate maximum symmetric defects at
\(b=1,2,4,8\) are \(3.514\times10^{-2}\),
\(2.828\times10^{-2}\), \(2.416\times10^{-2}\), and
\(4.244\times10^{-2}\). The sequence decreases through \(b=4\) and then
increases, while dose order also changes with scale. The tabulated aggregate
maxima are finite-horizon summaries over two source states and 96 maps, not
trajectory-wise monotonicity results. Four scales do not identify a response
law, beta function, or critical exponent. The one-ulp first-moment branch
changes zero of 96 maps at every scale; this bounds represented sensitivity
and does not prove exact invariance. Tables~\ref{rg:tab:closure-source-gate},
\ref{rg:tab:energy-closure-probe}, \ref{rg:tab:second-moment-dose}, and
\ref{rg:tab:closure-sensitivity-controls} distinguish these roles.

By Equation~\eqref{rg:eq:deterministic-scale-lumpability}, the admissible first-
and second-moment defects reject an autonomous deterministic map on the
complete row-energy vector for this intervention domain at all
\RGSourceClosureRejectedScaleCount{} of the \RGSourceClosureScaleCount{} tested scales. In
the zero-first-moment branch, every registered map also has equal scalar source
energy and unequal successor energy, so scalar closure for that map fails on
its paired states.

The measured parameter gauge resolves the symmetry structure rather than
counting equivalent representatives as independent interventions. Its three
commuting layer involutions generate \((\mathbb Z/2\mathbb Z)^3\). Each
layer action covers 20 parameter tensors and 98,816 elements; the full action
covers 60 tensors and 296,448 elements. The support is derived independently
from both source checkpoints and agrees between trajectories. The full
consistent action co-transforms the attached AdamW first moments, preserves
the source row map and energy, and yields zero successor defect for all 96 maps
at every scale. The layer-0 consistent action likewise agrees bitwise with
baseline at every scale. These are the exact null controls predicted by
Proposition~\ref{rg:prop:gauge-orbit-null}.

The full parameter-only action applies the same parameter involution but leaves
both raw moment tensors unchanged. It therefore preserves the proposed source
observation \((e,m,v)\), yet changes all 96 successor energies at every scale.
The full moment-only partner flips the first moments on the same support, an
elementwise dose \(2\lvert m\rvert\), while leaving parameters fixed. Its
successor vector is bitwise identical to the parameter-only successor at every
scale, exactly as Corollary~\ref{rg:cor:paired-gauge-successor} predicts. It is a
paired orbit representative, not an independent physical direction. At
layer 0, the parameter-only action also differs from baseline for all 96 maps
at every scale, while the corresponding consistent action remains exact.
Tables~\ref{rg:tab:gauge-closure-control} and
\ref{rg:tab:gauge-product-decomposition} report these direct comparisons.

The sealed direct verdict for energy-vector closure is
\RGSourceClosureEnergyDecision{}, and the raw-moment augmentation verdict is
\RGSourceClosureRawMomentDecision{}. The latter is stored independently at
\codepath{gauge_control.raw_moment_augmentation_closure_status} and agrees
with the conclusion field. The uniqueness or minimality status is
\RGSourceClosureAugmentationStatus{}. Thus Lemma~\ref{rg:lem:augmented-fiber-witness}
rejects raw \((e,m,v)\) on the tested gauge domain without claiming that every
moment component is needed. A gauge-covariant optimizer coordinate, a quotient
state, or another statistic may still close on a stated domain. The
intervened states also need not lie on the unmodified training orbit, so
closure on a smaller reachable manifold remains undecided. The specified
complete-state kernel remains the unconditional predictive object.

\subsubsection{Executed staged predictive-state closure gate}
\label{rg:sec:staged-closure-results}

The staged gate uses \RGSourceStageOneTrajectoryCount{} independently seeded update-16
sources, with T0 and T1 fixed as design and T2 and T3 held out. It tests
\RGSourceStageOneCandidateCount{} prespecified candidates at block sizes
\(b\in\{\RGSourceStageOneScaleList{}\}\). The candidates are the complete 48-entry
energy vector; that vector together with the complete raw AdamW state and
exogenous phase; and all 48 observed row maps together with the complete
AdamW state after canonicalizing the measured layer-sign gauge and the same
phase. Equality here is equality of content fingerprints over every declared
coordinate, not proximity under a fitted metric.

The first two candidates are challenged by the full parameter-only sign
action. The strongest candidate uses a separate probe-null embedding branch.
For each source, the runner selects the smallest token occurring in the
immediate next-batch inputs but absent from every fixed-probe input, then adds
\(0.125\) to coordinate zero of that token's input-embedding row. The
intervention changes exactly one model coordinate. The fixed probe does not
read it, and neither the optimizer state nor phase is changed. Consequently,
the complete observed row maps, gauge-canonical optimizer state, and phase
remain bitwise identical at the source even though the complete transition
state differs. The next batch reads the selected token once or twice,
depending on the trajectory.

The forward-observation trust boundary is explicit. The runner performs the
model forward remeasurements and seals the row-map hashes and source-energy
array. The analyzer authenticates those runner outputs rather than repeating
their forward passes. It independently reloads source and rewritten
checkpoints, reconstructs candidate fingerprints from the authenticated
measurements, replays the frozen token selection and exact one-coordinate
tensor dose, compares continuation arrays and terminal checkpoint state, and
derives all scale, split, candidate, control, promotion, and terminal fields.

Every candidate has an exact same-fiber source in each trajectory. For every
candidate, all 48 successor maps differ at each of the five scales in both
design trajectories and again in both held-out trajectories. For the
strongest candidate, the maximum symmetric relative defect over a
trajectory-scale comparison ranges from \(3.119\times10^{-4}\) to
\(2.931\times10^{-2}\). Table~\ref{rg:tab:staged-predictive-closure} reports the
split decisions. Thus Lemma~\ref{rg:lem:augmented-fiber-witness} rejects every
prespecified reduced state on this intervention domain without relying on a
tolerance or a trained predictor.

Three independent controls delimit that conclusion. Unchanged continuation
reproduces all 33 recorded states and all 48 maps bitwise for every trajectory.
The consistent complete-state gauge quotient has equal canonical source and
final transition states and zero differing successor maps at all five scales,
giving status \RGSourceStageOneGaugeStatus{} on this tested domain. Direct 16-step
continuation is bitwise identical to an 8-step continuation, save and reload,
followed by another 8 steps.

The promotion count is \RGSourceStageOnePromotedCount{} and the terminal stage status
is \RGSourceStageOneStatus{}, recorded literally as
\codepath{COMPLETED_STOPPED_AT_CLOSURE}. Therefore no local reduced flow,
critical surface, critical exponent, or empirical universality calculation is
licensed by this stage. Those downstream fits were not performed. This is the
executed stopping decision, not a pending confirmation program. The
intervention pair need not belong to the unmodified training orbit, so the
result does not rule out closure on a smaller invariant reachable manifold.
Nor does the exact complete-state quotient control prove a nontrivial
reduction; it verifies the represented symmetry and deterministic replay on
the four tested sources.

\subsubsection{Implemented PLGA checkpoints}
\label{rg:sec:plga-checkpoint-results}

PLGA diagnostics were evaluated at initialization, burn-in, design end, and
the final state for every registered head. The median constant-input defect
norm changed from \(\RGSourceConfirmInitialPLGADefectMedian{}\) initially to
\(\RGSourceConfirmFinalPLGADefectMedian{}\) finally. Every one of the 192 maps at
every checkpoint exceeded the direct \(10^{-6}\) defect tolerance. Even the
smallest final defect was \(3.202\times10^{-5}\). Direct face preservation
therefore failed at every checkpoint
(Table~\ref{rg:tab:confirmation-plga}). The largest sampled operator-norm
estimate was \(\RGSourceConfirmSampledOperatorMaximum{}\), attained at burn-in; it is
a
grid-based lower bound on the segment supremum in
Theorem~\ref{rg:thm:plga-rg-transfer}, not an upper certificate. The five
structural residuals in Table~\ref{rg:tab:confirmation-plga-structural} also
show that the jointly sufficient parameter conditions are not satisfied.

Constructed constrained heads provide the positive control: their maximum
constant-input defect was \(\RGSourceConfirmControlDefectMaximum{}\). The contrast
with the trained heads rejects pointwise row-constant face preservation for
the implemented PLGA maps at the measured checkpoints. It does not reject the
conditional transfer theorem; it rejects its zero-defect premise on this
domain.

\subsubsection{Implementation and numerical integrity}

Identity evaluations are kept separate from empirical decisions. Exact
chronological block replay had maximum relative residual
\(\RGSourceConfirmSemigroupResidual{}\), and the positive-tail cumulative-log
telescoping check had maximum relative residual
\(\RGSourceConfirmCumulativeResidual{}\). The PLGA response-plus-defect
decomposition residual was at most
\(\RGSourceConfirmDecompositionResidual{}\). These are integration-test results,
not experimental support for mathematical identities.

For the final LayerNorm gate-shape factorization, the well-conditioned
full-map backward residual was at most \(\RGSourceConfirmGateShapeResidual{}\), below
the frozen \(2\times10^{-6}\) tolerance. The absolute quotient-energy
residual was \(\RGSourceConfirmGateShapeAbsoluteResidual{}\). Its symmetric relative
counterpart reached \(\RGSourceConfirmGateShapeSymmetricResidual{}\), as expected
when the quotient norm approaches or reaches zero; that ill-conditioned
relative quantity is descriptive and was not an integrity gate.

\subsubsection{Empirical decision ledger}

Table~\ref{rg:tab:evidence-decision-ledger} is generated jointly from the
retained summary, current analysis, observer record, stage-resolved replay,
source-remeasured matched-fiber tests, and the staged candidate-state gate.
It keeps implementation identities outside the empirical claim ledger.

The registered negative and inconclusive outcomes refine the scope of the
theory. The exact affine state organization applies to every exact
branch, while the interval RG carries unresolved observer states without
inventing a source. No uniform PLDR phase or asymptotic universality class
follows on this domain. The PLGA measurements further show that downstream face
preservation is an additional parameter-dependent property rather than a
consequence of upstream row-map geometry. The staged closure decision is
terminal: every reduced candidate is rejected in both splits, so no downstream
reduced criticality or universality fit enters the evidence ledger.

\FloatBarrier
\setcounter{topnumber}{4}
\par\medskip\noindent
The closure tests motivate retaining a complete conditional law.
Chapter~\ref{ch:retained-law} develops that model-wide object and states
what exact elimination and approximate predictive reduction preserve.

\part{Complete laws and consuming-corpus training}
\chapter{The law retained under elimination}
\label{ch:retained-law}
This chapter defines the model-wide law retained under decoder, head,
observation and training-state elimination. It links complete-state kernels
to inference emissions and states compatibility conditions and finite error
budgets for smaller predictive descriptions.

\section{The object being renormalized}
\label{model:sec:scope}

A renormalization group (RG) description specifies a scale, retained
variables, a map and a law or dynamical state on which the map acts.
A power-law operation inside a network does not select these objects.
PLDR-LLMs use generated operators coupling queries and keys in every
attention head \cite{gokden2021,gokden2024,gokden2025,gokden2026soc,gokden2026foundations}.
Heads interact through output mixing, residual paths, normalization and
feed-forward networks. Their separate scalar norms therefore do not
specify a model-wide predictive state.

The objective is a comprehensive renormalization theory connecting training
and inference, conditioned on a specified pretraining-data law. We construct
that connection on an augmented training process and its joint inference
emission. The exact maps preserve the full law. Practical reductions carry
explicit closure defects, conditional Taylor remainders and a hierarchy of
connected covariance sectors. This formulation supplies a common language
for regular, heterogeneous and conditionally critical regimes; the
empirical scope of each regime is stated separately.
Sections~\ref{model:sec:maps}, \ref{model:sec:reduction} and \ref{model:sec:response} construct the maps, quantify closure error and derive predictive response, respectively.

\subsection{From a complete law to a predictive theory}
\label{model:sec:law-route}
The scientific chain begins with the ordered native PLGA computation,
then its complete source-conditioned training state, compatible scale
maps, the architecture-specific sign symmetry, training-selected regimes
and predictive emission. Four categories distinguish the resulting exact
identities, conditional reductions, native symmetry class and completed
measurements.
Let $S_t=(\theta_t,m_t,v_t,\sigma_t,\mathcal R_t,\xi_t)$ contain the
parameters, both Adam moments, scheduler and bias clocks, remaining corpus,
and any additional sampler or stochastic-program state. The conditioning
specification fixes the corpus law or realized corpus, tokenizer, supervised
block construction, initialization ensemble, optimizer and observation law.
The transition $K_t$ and proper-prefix emission $\mathcal E$ then define
the complete chronological experiment.

The complete conditioned law is
\begin{equation}
 \mathbb P_{\mathcal C_N}(dS_{0:T},dy)
 =\rho_{0,N}(dS_0)\prod_{k<T}K_{k,N}^{\mathcal C_N}(S_k,dS_{k+1})
   \mathcal E_N^{\mathcal D_{\rm ev}}(S_T,dy).
 \label{model:eq:canonical-conditioned-law}
\end{equation}
Conditioning on a realized corpus and order differs from averaging over
corpora. Every exact scale map in this monograph acts on this same joint law;
its compatibility retains shared randomness, source insertions and aligned
block boundaries. A separately trained smaller network requires an
additional model-matching hypothesis.

\begin{table}[htbp]\centering\small
\caption{Notation for the conditioned family and its scales.}
\label{model:tab:canonical-notation}
\begin{tabular}{@{}p{.25\textwidth}p{.68\textwidth}@{}}
\toprule Symbol & Meaning\\\midrule
$\mathcal D_{\rm tr},\mathcal D_{\rm ev},\mathcal C_N$ & Training law, evaluation law and declared conditioning.\\
$S_k,\rho_k,K_k,\mathcal E$ & Complete state, its law, transition and inference emission.\\
$P^{\rm pot},g,J$ & Learned row potential, training control and statistical reweighting source; these have distinct roles.\\
$N,L,d$ & Head count, decoder depth and head dimension.\\
$k,h_N,\tau=kh_N$ & Update count, physical time per update and physical age.\\
$b_{\rm doc},b_t,B_N$ & Document scale, temporal blocking factor and optimizer batch size.\\
$\ell,M_N,q_N$ & Proper-prefix length, source resource and consumed fraction.\\
$\kappa_\chi,\zeta,\alpha$ & Conditional susceptibility exponent, update-clock gap exponent and low-frequency forcing exponent.\\
\bottomrule
\end{tabular}
\end{table}

\begin{table}[htbp]\centering\scriptsize
\caption{Status of the model-wide construction. Each row separates an exact map, its additional analytical hypotheses, completed evidence and an unresolved property.}
\label{model:tab:theory-status}
\begin{tabular}{@{}p{.22\textwidth}p{.23\textwidth}p{.23\textwidth}p{.23\textwidth}@{}}
\toprule Exact statement & Conditional result & Completed measurement & Open property\\\midrule
Complete-state composition (Section~\ref{model:sec:maps}) & Evolving-law successor and emission bound (Proposition~\ref{model:prop:law-closure}) & Recorded single-pass paths (Section~\ref{model:sec:matched-clock-results}) & Economical autonomous training state\\[4pt]
Row and temporal-energy maps (Propositions~\ref{model:prop:collective-clock}, \ref{model:prop:temporal-energy}) & Positive-energy flux-error bound (Proposition~\ref{model:prop:finite-flux-forecast}) & Full common-duration path grid (Section~\ref{model:sec:equal-time-flux}) & Incoming forecast and free rollout\\[4pt]
Native sign symmetry (Proposition~\ref{model:prop:native-head-sign-symmetry}) & Covariance-conditioned mixture limit (Theorem~\ref{model:thm:native-sign-gaussian-mixture}) & Whole-initialization width diagnostics (Section~\ref{model:sec:critical-independent-results}) & Native invariant-sector convergence\\[4pt]
Fixed-reference projection (Proposition~\ref{model:prop:finite-density}) & Error relative to the fluctuation scale & State-conditioned predictive reduction (Section~\ref{model:sec:categorical-scale-results}) & Persistence of a critical mode in inference\\[4pt]
Finite-population transport (Proposition~\ref{model:prop:consuming-bridge}) & Specified physical-clock limit (Corollary~\ref{model:cor:consuming-clock}) & Matched-memory consuming-source family (Section~\ref{model:sec:matched-clock-results}) & Critical exponents and endogenous critical selection\\
\bottomrule\end{tabular}\end{table}

\small
\begin{longtable}{@{}p{.17\textwidth}p{.39\textwidth}p{.36\textwidth}@{}}
\caption{Principal claims on the common conditioned law. The detailed view adds replication units, costs and reconstruction routes from the same maintained records.}\label{model:tab:claim-index-main}\\
\toprule Claim & Assumptions and written result & Complete evidence and remaining obligation\\\midrule\endfirsthead
\toprule Claim & Assumptions and written result & Complete evidence and remaining obligation\\\midrule\endhead
\hyperref[model:claim:C1]{C1: Compatible complete laws} & One conditioned law; full parameters, optimizer memory, remaining identities, clocks and shared observation measure. Written basis: \ref{model:sec:maps}, \ref{model:sec:data-resource}. & Recorded single-pass histories and native replay. Complete evidence: \ref{model:sec:matched-clock-results}. An economical retained state needs an additional successor law.\\[5pt]
\hyperref[model:claim:C2]{C2: Native signed class} & Head-sign invariant initialization and optimizer, uniformly bounded signed operators, convergent orbit-covariance law. Written basis: \ref{model:prop:native-head-sign-symmetry}, \ref{model:thm:native-sign-gaussian-mixture}. & Complete operator observations distinguish signed and invariant fluctuations. Complete evidence: \ref{model:sec:critical-independent-results}. Trained covariance convergence; deterministic Gaussian behavior additionally needs covariance concentration.\\[5pt]
\hyperref[model:claim:C3]{C3: Conditional training reduction} & Evolving-law successor defect, coarse stability, initial-law discrepancy and emission error. Written basis: \ref{model:prop:law-closure}, \ref{model:prop:finite-optimizer}. & Moment-only interventions and source-return observations retain memory and representation errors. Complete evidence: \ref{model:app:finite-native-transport}. Held-out multistep successor accuracy and total cost advantage for an autonomous reduction.\\[5pt]
\hyperref[model:claim:C4]{C4: Finite collective row flow} & Nonzero endpoint energies, fixed averaging, actual chronological increments; derivative bounds additionally need small relative increments. Written basis: \ref{model:prop:collective-clock}, \ref{model:lem:finite-defect}, \ref{model:cor:row-path-error}, \ref{model:prop:finite-flux-forecast}, \ref{model:prop:energy-cocycle}, \ref{model:prop:temporal-energy}. & Eight single updates, sixteen fixed-count continuations and sixteen equal-duration continuations resolve finite transport and energy-coordinate blocking. Every derivative diagnostic is retained. Complete signed temporal energy is reconstructed independently by ordered pair products on all 64 equal-duration cells. Complete evidence: \ref{model:sec:collective-flux-results}, \ref{model:sec:row-path-results}, \ref{model:app:row-quadrature}, \ref{model:sec:equal-time-flux}. A flux successor law and uniform accumulated generator-remainder control.\\[5pt]
\hyperref[model:claim:C5]{C5: Source and physical clocks} & Specified width, resource and time family and matched Adam memory; reference covariance uses a frozen population and deterministic transports. Written basis: \ref{model:prop:matched-memory-force}, \ref{model:prop:consuming-bridge}, \ref{model:cor:consuming-clock}. & 48 paths at four widths and four controls; complete 64 field and 26 cache cells. Complete evidence: \ref{model:sec:matched-clock-results}. A trained limit and native reference-remainder estimates; independent corpus variability.\\[5pt]
\hyperref[model:claim:C6]{C6: Predictive and cache reduction} & Common reference for nested observation; state-specific calibration, complete context-risk and operator-displacement budgets. Written basis: \ref{model:prop:finite-density}, \ref{model:prop:tail-risk}. & All stated state-calibrated aggregate targets pass on the two disjoint state panels; unchanged-initial-cache controls and context exceptions remain visible. Complete evidence: \ref{model:sec:cache-state-results}, \ref{model:app:cache-state-details}. Population guarantees and justified downstream sensitivity; workload-specific timing does not establish general throughput.\\[5pt]
\hyperref[model:claim:C7]{C7: Conditional critical scaling} & Declared resource limit, physical response and forcing scales, and nonvanishing predictive coupling. Written basis: \ref{model:cor:colored-exponents}, \ref{model:prop:selection-window}. & Finite kinetic observations and covariance/response controls delimit the applicable conditional theory. Complete evidence: \ref{model:sec:critical-independent-results}. Native critical exponents, a critical surface, restoring feedback and inference persistence remain unestablished.\\[5pt]
\bottomrule\end{longtable}\normalsize

The principal outcomes have a short dependency chain. Native sign
invariance and the operator envelope imply the regular signed class
(Proposition~\ref{model:prop:native-head-sign-symmetry} and
Theorem~\ref{model:thm:native-sign-gaussian-mixture}); complete operator
observations test its finite normalization
(Section~\ref{model:sec:critical-sign-checks}). Full-state covariance transport
and specified clocks lead to the finite kinetic predictions tested on the
independent family (Section~\ref{model:sec:critical-independent-results}).
Fixed-reference density projection then gives compatible predictive scales
(Propositions~\ref{model:prop:finite-density}--\ref{model:prop:tail-risk}), evaluated
at three training ages and on three disjoint context panels in
Section~\ref{model:sec:categorical-scale-results}. The complete-state law is the
common foundation of these outcomes. A finite-population chronological
covariance and its clock limit additionally describe how consuming a
source modifies reference fluctuations (Section~\ref{model:sec:consuming-bridge}).
The matched physical-clock family in Section~\ref{model:sec:matched-clock-results}
then measures native training and inference at a fixed resource fraction.

Exact composition need not give an economical representation. An observation
map needs a separate successor law before it becomes an autonomous reduced
dynamical state. Its approximation is assessed under the evolving fine law.
For example, if $\bar K_t$ is $L_t$-Lipschitz in
$W_1$, its law-averaged defect is
\[
 d_t=\int W_1(c_\#K_t(s,\cdot),\bar K_t(c(s),\cdot))\,\rho_t(ds).
\]
The projected-law discrepancy obeys $e_{t+1}\le L_te_t+d_t$;
iteration gives Proposition~\ref{model:prop:law-closure}, including
initial-law error and the subsequent emission error.
A small empirical moment score alone estimates neither $d_t$ nor $L_t$.

Temporal, decoder, document and linear observation maps act on this
same declared experiment. Commuting diagrams require the indicated
compatibility assumptions. Changing native width defines a family of
experiments; it is not an exact projection to a smaller PLDR network.
In a single pass, a remaining-block count records resource age but
generally does not identify the remaining-block law. This is why the
state retains the resource itself. The PLDR collective variables then
retain shared learned metric fields, common and transverse row sectors,
body and generator interactions, optimizer memory and their predictive
projections. The closure and visibility bounds specify which of those
variables a validated reduction may discard.

\begin{figure}[htbp]\centering
\begin{tikzpicture}[>=Stealth,
 box/.style={draw=MidnightBlue!60,rounded corners,align=center,
 text width=5.0cm,minimum height=1.1cm,font=\small},
 node distance=1.15cm and 1.1cm]
\node[box] (state) {Single-pass corpus law and complete state\\parameters, moments, remaining blocks};
\node[box,right=of state] (law) {Chronological and decoder maps\\exact joint-law composition};
\node[box,below=of state] (reduced) {Finite effective variables\\closure defect and acquisition cost};
\node[box,right=of reduced] (emission) {Native proper-prefix emission\\predictive norm and visible covariance};
\node[box,below=of reduced] (physical) {Specified physical source\\Gram inversion and fitted common rows};
\node[box,right=of physical] (transfer) {Conditional scale transfer\\relative moments and RG intertwining};
\draw[->] (state)--(law);
\draw[->] (state)--(reduced);
\draw[->] (law)--(emission);
\draw[->] (reduced)--(emission);
\draw[->] (physical)--(transfer);
\draw[->] (emission)--(transfer);
\end{tikzpicture}
\caption{Dependencies of the conditional theory. Exact composition,
finite predictive closure, physical readout accuracy and fixed-point
intertwining have separate hypotheses. The completed physical
observations test the lower observation route; the single-pass native
trajectories test training-selected emissions and conditional memory.
A critical training-selection mechanism additionally requires the
shrinking-window conditions of Proposition~\ref{model:prop:selection-window}.}
\label{model:fig:conditional-dependencies}
\end{figure}
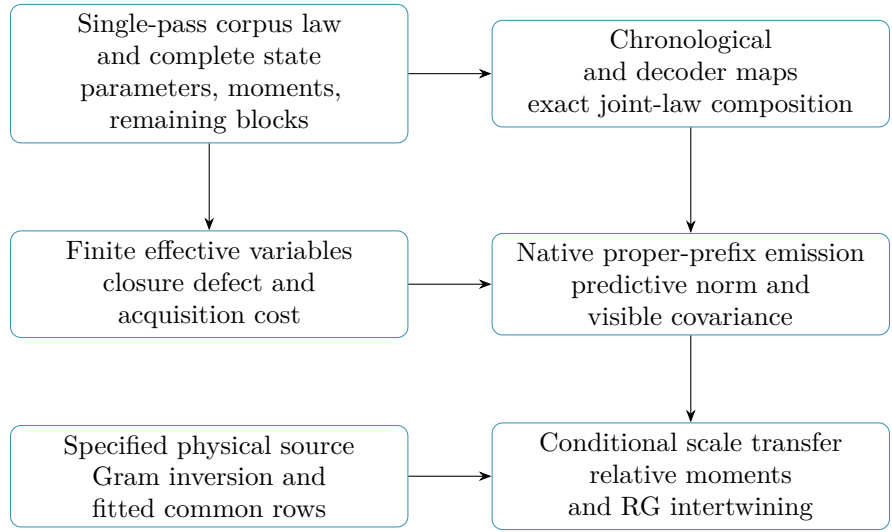

The primary training law is a single pass through a specified RefinedWeb
corpus \cite{penedo2023}. Its conditioning variables include tokenization,
document selection, target-block construction, the nonrepeated ordering,
the consumed fraction and the complete learning-rate schedule. Distinct
source target positions, rather than distinct vocabulary symbols, define
nonrepetition. The remaining corpus is part of the augmented RG state
in Section~\ref{model:sec:data-resource}. Repeated sampling from a small corpus
provides an auxiliary data-law comparison; fitted coefficients under that
law are not transferred to the single-pass process without matched evidence.
Fresh length-qualified evaluation cohorts and recorded controlled training
do not reconstruct the historical ordered pretraining stream of released
checkpoints. The two released models characterize trained emissions, while
controlled initializations permit recorded training and optimizer interventions.
Section~\ref{model:sec:onepass-results} gives the matched data-law comparison, and Section~\ref{model:sec:methods} specifies the released-model observations.

``Model-wide'' means that the analytical state can retain every forward
tensor, parameter, optimizer coordinate and source. The experiments observe
all decoders and heads and terminate at the full vocabulary distribution.
The 174-dimensional field, head-gain sources and 36 common training
coordinates, together with the bounded per-head entropy and row fields,
are specified projections of that state. They are not a
measurement of every tensor entry or the full parameter Hessian.
These projections are defined in Sections~\ref{model:sec:joint-observation-methods}, \ref{model:sec:whole-model-response-results} and \ref{model:sec:controlled-training-methods}, respectively.

The empirical argument includes complementary finite tests. Matched single-pass paths
measure training selection; moment-only interventions test the need for
optimizer memory; source selection and common-source return test the
representation and transport of full-vocabulary effects. The return
projection in Proposition~\ref{model:prop:projection-obstruction} separates a
fixed dictionary's irreducible component from the excess error of its
acquired coefficients. This makes representation adequacy, successor-law
closure and acquisition cost distinct obligations. The common claim index records these obligations and their acquisition costs.
The matched paths, moment interventions and common-source return are detailed in Sections~\ref{model:sec:matched-onepass}, \ref{model:sec:optimizer-results} and \ref{model:sec:finetuning-return-results}.

\subsection{Contributions and related work}
The physical row quotient, realized energy cocycle, optimizer-compatible
sign gauges and limits of scalar row closure are developed in Parts I
and II. Their \emph{data-conditioned, model-wide} connection to coupled decoder
sources and predictive observations, including finite-corpus transport,
optimizer memory, measured training selection and quantitative
reduction errors. Learned-operator caching and benchmark preservation
under freezing originate in the PLDR inference study
\cite{gokden2025}; here we quantify causal-prefix distinctions,
conditional operator errors and answer-margin preservation.

The maintained claim index in Table~\ref{model:tab:claim-index-main} connects
these architectural contributions to their assumptions, stand-alone
arguments and complete evidence. The same records generate the detailed
replication and cost view in Appendix~\ref{model:app:evidence-scope}.

The dependency route in Section~\ref{model:sec:law-route} connects these
contributions to the decisive native measurements. The \hyperref[model:app:supplement-guide]{Guide to the monograph}
(page~\pageref{model:app:supplement-guide}) locates complete resolution tables,
resource inventories and detailed proofs without interrupting that route.

Classical RG and its probability formulation identify scale transformations
and limiting laws \cite{wilson1974,jona2001}. Variational RG mappings and
information-preserving coarse graining concern explicit retained information
\cite{mehta2014,koch2018}; their transformations do not establish that the
PLGA generator is a physical RG map. Effective neural-network theories
likewise specify an ensemble and scale \cite{roberts2022,erdmenger2022}.
Wilsonian descriptions of neural-network distributions and finite-width
corrections also precede this construction \cite{halverson2021}.
Infinite-head attention limits \cite{hron2020} and adaptive-optimizer limits
\cite{yang2023adaptive} provide comparison classes with their own observables,
normalizations and horizons. The native signed-operator result in Section~\ref{model:sec:native-sign-limit}
does not identify those predictive-function limits.
Infinite-width kernel and feature-learning limits depend on parameterization
and learning-rate scaling \cite{jacot2018,yang2021}; those hypotheses cannot
be imported unchanged into a learned power generator.

Token-representation rank collapse in pure self-attention concerns
network outputs with residual and feed-forward mechanisms removed
\cite{dong2021rankcollapse}. The PLDR row quotient concerns an internal
generated operator in a decoder retaining those mechanisms. These
state spaces and hypotheses differ; the former result does not
establish the latter training flow.

Decoder composition and source-jet composition are structural lemmas with
no probabilistic content. Document cumulant flow and the Gaussian class use
classical independent-sum probability \cite{kallenberg2021}. Training kernel
composition is Chapman--Kolmogorov; exact reduced closure is lumpability
\cite{kemeny1976}. Eliminating unresolved coordinates generally creates
memory \cite{chorin2002}. The predictive metric is the Fisher information
of model probabilities \cite{amari1998,martens2020,kunstner2019}. The
architecture-specific content resides in the native parameterization and
independent head-sign action, its optimizer equivariance, cross-head
covariance constraints, and measured transport to predictive emission.
Conditional-expectation, total-covariance and Markov-inequality identities
are classical tools; their PLDR observation laws and empirical domains are
specified here.

Tensor concentration, predictive sensitivity and text susceptibility have
different variables and scales. The Gaussian independent-document class
is shared by bounded observations of other finite models. A persistent
shared component has another block normalization, while a critical width
limit additionally needs compatible normalization and a singular connected
sector. The finite evidence in Sections~\ref{model:sec:critical-independent-results}
and \ref{model:sec:scaling-results} does not establish that latter limit or
endogenous attraction to its critical surface.

The model-wide state retains the variables needed to specify the complete
conditioned training and inference process. Exact map compatibility,
conditional limiting statements and measured finite reductions have
different evidential roles. The signed-operator concentration and
Gaussian-mixture class require their stated ensemble and convergence
hypotheses. Finite row transport and state-calibrated predictive reductions
are supported on the declared observation domains. Establishing an
economical reduced training law, an independently identified critical
surface, endogenous approach to that surface, and the survival of its
critical fluctuations in inference requires further analysis and
independent confirmation.

\section{Exact model-wide renormalization maps}
\label{model:sec:maps}

\subsection{Decoder elimination with retained source insertions}
For a decoder boundary $u$, write a block as $(F,w)$, where $F$ is its
actual nonlinear boundary map and $w(u;j)$ sums source couplings to all
observables inside the block. The associated transfer operator acts on a
terminal function $f$ by
\begin{equation}
 (\mathcal T_{F,w}f)(u)=e^{w(u;j)}f(F(u)).\label{model:eq:transfer}
\end{equation}
Integrating this function against the embedding law induced by $\D$ gives
the complete source partition function. The construction is valid also
when deterministic intermediate tensors are regarded as delta constraints
in a joint graph measure.

\begin{lemma}[Exact elimination of decoder boundaries]\label{model:thm:graph}
Chronologically compose two blocks by
\begin{equation}
 (F_2,w_2)\circ(F_1,w_1)
   =\left(F_2\circ F_1,\ w_1+w_2\circ F_1\right).\label{model:eq:weighted-block}
\end{equation}
This operation is associative with identity $(\operatorname{id},0_{\mathrm{fun}})$.
It preserves every source partition function and every terminal
prediction of the fine computation. Grouping a finite decoder chain into
blocks and subsequently regrouping those blocks gives the same result as
the corresponding single grouping of the fine chain.
\end{lemma}
\begin{proof}
Substitution in~\eqref{model:eq:transfer} gives the endpoint
$F_2(F_1(u))$ and the exponent $w_1(u)+w_2(F_1(u))$. For three blocks,
either association order has endpoint $F_3(F_2(F_1(u)))$ and exponent
\[
 w_1(u)+w_2(F_1(u))+w_3(F_2(F_1(u))).
\]
The identity follows by substitution. The equality holds pointwise in
$u$ and the sources, hence also after integration against any input law.
Induction proves the grouping statement for every finite chain.
\end{proof}

This exact map is determined from the executable decoder functions, rather
than from ratios of observed endpoint energies. It generally creates a
more complicated effective block; it does not assert that the block is
another PLDR decoder with fewer weights. A closed law or transfer
operator is a larger state space than a fixed architectural ansatz.

\subsection{Joint-law coarse observation and data blocking}
Let $C:\R^p\to\R^r$ be a fixed linear observation, such as retaining or
averaging selected head fields. Let $\nu_b$ be the joint law of $b$ ordered
fine fields $\varphi_1,\ldots,\varphi_b$. For a scale exponent $H$, define
\begin{equation}
 B_{b,C,H}(\varphi_1,\ldots,\varphi_b)
       =b^{-H}C\sum_{i=1}^b\varphi_i,
 \qquad \RG_{b,C,H}\nu_b=(B_{b,C,H})_\#\nu_b.\label{model:eq:general-map}
\end{equation}
This is an exact reduction of a law and retains the input correlations
through $\nu_b$. In the independent-context case,
$\nu_b=\mu^{\otimes b}$ and we abbreviate the map by $\RG_{b,C,H}\mu$.

\begin{theorem}[Compatible law RG]\label{model:thm:law-rg}
On aligned blocks, with compatible linear maps $C,D$,
\begin{equation}
 B_{c,D,H}\bigl(B_{b,C,H}(\cdot),\ldots,B_{b,C,H}(\cdot)\bigr)
       =B_{bc,DC,H}(\cdot).\label{model:eq:block-semigroup}
\end{equation}
The same identity holds for pushforward laws if their complete joint
block dependence is retained. For independent draws from one law with
log moment generating function $W$,
\begin{equation}
 (\RG_{b,C,H}W)(j)=b\,W(b^{-H}C^{\mathsf T}j).\label{model:eq:rg-cgf}
\end{equation}
In particular,
$\RG_{c,D,H}\RG_{b,C,H}W=\RG_{bc,DC,H}W$.
\end{theorem}
\begin{proof}
The left side of~\eqref{model:eq:block-semigroup} is
\[
 c^{-H}D\sum_{a=1}^c b^{-H}C\sum_{i=1}^b\varphi_{a,i}
 =(bc)^{-H}DC\sum_{a,i}\varphi_{a,i}.
\]
Pushforwards preserve composition, proving the dependent-law statement.
For independent fields the moment generating function factors into $b$
identical factors,
\[
 \E\exp\left(j^{\mathsf T}b^{-H}C\sum_i\varphi_i\right)
 =\left[\E\exp\left((b^{-H}C^{\mathsf T}j)^{\mathsf T}\varphi\right)\right]^b.
\]
Taking logarithms proves~\eqref{model:eq:rg-cgf}. A second substitution gives
$bcW((bc)^{-H}(DC)^{\mathsf T}j)$, as required.
\end{proof}

If consecutive block fields remain dependent, replacing their joint law by
the product of their marginals changes the RG. A marginal alone does not
determine a dependent block sum. The appropriate closed state is the
finite joint law or, for unbounded sequences, its consistent family of
joint laws. This is the probabilistic reason that correlations between
heads and correlations between text positions must be kept separately.

\subsection{What head or state elimination retains}
The graph map and the law map are compatible: compute the same retained
fields using Lemma~\ref{model:thm:graph}, then apply any linear observation and
the law map of Theorem~\ref{model:thm:law-rg}. The result is identical to observing
and blocking the full fine computation. This statement follows pointwise
before averaging over $\D$ and includes source derivatives wherever they
exist.

There is also a useful closure warning. For a linearized boundary state
split into retained $r_t$ and unresolved $u_t$,
\begin{equation}
 r_{t+1}=Ar_t+Bu_t,\qquad u_{t+1}=Cr_t+Du_t,\label{model:eq:unresolved}
\end{equation}
direct elimination gives, for $t\geq0$,
\begin{equation}
 r_{t+1}=Ar_t+BD^tu_0+
             \sum_{k=0}^{t-1}BD^{t-1-k}Cr_k.\label{model:eq:memory}
\end{equation}
Indeed, induction gives
$u_t=D^tu_0+\sum_{k=0}^{t-1}D^{t-1-k}Cr_k$; substituting it proves the
formula. The memory kernels $BD^sC$ vanish only under additional structure.
This elementary finite counterpart of projection-based memory
\cite{chorin2002} explains why a collection of head norms cannot generally
replace the full state in a predictive recurrence.

\section{Training dynamics and the inference law they select}
\label{model:sec:training}

\subsection{An augmented state and two different scale coordinates}
Let $s_t=(\theta_t,o_t,r_t,t)$ contain every trainable parameter, the
optimizer variables, the data-sampler state, and the step counter. For
Adam-type updates, $o_t$ includes both moment tensors; the counter fixes
bias correction and the learning-rate schedule. A sampled training
innovation $\zeta_t$ determines a complete update
\begin{equation}
 s_{t+1}=U_t(s_t,\zeta_t),\qquad
 (K_tf)(s)=\E[f(U_t(s,\zeta_t))\mid s_t=s].\label{model:eq:training-kernel}
\end{equation}
The conditional innovation law includes the batch selection rule and any
stochastic model operations. Deterministic shuffled streams can instead
retain their permutation, cursor, and random-generator state. All formulas
below apply to a specified training protocol and require their displayed
integrals to exist. They do not posit an equilibrium Boltzmann law for
AdamW \cite{kingma2015adam,loshchilov2019adamw}.

For forward kernels our convention is
\[
 K_{s,u}(x,B)=\int K_{t,u}(y,B)\,K_{s,t}(x,dy),\qquad s<t<u.
\]
The associated observable operator is $(Kf)(x)=\int f(y)K(x,dy)$.
Thus chronological observable products are $K_{s,t}K_{t,u}$, whereas
state maps act with the later map on the left, $F_{t,u}\circ F_{s,t}$.
This also fixes the order of the affine edges in
Equation~\eqref{eq:overview-compose}.

Write $\rho_t$ for the distribution of the augmented state across training
realizations. The inference emission kernel is
\begin{equation}
 (\mathcal E f)(s)=\E_{(x,y)\sim\D}
                f\bigl(\Phi_{\theta(s)}(x,y)\bigr),\qquad
 \mu_t=\rho_t\mathcal E,
 \qquad \rho_{t+1}=\rho_tK_t.\label{model:eq:emission}
\end{equation}
Here inference contexts are independent of training randomness
conditional on $s$. The observation can be the entire forward trace.
For a single realized checkpoint, conditioning on $s_t$ recovers the
fixed-weight inference law of Section~\ref{model:sec:model}. The data used to
construct $K_t$ and to define $\mathcal E$ must be declared separately,
even when both are drawn from the same source corpus.

The reference-family single-pass experiments, including source selection
and finite optimizer transport, use the native all-nonpadding-target
objective, coordinatewise gradient-value clipping at one, AdamW moments
$(\beta_1,\beta_2)=(0.9,0.95)$, offset $10^{-5}$ and weight decay $0.1$.
The controlled external-target family uses global gradient-norm clipping
at radius one, the same moment coefficients, offset $10^{-8}$ and
weight decay $0.01$. Applied rates and scheduler phase are specified for
each family in Sections~\ref{model:sec:onepass-corpus-methods}
and \ref{model:sec:critical-onepass-results}. Writing $g_t$ for the
family-specific clipped gradient, the
coordinate update is
\begin{alignat}{2}
 m_{t+1}&=\beta_1m_t+(1-\beta_1)g_t, &\quad v_{t+1}&=\beta_2v_t+(1-\beta_2)g_t^2,\notag\\
 \theta_{t+1}&=(1-\eta_t\lambda)\theta_t-
 \eta_t\frac{m_{t+1}/(1-\beta_1^{k_t+1})}
 {\sqrt{v_{t+1}/(1-\beta_2^{k_t+1})}+\epsilon}.
 \label{model:eq:family-adamw}
\end{alignat}
Here $k_t$ is the relevant moment counter; resetting it changes the law.
The clipping operation acts on the full native gradient before any
retained-coordinate projection. These objective, clipping and offset
conventions belong to the definition of $K_t$, rather than universal
constants of the RG construction.

Optimization time $t$ changes the learned system. A temporal block length
$m$ changes its resolution, and the document scale $b$ changes the
observation law. These are three distinct coordinates. In particular,
executing $m$ optimizer steps is not a proof that the learned weights
follow the beta function of independent-document averaging.

\begin{theorem}[Exact temporal elimination and inference compatibility]
\label{model:thm:training-block}
Define $K_{t:t+m}=K_tK_{t+1}\cdots K_{t+m-1}$. Then
\begin{equation}
 K_{t:t+m+n}=K_{t:t+m}K_{t+m:t+m+n},\qquad
 \mu_{t+m}=\rho_tK_{t:t+m}\mathcal E.\label{model:eq:time-block}
\end{equation}
Eliminating internal training times preserves every terminal inference
law. The same result holds for path-source insertions: replacing $K_t$ by
\begin{equation}
 (T_{t,j}f)(s)=\E\left[
 e^{j^{\mathsf T}a_t(s,\zeta_t)}f(U_t(s,\zeta_t))\mid s\right]
 \label{model:eq:training-source}
\end{equation}
preserves the generating function of the sum of retained training
observations $a_t$ under regrouping. Forward-graph elimination inside
$\mathcal E$ and linear coarse observation at its output commute with
this regrouping.
\end{theorem}
\begin{proof}
Successive conditional expectations integrate the joint transition law
over every internal state. Associativity of integration, for nonnegative
integrands or absolutely integrable signed integrands, gives the first
identity in~\eqref{model:eq:time-block}. Applying the composed operator to
$\mathcal E f$ proves the second identity for every bounded measurable
test function $f$, hence equality of the terminal laws. With the
insertions, the integrand contains the product of the step exponentials,
which equals the exponential of their summed source couplings. Regrouping
the same integrals preserves this product. Exact decoder elimination
leaves $\Phi_\theta(x,y)$ unchanged pointwise for every $\theta,x,y$.
Applying a linear observation to that same value also leaves the
identities valid, proving the stated compatibility.
\end{proof}

This is a temporal RG on transition or weighted transfer operators. It
is exact on the same already augmented state space. The composed kernel
may belong to a richer functional class than the one-step optimizer ansatz.
Augmentation enlarges a weights-only description; composition itself does
not enlarge the complete state. For nonstationary protocols it is an indexed
composition law, not a time-homogeneous semigroup on weights alone.
With discrete sampler and counter coordinates fixed, wherever the update
is differentiable in its continuous coordinates, its tangent and source
derivatives compose
as $J_{2:0}=J_{2:1}J_{1:0}$ and
$B_{2:0}=J_{2:1}B_{1:0}+B_{2:1}$ by the chain rule. These derivatives
include the optimizer coordinates and mixed parameter derivatives.
Differentiability must be checked at clipping boundaries and at singular
moment coordinates; the kernel identities themselves do not need it.

\subsection{When a smaller training state closes}
Let $c(s)$ be a deterministic retained coordinate and let $P$ be its
pushforward kernel. An exact autonomous reduced transition requires
\begin{equation}
 K_t(s,c^{-1}(A))=\overline K_t(c(s),A)
 \quad\hbox{for every measurable }A.\label{model:eq:lumpability}
\end{equation}
\begin{proposition}[Closure criterion]\label{model:prop:training-closure}
Condition~\eqref{model:eq:lumpability} implies
$K_tP=P\overline K_t$ and therefore
\begin{equation}
 \rho_tK_{t:t+m}P=(\rho_tP)
                  \overline K_t\cdots\overline K_{t+m-1}.
\end{equation}
If two fine states with the same retained coordinate give different
next-step retained laws, no such reduced kernel can describe every
initial fine-state law.
\end{proposition}
\begin{proof}
Equality on all measurable events is equality of the two kernels.
Substitute $K_tP=P\overline K_t$ repeatedly in the product, starting at
the rightmost factor, to obtain the displayed identity. Conversely,
initialize at either of the two fine states. Their projected initial
laws coincide, while their projected next-step laws differ. One reduced
kernel applied to that common initial law cannot produce both outputs.
\end{proof}

For example, the same weights and gradient with different first-moment
optimizer tensors generally produce different AdamW updates. Thus a
weights-only or head-energy-only state does not automatically satisfy
\eqref{model:eq:lumpability}. The exact alternatives are the full augmented
state or the projected path law with its induced memory. For the
stochastic inference kernel $\mathcal E$, an identity
$K_t\mathcal E=\mathcal E Q_t$ would likewise be sufficient for autonomous
inference-law evolution, but is an additional condition rather than a
consequence of decoder composition.
Section~\ref{model:sec:optimizer-results} tests this distinction with paired moment interventions, while Section~\ref{model:sec:conditional-memory-law} constructs the exact memory alternative.

\section{Predictive reduction and training selection of inference regimes}
\label{model:sec:reduction}

The exact state is the coupled pair of a training transition kernel and an
inference emission. Fix laws $\D_{\rm tr}$ and $\D_{\rm ev}$, including their
crop and target rules, and write
\[
 S_t=(\theta_t,o_t,r_t,t),\qquad \rho_{t+1}=\rho_tK_t^{\D_{\rm tr}},
 \qquad \mu_t=\rho_t\mathcal E^{\D_{\rm ev}}.
\]
Here $o_t$ includes optimizer moments and $r_t$ includes required sampler
state. The emission can retain the entire graph trace, generated operators
and source derivatives jointly. Its finite observation is a specified
pushforward. Composition of this state is exact; replacing it by a few
collective variables introduces a separate approximation. The bounds below
make that approximation part of the theory.

\subsection{Compatible approximate kernels}
Let $c:S\to Q$ be a fixed retained state and $P$ its deterministic
pushforward kernel. A proposed transition $\overline K_t$ on $Q$ and
emission $\overline{\mathcal E}$ must predict the effects of both retained
and eliminated coordinates.

This is the Markov-contraction telescoping construction \cite{kemeny1976,hairer2021}.
\begin{proposition}[Finite-horizon closure error]\label{model:prop:closure-error}
Use $\|\alpha-\beta\|_{\rm TV}=\sup_A|\alpha(A)-\beta(A)|$ for probability
measures. Suppose the kernels are Markov and
\[
 \sup_s\|K_tP(s,\cdot)-P\overline K_t(s,\cdot)\|_{\rm TV}\le\delta_t,
 \qquad
 \sup_s\|\mathcal E(s,\cdot)-P\overline{\mathcal E}(s,\cdot)\|_{\rm TV}
 \le\varepsilon_E.
\]
For any initial probability law $\rho$ and $m$ updates,
\begin{equation}
 \|\rho K_{0:m}\mathcal E-
       \rho P\overline K_{0:m}\overline{\mathcal E}\|_{\rm TV}
 \le\varepsilon_E+\sum_{t=0}^{m-1}\delta_t.\label{model:eq:closure-error}
\end{equation}
Empty transition products denote identity kernels.
\end{proposition}
\begin{proof}
Insert $P$ successively at the intermediate boundaries:
\[
 \rho K_{0:m}P-\rho P\overline K_{0:m}
 =\sum_{t=0}^{m-1}\rho K_{0:t}
 (K_tP-P\overline K_t)\overline K_{t+1:m}.
\]
Each difference is a signed kernel of mass zero. Integration against a
probability measure and application of a Markov kernel contract its total
variation. The triangle inequality bounds the sum by $\sum_t\delta_t$.
Add and subtract $\rho K_{0:m}P\overline{\mathcal E}$ in the output
difference. The first resulting term is bounded by $\varepsilon_E$; the
other is the preceding difference after another contracting Markov kernel.
\end{proof}

The zero-defect case is exact lumpability \cite{kemeny1976}. Total variation
between nearby unequal point masses is one, so a Euclidean prediction
error does not estimate this bound. There is a useful metric counterpart.

The metric form uses Lipschitz propagation of Markov kernels \cite{hairer2021}.
\begin{proposition}[Metric closure error]\label{model:prop:metric-closure}
Let the retained state and observation spaces be Polish metric spaces,
and let all compared laws have finite first moments. Suppose the
corresponding one-step defects in Wasserstein distance $W_1$ are bounded
by $\delta_t$, the emission defect by $\varepsilon_E$, and the coarse
kernels and emission satisfy
\[
 W_1(\overline K_t(q,\cdot),\overline K_t(q',\cdot))\le L_t d_Q(q,q'),
 \quad
 W_1(\overline{\mathcal E}(q,\cdot),\overline{\mathcal E}(q',\cdot))
 \le L_Ed_Q(q,q').
\]
Then the output error is at most
\begin{equation}
 \varepsilon_E+L_E\sum_{t=0}^{m-1}\delta_t
             \prod_{k=t+1}^{m-1}L_k.\label{model:eq:metric-closure}
\end{equation}
\end{proposition}
\begin{proof}
For any coupling of two input laws, couple the conditional kernel outputs
with costs arbitrarily close to their $W_1$ distances. Integrating the
Lipschitz bound and taking the infimum gives contraction by $L_t$ for the
kernel. Equivalently, the dual action maps a 1-Lipschitz test function to
an $L_t$-Lipschitz function. Insert one coarse boundary at a time as in the
preceding proof. The defect introduced at time $t$ traverses each later
kernel and the emission, giving its displayed product. Sum by the triangle
inequality and add the direct emission defect.
\end{proof}

The TV and uniform metric statements use domain-wide defects. A finite
trajectory residual is an empirical prediction error, not a uniform
certificate. The evolving-law theorem in Section~\ref{model:sec:evolving-law-closure}
instead integrates each
successor defect against the fine law that actually evolves. Source-weighted
transfer operators require their own norm factors because they are not
Markov contractions. Unresolved optimizer coordinates and earlier coarse
states can be retained as conditional memory
(Section~\ref{model:sec:conditional-memory-law}).

\subsection{Closure under the evolving training law}
\label{model:sec:evolving-law-closure}
A consuming corpus gives a time-dependent state distribution even at a
constant learning rate. Uniform defects can be stronger than needed for
prediction under that distribution. The following transport argument
integrates each defect under the fine law that actually generates the
successor, while retaining the coarse transport factors.

\begin{proposition}[Evolving-law successor and emission error]
\label{model:prop:law-closure}
Let $S_t$ be a Markov process on a standard Borel space, with kernels
$K_t$ and laws $\rho_t$. Let $c:S\to Q$ be measurable, $P$ its
deterministic pushforward kernel, and $\bar K_t$ proposed Markov kernels
on a Polish metric space $(Q,d_Q)$. Write
$\bar\rho_{t+1}=\bar\rho_t\bar K_t$. Assume finite first moments of
the compared laws, measurability and integrability of the displayed discrepancies, and
\[
 W_1(\bar K_t(q,\cdot),\bar K_t(q',\cdot))\le L_t d_Q(q,q'),
 \qquad L_t\ge0.
\]
Define
\[
 d_t=\int W_1\bigl(K_tP(s,\cdot),\bar K_t(c(s),\cdot)\bigr)\rho_t(ds),
 \qquad e_0=W_1(\rho_0P,\bar\rho_0).
\]
For a Polish observation space, let $\mathcal E$ be a measurable
fine-state emission kernel and $\bar{\mathcal E}$ an
$L_E$-Lipschitz retained emission in $W_1$, $L_E\ge0$. Assume
\[
 a_m=\int W_1\bigl(\mathcal E(s,\cdot),
                  \bar{\mathcal E}(c(s),\cdot)\bigr)\rho_m(ds)<\infty.
\]
With empty products equal to one,
\begin{equation}
 W_1(\rho_m\mathcal E,\bar\rho_m\bar{\mathcal E})
 \le a_m+L_E\left[e_0\prod_{k=0}^{m-1}L_k+
       \sum_{j=0}^{m-1}d_j\prod_{k=j+1}^{m-1}L_k\right].
 \label{model:eq:law-closure}
\end{equation}
\end{proposition}
\begin{proof}
Set $e_t=W_1(\rho_tP,\bar\rho_t)$. Insert the intermediate law
$\rho_tP\bar K_t$. For any 1-Lipschitz test function on $Q$, the
pointwise difference of conditional expectations is bounded by the
corresponding conditional $W_1$ distance. Integration against $\rho_t$
and the Kantorovich dual formula give
$W_1(\rho_tK_tP,\rho_tP\bar K_t)\le d_t$.
The moment and integrability assumptions justify these integrals.
The dual action of $\bar K_t$ maps a 1-Lipschitz function to an
$L_t$-Lipschitz function, so
$W_1(\rho_tP\bar K_t,\bar\rho_t\bar K_t)\le L_te_t$.
The triangle inequality yields $e_{t+1}\le L_te_t+d_t$.
Induction, multiplying only by nonnegative factors, gives the
product-and-sum expression in brackets. The same integrated dual
argument bounds the fine-to-retained emission discrepancy by $a_m$;
emission Lipschitz propagation bounds the remaining discrepancy by
$L_Ee_m$. Their sum proves the claim, including the empty horizon.
\end{proof}

The constants may exceed one. Defects are averaged against $\rho_t$,
not against an unrelated calibration distribution. Neither a rollout
root-mean-square error nor an estimated transport constant alone
certifies the hypotheses. In particular, a $W_1$ bound on risk does
not imply the stronger $L^2$ fluctuation control needed to transfer a
critical exponent.

\begin{corollary}[Observable residuals for deterministic coarse maps]
\label{model:cor:empirical-law-closure}
Suppose $\bar K_j(q,\cdot)=\delta_{F_j(q)}$, with $F_j$ an
$L_j$-Lipschitz map on a Euclidean retained space. Then
\begin{equation}
 d_j=\E\norm{c(S_{j+1})-F_j(c(S_j))}.
 \label{model:eq:deterministic-law-defect}
\end{equation}
For any finite collection of paired paths $q_j^{(i)}$, initialize
$\bar q_0^{(i)}=q_0^{(i)}$ and iterate
$\bar q_{j+1}^{(i)}=F_j(\bar q_j^{(i)})$. Define
$r_j^{(i)}=\norm{q_{j+1}^{(i)}-F_j(q_j^{(i)})}$.
The empirical pairing cost satisfies
\begin{equation}
 \frac1B\sum_i\norm{q_m^{(i)}-\bar q_m^{(i)}}
 \le\sum_{j<m}\left(\frac1B\sum_i r_j^{(i)}\right)
                    \prod_{j<k<m}L_k.
 \label{model:eq:empirical-law-closure}
\end{equation}
It upper-bounds the $W_1$ discrepancy of the two empirical retained
laws. If $q$ contains the standardized cohort risk
$q_1=(R-\mu_R)/s_R$, $s_R>0$, the corresponding empirical mean-risk
error is at most $s_R$ times the right-hand side.
\end{corollary}
\begin{proof}
The only coupling to a point mass pairs every successor with that
point. Its mean distance is the Wasserstein distance, proving the
first identity by conditioning and the tower property. For each path,
the triangle inequality gives
\[
 \norm{q_{j+1}^{(i)}-\bar q_{j+1}^{(i)}}
 \le r_j^{(i)}+L_j\norm{q_j^{(i)}-\bar q_j^{(i)}}.
\]
Induct and average to obtain the displayed bound. Pairing equal path
indices is an admissible empirical coupling. Coordinate projection is
1-Lipschitz, and conversion to risk multiplies distances by $s_R$.
\end{proof}

The empirical inequality needs no independence assumption. Estimating
the expectation in Equation~\eqref{model:eq:deterministic-law-defect} does:
reset branches sample the specified conditional successor law,
while their incoming initialization remains the outer random unit.
A fitted $F_j$ must be fixed independently of the paths used to
estimate these defects. Transfer between incoming initialization
strata requires a separate check of the changed conditional law. The full-vocabulary
risk coordinate still requires a native forward pass at the initial
state; its inclusion does not establish cheap inference or closure
of its fluctuation law. Section~\ref{model:sec:law-closure-results} implements conditional path
prediction under the same consuming law.

For a fixed weighted evaluation cohort, another useful emission is
the vector of gauge-centered full-vocabulary logits with its weighted
root-mean-square Euclidean norm. Its mean target NLL is
$\sqrt2$-Lipschitz. Indeed, the categorical log-loss gradient is
$p-e_y$, whose squared norm is at most two; integrate along the
logit segment, average with the nonnegative normalized target weights,
and apply Cauchy--Schwarz. Thus Equation~\eqref{model:eq:law-closure}
also supplies a risk bound multiplied by $\sqrt2$ for that emission.
The target and proper-prefix rules remain part of the specified
observation law.

\subsection{Exact closure with retained conditional memory}
\label{model:sec:conditional-memory-law}
There is an exact alternative when a finite retained coordinate fails
the autonomous-kernel criterion. It uses the unresolved conditional
law, which can be much larger than a finite parameter vector.
This is the conditional-probability counterpart of retaining memory
after elimination \cite{kallenberg2021,chorin2002}.

\begin{proposition}[A conditional-memory training and emission law]
\label{model:prop:conditional-memory}
Let $S_t$ be a Markov process on a standard Borel space with kernels
$K_t$, and let $q_t=c(S_t)$ for a measurable retained map into a
standard Borel space. Define the regular conditional law
$\pi_t(ds)=\Pr(S_t\in ds\mid q_0,\ldots,q_t)$.
Let $P_s$ be a measurable evaluation-context kernel. Conditional on
$S_t=s$, draw $(X,Y)$ from $P_s$ independently of the retained
history, and define
$(\mathcal E f)(s)=\int f(\Phi_{\theta(s)}(x,y))\,P_s(dx,dy)$,
where the parameter coordinate $\theta(s)$ and emission are measurable.
For bounded measurable test functions on the respective spaces,
\begin{align}
 \E[f(q_{t+1})\mid q_0,\ldots,q_t]
 &=\int K_t(f\circ c)(s)\,\pi_t(ds),
 \label{model:eq:conditional-memory}\\
 \E[f(\Phi_{\theta_t}(X,Y))\mid q_0,\ldots,q_t]
 &=\int (\mathcal E f)(s)\,\pi_t(ds).
 \label{model:eq:conditional-memory-emission}
\end{align}
The prediction law is $\widetilde\pi_{t+1}=\pi_tK_t$;
conditioning it on $c(S_{t+1})=q_{t+1}$ gives $\pi_{t+1}$,
for almost every realized retained history. Consequently the
conditional-law state, with the phase $t$, has a closed update and
emission. Chronological blocking is compatible with this update
when it retains the same observation information.
\end{proposition}
\begin{proof}
The Markov property gives the next conditional expectation at $S_t=s$
as $K_t(f\circ c)(s)$. The tower property, conditioning again on the
retained history, gives the first formula. Independent evaluation
contexts conditional on $S_t$ give the second by the same argument.
Integration of $K_t$ against $\pi_t$ is the predictive law of
$S_{t+1}$. Regular conditional probabilities exist on standard Borel
spaces; its disintegration under $c$ is precisely the stated update
after the new observation. Applying these identities successively
integrates the original joint law over the same internal states,
so regrouping steps while retaining the same observation information
leaves it unchanged. If only block-boundary observations are retained,
the corresponding conditional law is obtained by marginalizing the
internal observations as well.
\end{proof}

The proposition is an exact law description, not an established
finite-dimensional training reduction. A useful finite approximation
requires accurate vocabulary emission, collective covariance in fixed units,
and retained successor laws, together with a measured acquisition budget.
For a fixed emission dictionary, Proposition~\ref{model:prop:projection-obstruction}
separates representation error from coefficient excess. Reducing the latter
cannot compensate for an unresolved orthogonal component. Neither component
alone estimates a reduced-kernel closure defect. The metric bounds of Section~\ref{model:sec:reduction}
propagate measured-domain or uniform one-step defects as appropriate.
The nested row maps address the first two requirements. The shared
loss-force and Adam-state measurements specify why additional moments,
clipping information or conditional memory enter the third.
Finite source panels do not bound a supremum over unobserved states.
Sections~\ref{model:sec:row-projection-results}, \ref{model:sec:gradient-projection-results} and \ref{model:sec:optimizer-results} report the corresponding row, loss-gradient and optimizer-state tests.

\subsection{Required accuracy along an increasing horizon}
\begin{proposition}[Finite stability accumulation]
\label{model:prop:constant-stability}
If $e_0=0$, $L\ge0$, and $e_{t+1}\le Le_t+d$ for all $t$, then
\begin{equation}
 e_T\le d\sum_{j=0}^{T-1}L^j.
 \label{model:eq:constant-stability}
\end{equation}
\end{proposition}
\begin{proof}
The empty sum proves the claim at zero. If it holds at $T$, multiplying
by $L\ge0$ and adding $d$ gives
$e_{T+1}\le d(1+L\sum_{j<T}L^j)=d\sum_{j<T+1}L^j$.
Induction proves the assertion, including $L=0$ and $L=1$.
\end{proof}

Suppose the same recurrence controls a coupled $L^2$ error, and a native
fluctuation has scale $a_N=N^{(\kappa-1)/2}$ with a nonzero limiting
variance in the specified direction. A sufficient dynamical accuracy
condition for transfer is
$d_N\sum_{j<T_N}L_N^j=o(a_N)$, together with an $o(a_N)$ emission error.
If that sum is $O(N^\zeta)$, it suffices that
$d_N=o(N^{(\kappa-1)/2-\zeta})$. This follows by division by $a_N$ and
the covariance bound of Proposition~\ref{model:prop:critical-closure}.
Matching uncentered limiting laws also requires mean error $o(a_N)$.
A law-averaged $W_1$ bound supplies no $L^2$ coupling without an additional
hypothesis. The fixed finite operating target in the experiments is
therefore distinct from this asymptotic accuracy requirement.

\FloatBarrier
\par\medskip\noindent
Chapter~\ref{ch:single-pass-resource} specifies the data component of
that law for a consuming corpus. Retaining the remaining resource is what
allows chronological training predictions to respect a single pass.

\chapter{A single pass as a finite data resource}
\label{ch:single-pass-resource}
This chapter treats a single training pass as consumption of a finite
corpus resource. It defines the state for sampling without replacement,
clarifies supervised-position exposure, and derives the covariance and
response consequences of finite-population sampling and document dependence.

\section{The pretraining data resource in the scale flow}
\label{model:sec:data-resource}

The underlying text distribution, the sampled corpus, and the rule that
presents that corpus to the optimizer are distinct parts of the training
law. A finite empirical distribution sampled repeatedly and a single
pass through a larger corpus need not select the same inference regime.
Deterministic row identities hold in either case. Empirical drift,
innovation covariance, collapse times, and generalization require the
data law in their conditioning variables.

\subsection{An exact state for sampling without replacement}

Let $\mathcal C_M=(x_1,\ldots,x_M)$ be a specified corpus of target
blocks. Each block contains its input context and its supervised target
positions. Distinct blocks may contain the same words; nonrepetition
means that a source target position is supervised at most once. A
boundary token may be the final target of one block and the initial
context token of the next. Let $R_t\subseteq\{1,\ldots,M\}$ denote
the unconsumed block indices immediately before update $t+1$. The state
\begin{equation}
 U_t=(\theta_t,m_t,v_t,\lambda_t,R_t,\zeta_t)
 \label{model:eq:data-augmented-state}
\end{equation}
contains the native weights, optimizer memories, imposed drive, and
additional execution state $\zeta_t$, including the numerical implementation
and any required random-number state, with unrevealed data-shuffle
randomness marginalized in the stochastic convention. In a
probabilistic description that integrates over the unrevealed part of
a uniformly shuffled stream, the next ordered batch is uniform over
distinct indices in $R_t$. In a description that conditions on the
entire realized shuffle, that choice is deterministic. These are
different conditioning conventions, not different training programs.

For the former convention, write $\Phi$ for the complete native update,
including its target mask, clipping and optimizer arithmetic. With
$r_t=|R_t|$ and batch size $B\leq r_t$, its transition kernel is
\begin{equation}
 K(U_t,\cdot)=\frac{1}{(r_t)_B}
 \sum_{\substack{(i_1,\ldots,i_B)\in R_t^B\\i_j\text{ distinct}}}
 \delta_{\Phi(U_t;i_1,\ldots,i_B)}(\cdot),
 \qquad (r)_B=r(r-1)\cdots(r-B+1).
 \label{model:eq:single-pass-kernel}
\end{equation}
The executed studies condition on their recorded permutations;
successive batches are not treated as independent sampling
replications. The map removes the selected indices from $R_t$. A declared numerical
failure can be represented by an absorbing recorded outcome. Omitting
$R_t$ generally destroys the Markov description: the same weights and
optimizer memories can have different remaining data. Compatible
temporal RG maps therefore compose this augmented kernel. Inference
is its trained-weight marginal followed by the native inference
emission; it does not continue the corpus-consumption process.

\begin{proposition}[Consumption and local stationarity]
\label{model:prop:consumption-stationarity}
Let $M,B$ be positive integers with $B$ dividing $M$. Suppose a
corpus has $M$ indexed blocks and every transition with a nonempty remaining corpus consumes $B$ distinct
remaining blocks. Exhausted states remain exhausted. Every invariant probability law of the augmented
process is supported on the exhausted states. Before exhaustion,
let $\pi_t$ be the uniform probability law on the remaining block
indices. Along any realized path, for an integer $h\ge0$ with
$hB<r_t$,
\begin{equation}
 \|\pi_{t+h}-\pi_t\|_{\rm TV}=\frac{hB}{r_t}.
 \label{model:eq:remaining-law-drift}
\end{equation}
For the induced law of block contents or any deterministic block
observable, the same expression is an upper bound. If $\pi_t^{(B)}$
is the uniform ordered next-batch law, then
\begin{equation}
 \|\pi_{t+h}^{(B)}-\pi_t^{(B)}\|_{\rm TV}
 =1-\frac{(r_t-hB)_B}{(r_t)_B}
 \leq \min\{1,hB^2/r_t\}.
 \label{model:eq:remaining-batch-law-drift}
\end{equation}
\end{proposition}
\begin{proof}
The bounded function $r(U)$ has one-step decrement
$B\mathbf1_{r(U)>0}$. Its mean decrement under an invariant
probability law is zero, which forces $r=0$ almost surely.
For the second statement, put $r=r_t$ and $k=hB$. There are $k$
removed indices and $r-k$ retained indices. Half the sum of the
absolute changes in their probabilities is
\[
 \frac12\left(\frac{k}{r}
 +(r-k)\left(\frac1{r-k}-\frac1r\right)\right)=\frac{k}{r}.
\]
A deterministic pushforward cannot increase total variation: the
preimage of each event is an event in the original index space.
The later ordered batch law is the earlier one conditioned on
avoiding the $k$ removed indices. Total variation between a law
and its conditioning on an event of positive probability is the
probability of the complement. There are $(r-k)_B$ allowed ordered
batches among $(r)_B$ possibilities. A union bound over the $B$
positions, each with probability $k/r$ of being removed, gives
the stated upper bound.
\end{proof}

This proposition concerns the full data-consumption process. It
allows a projected predictive observable to have a nearly stationary
finite window while the corpus changes. In a local window of
$\tau_N$ updates, the sufficient resource condition
$B\tau_N/r_t\to0$ makes the remaining one-block law stable in the
metric of Equation~\eqref{model:eq:remaining-law-drift}. The next-batch
law retains the additional factor $B$ in
Equation~\eqref{model:eq:remaining-batch-law-drift}. Neither marginal
statement replaces an entire window by independent draws; a
sufficient joint-path coupling condition is instead
$(B\tau_N)^2/r_t\to0$, by
Proposition~\ref{model:prop:data-path-coupling}. A stationary critical
scaling interpretation additionally needs control of the optimizer,
drive and emission on that window. A positive fixed
learning-rate floor alone supplies none of these controls. Conversely,
a finite training transition can be studied without asserting
stationarity, provided its time, remaining-data and drive coordinates
are retained. An exhausted endpoint is a finite stopping state,
not evidence of an invariant nonterminal training regime.

\begin{proposition}[Finite-population innovation]
\label{model:prop:finite-population-innovation}
Fix the current augmented state and a remaining population of $r>1$
vectors $h_i\in\R^p$. Let
\[
 \bar h=\frac1r\sum_i h_i,\qquad
 \Sigma_R=\frac1r\sum_i(h_i-\bar h)(h_i-\bar h)^{\mathsf T}.
\]
For a uniformly chosen batch of $B$ distinct indices, with
$1\leq B\leq r$, its mean $\widehat h_B$ satisfies
\begin{equation}
 \E[\widehat h_B\mid U]=\bar h,\qquad
 \Cov(\widehat h_B\mid U)
 =\frac{r-B}{B(r-1)}\Sigma_R.
 \label{model:eq:finite-population-covariance}
\end{equation}
\end{proposition}
\begin{proof}
Let $I_i$ indicate membership in the batch and set $a_i=h_i-\bar h$.
Then $\E I_i=B/r$ and, for $i\ne j$,
$\E I_iI_j=B(B-1)/(r(r-1))$. The expectation follows by summing
the first identity. Since $\sum_i a_i=\bm0$,
$\sum_{i\ne j}a_i a_j^{\mathsf T}=-\sum_i a_i a_i^{\mathsf T}$.
Expanding the second moment of $B^{-1}\sum_i I_i a_i$ consequently gives
\[
 \frac1{B^2}\left(\frac Br-
 \frac{B(B-1)}{r(r-1)}\right)\sum_i a_i a_i^{\mathsf T}
 =\frac{r-B}{B(r-1)}\Sigma_R.
\]
\end{proof}

The proposition applies to additive observables at a frozen state. It
does not identify the covariance of clipped Adam increments by
substitution: clipping, memory, normalization and native arithmetic
remain inside $\Phi$. Even the unmodified all-target gradient is a
ratio when blocks have different numbers of valid targets.

\begin{proposition}[Masked-target ratio]
\label{model:prop:masked-target-ratio}
At fixed weights, let $h_i$ be the sum of the valid-target gradients
in block $i$ and let $n_i\geq0$ be its valid-target count. Assume
$\bar n=r^{-1}\sum_i n_i>0$ and that every admitted batch has
$B^{-1}\sum_{i\in\mathcal B}n_i\geq n_*>0$. Put
\[
 g_* = \frac{\bar h}{\bar n},\qquad
 a_i=h_i-n_i g_*,\qquad
 \Sigma_a=\frac1r\sum_i a_i a_i^{\mathsf T}.
\]
Define the masked batch gradient before clipping and floating-point
evaluation by
\[
 g_{\mathcal B}=\frac{\sum_{i\in\mathcal B}h_i}{\sum_{i\in\mathcal B}n_i}.
\]
It satisfies
\begin{equation}
 \E\norm{g_{\mathcal B}-g_*}^2
 \leq \frac{r-B}{B(r-1)n_*^2}\tr\Sigma_a.
 \label{model:eq:masked-gradient-population-bound}
\end{equation}
It is unbiased for $g_*$ if all $n_i$ are equal and positive.
Unequal counts do not in general give that unbiasedness.
\end{proposition}
\begin{proof}
The residuals have mean zero, and
\[
 g_{\mathcal B}-g_*=
 \frac{B^{-1}\sum_{i\in\mathcal B}a_i}
 {B^{-1}\sum_{i\in\mathcal B}n_i}.
\]
Bound the squared reciprocal denominator by $n_*^{-2}$ and apply
Proposition~\ref{model:prop:finite-population-innovation} to the numerator.
Equal counts make the denominator constant and give unbiasedness.
For a counterexample with unequal counts, take $r=2$, $B=1$,
$(h_1,h_2)=(0,2)$ and $(n_1,n_2)=(1,2)$. Then $g_*=2/3$ while
$\E g_{\mathcal B}=1/2$.
\end{proof}

Conditional innovations in the RG equations are consequently defined
by the exact conditional mean of the implemented update, with the
sampling convention specified. A reduction to a fixed-denominator
gradient covariance requires its own hypothesis. At the final full
batch of a pass, the sampling covariance in
\eqref{model:eq:finite-population-covariance} vanishes. Earlier optimizer
memory and randomness in the trained state need not vanish. This
externally determined exhaustion effect is not an endogenous critical
attractor.

\subsection{The supervised source position as an exposure unit}

Distinct randomly cropped input windows can supervise overlapping
source positions. Consequently, the number of distinct windows does
not by itself measure target reuse. The objective and its target mask
are part of the data presentation law.

\begin{proposition}[Expected source-position occupancy]
\label{model:prop:source-position-occupancy}
Let $\mathcal J$ be a finite set of indexed source positions. Draw
$K$ crops independently from one fixed crop law. A crop supervises
position $j$ at most once, with probability $p_j$. Write $H_j$ for
its total number of supervised occurrences, $H=\sum_jH_j$ for all
target events, and $V=\sum_j\mathbf1_{H_j>0}$ for distinct supervised
positions. Then
\begin{align}
 \E H&=K\sum_{j\in\mathcal J}p_j,\qquad
 \E V=\sum_{j\in\mathcal J}\bigl[1-(1-p_j)^K\bigr],
 \label{model:eq:source-position-occupancy}\\
 \E(H-V)&=\sum_{j\in\mathcal J}
 \bigl[Kp_j-1+(1-p_j)^K\bigr].\nonumber
\end{align}
No independence between positions within one crop is required.
For a single pass through blocks with disjoint supervised positions,
$H=V$ on every realized path, including with a deterministic target
mask.
\end{proposition}
\begin{proof}
For a fixed position, the crop inclusion indicators are independent
Bernoulli variables with parameter $p_j$. Hence
$\E H_j=Kp_j$ and $\mathbb P(H_j=0)=(1-p_j)^K$.
Sum these identities and subtract. Disjoint supervised block
positions each occur zero or one times under sampling without
replacement, which proves the final statement.
\end{proof}

For uniform resampling from $D$ documents with crop offsets
$0,\ldots,O$ and $L$ targets per crop, an unmasked target at document
position $j\in\{1,\ldots,O+L\}$ has inclusion probability
\begin{equation}
 p_j=\frac{\min(O,j-1)-\max(0,j-L)+1}{D(O+1)}.
 \label{model:eq:overlapping-target-inclusion}
\end{equation}
The numerator counts offsets whose target interval contains $j$.
The external-last-target objective instead has $p_j=1/[D(O+1)]$
for $j=L,\ldots,O+L$, and zero elsewhere. Masked source positions
have $p_j=0$. The executed repeated corpus has $D=3072$, $O=448$
and $L=64$, with no padding tokens in its training target positions.
Its all-target objective therefore has at most $3072\cdot512$
distinct supervised source positions, while the last-target
objective has at most $3072\cdot449$. Even when different crops
are selected, their target intervals can overlap. Equation~
\eqref{model:eq:source-position-occupancy} quantifies this exposure; it
is not an effective-sample-size formula for nonlinear training.
Reused input positions can also affect a gradient when its supervised
target is new. Supervised-position occupancy is consequently one
coordinate of data presentation, alongside the context and crop law.
Section~\ref{model:sec:scaling-results} specifies this repeated-corpus family and its target construction.

\subsection{When resampling and a single pass are close}

\begin{proposition}[Finite-path sampling comparison]
\label{model:prop:data-path-coupling}
Fix one population of $M$ blocks, a common initial-state law and a
common training program. Compare $K\leq M$ independent uniform draws
with $K$ draws without replacement. Let $P_K$ and $Q_K$ be the laws
of any common measurable observation of the resulting training paths.
For total variation defined as $\sup_A|P(A)-Q(A)|$,
\begin{equation}
 \norm{P_K-Q_K}_{\rm TV}
 \leq 1-\frac{(M)_K}{M^K}
 \leq \min\left\{1,\frac{K(K-1)}{2M}\right\}.
 \label{model:eq:training-path-sampling-coupling}
\end{equation}
\end{proposition}
\begin{proof}
The independent index sequence contains no duplicate with probability
$(M)_K/M^K$. Conditional on this event, it is uniform over ordered
distinct sequences. Couple the without-replacement sequence to it on
that event. On its complement, draw a separate uniform ordered
distinct sequence. The second marginal has the required law, and the
two paths agree whenever the indices agree if their other randomness
is coupled identically. The coupling inequality gives the first bound.
A union bound over the $\binom K2$ equal-index events gives the second.
A common measurable observation cannot increase total variation.
\end{proof}

For $K\le r$ uniform ordered draws from $r$ distinct remaining indices,
the exact total-variation distance between sampling with and without
replacement is $1-\prod_{j=0}^{K-1}(1-j/r)$. Conditional on having no
repeated index, the replacement law is the without-replacement law, so
this distance is the probability of at least one collision. A union bound
over draw pairs gives $K(K-1)/(2r)$. Common deterministic or stochastic
emission can only decrease the distance.

In the recorded branch family there are 4,194,304 eligible blocks, with
3,932,160 and 3,145,728 remaining at incoming updates 8,192 and 32,768.
A 64-update batch-32 history has $K=2,048$, giving index-law distances
0.413255 and 0.486490. These are not measured prediction-law distances or
proofs against local stationarity. They show why a small replacement error
cannot be inferred from this particular whole-window bound for those
histories. The finite conditional analysis retains the resource law
explicitly; application of a stationary forcing model requires additional
justification.
The branch acquisition and conditional forecasts are detailed in Section~\ref{model:sec:law-closure-results}.

Thus $K=o(\sqrt M)$ is a sufficient condition for this particular
whole-path comparison to vanish. The proposition provides no useful
equivalence guarantee at a fixed positive consumed fraction $K/M$.
It also compares sampling rules on the \emph{same} population. It
does not remove the change in empirical distribution when a small
corpus is replaced by a much larger one.

For a single-pass head-count limit, the corpus resource must grow
with the selected training horizon: $M_N\geq B T_N$. The consumed
fraction
\begin{equation}
 q_N(t)=\frac{Bt}{M_N}
 \label{model:eq:corpus-consumption-coordinate}
\end{equation}
is an external coordinate alongside the learning-rate phase and the
body and generator clocks. Holding only a nominal update count or
learning rate fixed does not hold all these coordinates fixed.
Coefficients of a collective limiting equation can depend on the
underlying text law, the finite corpus, the remaining empirical law
and $q_N$. A critical exponent requires a stated joint limiting
construction and control of these variables. A power fitted under
one repeated-data law cannot be transferred to a single pass by
renaming its update axis.

For example, specify a family transformation by
$N\mapsto bN$, $M\mapsto b^{\mu_D}M$,
$B\mapsto b^{\beta_B}B$ and $T\mapsto b^{z_T}T$, with
integer realizations of the counts. Direct substitution gives
\begin{equation}
 q\mapsto b^{\beta_B+z_T-\mu_D}q,\qquad
 \frac{dq}{d\log b}=(\beta_B+z_T-\mu_D)q.
 \label{model:eq:data-consumption-scale-flow}
\end{equation}
The continuous expression describes the corresponding power-law
family; the discrete factors compose by multiplication.
Keeping a nonzero consumed fraction fixed requires
$\mu_D=\beta_B+z_T$. A smaller data-resource power would eventually
violate the single-pass constraint $BT\le M$. These are imposed
family-scaling conventions, not measured critical exponents.
With fixed batch size, the joint clock map
$(N,T)\mapsto(bN,bT)$ consequently requires $M\mapsto bM$
if it is also to preserve the consumed fraction.

\subsection{Temporal data aggregation and the response cutoff}

The conditional remaining-population covariance and a covariance
averaged over an entire random ordering answer different questions.
A fixed linear response makes the latter distinction explicit.

\begin{proposition}[Single-pass source transport]
\label{model:prop:single-pass-source-transport}
Let $h_1,\ldots,h_M\in\mathbb R^d$ be fixed, with $M>1$,
mean $\mu$ and population covariance
$\Sigma=M^{-1}\sum_i(h_i-\mu)(h_i-\mu)^{\mathsf T}$.
Partition the first $BT$ positions of a uniform permutation into
$T$ consecutive batches of size $B$, where $BT\le M$, and let
$Y_t$ be the average within batch $t$. For deterministic matrices
$A_t:\mathbb R^d\to\mathbb R^k$, put
$Z=\sum_{t=1}^T A_t(Y_t-\mu)$. Then
\begin{align}
 \Cov(Y_t,Y_s)
 &=\left(\frac{M}{B(M-1)}\mathbf1_{t=s}
                      -\frac1{M-1}\right)\Sigma,
 \label{model:eq:onepass-temporal-source-covariance}\\
 \Cov Z
 &=\frac{M}{B(M-1)}\sum_t A_t\Sigma A_t^{\mathsf T}
  -\frac1{M-1}\left(\sum_t A_t\right)\Sigma
                         \left(\sum_t A_t\right)^{\mathsf T}.
 \label{model:eq:onepass-transport-covariance}
\end{align}
Averaging $b$ consecutive batches is exactly the same random vector
as directly forming batches of size $bB$. Consequently these data
aggregation maps compose by multiplication of their integer scale
factors whenever the partitions are defined.
\end{proposition}
\begin{proof}
A single sampled centered vector has covariance $\Sigma$.
For two distinct positions of the permutation, summing over all
ordered unequal population indices gives covariance
$-\Sigma/(M-1)$, because the sum of the centered population is zero.
A batch average has $B$ diagonal terms and $B(B-1)$ unequal terms,
divided by $B^2$. Distinct batches have $B^2$ unequal terms.
This proves Equation~\eqref{model:eq:onepass-temporal-source-covariance}.
Expand $\Cov Z=\sum_{t,s}A_t\Cov(Y_t,Y_s)A_s^{\mathsf T}$
to obtain Equation~\eqref{model:eq:onepass-transport-covariance}.
An average of $b$ averages of $B$ vectors is their direct average
with weight $1/(bB)$, pointwise in the recorded permutation.
Applying this identity successively proves composition of the
aggregation maps and hence the same identity for their joint laws.
\end{proof}

The temporal coefficient matrix in
Equation~\eqref{model:eq:onepass-temporal-source-covariance} has eigenvalue
$(M-BT)/(B(M-1))$ in the constant temporal direction and
$M/(B(M-1))$ on its orthogonal complement. A full pass removes the
constant temporal source fluctuation. Responses with unequal
transport weights can still retain variation from the ordering.
Neither covariance closure nor this exact data aggregation identity
asserts that the source law is Gaussian or that native Adam steps
can be replaced by one update on an averaged batch.

\begin{corollary}[Data resource relative to response memory]
\label{model:cor:onepass-response-memory}
In Proposition~\ref{model:prop:single-pass-source-transport}, let
$A_t=w_t I$ with real weights not all zero, and define
\[
 T_{\rm eff}=\frac{(\sum_t w_t)^2}{\sum_t w_t^2}.
\]
Relative to independent sampling with replacement from the same
population and batch size, the transported covariance is multiplied
by
\begin{equation}
 \frac{M-BT_{\rm eff}}{M-1}.
 \label{model:eq:onepass-memory-cutoff}
\end{equation}
For an exponential response $w_t=\eta r^{T-t}$, with
$\eta\ne0$ and $0\le r<1$,
\begin{equation}
 T_{\rm eff}=\frac{1+r}{1-r}\frac{1-r^T}{1+r^T}.
 \label{model:eq:onepass-effective-memory}
\end{equation}
At $r=1$, the exact value is $T_{\rm eff}=T$.
\end{corollary}
\begin{proof}
Independent replacement batches have covariance $\Sigma/B$ and
zero cross covariance. Their transported covariance is therefore
$(\sum_t w_t^2)\Sigma/B$. Substitute $A_t=w_tI$ in
Equation~\eqref{model:eq:onepass-transport-covariance} and factor this
matrix to obtain Equation~\eqref{model:eq:onepass-memory-cutoff}.
Cauchy--Schwarz gives $T_{\rm eff}\le T\le M/B$, so the factor
is nonnegative. The two finite geometric sums for $\sum_t w_t$
and $\sum_t w_t^2$ give Equation~\eqref{model:eq:onepass-effective-memory}.
Constant nonzero weights give $T_{\rm eff}=T$ directly.
\end{proof}

If $r_N=\exp(-1/\tau_N)$, $\tau_N\to\infty$ and
$T_N/\tau_N\to\infty$, then $T_{\rm eff}\sim2\tau_N$.
Thus a data resource $M_N\gg B\tau_N$ makes this particular
linear-response covariance correction vanish. The near-integrator
regime $T_N/\tau_N\to0$ instead gives $T_{\rm eff}\sim T_N$.
At $r=1$ exactly, a consumption fraction
$1-q_N\asymp N^{-\omega}$ multiplies any otherwise matched
variance-based power $N^\gamma$ by a factor of order $N^{-\omega}$,
provided $M_N\to\infty$. The resulting power $\gamma-\omega$
can therefore reflect the selected data resource. A complete pass
has zero source variance in this constant-weight example.

More explicitly, if $T,\tau,M\to\infty$ with
$T/\tau\to x\in(0,\infty)$ and $BT/M\to q$, the covariance
ratio converges to the scaling function
\begin{equation}
 \mathcal F_q(x)=1-\frac{2q}{x}\tanh(x/2).
 \label{model:eq:data-memory-scaling-function}
\end{equation}
Indeed, $(1+r)/(1-r)=\coth(1/(2\tau))\sim2\tau$ and
$(1-r^T)/(1+r^T)=\tanh(T/(2\tau))$. Substitution into
Equation~\eqref{model:eq:onepass-memory-cutoff} gives the limit.
It tends to $1-q$ as $x\downarrow0$ and to $1$ as
$x\to\infty$. At a complete pass, its small-$x$ form is
$\mathcal F_1(x)=x^2/12+O(x^4)$. This quadratic cutoff is a
property of the source constraint and linear response, not an
identified PLDR critical exponent. Conversely, alternating weights
$w_t=(-1)^t$ over an even number of batches have $T_{\rm eff}=0$.
Their covariance ratio is $M/(M-1)$, even at a complete pass.
Low-frequency cumulative response and alternating response therefore
have different sensitivity to this same data constraint.

These are exact source and deterministic-transport statements.
They average the permutation at fixed population and coefficients.
The native measurements condition on a recorded stream and vary
complete model initializations. Their gradients, clipping,
nonpadding denominators and Adam transport also evolve with the
history. Equation~\eqref{model:eq:onepass-memory-cutoff} is therefore
not a multiplicative correction to the measured native
susceptibilities. It identifies a source of altered temporal
fluctuations that a joint data, time and model-size limit must control.

\subsection{Averaging over corpora and retaining document dependence}

The fixed-corpus source constraint has a different meaning after
averaging the corpus itself under the pretraining-data distribution.
The following identity makes the extra fluctuation sector explicit.
When the corpus is random, the full training state also retains that
corpus, or an equivalent conditional law of its unrevealed contents.
Its same realization is shared throughout the path. One therefore
averages the composed corpus-conditioned path law; independently
redrawing the corpus at each transition defines another process.
Discarding its persistent information generally introduces memory,
as for other unresolved training coordinates.

\begin{proposition}[Corpus-conditioned and distribution-averaged transport]
\label{model:prop:corpus-conditioning}
Let $\mathcal C=(H_1,\ldots,H_M)$ be a random corpus of
square-integrable vectors, $M>1$, and let a uniform permutation
be independent of it. Define
\[
 \mu_{\mathcal C}=\frac1M\sum_iH_i,\qquad
 \Sigma_{\mathcal C}=\frac1M\sum_i
 (H_i-\mu_{\mathcal C})(H_i-\mu_{\mathcal C})^{\mathsf T},
 \qquad \bar\mu=\E\mu_{\mathcal C}.
\]
Use the batches and deterministic matrices of
Proposition~\ref{model:prop:single-pass-source-transport}, and put
$S=\sum_tA_t$ and $Z=\sum_tA_t(Y_t-\bar\mu)$. Then
\begin{align}
 \Cov Z={}&\frac{M}{B(M-1)}\sum_t
 A_t(\E\Sigma_{\mathcal C})A_t^{\mathsf T}
 -\frac1{M-1}S(\E\Sigma_{\mathcal C})S^{\mathsf T}\nonumber\\
 &\quad+S\Cov(\mu_{\mathcal C})S^{\mathsf T}.
 \label{model:eq:corpus-averaged-transport}
\end{align}
If the $H_i$ are independent and identically distributed with
covariance $V$, this reduces exactly to
\begin{equation}
 \Cov Z=\frac1B\sum_tA_tVA_t^{\mathsf T}.
 \label{model:eq:iid-corpus-transport}
\end{equation}
\end{proposition}
\begin{proof}
Conditionally on $\mathcal C$, the transport covariance is
Equation~\eqref{model:eq:onepass-transport-covariance} and the conditional
mean of $Z$ is $S(\mu_{\mathcal C}-\bar\mu)$.
The law of total covariance gives
Equation~\eqref{model:eq:corpus-averaged-transport}. For independent
identically distributed vectors,
$\Cov(\mu_{\mathcal C})=V/M$ and
$\E\Sigma_{\mathcal C}=(M-1)V/M$.
Substitution cancels the two terms containing $SVS^{\mathsf T}$
and proves Equation~\eqref{model:eq:iid-corpus-transport}.
\end{proof}

In particular, a complete-pass sum is constant under permutations
of one fixed corpus, while its value can vary across corpora.
Discarding the last term of
Equation~\eqref{model:eq:corpus-averaged-transport} would incorrectly
remove that distribution-level uncertainty. These identities concern
fixed source features and deterministic transport. Adaptive native
response coefficients remain inside the complete training law.

\begin{corollary}[Independent-corpus training path law]
\label{model:cor:iid-corpus-path}
Let $X_1,\ldots,X_M$ be independent blocks with common law $\D$,
and permute them uniformly. Assume that the corpus, permutation,
and combined initial-state and execution randomness are mutually
independent. Consider the recorded parameters, optimizer variables and emissions
of a training program that accesses only its successive batches.
Through every $T$ with $BT\le M$, these coordinates have the same
joint path law as that program driven by $BT$ fresh independent
blocks from $\D$, after averaging over corpus and permutation.
The recorded path here excludes unrevealed corpus coordinates.
\end{corollary}
\begin{proof}
Conditional on any fixed permutation, the selected $BT$ coordinates
are distinct members of an independent family, each with law $\D$.
Their joint law is $\D^{\otimes BT}$ and does not depend on that
permutation. Averaging preserves it. Couple the independent initial
state and execution randomness identically, and apply the same
measurable training-path map to these inputs. The resulting path
laws are equal.
\end{proof}

The equality permits adaptive gradients, optimizer memory and
clipping. It concerns the specified causal program coordinates,
rather than identifying the augmented remaining-corpus states.
It does not condition on the realized corpus, its remaining
population or its full order. It also does not identify resampling
from one finite random corpus with drawing fresh independent data.
Thus corpus-conditional exhaustion and a distribution-averaged
fresh-stream description are compatible statements with different
random units. The corollary's independent-block hypothesis is not
implied by distinct source positions in text.

For a concrete document dependence, let $D,m$ be positive integers
and $M=Dm>1$ blocks be grouped
into $D$ independent documents of $m$ blocks each, and write
$H_{d,j}=C_d+\epsilon_{d,j}$. Assume all common vectors $C_d$ and
all block residuals $\epsilon_{d,j}$ are mutually independent and
centered, with covariances $V_C$ and $V_\epsilon$ respectively.
Let $V=V_C+V_\epsilon$ and
$Q=(m-1)V_C/(M-1)$. Under an independent uniform block permutation,
two distinct positions come from the same document with probability
$(m-1)/(M-1)$. Their covariance is consequently $Q$, while a single
position has covariance $V$. Summing the diagonal and unequal terms
within and between batches gives
\begin{align}
 \Cov(Y_t,Y_s)&=\frac{\mathbf1_{t=s}}B(V-Q)+Q,\nonumber\\
 \Cov\Bigl(\sum_tA_tY_t\Bigr)
 &=\frac1B\sum_tA_t(V-Q)A_t^{\mathsf T}+SQS^{\mathsf T}.
 \label{model:eq:document-common-source}
\end{align}
This is an explicit distribution-averaged example with nonrepeated
positions and a retained common document sector. It is not an
assumed covariance model for RefinedWeb. The executed corpus contains
eight disjoint blocks per selected document, so unique-position
counts are not independent-document or independent-gradient counts.
The large-corpus native initialization comparisons fix that corpus
and its recorded ordering; they estimate neither an ensemble of new
corpora nor the document covariance in
Equation~\eqref{model:eq:document-common-source}.
Section~\ref{model:sec:onepass-corpus-methods} gives the executed crop, block and permutation construction.

\subsection{Loss drift and the distinction between fitting and prediction}

\begin{proposition}[Risk drift of a native update]
\label{model:prop:data-conditioned-risk-drift}
Let $L$ be a twice differentiable evaluation risk whose Hessian has
operator norm at most $\beta$ on every segment reached by a native
weight increment $\Delta\theta$. Conditional on the complete current
state, suppose the increment has finite second moment, mean $b$ and
covariance $V$. Then
\begin{equation}
 \left|\E[L(\theta+\Delta\theta)-L(\theta)\mid U]
       -\ip{\nabla L(\theta)}b\right|
 \leq \frac\beta2\bigl(\norm b^2+\tr V\bigr).
 \label{model:eq:population-risk-native-drift}
\end{equation}
\end{proposition}
\begin{proof}
Taylor's formula with integral remainder gives
\[
 L(\theta+d)-L(\theta)=\ip{\nabla L(\theta)}d+
 \int_0^1(1-s)d^{\mathsf T}\nabla^2L(\theta+sd)d\,ds.
\]
The absolute remainder is at most $\beta\norm d^2/2$.
Take the conditional expectation and use
$\E\norm{\Delta\theta}^2=\norm b^2+\tr V$.
\end{proof}

This bound uses the actual native increment, including optimizer
memory, evaluated by the stipulated smooth real-arithmetic risk.
A risk evaluated by a floating-point forward program retains a
separate arithmetic discrepancy; the algebraic finite-corner identity
in Proposition~\ref{model:prop:finite-source-corners} applies directly
to that program. Contraction of row contrast imposes no sign on
$\ip{\nabla L}b$ for a different evaluation law. A rise in held-out
loss therefore does not by itself identify critical slowing, and a
train--held-out gap does not by itself identify the cause of a rise
in both losses. Frozen-state risk on consumed training blocks,
held-out risk and the preceding online loss are separate observations.

\begin{proposition}[Finite logit risk decomposition]
\label{model:prop:finite-logit-risk}
Let $z,z'\in\R^V$ be vocabulary logits, $d=z'-z$,
$p=\softmax(z)$, $p'=\softmax(z')$, and let $e_y$ denote the
unit vector of target $y$. The categorical log loss obeys
\begin{equation}
 -\log p'_y+\log p_y
 =\KL(p\|p')+\ip{p-e_y}d.
 \label{model:eq:finite-logit-risk}
\end{equation}
Both terms on the right are invariant under adding an arbitrary
constant to every coordinate of $d$. The identity also holds after
any fixed weighted context average and conditional expectation.
\end{proposition}
\begin{proof}
Put $A=\log\sum_i e^{z_i+d_i}-\log\sum_i e^{z_i}$.
The left side is $A-d_y$, while
$\KL(p\|p')=A-\sum_i p_i d_i$. Subtraction proves the identity.
A constant shift leaves $p'$ unchanged and has zero inner product
with $p-e_y$. Averaging preserves the equality whenever the stated
expectations exist.
\end{proof}

This finite log-score identity uses the actual emitted logit change.
It needs neither a small parameter displacement nor a differentiable
map from weights to logits. A decrease in evaluation loss requires
the target-directed score term $\ip{p-e_y}d$ to be negative enough
to exceed the nonnegative predictive KL in magnitude. Small
predictive distance alone does not determine the loss change.
For retained native logits, the real-valued identity and the numerical
log-softmax reduction have the same explicit arithmetic boundary as
the other predictive identities. It complements the weight-space
risk bound without substituting a real-arithmetic optimizer for the
executed update.

An elementary likelihood calculation illustrates the distinction.
Let an evaluation label be Bernoulli with parameter $p\in(0,1)$,
and let a finite training sample have frequency
$\widehat p\in(0,1)$ with $\widehat p\ne p$. Along
$q(s)=p+s(\widehat p-p)$, $0\leq s\leq1$, the cross-entropy
$\ell_a(q)=-a\log q-(1-a)\log(1-q)$ satisfies
\[
 \frac{d}{ds}\ell_{\widehat p}(q(s))
 =-\frac{(1-s)(\widehat p-p)^2}{q(s)(1-q(s))},\qquad
 \frac{d}{ds}\ell_p(q(s))
 =\frac{s(\widehat p-p)^2}{q(s)(1-q(s))}.
\]
Fitting the empirical frequency decreases training risk while
increasing evaluation risk by the final amount
$\KL(\operatorname{Bern}(p)\|\operatorname{Bern}(\widehat p))$.
These are smooth, finite-dimensional changes with no critical
singularity. In a PLDR-LLM, the row-stabilization and predictive-margin conditions developed in
Section~\ref{model:sec:inference-mechanism} must therefore be checked together with
the data-conditioned training law.

\FloatBarrier
\par\medskip\noindent
Chapter~\ref{ch:predictive-transport} combines this source law with the
optimizer and forward computation. It tracks how component motion and
training variability propagate into predictive drift and response.

\chapter{Training sources, potential activity, and predictive transport}
\label{ch:predictive-transport}
This chapter follows training sources from parameter and potential motion
to predictive observations. Exact finite transport, derivative response,
conditional mixtures and chronological covariance give complementary
budgets for operator replacement and prediction along a training path.

\section{Potential activity, training clocks and endogenous relaxation}
\label{model:sec:potential-dynamics}

The potential tensor is a useful observation of training because it
couples a positive, data-dependent base to learned exponents. It is not
itself a distribution of critical exponents. We retain the joint
coordinates needed to distinguish movement of that base, movement of the
exponents, optimizer memory and the imposed training clock. Throughout
this section, the input $x$ is fixed and excluded from gradient updates.
Write $M_t=\Alm(\theta_t;x)>\bm0$, $P_t=P(\theta_t)$ and
$q_t=\log V_t=P_t\odot\log M_t$, flattening any specified collection of
layers and heads into a finite coordinate vector.

\begin{proposition}[Potential increments and compatible temporal blocking]
\label{model:prop:potential-block}
Let $u_t=(P_{t+1}-P_t)\odot\log M_t$ and
$v_t=P_{t+1}\odot(\log M_{t+1}-\log M_t)$. For fixed nonnegative
coordinate weights $w_i$ summing to one, define
$\langle a,b\rangle_w=\sum_iw_i a_i b_i$. Then
\begin{align}
 \Delta q_t&=u_t+v_t,\label{model:eq:potential-split}\\
 \|\Delta q_t\|_w^2
 &=\|u_t\|_w^2+\|v_t\|_w^2+2\langle u_t,v_t\rangle_w.
 \label{model:eq:potential-energy}
\end{align}
For an interval $I=\{a,\ldots,b-1\}$, the blocked increment is
\begin{align}
 Q_I&=\sum_{t\in I}\Delta q_t=q_b-q_a,\\
 \|Q_I\|_w^2&=\sum_{s,t\in I}\langle\Delta q_s,\Delta q_t\rangle_w.
 \label{model:eq:potential-block-energy}
\end{align}
Consequently interval composition is associative for the signed
increments, while their norm cannot generally be reconstructed from the
individual squared norms alone.
\end{proposition}
\begin{proof}
Subtract $P_t\odot\log M_t$ from $P_{t+1}\odot\log M_{t+1}$ and
insert $P_{t+1}\odot\log M_t$. This proves the first identity.
Expanding each coordinate square and summing with weights gives the
second. The interval sum telescopes. Expanding the square of that sum
retains every pair of time indices, giving the final formula.
Associativity follows from regrouping a finite sum.
\end{proof}

\subsection{Endpoint allocation and the native operator}
\begin{proposition}[Symmetric potential increments and allocation error]
\label{model:prop:potential-symmetric}
Let \(M,M'\) have strictly positive coordinates, let \(P,P'\) be real
coordinate vectors, and write \(b=\log M\), \(b'=\log M'\),
\(q=P\odot b\), \(q'=P'\odot b'\). Define
\[
 \Delta P=P'-P,\quad \Delta b=b'-b,\quad c=\Delta P\odot\Delta b,
 \qquad
 u_*=\Delta P\odot\frac{b+b'}2,\quad
 v_*=\frac{P+P'}2\odot\Delta b .
\]
Then
\begin{align}
 q'-q&=u_*+v_*, \label{model:eq:symmetric-increment}\\
 q'-q&=u_0+v_0,\qquad
 u_0=\Delta P\odot b,\quad v_0=P'\odot\Delta b,\\
 u_*-u_0&=\tfrac12c,\qquad v_*-v_0=-\tfrac12c. \label{model:eq:potential-allocation-error}
\end{align}
For fixed nonnegative weights \(w_i\) summing to one,
\[
 \norm{q'-q}_w^2=\norm{u_*}_w^2+\norm{v_*}_w^2
                       +2\langle u_*,v_*\rangle_w .
\]
In particular,
\[
 \big|\norm{u_*}_w-\norm{u_0}_w\big|\leq\tfrac12\norm{c}_w,
 \quad
 \big|\norm{v_*}_w-\norm{v_0}_w\big|\leq\tfrac12\norm{c}_w.
\]
If \(\norm{\Delta P}_\infty\leq C_P h\) and
\(\norm{\Delta b}_w\leq C_b h\), the allocation discrepancy is at
most \(C_PC_bh^2/2\) in each component norm.
\end{proposition}
\begin{proof}
Expanding the two symmetric terms cancels their mixed endpoint products
and leaves \(P'\odot b'-P\odot b\).
Subtracting the one-sided expressions gives \(c/2\) and \(-c/2\).
Expansion of each squared coordinate and weighted summation proves the
energy identity. The reverse triangle inequality proves both norm bounds.
Finally,
\(\norm{\Delta P\odot\Delta b}_w
 \leq\norm{\Delta P}_\infty\norm{\Delta b}_w\).
\end{proof}

The symmetric allocation reverses sign when the two endpoints are exchanged.
Equivalently it assigns half of the exact mixed corner
\[
 P'\odot b'-P'\odot b-P\odot b'+P\odot b
       =\Delta P\odot\Delta b
\]
to each coordinate change. This is a transparent algebraic convention, not
an independent parameter intervention. In the native program \(M\) can
itself depend on parameters included in \(P\); mixed observation corners
need not correspond to separately realizable training states.

This refinement does not change the compatible block object:
\(q_b-q_a=\sum_{t=a}^{b-1}\Delta q_t\).
Block energy still contains every cross-time product. Excursion thresholding
and taking norms do not generally commute with blocking. An RG proposal
built only from event sizes must therefore establish its own compatibility
and closure, rather than inheriting them from signed increments.

\subsection{A signed-area completion for component blocking}
The signed total increment telescopes, but its separate symmetric allocations
carry an additional path coordinate. The following finite construction is
the antisymmetric second-order path-composition identity associated with
iterated-integral algebra \cite{chen1957}, specialized here to the native
product observation $q=P\odot b$. Its use does not assert predictive closure
of the retained path coordinates.

\begin{proposition}[Compatible blocking of potential components]
\label{model:prop:potential-area}
For an interval $I=\{a,\ldots,b-1\}$, set $d_t=P_{t+1}-P_t$,
$e_t=b_{t+1}-b_t$, $d_I=\sum_{t\in I}d_t$, $e_I=\sum_{t\in I}e_t$ and
\[
 \mathcal A_I=\frac12\sum_{a\le s<t<b}
       (d_t\odot e_s-d_s\odot e_t).
\]
Then the accumulated symmetric components obey
\begin{align}
 \sum_{t\in I}u_{*,t}&=d_I\odot\frac{b_a+b_b}{2}+\mathcal A_I,\\
 \sum_{t\in I}v_{*,t}&=\frac{P_a+P_b}{2}\odot e_I-\mathcal A_I.
 \label{model:eq:potential-area-components}
\end{align}
For chronologically adjacent intervals $I,J$ their retained coordinates
compose as
\begin{equation}
 (d,e,A)\star(\widetilde d,\widetilde e,\widetilde A)
 =\left(d+\widetilde d,e+\widetilde e,
 A+\widetilde A+\frac{\widetilde d\odot e-d\odot\widetilde e}{2}\right).
 \label{model:eq:potential-area-compose}
\end{equation}
This operation is associative, with identity $(\bm0,\bm0,\bm0)$ and inverse
$(-d,-e,-A)$. For scalar coordinates and starting values $(p,b)$, put
\[
 U_b(d,e,A)=d(b+e/2)+A,\qquad
 V_p(d,e,A)=(p+d/2)e-A.
\]
Writing $X=(d,e,A)$ and $Y=(D,E,B)$, one has
\begin{align}
 U_b(X\star Y)&=U_b(X)+U_{b+e}(Y),\\
 V_p(X\star Y)&=V_p(X)+V_{p+d}(Y),\\
 U_b(X)+V_p(X)&=(p+d)(b+e)-pb.
 \label{model:eq:potential-area-allocation}
\end{align}
Exchanging the two blocks changes their composed area by $De-dE$.
Tensor allocations obey these identities coordinatewise. Under the scale normalization
$(d,e,A)\mapsto(r d,s e,rs A)$ it remains compatible with composition.
For either accumulated component, dropping $\mathcal A_I$ changes its
weighted norm by at most $\|\mathcal A_I\|_w$.
\end{proposition}
\begin{proof}
Expand $b_t=b_a+\sum_{s<t}e_s$. The sum of the first symmetric
component becomes
$d_I\odot b_a+\tfrac12\sum_t d_t\odot e_t+
\sum_{s<t}d_t\odot e_s$.
The coarse endpoint component has the same first two terms and
$\tfrac12\sum_{s<t}(d_t\odot e_s+d_s\odot e_t)$.
Their difference is $\mathcal A_I$. The second identity follows by
subtracting the first from the exact total increment.
Split the strictly ordered index pairs of $I\cup J$ into pairs within
$I$, within $J$, and one index in each. The cross pairs give the last
term in Equation~\eqref{model:eq:potential-area-compose}. For three intervals,
either association includes their three internal areas and the same
three cross-pair terms, proving associativity. Bilinearity makes every
area term acquire the factor $rs$, proving scale compatibility.
Substitution proves both identity and inverse statements. Expanding the
three allocation expressions gives Equation~\eqref{model:eq:potential-area-allocation};
the initial endpoints shift by the first block's displacements.
Subtracting the two cross terms gives $De-dE$. A one-step block has
area zero, so induction gives the accumulated allocations for every
finite path. Finally
apply the reverse triangle inequality to each component and its signed
area correction.
\end{proof}

If the normalization factors satisfy $r_{uv}=r_ur_v$ and
$s_{uv}=s_us_v$, grouping adjacent blocks and applying these dilations
gives compatible nested temporal maps. The starting values $(P_a,b_a)$
are also needed to recover the absolute allocations. Selecting dilation
factors alone does not determine physical scaling exponents.

The positive-base and exponent path therefore has an exact finite block
state beyond endpoint displacement alone. A scalar activity trace loses
this signed order information. The area measures component transport,
not spatial propagation, a physical volume, or a thermodynamic exponent.
A small one-update allocation discrepancy does not imply a uniformly
small coarse-interval area. Both are measured at declared temporal scales.
Sections~\ref{model:sec:potential-results} and \ref{model:sec:finite-metric-results} give the activity measurements and finite predictive metric budgets.

The downstream operator has its own finite transport identity. If
$G_t=a_tV_t+b_t^a$, then
\begin{equation}
 \Delta G_t=(\Delta a_t)V_t+a_{t+1}\Delta V_t+\Delta b_t^a,
 \qquad \Delta V_t=V_t\odot\bigl(e^{\Delta q_t}-\bm1\bigr).
 \label{model:eq:potential-operator-transport}
\end{equation}
Here the exponential is applied coordinatewise, and $\bm1$ is the
vector of ones of the same dimension as $V_t$.
Adding and subtracting $a_{t+1}V_t$ proves the first identity; the
second follows from $V_{t+1}=V_t\odot e^{\Delta q_t}$.
A potential excursion can therefore be attenuated, amplified or opposed
by its learned coupling into $G$. Metric, base, exponent, potential and
curvature increments are distinct observations of the same training
state, with separate predictive projections.

The two contributions in Equation~\eqref{model:eq:potential-energy} are a
pathwise decomposition, not independent mechanisms: training changes
both through the same loss and optimizer. The base contribution includes
changes in the learned left multiplier, bias and metric generator, as
well as changes propagated from the body. Row collapse therefore does
not imply a stationary positive base. Its common row can keep moving.
The fraction $\sum_t\|v_t\|_w^2/
\sum_t(\|u_t\|_w^2+\|v_t\|_w^2)$ reports allocation between these
coordinates; it is not the fraction of total activity unless the signed
cross contribution is also accounted for.

\begin{proposition}[Clock-conditioned potential velocity]
\label{model:prop:potential-clock}
Let $q(\theta;x)$ be twice continuously differentiable on a region
containing a parameter step $\theta_{t+1}=\theta_t+\alpha_t b_t$,
where $\alpha_t>0$. If the operator norm of its second derivative is
at most $H$ along that segment, then
\begin{equation}
 \left\|\frac{\Delta q_t}{\alpha_t}
       -Dq(\theta_t;x)b_t\right\|
 \leq \frac{H\alpha_t}{2}\|b_t\|^2.
 \label{model:eq:potential-clock-bound}
\end{equation}
Writing $r_t=\|\Delta q_t\|/\alpha_t$, an activity threshold obeys
$\|\Delta q_t\|>c$ if and only if $r_t>c/\alpha_t$.
For an interval $I$, let $A_I=\sum_{t\in I}\alpha_t$. Its velocity is
$Q_I/A_I=\sum_{t\in I}(\alpha_t/A_I)(\Delta q_t/\alpha_t)$.
Adjacent interval velocities compose with weights proportional to
$A_I$.
\end{proposition}
\begin{proof}
The integral second-order Taylor remainder is bounded by
$H\|\alpha_t b_t\|^2/2$. Divide by the positive step size.
The threshold equivalence is division of a strict inequality by a
positive number. Substituting the definition of each velocity into its
weighted average proves the interval identity. Addition of the
numerators and denominators proves the composition rule.
\end{proof}

This proposition applies to the parameter component of the augmented
optimizer step; $b_t$ still depends on moments and the unconsumed corpus.
Dividing activity by the learning rate is a diagnostic for imposed
amplitude changes. It neither removes optimizer memory nor converts
training into a stationary process. Linear warm-up and annealing can
change a threshold excursion even for a constant underlying velocity.
The experiment therefore observes both clocks and retains its declared
phase boundaries. Event extraction is a nonlinear observation and need
not commute with temporal blocking in
Proposition~\ref{model:prop:potential-block}.
Section~\ref{model:sec:potential-excursions} specifies the two observed clocks, phase windows and correlated controls.

The native positive base also supplies a specific local amplification
mechanism. For a preactivation $z$ and fixed exponent $p$,
\begin{equation}
 \partial_z\{p\log[f(z)+\epsilon_A]\}
 =p\frac{f'(z)}{f(z)+\epsilon_A},\qquad
 f(z)=z^2\sigma(z).
 \label{model:eq:potential-base-gain}
\end{equation}
Near zero, $f(z)=z^2/2+O(z^3)$. The leading tangent observation along a
preactivation velocity, with $c=p\dot z$ fixed, is therefore
$w_\epsilon(z)=|cz|/(\epsilon+z^2/2)$. This is a local observation
approximation, not a closed equation for native training.

\begin{proposition}[Regularized positive-base amplification]
\label{model:prop:potential-base-amplification}
For $\epsilon>0$ and $c\ne0$,
$\sup_z w_\epsilon(z)=|c|/\sqrt{2\epsilon}$.
If a random preactivation $Z$ has a density continuous at zero with
$\rho(0)>0$, the zero-offset observation
$w_0(Z)=2|c|/|Z|$ satisfies
\begin{equation}
 \Pr(w_0(Z)>u)\sim\frac{4|c|\rho(0)}{u}
 \quad(u\longrightarrow\infty).
 \label{model:eq:potential-base-tail}
\end{equation}
Independent preactivations already suffice for this tail. Every positive
offset instead imposes the finite upper bound above.
\end{proposition}
\begin{proof}
For $z>0$, differentiation of $|c|z/(\epsilon+z^2/2)$ gives a derivative
with the sign of $\epsilon-z^2/2$. Symmetry yields maxima at
$|z|=\sqrt{2\epsilon}$, with the stated value. At zero offset,
$w_0(Z)>u$ is equivalent to $|Z|<2|c|/u$, up to the null event $Z=0$.
Integrating the continuous density over that shrinking interval gives
Equation~\eqref{model:eq:potential-base-tail}. No dependence assumption among
separate draws was used.
\end{proof}

Thus a regularized observation can have a broad amplitude range without
a collective critical point. Each native training branch fixes its specified positive offset. Raising it only during passive reconstruction
changes an observation of the recorded path; it does not predict the
outcome of retraining a different architecture. The empirical entry-panel diagnostics in Section~\ref{model:sec:potential-long-results} test concentration of activity at small bases without
assigning the idealized tangent tail exponent to native avalanches.

\begin{proposition}[A power potential does not identify criticality]
\label{model:prop:potential-tail-counterexample}
Fix a scalar base $m>1$. If an exponent $P$ has exponential survival
probability $\Pr(P>p)=e^{-\lambda p}$ for $p\geq0$, then the potential
$V=m^P$ has the exact power-law survival probability
\begin{equation}
 \Pr(V>z)=z^{-\lambda/\log m},\qquad z\geq1.
 \label{model:eq:potential-induced-tail}
\end{equation}
Independent exponents at successive times give this marginal law
without temporal propagation. For a fixed threshold with exceedance
probability $\pi\in(0,1)$, the duration $T$ of an exceedance run,
conditional on a run starting, satisfies
$\Pr(T=k)=(1-\pi)\pi^{k-1}$ for $k\geq1$.
\end{proposition}
\begin{proof}
Since $m>1$, $m^P>z$ is equivalent to $P>\log z/\log m$.
Substitution gives Equation~\eqref{model:eq:potential-induced-tail}.
Independence makes threshold exceedances Bernoulli variables. Given the
first exceedance, a run of length $k$ requires $k-1$ further exceedances
and one failure, giving the stated probability.
\end{proof}

Similarly, Gaussian exponents at a fixed base give a lognormal potential.
These examples concern marginal potential values, whereas the observed
excursion sizes are functions of the entire update path. They distinguish
three claims: a learned power operation, a heavy-tailed observable, and
an interacting critical process. Even power-law excursion statistics and
scaling forms can occur in noncritical stochastic systems
\cite{touboul2017}. Tail fitting therefore uses likelihood and
simulation diagnostics, with alternative distributions and explicit
serial-dependence limitations, rather than a straight-line fit to a
logarithmic histogram \cite{clauset2009}.

\begin{proposition}[Residual motion after removing gradient forcing]
\label{model:prop:potential-zero-gradient}
Consider AdamW at optimizer time $t$ with first and second moments
$m_t,v_t$, coefficients $\beta_1,\beta_2\in(0,1)$, positive denominator
offset $\epsilon$, and weight decay $\lambda$. After setting each new
loss gradient explicitly to zero, the next $k$ steps satisfy
\begin{alignat}{2}
 m_{t+k}&=\beta_1^k m_t,&\quad v_{t+k}&=\beta_2^k v_t,\\
 \theta_{t+k+1}
 &=(1-\alpha_{t+k}\lambda)\theta_{t+k}
 -\alpha_{t+k}
 \frac{\beta_1^{k+1}m_t/(1-\beta_1^{t+k+1})}
 {\sqrt{\beta_2^{k+1}v_t/(1-\beta_2^{t+k+1})}+\epsilon}.
 \label{model:eq:potential-passive-adam}
\end{alignat}
Products, divisions and square roots in the second line are coordinatewise.
Resetting both moments removes its second term; skipping the optimizer
entirely leaves the parameter-dependent potential fixed.
\end{proposition}
\begin{proof}
With zero new gradient, the moment recursions multiply their incoming
values by $\beta_1$ and $\beta_2$ at each step. Induction yields the
first line. Insert those values into the bias-corrected AdamW update at
time $t+k+1$. Zero incoming moments remain zero, while an unchanged
parameter vector gives the same deterministic fixed-input forward map.
\end{proof}

This intervention isolates passive optimizer motion. It does not emulate
relaxation driven by the native loss gradient. A sandpile interpretation
would additionally need an identified state variable, redistribution and
dissipation law, and a separation between external drive and collective
relaxation. Self-organization would require evidence that feedback
attracts the relevant control variable towards a critical window, with a
specified size and remaining-corpus limit, rather than merely traversing
that window under an imposed schedule. These are mechanism questions;
entrywise positivity and a threshold applied during analysis do not
supply their answers. Smooth finite native updates do not exclude an
emergent collective transition. The completed measurements in Section~\ref{model:sec:potential-memory-results} delimit
which parts of this mechanism are presently resolved.

\subsection{Transport under actual training interventions}
\label{model:sec:potential-intervention-law}
An offset changed inside native training belongs to the transition kernel,
not only to the observation. Let $F_{\epsilon,t}(s,z)$ denote the complete
augmented update at offset $\epsilon$ and matched next source block $z$.
The state includes both Adam moments, counters and the scheduler.

\begin{proposition}[Coupled offset and optimizer-state transport]
\label{model:prop:potential-intervention-transport}
Consider two paths with the same source sequence and two fixed positive
offsets $\epsilon,\epsilon'$. Suppose nonnegative constants $L_t,B_t$
satisfy, on a region containing both paths,
\[
 \|F_{\epsilon,t}(s,z)-F_{\epsilon',t}(s',z)\|
 \le L_t\|s-s'\|+B_t|\epsilon-\epsilon'|,
\]
and an emission $H_\epsilon$ satisfies the analogous bound with nonnegative constants
$L_H,B_H$. Set $D_0=\|s_0-s'_0\|$ and
$\Phi_{b:a}=\prod_{j=a}^{b-1}L_j$, with empty product one. Then
\[
 \|H_\epsilon(s_T)-H_{\epsilon'}(s'_T)\|
 \le L_H\left(\Phi_{T:0}D_0+
 |\epsilon-\epsilon'|\sum_{j=0}^{T-1}\Phi_{T:j+1}B_j\right)
 +B_H|\epsilon-\epsilon'|.
\]
Resetting only the first Adam moment changes $D_0$ even when parameters,
second moments and all forward tensors initially agree.
\end{proposition}
\begin{proof}
Apply the assumed update inequality to the coupled states to obtain
$D_{t+1}\le L_tD_t+B_t|\epsilon-\epsilon'|$.
Induction expands this recursion into its chronological product and sum.
Apply the emission inequality at time $T$. The first-moment coordinate is
part of the state norm, so resetting it need not give zero initial distance.
\end{proof}

These are conditional stability constants, not estimates obtained from
activity correlation. At a single gate with fixed incoming activation and
parameters, the direct log-potential difference is exactly
$P\odot\log(1+(\epsilon'-\epsilon)/M)$, where
$M=\operatorname{iSwiGLU}(Z)+\epsilon$ and both bases are positive.
Its magnitude can be large for small $M$ while its image in $V$, $G$ or
predictive logits is small. With all gates changed, subsequent decoder
inputs and loss gradients also change. The coupled bound retains those
paths and the independent initial optimizer-state perturbation.
Section~\ref{model:sec:potential-factorial-results} measures both interventions
and their interaction under the same unused RefinedWeb suffix.

\subsection{Finite categorical transport of an intervention}
The tensor-to-emission connection has an exact finite form that retains
the changing predictive metric. For fixed evaluation context, let
$\ell_t^j$ denote the full-vocabulary logits of branch $j\in\{0,1\}$.
The source sequence, complete incoming state and native offset specify
each branch. Define $d_j=\ell_{t+1}^j-\ell_t^j$ and $e=d_1-d_0$.

The categorical identity is the exponential-family KL/Bregman relation
for the log-partition function, followed by the integral Taylor formula
\cite{nielsen2020}. The specialization below retains both the signed
PLDR intervention and its changing predictive geometry.

\begin{proposition}[Signed finite predictive transport]
\label{model:prop:finite-categorical-transport}
For finite logits $\ell\in\R^V$, let
$A(\ell)=\log\sum_i e^{\ell_i}$, $p(\ell)=\softmax(\ell)$ and
$F(\ell)=\diag p(\ell)-p(\ell)p(\ell)^{\mathsf T}$. Set
\[
 \overline F_j=2\int_0^1(1-s)F(\ell_t^j+s d_j)\,ds,
 \qquad K_j=\KL(p(\ell_t^j)\|p(\ell_{t+1}^j)).
\]
Then
\begin{align}
 K_j&=\tfrac12d_j^{\mathsf T}\overline F_jd_j,\\
 K_1-K_0&=e^{\mathsf T}\overline F_0d_0
       +\tfrac12e^{\mathsf T}\overline F_0e
       +\tfrac12d_1^{\mathsf T}(\overline F_1-\overline F_0)d_1.
 \label{model:eq:finite-kl-contrast}
\end{align}
The matrices $\overline F_j$ are positive semidefinite and annihilate
the constant-logit direction. The second term in the contrast is
nonnegative; the first and third need not have a fixed sign.
Summing the identity over updates gives the integrated predictive
contrast. Independent time-dependent constant-logit shifts in either
branch leave every displayed scalar unchanged.
\end{proposition}
\begin{proof}
Direct substitution in categorical KL gives
$K_j=A(\ell_t^j+d_j)-A(\ell_t^j)-p(\ell_t^j)^{\mathsf T}d_j$.
Differentiating the finite exponential sum gives
$\nabla A=p$ and $D^2A=F$. The integral Taylor formula therefore
proves the first identity. In its difference substitute $d_1=d_0+e$,
add and subtract $d_1^{\mathsf T}\overline F_0d_1/2$, and expand
the symmetric quadratic form. This gives the second identity.
For every vector $v$, $v^{\mathsf T}Fv$ is the categorical variance
of its coordinates, hence is nonnegative and vanishes on constants.
Integration preserves these properties. Finite summation proves the
integrated identity. Softmax is invariant to a constant shift; the
remaining constant components are killed by both covariance matrices.
\end{proof}

The straight logit segments define finite observation geometry; they
are not interpolated native training trajectories. Equation~\eqref{model:eq:finite-kl-contrast}
is an allocation of an actual paired response, not an identified causal
mediation path. It explains the information required in a predictive
collective state: signed logit increments and their evolving categorical
geometry, with the source and optimizer coordinates that determine them.
The local Fisher approximation replaces $\overline F_j$ by $F(\ell_t^j)$
and must bound the resulting remainder. A decrease in a potential norm
alone fixes none of the three contrast terms. At frozen inference the
same identity applies to two declared inference interventions; optimizer
activity is then absent from the intervened state.

\subsection{A finite budget for metric replacement}
\label{model:sec:metric-budget}

Finite predictive transport permits a changing categorical metric.
The following budget controls its replacement by the initial Fisher
metric, using the oscillation of the full-vocabulary logit increment.
The Taylor comparison has the modified self-concordant antecedent in
Bach's Proposition~1 \cite{bach2010selfconcordant}. The specialization here
uses categorical logit oscillation, its gauge invariance and the PLDR
predictive observation budget. It requires only finite logits and is invariant to
additive logit gauges. It provides an observation error budget, not a
forecast of logits that have not been acquired.

For $A(s)=\log\sum_i e^{\ell_i+sd_i}$, differentiation gives
$A''(s)=\Var_{p_s}(d)$ and
$A'''(s)=\E_{p_s}(d-\E_{p_s}d)^3$.
Since $|d_i-\E_{p_s}d|\le r$, the path satisfies
$|A'''(s)|\le rA''(s)$. Integrating the logarithmic derivative
when $d$ is nonconstant, and then the integral Taylor remainder, yields
the exponential comparison below. Constant increments have zero divergence.
The proof also gives a direct weighted-variance derivation.

\begin{proposition}[Finite metric replacement budget]
\label{model:prop:metric-budget}
Let $p=\softmax(\ell)$, $d=\ell'-\ell$, and
$p_s=\softmax(\ell+sd)$ for $s\in[0,1]$. Put
\[
 r=\max_i d_i-\min_i d_i,\qquad
 Q=\tfrac12\Var_p(d),\qquad K=\KL(p\|p_1).
\]
Define $a_\pm(0)=1$ and, for $r>0$,
\[
 a_+(r)=\frac{2(e^r-1-r)}{r^2},\qquad
 a_-(r)=\frac{2(e^{-r}-1+r)}{r^2}.
\]
Then
\begin{equation}
 a_-(r)Q\le K\le a_+(r)Q.
 \label{model:eq:metric-budget}
\end{equation}
In particular, with
$\eta(r)=\max\{a_+(r)-1,\,1-a_-(r)\}$,
$|K-Q|\le\eta(r)Q$. As $r\to0$, $\eta(r)=r/3+O(r^2)$.
For two branches $j=0,1$, a finite contrast satisfies
\begin{equation}
 a_-(r_1)Q_1-a_+(r_0)Q_0
 \le K_1-K_0\le
 a_+(r_1)Q_1-a_-(r_0)Q_0.
 \label{model:eq:contrast-budget}
\end{equation}
The bounds do not require a lower bound on the minimum token probability.
\end{proposition}

\begin{proof}
The exponential tilt gives
\[
 \frac{p_{s,i}}{p_i}=\frac{e^{sd_i}}{\sum_jp_je^{sd_j}},
 \qquad e^{-sr}\le\frac{p_{s,i}}{p_i}\le e^{sr}.
\]
For any finite scalar vector $v$,
$\Var_p(v)=\min_c\sum_i p_i(v_i-c)^2$.
Pointwise comparison of the two weighted square sums, followed by
minimization, therefore yields
\[
 e^{-sr}\Var_p(v)\le\Var_{p_s}(v)\le e^{sr}\Var_p(v).
\]
For $A(\ell)=\log\sum_i e^{\ell_i}$, direct substitution gives
$K=A(\ell+d)-A(\ell)-\nabla A(\ell)^{\mathsf T}d$.
Since $D^2A=\diag(p)-pp^{\mathsf T}$, integral Taylor expansion gives
$K=\int_0^1(1-s)\Var_{p_s}(d)\,ds$.
Integrating the preceding variance comparison proves
\eqref{model:eq:metric-budget}, because
$2\int_0^1(1-s)e^{\pm sr}\,ds=a_\pm(r)$.
At $r=0$, $d$ is constant and $K=Q=0$. Expanding the exponential
proves the small-$r$ statement. Subtract the two branch intervals to
obtain \eqref{model:eq:contrast-budget}. A separate constant added to either
endpoint changes $d$ only by a constant and leaves its oscillation,
variance and categorical probabilities unchanged.
\end{proof}

\begin{corollary}[Controlled segment refinement]
\label{model:cor:refinement}
For an integer $m\ge1$, set $c_j=(j+1/2)/m$,
$w_j=(1-c_j)/m$, $j=0,\ldots,m-1$, and
\[
 T_m=\sum_{j=0}^{m-1}w_j\Var_{p_{c_j}}(d).
\]
Then
\begin{equation}
 e^{-r/(2m)}T_m\le K\le e^{r/(2m)}T_m.
 \label{model:eq:refinement}
\end{equation}
Intervals from successive refinements may be intersected. Their widths
converge to zero for fixed finite endpoints. Summing branch intervals
over updates and subtracting the two sums gives a valid interval for
the difference of summed step KL values.
\end{corollary}

\begin{proof}
For $s\in[j/m,(j+1)/m]$, comparison to $p_{c_j}$ gives the same
variance bounds with factors $e^{\pm |s-c_j|r}$, hence with the
uniform factors $e^{\pm r/(2m)}$. The integral of $(1-s)$ over that
segment equals $w_j$. Integrate and sum. Continuity on the compact
segment makes $T_m$ a convergent weighted Riemann sum; the two
exponential factors converge to one. Intersection preserves coverage,
as do finite sums and interval subtraction.
\end{proof}

For a prescribed relative tolerance $\varepsilon>0$, the choice
\[
 m\ge\max\left\{1,\left\lceil\frac{r}{2\log(1+\varepsilon)}\right\rceil\right\}
\]
therefore gives $|T_m/K-1|\le\varepsilon$ whenever $K>0$.
Indeed the exponential factor in \eqref{model:eq:refinement} is at most
$1+\varepsilon$, and the reciprocal interval has the same two factors.
Constant increments instead have $K=T_m=0$.

These are real-arithmetic results. Evaluating them in float64 and checking
against endpoint KL does not turn them into rigorous floating-point
interval certificates. Nor is a straight logit segment a native training
trajectory through interpolated parameters. The exact KL/Bregman identity
is classical; the bound supplies a transparent sufficient
budget for its PLDR observation approximation.

For aligned temporal blocking, retain the endpoint logit class modulo
constants and its signed increment $d_I=\ell_b-\ell_a$ alongside the
potential block $(d_I^P,d_I^b,\mathcal A_I)$. Adjacent intervals compose
by $d_{I\cup J}=d_I+d_J$ and the area law
\eqref{model:eq:potential-area-compose}; their intermediate endpoints must agree.
The same finite metric bound is then evaluated on $(\ell_a,d_{I\cup J})$
at each scale. The interval bounds constrain a compatible endpoint
observation, while summing step KL constrains a different path functional.
Neither operation supplies the unknown next logit increment. The measured
scale flow therefore retains its source, optimizer and observation
conditioning when used in the joint-law construction.
Section~\ref{model:sec:finite-metric-results} evaluates these endpoint and path budgets on the native predictive probes.

\section{Transported sources and predictive response}
\label{model:sec:response}

Probability sources $j$ in $W$ reweight observations. A separate,
instrumental source $\eta\in\R^q$ perturbs the forward computation itself.
For example, $\eta$ can perturb model weights, embeddings, or any specified
generated tensor. We write $F_\ell(X;\eta)$ and hold its definition fixed
when coarse graining. This distinction prevents a data susceptibility from
being confused with a parameter Hessian.

\subsection{The complete derivative cocycle}
At an unperturbed context $x$, let
\begin{equation}
 J_\ell(x)=\partial_X F_\ell(X_{\ell-1};0),\qquad
 B_\ell(x)=\partial_\eta F_\ell(X_{\ell-1};0).
\end{equation}
The shared source space includes all head sources, even when only a subset
is inserted directly at a particular layer. Its tangent propagation is
\begin{equation}
 \delta X_\ell=J_\ell\delta X_{\ell-1}+B_\ell\eta.\label{model:eq:tangent}
\end{equation}
Each $J_\ell$ includes the derivative through the Gram, the complete metric
generator, the power map, attention, output mixing, both residual paths,
both normalizations, and the feed-forward map.

\begin{lemma}[Source-preserving derivative RG]\label{model:thm:jet}
For two chronological tangent blocks,
\begin{equation}
 (J_2,B_2)\circ(J_1,B_1)
       =(J_2J_1,\ J_2B_1+B_2).\label{model:eq:jet-compose}
\end{equation}
This composition is associative. For the block $[a,b]$ its coefficients are
\begin{align}
 J_{b:a}&=J_b\cdots J_a,\\
 B_{b:a}&=\sum_{k=a}^b J_b\cdots J_{k+1}B_k,\label{model:eq:source-product}
\end{align}
where an empty product is the identity. They are exactly the derivatives
of the nonlinear block in Lemma~\ref{model:thm:graph}. Thus differentiating
then blocking agrees with blocking then differentiating.
\end{lemma}
\begin{proof}
Substitute $\delta X_1=J_1\delta X_0+B_1\eta$ into the second tangent
equation to obtain~\eqref{model:eq:jet-compose}. For three blocks, both
association orders give
\[
 (J_3J_2J_1,\ J_3J_2B_1+J_3B_2+B_3).
\]
Induction proves~\eqref{model:eq:source-product}. The chain rule for
$F_b(\cdots F_a(X;\eta)\cdots;\eta)$ produces the same product for the
boundary derivative and one term for each occurrence of the shared source.
This proves the derivative identity. Smoothness is supplied by
Proposition~\ref{model:prop:regularity}.
\end{proof}

The matrices generally do not commute. A norm or eigenvalue calculated
separately at each head does not reconstruct the transported sum. Nor does
$\E_\D(J_2J_1)$ usually equal $(\E_\D J_2)(\E_\D J_1)$: both factors
depend on the same context. The exact dataset state therefore retains the
law of the context-dependent jets. Averages may be taken after the
chronological composition, with their estimation uncertainty retained.

For reference, the PLGA differential itself has the explicit local chain
\begin{align}
 \delta\Alm&=f'(WA+b)\odot\delta(WA+b),\\
 \delta P^{\mathrm{pot}}
 &=P^{\mathrm{pot}}\odot\left(P\odot\frac{\delta\Alm}{\Alm}
                                  +\log\Alm\odot\delta P\right),\\
 \delta\G&=(\delta a)P^{\mathrm{pot}}+a\,\delta P^{\mathrm{pot}}+\delta b^a,\\
 \delta S&=\frac{(\delta Q)\G K^{\mathsf T}
                  +Q(\delta\G)K^{\mathsf T}+Q\G(\delta K)^{\mathsf T}}{\sqrt d},
 \label{model:eq:plga-differential}
\end{align}
where all entrywise divisions are well defined. For an attention row $t$,
$\delta t=(\diag(t)-tt^{\mathsf T})\delta S$. These formulas make explicit
the distinction between an operator perturbation and the response after
the rest of the model has acted.

\subsection{A source metric at the predictive output}
Let $z_x(\eta)$ be the final vocabulary logits and
$\mathcal J_x=\partial_\eta z_x(0)$ their complete source Jacobian. Set
\begin{equation}
 F(p)=\diag(p)-pp^{\mathsf T},\qquad
 \Hess_x=\mathcal J_x^{\mathsf T}F(p_x)\mathcal J_x,
 \qquad \Hess_\D=\E_\D\Hess_x.\label{model:eq:fisher}
\end{equation}
$\Hess_x$ is the pullback of categorical Fisher information into the
specified intervention coordinates. Its dimension need only equal the
number of tested sources, although its computation uses the full decoder
and vocabulary.

\begin{theorem}[Local whole-model prediction]\label{model:thm:kl}
For a smooth source map and a fixed context,
\begin{equation}
 \KL\bigl(p_x(0)\,\|\,p_x(\eta)\bigr)
       =\tfrac12\eta^{\mathsf T}\Hess_x\eta+O(\norm{\eta}^3).
       \label{model:eq:kl-response}
\end{equation}
$\Hess_x$ is positive semidefinite. Adding any source-dependent constant
to every vocabulary logit leaves it unchanged. Under an invertible source
change $\eta=T\xi$, it transforms by $\Hess_x\mapsto T^{\mathsf T}\Hess_xT$.
For finite $\D$ and a fixed compact sufficiently small source neighborhood,
the remainder can be bounded uniformly over contexts; averaging gives
\begin{equation}
 \E_\D\KL\bigl(p_x(0)\,\|\,p_x(\eta)\bigr)
 =\tfrac12\eta^{\mathsf T}\Hess_\D\eta+O(\norm{\eta}^3).
 \label{model:eq:ensemble-kl}
\end{equation}
\end{theorem}
\begin{proof}
Let $A(z)=\log\sum_v e^{z_v}$ and $p_0=\softmax z_x(0)$. Up to a
constant independent of $\eta$, the KL divergence equals
$A(z_x(\eta))-p_0^{\mathsf T}z_x(\eta)$. At zero source its gradient with
respect to $z$ vanishes, and its Hessian is $F(p_0)$. The second derivative
of the composite therefore has no contribution from second derivatives
of $z_x$ multiplied by a nonzero first derivative; it is exactly
$\mathcal J_x^{\mathsf T}F(p_0)\mathcal J_x$. Taylor's theorem proves
\eqref{model:eq:kl-response}.

For any logit direction $v$,
\[
 v^{\mathsf T}F(p_0)v
 =\sum_i p_{0i}(v_i-\bar v)^2\geq0,
 \qquad \bar v=\sum_i p_{0i}v_i.
\]
This proves positivity and shows that a constant direction is annihilated.
It proves the logit-gauge statement as well. The source change replaces
$\mathcal J_x$ by $\mathcal J_xT$, giving the congruence. On a compact
source neighborhood the third derivatives for each context are bounded;
there are finitely many contexts, so their maximum is finite. Integrating
the uniform remainder proves~\eqref{model:eq:ensemble-kl}.
\end{proof}

The theorem specifies a pointwise response field $x\mapsto\Hess_x$ and
an ensemble coefficient $\Hess_\D$. It does not replace the former by the
latter in a prediction about one context. We estimate the former on each
validation context and test new simultaneous source directions. Source
spectra depend on the stated coordinate units; a small eigenvalue of
$\Hess_x$ means weak predictive sensitivity in that direction and is not
itself a divergent statistical susceptibility.
Section~\ref{model:sec:whole-model-response-results} specifies the source directions, context splits and finite-difference checks.

Our executed source is dimensionless fractional gain at every head:
\begin{equation}
 \G^{(\ell a)}\longmapsto (1+\eta_{\ell a})\G^{(\ell a)},
 \qquad \eta\in\R^{5\times14}.\label{model:eq:measured-source}
\end{equation}
It is implemented equivalently by multiplying $Q\G$ before attention.
All downstream states and generated operators are recomputed. Thus these
70 sources cover the whole model in a specified intervention family;
they do not span arbitrary changes of all $70\times64^2$ operator entries.

\subsection{Inference response along a realized optimizer update}
\begin{corollary}[Full-parameter training displacement]\label{model:cor:training-response}
Condition on an augmented training state and a realized training batch,
and let $\Delta\theta$ be its actual optimizer-induced parameter change.
For an independently chosen inference context set
$v_x=D_\theta z_\theta(x)[\Delta\theta]$. Along
$\theta(\epsilon)=\theta+\epsilon\Delta\theta$,
\begin{equation}
 \KL(p_\theta(\cdot\mid x)\|p_{\theta(\epsilon)}(\cdot\mid x))
 =\frac{\epsilon^2}{2}v_x^{\mathsf T}F(p_\theta)v_x+O(\epsilon^3).
 \label{model:eq:training-response}
\end{equation}
The coefficient includes every parameter direction present in the update
and is unchanged by exact decoder regrouping.
\end{corollary}
\begin{proof}
With state and batch fixed, $\Delta\theta$ is a fixed vector. The smooth
one-dimensional source $\epsilon\mapsto z_{\theta+\epsilon\Delta\theta}(x)$
has derivative $v_x$. The log-sum-exp Taylor expansion has vanishing
first-order KL term and Hessian $F(p_\theta)$, so its quadratic term is
the displayed categorical variance; bounded local third derivatives give
the remainder. Exact decoder regrouping preserves this source map
pointwise, hence also its derivative and quadratic coefficient.
\end{proof}

This corollary connects the optimizer dynamics to the inference law
without differentiating through the optimizer itself. It predicts a
realized update after that update has been calculated; it does not
predict an unseen training gradient or replace the full parameter-space
Fisher matrix by a measured scalar. Section~\ref{model:sec:training-results}
uses matrix-free directional differentiation to evaluate the coefficient.

\subsection{An invariant deductive sector}
\begin{corollary}[Complete operator freezing]\label{model:cor:freezing}
Suppose that for every context in a specified support, the generated
operator at each layer and head equals a fixed matrix $G^*_{\ell a}$.
Replacing all operator generators by those matrices preserves every
decoder boundary and final prediction on that support. In a centered
complete observation containing the operators, these coordinates are
identically zero, and the joint law factors as a Dirac law on them times
the remaining observation law. Under independent-context blocking its
Gaussian limit has the corresponding zero-variance sector.
\end{corollary}
\begin{proof}
The original and replaced models have the same embedding state. If their
inputs to a layer agree, the query, key, and value projections agree, and
the assumed operator equals its replacement. The attention, output
projection, residual paths, and feed-forward map therefore give the same
next boundary. Induction over the layers proves equality of predictions.
Subtracting the mean of a constant coordinate makes it zero. The claimed
factorization then follows directly from the definition of the joint
law, and Theorem~\ref{model:thm:clt} gives the degenerate Gaussian limit.
\end{proof}

This is the full-graph form of the operator-freezing implication for PLGA
\cite{gokden2026foundations}. Its hypothesis is exact; the experiment
instead measures the error of the replacement on a finite set. Variation
of attention and predictions can remain substantial because queries,
keys, values, and decoder states continue to depend on the context.
Section~\ref{model:sec:controlled-inference-results} gives the frozen-operator
comparison on the fresh released-model evaluation cohort.

\section{Conditional training-mixture transport}
\subsection{Training variability and conditional fluctuation classes}
All fields in this subsection use a common observation coordinate system.
For a training state $s$, let $m_s=\E_\D\Phi_{\theta(s)}$ and
$C_s=\Cov_\D(\Phi_{\theta(s)})$. Define
\begin{equation}
 \overline m_t=\E_{\rho_t}m_s,\qquad
 A_t=\E_{\rho_t}C_s,\qquad
 B_t=\Cov_{\rho_t}(m_s).\label{model:eq:training-sectors}
\end{equation}
$A_t$ is the average within-checkpoint covariance. $B_t$ is the
between-training-state covariance of inference means. Both are positive
semidefinite; their distinction survives even when every checkpoint is
evaluated on independent documents.

\begin{theorem}[Exact training--inference susceptibility decomposition]
\label{model:thm:training-mixture}
Draw one training state $s\sim\rho_t$ and then $b$ conditionally independent
contexts for that same state. Center the fields by $\overline m_t$ and set
$Z_b=b^{-1/2}\sum_{i=1}^b(\Phi_{\theta(s)}(x_i,y_i)-\overline m_t)$.
If the second moments are finite, then
\begin{equation}
 \Cov(\Phi)=A_t+B_t,\qquad
 \Cov(Z_b)=A_t+bB_t.\label{model:eq:training-mixture}
\end{equation}
If instead the training state is independently redrawn for each context,
the covariance of $Z_b$ is $A_t+B_t$ at every $b$.
\end{theorem}
\begin{proof}
Write each centered field as
$(\Phi_{\theta(s)}-m_s)+(m_s-\overline m_t)$. The first term has
conditional mean zero, so its cross covariance with the second vanishes
after conditioning on $s$. This proves the first identity. For the
shared-state block, the sum of the first terms has conditional covariance
$bC_s$, while the sum of the second terms is $b(m_s-\overline m_t)$.
Divide the resulting covariance $bA_t+b^2B_t$ by $b$ to obtain the second
identity. Independent redraws of both state and context remove all
off-diagonal covariances, yielding the last statement.
\end{proof}

More information than covariance is retained in the exact generating
function. If $W_s(j)=\log\E_\D e^{j^{\mathsf T}(\Phi_{\theta(s)}-m_s)}$,
then the shared-state block has
\begin{equation}
 W_{t,b}(j)=\log\E_{\rho_t}\exp\left[
 \sqrt b\,j^{\mathsf T}(m_s-\overline m_t)
       +bW_s(j/\sqrt b)\right].\label{model:eq:mixture-cgf}
\end{equation}
This follows by conditional factorization before averaging over training
states. Averaging the $W_s$ themselves would give a different law.

\begin{proposition}[A conditional universality statement]
\label{model:prop:conditional-universality}
For each fixed finite checkpoint with finite second moments, blocks
centered by $m_s$ obey the multivariate independent-context CLT. For a
finite mixture of checkpoints, their unwhitened limiting law is
$\sum_s\pi_s\mathcal N(\bm0,C_s)$. If the active covariances are nonsingular
on a common retained space, whitening separately by $C_s^{-1/2}$ gives
the common standard Gaussian limiting law. Global centering without
removing a nonzero $B_t$ retains the $bB_t$ sector in
Equation~\eqref{model:eq:training-mixture}.
\end{proposition}
\begin{proof}
Condition on a checkpoint and apply the finite-dimensional CLT to its
centered independent fields. The characteristic functions of a finite
mixture are the corresponding weighted finite sum, so their limits are
the stated Gaussian mixture. Applying $C_s^{-1/2}$ makes each conditional
limiting covariance the identity, and their mixture is then the same
standard Gaussian. The globally centered covariance is exactly the
preceding theorem. Neither this growing sector nor a Gaussian mixture
requires long-range interaction between inference contexts.
\end{proof}

Thus training can change means, covariances, and predictive response
without changing a conditional independent-context universality class.
The training-path universality class is a separate question about the
coarse path law generated by the $K_t$. There is no assertion that its
fixed point is Gaussian. These conclusions connect training and inference
while keeping their distinct sources of fluctuation identifiable.

\section{Training-to-inference drift and response}
\label{model:app:emission-transport}
\subsection{The conditional drift of the inference law}
For a smooth collective observable define
$a(\theta)=\E_{\D_{\rm ev}}f(\Xi_\theta)$. Conditioning on $S=s$ fixes the
distribution of the next AdamW displacement, including moment estimates,
clipping, sampling and weight decay \cite{kingma2015adam,loshchilov2019adamw}.

Taylor expansion of a smooth observable supplies the local drift; its use for predictive geometry follows the finite-parameter setting of \cite{martens2020}.
\begin{proposition}[Training-to-inference drift]\label{model:prop:inference-drift}
Suppose $a$ is $C^3$ on each relevant displacement segment, with third
derivative operator norm bounded by $M_3$. If
$\E[\|\Delta\theta\|^3\mid s]<\infty$, then
\begin{align}
 \E[a(\theta+\Delta\theta)-a(\theta)\mid s]
 &=Da(\theta)\E[\Delta\theta\mid s]
 +\tfrac12\E[D^2a(\theta)[\Delta\theta,\Delta\theta]\mid s]+R_s,
 \label{model:eq:inference-drift}\\
 \|R_s\|&\le\tfrac16M_3\E[\|\Delta\theta\|^3\mid s].\nonumber
\end{align}
\end{proposition}
\begin{proof}
Taylor's formula along each segment has first two terms as displayed and
integral remainder bounded by $M_3\|\Delta\theta\|^3/6$. Take conditional
expectations and use the triangle inequality. Under a fixed finite token
law, differentiating the expectation defining $a$ is differentiation of a
finite weighted sum. For other laws, interchange of differentiation and
expectation is an additional hypothesis.
\end{proof}

This is a local prediction of a distribution of unseen updates when its
coefficients are estimated from other minibatches. It is not an autonomous
beta function of $a$ alone. The time resolution $m$ of training-kernel
blocking, optimizer time $t$, document block size $b$, decoder grouping,
and architecture size $N$ remain separate scales.

\subsection{One perturbation identity for coupled observations}
\begin{lemma}[Signed covariance perturbation]\label{model:lem:master-perturbation}
Under one specified conditioning law, let $X\in\R^d$ and $e\in\R^k$ have finite second moments, and let $A\in\R^{k\times d}$ and $b\in\R^k$ be deterministic. Set $Y=b+AX+e$, $\Sigma_X=\Cov X$, $\Sigma_e=\Cov e$, $C_{Xe}=\Cov(X,e)$, and $a^2=\tr\Sigma_e$. Then
\begin{align}
 \Cov Y-A\Sigma_XA^{\mathsf T}
 &=AC_{Xe}+C_{Xe}^{\mathsf T}A^{\mathsf T}+\Sigma_e,\label{model:eq:master-perturbation}\\
 \|\Cov Y-A\Sigma_XA^{\mathsf T}\|_{\rm op}
 &\le 2\|A\|_{\rm op}\sqrt{\tr\Sigma_X}\,a+a^2.\label{model:eq:master-perturbation-bound}
\end{align}
For scalar $X,e$, with $A=1$, $N>0$ and $\chi_N=N\Var X>0$,
\begin{align}
 N\Var(X+e)-\chi_N&=N\{2\Cov(X,e)+\Var e\},\label{model:eq:master-signed}\\
 |N\Var(X+e)-\chi_N|&\le2\sqrt{N\chi_N}\,a+Na^2.\label{model:eq:master-susceptibility}
\end{align}
If $\chi_N\sim C N^\kappa$ with $C>0$, then $a=o(N^{(\kappa-1)/2})$ suffices to preserve that asymptotic. For a specified coupling, preserving the same leading asymptotic is equivalent to
\[
 2\Cov(X,e)+\Var e=o(\Var X).
\]
\end{lemma}
\begin{proof}
Center $X$ and $e$ separately and expand the outer product of $A(X-\E X)+(e-\E e)$. The constant $b$ disappears, giving Equation~\eqref{model:eq:master-perturbation}. For unit vectors $u,v$, Cauchy--Schwarz bounds the corresponding bilinear form of $AC_{Xe}$ by $\|A\|_{\rm op}(\tr\Sigma_X)^{1/2}(\tr\Sigma_e)^{1/2}$. Each ordered mixed term satisfies this bound, and $\|\Sigma_e\|_{\rm op}\le\tr\Sigma_e=a^2$. The triangle inequality proves Equation~\eqref{model:eq:master-perturbation-bound}.

In the scalar case, multiply the exact variance expansion by $N$ and bound $|\Cov(X,e)|$ by $\sqrt{\Var X}\,a$. Dividing the bound by $\chi_N$ gives $2r_N+r_N^2$, where $r_N=a/\sqrt{\Var X}$. Under the stated small-error assumption this tends to zero. Dividing the exact signed identity by $\chi_N$ proves both directions of the final equivalence.
\end{proof}

Centering can only decrease the $L^2$ norm, so a bound on $\|e\|_{L^2}$ is a sufficient substitute for $a$. It can be conservative: a deterministic displacement does not change covariance, and $e=-2X$ preserves scalar variance through signed cancellation. The latter has a nonvanishing relative centered error. Preserving an exponent while allowing a different leading constant is weaker than the equivalence above.

\begin{corollary}[Empirical normalization]\label{model:cor:master-empirical}
For $N>0$ and paired vectors $q_i,\widehat q_i=q_i+\Delta_i$ in a real Hilbert space, $1\le i\le s$ with $s>1$, define
\[
 \chi=\frac{N}{s-1}\sum_i\|q_i-\bar q\|^2,\quad
 \widehat\chi=\frac{N}{s-1}\sum_i\|\widehat q_i-\overline{\widehat q}\|^2,\quad
 \chi_\Delta=\frac{N}{s-1}\sum_i\|\Delta_i-\bar\Delta\|^2.
\]
Then
\[
 |\widehat\chi-\chi|\le2\sqrt{\chi\chi_\Delta}+\chi_\Delta,
 \qquad
 \chi_\Delta\le\frac{Ns}{s-1}\frac1s\sum_i\|\Delta_i\|^2.
\]
\end{corollary}
\begin{proof}
In the product Hilbert space set $x_i=\sqrt{N/(s-1)}(q_i-\bar q)$ and $d_i=\sqrt{N/(s-1)}(\Delta_i-\bar\Delta)$. Expanding $\|x+d\|^2-\|x\|^2$ and applying Cauchy--Schwarz gives the first inequality. The identity $\sum_i\|\Delta_i-\bar\Delta\|^2=\sum_i\|\Delta_i\|^2-s\|\bar\Delta\|^2$ proves the second. No independence is needed for these deterministic inequalities.
\end{proof}

The same expansion applies to a parameter-update covariance, a coupled row reduction, accumulated prediction error or a linearized inference emission. Differentiability, time propagation, distributional convergence and nonzero inference overlap remain additional hypotheses in their respective applications. A common covariance proof does not supply them.

There is an exact covariance version of the drift, which specifies how
training selects the fluctuation coefficients of inference.
\begin{proposition}[Covariance transport under a parameter update]
\label{model:prop:covariance-drift}
Let $\Phi=\Phi_\theta(X)$, $\delta\Phi=\Phi_{\theta+\Delta}(X)-\Phi_\theta(X)$
for the same $X\sim\D_{\rm ev}$, and assume finite second moments. Then
\begin{equation}
 \Sigma_{\theta+\Delta}-\Sigma_\theta
 =\Cov(\Phi,\delta\Phi)+\Cov(\delta\Phi,\Phi)+\Cov(\delta\Phi).
 \label{model:eq:covariance-drift}
\end{equation}
If $\theta\mapsto\Phi_\theta$ is differentiable in mean square, its first variation is
$D\Sigma_\theta[\Delta]=\Cov(\Phi,D\Phi_\theta[\Delta])+
\Cov(D\Phi_\theta[\Delta],\Phi)$.
\end{proposition}
\begin{proof}
Lemma~\ref{model:lem:master-perturbation}, with $A=I$, $b=\bm0$,
$X=\Phi$ and $e=\delta\Phi$, gives the identity under the fixed
observation law. For a scalar step $h$, mean-square differentiability gives
$\Phi_{\theta+h\Delta}-\Phi_\theta=hD\Phi_\theta[\Delta]+o(h)$ in $L^2$.
Cauchy--Schwarz makes covariance continuous in its $L^2$ arguments;
thus the two mixed terms divided by $h$ converge to the stated
first variation and the error covariance divided by $h$ tends to zero.
\end{proof}

The observable can be an operator entry, attention field, predictive
entropy, or a product needed for a covariance coefficient. Thus the same
drift construction transports all selected means, covariances and response
fields. No sign of the covariance drift is forced by the architecture.
Training contraction of a generated operator and growth of an active
predictive covariance are compatible outcomes of different components.

\subsection{PLGA attenuation and the error of an effective operator}
For the positive base $u$ and exponent $p$ in the native generator,
\[
 d(u^p)=u^p(p\,du/u+\log(u)\,dp),\qquad
 G=a\,u^{\odot p}+b^a.
\]
These derivatives, the Gram derivative and the decoder source transport
specify the PLGA contribution to Equation~\eqref{model:eq:inference-drift}.
The offset ensures smoothness but does not bound the derivatives by a
small constant. An apparent tensor plateau therefore requires an input
dispersion measurement and a transported output error.

\begin{proposition}[Operator dispersion and predictive approximation]
\label{model:prop:operator-approximation}
If a generator $g$ between finite-dimensional Hilbert spaces is
$L_g$-Lipschitz on the support of a square-integrable random input $D$, then
\begin{equation}
 \E\|g(D)-\E g(D)\|^2\le L_g^2\E\|D-\E D\|^2.
 \label{model:eq:operator-dispersion}
\end{equation}
Consider a decoder chain with an operator argument $G_\ell$ and a
replacement $\overline G_\ell$. On all fine, replaced and intervening
states, assume its boundary map is $L_\ell$-Lipschitz in the boundary
input at fixed operator and $a_\ell$-Lipschitz in the operator. Let
$d_\ell(x)=\|G_\ell(X_{\ell-1}(x))-\overline G_\ell\|$ at the fine state,
and let the final logit map be $L_z$-Lipschitz. Then
\begin{equation}
 \|z(x)-\overline z(x)\|\le
 L_z\sum_{\ell=1}^L a_\ell d_\ell(x)
       \prod_{k=\ell+1}^L L_k.\label{model:eq:operator-output}
\end{equation}
With Euclidean logit norm, the predictive KL is at most one half the
square of the right side.
\end{proposition}
\begin{proof}
For an independent copy $D'$, Hilbert-space variance is
$\E\|g(D)-g(D')\|^2/2$. Apply the Lipschitz inequality and the same
identity to $D$, proving the first claim. At each decoder add and subtract
the output at the fine boundary and replacement operator. The discrepancy
$e_\ell$ obeys $e_\ell\le a_\ell d_\ell+L_\ell e_{\ell-1}$, with $e_0=0$.
Iterate this inequality and apply the final logit Lipschitz bound. Finally,
the Hessian of log-sum-exp is categorical covariance $F(p)$, and
$v^{\mathsf T}F(p)v\le\sum_i p_i v_i^2\le\|v\|_2^2$. The integral
second-order remainder for its Bregman divergence is therefore at most
$\|z-\overline z\|_2^2/2$; this divergence is the specified KL.
\end{proof}

The factor $1/2$ is a conservative Euclidean KL bound. The categorical
Hessian argument in Proposition~\ref{prop:row-predictive-bridge} gives
\[
 \KL\!\left(\softmax z\,\Vert\,\softmax\overline z\right)
 \le\tfrac14\|z-\overline z\|_2^2.
\]
Thus the same proved logit-error budget in
Equation~\eqref{model:eq:operator-output} also gives a factor $1/4$.
Only the categorical step is being reused: the layer discrepancies here
are evaluated at the fine states, whereas the bridge uses its specified
hybrid inputs. All fine, replaced and intervening states must still lie
in the assumed Lipschitz domains.

The Lipschitz constants are hypotheses, not estimates inferred from a
small norm coefficient of variation. The completed direct matrix and
predictive freezing measurements assess the approximation at its two
ends. The paired training branches change generator parameters or
optimizer moments; freezing generator parameters still allows the
operator to depend on its input.
The frozen-inference comparisons are in Sections~\ref{model:sec:pretrained-freezing-results} and \ref{model:sec:inference-results}; Section~\ref{model:sec:optimizer-results} details the paired moment perturbations.

\subsection{A curvature-dependent response window}
For a context and specified displacement let
$k_x(e)=\KL(p_\theta(\cdot\mid x)\|p_{\theta+e\Delta}(\cdot\mid x))$,
$v_x=z_x'(0)$ and $q_x=v_x^{\mathsf T}F(p_x)v_x$.
\begin{proposition}[Local response, numerical resolution and cubic term]
\label{model:prop:response-window}
Assume $z_x$ is $C^3$, $q_x>0$, and $|k_x'''(e)|\le M_x$ on the
displacement interval. If $|\widehat k_x-k_x|\le d_x$ and
$\widehat q_x>0$, then for $e\ne0$
\begin{align}
 \left|\frac{k_x(e)}{e^2q_x/2}-1\right|&\le\frac{M_x|e|}{3q_x},
 \label{model:eq:local-window}\\
 \left|\frac{\widehat k_x(e)}{e^2\widehat q_x/2}-1\right|
 &\le\left|\frac{q_x}{\widehat q_x}-1\right|
 +\frac{M_x|e|}{3\widehat q_x}+\frac{2d_x}{e^2\widehat q_x}.
 \label{model:eq:numerical-window}
\end{align}
At $e=0$, or when a curvature denominator is unresolved, retain the
absolute bound $|k_x(e)-e^2q_x/2|\le M_x|e|^3/6$; the relative
quotients are used only with their stated positive denominators.
Writing $w_x=z_x''(0)$, the cubic derivative is
\begin{equation}
 k_x'''(0)=\E_{p_x}[(v_x-\E_{p_x}v_x)^3]
                 +3\Cov_{p_x}(v_x,w_x).\label{model:eq:response-cubic}
\end{equation}
\end{proposition}
\begin{proof}
Write $k_x(e)=A(z_x(e))-p_x^{\mathsf T}z_x(e)+\mathrm{constant}$, with
$A(z)=\log\sum_i e^{z_i}$. At zero, $k_x(0)=k_x'(0)=0$ and
$k_x''(0)=q_x$. Taylor's formula bounds the remainder by $M_x|e|^3/6$;
division gives the first inequality. Add and subtract $k_x(e)$ and
$e^2q_x/2$, and divide by $e^2\widehat q_x/2$, for the second. In the
third composite derivative the coefficient of $z_x'''(0)$ cancels since
$\nabla A(z_x(0))=p_x$. The remaining terms are
$D^3A[v_x,v_x,v_x]+3v_x^{\mathsf T}F(p_x)w_x$. Derivatives of the finite
categorical generating function identify them with the displayed central
third moment and covariance.
\end{proof}

A tested amplitude grid is not a certified continuous Taylor radius.
Numerical error grows relative to a quadratic signal as $e$ becomes too
small; nonlinear error can grow as $e$ increases. For $q_x=0$ use absolute
errors rather than dividing by a vanishing quadratic prediction. The
predictive Fisher matrix is formed with model probabilities over the
vocabulary; it is distinct from an empirical outer product of training
loss gradients \cite{amari1998,martens2020,kunstner2019}.

For a direction $v$, let $\mu_v=\E_\D q_v(X)>0$ and assume finite variance.
An independent calibration mean from $n$ contexts satisfies
\begin{equation}
 \E\left[\left(\frac{\widehat\mu_v-\mu_v}{\mu_v}\right)^2\right]
 =\frac{\Var(q_v)}{n\mu_v^2}.\label{model:eq:response-sampling}
\end{equation}
Indeed it is unbiased and has variance $\Var(q_v)/n$. Independent
calibration and test means have difference variance
$\Var(q_v)(n_{\rm cal}^{-1}+n_{\rm test}^{-1})$. These are mean-estimation
identities, not pointwise predictions or unbiased-ratio claims. Context
heterogeneity remains part of the response field even when its mean
coefficient is well defined.

\subsection{Three fluctuation sectors and contrasts}
Use common coordinates and remove deterministic position means. A
conditional representation of a segment field is
\begin{equation}
 Y_i=m_S+a_S(U)+\epsilon_i,\quad
 \E[a_S(U)\mid S]=\bm0,\quad \E[\epsilon_i\mid S,U]=\bm0.
 \label{model:eq:three-sector-model}
\end{equation}
Here $S$ is a training state and $U$ is a document variable. Set
$B_{\rm tr}=\Cov(m_S)$,
$B_{\rm doc}=\E\Cov(a_S(U)\mid S)$, and, when the averaged residual is
second-order stationary,
$A_\epsilon=\E\Cov(\epsilon_i\mid S,U)$ and
$\Gamma(r)=\E\Cov(\epsilon_i,\epsilon_{i+r}\mid S,U)$.

The law of total covariance and the covariance-of-a-sum identity \cite{kallenberg2021} give the following specialization.
\begin{proposition}[Hierarchical susceptibility]\label{model:prop:three-sectors}
Under these assumptions and finite second moments,
\begin{equation}
 \chi_b=bB_{\rm tr}+bB_{\rm doc}+A_\epsilon+
 \sum_{r=1}^{b-1}(1-r/b)[\Gamma(r)+\Gamma(r)^{\mathsf T}].
 \label{model:eq:three-sectors}
\end{equation}
Every contrast $\sum_i d_iY_i$ with $\sum_i d_i=0$ cancels both shared
components exactly. Without residual stationarity the residual term is
$b^{-1}\E\sum_{i,j}\Cov(\epsilon_i,\epsilon_j\mid S,U)$.
\end{proposition}
\begin{proof}
Conditional zero means remove the cross covariances between the residual
sum and each shared component. The conditional centering of $a_S$ also
removes its covariance with $m_S$. The shared components contribute
$b^2B_{\rm tr}$ and $b^2B_{\rm doc}$ to the covariance of the sum.
Divide by $b$ and count $b-r$ ordered pairs at lag $r$ to obtain the
formula. In a contrast each shared component is multiplied by
$\sum_i d_i=0$, proving the cancellation.
\end{proof}

For fixed weights, $B_{\rm tr}=\bm0_{\mathrm{mat}}$. The document variable is not uniquely
identified by a finite covariance array. The residual $\Gamma$ remains
in the retained theory; subtracting the same finite document mean can
create negative sample correlations and does not estimate the latent
residual directly. Disjoint block contrasts remove a common component
without that interpretation.

This is a mean-square averaging criterion \cite{kallenberg2021}.
\begin{proposition}[Shared-component scaling class]\label{model:prop:shared-class}
Let $Y_i=M+\epsilon_i$, with $M$ centered and square integrable,
$\E[\epsilon_i\mid M]=\bm0$, and
$\E\|b^{-1}\sum_{i=1}^b\epsilon_i\|^2\to0$. Then
$b^{-1}\sum_iY_i\to M$ in mean square. The limiting law need not be
Gaussian. A stationary residual with absolutely summable matrix
covariances satisfies the stated mean-square condition.
\end{proposition}
\begin{proof}
The difference between the normalized sum and $M$ is exactly the
residual average. This proves the first assertion. Under covariance
summability its squared mean norm is bounded by a constant divided by
$b$, using the covariance-of-a-sum formula. Arbitrary square-integrable
choices of $M$ show that its law need not be Gaussian.
\end{proof}

The independent finite-variance class has $H=1/2$ after conditioning on
the shared state. A surviving common component instead has $H=1$ under
global centering. Summable residual covariance gives bounded residual
susceptibility; a residual Gaussian limit requires additional mixing or
other central-limit hypotheses. The nonsummable tail class of
Proposition~\ref{model:prop:longrange} has $H=1-\gamma/2$ in its nonzero sector.
These are distinct classes under declared probability blockings, not
interchangeable critical exponents of a width limit.

\subsection{A predictive coordinate for regime selection}
An entropy plateau alone need not imply a constant vocabulary prediction.
A coordinate that uses the whole emission is
\[
 I_{\rm ctx}(\theta)=\E_{X,X'\,\mathrm{iid}\sim\D_X}
 \KL(p_\theta(\cdot\mid X)\|p_\theta(\cdot\mid X')).
\]
It is defined at fixed weights and can be transported by the same training
drift construction, using pairs of contexts as the observation law.

\begin{proposition}[Predictive context dispersion]\label{model:prop:context-dispersion}
For finite vocabulary probabilities with $p_i(X)\ge\epsilon>0$ almost
surely, let $\Sigma_p=\Cov(p(X))$. Then
\begin{equation}
 2\tr\Sigma_p\le I_{\rm ctx}\le\epsilon^{-1}\tr\Sigma_p.
 \label{model:eq:context-dispersion}
\end{equation}
In particular $I_{\rm ctx}=0$ if and only if the predictive probability
vector is almost surely constant. This imposes no constancy condition on
an intermediate operator or on every internal representation.
\end{proposition}
\begin{proof}
For negative categorical entropy $h(p)=\sum_i p_i\log p_i$,
$D^2h(p)=\diag(1/p_i)$. On the tangent space $\sum_i v_i=0$,
Cauchy--Schwarz gives
\[
 \sum_i v_i^2/p_i\ge\Bigl(\sum_i|v_i|\Bigr)^2
                         \ge2\sum_i v_i^2.
\]
For the second inequality, the positive and negative parts of $v$
have equal sum $a$, and each has sum of squares at most $a^2$.
The upper quadratic-form bound is $\epsilon^{-1}\sum_i v_i^2$.
Integrating the second-order Taylor remainder of $h$ along the
segment between $p,q$ therefore gives
\[
 \|p-q\|_2^2\le\KL(p\|q)\le\|p-q\|_2^2/(2\epsilon).
\]
Independent copies satisfy
$\E\|p(X)-p(X')\|_2^2=2\tr\Sigma_p$; average the bounds.
Zero covariance trace is equivalent to almost sure constancy.
For two equiprobable binary vectors $(1/2+a,1/2-a)$ and
$(1/2-a,1/2+a)$, the ratio $I_{\rm ctx}/\tr\Sigma_p$ tends to two
as $a\to0$, establishing sharpness of the lower constant.
\end{proof}

Every fixed finite call has a positive minimum probability, though there
need not be a useful uniform lower bound across widths or training times.
The context-permutation measurements in
Section~\ref{model:sec:controlled-training-methods} are finite cohort estimates of
predictive context dispersion. They use all vocabulary probabilities and
therefore distinguish a nearly context-insensitive output regime from an
active predictive regime even when both have small operator dispersion.

\section{Chronological covariance and conditional prediction}
\label{model:app:chronological-details}
\subsection{A common chronological covariance identity}

Fix a conditioning sigma-field $\mathcal G$ specifying the declared external information. A complete incoming state and the data law are one useful choice. Write $\Cov(X,Y\mid\mathcal G)$ for the centered conditional cross covariance. All random vectors and transported summands below have finite second moments.

\begin{proposition}[Conditional chronological covariance]\label{model:prop:master-chronological}\label{model:prop:law-innovation-transport}
For $0\le j<t$, let
\[
 q_{j+1}=J_jq_j+b_j+\eta_j,
\]
where $J_j,b_j$ are deterministic conditional on $\mathcal G$. Define
\[
 \Phi_{t,j}=J_{t-1}\cdots J_j,\qquad \Phi_{t,t}=I,
 \quad U=(q_0^{\mathsf T},\eta_0^{\mathsf T},\ldots,\eta_{t-1}^{\mathsf T})^{\mathsf T},
\]
\[
 L_t=[\Phi_{t,0}\ \Phi_{t,1}\ \cdots\ \Phi_{t,t}],
 \qquad d_t=\sum_{j<t}\Phi_{t,j+1}b_j.
\]
Then $q_t=L_tU+d_t$. With $\mu_U=\E[U\mid\mathcal G]$ and $C_U=\Cov(U\mid\mathcal G)$,
\begin{align}
 \E[q_t\mid\mathcal G]&=L_t\mu_U+d_t,\label{model:eq:master-mean}\\
 \Cov(q_t\mid\mathcal G)&=L_tC_UL_t^{\mathsf T},\label{model:eq:master-cov}\\
 \Cov(q_t)&=\E[L_tC_UL_t^{\mathsf T}]+\Cov(L_t\mu_U+d_t).\label{model:eq:master-outer}
\end{align}
In particular, letting $C_0=\Cov(q_0\mid\mathcal G)$, $H_j=\Cov(q_0,\eta_j\mid\mathcal G)$ and $Q_{ij}=\Cov(\eta_i,\eta_j\mid\mathcal G)$, the conditional covariance is
\begin{align}
 \Phi_{t,0}C_0\Phi_{t,0}^{\mathsf T}
 &+\sum_{j<t}\left(\Phi_{t,0}H_j\Phi_{t,j+1}^{\mathsf T}
     +\Phi_{t,j+1}H_j^{\mathsf T}\Phi_{t,0}^{\mathsf T}\right)\nonumber\\
 &+\sum_{i,j<t}\Phi_{t,i+1}Q_{ij}\Phi_{t,j+1}^{\mathsf T}.\label{model:eq:law-innovation-covariance}
\end{align}
Consecutive temporal regrouping preserves these identities when the full transported forcing and its joint law are retained.
\end{proposition}
\begin{proof}
Induction in the recurrence gives the displayed expression for $q_t$. Since $L_t,d_t$ are $\mathcal G$-measurable, conditional expectation gives Equation~\eqref{model:eq:master-mean}. Subtracting that expectation yields $L_t(U-\mu_U)$. Its conditional outer product gives Equation~\eqref{model:eq:master-cov}. Partitioning $C_U$ into its initial and forcing blocks gives Equation~\eqref{model:eq:law-innovation-covariance}, including both orientations of each cross term.

For completeness, write $m_t=\E[q_t\mid\mathcal G]$. The decomposition
$q_t-\E q_t=(q_t-m_t)+(m_t-\E q_t)$ has zero expected mixed outer products because $\E[q_t-m_t\mid\mathcal G]=\bm0$. Taking expectations of its outer product proves Equation~\eqref{model:eq:master-outer}. Associativity of ordered matrix products and distributivity over finite sums show that composing consecutive intervals gives the same $L_tU+d_t$. Retaining the joint initial/forcing law therefore preserves its covariance under regrouping.
\end{proof}

For deterministic $J_j,b_j$, use a fixed conditioning law. Zero offsets give the nonstationary linear-response formula. A deterministic incoming coordinate removes initial covariance and initial/forcing cross terms; it does not remove correlations between distinct forcing intervals. For a finite-horizon Markov representation, sample $q_0$ and the entire
forcing sequence jointly under the conditioning law. The enlarged state
retains $q_j$, the remaining sequence and $j$; its next transition applies
the recurrence and deletes the used entry. This is a deterministic
measurable update, hence Markov. It does not show that $q_j$ alone is
Markov or that the retained forcing law is economical to predict.
Replacing the conditional joint law by the product of its initial and
individual forcing marginals deletes $H_j$ and the terms $Q_{ij}$ with
$i\ne j$; it changes the model unless those covariances already vanish.

Native Jacobians can be random and path dependent. Conditioning on their entire history permits conditional factorization only with the corresponding conditional moments and the outer covariance in Equation~\eqref{model:eq:master-outer}. Such conditioning can include future information. A predictive construction must specify what is available at the incoming time. A deterministic fitted response instead leaves a residual law to be estimated, and a native tangent approximation must retain its nonlinear remainder.

For example, take $J=X$ with $X\in\{1,2\}$ equiprobably.
Then $\Var(JX)=9/4$, whereas $\E[J^2]\Var(X)=5/8$.
Conditioning on $J$ makes the conditional variance zero, leaving
$9/4$ in the covariance of the conditional mean. The outer term is
therefore necessary even in a two-point law.

\subsection{Covariance blocking and finite predictive memory}
A second-order effective state can retain the joint fluctuations of a
finite set of training-time emissions. This state is sufficient for
linear covariance transport and optimal affine prediction; it need not
be sufficient for the nonlinear successor law. The distinction permits
finite tests without an unmeasured white-noise assumption.

\begin{proposition}[Covariance RG and affine memory]
\label{model:prop:affine-memory}
Let $Z=(X,Y)$ be a centered finite-second-moment random vector,
conditionally on a fixed complete incoming state and data law. Write
its covariance as
\[
 C=\begin{pmatrix}C_{XX}&C_{XY}\\C_{YX}&C_{YY}\end{pmatrix}.
\]
For deterministic linear block maps $B,D$, covariance blocking is
$\mathcal C_B(C)=BCB^{\mathsf T}$ and satisfies
$\mathcal C_D\circ\mathcal C_B=\mathcal C_{DB}$. If $C_{XX}$ is
positive definite, put $K=C_{YX}C_{XX}^{-1}$ and
$S=C_{YY}-C_{YX}C_{XX}^{-1}C_{XY}$. Then
\begin{equation}
 \Cov(Y-KX,X)=\bm0_{\mathrm{mat}},\qquad
 \E\norm{Y-HX}^2=\tr S+
 \tr\bigl((H-K)C_{XX}(H-K)^{\mathsf T}\bigr)
 \label{model:eq:affine-memory}
\end{equation}
for every matrix $H$ of matching size. Thus $KX$ is the unique
best affine predictor of the centered $Y$ from $X$; $S$ is its
residual covariance. With nonzero means the predictor becomes
$\E Y+K(X-\E X)$.
For nested linear observation spans, the optimum mean-square error
cannot increase when more observations are retained. If the joint
law is Gaussian, this affine predictor is also the conditional mean
and its residual is independent of $X$.
\end{proposition}
\begin{proof}
Centering commutes with deterministic linear maps, so expanding the
outer product gives $\Cov(BZ)=BCB^{\mathsf T}$. Associativity of
matrix multiplication proves composition. Direct expansion gives
$\Cov(Y-KX,X)=C_{YX}-KC_{XX}=\bm0_{\mathrm{mat}}$ and
$\Cov(Y-KX)=S$. Write
$Y-HX=(Y-KX)+(K-H)X$. Orthogonality removes the cross terms in
its squared norm, proving Equation~\eqref{model:eq:affine-memory}.
Positive definiteness gives uniqueness. An enlarged observation
span contains every predictor available in the smaller span, so its
minimum error is no larger. For a Gaussian vector the residual and
$X$ are jointly Gaussian with zero cross covariance, hence independent:
their joint characteristic function factors because its quadratic
form has no mixed term. Its centered conditional expectation is
therefore zero, giving the conditional mean formula.
\end{proof}

A pseudoinverse extends the projection to singular $C_{XX}$ on its
range, with uniqueness only of the predicted random variable.
The positive-definite statement suffices here; fitted predictors use
an explicitly stated ridge penalty. For any fitted center and matrix,
its error contains estimation and regularization errors in addition
to $\tr S$. An oracle inequality for nested spans therefore does not
promise improvement by every fitted history predictor. Orthogonality
to a finite linear span is also weaker than conditional mean zero
relative to the complete native filtration. Neither implies that
successive native risk increments are martingale innovations.

For the risk path $(R_1,R_4,R_{16},R_{64})$, differencing and
cumulative summation are inverse triangular linear maps at a fixed
incoming risk. Keeping the full increment covariance therefore
preserves every covariance of these four emissions under aligned
temporal blocking. Dropping its off-diagonal terms defines a distinct
approximation with a measurable error. The tests in
Section~\ref{model:sec:memory-results} freeze this covariance and the
finite-memory predictors before fresh continuation paths are drawn.

\subsection{Finite prediction sets for a conditional path law}
Fix a complete incoming state $s$, the unrevealed remaining-data law,
and the evaluation cohort. Let $Y=(R_{h_1},\ldots,R_{h_J})$ be its
future risk vector at specified horizons. A fitted joint path law can
be represented as the innovation-history state above. Its prediction
set can also have a finite-sample guarantee that does not require a
Gaussian or autonomous-coordinate approximation. The rank argument
is the standard split-conformal construction \cite{shafer2008}.

\begin{proposition}[A simultaneous risk-path prediction tube]
\label{model:prop:conditional-risk-tube}
Condition on $s$ and on a development sample independent of the
calibration and future paths. Let $a\in\R^J$ and $b_j>0$ be fixed
functions of that development sample. Assume that $n$ fresh
calibration vectors $Y_1,\ldots,Y_n$ and one future vector $Y_*$
are exchangeable under this conditional law. Define
\[
 T(y)=\max_{1\le j\le J}\frac{|y_j-a_j|}{b_j},\qquad
 t_* = \max_{1\le i\le n}T(Y_i),\qquad
 \mathcal P=\prod_{j=1}^J[a_j-t_*b_j,a_j+t_*b_j].
\]
Then
\begin{equation}
 \Pr(Y_*\in\mathcal P\mid s,\text{development})\ge\frac{n}{n+1}.
 \label{model:eq:conditional-risk-tube}
\end{equation}
The probability averages over calibration and future paths. It does
not assert that coverage conditional on a realized calibration set
is at least the displayed value.
\end{proposition}
\begin{proof}
The $n+1$ scores are exchangeable under the stated conditioning.
For any score tuple, at most one coordinate is strictly larger than
all the others. The sum of the indicators of these $n+1$ events is
therefore at most one. Their expectations are equal by exchangeability,
so the probability that the future score strictly exceeds every
calibration score is at most $1/(n+1)$. Its complementary event is
exactly membership in $\mathcal P$. Ties are included in the prediction
set and can only increase coverage. The maximum in $T$ makes this a
joint assertion about all $J$ coordinates of one future path.
\end{proof}

The unit is an entire reset branch. Nested horizons within it need
not be independent, and multiple validation branches do not inherit
a simultaneous guarantee from Equation~\eqref{model:eq:conditional-risk-tube}.
The incoming model, optimizer and remaining corpus define the
conditioning stratum. A fitted coefficient shared between different
strata requires additional evidence. This construction supplies a
finite predictive statement under a consuming law; it makes no
stationary or critical limiting assumption.
Section~\ref{model:sec:law-closure-results} gives the executed development, calibration and validation branches and their joint path coverage.

\FloatBarrier
\par\medskip\noindent
Chapter~\ref{ch:shared-collectives} organizes these transported effects
into shared and headwise collective variables. Its sign symmetry and
selection analysis identify which fluctuations can survive the chosen
conditioning and observation.

\chapter{Shared collectives, sign symmetry, and training selection}
\label{ch:shared-collectives}
This chapter studies the collective variables through which single-pass
training selects an inference law. It separates signed and invariant
sectors, develops the native head-sign symmetry, and identifies shared-path
fluctuations, adaptive forcing and finite control windows.

\section{Collective variables linking training and prediction}
\label{model:sec:collective-summary}

The retained state must distinguish the common learned operator from
its transverse row fluctuations and from the network that uses it.
For rows $x_i$ of one metric matrix, define
$\mu=d^{-1}\sum_i x_i$ and
$C^{\rm row}=d^{-1}\sum_i(x_i-\mu)(x_i-\mu)^{\mathsf T}$.
Under a native shared row map $F$, its exact moment transport is
\[
 \mu'=F(\mu)+b,\qquad
 C^{\rm row\prime}=JC^{\rm row}J^{\mathsf T}+B+B^{\mathsf T}+C_e.
\]
Here $J=DF(\mu)$, $b$ is the mean Taylor residual, $B$ the
linear--residual cross covariance, and $C_e$ the centered residual
covariance. Proposition~\ref{model:prop:metric-row-transport} proves the
identity and bounds all residual terms. A local contraction of $J$
can suppress row contrast while the common coordinate $\mu$ changes
under training. The downstream PLGA map additionally retains its
constant-input defect (Section~\ref{model:sec:row-correspondence}).

The corresponding training source is the complete AdamW update,
including both moments, bias correction and the derivative of global
clipping. Its ordered tangent products and the full-graph emission
Jacobian transmit common and transverse parameter perturbations to
vocabulary logits (Proposition~\ref{model:prop:adam-tangent}). A small
fixed-weight input derivative through the metric generator therefore
does not imply a small parameter-learning derivative. A generator
increment and a body increment also have a finite interaction:
\[
 \Delta R_{ab}=\Delta R_{\rm full}-\Delta R_a-\Delta R_b.
\]
Each corner uses the same coupled optimizer displacement. The exact
finite source identities and their covariance are proved in
Proposition~\ref{model:prop:finite-source-corners}; no sign of the complete
risk change follows from the isolated corners.

A useful hierarchy consequently retains common metric coordinates,
positive-base and exponent paths with their signed component area,
row covariance and selected exceptional rows, predictive source
coordinates, optimizer information, drive and remaining-data state.
Nested row projections have quantified complete-graph emission
errors (Section~\ref{model:sec:row-projection}). Their use in a training
recurrence still requires the successor-law control of
Proposition~\ref{model:prop:law-closure}. When those finite coordinates
are inadequate, conditional memory supplies an exact larger state.
Section~\ref{model:sec:conditional-memory-law} gives the retained-memory kernel, and Section~\ref{model:sec:memory-results} tests finite history forecasts.

Source selection supplies an additional finite probe of these collectives.
Section~\ref{model:sec:finetuning-theory} distinguishes paired operator
invariance, predictive source rank and coupled dynamical modes, and
gives a simultaneous fluctuation-visibility bound. Its native measurements
and common-source return tests are in
Section~\ref{model:sec:finetuning-results}.

The limiting classes depend on which connected sector survives this
transport. Independent finite-variance document blocks have Gaussian
fluctuations at normalization $b^{-1/2}$; a persistent shared field
survives at normalization $b^{-1}$. For conditional covariance
$C(r)\sim Mr^{-\gamma}$ with $0<\gamma<1$, the second-moment
normalization is $b^{-(1-\gamma/2)}$. Full proofs and the Gaussian
marginal action are in Section~\ref{model:sec:flow}. Regular trained-head
and singular critical limits need their separately specified
cumulant, resource, drive and horizon hypotheses. The classification in Section~\ref{model:sec:theory-synthesis} collects those
obligations alongside the scope of the completed measurements.

\section{Signed and invariant sectors of the single-pass flow}
\label{model:sec:singlepass-critical-observations}

A critical limit concerns the law selected by training as well as the
observation of that law. Fix the pretraining corpus, its realized block
order, the initial shared generator and the evaluation context. The
remaining initialization is the random variable $\xi_N$. At update $T$,
write $\mu_{N,T,g}$ for the conditional law of the complete state when
the generator learning rate is $\eta_0g$. The other parameter rate is
$2\eta_0/N$ in the controlled family. Thus $g$ is a multiplier of the
reference generator rate; the ratio of generator to other rate is
$gN/2$. The two accumulated rates are $\eta_0gT$ and $2\eta_0T/N$.
The remaining corpus and the consumed fraction accompany these clocks.
An inference observation is a pushforward of $\mu_{N,T,g}$ at the
selected weights, with no additional training update.

\begin{table}[htbp]\centering\small
\caption{Observation and replication conventions. The independent unit $s$
is a complete initialization at fixed corpus, source order and initial shared
state. Every covariance centers across $s$ at fixed observation coordinates.}
\label{model:tab:singlepass-observation-units}
\begin{tabular}{@{}>{\raggedright\arraybackslash}p{.17\textwidth}
>{\raggedright\arraybackslash}p{.29\textwidth}
>{\raggedright\arraybackslash}p{.34\textwidth}
>{\raggedright\arraybackslash}p{.13\textwidth}@{}}
\toprule Field & Native observation & Covariance units & Head-sign action\\\midrule
Row contrast & Centered row energy divided by floored total energy &
Average heads first; $N D^{-1}\tr\Cov_s$ over fixed context/decoder coordinates & Invariant\\[3pt]
Common vector & Full metric row centroid &
Componentwise centering; $N D^{-1}\tr\Cov_s$ after head averaging & Invariant\\[3pt]
Signed operator & Full $G_a$ in fixed Frobenius units &
Separate signed mean from invariant power; per-entry covariance & Headwise sign\\[3pt]
Prediction & $\phi(p)=2\sqrt p$ on the complete vocabulary &
$N\tr\Cov_s\phi$, context average, no vocabulary divisor & Invariant\\[3pt]
Shared displacement & $a_T-q^Ta_0$, $q=1-\eta\lambda$ &
$D^{-1}\tr\Cov_s$ in fixed parameter coordinates, all temporal cross terms & Invariant\\
\bottomrule\end{tabular}
\end{table}

\subsection{A native head-sign quotient for training and inference}
The metric and operator observations also carry representation symmetries.
A separate sign can be assigned to every native head, without changing
its metric learner or its prediction. This action leaves the stipulated
initial shared metric networks fixed, unlike a transformation that changes
their terminal LayerNorm representative.

\begin{proposition}[Native head-sign symmetry and operator covariance]
\label{model:prop:native-head-sign-symmetry}
Use the real-arithmetic native architecture of Equations~\eqref{model:eq:gram}
and~\eqref{model:eq:plga}, with dropout absent. For each decoder and head,
choose $\zeta_{\ell a}\in\{-1,1\}$ and transform
\begin{equation}
 (W^Q_{\ell a},b^Q_{\ell a},a_{\ell a},b^a_{\ell a})
 \longmapsto
 \zeta_{\ell a}(W^Q_{\ell a},b^Q_{\ell a},a_{\ell a},b^a_{\ell a}),
 \label{model:eq:native-head-sign-action}
\end{equation}
leaving all other parameters fixed. These actions form
$\Gamma_N=(\mathbb Z/2\mathbb Z)^{LN}$. They preserve the Gram matrices,
metrics $A$, potential matrices, attention weights, decoder states and
complete predictive law, while sending
$(Q_{\ell a},\G^{(\ell a)})$ to
$(\zeta_{\ell a}Q_{\ell a},\zeta_{\ell a}\G^{(\ell a)})$.
Co-transforming first Adam moments by the parameter signs and retaining
second moments, update indices and source state makes every clipped
AdamW update equivariant. Coordinate rates and decay coefficients may
differ between parameter groups.

Suppose the conditional initial law, given the source stream and fixed
shared initialization, is invariant under $\Gamma_N$. At every finite
training time, for distinct heads of one decoder,
\begin{gather}
 \E\G^{(\ell a)}=\bm0_{d\times d},\quad
 \E\bigl[\operatorname{vec}\G^{(\ell a)}
       (\operatorname{vec}\G^{(\ell b)})^{\mathsf T}\bigr]=\bm0_{d^2\times d^2}
       \quad(a\ne b),\notag\\
 \chi_N^G
 :=\frac{N}{d^2}\tr\Cov\!\left(\frac1N\sum_a
                   \operatorname{vec}\G^{(\ell a)}\right)
 =\frac1N\sum_a\E\frac{\|\G^{(\ell a)}\|_F^2}{d^2}.
 \label{model:eq:native-operator-sign-covariance}
\end{gather}
Assume the indicated second moments are finite. Fixed context and decoder
averaging preserves the last identity. The same sign argument eliminates
cross covariances between signed operators in distinct decoders.
\end{proposition}
\begin{proof}
RoPE is linear in the query. Changing its projection sign therefore
changes $Q$ by that sign, and
$(\zeta Q)^{\mathsf T}(\zeta Q)=Q^{\mathsf T}Q$ preserves the
metric input and every deterministic metric-learner output.
The potential is unchanged. Changing both output couplings in
$\G=aP^{\mathrm{pot}}+b^a$ changes $\G$ by the same sign. Thus
$(\zeta Q)(\zeta\G)K^{\mathsf T}=Q\G K^{\mathsf T}$, preserving
attention and the head output. The output projection and residual
updates preserve the decoder state. Induction over decoders proves
invariance of all later states and the final emission. Sign actions
on different heads commute and each squares to the identity.

The parameter transformation is a diagonal orthogonal matrix $S$.
Loss invariance gives $\nabla\ell(S\theta)=S\nabla\ell(\theta)$.
Global clipping commutes with $S$ because it preserves the gradient
norm. The first moment recursion co-transforms by $S$; the squared
gradient and the second moment recursion are unchanged. Bias correction,
the positive denominator offset, coordinate learning rates and weight
decay all commute with this sign action. Induction proves equivariance
of the full update and invariance of an invariant initial law under any
fixed valid source stream.

For the first moment, change variables under a sign flip of head $a$:
its expectation equals its negative and is zero. For the ordered
cross moment, flip $a$ alone. The second factor is unchanged, so this
expectation also equals its negative. Expanding the covariance of the
head mean now leaves only the diagonal terms, proving
Equation~\eqref{model:eq:native-operator-sign-covariance}.
A flip at one decoder preserves all later hidden inputs, giving the same
argument for distinct decoders. The stipulated moment condition justifies
these expectations and expansions.
\end{proof}

The symmetric real Xavier initialization and its positive variance
rescalings preserve this head-sign law. Conditioning on the initial
shared metric networks preserves it as well, since those parameters are
not changed by Equation~\eqref{model:eq:native-head-sign-action}.
\begin{corollary}[Uniform concentration of the signed mean operator]
\label{model:cor:head-sign-uniform-concentration}
Fix the head dimension $d$, positive offsets $\epsilon_A,\epsilon_o$,
weight decay $\lambda>0$, zero initial Adam moments and
$0<\beta_1<1$, $\beta_1^2<\beta_2<1$. Assume the conditional initial
law has the head-sign invariance above. Let the initial power-layer
parameters and terminal metric LayerNorm affine parameters have entries
bounded by one common $K_0$, independently of width. Each of their
coordinate rates may depend on width and update, provided
$0\le\lambda\eta_{N,k,j}\le1$. Along finite real-arithmetic paths,
there is a constant $B_G$ independent of width, update count and valid
source stream such that
\begin{equation}
 \chi_{N,T}^G\le B_G^2,\qquad
 \frac1{d^2}\E\left\|\frac1N\sum_{a=1}^N
                  G_{\ell a,T}\right\|_F^2\le\frac{B_G^2}{N}.
 \label{model:eq:head-sign-uniform-concentration}
\end{equation}
Consequently the signed mean operator converges to zero in mean square
along every such sequence $T_N$, including increasing single-pass
horizons. Its defined head susceptibility cannot diverge as a positive
power of $N$ under these hypotheses.
\end{corollary}
\begin{proof}
The weighted Cauchy--Schwarz bound for the bias-corrected Adam force gives
$|u_{k,j}|\le C_\beta$, with
$C_\beta=[(1-\beta_2)(1-\beta_1^2/\beta_2)]^{-1/2}$.
The coordinate recursion
$\theta'=(1-\lambda\eta)\theta-\eta u$ preserves
$|\theta|\le K:=\max\{K_0,C_\beta/\lambda\}$, since
$(1-\lambda\eta)K+\eta C_\beta\le K$.
A normalized row entering the last affine LayerNorm has Euclidean norm
at most $\sqrt d$, so every entry of $A$ is at most $(\sqrt d+1)K$
in absolute value. Put
\[
 M=d(\sqrt d+1)K^2+K,\quad
 L=\max\{|\log\epsilon_A|,|\log(M^2+\epsilon_A)|\},\quad
 B_G=dK e^{KL}+K.
\]
The preactivation $WA+b$ has entries bounded by $M$.
The native activation $f(z)=z^2\sigma(z)$ obeys $0\le f(z)\le z^2$,
so its offset power base lies in $[\epsilon_A,M^2+\epsilon_A]$.
The bounded exponent gives a powered entry at most $e^{KL}$.
Final multiplication and bias therefore give $|G_{ij}|\le B_G$.
The sign-covariance identity bounds $\chi_{N,T}^G$ by $B_G^2$.
The mean operator has zero expectation, so its second moment is its
covariance trace and the second inequality follows. All constants are
independent of $N$ and $T$, proving the sequence statement.
\end{proof}

This bound is conservative and does not estimate the measured operator
magnitude. It applies to the signed first operator moment. Sign-invariant
observations such as $A$, operator energy, $QG$ and the predictive law can
retain shared correlations. The corollary implies neither their
concentration nor a Gaussian law for the signed mean, nor uniform control
of derivatives with respect to a control. It constrains which observable
can carry a divergent head susceptibility; it does not exclude a
transition in the complete quotient law.

The covariance here varies the trained initialization at a fixed
context, followed by the declared context average. Varying prompts at
one fixed checkpoint defines a different covariance. The parameter
sign action does not turn those prompts into an initialization ensemble.

A finite pseudorandom sample need not be closed under the sign group,
and its unbiased sample covariance need not satisfy the population
identity exactly. Alternatively, any state law can be averaged over its
finite sign orbit. This symmetrized law satisfies the identity and has
exactly the same predictive law and row observations as the original
law. Orbit representatives are not independent training realizations.
Section~\ref{model:sec:critical-sign-checks} reports the native sign actions, exact orbit quadratures and finite-mixture checks.

This symmetry gives a compatible quotient of the complete state maps.
Let $\pi_N$ map a parameter, optimizer and source state to its
$\Gamma_N$ orbit. Equivariance gives a well-defined successor
$\overline\Phi_N$ with
$\pi_N\Phi_N=\overline\Phi_N\pi_N$, while invariance gives an
emission $\overline F_N$ with $F_N=\overline F_N\pi_N$.
Indeed two representatives of one orbit have successors in the same
orbit and identical emissions, which proves well-definedness directly.
If native operator caches are used, their $G$ entries carry the same
head signs; the cached metric, potential, keys and values remain
unchanged. The corresponding cached emission has the same quotient.
The sign group is discrete, so this argument supplies no tangent zero mode.

Head observations such as $(A,QG,K,V)$ are invariant, so their empirical
head measures and ordinary observation or blocking maps are defined on
this quotient. This does not assert that a finite collection of their
moments closes the optimizer dynamics. The signed operator mean alone
is not an invariant collective coordinate. Its susceptibility can vary
with the one-head power in
Equation~\eqref{model:eq:native-operator-sign-covariance}, even though every
cross-head covariance vanishes by symmetry. Conversely, vanishing
signed cross covariances do not make the heads independent: invariant
quadratic fields and row fields can retain the same random shared state.
A physical criticality interpretation must therefore use an invariant
observation or specify a justified representative convention and its
visibility in the emission.

The limiting signed law is described by
Theorem~\ref{model:thm:native-sign-gaussian-mixture}. Conditional on the invariant
orbit, independent head signs give a Gaussian limit for the
$N^{-1/2}$-normalized operator sum when its invariant covariance law
converges. The limiting covariance may be random and remain coupled to
the invariant training and inference state. Signed cross-head covariance
cancellation therefore does not imply independence of trained magnitudes
or a unique limiting law for the predictive sector. The complete statement,
finite-head error bounds and stand-alone proofs appear in
Theorem~\ref{model:thm:native-sign-gaussian-mixture} and
Section~\ref{model:sec:initialization-limit}.

\subsection{Predictive law fluctuations}
A finite predictive observation can avoid selecting a logit projection.
For a next-token probability vector $p$, define
$\phi(p)=2(\sqrt{p_1},\ldots,\sqrt{p_V})$. This is a fixed embedding of
the complete predictive law. Its covariance trace is in vocabulary-summed
units; it is not divided by $V$. The following contraction is the
square-root case of divergence contraction under an observation
channel \cite{csiszarshields2004}; the finite proof is included here.

\begin{proposition}[Predictive susceptibility and categorical observation]
\label{model:prop:predictive-hellinger}
Let $p$ be a random probability vector on a finite vocabulary, with
$p'$ an independent copy under the same conditional training law. Put
\[
 H^2(p,q)=\tfrac12\sum_v(\sqrt{p_v}-\sqrt{q_v})^2,
 \qquad \chi_N^\phi=N\tr\Cov(\phi(p)).
\]
Then
\begin{equation}
 \chi_N^\phi=4N\E H^2(p,p'),\qquad 0\le\chi_N^\phi\le4N.
 \label{model:eq:predictive-hellinger}
\end{equation}
For any column-stochastic matrix $K$ shared by every realization,
$\chi_N^{\phi\circ K}\le\chi_N^\phi$. These categorical observations
compose by ordinary matrix multiplication.
For $S\ge2$ complete training realizations, the unbiased sample trace
has the exact identity
\begin{equation}
 \widehat\chi_N^\phi
 =\frac{N}{S-1}\sum_{s=1}^S\|\phi(p_s)-\overline\phi\|^2
 =\frac{8N}{S(S-1)}\sum_{s<r}H^2(p_s,p_r),
 \label{model:eq:sample-predictive-hellinger}
\end{equation}
and obeys the same upper bound and channel contraction. All formulas
remain valid after averaging over a fixed context distribution.
\end{proposition}
\begin{proof}
For any square-integrable vector $X$ and an independent copy $X'$,
expansion gives
$\E\|X-X'\|^2=2\E\|X-\E X\|^2$.
Here $\|\phi(p)-\phi(q)\|^2=8H^2(p,q)$.
Also
$H^2(p,q)=1-\sum_v\sqrt{p_vq_v}$ lies in $[0,1]$:
nonnegativity follows from its squared-distance definition, and the
upper bound from nonnegative affinity. This proves
Equation~\eqref{model:eq:predictive-hellinger}.
For each output coordinate $j$, Cauchy--Schwarz gives
\[
 \sqrt{\Bigl(\sum_vK_{jv}p_v\Bigr)
             \Bigl(\sum_vK_{jv}q_v\Bigr)}
 \ge\sum_vK_{jv}\sqrt{p_vq_v}.
\]
Sum over $j$ and use $\sum_jK_{jv}=1$ to show
$H^2(Kp,Kq)\le H^2(p,q)$. Taking expectations proves contraction;
$K_2(K_1p)=(K_2K_1)p$ proves composition.
Finally, expansion of the finite double sum yields
$\sum_{s<r}\|X_s-X_r\|^2=S\sum_s\|X_s-\overline X\|^2$.
Substitution proves Equation~\eqref{model:eq:sample-predictive-hellinger}.
There are $S(S-1)/2$ pairs, each with squared Hellinger distance at
most one, giving the bound $4N$. Pairwise channel contraction gives
the sample contraction. Fixed context averaging preserves each identity
and inequality.
\end{proof}

This susceptibility measures variation of emitted distributions across
training initializations. A pulse instead measures propagation of a
specified state displacement under matched subsequent data. The two
observations need not be proportional. An extensive predictive trace
can reflect an extensive common sector without identifying an
intrinsic critical instability. A common categorical observation can
also erase that sector; the finite contraction has no converse without
an additional visibility condition.

The emission normalization also matters. Let $Z_1,\ldots,Z_N$ be
independent standard Gaussian head variables, and define a binary
readout by
$L_N=N^{-1/2}\sum_aZ_a$ and $p_N=(\sigma(L_N),1-\sigma(L_N))$,
where $\sigma(z)=(1+e^{-z})^{-1}$. The characteristic function of
$L_N$ is $[\exp(-u^2/(2N))]^N=\exp(-u^2/2)$, so its law is the same
standard Gaussian at every width. Consequently
$\chi_N^\phi=Nc$ with the fixed constant
$c=\tr\Cov(\phi(p_1))>0$, although the head variables are independent
and their mean has variance $1/N$. Positivity follows because the
bounded vector $\phi(p_1)$ is not almost surely constant.
This calculation supplies a regular random-readout reference.
Comparing the initial and trained predictive laws in identical units
is therefore necessary when interpreting an extensive predictive trace.
It does not identify the trained native readout with a Gaussian sum.

The symmetry class is conditional on its invariant orbit law. Under the
uniform envelope and convergence of that law's covariance,
Theorem~\ref{model:thm:native-sign-gaussian-mixture} gives a conditional Gaussian
limit for $\sqrt N$ times the signed operator mean. A random limiting
covariance yields a Gaussian mixture; a deterministic Gaussian additionally
requires covariance concentration. At zero generator control,
Corollary~\ref{model:cor:zero-control-regular-rate} gives a strictly positive
one-head power bound and an exact order $N^{-1/2}$ RMS rate.
Sign-invariant common fields and predictions do not inherit cancellation
of cross-head covariance. The full shared-force and covariance arguments
in Section~\ref{model:app:singlepass-selection} explain how their finite
training fluctuations survive that signed concentration.

\section{Training selection under a single-pass corpus law}
\label{model:sec:training-selection-main}
\begin{table}[htbp]\centering\small
\caption{Training-family roadmap. Trajectories are grouped by their
source, schedule and clock law; inner observations do not increase the
number of initialization identities.}\label{model:tab:training-family-roadmap}
\begin{tabular}{@{}p{.24\textwidth}p{.21\textwidth}p{.45\textwidth}@{}}
\toprule Family & Recorded size & Interpretation\\\midrule
Single-pass kinetic family & 384 trajectories; 1,228,800 updates &
Proper-prefix objective; constant update-clock schedule. Coarse and independent acquisition panels. The
independent component has 216 paths and 884,736 updates.\\[3pt]
Matched physical clock & 48 trajectories; 76,800 updates &
Proper-prefix objective; width-dependent rates and Adam memory, matched physical age and
consumed fraction; three initialization identities per condition.\\[3pt]
Source/schedule comparison family & 54 trajectories &
All-position training objective; separate source and update-count schedule interventions and
conditioning; not pooled with either family above.\\[3pt]
Chronological row continuations & 16 paths; 128 added updates &
Two retained identities from the matched-clock family; four aligned block
lengths and 64 path/scale cells.\\[3pt]
Equal physical-duration continuations & 16 paths; 1,600 executed updates &
Same two identities; 128 prefix replay updates and 1,472 additional updates.
Common duration $1/16$, with every aligned physical block retained.\\\bottomrule
\end{tabular}
\end{table}

The complete coarse and independent families contain 384 native trajectories
and 1,228,800 updates. The independent family uses six fresh initialization
identities, a disjoint master-corpus partition, $N=4,8,14,24$ heads, and
nine generator controls $g=0,0.75,1,1.25,1.5,1.75,2,2.5,3$.
Every path completes 4,096 updates on 131,072 distinct supervised blocks,
consuming $1/16$ of its population. The largest model has 245.1 million
parameters. Both families use five decoders, head dimension 64, the
recorded AdamW recipe and proper-prefix observations. Their complete
conditioning, observation units and ledger are given in
Sections~\ref{model:sec:statistics} and \ref{model:sec:critical-independent-results}.

\subsection{A finite kinetic control scale}
The independent endpoint row-susceptibility peaks occur at
$g=1.25,1.5,1.25,1.25$ for increasing width. Their intensive variances are
$0.0416$, $0.0374$, $0.0520$, and $0.0317$, and interpolated half-height widths are
$0.617$, $0.558$, $0.593$, and $0.705$. The peak has a large common covariance component;
its control window does not narrow systematically with width.
Complete profiles and grid brackets are retained in
Table~\ref{model:tab:critical-refinement-peaks}.

\begin{table}[htbp]\centering\small
\caption{Independent single-pass endpoints. Predictions were fixed from the
coarse family before independent training. The 24-head prediction uses the
separately specified largest-coarse-width reference. Bracket membership is
not a confidence interval.}
\label{model:tab:main-kinetic-selection}
\begin{tabular}{@{}rrrrl@{}}\toprule
Heads & Half-row crossing & Half-height width & Frozen clock error & In grid bracket\\\midrule
4 & 1.1307 & 0.617 & $+7.78\%$ & yes\\
8 & 1.2490 & 0.558 & $-3.42\%$ & yes\\
14 & 1.1754 & 0.593 & $+6.39\%$ & no\\
24 & 1.2632 & 0.705 & $-1.01\%$ & yes\\
\bottomrule\end{tabular}
\end{table}

The square-root-time forecast locates the endpoint half-row crossings
within eight percent in every width. Its 14-head prediction exceeds the
upper grid bracket by $4.70\times10^{-4}$; that criterion is retained.
The linear-clock errors are $-52.3\%$ to $-46.6\%$.
Across all 32 recorded times beyond the coarse horizon, frozen
square-root forecasts have relative RMS errors
$7.09\%,2.03\%,4.11\%,1.90\%$.
These times are correlated observations of the same six paths per cell.
A descriptive fit gives powers between $-0.523$ and $-0.480$ over the
resolved age range, with visible trailing-window dependence.
The result supports a finite kinetic scale, not an identified critical
exponent. Proposition~\ref{model:prop:kinetic-control-peak} proves why a regular
common-clock model can generate both size-dependent peak height and a
moving control window without a critical surface.
Section~\ref{model:sec:critical-time-predictions} specifies the frozen clock predictions and their complete time comparisons.

\begin{figure}[htbp]\centering
\includegraphics[width=.96\textwidth]{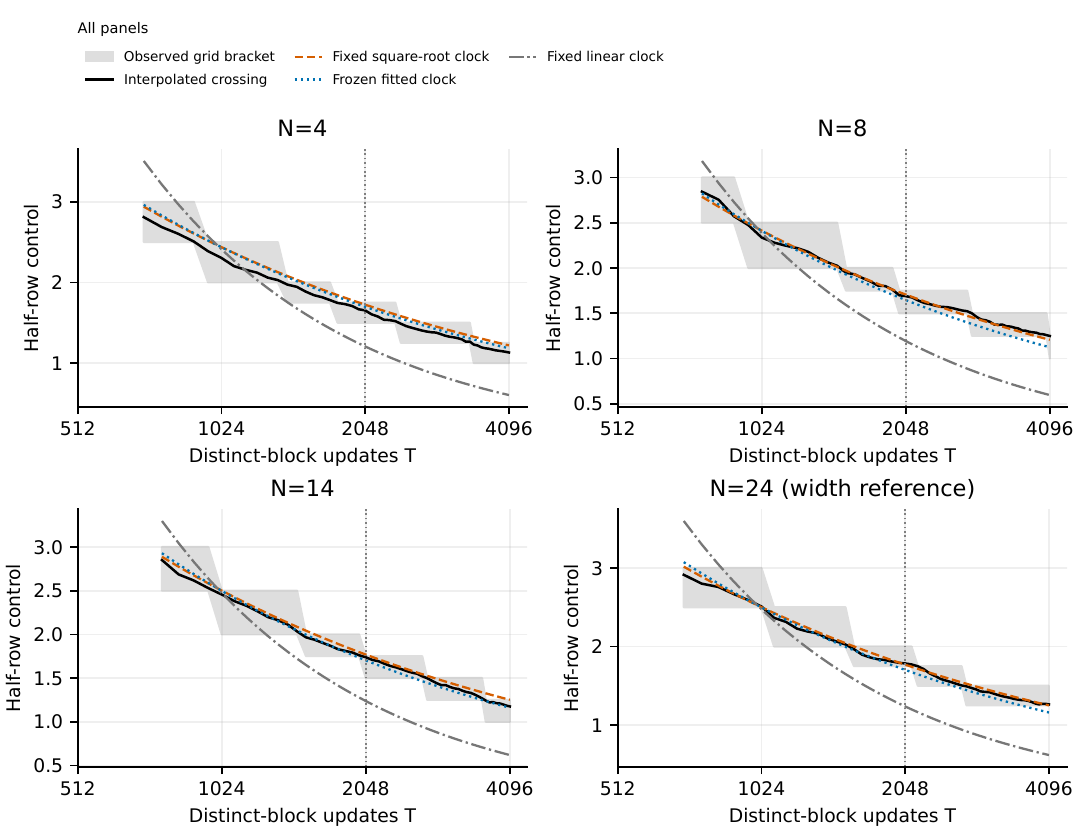}
\caption{Frozen time predictions and complete independent crossing profiles.
Shading gives the observed control-grid brackets. The widths share the
source convention and fixed shared initialization; independent complete
model initializations provide conditional replication.}
\label{model:fig:main-training-selection}
\end{figure}

\subsection{What training concentrates and what it retains}
At $g=3$, centered metric energies at the endpoint are
$1.013\times10^{-3},7.444\times10^{-5},1.229\times10^{-4},2.390\times10^{-5}$,
while total energies remain between $1.019$ and $1.027$. This is absolute
row-contrast suppression, rather than growth of its denominator.
The complete common-centroid per-coordinate variances nevertheless remain
$1.024,1.039,1.030,1.017$. Their scalar RMS amplitudes have much smaller
variances. Retaining only that amplitude loses the mobile common direction.
The signed-operator susceptibilities at the same control are
$10.52,10.33,10.57,10.78$, consistent with the bounded signed sector.
Section~\ref{model:sec:critical-absolute-observations} reports the absolute row and common-vector measurements; Section~\ref{model:sec:critical-sign-checks} gives the signed-operator checks.

The complete shared-parameter covariance also identifies a finite transport
scale. At $g=2$, endpoint per-coordinate variances are
$1.070,1.105,1.123,1.103$ times $10^{-3}$ and account for
$99.92$--$99.96\%$ of mean squared decay-corrected displacement.
Across positive controls, signed cross-block terms on the 256-update
partition contribute $2.55$--$3.61\%$ of endpoint variance.
The exact accumulation law retains those terms and within-block dynamics;
approximate quadratic-rate, linear-time accumulation over part of the
finite range does not establish independent one-update forcing.
Section~\ref{model:sec:critical-shared-accumulation} gives the full parameter accumulation and its signed chronological covariance budget.

Held-out next-token NLL decreases from approximately $10.38$ initially to
$6.32$--$6.41$ at $g=0$, $6.55$--$6.66$ at the row peaks, and
$6.70$--$6.78$ at $g=3$ on the fixed 128-context training-observation panel.
Predictive improvement therefore spans several row regimes; the row peak
does not select the smallest measured risk. Schedule and optimizer
interventions in the matched single-pass and conditional continuation
studies further separate risk selection, row contrast and incoming memory
(Sections~\ref{model:sec:matched-onepass}, \ref{model:sec:law-closure-results}).
Their full outcomes, complete covariance vectors and uncertainty
conventions are given in Appendices~\ref{model:sec:statistics}
and \ref{model:app:critical-independent-details}. The inference reductions in
Sections~\ref{model:sec:categorical-scale-results} and
\ref{model:sec:operator-cache-results} are evaluated at these recorded trained states.

\section{Shared-path kinetics and finite control observations}
\label{model:app:singlepass-selection}
\subsection{Finite-time continuity and limiting control fields}
\begin{proposition}[Finite-time native control continuity]
\label{model:prop:finite-training-continuity}
Fix a finite native architecture, finite update count and fixed valid
single-pass token stream. Use the dropout-free real-arithmetic program,
strictly positive LayerNorm and power-base offsets, continuous global
norm clipping and an Adam denominator offset $\epsilon_o>0$. Let the
finite coordinate learning rates depend continuously on a control $g$,
and use the same almost surely finite initial state for all controls.
Then the state after $T$ updates is continuous in $g$ for every such
initial state. Conditional means and covariance entries of the bounded
row and attention fields in Equation~\eqref{model:eq:conditional-head-fields},
and of every finite-vocabulary probability coordinate, are continuous
in $g$. Their fixed-width susceptibilities have no finite-control
divergence.
\end{proposition}
\begin{proof}
At any finite state, the positive offsets make the native forward map
and loss gradient continuous. Clipping is the continuous map
$z\mapsto z\min\{1,C/\|z\|\}$, with value $\bm0$ at $z=\bm0$.
Both Adam moment recursions are continuous, the second moment remains
nonnegative, and division by $\sqrt{\widehat v}+\epsilon_o$ is continuous
on that domain. Bias-correction denominators are strictly positive at
each of the finitely many positive update indices. Weight decay and the
control-dependent finite rates preserve continuity. Induction through
the fixed token stream proves state continuity and finiteness. The row
ratio uses a positive denominator floor and lies in $[0,1]$; normalized
attention entropy and probability coordinates also lie in $[0,1]$.
Dominated convergence therefore applies to their conditional first
and second moments. Covariance is the second moment minus a product
of first moments. Multiplication by the fixed width preserves
continuity and boundedness.
\end{proof}

This statement concerns the mathematical native program. Finite
precision has its own execution law, discussed in
Section~\ref{model:sec:numerical-response}. A critical control can emerge
only in a specified limit, such as $(N,T_N,M_N)\to\infty$ with a
specified consumed fraction. Neither a steep finite curve nor a
noncommuting time and control limit specifies that thermodynamic
family. In particular the zero-generator control freezes a parameter
subsystem. Unit response in those coordinates is then a consequence
of the update definition; identifying a fluctuating critical mode
requires the response and forcing conditions of
Section~\ref{model:sec:colored-collectives}.

Before a thermodynamic limit, fix the pretraining distribution $\mathcal D$
and specify corpus construction, tokenizer and supervised-block law for every
$M_N$. State whether the law conditions on each realized corpus or averages
over corpus draws. The retained finite master corpus does not itself realize
an infinite-resource sequence.

\subsection{Compact shared paths at matched single-pass clocks}
There is a joint size--time family in which existence of subsequential
shared-path laws follows directly from the native update, without a
stationary or independent-gradient assumption. Set
$g_N=2r/N$, $T_N=\lfloor N\tau_*\rfloor$, and take at least
$B_NT_N$ distinct source blocks. Both coordinate rate groups are then
bounded by $c/N$, with $c=6\times10^{-4}\max\{1,r\}$.
For a resource law $M_N=m_0N$ and fixed batch size, the consumed
fraction has a fixed finite limit. The complete source and optimizer
state still determine each gradient. Head count is the size coordinate: the
dense body has $O(N^2)$ parameters, while the supervised positions in
this resource family grow as $O(N)$. Their ratio therefore decreases as
$O(N^{-1})$ and remains part of the declared limit.

This finite-drift-clock family differs from $T_N\asymp N$ with
$\eta_{\rm gen}\asymp N^{-1/2}$, which keeps the quadratic generator clock
$\eta_{\rm gen}^2T_N$ of order one while its drift clock diverges as
$N^{1/2}$. That alternative requires additional control of the mean force,
weight-decay balance and temporal covariance. The following compactness
bound does not apply to it. Even at rates $N^{-1}$, an $O(N^{-1})$ variance
conclusion needs an integrated temporal-covariance bound. A fixed nonzero
consumed fraction also does not imply independent sampling: the collision
bound $(BT_N)^2/(2M_N)$ need not tend to zero.

\begin{proposition}[Compactness of the shared training paths]
\label{model:prop:shared-path-compactness}
Work in real arithmetic with zero initial Adam moments, fixed
$0<\beta_1<1$, $\beta_1^2<\beta_2<1$, denominator offset
$\epsilon_o>0$ and weight decay $\lambda\ge0$. Consider a fixed-dimensional
parameter subset $a_N\in\R^D$ with the same deterministic initial value
$a_0$ at every width. Suppose every participating coordinate has rate
$0\le\eta_{N,k,j}\le c/N$ and
$0\le\lambda\eta_{N,k,j}\le1$. No moment resets are made.
The gradients may depend on the entire model and its remaining source,
but are finite along each executed real-arithmetic path.
Linearly interpolate the parameter path at times $k/N$, and keep it
constant after $T_N/N$ up to $\tau_*$.
Put
\[
 C_\beta=\bigl[(1-\beta_2)(1-\beta_1^2/\beta_2)\bigr]^{-1/2},
 \qquad K=\|a_0\|_\infty+cC_\beta\tau_*.
\]
Every interpolated path obeys
\begin{align}
 \sup_{0\le\tau\le\tau_*}\|a_N(\tau)\|_\infty&\le K,\notag\\
 \|a_N(\tau)-a_N(\sigma)\|_\infty
 &\le c(\lambda K+C_\beta)|\tau-\sigma|.
 \label{model:eq:shared-path-bound}
\end{align}
Consequently the conditional path laws have weakly convergent
subsequences in $C([0,\tau_*];\R^D)$. The conclusion is uniform over
all valid single-pass source streams of the declared lengths.
\end{proposition}
\begin{proof}
For one coordinate, expansion of the two moments and weighted
Cauchy--Schwarz give
\[
 \left|{\widehat m_k\over\sqrt{\widehat v_k}+\epsilon_o}\right|
 \le {1-\beta_1\over\sqrt{1-\beta_2}}
       {\sqrt{1-\beta_2^k}\over1-\beta_1^k}
       \sqrt{\sum_{i=0}^{k-1}(\beta_1^2/\beta_2)^i}
 \le C_\beta.
\]
If $\widehat v_k=0$, all contributing gradients and the first moment
are zero, so the same bound holds. The last inequality uses
$1-\beta_1^k\ge1-\beta_1$, $1-\beta_2^k\le1$ and the convergent
geometric sum. The coordinate update therefore satisfies
\[
 |a_{N,k+1,j}|\le |a_{N,k,j}|+\eta_{N,k,j}C_\beta,
 \qquad
 |a_{N,k+1,j}-a_{N,k,j}|
 \le\eta_{N,k,j}(\lambda K+C_\beta).
\]
Summing the first inequality over at most $N\tau_*$ updates gives
$K$. Multiplying the second by $N$ bounds the slope of each
interpolated segment; the final constant segment also obeys it.
This proves Equation~\eqref{model:eq:shared-path-bound}.
The uniformly bounded and uniformly Lipschitz paths lie in a compact
subset of the continuous-path space by Arzel\`a--Ascoli. Probability
laws on a compact metric space are sequentially weakly compact,
which gives the stated subsequences. No step of the argument averages
or resamples the source stream.
\end{proof}

\begin{corollary}[Joint finite head and predictive observation limits]
\label{model:cor:joint-observation-compactness}
Use the normalized native initialization, head dimension $d$ fixed
independently of $N$, a fixed positive power-base offset and positive
weight decay, together with
the matched-clock family above. At any fixed finite panel of contexts,
decoders and normalized times, let $U_{N,a}$ retain the row ratio,
attention entropy, common centroid and complete PLGA operator of a head.
Let $\nu_N=N^{-1}\sum_a\delta_{U_{N,a}}$ be its empirical head measure;
a separate measure is retained for each panel entry. Include the full
predictive probability vector at each entry.
The joint laws of the shared parameter path, these empirical measures
and the predictive vectors have weakly convergent subsequences.
Along such a subsequence, continuous head averages and their unscaled
conditional first and second moments converge jointly with the
predictive observations.
\end{corollary}
\begin{proof}
The final metric-learner LayerNorm bounds every entry of $A$ by a
constant independent of its incoming Gram matrix. Indeed a normalized
row has Euclidean norm at most $\sqrt d$, and the affine parameters
are uniformly bounded by their AdamW recursion. The normalized initial
PLGA entries lie in a fixed bounded interval, and their coordinate
recursions have the same uniform bound. The positive power-base offset
then bounds every entry of $G$ uniformly in width and time, as proved
explicitly in Proposition~\ref{model:prop:generator-envelope}. The row ratio
and normalized entropy lie in $[0,1]$, and the common centroid is an
average of bounded rows. Thus $U_{N,a}$ lies in one fixed compact box
$K_U$. The space of probability measures on $K_U$, with its weak
topology, is compact. Each predictive vector lies in the compact
finite-vocabulary simplex. Their finite product with the compact
shared-path set from Proposition~\ref{model:prop:shared-path-compactness}
is compact, proving subsequential joint convergence.
For continuous $f$ on $K_U$, the map
$\nu\mapsto\int f\,d\nu$ is bounded and continuous. Finite products
of these averages are also bounded and continuous. Applying weak
convergence to those functions proves convergence of their first and
second moments, including cross moments with bounded continuous
predictive observations.
\end{proof}

These conclusions establish existence of compatible subsequential
observation laws in a specified single-pass normalization. They leave
uniqueness, a closed finite successor kernel, convergence rates and a
singular control field to be determined. In particular, convergence of
an unscaled covariance does not determine the behavior after multiplying
it by $N$. Raw optimizer moments and full-body path regularity are
outside the shared-path compactness statement. Finite-precision
comparisons retain the arithmetic policy and relative fluctuation-error
conditions of Section~\ref{model:sec:numerical-response}.

\subsection{Complete common fields}
The scalar row ratio observes the relative centered energy of
$A_{\ell a}\in\R^{d\times d}$, with $d$ fixed independently of $N$.
Write $P_d=I_d-d^{-1}\mathbf1\mathbf1^{\mathsf T}$, where $I_d$ is the
$d\times d$ identity and $\mathbf1\in\R^d$ is the vector of ones.
The common centroid, written as a row vector,
$c_{\ell a}=d^{-1}\mathbf1^{\mathsf T}A_{\ell a}\in\R^{1\times d}$ and the
head-mean operator $N^{-1}\sum_aG_{\ell a}$ retain additional information.
With angle brackets denoting the mean over all $d^2$ matrix entries,
orthogonality of row centering gives
\begin{equation}
 \langle A_{\ell a}^{\odot2}\rangle
 =\langle(P_dA_{\ell a})^{\odot2}\rangle
   +\|c_{\ell a}\|^2/d.
 \label{model:eq:critical-common-energy}
\end{equation}
The proof is expansion of
$A=P_dA+\mathbf1c$, whose cross inner product vanishes because
$\mathbf1^{\mathsf T}P_dA=\bm0^{\mathsf T}$. Concentration of $\|c\|$ does not
imply concentration of the vector $c$: random rotations of a
nonzero vector preserve its norm. The same issue applies to an
operator RMS. The conditional head susceptibility and covariance
blocking formulas apply to all coordinates of these vector fields,
with the coordinate normalization fixed before size comparisons.
The recorded experiments specialize these formulas to $d=64$.

\subsection{Creation of a shared fluctuation by the first adaptive update}
The fixed initial shared state does not imply that the learned shared
state concentrates. In the native control convention, the rate
$\eta_0g$ applies both to the shared residual metric networks and to
head-specific PLGA parameters. The following statement concerns only
the fixed-dimensional shared residual networks. The clipped gradient
still includes the effect of the complete model through its common
clipping factor.

\begin{proposition}[First-step shared force and sign normalization]
\label{model:prop:first-shared-force}
Let $a_0\in\R^D$ be the same shared parameter vector in every training
realization. Start AdamW with zero moments, positive denominator offset
$\epsilon_o$, positive learning rate $\eta_N$ and weight decay
$\lambda$. Let $h_N$ be its clipped shared loss gradient at the first
update, with every coordinate participating. In real arithmetic,
\begin{equation}
 a_1-a_0=-\eta_N\left(\lambda a_0+u_N\right),\qquad
 (u_N)_j={(h_N)_j\over |(h_N)_j|+\epsilon_o}.
 \label{model:eq:first-shared-force}
\end{equation}
Put $s_N=\operatorname{sign}(h_N)$, with sign zero equal to zero, and
$d_N=D^{-1}\sum_j\E|(u_N)_j-(s_N)_j|$. Then
\begin{align}
 D^{-1}\tr\Cov(a_1)&=\eta_N^2D^{-1}\tr\Cov(u_N),\\
 \left|D^{-1}\tr\Cov(u_N)-D^{-1}\tr\Cov(s_N)\right|&\le4d_N.
 \label{model:eq:first-shared-force-bound}
\end{align}
For nonzero $h$, the coordinate difference is
$\epsilon_o/(|h|+\epsilon_o)$; at zero it vanishes. For an empirical
ensemble of $S\ge2$ realizations with unbiased covariance divisor,
the second bound is $4S\widehat d_N/(S-1)$.
\end{proposition}
\begin{proof}
The first bias-corrected moments are $h_N$ and $h_N^{\odot2}$.
Inserting them into AdamW gives Equation~\eqref{model:eq:first-shared-force}.
The deterministic common vector $\lambda a_0$ contributes no covariance,
which proves the first identity. Both $u$ and $s$ lie in $[-1,1]$
coordinatewise. Therefore
$|\E(u_j^2-s_j^2)|\le2\E|u_j-s_j|$ and
$|(\E u_j)^2-(\E s_j)^2|\le2\E|u_j-s_j|$.
Subtract the squared means from the second moments, add over coordinates
and divide by $D$ to obtain the bound. The coordinate error follows by
subtracting $h/(|h|+\epsilon_o)$ from its sign. The same calculation
with empirical expectations proves the finite bound after multiplication
by $S/(S-1)$.
\end{proof}

For recorded arithmetic, write the displacement as its real formula plus
$e$, and let $\widehat V$ be the per-coordinate sample covariance trace
with one fixed divisor. Centering this identity over realizations and
applying Cauchy--Schwarz to its cross term gives
\begin{equation}
 \bigl|\widehat V(\Delta a)-\eta_N^2\widehat V(u_N)\bigr|
 \le 2\eta_N\sqrt{\widehat V(u_N)\widehat V(e)}+\widehat V(e).
 \label{model:eq:first-force-arithmetic-budget}
\end{equation}
A measured arithmetic residual can therefore be compared with the
fluctuation scale itself, as well as with the size of the parameter vector.

Gradient magnitude alone does not determine the scale of a shared
parameter fluctuation under Adam. Its sign distribution and its size
relative to the denominator offset enter the first update. A fixed
learning rate and a learning rate proportional to $N^{-1}$ also give
different first-step covariance scales. These are finite algorithmic
mechanisms. Relating their accumulation to a critical training limit
requires the subsequent moment, source and response laws; the first-step
identity makes no stationarity assumption.

\subsection{Exact accumulation of shared fluctuations}
The first adaptive step identifies a source of common variation, but its
accumulation depends on temporal covariances. There is an exact finite
budget for these covariances that does not require independent batches
or equilibrium. Covariance traces are divided by the fixed number $D$ of shared
coordinates; the inner product is the Euclidean inner product.

\begin{proposition}[Shared-path covariance budget]
\label{model:prop:shared-path-covariance-budget}
Let $a_k\in\R^D$ be a square-integrable random parameter path with a
deterministic initial value, and fix $q\in(0,1]$. For recorded times
$0=t_0<t_1<\cdots<t_m=T$, define
\[
 b_i=a_{t_i}-q^{t_i-t_{i-1}}a_{t_{i-1}},\quad
 K_{ij}=D^{-1}\E\langle b_i-\E b_i,b_j-\E b_j\rangle,
 \quad w_i=q^{T-t_i}.
\]
Then
\begin{align}
 a_T-q^Ta_0&=\sum_{i=1}^m w_ib_i,\notag\\
 D^{-1}\tr\Cov(a_T)&=
 \sum_{i=1}^m w_i^2K_{ii}+2\sum_{i<j}w_iw_jK_{ij}.
 \label{model:eq:shared-path-covariance-budget}
\end{align}
The matrix $K$ is positive semidefinite. The same identities hold
exactly for an empirical ensemble using divisor $S-1$ throughout.
Writing $d_s=a_{T,s}-q^Ta_0$ and $\widehat V$ for its per-coordinate
sample covariance trace, the finite decomposition is
\begin{equation}
 \frac1S\sum_s\frac{\|d_s\|^2}{D}
 =\left(\frac{\|\overline d\|^2}{D}-\frac{\widehat V}{S}\right)
   +\widehat V.
 \label{model:eq:shared-mean-budget}
\end{equation}
For independent, identically distributed realizations, the parenthesized
quantity is an
unbiased estimate of $D^{-1}\|\E d\|^2$, and may be negative.
\end{proposition}
\begin{proof}
Substitute the definition of $b_i$ into the weighted sum. The negative
term for $a_{t_{i-1}}$ cancels its preceding positive term, leaving only
$a_T-q^Ta_0$. Centering removes the deterministic initial value.
Expansion of the squared norm of the centered sum gives the covariance
formula. For every $v\in\R^m$,
$v^{\mathsf T}Kv=D^{-1}\E\|\sum_i v_i(b_i-\E b_i)\|^2\ge0$.
These are algebraic identities and remain valid under finite sample
averaging with one common divisor. Finally,
$S^{-1}\sum_s\|d_s\|^2/D=
\|\overline d\|^2/D+(S-1)\widehat V/S$ gives
Equation~\eqref{model:eq:shared-mean-budget}. Independence gives
$\E\|\overline d\|^2/D=\|\E d\|^2/D+V/S$ and
$\E\widehat V=V$, proving unbiasedness.
\end{proof}

For constant-rate AdamW, take $q=1-\eta\lambda>0$. In real arithmetic,
$b_i=-\eta\sum_{k=t_{i-1}}^{t_i-1}q^{t_i-1-k}u_k$, where $u_k$ is
the bias-corrected normalized Adam force. With recorded arithmetic,
the same definition of $b_i$ retains the accumulated arithmetic
residual and Equation~\eqref{model:eq:shared-path-covariance-budget} remains
an identity for the recorded parameters. Its off-diagonal contribution
is signed and depends on the recording cadence. Omitting it is an
independence assumption that can be checked directly.

To see when a quadratic control clock is appropriate, consider the
special case $q=1$ and individual updates. Suppose the normalized
force covariance is stationary under the stipulated conditional law,
with $c_r=D^{-1}\E\langle u_0-\E u_0,u_r-\E u_r\rangle$ and
$\sum_{r\ge1}|c_r|<\infty$. Direct grouping by lag gives
\begin{equation}
 \frac{\tr\Cov(a_T)}{D\eta^2T}
 =c_0+2\sum_{r=1}^{T-1}(1-r/T)c_r
 \longrightarrow c_0+2\sum_{r\ge1}c_r.
 \label{model:eq:quadratic-force-clock}
\end{equation}
The limit follows by dominated convergence on the absolutely summable
series. Thus $\eta^2T$ is a variance clock under these additional
conditions. A finite moving row threshold consistent with $g^2T$
does not establish those conditions, identify the coefficient, or prove
a diffusion limit. In particular, variation across model initializations
at one fixed source order is distinct from fresh-batch innovation noise.

The fixed denominator offset remains part of the width family.
Section~\ref{model:sec:optimizer-scales} gives distinct adaptive limits when
the clipped-gradient scale changes relative to that offset. Finite
sign-force agreement locates the observed family relative to its actual
denominators; it does not remove them from the thermodynamic
normalization.

The readout normalization can itself prevent a shared gradient from
concentrating. In the Gaussian construction above, put
$L_N(a)=aN^{-1/2}\sum_{h=1}^NZ_h=aZ_N$, use loss
$\ell(L)=\log(1+e^{-L})$, and start at a deterministic scalar $a_0$.
Then
\[
 \partial_a\ell(L_N(a_0))=-\frac{Z_N}{1+e^{a_0Z_N}},
 \qquad Z_N\sim\mathcal N(0,1)
\]
at every width. This nonconstant shared gradient has the same finite,
positive variance at every $N$, before any adaptive normalization.
Indeed its absolute value is bounded by $|Z_N|$, and it takes both
signs with positive probability. This is an explicit regular reference,
not a limit theorem for the native PLDR gradient. It shows why the
shared-gradient scale must be measured from the complete native
computational graph instead of inferred from a head-average formula.

\subsection{Separating timing fluctuations from a critical control}
Let $Q_N(x)$ be a head-mean row field and let $Z\in\{0,1\}$ assign one
label to the complete training realization. For example, the measured
label records whether that realization's context-averaged row ratio
has fallen below half its own initial value. The label is shared by
all of its contexts. Write $p=\Pr(Z=1)$,
$m_z(x)=\E[Q_N(x)\mid Z=z]$, and
$V_z(x)=\Cov(Q_N(x)\mid Z=z)$.

\begin{proposition}[Conditional timing-mixture decomposition]
\label{model:prop:critical-timing-mixture}
For $0<p<1$ and square-integrable vector observations,
\begin{equation}
 N\Cov(Q_N(x))=
 N\bigl[(1-p)V_0(x)+pV_1(x)\bigr]
 +Np(1-p)(m_1(x)-m_0(x))(m_1(x)-m_0(x))^{\mathsf T}.
 \label{model:eq:critical-timing-mixture}
\end{equation}
If a class is empty, its weight is zero and the within-class term is
the full covariance. The corresponding finite-sample decomposition is
exact when both terms use the same divisor $S-1$, even when the labels
are functions of the observed realizations.
\end{proposition}
\begin{proof}
Write $Q_N-\E Q_N=(Q_N-m_Z)+(m_Z-\E Q_N)$.
The conditional mean of the first term given $Z$ is zero, so the two
cross terms vanish after conditioning. The remaining between-class
covariance equals $p(1-p)(m_1-m_0)(m_1-m_0)^{\mathsf T}$ by inserting
$\E Q_N=(1-p)m_0+pm_1$. For a finite sample use the class sample
means in the same expansion. Residuals sum to zero within each class,
so the cross sums vanish before dividing by $S-1$.
\end{proof}

A growing susceptibility and a narrowing control window can both
arise from a regular single-clock evolution. The next construction
makes this ambiguity exact; it is an observation-law example, not an
identification of the native PLDR update with a scalar logistic flow.

\begin{proposition}[Kinetic scaling of a finite control peak]
\label{model:prop:kinetic-control-peak}
Let $p>0$, let $r(u,\xi)$ be a bounded random scalar field, and suppose
$V(u)=\Var_\xi r(u,\xi)$ has an interior maximum at $u_*>0$ and
half-height points $0<u_-<u_*<u_+$. Define an exchangeable family by
$U_{N,a}(T,g)=r(g^pT,\xi)$ for every head, with $T>0$ and $g\ge0$.
Then
\begin{alignat}{2}
 \chi_{N,T}(g)&=NV(g^pT), &\quad g_*&=(u_*/T)^{1/p},\notag\\
 w_{1/2}&=(u_+^{1/p}-u_-^{1/p})T^{-1/p}, &\quad \max_g\chi_{N,T}&=NV(u_*).
 \label{model:eq:kinetic-control-peak}
\end{alignat}
Consequently a chosen horizon $T_N=cN^z$ produces a peak of order $N$
and a window of order $N^{-z/p}$, with a finite nonzero intensive
variance at its maximum. These powers alone do not identify a
positive limiting critical control or an intrinsic relaxation exponent.
The cases $p=1$ and $p=2$ give linear and quadratic control clocks.
The construction permits smooth, bounded trajectories that move from
large to small row values for $g>0$.
\end{proposition}
\begin{proof}
The head mean is exactly $r(g^pT,\xi)$, hence its susceptibility is
$NV(g^pT)$. The increasing change of variable $u=g^pT$ maps the maximum
and half-height points to the stated controls. Substituting $T_N$
proves the scaling formulas; $z$ was chosen in the observation horizon.
For an explicit smooth example, let $\xi$ select $a$ or $b$ with equal
probability, where $0<a<b$, and set
$r(u,\xi)=(1+e^{u-\xi})^{-1}$. With $m=(a+b)/2$ and
$d=(b-a)/2$, direct subtraction gives
\[
 V(u)=\frac14\left(\frac{\sinh d}
                             {\cosh(u-m)+\cosh d}\right)^2.
\]
This smooth function has its unique maximum at $m$, tends to zero
at either infinite endpoint, and has two finite symmetric half-height
points. Choosing $m$ larger than their distance from the maximum
makes both points positive. Both row trajectories decrease smoothly,
and their maximal variance is $\tfrac14\tanh^2(d/2)>0$.
\end{proof}

The clock matching in this construction is explicit. For its quadratic
case $u=g^2T$, take a fixed generator/body rate ratio $r_0>0$,
$g_N=2r_0/N$ and the finite-clock horizon
$T_N=\lfloor N\tau_*\rfloor$. Then
\begin{equation}
 u_N=g_N^2T_N=\frac{4r_0^2\tau_*}{N}+O(N^{-2})\longrightarrow0.
 \label{model:eq:kinetic-clock-matching}
\end{equation}
Keeping $u_N$ bounded above and away from zero at this fixed ratio requires
$T_N$ of order $N^2$. For a fixed batch and a nonzero finite consumed
fraction, that choice also requires a source population of order $N^2$;
both accumulated learning-rate clocks are then of order $N$.
These statements follow by substitution into the declared clocks.
They distinguish the compact finite-clock family from the quadratic
observation window. The measured finite $g^2T$ approximation supplies
no native long-horizon limit theorem at the latter scaling.

Equation~\eqref{model:eq:critical-timing-mixture} identifies the fraction of a
finite peak associated with differing training outcomes. It does not
classify every endogenous shared mode as regular: a genuine transition
can also produce phase coexistence. Equation~\eqref{model:eq:kinetic-control-peak}
shows why size scaling must be assessed together with peak-location
drift, both optimizer clocks, source depletion and a resolved response
observation. Universality additionally concerns limiting joint laws and
predictions under specified changes of microscopic parameters, rather
than agreement of a finite fitted slope with a familiar exponent.

\FloatBarrier
\par\medskip\noindent
Comparing those collectives across widths requires a specified clock.
Chapter~\ref{ch:physical-clocks} matches optimizer memory and finite corpus
exposure, then derives the finite matrix flux used to measure their motion.

\chapter{Physical clocks and finite matrix flux}
\label{ch:physical-clocks}
This chapter places optimizer memory, response and row transport on
explicit physical clocks. It combines a uniform matched-memory force
bound with consuming-population forcing and exact finite matrix increments,
including the distinction between finite cross terms and derivatives.

\section{A uniform force envelope for matched optimizer memory}
\label{model:sec:matched-memory-bound}
A joint size--time family must specify optimizer memory in the same time
coordinate as its learning rates and resource consumption. We use the
standard bias-corrected Adam moments \cite{kingma2015adam}. The following
bound supplies one prerequisite for such a family. It asserts neither
convergence of the trained law nor a critical limit.

\begin{proposition}[Matched-memory adaptive-force bound]
\label{model:prop:matched-memory-force}
Let $\gamma_1,\gamma_2>0$ and $2\gamma_1>\gamma_2$. For $h>0$ set
$\beta_i=\exp(-\gamma_i h)$. Given any finite real scalar gradient
sequence $g_1,\ldots,g_t$, initialize $m_0=v_0=0$ and use
\[
 m_t=\beta_1m_{t-1}+(1-\beta_1)g_t,\qquad
 v_t=\beta_2v_{t-1}+(1-\beta_2)g_t^2.
\]
For $t\geq1$, let $\widehat m_t=m_t/(1-\beta_1^t)$,
$\widehat v_t=v_t/(1-\beta_2^t)$, and $\epsilon>0$. Then
\begin{alignat}{2}
 \left|\frac{\widehat m_t}{\sqrt{\widehat v_t}+\epsilon}\right|
 &\leq C_h,
 &\quad C_h&=\frac{1-\beta_1}
 {\sqrt{(1-\beta_2)(1-\beta_1^2/\beta_2)}},\label{model:eq:matched-force}\\
 \lim_{h\downarrow0} C_h
 &=\frac{\gamma_1}{\sqrt{\gamma_2(2\gamma_1-\gamma_2)}}.
 \label{model:eq:matched-force-limit}
\end{alignat}
In particular, for any fixed $h_{\max}>0$, these forces are uniformly
bounded over $0<h\leq h_{\max}$, all $t\geq1$, and all gradient sequences.
For a scalar decayed coordinate updated by
$\theta^+=(1-\lambda\eta)\theta-\eta f$, with $\lambda>0$,
$0\leq\lambda\eta\leq1$ and the above force, the interval
$|\theta|\leq\max(|\theta_0|,\overline C/\lambda)$ is invariant,
where $\overline C=\sup_{0<h\leq h_{\max}}C_h<\infty$.
\end{proposition}

\begin{proof}
Put $r=\beta_1^2/\beta_2\in(0,1)$. Expanding both moment recurrences
and applying weighted Cauchy--Schwarz gives, when $v_t>0$,
\[
 \frac{|m_t|}{\sqrt{v_t}}
 \leq\frac{1-\beta_1}{\sqrt{1-\beta_2}}
       \left(\sum_{j=0}^{t-1}r^j\right)^{1/2}.
\]
Bias correction therefore bounds the absolute normalized force by
\[
 C_h\frac{\sqrt{(1-\beta_2^t)(1-r^t)}}{1-\beta_1^t}.
\]
For $u,v\geq0$, $(1-u^2)(1-v^2)\leq(1-uv)^2$, since the difference
of the right and left sides is $(u-v)^2$. Taking
$u=\beta_2^{t/2}$ and $v=r^{t/2}$ proves that the last fraction is
at most one. A positive denominator offset can only decrease the force.
If $v_t=0$, every contributing gradient is zero, hence $m_t=0$ and
the force is zero. Finally, $1-e^{-ah}\sim ah$ gives
\eqref{model:eq:matched-force-limit}. Extending $C_h$ by this finite limit
makes it continuous on $[0,h_{\max}]$, so its supremum is finite.
For the decayed coordinate,
$|\theta^+|\leq(1-\lambda\eta)|\theta|+\eta\overline C$,
which proves the invariant interval by induction.
\end{proof}

The bound assumes standard zero initial moments, or moments generated by the
same admitted history. Arbitrarily intervened nonzero moments require their
own initial envelope. It applies to coordinates with the stated decay rule;
it does not bound all network derivatives, establish floating-point
stability, or justify replacing a consuming data stream by stationary noise.

\paragraph{Finite formula verification.}
With $\gamma_1=-1792\log(0.9)$, $\gamma_2=-1792\log(0.95)$ and
$h=(128N)^{-1}$ for $N\in\{4,8,14,24\}$, direct finite sums were checked
on 4,096 fixed-seed random scalar gradient sequences at four ages per
step size and on 16 extremizing sequences proportional to
$(\beta_1/\beta_2)^j$ in reverse chronological order. The maximum
extremizer discrepancy is $4.45\times10^{-16}$ and the largest excess over
the finite-age bound is below $3.34\times10^{-16}$. The limiting envelope is
approximately $1.165107$. These finite arithmetic checks accompany the
proof and do not constitute a native training experiment or a critical
limit. The bound permits a matched-memory family without a divergent
coordinate force; it does not establish compact network derivatives,
uniqueness of the state-law limit or critical selection.
\begin{table}[htbp]\centering\small
\caption{Finite scalar verification of the matched-memory force envelope. These are evaluated step sizes, not a newly trained model family. All 4,096 random sequences and 16 extremizing sequences obey the finite-age formula within rounding tolerance.}
\label{model:tab:matched-memory-check}
\begin{tabular}{@{}rrrrr@{}}
\toprule $N$ & $h^{-1}$ & $\beta_1$ & $\beta_2$ & $C_h$\\\midrule
4 & 512 & 0.691590 & 0.835666 & 1.163382\\
8 & 1024 & 0.831619 & 0.914148 & 1.164673\\
14 & 1792 & 0.900000 & 0.950000 & 1.164965\\
24 & 3072 & 0.940390 & 0.970522 & 1.165058\\
\bottomrule\end{tabular}\end{table}

\section{Physical response and the finite resource}
\label{model:sec:physical-clock}
A relaxation exponent depends on the clock in which it is measured.
Matching optimizer memory to a decreasing step $h_N$ makes this choice
essential. A step eigenvalue near one by itself identifies neither a
vanishing physical restoring force nor a divergent susceptibility.

\begin{proposition}[Clock-covariant scalar response]
\label{model:prop:physical-clock}
Let $h_N,\kappa_N>0$ and $\lambda_N=e^{-\kappa_Nh_N}$.
For the stable scalar update $z_{k+1}=\lambda_Nz_k$,
\begin{equation}
 \tau_N^{\rm updates}=-\frac1{\log\lambda_N}
     =\frac1{\kappa_Nh_N},\qquad
 \tau_N^{\rm physical}=-\frac{h_N}{\log\lambda_N}
     =\frac1{\kappa_N}.
 \label{model:eq:two-response-clocks}
\end{equation}
If a bi-infinite sequence of iid centered innovations has finite variance
$a_N=\sigma_N^2(1-e^{-2\kappa_Nh_N})/(2\kappa_N)$, the stationary
variance of the causal strictly stationary solution of
$z_{k+1}=\lambda_Nz_k+\xi_k$ is
\begin{equation}
 \operatorname{Var}(z_k)=\frac{\sigma_N^2}{2\kappa_N}.
 \label{model:eq:clock-invariant-variance}
\end{equation}
Consequently $h_N\asymp N^{-a}$ and $\kappa_N\asymp N^{-b}$ give
relaxation orders $N^{a+b}$ in updates and $N^b$ in physical time.
When $\kappa_Nh_N\to0$ and $\sigma_N^2\asymp N^{-v}$,
the innovation variance has order $N^{-(a+v)}$ and the stationary
variance has order $N^{b-v}$.
\end{proposition}
\begin{proof}
Taking the logarithm of $\lambda_N$ gives
\eqref{model:eq:two-response-clocks}. The stationary solution is the
$L^2$-convergent sum $\sum_{j\geq0}\lambda_N^j\xi_{k-1-j}$.
The common innovation law makes this causal solution strictly stationary.
Independence and centering remove all cross terms, so its variance is
$a_N\sum_{j\geq0}\lambda_N^{2j}=a_N/(1-\lambda_N^2)$.
Substitution proves \eqref{model:eq:clock-invariant-variance}.
Pairwise orthogonal centered innovations with constant variance suffice
for the same second-moment calculation, but do not imply strict
stationarity without an invariant joint innovation law.
The scaling orders follow by multiplication and
$1-e^{-2\kappa_Nh_N}\sim2\kappa_Nh_N$.
\end{proof}

For the normalized mode $z_N=\sqrt N\,\bar u_N$, the variance in
\eqref{model:eq:clock-invariant-variance} is its susceptibility. The
white-innovation specialization of Corollary~\ref{model:cor:colored-exponents}
then has update-gap exponent $\zeta=a+b$ and innovation exponent
$\upsilon=a+v$, hence $\zeta-\upsilon=b-v$. The artificial clock
power cancels. Stationary forcing and scalar stability are hypotheses
of this comparison; a nonstationary or nonnormal native process retains
its chronological propagator and full forcing covariance.

An executed scalar control fixes $\kappa=0.7$ and
$h_N=(128N)^{-1}$ at $N=4,8,14,24,96,384$.
Its update relaxation grows linearly while its physical relaxation
remains $1/0.7$. Direct discrete propagation for one physical unit
agrees with the matrix exponential, and the exact innovation
normalization keeps stationary variance fixed. These are finite
analytical controls, distinct from native model observations.
\begin{table}[htbp]\centering\small
\caption{Executed scalar clock control with physical decay $0.7$ and unit diffusion variance. Growing update relaxation coexists with fixed physical response and stationary variance.}
\label{model:tab:physical-clock-check}
\begin{tabular}{@{}rrrrr@{}}
\toprule $N$ & $h_N^{-1}$ & $\tau^{\rm updates}$ & $\tau^{\rm physical}$ & Stationary variance\\\midrule
4 & 512 & 731.4286 & 1.42857143 & 0.71428571\\
8 & 1024 & 1462.8571 & 1.42857143 & 0.71428571\\
14 & 1792 & 2560.0000 & 1.42857143 & 0.71428571\\
24 & 3072 & 4388.5714 & 1.42857143 & 0.71428571\\
96 & 12288 & 17554.2857 & 1.42857143 & 0.71428571\\
384 & 49152 & 70217.1429 & 1.42857143 & 0.71428571\\
\bottomrule\end{tabular}\end{table}

The source resource has a separate scaling requirement. With fixed
batch $B$, remaining fraction $1-q_N$ bounded away from zero,
$M_N\asymp N^m$ blocks and a window
$\tau_N^{\rm updates}\asymp N^\zeta$, the sufficient independent-source
path-coupling condition of Proposition~\ref{model:prop:data-path-coupling} is
\[
 \frac{(B\tau_N^{\rm updates})^2}{(1-q_N)M_N}\longrightarrow0.
\]
Its ratio has order $N^{2\zeta-m}$, so this sufficient condition holds
in the stated power-law comparison exactly when $m>2\zeta$.
In particular, a linear-resource family with $T_N=128N$,
$M_N=16384N$, $B=32$ and $q_N=1/4$ satisfies it only for $\zeta<1/2$.
Even the regular fixed-$\kappa$ mode above has $\zeta=1$ in updates
and does not qualify through this condition. Failure of a sufficient
coupling bound does not rule out a different approximation; it does
prevent using that bound to justify stationary independent forcing.
A consuming source is therefore retained in the complete nonstationary
law \eqref{model:eq:complete-conditioned-law}. The finite controls also
evaluate the exact sampling distance and its collision bound at
16 population/window pairs, without imposing a native noise model.
Section~\ref{model:sec:matched-clock-results} gives the matched physical-clock acquisition and its width-dependent outcomes.

\section{Chronological forcing from a consuming finite population}
\label{model:sec:consuming-bridge}
A linear resource need not approximate stationary independent forcing in
physical time. The following reference law retains its depletion covariance
exactly. It is useful for a frozen-state response or a specified linear
surrogate. Adaptive gradients evaluated at changing native parameters do
not satisfy its frozen-population assumption automatically.

\begin{proposition}[Finite-population chronological covariance]
\label{model:prop:consuming-bridge}
Let $u_1,\ldots,u_M\in\R^d$ have mean zero and
$C=M^{-1}\sum_i u_i u_i^\top$. Draw a uniform permutation $\pi$ of
these $M>1$ identities. For integers $B\geq1$, $T\geq0$ and $BT\leq M$, set
$\eta_k=B^{-1}\sum_{r=1}^B u_{\pi(Bk+r)}$, $0\leq k<T$.
Let deterministic matrices $D_k:\R^d\to\R^p$ represent the full
chronological transport of each forcing increment to the final time.
Then $Z_T=\sum_{k<T}D_k\eta_k$ has mean zero and covariance
\begin{equation}
 \Cov(Z_T)=\frac{1}{M-1}\left[
 \frac{M}{B}\sum_{k<T}D_kCD_k^\top-
 \left(\sum_{k<T}D_k\right)C\left(\sum_{k<T}D_k\right)^\top
 \right].\label{model:eq:consuming-bridge}
\end{equation}
In particular, an unweighted average of all $M$ identities has zero
variance. The displayed covariance is positive semidefinite even when
the $D_k$ contain rotations or signed directions.
\end{proposition}
\begin{proof}
Uniformity gives $\E u_{\pi(i)}u_{\pi(i)}^\top=C$.
For distinct positions $i,j$, sampling ordered distinct pairs gives
\[
 \E u_{\pi(i)}u_{\pi(j)}^\top
 =\frac{\sum_{a\ne b}u_au_b^\top}{M(M-1)}
 =-\frac{C}{M-1},
\]
because $\sum_a u_a=\bm0$. Therefore
$\Cov(\eta_k)=(M-B)C/[B(M-1)]$ and
$\Cov(\eta_k,\eta_l)=-C/(M-1)$ for $k\ne l$.
Expanding $\Cov(Z_T)$ with these diagonal and cross terms and collecting
all ordered pairs proves the formula. Positivity follows from its
construction as $\E Z_TZ_T^\top$. At full consumption, equal weights
sum the entire centered population, giving zero identically.
\end{proof}

This forcing law is compatible with aligned temporal blocking. If the
transport is constant on each group of $b$ batches, replace that group's
forcing by their average, set $B'=bB$, and multiply its transport matrix
by $b$. The grouped $Z_T$ is unchanged, so the same covariance formula
holds with $T'=T/b$ when $b$ divides $T$. In a general chronological
linear recursion, each group instead retains its internally transported
sum and its cross-group covariance. Discarding that dependence introduces
a closure defect. Thus resource consumption, chronological transport and
observation reduction refer to one specified law.

\begin{corollary}[Physical-clock bridge correction]
\label{model:cor:consuming-clock}
Consider a sequence of these populations with $h_N\downarrow0$,
$M_Nh_N\to m>0$, $T_Nh_N\to\tau$, fixed $B$, $B\tau<m$,
and $C_N\to C$. Suppose
$D_{k,N}=h_NF(kh_N)$ for a continuous matrix function
$F:[0,\tau]\to\R^{p\times d}$, extending continuously to a common
neighborhood if the endpoints vary. Then
\begin{equation}
 h_N^{-1}\Cov(Z_{T_N})\longrightarrow
 \frac1B\int_0^\tau F(s)CF(s)^\top\,ds
 -\frac1m\left(\int_0^\tau F(s)\,ds\right)
 C\left(\int_0^\tau F(s)\,ds\right)^\top.
 \label{model:eq:consuming-clock}
\end{equation}
This is a second-moment limit; a Gaussian process limit requires
additional population and maximum-increment conditions.
\end{corollary}
\begin{proof}
Substitute $D_{k,N}=h_NF(kh_N)$ into
Equation~\eqref{model:eq:consuming-bridge} and divide by $h_N$.
The first term becomes $M_N/[B(M_N-1)]$ times the Riemann sum
$h_N\sum_k F(kh_N)C_NF(kh_N)^\top$.
The second has coefficient $[h_N(M_N-1)]^{-1}$ multiplying the
product of the two corresponding first-moment Riemann sums.
Continuity, finite matrix dimension and $C_N\to C$ give the stated
limits. No exchange of random infinite sums is required.
\end{proof}

For $F=I$, the limit is $(\tau/B)(1-B\tau/m)C$.
Thus a fixed consumed fraction leaves an order-one correction in the
normalized fluctuation covariance even as each update becomes small.
For a scalar stable response $F(s)=e^{-\kappa(\tau-s)}$, it is
\[
 C\left[\frac{1-e^{-2\kappa\tau}}{2B\kappa}
 -\frac{(1-e^{-\kappa\tau})^2}{m\kappa^2}\right].
\]
This finite-resource mechanism changes fluctuation amplitude without
requiring a vanishing restoring rate. In the measured family
$M_Nh_N=128$, $B=32$ and $\tau=1$, the unweighted reference correction
is $1-q=3/4$. This value is a derived reference-law prediction, not
an estimate of native gradient noise. Conditioning on a realized order
removes the permutation randomness; then training uncertainty instead
comes from the declared initialization law. Random source order is a
separate averaging operation on the same complete law.
The measured resource and clock values for this family are specified in Section~\ref{model:sec:matched-clock-results}.

If a native transported fluctuation admits a coupled decomposition
$X_N=Z_N+R_N$, the reference limit transfers when
$\|R_N-\E R_N\|_{L^2}=o(\sqrt{h_N})$ and
$\|Z_N\|_{L^2}=O(\sqrt{h_N})$. Indeed, covariance expansion and
Cauchy--Schwarz bound the covariance difference in operator norm by
$2\|Z_N\|_{L^2}\|R_N-\E R_N\|_{L^2}
 +\|R_N-\E R_N\|_{L^2}^2=o(h_N)$.
For adapted transport matrices or gradients, the unresolved dependence
belongs to $R_N$ or to the full chronological forcing covariance.
The native experiment does not assume that this remainder is negligible.

The finite controls enumerate all 24 or 720 permutations at population
sizes four and six with vector-valued observations and signed rectangular
transport matrices. All four covariance identities agree within
$1.78\times10^{-15}$. At full consumption the unweighted centered sum is
exactly zero. For $\kappa=0.7$, the scalar physical-clock covariance
converges from $0.01275917$ at $N=4$ to $0.01277625$ at $N=384$;
the derived limit is $0.01277643$. These are executed mathematical
controls, separate from native training observations.

\section{Finite row flux and the collective clock}
\label{model:sec:collective-clock}
Matching optimizer memory and coordinate rates does not specify the
speed of a generated matrix observable. A finite row-energy balance
retains the signed cross terms, quadratic increments and normalization
changes induced by the complete native update.

\begin{proposition}[Finite row flux and compatible blocking]
\label{model:prop:collective-clock}
Let $A,D\in\R^{d\times d}$ and
$\Pi=I-\mathbf1\mathbf1^\top/d$. Suppose
$E=\|A\|_F^2>0$ and $E'=\|A+D\|_F^2>0$, and set
$u(A)=\|\Pi A\|_F^2/E$. Define
\begin{align}
 \ell(A,D)&=\frac{2\langle\Pi A,\Pi D\rangle_F
                    -2u(A)\langle A,D\rangle_F}{E'},\\
 q(A,D)&=\frac{\|\Pi D\|_F^2-u(A)\|D\|_F^2}{E'}.
 \label{model:eq:row-fluxes}
\end{align}
The finite row map is exactly
\begin{equation}
 u(A+D)=u(A)+\ell(A,D)+q(A,D).
 \label{model:eq:finite-collective-clock}
\end{equation}
The same equality holds after fixed normalized averaging over heads,
decoders and contexts. Chronological blocking is compatible when the
blocked displacement is the sum of its actual successive matrix
increments; its quadratic terms retain their cross-time inner products.
\end{proposition}
\begin{proof}
Expand $\|\Pi(A+D)\|_F^2$ and
$E'=E+2\langle A,D\rangle_F+\|D\|_F^2$.
Subtract $u(A)E'$ from the expanded numerator. The constant terms
cancel because $u(A)E=\|\Pi A\|_F^2$, leaving precisely the
numerators of $\ell$ and $q$. Division by $E'>0$ proves the formula.
Fixed weighted averaging preserves an equality term by term.
For a sequence $A_{k+1}=A_k+D_k$, both the sum of the one-step
row increments and the formula with $D=\sum_kD_k$ give
$u(A_T)-u(A_0)$. Expanding the blocked squared norm includes
$2\langle D_i,D_j\rangle_F$ for $i<j$, and likewise after applying
$\Pi$. Thus omitting those terms generally changes the blocked map.
\end{proof}

\begin{lemma}[Finite cross term and differential error]
\label{model:lem:finite-defect}
In Proposition~\ref{model:prop:collective-clock}, put
$L=E'\ell(A,D)$ and $Q=E'q(A,D)$. On the nonzero-matrix domain,
\begin{align}
 \mathrm D u(A)[D]&=L/E,\label{model:eq:row-matrix-derivative}\\
 u(A+D)-u(A)-\mathrm D u(A)[D]
 &=\frac{Q}{E'}-\frac{E'-E}{E'}\,\mathrm D u(A)[D].
 \label{model:eq:row-derivative-defect}
\end{align}
If $\delta=\|D\|_F/\|A\|_F<1$, then
\begin{equation}
 \left|u(A+D)-u(A)-\mathrm D u(A)[D]\right|
 \leq\frac{3\delta^2+\delta^3}{(1-\delta)^2}.
 \label{model:eq:row-derivative-bound}
\end{equation}
For a differentiable generator $A(\theta)$, write
$D=A(\theta+v)-A(\theta)$ and $Jv=\mathrm D A(\theta)[v]$.
The parameter directional derivative is $\mathrm D u(A)[Jv]$.
Replacing the matrix directional derivative by this quantity adds at most
$\|D-Jv\|_F/\|A\|_F$ to this bound. If the generator derivative
is $\Lambda$-Lipschitz along the parameter segment in the operator norm
induced by the specified parameter and Frobenius norms, that additional term is at most
$\Lambda\|v\|^2/(2\|A\|_F)$.
\end{lemma}
\begin{proof}
The quotient rule gives \eqref{model:eq:row-matrix-derivative}.
Subtracting $L/E$ from the exact finite increment $(L+Q)/E'$
gives \eqref{model:eq:row-derivative-defect}. Orthogonality of $\Pi$ gives
$0\leq u\leq1$ and
\[
 \nabla u(A)=\frac{2(\Pi A-uA)}{E},\qquad
 \|\nabla u(A)\|_F^2=\frac{4u(1-u)}{E}\leq\frac1E.
\]
Hence $|\mathrm D u(A)[D]|\leq\delta$.
Both $\|\Pi D\|_F^2$ and $u\|D\|_F^2$ lie between zero and
$\|D\|_F^2$, so $|Q|\leq E\delta^2$.
Expansion and the reverse triangle inequality give
$|E'-E|\leq E(2\delta+\delta^2)$ and
$E'\geq E(1-\delta)^2$. Substitution into
\eqref{model:eq:row-derivative-defect} proves \eqref{model:eq:row-derivative-bound}.
Linearity of the directional derivative and its gradient norm bound
control the additional generator remainder. Finally,
\[
 D-Jv=\int_0^1[\mathrm D A(\theta+sv)-\mathrm D A(\theta)]v\,ds
\]
has norm at most $\int_0^1\Lambda s\|v\|^2 ds$ under the
stated Lipschitz hypothesis.
\end{proof}

The finite cross term $\ell=L/E'$ has the outgoing energy denominator;
it is distinct from the matrix directional derivative $L/E$.
In the accompanying experimental data, the entry
\texttt{linear\_flux} stores the average finite cross term
$\bar\ell=\overline{L/E'}$ over heads, decoder layers and assessment
contexts. These measurements are reported in
Section~\ref{model:sec:collective-flux-results}, in the $\bar\ell$ column of
Table~\ref{model:tab:collective-row-flux}.
The finite cross term, the matrix directional derivative and the parameter
directional derivative $\mathrm D u(A)[Jv]$ have different mathematical roles.
For example, with
$A=\left(\begin{smallmatrix}1&0\\0&0\end{smallmatrix}\right)$ and
$D=\left(\begin{smallmatrix}0&0\\1&0\end{smallmatrix}\right)$,
$u(A)=1/2$, $u(A+D)=0$, $\ell=-1/2$, whereas
$\mathrm D u(A)[D]=-1$. This exact comparison has $\delta=1$,
outside the strict premise of \eqref{model:eq:row-derivative-bound}.
Fixed nonnegative normalized averaging preserves the finite identities
and the averaged absolute-error bound. None of these formulas identifies
a realized generator displacement with its parameter linearization.

\begin{corollary}[Pathwise differential approximation and its scale]
\label{model:cor:row-path-error}
Let $A_{k+1}=A_k+D_k\ne\bm0_{\mathrm{mat}}$, $0\leq k<T$, with $A_0\ne\bm0_{\mathrm{mat}}$ and
$\delta_k=\|D_k\|_F/\|A_k\|_F\leq r<1$.
Define the recorded-path derivative sum
$V_m=\sum_{k<m}\mathrm D u(A_k)[D_k]$. Then
\begin{equation}
 \max_{0\leq m\leq T}|u(A_m)-u(A_0)-V_m|
 \leq\sum_{k<T}b(\delta_k)
 \leq\frac{3+r}{(1-r)^2}\sum_{k<T}\delta_k^2,
 \qquad b(x)=\frac{3x^2+x^3}{(1-x)^2}.
 \label{model:eq:row-path-error}
\end{equation}
For parameter increments $v_k$, replacing each summand by
$\mathrm D u(A_k)[J_kv_k]$ adds
$\sum_{k<T}\|D_k-J_kv_k\|_F/\|A_k\|_F$ to the bound.
Consequently, for a family with a uniform $r<1$, vanishing summed squared
relative matrix increments and vanishing summed generator remainders
suffice for a uniform-in-time parameter-derivative approximation.
For accuracy $o(s_N)$ at a specified fluctuation scale $s_N>0$,
it is sufficient that the combined bound is $o(s_N)$. No converse is asserted.
\end{corollary}
\begin{proof}
The reverse triangle inequality gives
$\|A_{k+1}\|_F\geq(1-r)\|A_k\|_F>0$ inductively.
Telescope the actual row increments up to each $m$. Subtract $V_m$,
apply Lemma~\ref{model:lem:finite-defect} to each edge, and use the triangle
inequality. All $b(\delta_k)$ are nonnegative, so their prefix sums
are bounded by the full sum. Since $\delta_k\leq r<1$,
$b(\delta_k)\leq(3+r)\delta_k^2/(1-r)^2$.
Apply the generator remainder bound on each edge and sum to obtain
the parameter statement. Dividing the resulting deterministic bound
by $s_N$ proves the final implication.
\end{proof}

To see why a converse fails, take $A_{k+1}=\tfrac32 A_k$,
$D_k=\tfrac12 A_k$, with any $A_0\ne\bm0_{\mathrm{mat}}$. Homogeneity gives
$u(A_{k+1})=u(A_k)$ and $\mathrm D u(A_k)[D_k]=0$.
Every prefix error is zero, while $\delta_k=1/2$ and $b(1/2)=7/2$.
An identity generator has zero generator remainder. Even a one-step
family therefore has error $o(s_N)$ for every positive scale sequence,
without requiring its positive budget to be $o(s_N)$.

The same proof permits fixed nonnegative normalized averaging, with
the average of the individual bounds on the right. On a clock with
$T_N=O(h_N^{-1})$, a uniform $\delta_{k,N}=O(h_N)$ gives an
$O(h_N)$ matrix-derivative error. A uniform
$\delta_{k,N}=O(\sqrt{h_N})$ provides only an $O(1)$ bound.
Neither power count controls the generator remainder. For random paths,
a sufficient condition for an $L^2$ version is that the right-hand error budget is
$o(s_N)$ in $L^2$; pointwise convergence alone does not transfer a
covariance limit. These are sufficient conditions, not converses.
The sums use realized matrix displacements and do not predict a successor
from incoming reduced coordinates. Their empirical role is to test the
precision of the differential description at an explicitly chosen scale.
Section~\ref{model:sec:row-path-results} reports the multistep errors; Appendix~\ref{model:app:row-quadrature} gives the separate realized-path derivative integration diagnostic.

The finite cross and quadratic terms have a direct scale interpretation. Since
$0\leq u(A)\leq1$ and $\|\Pi D\|_F\leq\|D\|_F$,
\[
 |q(A,D)|\leq\frac{\|D\|_F^2}{E'},\qquad
 |\ell(A,D)|\leq
 \frac{2(\sqrt{u(A)}+u(A))\sqrt E\,\|D\|_F}{E'}.
\]
With $E'/E$ bounded below uniformly over the time window, relative matrix increments
$\|D\|_F/\sqrt E=O(h_N)$ give $q=O(h_N^2)$.
Their accumulated quadratic contribution over $O(h_N^{-1})$ updates
vanishes. Increments of order $\sqrt{h_N}$ instead permit $q=O(h_N)$
and a finite accumulated contribution. This power counting does not
assume independent increments, a Gaussian limit or stationarity.
The signed cross-time terms remain in the chronological law.

For the averaged row coordinate, the physical-clock increment is
$(\bar\ell_{k,N}+\bar q_{k,N})/h_N$.
Convergence of the initial coordinate and of the two integrated flux
sums gives a row-flow limit by telescoping the exact map.
The optimizer force envelope does not by itself control these generated
matrix fluxes. In the matched family, block rates per physical unit
are $0.5376g$ for generator coordinates and $0.0768$ for the remaining
coordinates; the full model response between parameters and $A$
still depends on width, incoming state and the source block.
Sections~\ref{model:sec:collective-flux-results} and \ref{model:sec:row-path-results} detail the one-step and multistep measurements.

The finite fluxes are effective transition variables of the complete
conditioned law. Measuring them uses native predecessor and successor
fields. An economical autonomous row theory would additionally require
a successor law for these variables with the closure and emission
control of Proposition~\ref{model:prop:law-closure}. The exact energy map
and its finite measurements do not assume that extra closure.

\begin{proposition}[Finite-flux forecast error on a positive-energy domain]
\label{model:prop:finite-flux-forecast}
Let $\Pi$ be an orthogonal row-centering projector,
$E=\|A\|_F^2>0$, $E'=\|A+D\|_F^2>0$, and
$u=\|\Pi A\|_F^2/E$. Define the dimensionless transition coordinates
\[
 a=\frac{2\langle\Pi A,\Pi D\rangle_F-2u\langle A,D\rangle_F}{E},
 \qquad b=\frac{\|\Pi D\|_F^2-u\|D\|_F^2}{E},
 \qquad c=\frac{E'-E}{E}.
\]
Then $\Delta u=(a+b)/(1+c)$. Let $(\widehat a,\widehat b,\widehat c)$
be any forecast with $1+\widehat c>0$ and let
$\widehat f=(\widehat a+\widehat b)/(1+\widehat c)$.
If $1+c\ge\varepsilon>0$, then
\begin{equation}
 |\Delta u-\widehat f|
 \le \frac{|a-\widehat a|+|b-\widehat b|
                  +|\widehat f|\,|c-\widehat c|}{\varepsilon}.
 \label{model:eq:finite-flux-forecast}
\end{equation}
If additionally $1+\widehat c\ge\varepsilon$ and
$|\widehat a+\widehat b|\le M$, the right-hand side is at most
\[
 \varepsilon^{-1}(|a-\widehat a|+|b-\widehat b|)
 +M\varepsilon^{-2}|c-\widehat c|.
\]
Fixed nonnegative normalized averaging preserves the bound, with the
average of the individual right-hand sides.
\end{proposition}
\begin{proof}
Expand the centered and total squared norms, subtract $uE'$ from the
centered endpoint energy, and divide by $E'$. This gives
$\Delta u=(a+b)/(1+c)$. Subtraction of the forecast gives the exact identity
\[
 \Delta u-\widehat f
 =\frac{(a-\widehat a)+(b-\widehat b)
                     -(c-\widehat c)\widehat f}{1+c}.
\]
The triangle inequality and $1+c\ge\varepsilon>0$ prove the first bound.
The extra assumptions give $|\widehat f|\le M/\varepsilon$, proving the
second. For fixed weights, apply the triangle inequality to the weighted
sum and use their nonnegativity.
\end{proof}

A forecast must be computed from incoming information. Reconstructing
$(a,b,c)$ from an observed successor verifies the finite balance but does
not test forecastability. These three quantities are joint transition
variables; they are not, by themselves, an autonomous incoming state.
A predictive kernel must retain their conditional dependence, unresolved
memory and source-consumption law. Replacing that kernel by conditional
means can erase the fluctuations whose scaling is to be studied.

The bound controls the row component of a local closure defect. It does
not supply the error of other collective coordinates or of the predictive
emission. If a reduced transition is $L_k$-Lipschitz and its complete
local defect along the evolving fine law is bounded by $d_k$, the usual
coupled error recursion is $e_{k+1}\le L_ke_k+d_k$. Hence
\[
 e_m\le e_0\prod_{k<m}L_k+
       \sum_{j<m}d_j\prod_{j<k<m}L_k.
\]
Local measurements at observed incoming states must therefore be
separated from a free reduced rollout and its stability domain. Transfer
of a fluctuation law at scale $s_N$ additionally requires the relevant
centered coupled error to be $o(s_N)$, in $L^2$ when covariance is claimed,
and a predictive emission that retains the limiting mode.

\begin{proposition}[Chronological energy-coordinate map]
\label{model:prop:energy-cocycle}
Let $E_i=\|A_i\|_F^2>0$ and $C_i=\|\Pi A_i\|_F^2$ along a
finite path. For consecutive endpoints define
$r_{ij}=E_j/E_i>0$ and $q_{ij}=(C_j-C_i)/E_i$.
Then the row fraction transforms as
\[
 u_j=\frac{u_i+q_{ij}}{r_{ij}},\qquad
 (q_{ik},r_{ik})=(q_{ij}+r_{ij}q_{jk},r_{ij}r_{jk}).
\]
The product $(q,r)\star(\widetilde q,\widetilde r)
=(q+r\widetilde q,r\widetilde r)$ is associative, with identity $(0,1)$.
For an actual transition, the finite-flux coordinates obey
$r=1+c$ and $q=a+b+uc$.
\end{proposition}
\begin{proof}
Divide $C_j=C_i+E_iq_{ij}$ by $E_j=E_ir_{ij}$.
Substitution of $C_k-C_i=(C_j-C_i)+(C_k-C_j)$ and cancellation
of the positive intermediate energy give the displayed product.
Both bracketings of three factors have first coordinate
$q_1+r_1q_2+r_1r_2q_3$ and second coordinate $r_1r_2r_3$.
The identity follows by substitution. Finally, expansion of
$C_j-C_i=2\langle\Pi A_i,\Pi(A_j-A_i)\rangle_F+
\|\Pi(A_j-A_i)\|_F^2$ gives $q=a+b+uc$.
\end{proof}

These scale maps retain the realized joint transition law. They give an
exact two-coordinate transport representation without closing its
conditional law on $u$ alone. The physical domain of an individual action
is $r>0$ and $0\le u+q\le r$. Composition along an actual matrix path
preserves this domain. Forecasted coordinates require their own admission
and error control. Averaging $q$ and $r$ before taking their quotient does
not in general preserve the mean row fraction. Thus statistical
coarse graining must retain the relevant joint distribution or bound the
error caused by replacing it.

Unlike a quotient by centered energy, this representation admits $C_i=0$
whenever the total energy $E_i$ is positive. At a zero total matrix the
row fraction is undefined and the unnormalized energies must be retained.
The product algebra is a specialization of chronological affine
composition; its role here is to identify the finite row-fraction
coordinates and their joint-law and error requirements.

\subsection{Signed temporal energy and compatible scale summaries}
\label{model:sec:temporal-energy}
For a chronological generated-matrix path, blocking combines displacements
before computing their squared norm. The following identity makes the
resulting temporal contribution explicit without assuming a stationary
source, independent increments or a small update.

\begin{proposition}[Finite temporal-energy accounting]
\label{model:prop:temporal-energy}
Let $A_i,\ldots,A_j$ belong to a real inner-product space, let
$m=j-i\geq1$, and suppose $E_i=\|A_i\|^2>0$. Put $D_k=A_{k+1}-A_k$ and
\begin{equation}
 S_{ij}=\frac{\sum_{k=i}^{j-1}\|D_k\|^2}{E_i},\qquad
 B_{ij}=\frac{\|A_j-A_i\|^2}{E_i},\qquad
 X_{ij}=\frac{2\sum_{i\leq k<\ell<j}\langle D_k,D_\ell\rangle}{E_i}.
 \label{model:eq:temporal-energy}
\end{equation}
Then $B_{ij}=S_{ij}+X_{ij}$, $S_{ij}\geq0$, and
\begin{equation}
 -S_{ij}\leq X_{ij}\leq(m-1)S_{ij}.
 \label{model:eq:temporal-bounds}
\end{equation}
Unnormalized block summaries $(V,Q,H)$, with
$V=\sum_kD_k$, $Q=\sum_k\|D_k\|^2$ and
$H=2\sum_{k<\ell}\langle D_k,D_\ell\rangle$, combine by
\begin{equation}
 (V,Q,H)\odot(\widetilde V,\widetilde Q,\widetilde H)
 =(V+\widetilde V,Q+\widetilde Q,
 H+\widetilde H+2\langle V,\widetilde V\rangle).
 \label{model:eq:temporal-merge}
\end{equation}
This operation is associative. Normalization of the merged block uses
its incoming energy; separately normalized summaries cannot simply be added.
\end{proposition}
\begin{proof}
Telescoping gives $A_j-A_i=\sum_{k=i}^{j-1}D_k$. Expanding its squared
norm separates the diagonal from the ordered off-diagonal pairs and gives
$B=S+X$. Nonnegativity of squared norms proves $S\geq0$ and $X=B-S\geq-S$.
The triangle inequality followed by scalar Cauchy--Schwarz gives
$\|\sum_kD_k\|^2\leq(\sum_k\|D_k\|)^2\leq m\sum_k\|D_k\|^2$.
Division by $E_i>0$ proves the upper bound. Splitting the ordered pairs
into those within each block and those between blocks gives
\eqref{model:eq:temporal-merge}. Both parenthesizations of three blocks produce
the sum of their three $H$ coordinates plus twice the three pairwise
inner products of their $V$ coordinates. Vector and scalar addition are
associative, so all coordinates agree. The normalization statement follows
directly from the definition of $E_i$.
\end{proof}

The theorem applies to Frobenius matrices, including their row-centered
projections. Positive or negative $X$ is compatible with a positive
semidefinite increment Gram matrix. Coherent drift contributes to $X$,
so its sign alone does not diagnose a centered forcing covariance or a
critical mode. The summary is unchanged by permuting a fixed list of
increments. It is consequently a diagnostic attached to the chronological
path, not a sufficient incoming state for its future law.

There are four distinct quantities in this description. For a single row transition, write
$L=2\langle\Pi A,\Pi D\rangle_F-2u\langle A,D\rangle_F$ and
$Q_{\rm row}=\|\Pi D\|_F^2-u\|D\|_F^2$. The exact row change is $(L+Q_{\rm row})/E'$,
the finite row cross contribution is $L/E'$, and the matrix directional
derivative is $L/E$. A parameter derivative additionally replaces $D$
by the linearized generator displacement. None is the cross-time energy
$X$ in \eqref{model:eq:temporal-energy}. Positive total energy suffices for
the exact identities; differential approximation requires the additional
relative-displacement assumptions in Corollary~\ref{model:cor:row-path-error}.
Section~\ref{model:sec:equal-time-flux} measures these finite row and cross-time contributions at matched physical durations.

\FloatBarrier
\par\medskip\noindent
Chapter~\ref{ch:single-pass-evidence} reports single-pass training and
potential observations under specified data and schedule laws. These results
give concrete settings for the source and clock distinctions developed here.

\chapter{Single-pass training and potential observations}
\label{ch:single-pass-evidence}
This chapter presents the primary single-pass experiments and their
predictive observations. It compares data laws, learning schedules, prefix
risks, row-map input dependence and potential activity, while distinguishing
fitting, operator concentration and inference fidelity.

\section{What the completed experiments establish}
\label{model:sec:primary-results}

\subsection{Single-pass training selects the inference-visible regime}
The primary corpus consists of 524,288 distinct document-content hashes
from sixteen RefinedWeb strata. Its 513-token crops give 4,194,304
nonrepeated 64-target-position blocks. Batch 32 permits 131,072 updates;
execution refuses exhaustion instead of wrapping the corpus. Selected
training and observation documents have no exact content-hash overlap.
The theory conditions on this realized source law, tokenizer, target
construction, source ordering and optimizer. It does not recover the
unrecorded pretraining stream of released checkpoints.

The source/schedule comparison family contains 54 single-pass paths and 2,555,904
updates, with 284 checkpoint observations. Twelve paths use constant
rates, forty complete compact cosine/floor cycles, and two follow longer
source-recipe schedule prefixes. All use five decoders and 64-dimensional
heads; width, objective and parameter-group rates are specified with each
family in Section~\ref{model:sec:onepass-results}. Sixteen matched
constant/cosine checkpoint pairs favor cosine training in external-target
risk while retaining greater row contrast. This finite relation between
drive, row geometry and prediction does not identify a critical surface.
Constant-rate paths can lose predictive quality late in training even
when no source position is reused. The auxiliary repeated-corpus
comparison changes corpus content and size as well as presentation,
so its forty paired checkpoint outcomes do not isolate repetition alone.

A row-concentrated operator still has a mobile common sector. Native
proper-prefix interventions and downstream vocabulary measurements test
that sector separately from transverse row energy
(Sections~\ref{model:sec:onepass-results} and \ref{model:sec:inference-mechanism}).
The update-resolved potential and actual training interventions in
Section~\ref{model:sec:potential-results} supply a direct test of the chain
$(A,M,P,V,G)\longrightarrow p_\theta(\cdot\mid x)$ under single-pass
training. They retain all heads and decoders in the stated observations.

\subsection{Optimizer memory and finite predictive reduction}
A separate paired optimizer experiment fixes initial parameters and
perturbs first or second moments at two widths and two remaining-source
sequences. All incoming logits agree. After one native update the
absolute antisymmetric risk response spans
$2.86\times10^{-5}$--$8.03\times10^{-4}$ nats across its eight
width/source/moment cells. Complete first-step parameter and moment
predictions satisfy the declared arithmetic criterion. This supports
retaining optimizer state in the model-wide kernel. Finite pulse and
arithmetic-control results, with all amplitudes and costs, are in
Section~\ref{model:sec:optimizer-results}.

Exact joint-law composition is compatible with substantial error in a
chosen finite dictionary. The four frozen-coordinate forecasts do not
meet their uniform 0.01-nat mean-risk criterion on the stated transfer
domain. Same-state conditional moment predictions have a narrower tested
scope, and their acquisition and refresh costs remain explicit.
The fixed-dictionary projection identity separates irreducible
representation error from coefficient-transport error; simply refitting
coefficients cannot remove the former. The complete conditional-memory
law in Proposition~\ref{model:prop:conditional-memory} retains the unresolved
state distribution and remains exactly closed. These findings determine
which mathematical object is empirically supported; they do not turn an
expensive complete law into an autonomous economical surrogate.
Sections~\ref{model:sec:law-closure-results}, \ref{model:sec:memory-results},
\ref{model:sec:moment-results} and \ref{model:sec:finetuning-results} retain every
forecast, source intervention, uncertainty unit and cost comparison.

\subsection{Inference-law transport and its conditioning}
Two matched single-pass adaptations of the validation-selected released
base use the same 32,768 technical proper-prefix examples. Full-parameter
validation selects the incoming state. Rank-four validation selects the
completed one-pass state; the technical test NLL change over the complete
prefix panel is $-0.00227$ nats, with the stated family-adjusted paired
interval $[-0.00381,-0.00076]$. Other domains do not establish an
improvement under that uncertainty analysis. The factor coordinates and
their optimizer state are needed during training even though their merged
native weights define inference (Section~\ref{model:sec:singlepass-released}).

Known Ising and Potts source laws provide a separate calibration of
architectural visibility, magnetic readout and generated-law fidelity.
The completed physical adaptations improve aggregate held-out next-spin
loss and accuracy, with class- and seed-dependent size transfer. Exact
small-configuration laws and 138 generated spatial cells separate mean
profiles, connected covariance and thermal contrasts. Finite task gains
coexist with an almost input-invariant metric generator. These physical
source and finite-corpus adaptation laws are distinct from RefinedWeb
pretraining. They test the conditional inference-transfer bounds, not
native PLDR critical exponents. Detailed theory and every physical
comparison are in Sections~\ref{model:sec:physical-calibration} and
\ref{model:sec:readout-budget}, with their associated measurement sections.
Sections~\ref{model:sec:physical-results} and \ref{model:sec:readout-results} give those physical calibration and readout measurements.

\section{Single-pass training and the selected inference law}
\label{model:sec:onepass-results}

The primary controlled training law consumes each selected source
position once. This makes corpus size, the consumed fraction and the
remaining empirical distribution explicit coordinates of the theory
in Section~\ref{model:sec:data-resource}. The completed primary design
contains 54 training paths with 2,555,904 updates: twelve constant-rate
paths, forty complete compact cosine/floor paths, and two longer
source-recipe schedule prefixes. It retains 284 checkpoint states,
94 task/generation states, forty paired data-law checkpoint comparisons,
and sixteen paired constant/cosine checkpoint comparisons. Complete
proper-prefix risks, local row derivatives, decoder interventions and
192 conditional native optimizer steps test the relevant reduction
mechanisms. Their designs and denominators are stated in
Sections~\ref{model:sec:onepass-causal-drives},
\ref{model:sec:onepass-row-input-results},
\ref{model:sec:onepass-decoder-results} and
\ref{model:sec:onepass-native-risk}. The
numerical conclusions concern these executed paths and independently
reconstructed observations, conditional on the fixed corpus, stream
and shared-generator initialization.

\subsection{Corpus, native adaptation and matched data laws}
\label{model:sec:onepass-corpus-methods}

The single-pass corpus contains 524,288 distinct document content
hashes from sixteen recorded RefinedWeb source strata. Its selection
excludes all 4,608 documents in the smaller training/evaluation pool.
A uniformly selected 513-token crop from each qualified document gives
eight consecutive blocks with 64 target positions per block. There
are 4,194,304 blocks. Each run uses a prefix of one uniform permutation
of these block indices, with 32 blocks per update. The resulting
capacity is 131,072 updates, or 268,435,456 supervised source positions
for the all-target objective. Native execution rejects an exhausted
stream instead of wrapping it. Exact document-content deduplication
does not assert semantic deduplication. Common words and token values
occur naturally; only the selected source positions are nonrepeated.
The complete document selection, tokenization, crop construction and
block stream were independently reconstructed from the read-only
corpus. A separate complete hash intersection check finds no exact
document overlap with the 2,048-document short or 1,024-document
long observation cohorts. This establishes separation of these
finite document sets, not absence of semantically similar text.

The matched paths use five native decoders, head dimension 64,
$N=14$ heads and eight shared metric residual units of width 170.
Their batch size is 32 and context length is 64, with native float32
training and TF32 disabled. Both source recipes use AdamW with
$(\beta_1,\beta_2)=(0.9,0.95)$, epsilon $10^{-5}$,
weight decay $0.1$, coordinatewise gradient clipping at one, and
all nonpadding targets. Source Near recipe~1 uses peak rate
$0.0012$ and 2,000 linear warmup updates. Source Below recipe~1
uses peak rate $0.0006$ and 6,000 warmup updates. Both parameter
groups receive the recipe rate. Cosine annealing reaches ten percent
of the peak at update 250,000. Each compared path stops deliberately
at 65,536 updates, before that cosine endpoint. These are matched
schedule prefixes, not observations of a completed annealing cycle.
The source names identify settings from the reference study
\cite{gokden2026soc}; they do not assign a thermodynamic phase to
this smaller-context adaptation. The reference pretraining used
context length 1,024 and a different arithmetic and distributed-training
configuration; its historical exposure is not reconstructed by these
controlled paths.

For each recipe, the single-pass and repeated-corpus paths share
byte-identical initial parameters, optimizer settings and the same
applied learning-rate function. The repeated law resamples crops
from 3,072 fixed documents. The single-pass law uses the larger
corpus and disjoint blocks just described. The comparison changes
both the empirical corpus and its presentation rule, so it does not
isolate repetition frequency alone. One initialization is compared
per recipe at updates 32,768 and 65,536. No initialization-ensemble
uncertainty is estimated from this one pair.

Held-out risk uses the same 512 fixed 64-token contexts and targets
under both training laws. Training risk uses 1,024 recorded crops
from the repeated training law, or a fixed sample of 1,024 already
consumed blocks at each single-pass checkpoint. The training cohorts
therefore have different membership, and the single-pass seen-block
cohort changes with time. Their contrast describes fit under the
respective laws; the identical held-out cohort is the primary
prediction comparison. Full vocabulary logits and their native
cross-entropy reductions are retained and independently replayed.

\subsection{Measured supervised-position reuse}

Table~\ref{model:tab:onepass-source-reuse} counts exact document/offset
positions in the recorded training streams. All nine selected prefix
pairs at $N=14$ and initialization 640101 were reconstructed: five
constant-objective horizons and two horizons for each source recipe.
The two source recipes have identical sampling prefixes, so their
exposure counts form two groups rather than four independent
replications. Integer counts reproduce all eighteen saved native
vocabulary target histograms exactly. The reconstruction adds no
training updates.

\begin{table}[htbp]
\centering\small
\caption{Supervised-position exposure in completed prefixes. Unique
means the number $V$ of exact source positions supervised at least
once. Reused is $100(H-V)/H$, the percentage of target events after
the first exposure at that position. ``Last'' supervises one target
per crop; ``all'' supervises all 64 nonpadding targets. Shared
sampling prefixes do not create independent data replications.}
\label{model:tab:onepass-source-reuse}
\begin{tabular}{lrrrrr}
\toprule
Targets & Updates & Repeated: unique & Reused (\%) & Single pass: unique & Reused (\%)\\
\midrule
Last & 8192 & 238,746 & 8.93 & 262,144 & 0\\
Last & 32768 & 734,243 & 29.98 & 1,048,576 & 0\\
Last & 65536 & 1,077,695 & 48.61 & 2,097,152 & 0\\
Last & 98304 & 1,237,826 & 60.65 & 3,145,728 & 0\\
Last & 131072 & 1,313,177 & 68.69 & 4,194,304 & 0\\
All & 32768 & 1,567,290 & 97.66 & 67,108,864 & 0\\
All & 65536 & 1,571,128 & 98.83 & 134,217,728 & 0\\
\bottomrule
\end{tabular}
\end{table}

At 65,536 updates, the all-target reference recipes have 134,217,728
target events under either law. Repeated sampling supervises only
1,571,128 distinct positions, and 98.83\% of events reuse a position.
The largest exposure of one position is 145. In contrast, every
single-pass target event has a distinct source position. The
constant-objective control at the same update count has only
2,097,152 target events. Under repeated sampling it supervises
1,077,695 distinct positions, giving 48.61\% repeated events and
maximum exposure ten. Even its 131,072-update endpoint has 68.69\%
repeated target events, below the reference recipes' earlier
all-target exposure fraction.

Input-context reuse follows the same crop stream for both objectives
under each data law. In the unmasked repeated corpus, shifting each
input position forward by one gives a bijection to the all-target
positions of the same crops. Their occurrence counts therefore agree,
so 98.83\% of input-position events are repeated at update 65,536
for both objectives. The separate supervised-target count reflects
the loss function. Thus equal update counts do not mean equal target
budgets or equal supervised-target exposure to repetition. Overlapping crops matter: the repeated
all-target law has 1,077,695 distinct crops at update 65,536, but
those crops draw their targets from almost the same finite set of
positions. Proposition~\ref{model:prop:source-position-occupancy} gives
the corresponding exact occupancy expectation under the sampling
law. These counts quantify the two executed presentations. They do
not convert correlated training events into independent samples,
or isolate repetition frequency from the simultaneous change in
corpus content and size.

\subsection{Constant-rate comparison under a single pass}
\label{model:sec:onepass-constant-data-law}

The completed constant-rate control compares both data laws at
$N=4,8,14$, with four paired initializations at 8,192 and 32,768
updates. Two initializations at each of $N=4,14$ additionally reach
65,536, 98,304 and 131,072 updates. This gives twelve paired paths
and 36 saved checkpoint pairs, conditional on the shared-generator initialization
and fixed data stream. Repeated checkpoints from a path are not
independent training replications. This control supervises the external
target only, uses generator rate $0.0003$ and body rate $0.0006/N$,
AdamW coefficients $(0.9,0.95)$, epsilon $10^{-8}$, decay $0.01$
and norm clipping at one. Its objective and optimizer therefore differ
from the all-target source recipes.

Single-pass held-out NLL is lower in 21 of the 36 checkpoint pairs
and higher in fifteen. Six of the twelve head-count/horizon mean
contrasts favor each law. These descriptive counts directly exclude
a uniform single-pass improvement on the selected panel; they are
not significance tests. For example, at $N=8$ and 32,768 updates,
the single-pass minus repeated mean difference is $0.0811$ nats,
with exact empirical paired-resampling percentiles
$[0.00156,0.1550]$. At $N=14$ and 8,192 updates it is
$-0.0579$ nats, with percentiles $[-0.0733,-0.0447]$.
Both contrasts use four paired initializations. Their intervals
describe empirical resampling, with the qualifications in
Section~\ref{model:sec:statistics}.

Every long path has higher held-out NLL at 131,072 than at 32,768
updates under both data laws. For the single-pass paths, the increases
are $0.9900$ and $1.0121$ nats at $N=4$, and $0.2379$ and $0.3052$
at $N=14$. At the complete-pass endpoint, mean single-pass NLL is
$8.5225$ at $N=4$ and $7.5736$ at $N=14$, compared with $8.5799$
and $7.6133$ under repeated sampling. Thus both endpoint means improve
modestly under a single pass while the late loss increase persists.
Mean context KL at these single-pass endpoints is
$6.89\times10^{-10}$ and $5.66\times10^{-4}$, respectively.
The larger nonrepeated corpus does not by itself preserve a useful
context-dependent predictive law under this constant-rate objective.

\begin{table}[htbp]
\centering\small
\caption{All single-pass constant-rate collective and risk conditions.
The seeds column gives the actual initialization count. Mean row contrast
uses the native residual metric $A$, and $\chi_C/N$ is the
across-initialization variance of its common centroid. Training risk
uses the recorded sample of already consumed blocks; held-out risk
uses the same fixed 512 contexts throughout. Both risks score the
external target in nats.}
\label{model:tab:onepass-constant-summary}
\begin{tabular}{rrrrrrr}
\toprule
$N$ & Updates & Seeds & $\overline R$ & $\chi_C/N$ & Train last & Held last\\
\midrule
4 & 8192 & 4 & 0.273 & 0.638 & 7.24 & 7.41\\
4 & 32768 & 4 & $4.62\!\times\!10^{-5}$ & 0.94 & 7.47 & 7.48\\
4 & 65536 & 2 & $7.06\!\times\!10^{-5}$ & 0.786 & 7.48 & 7.5\\
4 & 98304 & 2 & 0.0113 & 0.636 & 7.94 & 8.09\\
4 & 131072 & 2 & 0.0803 & 0.509 & 8.94 & 8.52\\
8 & 8192 & 4 & 0.326 & 0.609 & 7.13 & 7.31\\
8 & 32768 & 4 & 0.00784 & 0.9 & 7.48 & 7.5\\
14 & 8192 & 4 & 0.336 & 0.597 & 7.06 & 7.26\\
14 & 32768 & 4 & $3.22\!\times\!10^{-5}$ & 0.916 & 7.28 & 7.31\\
14 & 65536 & 2 & $8.33\!\times\!10^{-7}$ & 0.719 & 7.36 & 7.41\\
14 & 98304 & 2 & $5.29\!\times\!10^{-6}$ & 0.601 & 7.41 & 7.55\\
14 & 131072 & 2 & $6.97\!\times\!10^{-5}$ & 0.549 & 7.79 & 7.57\\
\bottomrule
\end{tabular}
\end{table}

This complete comparison separates data-law conditioning from a
claim that repetition explains every late loss increase. The latter
increase also occurs when the path never reuses a supervised source
position. The risk-drift and finite source-corner identities retain
the optimizer's direction, body contribution and mixed response,
which are needed even after source reuse is removed. These results
motivate interpreting the single-pass regimes on their own terms;
the all-target source-recipe comparison in
Section~\ref{model:sec:onepass-matched-data-law} supplies a distinct
response to the changed data law.

\subsection{Completed cosine cycles across heads and initializations}
\label{model:sec:onepass-compact-family}

The compact single-pass family contains forty training paths, each
with a cosine endpoint at update 32,768 and an additional 8,192
updates at ten percent of its peak rate. Table~\ref{model:tab:onepass-design}
gives all five settings. Each head-count/recipe condition has four
nonshared initialization identities, with the shared metric-learner
initialization and the entire block stream fixed. Near~1 and Below~1
each use $N=4,8,14$; Near~2 and Below~2 each use $N=4$; the
schedule-only controlled family uses $N=4,14$. Every path has the
same five-decoder, head-dimension-64 architecture convention,
batch size 32, context length 64 and float32 arithmetic described
above. Source names identify optimizer settings, not inferred phases.

\begin{table}[htbp]
\centering\small
\caption{The complete compact schedule family. The body peaks are
listed in the same head-count order as the second column. Every
head-count/recipe condition has four initialization identities and
40,960 executed updates. Source recipes supervise all nonpadding
targets and use the source AdamW configuration. The controlled
recipe retains its external-target objective and optimizer and changes
only the imposed rate history.}
\label{model:tab:onepass-design}
\begin{tabular}{llrrrl}
\toprule
Recipe & $N$ & Paths & Warmup & $\eta_G$ peak & $\eta_B$ peaks\\
\midrule
Controlled & 4,14 & 8 & 2000 & 0.0003 & 0.00015, $4.29\!\times\!10^{-5}$\\
Near 1 & 4,8,14 & 12 & 2000 & 0.0012 & 0.0012, 0.0012, 0.0012\\
Below 1 & 4,8,14 & 12 & 6000 & 0.0006 & 0.0006, 0.0006, 0.0006\\
Near 2 & 4 & 4 & 1000 & 0.001 & 0.001\\
Below 2 & 4 & 4 & 2000 & 0.0003 & 0.0003\\
\bottomrule
\end{tabular}
\end{table}

The six saved states per compact path are its warmup endpoint,
8,192, 16,384, 24,576, 32,768 and 40,960 updates. Thus the
compact family contributes 240 checkpoint states. Together with
eight states from the two source-horizon prefixes and 36 states from
the twelve constant-rate paths, the primary single-pass analysis has
284 states and eighty head-count/recipe/drive/time conditions.
Native float32 and float64 forward observations give 160 separately
retained collective conditions. They are two implemented observations
of fixed trained weights, rather than two training ensembles.

\subsection{Row contrasts, common fields and predictive risk}

The row statistic $R$ is the normalized centered energy of the
residual learner output $A$, before its positive PLGA transform.
The common centroid $C$ averages the metric-row centroids over heads.
Its susceptibility is $N$ times the unbiased initialization variance
at each fixed context and decoder, averaged in mean-square-column
units. Hence $\chi_C/N$ is the measured common-coordinate variance.
The effective head count is the average one-head variance divided by
the variance of the head mean. It describes the declared covariance
sector; it is not a number of independent training replications.

\begin{figure}[htbp]
\centering
\includegraphics[width=\textwidth]{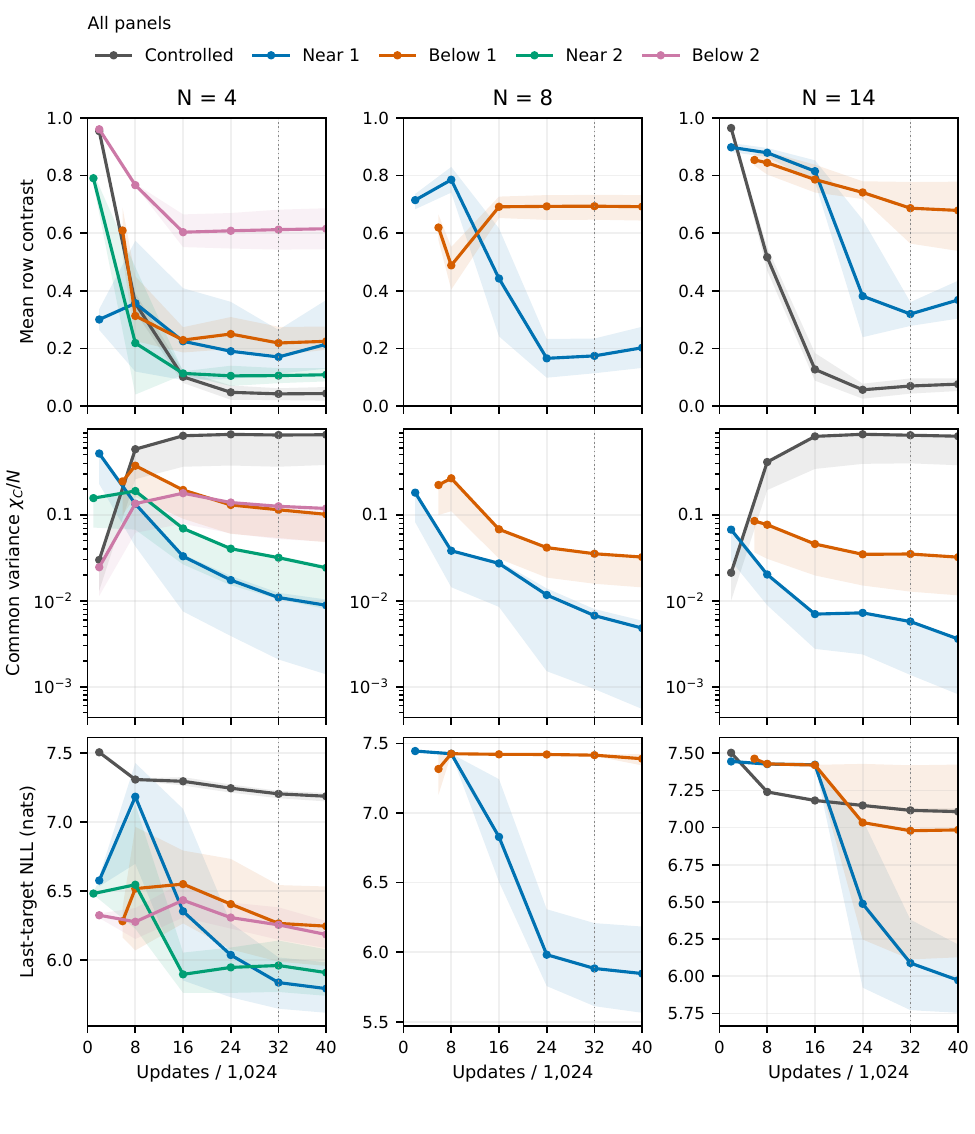}
\caption{The complete compact single-pass family. Columns specify
head count; rows show row contrast, common centroid variance and
external-target held-out NLL. Row contrast uses a linear
ordinate from zero to one. Common variance uses a logarithmic
ordinate with shared limits; every displayed mean and resampling
bound is strictly positive. Shading gives empirical initialization
resampling percentiles without asserted population or simultaneous
coverage. The dotted line marks the imposed cosine endpoint.}
\label{model:fig:onepass-compact-trajectories}
\end{figure}

Tables~\ref{model:tab:onepass-compact-collectives} and
\ref{model:tab:onepass-compact-risk} retain all twenty endpoint conditions.
Figure~\ref{model:fig:onepass-compact-trajectories} also shows every saved
compact time. Its lines join checkpoint means and impose no temporal
power fit. The empirical bands enumerate all 256 ordered resamples
of the four initialization identities under the fixed corpus and stream.

\begin{table}[htbp]
\centering\small
\caption{All compact endpoint collective conditions in native
float32. Each row uses four initialization identities. Absolute row
energy retains the numerator of the normalized contrast, in fixed
mean-square-entry units. Common variance and effective head count
refer to across-initialization covariance at fixed evaluation contexts.}
\label{model:tab:onepass-compact-collectives}
\begin{tabular}{lrrrrrr}
\toprule
Recipe & $N$ & Updates & $\overline R$ & Row energy & $\chi_C/N$ & $N_{\rm eff}$\\
\midrule
Controlled & 4 & 32768 & 0.0429 & 0.0382 & 0.85 & 1.01\\
Controlled & 4 & 40960 & 0.0441 & 0.0392 & 0.856 & 1.01\\
Controlled & 14 & 32768 & 0.0701 & 0.0628 & 0.846 & 1.02\\
Controlled & 14 & 40960 & 0.0767 & 0.0686 & 0.823 & 1.02\\
Near 1 & 4 & 32768 & 0.171 & 0.00463 & 0.0109 & 1.02\\
Near 1 & 4 & 40960 & 0.215 & 0.00445 & 0.00888 & 1.02\\
Near 1 & 8 & 32768 & 0.175 & 0.003 & 0.00673 & 1.01\\
Near 1 & 8 & 40960 & 0.203 & 0.00268 & 0.00481 & 1.01\\
Near 1 & 14 & 32768 & 0.32 & 0.00304 & 0.00574 & 1.02\\
Near 1 & 14 & 40960 & 0.369 & 0.00269 & 0.0036 & 1.02\\
Below 1 & 4 & 32768 & 0.219 & 0.0339 & 0.115 & 1.01\\
Below 1 & 4 & 40960 & 0.225 & 0.0315 & 0.101 & 1.01\\
Below 1 & 8 & 32768 & 0.694 & 0.0839 & 0.0353 & 1.04\\
Below 1 & 8 & 40960 & 0.692 & 0.0758 & 0.0323 & 1.04\\
Below 1 & 14 & 32768 & 0.687 & 0.0803 & 0.0351 & 1.05\\
Below 1 & 14 & 40960 & 0.679 & 0.0717 & 0.0322 & 1.05\\
Near 2 & 4 & 32768 & 0.106 & 0.00547 & 0.0318 & 1.01\\
Near 2 & 4 & 40960 & 0.109 & 0.00494 & 0.0242 & 1.01\\
Below 2 & 4 & 32768 & 0.613 & 0.209 & 0.126 & 1.04\\
Below 2 & 4 & 40960 & 0.616 & 0.2 & 0.119 & 1.03\\
\bottomrule
\end{tabular}
\end{table}

\begin{table}[htbp]
\centering\small
\caption{All compact endpoint predictive conditions, averaged over
their four selected initializations. Train and held-out ``all'' score
the parallel full-window targets; ``last'' scores the external target
on 512 held-out contexts. The proper panel uses a separate fixed set
of eight contexts at prefix lengths 16, 32 and 48. Its mean is a
different estimand from the complete 32-context, 64-position causal
panel of Section~\ref{model:sec:onepass-causal-drives}.}
\label{model:tab:onepass-compact-risk}
\begin{tabular}{lrrrrrr}
\toprule
Recipe & $N$ & Updates & Train all & Held all & Held last & Proper panel\\
\midrule
Controlled & 4 & 32768 & 7.28 & 7.32 & 7.2 & 7.52\\
Controlled & 4 & 40960 & 7.27 & 7.3 & 7.19 & 7.45\\
Controlled & 14 & 32768 & 7.36 & 7.4 & 7.12 & 7.54\\
Controlled & 14 & 40960 & 7.36 & 7.4 & 7.11 & 7.48\\
Near 1 & 4 & 32768 & 4.47 & 4.52 & 5.84 & 11.7\\
Near 1 & 4 & 40960 & 4.16 & 4.24 & 5.79 & 13.8\\
Near 1 & 8 & 32768 & 3.47 & 3.51 & 5.88 & 13.7\\
Near 1 & 8 & 40960 & 3.19 & 3.25 & 5.85 & 13.7\\
Near 1 & 14 & 32768 & 4.22 & 4.26 & 6.09 & 14.1\\
Near 1 & 14 & 40960 & 3.61 & 3.68 & 5.97 & 14.9\\
Below 1 & 4 & 32768 & 4.47 & 4.52 & 6.27 & 17.8\\
Below 1 & 4 & 40960 & 4.2 & 4.27 & 6.25 & 19.2\\
Below 1 & 8 & 32768 & 7.41 & 7.42 & 7.42 & 7.67\\
Below 1 & 8 & 40960 & 7.36 & 7.39 & 7.39 & 7.64\\
Below 1 & 14 & 32768 & 6.52 & 6.55 & 6.98 & 9.28\\
Below 1 & 14 & 40960 & 6.34 & 6.38 & 6.98 & 9.36\\
Near 2 & 4 & 32768 & 4.36 & 4.42 & 5.96 & 12.7\\
Near 2 & 4 & 40960 & 4.15 & 4.22 & 5.91 & 14.5\\
Below 2 & 4 & 32768 & 3.88 & 3.93 & 6.26 & 16\\
Below 2 & 4 & 40960 & 3.7 & 3.77 & 6.19 & 17.1\\
\bottomrule
\end{tabular}
\end{table}

The endpoint means span $0.0429\leq\overline R\leq0.694$ and
$1.007\leq N_{\rm eff}\leq1.047$. The common sector remains strongly
correlated across heads throughout this panel. For example, the
controlled cosine endpoint has $\chi_C/N=0.850$ at $N=4$ and
$0.846$ at $N=14$. Multiplying these almost equal intensive
variances by $N$ produces an increasing susceptibility without
identifying a critical divergence. Near~1 instead has smaller common
variances, $0.0109$, $0.00673$ and $0.00574$ at $N=4,8,14$.
These are finite, drive-conditioned covariance measurements on three
head counts and four initializations, with a fixed shared generator.
They do not determine a universality class.

During the positive floor, mean absolute row energy decreases in all
eight source-recipe/head-count conditions, whereas normalized row
contrast increases in six of them. Both quantities increase in the
two controlled conditions. In particular, the Near~1 $N=14$ mean
contrast rises from $0.320$ to $0.369$ while its absolute row energy
falls from $0.00304$ to $0.00269$. Retaining the absolute numerator
and the common field is therefore necessary when interpreting a
normalized collapse statistic. The ensemble endpoint means also
differ substantially from the strongly concentrated source-horizon Near endpoint examined in
Section~\ref{model:sec:onepass-prefix-results}.

Prediction depends on the observation law. Near~1 has lower
external-target loss than Below~1 at all six matched compact endpoint
conditions. At $N=8$, for example, their cosine-endpoint mean losses
are 5.884 and 7.416 nats. Yet the corresponding eight-context proper
panel losses are 13.677 and 7.672 nats. Improved parallel full-window
fit does not guarantee transfer to shorter causal prefixes. The
Near~1 $N=4$ external-target ordering also reverses with time: its
mean loss exceeds Below~1 by 0.664 nats at 8,192 updates and is
lower by 0.197 nats at 16,384. All 34 matched Near/Below recipe
contrasts, including the cross-recipe $N=4$ comparisons, retain
this time dependence in the numerical record. Unequal warmup
endpoints are not interpolated into a matched comparison.

The floor reduces mean external-target loss in nine of the ten
conditions; the remaining Below~1 $N=14$ change is $+0.00548$
nats. It also reduces full-window held-out loss in nine conditions,
with a $+0.000451$-nat change in the controlled $N=14$ condition.
However, proper-panel risk increases in seven conditions, including
all four Near conditions. These comparisons use the same selected
cohort within each observation law. They show that extra updates at
a small positive rate do not uniformly repair the prefix discrepancy.
The complete 32-context causal measurements in
Section~\ref{model:sec:onepass-causal-drives} test that discrepancy with all target positions rather than the three-length panel.

Initialization variation is retained in these means. At the Below~1
$N=14$ floor endpoint, initialization 640101 has full-window held-out
risk 3.274 nats and proper-panel risk 14.368 nats. The other three
initializations have full-window risks between 7.382 and 7.424 and
proper-panel risks between 7.689 and 7.701. Thus a condition mean
can combine different finite training outcomes. Heads or evaluation
contexts cannot replace additional independent initialization identities
when assessing this variation.

\subsection{Paired constant and cosine rate histories}
\label{model:sec:onepass-paired-schedules}

The schedule-only comparison pairs the controlled constant-rate and
compact cosine paths at $N=4,14$ and updates 8,192 and 32,768.
All four initializations enter each condition. This gives sixteen
checkpoint pairs from eight trajectory pairs. Each pair has identical
initial parameters, corpus, block ordering, training objective,
clipping rule, Adam coefficients, peak rates and fixed evaluation
cohorts. The controlled cosine path uses 2,000 warmup updates and
the 32,768-update horizon; the constant path applies its peak rates
from its first update. The first cosine update still advances the
Adam state at zero parameter rate. The two histories therefore retain
different optimizer states even when compared at the same update count.

\begin{table}[htbp]
\centering\small
\caption{All paired schedule-only external-target risk conditions,
in nats on the same 512 held-out contexts. Differences are cosine
minus constant. Each empirical range enumerates the 256 ordered
whole-initialization paired resamples. The final column counts the
four paired identities with a negative measured difference; it does
not give a significance-test probability.}
\label{model:tab:onepass-schedule-risk}
\begin{tabular}{rrrrrrr}
\toprule
$N$ & Updates & Constant & Cosine & $\Delta L$ & Empirical range & $\Delta L<0$\\
\midrule
4 & 8192 & 7.40538 & 7.30651 & -0.09887 & [-0.13075, -0.06698] & 4/4\\
4 & 32768 & 7.48017 & 7.20300 & -0.27717 & [-0.35672, -0.15669] & 4/4\\
14 & 8192 & 7.25692 & 7.23975 & -0.01717 & [-0.02753, -0.00680] & 4/4\\
14 & 32768 & 7.31367 & 7.11578 & -0.19789 & [-0.20749, -0.19052] & 4/4\\
\bottomrule
\end{tabular}
\end{table}

\begin{table}[htbp]
\centering\small
\caption{All paired schedule collective conditions in both
implemented forward arithmetics. Differences are cosine minus
constant. $E_{\rm row}$ is absolute centered row energy;
$\widehat I_{\rm ctx}$ is the fixed paired-context predictive KL.
The complete paired mean and susceptibility resampling distributions
are retained with the numerical results.}
\label{model:tab:onepass-schedule-collectives}
\begin{tabular}{rrrrrrr}
\toprule
$N$ & Updates & Bits & $\Delta\langle R\rangle$ & $\Delta\langle E_{\mathrm{row}}\rangle$ & $\Delta\chi_C/N$ & $\Delta\widehat I_{\mathrm{ctx}}$\\
\midrule
4 & 8192 & 32 & $8.36\times10^{-2}$ & $8.08\times10^{-2}$ & $-5.72\times10^{-2}$ & $0.159$\\
4 & 8192 & 64 & $8.36\times10^{-2}$ & $8.08\times10^{-2}$ & $-5.72\times10^{-2}$ & $0.159$\\
4 & 32768 & 32 & $4.28\times10^{-2}$ & $3.82\times10^{-2}$ & $-9.02\times10^{-2}$ & $0.326$\\
4 & 32768 & 64 & $4.28\times10^{-2}$ & $3.82\times10^{-2}$ & $-9.02\times10^{-2}$ & $0.326$\\
14 & 8192 & 32 & $0.181$ & $0.179$ & $-0.185$ & $7.85\times10^{-2}$\\
14 & 8192 & 64 & $0.181$ & $0.179$ & $-0.185$ & $7.85\times10^{-2}$\\
14 & 32768 & 32 & $7.01\times10^{-2}$ & $6.27\times10^{-2}$ & $-7.01\times10^{-2}$ & $0.176$\\
14 & 32768 & 64 & $7.01\times10^{-2}$ & $6.27\times10^{-2}$ & $-7.01\times10^{-2}$ & $0.176$\\
\bottomrule
\end{tabular}
\end{table}

The cosine history lowers external-target held-out loss in all sixteen
checkpoint pairs. At 32,768 updates, its paired mean improvement is
0.2772 nats at $N=4$ and 0.1979 nats at $N=14$. All four empirical
percentile ranges in Table~\ref{model:tab:onepass-schedule-risk} lie below
zero. This is a consistent measured benefit for the specified
external-target control and fixed stream; the four repeated checkpoint
conditions are not sixteen independent training experiments.

Every collective condition has higher row contrast and absolute row
energy under cosine, lower common-coordinate variance, and higher
paired-context predictive KL. Both forward arithmetics give the same
displayed signs and values. Better external-target prediction can
therefore accompany less row concentration and greater context
sensitivity. A collapse statistic alone cannot order these predictive
outcomes. The comparison changes the complete rate history, including
its accumulated parameter motion and optimizer feedback, so it does
not isolate one instantaneous learning rate.

The risk benefit also has a specified scoring domain. At $N=4$ and
8,192 updates, the proper-panel mean difference is $+0.0792$ nats,
with empirical percentiles $[0.0363,0.1557]$, although external-target
risk improves in every paired identity. At the cosine endpoint the
proper-panel difference is negative in all eight pairs, averaging
$-0.2807$ and $-0.3734$ nats at $N=4,14$. All-target held-out
mean differences are negative in all four conditions, but three
conditions contain an individual pair with the opposite sign.
Training-risk reductions likewise retain their complete per-identity
values. The data thus support the particular schedule benefit in
Table~\ref{model:tab:onepass-schedule-risk}, with explicit limits on its
extension across time, targets and objectives.

\subsection{Temporal observations at the positive rate floor}
\label{model:sec:onepass-floor}

Every compact path contributes both halves of its 8,192-update
floor, giving eighty 4,096-update windows. The constant applied
rate in these windows does not make the remaining corpus stationary.
Each window contains 64 probes on the regular 64-update grid;
every-update shared-parameter increments remain a separate observation.
Table~\ref{model:tab:onepass-floor-temporal} reports the complete ranges
across the eight half-windows per head-count/recipe condition.

\begin{table}[htbp]
\centering\footnotesize
\caption{Complete positive-floor temporal panel. ``Drift'' is the
variance explained by the linear temporal trend divided by the
centered variance of the indicated fixed-context observation.
Lag 64 is its centered lag-one correlation on the 64-update grid.
Each interval gives the minimum and maximum over all four
initializations and both floor halves; it is not a confidence interval.
Any undefined constituent prevents an asserted complete numerical range.}
\label{model:tab:onepass-floor-temporal}
\begin{tabular}{lrrllll}
\toprule
Recipe & $N$ & Halves & Common drift & Loss drift & Common lag 64 & Loss lag 64\\
\midrule
Controlled & 4 & 8 & [0.46, 0.666] & [0.0672, 0.27] & [0.928, 0.956] & [0.4, 0.613]\\
Controlled & 14 & 8 & [0.368, 0.671] & [0.0832, 0.253] & [0.926, 0.955] & [0.466, 0.593]\\
Near 1 & 4 & 8 & [0.421, 0.915] & [0.105, 0.384] & [0.886, 0.957] & [0.261, 0.571]\\
Near 1 & 8 & 8 & [0.444, 0.751] & [0.113, 0.259] & [0.837, 0.939] & [0.269, 0.626]\\
Near 1 & 14 & 8 & [0.47, 0.682] & [0.0842, 0.276] & [0.848, 0.936] & [0.343, 0.438]\\
Below 1 & 4 & 8 & [0.599, 0.815] & [0.0832, 0.301] & [0.926, 0.961] & [0.194, 0.469]\\
Below 1 & 8 & 8 & [0.291, 0.996] & [0.157, 0.45] & [0.72, 0.968] & [0.479, 0.852]\\
Below 1 & 14 & 8 & [0.558, 0.998] & [0.112, 0.431] & [0.931, 0.968] & [0.304, 0.843]\\
Near 2 & 4 & 8 & [0.584, 0.882] & [0.127, 0.426] & [0.898, 0.96] & [0.312, 0.544]\\
Below 2 & 4 & 8 & [0.506, 0.652] & [0.093, 0.247] & [0.929, 0.946] & [0.242, 0.395]\\
\bottomrule
\end{tabular}
\end{table}

On this corpus the consumed fraction increases from $1/4$ to
$5/16$ during the full floor interval. The two halves remove
$1/24$ and $1/23$ of the respective incoming remaining populations.
Under the unrevealed-data convention, these are also the exact
total-variation changes of the uniform laws on remaining block indices in Proposition~\ref{model:prop:consumption-stationarity}.
They refer to source identities, not a measured change of the emitted
predictive law. A projection may attenuate that change, while learning
can introduce further drift. The sampling resource and the observed
model dynamics therefore require separate stationarity controls.

The complete phase-local analysis retains 1,718 windows over all
54 paths, without pooling across a warmup, annealing or floor
boundary. Row-excursion intervals are recorded at each of the five
specified thresholds, including censoring. A threshold interval is
an observation of the finite row field. No threshold defines a
critical surface, and no such interval is relabeled as an avalanche.

Across the eighty floor halves, linear drift explains between 0.291
and 0.998 of the common-centroid temporal variance, and between
0.0672 and 0.450 of the loss variance. The corresponding centered
lag-64 correlations range from 0.720 to 0.968 for the centroid and
from 0.194 to 0.852 for loss. Persistent finite-window correlations
therefore coexist with substantial deterministic trend. The retained
detrended covariances separate that trend contribution but do not
establish a stationary fluctuation process or a diverging relaxation time.

The threshold observations are also sensitive to their definition.
For the forty compact paths, threshold $R=0.1$ yields 171 observed
above-threshold intervals, of which 121 have neither boundary
censored. At $R=10^{-5}$ there are 41 intervals and none has both
boundaries observed; every path starts and ends above that threshold.
The intermediate thresholds and all constant and source-horizon paths
remain in the complete numerical record. Neither merging censored
intervals with observed durations nor selecting a favorable threshold
would supply an avalanche distribution. Together with the changing
data resource and imposed rate history, these measurements support
finite transient dynamics under the specified drive. They provide
no measured critical exponent or endogenous self-organized transition.

\subsection{Complete matched data-law comparison}
\label{model:sec:onepass-matched-data-law}

Tables~\ref{model:tab:onepass-repetition-loss} and
\ref{model:tab:onepass-repetition-geometry} collect all forty selected
checkpoint pairs from fourteen paired training paths. Single-pass
held-out loss is lower in 24 pairs and higher in sixteen. Nine of
the sixteen condition means favor the single-pass law and seven
favor the repeated law. These counts combine different objectives,
head counts and repeated times for descriptive completeness, without
assigning them an independent-trial interpretation. The primary
single-pass theory therefore uses its own measured coefficients
and regimes. The matched repeated law quantifies data conditioning;
it does not supply transferable clock powers or training-noise
coefficients merely because the architecture is shared.

\begin{table}[htbp]
\centering\small
\caption{Complete paired data-law external-target comparison. Every
head-count/horizon condition is retained. $\Delta L$ is single-pass
minus repeated risk on the common held-out cohort; improved counts
paired initialization identities with a negative difference. Repeated
times from a path are not independent replications. The source-recipe
rows retain one paired initialization each.}
\label{model:tab:onepass-repetition-loss}
\begin{tabular}{lrrrrrrr}
\toprule
Settings & $N$ & Updates & Seeds & Repeated & Single pass & $\Delta L$ & Improved\\
\midrule
Const. & 4 & 8192 & 4 & 7.44 & 7.41 & -0.033 & 2/4\\
Const. & 4 & 32768 & 4 & 7.46 & 7.48 & 0.0179 & 2/4\\
Const. & 4 & 65536 & 2 & 7.49 & 7.5 & 0.006 & 0/2\\
Const. & 4 & 98304 & 2 & 7.86 & 8.09 & 0.23 & 1/2\\
Const. & 4 & 131072 & 2 & 8.58 & 8.52 & -0.0574 & 2/2\\
Const. & 8 & 8192 & 4 & 7.31 & 7.31 & 0.00532 & 2/4\\
Const. & 8 & 32768 & 4 & 7.42 & 7.5 & 0.0811 & 1/4\\
Const. & 14 & 8192 & 4 & 7.31 & 7.26 & -0.0579 & 4/4\\
Const. & 14 & 32768 & 4 & 7.36 & 7.31 & -0.0496 & 3/4\\
Const. & 14 & 65536 & 2 & 7.51 & 7.41 & -0.0926 & 2/2\\
Const. & 14 & 98304 & 2 & 7.53 & 7.55 & 0.0227 & 0/2\\
Const. & 14 & 131072 & 2 & 7.61 & 7.57 & -0.0397 & 2/2\\
Near 1 & 14 & 32768 & 1 & 7.01 & 5.5 & -1.51 & 1/1\\
Near 1 & 14 & 65536 & 1 & 9.14 & 5.09 & -4.04 & 1/1\\
Below 1 & 14 & 32768 & 1 & 5.98 & 7.42 & 1.44 & 0/1\\
Below 1 & 14 & 65536 & 1 & 8.89 & 5.18 & -3.72 & 1/1\\
\bottomrule
\end{tabular}
\end{table}

\begin{table}[htbp]
\centering\small
\caption{Complete paired data-law geometry on the same selected states.
Row contrast is in native float32. Susceptibilities condition on the
shared initialization and stream under each data law. A one-initialization
source-recipe row has no measured initialization covariance.}
\label{model:tab:onepass-repetition-geometry}
\begin{tabular}{lrrrrrr}
\toprule
Settings & $N$ & Updates & $R_{\rm repeated}$ & $R_{\rm single}$ & $\chi_{C,\rm repeated}/N$ & $\chi_{C,\rm single}/N$\\
\midrule
Const. & 4 & 8192 & 0.255 & 0.273 & 0.692 & 0.638\\
Const. & 4 & 32768 & $8.05\!\times\!10^{-6}$ & $4.62\!\times\!10^{-5}$ & 0.929 & 0.94\\
Const. & 4 & 65536 & $4.03\!\times\!10^{-6}$ & $7.06\!\times\!10^{-5}$ & 0.796 & 0.786\\
Const. & 4 & 98304 & 0.000439 & 0.0113 & 0.675 & 0.636\\
Const. & 4 & 131072 & 0.0137 & 0.0803 & 0.572 & 0.509\\
Const. & 8 & 8192 & 0.22 & 0.326 & 0.722 & 0.609\\
Const. & 8 & 32768 & 0.000184 & 0.00784 & 0.906 & 0.9\\
Const. & 14 & 8192 & 0.314 & 0.336 & 0.599 & 0.597\\
Const. & 14 & 32768 & 0.000143 & $3.22\!\times\!10^{-5}$ & 0.897 & 0.916\\
Const. & 14 & 65536 & $6.83\!\times\!10^{-6}$ & $8.33\!\times\!10^{-7}$ & 0.774 & 0.719\\
Const. & 14 & 98304 & $1.8\!\times\!10^{-5}$ & $5.29\!\times\!10^{-6}$ & 0.72 & 0.601\\
Const. & 14 & 131072 & 0.0255 & $6.97\!\times\!10^{-5}$ & 0.604 & 0.549\\
Near 1 & 14 & 32768 & 0.156 & $1.41\!\times\!10^{-5}$ & Undefined & Undefined\\
Near 1 & 14 & 65536 & $3.63\!\times\!10^{-6}$ & $3.58\!\times\!10^{-14}$ & Undefined & Undefined\\
Below 1 & 14 & 32768 & 0.138 & 0.518 & Undefined & Undefined\\
Below 1 & 14 & 65536 & $8.71\!\times\!10^{-8}$ & 0.0302 & Undefined & Undefined\\
\bottomrule
\end{tabular}
\end{table}

\subsection{Prediction improves at the longer matched horizon}

Table~\ref{model:tab:onepass-source-risk} reports every selected risk.
At 65,536 updates, the single-pass held-out last-target NLL is
5.094 and 5.177 nats for the Near and Below settings, respectively,
compared with 9.136 and 8.893 under the repeated law. Both
single-pass paths improve from 32,768 to 65,536 on this held-out
cohort. The benefit is not uniform over all selected times:
at 32,768, the Below setting has last-target NLL 7.420 with
single-pass data and 5.979 with repeated data. A claim that a single
pass is always better at each update would therefore be unsupported.

\begin{table}[htbp]
\centering\small
\caption{Matched data-law risk in nats. ``All'' is weighted over all
64 nonpadding targets of a parallel full-window forward; ``last''
is the external target after the input window. All held-out rows use
the same 512 contexts. Training cohorts follow their separately
specified data laws. One initialization per recipe is retained.}
\label{model:tab:onepass-source-risk}
\begin{tabular}{lrlrrr}
\toprule
Settings & Updates & Data & Train all & Held all & Held last\\
\midrule
Near recipe 1 & 32768 & Single pass & 5.550 & 5.626 & 5.500\\
Near recipe 1 & 32768 & Repeated & 3.790 & 6.893 & 7.014\\
Near recipe 1 & 65536 & Single pass & 5.308 & 5.353 & 5.094\\
Near recipe 1 & 65536 & Repeated & 1.777 & 8.929 & 9.136\\
Below recipe 1 & 32768 & Single pass & 7.410 & 7.428 & 7.420\\
Below recipe 1 & 32768 & Repeated & 4.737 & 6.072 & 5.979\\
Below recipe 1 & 65536 & Single pass & 5.304 & 5.327 & 5.177\\
Below recipe 1 & 65536 & Repeated & 0.854 & 8.922 & 8.893\\
\bottomrule
\end{tabular}
\end{table}

The repeated Near setting reduces all-target training risk from
3.790 to 1.777 while its held-out risk rises from 6.893 to 8.929.
The repeated Below setting reduces training risk from 4.737 to
0.854 while held-out risk rises from 6.072 to 8.922. This is
consistent with increasing fit to the small empirical law and
worsening transfer to held-out contexts. The single-pass all-target
train--held-out gaps at 65,536 are only 0.044 and 0.023 nats on
their specified cohorts. These observations support explicit data-law
conditioning of regime selection. They do not identify all causes
of the difference, estimate an unseen population risk, or imply a
critical singularity.

\subsection{Row concentration and proper-prefix prediction}
\label{model:sec:onepass-prefix-results}

The mean native row statistic $R$ averages the normalized centered
energy of the residual learner output $A$ in
Equation~\eqref{model:eq:conditional-head-fields}, before the positive
PLGA transform, over all observed contexts, decoders and heads. Table~\ref{model:tab:onepass-source-rows} also retains absolute
centered energy, separating the numerator from its normalization.
The single-pass Near path has $R=1.409\times10^{-5}$ at 32,768
and $3.578\times10^{-14}$ at 65,536. The Below path has
$R=0.518$ and $0.0302$ at those same times. Thus the longer
single-pass endpoints have similar external-target loss while
retaining very different row geometry. The apparent smallness of
an inter-context operator-dispersion summary alone would not supply
this distinction. In particular, variance across fixed-length training
windows does not bound a change in the conditional operator law across
prefix lengths. The input domain in a contraction statement must include
the prefixes used by inference.

\begin{table}[htbp]
\centering\small
\caption{Complete matched metric-row observations. $R$ is the mean
native normalized row contrast. Absolute row energy uses fixed
mean-square-entry coordinates and float64 accumulation from each
native float32 forward. A small native value does not assert exact
row equality in real arithmetic.}
\label{model:tab:onepass-source-rows}
\begin{tabular}{lrlrr}
\toprule
Settings & Updates & Data & $R$ & Absolute row energy\\
\midrule
Near recipe 1 & 32768 & Single pass & $1.41\!\times\!10^{-5}$ & $6.75\!\times\!10^{-9}$\\
Near recipe 1 & 32768 & Repeated & 0.156 & 0.000265\\
Near recipe 1 & 65536 & Single pass & $3.58\!\times\!10^{-14}$ & $2.49\!\times\!10^{-17}$\\
Near recipe 1 & 65536 & Repeated & $3.63\!\times\!10^{-6}$ & $5.18\!\times\!10^{-13}$\\
Below recipe 1 & 32768 & Single pass & 0.518 & 0.0154\\
Below recipe 1 & 32768 & Repeated & 0.138 & 0.00188\\
Below recipe 1 & 65536 & Single pass & 0.0302 & 0.000129\\
Below recipe 1 & 65536 & Repeated & $8.71\!\times\!10^{-8}$ & $1.01\!\times\!10^{-11}$\\
\bottomrule
\end{tabular}
\end{table}

To test the inference significance of this geometry, a further fixed
cohort of 32 held-out contexts retains the full vocabulary logits
at every one of the 64 proper prefixes. Each target is scored both
from the full-window forward and from its actual available prefix.
The two programs agree bytewise at the final prefix. Target-log-score
and KL bounds from Proposition~\ref{model:prop:causal-risk-transfer} hold
for all 2,048 targets in every selected state; an independent extended
precision calculation checks the probability reductions separately
from native float32 cross entropy.

Table~\ref{model:tab:onepass-source-causal} gives the complete comparison.
For the single-pass Near setting at 65,536, proper-prefix risk is
5.382 nats, the proper-minus-full risk gap is
$1.15\times10^{-8}$ nats and the maximum measured forward KL is
$1.98\times10^{-12}$. For the single-pass Below setting, proper
risk is 5.745 nats while full-window risk is 5.350, a gap of
0.395 nats; the maximum prefix KL is 3.991. The external target
alone does not reveal this earlier-position defect, since it is
outside the input window in both computations. The retained global
metric can depend on the additional suffix at earlier positions.

\begin{table}[htbp]
\centering\small
\caption{Full-window and proper-prefix risk on the same 32 held-out
contexts and all 64 targets. Risks and KL divergences use mathematical
softmax scores of the retained finite logits, reduced in float64 and
independently checked with extended precision. Native cross-entropy
rounding differences are recorded separately. Maximum KL is over
all 2,048 measured target positions.}
\label{model:tab:onepass-source-causal}
\begin{tabular}{lrlrrrr}
\toprule
Settings & Updates & Data & Full & Proper & $L_P-L_F$ & Maximum KL\\
\midrule
Near recipe 1 & 32768 & Single pass & 5.689 & 5.689 & $5.77\!\times\!10^{-9}$ & $9.97\!\times\!10^{-12}$\\
Near recipe 1 & 32768 & Repeated & 6.952 & 6.952 & $-1.02\!\times\!10^{-6}$ & $4.15\!\times\!10^{-8}$\\
Near recipe 1 & 65536 & Single pass & 5.382 & 5.382 & $1.15\!\times\!10^{-8}$ & $1.98\!\times\!10^{-12}$\\
Near recipe 1 & 65536 & Repeated & 9.084 & 9.084 & $-3.37\!\times\!10^{-9}$ & $1.15\!\times\!10^{-11}$\\
Below recipe 1 & 32768 & Single pass & 7.431 & 7.431 & $3.42\!\times\!10^{-7}$ & $6.89\!\times\!10^{-7}$\\
Below recipe 1 & 32768 & Repeated & 6.100 & 6.100 & $-4.12\!\times\!10^{-5}$ & $7.96\!\times\!10^{-5}$\\
Below recipe 1 & 65536 & Single pass & 5.350 & 5.745 & 0.395 & 3.99\\
Below recipe 1 & 65536 & Repeated & 9.011 & 9.011 & $3.36\!\times\!10^{-8}$ & $1.75\!\times\!10^{-11}$\\
\bottomrule
\end{tabular}
\end{table}

\begin{figure}[htbp]
\centering
\includegraphics[width=\textwidth]{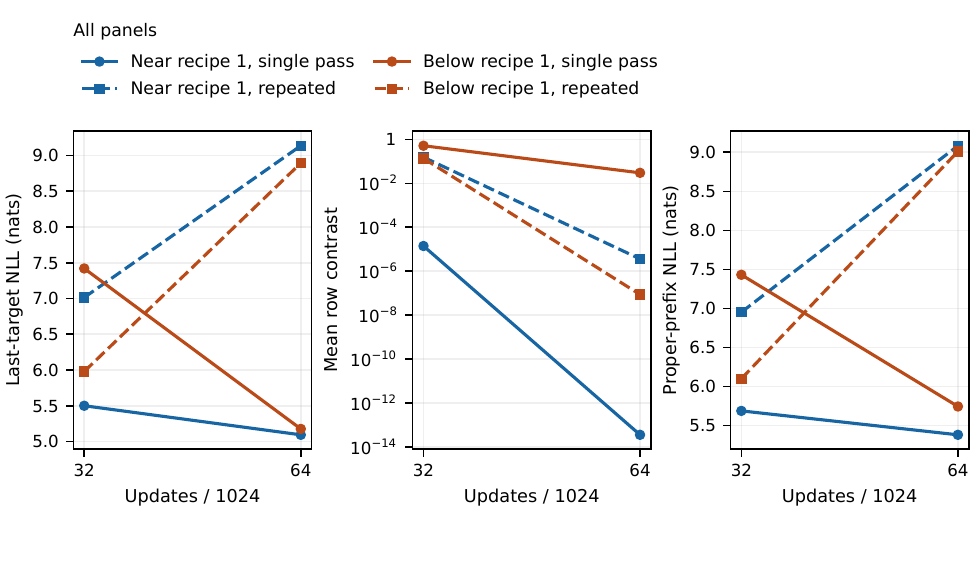}
\caption{The complete matched source-recipe comparison. Left:
external-target held-out risk on 512 contexts. Center: mean metric-row
contrast. Right: proper-prefix all-target risk on the separate fixed
32-context cohort. Lines connect saved observations and do not
identify an unobserved transition time. The reference setting names
describe optimizer recipes; the figure does not classify critical
phases.}
\label{model:fig:onepass-source-comparison}
\end{figure}

These observations support the separation of predictive fit and
operator stabilization in Section~\ref{model:sec:inference-mechanism}.
Under the single-pass law, the Near setting combines strong row
concentration, improved held-out prediction and small measured
prefix defect. The Below setting improves its full-window and
external-target loss at the longer horizon while retaining a larger
row contrast and a measurable prefix defect. Under the repeated
law, even small row contrast and a small prefix defect coexist with
large held-out risk. Correct task margins remain an additional
requirement. Row concentration can help transfer a learned predictive
law to autoregressive inference; it cannot replace learning that law
from the specified corpus or establish reasoning accuracy by itself.

\subsection{Proper-prefix risks under constant rates and completed annealing}
\label{model:sec:onepass-causal-drives}

Table~\ref{model:tab:onepass-constant-causal} completes the all-position
causal comparison for the constant-rate control. It uses initialization
640101 at $N=4,14$ and updates 32,768 and 131,072 under both data
laws. The 32 contexts and 2,048 targets are identical to those in
Table~\ref{model:tab:onepass-source-causal}. These eight states have
absolute proper-minus-full mean gaps below $3.7\times10^{-5}$ nats;
the largest maximum KL over any selected target is 0.00305 nats.
At $N=4$, the single-pass proper risk rises from 7.5072 to 8.4932
nats despite retaining a small prefix defect. The corresponding
$N=14$ proper risks are 7.5376 and 7.5190 on this separate finite
cohort. They need not have the same time ordering as the external-target
risks on 512 contexts. The near equality of the two causal scoring
programs therefore does not ensure good prediction. The small context
KL at the long constant-rate endpoints supplies a complementary check
that prefix stability alone can accompany weak use of context.

\begin{table}[htbp]
\centering\small
\caption{Complete constant-rate causal-risk panel, with one matched
initialization per head count. Full and proper risks use the same
32 contexts and all 64 targets under each data law. Maximum KL and
mathematical risk reductions have the definitions of
Table~\ref{model:tab:onepass-source-causal}.}
\label{model:tab:onepass-constant-causal}
\begin{tabular}{rrlrrrr}
\toprule
$N$ & Updates & Data & Full & Proper & $L_P-L_F$ & Maximum KL\\
\midrule
4 & 32768 & Single pass & 7.50721 & 7.50722 & $6.16\!\times\!10^{-6}$ & 0.000353\\
4 & 32768 & Repeated & 7.51182 & 7.51182 & $1.06\!\times\!10^{-8}$ & $1.32\!\times\!10^{-10}$\\
4 & 131072 & Single pass & 8.49322 & 8.49322 & $-4.47\!\times\!10^{-9}$ & $1.41\!\times\!10^{-14}$\\
4 & 131072 & Repeated & 8.51514 & 8.51514 & $-1.67\!\times\!10^{-11}$ & $4.53\!\times\!10^{-14}$\\
14 & 32768 & Single pass & 7.53760 & 7.53759 & $-1.08\!\times\!10^{-5}$ & 0.00304\\
14 & 32768 & Repeated & 7.57660 & 7.57657 & $-3.62\!\times\!10^{-5}$ & 0.000552\\
14 & 131072 & Single pass & 7.51901 & 7.51901 & $-3.8\!\times\!10^{-10}$ & $4.32\!\times\!10^{-13}$\\
14 & 131072 & Repeated & 7.56326 & 7.56325 & $-1.49\!\times\!10^{-5}$ & 0.000205\\
\bottomrule
\end{tabular}
\end{table}

The completed compact schedules change the annealing horizon to
32,768 updates, then run 8,192 further updates at ten percent of the
peak rate. Table~\ref{model:tab:onepass-compact-causal} reports every
selected all-position causal state from this family: the two source
recipe~1 settings at $N=14$, initialization 640101, and both the
cosine endpoint and the end of the rate floor. Their full-window
risks are low, but their proper-prefix risks and defects are large.
Near proper risk is 17.5955 and 17.2763 nats at the two times;
Below proper risk is 12.8919 and 14.1774. The corresponding
proper-minus-full gaps are 13.9129, 13.6481, 8.9201 and 10.8428
nats. The constant-rate floor does not remove this measured defect
within its observed duration.

\begin{table}[htbp]
\centering\small
\caption{All-position causal risk for the completed compact source
recipe~1 schedules. Both recipes use a single pass and one matched
initialization at $N=14$. The 32-context cohort is identical to that
in Tables~\ref{model:tab:onepass-source-causal} and
\ref{model:tab:onepass-constant-causal}. Update 32,768 is the cosine
endpoint; update 40,960 follows 8,192 updates at the rate floor.}
\label{model:tab:onepass-compact-causal}
\begin{tabular}{lrrrrr}
\toprule
Recipe & Updates & Full & Proper & $L_P-L_F$ & Maximum KL\\
\midrule
Near 1 & 32768 & 3.68264 & 17.59553 & 13.9 & 29.4\\
Near 1 & 40960 & 3.62823 & 17.27629 & 13.6 & 26.1\\
Below 1 & 32768 & 3.97181 & 12.89189 & 8.92 & 28.1\\
Below 1 & 40960 & 3.33455 & 14.17739 & 10.8 & 27\\
\bottomrule
\end{tabular}

\end{table}

A secondary comparison isolates the changed imposed horizon in
these two initialization-matched source-recipe pairs. Their initial
parameters, shared learner, corpus, complete sampling prefix through
32,768 updates, warmup rates and common checkpoint evaluation
cohorts agree. All other optimizer-profile fields agree within each
recipe. The longer schedule has horizon 250,000; the compact schedule
has horizon 32,768. At the matched update 32,768, their applied Near
rates are 0.00115950 and 0.00012000, and their applied Below rates
are 0.00058412 and 0.00006000. The observed difference includes the
entire changed post-warmup drive history, rather than a one-step
rate perturbation. This comparison adds no native training, corpus
replication or untouched population confirmation.

At this matched update, shortening the horizon changes Near
proper risk from 5.6887 to 17.5955 nats and Below proper risk
from 7.4315 to 12.8919. Near mean native row contrast changes
from $1.409\times10^{-5}$ to 0.270. Below row contrast remains
large in both paths, 0.518 and 0.511, while its causal risk changes
substantially. The compact endpoints at 40,960 retain contrasts
0.314 and 0.477, respectively. Thus neither the source recipe name
nor update count specifies the inference regime. The complete drive
history, learned full-graph response and observed input domain remain
necessary coordinates.

In these compact states, low full-window risk does not transfer to
proper-prefix inference. Earlier targets can influence the global
metric through input positions that are available to a parallel
forward but absent from their actual prefix. This route is absent
for the external target after the entire input window, which explains
why last-target risk alone does not measure the defect. The exact
causal-risk identity retains both the learned predictive fit and the
operator dependence on the observation domain. It accommodates the
large measured schedule effect without treating annealing-induced
regime selection as self-organized criticality.

\subsection{The learned row map over the inference input domain}
\label{model:sec:onepass-row-input-results}

A completed mechanism extension examines the same two recipes and
both matched single-pass checkpoints at proper-prefix lengths
16, 32, 48 and 64. It retains the normalized input Gram and every
shared residual stage in all five decoders and fourteen heads on
eight fixed held-out contexts. The 77,496,320 retained endpoint
entries for logits, metrics and operators reproduce their independently
verified prefix observations bytewise. Thus passive stage capture
does not change the measured native emission.

At every length and decoder, the full $64\times64$ Jacobian of the
composed eight-unit row map is measured at eighteen fixed points:
two contexts, three heads and three rows. This gives 1,440 local
Jacobians over all sixteen checkpoint/length conditions. Exact
float64 lifts of the saved native parameters and input vectors are
differentiated by automatic differentiation and independently by
explicit affine, SiLU, product, residual and layer-normalization
formulas. The largest relative Frobenius discrepancy is
$4.66\times10^{-14}$. Native float32 replay of all residual
stages is checked separately.

\begin{table}[htbp]
\centering\small
\caption{Complete local row-map derivative panel for the matched
single-pass checkpoints. A contracting point has composed-map
spectral norm below one. Each condition contains eighteen points
in each of five decoders. These are local input derivatives, not
uniform domain bounds or augmented training multipliers.}
\label{model:tab:onepass-prefix-jacobians}
\begin{tabular}{lrrrr}
\toprule
Settings & Updates & Length & Contracting points & Maximum $\|DF\|_2$\\
\midrule
Near recipe 1 & 32768 & 16 & 90/90 & 0.0257\\
Near recipe 1 & 32768 & 32 & 90/90 & 0.0172\\
Near recipe 1 & 32768 & 48 & 90/90 & 0.0157\\
Near recipe 1 & 32768 & 64 & 90/90 & 0.0164\\
Near recipe 1 & 65536 & 16 & 90/90 & $1.69\!\times\!10^{-5}$\\
Near recipe 1 & 65536 & 32 & 90/90 & $1.99\!\times\!10^{-5}$\\
Near recipe 1 & 65536 & 48 & 90/90 & $2.02\!\times\!10^{-5}$\\
Near recipe 1 & 65536 & 64 & 90/90 & $2.01\!\times\!10^{-5}$\\
Below recipe 1 & 32768 & 16 & 20/90 & 4.54\\
Below recipe 1 & 32768 & 32 & 22/90 & 3.24\\
Below recipe 1 & 32768 & 48 & 28/90 & 3.7\\
Below recipe 1 & 32768 & 64 & 28/90 & 4.15\\
Below recipe 1 & 65536 & 16 & 72/90 & 3.19\\
Below recipe 1 & 65536 & 32 & 72/90 & 3.19\\
Below recipe 1 & 65536 & 48 & 72/90 & 3.31\\
Below recipe 1 & 65536 & 64 & 72/90 & 3.31\\
\bottomrule
\end{tabular}
\end{table}

Every selected Near-recipe Jacobian contracts at both horizons and
all four lengths. Its largest norm across the selected points is
0.0257 at 32,768 and $2.02\times10^{-5}$ at 65,536. The
Below setting has only 20--28 contracting points out of 90 per
length at 32,768, with norms as large as 4.545. At 65,536,
72 of 90 points contract at each length. All eighteen points
in decoder four remain expansive; the largest norm is 3.313.
The other four decoders contract on the entire selected point panel.

The finite prefix secants give a complementary observation.
For each decoder, compare a shorter-prefix normalized Gram with
its full-window counterpart, apply the shared row map to each,
and divide the resulting root mean squared separation by the
incoming separation. On the Near endpoint at 65,536, every
measured ratio is at most $7.28\times10^{-6}$. On the Below
endpoint, decoder four retains a ratio of 0.573--0.583, while
the other four decoders have ratios at most 0.0326. The
full-window comparison with itself has zero incoming separation
and its ratio is undefined, rather than a measured zero gain.
All stage energies, ratios and definition states are retained.

These observations locate a remaining weakly attenuated prefix
route in the Below endpoint and support the input-transport
mechanism in Proposition~\ref{model:prop:shared-row-input-contraction}.
They do not replace the full-graph predictive defect measurement:
upstream states, downstream response and nonlinear interactions
also contribute. In particular, neither the largest derivative
on a finite point set nor a finite-cloud secant is a supremum
Lipschitz constant. Their successful local reconstruction does
not establish a training critical eigenmode.

\subsection{Decoder interventions distinguish prefix consistency from fit}
\label{model:sec:onepass-decoder-results}

A further completed intervention panel tests the two single-pass
endpoints at 65,536 updates on the same 32 held-out contexts and
all 64 target positions. It replaces each decoder's operator
individually, or all five together, by that decoder's previously
measured 64-context calibration mean. The mean is accumulated in
float64 and cast to native float32. Every unmodified decoder still
computes its own context-dependent operator. Two additional
interventions project the metric rows in decoder four, or in all
five decoders, onto their respective centroids. This gives eight
interventions per model. The panel was selected after the prefix
and Jacobian observations; all selected interventions are retained.
No model is retrained and no additional training replica is created.

Every fixed subset replays the native logits bytewise when supplied
with its own context operators. Restoring each intervention restores
the native logits bytewise. Applying the intervention to all five
decoders agrees bytewise with the separately qualified implementation
that acts on the entire stack. The complete full-window and proper-prefix logits are
retained, together with the calibration tensors and full control
logits. Native target losses and independently accumulated
mathematical score and KL differences are reconstructed separately.

\begin{table}[htbp]
\centering\footnotesize
\caption{All decoder interventions on the single-pass Near endpoint.
Risk is measured in nats on the fixed 2,048-target cohort.
$\Delta L_{\rm proper}$ is the change relative to native proper-prefix
risk. ``Rows'' means centroid projection of the selected metric
rows. The maximum KL compares full-window with proper-prefix
predictions within the same intervention, not with the native model.}
\label{model:tab:onepass-decoder-near}
\begin{tabular}{lrrrrr}
\toprule
Intervention & $L_{\rm full}$ & $L_{\rm proper}$ & $\Delta L_{\rm proper}$ & $L_{\rm proper}-L_{\rm full}$ & Max. KL\\
\midrule
Native & 5.3822 & 5.3822 & 0 & $1.15\!\times\!10^{-8}$ & $1.98\!\times\!10^{-12}$\\
Fixed decoder 1 & 5.3822 & 5.3822 & 0 & $1.15\!\times\!10^{-8}$ & $1.98\!\times\!10^{-12}$\\
Fixed decoder 2 & 5.3822 & 5.3822 & $1.01\!\times\!10^{-8}$ & $8.3\!\times\!10^{-9}$ & $1.92\!\times\!10^{-12}$\\
Fixed decoder 3 & 5.3822 & 5.3822 & $2.14\!\times\!10^{-10}$ & $1.19\!\times\!10^{-8}$ & $1.57\!\times\!10^{-12}$\\
Fixed decoder 4 & 5.3822 & 5.3822 & $1.4\!\times\!10^{-9}$ & $-3.15\!\times\!10^{-9}$ & $1.2\!\times\!10^{-12}$\\
Fixed decoder 5 & 5.3822 & 5.3822 & $-1.52\!\times\!10^{-9}$ & $-8.19\!\times\!10^{-9}$ & $1.87\!\times\!10^{-12}$\\
All fixed & 5.3822 & 5.3822 & $-1.63\!\times\!10^{-8}$ & $2.22\!\times\!10^{-10}$ & $1.22\!\times\!10^{-12}$\\
Decoder 4 rows & 5.3822 & 5.3822 & $1.98\!\times\!10^{-8}$ & $1.35\!\times\!10^{-8}$ & $1.96\!\times\!10^{-12}$\\
All rows & 5.3822 & 5.3822 & $1.3\!\times\!10^{-8}$ & $3.59\!\times\!10^{-9}$ & $2.36\!\times\!10^{-12}$\\
\bottomrule
\end{tabular}
\end{table}

\begin{table}[htbp]
\centering\footnotesize
\caption{All decoder interventions on the single-pass Below endpoint,
with the same definitions and target cohort as
Table~\ref{model:tab:onepass-decoder-near}. A reduced prefix defect is
reported together with the change in predictive risk.}
\label{model:tab:onepass-decoder-below}
\begin{tabular}{lrrrrr}
\toprule
Intervention & $L_{\rm full}$ & $L_{\rm proper}$ & $\Delta L_{\rm proper}$ & $L_{\rm proper}-L_{\rm full}$ & Max. KL\\
\midrule
Native & 5.3503 & 5.7451 & 0 & 0.395 & 3.99\\
Fixed decoder 1 & 5.3503 & 5.7451 & $-1.05\!\times\!10^{-7}$ & 0.395 & 3.99\\
Fixed decoder 2 & 5.3503 & 5.7451 & $-1.2\!\times\!10^{-7}$ & 0.395 & 3.99\\
Fixed decoder 3 & 5.3503 & 5.7451 & $-2.9\!\times\!10^{-8}$ & 0.395 & 3.99\\
Fixed decoder 4 & 5.8331 & 5.8331 & 0.0879 & $-7.74\!\times\!10^{-8}$ & $1.11\!\times\!10^{-10}$\\
Fixed decoder 5 & 5.3503 & 5.7451 & $-1.88\!\times\!10^{-8}$ & 0.395 & 3.99\\
All fixed & 5.8331 & 5.8331 & 0.0879 & $2.19\!\times\!10^{-8}$ & $1.77\!\times\!10^{-12}$\\
Decoder 4 rows & 6.8047 & 6.7851 & 1.04 & -0.0197 & 0.0244\\
All rows & 6.8047 & 6.7851 & 1.04 & -0.0197 & 0.0244\\
\bottomrule
\end{tabular}
\end{table}

The Near endpoint preserves proper risk to within
$2.0\times10^{-8}$ nats under every fixed-operator or row
intervention. Its maximum full-to-proper KL remains below
$2.4\times10^{-12}$. For the Below endpoint, individually fixing
decoders one, two, three or five leaves the risk gap near 0.395
nats and maximum KL near 3.991. Fixing decoder four reduces
maximum KL to $1.11\times10^{-10}$ and the absolute mean risk
gap below $7.8\times10^{-8}$ nats. Fixing all decoders reduces
maximum KL further to $1.77\times10^{-12}$. Together with the
input-domain measurements, the individual interventions locate the
dominant measured prefix-dependent route in decoder four. They do
not prove that this route is unique under every input or intervention.

Prefix consistency and predictive fit have different outcomes.
Fixing decoder four increases the Below endpoint's full-window
risk from 5.3503 to 5.8331 nats, while proper risk increases from
5.7451 to 5.8331. Equation~\eqref{model:eq:causal-intervention-risk}
resolves the latter change into a 0.4827-nat increase in full-window
risk and a 0.3948-nat decrease in the signed prefix gap. Projecting
decoder four's metric rows reduces maximum prefix KL to 0.0244
but increases proper risk to 6.7851; projecting all five gives
essentially the same measured result. These interventions support
the conditional mechanism: concentration can preserve a learned
predictive computation when its operator and input-domain
conditions hold. Forcing concentration does not by itself learn
that computation or guarantee useful autoregressive prediction.

\subsection{Conditional native risk transport}
\label{model:sec:onepass-native-risk}

A completed panel measures 192 conditional native steps at six incoming
states from four single-pass trajectories. Every trajectory carries
initialization label 640101. At each state, the complete incoming
weights, Adam moments and schedule state are restored before each of
32 independent conditional draws. A draw selects 32 distinct blocks
uniformly from the unconsumed population, excluding every block already
used by that path. This is the unrevealed-data convention of
Equation~\eqref{model:eq:single-pass-kernel}. The draws are counterfactual
one-step branches at a fixed state. They add neither a training
initialization nor a repeated-data continuation.

The held-out law contains the same 512 fixed external-target contexts.
Forward batches contain 32 contexts and use CPU float32, with mathematical
softmax risks reduced in float64. Geometry uses native float32 matrices
$A$ with float64 construction of row fields. The full, generator and body corners
use the displacements of the same fully clipped Adam update, under the
partition specified in Section~\ref{model:sec:model}. Every selected parameter,
optimizer and scheduler update was independently reconstructed bytewise.
Baseline and all three first/last draw emissions were replayed bytewise;
every retained target risk and predictive KL was independently recomputed
in extended precision. These are CPU updates at GPU-trained incoming
states, with no asserted bytewise equivalence between CPU and GPU steps.

Tables~\ref{model:tab:onepass-native-risk}--\ref{model:tab:onepass-native-kl} retain
every selected state. The finite changes satisfy
$\Delta=\Delta_G+\Delta_B+\Delta_{GB}$ by
Proposition~\ref{model:prop:finite-source-corners}. The displayed means and
standard errors concern conditional minibatch variation at each frozen
state. They are separate from uncertainty over training initializations,
evaluation contexts or corpora.

\begin{table}[htbp]
\centering\small
\caption{Conditional native risk transport on the fixed 512-context
law. $L$ is the incoming mathematical softmax risk in nats. The final
two columns display $10^5$ times the sample mean risk change of the
full and body corners, with one conditional minibatch standard error.
Each row uses all 32 selected draws. Repeated times retain the same
training initialization.}
\label{model:tab:onepass-native-risk}
\begin{tabular}{lrrrrr}
\toprule
Recipe & $N$ & $t$ & $L$ & $10^5\Delta$ & $10^5\Delta_B$\\
\midrule
Constant & 4 & 65,536 & 7.4989 & $-51.5\pm3.12$ & $-51.5\pm3.12$\\
Constant & 4 & 98,304 & 8.0447 & $-2.81\pm0.447$ & $-2.81\pm0.447$\\
Constant & 14 & 65,536 & 7.4788 & $716\pm24.8$ & $690\pm29.9$\\
Constant & 14 & 98,304 & 7.5692 & $183\pm10.9$ & $186\pm10.8$\\
Near 1 & 14 & 65,536 & 5.0941 & $-158\pm32.5$ & $-114\pm32.1$\\
Below 1 & 14 & 65,536 & 5.1767 & $193\pm63.5$ & $-208\pm27.6$\\
\bottomrule
\end{tabular}
\end{table}

\begin{table}[htbp]
\centering\small
\caption{Generator and mixed contributions to the same complete
native steps as Table~\ref{model:tab:onepass-native-risk}. Both means and
standard errors are multiplied by $10^5$. The mixed finite difference
is evaluated from all four corners of one realized update; the partial
corners are not independently optimized paths.}
\label{model:tab:onepass-native-components}
\begin{tabular}{lrrrr}
\toprule
Recipe & $N$ & $t$ & $10^5\Delta_G$ & $10^5\Delta_{GB}$\\
\midrule
Constant & 4 & 65,536 & $4.46\!\times\!10^{-5}\pm6.31\!\times\!10^{-5}$ & $8.87\!\times\!10^{-5}\pm0.000104$\\
Constant & 4 & 98,304 & $0.000592\pm6.57\!\times\!10^{-5}$ & $-0.000565\pm0.000119$\\
Constant & 14 & 65,536 & $130\pm15$ & $-105\pm29.8$\\
Constant & 14 & 98,304 & $-5.88\pm0.54$ & $2.46\pm1.36$\\
Near 1 & 14 & 65,536 & $-45.5\pm5.48$ & $1.95\pm0.28$\\
Below 1 & 14 & 65,536 & $-21\pm24.2$ & $422\pm43$\\
\bottomrule
\end{tabular}
\end{table}

\begin{table}[htbp]
\centering\small
\caption{Mean forward predictive KL from the incoming model to each
updated corner, in nats, over the same 32 conditional draws and 512
contexts. Its magnitude is distinct from the signed target-risk change.}
\label{model:tab:onepass-native-kl}
\begin{tabular}{lrrrrr}
\toprule
Recipe & $N$ & $t$ & Generator KL & Body KL & Full KL\\
\midrule
Constant & 4 & 65,536 & $8.35\!\times\!10^{-15}$ & $4.68\!\times\!10^{-5}$ & $4.68\!\times\!10^{-5}$\\
Constant & 4 & 98,304 & $2.03\!\times\!10^{-15}$ & $3.97\!\times\!10^{-7}$ & $3.97\!\times\!10^{-7}$\\
Constant & 14 & 65,536 & 0.000536 & 0.00295 & 0.00315\\
Constant & 14 & 98,304 & $4.17\!\times\!10^{-7}$ & 0.000446 & 0.000446\\
Near 1 & 14 & 65,536 & $2.68\!\times\!10^{-5}$ & 0.00216 & 0.00223\\
Below 1 & 14 & 65,536 & 0.00386 & 0.00283 & 0.0101\\
\bottomrule
\end{tabular}
\end{table}

For the constant-rate $N=4$ path, the full mean increments are
$-5.14959\times10^{-4}$ and $-2.80621\times10^{-5}$ nats at
65,536 and 98,304 updates. The body accounts for almost all of both
predictive changes; generator-only mean KL is below $8.4\times10^{-15}$
nats. Yet the same completed path has external-target held-out risks
7.4989, 8.0447 and 8.5194 at 65,536, 98,304 and 131,072 updates.
A local conditional decrease and an increasing longer trajectory can
coexist. Integrating the conditional drift requires the evolving law
of weights, optimizer memory and remaining data; two frozen means
cannot replace that law.

Both tested constant-rate $N=14$ states have positive full increments
in all 32 sampled branches. Their means are $7.16021\times10^{-3}$
and $1.82833\times10^{-3}$ nats at 65,536 and 98,304 updates,
respectively. The body supplies the largest contribution at each
state. The generator sample mean changes from positive at the earlier
state to negative at the later state. The negative mixed contribution
at 65,536 partly offsets the two positive partial means. Thus even
the observed component balance depends on the incoming training state.

The earlier $N=14$ state also separates row concentration from the
sign of the predictive step. Its incoming mean row contrast is
$1.4932\times10^{-6}$. The independently replayed first and last
full corners reduce this statistic to $1.3402\times10^{-6}$ and
$1.3295\times10^{-6}$, while each increases held-out risk. Equation~\eqref{model:eq:population-risk-native-drift}
retains the target-risk gradient and the complete optimizer increment,
which the scalar row statistic does not determine.

At the source Near recipe~1 endpoint, the full, body and generator
sample means are all negative. The generator mean is
$-4.54568\times10^{-4}$ nats with conditional standard error
$5.47968\times10^{-5}$. Its first and last replayed corners retain
mean row contrasts $2.1693\times10^{-14}$ and
$2.1551\times10^{-14}$, compared with the incoming
$2.1795\times10^{-14}$. These geometry summaries retain the panel's
float64 field construction, distinct from the wholly native row reduction
in Table~\ref{model:tab:onepass-source-rows}. Generator-parameter learning can therefore
change prediction while row concentration persists and the incoming
row-input map is strongly contracting on the tested domain.
This supports the distinction between input and parameter transport in
Proposition~\ref{model:prop:transverse-common-modes}. The generator partition
includes downstream PLGA parameters, so its risk benefit is not
assigned to one isolated common-metric coordinate.

At the source Below recipe~1 endpoint, the full mean change is
positive while the sum of the two partial means is negative. The
mixed term, $4.22079\times10^{-3}$ nats, reverses their sum.
The generator sample mean itself is small relative to its conditional
standard error. This observation supports retaining the coupled
finite difference in Proposition~\ref{model:prop:finite-source-corners}.
It identifies a joint generator--body contribution at this incoming
state, without localizing that training contribution to one decoder
or identifying a critical instability.

The finite logit identity further distinguishes mechanisms for a loss
increase. Subtracting mean forward KL from mean full risk change in
Equation~\eqref{model:eq:finite-logit-risk} gives the target-score terms
$4.01355\times10^{-3}$ and $1.38255\times10^{-3}$ nats in the
two constant-rate $N=14$ states. Both terms are unfavorable. In
contrast, the Below endpoint has a favorable mean target-score term
of $-8.14700\times10^{-3}$ nats, outweighed by its
$1.00792\times10^{-2}$-nat mean predictive KL. The Near endpoint
has target-score term $-3.80380\times10^{-3}$ and a smaller KL
of $2.22787\times10^{-3}$, leaving an improvement. The Below
mixed risk term decomposes into $3.38347\times10^{-3}$ nats of
full-minus-partial KL and $8.37316\times10^{-4}$ of mixed
target-score change. These are exact algebraic decompositions of the
checked finite emissions. They neither estimate a parameter-space
Hessian nor determine the effect of a learning-rate intervention.

The six states support a model-wide risk law with explicit data
conditioning, optimizer memory and coupled parameter sources.
Their different balances do not support a common component-dominance
approximation across the tested settings. The observations concern
the selected incoming states and conditional update laws; they do not
supply thermodynamic exponents or an endogenous critical mode.

\subsection{Task margins under the single-pass training laws}
\label{model:sec:onepass-task-ensemble}

The task and generation panel includes 94 selected single-pass states.
All forty compact paths are measured at updates 32,768 and 40,960;
both source-horizon paths are measured at 65,536; and every
constant-rate path is measured at its final selected update. These
states form 27 conditions with explicitly retained initialization counts.
In particular, the constant-rate $N=4,14$ task observations at
32,768 use identities 640103 and 640104. Their longer counterparts,
640101 and 640102, are measured at 131,072. The four-identity
core observation at 32,768 is therefore not a four-identity task
measurement. The $N=8$ constant-rate task endpoint uses all four
identities at 32,768.

Each state uses the same 320-question observation law: 64 ARC-Easy
questions, 64 ARC-Challenge questions, and 32 counterfactual pairs
at each of three composition depths. The native, calibrated-operator,
centroid-row-projection and permuted-operator modes retain the same
candidate answers and prompts. Tables~\ref{model:tab:onepass-task-native}--
\ref{model:tab:onepass-task-permuted} report every condition. ``Base correct''
always counts positive native correct-answer margins, including in
the intervention tables. Decision changes count all 320 questions;
certified preservation requires the native correct margin to exceed
twice the largest measured candidate-score error. Pairs correct
counts the 96 counterfactual pairs for which both answer decisions
are correct, rather than merely changed in the expected direction.
Noninteger values are means across the selected initialization identities.

\begin{table}[htbp]
\centering\footnotesize
\caption{Complete native single-pass task panel. ``Cosine'' has
horizon 32,768; ``Source'' has horizon 250,000 and stops at the
reported prefix; ``Const.'' has no imposed annealing. Seeds gives
the actual number of selected task identities. ARC columns are
percentages on the 64-question subsets, not the complete reference
test splits evaluated in Section~\ref{model:sec:inference-results}.}
\label{model:tab:onepass-task-native}
\begin{tabular}{lrrrrrrr}
\toprule
Settings & $N$ & Updates & Seeds & Base correct & ARC-E (\%) & ARC-C (\%) & Pairs correct\\
\midrule
Const. Controlled & 4 & 32768 & 2 & 119.5 & 20.3 & 17.2 & 1\\
Const. Controlled & 4 & 131072 & 2 & 122 & 21.1 & 19.5 & 0\\
Const. Controlled & 8 & 32768 & 4 & 121.75 & 20.7 & 18.4 & 1.25\\
Const. Controlled & 14 & 32768 & 2 & 120.5 & 20.3 & 18.8 & 2\\
Const. Controlled & 14 & 131072 & 2 & 122.5 & 24.2 & 18 & 0\\
Cosine Controlled & 4 & 32768 & 4 & 124 & 23 & 20.3 & 2.75\\
Cosine Controlled & 4 & 40960 & 4 & 123 & 22.3 & 21.9 & 2.25\\
Cosine Controlled & 14 & 32768 & 4 & 122 & 21.5 & 22.3 & 1.75\\
Cosine Controlled & 14 & 40960 & 4 & 121.5 & 21.5 & 21.1 & 1\\
Cosine Near 1 & 4 & 32768 & 4 & 125.25 & 25.8 & 19.1 & 0.5\\
Cosine Near 1 & 4 & 40960 & 4 & 122 & 21.5 & 18.8 & 1.5\\
Cosine Near 1 & 8 & 32768 & 4 & 123 & 22.7 & 18.8 & 1.25\\
Cosine Near 1 & 8 & 40960 & 4 & 125.25 & 25.8 & 20.7 & 1.75\\
Cosine Near 1 & 14 & 32768 & 4 & 125 & 25 & 20.7 & 0\\
Cosine Near 1 & 14 & 40960 & 4 & 121.75 & 22.7 & 19.5 & 0.75\\
Cosine Below 1 & 4 & 32768 & 4 & 125 & 23.8 & 20.3 & 1.25\\
Cosine Below 1 & 4 & 40960 & 4 & 122.75 & 23.4 & 19.1 & 0.75\\
Cosine Below 1 & 8 & 32768 & 4 & 123.75 & 20.3 & 23 & 0\\
Cosine Below 1 & 8 & 40960 & 4 & 125 & 22.3 & 23 & 0\\
Cosine Below 1 & 14 & 32768 & 4 & 123.5 & 21.5 & 21.5 & 0\\
Cosine Below 1 & 14 & 40960 & 4 & 124.5 & 23 & 21.5 & 0\\
Cosine Near 2 & 4 & 32768 & 4 & 122 & 21.9 & 18 & 2\\
Cosine Near 2 & 4 & 40960 & 4 & 122 & 21.5 & 18 & 2\\
Cosine Below 2 & 4 & 32768 & 4 & 121.5 & 22.7 & 18.4 & 1.5\\
Cosine Below 2 & 4 & 40960 & 4 & 124.5 & 23.8 & 18.4 & 3\\
Source Near 1 & 14 & 65536 & 1 & 122 & 25 & 17.2 & 0\\
Source Below 1 & 14 & 65536 & 1 & 128 & 28.1 & 21.9 & 0\\
\bottomrule
\end{tabular}
\end{table}

\begin{table}[htbp]
\centering\footnotesize
\caption{All calibrated-operator task interventions on the same
single-pass states. Changed and certified counts refer to the native
decisions and positive correct margins, respectively. Calibration uses
the separately fixed operator cohort; the task questions do not choose
the replacement operators.}
\label{model:tab:onepass-task-calibration}
\begin{tabular}{lrrrrrrr}
\toprule
Settings & $N$ & Updates & Seeds & Base correct & Changed & Certified & Pairs correct\\
\midrule
Const. Controlled & 4 & 32768 & 2 & 119.5 & 0 & 119.5 & 1\\
Const. Controlled & 4 & 131072 & 2 & 122 & 0 & 121.5 & 0\\
Const. Controlled & 8 & 32768 & 4 & 121.75 & 0 & 120.75 & 1.25\\
Const. Controlled & 14 & 32768 & 2 & 120.5 & 0 & 120.5 & 2\\
Const. Controlled & 14 & 131072 & 2 & 122.5 & 0.5 & 121.5 & 0\\
Cosine Controlled & 4 & 32768 & 4 & 124 & 0.75 & 122.25 & 2.75\\
Cosine Controlled & 4 & 40960 & 4 & 123 & 0.5 & 121 & 2.25\\
Cosine Controlled & 14 & 32768 & 4 & 122 & 4 & 114.75 & 1.75\\
Cosine Controlled & 14 & 40960 & 4 & 121.5 & 3 & 115 & 1.5\\
Cosine Near 1 & 4 & 32768 & 4 & 125.25 & 67.25 & 11.75 & 0.5\\
Cosine Near 1 & 4 & 40960 & 4 & 122 & 84.5 & 11 & 0.5\\
Cosine Near 1 & 8 & 32768 & 4 & 123 & 71.5 & 10.25 & 0\\
Cosine Near 1 & 8 & 40960 & 4 & 125.25 & 98.25 & 8.25 & 0.5\\
Cosine Near 1 & 14 & 32768 & 4 & 125 & 115 & 12.25 & 0.25\\
Cosine Near 1 & 14 & 40960 & 4 & 121.75 & 97 & 8.25 & 0.75\\
Cosine Below 1 & 4 & 32768 & 4 & 125 & 101.25 & 6.25 & 0\\
Cosine Below 1 & 4 & 40960 & 4 & 122.75 & 80.25 & 4.75 & 0\\
Cosine Below 1 & 8 & 32768 & 4 & 123.75 & 2.25 & 121.75 & 0\\
Cosine Below 1 & 8 & 40960 & 4 & 125 & 1.75 & 120.75 & 0\\
Cosine Below 1 & 14 & 32768 & 4 & 123.5 & 23.5 & 87 & 0.5\\
Cosine Below 1 & 14 & 40960 & 4 & 124.5 & 25 & 86.25 & 0.25\\
Cosine Near 2 & 4 & 32768 & 4 & 122 & 110.25 & 3.5 & 0.75\\
Cosine Near 2 & 4 & 40960 & 4 & 122 & 102.75 & 7.5 & 0\\
Cosine Below 2 & 4 & 32768 & 4 & 121.5 & 90.25 & 14.25 & 0.25\\
Cosine Below 2 & 4 & 40960 & 4 & 124.5 & 84 & 18.5 & 0\\
Source Near 1 & 14 & 65536 & 1 & 122 & 0 & 122 & 0\\
Source Below 1 & 14 & 65536 & 1 & 128 & 22 & 58 & 0\\
\bottomrule
\end{tabular}
\end{table}

\begin{table}[htbp]
\centering\footnotesize
\caption{All centroid-row-projection task interventions, with the
same state selection and definitions as
Table~\ref{model:tab:onepass-task-calibration}.}
\label{model:tab:onepass-task-rows}
\begin{tabular}{lrrrrrrr}
\toprule
Settings & $N$ & Updates & Seeds & Base correct & Changed & Certified & Pairs correct\\
\midrule
Const. Controlled & 4 & 32768 & 2 & 119.5 & 0 & 119.5 & 1\\
Const. Controlled & 4 & 131072 & 2 & 122 & 1 & 121.5 & 0\\
Const. Controlled & 8 & 32768 & 4 & 121.75 & 0 & 121.25 & 1.25\\
Const. Controlled & 14 & 32768 & 2 & 120.5 & 0 & 120.5 & 2\\
Const. Controlled & 14 & 131072 & 2 & 122.5 & 0.5 & 121.5 & 0\\
Cosine Controlled & 4 & 32768 & 4 & 124 & 1.5 & 121.75 & 2.5\\
Cosine Controlled & 4 & 40960 & 4 & 123 & 0.75 & 120.75 & 2.25\\
Cosine Controlled & 14 & 32768 & 4 & 122 & 5 & 113 & 2\\
Cosine Controlled & 14 & 40960 & 4 & 121.5 & 3 & 113 & 1.25\\
Cosine Near 1 & 4 & 32768 & 4 & 125.25 & 64.5 & 14.25 & 0.25\\
Cosine Near 1 & 4 & 40960 & 4 & 122 & 86.75 & 17 & 0.5\\
Cosine Near 1 & 8 & 32768 & 4 & 123 & 67.5 & 17 & 0\\
Cosine Near 1 & 8 & 40960 & 4 & 125.25 & 80.75 & 7.75 & 0.5\\
Cosine Near 1 & 14 & 32768 & 4 & 125 & 85 & 3.25 & 0\\
Cosine Near 1 & 14 & 40960 & 4 & 121.75 & 85.25 & 4.25 & 0.75\\
Cosine Below 1 & 4 & 32768 & 4 & 125 & 74.5 & 12 & 0.25\\
Cosine Below 1 & 4 & 40960 & 4 & 122.75 & 77.75 & 7.25 & 0\\
Cosine Below 1 & 8 & 32768 & 4 & 123.75 & 2 & 121.25 & 0\\
Cosine Below 1 & 8 & 40960 & 4 & 125 & 0 & 123.5 & 0\\
Cosine Below 1 & 14 & 32768 & 4 & 123.5 & 13.75 & 86.25 & 0\\
Cosine Below 1 & 14 & 40960 & 4 & 124.5 & 11.5 & 89.25 & 0.5\\
Cosine Near 2 & 4 & 32768 & 4 & 122 & 104 & 7.25 & 1\\
Cosine Near 2 & 4 & 40960 & 4 & 122 & 104.25 & 4.75 & 0.25\\
Cosine Below 2 & 4 & 32768 & 4 & 121.5 & 64.5 & 27.25 & 0\\
Cosine Below 2 & 4 & 40960 & 4 & 124.5 & 65 & 30.75 & 0\\
Source Near 1 & 14 & 65536 & 1 & 122 & 0 & 122 & 0\\
Source Below 1 & 14 & 65536 & 1 & 128 & 22 & 42 & 0\\
\bottomrule
\end{tabular}
\end{table}

\begin{table}[htbp]
\centering\footnotesize
\caption{All permuted-operator task interventions, with the same
state selection and definitions as
Table~\ref{model:tab:onepass-task-calibration}. This intervention tests the
learned operator coordinates rather than just their dispersion.}
\label{model:tab:onepass-task-permuted}
\begin{tabular}{lrrrrrrr}
\toprule
Settings & $N$ & Updates & Seeds & Base correct & Changed & Certified & Pairs correct\\
\midrule
Const. Controlled & 4 & 32768 & 2 & 119.5 & 25.5 & 83.5 & 1.5\\
Const. Controlled & 4 & 131072 & 2 & 122 & 1 & 121 & 0\\
Const. Controlled & 8 & 32768 & 4 & 121.75 & 26.75 & 73.25 & 2.75\\
Const. Controlled & 14 & 32768 & 2 & 120.5 & 30.5 & 70.5 & 2\\
Const. Controlled & 14 & 131072 & 2 & 122.5 & 4.5 & 113.5 & 0\\
Cosine Controlled & 4 & 32768 & 4 & 124 & 40.75 & 62.25 & 3.5\\
Cosine Controlled & 4 & 40960 & 4 & 123 & 41.25 & 62.75 & 3.75\\
Cosine Controlled & 14 & 32768 & 4 & 122 & 43 & 58.5 & 2.25\\
Cosine Controlled & 14 & 40960 & 4 & 121.5 & 31.5 & 67 & 2\\
Cosine Near 1 & 4 & 32768 & 4 & 125.25 & 56.25 & 6.25 & 0.25\\
Cosine Near 1 & 4 & 40960 & 4 & 122 & 84.25 & 4.25 & 0.5\\
Cosine Near 1 & 8 & 32768 & 4 & 123 & 90 & 4.75 & 0.25\\
Cosine Near 1 & 8 & 40960 & 4 & 125.25 & 94.5 & 9 & 0.25\\
Cosine Near 1 & 14 & 32768 & 4 & 125 & 90.25 & 6.75 & 0\\
Cosine Near 1 & 14 & 40960 & 4 & 121.75 & 82.25 & 11.5 & 1.25\\
Cosine Below 1 & 4 & 32768 & 4 & 125 & 84.75 & 6 & 0\\
Cosine Below 1 & 4 & 40960 & 4 & 122.75 & 84 & 5.25 & 0\\
Cosine Below 1 & 8 & 32768 & 4 & 123.75 & 2.75 & 119.75 & 0\\
Cosine Below 1 & 8 & 40960 & 4 & 125 & 7.25 & 114.5 & 0\\
Cosine Below 1 & 14 & 32768 & 4 & 123.5 & 16.5 & 83.25 & 0.5\\
Cosine Below 1 & 14 & 40960 & 4 & 124.5 & 18.75 & 87 & 0.25\\
Cosine Near 2 & 4 & 32768 & 4 & 122 & 89.5 & 5 & 0\\
Cosine Near 2 & 4 & 40960 & 4 & 122 & 93.75 & 2.25 & 0\\
Cosine Below 2 & 4 & 32768 & 4 & 121.5 & 85.25 & 10.75 & 0\\
Cosine Below 2 & 4 & 40960 & 4 & 124.5 & 97.75 & 23.5 & 0\\
Source Near 1 & 14 & 65536 & 1 & 122 & 41 & 44 & 1\\
Source Below 1 & 14 & 65536 & 1 & 128 & 26 & 29 & 1\\
\bottomrule
\end{tabular}
\end{table}

Across all 27 native conditions, mean strictly correct counts range
from 119.5 to 128 of 320. ARC-Easy condition accuracies range from
20.31\% to 28.13\%, and ARC-Challenge from 17.19\% to 23.05\%.
The composition observation is more restrictive than a single answer
preference: only zero to three of the 96 counterfactual pairs are
correct on both variants, in condition means. A fixed preference can
answer one member of a balanced counterfactual pair correctly without
using the changed premise. Positive signed premise-response fractions
are retained separately at each depth and do not substitute for
both-variant correctness.

The matched task outcomes have no universal Near/Below ordering.
Near~1 has better external-target loss than Below~1 at every compact
endpoint, but its ARC-Challenge accuracy is lower at all six of
those conditions. Its ARC-Easy accuracy is higher in four and lower
in two. Both $N=4$ Near~2 endpoints have lower measured ARC-Easy
and ARC-Challenge accuracy than their Below~2 counterparts. These
small, fixed-cohort differences do not rank population reasoning
ability. They do show why task performance cannot be inferred from
a source-recipe name or from the external-target loss ordering.

The interventions resolve several different forms of stability.
Calibrated operators change at most 0.5 decisions per condition mean
in the constant-rate panel; row projection changes at most one.
The controlled cosine conditions remain relatively stable, with at
most four and five changed decisions, respectively. In contrast,
the six Near~1 compact endpoint conditions change 67.25--115
answers under calibration and 64.5--86.75 under row projection.
Only 8.25--12.25 native correct margins are certified under
calibration in these conditions. The Near~2 endpoints likewise
change more than one hundred answers under either reduction.
The calibrated operator from one fixed domain consequently need
not preserve task decisions on the actual prompt domain.

Large row contrast does not force large task sensitivity to every
operator intervention. Below~1 at $N=8$ has mean $R$ near 0.69,
but calibration changes only 2.25 and 1.75 answers at the two
endpoints; row projection changes two and zero. Its external-target
loss remains above 7.39 nats and no counterfactual pair is correct
on both variants. Weak sensitivity can accompany weak learned
prediction. Conversely, at the source-horizon Near endpoint,
calibration and row projection preserve all 320 decisions and
certify all 122 positive native correct margins. Permutation changes
41 decisions and retains 44 certificates. The measured learned
coordinates therefore matter even when their replacement by the
calibrated operator is harmless. The corresponding Below endpoint
changes 22 decisions under each reduction and has 58 and 42
certificates. All four modes and every selected condition remain
in the tables, including interventions that alter previously
incorrect decisions.

The generation law uses 100 fixed 64-token prompts and two fixed
streams of sampling uniforms per state, with a maximum of 32 new
tokens. Of the 18,800 selected continuations, only one ends at EOS
before reaching that limit. Thus nearly all reported lengths are
right-censored by the fixed generation horizon. The condition-mean
RMS-normalized $G_{\mathrm{LM}}$ discrepancy between the prompt and
the first generated continuation is about $1.80\times10^{-8}$ at
the source-horizon Near endpoint and $0.00602$ at the Below endpoint.
For the compact Near~1 endpoints it ranges from $0.00965$ to
$0.0159$, and for Near~2 from $0.0893$ to $0.0963$. The full
records also retain the second continuation, continuation-to-continuation
comparisons, absolute errors and mean-normalized errors. This
finite operator stability concerns the specified generated inputs;
it is neither a cached-generation execution test nor a correctness
assessment of the generated text.

The generation observations retain the selected token paths, stopping
counts and changes of native operators and internal tensors. They
describe behavior under their finite prompt law. Entropy, output length
and operator stability are not substitutes for task correctness or
coherence. Likewise, preserving a model's answer under operator
replacement certifies preservation of that measured computation; it
does not establish broad reasoning competence or a native critical class.

\section{Potential and deductive-tensor activity during single-pass training}
\label{model:sec:potential-results}

\subsection{Matched early-training paths and observation units}
\label{model:sec:potential-methods}
The completed early-path assay comprises eighteen native PLDR runs:
2, 4 and 8 heads per layer, two full initialization seeds, and three
learning-rate schedules. Five decoder layers, 64-dimensional heads,
eight residual metric units and the variance-normalized initialization
family are retained. The parameter counts are 12,069,906, 23,423,332
and 52,032,274. Each run performs 2,048 updates at batch 32, consuming
65,536 distinct RefinedWeb blocks with 64 input tokens and one external
next-token target. No block repeats within a path. Model widths and
schedule arms share the same block order within each seed. The fixed
probe documents are disjoint from the training corpus.

The peak rate is $8\times 10^{-4}$. The three schedules are constant,
256-step linear warm-up followed by a plateau, and the same warm-up
followed by cosine annealing to ten percent of peak. All use AdamW with
$(\beta_1,\beta_2)=(0.9,0.95)$, $\epsilon=10^{-5}$, decay 0.1 and
value clipping at 1. The schedules start with a positive applied rate;
there is no zero-rate initial optimizer update in this assay.
This is an early pretraining family with a proper-prefix target, not a
reconstruction of the long parallel-target reference runs.

After every optimizer update, two fixed unseen inputs expose the complete
potential-increment decomposition and the curvature increment. Full
matrix coordinates determine their RMS statistics. A fixed panel of 128
entries per layer and head retains the base logarithms and learned
exponents for spatial concentration diagnostics. The two probes, matrix
entries and excursion events are observations of a training realization,
not independent training replicates. Sixty-four disjoint fixed inputs
supply initial and endpoint external-target risk measurements.

All eighteen paths complete in 3,046 summed GPU seconds, using at most
5.5 GB of allocated device memory per path. They execute 36,864 training
updates and 75,497,472 input-token presentations across the matched
paths. This latter total counts repeated presentation across distinct
models; it is not a count of globally unique text. The all-coordinate
finite identities are reconstructed from native float 32 tensors using
float 64 arithmetic. No exact equality of real native tensors is inferred
from a native-precision row contrast.

\subsection{Base motion dominates the measured potential decomposition}
Let $a_t=\|\Delta q_t\|_w$ with uniform weights over all layer, head
and matrix coordinates of the first fixed probe. Over updates 257--2048,
the base contribution fraction in Equation~\eqref{model:eq:potential-energy}
is 0.99843--0.99985 across the complete eighteen-path family. The signed
cross contribution, divided by the sum of the two separate squared
contributions, lies between $-1.04\times 10^{-4}$ and
$6.11\times 10^{-5}$. Exponent motion therefore contributes little to
this particular log-potential observation even though the exponents
remain learned parameters. The median row-contrast ratio over these
windows is 0.844--0.976, so this early family does not by itself assess
activity on a strongly collapsed row manifold.

\begin{figure}[htbp]\centering
\includegraphics[width=\textwidth]{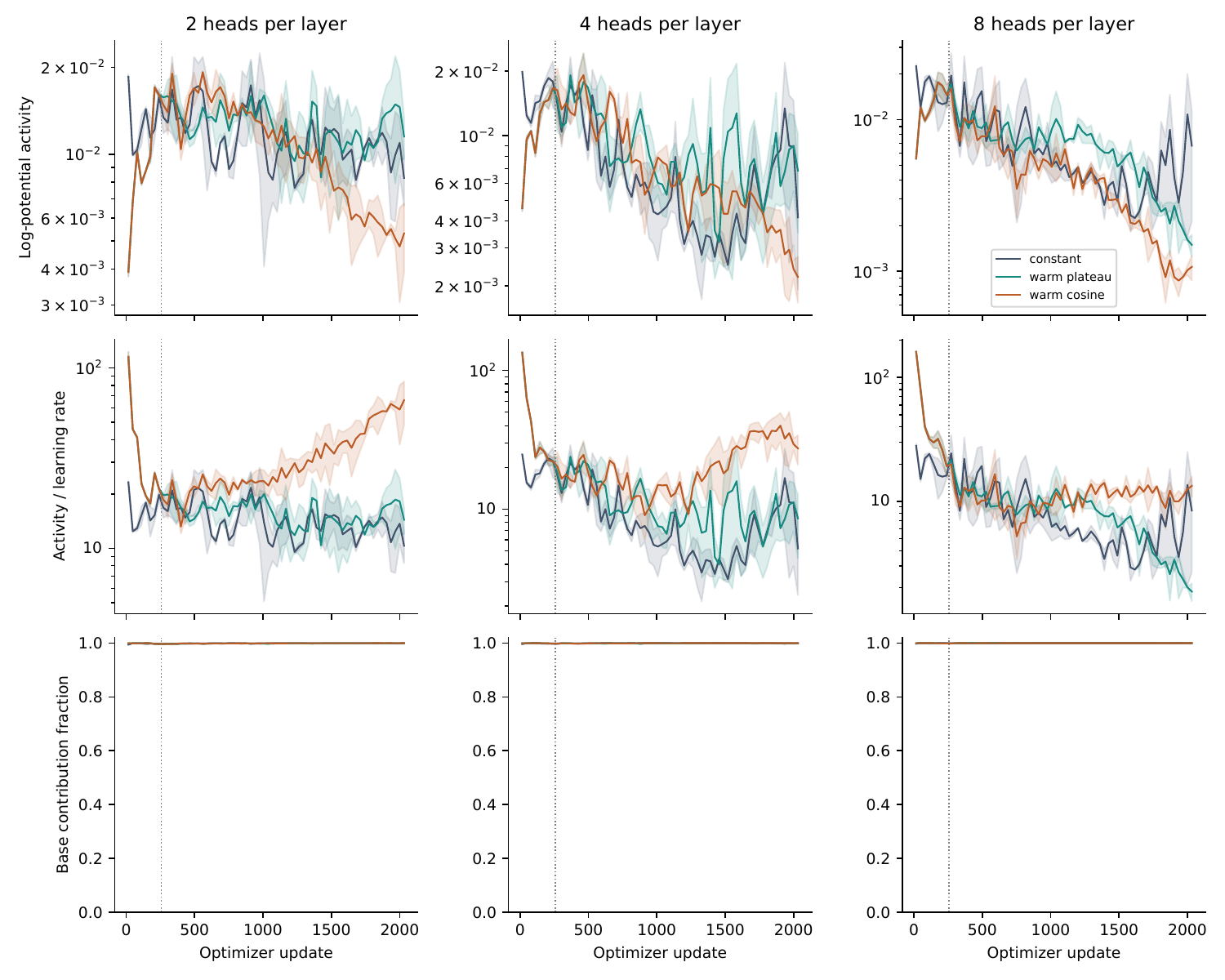}
\caption{Potential activity, its learning-rate-normalized version and
base-contribution fraction in the complete early-path family. Lines are
the mean of two seeds and bands span them. Nonoverlapping 32-update
means are used only for display; every excursion statistic uses its
specified raw cadence. The vertical dotted line ends warm-up.}
\label{model:fig:potential-activity}
\end{figure}

The retained entry panel resolves an additional mechanism. Its median
participation fraction,
$(\sum_i\Delta q_i^2)^2/(N_{\rm panel}\sum_i\Delta q_i^4)$,
is 0.0016--0.0083 across the paths. Entries for which either endpoint
has $M<10^{-6}$ account for 45.7--63.7\% of the accumulated sampled
squared activity. Passive reconstruction with a base offset of
$10^{-6}$, holding the recorded weights and activations fixed, retains
45.6--65.7\% of that squared activity. This supports sensitivity to
small positive bases, as predicted by
Equation~\eqref{model:eq:potential-base-gain}. It does not identify a native
zero-offset limit or predict how changing the offset during training
would alter the learned state.

\subsection{Excursions and correlated controls}
\label{model:sec:potential-excursions}
An excursion is a maximal consecutive run above a specified activity
threshold. Its duration is the number of optimizer updates and its size
is the integrated excess above threshold. Runs touching an observation
window boundary are retained with censoring flags and excluded from
complete-event fits. The analysis keeps four training phases, whole and
post-warm-up windows, both fixed probes, quantiles 0.8,0.9 and 0.95,
and temporal bins of 1,2 and 4 updates. It also applies thresholds frozen
from updates 257--512 to later windows. These alternative observations
are retained together; a favorable threshold does not define a critical
surface.

The primary descriptive comparison uses the first fixed probe, the post-warm-up window,
quantile 0.9 and single-update bins, for both $a_t$ and $a_t/\alpha_t$.
The window includes annealing and parameter drift, so its tail is a
nonstationary mixture. A continuous Pareto tail uses maximum likelihood
on the inclusive support $T(x_{\min})=\{x_i:x_i\ge x_{\min},\ x_i>0,
\ x_i\text{ finite}\}$, with $\widehat\alpha=1+|T|/
\sum_{x_i\in T}\log(x_i/x_{\min})$. The KS search considers at most
128 eligible distinct cutoff values, at equally spaced ranks including
the endpoints, each retaining at least 50 complete tail events. This is
a deterministic approximation to an exhaustive cutoff search; ties are
never split. All diagnostics and alternative likelihoods use the same support.
The 199-replicate semiparametric bootstrap refits the cutoff. Exponential
and conditional lognormal alternatives use the same selected support.
Its iid goodness-of-fit probability is a diagnostic under an assumption
that the training events do not satisfy; it is not calibrated evidence
for a critical process.

\begin{table}[htbp]\centering\small
\caption{Complete early-path excursion comparison. Each slash separates the two initialization seeds. Events use the first fixed probe, updates 257--2048, the within-window 90th percentile and single-update bins. $\alpha$ is a conditional continuous-Pareto density exponent; $p_{\rm iid}$ is an iid semiparametric goodness-of-fit diagnostic. A dash means fewer than 50 complete events, so no tail fit is reported. These are nonstationary-window descriptions, not native critical exponents.}
\label{model:tab:potential-early}
\begin{tabular}{@{}rlrrr@{}}
\toprule Heads & Schedule & Events & $\alpha$ & $p_{\rm iid}$\\\midrule
2 & Constant & 86/83 & 1.741/1.793 & 0.025/0.965\\
2 & Warm-up, plateau & 61/87 & 1.466/1.877 & 0.025/0.130\\
2 & Warm-up, cosine & 72/81 & 1.646/1.717 & 0.045/0.015\\
4 & Constant & 44/26 & --/-- & --/--\\
4 & Warm-up, plateau & 73/26 & 1.776/-- & 0.025/--\\
4 & Warm-up, cosine & 29/38 & --/-- & --/--\\
8 & Constant & 31/30 & --/-- & --/--\\
8 & Warm-up, plateau & 48/39 & --/-- & --/--\\
8 & Warm-up, cosine & 22/45 & --/-- & --/--\\
\bottomrule\end{tabular}\end{table}

The primary raw-activity windows contain 22--87 complete events per path.
Seven paths have enough events for a tail fit; two have
$p_{\rm iid}\geq 0.1$. Their fitted density exponents across all seven
fits span 1.466--1.877. Learning-rate normalization changes the event
sets: ten paths admit a fit and three have $p_{\rm iid}\geq 0.1$.
Neither comparison establishes a common size-dependent exponent.
Shorter within-phase windows and temporal coarsening further limit the
number of complete events. An insufficient event count is not evidence
against every possible tail law.

All eighteen raw paths have longer mean excursions than each of 199
whole-window permutations and each of 199 permutations restricted to
64-update blocks. Their activity has lag-one rank correlation
0.770--0.947. Iterative amplitude-adjusted Fourier surrogates retain the
observed amplitudes and approximately the spectrum. Only one of eighteen
raw comparisons has an unadjusted upper-tail rank at most 0.05 for excess
mean duration, with minimum 0.045. For the declared family of eighteen raw comparisons,
Bonferroni adjustment multiplies each rank by eighteen and caps it at one;
the smallest adjusted descriptive rank is 0.81. Counting both clocks as
one family gives 36 comparisons. With $B=199$, the plus-one rank
$(1+\#\{T_b\ge T_{\rm obs}\})/(B+1)$ has floor 0.005, exceeding
$0.05/18$ and $0.05/36$. At least 359 or 719 surrogates, respectively,
are needed merely to resolve those thresholds. These are resolution
floors, not power guarantees. IAAFT ranks remain descriptive because
exchangeable-null calibration has not been established for the
nonstationary adaptive training law.
These controls distinguish persistent activity from independently ordered
noise, while showing that the persistence statistic does not isolate
collective propagation beyond a correlated surrogate.

\begin{figure}[htbp]\centering
\includegraphics[width=\textwidth]{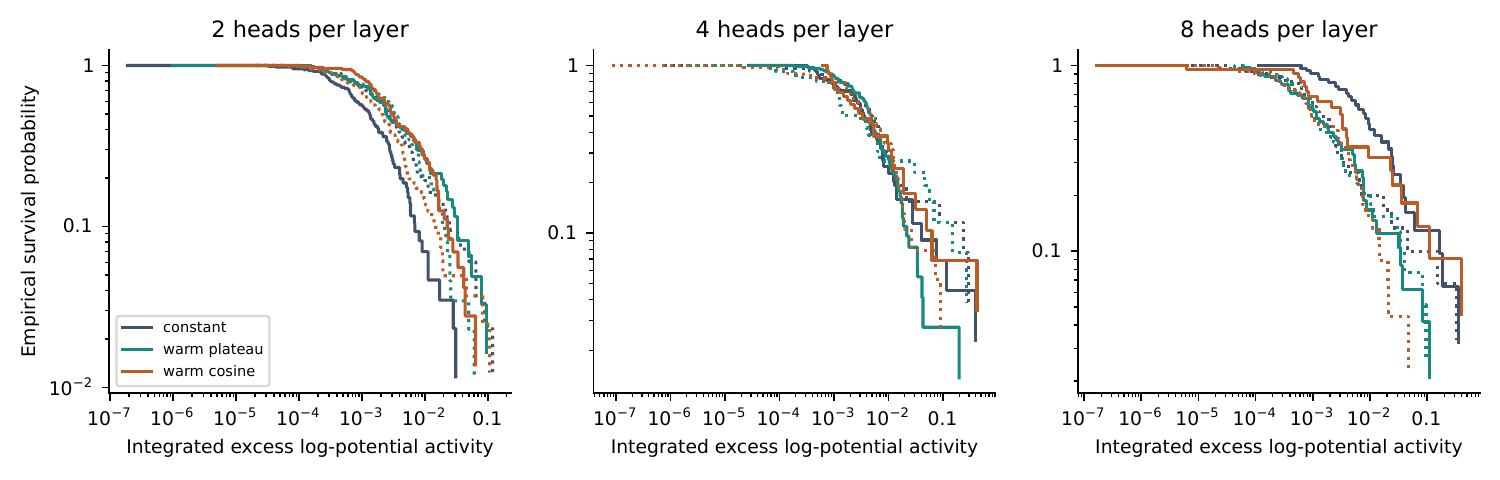}
\caption{Complete primary excursion-size survival curves. Each path is
drawn separately; solid and dotted curves distinguish the two seeds.
No event distribution is pooled across widths or initialization seeds.
The curves describe finite threshold observations of training activity.}
\label{model:fig:potential-excursions}
\end{figure}

\subsection{Activity after long pretraining and row collapse}
\label{model:sec:potential-long-results}
Four further paths continue the two available long scheduled single-pass
14-head endpoints from update 65,536. Each endpoint supplies a native-clock
continuation and a branch annealing from its incoming rate to ten percent
over 2,048 additional updates. The complete incoming parameters and Adam
state are retained. These are conditional comparisons at two trained
states with the same initialization seed, not independent pretraining
replicates. The original all-nonpadding-position training objective is
retained; fixed probes always score an external next token.
The producer reconstructs the original sampling prefix, checks exact
agreement with its record, and consumes only the unused suffix. There is
no repeated block in any combined 67,584-update trajectory.

The incoming first-probe row-contrast ratios are
$2.31\times 10^{-14}$ and 0.0313. We call these strongly and partially
row-collapsed states without assigning either a criticality class. The
strongly collapsed state stays at native-precision row contrast under
both drives. Its base-contribution fractions are 0.99252 and 0.99278;
the partially collapsed state's fractions are 0.99280 and 0.99373.
Thus base motion remains dominant after row collapse. In the late entry
panels, small bases $M<10^{-6}$ account for 81.8--88.9\% of accumulated
sampled squared activity. Raising the offset only in passive readout
reconstruction to $10^{-6}$ retains 9.4--18.7\% of that activity.

\begin{figure}[htbp]\centering
\includegraphics[width=\textwidth]{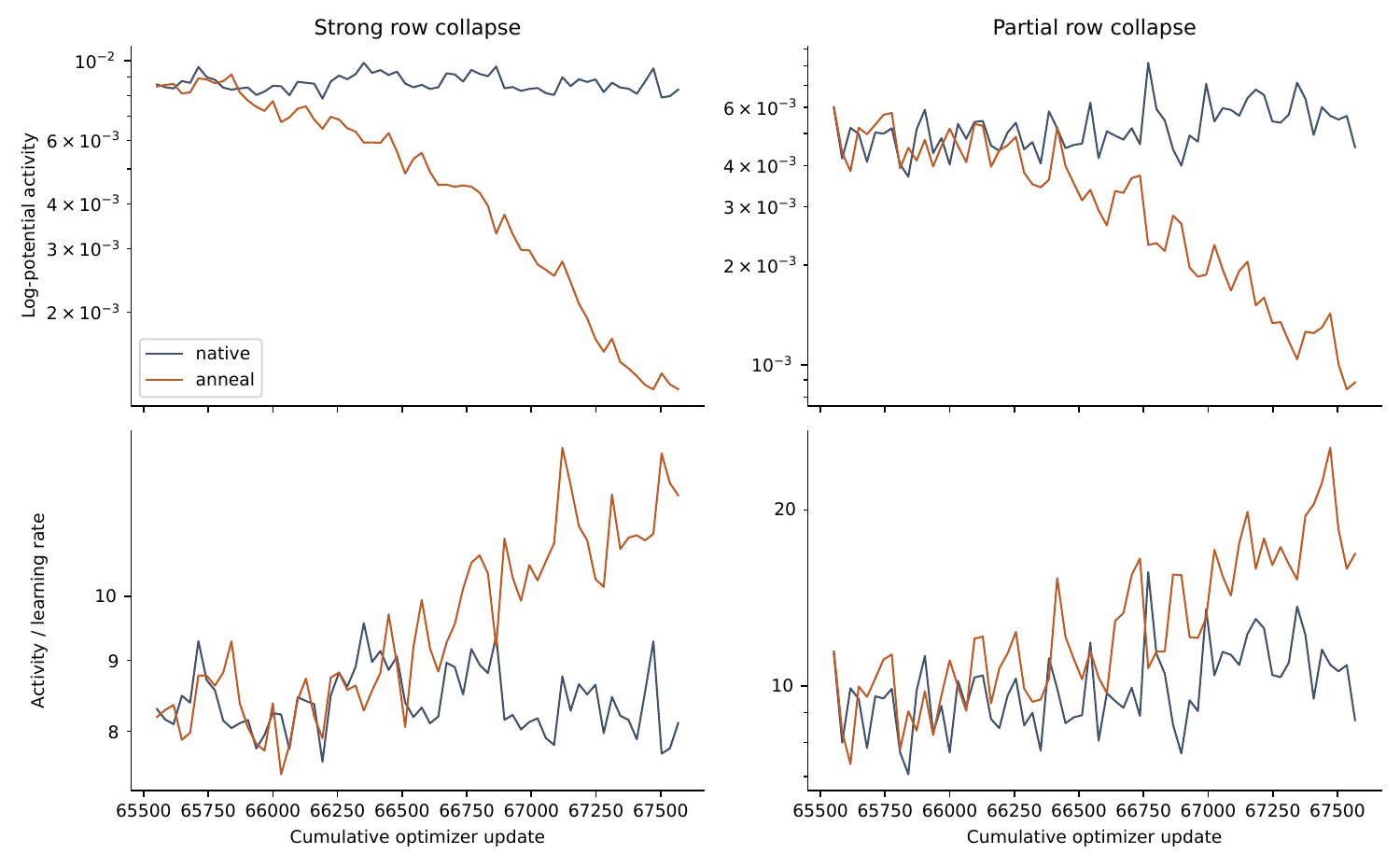}
\caption{Complete single-pass continuation comparison at the two long
pretraining states. The native branch retains the incoming schedule;
the annealed branch changes only its subsequent applied rates. Both
consume the same previously unused block suffix. The plotted 32-update
averages do not replace the update-resolved statistical observations.}
\label{model:fig:potential-continuations}
\end{figure}

The expanded observation records native $A,M,P,V,G$, including every
matrix coordinate for RMS activity and the fixed 128-entry panel per
head. This separates movement of the row-collapsed metric from the
nonlinear positive-base amplification and its subsequent coupling into
attention. All four continuations complete, requiring 1,375 summed GPU
seconds and at most 10.5 GB of allocated device memory per branch.
Together the early and continuation families execute 45,056 additional
training updates on the two GPUs.

\subsection{Which deductive tensor is informative?}
For each tensor $X$, compare its update RMS with the predictive
$\KL(p_{\theta_t}(\cdot\mid x)\|p_{\theta_{t+1}}(\cdot\mid x))$
on the first fixed input. The learning-rate-normalized comparison uses
activity divided by $\alpha_t$ and KL divided by $\alpha_t^2$, as
motivated by the finite-step tangent and quadratic predictive-response
laws. Neither transformation removes the changing native state.

\begin{table}[htbp]\centering\small
\caption{Complete late-branch predictive association. Each entry is raw/rate-normalized Spearman correlation between the named tensor activity and the fixed-probe predictive step KL. The normalized comparison divides tensor activity by $\alpha_t$ and KL by $\alpha_t^2$. These are within-path descriptions, not independent-seed estimates or causal effects.}
\label{model:tab:deductive-predictive}
\begin{tabular}{@{}lrrrr@{}}
\toprule & \multicolumn{2}{c}{Strong row collapse} & \multicolumn{2}{c}{Partial row collapse}\\
Tensor & Native clock & Anneal & Native clock & Anneal\\\midrule
$A$ & 0.020/0.017 & 0.877/0.000 & 0.109/0.112 & 0.214/0.411\\
$M$ & 0.110/0.104 & 0.876/-0.096 & 0.133/0.134 & 0.237/0.030\\
$P$ & 0.020/0.010 & 0.891/-0.020 & 0.147/0.152 & 0.190/0.320\\
$V$ & 0.037/0.036 & 0.757/0.044 & 0.011/0.013 & 0.165/0.169\\
$G$ & 0.044/0.042 & 0.840/0.056 & 0.018/0.021 & 0.189/0.215\\
$\log V$ & -0.020/-0.024 & 0.876/0.029 & 0.070/0.073 & 0.211/0.487\\
\bottomrule\end{tabular}\end{table}

In the annealed strongly collapsed branch, raw rank correlations are
0.757--0.891 across the six observations in
Table~\ref{model:tab:deductive-predictive}. They fall to between $-0.096$ and
0.056 after rate normalization. This association is therefore sensitive to clock normalization; the
comparison alone does not identify schedule-mediated causal pathways. That conclusion does not extend
uniformly to the partially collapsed branch: its normalized log-potential
correlation is 0.487, with metric and exponent correlations 0.411 and
0.320. These retained finite associations motivate observing the joint
transition and predictive projection rather than selecting a tensor
solely for an attractive tail fit. They are not evidence of a causal
branching law.

The complete tensor-tail comparisons also differ by state and schedule.
For example, the learned-exponent activity in the native strongly
collapsed branch has 82 complete events, fitted $\alpha=2.024$ and
$p_{\rm iid}=0.150$. Curvature activity in the native partially
collapsed branch has 92 events, fitted $\alpha=1.730$ and
$p_{\rm iid}=0.495$. The corresponding strongly collapsed curvature
fit has $p_{\rm iid}=0.045$; annealing changes the partially collapsed
curvature fit to $p_{\rm iid}=0.020$. All tensor, clock and threshold
outcomes are retained in the compact evidence. These examples identify
finite candidate tails without selecting a shared critical exponent.
All eight primary late comparisons of log-potential or curvature mean
duration have IAAFT upper-tail ranks above 0.19, so this statistic supplies
no calibrated evidence of excess persistence beyond the correlated controls.

\subsection{Residual activity carried by the optimizer}
\label{model:sec:potential-memory-results}
At both incoming long-training states and the two selected early
warm-up/cosine endpoints, 64 no-data observation steps compare frozen
weights, zero new gradients with retained moments, and zero new gradients
with reset moments. The two moving branches retain weight decay and the
incoming learning rate. They execute 512 optimizer-only updates in total
and consume no corpus blocks.

\begin{table}[htbp]\centering\small
\caption{Zero-gradient intervention over 64 observation steps. The entries are summed log-potential activity. Frozen weights produce zero activity for every recorded deductive tensor. Resetting moments retains native weight decay and the incoming learning rate.}
\label{model:tab:potential-relaxation}
\begin{tabular}{@{}lrrr@{}}
\toprule Incoming state & Retained moments & Reset moments & Ratio\\\midrule
Strong row collapse & 0.16773 & 0.00467 & 35.94\\
Partial row collapse & 0.05149 & 0.00112 & 46.06\\
Early, 2 heads & 0.07354 & 0.00418 & 17.57\\
Early, 8 heads & 0.01918 & 0.00108 & 17.76\\
\bottomrule\end{tabular}\end{table}

Frozen weights give exactly zero repeated-call increments for all five
native deductive tensors in this execution. With explicit zero gradients,
retained moments produce 17.57--46.06 times the integrated log-potential
activity of reset moments. The coordinatewise exponent update predicted
by Equation~\eqref{model:eq:potential-passive-adam} agrees with native execution
to maximum absolute error $3.59\times 10^{-8}$. This verifies passive
optimizer memory as a source of continuing potential motion after new
gradient forcing is removed. It does not establish, or rule out, a
separate collective relaxation process under the native loss gradient.

\begin{figure}[htbp]\centering
\includegraphics[width=\textwidth]{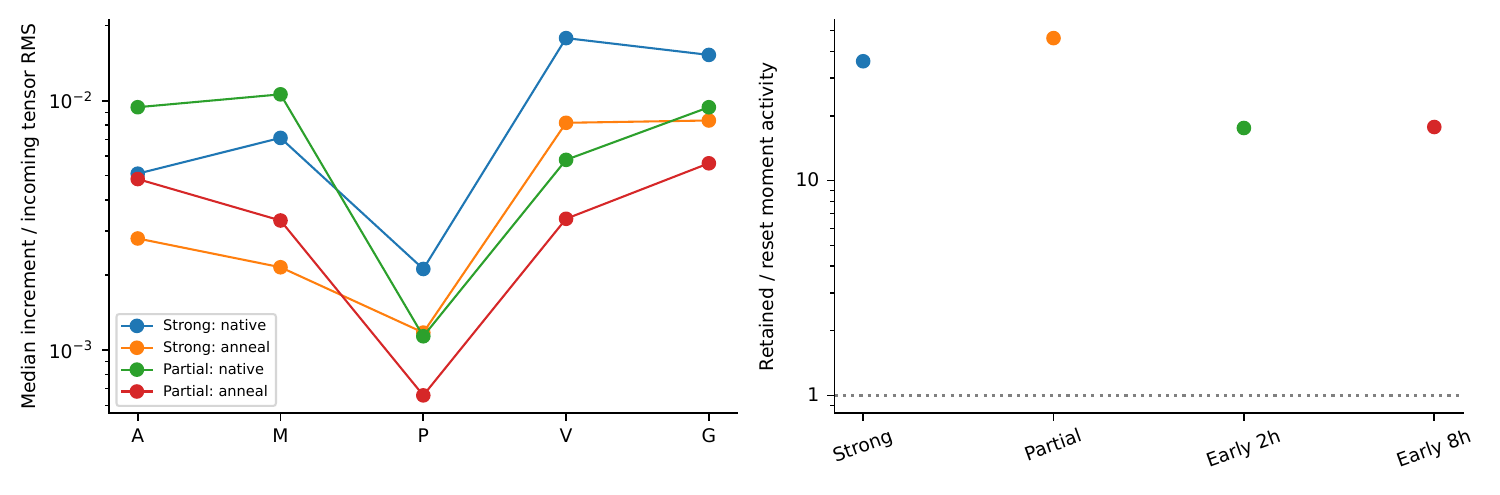}
\caption{Left: median native tensor increments divided by their incoming
RMS, for every long-training branch. Right: the complete retained-moment
to reset-moment activity ratios under zero new gradients. Tensor
normalization is an observation convention; its amplitudes are not
thermodynamic susceptibilities.}
\label{model:fig:deductive-comparison}
\end{figure}

\subsection{Endpoint convention and compatible component blocking}
The fixed first-probe panel of 128 coordinates per layer and head is
reconstructed at both endpoints in float64 from saved float32 base
logarithms and exponents. The early
windows use updates 257--2048; the continuations use every recorded
increment. Proposition~\ref{model:prop:potential-symmetric} applies independently
of the all-coordinate one-sided fractions above.

\begin{table}[htbp]\centering\small
\caption{Endpoint allocation on every retained first-probe coordinate panel. Ranges are across paths, not confidence intervals. Each panel uses 128 entries per layer/head.}
\label{model:tab:potential-endpoint-panels}
\begin{tabular}{@{}lrrr@{}}
\toprule Panel family & Forward base (\%) & Reverse base (\%) & Symmetric base (\%)\\\midrule
18 early paths & 99.8331--99.9855 & 99.8331--99.9856 & 99.8338--99.9856\\
4 continuations & 99.2068--99.3277 & 99.2044--99.3262 & 99.2057--99.3270\\
\bottomrule\end{tabular}\end{table}

The largest forward-to-symmetric change of the base fraction is
$1.161\times10^{-5}$, or 0.001161 percentage points. The largest relative
vector reconstruction residual is $1.22\times10^{-19}$. Thus base motion
dominates this measured decomposition under all three endpoint conventions.
The signed cross term is retained; these fractions divide by the sum of
the separate component energies, not by total energy. The arithmetic
residual refers to saved coordinates and is not an enclosure of an
exact-real native program.

The compatible component map of Proposition~\ref{model:prop:potential-area} is
checked on the same 22 paths at block lengths
$1,2,4,8,16,32,64,128,256$. Each length partitions the entire chosen
window, with no discarded terminal interval. Independent prefix sums
of all strictly ordered exponent/base increments reconstruct the area;
coarse endpoints reconstruct the two allocations separately. All 198
cells satisfy both component identities and composition of adjacent
half-blocks. The largest absolute component-identity residual is
$1.81\times10^{-16}$ in the saved-coordinate arithmetic.

\begin{table}[htbp]\centering\small
\caption{Signed-area correction relative to the blocked log-potential increment, $\|\mathcal A_I\|/\|q_b-q_a\|$, aggregated in squared norm over all aligned blocks and fixed coordinates. Ranges are across paths.}
\label{model:tab:potential-block-area}
\begin{tabular}{@{}lrrr@{}}
\toprule Family & 16 updates & 64 updates & 256 updates\\\midrule
18 early paths & 0.00684--0.02483 & 0.01936--0.07546 & 0.03332--0.16817\\
4 continuations & 0.00330--0.00751 & 0.01039--0.02440 & 0.02638--0.05469\\
\bottomrule\end{tabular}\end{table}

At 256 updates the area norm is 3.3--16.8\% of the blocked total-increment
norm in early paths and 2.6--5.5\% in the continuations. Thus a small
single-update convention discrepancy does not remove path-order
information over longer intervals. Retaining the area gives exact
component blocking, with a measurable norm error if it is omitted.
These are temporal scales on fixed trajectories, not native-width
critical exponents. Predicting the area at a new training state still
requires the conditional successor law.

\subsection{Native offset, optimizer memory and the source suffix}
\label{model:sec:potential-factorial-results}
The four incoming states used for the optimizer-only controls also define
paired $2\times2$ native training experiments. Each state crosses positive
base offsets $10^{-9}$ and $10^{-6}$ with retained or reset first Adam
moments. Second moments, bias counters, decay and scheduler phase are
preserved. Each branch computes its own native loss gradients. This
changes the executed model, rather than only the passive readout of a
saved trajectory.

Two disjoint source suffixes, A and B, are used at each fixed incoming
state. Each branch performs 128 batch-32 updates on 4,096 distinct
RefinedWeb blocks; its four interventions share their source suffix.
Suffix A takes the next unused blocks of the recorded permutation.
Suffix B reserves those blocks and takes the following 4,096, starting
from the same incoming model, optimizer and scheduler. No update occurs
on a reserved block, so source rank and optimizer time remain distinct
coordinates. Both complete consumed prefixes and the absence of reuse
within either path are checked. The two suffixes share a finite corpus
and are not independent model initializations.

The early states retain their positive terminal floor rate and
external-next-token objective; the two long states continue their saved
schedule and all-nonpadding-target objective. The four states therefore
form conditional interventions, not a controlled thermodynamic size
family. Every evaluation uses a proper 64-token prefix and external
target. All-coordinate potential components and five native tensor
summaries accompany full-vocabulary step logits on two fixed inputs.
The first input defines the primary activity and KL. A fixed 64-input
cohort defines origin and endpoint NLL; immediate offset changes are
measured before any optimizer update.

Together the two suffix families contain 32 scientific branches and
4,096 updates. Each has a 64-step width-14 profile with replay and four
128-step unobserved replays, adding 640 separately classified updates.
For suffix B, the profile takes 24.9 seconds and the four case jobs,
including replay and serialization, take 375 summed GPU-job seconds,
with peak allocation 10.63 GiB. One worker runs on each RTX 4090.
Unchanged-offset forward and gradient comparisons agree bitwise;
saved replay arrays and complete-state digests agree. The full per-corner
measurements appear in Appendix~\ref{model:app:factorial-corners}.

\begin{table}[htbp]\centering\small
\caption{Paired offset effects on two disjoint RefinedWeb suffixes, conditional on the same incoming checkpoint. Each cell is a fixed-state contrast, with no population confidence interval. $Q$ is integrated log-potential activity, $G$ integrated curvature activity and $K$ summed successive-prediction KL on the first probe; NLL averages the fixed 64-target proper-prefix cohort. The raised/native ratios retain first Adam moments. Early and long-state objectives and schedules differ.}
\label{model:tab:factorial-source-replication}
\begin{tabular}{@{}llrrrr@{}}
\toprule Incoming state & Suffix & $Q$ ratio & $G$ ratio & $K$ ratio & $\Delta$NLL\\\midrule
Strong, 14 heads & A & 0.4091 & 0.7404 & 0.9964 & -0.003415\\
Strong, 14 heads & B & 0.4082 & 0.7379 & 0.9964 & +0.006405\\
Partial, 14 heads & A & 0.2949 & 0.9664 & 0.7187 & +0.034539\\
Partial, 14 heads & B & 0.3293 & 0.8793 & 1.0585 & +0.014187\\
Early, 2 heads & A & 0.4877 & 0.6741 & 0.6349 & +0.010121\\
Early, 2 heads & B & 0.7109 & 0.7902 & 0.8949 & +0.011929\\
Early, 8 heads & A & 0.3235 & 0.3939 & 1.0000 & $-1.91\times10^{-7}$\\
Early, 8 heads & B & 0.2777 & 0.2828 & 1.0000 & $+5.31\times10^{-8}$\\
\bottomrule\end{tabular}\end{table}

Raising the offset with first moments retained reduces integrated
log-potential activity in every state on both suffixes. The ratios range
from 0.295 to 0.488 on A and 0.278 to 0.711 on B. Thus the directional
activity response carries to disjoint unused blocks at the same incoming
states. Predictive coupling requires additional source information.
On the primary probe at the partially concentrated checkpoint, the KL ratio is 0.719 on A
and 1.058 on B. At the strongly concentrated checkpoint both ratios are
about 0.996, while the eight-head predictive changes remain near native
arithmetic resolution. These finite observations support the joint
source/state description, without assigning an offset a universal
relevance class. NLL effects are conditional on the fixed cohort; the
near-zero eight-head differences are not substantive quality changes.

\begin{figure}[htbp]\centering
\includegraphics[width=\textwidth]{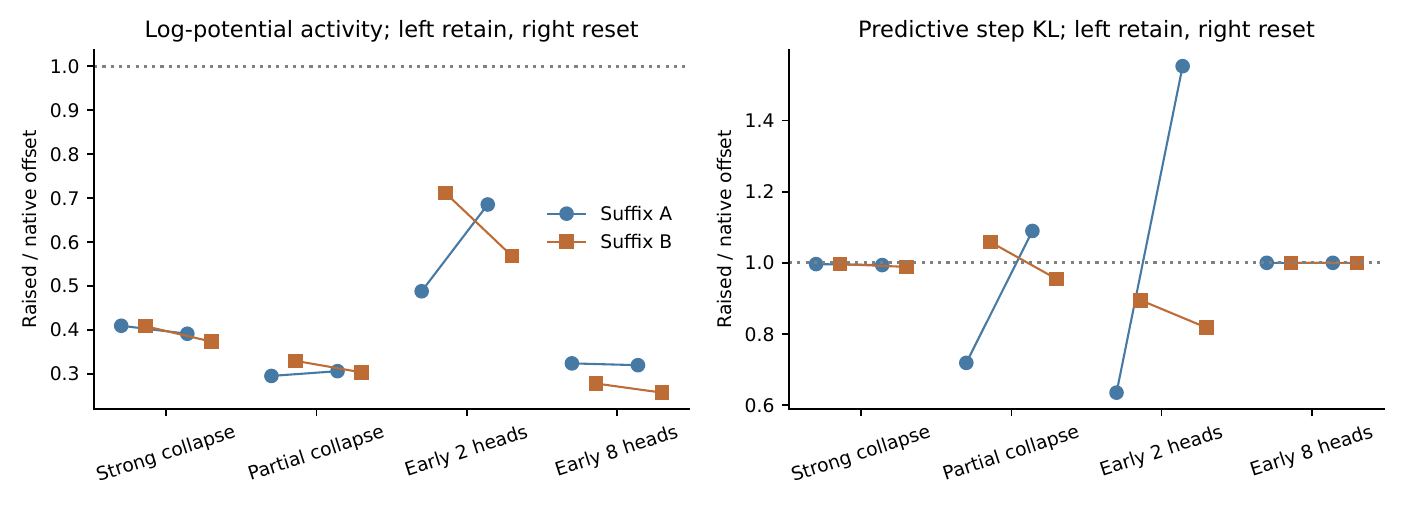}
\caption{All paired offset effects under retained and reset first moments
on both disjoint source suffixes. Points are fixed-state conditional
contrasts. Connecting segments compare the two moment arms, not
independent replicate means. Potential activity is suppressed in every
arm; the primary-probe predictive direction also depends on the source.}
\label{model:fig:potential-factorial}
\end{figure}

The first-moment intervention distinguishes loss-driven training from
passive optimizer relaxation. At the native offset the reset/retain
integrated-activity ratio is 0.913--1.040 on A and 0.922--1.085 on B,
compared with the much larger passive contrast after setting all new
gradients to zero. At the partially concentrated checkpoint the offset
contrast in summed KL changes from $-0.173898$ to $+0.060501$ after
moment reset on A, and from $+0.040271$ to $-0.035078$ on B. The
corresponding interactions are $+0.234399$ and $-0.075348$.
All four corners and both source outcomes remain in the evidence.

\subsection{Resolving the finite predictive contrast}
Proposition~\ref{model:prop:finite-categorical-transport} gives a signed account
of this source dependence. Its categorical covariances are evaluated
without constructing dense vocabulary matrices: a quadratic form is a
weighted coordinate variance, and its cross form a weighted covariance.
Gauss--Legendre integration at orders 8 and 16 is compared with direct
endpoint KL for every update of all sixteen offset pairs, including both
moment arms and both suffixes. The 2,048 step comparisons use all 32,000
vocabulary logits on the primary probe. The largest discrepancy across
endpoint KL, the two quadrature orders and the signed identity is
$2.38\times10^{-15}$ in float64 reconstruction of saved float32 logits.
No higher quadrature order is needed in these observations. This
reconstruction uses the same saved trajectories and supplies an algebraic
correspondence check, not additional independent training replications.

\begin{table}[htbp]\centering\small
\caption{Finite categorical decomposition of the summed offset contrast in predictive step KL at the fixed partially concentrated incoming state. The two suffixes are disjoint; each moment arm is a paired native intervention. Terms are signed observation allocations from Equation~\eqref{model:eq:finite-kl-contrast}, with the first fixed probe and all 32,000 vocabulary logits. No population interval or causal mediation claim is assigned.}
\label{model:tab:finite-kl-transport}
\begin{tabular}{@{}llrrrr@{}}
\toprule Suffix & Moment & Cross & Square & Metric & Total\\\midrule
A & Retain & -0.746223 & +0.591784 & -0.019458 & -0.173898\\
A & Reset & -0.575158 & +0.625352 & +0.010307 & +0.060501\\
B & Retain & -0.509696 & +0.583481 & -0.033515 & +0.040271\\
B & Reset & -0.329823 & +0.328493 & -0.033748 & -0.035078\\
\bottomrule\end{tabular}\end{table}

At the partially concentrated state, the cross term is negative and
the squared-increment term positive in all four comparisons. Their
balance and the signed metric change determine the total response.
The decomposition preserves the actual finite response without imposing
a frozen predictive metric. It is an observation-level allocation;
it does not identify which internal tensor causally mediates that
response or establish a closed successor law for these coordinates.

The signed-area map is also tested on every one of the sixteen B
branches at eight aligned lengths from 1 to 128 updates, adding 128
scale cells. The largest component residual is $3.82\times10^{-17}$.
At length 128 the area norm relative to the total blocked increment
ranges from 0.000276 to 0.038087 across these fixed branches. Independent
trapezoidal reconstruction of these cells and the 198 longer-path cells
agrees, with maximum component residual $1.81\times10^{-16}$. These
are path and observation-scale measurements, with no independent seed
interval or native critical exponent assigned.

\subsection{Finite metric budgets on both predictive probes}
\label{model:sec:finite-metric-results}
The metric-replacement bound in Proposition~\ref{model:prop:metric-budget}
is evaluated on every saved scientific branch of the two source-suffix
experiments. There are four incoming states, four intervention arms,
two suffixes and two fixed probes. The 129 endpoints per branch permit
aligned lengths $b\in\{1,2,4,8,16,32,64,128\}$: 512 scale cells and
16,320 endpoint spans. Each span uses the complete 32,000-token
vocabulary. These are observations of the same 32 branches, with no
additional independent training replication.

Every computed KL lies in its derived interval, with zero observed
violation at tolerance $2\times10^{-12}(1+|K|)$. Independent endpoint
gauge shifts change the checked scalars by at most
$3.553\times10^{-15}$. These are float64 checks of bounds proved in real
arithmetic, rather than floating-point interval certificates.

\begin{table}[htbp]\centering\small
\caption{Finite metric replacement on retained native endpoints.
Relative errors are $100|K/Q-1|$, with $Q=\Var_p(d)/2$ and
$Q>10^{-12}$ in every listed span. The final column counts spans whose
theoretical relative envelope is at most 1\%. Medians and maxima
summarize fixed recorded observations, without population intervals.
The complete eight-scale grid is included in the compact evidence.}
\label{model:tab:metric-budget}
\begin{tabular}{@{}rrrrrr@{}}
\toprule Probe & Block & Spans & Median (\%) & Maximum (\%) & Envelope $\le1\%$\\\midrule
1 & 1 & 4,096 & 0.234 & 11.274 & 711 \\
1 & 8 & 512 & 1.543 & 29.394 & 0 \\
1 & 128 & 32 & 2.909 & 27.792 & 0 \\
2 & 1 & 4,096 & 0.198 & 7.529 & 613 \\
2 & 8 & 512 & 1.305 & 22.404 & 0 \\
2 & 128 & 32 & 2.807 & 72.708 & 0 \\
\bottomrule\end{tabular}

\end{table}

The observation scale determines the cost of replacing the metric.
The largest local-Fisher error is 11.27\% on one-step probe-1 spans
and 72.71\% on probe-2 128-step spans. The exponential envelope
covers this variation without assuming that potential concentration
makes the predictive geometry constant. Summed step KL and KL across
a blocked endpoint pair are different observables; KL does not telescope.

Corollary~\ref{model:cor:refinement} supplies a controlled refinement when
the initial metric intervals do not resolve a paired contrast. On all
eight partially concentrated-state comparisons, the common grid
$m\in\{1,4,16,64\}$ gives respectively zero, one, seven and eight
resolved signs at a $10^{-8}$-nat threshold. Intersecting valid intervals
preserves the bound. This computation observes acquired logits; it is
neither an autonomous prediction nor a cheaper alternative to endpoint
KL already available from those logits.

\begin{table}[htbp]\centering\small
\caption{Every partial-state contrast after 64-segment refinement.
$\Delta K$ is raised-offset minus native summed step KL over 128 updates.
Intervals are deterministic real-arithmetic bounds evaluated in float64,
not statistical confidence intervals. Both probes share each native
branch, and neither suffix is an independent trained-model replicate.}
\label{model:tab:refined-contrasts}
\begin{tabular}{@{}llrrl@{}}
\toprule Suffix & First moment & Probe & $\Delta K$ (nats) & Refined interval (nats)\\\midrule
A & Retain & 1 & -0.173898 & $[-0.180492,\,-0.167311]$ \\
A & Retain & 2 & -0.081538 & $[-0.086973,\,-0.076105]$ \\
A & Reset & 1 & +0.060501 & $[+0.050602,\,+0.070407]$ \\
A & Reset & 2 & +0.087512 & $[+0.080587,\,+0.094442]$ \\
B & Retain & 1 & +0.040271 & $[+0.030354,\,+0.050194]$ \\
B & Retain & 2 & -0.089292 & $[-0.096301,\,-0.082290]$ \\
B & Reset & 1 & -0.035078 & $[-0.046268,\,-0.023882]$ \\
B & Reset & 2 & -0.038448 & $[-0.045447,\,-0.031450]$ \\
\bottomrule\end{tabular}
\end{table}

With retained moments, probe 1 changes from a negative offset contrast
on suffix A to a positive contrast on B. Probe 2 remains negative on
both suffixes. With reset moments, both probes change from positive to
negative across the suffixes. Thus the finite mechanism is conditional
on incoming state, remaining source and predictive observation. A sign
on one context cannot be promoted to a universal emission response.

\begin{figure}[htbp]\centering
\includegraphics[width=.94\textwidth]{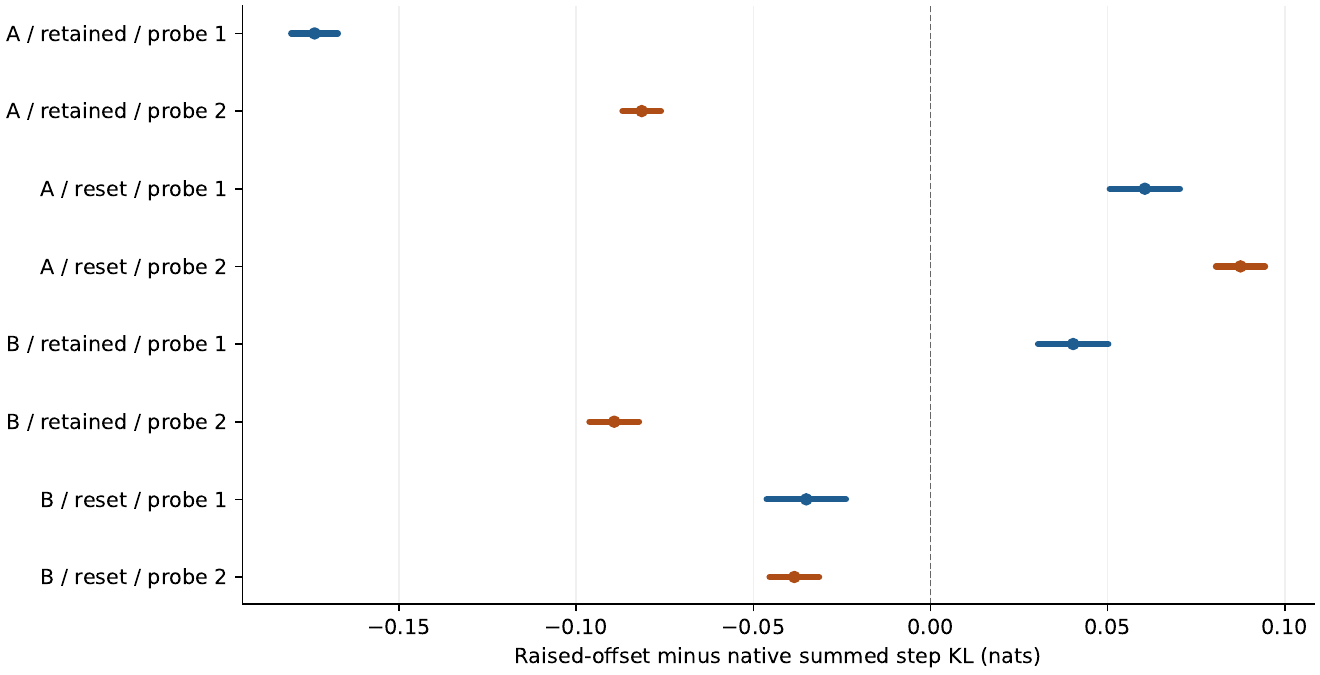}
\caption{Controlled bounds for all eight partial-state predictive contrasts.
Segments give the 64-interval enclosures; dots give direct summed endpoint
calculations. Blue and orange distinguish the two fixed probes.}
\label{model:fig:finite-metric-intervals}
\end{figure}

Predictive motion also differs from task improvement. For target $y$,
direct substitution gives
\[
 -\log p'_y+\log p_y
 =\KL(p\|p')+(p-e_y)^{\mathsf T}(\ell'-\ell).
\]
The signed target term is retained when comparing external-target risk.

The measurements support a finite, jointly observed activity theory:
small positive bases amplify movement, the schedule changes the
observation clock, optimizer memory carries residual motion, and the
curvature and predictive projections can suppress or preserve different
parts of that activity. They do not identify a native toppling rule,
endogenous attraction to a critical window, or universal avalanche
exponents. No conclusion about those limits is obtained by treating
coordinates, events or the two conditional long-training branches as
independent model realizations.

\section{Matched single-pass training across width and schedule}
\label{model:sec:matched-onepass}
The finite theory is tested on a controlled family with head counts
$N\in\{2,4,8,14\}$, residual width $64N$, head dimension 64, depth five
and generator width 170. Three seeds jointly vary full initialization
and source ordering. Native initialization is followed by the specified
variance normalization. At every width and seed, a plateau schedule and
a cosine schedule share the initial parameters, source blocks and
128-update linear warm-up. Both use peak rate $8\times10^{-4}$ for all
parameters; cosine decay ends at a positive 10\% floor at update 1,024.
AdamW coefficients are $(0.9,0.95)$, denominator offset $10^{-5}$,
weight decay 0.1 and coordinatewise gradient clipping at one.

Training, as well as evaluation, supplies a proper 64-token prefix and
supervises its one external next token from the last logit. Table~\ref{model:tab:training-families}
puts this family beside the central control family; identical target
construction does not imply identical optimizer dynamics.

\begin{table}[htbp]\centering\small
\caption{The two central single-pass optimizer families. Both supervise one
external next token per 64-token prefix. $N$ is head count. The separate
source-recipe comparison in the statistical-methods appendix uses all nonpadding targets;
its objective is not identified with either family here.}
\label{model:tab:training-families}
\begin{tabular}{@{}>{\raggedright\arraybackslash}p{.19\textwidth}
>{\raggedright\arraybackslash}p{.38\textwidth}
>{\raggedright\arraybackslash}p{.36\textwidth}@{}}
\toprule & Central control family & Matched schedule family\\\midrule
Rates & Generator and head PLGA $3\times10^{-4}g$; other parameters
$6\times10^{-4}/N$; constant & All parameters peak $8\times10^{-4}$;
128-update warm-up; plateau or cosine to a 10\% floor.\\[4pt]
AdamW & $(0.9,0.95)$; offset $10^{-8}$; decay $0.01$ &
$(0.9,0.95)$; offset $10^{-5}$; decay $0.1$.\\[4pt]
Clipping & Global gradient norm at one & Each gradient coordinate at one.\\[4pt]
Replication & Six complete conditional initializations; fixed source order
and initial shared generator & Three joint initialization/order seeds;
paired schedules.\\
\bottomrule\end{tabular}
\end{table}

All 24 paths perform 1,024 batch-32 updates, totaling 24,576 scientific
updates. Each consumes 32,768 distinct RefinedWeb blocks from one fixed
4,194,304-block resource, a fraction $1/128$. Evaluation uses 128 separate
documents and proper 64-token prefixes with one external target. Eight
fixed contexts provide full-vocabulary logits, all-entry row ratios and
a fixed 32-entry potential panel per layer/head at steps
$0,32,\ldots,1024$. The three seeds replicate initialization/order under
one realized corpus; widths, schedules, times and contexts do not add
independent corpus draws. The complete protocol and resource checks are
in Appendix~\ref{model:app:matched-reproduction}.

\begin{table}[htbp]\centering\small
\caption{Every paired schedule contrast after 1,024 single-pass updates.
Differences are cosine minus plateau. NLL averages 128 fixed external
targets; row ratio averages eight fixed contexts and all matrix entries.
Activity ratios compare the mean RMS of selected log-potential increments
over the final eight 32-update intervals. Velocity first divides each
increment by its accumulated learning rate. These are three paired seed
outcomes per width, with no population confidence interval.}
\label{model:tab:matched-pairs}
\begin{tabular}{@{}rrrrrr@{}}
\toprule Heads & Seed & $\Delta$NLL & $\Delta$row ratio & Activity ratio & Velocity ratio\\\midrule
2 & 914241 & -0.03479 & +0.06391 & 0.6280 & 4.2163\\
2 & 914242 & -0.08784 & +0.12794 & 0.6451 & 4.3133\\
2 & 914243 & -0.02041 & +0.17357 & 0.4255 & 2.8659\\
4 & 914241 & -0.19051 & +0.06229 & 0.7938 & 5.3879\\
4 & 914242 & -0.10194 & +0.04926 & 0.5747 & 3.8593\\
4 & 914243 & -0.11753 & -0.01149 & 0.7164 & 4.8903\\
8 & 914241 & -0.22458 & -0.00433 & 1.4665 & 9.8083\\
8 & 914242 & -0.06693 & +0.00032 & 0.5205 & 3.7187\\
8 & 914243 & -0.26623 & -0.00252 & 0.4592 & 3.2049\\
14 & 914241 & -0.31728 & +0.00747 & 0.2205 & 1.5427\\
14 & 914242 & -0.11057 & -0.00713 & 0.5520 & 3.5500\\
14 & 914243 & -0.19046 & +0.01155 & 0.7779 & 5.2104\\
\bottomrule\end{tabular}
\end{table}

The saved warm-up observations agree bitwise for every paired schedule
through update 128. Final external-target NLL is lower under cosine in
12 of twelve width/seed comparisons, with improvements of
$0.0204$--$0.3173$ nats per target. Row contrast is greater under cosine
in 8 comparisons and smaller in 4. Thus this finite predictive
improvement does not select a universal direction of row concentration.
The complete native outcomes appear in Appendix~\ref{model:app:matched-complete}.

Late potential block activity is smaller under cosine in 11 pairs,
with cosine/plateau ratios spanning $0.221$--$1.467$ over all twelve.
After dividing each block increment by its accumulated learning rate,
the corresponding velocity ratios are all above one, spanning
$1.543$--$9.808$. This supports the clock-conditioned interpretation in
Proposition~\ref{model:prop:potential-clock}: an imposed rate changes observed
activity amplitude, while the evolving state and optimizer still affect
its normalized motion. These are matched schedule interventions, not an
identified restoring drift toward a critical surface.

The scientific paths require 2,072.7 summed recorded GPU-job seconds and
at most 9.15 GiB allocated device memory per job on the two RTX 4090s.
Recorded job time includes initialization, observations and checkpoint
writing; initial admission and final artifact hashing are separate.
The independently classified widest-model qualification and replay add
64 updates. All observed symmetric product increments agree with their
finite identity to maximum absolute discrepancy
$2.220\times10^{-16}$ in float64 reconstruction.

\begin{figure}[htbp]\centering
\includegraphics[width=\textwidth]{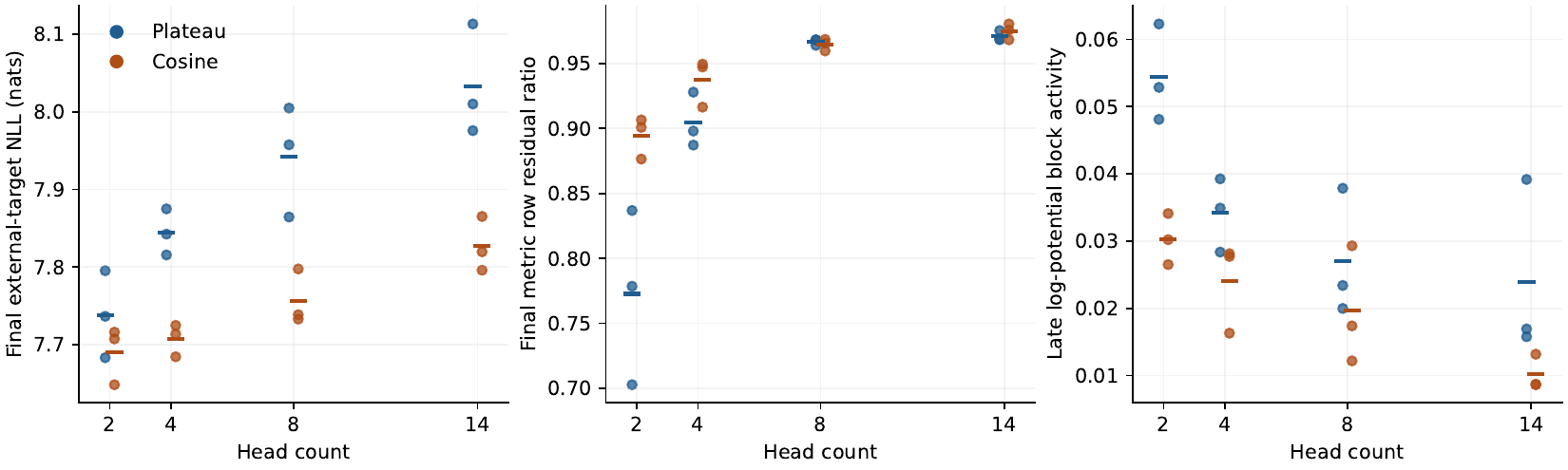}
\caption{Complete final-state observations for the matched family.
Dots show individual initialization/order seeds; short bars show their
arithmetic means. Both schedules use the same source realization and
paired initialization at each width.}
\label{model:fig:matched-onepass}
\end{figure}

\subsection{Predictive metric transport on newly trained paths}
Every one of the 24 paths is checked at all six aligned observation-block
lengths, corresponding to 32--1,024 native updates. Eight contexts give
1,152 scale cells and 12,096 endpoint spans. This is a fresh native family
with a frozen size/schedule grid, separate from the fixed incoming-state
factorial contrasts. The mathematical endpoint bound does not require
that these paths remain in one row regime.

\begin{table}[htbp]\centering\small
\caption{Local-Fisher error across the complete matched family.
Relative error is $100|K/Q-1|$ for $Q=\Var_p(d)/2>10^{-12}$.
The last column counts spans with a theoretical envelope at most 1\%.
All spans obey the finite metric bound at the declared float64 tolerance.
The scalar summaries pool fixed observations and supply no extra model
replication or population interval.}
\label{model:tab:matched-metric}
\begin{tabular}{@{}rrrrr@{}}
\toprule Updates per span & Spans & Median error (\%) & Maximum error (\%) & Envelope $\le1\%$\\\midrule
32 & 6,144 & 5.892 & 306.108 & 0\\
64 & 3,072 & 9.619 & 412.214 & 0\\
128 & 1,536 & 6.049 & 472.376 & 0\\
256 & 768 & 10.042 & 460.248 & 0\\
512 & 384 & 155.139 & 554.634 & 0\\
1024 & 192 & 385.965 & 484.113 & 0\\
\bottomrule\end{tabular}

\end{table}

The prescribed-tolerance rule is evaluated in a complete follow-up
calculation on all 192 full-horizon endpoint pairs. With
$\varepsilon=0.01$, the rule requires 441--527 segments per job and
encloses every direct KL. The maximum observed relative midpoint error is
$1.994\times10^{-6}$, within the proved 1\% envelope. The calculation takes
16.54 seconds on the saved logits. It refines a finite observation
of the same trajectories; no additional training or independent replication
is counted. As elsewhere, numerical evaluation is in float64 and does not
constitute rigorous floating-point interval arithmetic.

The endpoint intervals support finite predictive transport at every
measured scale while quantifying the cost of holding the metric fixed.
They observe the realized logit endpoints and do not predict an unknown
successor. The row/common decomposition, signed potential path state,
optimizer/source coordinates and predictive metric therefore remain
separate retained objects in the model-wide law. The finite family
identifies schedule-selected inference observations under a declared
consuming source law. Its fixed horizon, four small widths and one corpus
do not establish stationarity, a singular thermodynamic limit or a native
critical exponent.

\FloatBarrier
\par\medskip\noindent
Chapter~\ref{ch:width-continuations} extends the evidence to independent
width families and native continuations. Matched clocks and chronological
flux measurements test how the finite patterns change with size and time.

\chapter{Independent width families and chronological continuations}
\label{ch:width-continuations}
This chapter examines independent single-pass width families, training-age
comparisons and chronological continuations. The measurements resolve
conditional fluctuation profiles, shared-parameter accumulation, sign-orbit
budgets and finite row transport over matched physical durations.

\section{Conditional single-pass fluctuation profiles}
\label{model:sec:critical-onepass-results}

The controlled native family fixes five decoders, 64-dimensional heads,
eight residual units in each metric learner and shared hidden width 170.
The residual width is $64N$. The base grid uses
$N\in\{2,4,8,14\}$, generator multipliers
$g\in\{0,1/4,1/2,1,2,4,8\}$, and six independent remaining
initializations for each cell. Shape-aware variance normalization is
applied before fixing the same initial shared residual metric networks
across every width and realization. Head-specific PLGA parameters remain
part of the varying initialization. Both those parameters and the shared
metric networks use rate $3\times10^{-4}g$; the remaining parameters use
rate $6\times10^{-4}/N$. AdamW has moment coefficients $(0.9,0.95)$,
denominator offset $10^{-8}$, weight decay $0.01$, and global gradient
norm clipping at one. Rates are constant. Training uses float32 without
TF32; observation reductions use float64.

The clocks and source normalization are
\begin{equation}
 \tau_{\rm gen}=3\times10^{-4}gT,\quad
 \tau_{\rm body}=6\times10^{-4}T/N,\quad
 v_{\rm clock}=(3\times10^{-4}g)^2T,\quad f=32T/M.
 \label{model:eq:singlepass-reported-clocks}
\end{equation}
Here $M=2,097,152$ blocks and batch size is 32. Holding $g,T$ fixed
keeps the generator clock fixed and changes the body clock. The quadratic
quantity becomes a variance clock only under the covariance assumptions
in Equation~\eqref{model:eq:quadratic-force-clock}.

Each path consumes 2,048 batches of 32 distinct token blocks from one
fixed, uniformly permuted half of the deduplicated RefinedWeb master
corpus. Its finite population contains 262,144 documents and 2,097,152
blocks. Each block supplies a 64-token prefix and one external next-token
target. Thus a trajectory supervises 65,536 distinct positions, visits
4,194,304 prefix tokens, and consumes $1/32$ of its block population.
The objective supervises one target per prefix. The source order is
shared across controls, widths and initialization seeds. The other
master-corpus half is disjoint by document identity. Conditioning on these
finite populations differs from averaging over new master-corpus draws.
The 128 evaluation documents are disjoint from the complete master corpus
by content hash.

The primary field is the native relative centered row energy $R$;
normalized attention entropy is a separate secondary field. Every 64
updates, 64 fixed contexts retain every head in every decoder.
At each fixed context and decoder, the head mean is centered across the
six training realizations before computing its unbiased sample variance.
We report that intensive variance, its susceptibility after multiplication
by $N$, the mean one-head variance, and the off-diagonal covariance in the
same units. Heads, contexts, decoders and observation times are dependent
coordinates of each realization. They are not additional training
replicas. A joint descriptive bootstrap resamples the complete seed block
10,000 times, preserving its control, width, time and source coupling.
Unresolved or boundary peaks remain explicit.

Let $\widehat m_{N,T}(g)$ be the empirical row mean over the complete
initializations, contexts, decoders and heads. The half-mean control
crossings solve
\begin{equation}
 \widehat m_{N,T}(g)/\widehat m_{N,T}(0)=1/2
 \label{model:eq:half-row-control}
\end{equation}
by linear interpolation between neighboring sampled controls; all
downward crossings and their brackets are retained. The denominator
is the zero-generator mean at the same training age. Susceptibility
half-height widths instead use half the sampled susceptibility maximum.
The separate within-run timing label compares that run's mean with
half its own initial mean, as in
Proposition~\ref{model:prop:critical-timing-mixture}.

At saved milestones the predictive observation is the complete vocabulary
law on eight fixed contexts. Its covariance trace uses
$\phi(p)=2\sqrt p$ and sums vocabulary coordinates as in
Proposition~\ref{model:prop:predictive-hellinger}. Initialization and trained
states retain the same units. Independent endpoint observations record
complete common-centroid vectors, complete head-mean operators and eight
fixed orthonormal operator projections per head. They replay the recorded
scalar fields and logits at the saved weights and add no training update.
The common-vector and operator traces use per-coordinate units, averaged
over contexts and decoders. Their complete directions distinguish common
field concentration from concentration of its scalar amplitude.

\subsection{Row contrast, collective directions and training age}
All 168 declared paths completed their 2,048 updates, giving 344,064
scientific updates without a numerical failure. The native replay
qualification matched the full model, Adam and random state bit for bit.
Independent reconstruction checked every reported scalar cell and the
unique source block identities. Complete endpoint observations reproduced
the recorded row fields and logits exactly.

\begin{figure}[htbp]
\centering
\includegraphics[width=\textwidth]{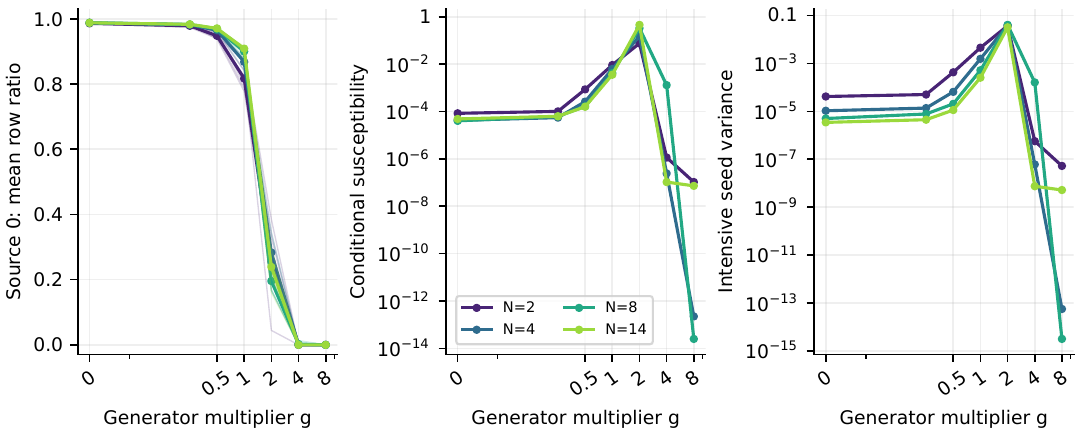}
\caption{Completed single-pass row profiles at 2,048 updates.
Faint curves in the left panel retain the six initialization realizations
at each width. The other panels use context-conditioned replica
covariances with divisor five. Lines join measured controls; they are
not fitted singular curves. A susceptibility increase accompanied by an
order-one intensive variance identifies a common fluctuation component,
without determining whether its limiting law is critical.}
\label{model:fig:critical-coarse-profile}
\end{figure}

At $T=2048$, every sampled row-susceptibility maximum occurs at $g=2$.
For $N=2,4,8,14$, the heights are respectively
$0.0751,0.155,0.329,0.455$, while the corresponding intensive variances
are $0.0376,0.0388,0.0412,0.0325$. The fitted log-size slope of the peak
heights is $0.948$. The intensive values show why this slope cannot be
assigned a critical exponent: the observed peak is compatible with a
common component of nonvanishing variance. At $g=0$, the row ratio stays
near $0.99$ and its susceptibility is between $4.03\times10^{-5}$ and
$8.37\times10^{-5}$. Figure~\ref{model:fig:critical-coarse-profile} retains
both variance normalizations and each realization's mean row profile.

The half-mean row crossing moves toward smaller generator multipliers as
training proceeds. Fits to all resolved positive-control brackets from
256 through 2,048 updates give control-versus-time powers
$-0.532,-0.522,-0.550,-0.554$ for the four widths. A separate milestone estimator uses only
$T=256,512,1024,2048$. Joint resampling of complete seeds gives its
median slopes $-0.542,-0.541,-0.560,-0.558$ and respective descriptive
95\% percentile intervals
$[-0.571,-0.511]$, $[-0.555,-0.525]$,
$[-0.568,-0.552]$ and $[-0.565,-0.550]$.
The coefficient and the interpolation uncertainty depend on the sampled
grid. These are kinetic drift estimates under the fixed source law,
not independent time replicates or universal exponents.
Proposition~\ref{model:prop:kinetic-control-peak} gives an explicit regular
observation law with extensive peaks and a moving control window.
The coarse half-height widths are limited by the control spacing, so
their apparent similarity across sizes cannot resolve a limiting window.

The row decrease corresponds to absolute contrast loss. At $g=4$, mean
total metric energy lies between $1.029$ and $1.034$ across the four
widths, whereas mean centered energy lies between
$5.1\times10^{-6}$ and $2.58\times10^{-3}$.
The pointwise positive-energy identity
$\Delta\log R=\Delta\log C-\Delta\log E$
separates centered energy $C$ from total energy $E$.
The mean denominator change is only $0.0275$--$0.0331$ in logarithmic
units, compared with centered-energy changes from $-25.6$ to $-23.3$.
Every recorded initial/final pair at this control is strictly positive.
Thus denominator growth does not account for the observed suppression.

\begin{figure}[htbp]
\centering
\includegraphics[width=\textwidth]{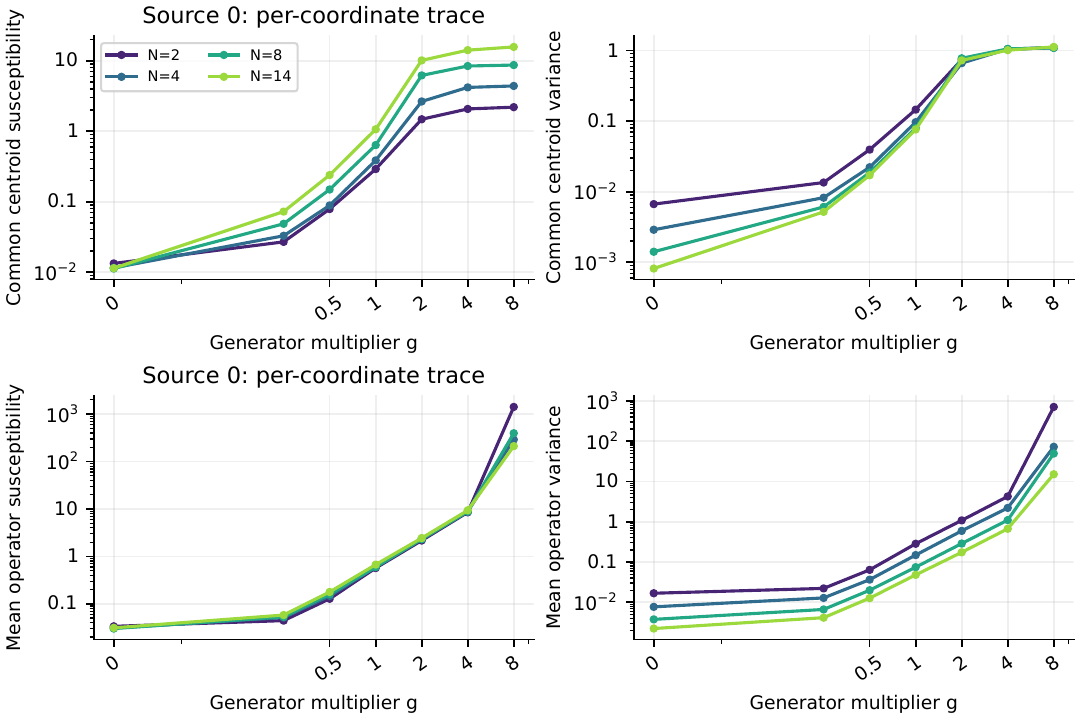}
\caption{Complete common-centroid and head-mean operator observations at
the 168 trained endpoints. Each row uses a fixed per-coordinate covariance
trace, averaged over the same contexts and decoders. Susceptibility
multiplies that trace by $N$. The common-vector direction retains
substantial replica variation in the collapsed region even as its
amplitude concentrates. The first moment of the complete operator has
a different width dependence. These observations require distinct
visibility conditions when connecting a training collective to inference.}
\label{model:fig:critical-complete-fields}
\end{figure}

The complete common centroid remains random after this contrast loss.
At $g=4$, its per-coordinate head-mean variances are
$1.039,1.049,1.055,1.012$ for increasing $N$, giving approximately
extensive common-vector susceptibilities. Its scalar RMS fluctuates
much less. The orientation retained by the complete vector therefore
matters. In contrast, the complete head-mean PLGA operator has
susceptibilities $8.50,8.83,8.80,9.37$ in per-coordinate operator units
at the same control. Its intensive variance decreases substantially
with width. Figure~\ref{model:fig:critical-complete-fields} shows this
difference between observations of the same native states.
Concentration of one projected statistic does not imply concentration
of the joint empirical head law or its nonlinear observations.

The operator normalization has an exact native explanation in
Proposition~\ref{model:prop:native-head-sign-symmetry}. Twelve paired native
CPU cases, using two widths, three initializations and two head-sign
patterns, verified eight successive single-pass updates each.
The recorded forward-tensor differences were exactly zero. Transformed
gradients and saved final model and Adam states agreed byte for byte;
an independent reconstruction checked the complete saved states and RNG.
These checks cost 192 replay updates and add no scientific trajectory. A separate inference check used the 24 trained endpoints at all four
widths, $g\in\{0,2,8\}$ and the first two declared seeds. Its 72 CPU
forward calls gave 48 paired sign comparisons. The recorded metric,
operator, attention and next-token logit tensors agreed byte for byte
after the stipulated transformation; the original parameters and RNG
were preserved. Both branches used the same saved weights on CPU,
and the complete predictive comparison arrays were retained. These
inference checks execute zero training updates. Across all 28 original coarse
endpoint cells, the ratio of the measured operator susceptibility to
mean one-head operator power lies between $0.965$ and $1.057$.
Balanced orthogonal sign quadrature reconstructs the corresponding
projected orbit covariance with maximum absolute error
$2.28\times10^{-13}$. The orbit calculation uses population denominators
and equivalent parameter representatives; it does not augment the six
independent initializations. Thus the signed operator statistic has a
specific symmetry normalization, while invariant row and predictive
observations remain the relevant tests of an observable critical law.

\subsection{A regular conditional signed-operator class}
\begin{figure}[htbp]
\centering
\includegraphics[width=\textwidth]{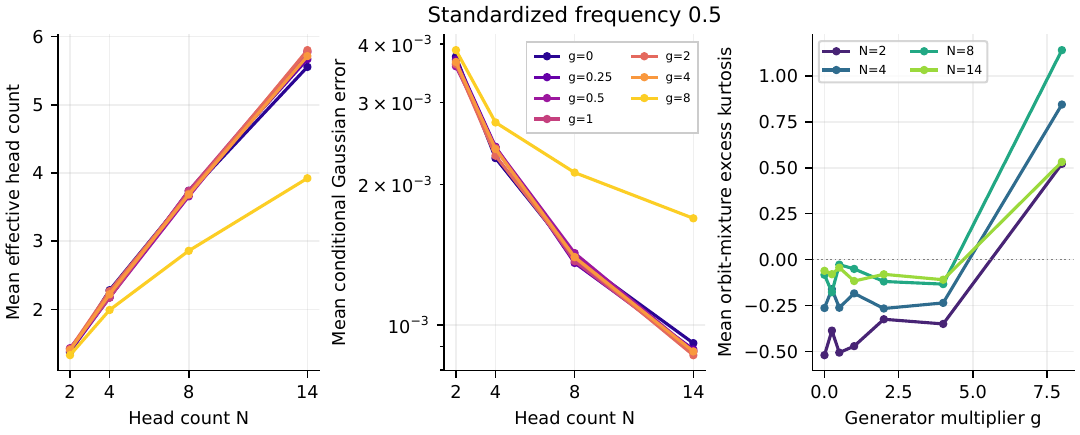}
\caption{Finite conditional signed-operator observations. Effective
head counts quantify the concentration of projection power among heads.
The middle panel compares exact orbit characteristic functions with a
variance-matched Gaussian at standardized frequency $1/2$.
The right panel retains the fourth cumulant of the finite orbit mixture,
normalized separately at each fixed context, decoder and projection
before averaging. All signs refer to equivalent parameter
representatives; no exponent is fitted and no new native path is added.}
\label{model:fig:critical-coarse-sign-limit}
\end{figure}

The native sign law also gives the Gaussian-mixture limit in
Theorem~\ref{model:thm:native-sign-gaussian-mixture}. At every trained endpoint,
we evaluated its exact sign-orbit characteristic function on the eight
fixed orthonormal operator projections, retaining each context and
decoder separately. At $g=2$, the panel-mean errors against the
variance-matched Gaussian at standardized frequency $1/2$ are
$0.003629,0.002306,0.001395,0.000861$ for increasing head count.
The corresponding mean effective-head fractions are
$0.710,0.567,0.462,0.414$. All comparisons inside the stated domain
obey Equation~\eqref{model:eq:sign-characteristic-bound}.
Independent enumeration of every sign on the first fixed panel entry
through $N=14$ reproduces the second and fourth moments and the
characteristic functions; its maximum fourth-moment error is
$2.27\times10^{-14}$. Figure~\ref{model:fig:critical-coarse-sign-limit}
retains every coarse control.

The finite uniform mixture of the six saved sign orbits can have
positive excess kurtosis despite negative within-orbit fourth
cumulants. At $g=8$, its panel-mean excesses are
$0.521,0.844,1.140,0.532$ across the four widths.
Equation~\eqref{model:eq:sign-mixture-fourth-budget} resolves them into a
positive variance-mixture contribution and a negative finite-sign
contribution, with maximum relative budget residual
$7.19\times10^{-16}$. These are conditional symmetry averages of the
same 168 native checkpoints, with no extra independent training
realizations. The regular signed-operator law leaves the invariant
row, common and predictive fluctuation laws to be assessed separately.

\subsection{Adaptive shared force and predictive observations}
\label{model:sec:critical-shared-force}
\begin{figure}[htbp]
\centering
\includegraphics[width=\textwidth]{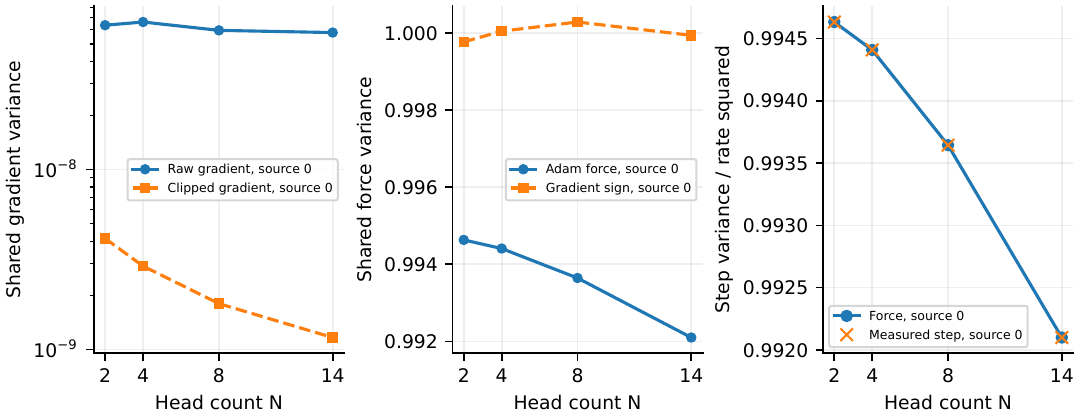}
\caption{Native creation of shared-state variation at the first adaptive
update, reconstructed from the complete unit-control initialization
family. Global clipping reduces the gradient magnitude as width grows,
while the normalized Adam force remains close to the gradient-sign
variance. The right panel compares the force variance with measured
parameter-step variance divided by the squared rate. The measured
arithmetic covariance budget is retained. This observation establishes
a finite native mechanism, without assuming stationary accumulation.}
\label{model:fig:critical-first-force}
\end{figure}

The shared metric networks have 2,648,640 coordinates at every width.
Reconstructing the first 64 updates of all 24 unit-control paths gives
bitwise agreement with the recorded native prefixes and costs 1,536
replay updates; these are not additional scientific trajectories.
The first-step normalized-force variances per shared coordinate are
$0.9946,0.9944,0.9936,0.9921$ as $N$ increases.
The corresponding actual parameter-step variances are approximately
$8.95,8.95,8.94,8.93$ times $10^{-8}$, consistent with multiplication
by $(3\times10^{-4})^2$ as in
Proposition~\ref{model:prop:first-shared-force}.
The RMS arithmetic remainder is less than $8.9\times10^{-6}$ of the
ideal fluctuation scale. Figure~\ref{model:fig:critical-first-force} retains
the raw, clipped and normalized quantities separately.
Coordinates are averaged measurements of each realization, not extra
independent initialization replicates.

The predictive variance in the full-vocabulary square-root embedding
is already about $0.00796$ at initialization at every width.
At the trained $g=4$ endpoints it is
$0.0303,0.0318,0.0416,0.0515$, in the same vocabulary-summed units.
The initial random-readout reference and the trained law therefore both
matter for interpreting predictive susceptibility.
Held-out NLL at $g=0$ is $6.472,6.453,6.406,6.385$, compared with
$6.789,6.806,6.753,6.713$ at the row-fluctuation maximum $g=2$.
Row suppression and predictive risk select different control profiles
within this finite training budget. The row maximum is not identified
with a task optimum or a self-organized training transition.
The complete endpoint table, time profiles and predictive curves appear
in Appendix~\ref{model:app:critical-onepass-details}.

\section{Independent size and training-age comparison}
\label{model:sec:critical-independent-results}

The independent unit is a complete initialization/path, with six fresh
initialization identities per width/gain cell. The independent family uses
the disjoint second partition of one master corpus, $N=4,8,14,24$, and
$g=0,0.75,1,1.25,1.5,1.75,2,2.5,3$. Its architecture, arithmetic,
optimizer, source ordering convention and fixed initial shared generator
match the controlled family above. Context panels and design pairings are
shared; heads and contexts do not add initialization replicas. Each path runs for 4,096 updates,
consuming 131,072 distinct supervised blocks, or $1/16$ of its population,
and processing 8,388,608 input-prefix token positions. The 216 declared
paths completed 884,736 scientific updates without a numerical failure.
The two families together contain 384 paths and 1,228,800 scientific
updates. They condition on two disjoint parts of one master corpus;
they do not represent independent master-corpus draws.

The four model sizes contain approximately $23.4,52.0,109.7,245.1$
million parameters. Recorded worker times sum to 30.17 hours for the
independent family on two RTX 4090 GPUs. Median path times are
$271,328,523,890$ seconds, respectively; the largest recorded CUDA
allocation is 16.33 GiB. The passive recorder retains all 2,648,640
shared metric-network coordinates at initialization, after the first
update and every 256 updates. A separate 128-update native comparison
with and without this recorder preserved all original observations,
model, optimizer and RNG states byte for byte. All final Adam states
are retained for the independent family. Independent scalar reconstruction
checked all 9,360 cells and the distinct source-block identities; its
maximum absolute reduction discrepancy is $8.89\times10^{-16}$.

\subsection{A reproduced finite-time fluctuation region}
At $T=2048$, sampled row-susceptibility maxima occur at
$g=2,1.75,2,2$ for increasing $N$. At $T=4096$ they occur at
$g=1.25,1.5,1.25,1.25$. Every maximum is interior and has observed
half-height brackets on both sides. The endpoint peak heights are
$0.1665,0.2992,0.7277,0.7598$, with intensive variances
$0.0416,0.0374,0.0520,0.0317$. The corresponding cross-head covariances
are $0.0373,0.0362,0.0515,0.0314$. These measurements reproduce a
finite training-age region with substantial shared variation, including
at the new largest width.

The following profiles show normalized row fractions and their ensemble
variance over the six whole initialization paths; multiplying by $N$ defines
susceptibility. Neither the fraction nor its susceptibility is absolute row
energy or a predictive fidelity measure. Control brackets describe grid
resolution; the displayed initialization spread describes the fixed-panel
training ensemble.

\begin{figure}[htbp]
\centering
\includegraphics[width=\textwidth]{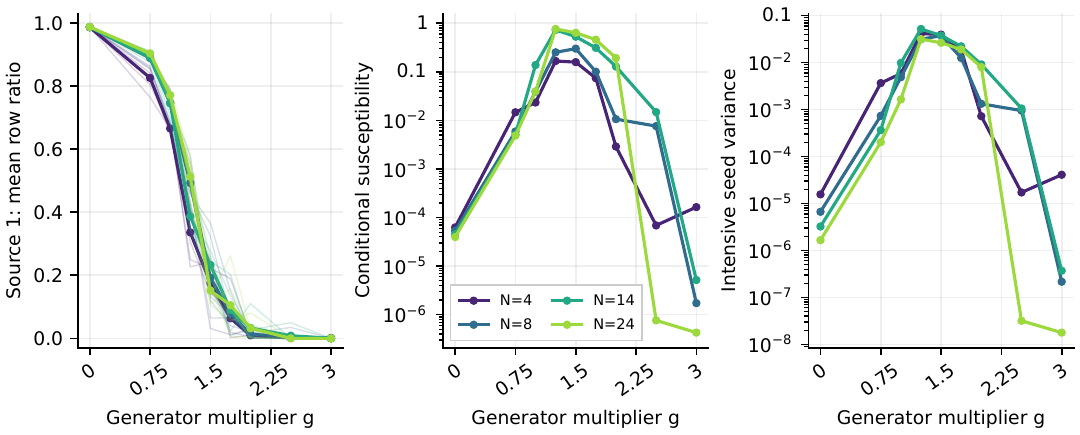}
\caption{Independent single-pass row profiles at 4,096 updates. All six
initializations remain visible, and the intensive variance accompanies
its susceptibility normalization. The sampled maximum shifts between
neighboring controls across widths. Complete paths, rather than heads
or contexts, are the training realizations.}
\label{model:fig:critical-independent-profiles}
\end{figure}

The interpolated half-height widths at $T=4096$ are
$0.617,0.558,0.593,0.705$. Their sequence does not resolve systematic
narrowing with head count. Table~\ref{model:tab:critical-refinement-peaks}
retains the measured control segments behind each interpolant. These
segments express grid resolution and are distinct from initialization
uncertainty. Doubling training age moves and narrows the finite control
region, while the four-width comparison does not identify a limiting
critical window.

Peak-height log-size slopes are $1.035$ at $T=2048$ and $0.922$ at
$T=4096$. They are descriptive. Fitting only the first three widths
and predicting $N=24$ at the latter time gives $1.253$ from a power law,
compared with the observed $0.760$. The regular finite-size reference
predicts $0.695$; the shared-sector references predict $1.152$ and
$1.277$ with nonnegative and signed finite corrections, respectively.
At $T=2048$ the corresponding predictions are $1.254$, $0.674$,
$1.110$ and $1.123$, against $1.016$. The model ordering therefore
depends on training age. Six initializations and four widths do not
select one asymptotic class from those finite fits.

\subsection{Frozen time predictions}
\label{model:sec:critical-time-predictions}
\begin{figure}[htbp]
\centering
\includegraphics[width=\textwidth]{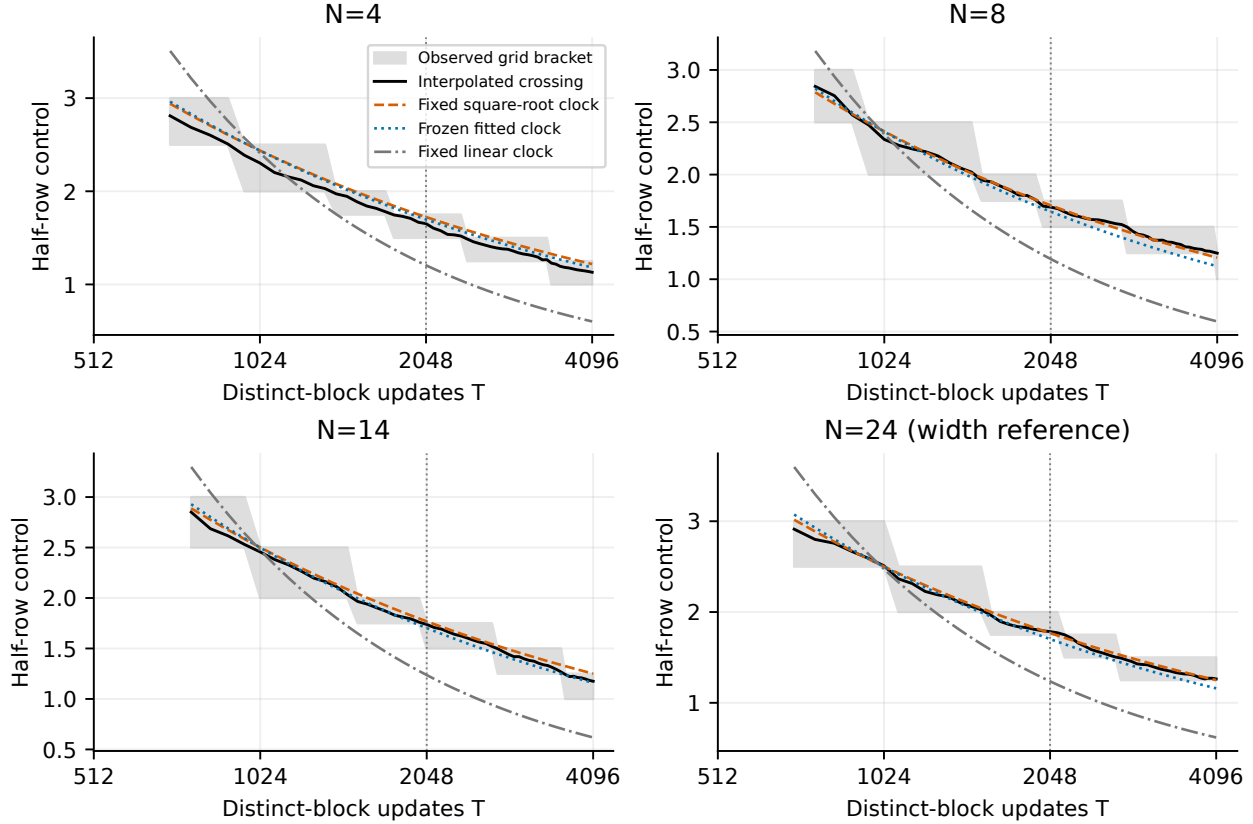}
\caption{Predictions fixed before independent training. Shading retains
the observed control-grid bracket; it is not a confidence interval.
The vertical line marks the largest coarse training horizon. All
resolved independent crossing times are retained, and the new largest
width uses the separately identified reference.}
\label{model:fig:critical-independent-crossings}
\end{figure}

All time-prediction coefficients were fixed using the complete coarse
family before any independent training. The 24-head time prediction
inherits the largest coarse width as an explicitly separate saturation
reference. No independent observation is used to refit these
coefficients. Equation~\eqref{model:eq:half-row-control} gives endpoint
crossings $1.1307,1.2490,1.1754,1.2632$ for increasing width.
The fixed square-root clock predicts them with signed relative errors
$+7.78\%,-3.42\%,+6.39\%,-1.01\%$. The $N=4,8,24$ predictions fall within their observed control
brackets. The $N=14$ prediction exceeds its upper bracket endpoint
by $4.70\times10^{-4}$.
The corresponding linear-clock errors are between $-52.3\%$ and
$-46.6\%$. The complete coefficient-frozen comparison appears in
Table~\ref{model:tab:critical-frozen-crossings} and
Figure~\ref{model:fig:critical-independent-crossings}.

As a descriptive fit to all resolved 64-update observations, the
independent control-time powers are $-0.518,-0.480,-0.523,-0.498$.
The first resolved positive brackets occur at update 704 or 768,
depending on width; earlier controls outside the sampled range remain
unresolved. The 53 or 54 resolved points per width are correlated
observations of the same six trajectories. Thus an approximately
square-root motion is reproducible over this finite age range. It
supports a kinetic control scale and does not establish a critical
exponent, a diffusion coefficient or self-organized attraction to a
critical surface. The exact chronological covariance budget in
Section~\ref{model:sec:critical-shared-accumulation} retains the additional information needed to interpret that scale.
Across all 32 recorded times beyond the coarse training horizon, the
fixed square-root predictions have relative root mean square errors
$7.09\%,2.03\%,4.11\%,1.90\%$ for increasing width. The frozen
empirical-clock errors are $4.55\%,7.17\%,2.62\%,4.78\%$, while the
linear-clock errors are $38.8$--$43.7\%$. The relative ordering of the
square-root and empirically fitted clocks depends on width. These
scores use correlated times from the same six paths per cell.

Joint seed resampling resolves half-height widths in 99.99\% of
recorded endpoint draws. Its descriptive 95\% width intervals at
$T=4096$ are $[0.507,0.701]$, $[0.313,0.607]$, $[0.362,0.848]$ and
$[0.317,0.810]$. The peak control varies across neighboring grid values;
these intervals do not establish a shrinking thermodynamic window.
Exact enumeration of all 46,656 ordered empirical resamples resolves every
profile except the six single-support draws, giving probability $7775/7776$.
The weighted inverse-CDF width intervals appear in
Table~\ref{model:tab:exact-replica-widths}; they retain the finite-width interpretation.
The corresponding sampled-peak slope interval is $[0.704,1.183]$,
and the paired withheld power-prediction/observation interval is
$[0.874,3.025]$. Tables~\ref{model:tab:critical-independent-uncertainty}
and~\ref{model:tab:critical-independent-drift-windows} retain all declared
endpoint profiles and every trailing milestone fit. For the last four
times, the drift median ranges from $-0.608$ at $N=14$ to $-0.436$ at
$N=24$. Thus the useful square-root forecast is a finite kinetic
approximation with visible width and window dependence.

The binary timing decomposition also retains fluctuation structure
within each label. At the endpoint sampled peaks, its between-label
fractions are $0,0,0.096,0.247$ for increasing width. In particular,
the largest-width peak is not exhausted by dividing runs at half their
initial row mean. Continuous timing variation, decoder structure and
within-label common fields remain in the measured covariance.

\subsection{Complete shared-parameter accumulation}
\label{model:sec:critical-shared-accumulation}
All 36 complete shared-coordinate ensembles reconstruct their final
vectors from the saved native states. Initial gradients agree exactly
across the control-paired paths. At $g=2$, first normalized-force
variances are $0.9940,0.9933,0.9917,0.9910$ across increasing width;
raw gradient variances remain between $5.02\times10^{-8}$ and
$6.27\times10^{-8}$. The normalization of the adaptive force therefore
preserves an order-one conditional fluctuation at the first update.

At $T=4096,g=2$, the per-coordinate shared-parameter variances are
$1.070,1.105,1.123,1.103$ times $10^{-3}$. They account for
$99.92$--$99.96\%$ of the recorded mean squared decay-corrected
displacement. The corrected squared-mean estimates are only
$5.72,4.30,9.05,4.47$ times $10^{-7}$.
Across all 32 positive-control endpoint cells, eight corrected
squared-mean estimates are negative. These finite-sample estimates are
retained with their signs; a small estimate does not prove a zero
population mean.

For the registered $g=2$ path, dividing the variance by
$(\eta_0g)^2T$ gives values $0.870$--$0.931$ at $T=1024$,
$0.852$--$0.911$ at $T=2048$ and $0.726$--$0.762$ at $T=4096$.
At the endpoint, this ratio is $0.883$--$0.927$ for $g=1.25$ and
$0.420$--$0.456$ for $g=3$. Approximate quadratic-rate, linear-time
variance accumulation is consequently useful over part of the finite
range, with visible control and age dependence. This supplies a
measured shared-parameter scale accompanying the kinetic row crossing.
It does not identify an autonomous diffusion law.

The complete 256-update covariance partition retains a positive net
cross-block contribution in every positive-control endpoint cell.
Its fraction of the endpoint variance lies between $2.55\%$ and
$3.61\%$. At $g=2$ the four fractions are
$2.88\%,2.64\%,2.88\%,3.61\%$. The diagonal terms include all
within-block dynamics, and the net cross term can conceal cancellation
between its signed entries. Neither this small net fraction nor the
approximately linear variance growth demonstrates independent
one-update forcing. These are covariances over the initialized
trajectories at one fixed source order.

\begin{figure}[htbp]\centering
\includegraphics[width=\textwidth]{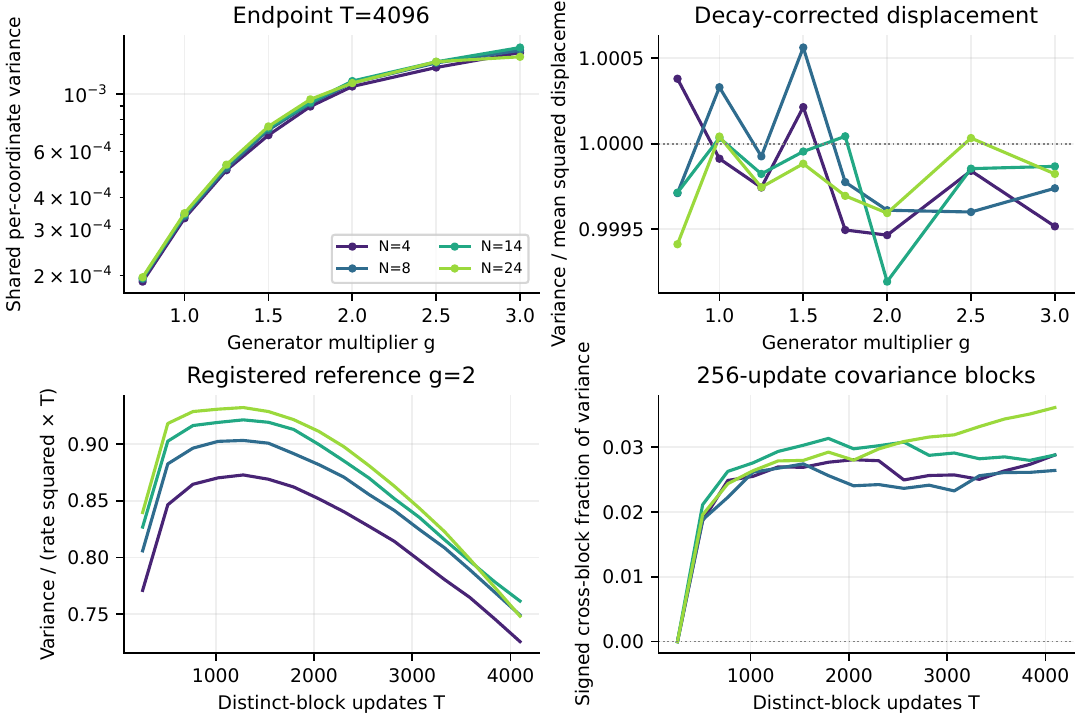}
\caption{Complete shared-parameter accumulation in the independent
family. The displacement is corrected for the registered weight decay.
All positive controls appear at the endpoint, while the time panels use
the fixed reference $g=2$. The signed cross-block fraction refers to the
same 256-update partition as the exact covariance budget.}
\label{model:fig:critical-independent-shared-paths}
\end{figure}

The pairwise variance, squared-mean and complete block identities have
maximum absolute errors below $1.09\times10^{-18}$ in per-coordinate
variance units. The coordinate displacement telescope has maximum
absolute error $1.64\times10^{-15}$. Table~\ref{model:tab:critical-shared-budgets}
and Figure~\ref{model:fig:critical-independent-blocks} retain every endpoint
budget and the complete reference correlation matrices.

\subsection{Absolute contrast and inference observations}
\label{model:sec:critical-absolute-observations}
The independent family reproduces the separation between centered
contrast and total metric energy. At $g=3,T=4096$, mean centered
energies are $1.013\times10^{-3}$, $7.444\times10^{-5}$,
$1.229\times10^{-4}$ and $2.390\times10^{-5}$ across increasing
width, while total energies remain between $1.019$ and $1.027$.
At $N=24$, the mean log centered-energy change is $-27.569$ and the
mean log denominator change is $0.0233$. Every pair in this cell is
strictly positive and above the denominator floor. The log-budget
identity is reconstructed to maximum absolute error
$1.40\times10^{-14}$ across the complete family. This directly supports
absolute row-contrast suppression under single-pass training.

At the endpoint row peaks, the leading eigenvalue accounts for
$46.4\%,59.8\%,50.1\%,55.4\%$ of the five-decoder row-covariance
trace. Multiple decoder directions therefore contribute to the
measured fluctuation region. Near complete collapse, a leading
fraction close to one can instead accompany a very small trace:
at $N=24,g=3$, the fraction exceeds $0.99999$, while susceptibility
is only $4.27\times10^{-7}$. An eigenvalue fraction must be interpreted
with its absolute fluctuation scale.

The complete common-centroid vectors retain large conditional
variation after this contrast suppression. At $g=3$, their
per-coordinate variances are $1.024,1.039,1.030,1.017$ for increasing
width. Their scalar RMS-amplitude variances are only
$3.73,3.42,4.46,2.51$ times $10^{-4}$. The complete direction therefore
remains relevant even when the amplitude varies little. At zero
generator control, the common-vector susceptibilities instead lie
between $0.0114$ and $0.0126$ over the four measured widths.

The complete signed mean operator has endpoint susceptibilities
$10.52,10.33,10.57,10.78$ at $g=3$. At $g=0$ the values are
$0.0326,0.0309,0.0305,0.0305$. Its approximately bounded susceptibility
in each of these controls is consistent with the native sign law and
the zero-control regular-rate bound. It has different units and width
dependence from the common-vector susceptibility, which reaches
$24.42$ at $N=24,g=3$. Figure~\ref{model:fig:critical-independent-fields}
retains the complete control profiles. The separate 216 native
endpoint replays reproduce the recorded scalar fields and complete
next-token logits exactly, with no additional training update.

The full-vocabulary predictive variance has initial values between
$0.00795$ and $0.00797$. At $g=3,T=4096$, its intensive values are
$0.0974,0.0809,0.0949,0.0760$. At the respective row peaks they are
$0.0979,0.0979,0.1399,0.1267$, compared with
$0.1118,0.1018,0.0933,0.1025$ at zero generator control.
Thus large predictive variation also occurs outside the row peak.
Its susceptibility includes a factor of $N$ and must retain this
nonzero initialization reference and the same embedding units.
Figure~\ref{model:fig:critical-independent-predictions} includes all controls.
The independent Hellinger pair identity has maximum normalized error
$7.51\times10^{-16}$.

Mean held-out next-token NLL on the fixed 128-context panel decreases
from approximately $10.38$ initially to $6.32$--$6.41$ at $g=0$,
$6.55$--$6.66$ at the row peaks and $6.70$--$6.78$ at $g=3$.
These finite training paths therefore improve prediction across
multiple row regimes. The evaluation panel, one-target objective and
short single-pass horizons delimit this result. The row fluctuation
maximum does not select the smallest measured held-out loss.

\subsection{Independent native sign-orbit checks}
\label{model:sec:critical-sign-checks}
Six trained endpoints at the new largest width, using $g=0,1.25,3$
and the first two declared initializations, passed the separate native
sign action. Two sign patterns give 12 comparisons in 18 CPU forward
calls. Both branches use the same saved weights on CPU. The recorded
metric, operator, attention and next-token logit tensors agree byte
for byte after the stated transformation; applying the action twice
restores the parameters and preserves the RNG. These comparisons add
zero training updates.

Across all 36 independent endpoint cells, the original unbiased
operator susceptibility divided by mean one-head power lies between
$0.9583$ and $1.0592$. The exact projected sign quadrature reconstructs
its orbit second moment with maximum absolute error
$1.78\times10^{-15}$ in the declared projection units. The separate
pairwise reconstruction of complete collective covariances has maximum
absolute error $8.89\times10^{-15}$, and the common/centered energy
identity has maximum normalized error $3.26\times10^{-16}$.

At $g=2$, the mean conditional characteristic-function errors against
the variance-matched Gaussian at standardized frequency $1/2$ are
$0.002433,0.001384,0.000854,0.000523$ for increasing width. Mean
effective-head counts are $2.16,3.71,5.83,9.35$; the fixed projections
therefore retain appreciable finite-head corrections. Every comparison
inside the theorem's stated domain satisfies its bound. Independent
sign enumeration through $N=14$ reproduces the fourth moment with
maximum absolute error $2.76\times10^{-14}$. The full finite-mixture
fourth-cumulant budget has maximum relative residual
$7.66\times10^{-16}$.
At $N=24,g=2$, its mean excess is $-0.0314$, combining a positive
variance-mixture contribution $0.2065$ with a negative sign contribution
$-0.2380$. Figure~\ref{model:fig:critical-independent-sign-limit} and its
complete table retain all controls. These are exact finite symmetry
averages of the six saved orbits and supply no new training realizations.
They support the regular signed-operator normalization without
identifying a unique limiting covariance law.

The exploratory sign-invariant operator energy has susceptibilities
$0.897,0.905,1.070,0.788$ at $g=3$. The corresponding per-coordinate covariance traces
of all 64 ordered products in the fixed eight-projection panel give
susceptibilities between $98.7$ and $104.3$. These quadratic fields
have their own squared observation units. Their finite width behavior
also differs from the complete common-vector field; a selected moment
panel cannot determine concentration of the full empirical head law.

\section{Measured single-pass dynamics on a matched physical clock}
\label{model:sec:matched-clock-results}

We measure a native family that fixes physical training time, optimizer
memory and source-consumption fraction together. Head counts are
$N\in\{4,8,14,24\}$, generator controls are $g\in\{0,1,1.5,2\}$,
and each condition has three independent complete wide initializations.
Controls and ages share each initialization, corpus order and common
metric-generator initialization. Thus the experiment has 48 trajectories;
its 64 contexts and four observation ages are paired measurements.
It conditions on one realized RefinedWeb corpus and order.

At width $N$, the physical step is $h_N=(128N)^{-1}$, the horizon is
$T_N=128N$, and the nested reservoir contains $M_N=16384N$ distinct
master documents. A batch uses $B=32$ previously unused document blocks,
each consisting of the first 64 tokens as context and the following token
as its sole supervised target. Consequently $BT_N/M_N=1/4$ and
$M_Nh_N=128$. The entire family uses 76,800 scientific optimizer updates.
Every trajectory traverses distinct blocks once. Corpus identities and
order are fixed before execution; the nested resource family also
specifies which documents would remain after the observed endpoint.
The held-out pool contains 16 calibration and 64 assessment documents,
content-disjoint from training and previously reserved panels.

The five-decoder architecture has head dimension 64. AdamW uses
$\beta_{1,N}=0.9^{14/N}$, $\beta_{2,N}=0.95^{14/N}$,
generator rate $0.0003g\,14/N$, and remaining-parameter rate $0.0006/N$.
The denominator offset is $10^{-8}$, decoupled decay is $0.01$ and
global gradient clipping has threshold one. Both moment tensors start
at zero. Thus both rates per unit
physical time and the exponential memory times are fixed across widths.
The normalization of the complete wide initialization is the same as
in the preceding native families. Observations occur at
$\tau\in\{0,1/4,1/2,1\}$. At each age we retain native logits over
all 32,000 vocabulary entries and every decoder/head row field.

For a head matrix $A$, the measured row coordinate is
$\|A-\mathbf 1\bar a^\top\|_F^2/\|A\|_F^2$, where $\bar a$
is its row mean. The stored zero-denominator safeguard is $10^{-30}$.
The reported row coordinate averages the five decoders and $N$ heads.
The operator coordinate averages $\|G\|_F/64$ over the same indices.
For each fixed context, susceptibilities are $N$ times the unbiased
variance across the three complete initialization identities, followed
by context averaging. Predictive variance uses the full-vocabulary
embedding $2\sqrt p$. External-target NLL instead evaluates the
held-out next token. These are different observables of the same
conditioned training/emission law.

At $N=8,24$ and $g=0,1.5$, a 16-context mean operator cache is calibrated
at every measured state. Each later state also receives its own unchanged
initial cache. The initialization cell is reported once per width;
all 26 distinct state/age/policy cells remain in the result tables.
Centered predictive RMS and KL assess fidelity to native predictions,
while the signed NLL change assesses target loss. The fixed targets are
$R\leq0.25$ and mean KL $\leq0.03$ nats. Individual-context exceedances
remain explicit. If $V_c>0$ is the native seed variance and $V_{e,c}$
the residual seed variance at fixed context $c$, the aggregate obeys
\[
 R^2=\frac{\sum_c V_{e,c}}{\sum_c V_c}
 =\sum_c\frac{V_c}{\sum_jV_j}\,r_c^2,
 \qquad r_c^2=V_{e,c}/V_c.
\]
Thus the reported RMS is weighted by native predictive variance;
it is not the unweighted mean of context ratios. The independent
replication unit is the complete initialization, conditional on this
fixed context panel. These acquisitions do not measure latency.

\begin{table}[htbp]\centering\small
\caption{Completed matched-clock endpoints at physical age one. Means and susceptibilities condition on the frozen corpus/order and common generator. Each cell uses three complete wide initializations and 64 fixed contexts. $\chi_{\rm row}$ and $\chi_{\rm pred}$ use the unbiased seed divisor two, then average contexts. NLL is in nats.}
\label{model:tab:matched-clock-endpoints}
\begin{tabular}{@{}rrrrrr@{}}
\toprule $N$ & $g$ & Row contrast & $\chi_{\rm row}$ & $\chi_{\rm pred}$ & Target NLL\\\midrule
4 & 0 & 0.98745 & $1.19\mathord{\times}10^{-5}$ & 0.20349 & 7.79289\\
4 & 1 & 0.51409 & 0.00123 & 0.06227 & 7.83244\\
4 & 1.5 & 0.03466 & 0.00579 & 0.04404 & 7.85363\\
4 & 2 & $2.22\mathord{\times}10^{-4}$ & $1.84\mathord{\times}10^{-6}$ & 0.04832 & 7.84221\\
8 & 0 & 0.98741 & $1.39\mathord{\times}10^{-5}$ & 0.47548 & 7.71573\\
8 & 1 & 0.83087 & 0.00379 & 0.23015 & 7.82309\\
8 & 1.5 & 0.40805 & 0.02609 & 0.19239 & 7.84719\\
8 & 2 & 0.04625 & $6.79\mathord{\times}10^{-4}$ & 0.22221 & 7.82769\\
14 & 0 & 0.98836 & $1.07\mathord{\times}10^{-5}$ & 1.08051 & 7.49746\\
14 & 1 & 0.92346 & 0.00109 & 1.15266 & 7.71479\\
14 & 1.5 & 0.75176 & 0.02651 & 1.02704 & 7.71452\\
14 & 2 & 0.36871 & 0.25460 & 0.79131 & 7.76598\\
24 & 0 & 0.98825 & $9.67\mathord{\times}10^{-6}$ & 2.08799 & 7.37393\\
24 & 1 & 0.94986 & $3.77\mathord{\times}10^{-4}$ & 2.41481 & 7.54268\\
24 & 1.5 & 0.89595 & $8.03\mathord{\times}10^{-4}$ & 2.47294 & 7.61296\\
24 & 2 & 0.77225 & 0.01883 & 2.32881 & 7.66309\\
\bottomrule\end{tabular}\end{table}
\begin{figure}[htbp]\centering
\includegraphics[width=\textwidth]{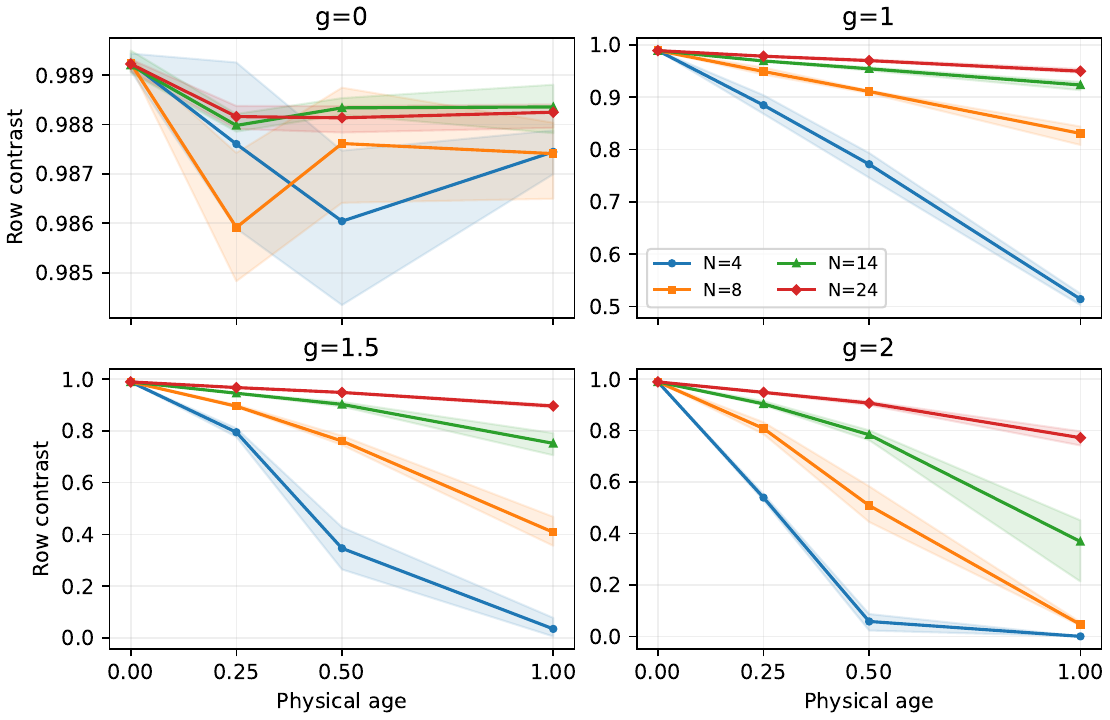}
\caption{Completed row dynamics under a matched physical clock.
Each line is a mean over three complete wide initializations and 64
fixed contexts. Shading is the minimum-to-maximum seed range, not a
confidence interval. Controls and ages are paired.}
\label{model:fig:matched-clock}
\end{figure}
\begin{figure}[htbp]\centering
\includegraphics[width=\textwidth]{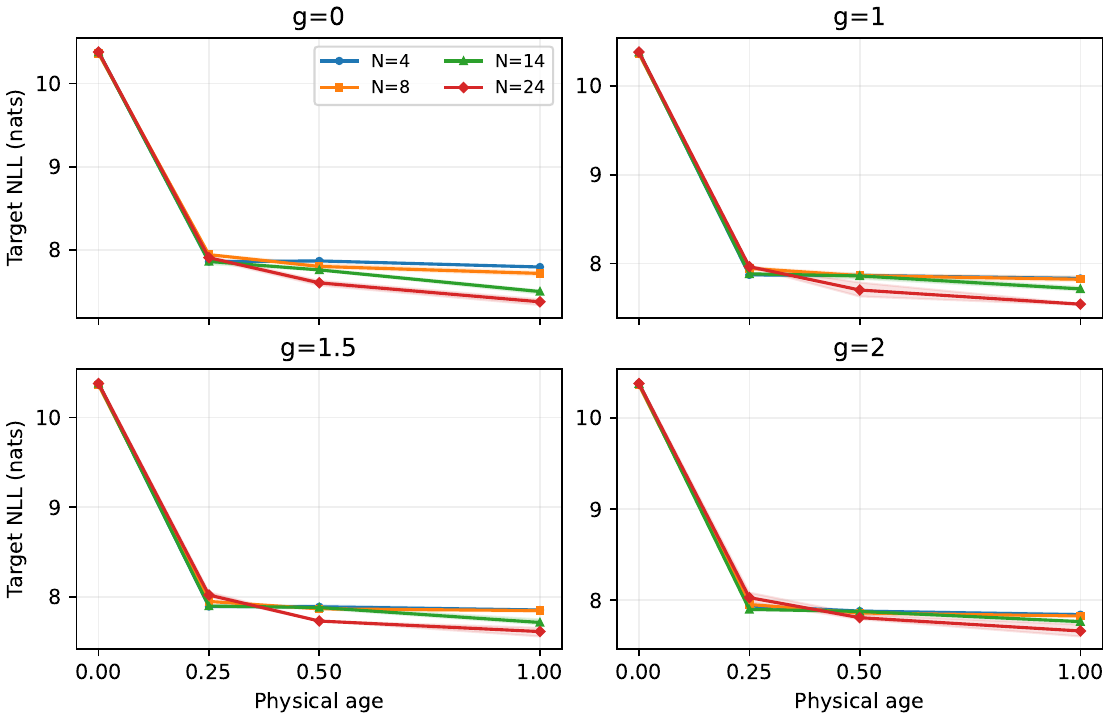}
\caption{External-target loss along the same matched-clock trajectories.
Means and seed ranges use the same fixed assessment contexts as
Figure~\ref{model:fig:matched-clock}. Target loss is distinct from fidelity
to native predictions under an operator-cache intervention.}
\label{model:fig:matched-clock-nll}
\end{figure}

At physical age one, the zero-generator controls retain row coordinates
between $0.9874$ and $0.9884$. Increasing generator control gives a
width-dependent response. At $g=1.5$, endpoint coordinates are
$0.0347$, $0.4081$, $0.7518$ and $0.8959$ in increasing width order;
at $g=2$ they are $0.00022$, $0.0463$, $0.3687$ and $0.7723$.
Thus matching coordinate rates, exponential memory and consumed fraction
does not collapse the finite row curves onto a common physical-time
trajectory. The observation is consistent with retaining the generated
matrix fluxes in Proposition~\ref{model:prop:collective-clock}; a uniform
coordinate-force envelope alone does not fix them. The largest recorded
adaptive force is $1.15593$.

External-target NLL decreases in all sixteen width/control conditions,
from initial values $10.3604$--$10.3805$ to endpoints
$7.3739$--$7.8536$ nats. At each measured width the $g=0$ condition
has the lowest endpoint mean NLL on this panel, while positive generator
control drives substantially stronger row suppression. Row contrast and
predictive quality therefore measure distinct aspects of training.
The endpoint predictive susceptibilities span $0.0440$--$2.4729$;
three initialization identities per condition support a finite paired
comparison, not a thermodynamic exponent fit. The matched-clock family
is distinct from the kinetic-scaling family in
Section~\ref{model:sec:singlepass-critical-observations}.

All fourteen state-calibrated cache cells meet both fixed aggregate
accuracy targets. Relative centered RMS ranges from $0.00945$ to
$0.19509$, and mean KL is at most $0.000696$ nats. The signed target
NLL changes range from $-0.001355$ to $0.000927$ nats.
Individual-context behavior remains heterogeneous: up to eight of
64 contexts exceed the RMS target, with a maximum of $0.77479$.
These measurements include initial, intermediate and trained states
across different row regimes. Six of twelve unchanged initial-cache
cells meet the aggregate targets; their maximum RMS is $1.04354$ and
maximum mean KL is $0.07441$ nats. The full policy comparison and its
operator scatter, displacement and signed-cross budgets appear in
Tables~\ref{model:tab:matched-clock-cache} and~\ref{model:tab:matched-clock-operator}.
The evidence supports state-conditioned inference reduction within the
specified family, without identifying row suppression as its sole cause.

The 48 trajectories require $3.391$ aggregate GPU worker-hours on two
24-GiB RTX 4090 devices, with maximum allocated memory $16.329$ GiB.
They add 2,400 scientific observation forwards alongside the 76,800
training forwards. Separate largest-width execution qualification and
current-source replay each use three optimizer updates and six forwards;
these are excluded from scientific updates. Independent CPU reduction
of every retained field and cache cell agrees within $5.60\times10^{-15}$.

\section{Finite row-energy transport in recorded native continuations}
\label{model:sec:collective-flux-results}
We resolve the row map into its signed cross and quadratic increments
using eight recorded native continuations of the matched-clock endpoints:
four widths, $g\in\{1.5,2\}$, and the first retained complete
initialization. Each continuation consumes exactly the next 32 unused
source documents and advances the full incoming AdamW state once.
The readout uses the first eight assessment contexts already paired to
the parent trajectory. This adds eight scientific updates and 40 native
forwards, including finite parameter-displacement controls. The analysis
is an exact decomposition constructed from the recorded predecessor and
successor fields, rather than an independently forecast endpoint.

\begin{table}[htbp]\centering\small
\caption{Exact finite row-flux decomposition of eight recorded next-block updates. Each row averages eight paired contexts, five decoders and all heads of one retained initialization. $\Delta\bar u=\bar\ell+\bar q$, where $\bar\ell$ is the finite cross term with outgoing-energy denominator; $s=\overline{\|D\|_F^2/\|A\|_F^2}$. These are transition measurements, not independently predicted endpoints.}
\label{model:tab:collective-row-flux}
\begin{tabular}{@{}rrrrrr@{}}
\toprule $N$ & $g$ & $\Delta\bar u$ & $\bar\ell$ & $\bar q$ & $s$\\\midrule
4 & 1.5 & $7.260\mathord{\times}10^{-3}$ & $-3.917\mathord{\times}10^{-3}$ & $1.118\mathord{\times}10^{-2}$ & $2.380\mathord{\times}10^{-2}$\\
4 & 2 & $-2.542\mathord{\times}10^{-9}$ & $-1.956\mathord{\times}10^{-8}$ & $1.702\mathord{\times}10^{-8}$ & $1.884\mathord{\times}10^{-3}$\\
8 & 1.5 & $2.204\mathord{\times}10^{-4}$ & $-5.089\mathord{\times}10^{-3}$ & $5.309\mathord{\times}10^{-3}$ & $9.040\mathord{\times}10^{-3}$\\
8 & 2 & $1.540\mathord{\times}10^{-3}$ & $3.392\mathord{\times}10^{-4}$ & $1.201\mathord{\times}10^{-3}$ & $2.064\mathord{\times}10^{-3}$\\
14 & 1.5 & $-4.832\mathord{\times}10^{-4}$ & $-1.838\mathord{\times}10^{-3}$ & $1.354\mathord{\times}10^{-3}$ & $3.639\mathord{\times}10^{-3}$\\
14 & 2 & $9.604\mathord{\times}10^{-4}$ & $-8.888\mathord{\times}10^{-4}$ & $1.849\mathord{\times}10^{-3}$ & $4.758\mathord{\times}10^{-3}$\\
24 & 1.5 & $-3.885\mathord{\times}10^{-5}$ & $-1.208\mathord{\times}10^{-4}$ & $8.191\mathord{\times}10^{-5}$ & $1.059\mathord{\times}10^{-3}$\\
24 & 2 & $5.432\mathord{\times}10^{-5}$ & $-2.717\mathord{\times}10^{-4}$ & $3.260\mathord{\times}10^{-4}$ & $1.618\mathord{\times}10^{-3}$\\
\bottomrule\end{tabular}
\end{table}
Across all 4,000 head/decoder/context pairs, the finite identity in
Proposition~\ref{model:prop:collective-clock} agrees with directly measured
row changes within $4.57\times10^{-16}$. An independent reconstruction using
the complementary common-row energy agrees within $1.56\times10^{-13}$.
These 4,000 pairs are inner measurements of eight states, not independent
training replicas. Incoming fields replay exactly and each continuation
restores its actual post-update state after the parameter controls.

The quadratic contribution is positive in every averaged cell and is
comparable to, or larger than, the signed cross contribution. For example,
$N=4$, $g=1.5$ has $\bar\ell=-0.00392$ and $\bar q=0.01118$,
yielding a net row increase of $0.00726$; at $N=24$, $g=1.5$,
$\bar\ell=-0.000121$ and $\bar q=0.0000819$ yield a small decrease.
At $N=4$, $g=2$, both contributions are small in absolute units and
partly cancel. The relative squared matrix increments span
$0.00106$--$0.02380$. Thus the native finite transition requires its
quadratic and normalization contributions even when the parameter
update obeys a uniform force envelope. The measurement validates the
complete finite energy budget for these transitions. It does not supply
a closed successor law for the fluxes or establish their limiting
scaling. Chronological compatibility remains the exact telescoping and
cross-time statement of Proposition~\ref{model:prop:collective-clock}.

The finite cross term $\bar\ell$ uses the outgoing energy denominator;
its stored analysis key is \texttt{linear\_flux}. It is distinct from
both a matrix directional derivative and a parameter derivative.
The exact finite terms reconstruct all measured increments. A separate
integration of recorded parameter-path derivatives meets the absolute
increment tolerance $10^{-5}$ in three of eight cases at 32 intervals.
All cases and six nested resolutions appear in
Section~\ref{model:app:row-quadrature}. These diagnostics delimit the tested
approximation; the finite energy identity requires neither differentiation
nor a quadrature convergence assumption.

\section{Executed multistep chronological transport}
\label{model:sec:row-path-results}
Sixteen further native continuations test finite chronological transport
on the two retained initialization identities not used in the one-step
row diagnostic. The design includes every width $N\in\{4,8,14,24\}$,
both controls $g\in\{1.5,2\}$ and both identities. Each path starts at
the matched physical age one, restores its actual parameters, Adam
moments, bias clock and random state, and consumes eight consecutive
next-unused batches of 32 distinct RefinedWeb documents. All paths use
the same eight assessment contexts as the one-step diagnostic. The
protocol fixes the complete grid and block lengths $1,2,4,8$ before
acquisition. The added physical duration is $8/(128N)$, so these
continuation windows have different physical lengths across widths.
They measure finite blocking at fixed update counts and do not estimate
equal-time width scaling. No future flux is an incoming predictor.

The experiment adds 128 scientific updates and 272 native forwards,
with zero new independently pretrained models. The largest workload
runs first and qualifies the exact observation/continuation program;
it is counted once in the scientific total. Aggregate worker time is
122.84 seconds, with peak allocated GPU memory 16.325 GiB. One worker
per RTX 4090 is used. These costs include the short worker's loading,
continuation, observation and saved-array production; they exclude
protocol hashing and the already completed parent pretraining.

\begin{table}[htbp]\centering\small
\caption{All eight-update row-path outcomes. Seed labels 2 and 3 are the two retained initialization identities, paired across conditions. Each entry averages eight contexts, five decoders and all heads. $L_8=\sum_k\bar\ell_k$ and $Q_8=\sum_k\bar q_k$ are accumulated finite terms; $V_8$ is the matrix directional derivative sum. No parameter derivative or successor forecast is used.}\label{model:tab:row-path-endpoints}
\setlength{\tabcolsep}{3pt}
\begin{tabular}{@{}rrrrrrr@{}}
\toprule $N$ & $g$ & Seed & $\Delta\bar u$ & $L_8$ & $Q_8$ & $V_8$\\\midrule
4 & 1.5 & 2 & $8.195\mathord{\times}10^{-3}$ & $-0.1019$ & $0.1101$ & $-0.1019$\\
4 & 1.5 & 3 & $7.592\mathord{\times}10^{-3}$ & $-9.059\mathord{\times}10^{-3}$ & $1.665\mathord{\times}10^{-2}$ & $-9.059\mathord{\times}10^{-3}$\\
4 & 2 & 2 & $3.933\mathord{\times}10^{-4}$ & $-5.642\mathord{\times}10^{-4}$ & $9.575\mathord{\times}10^{-4}$ & $-5.653\mathord{\times}10^{-4}$\\
4 & 2 & 3 & $4.608\mathord{\times}10^{-6}$ & $-1.967\mathord{\times}10^{-6}$ & $6.574\mathord{\times}10^{-6}$ & $-1.965\mathord{\times}10^{-6}$\\
8 & 1.5 & 2 & $3.318\mathord{\times}10^{-3}$ & $-3.629\mathord{\times}10^{-2}$ & $3.961\mathord{\times}10^{-2}$ & $-3.633\mathord{\times}10^{-2}$\\
8 & 1.5 & 3 & $-9.776\mathord{\times}10^{-3}$ & $-6.853\mathord{\times}10^{-2}$ & $5.875\mathord{\times}10^{-2}$ & $-6.853\mathord{\times}10^{-2}$\\
8 & 2 & 2 & $6.963\mathord{\times}10^{-3}$ & $-3.485\mathord{\times}10^{-3}$ & $1.045\mathord{\times}10^{-2}$ & $-3.469\mathord{\times}10^{-3}$\\
\bottomrule\end{tabular}\end{table}

All 64 path/scale cells satisfy the exact finite map and aligned
chronological composition. The largest float64 residual is
$5.94\times10^{-16}$ across 120,000 aligned block matrix pairs; the
64,000 one-step pairs are contained in that total. Independent
common-row-energy reduction and explicit cross-time pair products
reproduce the reported quantities within $4.45\times10^{-14}$.
These are deterministic reconstruction accuracies on the recorded
float32 native fields, not enclosures of the native real-arithmetic
program. Every incoming row field replays its parent measurement.

The accumulated quadratic term is positive in all sixteen averaged
paths. At $N=4,g=1.5$, initialization 2, the cross contribution
$-0.101882$ and quadratic contribution $0.110077$ give the much
smaller row increment $0.0081953$. The matrix directional derivative
sum is $-0.101899$. Keeping the finite quadratic term therefore changes
the sign of the accumulated prediction for this recorded path.
Across all paths, the absolute error of the averaged matrix-derivative
sum ranges from $6.57\times10^{-6}$ to $0.110095$.
All one-step matrix pairs obey $\delta<1$, with maximum $0.85606$,
and all pathwise errors satisfy Corollary~\ref{model:cor:row-path-error}.
The derivative sum has the opposite sign from the exact increment in
7 of 16 paths. The sufficient averaged bounds range from $0.0188183$
to $4.86111$. Their validity and the condition $\delta<1$ therefore
do not establish a quantitatively small error or a vanishing width-limit
error. This comparison uses realized displacements and is retrospective.

At block length eight, the mean normalized cross-time matrix energy
ranges from $-0.07036$ to $0.07081$. This signed contribution is part
of the blocked squared displacement; it cannot be replaced by a sum
of individual squared increments. Complete block outcomes, separate
initialization rows, error budgets and target NLL are retained in
Section~\ref{model:app:row-path-details}. The executed paths support exact
finite transport and its conditional differential error budget. They
do not identify an autonomous flux law or add population evidence for
critical exponents.

\section{Equal physical-duration finite transport}
\label{model:sec:equal-time-flux}
The two retained initialization identities are continued from physical age
one for $T_N^{\rm cont}=8N$ updates at each $N\in\{4,8,14,24\}$ and
$g\in\{1.5,2\}$. Each path has duration $1/16$ on the matched clock
$h_N=1/(128N)$. The consumed reservoir fraction increases from $1/4$
to $17/64$ at every width. The remaining-source cursor, optimizer moments, bias clock
and random state are restored. Every supervised batch contains 32 next-unused
RefinedWeb blocks and no block repeats within a trajectory. The same eight
proper-prefix assessment contexts are retained. The complete grid, duration,
aligned block lengths $N,2N,4N,8N$ and numerical tolerance were frozen before
acquisition. This is a finite comparison conditioned on the shared corpus
order and assessment panel, with two initialization identities per condition.
It adds no independently pretrained model.

All 1,600 continuation updates and 3,216 native forward calls were executed.
The first eight updates of each path replay the corresponding fixed-count
window. Consequently 128 executed updates are replay and 1,472 extend the
observed source histories. Scientific totals must not count the overlap twice.
The widest complete scientific path qualifies the actual workload first and
is counted once. Full native matrices, source indices, clocks and target NLL
are retained outside the compact source package.

All 16 paths and 64 aligned scale cells satisfy the finite transport
and chronological energy-map identities. The largest reconstruction
residual is $2.5535\mathord{\times}10^{-15}$; independent common-row-energy reduction differs by at most $3.2969\mathord{\times}10^{-11}$.
The census contains 1,210,560 matrix pairs, including 1,090,560 one-step pairs. The minimum observed outgoing/incoming energy ratio is 0.97323.
The accumulated quadratic contribution is positive in 16 of 16 paths. The averaged derivative sum has the opposite sign to the exact increment in 4 of 16 paths; its absolute error ranges from $1.5285\mathord{\times}10^{-5}$ to 0.40378.
Where applicable, the conservative averaged path budgets range from 0.16282 to 725.53. These budgets do not imply a small differential error.
Aggregate worker time is 485.62 seconds and peak allocated memory is 16.325 GiB. These measurements include native loading, continuation, observation and raw-array production, and exclude parent pretraining and external protocol hashing.
The maximum matrix discrepancy on the 128 replayed prefix updates is 0.
\begin{table}[htbp]\centering\scriptsize
\caption{Complete equal-duration endpoints. Seeds 2 and 3 denote the two retained initialization identities. $V$ is the recorded matrix-derivative sum and $Q$ is the accumulated finite quadratic term. The NLL is the proper-prefix mean on the same eight contexts.}\label{model:tab:equal-time-flux-endpoints}
\begin{tabular}{@{}rrrrrrrr@{}}\toprule $N$ & $g$ & Seed & $\Delta\bar u$ & $V$ & $Q$ & NLL before & NLL after\\\midrule
4 & 1.5 & 2 & -0.010144 & -0.41392 & 0.40383 & 8.8246 & 8.8799\\
4 & 1.5 & 3 & 0.017474 & -0.098535 & 0.11624 & 8.8041 & 8.8696\\
4 & 2 & 2 & 0.0018259 & -0.02336 & 0.025193 & 8.7945 & 8.8846\\
4 & 2 & 3 & $-3.5566\mathord{\times}10^{-8}$ & $-1.5321\mathord{\times}10^{-5}$ & $1.5285\mathord{\times}10^{-5}$ & 8.8078 & 8.8324\\
8 & 1.5 & 2 & -0.039542 & -0.42833 & 0.38874 & 8.8554 & 8.8817\\
8 & 1.5 & 3 & -0.12138 & -0.49867 & 0.37723 & 8.8174 & 8.8668\\
8 & 2 & 2 & 0.00095385 & -0.093625 & 0.094406 & 8.7753 & 8.8497\\
8 & 2 & 3 & 0.0067921 & -0.19279 & 0.19945 & 8.7907 & 8.897\\
14 & 1.5 & 2 & -0.027214 & -0.13667 & 0.10944 & 8.7848 & 8.8992\\
14 & 1.5 & 3 & -0.0085484 & -0.071249 & 0.062693 & 8.684 & 8.8672\\
14 & 2 & 2 & -0.049624 & -0.33099 & 0.28142 & 8.6987 & 8.8106\\
14 & 2 & 3 & -0.081913 & -0.39941 & 0.31751 & 8.8173 & 8.8733\\
24 & 1.5 & 2 & -0.01192 & -0.023986 & 0.012066 & 8.8335 & 8.72\\
24 & 1.5 & 3 & -0.0050393 & -0.017585 & 0.012545 & 8.91 & 8.7037\\
24 & 2 & 2 & -0.027747 & -0.084894 & 0.057146 & 8.8663 & 8.8777\\
24 & 2 & 3 & -0.036733 & -0.1025 & 0.065759 & 8.9831 & 8.7079\\
\bottomrule\end{tabular}\end{table}

All increments satisfy the differential-bound premise on 15 of the 16
complete paths. At $N=4,g=1.5$, initialization 3, the maximum relative
matrix increment is $1.17343$, so no whole-path application of
Corollary~\ref{model:cor:row-path-error} is asserted for that case. The
positive-energy finite map and finite-flux error formula still apply.
The complete budget table retains this domain distinction. The independent
comparison uses the common-row-energy formula, with absolute tolerance
$3\times10^{-12}$ times the larger of one and the compared magnitude; its
largest absolute difference occurs in a conservative accumulated bound.

The evidence tests finite energy transport and associative chronological
blocking at a common physical duration. It does not estimate a limiting
critical exponent or an autonomous conditional flux law. The finite-flux
sensitivity check perturbs the realized coordinates synthetically; its
certificate tests the error formula, not accuracy of an incoming predictor.
The conditional kernel and predictive-emission requirements in
Proposition~\ref{model:prop:finite-flux-forecast} remain separate. All paths and all
64 aligned scale cells, including approximation errors, are retained in the
compact evidence records and Appendix~\ref{model:app:row-path-details}.

A frozen synthetic perturbation of each of the $1{,}090{,}560$ one-step
flux triples satisfies Proposition~\ref{model:prop:finite-flux-forecast}
within $4.15\times10^{-17}$ in the computed error-minus-bound diagnostic.
This checks numerical sensitivity on the positive-energy domain and
provides no measurement of trained forecast accuracy.

The declared signed temporal-energy outcome uses Eq.~\eqref{model:eq:temporal-energy}.
The acquisition protocol named the outcome in prose. The precise
incoming-energy convention was specified during post-acquisition
reconstruction, before the complete table was generated; it is not
claimed as a separately frozen normalization rule. We first normalize
each coordinate and aligned block, then average over all eight fixed
contexts, five layers, $N$ heads and aligned blocks of $fN$ updates,
where $f\in\{1,2,4,8\}$ and the physical duration is $f/128$.
The reconstruction compares endpoint subtraction with an independent
streaming sum of ordered pair products and checks the pointwise bounds.
It is a reproducibility measurement on existing paths and contributes
zero new optimizer updates, model calls or initialization draws.

The signed temporal contribution is positive in 53 cells and negative in 11, with range -0.57208 to 0.3561.
All 120,000 coordinate-block observations satisfy the temporal-energy identity and bounds. Independent ordered pair accumulation agrees with endpoint subtraction within $1.7858\mathord{\times}10^{-14}$ in scaled units.
Table~\ref{model:tab:equal-time-temporal} reports all $S,B,X$ outcomes. These uncentered path energies include coherent drift and do not estimate a stationary forcing covariance.

The observed temporal contribution supports retaining the full signed
finite-energy map. It provides no justification for a diagonal-in-time
energy approximation, and neither its 64 cells nor its internal
coordinate-block observations enlarge the initialization ensemble.

\FloatBarrier
\par\medskip\noindent
Chapter~\ref{ch:predictive-resolution} turns from the evolution of these
states to reductions of their predictive computation. It defines the
resolution, context law and conditional risk needed to assess a cached or
coarsened operator.

\part{Predictive reduction, cached operators, and adaptation}
\chapter{Predictive resolution and conditional operator risk}
\label{ch:predictive-resolution}
This chapter develops predictive reduction at a declared observation
resolution. Nested categorical maps, context-conditioned risks and
calibrated operator caches separate lost predictive information from
estimation error and from the cost of replacing native operator generation.

\section{Nested predictive resolutions and a finite error budget}
\label{model:sec:nested-resolution}

The vocabulary resolution is an observation scale. Its maps can be
compatible at every training state even when a fixed resolution has
different relative accuracy at different training times. Fix a finite
vocabulary $V$, a strictly positive reference distribution $r$, and nested
retained sets $I_k\subset I_\ell\subsetneq V$. Write $J_k=V\setminus I_k$.
The elimination $K_k$ retains $p_i$ for $i\in I_k$ and one mass
$p(J_k)$. The lift $R_k$ uses the reference conditional on $J_k$:
\[
 (\Pi_kp)_i=(R_kK_kp)_i=
 \begin{cases}
 p_i,&i\in I_k,\\
 p(J_k)r_i/r(J_k),&i\in J_k.
 \end{cases}
\]
The reference and retained sets may depend on the declared context, but
must be fixed across the conditional training ensemble. All resolutions
use conditionals of the same $r$.

\subsection{Conditional density and compatible time transport}
\begin{proposition}[Finite reference-density projection]
\label{model:prop:finite-density}
Let $\mathcal P$ be a partition of a finite vocabulary into nonempty
blocks, and let $p,r$ be strictly positive probability vectors. For the
block $B(i)$ containing $i$, define
\[
 (\Pi_{\mathcal P}p)_i=r_i\frac{p(B(i))}{r(B(i))},\qquad
 f_i=p_i/r_i,\qquad \|h\|_r^2=\sum_i r_i h_i^2.
\]
Then
\begin{equation}
 A_{\mathcal P}f:=\Pi_{\mathcal P}p/r
       =\E_r[f\mid\sigma(\mathcal P)].
 \label{model:eq:finite-density-interpretation}
\end{equation}
For a refinement $\mathcal Q$ of $\mathcal P$, both with the same reference,
\begin{align}
 \Pi_{\mathcal P}\Pi_{\mathcal Q}p
   &=\Pi_{\mathcal Q}\Pi_{\mathcal P}p=\Pi_{\mathcal P}p,\label{model:eq:partition-absorption}\\
 \|f-A_{\mathcal P}f\|_r^2
   &=\|f-A_{\mathcal Q}f\|_r^2
     +\|A_{\mathcal Q}f-A_{\mathcal P}f\|_r^2,\label{model:eq:density-risk-flow}\\
 D_{\rm KL}(p\|\Pi_{\mathcal P}p)
   &=D_{\rm KL}(p\|\Pi_{\mathcal Q}p)
     +D_{\rm KL}(\Pi_{\mathcal Q}p\|\Pi_{\mathcal P}p).
 \label{model:eq:partition-kl-flow}
\end{align}
For a random proper-prefix emission $p_t$ of the complete single-pass
state and any conditioning sigma-field $\mathcal H$, a fixed $r$ and
partition also give
\begin{equation}
 \E[\Pi_{\mathcal P}p_{t+b}\mid\mathcal H]
 =\Pi_{\mathcal P}\E[p_{t+b}\mid\mathcal H],\qquad
 \Pi_{\mathcal P}(p_{t+b}-p_t)
 =\Pi_{\mathcal P}p_{t+b}-\Pi_{\mathcal P}p_t.
 \label{model:eq:density-time-compatibility}
\end{equation}
The last formula uses the same linear map on signed increments.
\end{proposition}
\begin{proof}
The reference-weighted mean of $f$ on a block $B$ is
$\sum_{i\in B}r_i(p_i/r_i)/r(B)=p(B)/r(B)$, proving
Equation~\eqref{model:eq:finite-density-interpretation}. Fine projection
preserves every coarse block mass. A coarse projected density is already
constant on each fine block. These facts give both absorption identities.
For any block-constant $h$, one has
$\sum_{i\in B}r_i(f_i-(A_{\mathcal Q}f)_i)h_i=0$ on every
$\mathcal Q$ block. Take $h=A_{\mathcal Q}f-A_{\mathcal P}f$,
expand the square, and sum to obtain the orthogonal risk identity.

Put $q=\Pi_{\mathcal Q}p$ and $u=\Pi_{\mathcal P}p$.
The ratio $q_i/u_i$ is constant on each fine block, and $p,q$ give
that block the same mass. Hence
$\sum_i p_i\log(q_i/u_i)=\sum_i q_i\log(q_i/u_i)$.
Expanding $\log(p_i/u_i)$ proves the KL identity. Finally,
$\Pi_{\mathcal P}$ is a deterministic finite matrix, with entries
$r_i/r(B)$ for $i,j\in B$ and zero otherwise. Finite linearity of
conditional expectation and subtraction proves
Equation~\eqref{model:eq:density-time-compatibility}. Probabilities are bounded,
so these conditional expectations exist.
\end{proof}

This applies established conditional-expectation algebra
\cite{kallenberg2021} to the training-selected predictive law. The
reference measure is on vocabulary coordinates, whereas the outer
expectation averages the declared training ensemble. They are different
probability spaces. Successive density increments have orthogonal
reference-weighted energies, and successive KL decreases have a
nonnegative, additive scale budget. Neither energy is automatically the
centered Hellinger fluctuation measured in
Section~\ref{model:sec:categorical-scale-results}. The singletons-and-tail
partition recovers $\Pi_k$ without changing its observation definition.

Taking $\mathcal H=\sigma(S_t)$ connects this projection exactly to the
complete chronological kernel, including optimizer and remaining-resource
state. It does not express the result as a function of $\Pi_kp_t$ alone.
A reduced successor still requires the closure bounds of
Section~\ref{model:sec:reduction}. A reference or partition refitted separately
at each time or resolution generally loses these fixed-map identities.

\begin{proposition}[Compatible lifts and an exact KL refinement budget]
\label{model:prop:nested-kl}
For strictly positive probability vectors $p,r$ and the nested sets above,
\begin{gather}
 \Pi_k\Pi_\ell p=\Pi_\ell\Pi_kp=\Pi_kp,
 \qquad K_k=K_{k\leftarrow\ell}K_\ell,\label{model:eq:absorb}\\
 D_{\rm KL}(p\|\Pi_kp)
 =D_{\rm KL}(p\|\Pi_\ell p)
  +D_{\rm KL}(\Pi_\ell p\|\Pi_kp).\label{model:eq:kl-chain}
\end{gather}
Here $K_{k\leftarrow\ell}$ adds the newly deleted coordinates to the
tail mass. Thus the KL reconstruction error decreases when more tokens
are retained. If $v_k$ is the population variance trace of
$2\sqrt{K_kp}$, or its unbiased empirical analogue under the same
replica divisor and context weights, then
\[
 0\le v_k\le v_\ell\le v_p.
\]
These statements remain valid at any selected training time and under
any fixed positive weighting of contexts.
\end{proposition}
\begin{proof}
The finer lift preserves every $I_k$ coordinate and the total mass on
$J_k$, so the coarser lift applied afterward gives $\Pi_kp$. Conversely,
$\Pi_kp$ already has the reference conditional distribution on $J_\ell$;
applying the finer lift preserves it. Summing deleted coordinates proves
the identity for $K$.

Set $q=\Pi_\ell p$ and $u=\Pi_kp$. Expanding the logarithm gives
\[
 D_{\rm KL}(p\|u)-D_{\rm KL}(p\|q)
 =\sum_{i\in V}p_i\log(q_i/u_i).
\]
On $I_\ell$, $p_i=q_i$. On $J_\ell$, $q_i/u_i$ is constant because
both $q$ and $u$ are proportional to $r$ there. The two distributions
also give $J_\ell$ the same mass. Hence the last sum is unchanged if
$p$ is replaced by $q$, giving the claimed KL identity. Nonnegativity
of KL proves monotonicity.

For any partition channel and two probability vectors $a,b$,
Cauchy--Schwarz on each partition block $B$ gives
$\sum_{i\in B}\sqrt{a_ib_i}\le\sqrt{a(B)b(B)}$.
Using unit total mass, this implies contraction of
$\|2\sqrt a-2\sqrt b\|^2$ under aggregation. For a square-integrable
random vector $X$ and an independent copy $X'$,
$\mathbb E\|X-\mathbb EX\|^2=\tfrac12\mathbb E\|X-X'\|^2$.
The empirical counterpart is the sum over unordered replica pairs
with divisor $S(S-1)$. Apply the contraction successively to
$K_\ell$ and $K_{k\leftarrow\ell}$. Summation over the same fixed
context weights preserves every inequality.
\end{proof}

This is a composition law for acquired predictive observations. It
does not supply a transition kernel for a smaller optimizer state or
a block map to a separately trained smaller architecture. The same
reference across resolutions is essential to the absorption and KL
identities; independently refitted tail laws need not obey them.

\begin{proposition}[Finite empirical and population tail-risk bounds]
\label{model:prop:tail-risk}
Let $p^{(s,c)}$ be positive probability vectors for $S\ge2$ complete
replicas and $C\ge1$ fixed equally weighted contexts. Set
$e^{(s,c)}=2\sqrt{p^{(s,c)}}-2\sqrt{\Pi_kp^{(s,c)}}$ and
\[
 U_k=\frac1{SC}\sum_{s,c}\|e^{(s,c)}\|^2,\qquad
 B_k=\frac1C\sum_c\left\|\frac1S\sum_s e^{(s,c)}\right\|^2.
\]
With the unbiased centered residual variance $V_{e,k}$ and the mean
KL error $\overline D_k$, one has
\begin{equation}
 V_{e,k}=\frac{S}{S-1}(U_k-B_k)
 \le\frac{S}{S-1}U_k
 \le\frac{4S}{S-1}\overline D_k.
 \label{model:eq:empirical-kl-budget}
\end{equation}
For the population law the corresponding inequalities are
$v_{e,k}\le\mathbb E\|e_k\|^2\le4\mathbb E D_{\rm KL}(p\|\Pi_kp)$,
with the same context averaging. If $v_p>0$, a sufficient condition
for $\sqrt{v_{e,k}/v_p}\le\varepsilon$ is
$4\mathbb E D_{\rm KL}(p\|\Pi_kp)\le\varepsilon^2v_p$.
The empirical condition uses the additional factor $S/(S-1)$.
\end{proposition}
\begin{proof}
Expanding the centered sum of squares yields
$\sum_s\|e^{(s,c)}-\bar e^{(c)}\|^2
=\sum_s\|e^{(s,c)}\|^2-S\|\bar e^{(c)}\|^2$.
Divide by $(S-1)C$ and sum over contexts to obtain the equality and
first inequality. For positive $p,q$, write
$A=\sum_i\sqrt{p_iq_i}\in(0,1]$. Jensen's inequality for $\log$
gives $D_{\rm KL}(p\|q)\ge-2\log A\ge2(1-A)$.
Since $\|2\sqrt p-2\sqrt q\|^2=8(1-A)$, its value is at most
$4D_{\rm KL}(p\|q)$. Apply this pointwise and average. In the
population case, centering subtracts the squared mean, giving the
same argument without the unbiased empirical factor. Dividing by
positive $v_p$ proves the sufficient relative-error condition.
\end{proof}

The KL certificate can be conservative because it includes the mean
reconstruction bias. A finite centered-error target therefore requires
its own measured residual variance. At zero native variance the relative
criterion is undefined; the absolute identities still apply. A tolerance
or numerical floor cannot replace this distinction.

For a coupled fluctuation-scale limit, the same reasoning gives the
sufficient population condition
$\mathbb E D_{\rm KL}(p_N\|\Pi_{k_N}p_N)=o(v_{p,N})$.
It forces $v_{e,k_N}/v_{p,N}\to0$. Expanding the centered norm of
$2\sqrt{p_N}=2\sqrt{\Pi_{k_N}p_N}+e_{k_N}$ then shows that the
retained and full variance traces have ratio tending to one. This
condition does not identify the native fluctuation law, establish a
critical exponent, or remove acquisition and dictionary costs.

\section{The context law of a predictive observation}
\label{model:sec:context-risk-law}
The vocabulary projection is evaluated under a declared context law.
Its fluctuation error can vary across that law even when the dictionary
is fixed. This variation is separate from the document-latent covariance
in Equations~\eqref{model:eq:latent}--\eqref{model:eq:latent-conditional}: there the
summands are positions in a random document, whereas here each context
indexes an across-initialization predictive variance.

\begin{proposition}[Context-resolved fluctuation risk]
\label{model:prop:context-risk}
Let $x_{s,c}$ and $\widehat x_{s,c}$ be vectors in a real Hilbert space,
for replicas $s=1,\ldots,S$, $S\geq2$, and a nonempty finite context panel.
Write $e_{s,c}=x_{s,c}-\widehat x_{s,c}$ and let $V_{p,c}$ and $V_{e,c}$
be their centered empirical variances with divisor $S-1$.
Suppose $V_{p,c}>0$ for all $c$. For context weights $\mu_c\geq0$,
$\sum_c\mu_c=1$, define
\begin{equation}
 R_c^2=\frac{V_{e,c}}{V_{p,c}},\qquad
 R_\mu^2=\frac{\sum_c\mu_cV_{e,c}}{\sum_c\mu_cV_{p,c}},\qquad
 w_c=\frac{\mu_cV_{p,c}}{\sum_d\mu_dV_{p,d}}.
 \label{model:eq:context-risk-weights}
\end{equation}
For uniform context weights, $R_\mu^2$ equals the ratio of the two
context-averaged variances. Then $R_\mu^2=\sum_cw_cR_c^2$. For every $\tau>0$,
\begin{equation}
 \sum_{c:R_c>\tau}w_c\leq
 \min\{1,R_\mu^2/\tau^2\}.
 \label{model:eq:context-risk-tail}
\end{equation}
These are finite-array statements; the replicas and contexts need not be
independent for the identities. Their statistical interpretation requires
specifying the sampling law separately.
\end{proposition}
\begin{proof}
The panel is nonempty and at least one context weight is positive, so
the denominator is positive. For uniform weights the common nonzero
panel-size divisor cancels between the two means. Substitution of $w_c$ and $R_c^2$ cancels
each $V_{p,c}$, giving the equality. All summands are nonnegative.
For $A=\{c:R_c>\tau\}$,
$\tau^2\sum_{c\in A}w_c\leq\sum_{c\in A}w_cR_c^2
\leq R_\mu^2$; also $\sum_cw_c=1$. Division by $\tau^2$ proves the bound.
\end{proof}

The tail bound is Markov's inequality for the variance-weighted finite
context law. It does not bound the largest error by $R_\mu$, or identify
the unweighted context count with the weighted mass. If a context has
zero native variance, its relative error is undefined; absolute variances
must be retained and the positive-variance identity applied on its stated
domain. In the panels of Section~\ref{model:sec:context-risk-results},
every measured native variance is positive.
For the Hellinger embedding $x=2\sqrt p$, this proposition applies to
exactly the fluctuation norm of Proposition~\ref{model:prop:tail-risk}.
A constant finite target bounds a finite observation error. Limit-law
inheritance additionally needs that error to vanish relative to the
claimed fluctuation scale in each retained direction.

\section{Equivariant operator elimination at frozen inference}
\label{model:sec:operator-cache-law}
The training law retains the common learned operator, even when its
realization varies little with the inference input. This gives a useful
computational reduction distinct from projecting a completed vocabulary
prediction. Fix a trained state $\theta$, a calibration panel
$X_1,\ldots,X_m$ of proper prefixes, and weights $a_j\geq0$ with
$\sum_j a_j=1$. Define a cache for every decoder and head by
\begin{equation}
 \overline G_{\ell h}(\theta)
   =\sum_{j=1}^m a_jG_{\ell h}(\theta,X_j).
 \label{model:eq:operator-cache-mean}
\end{equation}
The cached network recomputes queries, keys, values, residuals and
feed-forward blocks on each new prefix, using $\overline G_{\ell h}$
in place of native operator generation. Its calibration panel is fixed
before assessment and contains neither assessment documents nor their
targets. Operator freezing and G caching originate in
\cite{gokden2025,gokden2026foundations}; the construction here conditions
the cache on the recorded single-pass training ensemble and measures
its retained predictive fluctuations and actual model-call cost.

\begin{proposition}[Cache equivariance and decoder error]
\label{model:prop:equivariant-cache}
Under the native head-sign action of
Proposition~\ref{model:prop:native-head-sign-symmetry}, a cache constructed by
\eqref{model:eq:operator-cache-mean} transforms as
$\overline G_{\ell h}\mapsto\sigma_{\ell h}\overline G_{\ell h}$.
Co-transforming the query projection and this cache leaves the cached
predictions unchanged.

For a fixed prefix, write the native and cached decoder states as
$u_\ell=F_\ell(u_{\ell-1})$ and
$\widehat u_\ell=\widehat F_\ell(\widehat u_{\ell-1})$, with equal inputs.
Suppose $\widehat F_\ell$ is $L_\ell$-Lipschitz on the compared states
and the same-input substitution obeys
$\|F_\ell(u_{\ell-1})-\widehat F_\ell(u_{\ell-1})\|\leq d_\ell$.
If the vocabulary-centered final logit map is $c$-Lipschitz, put
\[
 D=c\sum_{\ell=1}^L d_\ell\prod_{j=\ell+1}^L L_j.
\]
For the native and cached next-token laws $p$ and $\widehat p$,
\begin{equation}
 \|2\sqrt p-2\sqrt{\widehat p}\|_2\leq D/\sqrt2,
 \qquad \KL(p\|\widehat p)\leq D^2/4.
 \label{model:eq:cache-emission-bound}
\end{equation}
For an external target $y$, their negative-log-probability difference has
absolute value at most $\sqrt2 D$.
\end{proposition}
\begin{proof}
The native sign action preserves each query Gram, generated metric and
hidden state and sends $G_{\ell h}(\theta,X_j)$ to
$\sigma_{\ell h}G_{\ell h}(\theta,X_j)$. Linearity of the fixed weighted
sum proves cache equivariance. The query also receives the same sign,
so $Q_{\ell h}\overline G_{\ell h}$ is invariant. Induction over decoders
therefore preserves all cached hidden states and final predictions.

Insert $\widehat F_\ell(u_{\ell-1})$ between the two decoder outputs.
The triangle inequality gives
$e_\ell\leq d_\ell+L_\ell e_{\ell-1}$, $e_0=0$. Iteration and the
readout bound give centered logit error at most $D$.
For $p=\softmax(z)$, the Hessian of log-sum-exp is
$H=\diag(p)-pp^{\mathsf T}$. Its row absolute sum is
$2p_i(1-p_i)\leq1/2$, so its symmetric operator norm is at most $1/2$.
The Jacobian of $\phi(z)=2\sqrt{\softmax(z)}$ satisfies
$D\phi(z)^{\mathsf T}D\phi(z)=H$. Integrating along the centered logit
segment proves the first bound. The Bregman formula for log-sum-exp,
with the same Hessian bound and $\int_0^1(1-t)\,dt=1/2$, proves the KL
bound. Finally, the gradient of $-\log p_y$ is $p-e_y$, has norm at
most $\sqrt2$, and annihilates the constant logit direction. Integrating
it along that segment proves the target-risk bound.
\end{proof}

The $d_\ell$ can be bounded by a same-state operator error and its native
PLGA sensitivity as in Proposition~\ref{model:prop:prefix-defect-bound}.
The proposition makes these domains and constants explicit; the finite
experiment in Section~\ref{model:sec:operator-cache-results} measures
the resulting emission errors directly and
does not estimate uniform decoder Lipschitz constants. In particular,
small row contrast alone is not its premise.

The conditional signed class survives this cache construction under its
stated limiting hypotheses. Convex averaging preserves the uniform
operator envelope, and cache equivariance preserves the independent
head-sign law. Theorem~\ref{model:thm:native-sign-gaussian-mixture} therefore
applies to cached operators when their invariant covariance laws
converge. Its limiting covariance need not equal the native covariance.
Prediction fidelity is a separate, measured requirement.

For a complete ensemble of trained states, set $x=2\sqrt p$,
$\widehat x=2\sqrt{\widehat p}$ and $e=x-\widehat x$. At each fixed
context the exact centered budget is
\begin{equation}
 V_x=V_{\widehat x}+V_e+2\Cov(\widehat x,e),
 \qquad
 |V_{\widehat x}-V_x|\leq2\sqrt{V_xV_e}+V_e.
 \label{model:eq:cache-covariance-budget}
\end{equation}
Here variances are squared Hilbert norms and covariance is the scalar
inner product. To verify the identities, center $x=\widehat x+e$ and
expand its squared norm. Equivalently expand
$\widehat x=x-e$ and apply Cauchy--Schwarz to obtain the bound.
The same proof uses empirical sums with a common divisor $S-1$.
Proposition~\ref{model:prop:context-risk} then supplies the context-weighted
relative error. A checkpoint-specific cache can preserve substantial
between-checkpoint fluctuations while eliminating within-checkpoint
operator generation. A single cache pooled over independently trained
states would define a different observation and has no such guarantee.

\section{Conditional risk of a calibrated operator}
\label{model:sec:cache-risk-law}
Cache accuracy is a property of a specified trained state and prefix law.
It depends on operator dispersion, displacement of the calibration mean,
and propagation through the decoder. The following decomposition gives
these quantities separate roles in the training-to-inference map.

\begin{proposition}[Conditional calibration risk]
\label{model:prop:cache-risk}
Fix a state $\theta$ and a square-integrable operator $G_\theta(X)$ in a
real Hilbert space. Write $m_\mu=\E_\mu G_\theta$,
$V_\mu=\E_\mu\norm{G_\theta-m_\mu}^2$ for an evaluation law $\mu$,
and $m_\nu,V_\nu$ for a calibration law $\nu$.
Let $X\sim\mu$ and $Y_1,\ldots,Y_m\sim\nu$ be independent conditional
on $\theta$, with the calibration prefixes identically distributed.
For deterministic real weights $a_j$ with $\sum_j a_j=1$ and
$\widehat G=\sum_j a_jG_\theta(Y_j)$,
\begin{equation}
 \E\norm{G_\theta(X)-\widehat G}^2
 =V_\mu+\norm{m_\mu-m_\nu}^2+V_\nu\sum_j a_j^2.
 \label{model:eq:cache-population-risk}
\end{equation}
For a fixed empirical panel, nonnegative weights $w_i$ summing to one,
$\bar G_w=\sum_iw_iG_i$, and any fixed cache $c$,
\begin{equation}
 \sum_iw_i\norm{G_i-c}^2
 =\sum_iw_i\norm{G_i-\bar G_w}^2+\norm{\bar G_w-c}^2.
 \label{model:eq:cache-empirical-risk}
\end{equation}
The finite identity requires no sampling independence.
\end{proposition}
\begin{proof}
Decompose the population residual into $G_\theta(X)-m_\mu$,
$m_\mu-m_\nu$, and $-(\widehat G-m_\nu)$.
The first and third terms have zero conditional mean, and independence
annuls their expected inner product. Distinct centered calibration terms
also have zero expected inner product, so the cache variance is
$V_\nu\sum_j a_j^2$. Expanding the squared norm proves the first identity.
For the second, write $G_i-c=(G_i-\bar G_w)+(\bar G_w-c)$.
The weighted cross term vanishes since $\sum_iw_i(G_i-\bar G_w)=\bm0$.
\end{proof}

Uniform independent calibration has variance contribution $V_\nu/m$.
For dependent calibration or evaluation, the general centered expansion is
\begin{align}
 \E\norm{G_\theta(X)-\widehat G}^2
 &=V_\mu+\E\norm{\widehat G-\E\widehat G}^2
   +\norm{m_\mu-\E\widehat G}^2\notag\\
 &\quad-2\E\ip{G_\theta(X)-m_\mu}{\widehat G-\E\widehat G}.
 \label{model:eq:cache-dependent-risk}
\end{align}
Indeed, subtract each mean before expanding; deterministic mean terms
have zero expected inner products with centered terms. Sampling distinct
documents does not establish conditional independence. For example, a
uniform size-$m$ sample, $1\le m\le M$, without replacement from $M>1$ fixed calibration
vectors has cache variance $V_\nu(M-m)/(m(M-1))$, where $V_\nu$ uses
divisor $M$. Each centered draw has variance $V_\nu$ and distinct draws
have inner-product expectation $-V_\nu/(M-1)$; expanding the mean proves
the formula. A coupled assessment draw still retains the cross term in
\eqref{model:eq:cache-dependent-risk}. The executed fixed-panel measurements in Section~\ref{model:sec:cache-risk-results} use \eqref{model:eq:cache-empirical-risk} and do not test a population
$1/m$ law.

\begin{corollary}[Operator risk controls finite emission risk]
\label{model:cor:cache-emission-risk}
On the compared domain, assume the same-input defect at layer $\ell$
is at most $A_\ell\norm{G_\ell(X)-c_\ell}$, with $A_\ell\ge0$.
Let the cached decoder Lipschitz constants $L_k$ and the
vocabulary-centered logit constant $c_*$ be finite and uniform on that
domain. Set $b_\ell=c_*A_\ell\prod_{k>\ell}L_k$. Then
\begin{equation}
 \E\KL(p_X\|\widehat p_X)
 \le\frac14\left(\sum_\ell b_\ell
       \sqrt{\E\norm{G_\ell(X)-c_\ell}^2}\right)^2.
 \label{model:eq:cache-emission-risk}
\end{equation}
\end{corollary}
\begin{proof}
Insertion of the cached decoder at the native input gives
$e_\ell\le A_\ell\norm{G_\ell-c_\ell}+L_\ell e_{\ell-1}$, with $e_0=0$.
Induction bounds the centered-logit error by
$D(X)=\sum_\ell b_\ell\norm{G_\ell(X)-c_\ell}$.
The log-sum-exp Hessian $H=\diag(p)-pp^{\mathsf T}$ is symmetric with
absolute row sums $2p_i(1-p_i)\le1/2$, so $\norm H_{\rm op}\le1/2$.
Its Bregman integral gives $\KL(p_X\|\widehat p_X)\le D(X)^2/4$.
Taking expectations and applying the $L^2$ triangle inequality proves
the displayed bound.
\end{proof}

All expectations here condition on the recorded state. Along training,
both the operator law and its downstream sensitivity change. Row-contrast
suppression alone bounds neither its common coordinate nor those
sensitivities. A cache measurement at a trained endpoint therefore
establishes finite conditional approximation, without identifying when
or why that approximation became accurate. The finite empirical
logit/operator ratios in Section~\ref{model:sec:cache-risk-results}
describe the observed secants; they are not
estimates of uniform $A_\ell,L_k$ or a certificate for unseen prefixes.

\begin{proposition}[Transport between cache states]
\label{model:prop:cache-state-transport}
Fix the assessment panel and its normalized weights at two states $s,t$.
Let $m_s,m_t$ be its mean operators, $V_t$ its scatter at $t$, and $c_s$
a fixed cache calibrated at $s$. Set $\Delta=m_t-m_s$ and $a=m_s-c_s$.
Then, in the same Hilbert norm as Equation~\eqref{model:eq:cache-empirical-risk},
\begin{align}
 R_t(c_s)&=V_t+\norm\Delta^2+\norm a^2+2\ip\Delta a,
 \label{model:eq:cache-state-transport}\\
 \sqrt{R_t(c_s)}&\leq\sqrt{V_t} + \norm\Delta+\norm a.
 \label{model:eq:cache-state-bound}
\end{align}
For a recalibrated cache $c_t$, $R_t(c_t)=V_t+\norm{m_t-c_t}^2$.
No independence, stationarity, monotone training trend, or unchanged
query distribution is assumed.
\end{proposition}
\begin{proof}
At state $t$, split each residual as $G_t(X_i)-m_t+(m_t-c_s)$.
Its centered weighted sum is zero. The squared risk therefore equals
$V_t+\norm{m_t-c_s}^2$. Substitute $m_t-c_s=\Delta+a$ and expand its
squared norm. The norm triangle inequality and
$\sqrt{x^2+y^2}\leq x+y$ for $x,y\geq0$ give the stated bound.
Replacing $c_s$ by $c_t$ proves the final identity.
\end{proof}

The state change includes the query representations induced by all trained
parameters. Even a frozen metric generator need not have a constant
mean output under that changed input law. The signed cross term can either
increase or decrease the displacement, so operator drift is not itself a
monotone cache-error predictor. Applying
Corollary~\ref{model:cor:cache-emission-risk} with the corresponding layer risks
transports this budget to predictive risk when its uniform sensitivity
hypotheses hold. The measurements in Section~\ref{model:sec:cache-state-results} separately retain the exact
operator budget and predictive errors; they do not infer those uniform
sensitivities from observed secants.

Fidelity to the native emission is distinct from target-law cross entropy.
For positive vocabulary laws $p,q$ and any external target law $r$,
\begin{equation}
 \mathbb E_{Y\sim r}[-\log q(Y)+\log p(Y)]
     =\KL(r\|q)-\KL(r\|p).
 \label{model:eq:cache-target-risk}
\end{equation}
Subtracting the two KL sums cancels $\sum_y r_y\log r_y$ and proves the
identity, with $0\log0=0$. Only when $r=p$ does this difference equal
the nonnegative fidelity loss $\KL(p\|q)$. An intervention can therefore
lower external target NLL while moving away from the recorded native law.
Both quantities are measured in Section~\ref{model:sec:cache-state-results}.

\FloatBarrier
\par\medskip\noindent
Chapter~\ref{ch:cache-evidence} evaluates these reductions on executed
predictive and cache studies. Disjoint contexts, prefix transfer and changes
of training state test the scope of the conditional risk statements.

\chapter{Executed predictive resolution and cache transfer}
\label{ch:cache-evidence}
This chapter reports predictive-resolution and cached-operator experiments.
It compares temporal and categorical reductions, disjoint context panels,
calibration risk and state transfer, thereby distinguishing a successful
conditional cache from a state-independent replacement law.

\section{Predictive resolution along training and on disjoint contexts}
\label{model:sec:categorical-scale-results}

\subsection{Time-conditioned predictive scale}
The nested maps in Proposition~\ref{model:prop:nested-kl} act on the same
training-selected emission at every saved time. We evaluate the fixed
development dictionary on all 216 single-pass trajectories at
$T=2048,3072,4096$, retaining the five declared resolutions and every
width/control cell. These time slices share six complete initialization
replicas per cell. They are not additional independent training samples.
The two-time reduction and the additional-time check were specified in
separate analysis plans; the combined reduction reproduces their complete
panel without refitting the dictionary or accuracy target.

The 540 resolution cells and 432 adjacent-resolution cells include
20,736 path/context KL chains. The largest independently recomputed KL-chain residual
is $\ModelSourceNestedChainError$ at an absolute float64 tolerance of
$3\times10^{-12}$. Independent direct-pair reconstruction verifies every
cell and summary, with maximum cell discrepancy $\ModelSourceNestedDirectError$. These numerical checks support the implementation of the exact
identities; they are not interval-arithmetic error certificates.

At the 8,192-token scale, the centered-RMS target holds in 32 of 36 cells
at $T=2048$ and all 36 cells at each later time. The valid finite statement
therefore conditions resolution accuracy on training age as well as the
observation and source law. At $T=3072$, the actual relative centered RMS
is $0.124$--$0.227$, while the sufficient KL-based bound in
Equation~\eqref{model:eq:empirical-kl-budget} is $0.838$--$1.389$.
The exact bias subtraction explains why a useful centered observation
need not receive a sharp certificate from its uncentered KL bound.
All times and resolutions remain visible in Table~\ref{model:tab:nested-categorical}.

\subsection{A fixed dictionary on disjoint documents}
We acquire fresh proper-prefix predictions from 24 retained checkpoints:
$N\in\{8,24\}$, $g\in\{0,1.5\}$ and the six complete initialization
identities of the independent single-pass family at $T=4096$.
The 32 evaluation documents are distinct and disjoint from the training
master corpus and the original 128-document observation panel.
Each input supplies exactly a 64-token prefix; its external next token
is excluded from the native query Gram and prediction computation.

Before acquisition, we fix a context-independent reference by averaging
the original development reference over its eight contexts, then rank
vocabulary entries by that mean probability. This reference uses only the
six development initializations at $N=14,g=0,T=2048$. No prediction from
the 32 evaluation documents enters the dictionary. The same ordering,
reference and five resolutions apply to every evaluation cell. The
experiment tests a single shared reference on this declared finite
context panel; it does not estimate uniform accuracy over all prompts.

At 8,192 tokens, all four cells meet the unchanged 25\% centered-error
target. They retain $96.1$--$97.2\%$ of full predictive variance, with
relative centered RMS $0.176$--$0.197$. At 16,384 tokens the corresponding
ranges are $97.9$--$98.5\%$ and $0.129$--$0.144$. The complete
20-cell panel and its 3,072 adjacent-resolution path/context KL chains
satisfy the scale and risk identities; the maximum independently recomputed KL-chain
residual is $\ModelSourceContextSmallChainError$. The smaller retained alphabets and
their full errors are included in Table~\ref{model:tab:context-categorical}.

\begin{figure}[htbp]\centering
\includegraphics[width=\textwidth]{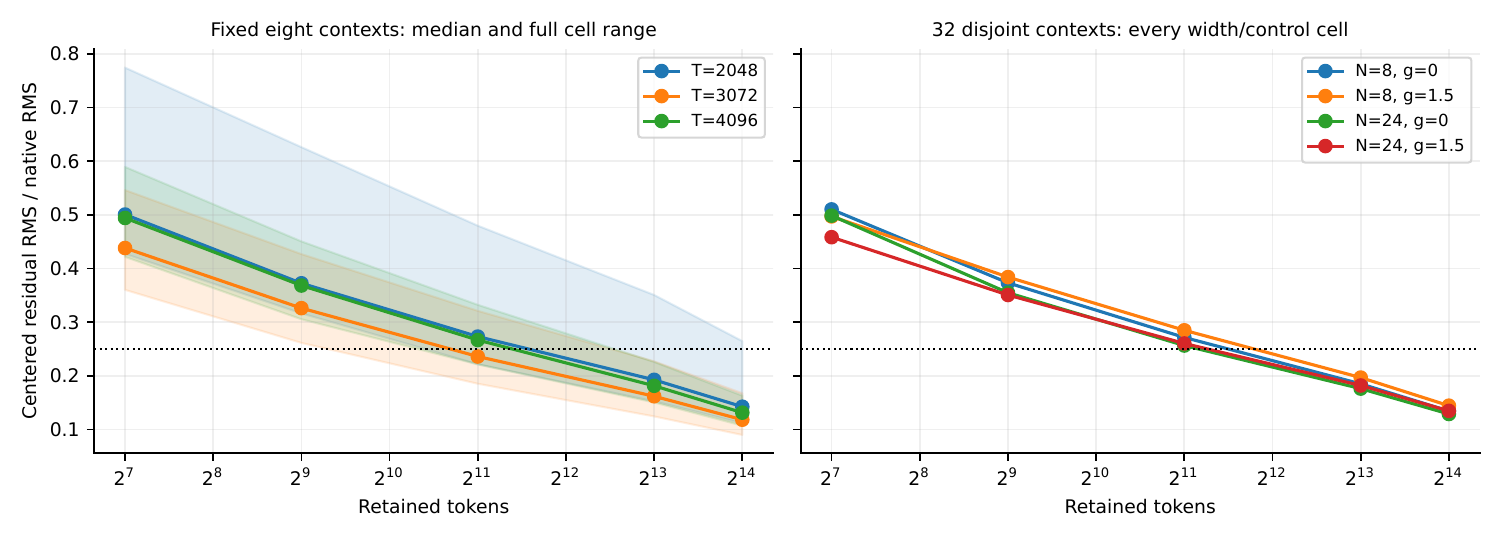}
\caption{Finite predictive resolution. Left: medians and full ranges over
36 width/control cells at three times of the same 216 paths. Right:
every cell of the 32-document disjoint-context experiment. Each cell uses
six complete initialization replicas; ranges are not confidence intervals.
The dotted line is the fixed relative centered-RMS target.}
\label{model:fig:categorical-scales}
\end{figure}

Acquisition makes \ModelSourceContextSmallForwards{} native forward calls, with
\ModelSourceContextSmallQualification{} additional same-input
qualification calls checking equality of logical float32 bytes. Worker
timers sum to \ModelSourceContextSmallWorkerSeconds{} seconds, including checkpoint loading, native execution
and serialization. The float32 evaluation logits occupy \ModelSourceContextSmallLogitBytes{} logical
bytes. Training updates are zero. The shared float64 reference and integer
ordering each contain 32,000 entries, totaling \ModelSourceCategoricalDictionaryBytes{} logical bytes.
The retained observations still require full native predictions and the
complete underlying model. These are measured predictive scale maps,
with acquisition and dictionary costs included, rather than a measured
speedup or a closed low-dimensional optimizer evolution.

\subsection{Transfer to an additional untouched document panel}
\label{model:sec:context-transfer}
A further acquisition fixes 64 additional RefinedWeb documents before
prediction. Their content hashes exclude the full training master corpus,
the original 128-document panel and all 32 documents above. It uses the
same 24 endpoint checkpoints, six initialization identities per cell,
context-independent development reference, token ordering, five resolutions
and 25\% centered-RMS target. Thus the changed variable is the finite
context panel. It adds no trained replica and no training update.

At 8,192 tokens all four width/control cells meet the frozen target.
Retained predictive variance is $\ModelSourceContextLargeRetained\%$ and relative
centered RMS is $\ModelSourceContextLargeRMS$. The absolute native variance is
$\ModelSourceContextLargeNativeVariance$, while centered residual variance is
$\ModelSourceContextLargeResidualVariance$. These are vocabulary-summed traces,
with the same replica divisor and context averaging in numerator and
denominator. Mean bias makes the uncentered errors larger, as shown for
every resolution in Table~\ref{model:tab:context-transfer}. The complete
per-cell values appear in Table~\ref{model:tab:context-transfer-full}.

The independent reduction checks \ModelSourceContextLargeCells{} cells and
\ModelSourceContextLargeChains{} path/context KL chains. Its maximum cell discrepancy
is $\ModelSourceContextLargeDirectError$ and its maximum KL-chain residual is
$\ModelSourceContextLargeChainError$ at the declared $3\times10^{-12}$ float64
tolerance. Acquisition uses \ModelSourceContextLargeForwards{} native forwards and
\ModelSourceContextLargeQualification{} repeatability forwards, with zero training
updates. Recorded worker wall times sum to \ModelSourceContextLargeWorkerSeconds{}
seconds, including loading and serialization; the saved logits contain
\ModelSourceContextLargeLogitBytes{} logical bytes. These acquisition counts, byte
counts and worker times are reconciled to the bound arrays and manifests.
Timing is recorded execution metadata, rather than a remeasurement by the
numerical verifier. The shared dictionary is unchanged.

The transfer result supports a finite observation scale under the declared
context law. It supplies neither uniform prompt fidelity nor a task-level
accuracy guarantee. In particular, the complete native prediction is still
needed to acquire the reduced observation. Proposition~\ref{model:prop:finite-density}
explains the scale identities and their compatibility with complete-state
time transport; the measured risk determines whether that exact projection
preserves useful finite fluctuations.

Together, the fixed-context time panel and the disjoint-context acquisition
support a finite, data-conditioned predictive resolution for the native
flow. Preservation of an asymptotic fluctuation normalization additionally
requires the vanishing-relative-risk condition following
Proposition~\ref{model:prop:tail-risk}; the finite target alone does not establish
that condition or a native critical exponent.

\subsection{Context-resolved risk on a frozen 128-document panel}
\label{model:sec:context-risk-results}
For equal context weights, Proposition~\ref{model:prop:context-risk} gives
\[
 R_{\rm agg}^2=\frac{\sum_cV_{e,c}}{\sum_cV_{p,c}}
             =\sum_c\frac{V_{p,c}}{\sum_dV_{p,d}}R_c^2.
\]
All measured $V_{p,c}$ are positive. Independent saved-logit
reconstruction of the 64-document panel gives 8, 5, 11 and 5 contexts
above the 0.25 target, in the condition order $(8,0),(8,1.5),(24,0),(24,1.5)$,
while all four aggregates meet it. The aggregate therefore describes a
native-variance-weighted error under the declared finite context law.

We test this interpretation on 128 further documents, rows 1248--1375 of
the same derived context corpus. Content hashes exclude the training master
corpus, the original 128-document observation set, and both the 32- and
64-document panels. The 24 checkpoints, development reference, vocabulary
order, five resolutions and target are unchanged. Before acquiring the
new predictions we fix the primary 8,192-token aggregate test and the
complete per-context reporting rule. Each document contributes one
64-token proper prefix. The six initialization identities remain the
training replicas; neither the added contexts nor paired controls
increase that count.

\begin{table}[htbp]\centering\small
\caption{Context-resolved errors at the fixed 8,192-token resolution. Each panel uses the same six trained replicas in each width/control cell. The exceedance counts are descriptive dependent observations, not independent prompt failure trials.}
\label{model:tab:context-risk-main}
\begin{tabular}{@{}rrrrrrr@{}}
\toprule Contexts & $N$ & $g$ & $R_{\rm agg}$ & Median $R_c$ & Maximum $R_c$ & $R_c>0.25$\\\midrule
64 & 8 & 0 & 0.18912 & 0.19515 & 0.36604 & 8/64\\
64 & 8 & 1.5 & 0.18850 & 0.20070 & 0.32116 & 5/64\\
64 & 24 & 0 & 0.17547 & 0.17100 & 0.30092 & 11/64\\
64 & 24 & 1.5 & 0.19302 & 0.19483 & 0.31857 & 5/64\\
128 & 8 & 0 & 0.19169 & 0.18772 & 0.33714 & 24/128\\
128 & 8 & 1.5 & 0.19708 & 0.20763 & 0.32582 & 22/128\\
128 & 24 & 0 & 0.18202 & 0.17469 & 0.46253 & 23/128\\
128 & 24 & 1.5 & 0.19178 & 0.19649 & 0.37593 & 12/128\\
\bottomrule\end{tabular}\end{table}

All four new aggregates meet the unchanged target, with relative centered
RMS $\ModelSourceContextRiskRMS$ and retained variance \ModelSourceContextRiskRetained\%.
Their context exceedance counts are 24, 22, 23 and 12 out of 128.
The largest observed context error is 0.46253 in the $(24,0)$ cell.
These observations support the context-weighted formulation and its
finite-panel accuracy statement. They do not support a uniform context
bound. The counts share trained replicas, so they are neither pooled as
independent trials nor converted into binomial confidence intervals.

\begin{figure}[htbp]\centering
\includegraphics[width=\textwidth]{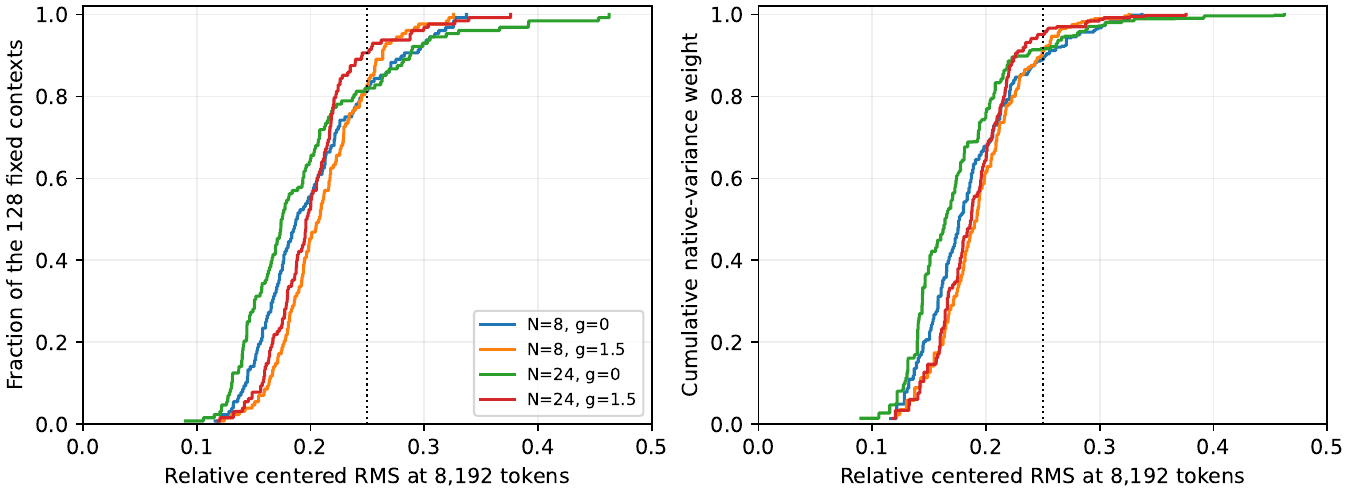}
\caption{Context variation on the fixed 128-document panel at 8,192 tokens.
The left curves give unweighted empirical fractions; the right curves use
native-variance weights. The vertical line is the unchanged 0.25 target.
Both plots are descriptive functions of the same six trained replicas per
cell, not estimates based on independent prompt failure trials.}
\label{model:fig:context-risk}
\end{figure}

At this resolution the native-variance weight of exceeding contexts is
\ModelSourceContextRiskWeightedMass\%, while Equation~\eqref{model:eq:context-risk-tail}
gives bounds $\ModelSourceContextRiskBound$. The deterministic bound is conservative;
it controls that weighted mass and makes no guarantee about the maximum.
Tables~\ref{model:tab:context-risk-scales} and \ref{model:tab:context-risk-full}
retain every tested scale, absolute variance and exceedance summary.
All 2,560 context/resolution values are retained in the numerical record.

Independent direct-pair and centered reductions check the 20 aggregate
cells, the context identity and \ModelSourceContextRiskChains{} adjacent-resolution
path/context KL chains. The maximum aggregate discrepancy is
$\ModelSourceContextRiskDirectError$ and the maximum KL-chain residual is
$\ModelSourceContextRiskChainError$, below the specified $3\times10^{-12}$
float64 tolerance. Acquisition uses \ModelSourceContextRiskForwards{} native forwards
and \ModelSourceContextRiskQualification{} repeatability forwards, with zero training
updates. Recorded worker times sum to \ModelSourceContextRiskWorkerSeconds{} seconds
on the two RTX 4090 queues, including loading and serialization.
Saved logits contain \ModelSourceContextRiskLogitBytes{} logical bytes. The complete
training states were checked against their acquisition hashes before each
checkpoint's new native observations. Offline reductions reread observations;
they do not claim another full optimizer-state verification.

\section{Executed elimination of native operator generation}
\label{model:sec:operator-cache-results}
We tested the computational realization of
Section~\ref{model:sec:operator-cache-law} at the 24 complete 4,096-update
single-pass endpoints with $N\in\{8,24\}$, $g\in\{0,1.5\}$ and the
six initialization identities 915101--915106. The frozen cache is the
arithmetic mean of each checkpoint's native operators on 16 calibration
documents. The evaluation panel contains 128 additional RefinedWeb
documents. Content hashes exclude the training master corpus, all
predictive observation panels described above, and overlap between
calibration and assessment. Each call sees a 64-token proper prefix;
token 65 is used only as an external target after prediction.

The protocol was fixed before acquisition: report every endpoint and
context, require aggregate centered RMS at most 0.25 and mean KL at most
0.03 nats in every condition, and require at least a 10\% median
steady-state latency reduction. These are finite operating targets.
The cached path uses the author's external-operator branch, retaining
native queries, keys, values, rotary encoding and causal attention.
The metric-generator layers and the power-generator computation are
bypassed. No KV cache is used. All weights remain resident, so the
comparison measures the effect of avoiding generator execution.

Every endpoint first passed exact same-context operator replay and exact
native restoration. Instrumented qualification observed all 40 native
residual-generator calls per forward and zero in the cached path.
Acquisition used one worker per RTX 4090, native float32 arithmetic with
TF32 disabled, and float64 cache averaging and analysis. At batch eight,
the experiment performed 48 calibration forwards, 768 assessment forwards (384 per path) and 1,920 timing forwards, plus 96 separately counted qualification
forwards. There were zero new optimizer updates or training replicas.
Worker timers, including loading and serialization, summed to
\ModelSourceCacheWorkerSeconds\ seconds.

\begin{table}[htbp]\centering\small
\caption{Executed operator elimination on 128 fresh proper-prefix documents. Each row uses six retained single-pass checkpoints and a separate 16-document calibration panel. $R$ is relative centered RMS, $V_q/V_p$ is the predictive variance ratio, and the final columns are medians across checkpoint timing summaries.}
\label{model:tab:operator-cache-main}
\begin{tabular}{@{}rrrrrrrr@{}}
\toprule $N$ & $g$ & $R$ & $V_q/V_p$ & Mean KL & Native (ms) & Cache (ms) & Saving (\%)\\\midrule
8 & 0 & 0.0416 & 1.00030 & 0.000091 & 6.43 & 2.36 & 63.04\\
8 & 1.5 & 0.1242 & 1.00497 & 0.000711 & 6.48 & 2.39 & 62.96\\
24 & 0 & 0.0419 & 1.00125 & 0.000069 & 18.99 & 6.63 & 65.12\\
24 & 1.5 & 0.0989 & 0.99961 & 0.000433 & 19.05 & 6.66 & 65.03\\
\bottomrule\end{tabular}\end{table}

All four conditions meet the frozen aggregate accuracy and latency
targets. The relative centered RMS is \ModelSourceCacheRMS, and mean KL is at most
\ModelSourceCacheMaxKL\ nats. Cached/native predictive variance ratios are
0.99961--1.00497. Ratios above one are permitted: operator substitution
is not a categorical coarse-graining channel, and the signed cross term
in Equation~\eqref{model:eq:cache-covariance-budget} is retained. This result
shows that substantial across-initialization predictive fluctuations can
survive removing input-dependent operator generation at these endpoints.
It does not require pooling the operators of different trained states.

The individual-context relative error exceeds 0.25 in respectively
0, 3, 0 and 1 of the 128 contexts in table order; maxima are
0.0883, 0.2742, 0.1017 and 0.2941. Mean external-target NLL changes are
$+0.00078,+0.00286,-0.00020,+0.00107$ nats. These describe paired
fidelity to the native next-token law, not improved language capability.
The six checkpoints are shared across contexts, and controls are paired;
none of the context counts is an independent-trial confidence interval.

Thirty synchronized timing pairs per endpoint followed ten warm-up pairs.
Order alternated between native and cached calls on the same device and
fixed input batch. The per-condition median reductions are
\ModelSourceCacheLatency\%, with all 24 checkpoint reductions between 62.7\%
and 65.2\%. Cache construction takes 0.117--0.218 seconds per checkpoint;
using its own measured per-request saving gives an amortization count of
11.33--54.37 batches of eight prefixes. The measured interval excludes
one-time cache installation, checkpoint loading and output serialization.
It covers final-token prefill at length 64, not autoregressive decoding
throughput or other batch sizes. Appendix~\ref{model:app:operator-cache-details}
retains every checkpoint summary, context distribution and covariance term.

\begin{figure}[htbp]\centering
\includegraphics[width=.98\textwidth]{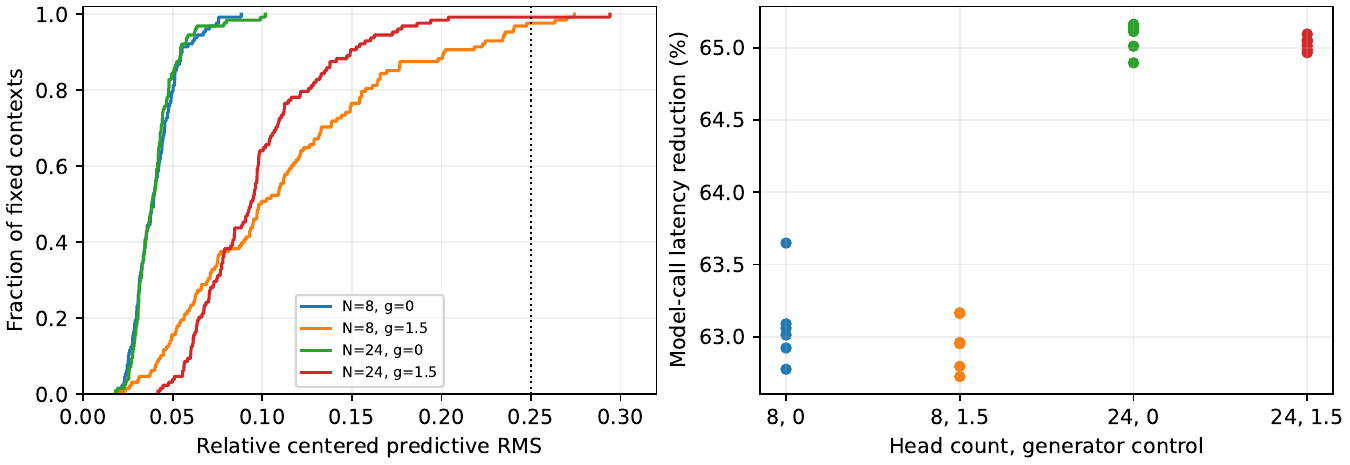}
\caption{Context-dependent predictive error and measured model-call savings.
Each right-panel point is a retained checkpoint, summarized from paired
repetitions. The left panel contains dependent observations under the same
six trained replicas per condition.}
\label{model:fig:operator-cache}
\end{figure}

This finite observation connects recorded single-pass training states to
inference execution: these endpoints admit an economical operator
substitution while retaining predictive fluctuations. The endpoint-only
comparison does not distinguish input stability already present at
initialization from stability acquired during training. It supplies neither an autonomous reduced training successor nor a native
critical exponent. Conditional critical inheritance still requires the
vanishing relative error and visible limiting mode of
Proposition~\ref{model:prop:critical-closure}. The observation projection and
operator elimination have different cost and error budgets, even though
both act on the same complete training-conditioned emission law.

\section{Conditional calibration risk and prefix transfer}
\label{model:sec:cache-risk-results}
The finite risk decomposition was assessed at all 24 selected single-pass
endpoints on a fresh 128-document RefinedWeb panel. Four disjoint sets of
16 documents supply calibration, and 64 separate documents supply
assessment. Their content hashes exclude the training corpus and all
other inference panels in the central lineage. We use the first 129
retained tokens of each selected document block. At prefix length $L$,
only the first $L$ tokens enter the model and the next token is an external
target. Lengths 32, 64 and 128 therefore define coupled observations on
the same documents, rather than independent context draws.

The protocol fixed every model, panel, arm and operating target before
acquisition. At length 64, each calibration panel gives nested cache
sizes $m=4,8,16$. At lengths 32 and 128, $m=16$ is fixed and two arms
compare a cache calibrated at that length with its unchanged length-64
cache. There are 28 arms per width/control condition and 112 aggregate
cells. All use the same six complete initialization identities. Neither
calibration panels nor prefix lengths add training replicas.

\begin{table}[htbp]\centering\footnotesize
\caption{Complete prefix/calibration groups on 64 fresh assessment documents. Ranges retain all four calibration panels and, at length 64, all three cache sizes. $L$ is assessment length, $C$ calibration length, $R$ relative centered RMS, and $n_{>.25}$ the individual-context exceedance count. Each row summarizes dependent arms, with six complete training seeds per arm.}
\label{model:tab:cache-risk-main}
\begin{tabular}{@{}rrrrrrr@{}}
\toprule $N$ & $g$ & $L$ & $C$ & Range of $R$ & Largest mean KL & Range of $n_{>.25}$\\\midrule
8 & 0 & 32 & 32 & 0.0455--0.0463 & 0.000149 & 0--0\\
8 & 1.5 & 32 & 32 & 0.1485--0.1582 & 0.001496 & 6--10\\
24 & 0 & 32 & 32 & 0.0419--0.0428 & 0.000095 & 0--0\\
24 & 1.5 & 32 & 32 & 0.1278--0.1485 & 0.001025 & 3--4\\
8 & 0 & 32 & 64 & 0.0457--0.0469 & 0.000151 & 0--0\\
8 & 1.5 & 32 & 64 & 0.1464--0.1516 & 0.001355 & 7--8\\
24 & 0 & 32 & 64 & 0.0420--0.0427 & 0.000095 & 0--0\\
24 & 1.5 & 32 & 64 & 0.1372--0.1420 & 0.000934 & 3--3\\
8 & 0 & 64 & 64 & 0.0448--0.0494 & 0.000105 & 0--0\\
8 & 1.5 & 64 & 64 & 0.1069--0.1307 & 0.000832 & 1--4\\
24 & 0 & 64 & 64 & 0.0396--0.0435 & 0.000069 & 0--0\\
24 & 1.5 & 64 & 64 & 0.1037--0.1172 & 0.000548 & 0--0\\
8 & 0 & 128 & 128 & 0.0445--0.0457 & 0.000129 & 0--0\\
8 & 1.5 & 128 & 128 & 0.0976--0.1089 & 0.000763 & 1--2\\
24 & 0 & 128 & 128 & 0.0421--0.0428 & 0.000093 & 0--0\\
24 & 1.5 & 128 & 128 & 0.1070--0.1122 & 0.000545 & 1--3\\
8 & 0 & 128 & 64 & 0.0473--0.0484 & 0.000144 & 0--0\\
8 & 1.5 & 128 & 64 & 0.1080--0.1176 & 0.000884 & 1--2\\
24 & 0 & 128 & 64 & 0.0434--0.0436 & 0.000096 & 0--0\\
24 & 1.5 & 128 & 64 & 0.1086--0.1128 & 0.000555 & 1--2\\
\bottomrule\end{tabular}\end{table}

All 112 cells meet the fixed centered-RMS target 0.25 and mean-KL target
0.03 nats. Their relative centered RMS spans \ModelSourceCacheRiskRMS\ and mean KL
is at most \ModelSourceCacheRiskMaxKL\ nats. Cached/native variance ratios range
from \ModelSourceCacheRiskRetained; signed covariance terms allow ratios above one.
Individual-context relative errors remain heterogeneous. Positive-control
arms have zero to ten exceedances among 64 contexts, with maximum
\ModelSourceCacheRiskMaxContext; every zero-control arm has none. In particular,
short-prefix aggregate success is compatible with appreciable tail error.
Mean target-NLL changes range from $-0.002045$ to $+0.007780$ nats across
all cells. These paired discrepancies measure fidelity to the native
predictor and do not establish improved language capability.

\begin{figure}[htbp]\centering
\includegraphics[width=\textwidth]{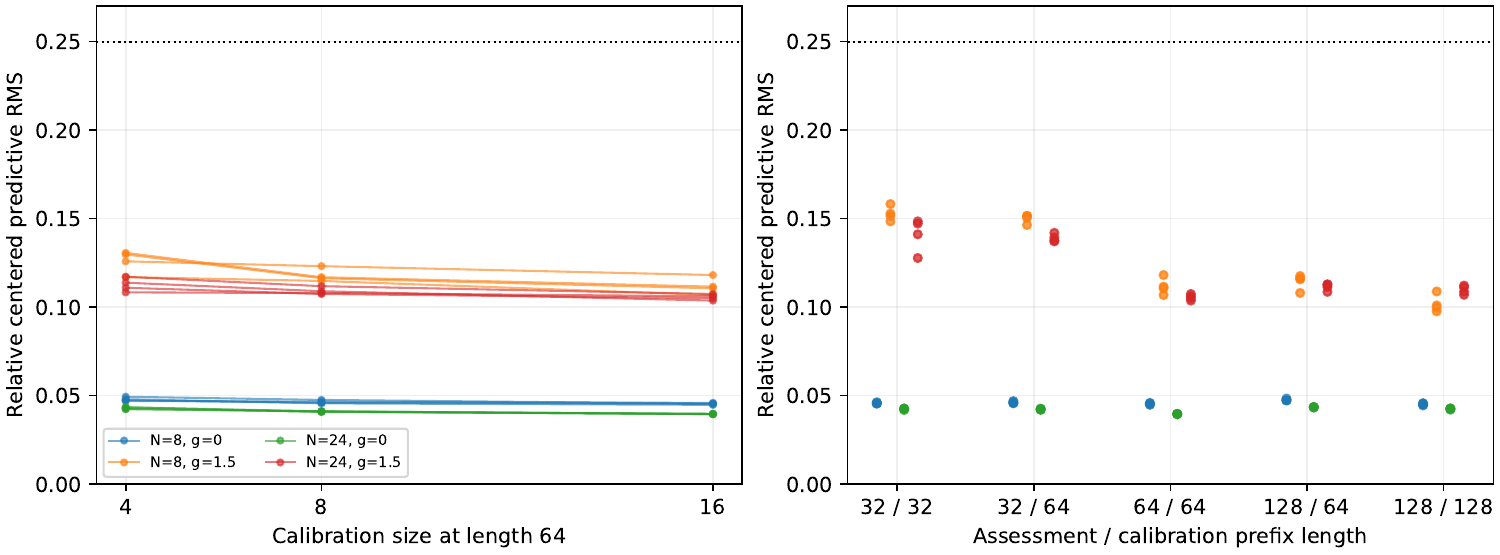}
\caption{Every calibration panel at length 64 (left), and all size-16
length-transfer arms (right). Lines connect nested cache sizes within
one panel; points at each length pair show all four panels. The dotted
line is the fixed aggregate target, not a population confidence bound.
The per-context exceptions are retained in Table~\ref{model:tab:cache-risk-main}
and the complete observation record.}
\label{model:fig:cache-risk}
\end{figure}

For each layer and state, the squared operator norm averages over its
head and matrix-entry coordinates. The mean of those risks over five
layers is reported in Table~\ref{model:tab:cache-risk-operator}.
The exact fixed-panel decomposition into assessment scatter and squared
cache displacement holds in every seed/layer/arm comparison, with maximum
absolute numerical discrepancy $2.44\times10^{-19}$. Across the aggregate
cells, scatter accounts for $74.42$--$93.26\%$ of total operator risk;
calibration displacement supplies the remainder. Different panel means
and length laws change that remainder. Nesting calibration samples alone
does not impose a monotone risk path, and these fixed panels do not
establish the population $1/m$ term in
Equation~\eqref{model:eq:cache-population-risk}.

The measurements support finite operator substitution over the declared
calibration and prefix domains, with the risk factors kept separate.
They do not estimate the uniform downstream constants in
Corollary~\ref{model:cor:cache-emission-risk} or identify an autonomous optimizer
successor. The common trained state remains the conditioning variable.
The full 112-cell predictive table, all operator budgets, and all
context/seed contrasts are retained in Section~\ref{model:app:cache-risk-details}
and the saved observation package.

Acquisition used one worker per RTX 4090 after a widest-model check.
Each state passed exact own-operator replay, restoration and generator
bypass at every tested length; a batch-one 128-token check preceded
batch-eight observations. There were 576 calibration, 576 native assessment,
5,376 cached assessment and 336 qualification forwards, totaling 6,864,
with zero optimizer updates and zero new training replicas. Summed worker
time was \ModelSourceCacheRiskWorkers\ seconds and peak allocated memory was
\ModelSourceCacheRiskPeakGiB\ GiB. The three-hour worker cap and 22 GiB memory
ceiling were respected. This experiment provides no additional latency
measurement; the timing claim remains the batch-eight, length-64 forward
workload of Section~\ref{model:sec:operator-cache-results}.

\section{Training selects the cache state}
\label{model:sec:cache-state-results}

This study asks whether a cache accurate at one training state remains
accurate after the state changes. Six paired initialization identities at
each of widths 8 and 24 supply the independent training units, conditional
on the fixed corpus, order and initial shared generator. Recalibrated and
initial caches are compared by centered RMS and mean KL on proper
prefixes. All eighteen recalibrated cells and all six zero-control
initial-cache cells meet the aggregate targets; the positive-control
initial-cache cells fail (Table~\ref{model:tab:cache-state-main}).
This establishes finite state-specific fidelity and a transport limitation,
with no added independent training replicas or arbitrary-context guarantee.
Whole-initialization paired resampling describes the fixed-panel,
inherited-corpus design. Contexts, prefix lengths and cache controls reuse
the same states and documents.

Small cache error is conditional on both the trained state and the
assessment law. Two paired measurements separate these conditions.
First, Table~\ref{model:tab:cache-initial-context} reconstructs the same
16-document calibration and 128-document assessment panel at all twelve
unique initial states and their 24 trained endpoints. The two controls
share each initial baseline. Every seed-wise endpoint-minus-initial
mean-KL contrast is positive in all four endpoint conditions; the complete
24 signed contrasts appear in Table~\ref{model:tab:cache-initial-pairs}.
Initial NLL is near $\log 32000=10.37349$, whereas trained NLL is
substantially lower. Small initial predictive error therefore does not
identify a training-created operator invariance or an equally competent
initial model. This fixed-panel comparison supplies no independent
context replication and does not isolate operator and sensitivity changes.
\begin{table}[htbp]\centering\small
\caption{Paired initialization and trained-state cache context on the same 128 assessment documents at length 64. Initial states are shared between controls; KL and NLL are in nats.}
\label{model:tab:cache-initial-context}
\begin{tabular}{@{}rrlrrrr@{}}
\toprule $N$ & Updates & $g$ & $R$ & Mean KL & Native NLL & Exceed\\\midrule
8 & 0 & shared & 0.038943 & 0.0000060 & 10.37319 & 0\\
8 & 4096 & 0 & 0.041604 & 0.0000909 & 7.14143 & 0\\
8 & 4096 & 1.5 & 0.124209 & 0.0007113 & 7.40437 & 3\\
24 & 0 & shared & 0.038714 & 0.0000060 & 10.38060 & 0\\
24 & 4096 & 0 & 0.041866 & 0.0000687 & 7.09867 & 0\\
24 & 4096 & 1.5 & 0.098890 & 0.0004332 & 7.38094 & 1\\
\bottomrule\end{tabular}\end{table}

Second, a completed state-transfer experiment uses 80 further
content-distinct RefinedWeb documents, disjoint from training and all
reserved observation panels. Sixteen calibrate and 64 assess caches at
proper-prefix lengths 32, 64 and 128. The target token remains outside
each model call. For each of six paired initialization identities at
$N=8,24$, we acquire its initial state and both 4,096-update endpoints
at $g=0,1.5$. The generator initialization, corpus and training source
order remain conditioned. Each state receives its own calibrated cache;
each trained endpoint also receives the unchanged cache of its paired
initial state at the same length. The declared hypotheses are the finite
risk and state-displacement identities and the fixed accuracy target for
state-specific recalibration. The initial-cache arm measures transport;
it carries no assumption of small displacement. All 30 cells are retained.

All eighteen recalibrated state/length cells meet centered RMS $0.25$
and mean KL $0.03$ nats: RMS ranges from $0.03273$ to $0.17044$ and
mean KL is at most $0.001684$ nats. The largest individual-context
exceedance count is six of 64. All six zero-control initial-cache cells
also meet both targets, with RMS $0.03915$--$0.05046$. For positive
control, the initial-cache RMS is $1.05765$--$1.20826$, with maximum
cell mean KL $0.08975$ nats. These outcomes distinguish accuracy at a
fixed state from portability of the same operator across trained states.
Table~\ref{model:tab:cache-state-main} summarizes every state and prefix range;
Tables~\ref{model:tab:cache-state-complete}--\ref{model:tab:cache-state-budget}
retain the full grid, context exceedances and signed transport budgets.
\begin{table}[htbp]\centering\footnotesize
\caption{State-conditioned cache transfer on 64 fresh assessment documents. Ranges cover lengths 32, 64 and 128. $R_t$ uses a cache recalibrated at the displayed state; $R_0$ uses its paired initial-state cache. The same six initialization identities occur in every row of a width.}
\label{model:tab:cache-state-main}
\begin{tabular}{@{}rlrrrr@{}}
\toprule $N$ & State & Range of $R_t$ & Largest mean KL$_t$ & Range of $R_0$ & Largest mean KL$_0$\\\midrule
8 & initial & 0.03331--0.04401 & 0.000008 & same & same\\
8 & $g=0$ & 0.04103--0.04892 & 0.000149 & 0.04199--0.05045 & 0.000158\\
8 & $g=1.5$ & 0.10562--0.17043 & 0.001684 & 1.05765--1.20825 & 0.089746\\
24 & initial & 0.03273--0.04366 & 0.000008 & same & same\\
24 & $g=0$ & 0.03797--0.04393 & 0.000098 & 0.03915--0.04435 & 0.000100\\
24 & $g=1.5$ & 0.12496--0.14581 & 0.001104 & 1.06936--1.10440 & 0.084305\\
\bottomrule\end{tabular}\end{table}

The operator measurements explain this separation in the tested family.
The per-layer norm is Frobenius norm divided by $\sqrt{Nd^2}$, and
reported squared risks average five layers and six seeds. At length 64,
positive-control squared mean drift from initialization is $2.59858$
and $2.70411$ at widths 8 and 24. Assessment scatter is only
$1.4330\times10^{-4}$ and $1.5079\times10^{-4}$, respectively. Recalibrated
mean displacement is $1.1689\times10^{-5}$ and $1.2859\times10^{-5}$.
The corresponding zero-control mean drifts are $6.1734\times10^{-6}$ and
$5.0515\times10^{-6}$. Thus the positive-control trajectory selects a
substantially different common operator while leaving a much smaller
within-state context dispersion on this panel. The signed cross term in
Equation~\eqref{model:eq:cache-state-transport} is measured explicitly; replacing
it by an unsigned sum would change the identity. Predictive secants
also vary with state and intervention amplitude, so these observations
are not uniform sensitivity estimates or a causal mediation proof.

In the following figure, operator dispersion is context centered at a fixed
checkpoint, whereas predictive KL compares paired native and cached prefix
laws. The relative centered-RMS statistic uses its stated predictive
normalization. Averages over layers, documents and repeated prefix lengths
retain the same six initialization identities per width.

\begin{figure}[htbp]\centering
\includegraphics[width=\textwidth]{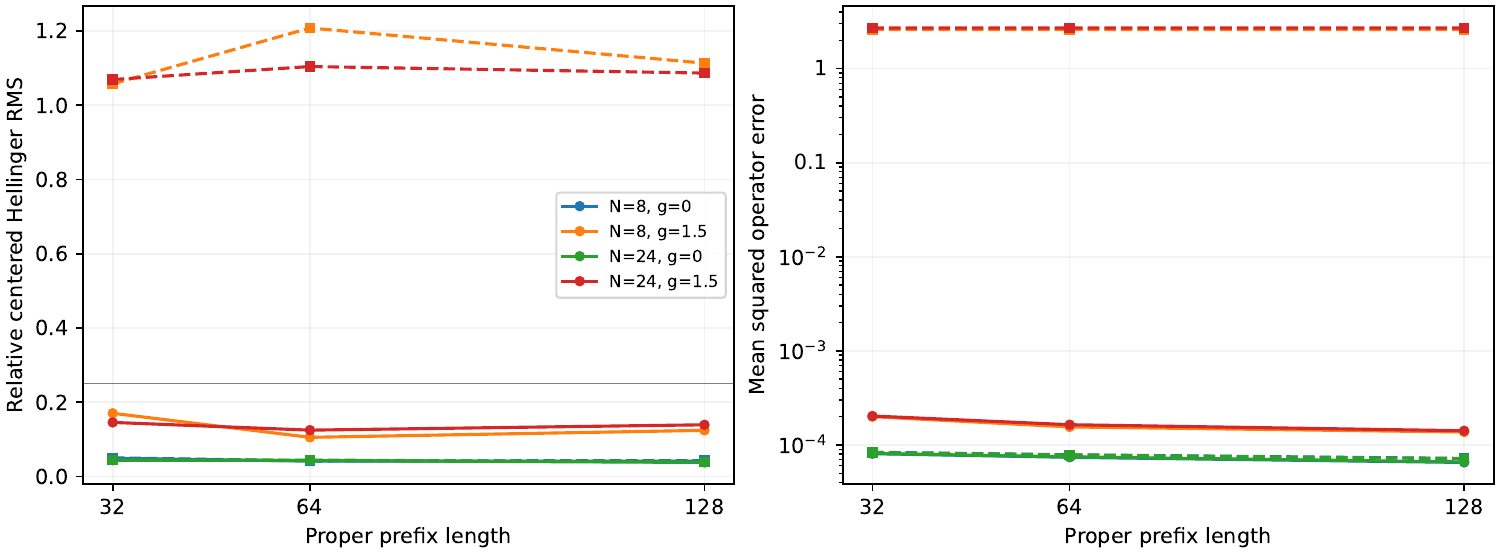}
\caption{State-specific recalibration (solid circles) and transport of the
paired initial cache (dashed squares) at trained endpoints. The predictive
and operator errors are separate observables. Colors identify width and
control; repeated prefix lengths share all trained states and documents.}
\label{model:fig:cache-state-transfer}
\end{figure}

The experiment executed 216 calibration, 864 native assessment,
864 recalibrated assessment, 576 initial-cache assessment and 504
qualification forwards, totaling 3,024. It added no optimizer updates
or independent trained replicas. Exact initial parameter identity,
own-operator replay, native restoration and generator bypass were
qualified for every state and length; longest-prefix batch-one checks
preceded batch-eight acquisition. Summed worker time was 851.0 seconds
on two RTX 4090 cards, with maximum allocated memory 1.154 GiB.
Independent raw-moment and pair-distance reconstructions have maximum
operator and predictive discrepancies below $2.06\times10^{-15}$ and
$2.69\times10^{-15}$. These are accuracy and transport measurements,
with no new timing, intermediate-training-age or generated-history claim.

\paragraph{Disjoint-context confirmation.}
The second panel reuses the same six paired initialization identities
per width, twelve initial states and twenty-four trained endpoints.
Its sixteen calibration and sixty-four assessment documents are disjoint
from each other, the preceding panel, reserved observations and training.
The state, proper-prefix lengths 32, 64 and 128, cache-policy and accuracy
definitions are unchanged. The panel tests context transfer and
recalibration variation; it adds no independent training realization.

All eighteen recalibrated cells and six zero-control initial-cache cells
pass centered RMS $0.25$ and mean KL $0.03$ nats; all six positive-control
initial-cache cells fail. Within the recalibrated positive-control group,
19 cell--context centered-RMS exceedances above $0.25$ remain, with maximum
individual-context KL approximately $0.07522$ nats. These occurrences can
reuse documents across cells and are not 19 independent training failures.
The result supports aggregate state-specific fidelity on the tested
distribution, with explicit context exceptions.

For recalibrated cells, centered RMS is $0.03250$--$0.14382$ and mean KL is
at most $0.001154$ nats; context exceedances reach six of 64.
The positive-control initial-cache cells exhibit the distinct mean-motion
contribution. Table~\ref{model:tab:cache-confirmation-main} retains both policies.
Tables~\ref{model:tab:cache-confirmation-complete}--\ref{model:tab:cache-confirmation-budget}
give all 30 predictive cells, native and residual variances, mean bias,
signed covariance, NLLs, context exceptions and operator budgets.
The fidelity targets quantify agreement with the native law.
Target NLL sometimes improves under the displaced initial cache;
Equation~\eqref{model:eq:cache-target-risk} explains why that improvement
does not certify predictive agreement.

\begin{table}[htbp]\centering\small
\caption{Disjoint-context confirmation on 64 additional assessment documents with 16 additional calibration documents. Ranges cover all three prefix lengths. The model identities and control pairing are unchanged.}
\label{model:tab:cache-confirmation-main}
\begin{tabular}{@{}rlrrrr@{}}
\toprule $N$ & State & State-cache RMS & Largest mean KL & Initial-cache RMS & Largest mean KL\\\midrule
8 & initial & 0.03323--0.04344 & 0.000008 & same & same\\
8 & $g=0$ & 0.04158--0.04840 & 0.000158 & 0.04278--0.05033 & 0.000172\\
8 & $g=1.5$ & 0.10698--0.13624 & 0.001153 & 1.06494--1.20889 & 0.084054\\
24 & initial & 0.03250--0.04339 & 0.000007 & same & same\\
24 & $g=0$ & 0.03770--0.04555 & 0.000102 & 0.03887--0.04553 & 0.000102\\
24 & $g=1.5$ & 0.11174--0.14382 & 0.000923 & 1.05043--1.05605 & 0.075829\\
\bottomrule\end{tabular}\end{table}

Whole-seed paired differences between panels, including each sign,
appear in Table~\ref{model:tab:cache-confirmation-contrasts}. The exact empirical
bootstrap enumerates all $6^6=46{,}656$ ordered resamples of complete
initialization identities. The panels are fixed, so these conditional
intervals do not include an outer corpus-distribution uncertainty.
The unchanged seeds, widths, controls and prefixes are paired observations
and do not become additional independent replicas.

The confirmation adds 3,024 calls with the same five-role allocation as
the state-transfer measurement, zero optimizer updates, and 596.5 summed
GPU-worker seconds. Maximum allocated memory is 1.154 GiB.
Independent reconstruction of predictive and operator quantities agrees
within $3.56\times10^{-15}$ and $1.88\times10^{-15}$, respectively.
Together the two disjoint panels support a state-conditioned effective
operator with small within-state context scatter, while explicitly
retaining the separate training-induced mean displacement.

\paragraph{Transfer of a fixed calibration to further contexts.}
A separately frozen sixteen-document assessment reuses the preceding
panel's length-64 caches without recalibration. The documents are
content-distinct from all calibration, assessment and training panels.
All six state-specific cells meet both accuracy targets, with centered
RMS $0.03765$--$0.13530$ and mean KL at most $0.00094001$ nats.
The two zero-control initial-cache cells also meet the targets. Both
positive-control initial-cache cells exceed them, with centered RMS
$1.11646$ and $1.16609$; all sixteen contexts in each cell exceed $0.25$.
The state-specific width-eight positive-control cell has two exceptions,
with maximum context RMS $0.31304$. These outcomes support conditional
aggregate transfer and retain the distinction from uniform context accuracy.

\begin{table}[htbp]\centering\small
\caption{Fixed-cache transfer to sixteen further documents. $R$ is relative centered predictive RMS; exceptions count contexts with $r_c>0.25$. $\Delta\mathrm{NLL}$ is cached minus native target loss in nats. Six complete initialization identities are paired within each width; contexts are not independent training replicas.}
\label{model:tab:fixed-cache-panel}
\begin{tabular}{@{}rllrrrr@{}}
\toprule $N$ & State & Cache & $R$ & Mean KL & Exceptions & $\Delta\mathrm{NLL}$\\\midrule
8 & initial & state & 0.03912 & 0.00000609 & 0 & -0.00025\\
8 & $g=0$ & state & 0.03869 & 0.00008235 & 0 & -0.00020\\
8 & $g=0$ & initial & 0.04302 & 0.00010026 & 0 & -0.00068\\
8 & $g=1.5$ & state & 0.13530 & 0.00094001 & 2 & +0.00617\\
8 & $g=1.5$ & initial & 1.16609 & 0.07203175 & 16 & -0.03179\\
24 & initial & state & 0.03954 & 0.00000622 & 0 & +0.00050\\
24 & $g=0$ & state & 0.03765 & 0.00005844 & 0 & +0.00062\\
24 & $g=0$ & initial & 0.03847 & 0.00006189 & 0 & -0.00002\\
24 & $g=1.5$ & state & 0.08938 & 0.00031354 & 0 & +0.00195\\
24 & $g=1.5$ & initial & 1.11646 & 0.05783725 & 16 & -0.05754\\
\bottomrule
\end{tabular}
\end{table}

The two positive-control initial-cache cells have target-loss changes
$-0.03179$ and $-0.05754$ nats despite their large native-law discrepancies.
Thus predictive fidelity, external-target loss and computational cost
remain separate observables. The measurement adds 192 assessment and
72 qualification forwards, totaling 264, with no optimizer updates or
new training replicas. Independent CPU reconstruction from every raw
logit archive agrees with the separately retained reduction within
$2.03\times10^{-15}$. No latency or generated-history claim follows from
this finite context transfer.

\section{Predictive visibility under categorical elimination}
\label{model:sec:categorical-visibility-results}

A finite observation reduction can be evaluated at the scale of native
predictive fluctuations. We fix a vocabulary ordering and conditional tail
law from the six development initializations at $N=14,g=0,T=2048$ on
the first corpus partition. The ordering ranks mean probability across the
eight fixed contexts. The tail dictionary keeps a separate distribution
for each context. It is fixed before evaluating the retained independent
family and is shared by all its widths and controls, including $N=24$.
The comparison reuses completed native endpoints; it acquires no new
training realization and performs no additional forward pass.

For each declared retained set of 128, 512, 2,048, 8,192 and 16,384 tokens,
$Kp$ preserves those token probabilities and one tail mass.
Proposition~\ref{model:prop:categorical-tail-visibility} supplies compatible
elimination maps, a fixed predictive lift and exact KL and variance budgets.
We test every one of the 216 endpoint paths, giving 36 width/control cells
with six independent initializations each. The native variance uses the
full vocabulary sum and the same eight contexts as the coarse variance.
The finite accuracy target is centered reconstruction RMS at most one
quarter of the native across-initialization RMS fluctuation. This numerical
target is fixed in the observation protocol and is not an asymptotic
vanishing-remainder condition.

At 8,192 retained tokens, all 36 cells meet the finite target. The
observation preserves 95.2--98.0\% of native predictive variance, with
relative centered RMS 0.150--0.226. At 16,384 tokens those ranges improve
to 97.5--98.9\% and 0.107--0.163. Table~\ref{model:tab:categorical-visibility}
retains all five declared resolutions, including the smaller alphabets
whose missing tail directions remain appreciable at this fluctuation scale.
An independent direct-pair reduction checks all 180 cells and the
conditional-tail KL identity. The measured observation is invariant under
the native head-sign action because the entire probability law is invariant.

\begin{figure}[htbp]\centering
\includegraphics[width=\textwidth]{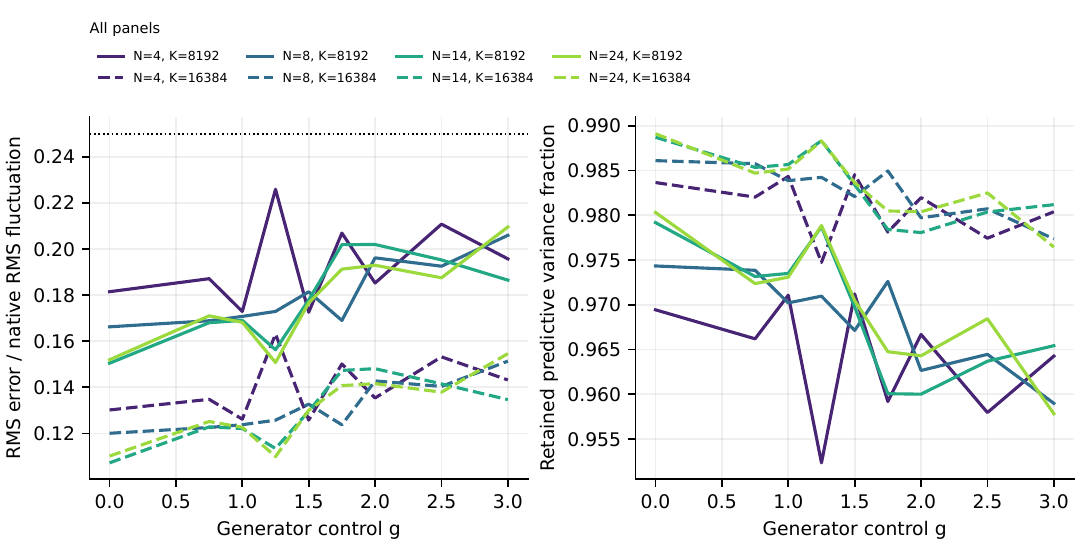}
\caption{Predictive visibility in the independent single-pass family at
$T=4096$. Solid curves retain 8,192 tokens; dashed curves retain 16,384.
The horizontal dotted line is the predeclared finite RMS target.
Each point summarizes six complete initializations with fixed context
weights. The dictionary is shared across all displayed cells.}
\label{model:fig:categorical-visibility}
\end{figure}

The uncentered reconstruction RMS at 8,192 tokens is 0.541--0.745
times the native RMS fluctuation. Its larger size retains the dictionary
bias, which centering removes. The mean KL errors in the table measure
full emission discrepancy separately from covariance preservation.

This supplies a tested finite predictive observation across the row regimes.
It does not provide a reduced optimizer successor law. Acquisition uses the
full native logits; the full-vocabulary tail dictionary is stored once per
context, in addition to the coarse probability vector. No forward-compute
saving, refresh-cost advantage or economical autonomous training closure is
inferred from its smaller observation dimension. Susceptibility multiplies
both native and retained variance by the same $N$, so the reported fractions
also describe finite susceptibility retention. Critical-exponent transport
would still require the limiting normalization and vanishing relative
remainder in Proposition~\ref{model:prop:vector-fluctuation-visibility}.

\FloatBarrier
\par\medskip\noindent
Chapter~\ref{ch:finite-adaptation} returns to evolving optimizer states.
It explains how finite pulses, memory and adaptation sources can change the
inference law whose frozen reductions were assessed here.

\chapter{Finite optimizer transport and autoregressive prediction}
\label{ch:finite-adaptation}
This chapter develops finite optimizer and source transport into
autoregressive prediction. It treats pulse amplitudes, retained memory,
prefix consistency, task margins and coupled adaptation sectors, keeping
finite interventions distinct from infinitesimal response claims.

\section{Finite native pulses and their response domain}
\label{model:app:finite-native-transport}
\label{model:sec:finite-pulse}

A derivative approximation needs a specified amplitude and time domain.
The complete-state training map provides a finite description even when
that approximation is inaccurate. Fix an incoming state $s$, an admissible
parameter direction $v$, and a without-replacement source sequence $\omega$
from its remaining corpus. Let $F_t(h,\omega)$ be the retained observation
after $t$ native updates from the parameter pulse $\theta+h v$, with the
same incoming moments, counters and source indices for every $h$.
Subsequent moments and both parameter families evolve normally. This
coupling is available because the sampling law here is independent of
the parameters, conditional on the remaining corpus. An adaptive sampler
would require its own common-randomness construction.

\begin{proposition}[Compatible finite-pulse law]
\label{model:prop:finite-pulse}
For any nonzero $h$, define observation-path vectors
\[
 X=F(0,\omega),\qquad
 O_h=\frac{F(h,\omega)-F(-h,\omega)}2,\qquad
 E_h=\frac{F(h,\omega)+F(-h,\omega)}2-X.
\]
Then $F(\pm h,\omega)=X+E_h\pm O_h$ exactly. Every deterministic
linear path map $B$ sends $(X,E_h,O_h)$ to $(BX,BE_h,BO_h)$;
these maps compose in the same order as the observation maps.
Under finite second moments, the full covariance at either sign is
\begin{align}
 \Cov(F(\pm h))={}&\Cov X+\Cov E_h+\Cov O_h
       +\Cov(X,E_h)+\Cov(E_h,X)\notag\\
 &\quad\pm\{\Cov(X,O_h)+\Cov(O_h,X)
            +\Cov(E_h,O_h)+\Cov(O_h,E_h)\}.
 \label{model:eq:pulse-covariance}
\end{align}
No differentiability or independence of these sectors is required.
If, additionally, $h\mapsto F(h,\omega)$ is three times continuously differentiable
on $[-a,a]$ with third derivative norm at most $M_3$ and second
derivative norm at most $M_2$, in the fixed path norm, then
\begin{alignat}{2}
 \left\|\frac{O_h}{h}-\partial_hF(0,\omega)\right\|
       &\le M_3h^2/6, &\quad \|E_h\|&\le M_2h^2/2,\label{model:eq:pulse-taylor}\\
 \left\|\frac{O_h}{h}-\frac{O_{h/2}}{h/2}\right\|
       &\le 5M_3h^2/24, &\quad &0<|h|\le a.\label{model:eq:pulse-halving}
\end{alignat}
Uniform bounds give the same conclusions in conditional $L^2$;
more generally square-integrable random derivative bounds suffice.
\end{proposition}
\begin{proof}
Adding and subtracting the two pulse definitions proves reconstruction.
Linearity proves their transport and associativity proves composition.
Apply Lemma~\ref{model:lem:master-perturbation} first to $X$ and
$E_h\pm O_h$, then expand the covariance of the latter sum with
the same lemma. This gives all ordered terms in
Equation~\eqref{model:eq:pulse-covariance}.
For the additional estimates, Taylor's integral remainder at zero
bounds the second-order remainder in $F(h)$ by $M_2h^2/2$ and the
third-order remainder after the quadratic polynomial by
$M_3|h|^3/6$. The symmetric average cancels the linear term;
the antisymmetric difference cancels the quadratic term. Taking norms
proves Equation~\eqref{model:eq:pulse-taylor}. Apply its first bound at
$h$ and $h/2$ and use the triangle inequality to obtain
$(1+1/4)M_3h^2/6$. Taking conditional $L^2$ norms gives the final
claim whenever the bounds are square-integrable.
\end{proof}

The finite law acts on the complete path observation. The derivative
bound is an additional condition on the composed native training and
emission map; a short-step bound does not control its growth with $t$.
Crossing a clipping or activation boundary and finite arithmetic require
explicit residuals. Agreement of two measured amplitudes is a domain
diagnostic, not an upper bound on $M_3$ or a proof of differentiability.
The pulse direction has zero incoming moment components, so it probes
one direction of augmented-state evolution, not its full Jacobian.
Random response products still require the conditional and outer-mean
sectors of Proposition~\ref{model:prop:master-chronological}.

This representation retains finite nonlinear response through $E_h$
and amplitude dependence through $O_h$. A temporal covariance or an
observed return over a finite window therefore does not, by itself,
identify a scalar relaxation gap. The executed paired native study in
Section~\ref{model:sec:directional-results} measures the response domain and
retains the signed finite covariance at all tested amplitudes.

\section{Finite optimizer memory as a source for inference}
\label{model:sec:finite-optimizer}
Finite response can intervene on the complete state, including optimizer
memory. An optimizer-only intervention leaves the incoming inference
law unchanged but can change its successor. At fixed parameters and
the same first minibatch, its clipped gradient $g$ is unchanged as well.
This observation gives an exact finite transport calculation without
differentiating clipping, the generator or the native training trajectory.
We use the Adam moments and decoupled decay conventions of
\cite{kingma2015adam,loshchilov2019adamw}.

\begin{proposition}[Finite optimizer-state transport]
\label{model:prop:finite-optimizer}
Fix a batch, parameters $\theta$, counter $k\ge0$, and coordinate rates
$\eta_i$, with $0\le\beta_j<1$ and denominator offset $\epsilon_o>0$.
Put $a_j=1-\beta_j^{k+1}$ and
\[
 M=\beta_1m+(1-\beta_1)g,\qquad
 V=\beta_2v+(1-\beta_2)g^{\odot2},\qquad
 D_i=\sqrt{V_i/a_2}+\epsilon_o,
\]
where $v_i\ge0$ and all expressions denote real arithmetic. Consider
the update $\theta_i^+=(1-\eta_i\lambda_d)\theta_i-
\eta_i M_i/(a_1D_i)$.
An incoming first-moment displacement $m\mapsto m+u$ gives
\begin{equation}
 \Delta M=\beta_1u,\quad\Delta V=\bm0,\qquad
 \Delta\theta_i^+=-\frac{\eta_i\beta_1u_i}{a_1D_i}.
 \label{model:eq:finite-first-moment}
\end{equation}
An incoming second-moment displacement $v_i\mapsto e^{h d_i}v_i$
gives
\begin{alignat}{2}
 \Delta M&=\bm0,&\quad \Delta V_i&=\beta_2v_i(e^{h d_i}-1),\notag\\
 D_i(h)&=\sqrt{[V_i+\Delta V_i]/a_2}+\epsilon_o,&\quad \Delta\theta_i^+&=\frac{\eta_iM_i}{a_1}
           \left(\frac1{D_i}-\frac1{D_i(h)}\right).
 \label{model:eq:finite-second-moment}
\end{alignat}
The multiplicative intervention preserves the zero support of $v$.
For both interventions the incoming inference emission is identical;
the outgoing emission is the full graph evaluated at
$\theta^++\Delta\theta^+$.
\end{proposition}
\begin{proof}
The gradient and clipping use only the unchanged parameters, batch
and stochastic-program choices. Subtracting the first-moment recursions
therefore gives $\Delta M=\beta_1u$, while the second-moment recursion
is unchanged. Subtract the parameter updates to obtain
Equation~\eqref{model:eq:finite-first-moment}. For the second intervention,
subtracting the affine second-moment recursion gives the displayed
$\Delta V$; $M$ and the decayed incoming parameter term are unchanged.
Taking the difference of their two positive-denominator update terms
gives Equation~\eqref{model:eq:finite-second-moment}. Since $e^{h d_i}>0$,
zero and positive second moments retain their supports. The incoming
network uses $\theta$ alone. Evaluating the outgoing network uses the
two calculated parameter values and proves the emission assertion.
\end{proof}

\begin{corollary}[Obstruction to parameter-only closure]
\label{model:cor:optimizer-closure-obstruction}
Suppose $\beta_1>0$ and $\eta_i>0$ for some coordinate $i$ with
$u_i\ne0$. The two complete states in
Equation~\eqref{model:eq:finite-first-moment} have the same incoming parameters
and different outgoing parameters under every shared first batch.
Thus the parameter projection is not exactly lumpable on a domain
containing those two states for a deterministic batch law. For a common
random batch law the conclusion also holds if the outgoing coordinate
has finite first moment under both states.
\end{corollary}
\begin{proof}
The denominator in Equation~\eqref{model:eq:finite-first-moment} is strictly
positive, so its displacement is nonzero with the same sign for every
batch. For a deterministic batch, identical incoming projected states
have different outgoing point masses, contradicting lumpability. For
a random batch the coupled outgoing difference has that fixed strict
sign. Its finite expectation is consequently nonzero, so the two
outgoing marginal laws cannot coincide.
\end{proof}

The corollary concerns a domain of admissible optimizer interventions,
not the frequency of such states in naturally sampled training histories.
A coarser predictive projection may erase the parameter displacement;
visibility must therefore be measured separately. After the first
update, changed parameters generally change subsequent gradients.
Equations~\eqref{model:eq:finite-first-moment}--\eqref{model:eq:finite-second-moment}
do not provide an autonomous eight-update moment law. The complete-state
kernel and its finite pulse sectors provide that chronological description.
Numerical implementations require the explicit arithmetic residual of
Section~\ref{model:sec:numerical-response}. The experiment in
Section~\ref{model:sec:optimizer-results} tests the finite state prediction and
then measures its inference visibility on fresh single-pass continuations.

\section{From operator stabilization to autoregressive inference}
\label{model:sec:inference-mechanism}

An explanation of the reference association between near-critical
pretraining, row concentration and reasoning must connect three
different observations: the learned metric, the causal predictive law,
and correct answers on a declared task distribution. Here reasoning
means performance on specified inference tasks whose answers depend
on the supplied premises. Grammatical continuation, a small row
statistic and a small deductive-output variation are separate
observations. None is a definition of task correctness.

\subsection{The global metric and prefix consistency}
Let $z_t(x_{1:n})$ denote the vocabulary logits at position $t$ in a
parallel native forward on $n$ input tokens. Autoregressive generation
instead uses $z_t(x_{1:t})$. The native metric generator contains
\begin{equation}
 B_l(x_{1:n})=\sum_{j=1}^n q_{l,j}q_{l,j}^{\mathsf T},\qquad
 A_l=F_l(\LN(B_l)),\qquad G_l=\mathcal G_l(A_l).
 \label{model:eq:global-prefix-metric}
\end{equation}
The causal mask restricts the subsequent attention sum. It does not
restrict the positions entering $B_l$. Consequently the mask alone
does not prove
$z_t(x_{1:n})=z_t(x_{1:t})$. For $t<n$, the next target $x_{t+1}$
is itself among the input tokens of the parallel forward. Fixed-length
suffix interventions directly test this dependence without changing
the true next target or any prefix token.

\begin{proposition}[Full-window to proper-prefix risk transfer]
\label{model:prop:causal-risk-transfer}
For one context, let $z_j^F=z_j(x_{1:n})$ and
$z_j^P=z_j(x_{1:j})$ be finite vocabulary-logit vectors. Set
$p_j^F=\softmax(z_j^F)$, $p_j^P=\softmax(z_j^P)$, and
\[
 v_j=z_j^P-z_j^F,\qquad
 \omega_j=\max_a v_{j,a}-\min_a v_{j,a}.
\]
For every target $y$,
\begin{align}
 |\log p_j^P(y)-\log p_j^F(y)|&\le\omega_j,
 \label{model:eq:causal-score-bound}\\
 \KL(p_j^F\Vert p_j^P)&\le
 \min\{\omega_j,\omega_j^2/8\}.
 \label{model:eq:causal-kl-bound}
\end{align}
For any nonnegative target weights $a_j$ with $\sum_j a_j>0$,
the correspondingly weighted full-window and proper-prefix cross
entropies satisfy
\begin{equation}
 |L_P-L_F|\le
 \frac{\sum_j a_j\omega_j}{\sum_j a_j}.
 \label{model:eq:causal-risk-bound}
\end{equation}
These statements are invariant under adding any scalar to either
logit vector.
\end{proposition}
\begin{proof}
Suppress $j$ and put
$A(s)=\log\sum_a\exp(z_a^F+s v_a)$.
The identity
$\log p^P(y)-\log p^F(y)=v_y-[A(1)-A(0)]$
and
$\min_a v_a\le A(1)-A(0)\le\max_a v_a$
prove Equation~\eqref{model:eq:causal-score-bound}. Also
\[
 \KL(p^F\Vert p^P)=A(1)-A(0)-A'(0)
 =\int_0^1(1-s)A''(s)\,ds.
\]
Here $A''(s)$ is the variance of $v$ under
$\softmax(z^F+s v)$. Subtract the midpoint of the interval
containing its coordinates. Its absolute value is at most
$\omega/2$, so its variance is at most $\omega^2/4$.
Integration gives $\omega^2/8$; the range bound on
$A(1)-A(0)$ and on $A'(0)$ gives the other bound $\omega$.
The triangle inequality for the weighted target losses proves
Equation~\eqref{model:eq:causal-risk-bound}. Scalar shifts change neither
probabilities nor the range of $v$.
\end{proof}

The target weights may be the native nonpadding mask, conditional
on the realized context. Native floating-point cross entropy is a
separate arithmetic observation: its discrepancy from the log scores
in the proposition must also be included when applying a bound to
reported native losses. The external target at $j=n$ has the same
available prefix in both forwards; positions $j<n$ test the effect
of the additional suffix through the global metric.

For a proper conditional predictor $q(\cdot\mid X_{1:j})$ under a
specified token law, direct conditioning gives
\[
 \E[-\log q(X_{j+1}\mid X_{1:j})]
 =H(X_{j+1}\mid X_{1:j})+
 \E\KL\bigl(p_*(\cdot\mid X_{1:j})\Vert
                  q(\cdot\mid X_{1:j})\bigr).
\]
This identity concerns an unmasked conditional risk, or the
corresponding explicitly conditioned target law. A full-window
prediction with access to $X_{j+1}$ through its metric need not obey
that conditional-entropy lower bound. For example, a uniform binary target
independent of its prefix has entropy $\log 2$. A predictor that
observes that target and assigns it probability $1-\epsilon$, with
$0<\epsilon<1/2$, has loss $-\log(1-\epsilon)<\log 2$.
Thus assessing transfer from the training objective to inference
requires prefix stability as well as fit. Small row contrast supports
this transfer only with control of the input-dependent common
operator and the downstream logit response. Task competence further
requires correct margins on the specified task law.

\begin{proposition}[Shared row contraction and input stability]
\label{model:prop:shared-row-input-contraction}
Let $U(X)\in\mathbb R^{m\times d}$ be a square-integrable input
row cloud at a fixed checkpoint and decoder layer. Suppose the
shared row map $F=f_k\circ\cdots\circ f_1$ has Lipschitz constant
$K=\prod_{j=1}^k K_j$ on the encountered domains, with each $f_j$
$K_j$-Lipschitz. Write $A_i(X)=F(U_i(X))$. If
$\Psi:A\mapsto G$ is $M$-Lipschitz in Frobenius norm on the
corresponding matrix domain, then
\begin{align}
 \frac1m\sum_i\norm{A_i-\overline A}^2
 &\le K^2\frac1m\sum_i\norm{U_i-\overline U}^2,
 \label{model:eq:shared-row-contraction}\\
 \E\norm{G-\E G}_F^2
 &\le M^2K^2\E\norm{U-\E U}_F^2.
 \label{model:eq:shared-input-contraction}
\end{align}
The first bound holds for every encountered cloud. The second is
conditional on the fixed checkpoint and input law.
\end{proposition}
\begin{proof}
Composition multiplies the Lipschitz bounds. For a finite cloud,
its centered mean squared norm equals
$(2m^2)^{-1}\sum_{i,j}\norm{A_i-A_j}^2$.
Apply the $K$-Lipschitz inequality to every pair and use the same
identity for $U$. This proves Equation~\eqref{model:eq:shared-row-contraction}
without replacing the nonlinear image of a centroid by the centroid
of its image. For two inputs $X,X'$, apply that Lipschitz inequality
to corresponding rows and sum, giving
$\norm{A(X)-A(X')}_F\le K\norm{U(X)-U(X')}_F$.
The $M$-Lipschitz bound then applies to their emitted metrics.
For independent copies, the expected squared pair difference is
twice the centered second moment. Squaring, averaging and applying
that identity proves Equation~\eqref{model:eq:shared-input-contraction}.
\end{proof}

The native residual metric units apply their linear, gated and
normalization operations to the last coordinate of every row with
shared weights. They therefore have the structure used in this
proposition. The subsequent power map supplies $\Psi$ on its
admissible domain. A product of small row-map bounds can suppress
both row differences and variation between contexts, producing a
learned operator that is reusable at inference. Absolute centered
row energy is used here; a normalized row ratio additionally needs
control of its denominator. Measured contraction of finitely many
row or input secants does not certify a uniform Lipschitz bound.
In particular, a small row statistic alone does not establish
Equation~\eqref{model:eq:shared-input-contraction} on unobserved inputs.

The orientation of local transport matters. For differentiable row
maps, let $x_j=f_j(x_{j-1})$ and $J_j=Df_j(x_{j-1})$. The chain
rule gives
\[
 DF(x_0)=J_k\cdots J_1,\qquad
 \norm{DF(x_0)}_2\le\prod_{j=1}^k\norm{J_j}_2.
\]
The right-hand side can be much larger than the norm of the product.
On a convex input domain $D$, a bound
$\sup_{x\in D}\norm{DF(x)}_2\le K_*$ implies a $K_*$-Lipschitz
bound for $F$: integrate $DF$ along the line segment between the two
inputs and apply the norm inequality to that integral. Thus
Proposition~\ref{model:prop:shared-row-input-contraction} also holds with
$K_*$ in place of the separate-unit product bound. The matrix product
retains orientation information discarded by scalar layer norms.
Its measurement at finitely many inputs still does not bound the
supremum over $D$.

The distinction between input contraction and training response is
also explicit in the differential. Write the learned row map as
$F_\theta$ and its normalized Gram input as $U_\theta(x)$. Then
\begin{equation}
 dA=D_UF_\theta\,dU_\theta
       +D_\theta F_\theta\,d\theta.
 \label{model:eq:row-input-training-differential}
\end{equation}
At fixed weights, an input intervention has $d\theta=\bm0$. A training
intervention changes both terms, with its parameter response governed
by the augmented optimizer dynamics. A small $D_UF_\theta$ therefore
can suppress input variation while leaving a learned common operator
sensitive to training. This provides compatible directions for row
concentration at inference and persistent common training modes. It
does not establish that the latter modes are critical.

For the native downstream PLGA map $G=\Psi_\theta(A)$, the same
chain rule gives
\begin{align}
 dG={}&D_A\Psi_\theta(A)D_UF_\theta(U)\,dU\nonumber\\
 &+\bigl[D_A\Psi_\theta(A)D_\theta F_\theta(U)
                  +D_\theta\Psi_\theta(A)\bigr]\,d\theta.
 \label{model:eq:plga-input-parameter-transport}
\end{align}
The parameter derivatives in the bracket hold the displayed function
input fixed. This formula follows by substituting
Equation~\eqref{model:eq:row-input-training-differential} into
$dG=D_A\Psi_\theta\,dA+D_\theta\Psi_\theta\,d\theta$.
A small row-input derivative attenuates the first route while leaving
both parameter routes available. The experimental generator partition
contains the residual learner and downstream PLGA factors; its finite
corner does not isolate these two routes. Body parameters additionally
change prediction through the query, key, value, residual and vocabulary
paths, even when their route through $G$ is attenuated. The full source
composition retains all these contributions.

\begin{proposition}[Contracting contrasts and a retained common mode]
\label{model:prop:transverse-common-modes}
Suppose a differentiable deterministic effective training map has
local coordinates $(r,c)$ around a fixed point $(\bm0,c_*)$, where
$r$ represents row contrasts and $c$ retains the other state
coordinates. Assume $T_r(\bm0,c)=\bm0$ near $c_*$ and
$\norm{D_rT_r(\bm0,c_*)}_2\le\rho<1$. Then its linearization has
block form
\begin{equation}
 DT(\bm0,c_*)=\begin{pmatrix}A&\bm0_{\mathrm{mat}}\\ B&D\end{pmatrix},
 \qquad \operatorname{spec}(DT)=\operatorname{spec}(A)
                            \cup\operatorname{spec}(D).
 \label{model:eq:transverse-common-blocks}
\end{equation}
All contrast multipliers lie strictly inside the unit disk, while
$D$ can have a multiplier equal to one. If $Dv=v$, the corresponding
full-state eigenvector is $(\bm0,v)$. At a fixed inference context, let
$z$ be the vocabulary logits, $q=\softmax z$, and
$w=D_cz(\bm0,c_*)v$. This mode changes the predictive law to first order
exactly when $w\notin\operatorname{span}\{\mathbf1\}$. Equivalently,
\begin{equation}
 w^{\mathsf T}\bigl(\diag q-qq^{\mathsf T}\bigr)w>0.
 \label{model:eq:common-mode-predictive-visibility}
\end{equation}

For a noisy transverse linearization
$r_{t+1}=A_t r_t+\xi_{t+1}$, suppose $A_t,r_t$ are measurable
with respect to the incoming history $\mathcal F_t$,
$\norm{A_t}_2\le\rho$,
$\E[\xi_{t+1}\mid\mathcal F_t]=\bm0$ and
$\E[\norm{\xi_{t+1}}^2\mid\mathcal F_t]\le\sigma^2$.
Then
\begin{equation}
 \E\norm{r_t}^2\le\rho^{2t}\E\norm{r_0}^2
       +\frac{\sigma^2(1-\rho^{2t})}{1-\rho^2}.
 \label{model:eq:transverse-noise-floor}
\end{equation}
\end{proposition}
\begin{proof}
Differentiating the identity $T_r(\bm0,c)=\bm0$ in $c$ gives the zero
upper-right block. The determinant of the block triangular matrix
$\lambda I-DT$ is the product of the two diagonal-block
determinants, which proves the spectral statement. The norm bound
on $A$ bounds every contrast eigenvalue in modulus by $\rho$.
If $Dv=v$, block multiplication gives
$DT(\bm0,v)=(\bm0,v)$. Differentiating softmax gives the predictive
variation $(\diag q-qq^{\mathsf T})w$, whose $i$th entry is
$q_i(w_i-\sum_jq_jw_j)$. Every $q_i$ is positive, so this vector
vanishes exactly when all $w_i$ are equal. Its associated quadratic
form is $\sum_iq_i(w_i-\sum_jq_jw_j)^2$, which is positive under
exactly the complementary condition. This proves
Equation~\eqref{model:eq:common-mode-predictive-visibility} and excludes
constant-logit gauge directions. For the stochastic recursion, the
conditional expectation of the cross term
$\langle A_t r_t,\xi_{t+1}\rangle$ vanishes. Hence
$\E\norm{r_{t+1}}^2\le\rho^2\E\norm{r_t}^2+\sigma^2$.
Iteration and the finite geometric series give
Equation~\eqref{model:eq:transverse-noise-floor}.
\end{proof}

This conditional local geometry makes contraction of contrasts
compatible with a slow, predictively visible common training mode.
It requires a justified effective state, including optimizer memory;
it is not inferred from the row statistic alone. Nor does a unit
multiplier alone identify thermodynamic criticality. A critical
interpretation still needs the limiting collective law and response
conditions stated in Section~\ref{model:sec:joint-scaling}.
The matrices $A,D$ describe a training linearization. They are
not the fixed-weight input Jacobian $D_UF_\theta$ in
Equation~\eqref{model:eq:row-input-training-differential}. The latter
can validate local input contraction without measuring the former.

\begin{proposition}[Prefix consistency with fixed operators]
\label{model:prop:fixed-operator-prefix}
Consider a finite decoder whose token embeddings, normalizations and
feed-forward layers act positionwise, whose positional coordinates
agree on equal prefixes, and whose attention is causally masked.
Replace each layer's global metric by a fixed matrix, independently
of tokens occurring after the prediction position. If the same
matrices are used in two forwards with equal prefixes of length $t$,
their hidden states and logits agree at every position up to $t$,
in exact arithmetic.
\end{proposition}
\begin{proof}
The embeddings and positional coordinates agree on the common prefix.
Suppose the incoming states agree there at a layer. Its prefix queries,
keys and values then agree. Multiplication by the same fixed metric
preserves this equality. At a position $j\le t$, the causal attention
sum uses only keys and values at positions at most $j$. Thus its
scores, normalized weights and output agree. The residual,
normalization and feed-forward operations preserve equality at each
position. Induction over the finite layers proves the assertion, and
the final positionwise vocabulary projection preserves it.
\end{proof}

The collapsed row family also retains the complete learned PLGA
transformation. Suppressing layer and head indices, substitution of
$A=\mathbf1c^{\mathsf T}$ in Equation~\eqref{model:eq:plga} gives
\begin{equation}
 G(c)=a\left[f\bigl((W\mathbf1)c^{\mathsf T}+b\bigr)
                  +\epsilon_A\right]^{\odot P}+b^a.
 \label{model:eq:collapsed-row-plga}
\end{equation}
The matrix biases and entrywise exponents remain in this expression.
There is no rank-one constraint on $G$: for example, the allowed
choice $a=\bm0_{d\times d}$, $b^a=G_*$ realizes any prescribed finite matrix $G_*$
on the collapsed row family. This is an exact statement of retained
architectural capacity, not an assertion that a trained checkpoint
uses this particular choice. Moreover, the common row $c$ can vary
between inputs unless its own transport is controlled.

The fixed operators may therefore be learned and highly nontrivial. In
particular, fixed $G_l$ leaves the scores
$q_{l,i}^{\mathsf T}G_l k_{l,j}$ dependent on the context through
queries and keys. Stabilization of the metric therefore permits
context-dependent attention and composition across layers. It does
not require token hidden states to become equal.

Row concentration does not by itself supply the hypothesis of
Proposition~\ref{model:prop:fixed-operator-prefix}. For example,
$A(x)=\mathbf1 c(x)^{\mathsf T}$ has zero centered row energy for
every $x$, while its common row can vary arbitrarily with the suffix.
A metric map that responds to $c(x)$ retains that variation. The
reduced inference state must therefore retain both the common-row
field and its predictive transport. It cannot replace them by the
centered row norm. Likewise a frozen generator can emit a
context-dependent metric through its input Gram matrix.

\begin{proposition}[A quantitative prefix defect]
\label{model:prop:prefix-defect-bound}
At a fixed input, let the prefix-state difference entering layer $l$
have norm $e_{l-1}$. Suppose its two native computations obey
\[
 e_l\le a_l e_{l-1}+b_l\delta_l,\qquad e_0=0,
\]
where $a_l,b_l\ge0$ and $\delta_l$ bounds their metric difference
in the declared matrix norm. If the final logit map is
$c$-Lipschitz from this state norm to the vocabulary infinity norm,
then
\begin{equation}
 \norm{z_t(x_{1:n})-z_t(x_{1:t})}_\infty
 \le c\sum_{l=1}^L b_l\delta_l\prod_{j=l+1}^L a_j
 =:\varepsilon_t.
 \label{model:eq:prefix-defect-bound}
\end{equation}
The corresponding difference in the negative log probability of any
one target is at most $2\varepsilon_t$. The predictive relative
entropy is at most
$\min\{2\varepsilon_t,\varepsilon_t^2/2\}$.
\end{proposition}
\begin{proof}
Repeated substitution in the layer recurrence gives the displayed
sum for $e_L$. Apply the final Lipschitz bound. If two logit vectors
differ by at most $\varepsilon$ in every coordinate, their
log-sum-exp values differ by at most $\varepsilon$, by monotonicity
and translation equivariance. Each log probability consequently
differs by at most $2\varepsilon$. This proves the target-loss bound.
Averaging the log-probability ratio against the first probability
vector proves the linear relative-entropy bound. For the quadratic
bound, put $v=z'-z$ and $H(z)=\log\sum_i\exp z_i$. The relative
entropy from $\softmax z$ to $\softmax z'$ is
$H(z')-H(z)-DH(z)v$. Along their line segment,
\[
 v^{\mathsf T}D^2H\,v=\Var_q(v_i)
 \le\frac{(\max_i v_i-\min_i v_i)^2}{4}\le\varepsilon^2.
\]
To see the variance bound, let the minimum and maximum be $a,b$.
The inequality $(v_i-a)(b-v_i)\ge0$, averaged against $q$, gives
$\Var_q(v_i)\le(\E_qv_i-a)(b-\E_qv_i)\le(b-a)^2/4$.
The integral Taylor remainder has weight $1-t$ on $0\le t\le1$;
its integral is $1/2$. This proves the quadratic bound and therefore
the stated minimum.
\end{proof}

The constants in this proposition concern the actual domain of the
comparison. They are not uniform bounds obtained by extrapolating a
finite intervention panel. A small operator norm difference can be
amplified by downstream transport; conversely a substantial matrix
change can be weakly visible to predictions. The full vocabulary
comparison is therefore retained alongside the metric comparison.

\subsection{The training objective and the inference law}
For a fixed sequence law, define the parallel and proper prefix risks
by averaging respectively
$-\log p_\theta(x_{t+1}\mid x_{1:n};t)$ and
$-\log p_\theta(x_{t+1}\mid x_{1:t})$ over the same positions and
targets. Proposition~\ref{model:prop:prefix-defect-bound} bounds their
absolute difference, when its hypotheses hold pointwise, by
\begin{equation}
 |R_{\rm parallel}-R_{\rm prefix}|\le2\E\varepsilon_t.
 \label{model:eq:parallel-prefix-risk}
\end{equation}
For a finite weighted law this is the triangle inequality followed
by summing the individual target-loss bounds with nonnegative
weights. Without a prefix bound, low parallel risk need not
bound proper prefix risk. A predictor given its target as additional
input can concentrate its probability on that target regardless of
the conditional uncertainty given the prefix. This observation
identifies an objective mismatch that must be measured; it does not
assert that a particular trained model exploits it.

The risk gap is also distinct from the information in the suffix.
Let $X$ be the available prefix, $Z$ the additional suffix and $Y$
the target under a specified joint law $P$. Write $E_{\rm prefix}$
for the expected conditional relative entropy from $P(Y\mid X)$
to the model's proper prediction, and $E_{\rm parallel}$ for that
from $P(Y\mid X,Z)$ to its parallel prediction. Whenever the
risks are finite,
\begin{equation}
 R_{\rm prefix}-R_{\rm parallel}
 =I_P(Y;Z\mid X)+E_{\rm prefix}-E_{\rm parallel}.
 \label{model:eq:prefix-risk-information}
\end{equation}
Indeed, each conditional cross entropy is the target conditional
entropy plus its corresponding relative entropy; subtracting gives
the identity. Thus the observed loss gap is not an estimator of
suffix mutual information without the two model-error terms. If
the model's predictions are exactly prefix consistent, its risk
gap vanishes even when the corpus suffix contains information
about the target. The measured predictive KL and risk observations
concern the learned model, not an unobserved oracle conditional law.

The external-target training law used in the constant-rate family
places its only target after all input tokens. It therefore does not
give that target to the forward through a later input position. Its
frozen-state external-target risk measures a different objective
from parallel all-position loss. A comparison of the reference
recipes must retain both observations explicitly.

An operator intervention can change both the parallel computation and
its prefix defect. For any two predictive computations $f,g$ on the
same weighted sequence/target law, set
$D(f)=R_{\rm prefix}(f)-R_{\rm parallel}(f)$. Direct subtraction gives
\begin{align}
 R_{\rm prefix}(g)-R_{\rm prefix}(f)
 &=R_{\rm parallel}(g)-R_{\rm parallel}(f)
   +D(g)-D(f).
 \label{model:eq:causal-intervention-risk}
\end{align}
Thus an intervention that removes the prefix defect improves proper
risk only when its change in the parallel predictive fit is small
enough. This is an exact risk decomposition, not a monotonicity
claim for row projection or fixed-operator replacement. Since $D$
is signed and averages target positions, a small $D$ alone can
also conceal cancellation; pointwise score bounds and prefix KL
remain separate measurements.

\begin{proposition}[Proper risk and multi-step predictive fidelity]
\label{model:prop:proper-risk-chain}
Let $P$ be a law of a finite token sequence conditional on an initial
prompt, and let $Q_\theta$ be the autoregressive law formed from the
model's proper prefix probabilities. Write $\Delta_t$ for the
expected excess conditional log loss at step $t$, relative to the
true conditional law of $P$. Then
\begin{equation}
 \KL(P\Vert Q_\theta)=\sum_t\Delta_t.
 \label{model:eq:proper-risk-chain}
\end{equation}
For any event $E$ determined by that continuation, including a
specified correctness or coherence event,
\begin{equation}
 |P(E)-Q_\theta(E)|\le
 \sqrt{\tfrac12\sum_t\Delta_t}.
 \label{model:eq:task-event-chain}
\end{equation}
The same statements hold after averaging a common prompt law.
\end{proposition}
\begin{proof}
Factor both sequence probabilities into their conditional prefix
probabilities. The logarithm of their ratio is the sum of the
conditional log ratios. Taking expectation under $P$ gives the
chain rule in Equation~\eqref{model:eq:proper-risk-chain}; each term is
the excess conditional log loss. The event difference is bounded
by total variation, and Pinsker's inequality gives
Equation~\eqref{model:eq:task-event-chain}. Conditioning first on the
prompt proves the corresponding joint-law statement.
\end{proof}

This result requires the proper conditional losses on the relevant
sequence law. A low marginal token loss or one measured final-token
loss does not bound every term. Nor does stability alone imply that
the target law assigns high probability to correct reasoning. For
task transfer, if a task's prefix distribution is absolutely
continuous with respect to the pretraining prefix distribution with
density at most $w$, and the conditional target laws agree, its
expected conditional relative entropy is at most $w$ times the
pretraining value. This follows by integrating the nonnegative
conditional relative entropy against that density. Without those
support and conditional-law assumptions, pretraining risk alone
does not give this transfer bound.

The predictive law also includes the decoding rule. The probabilities
in Proposition~\ref{model:prop:proper-risk-chain} are the declared proper
conditional probabilities; replacing them by a temperature transform
or nucleus truncation changes $Q_\theta$. In particular, truncation
can assign zero probability to tokens with positive target probability,
so a finite relative-entropy bound for the untruncated model does not
automatically transfer to that decoder.

There is a useful conditional stability bound when the retained
nucleus is fixed. If probability vectors $p,p'$ retain the same set
$S$ with masses $a=p(S),b=p'(S)\ge\tau>0$, their normalized
restrictions satisfy
\begin{equation}
 \left\lVert\frac{p\mathbf1_S}{a}
       -\frac{p'\mathbf1_S}{b}\right\rVert_1
 \le \frac{2}{\tau}\norm{(p-p')\mathbf1_S}_1.
 \label{model:eq:fixed-nucleus-stability}
\end{equation}
To prove it, insert $p'\mathbf1_S/a$, apply the triangle inequality,
and use $|a-b|\le\norm{(p-p')\mathbf1_S}_1$. The first difference
is at most that norm divided by $a$, and the normalization difference
is $|a-b|/a$. This yields the bound. Stability of $S$ itself requires
control of the ordering and cumulative-mass threshold margins.
Thus the nucleus boundary is an additional observation coordinate
when comparing generated continuations; it is not controlled solely
by a small deductive-output variation.

\subsection{Task margins and the retained critical coordinates}
\begin{proposition}[Task margin transport]
\label{model:prop:task-margin}
For a finite answer set with at least two candidates, let $s_a(x)$
be the proper autoregressive score of answer $a$, and let $a_*(x)$
be the specified correct answer. Define
\[
 m(x)=s_{a_*}(x)-\max_{a\ne a_*}s_a(x).
\]
If a reduced computation changes every candidate score by at most
$e(x)$, then its signed margin differs from $m(x)$ by at most
$2e(x)$. Every example with $m(x)>2e(x)$ consequently remains
correct under the reduction.
\end{proposition}
\begin{proof}
Both the correct score and the maximum competing score change by
at most $e(x)$: for the maximum, apply the pointwise bounds to every
candidate and interchange the two computations. Subtracting the two
scores and applying the triangle inequality bounds the margin change
by $2e(x)$. A margin greater than that bound remains positive.
\end{proof}

A stable incorrect margin remains incorrect as well. Operator
interventions test preservation of learned task margins; they cannot
supply missing task knowledge.

\begin{proposition}[Premise response and answer preference]
\label{model:prop:premise-routing-margin}
Consider two questions with the same two answer candidates and
opposite correct answers. Index them by $v=0,1$, with candidate $v$
correct on question $v$. Let $u_v=s_1(x_v)-s_0(x_v)$ be the
candidate-score difference, and define
\[
 b=\frac{u_0+u_1}{2},\qquad \Delta=u_1-u_0.
\]
The smaller of their two signed correct-answer margins is
\begin{equation}
 \min(m_0,m_1)=\frac{\Delta}{2}-|b|.
 \label{model:eq:premise-routing-margin}
\end{equation}
Both answers are therefore strictly correct exactly when
$\Delta>2|b|$. If each candidate score on question $v$ changes by
at most $e_v$, both remain strictly correct whenever
$\Delta/2-|b|>2\max(e_0,e_1)$.
\end{proposition}
\begin{proof}
The margins are $m_0=-u_0=\Delta/2-b$ and
$m_1=u_1=\Delta/2+b$. Their minimum is the displayed expression,
which is positive exactly under the stated inequality. Each margin
changes by at most $2e_v$ by Proposition~\ref{model:prop:task-margin}.
Subtracting the larger of these two bounds proves the retention
condition.
\end{proof}

The variable $b$ measures the pair's average answer preference;
$\Delta$ measures response in the direction required by the changed
premises. A small positive response need not overcome that preference.
A predictor choosing the same candidate on both questions scores
one of two individually while never answering both correctly.
Demonstrations can change both variables, so retaining every
specified demonstration count tests elicitation as well as stability.
For a fixed learned metric, premise information can still propagate
through the queries, keys, values and subsequent decoder layers.
Correct composition requires that this contextual computation
produce the appropriate margins. Contraction of the metric generator
alone imposes no sign or lower bound on $\Delta$.

The native vocabulary projection is untied from the token embedding.
For fixed input tokens, replacing its matrix and bias
$(W_{\rm out},b_{\rm out})$ by
$(\Pi W_{\rm out},\Pi b_{\rm out})$ for a vocabulary permutation
$\Pi$ leaves all upstream deductive tensors unchanged and gives
\[
 \widetilde z=\Pi z,\qquad
 \softmax(\widetilde z)=\Pi\softmax(z).
\]
The first identity follows from the final affine projection; the
second follows by permuting the numerator and sum in softmax.
Thus the same deductive tensors can accompany different token-label
predictions. This is an architectural dependence statement, not an
invariance of training on a fixed labeled data distribution. The
conditioned training law must also select the predictive readout and
contextual computation. A reduction eliminating them must retain
their conditional law rather than infer task correctness from the
metric alone.

For candidate sets of different lengths, use a common finite set of
candidate, token-position and decoder slots. Fill unused slots with
their fixed reference operators, so those slots contribute zero
operator error. This convention places every collection and its
reference in the same finite-dimensional normed space.

\begin{proposition}[Operator variance and retention of task margins]
\label{model:prop:operator-task-transfer}
Fix a checkpoint and a distribution of inference tasks. Let
$\mathbf G(X)$ be the finite collection of native operators needed
for their declared scores, and let $\widehat{\mathbf G}$ be the
corresponding fixed reference collection. Suppose a common bound on
all candidate score errors is
\[
 e(X)\le\Lambda\norm{\mathbf G(X)-\widehat{\mathbf G}},\qquad
 V_*:=\E\norm{\mathbf G(X)-\widehat{\mathbf G}}^2<\infty,
\]
with $\Lambda\ge0$. Let $m$ be the native signed correct-answer
margin. If for some $C,\alpha>0$,
$\Pr(0<m\le u)\le Cu^\alpha$ for every $u>0$, then the probability
that a strictly correct native answer loses its positive margin
under freezing is at most
\begin{equation}
 Cu^\alpha+\frac{4\Lambda^2 V_*}{u^2},\qquad u>0.
 \label{model:eq:operator-margin-transfer}
\end{equation}
In particular this loss probability is
$O(V_*^{\alpha/(\alpha+2)})$ as $V_*\downarrow0$ with fixed
$C,\alpha,\Lambda$. For squared Euclidean or Hilbert norms,
\begin{equation}
 V_*=\E\norm{\mathbf G-\E\mathbf G}^2
       +\norm{\E\mathbf G-\widehat{\mathbf G}}^2.
 \label{model:eq:operator-calibration-bias}
\end{equation}
\end{proposition}
\begin{proof}
By Proposition~\ref{model:prop:task-margin}, an originally positive margin
can cease to be positive only if $0<m\le2e$. For any $u>0$, this
event is contained in the union of $\{0<m\le u\}$ and
$\{e\ge u/2\}$. The margin hypothesis bounds the first probability.
Markov's inequality and $\E e^2\le\Lambda^2V_*$ bound the second
by $4\Lambda^2V_*/u^2$. If $V_*>0$, take
$u=V_*^{1/(\alpha+2)}$ to obtain the asserted order bound. If
$V_*=0$, the score error is zero almost surely and the loss
probability is zero. Finally expand the square after adding and
subtracting $\E\mathbf G$; the mixed inner product has expectation
zero. This gives Equation~\eqref{model:eq:operator-calibration-bias}.
\end{proof}

This proposition separates conditional operator variation from the
error of a finite calibration reference. A fixed reference preserves
useful computation when both are small in predictively visible
coordinates and the native answers have adequate margins. For a
single next-token prediction, the layer transport in
Proposition~\ref{model:prop:prefix-defect-bound} supplies the corresponding
logit error bound. For a candidate answer, the finite collection
includes every proper prefix used in its score, and the score bound
sums the constituent log-probability errors. The task law and all
constants must refer to that same collection.

At a fixed checkpoint, zero conditional operator variance means that
the operator is constant almost surely under the declared input law.
It can still differ between independently trained checkpoints. Thus
vanishing input variation and persistent training-seed susceptibility
can coexist. The common metric remains part of the retained training
state, even when its inference realization can be frozen. This is a
precise form of a training-selected fixed-operator regime.

\begin{proposition}[Fluctuation transfer through the inference emission]
\label{model:prop:inference-fluctuation-transfer}
For a specified joint data, drive, time and model-size family, let
$Y_N\in\R^d$ be centered training collective fluctuations and let
$a_N>0$ satisfy
\[
 a_N^{-2}\Cov(Y_N)\longrightarrow\Sigma_*.
\]
Suppose the retained inference observation $Z_N\in\R^k$ has
the representation $Z_N=b_N+A_NY_N+e_N$, where $b_N$ is
deterministic, the deterministic matrices $A_N$ converge in
operator norm to $A$, and $\E\|e_N\|^2=o(a_N^2)$. Then
\begin{equation}
 a_N^{-2}\Cov(Z_N)\longrightarrow A\Sigma_*A^{\mathsf T}.
 \label{model:eq:inference-fluctuation-transfer}
\end{equation}
If also $a_N^{-1}Y_N$ converges in distribution to $Y_*$, then
$a_N^{-1}(Z_N-\E Z_N)$ converges in distribution to $AY_*$.
In particular, if $a_N^2=N^{\kappa-1}$ and
$A\Sigma_*A^{\mathsf T}\ne\bm0_{k\times k}$, the nonzero retained sector of
$N\Cov(Z_N)$ has the same count exponent $\kappa$ as
$N\Cov(Y_N)$. If the limiting matrix vanishes, the leading
fluctuation sector is lost and the proposition supplies no
nonzero next-order exponent.
\end{proposition}
\begin{proof}
Finite dimensionality and the covariance limit imply
$\E\|Y_N\|^2=O(a_N^2)$. Set $\widetilde e_N=e_N-\E e_N$.
Lemma~\ref{model:lem:master-perturbation} bounds the covariance discrepancy by
\[
 2\|A_N\|_{\rm op}\sqrt{\tr\Cov(Y_N)}
       \|\widetilde e_N\|_{L^2}+\|\widetilde e_N\|_{L^2}^2=o(a_N^2).
\]
Here $A_N\to A$ ensures bounded operator norms, and centering contracts
$L^2$. Divide by $a_N^2$ and pass to the covariance limit to prove
Equation~\eqref{model:eq:inference-fluctuation-transfer}.
Moreover $a_N^{-1}\widetilde e_N\to\bm0$ in probability by its
second-moment bound. Convergence in distribution of
$a_N^{-1}Y_N$, together with $A_N\to A$, then gives the stated
limiting emission law. Multiplication of the covariance limit by
$N a_N^2=N^\kappa$ proves the exponent statement.
\end{proof}

The matrices $A_N$ specify the full retained inference emission,
including downstream transport and the logit gauge quotient when
logits are used. For a nonlinear network the small-error hypothesis
must control its nonlinear remainder, omitted coordinates and
observation arithmetic in these same units. A finite-checkpoint
Jacobian alone does not establish that hypothesis. Identifying $\Sigma_*$ with the covariance of $Y_*$ additionally
requires control of second moments in the distributional limit,
for example uniform integrability of $\|Y_N/a_N\|^2$. Truncating
the squared coordinates and then removing the truncation gives
that identification. A limiting law may also lose information
under the projection $A$ even when one covariance exponent survives. If the observation itself carries a
factor $N^{-\eta}$, and its remainder is controlled at that smaller
scale, direct multiplication changes the displayed susceptibility
power to $\kappa-2\eta$. Fixed observation units are consequently
part of an exponent-transfer claim. None of these conclusions
identifies a native critical exponent without an independently
established training critical family.

If an independently established critical family has
$V_*=O(|r|^p)$ and uniform transport and margin constants, the bound
in Equation~\eqref{model:eq:operator-margin-transfer} gives the inherited
upper rate $O(|r|^{p\alpha/(\alpha+2)})$. It is a bound on loss of
already correct answers, not an equality defining a critical
exponent of reasoning. The small-margin law and predictive
transport can change under scale flow; they therefore belong to
the inference part of a comprehensive theory.

The resulting training-to-inference description retains a joint
field containing row contrast, common metric variation, prefix
defect, proper predictive risk and task margins, with the optimizer
state or its justified conditional memory. Row concentration can
suppress one source of global metric variability. Stabilization of
the common metric can then make the parallel objective consistent
with autoregressive prediction, while the learned query, key, value
and feed-forward paths retain useful context dependence. This supplies the analytical connection tested by the fixed-checkpoint
interventions in Section~\ref{model:sec:inference-results}.
It does not identify the stable metric with a critical mode.

A claim of near-critical reasoning additionally requires a critical
surface in the conditioned training family, a compatible scale law
for its connected fluctuations, and predictive visibility of the
relevant retained modes. A claim of self-organization further
requires endogenous approach to that surface under the declared
drive. The inference task margins and proper-risk observations test
what predictive properties persist after training. The joint
theory therefore allows row contraction, persistent common modes
and task-sensitive contextual computation to occupy different
directions of the same model-wide law.

\section{Source selection, coupled sectors and fine-tuning}
\label{model:sec:finetuning-theory}

Fine-tuning changes the conditional training kernel and can change which
collective directions an inference observation resolves. This gives a precise
interpretation of source selection as a probe of a trained PLDR-LLM. It also
distinguishes a response that is already present at the incoming state from
one created by the subsequent adaptation. Throughout this section the native
architecture, target rule and arithmetic implementation are fixed. A width
family has the conditioning and clock conventions of
Section~\ref{model:sec:theory-synthesis}.
An exponent established for a fine-tuned family belongs to its combined
pretraining and adaptation law. Fine-tuning may reveal a previously weak
projection, generate a new slow sector, or move the state away from a
critical surface. Identifying an inherited exponent requires the additional
coupling and visibility conditions proved below; an adapted exponent alone
does not establish the same property of the incoming pretrained family.

\subsection{Finite-corpus epochs and parameterized adaptation}

Single-pass pretraining and repeated adaptation have different resource laws.
For a fixed finite adaptation corpus $D=(z_1,\ldots,z_M)$, retain $D$, the
epoch index, the current permutation and its cursor in the complete state.
At an epoch boundary the cursor resets and a new permutation is drawn from
the declared refill kernel. Configuration symmetries and supervised target
sites have their own recorded draws. The next update is measurable from
this augmented state and those draws. Conditioning on it therefore yields
a training kernel, and chronological composition remains exact across
refills. A depleted resource is not silently replaced by an independent
population draw. Repeated visits change optimization exposure without
increasing the number of independent source configurations. In particular,
the nonexhaustion argument for a strictly consumed single-pass resource
applies between refills; a monotone resource counter is no longer a global
Lyapunov variable for the repeated process.

A parameterization used for small-corpus adaptation is also part of this
state. A low-rank update~\cite{hu2021lora} represents an existing linear
matrix by $W+BA$.
It leaves the native inference architecture unchanged after merging, but
its factor dynamics are a different training kernel from an unconstrained
update of $W$.

\begin{proposition}[Inference equivalence does not close factor dynamics]
\label{model:prop:adaptation-factor-state}
For compatible real matrices $W,A,B$ and vector $x$,
$(W+BA)x=Wx+B(Ax)$. Replacing a factorized layer by its merged matrix in
a finite deterministic network therefore preserves its exact-real emission.
For a scalar parameter $w=w_0+ab$ and differentiable loss $\ell(w)$,
put $g=\ell'(w)$. A simultaneous gradient step of rate $\eta$ on $(a,b)$
induces
\begin{equation}
 w^+-w=-\eta g(a^2+b^2)+\eta^2g^2ab.
 \label{model:eq:adaptation-factor-update}
\end{equation}
Consequently, the merged weight need not determine its successor, even
when the incoming emission and loss gradient are identical.
\end{proposition}
\begin{proof}
Distributivity and associativity prove the matrix identity. Substituting
that identity successively at each affected layer preserves every later
node of a deterministic computation. The chain rule gives factor gradients
$bg$ and $ag$. Expanding $(a-\eta bg)(b-\eta ag)-ab$ proves
\eqref{model:eq:adaptation-factor-update}. For $c\ne0$, the factors $(ca,b/c)$
have the same product $ab$, but their squared-norm sum is
$c^2a^2+b^2/c^2$. For example, $(1,1)$ and $(2,1/2)$ have product one
and squared-norm sums two and $17/4$. With $g=1$ and $\eta\ne0$, their
successor merged weights differ by $-9\eta/4$. Thus an emission-preserving
quotient can discard information required by the next training step.
\end{proof}

The executed low-rank parameterization retains the factors and their AdamW
moments during training and exports merged native weights for inference.
Floating-point agreement of the two evaluation routes is separately
measured. The proposition explains why that agreement cannot identify
their training transition laws with an unconstrained native optimizer.
Sections~\ref{model:sec:released-adaptation} and \ref{model:sec:singlepass-released} give the repeated-corpus and single-pass adaptation protocols and merged-weight checks.

\subsection{The paired operator order parameter}

The normalized operator discrepancy in the PLDR-LLM criticality
study~\cite{gokden2026soc} compares metric tensors under paired inference
executions. Its numerator measures changes of the tensor entries, rather
than distances between rows of one tensor. Let $U,V\in\R^D$ denote the
flattened tensors under a specified coupling $\Pi$ of the two executions,
at a fixed trained state. Write
\begin{equation}
 \bar\mu_V=D^{-1}{\bf1}^{\mathsf T}\E_\Pi V,\qquad
 \mathcal O_\Pi^2=
 \frac{\E_\Pi\norm{U-V}^2}{D|\bar\mu_V|^2},
 \qquad \bar\mu_V\ne0.
 \label{model:eq:finetuning-order}
\end{equation}
The finite-cohort version replaces expectations by the stated empirical
average. The order parameter is undefined when its reference mean is zero.
We also retain its unnormalized numerator and a separate normalization by
$\E\norm V^2$. The latter is a different statistic.

\begin{proposition}[A paired discrepancy retains the input coupling]
\label{model:prop:paired-order-covariance}
For square-integrable $U,V$ under $\Pi$,
\begin{align}
 \E_\Pi\norm{U-V}^2
 &=\tr\Cov_\Pi(U)+\tr\Cov_\Pi(V)
   -2\tr\Cov_\Pi(U,V) \notag\\
 &\quad+\norm{\E_\Pi U-\E_\Pi V}^2.
 \label{model:eq:paired-order-covariance}
\end{align}
Independent identically distributed copies have numerator
$2\tr\Cov(U)$. Identically coupled copies have numerator zero for any
marginal covariance. With a finite nonzero denominator,
$\mathcal O_\Pi=0$ is equivalent to $U=V$ almost surely under that coupling.
\end{proposition}
\begin{proof}
Subtract the respective means from $U$ and $V$, expand the squared norm,
and take expectations. The cross terms with the deterministic mean
difference vanish by centering. The remaining cross inner product is the
trace of the paired cross covariance. Independence removes that covariance
for the first special case; equality of the coupled variables gives the
second. A nonnegative random variable has expectation zero precisely when
it is zero almost surely, which proves the final assertion.
\end{proof}

A nondegenerate input coupling also permits vanishing order under
ordinary stable relaxation. Let $X,Y$ be independent centered scalar
variables of variance one, $a\ne0$, and $0<q<1$. Set
$U_t=a+q^tX$ and $V_t=a+q^tY$. The proposition gives
$\mathcal O_t^2=2q^{2t}/a^2\to0$, while the fixed relaxation multiplier
$q$ stays separated from one. Its relaxation time $-1/\log q$ is finite.
Thus an order parameter approaching zero does not alone distinguish
regular contraction from a singular critical approach.

Consequently the coupling is part of the observable. A stochastic
generation pair, a cached operator comparison and a short/long prefix pair
need not have the same value at the same checkpoint. The executed
fine-tuning study uses prefixes of lengths 32 and 64 from the same untouched
document, and records both the residual metric $A$ and the energy-curvature tensor
$G_{\mathrm{LM}}$.
It is a test of the same discrepancy construction under an explicitly
different input coupling from the stochastic generation study.
Section~\ref{model:sec:finetuning-observation-results} specifies this paired-prefix observation law and its source comparisons.

The decomposition $A={\bf1}\mu^{\mathsf T}+Z$, with
${\bf1}^{\mathsf T}Z=\bm0^{\mathsf T}$, explains its relationship to row collapse.
Vanishing transverse energy $\norm Z^2$ leaves an input-dependent common
field $\mu$. Conversely, agreement of two inputs under $\Pi$ does not imply
that $Z$ vanishes within either tensor. Agreement for every observed
layer and head is a model-wide statement about those paired operators;
it does not constrain every body variable, optimizer mode, correlation
length or response gap. An order parameter approaching zero therefore
specifies a candidate limiting regime, without by itself identifying a
critical point. Even absolute paired agreement requires control of the
normalization when taking a size limit. This is consistent with retaining
common fields and predictive visibility in the model-wide theory.

\subsection{An exact finite source law}

Fix the incoming complete state $S_0$, including the remaining corpus and
optimizer memory. Partition an eligible portion of that resource into two
disjoint reservoirs, each containing at least $BT$ blocks. Draw a uniform
permutation of each reservoir independently of $\rho$. For each of the
$BT$ block positions in $T$ updates of batch size $B$, select the technical
reservoir with probability $\rho$, and otherwise select the narrative
reservoir. In each selected reservoir use its next unused block. This is
a single-pass law. The two permutations, their cursors and the selected
block history are part of the complete state.

\begin{proposition}[Finite fine-tuning is a source-mixture polynomial]
\label{model:prop:finite-source-polynomial}
Suppose the native update and observation, conditional on the selected
blocks and all other declared random variables, do not additionally depend
on $\rho$. Let $f$ be an integrable scalar observable of the resulting
complete path. For $m=BT$ and $c\in\{0,1\}^m$, let $f_c$ be its conditional
expectation given selection word $c$, averaging the other declared
randomness. Assume each $f_c$ is finite. Then
\begin{equation}
 F_T(\rho)=\E_\rho f
   =\sum_{c\in\{0,1\}^m}
       \rho^{|c|}(1-\rho)^{m-|c|}f_c .
 \label{model:eq:finite-source-polynomial}
\end{equation}
In particular, $F_T$ is a polynomial of degree at most $m$. For
$0<\rho<1$ its source derivative is
\begin{equation}
 F_T'(\rho)=
 \E_\rho\left[f\,\frac{|c|-m\rho}{\rho(1-\rho)}\right].
 \label{model:eq:source-score}
\end{equation}
\end{proposition}
\begin{proof}
Every binary word is feasible because either reservoir can supply all
$m$ blocks. Independence of the selection indicators assigns word $c$
the probability in~\eqref{model:eq:finite-source-polynomial}. Conditional on
that word, its path law and $f_c$ do not depend on $\rho$, including the
without-replacement dependence within each reservoir. The finite law of
total expectation proves the first identity. Differentiate its finitely
many polynomial terms. The logarithmic derivative of the word weight in
the open interval is $|c|/\rho-(m-|c|)/(1-\rho)$, giving the second identity.
\end{proof}

This exact construction is compatible with chronological kernel
composition and with subsequent observation pushforwards. It need not be
an economical representation: the displayed sum contains $2^{BT}$ words.
It also does not assert that a single path, coupled across $\rho$ by fixed
uniform draws, is a polynomial. That individual path changes when a draw
crosses a threshold. Finite source laws can have sharp crossovers while
remaining smooth after this averaging. Singular critical behavior requires
a specified joint limit in width, time and source resource. Source
depletion and the inherited optimizer clock remain explicit in that limit.

For a finite observable $y_\rho$, define the three-knot approximation
\begin{equation}
 \widehat y_\rho=(1-\rho)y_0+\rho y_1+
 4\rho(1-\rho)\left(y_{1/2}-\frac{y_0+y_1}{2}\right).
 \label{model:eq:source-quadratic}
\end{equation}
Substitution gives exact agreement at $0,1/2,1$. This is the unique
polynomial of degree at most two with those values: the difference of two
such polynomials has three distinct roots and hence is zero. Predictions
at other mixture values test an approximation to the conditional response.
The degree bound in Proposition~\ref{model:prop:finite-source-polynomial} does
not supply a small quadratic approximation error. That error is measured
at untouched mixture values under the stated document and anchor
acquisition law.
Section~\ref{model:sec:finetuning-observation-results} gives the acquisition split and complete unused-mixture predictions.

\subsection{Observable sectors and coupled critical surfaces}

Let $z_N,z_T,z_G$ be proper-prefix vocabulary logits following narrative,
technical and general-source continuation from a common incoming state.
With incoming predictive distribution $p_0$ and a fixed input law $\nu$,
define $D_N=z_N-z_G$, $D_T=z_T-z_G$ and
\begin{equation}
 \langle u,v\rangle_{p_0,\nu}=
 \E_{x\sim\nu}\left[
    \sum_j p_{0j}(x)u_j(x)v_j(x)
    -\Big(\sum_jp_{0j}(x)u_j(x)\Big)
     \Big(\sum_jp_{0j}(x)v_j(x)\Big)\right].
 \label{model:eq:finetuning-fisher}
\end{equation}
This positive semidefinite inner product on logits modulo their additive
gauge is fixed before the continuation. We use it as a fixed norm for
finite contrasts; identifying its square with a KL divergence requires
a separately controlled small-displacement expansion. The $2\times2$ Gram matrix of
$D_N,D_T$ measures whether two source effects are distinguishable in
prediction. A positive second eigenvalue identifies two noncollinear
finite effects. It is not the second eigenvalue of a native dynamical
operator. Cross-time canonical angles in this same inner product measure
the persistence of their span, with fixed contexts and weights.

Multiple critical sectors are mathematically possible in a coupled model.
They need not coincide with named architecture components. For example,
consider a symmetric local two-sector relaxation matrix
\begin{equation}
 \Gamma=\begin{pmatrix}a&g\\g&b\end{pmatrix},\qquad
 \delta_\pm=\frac{a+b\pm\sqrt{(a-b)^2+4g^2}}2 .
 \label{model:eq:coupled-finetuning-gaps}
\end{equation}
Its eigenvectors mix the two coordinates when $g\ne0$. The determinant
condition $ab-g^2=0$ can close one gap while the other remains positive.
If $\Gamma$ is positive semidefinite and both gaps vanish, then
$a=b=g=0$. Indeed their sum gives $a+b=0$; nonnegative diagonal entries
then give $a=b=0$, and $g^2\le ab$ gives $g=0$. Thus two vanishing gaps
impose stronger conditions than one. A single source-mixture coordinate
need not intersect such a locus unless other controls, symmetry or an
established selection mechanism enforce the additional conditions.

The native optimizer need not have this symmetric form. For example,
a recursion matrix $rR(\omega)$, with $R(\omega)$ a planar rotation, has
eigenvalues $r e^{\pm i\omega}$. Their moduli reach one together at
$r=1$. Symmetry and linked oscillatory coordinates can therefore enforce
simultaneous marginality along one control. Nonnormal transient response
also requires more information than eigenvalue moduli. These distinctions
matter when optimizer memory and the shared generator couple the native
collective coordinates.

For a stable frozen linear recursion $q_{t+1}=Mq_t+\eta_t$ with independent
centered forcing of covariance $Q$ and spectral radius below one, its
stationary covariance is
\begin{equation}
 C=\sum_{j\ge0}M^jQ(M^{\mathsf T})^j.
 \label{model:eq:finetuning-lyapunov}
\end{equation}
To verify this, iterate the recursion, use independence to remove cross
terms, and let the contribution of the initial state tend to zero. The
matrix-power series converges in finite dimension because the spectral
radius is less than one. Shifting its index gives $C=MCM^{\mathsf T}+Q$.
Nonstationary, colored or state-correlated forcing requires the conditional
and memory constructions developed above. PLDR's shared metric generator
couples all heads and layers; freezing that generator changes the recursion
and need not reveal an independent sector of the full native dynamics.
Source-response rank, a local symmetric example, and a fitted covariance
slope therefore have distinct scientific meanings.

\subsection{How mixed critical coordinates can obscure a finite fit}

If a justified limiting RG has relevant fields $u_1,\ldots,u_k$, a
corresponding conditional count-scaling hypothesis takes the form
\begin{equation}
 \chi_N=N^\kappa\mathcal F
 \left(u_1N^{\phi_1},\ldots,u_kN^{\phi_k},
       \tau/N^z,q,\mathcal C\right).
 \label{model:eq:finetuning-multifield-scaling}
\end{equation}
Here $q$ is a specified consumed-resource fraction and $\mathcal C$
collects the remaining conditioning. The fields and their scale
multipliers must come from that limiting theory. They are not identified
by the number of source contrasts measured in a finite experiment.
Fine-tuning traces a curve in these fields and can intersect different
critical surfaces at different values of its controls. Tuning one field
while keeping others near their required values can isolate a scaling
regime; an arbitrary data-mixture curve need not do so. With head count
as the declared size, count exponents do not separately identify a
spatial dimension or conventional correlation-length exponent.

Even two positive contributions illustrate the finite-window issue.
Suppose, as a conditional example, that
$\chi(r)=c_1|r|^{-\gamma_1}+c_2|r|^{-\gamma_2}$ with $c_i>0$ and
$r\ne0$. Direct differentiation gives
\begin{equation}
 -\frac{d\log\chi}{d\log|r|}=
 \frac{\gamma_1c_1|r|^{-\gamma_1}
       +\gamma_2c_2|r|^{-\gamma_2}}
      {c_1|r|^{-\gamma_1}+c_2|r|^{-\gamma_2}}.
 \label{model:eq:mixed-effective-exponent}
\end{equation}
The apparent exponent is a scale-dependent weighted mean and tends to
the larger exponent as $r\to0$ when the two powers differ. Observation
projections, vanishing amplitudes and signed cross covariances can further
change what a scalar measurement resolves. This supplies a mechanism by
which critical sectors, if present, could be difficult to identify in a
short range. Establishing them requires their own source, size, time and
observation tests; a drifting finite slope alone does not distinguish
this mechanism from regular finite response.

\subsection{A condition for inherited fluctuation exponents}

\begin{theorem}[Simultaneous visibility of several fluctuation sectors]
\label{model:thm:finetuning-mode-inheritance}
Let centered $X_N\in L^2(\Omega;\R^k)$ have positive covariance eigenvalues
$\lambda_{1,N}\ge\cdots\ge\lambda_{k,N}>0$. Suppose a centered observation
of the fine-tuned state is coupled to the incoming sectors by
$Y_N=A_NX_N+e_N\in L^2(\Omega;\R^d)$, where $d\ge k$,
$A_N$ is deterministic under the conditioning, and
\[
 a\norm v\le\norm{A_Nv}\le b\norm v\quad(v\in\R^k),
 \qquad 0<a\le b<\infty.
\]
Set $\epsilon_N=\norm{e_N}_{L^2}$. The ordered covariance eigenvalues of
$Y_N$ obey, for $1\le i\le k$,
\begin{equation}
 \max\{0,a\sqrt{\lambda_{i,N}}-\epsilon_N\}
 \le\sqrt{\lambda_i(\Cov Y_N)}
 \le b\sqrt{\lambda_{i,N}}+\epsilon_N .
 \label{model:eq:finetuning-mode-inheritance}
\end{equation}
If $\epsilon_N=o(\sqrt{\lambda_{k,N}})$, all $k$ eigenvalues retain their
orders of growth. In particular a covariance eigenvalue with a limiting
logarithmic power exponent retains that exponent. The same statement
applies to a declared susceptibility prefactor multiplying both spectra.
\end{theorem}
\begin{proof}
For a centered random vector $W$, define the operator
$T_W:v\mapsto v^{\mathsf T}W$ from its finite-dimensional coefficient space
to $L^2(\Omega)$. Its squared singular values are the eigenvalues of
$\Cov W$, since $T_W^*T_W=\Cov W$. Cauchy--Schwarz gives
$\norm{T_{e_N}}_{\rm op}\le\epsilon_N$. The variational characterization
of singular values and the triangle inequality therefore bound the change
of each singular value of $T_{A_NX_N}$ by $\epsilon_N$ when $T_{e_N}$ is
added. For completeness, on any subspace the norm of either operator on a
unit vector differs from that of their sum by at most $\epsilon_N$;
taking the min--max over subspaces yields this bound in both directions.

Write $C_N=\Cov X_N$. The nonzero eigenvalues of
$A_NC_NA_N^{\mathsf T}$ equal those of
$C_N^{1/2}A_N^{\mathsf T}A_NC_N^{1/2}$.
The assumed singular-value bounds put the latter matrix between
$a^2C_N$ and $b^2C_N$ in positive semidefinite order. The eigenvalue
min--max principle gives bounds $a^2\lambda_{i,N}$ and
$b^2\lambda_{i,N}$. Combining these bounds with the singular-value
perturbation bound proves~\eqref{model:eq:finetuning-mode-inheritance}.
The stated small-error hypothesis is sufficient for every $i$ because
$\lambda_{i,N}\ge\lambda_{k,N}$. The resulting upper and lower constant
multipliers preserve logarithmic powers, although they do not identify
leading amplitudes or scaling functions.
\end{proof}

\begin{corollary}[A finite source-span certificate]
\label{model:cor:finite-source-span}
Let $D_0,D_t:\R^2\to\mathcal H$ have the two incoming and current source
contrasts as columns in the Hilbert space defined by
\eqref{model:eq:finetuning-fisher}. For an arbitrary fitted matrix $M$, write
$D_t=D_0M+E$. Set $G_0=D_0^*D_0$, $G_t=D_t^*D_t$ and
$\epsilon=\norm E_{\mathrm{HS}}$. If the singular values of $M$ lie in
$[a,b]$, then
\[
 \max\{0,a\sqrt{\lambda_i(G_0)}-\epsilon\}
 \le\sqrt{\lambda_i(G_t)}
 \le b\sqrt{\lambda_i(G_0)}+\epsilon,\quad i=1,2.
\]
In particular $a\sqrt{\lambda_2(G_0)}>\epsilon$ certifies two resolved
current effect directions in this observation space.
\end{corollary}
\begin{proof}
Apply the singular-value variational argument to $D_0M$ and $D_t$.
Multiplication by $M$ changes each nonzero singular value by factors
between $a$ and $b$, while addition of $E$ changes each singular value
by at most $\norm E_{\rm op}\le\norm E_{\rm HS}$. Squared singular
values are the eigenvalues of the corresponding Gram matrix. These
arguments hold in the finite-rank subspaces of $\mathcal H$ and do not
require that the logit contrasts themselves are random centered sectors.
\end{proof}

For an empirical calibration law, define
$G_{0,\mathrm{cal}}=D_0^*D_0$ and
$C_{0t,\mathrm{cal}}=D_0^*D_t$ in that law. If the calibration Gram is
invertible, least squares gives
$M=G_{0,\mathrm{cal}}^{-1}C_{0t,\mathrm{cal}}$.
On a separate test law its residual Gram is exactly
\begin{equation}
 E^*E=G_t-M^{\mathsf T}C_{0t}-C_{0t}^{\mathsf T}M
           +M^{\mathsf T}G_0M.
 \label{model:eq:source-span-residual}
\end{equation}
This follows by expanding $(D_t-D_0M)^*(D_t-D_0M)$.
The two laws must remain separate in this calculation. Calibration at
successive states measures state-conditioned observation transport; it
is not autonomous prediction of future transport coefficients.

The assertion concerns ordered eigenvalues and does not identify individual
eigenvectors inside a degenerate sector. The weakest retained fluctuation sets the required error scale. Small
parameter displacement or good mean prediction alone does not control it.
A size-dependent observation matrix can suppress one mode, and strong
adaptation can replace the incoming fluctuation law altogether. Conversely,
even a scalar observation that hides a direction can coexist with a
well-conditioned vector observation that resolves it. Fine-tuning can thus
help establish exponents for a specified adapted family, but attributing
them to pre-existing critical sectors additionally requires an inheritance
bound of this kind. The finite experiments in Section~\ref{model:sec:finetuning-results}
measure source response,
its time dependence, component controls and conditional size behavior;
they do not assume this asymptotic hypothesis.

\subsection{Representation error and coefficient transport}
\label{model:sec:projection-transport}
We apply finite-dimensional orthogonal projection to the PLDR emission.
Fix the incoming complete state, a finite document panel, and incoming
categorical probabilities $p_0(j\mid x)>0$. Work in the finite Hilbert
space of logit contrasts modulo documentwise additive constants, with
\[
 \langle u,v\rangle_0=\frac1{|\mathcal X|}\sum_{x\in\mathcal X}
 \sum_j p_0(j\mid x)
 [u_j(x)-\bar u(x)][v_j(x)-\bar v(x)],\qquad
 \bar u(x)=\sum_jp_0(j\mid x)u_j(x).
\]
This fixed Fisher geometry measures finite logit contrasts; it is not
identified with an exact finite-change KL divergence. Let
$D:\R^r\to\mathcal H$ have the retained dictionary as its columns and
let $Y:\R^k\to\mathcal H$ have successor contrasts as its columns.
The dictionary, panel, and probabilities are held fixed in the following
statement. Coefficients are shared across all coordinates of that panel.

\begin{proposition}[Projection obstruction and coefficient excess]
\label{model:prop:projection-obstruction}
Let $P_D$ be the orthogonal projection onto $\operatorname{range}D$.
For every coefficient matrix $A\in\R^{r\times k}$,
\begin{equation}\label{model:eq:projection-split}
 \|Y-DA\|_{\mathrm{HS}}^2
 =\|(I-P_D)Y\|_{\mathrm{HS}}^2
  +\|P_DY-DA\|_{\mathrm{HS}}^2.
\end{equation}
Consequently, $\|(I-P_D)Y\|_{\mathrm{HS}}$ is the smallest attainable
residual within that dictionary, irrespective of the coefficient
acquisition rule. If $G=D^*D$ is positive definite and $C=D^*Y$, then
$A_*=G^{-1}C$, $P_DY=DA_*$, and, writing $H=Y^*Y$ and $R_*=Y-DA_*$,
\begin{align}
 R_*^*R_*&=H-C^*G^{-1}C,\label{model:eq:projection-gram}\\
 (Y-DA)^*(Y-DA)&=R_*^*R_*+(A-A_*)^*G(A-A_*).
 \label{model:eq:projection-matrix}
\end{align}
The general projection identity remains valid for a rank-deficient
finite dictionary; the inverse formula then requires the corresponding
orthogonal projection or Moore--Penrose inverse.
\end{proposition}
\begin{proof}
The range of a finite-dimensional linear map into a Hilbert space is
closed, so $P_D$ exists. Decompose each column of $Y-DA$ as
$(I-P_D)Y+(P_DY-DA)$. The first term is orthogonal to the range of $D$;
the second belongs to that range. Summing the Pythagorean identity over
columns proves~\eqref{model:eq:projection-split}. Each column of $P_DY$ lies
in the range of $D$, so some coefficient matrix attains the lower bound.
When $G$ is positive definite, $A_*=G^{-1}C$ satisfies
$D^*(Y-DA_*)=\bm0_{r\times k}$, identifying $DA_*$ with the projection. Expansion gives
$R_*^*R_*=H-C^*G^{-1}C$. Finally,
$Y-DA=R_*+D(A_*-A)$ and the cross matrices vanish by the normal equations,
which proves~\eqref{model:eq:projection-matrix}.
\end{proof}

An oracle fit on the assessment panel evaluates this finite-panel lower
bound. Such a fit is a descriptive diagnostic, not an out-of-sample
prediction or a population error bound. The excess term measures the
assessment-panel discrepancy of the chosen coefficients from the oracle;
it need not be purely sampling variance. Distribution change, conditioning,
and the coefficient model can all contribute.
Section~\ref{model:sec:projection-results} reports this diagnostic for the complete fixed-dictionary return comparison.

This distinction also constrains fluctuation-scale inheritance. For a
family with weakest retained fluctuation eigenvalue $\lambda_{k,N}>0$,
a fixed dictionary whose orthogonal residual is not
$o(\sqrt{\lambda_{k,N}})$ cannot supply a residual satisfying that
sufficient inheritance condition merely by improving its coefficients.
A finite-panel lower bound diagnoses a finite representation. It does not
establish the large-$N$ premise, the absence of other representations, or
the absence of a native critical mode.

\FloatBarrier
\par\medskip\noindent
Chapter~\ref{ch:conditional-moments} constructs conditional moment
forecasts for these evolving states. Compatible temporal maps, memory and
observation costs determine how such forecasts can be estimated and refreshed.

\chapter{Conditional moments, memory, and observation cost}
\label{ch:conditional-moments}
This chapter gives finite conditional-moment descriptions of training
continuations. It develops compatible blocking, covariance-aware scoring,
state adaptation and refresh costs, and specifies the numerical law under
which a smooth response approximation is meaningful.

\section{Compatible conditional moment maps and their predictive score}
\label{model:sec:moment-transport}

Fix the data law, a realized corpus, an incoming augmented state $s$,
and the evaluation kernel. Let $V_j$ collect specified full-graph
observations after $t_j$ further updates, and put
$D_j=V_j-V_{j-1}$. The vector $D=(D_1,\ldots,D_k)$ retains an entire
finite set of intervals. Its first two conditional moments form an
effective state for \emph{linear observations of this path}. They need
not close the next training transition. In particular, these are raw
increments; centering them by a state-conditional mean does not make
different intervals martingale innovations.

\begin{proposition}[Conditional moment blocking]
\label{model:prop:moment-blocking}
Suppose $D\in\R^d$ has finite second moment under the specified
conditional law, with mean $\mu$ and covariance $C$. For a deterministic
linear block map $B:\R^d\to\R^r$, define
\[
 \mathcal M_B(\mu,C)=(B\mu,BCB^{\mathsf T}).
\]
These are exactly the moments of $BD$, and
$\mathcal M_A\mathcal M_B=\mathcal M_{AB}$ whenever the dimensions
agree. For any $m\ge2$ observed paths the same identity holds for
their sample means and sample covariances with divisor $m-1$.
An estimated pair $(\widehat\mu,\widehat C)$ has exact transported
errors $B(\widehat\mu-\mu)$ and
$B(\widehat C-C)B^{\mathsf T}$.
\end{proposition}
\begin{proof}
Linearity gives $\E[BD]=B\mu$ and
$BD-B\mu=B(D-\mu)$. Expanding the outer product and taking
expectations gives $BCB^{\mathsf T}$. Associativity of matrix
multiplication proves composition. Replacing expectation by the
finite sum of centered outer products, with the common divisor
$m-1$, proves the sample statement. Subtracting the two transported
pairs proves the error formulas.
\end{proof}

The maps may sum adjacent time intervals, project onto risk or
predictive entropy, or combine decoder observations. They act on the
same joint path law, so they are compatible with the exact temporal
kernel construction. Regularizing a covariance separately at each
scale generally breaks this compatibility. We instead regularize
once and then transport the resulting matrix. This convention
distinguishes an estimation rule from the exact moment map.

\begin{proposition}[Common-origin endpoint covariance]
\label{model:prop:endpoint-sample}
Let $m\ge2$ scalar paths share an incoming value $R_0$ and let
$D_{ij}=R_{t_j,i}-R_{t_{j-1},i}$ for $0=t_0<\cdots<t_k$.
With $\widehat C_D$ their increment sample covariance, divisor $m-1$,
\begin{equation}
 \boldsymbol1^{\mathsf T}\widehat C_D\boldsymbol1
 =\frac1{m-1}\sum_{i=1}^m(R_{t_k,i}-\bar R_{t_k})^2.
 \label{model:eq:endpoint-sample}
\end{equation}
If the incoming values depend on $i$, the left side instead equals
the sample variance of $R_{t_k,i}-R_{0,i}$.
\end{proposition}
\begin{proof}
Telescoping gives $\sum_jD_{ij}=R_{t_k,i}-R_0$. Subtract its sample
mean: the common incoming value cancels, leaving
$R_{t_k,i}-\bar R_{t_k}$. The quadratic form of the centered outer
products is the sum of squares of these centered sums, with divisor
$m-1$. For varying origins the same argument centers endpoint minus
origin, proving the final statement. No independence assumption is used.
\end{proof}

The endpoint exposes one quadratic form of a temporal covariance.
It cannot identify all entries. For example, $I_2$ and
$\diag(3/2,1/2)$ are positive definite, distinct, and have the same
quadratic form on $(1,1)^{\mathsf T}$. To assess a full moment matrix
we use the Dawid--Sebastiani mean/covariance score
\cite{dawid1999dispersion}, which also distinguishes the predicted mean.
Its elementary propriety argument is included to specify exactly which
features of the path law are assessed.

\begin{proposition}[A score identifying the first two moments]
\label{model:prop:moment-score}
Let a vector $X\in\R^d$ have finite second moment, mean $\mu_*$ and
positive definite covariance $C_*$. For $C\succ\bm0_{d\times d}$, define
\[
 S(\mu,C;x)=\log\det C+(x-\mu)^{\mathsf T}C^{-1}(x-\mu).
\]
Then the excess expected score over $(\mu_*,C_*)$ is
\begin{align}
 \mathcal L(\mu,C)
 &=\log\frac{\det C}{\det C_*}
   +\tr(C^{-1}C_*)-d
   +(\mu-\mu_*)^{\mathsf T}C^{-1}(\mu-\mu_*)\ge0.
 \label{model:eq:moment-score-regret}
\end{align}
Equality holds exactly when $\mu=\mu_*$ and $C=C_*$. No Gaussian
distribution assumption on $X$ is required.
\end{proposition}
\begin{proof}
Write $X-\mu=(X-\mu_*)+(\mu_*-\mu)$ and expand the quadratic
form. The cross term has zero expectation; the remaining random
term is $\tr(C^{-1}C_*)$. Subtract the true-moment score, whose
expected quadratic term is $d$. If $\lambda_1,\ldots,\lambda_d>0$
are the eigenvalues of $C^{-1/2}C_*C^{-1/2}$, the first three terms
of the difference are $\sum_j(\lambda_j-\log\lambda_j-1)$.
Each is nonnegative because $\log u\le u-1$, with equality only
at $u=1$. The final quadratic term is nonnegative and vanishes
only at the true mean. All eigenvalues equal one precisely when
$C_*=C$, proving the equality statement.
\end{proof}

Coordinates and units must be fixed when comparing scores. A common
invertible change of coordinates adds the same log-determinant
constant to both forecasts, so their score difference is invariant.
Scores at different dimensions are separate targets; transformed-score
comparisons retain their specified observations~\cite{pic2025transformation}. Even a lower
score for the full vector does not imply a lower score at every
coarser scale or identify higher moments or a successor kernel.

\begin{remark}[Fixed mean error in a covariance comparison]
\label{model:rem:score-centering}
Let $x_1,\ldots,x_m\in\R^d$, $m\ge2$, have mean $\bar x$ and
sample covariance $\widehat C$ with divisor $m-1$.
Two positive definite covariance forecasts $C_A,C_B$ share a fixed mean
$\mu$. Put $P=C_A^{-1}-C_B^{-1}$ and $b=\bar x-\mu$. Their average
Dawid--Sebastiani score difference is exactly
\[
 \overline{S_A-S_B}=\log\frac{\det C_A}{\det C_B}
    +\frac{m-1}{m}\tr(P\widehat C)+b^{\mathsf T}Pb.
\]
Thus a common forecast mean still contributes through its error and each
forecast's precision. The three terms are signed; $P$ need not be positive.
\end{remark}
\begin{proof}
Expand $x_i-\mu=(x_i-\bar x)+b$ in the quadratic score difference.
The averaged cross terms vanish since the centered sample sums to zero.
The averaged centered outer product is $(m-1)\widehat C/m$.
Taking its trace against $P$ gives the stated formula. No independence
or Gaussian hypothesis is required for this finite identity.
\end{proof}

\begin{proposition}[Relative moment error survives linear blocking]
\label{model:prop:moment-precision}
With the preceding notation suppose
\[
 \norm{C_*^{-1/2}(C-C_*)C_*^{-1/2}}_{\rm op}\le\epsilon<1,
 \qquad \norm{C_*^{-1/2}(\mu-\mu_*)}\le\delta.
\]
For any full-row-rank $B:\R^d\to\R^r$, the transported pair
satisfies the same two bounds relative to the true moments of $BX$.
Its excess expected score obeys
\begin{equation}
 0\le\mathcal L(B\mu,BCB^{\mathsf T})
 \le\frac{r\epsilon^2}{2(1-\epsilon)^2}
       +\frac{\delta^2}{1-\epsilon},
 \label{model:eq:moment-score-bound}
\end{equation}
where the reference moments in $\mathcal L$ are also transported.
\end{proposition}
\begin{proof}
Set $C_{*,B}=BC_*B^{\mathsf T}\succ\bm0_{r\times r}$ and
$Q=C_{*,B}^{-1/2}BC_*^{1/2}$. Then $QQ^{\mathsf T}=I_r$.
Writing $E=C_*^{-1/2}(C-C_*)C_*^{-1/2}$ and
$v=C_*^{-1/2}(\mu-\mu_*)$, the two transported errors are
$QEQ^{\mathsf T}$ and $Qv$. Since $\norm Q_{\rm op}\le1$,
their norms do not increase. In whitened coordinates the forecast
covariance has eigenvalues $1+e_j$ with $|e_j|\le\epsilon$.
Its covariance contribution to \eqref{model:eq:moment-score-regret} is
\[
 \sum_{j=1}^r\left(\log(1+e_j)+(1+e_j)^{-1}-1\right).
\]
Each summand equals $\int_0^{e_j}u/(1+u)^2\,du$ and lies between
zero and $e_j^2/[2(1-\epsilon)^2]$. The inverse covariance has
operator norm at most $(1-\epsilon)^{-1}$, which bounds its
mean contribution by $\delta^2/(1-\epsilon)$. Summation proves
the result.
\end{proof}

This is a sufficient precision criterion for a hierarchy of finite
moment predictions. It requires covariance error relative to the
true fluctuation scales; a small absolute risk error does not
establish it. At a singular limiting covariance it applies only on
a specified positive definite subspace. The completed experiment in
Section~\ref{model:sec:moment-results} compares frozen moment forecasts
using this score, without treating empirical sample discrepancies
as a bound on the unknown population matrix.

\section{State adaptation, correlation shape and calibrated scale maps}
\label{model:sec:adaptive-moments}

The finite-resource law has nested randomness: a corpus $\mathcal C$,
a complete initialization identity $I$, and the unrevealed continuation.
The following iteration of total covariance specifies which collective
sectors remain when those conditioning variables are averaged
\cite{kallenberg2021}.

\begin{corollary}[Corpus, initialization and continuation sectors]
\label{model:cor:nested-moment-sectors}
Let $X\in\R^d$ have finite second moment. Write
$\mu_{\mathcal C,I}=\E[X\mid\mathcal C,I]$,
$\mu_{\mathcal C}=\E[X\mid\mathcal C]$ and $\mu=\E X$.
Then
\begin{align}
 \Cov(X)&=C_{\rm path}+C_{\rm init}+C_{\rm corpus},\label{model:eq:nested-moment-sectors}\\
 C_{\rm path}&=\E\Cov(X\mid\mathcal C,I),\nonumber\\
 C_{\rm init}&=\E[(\mu_{\mathcal C,I}-\mu_{\mathcal C})
                (\mu_{\mathcal C,I}-\mu_{\mathcal C})^{\mathsf T}],\nonumber\\
 C_{\rm corpus}&=\E[(\mu_{\mathcal C}-\mu)
                  (\mu_{\mathcal C}-\mu)^{\mathsf T}].\nonumber
\end{align}
Each sector is positive semidefinite. Every deterministic common linear
observation $B$ transports the three sectors by $C\mapsto BCB^{\mathsf T}$;
these sector maps compose under successive observations. The identity
also holds exactly for a specified finite empirical hierarchical law,
using that law's probability weights.
\end{corollary}
\begin{proof}
Decompose $X-\mu$ into $X-\mu_{\mathcal C,I}$,
$\mu_{\mathcal C,I}-\mu_{\mathcal C}$ and $\mu_{\mathcal C}-\mu$.
The first term has conditional mean zero given $(\mathcal C,I)$,
and the second has conditional mean zero given $\mathcal C$.
Conditioning each cross outer product on the appropriate information
therefore gives zero. Expanding the outer product of the sum yields
\eqref{model:eq:nested-moment-sectors}. Each term is an expectation of a
centered outer product and hence is positive semidefinite.
Linearity of conditional expectation proves the transport statement;
associativity proves composition. Finite probability weights give the
same argument with finite sums.
\end{proof}

State-dependent standardization must be retained inside these
expectations. Averaging normalized correlations is generally different
from averaging physical covariances. Counts of corpus draws,
initializations and conditional paths are distinct sampling axes;
increasing a branch count does not enlarge a thermodynamic head family.

A transferable covariance shape is a stronger hypothesis than compatible
moment blocking. For each incoming state $s$, write the conditional
increment covariance as $C_s=D_s R_s D_s$, where $D_s$ is its positive
diagonal standard-deviation matrix and $R_s$ its correlation matrix.
Even a common $R_s$ leaves the mean, scale and remaining corpus dependent
on $s$. A fitted moment hierarchy therefore records which of these
objects was transferred and which was estimated at the new state.

\begin{proposition}[A sufficient condition for covariance-shape transfer]
\label{model:prop:correlation-transfer}
Let $C_*=DRD\succ\bm0_{\mathrm{mat}}$, with $D$ positive diagonal and $R\succeq rI$ for $r>0$.
Let $\widehat C=\widehat D\widehat R\widehat D\succ\bm0_{\mathrm{mat}}$, with
$\widehat D$ positive diagonal. Suppose
\[
 \|D^{-1}\widehat D-I\|_{\rm op}\le a,\quad
 \|\widehat R-R\|_{\rm op}\le e,\quad
 \|D^{-1}(\widehat\mu-\mu_*)\|\le b.
\]
Then the relative covariance and mean errors in
Proposition~\ref{model:prop:moment-precision} are bounded by
\begin{equation}
 \epsilon=\frac{(2a+a^2)\|R\|_{\rm op}+(1+a)^2e}{r},
 \qquad \delta=\frac{b}{\sqrt r}.
 \label{model:eq:correlation-transfer-bound}
\end{equation}
If $\epsilon<1$, the same bounds and the resulting score-regret bound
hold after every deterministic full-row-rank linear observation map.
\end{proposition}
\begin{proof}
Put $E=D^{-1}\widehat D-I$. Since both diagonal matrices commute,
\[
 D^{-1}(\widehat C-C_*)D^{-1}
 =ER+RE+ERE+(I+E)(\widehat R-R)(I+E).
\]
The triangle and operator product inequalities give the numerator in
\eqref{model:eq:correlation-transfer-bound}. The matrix
$Q=C_*^{-1/2}DR^{1/2}$ obeys $QQ^{\mathsf T}=I$ and is orthogonal.
Thus the relative covariance error is orthogonally conjugate to
$R^{-1/2}D^{-1}(\widehat C-C_*)D^{-1}R^{-1/2}$.
Its norm is at most the preceding numerator divided by $r$.
Also $C_*^{-1/2}(\widehat\mu-\mu_*)=
QR^{-1/2}D^{-1}(\widehat\mu-\mu_*)$, giving $\delta$.
Apply Proposition~\ref{model:prop:moment-precision} to the transported pair.
\end{proof}

The condition exposes both an estimation cost and a possible state
mismatch. Similar empirical correlations alone certify neither $e$
nor the smallest population eigenvalue $r$. The completed transfer
study compares a frozen archived correlation with local shrinkage
and a diagonal control at equal adaptation budget. Its moment scores
assess those finite predictions; they do not estimate an autonomous
successor kernel or establish a universality class.
Section~\ref{model:sec:outer-results} specifies the outer-state acquisition and all matched covariance forecasts.

A separate finite reduction requires exchangeability, rather than an
accurate covariance model, using split-conformal calibration~\cite{shafer2008}. It extends the whole-path construction in
Proposition~\ref{model:prop:conditional-risk-tube} to all chosen scale maps.

\begin{proposition}[A calibrated tube and its linear scale images]
\label{model:prop:calibrated-scale-images}
Condition on an incoming state and on an independent adaptation sample.
Let $X_1,\ldots,X_n,X_{n+1}\in\R^d$ be exchangeable conditional on
that information. Fix from adaptation a center $m$ and positive
coordinate scales $a_j$. Define
\[
 T(x)=\max_j |x_j-m_j|/a_j,\qquad
 q=\max_{1\le i\le n}T(X_i),\qquad
 \mathcal T=\{x:T(x)\le q\}.
\]
Then $\Pr\{X_{n+1}\in\mathcal T\}\ge n/(n+1)$, with probability
averaged over calibration and the new path, conditional on the
incoming state and adaptation. For any fixed collection of compatible
linear maps $B$, with the same event probability, simultaneously
\begin{equation}
 BX_{n+1}\in B\mathcal T,
 \qquad |BX_{n+1}-Bm|\le q\,|B|a
 \label{model:eq:calibrated-scale-image}
\end{equation}
coordinatewise. Exact images compose: $A(B\mathcal T)=(AB)\mathcal T$.
Coordinate enclosures need not preserve this equality.
\end{proposition}
\begin{proof}
The $n+1$ scores are exchangeable. After independent random tie-breaking,
each index has equal probability of being the largest. Failure of
$T(X_{n+1})\le q$ requires that the new score be strictly largest, so
its probability is at most $1/(n+1)$. On the complementary event
$|X_{n+1}-m|\le qa$. Multiplying each coordinate by the absolute
matrix entries and summing proves the displayed enclosure for every
$B$ on that same event. Membership in the exact image and composition
follow from the definitions of image and matrix multiplication.
\end{proof}

No Gaussian or stationary training assumption enters this result.
Its event concerns an entire path; horizons are not separate independent
replicates. Its coverage is not conditional on the realized calibration
radius and does not assert a fixed useful width. If state adaptation
misses a drift or covariance sector, the empirical tube can widen.
This distinguishes calibrated finite predictive reduction from a
uniformly accurate portable mean law. The two constructions use the
same exact resource-conditioned paths and the same linear observations,
with different accuracy requirements.

For a fixed incoming state, the resulting finite prediction object retains
its fitted mean, covariance, calibrated set and conditioning information.
The moments and calibrated set must use the same coordinates: for risk
levels, apply the cumulative-sum map to the increment moments first.
At a linear observation $B$ these become
$(B\widehat\mu_s,B\widehat C_sB^{\mathsf T},B\mathcal T_s)$.
These maps compose for the fixed object. The conditioning information
includes the remaining corpus, complete optimizer state, observation
panel and adaptation/calibration design. Score error and rank coverage
refer to different properties of this object; neither asserts that its
mean and covariance alone determine the next training kernel.

\section{Training-state transport and the cost of retaining a forecast}
\label{model:sec:state-refresh-theory}

A fitted conditional path law has two distinct scale coordinates.
Observation blocking changes its resolution at a fixed incoming state;
training changes the incoming state itself. The former has the exact
moment map of Proposition~\ref{model:prop:moment-blocking}. The latter changes
the parameters, optimizer memory and remaining corpus, and requires a
separate transport estimate. The following bound makes that difference
quantitative in the fluctuation units used by the moment score.

\begin{proposition}[Reusing moments under conditional path-law drift]
\label{model:prop:state-moment-reuse}
Let $X,Y\in\R^d$ be coupled path-observation vectors at two specified
incoming states, with finite second moments. Write their moments as
$(\mu_0,C_0)$ and $(\mu_1,C_1)$, and assume $C_0\succ\bm0_{d\times d}$.
Suppose $\epsilon,\delta,\rho\ge0$ and a fixed forecast $(m,C)$, $C\succ\bm0_{d\times d}$, satisfy
\[
 \|C_0^{-1/2}(C-C_0)C_0^{-1/2}\|_{\rm op}\le\epsilon,
 \qquad \|C_0^{-1/2}(m-\mu_0)\|\le\delta,
 \qquad \E\|C_0^{-1/2}(Y-X)\|^2\le\rho^2.
\]
Put $\beta=2\rho+\rho^2$. If $\beta<1$, then $C_1\succ\bm0_{d\times d}$ and the
same forecast, reused at the second state, has relative errors bounded by
\begin{equation}
 \epsilon_1=\frac{\epsilon+\beta}{1-\beta},\qquad
 \delta_1=\frac{\delta+\rho}{\sqrt{1-\beta}}.
 \label{model:eq:state-moment-reuse}
\end{equation}
If also $\epsilon_1<1$, every deterministic full-row-rank observation
map has the score-regret bound~\eqref{model:eq:moment-score-bound} with
$(\epsilon_1,\delta_1)$ in place of $(\epsilon,\delta)$.
\end{proposition}
\begin{proof}
Let $W=C_0^{-1/2}(X-\mu_0)$ and
$Z=C_0^{-1/2}(Y-X)$. Then $\Cov W=I$,
$\|\E Z\|\le\rho$ and $\|\Cov Z\|_{\rm op}\le\rho^2$.
For unit vectors $u,v$, Cauchy--Schwarz gives
$|\Cov(u^{\mathsf T}W,v^{\mathsf T}Z)|\le\rho$.
Consequently the covariance expansion
\[
 C_0^{-1/2}C_1C_0^{-1/2}
 =I+\Cov(W,Z)+\Cov(Z,W)+\Cov Z
\]
differs from $I$ in operator norm by at most $\beta$.
In particular, $C_1\succeq(1-\beta)C_0\succ\bm0_{d\times d}$.
The forecast error relative to $C_0$ and the covariance drift together
give
$-(\epsilon+\beta)C_0\preceq C-C_1
\preceq(\epsilon+\beta)C_0$.
Congruence by $C_1^{-1/2}$ and
$C_0\preceq C_1/(1-\beta)$ prove the first bound.
Also
$\|C_0^{-1/2}(m-\mu_1)\|\le\delta+\rho$ and
$\|C_1^{-1/2}C_0^{1/2}\|_{\rm op}\le(1-\beta)^{-1/2}$,
which prove the second. Apply Proposition~\ref{model:prop:moment-precision}
to this forecast and the second state's true moments.
\end{proof}

The coupling is between complete conditional observation laws. A small
weight displacement, similar empirical correlations or a small change
in the consumed fraction alone supplies no bound on $\rho$.
Independent continuation ensembles estimate each state's moments but
do not certify this sufficient population coupling criterion. The
executed state-change experiment in Section~\ref{model:sec:refresh-results}
instead measures reuse and reacquisition directly. Its matched horizons
and observation panel keep the compared coordinates fixed.

\begin{proposition}[Resource deletion with a controlled number of source edits]
\label{model:prop:resource-edit-coupling}
Let $R$ contain $r>1$ indexed blocks, let $A\subseteq R$ have $k$
elements, and let $1\le\ell\le r-k$. There is a coupling of uniform
ordered distinct $\ell$-block sequences $I$ from $R$ and $J$ from
$R\setminus A$ whose Hamming distance $H$ satisfies
\begin{alignat}{2}
 \E H&=\frac{\ell k}{r},&\quad \Var(H)&=\ell\frac{k}{r}\left(1-\frac{k}{r}\right)
                \frac{r-\ell}{r-1},\label{model:eq:resource-edit-moments}\\
 \Pr\{I\ne J\}&=1-\frac{(r-k)_\ell}{(r)_\ell}
 =\|\mathcal L(I)-\mathcal L(J)\|_{\rm TV}.\nonumber
\end{alignat}
Let $F_0,F_1$ be parent and successor observation programs, and let $W$
be a fixed linear map into a Euclidean space. Suppose $c,d_0\ge0$,
$\|W(F_0(i)-F_0(j))\|\le c\,d_H(i,j)$ for all parent-admissible
sequences, and $\|W(F_1(j)-F_0(j))\|\le d_0$ for all
successor-admissible sequences. Then
\begin{equation}
 \left(\E\|W(F_1(J)-F_0(I))\|^2\right)^{1/2}
 \le d_0+c\sqrt{\Var(H)+(\E H)^2}.
 \label{model:eq:resource-edit-observation-bound}
\end{equation}
\end{proposition}
\begin{proof}
Draw $I$ uniformly. Keep every slot whose index is outside $A$ and
fill the remaining slots uniformly without replacement from the
unused indices in $R\setminus A$. The resulting $J$ has distinct
indices. Its law is invariant under every relabeling of $R\setminus A$:
the parent draw and the uniform refill both have this invariance.
Those relabelings act transitively on the ordered distinct sequences,
so $J$ is uniform. Exactly the slots of $I$ lying in $A$ change.
Their indicators have mean $k/r$ and pair mean
$k(k-1)/(r(r-1))$. Expanding the first and second moments of their
sum gives~\eqref{model:eq:resource-edit-moments}. The probability of no
changed slot is $(r-k)_\ell/(r)_\ell$. The successor sequence law
is the parent sequence law conditioned on avoiding $A$, so their
total variation equals the complement probability.
Finally the two program bounds and the triangle inequality give
$\|W(F_1(J)-F_0(I))\|\le d_0+cH$. The $L^2$ triangle
inequality and $\E H^2=\Var(H)+(\E H)^2$ prove
\eqref{model:eq:resource-edit-observation-bound}.
\end{proof}

This coupling can change few source slots even when the complete index
laws have total variation close to one. With $W=C_0^{-1/2}$, its
observation bound supplies a sufficient $\rho$ for
Proposition~\ref{model:prop:state-moment-reuse}, provided the two uniform
program-sensitivity bounds hold. The source-only experiment in
Section~\ref{model:sec:refresh-source-coupling} checks the edit law on the
actual consumed resources. It measures neither $c$ nor $d_0$ and
therefore does not certify native moment transfer. Parameter and
optimizer drift remain distinct from deletion of source indices.

Coverage loss under distribution drift belongs to the established theory
of conformal prediction beyond exchangeability \cite{barber2023}.
The following elementary change-of-test-law bound is applied to a complete
conditional training path, with an independently fitted score and parent
calibration held fixed. The model-specific content is the native state/path
specification, compatible observation-set images and the measured
reuse/recalibration comparison; the total-variation inequality is a
general statistical fact.

\begin{proposition}[Coverage under state change and renewed calibration]
\label{model:prop:state-tube-reuse}
Fix a parent state, a second state, and an adaptation-fitted score.
Let $P$ and $Q$ be the respective path laws. Let $\mathcal T$ be
the maximum-score tube from $n$ independent $P$ calibration paths,
and assume the fresh path is independent of this calibration.
If $\|P-Q\|_{\rm TV}\le v$, then
\begin{equation}
 \Pr_{P^{\otimes n}\otimes Q}\{Y\in\mathcal T\}
 \ge \frac n{n+1}-v.
 \label{model:eq:state-tube-tv}
\end{equation}
Alternatively, suppose a coupling $(X,Y)$ of $P,Q$, independent of
calibration, satisfies
$\Pr\{\|Y-X\|_\infty>a\}\le p$ for some $a\ge0$.
Then the enlarged tube $\mathcal T+[-a,a]^d$ contains $Y$ with
probability at least $n/(n+1)-p$. Exact images of either covered set
under any fixed collection of linear maps retain the same event.
If both adaptation and calibration are instead freshly drawn at the
second state, the bound $n/(n+1)$ holds there without a state-transfer
assumption.
\end{proposition}
\begin{proof}
For each realized tube, $Q(\mathcal T)\ge P(\mathcal T)-v$.
Average over calibration and apply the exchangeable-rank argument
of Proposition~\ref{model:prop:calibrated-scale-images} to obtain
\eqref{model:eq:state-tube-tv}. Under the coupling, the event
$X\in\mathcal T$ and $\|Y-X\|_\infty\le a$ implies membership
of $Y$ in the enlarged tube. The union bound proves the second
claim. The tube is a compact coordinate box, so its linear images
are compact and measurable; membership implies membership of every
image. Fresh adaptation fixes the score independently of the $Q$
calibration paths and fresh $Q$ assessment path. Those $n+1$ scores
are exchangeable, giving the final claim by the same rank argument.
\end{proof}

These probabilities average over calibration and a fresh complete
path, conditional on the stated incoming states and adaptation.
They do not assert coverage for each realized tube, for a selected
favorable state, or jointly over multiple assessment paths. In the
native experiment the successor transition is independent of the
parent fitting paths conditional on the parent complete state.
Conditioning further on that successor therefore leaves the parent
calibration law unchanged. A reused parent tube has no automatic
rank guarantee at the successor; independently recalibrating there
restores the conditional guarantee. This is the same distinction as
moment reuse versus reacquisition, expressed for whole-path events.

\begin{proposition}[Acquisition noise as a shared conditional sector]
\label{model:prop:acquisition-noise}
Let $m\ge1$ be an integer. At one fixed incoming state let $X_1,\ldots,X_m$, $X$ and $X'$ be
independent path vectors with common mean $\mu$ and finite covariance
$C$. Put $\widehat\mu=m^{-1}\sum_{i=1}^mX_i$ and
$Z=X-\widehat\mu$, $Z'=X'-\widehat\mu$. Then
\begin{alignat}{2}
 \Cov(\widehat\mu)&=C/m,&\quad \E\|\widehat\mu-\mu\|^2&=\tr(C)/m,\nonumber\\
 \Cov(Z)&=(1+1/m)C,&\quad \Cov(Z,Z')&=C/m.
 \label{model:eq:acquisition-sector}
\end{alignat}
Conditionally on the realized fitted mean, the respective two
covariances in the second line are $C$ and zero. Two independent
$m$-path fitted means instead have difference covariance $2C/m$.
All these identities commute with deterministic linear observation maps.
\end{proposition}
\begin{proof}
Center the $m$ adaptation vectors by $\mu$. Their cross covariances
vanish by independence, so the covariance of their average is $C/m$;
its expected squared norm is the trace. The future vectors are
independent of that average and of one another. Expanding the centered
outer products of $Z$ and $Z'$ therefore gives $(1+1/m)C$ and $C/m$.
Conditioning fixes $\widehat\mu$ and leaves the future law unchanged,
giving $C$ and zero. Independence of two fitted means adds their
covariances. Congruence by each linear map proves compatibility.
\end{proof}

The added covariance is acquisition uncertainty shared by all future
paths centered using one fit. Conditioning on that fit removes this
sector from the assessment Monte Carlo error. For a scalar observation
with $C_N=\chi_N/N$, the mean-estimation RMS is
$\sqrt{\chi_N/(Nm_N)}$. Making this error negligible relative to
$\sqrt{\chi_N/N}$ requires $m_N\to\infty$ when $\chi_N>0$.
A fixed budget changes the centered susceptibility averaged over fits to
$(1+1/m_N)\chi_N$ in this independent construction; it preserves a
power exponent when that factor has a positive finite limit, but need
not preserve its leading coefficient. This distinguishes common variation due to fitting from a new intrinsic critical mode. The experiment
uses two fit cohorts to expose this sampling axis, without treating
their limited realized spread as a population covariance certificate.
Section~\ref{model:sec:refresh-results} gives the nested budgets, successor-state assessment and acquisition costs.

For a reusable conditioning law, let $A$ be acquisition cost and let
$c_r<c_d$ be reduced and direct query costs at \emph{matched prediction
quality}. A saving over $Q$ queries requires
\begin{equation}
 A+Qc_r<Qc_d,\qquad Q>\frac{A}{c_d-c_r}.
 \label{model:eq:forecast-acquisition-cost}
\end{equation}
This follows by subtracting $Qc_r$ and dividing by $c_d-c_r>0$.
If state drift requires $J$ refits with acquisition costs $A_j$,
replace $A$ by $\sum_{j=1}^J A_j$. The inequality is an accounting
identity, not a speedup theorem: a native sample, an estimated mean
and a calibrated path set provide different statistical information.
The measured ledger includes restoration, native updates, vocabulary
observations, fitting, calibration and score evaluation, with independent
assessment listed as evaluation work.

\section{The numerical law and its smooth approximation}
\label{model:sec:numerical-response}

Precision policy is part of the implemented transition in
Equation~\eqref{model:eq:single-pass-kernel}: stored constants, parameter and
moment dtypes, forced casts, reduction order and the pulse application
convention can affect the outgoing law. Fixing this policy and a complete
incoming state defines a finite conditional experiment. Its exact kernel
and finite-pulse identities remain meaningful without a derivative of the
floating-point program. An idealized real-arithmetic map supplies a
separate approximation, with the following explicit error budget.

\begin{proposition}[Measured pulse error on a fixed observation space]
\label{model:prop:measured-pulse-error}
Let $F:[-a,a]\to\mathcal H$ take values in a real Banach space.
For $0<|h|\le a$, suppose its numerical observations satisfy
$\|\widehat F(u)-F(u)\|\le\epsilon_T$ at
$u=0,\pm h,\pm h/2$. With the pulse definitions of
Proposition~\ref{model:prop:finite-pulse},
\begin{alignat}{2}
 \|\widehat O_h-O_h\|&\le\epsilon_T,&\quad \|\widehat E_h-E_h\|&\le2\epsilon_T.
 \label{model:eq:pulse-arithmetic}
\end{alignat}
If $F$ is $C^3$ and $\sup_{|u|\le a}\|F'''(u)\|\le M_{3,T}$, then
\begin{align}
 \|\widehat O_h/h-F'(0)\|
 &\le M_{3,T}h^2/6+\epsilon_T/|h|,
 \label{model:eq:measured-derivative}\\
 \|\widehat O_h-2\widehat O_{h/2}\|
 &\le 5M_{3,T}|h|^3/24+3\epsilon_T.
 \label{model:eq:measured-halving}
\end{align}
The same inequalities hold in conditional $L^2$ if their hypotheses hold
in that norm. A deterministic bounded linear observation map $B$
commutes with the finite pulse parts and multiplies these error bounds
by at most its induced operator norm.
\end{proposition}
\begin{proof}
Put $e_u=\widehat F(u)-F(u)$. The odd error equals
$(e_h-e_{-h})/2$, and the even error equals
$(e_h+e_{-h})/2-e_0$. Their norms are bounded by $\epsilon_T$
and $2\epsilon_T$ by the triangle inequality. The third-order integral
Taylor remainder has norm at most $M_{3,T}|h|^3/6$ at either
sign. Subtraction cancels the constant and quadratic terms; division
by $h$ and addition of the odd numerical error give
Equation~\eqref{model:eq:measured-derivative}. Applying the exact remainder
bound at $h$ and $h/2$ gives
$M_{3,T}|h|^3(1+1/4)/6$ for the exact halving numerator.
The full odd error contributes $\epsilon_T$ and twice the half-radius
odd error contributes $2\epsilon_T$, proving
Equation~\eqref{model:eq:measured-halving}. The conditional statement uses
the triangle inequality in $L^2$. Linearity and boundedness prove
the final assertion.
\end{proof}

For an observation weighting $W\succ\bm0_{\mathrm{mat}}$, use
$\|z\|_W^2=z^{\mathsf T}Wz$. Between two weighted resolutions,
the relevant operator norm is
$\|W_{\rm out}^{1/2} B W_{\rm in}^{-1/2}\|_2$.
A reduction of the pulse radius decreases the smooth truncation term
while increasing the numerical term after division by $|h|$.
For positive $M_{3,T}$ and $\epsilon_T$, the unconstrained minimizer
of the bound in Equation~\eqref{model:eq:measured-derivative} is
$|h|=(3\epsilon_T/M_{3,T})^{1/3}$; it is admissible only inside
the assumed domain. Neither constant is certified by the native
experiment. A float32/float64 discrepancy measures implementation
sensitivity and is not an enclosure of real-arithmetic error.
Bitwise replay checks the repeatability of a chosen implementation.
Section~\ref{model:sec:numerical-response-results} details the finite-radius trajectory comparisons and their arithmetic controls.

\begin{proposition}[A local smooth domain of the native program]
\label{model:prop:native-smooth-domain}
Fix a finite source sequence, dropout-free native PLGA architecture,
positive normalization offsets, fixed optimizer counters and schedule,
and a parameter pulse interval. Replace floating-point operations by
real arithmetic. Suppose throughout a neighborhood of every trajectory
on that interval: each clipping branch has a strict margin; each
second-moment coordinate used in an Adam square root is positive;
and each metric energy used in an observed RMS is positive. Coordinates
that are identically constant may instead be removed before imposing
the positivity conditions. Then the finite trajectory observation is
$C^3$ in its pulse amplitude. On a compact pulse interval contained
in that domain, $M_{3,T}$ is finite.
\end{proposition}
\begin{proof}
The native elementwise positive power base is
$x\,\operatorname{SiLU}(x)+\varepsilon_A
=x^2/(1+e^{-x})+\varepsilon_A>0$.
Its power with a variable exponent is
$\exp(p\log(x^2/(1+e^{-x})+\varepsilon_A))$, a smooth function
of both arguments. Affine operations, SiLU, softmax and log-softmax
are smooth; positive offsets keep normalization denominators away
from zero locally. On an inactive clipping branch the gradient is
unchanged; on a strictly active branch its norm is nonzero and its
normalization is smooth. Positive second moments make the Adam square
roots smooth. Fixed counters make bias correction and drive factors
constants in the pulse variable. Induction through the finite sequence
therefore gives a smooth state map. Positive observed energies make
the RMS and logarithmic emission smooth as well. Continuity of its
third derivative on a compact interval gives a finite maximum.
\end{proof}

This is a sufficient local condition, not a uniform certificate for
trained models. In particular, an optimizer offset outside a square
root does not make that square root differentiable at zero. The native
SiLU and positive power-base activation have no real-arithmetic kink
at zero. Numerical underflow, clipped-gradient branches, degenerate
moments and nonlinear amplification must be distinguished when
interpreting a measured response. The experiments record clipping
norms, power-activation minima and zero second-moment counts; they do
not bound all third derivatives on the intervening pulse interval.
These diagnostics and their sampled domains are reported in Section~\ref{model:sec:numerical-response-results}.

\begin{proposition}[Chronological implementation and reduction error]
\label{model:prop:implementation-error}
Suppose $s_{t+1}=T_t(s_t)$ and
$\widehat s_{t+1}=\widehat T_t(\widehat s_t)$ are coupled on a declared
domain where $T_t$ is $L_t$-Lipschitz, $L_t\ge0$, and
$\|\widehat T_t(\widehat s_t)-T_t(\widehat s_t)\|\le\eta_t$.
For $e_0=\|\widehat s_0-s_0\|$,
\begin{equation}
 \|\widehat s_T-s_T\|
 \le e_0\prod_{j=0}^{T-1}L_j+
       \sum_{k=0}^{T-1}\eta_k\prod_{j=k+1}^{T-1}L_j.
 \label{model:eq:implementation-chronology}
\end{equation}
If an emission is $L_{\rm em}$-Lipschitz with implementation error
$\eta_{\rm em}$, its discrepancy is at most $L_{\rm em}$ times
this bound plus $\eta_{\rm em}$.
\end{proposition}
\begin{proof}
Add and subtract $T_t(\widehat s_t)$. The triangle inequality gives
$e_{t+1}\le L_te_t+\eta_t$. Induction gives the chronological
products in Equation~\eqref{model:eq:implementation-chronology}, with
empty products equal to one. The emission follows by the same
addition and subtraction.
\end{proof}

This specialization of the reduction recursion explains why a
per-observation engineering floor is not a propagated path-error bound.
For a weighted path, apply the emission bound at every retained time
and take the corresponding weighted norm. If a uniformly contractive
reduction has gap $1-\rho_N\asymp N^{-\zeta}$, its accumulated
per-step error is of order $N^\zeta\eta_N$ after the initial transient.
Preserving an inference susceptibility $\chi_N\sim cN^\kappa$
therefore requires, as a sufficient condition,
$\eta_N=o(N^{(\kappa-1)/2-\zeta})$, with the same control on
acquisition, numerical and emission errors in their respective units
(Proposition~\ref{model:prop:critical-closure}). Fixed finite-risk accuracy
alone does not establish this scale-dependent condition.

A thermodynamic family must accordingly specify its numerical policy
$\nu_N$ alongside size, corpus resource, clocks and observation units.
Changing $\nu_N$ is a change of the implemented law. Transferring an
idealized limiting class requires numerical and reduction errors small
relative to its connected fluctuation scale. Neither fixed arithmetic
nor an incoming-state amplitude check alone establishes that condition.

\FloatBarrier
\par\medskip\noindent
Chapter~\ref{ch:forecast-evidence} tests these forecasting constructions
on fresh continuations. Its frozen predictions and matched interventions
separate estimation cost, state transfer and numerical response error.

\chapter{Executed forecasts on fresh continuations}
\label{ch:forecast-evidence}
This chapter evaluates finite predictions on fresh single-pass paths.
The studies test conditional covariance and memory, temporal moment maps,
transfer across corpus and incoming-state draws, refresh budgets and finite
pulse response under stated numerical and inference observations.

\section{Finite conditional prediction under single-pass continuation}
\label{model:sec:law-closure-results}

This study tests finite risk-path prediction after restoring a complete
incoming training state: weights, both Adam moments, bias-correction
counter, rate groups and schedule phase, consumed blocks and remaining
corpus. The development fit and 32-document evaluation panel are held
fixed. Each observation is the external-target negative log likelihood,
jointly at horizons $1,4,16,64$; the score evaluates the whole path.
At each state, sixteen development, eight fresh calibration and sixteen
fresh validation branches use the same reset law. A branch draws
2,048 distinct blocks uniformly without replacement from the remaining
population for 64 batch-32 updates. Counterfactual branches may overlap
each other; an individual path does not reuse a consumed source position.
Two outer initialization identities supply eight paired incoming states.
Branches, widths, checkpoints and horizons do not add independent
training initializations.

All four fixed deterministic candidates fail the all-cell $0.01$-nat
mean-risk target (Table~\ref{model:tab:original-candidate-results}).
The subsequent state-conditioned tubes cover 117 of 128 validation paths,
with horizon-64 half-widths $0.0431$--$0.3472$ nats
(Table~\ref{model:tab:law-closure-risk}). These are separate prediction
objects. The pooled count is descriptive. Conditional on the incoming state
and development fit, the $8/9$ guarantee averages over calibration and one
fresh whole path. It is neither a confidence interval for mean risk nor a
guarantee conditional on each realized calibration set or simultaneous over
arbitrary states or future branches.

The evolving-law formulation marginalizes the unrevealed future ordering.
The native five-decoder architecture, head dimension 64, shared metric
width 170 and controlled constant-rate recipe are unchanged.

\subsection{Reduced-coordinate forecasts across incoming initializations}
The initial conditional-continuation protocol fixes four candidates:
identity, translation, row/predictive affine transport, and
optimizer-augmented affine transport. The first three retain 25
coordinates; the last retains 29, adding four Adam-moment summaries.
These summaries are distinct from the complete optimizer state.
Fits, coordinate centers and scales use only initialization identities
640101 and 640102, separately for each width, incoming time and
observed interval. The two fitted affine candidates use ridge
penalty 1 and spectral norm at most 1. Validation uses identities
640103 and 640104. Widths $4,14$, incoming updates 8,192 and 32,768,
two validation identities and horizons $1,4,16,64$ give 32 cells per
candidate, including eight at horizon 64. The two identities are the
outer validation units; the cells are paired observations.

\begin{table}[htbp]\centering\small
\caption{All four fixed reduced-coordinate forecasts. A hit means
absolute error of mean external-target risk at most 0.01 nats.
MAE averages absolute cell errors and is not a confidence interval.
The error unit is nats per external target on the fixed 32-document
evaluation panel. No candidate is selected using these validation results.}
\label{model:tab:original-candidate-results}
\begin{tabular}{@{}lrrrr@{}}
\toprule Candidate & Hits / 32 & Hits at 64 / 8 & MAE & MAE at 64\\\midrule
Identity & 14 & 3 & 0.01770 & 0.02697\\
Translation & 13 & 3 & 0.02131 & 0.03360\\
Row/predictive & 6 & 1 & 0.08840 & 0.10503\\
Optimizer-augmented & 7 & 1 & 0.08656 & 0.10191\\
\bottomrule
\end{tabular}

\end{table}

No candidate meets the 0.01-nat target across the tested domain.
The largest absolute cell errors are 0.06979, 0.08530, 0.99610 and
0.97406 nats, respectively. At horizon 64, the empirical mean-risk
telescoping bounds range from 0.3814 to 0.9848 for identity,
0.4247 to 0.9963 for translation, 0.7359 to 3.2318 for the
row/predictive model and 0.7331 to 3.2121 for the optimizer-augmented
model. These are valid empirical bounds for the retained-coordinate
paths; they do not certify useful prediction at the chosen tolerance
or a uniform population closure defect. Since each candidate is
deterministic and each stratum starts at a common retained state,
its conditional branch covariance is zero.

The following conditional path-law study changes the prediction
object. It uses sixteen initial branches at each complete incoming
state to fit that state's risk center and scale, including the states
previously used for the cross-initialization validation above.
It then uses eight new calibration branches and sixteen further
validation branches per state. These are fresh paths from the same
incoming states, with fits and tubes frozen before their respective
validation. They test same-state conditional path prediction; they
do not retest cross-initialization closure. All 256 initial branches
remain part of the original forecast experiment even when a subset
is reused as development data for this distinct conditional study.

\subsection{A specified state-conditioned path law}
The reported conditional prediction strata cross $N=4,14$, incoming
updates 8,192 and 32,768, and initialization identities 640103 and
640104. Thus there are eight incoming states and two outer
initialization identities, with widths and checkpoints paired within
identity. Every observation is the average external-next-token NLL
over the same 32 short-cohort evaluation documents, rows $[512,544)$.
The objective uses the target immediately after 64 input tokens.
Native training and evaluation use float32, full batch 32 and disabled
TF32; log-softmax, target losses and reported reductions use float64.

At each incoming state, sixteen development branches fix the center
and scale of the four-dimensional future-risk vector at horizons
$1,4,16,64$. The center is the mean development risk; each scale is
the sample standard deviation with divisor fifteen, floored at
$0.01$ nats to define a nondegenerate score. That floor is a score
normalization, not a statement of achieved prediction accuracy.
The empirical joint path law retains the correlated innovation
sequence. Its conditioner includes the complete incoming state;
coefficients are fitted within that stratum.

Eight entirely fresh calibration branches then determine the maximum
standardized whole-path score in
Proposition~\ref{model:prop:conditional-risk-tube}. The resulting tube is
frozen before sixteen further validation branches are executed at
each state. These 64 calibration and 128 validation branches are
independent of the development fit under the specified reset law.
For the $n=8$ instance of Proposition~\ref{model:prop:conditional-risk-tube},
hold the complete incoming state, remaining corpus, evaluation panel and
development fit fixed. Let $S_1,\ldots,S_8$ be the calibration whole-path
scores and $S_9$ the score of one fresh reset continuation. For
$H=\{1,4,16,64\}$, fixed development centers $a_h$ and positive scales
$b_h$, the score and threshold are
\[
 S_i=\max_{h\in H}\frac{|R_h^{(i)}-a_h|}{b_h},
 \qquad q=\max_{1\le i\le8}S_i.
\]
Under the stated conditional exchangeability, the rank of $S_9$ is
uniform among nine positions when there are no ties. A miss requires
$S_9>q$, hence rank nine. With ties, random ordering within tied groups
gives a uniform rank and the inclusive boundary remains conservative.

\begin{samepage}
Thus, with $\mathcal C$ denoting the fixed conditioning,
\[
 \mathbb P\!\left(\,|R_h^{(9)}-a_h|\le q b_h
           \text{ for every }h\in H\mid\mathcal C\right)
 \ge 1-\frac19=\frac89.
\]
\end{samepage}
The probability averages over the calibration sample and one future path.
Independence of the four horizons is not required. Only the threshold is
calibrated by the eight branches; another incoming state requires its own
development fit and calibration. This statement does not give conditional
coverage for every realized tube or simultaneous coverage of all sixteen
validation branches.

\begin{table}[htbp]\centering\small
\caption{Complete state-conditioned validation. Coverage counts whole
paths whose four risks all lie inside the frozen tube. Half-width and
signed center-minus-validation-mean error refer to horizon 64 and are
in nats per external target. Identities 3 and 4 denote the two paired
initializations. Each row conditions on one incoming state and one
fixed 32-document evaluation cohort.}
\label{model:tab:law-closure-risk}
\begin{tabular}{@{}rrrrrr@{}}
\toprule $N$ & Incoming update & Identity & Covered paths & Half-width at 64 & Mean error at 64\\\midrule
4 & 8,192 & 3 & 15/16 & 0.1951 & -0.0358\\
4 & 32,768 & 3 & 11/16 & 0.0431 & +0.0019\\
4 & 8,192 & 4 & 16/16 & 0.1410 & +0.0206\\
4 & 32,768 & 4 & 15/16 & 0.1444 & -0.0113\\
14 & 8,192 & 3 & 15/16 & 0.1166 & +0.0285\\
14 & 32,768 & 3 & 16/16 & 0.3472 & -0.0075\\
14 & 8,192 & 4 & 14/16 & 0.1004 & -0.0234\\
14 & 32,768 & 4 & 15/16 & 0.1385 & +0.0436\\
\bottomrule
\end{tabular}

\end{table}

The frozen tubes cover 117 of the 128 fresh risk paths, or 91.4\%,
with per-state counts from 11 to 16 of 16. At horizon 64 the
half-widths range from 0.0431 to 0.3472 nats; the largest absolute
center-minus-validation-mean discrepancy is 0.0436 nats. Thus the
finite stochastic prediction retains appreciable state-dependent
uncertainty. The narrowest horizon-64 tube belongs to the state with
11 covered paths. That realized count is reported directly; it is
not inconsistent with a guarantee that averages over fresh
calibration sets. The pooled count is descriptive, and neither it
nor the eight strata supplies an independent eight-initialization
population confidence statement.

\begin{figure}[htbp]\centering
\includegraphics[width=.96\textwidth]{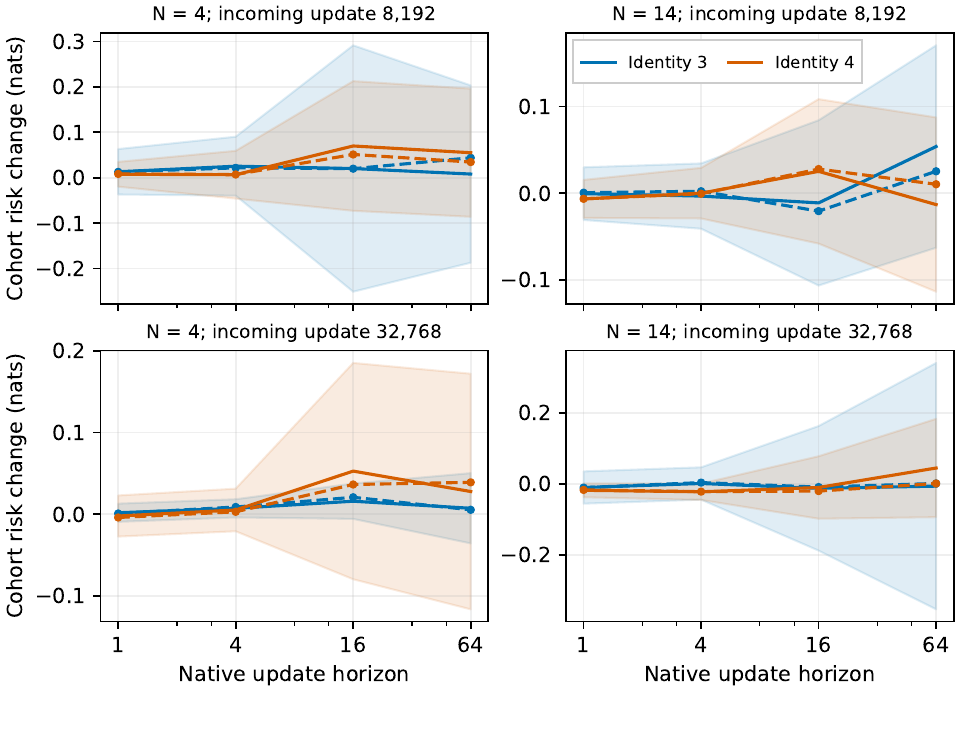}
\caption{Frozen conditional risk-path centers and prediction tubes
(solid curves and shading), with the fresh validation mean at each
horizon (dashed curves and points). Values are relative to each
incoming state's risk. The shading is a prediction set for the
four-horizon path of one branch, not a confidence band for its mean.
Connecting curves interpolate the four measured horizons.
The coverage counts use all individual paths, as reported in
Table~\ref{model:tab:law-closure-risk}.}
\label{model:fig:law-closure-risk}
\end{figure}

\subsection{Measured covariance of the risk-increment history}
The four observed intervals have lengths $1,3,12,48$ native updates.
For each fresh validation path, write its successive risk increments
as $\Delta_j R$. At fixed incoming state, the population-divisor
covariance of the sixteen recorded branches obeys
\[
 \Var(R_{64})=
 \sum_j\Var(\Delta_jR)+
 \sum_{i\ne j}\Cov(\Delta_iR,\Delta_jR).
\]
This is the $A_j=I$ specialization of
Proposition~\ref{model:prop:law-innovation-transport}. The measurements
retain the signed cross sum and the complete interval covariance,
not merely a sequence of marginal noise variances.

\begin{table}[htbp]\centering\small
\caption{Conditional risk-increment covariance from the sixteen fresh
validation branches per incoming state. Variances and cross sums are
in squared nats. The cross share is the signed cross sum divided by
$\Var(R_{64})$; it is not restricted to $[-1,1]$. All entries use
divisor sixteen and the same paired horizons and contexts.}
\label{model:tab:law-closure-innovation}
\begin{tabular}{@{}rrrrrrr@{}}
\toprule $N$ & Incoming update & Identity & $\Var(R_{64})$ & Diagonal sum & Cross sum & Cross share\\\midrule
4 & 8,192 & 3 & $4.25\times10^{-3}$ & $7.06\times10^{-3}$ & $-2.81\times10^{-3}$ & -0.662\\
4 & 32,768 & 3 & $4.61\times10^{-4}$ & $1.17\times10^{-3}$ & $-7.09\times10^{-4}$ & -1.539\\
4 & 8,192 & 4 & $1.11\times10^{-3}$ & $5.31\times10^{-3}$ & $-4.20\times10^{-3}$ & -3.790\\
4 & 32,768 & 4 & $1.78\times10^{-3}$ & $5.01\times10^{-3}$ & $-3.23\times10^{-3}$ & -1.810\\
14 & 8,192 & 3 & $4.14\times10^{-3}$ & $3.00\times10^{-3}$ & $1.14\times10^{-3}$ & 0.275\\
14 & 32,768 & 3 & $4.02\times10^{-3}$ & $6.59\times10^{-3}$ & $-2.57\times10^{-3}$ & -0.638\\
14 & 8,192 & 4 & $2.82\times10^{-3}$ & $3.16\times10^{-3}$ & $-3.40\times10^{-4}$ & -0.121\\
14 & 32,768 & 4 & $3.96\times10^{-3}$ & $6.12\times10^{-3}$ & $-2.16\times10^{-3}$ & -0.544\\
\bottomrule
\end{tabular}

\end{table}

Seven of the eight states have a negative signed cross-interval
sum, while the $N=14$, update-8,192 state of identity 640103 has a
positive sum. The cross shares range from $-3.790$ to $0.275$.
Consequently, replacing the conditional history by independent
interval marginals would generally change the endpoint risk
variance, and the direction of that change depends on the incoming
state. The retained innovation history captures both cancellation
and reinforcement under the single-pass update law.

The covariance transport is an exact finite reconstruction; predictive
coverage is a separate test using unseen paths. Together they give
a stochastic description of inference risk selected by short native
training continuations. The conditional coefficients depend on the
complete incoming state. These measurements supply no autonomous
training law in the displayed row and optimizer summaries, no
stationary approximation, and no native critical exponent. The
law-averaged error and conditional-memory constructions retain the
information needed to state those additional obligations precisely.

\subsection{Execution and verification scope}
The complete conditional study contains 448 scientific branches and
28,672 native updates. It includes the 256 branches of the original reduced-coordinate experiment at sixteen incoming states,
the 64 fresh
calibration branches and the 128 fresh validation branches above.
The sixteen development branches in each displayed stratum are a
subset of the 256, so they are counted once. These conditional
branches add no independent pretraining initialization to the
54-path primary training inventory.

Two reset qualifications executed 24 additional GPU updates.
Independent 64-update replays on both producing GPUs executed 128
further updates and reproduced all final model and Adam tensors and
the full-vocabulary logits at all five observed horizons byte for
byte. These are execution checks, separately counted from scientific
branches. Independent reductions reconstruct all source-block
selections, every reported risk and coverage decision, 160 retained
full-vocabulary horizon panels, and the signed covariance sums.
The finite rank proof was also checked on all $9!$ distinct-score
permutations and all $3^9$ tied-score tuples. The supporting record
retains every development and validation outcome; the displayed
finite prediction law uses its stated complete-state conditioner.

\section{Frozen covariance and history forecasts on fresh continuations}
\label{model:sec:memory-results}

The eight incoming states derive from two outer initialization identities,
with widths and checkpoints paired within identity. All comparisons condition
on their recorded remaining corpus and the same external-target risk panel.
Additional reset branches increase conditional path replication; they do not
increase the number of independently initialized models.

A same-state endpoint-variance forecast uses coefficients fixed before
its assessment paths are drawn. The eight incoming states, complete-state reset law, corpus,
batch size and evaluation documents are those of
Section~\ref{model:sec:law-closure-results}. The forty available paths per
state, consisting of sixteen initial, eight calibration and sixteen
validation paths from that conditional study, supply development data
for this distinct test. Their former split roles are preserved in the
preceding results. No coefficient is transferred to a new incoming
initialization. Thirty-two further branches per state provide 256
fresh paths and 16,384 native updates. Each branch consumes 2,048
distinct remaining RefinedWeb blocks; its observations are at native
horizons $0,1,2,4,8,16,32,64$.

\subsection{Endpoint variance and a covariance-diagonalization control}
For each state, let $D_i$ be the four-dimensional increment vector
from development path $i$ over intervals $[0,1]$, $[1,4]$, $[4,16]$
and $[16,64]$. The frozen covariance is
\[
 \widehat C=\frac1{39}\sum_{i=1}^{40}
 (D_i-\bar D)(D_i-\bar D)^{\mathsf T}.
\]
Proposition~\ref{model:prop:endpoint-sample} identifies the joint endpoint forecast
$\widehat V_{\rm joint}=\boldsymbol1^{\mathsf T}\widehat C\boldsymbol1$.
It is exactly the direct development-sample endpoint variance, not
an additional four-dimensional memory estimator. The
covariance-diagonalization control uses
$\widehat V_{\rm diag}=\tr\widehat C$ with the same marginal
variances. Both are frozen forecasts, compared with the endpoint
sample variance of the 32 fresh branches using divisor 31. This
is distinct from reconstructing a covariance from the paths being
predicted. The comparison measures the cost of dropping the signed cross-interval
covariances. It does not identify their individual entries or a forcing
spectrum. Neither forecast is assigned a population uncertainty
interval or an asymptotic scaling interpretation.

\begin{table}[htbp]\centering\small
\caption{Every frozen direct endpoint-variance forecast, its diagonal control and its fresh
endpoint measurement. All three variance columns are in
$10^{-3}$ squared nats per external target. The final column evaluates
the unchanged four-horizon prediction tubes of
Section~\ref{model:sec:law-closure-results} on the 32 new paths, with no
refitting or recalibration. Widths and checkpoints are paired within
the two initialization identities.}
\label{model:tab:memory-covariance}
\begin{tabular}{@{}rrrrrrr@{}}
\toprule $N$ & Update & Identity & Measured $V$ & Direct forecast & Diagonal forecast & Covered\\\midrule
4 & 8,192 & 3 & 3.670 & 3.621 & 7.777 & 32/32\\
4 & 32,768 & 3 & 0.940 & 0.867 & 1.504 & 17/32\\
4 & 8,192 & 4 & 1.926 & 2.484 & 7.771 & 32/32\\
4 & 32,768 & 4 & 1.969 & 2.151 & 5.858 & 32/32\\
14 & 8,192 & 3 & 3.624 & 3.224 & 3.587 & 27/32\\
14 & 32,768 & 3 & 4.234 & 4.792 & 6.531 & 32/32\\
14 & 8,192 & 4 & 2.497 & 2.814 & 4.013 & 27/32\\
14 & 32,768 & 4 & 3.104 & 4.801 & 6.098 & 30/32\\
\bottomrule
\end{tabular}

\end{table}

The joint/direct forecast has smaller absolute endpoint variance error in 7 of eight states. Across the eight equally weighted states, its mean absolute error is 0.000479 squared nats, compared with 0.002656 for the covariance-diagonalization control. Joint forecast-to-measurement ratios range from 0.890 to 1.547. The joint forecast equals the direct design endpoint sample variance. The comparison measures the cost of discarding signed cross terms, with finite estimation error still visible at individual states. This endpoint comparison identifies neither the individual covariance entries nor a stationary noise model.

The fixed tubes cover 229 of the 256 further four-horizon paths. The same calibration sets are reused across these predictions. As in Proposition~\ref{model:prop:conditional-risk-tube}, the coverage guarantee averages over calibration and a new path; it does not impose the nominal fraction on each realized set of 32 paths.

\subsection{Predictive memory has an estimation cost}
Three endpoint predictors are fixed for every incoming state. The
first uses its development mean risk at horizon 64. The second adds
a linear predictor of the current branch's risk at horizon 16. The
third uses that branch's risks at horizons 1, 4 and 16. Thus the
last two forecasts are issued at horizon 16, using only observations
available by that time. They are assessed on the same endpoint loss,
and their different information sets are explicit.

Each history coordinate is centered on its development mean and
divided by its development sample standard deviation, floored at
$0.01$ nats. With standardized history $x_i$ and centered endpoint
risk $y_i$, the fitted coefficient minimizes
\[
 \frac1{40}\sum_{i=1}^{40}(y_i-\beta^{\mathsf T}x_i)^2
       +0.1\norm\beta^2.
\]
The intercept is the development endpoint mean. The dimension,
scaling, penalty and complete list of predictors are fixed before
any fresh path. All errors below use the 32 new paths. The mean-risk
screen additionally retains the identity forecast at the incoming
risk and assesses the fixed state mean at each of the four matched
horizons against the $0.01$-nat resolution target.

\begin{table}[htbp]\centering\small
\caption{Endpoint mean-square errors of every fixed predictor, in
$10^{-3}$ squared nats. Last-risk uses horizon 16; history uses
horizons 1, 4 and 16. Hits count the four mean-risk cells within
$0.01$ nats for the fixed state-mean predictor. These are empirical
errors conditional on a complete state and the evaluation cohort,
not independent pretrained-model error bars.}
\label{model:tab:memory-forecasts}
\begin{tabular}{@{}rrrrrrr@{}}
\toprule $N$ & Update & Identity & Mean MSE & Last-risk MSE & History MSE & Mean hits / 4\\\midrule
4 & 8,192 & 3 & 3.583 & 3.579 & 4.134 & 3\\
4 & 32,768 & 3 & 1.001 & 0.970 & 0.953 & 4\\
4 & 8,192 & 4 & 1.868 & 1.801 & 2.105 & 4\\
4 & 32,768 & 4 & 1.924 & 1.870 & 1.900 & 3\\
14 & 8,192 & 3 & 4.060 & 5.775 & 5.707 & 3\\
14 & 32,768 & 3 & 4.224 & 3.780 & 4.594 & 3\\
14 & 8,192 & 4 & 2.636 & 2.429 & 2.482 & 3\\
14 & 32,768 & 4 & 3.105 & 3.361 & 3.651 & 3\\
\bottomrule
\end{tabular}

\end{table}

Across equally weighted states, endpoint MSE is 0.002800, 0.002946 and 0.003191 squared nats for the state mean, last-risk and three-risk history predictors, respectively. The last-risk predictor improves on the state mean in 6 of eight states; the history predictor does so in 3. Every fitted outcome is retained. An improvement here concerns conditional prediction after observing a partial path, not autonomous training prediction from an initial row summary.

The state-mean forecast meets the 0.01-nat target in 26/32 cells; the identity forecast meets it in 16/32. The largest absolute state-mean cell error is 0.02345 nats. The state-mean predictor therefore does not meet a uniform 0.01-nat criterion over this finite domain.

\begin{table}[htbp]\centering\small
\caption{Complete signed state-mean forecast errors in nats, forecast
minus fresh branch mean, at each of the four assessed horizons.
The source-bound numerical record additionally contains every
identity error and every individual history forecast.}
\label{model:tab:memory-risk-cells}
\begin{tabular}{@{}rrrrrrr@{}}
\toprule $N$ & Update & Identity & $h=1$ & $h=4$ & $h=16$ & $h=64$\\\midrule
4 & 8,192 & 3 & +0.00093 & +0.00262 & +0.01021 & +0.00528\\
4 & 32,768 & 3 & -0.00053 & +0.00126 & -0.00305 & -0.00950\\
4 & 8,192 & 4 & -0.00122 & +0.00020 & +0.00340 & +0.00154\\
4 & 32,768 & 4 & +0.00077 & +0.00501 & +0.01612 & -0.00402\\
14 & 8,192 & 3 & -0.00046 & -0.00217 & -0.00308 & +0.02345\\
14 & 32,768 & 3 & -0.00137 & -0.00240 & -0.00056 & -0.01103\\
14 & 8,192 & 4 & +0.00083 & +0.00128 & +0.00390 & -0.01475\\
14 & 32,768 & 4 & -0.00044 & -0.00034 & -0.01170 & +0.00989\\
\bottomrule
\end{tabular}

\end{table}

\begin{figure}[htbp]\centering
\includegraphics[width=.96\textwidth]{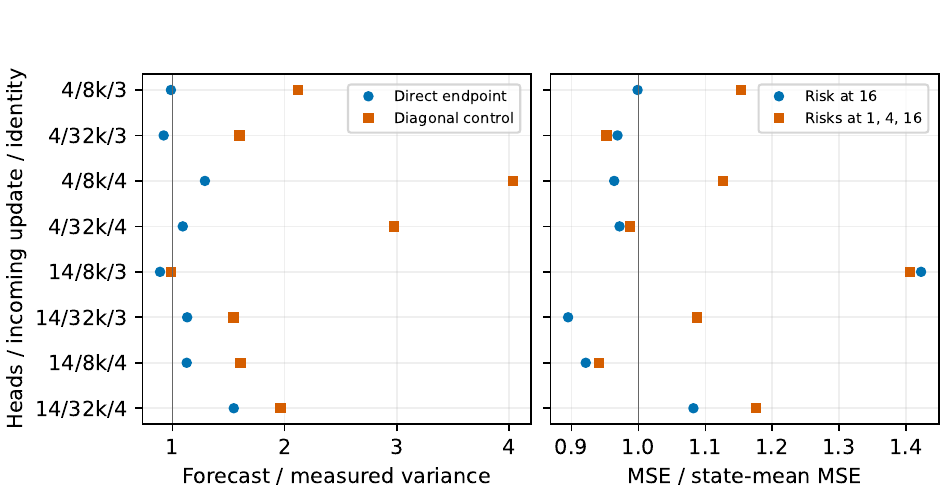}
\caption{All eight finite prediction strata. Left: each frozen
direct endpoint-variance forecast and its covariance-diagonalization control divided by the measured fresh endpoint variance.
Right: the endpoint MSE of each fitted memory predictor divided by
the state-mean MSE. The reference line denotes exact variance
agreement on the left and equal MSE on the right. Ratios are
descriptive comparisons; no population or simultaneous confidence
interval is implied.}
\label{model:fig:memory-validation}
\end{figure}

The finite conditional theory consequently retains the incoming
state, the joint covariance and the declared past observations as
separate information. Exact covariance transport does not make its
finite-sample estimate exact. The oracle improvement for nested
observation spans in Proposition~\ref{model:prop:affine-memory} does not
guarantee improvement of estimated ridge predictors. Residual
orthogonality in a fitted finite linear span is not a martingale
condition on the complete consuming-corpus process. These
 distinctions delimit the reduction supported by this experiment.

\subsection{Native and numerical qualification}
Independent reconstruction checks all 256 new branch arrays, their
block selections, source-position nonreuse, risks, frozen coefficients,
covariances and coverage decisions. The full-vocabulary sidecars
comprise eight horizons from one branch at every state, or 64
panels. Two separately executed 64-update native replays, one per
width, compare all 8,192,000 retained logit coordinates per replay
and the final complete model and Adam state bytewise. The 128 replay
updates are verification executions, separate from the 16,384
scientific updates. These checks add no independent pretrained
initialization or new evaluation document.

The mathematical qualification separately tests 64 four-step,
three-dimensional covariance paths with correlated initial states
and forcing. It also integrates the spectral formula for
$s(\omega)=(2|\sin(\omega/2)|)^\alpha$, with
$\alpha\in\{-1/2,0,1/2\}$ and gaps $10^{-2},10^{-3},10^{-4},10^{-5}$.
All twelve quadratures satisfy their declared checks; the relative
errors of the leading asymptotic decrease with the gap and are
below one percent at $10^{-5}$. The flat-spectrum integrals also
agree with $1/(1-\lambda^2)$. These are qualifications of the
conditional mathematics, not measurements of a native forcing
spectrum or evidence of native stationary critical behavior.

\section{Conditional moment predictions across temporal scales}
\label{model:sec:moment-results}

The endpoint statistic in Section~\ref{model:sec:memory-results} tests one
quadratic form of a covariance. We separately test a full temporal
moment prediction and its linear coarse observations. The conditional
law fixes the same eight complete incoming states: $N\in\{4,14\}$
heads, updates 8,192 and 32,768, and two paired body-initialization
identities. Here $N$ counts heads; it is neither the number of
branch replicates nor a geometric length. The outer family uses one shared metric-network initialization,
one realized corpus and one 32-document evaluation panel. Within
each branch, both generator and body parameters continue to learn.

All 72 available branches per state supply development data for this
distinct prediction object. They comprise the forty design paths and
32 further paths of the endpoint study; their outcomes in that study
remain intact. Before any new branch, the protocol fixes both moment
estimators, both observation vectors, every time-block map, the score,
and all reported states. Sixty-four new branches per state then give
512 scientific paths and 32,768 updates. Each path resets the weights,
both Adam moments, parameter-group settings and schedule, draws
2,048 distinct blocks from the remaining corpus, and executes 64
batch-32 updates. Counterfactual branches may share blocks; a single
path never reuses an incoming consumed or newly supervised position.

The native computations use float32 with TF32 disabled and 64-token
contexts with one external next-token target. At horizons
$0,1,4,16,64$, the observations are external-target risk $R_t$ and
predictive entropy $H_t$, each averaged over the same fixed evaluation
panel and computed from the complete vocabulary distribution. The
primary target is the four-vector of consecutive risk increments.
The secondary target concatenates that vector with the four entropy
increments. These are dimensions four and eight, respectively; no
unobserved terminal value from a new branch enters a forecast.

\subsection{Fixed estimation and compatible blocking}
For either increment vector, let $\widehat\mu$ and $\widehat C$ be
the development mean and sample covariance with divisor 71. Let $D$
be diagonal with $D_{jj}=\max\{\sqrt{\widehat C_{jj}},10^{-6}\}$,
in nats, and set
\begin{align}
 \widehat\Sigma_{\rm reg}
 &=0.75\widehat C+0.25\diag(\widehat C)+0.001D^2,\nonumber\\
 \widehat\Sigma_{\rm diag}
 &=\diag(\widehat C)+0.001D^2.
 \label{model:eq:moment-estimators}
\end{align}
Both candidates use $\widehat\mu$. The floor ensures positive
definiteness; the fixed shrinkage limits finite-sample covariance
variation but also introduces bias. It is not asserted to be a
consistent estimator at fixed shrinkage.

The three temporal resolutions retain four increments, sum them in
two consecutive pairs, or sum all four to the endpoint change.
The same operation acts separately on risk and entropy. Each
candidate at each scale is exactly
$(B\widehat\mu,B\widehat\Sigma B^{\mathsf T})$.
The native intervals have unequal lengths, so these are finite
coarsenings of one path law, not a fitted stationary scaling orbit.
We do not refit or add a new covariance floor after blocking.
Proposition~\ref{model:prop:moment-blocking} therefore supplies a compatible
finite hierarchy, even though its estimated moments have error.

The primary comparison is the paired difference in
$S(\mu,\Sigma;x)$ from Proposition~\ref{model:prop:moment-score},
regularized minus diagonal. A negative difference favors the
regularized forecast for that specified target. This score assesses
means and covariances without a Gaussian training-law premise. Since
both candidates use the same mean, their comparison measures the
predictive contribution of cross-covariances in the presence of that
fixed mean error; it does not compare mean estimators.
Tables~\ref{model:tab:moment-risk} and \ref{model:tab:moment-entropy} retain all
state and scale outcomes. Parentheses give the paired sample
standard error across 64 branches, conditional on the incoming
state, realized development data and evaluation panel. They are
Monte Carlo precision summaries, without guaranteed population or
simultaneous coverage. The new branches add no pretrained-model
initialization and no new evaluation document.

\begin{table}[htbp]\centering\small
\caption{Complete external-target risk score differences, with paired
Monte Carlo standard errors in parentheses. The dimensions are four,
two and one. Negative values favor the regularized covariance.
Identities 3 and 4 denote seeds 640103 and 640104.}
\label{model:tab:moment-risk}
\begin{tabular}{@{}rrrccc@{}}
\toprule $N$ & Update & Identity & Four intervals & Two intervals & Endpoint\\\midrule
4 & 8,192 & 3 & $-0.211\ (0.083)$ & $-0.098\ (0.086)$ & $-0.156\ (0.061)$\\
4 & 32,768 & 3 & $-0.119\ (0.136)$ & $-0.072\ (0.101)$ & $-0.049\ (0.065)$\\
4 & 8,192 & 4 & $-0.587\ (0.179)$ & $-0.207\ (0.072)$ & $-0.206\ (0.072)$\\
4 & 32,768 & 4 & $-0.387\ (0.136)$ & $-0.004\ (0.117)$ & $-0.012\ (0.107)$\\
14 & 8,192 & 3 & $-0.138\ (0.091)$ & $-0.130\ (0.044)$ & $-0.132\ (0.042)$\\
14 & 32,768 & 3 & $-0.060\ (0.088)$ & $-0.007\ (0.077)$ & $+0.002\ (0.054)$\\
14 & 8,192 & 4 & $-0.212\ (0.089)$ & $+0.045\ (0.075)$ & $+0.058\ (0.065)$\\
14 & 32,768 & 4 & $-0.311\ (0.090)$ & $-0.207\ (0.048)$ & $-0.115\ (0.038)$\\
\bottomrule
\end{tabular}

\end{table}

\begin{table}[htbp]\centering\small
\caption{Complete joint external-target risk and predictive-entropy
score differences under the same fixed protocol. The dimensions are
eight, four and two; parentheses have the same conditional meaning
as in Table~\ref{model:tab:moment-risk}.}
\label{model:tab:moment-entropy}
\begin{tabular}{@{}rrrccc@{}}
\toprule $N$ & Update & Identity & Four intervals & Two intervals & Endpoint\\\midrule
4 & 8,192 & 3 & $-1.035\ (0.154)$ & $-0.484\ (0.106)$ & $-0.604\ (0.074)$\\
4 & 32,768 & 3 & $-0.771\ (0.266)$ & $-0.346\ (0.132)$ & $-0.286\ (0.095)$\\
4 & 8,192 & 4 & $-1.647\ (0.283)$ & $-0.716\ (0.132)$ & $-0.550\ (0.107)$\\
4 & 32,768 & 4 & $-1.229\ (0.192)$ & $-0.460\ (0.117)$ & $-0.461\ (0.109)$\\
14 & 8,192 & 3 & $-1.512\ (0.385)$ & $-0.459\ (0.169)$ & $-0.404\ (0.078)$\\
14 & 32,768 & 3 & $-1.258\ (0.245)$ & $-0.721\ (0.160)$ & $-0.360\ (0.080)$\\
14 & 8,192 & 4 & $-1.682\ (0.286)$ & $-0.497\ (0.153)$ & $-0.265\ (0.098)$\\
14 & 32,768 & 4 & $-0.856\ (0.191)$ & $-0.579\ (0.126)$ & $-0.440\ (0.059)$\\
\bottomrule
\end{tabular}

\end{table}

For risk, the equally weighted score differences at four intervals, two intervals and the endpoint are -0.253, -0.085, -0.076. The regularized covariance has a lower score in 8/8, 7/8, 6/8 states, respectively. These are comparisons conditional on the eight recorded states, not eight independent pretrained models.

For joint risk and predictive entropy, the equally weighted score differences at four intervals, two intervals and the endpoint are -1.249, -0.533, -0.421. The regularized covariance has a lower score in 8/8, 8/8, 8/8 states, respectively. These are comparisons conditional on the eight recorded states, not eight independent pretrained models.

The risk comparison has one positive score difference at two intervals
and two at the endpoint. Several risk differences are also small
relative to their conditional Monte Carlo standard errors. Exact
compatibility of a forecast across resolutions therefore does not
imply a uniform empirical ranking or strict population dominance.

\begin{figure}[htbp]\centering
\includegraphics[width=.98\textwidth]{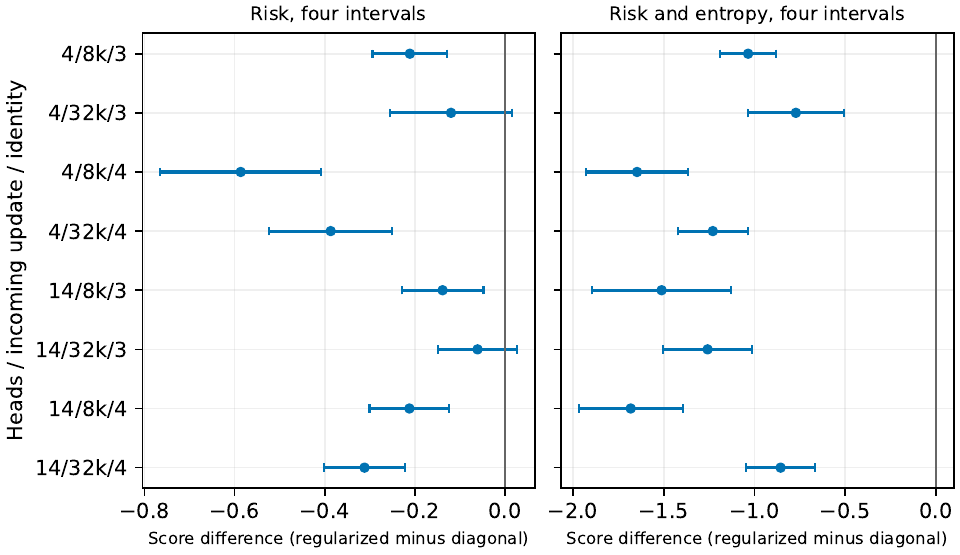}
\caption{Every four-interval score comparison. Bars show one paired
Monte Carlo standard error conditional on each fixed incoming state
and development fit. Risk is the external next-token observation;
entropy is that of the full-vocabulary predictive distribution.}
\label{model:fig:moment-scores}
\end{figure}

\subsection{Matrix discrepancies and the scope of the reduction}
The fresh sample covariance uses divisor 63. To give a separate
matrix diagnostic, at each scale set
$T_B^2=\diag(BD^2B^{\mathsf T})$ and report
\[
 L_F=\frac{\norm{T_B^{-1}(\widehat\Sigma_B-\widehat C_{{\rm fresh},B})
                     T_B^{-1}}_F}
              {\norm{T_B^{-1}\widehat C_{{\rm fresh},B}T_B^{-1}}_F}.
\]
The fine-resolution values appear in Table~\ref{model:tab:moment-matrix};
all coarser values and all branch scores are retained in the
numerical evidence. This diagnostic compares an estimated forecast
with another noisy estimate. It is not the relative operator-norm
bound against the population covariance required by
Proposition~\ref{model:prop:moment-precision}. Score and matrix-loss
rankings need not agree, because their estimands and weightings differ. In the fine-resolution comparison the regularized
covariance has a lower observed matrix discrepancy in all eight
states for each target. Its discrepancies range from $0.293$ to
$0.450$ for risk and $0.372$ to $0.566$ for joint risk/entropy;
these nonzero values retain the finite estimation cost.

\begin{table}[htbp]\centering\small
\caption{Every normalized Frobenius discrepancy for the fine temporal
covariance, with the declared design scaling. Smaller values mean
closer agreement with the fresh sample covariance, not certified
population accuracy.}
\label{model:tab:moment-matrix}
\begin{tabular}{@{}rrrcccc@{}}
\toprule & & & \multicolumn{2}{c}{Risk} & \multicolumn{2}{c}{Risk and entropy}\\
$N$ & Update & Identity & Regularized & Diagonal & Regularized & Diagonal\\\midrule
4 & 8,192 & 3 & 0.355 & 0.472 & 0.566 & 0.613\\
4 & 32,768 & 3 & 0.350 & 0.380 & 0.483 & 0.583\\
4 & 8,192 & 4 & 0.390 & 0.552 & 0.375 & 0.593\\
4 & 32,768 & 4 & 0.422 & 0.592 & 0.440 & 0.624\\
14 & 8,192 & 3 & 0.299 & 0.337 & 0.372 & 0.655\\
14 & 32,768 & 3 & 0.293 & 0.333 & 0.404 & 0.583\\
14 & 8,192 & 4 & 0.450 & 0.504 & 0.479 & 0.633\\
14 & 32,768 & 4 & 0.323 & 0.484 & 0.421 & 0.564\\
\bottomrule
\end{tabular}

\end{table}

The experiment makes the hierarchy of inference explicit. Common
incoming values give an exact direct endpoint-covariance identity.
A full-vector moment score assesses additional covariance directions.
Linear blocking carries the same estimated information to coarser
observations, while conditional score differences measure its
predictive value there. None of these operations identifies a
Markov kernel on risk and entropy, a stationary forcing spectrum,
or a portable law on unseen incoming states. They support an
effective finite path-moment description under the specified
single-pass law. Training-selected means and covariance sectors,
rather than a fitted thermodynamic exponent, are the retained variables.

\subsection{Execution and independent reconstruction}
One scientific worker owns each of the two RTX 4090 GPUs.
Peak allocated CUDA tensor memory is 2.54 GiB for four heads and
9.14 GiB for fourteen heads, below each GPU's 24-GiB capacity.
Two short reset qualifications execute 24 additional updates.
Two independent 64-update native replays add 128 verification
updates and reproduce the full model/Adam state and 5,120,000
logit coordinates per width bytewise. These executions are separate
from the 32,768 scientific updates. Independent finite-sum
reconstruction checks every sampled block, every moment fit and
all 48 state/target/scale cells. It evaluates the score through an
eigendecomposition, separately from the producer's Cholesky solve.
The eight scientific case runtimes range from 327.3 to 700.1 seconds. Independent reductions of all 40 retained full-vocabulary panels have maximum observed risk or entropy discrepancy $1.04\times10^{-14}$ nats. The maximum difference between independently recomputed scores and stored scores is $2.42\times10^{-12}$; the maximum covariance-blocking identity residual is $5.55\times10^{-17}$. These are observed arithmetic discrepancies, not certified rounding enclosures.

\subsection{Finite score contributions and conditioning}
Remark~\ref{model:rem:score-centering} separates the determinant, empirical
covariance and fixed-mean-error terms in every retained score difference.
The complete 48-cell reduction agrees with the direct scores within
$2.14\times10^{-14}$. The assessment covariance has divisor 63, so its
contribution uses the factor $63/64$. Table~\ref{model:tab:score-components}
gives equal-state averages without refitting any candidate. These signed
terms describe the observed score; they are not three nonnegative errors.

\begin{table}[htbp]\centering\small
\caption{Regularized-minus-diagonal score components on the retained
64-path assessment samples. Dimensions differ between targets and scales.}
\label{model:tab:score-components}
\begin{tabular}{@{}llrrrr@{}}
\toprule
Target & Intervals & Log determinant & Covariance & Mean error & Total\\
\midrule
Risk & Four & -0.211764 & -0.041890 & +0.000542 & -0.253111\\
Risk & Two & -0.491124 & +0.384691 & +0.021383 & -0.085050\\
Risk & Endpoint & -0.406883 & +0.316887 & +0.013740 & -0.076256\\
Risk/entropy & Four & -0.965532 & -0.284999 & +0.001881 & -1.248650\\
Risk/entropy & Two & -1.046695 & +0.494396 & +0.019612 & -0.532686\\
Risk/entropy & Endpoint & -1.065493 & +0.625904 & +0.018223 & -0.421366\\
\bottomrule
\end{tabular}

\end{table}

For the uniform empirical mixture over those eight states and their
64 assessment branches, endpoint-risk variance is $0.003348034$ within
states and $0.006051354$ between states, totaling $0.009399388$ squared
nats. The between-state share is $64.38\%$ for risk levels and
$7.46\%$ for risk increments. Incoming means therefore matter to the
mixture even when a conditional increment description is useful.
These states share one corpus and metric initialization; this is an
exact finite mixture calculation, not a new-corpus variance estimate.

\section{Finite prediction across corpus and complete initialization draws}
\label{model:sec:outer-results}

The finite moment law is also tested with newly drawn corpora, complete
initialization identities and evaluation documents. This experiment
changes several conditioning coordinates together. Its result describes
transfer to that declared domain; it does not isolate a corpus effect
from an initialization, training-age, adaptation-budget or
observation-panel effect.

\subsection{Executed single-pass design}
Two randomized corpus draws use the same sixteen RefinedWeb shard strata,
SentencePiece tokenizer and length eligibility of at least 513 tokens.
Each draw retains 4,096 exact-content-distinct documents per stratum,
65,536 in total, and a uniform 513-token crop per document. All 64 new
observation documents are reserved before either training corpus is
sampled. Thirty-two provide the fixed risk/entropy panel and thirty-two
provide operator donors. Each panel retains the first 65 tokens of its
uniform 513-token crop, with no added beginning/end tokens. The panels
are disjoint from both new corpora and
from the 524,288-document corpus used in Section~\ref{model:sec:onepass-results}.
The two training draws share 4,697 document hashes; independent draws
need not be disjoint. Their overlaps with the larger corpus are 37,381
and 37,567 documents. This is a specified finite, length-conditioned
text law, without a claim of semantic deduplication.

For each corpus, two independent complete initialization identities are
paired across $N=4,14$ heads. A new metric-network draw is shared only
within its width pair; separate body seeds also change the wide PLGA
parameters. Component hashes verify four distinct body, generator and
metric identities at each width. Both generator and body parameters
continue to learn. The five decoders, 64-dimensional heads and metric
width 170 are fixed. Shape-aware variance normalization follows
Proposition~\ref{model:prop:shape-initialization}. The constant generator rate
is $3\times10^{-4}$ and body rate $6\times10^{-4}/N$; AdamW uses
$(0.9,0.95)$ moments, offset $10^{-8}$, decay $0.01$ and norm clipping
at one. Execution is float32, without TF32 or microbatch substitution.

Each of eight primary paths consumes 65,536 distinct blocks in 2,048
batch-32 updates, using one external target per 64-token context.
The incoming consumed fraction is $1/8$. Fifty-six reset continuations
per state then use 2,048 distinct remaining blocks in 64 updates each.
The split is sixteen adaptation paths, eight calibration paths and
32 assessment paths. Every path restores the complete model, both Adam
moments, optimizer settings and phase. Counterfactual paths can overlap;
no path reuses its own supervised source position. There are 16,384
primary and 28,672 continuation updates, totaling 45,056 scientific
updates. These are eight training paths and 448 conditional branches,
including 256 assessment branches, not 456 independent pretrained models.
The outer design has two corpus draws and four nested initialization
identities; the evaluation documents remain fixed across states.

All observations use proper prefixes. The future risk and entropy
vectors are measured at $0,1,4,16,64$ updates, with raw intervals of
lengths $1,3,12,48$. The block population is 524,288; 458,752 indices
remain at the incoming state. Its exact without-replacement path law
is used throughout. No stationary kernel, homogeneous time block or
native critical exponent is inferred from these four intervals. The
exact with/without-replacement index-law distances for these horizons
are $0.001081$, $0.017563$, $0.248183$ and $0.989704$.
They bound a common observation-law distance from above and do not
measure that distance. In particular, this whole-window comparison
does not justify replacing the 64-update path by independent draws.

Table~\ref{model:tab:outer-primary} describes the initial and incoming
states on the common evaluation panel. For each decoder, the row ratio
is the average over documents and heads of
$\|P_\perp A\|_F^2/\|A\|_F^2$, with the native mean-squared denominator
floored at $10^{-30}$. The table gives the range over all five decoders;
Table~\ref{model:tab:outer-rows} retains every decoder value at both times.
These are descriptive measurements from the recorded primary paths,
not an additional hypothesis test or an exponent fit.

\begin{table}[htbp]\centering\small
\caption{Descriptive training changes on the fixed new text panel.
Risk is external-target cross-entropy in nats; incoming row ratios
range over the five decoders. All primary paths have 2,048 updates.}
\label{model:tab:outer-primary}
\begin{tabular}{@{}rrrcccc@{}}
\toprule
$N$ & Corpus & Init. & Initial risk & Incoming risk & Min. row ratio & Max. row ratio\\
\midrule
4 & 1 & 1 & 10.3783 & 7.6205 & 0.8514 & 0.9251\\
4 & 1 & 2 & 10.3683 & 7.4122 & 0.8138 & 0.9310\\
4 & 2 & 1 & 10.3827 & 7.4009 & 0.8111 & 0.9279\\
4 & 2 & 2 & 10.3775 & 7.3677 & 0.7621 & 0.9128\\
14 & 1 & 1 & 10.3855 & 7.4763 & 0.8836 & 0.9177\\
14 & 1 & 2 & 10.3727 & 7.4251 & 0.8918 & 0.9293\\
14 & 2 & 1 & 10.3720 & 7.3713 & 0.8875 & 0.9418\\
14 & 2 & 2 & 10.3807 & 7.2882 & 0.8841 & 0.9230\\
\bottomrule
\end{tabular}

\end{table}

Risk decreases from $10.368$--$10.386$ nats initially to
$7.288$--$7.620$ nats in all eight trained states. Their forty incoming
decoder row ratios range from $0.762$ to $0.942$. The tested incoming
regime therefore has substantial row contrast. Its conditional moment
and inference predictions do not presume proximity to a collapsed row
manifold.

\subsection{Frozen transfer and equal-budget adaptation}
Before any new training, the archived 72-path fits at each of four
incoming states per width determine an equal-state average increment
mean and an equal-state average correlation matrix $\bar R_N$.
Those states use incoming updates 8,192 and 32,768. Two zero-branch
mean predictions are retained: no change and this frozen mean increment.
Their training-age and data-resource mismatch is part of the declared
transfer test. Neither uses a continuation from the new state.

A separate comparison gives every covariance candidate the same
sixteen adaptation paths. Write $\widehat\mu_s,\widehat C_s$ for their
increment mean and divisor-15 covariance, and
$D_s=\diag(\max\{\sqrt{\widehat C_{s,jj}},10^{-6}\})$.
The local and diagonal estimates follow
\eqref{model:eq:moment-estimators}. The transferred candidate is
\begin{equation}
 \widehat C_{s,\mathrm{transfer}}
 =D_s\{0.75\bar R_N+0.25\diag(\bar R_N)+0.001I\}D_s.
 \label{model:eq:outer-transferred-covariance}
\end{equation}
The local acquisition cost is $16\cdot64=1,024$ native updates per
incoming state, for every covariance candidate. Eight calibration paths
cost another 512 updates; 32 assessment paths cost 2,048 updates and
are evaluation overhead. Archive and incoming-state construction are
separate costs. Thus this equal-budget transfer comparison establishes
no acquisition-cost advantage over local covariance estimation.
the acquisition account in claim~\hyperref[model:claim:S1]{S1} joins conditioning and cost near the
central claims; Equation~\eqref{model:eq:forecast-acquisition-cost} states the
additional matched-quality requirement for amortization.

All three use $\widehat\mu_s$. Each is regularized once at the finest
resolution and pushed forward by the same block maps. This comparison
tests transferred correlation shape with local mean and scale adaptation;
it is not zero-branch covariance portability. The eight calibration
paths set the maximum whole-risk-path standardized error, with means
and scales frozen from adaptation. No assessment observation tunes any
coefficient, covariance, scale or calibration radius.

The frozen primary endpoint is the four-dimensional risk-increment score.
Joint risk/entropy and coarser observations are secondary. The two primary
comparisons are local minus diagonal and transfer minus diagonal; transfer
minus local is supplementary. Table~\ref{model:tab:outer-endpoint-summary} reports
all equal-state means and observed favorable counts before the statewise
results. Counts describe signs, without a simultaneous significance claim.

\begin{table}[htbp]\centering\small
\caption{Equal-state mean score differences and favorable-state counts.
The primary target remains four-interval risk. Dimensions and resolutions
are separate prediction targets.}
\label{model:tab:outer-endpoint-summary}
\begin{tabular}{@{}lllcc@{}}
\toprule
Role & Target & Intervals & Local $-$ diagonal & Transfer $-$ diagonal\\
\midrule
Primary & Risk & Four & $-0.117415$ (4/8) & $-0.145059$ (6/8)\\
Secondary & Risk & Two & $-0.035452$ (5/8) & $-0.070289$ (6/8)\\
Secondary & Risk & Endpoint & $-0.069830$ (6/8) & $-0.078662$ (6/8)\\
Secondary & Risk/entropy & Four & $-1.098523$ (8/8) & $-0.633682$ (8/8)\\
Secondary & Risk/entropy & Two & $-0.786000$ (8/8) & $-0.278946$ (8/8)\\
Secondary & Risk/entropy & Endpoint & $-0.454969$ (8/8) & $-0.439758$ (8/8)\\
\bottomrule
\end{tabular}

\end{table}

Tables~\ref{model:tab:outer-risk} and \ref{model:tab:outer-joint} give all finest-scale
scores. Table~\ref{model:tab:outer-coarse} retains every coarser state/target
comparison. The complete numerical record also retains the transferred
minus local comparison at those scales. A negative difference favors
the first candidate; parentheses give paired sample Monte Carlo standard
errors across the 32 assessment branches conditional on the incoming
state, adaptation and evaluation panel. They are not outer-population
or simultaneous confidence intervals.

\begin{table}[htbp]\centering\small
\caption{Primary: four-interval risk scores at every new incoming state. Corpus
and initialization labels enumerate the two nested draws; $N$ is head count.}
\label{model:tab:outer-risk}
\begin{tabular}{@{}rrrccc@{}}
\toprule
$N$ & Corpus & Init. & Local $-$ diagonal & Transfer $-$ diagonal & Transfer $-$ local\\
\midrule
4 & 1 & 1 & $+0.512$ (0.223) & $-0.249$ (0.176) & $-0.761$ (0.308)\\
4 & 1 & 2 & $-0.763$ (0.191) & $-0.632$ (0.173) & $+0.131$ (0.155)\\
4 & 2 & 1 & $-0.446$ (0.262) & $-0.243$ (0.145) & $+0.204$ (0.248)\\
4 & 2 & 2 & $+0.146$ (0.354) & $+0.094$ (0.218) & $-0.052$ (0.285)\\
14 & 1 & 1 & $-0.098$ (0.142) & $-0.026$ (0.100) & $+0.072$ (0.164)\\
14 & 1 & 2 & $+0.124$ (0.119) & $+0.054$ (0.111) & $-0.070$ (0.065)\\
14 & 2 & 1 & $+0.254$ (0.249) & $-0.124$ (0.099) & $-0.379$ (0.240)\\
14 & 2 & 2 & $-0.669$ (0.260) & $-0.034$ (0.135) & $+0.635$ (0.277)\\
\bottomrule
\end{tabular}

\end{table}

\begin{table}[htbp]\centering\small
\caption{Secondary: the complete comparison for the eight-dimensional joint
risk/entropy increment vector. Different target dimensions are scored
separately.}
\label{model:tab:outer-joint}
\input{content/model/generated/outer-fine-risk_entropy.tex}

\end{table}

For the primary four-interval risk target, local shrinkage has a lower observed fine-scale score than the diagonal control in 4 of eight states, and transferred correlation in 6 of eight. Their equal-state mean score differences are $-0.117415$ and $-0.145059$, respectively.
For the secondary joint risk/entropy target, local shrinkage has a lower observed fine-scale score than the diagonal control in 8 of eight states, and transferred correlation in 8 of eight. Their equal-state mean score differences are $-1.098523$ and $-0.633682$, respectively.
The realized calibrated whole-risk-path tubes contain 237 of the 256 assessment paths. This fraction describes the realized calibration sets; the rank guarantee averages over calibration.

For the joint vector, both covariance candidates also improve the
observed score at all eight states after either coarsening. Local
minus diagonal equal-state differences are $-0.786$ at two intervals
and $-0.455$ at the endpoint; transferred minus diagonal differences
are $-0.279$ and $-0.440$. Local shrinkage has the lower equal-state
joint score, but its per-state ordering against transfer also varies.
Risk-only comparisons are less uniform: local shrinkage is favored
in five states at two intervals and six at the endpoint, while transfer
is favored in six at each scale. These outcomes support finite joint
moment prediction with explicit state adaptation. They do not support
one covariance candidate dominating at every target and state.
The common means, finite adaptation budget and signed score decomposition
in Remark~\ref{model:rem:score-centering} are part of this interpretation.

\begin{figure}[htbp]\centering
\includegraphics[width=.98\textwidth]{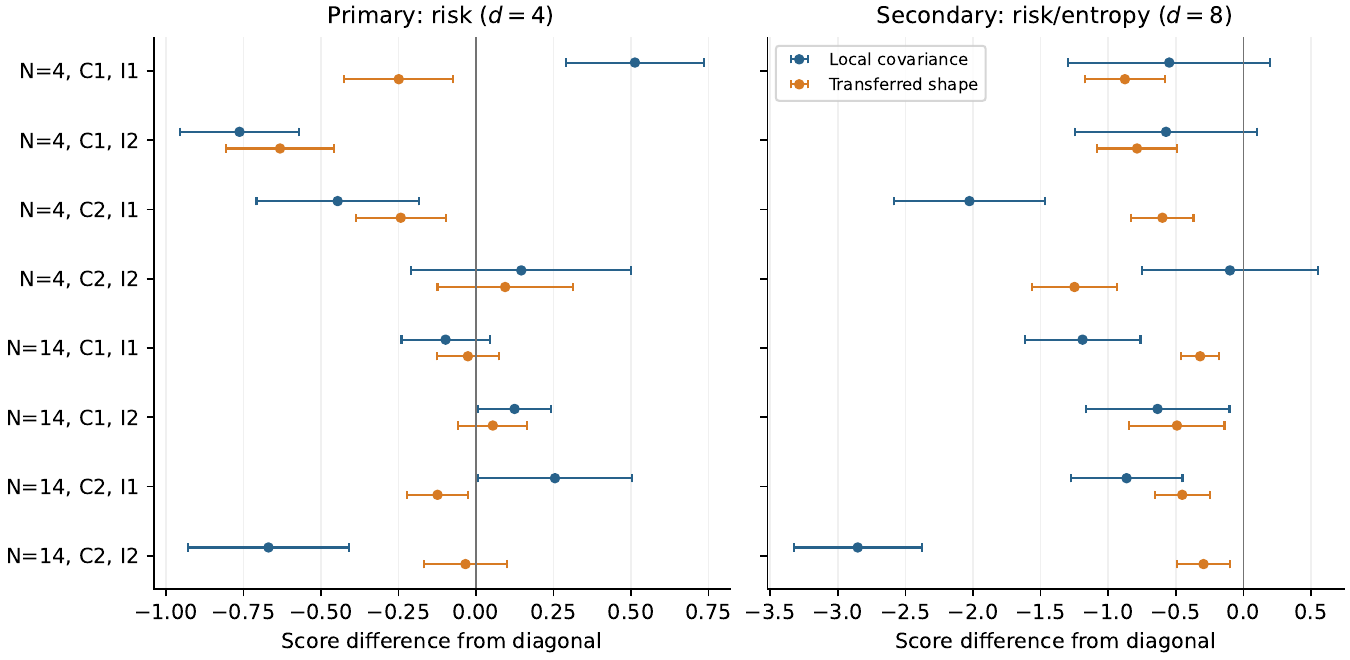}
\caption{Primary risk (left) and secondary joint risk/entropy (right)
comparisons with the diagonal control. Bars
show one conditional paired Monte Carlo standard error. The transferred
correlation is frozen from the archived states; local moments and scales
use the same sixteen adaptation paths for every candidate.}
\label{model:fig:outer-scores}
\end{figure}

\subsection{Mean accuracy, calibration and inference}
Table~\ref{model:tab:outer-means} reports the maximum absolute error of each
mean prediction over all four future horizons, against the assessment
sample mean, together with realized whole-path coverage. The $0.01$-nat
operational target applies to mean accuracy. It is not a required tube
width or a critical fluctuation scale. Calibration coverage can therefore
coexist with a measurable mean error or a wide prediction tube. Linear
images of every covered assessment path satisfy the corresponding
coordinate enclosures of Proposition~\ref{model:prop:calibrated-scale-images};
all coarse coverage counts and widths are retained in the result object.
The enclosure check allows $10^{-12}$ nats for floating-point comparison;
it is not a certified rounding enclosure.

\begin{table}[htbp]\centering\small
\caption{Maximum mean-risk errors in nats, whole-path coverage among
32 assessment paths, and the endpoint half-width of the realized tube.
The first two predictors use zero adaptation branches; ``Adapted'' uses
sixteen. Coverage uses eight further calibration paths.}
\label{model:tab:outer-means}
\input{content/model/generated/outer-means-coverage.tex}

\end{table}

All eight zero-branch no-change and archived-mean predictions exceed
$0.01$ nats in maximum mean-path error. Sixteen-path adaptation meets
that target in five states, with maximum errors over the eight states
ranging from $0.00369$ to $0.02677$ nats. The realized endpoint tube
half-widths range from $0.120$ to $0.275$ nats, and whole-path coverage
ranges from 26 to 32 out of 32 within a state. The calibrated result
is consequently a finite path prediction with measured width; it is
not a uniform $0.01$-nat mean law. All mean forecasts and assessment paths are retained in this comparison.

The exact empirical hierarchy samples a corpus label, an initialization
label within that corpus, and an assessment branch, each uniformly.
Corollary~\ref{model:cor:nested-moment-sectors} splits its endpoint-risk
covariance into within-state, between-initialization and between-corpus terms (Table~\ref{model:tab:outer-hierarchy}).
Population divisors specify this finite empirical law. Two corpora do
not provide a precise population variance estimator, and no branch is
counted as an independent corpus. Within-state continuation variance accounts for $22.8\%$ of the
four-head total and $43.3\%$ of the fourteen-head total in this
empirical law. The remaining terms retain variation that fixed-state
branching conditions away; their sum is not an isolated causal effect
of the corpus.

\begin{table}[htbp]\centering\small
\caption{Endpoint-risk variance of the declared empirical hierarchy,
in squared nats. Each row is an exact finite covariance decomposition,
not a thermodynamic susceptibility fit.}
\label{model:tab:outer-hierarchy}
\begin{tabular}{@{}rrrrr@{}}
\toprule
$N$ & Within state & Initialization & Corpus & Total\\
\midrule
4 & 0.003746 & 0.004633 & 0.008036 & 0.016415\\
14 & 0.003577 & 0.000036 & 0.004654 & 0.008268\\
\bottomrule
\end{tabular}

\end{table}

At each incoming state, full-vocabulary inference is evaluated at proper
prefix lengths $1,4,16,64$. The two interventions either fix every
learned operator to its average on the separate donor panel or replace
each emitted metric row by its centroid. Every declared length and state
is retained in Table~\ref{model:tab:outer-inference}, including risk changes
and forward KL. These are predictive fidelity measurements on 32 fixed
documents. They neither measure task competence nor identify which
critical properties survive an unmeasured thermodynamic limit.

At prefix one both interventions preserve the logits exactly: causal
attention has a single available key. Across all longer prefixes and
states, donor-operator replacement has mean forward KL at most
$1.454\times10^{-3}$ nats, and row projection at most
$8.736\times10^{-3}$ nats. Signed mean-risk changes range from
$-0.01009$ to $0.01600$ nats for the former and from $-0.01742$ to
$0.02744$ nats for the latter. Individual target-score changes are
larger, reaching $0.410$ and $0.818$ nats. Small panel-average KL and
substantial row contrast can thus coexist. The full predictive emission
and its observation domain are necessary to assess an operator reduction.

\subsection{Execution and verification}
One scientific worker owns each RTX 4090 GPU. The two 256-update
qualifications separately establish equality of native and adapter loss
and all 690 parameter gradients, complete reset and replay. Each
scientific case also replays its first continuation. Independently
implemented 64-update replays at both widths check an assessment
continuation against retained logits, losses and complete parameter/Adam
digests. Two complete 2,048-update primary paths are also replayed
from their initialization seeds, without loading trained weights.
All losses, boundary logits and the complete final parameter/Adam
digests agree bitwise, including the fourteen-head path replayed
on the other GPU. Separate reexecution checks all sixteen primary
boundary emissions against 16,384,000 retained logits bitwise; an
independent NumPy reduction of the emitted matrices reconstructs all
eighty decoder row fields within $10^{-12}$. Qualifications and replays use 5,272 additional
verification updates and add no independent observations.
Scientific case runtimes range from 425.4 to 899.6 seconds, totaling 88.1 GPU-worker minutes. Peak allocated CUDA tensor memory is 2.54 GiB at four heads and 9.15 GiB at fourteen heads. These counts include per-case observations and the recorded replay, while the 45,056 scientific updates exclude qualification and replay.

All 131,136 retained document crops, comprising 67,244,096 token
coordinates, are reconstructed directly from the read-only Arrow sources.
Independent reduction checks 2,293,760,000 retained branch logit
coordinates, all 48 state/target/scale cells and all three pairwise
covariance comparisons, together with initialization separation,
source-position nonreuse, fitting splits and inference risk transport.

\section{Acquisition budget and prediction after a training-state change}
\label{model:sec:refresh-results}

The state-transport theory is tested on all eight complete incoming
states of Section~\ref{model:sec:outer-results}, without selecting favorable
states. Each is advanced by 64 native batch-32 updates on a newly fixed
without-replacement path. This gives eight successors at update 2,112.
At every parent and successor, two independent fit cohorts each contain
sixteen adaptation and eight calibration paths. A separate common
assessment cohort contains 32 paths. Every continuation has 64 updates
and observations at $0,1,4,16,64$. Training uses one external target per 64-token context. The full-vocabulary
risk/entropy panel, native program, constant rates and optimizer settings
of Section~\ref{model:sec:outer-results} are held fixed. Both body and generator parameters continue to learn.

There are $16\{2(16+8)+32\}=1,280$ continuation paths and eight
state-advance paths, totaling $81,920+512=82,432$ scientific updates.
The sixteen states arise from the same eight incoming lineages, two
corpus draws and four nested complete initialization identities.
They are not sixteen independent model trainings. The two fit cohorts
share their state's assessment paths. All 512 unique assessment paths
are new; no inspected assessment path is fitted or reassessed as fresh.

The parent has 458,752 remaining block indices. Its successor has
456,704 and consumed fraction $2,112/16,384=0.12890625$.
Every path excludes its complete consumed prefix; every successor
continuation also excludes the 2,048 blocks of its state-advance path.
Counterfactual paths may overlap one another. The same fixed evaluation
documents remain excluded from training. Replays restore the complete
parameters, Adam moments, counters and schedule state.

\subsection{Frozen candidates, estimands and costs}
Within each fit cohort, nested budgets $m=4,8,16$ use the first $m$
adaptation paths. Local covariance, diagonal covariance and archived
correlation transfer use the same fitted mean and the same fine-scale
shrinkage and floors as Section~\ref{model:sec:outer-results}. Regularization
is applied once; all coarser means and covariances are its linear images.
The zero-adaptation controls retain the archive's mean and physical
covariance, or use zero mean increment with that same covariance.
Their archive covariance is regularized by
$0.75C_A+0.25\diag(C_A)+0.001\diag(d_A^2)$, with
$d_{A,j}=\max(\sqrt{C_{A,jj}},10^{-6})$.

At successors, every parent mean, covariance and calibration tube is
also reused without refitting and compared with its independently
reacquired counterpart. These predictions concern increments from the
newly observed incoming risk. Reanchoring a parent increment forecast
at the successor's risk is explicit; it is not a fitted autonomous
successor kernel. The parent fit and its calibration are independent
of the state-advance path conditional on the parent complete state.
Locally renewed calibration has the rank guarantee of
Proposition~\ref{model:prop:state-tube-reuse}; parent reuse needs an additional
law-transfer condition.

The primary endpoint remains the four-risk-increment moment score.
Local minus diagonal and transfer minus diagonal assess covariance
candidates at each budget. Within each candidate, budget minus
sixteen-path score measures acquisition sensitivity, and parent reuse
minus successor refit measures state transfer. Joint risk/entropy and
coarser projections are secondary. All state, fit, candidate, budget
and scale combinations are retained. Mean-path accuracy and whole-path
coverage are separate endpoints.

The frozen engineering gate requires a smaller budget to have maximum
empirical mean-path error at most $0.01$ nats and primary score excess
over its sixteen-path counterpart at most $0.05$, with lower measured
acquisition plus score-query cost, at every state and fit cohort.
The uncertainty version applies those thresholds to the upper endpoints
of 95\% percentile intervals from 2,000 whole-path bootstrap resamples.
The common assessment indices preserve pairing in score differences.
These conditional finite-sample diagnostics are neither simultaneous
confidence statements nor certificates of population error. Assessment
Monte Carlo standard errors also condition on the realized fit and panel;
the second fit cohort exposes some fitting variability without estimating
its population tails.

\begin{table}[htbp]\centering\small
\caption{Primary four-risk-increment scores: equal-state/fit means and
observed favorable counts. Each row has eight states and two fits per
state. Counts are descriptive signs among sixteen dependent fit cells.}
\label{model:tab:refresh-primary}
\begin{tabular}{@{}lrcc@{}}
\toprule
State & Paths $m$ & Local $-$ diagonal & Transfer $-$ diagonal\\
\midrule
Parent & 4 & $+8.126$ (0/16) & $-0.113$ (8/16)\\
Parent & 8 & $+1.466$ (3/16) & $-0.100$ (11/16)\\
Parent & 16 & $+0.019$ (12/16) & $-0.130$ (15/16)\\
Successor & 4 & $+6.308$ (0/16) & $-0.300$ (11/16)\\
Successor & 8 & $+0.820$ (4/16) & $-0.197$ (12/16)\\
Successor & 16 & $-0.239$ (14/16) & $-0.194$ (13/16)\\
\bottomrule
\end{tabular}

\end{table}

\begin{figure}[htbp]\centering
\includegraphics[width=.98\textwidth]{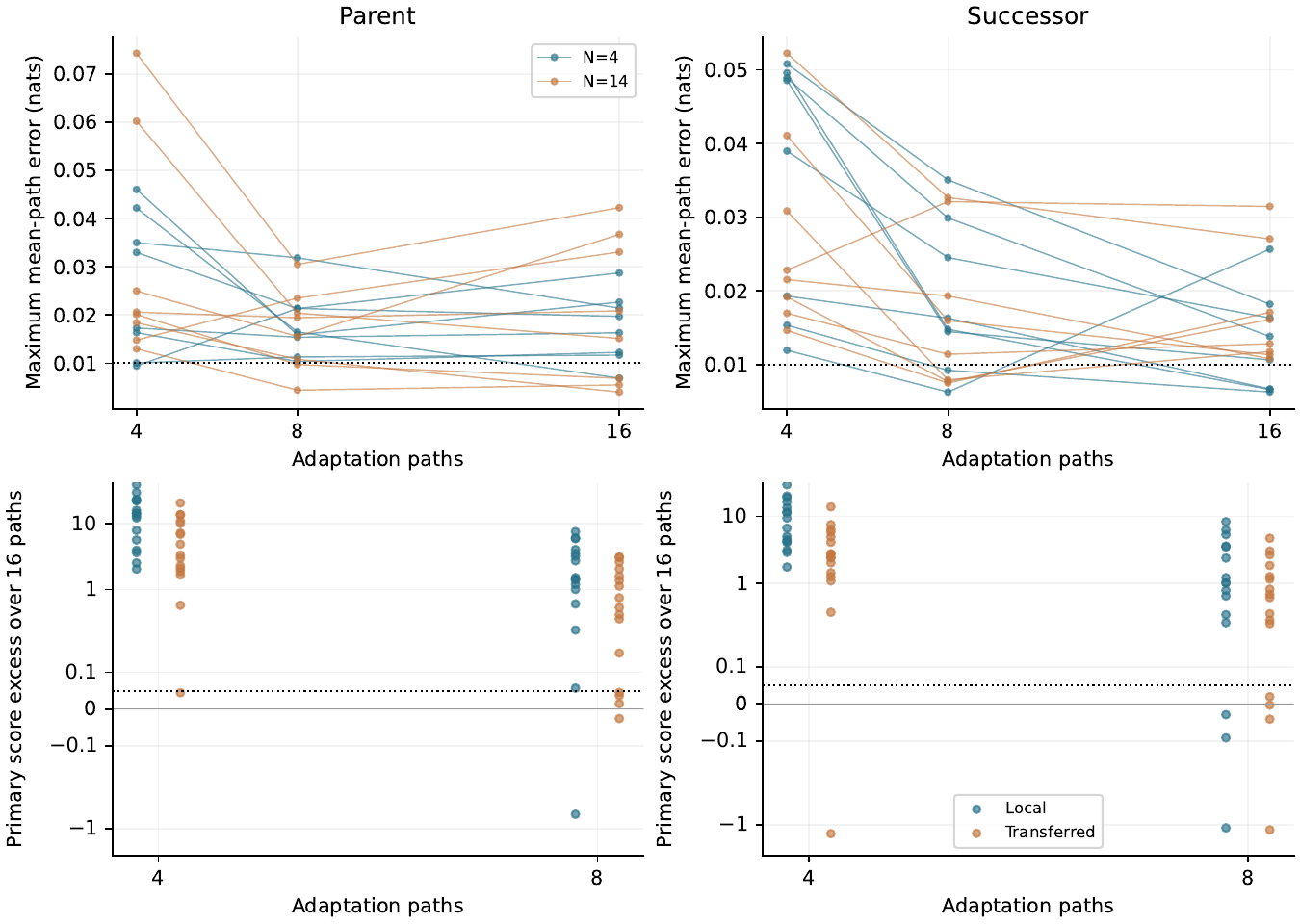}
\caption{Every state/fit outcome across adaptation budgets. Top: maximum
empirical mean-path error; connected points share a nested fit cohort,
and colors denote head count.
Bottom: primary score excess over the same candidate at sixteen paths,
with a symmetric logarithmic vertical scale outside $[-0.2,0.2]$.
Dotted lines give the frozen engineering tolerances. Repeated points
and paired widths are not independent model replicates.}
\label{model:fig:refresh-budget}
\end{figure}

\begin{table}[htbp]\centering\small
\caption{Renewed mean fits, reused parent fits at the successor, and
whole-path calibration. Mean-target counts
are among sixteen state/fit cells. Each coverage column assesses 256
unique paths across eight states; the two columns reuse those paths
with independent fitting and calibration. Half-widths are endpoint
ranges in nats. No pooling as 512 independent paths is implied.}
\label{model:tab:refresh-calibration}
\input{content/model/generated/refresh-calibration-summary.tex}

\end{table}

\begin{table}[htbp]\centering\small
\caption{Primary scores for frozen parent reuse relative to matched-budget
successor reacquisition. Negative favors reuse. Each row averages all
sixteen successor/fit cells; signs do not establish population dominance.}
\label{model:tab:refresh-reuse}
\input{content/model/generated/refresh-reuse-summary.tex}

\end{table}

\subsection{Measured prediction and the state dependence of the retained law}
At sixteen adaptation paths, the primary local-minus-diagonal score is
$+0.019$ on average at parents despite twelve favorable fit cells out of
sixteen; at successors it is $-0.239$, with fourteen favorable cells.
Transferred correlation gives $-0.130$ and $-0.194$, favorable in fifteen
and thirteen cells, respectively. These counts and equal-cell means
answer different questions. The four-path local covariance is worse than
the diagonal control in every cell, with mean excesses $8.126$ and
$6.308$. Covariance estimation therefore retains a visible budget cost.
For the secondary fine joint risk/entropy target at sixteen paths,
transferred correlation is favorable in all sixteen cells at each age,
with mean differences $-0.725$ and $-1.236$. Local covariance gives
$-1.096$ and $-0.587$, favorable in thirteen and eleven cells.
The complete coarser outcomes remain in Appendix~\ref{model:app:refresh-details}.

The state change has a larger measured effect than these within-state
covariance comparisons. At sixteen paths, successor reacquisition lowers
the primary score relative to parent reuse in every one of the sixteen
successor/fit cells for each covariance candidate. Mean reuse excesses
are $16.949$ for local covariance, $15.645$ for diagonal covariance and
$16.021$ for transferred correlation. The respective conditional bootstrap
intervals exclude zero on the positive side in fourteen, fifteen and
fifteen cells; those intervals are not simultaneous statements or estimates
of uncertainty across new corpora. Lower-budget reuse remains mixed
(Table~\ref{model:tab:refresh-reuse}). Reanchoring at the measured successor risk
has already removed a trivial error in the initial level.

At sixteen paths, the two renewed successor calibrations cover $222/256$
and $230/256$ assessment paths. Reusing their parent counterparts gives
$92/256$ and $125/256$ on those same respective paths. Renewed endpoint
half-widths are $0.098$--$0.230$ nats. This comparison supports retaining
the incoming state in calibration and agrees with the need for an
additional transfer condition in Proposition~\ref{model:prop:state-tube-reuse}.
It does not turn rank coverage into a guarantee for each realized set.
Renewed successor mean errors range from $0.00630$ to $0.03146$ nats,
compared with $0.01348$--$0.18670$ for parent-mean reuse. Only three of
sixteen renewed successor means and four of sixteen renewed parent means
meet the separate $0.01$-nat target. Thus no tested budget establishes a
uniformly accurate mean law. The two cohorts and nested budget curves
exhibit the fitting variation distinguished by
Proposition~\ref{model:prop:acquisition-noise}; realized errors need not decrease
monotonically along a nested sample.

No four-path fit meets the joint empirical engineering gate. At eight
paths, each covariance candidate meets it in three of 32 state/fit cells.
Neither smaller budget passes the uncertainty gate in any cell, despite
lower acquisition cost throughout. No smaller-budget candidate therefore
qualifies on the declared domain (Table~\ref{model:tab:refresh-gates}). Both
zero-adaptation means also miss the mean target at every parent and
successor (Table~\ref{model:tab:refresh-zero-controls}). Favorable secondary
scores do not change these primary and mean-accuracy decisions.

The supported retained object is a family of conditional moment forecasts
and calibrated sets indexed by the incoming complete state and acquisition
information. Observation blocking has exact compatible maps within each
member. Moving between members requires a law-transport bound or fresh
estimation and calibration. The positive successor-refitting comparison
establishes that finite measured distinction; it does not supply an
autonomous transition for the moment pair. Parameter, optimizer and
resource changes occur together here, so their separate contributions
to the native prediction change are not identified.

\begin{table}[htbp]\centering\small
\caption{Smaller-budget engineering gates over all 32 state/fit cells.
The uncertainty gate uses conditional whole-path percentile intervals;
qualification requires every declared cell. A failed or unresolved
cell does not change the frozen tolerance or the primary endpoint.}
\label{model:tab:refresh-gates}
\begin{tabular}{@{}rlccc@{}}
\toprule
$m$ & Covariance & Empirical gates & Uncertainty gates & Uniformly qualified\\
\midrule
4 & Local & 0/32 & 0/32 & No\\
4 & Diagonal & 0/32 & 0/32 & No\\
4 & Transferred & 0/32 & 0/32 & No\\
8 & Local & 3/32 & 0/32 & No\\
8 & Diagonal & 3/32 & 0/32 & No\\
8 & Transferred & 3/32 & 0/32 & No\\
\bottomrule
\end{tabular}

\end{table}

\begin{table}[htbp]\centering\small
\caption{Measured acquisition ranges across widths, states and fit cohorts.
Adaptation seconds include restoration, native updates, vocabulary
observation, branch serialization and fitting. Calibration uses eight
additional paths. Assessment, incoming-state construction and archived-fit
construction are separate costs.}
\label{model:tab:refresh-costs}
\input{content/model/generated/refresh-costs.tex}

\end{table}

At each state a sixteen-path fit costs 1,024 native updates and its
calibration another 512. Four- and eight-path fits cost 256 and 512
updates, respectively. The 32-path assessment uses 2,048 updates and
serves evaluation. Nested budgets reuse measured paths; their costs
describe acquiring that prefix, rather than three separate acquisitions.
Both local and transferred candidates still share the same acquisition
at each budget. Fitting time covers the shared routine producing all three
covariance candidates. Timed score queries act on 32 already observed paths;
their native production remains assessment work. All forecasts also require
the incoming risk observation.
Equation~\eqref{model:eq:forecast-acquisition-cost} is therefore evaluated only
at matched target and prediction quality; this study does not compare
the total training cost of an autonomous reduced model with native training.

The scientific experiment used 177.1 GPU-worker minutes and 89.1 elapsed launcher minutes. Peak allocated CUDA tensor memory was 2.54 GiB at four heads and 9.15 GiB at fourteen heads. The 82,432 scientific updates exclude 1,024 bitwise replay updates. Full-vocabulary reconstruction checked 6,553,600,000 logit coordinates and 89,088 individual scores, with maximum score discrepancy \ensuremath{5.5\times10^{-10}}.

The short end-to-end qualification uses two parent/successor lineages,
four-update paths and separate seeds. Its 840 qualification updates
and sixteen replay updates contribute no scientific observations.
Complete sampling, state and source bindings precede the scientific
outcomes. An independent finite-sum reducer reconstructs every saved
vocabulary observation, fit, tube and paired score from the native records.
Appendix~\ref{model:app:refresh-details} retains all primary statewise comparisons
and every target/resolution summary; the compact numerical record retains
every individual score, bootstrap interval, mean error and coverage event.

\subsection{Source deletion under a matched sequence coupling}
\label{model:sec:refresh-source-coupling}
A separate CPU experiment uses the actual eight remaining-index sets and
state-advance deletions. For each case, 4,096 independently seeded pairs
couple a uniform 2,048-block parent continuation to a uniform successor
continuation, using Proposition~\ref{model:prop:resource-edit-coupling}.
The source sequences are saved in full. A direct reconstruction checks
every index, absence of repetition, exclusion of the consumed prefix,
exclusion of the deleted blocks at the successor, and preservation of
every surviving parent slot. Exact finite probability enumeration also
checks the target-uniformity construction on a small population.

Here $r=458,752$, $k=\ell=2,048$. The exact sequence-law total
variation is $0.9998973184$, although the derived mean number of replaced
slots is only $9.142857$, with variance $9.061426$. The observed case
means range from $9.0537$ to $9.1641$, with Monte Carlo standard errors
$0.0458$--$0.0485$ (Table~\ref{model:tab:resource-coupling}). Four of the
32,768 coupled pairs need no edit. The total-variation value is calculated
from the exact population counts, not estimated to that precision from
these rare no-edit events.

\begin{table}[htbp]\centering\small
\caption{Complete source-coupling outcomes, with 4,096 pairs per case.
Case identifiers match the native lineages; the random units here are
source-sequence pairs, not trained models or risk observations. Variance
uses the sample divisor. All cases have the same theoretical edit moments.}
\label{model:tab:resource-coupling}
\input{content/model/generated/resource-coupling.tex}

\end{table}

This result supports a source-level description through edit distance
when total variation of entire index sequences is nearly maximal.
Its effect on risk still depends on the uniform program-sensitivity and
state-drift bounds in~\eqref{model:eq:resource-edit-observation-bound}; neither
constant is measured by this experiment. The 32,768 pairs contain 65,536
source sequences, are generated in 18.3 CPU-worker seconds, and add zero native
training updates. They are separate from all fitted and assessed native
continuations above.

\section{Native finite-pulse transport on a consuming corpus}
\label{model:sec:directional-results}
\subsection{Single-pass design and observation law}
The paired experiment conditions on the eight complete incoming states
of Section~\ref{model:sec:outer-results}, at update 2,048. Each of four
fresh source sequences has 128 batch-32 updates. The zero arm and four
signed amplitudes in each of two parameter directions use exactly the
same sequence after complete-state restoration. Thus the study contains
$8\cdot4\cdot9=288$ native continuations and 36,864 updates.
The directions are normalized negative external-target gradients on the
32 document-excluded donor contexts, separately within the generator
and body. At the pulse, the relative displacement multiplies the
incoming family parameter norm, at $h=3\times10^{-5}$ and $h/2$.
Adam moments, counters, learning rates
and remaining blocks are fixed; both parameter families and moments
evolve during continuation.

The observation grid is
\[
 \mathcal T=\{0,1,2\}\cup\{4j:1\le j\le32\},\qquad |\mathcal T|=35.
\]
Observations at these times include full-vocabulary risk
and entropy on the 32 fixed evaluation documents and the logarithms
$\log(1+\mathrm{RMS})$ of the common and transverse metric components
at each of five layers. The twelve coordinates use unit scales and the exact path norm
\begin{equation}
 \|z\|_{\mathcal T}^2=\frac1{35\cdot12}
 \sum_{t\in\mathcal T}\sum_{a=1}^{12}z_{t,a}^2.
 \label{model:eq:directional-observation-norm}
\end{equation}
Every saved time and coordinate has equal weight. This is a discrete
observation measure, not quadrature weighted by elapsed update time.
The metric is the residual learner's output $A$, before the positive
activation. With $P=\mathbf1\mathbf1^{\mathsf T}/64$, its two RMS
values are the square roots of the context/head averages of
$\|PA\|_F^2/64^2$ and $\|(I-P)A\|_F^2/64^2$. Thus the common
and transverse components retain their distinct physical meanings.
All evaluation targets are external to their 64-token prefixes.
Every continuation avoids the 65,536 blocks consumed before its incoming
state and uses 4,096 distinct new source blocks. Counterfactual arms
share source indices intentionally; they are paired interventions,
not independent model samples.

The generator rate is $3\times10^{-4}$ and the body rate is
$6\times10^{-4}/N$. AdamW uses $(\beta_1,\beta_2)=(0.9,0.95)$,
$\epsilon=10^{-8}$, decay $0.01$ and global-norm clipping at one.
Arithmetic is float32 without TF32, with float64 observation reductions.
Fixed update age consequently does not match both nominal learning-rate
clocks across widths. The statistical units remain conditional source
sequences within the eight incoming states; heads, layers, pulse signs
and repeated observations are dependent measurements.

Native execution, including restoration, direction construction, dense
observation and complete zero-path replays, took 49.5 elapsed minutes on
two RTX 4090 GPUs. The maximum allocated and reserved memory per worker
was 9.56 and 11.64 GiB, respectively. Protocol preparation and independent
terminal reduction are separate from this native-execution time.

\subsection{Finite-amplitude response}
Write $X$, $E_h$ and $O_h$ as in Proposition~\ref{model:prop:finite-pulse},
with $F$ the full twelve-coordinate observation path. The fixed primary
amplitude diagnostic is
\begin{equation}
 d_h=\frac{\|O_h-2O_{h/2}\|_{\mathcal T}}{\|2O_{h/2}\|_{\mathcal T}},
 \qquad d_h\le0.10,\qquad \|O_{h/2}\|_{\mathcal T}>5\times10^{-7}.
 \label{model:eq:native-pulse-domain}
\end{equation}
The RMS convention divides the squared Euclidean norm by the number
of recorded time-coordinate entries. The denominator is positive in
all resolved cells. A half-amplitude norm at or below the floor is
unresolved and cannot pass, including when both odd responses vanish. The floor is a fixed numerical resolution criterion;
exact zero-arm replay separately checks reproducibility. An incoming-state
comparison or agreement at one time cannot certify the whole path.
The primary rule, every amplitude and all 64 state/family/source cells
are retained.

All 64 half-amplitude signals exceed the fixed resolution floor, but
none meets the whole-path halving condition. Generator discrepancies
range from $0.808$ to $1.481$, and body discrepancies from $0.805$ to
$1.332$. Their half-amplitude path RMS ranges are $0.0122$--$0.0354$
and $0.0182$--$0.0314$, respectively. Thus a uniform linear response
domain over 128 updates is not resolved at these amplitudes and incoming
states. This conclusion concerns the complete path norm and preserves
all state, direction and source outcomes; it is not a claim that no
shorter-time or real-arithmetic derivative can exist.

\begin{table}[htbp]\centering\small
\setlength{\tabcolsep}{4pt}
\caption{Complete state-level amplitude-domain outcomes. Each family has
four conditional source comparisons. Ranges include every source;
Appendix~\ref{model:app:directional-details} lists the complete 64-cell vector.}
\label{model:tab:directional-primary}
\input{content/model/generated/directional-primary.tex}
\end{table}

The ratio $\|E_{h/2}\|/\|O_{h/2}\|$ compares the symmetric displacement
with the antisymmetric response. Keeping both gives the exact finite
law; dropping the even term incurs the measured error $E_{h/2}$ in
either signed reconstruction. This error is distinct from amplitude
halving, which compares two odd sectors. Neither operation estimates
an unrestricted augmented-state Jacobian.

\begin{figure}[htbp]\centering
\includegraphics[width=.98\textwidth]{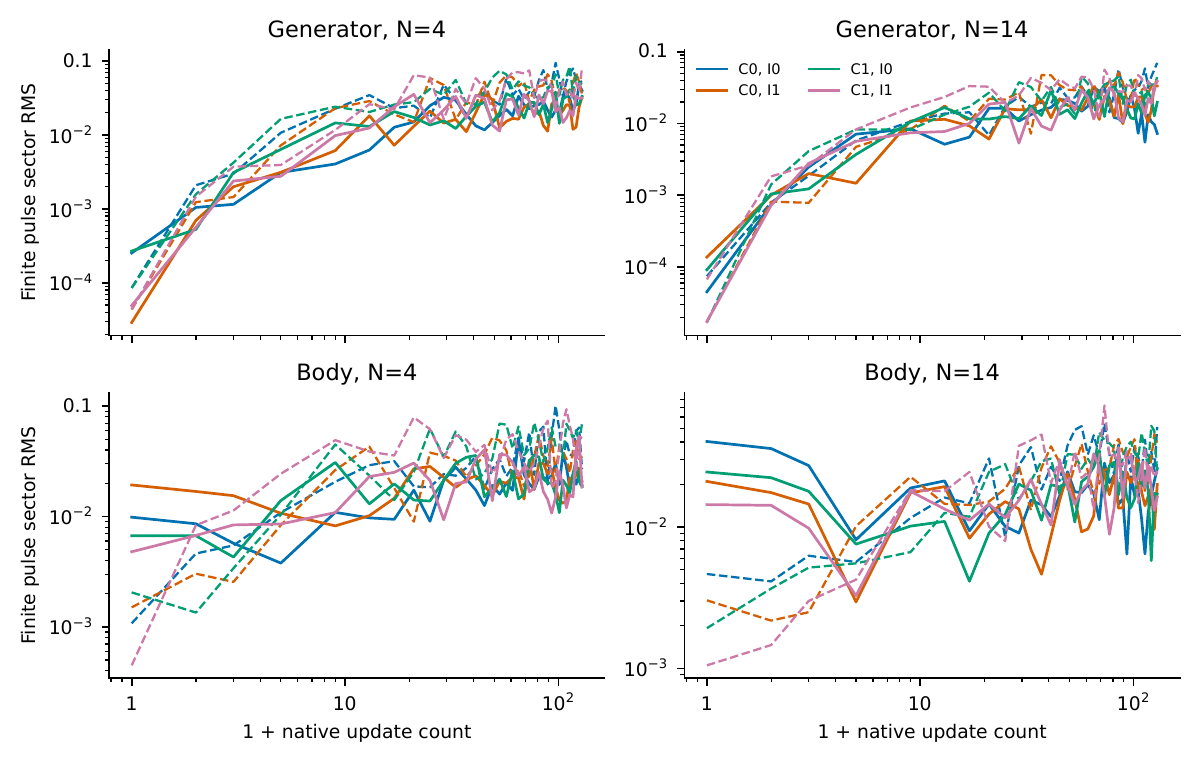}
\caption{Finite-pulse path components at relative amplitude
$1.5\times10^{-5}$. Solid curves show odd-sector RMS across the twelve
coordinates; dashed curves show even-sector RMS. Each curve is the
pointwise median of four source sequences, conditional on the indicated
corpus C and initialization I at width $N$. It is descriptive; taking
a pointwise median does not create an independent trajectory. The time
axis is $1+t$ to display the incoming pulse at $t=0$.}
\label{model:fig:directional-response}
\end{figure}

The measured even-to-odd path ratios are $1.006$--$3.105$ for the
generator direction and $0.906$--$2.699$ for the body direction
(Table~\ref{model:tab:directional-parity}). The even sector is therefore
comparable to, or larger than, the odd sector on every tested path.
Figure~\ref{model:fig:directional-response} shows their temporal magnitudes.
Both signs reconstruct from $X+E_h\pm O_h$; their complete endpoint
covariance traces agree with the six-term expansion to absolute error
at most $5.0\times10^{-16}$ over all 64 state/family/radius/sign cases.
This supplies a measured reason to retain the full finite-pulse law
instead of reducing a signed response to a single derivative coordinate.

\begin{table}[htbp]\centering\small
\caption{Finite-pulse sector ratios at relative amplitude
$h/2=1.5\times10^{-5}$. Each range contains all four source-path
values of $\|E_{h/2}\|/\|O_{h/2}\|$ for the indicated complete state.}
\label{model:tab:directional-parity}
\input{content/model/generated/directional-parity.tex}
\end{table}

\subsection{Signed temporal and pulse covariance}
At one fixed state and arm, let $r_j$ be proper-prefix risk at the
$j$th recorded time and $\delta_j=r_{j+1}-r_j$. The initial $r_0$
is identical across its four source sequences. With sample covariance
using divisor three,
\begin{equation}
 \widehat{\Var}(r_{\rm final})
 =\sum_{i,j}\widehat{\Cov}(\delta_i,\delta_j).
 \label{model:eq:native-chronological-sum}
\end{equation}
This is the deterministic-initial-state specialization of
Proposition~\ref{model:prop:master-chronological}, with identity response
and empirical forcing equal to the recorded increments. It is an exact
account of measured paths. Using those same increments does not create
a predictive forcing law. In particular, the diagonal sum alone discards
the cross-interval covariance.

At the endpoint, every signed twelve-coordinate pulse also supplies the
coupling $Y=X+e$. Lemma~\ref{model:lem:master-perturbation} retains both
ordered $X$--$e$ cross terms. Decomposing $e=E_h\pm O_h$ gives the
six trace contributions in Equation~\eqref{model:eq:pulse-covariance}.
The trace depends on the declared unit coordinate scales; only the
risk-only temporal comparison has units of nats squared. These are
conditional empirical covariances of four source sequences, without a
population confidence interval or an independent initialization claim.
The twelve-coordinate sample covariance has rank at most three; it does
not resolve a population forcing spectrum.

For the zero arm, retaining only interval-diagonal variance overestimates
endpoint risk variance by factors $1.181$--$14.560$ across the eight
states (Table~\ref{model:tab:directional-chronological}). Every zero-arm
cross-interval sum is negative. Across all 72 state/arm comparisons,
Equation~\eqref{model:eq:native-chronological-sum} reconstructs endpoint
variance to absolute error at most $1.1\times10^{-17}$ nats squared.
Across the 64 nonzero endpoint pulses, the covariance trace changes
have 35 negative and 29 positive signs, spanning $-0.15815$ to
$0.28702$ in the declared coordinate norm. The complete matrix
perturbation identity agrees to maximum entrywise error
$5.6\times10^{-17}$. A model retaining only the nonnegative error
covariance would miss the observed negative changes.

These reconstructions validate the finite-sample accounting of the
native records. They do not constitute a held-out forecast, a native
Jacobian estimate or a relaxation-gap measurement. The state-conditioned
moment predictions in Sections~\ref{model:sec:outer-results} and
\ref{model:sec:refresh-results} supply the separate predictive evidence.
Here, the full pulse and chronological law specifies the response domain
and retains the covariance sectors required to connect the complete
training state to its proper-prefix inference observations. Float32
amplification, activation or clipping boundaries, and smooth nonlinear
drift are not separated by this experiment.

\begin{table}[htbp]\centering\small
\setlength{\tabcolsep}{4pt}
\caption{Risk covariance under the zero pulse. The three covariance
columns are in $10^{-6}$ nats squared. The final column is the
interval-diagonal sum divided by endpoint variance. All nine arms,
including the complete signed cross sums, are retained in the compact data.}
\label{model:tab:directional-chronological}
\input{content/model/generated/directional-chronological.tex}
\end{table}

\section{Finite response, arithmetic and inference visibility}
\label{model:sec:numerical-response-results}

\subsection{A fixed small-radius trajectory assessment}
The response domain in Section~\ref{model:sec:numerical-response} was assessed
on new without-replacement source sequences. A separate incoming-state
radius diagnostic selected $h=3\times10^{-7}$ before these sequences
were evaluated. The two complete update-2,048 states have $N=4,14$,
corpus zero and complete initialization identity zero from
Section~\ref{model:sec:outer-results}. At each width, four independent
conditional source draws each provide eight batch-32 updates, excluding
the 65,536 previously consumed blocks. Zero and $\pm h,\pm h/2$
pulses in each of the generator and body directions give nine arms.
All arms and both numerical implementations at a width share the same
source indices and incoming complete state. There are 72 native
float32 paths and 72 matched arithmetic-control paths, with 576 updates
in each implementation. Four complete zero-path replays add 32 updates.
No trajectory repeats a supervised block within its single pass.

The donor-gradient directions, family parameter normalization and twelve
observation coordinates are those of Section~\ref{model:sec:directional-results}.
The pulse is applied by materializing its displacement tensor and then
adding that tensor in the parameter dtype. The retained times are
$\mathcal T_8=\{0,1,2,4,8\}$, with
$\|z\|_{\mathcal T_8}^2=(5\cdot12)^{-1}\sum_{t,a}z_{t,a}^2$.
The complete eight-update path is primary. The relative tolerance is
$0.10$ and the half-amplitude signal must exceed $5\times10^{-7}$.
Every state, direction and source remains in the assessment.
Shorter-prefix norms describe where the domain changes; selecting a
successful prefix does not replace the primary target.

The arithmetic control promotes the same stored float32 parameters,
Adam moments, fixed donor directions and cached rotary constants to
float64. It removes the forced-float32 softmax and rotary-input casts
in isolated in-memory modules, preserving the stored constants exactly.
It is a second implemented law, not a float64 pretraining history or
a certified real-arithmetic solution. Both implementations restore the
complete state before each arm. Every zero replay is bitwise exact,
including final parameters, both moments and schedule counters.

Table~\ref{model:tab:numerical-response-prefixes} shows the complete prefix
counts. All four incoming state/direction observations pass the
instantaneous comparison in each implementation; their repetition
under four source labels accounts for the count sixteen at time zero.
Those are not sixteen independent incoming-state measurements.
All full-path signals are resolved, but none of the sixteen native
comparisons or sixteen controls meets the complete-path tolerance.
The native discrepancy range is $0.553376$--$5.356736$ and the
control range is $0.375120$--$3.056367$.
Table~\ref{model:tab:numerical-response-all} retains every value.
Thus the experimentally supported finite law retains both pulse sectors;
instantaneous agreement does not supply a training-response derivative
on this eight-update domain.

\begin{table}[htbp]\centering\small
\caption{Small-radius domain over prefixes of the fixed saved-time grid.
Each column has sixteen state/direction/source cells, all with resolved
signals. The eight-update row is the primary assessment; time zero
repeats the same incoming observation across source labels.}
\label{model:tab:numerical-response-prefixes}
\begin{tabular}{rcc}
\toprule Largest saved time & Float32 passes / 16 & Float64 passes / 16\\
\midrule
0 & 16 & 16\\
1 & 1 & 1\\
2 & 0 & 0\\
4 & 0 & 0\\
8 & 0 & 0\\
\bottomrule
\end{tabular}

\end{table}

\begin{table}[htbp]\centering\small
\setlength{\tabcolsep}{4pt}
\caption{Every complete eight-update discrepancy at $h=3\times10^{-7}$.
The four source sequences are conditional replicates, paired across
signs and numerical implementations.}
\label{model:tab:numerical-response-all}
\input{content/model/generated/numerical-response-complete.tex}
\end{table}

\begin{figure}[htbp]\centering
\includegraphics[width=.98\textwidth]{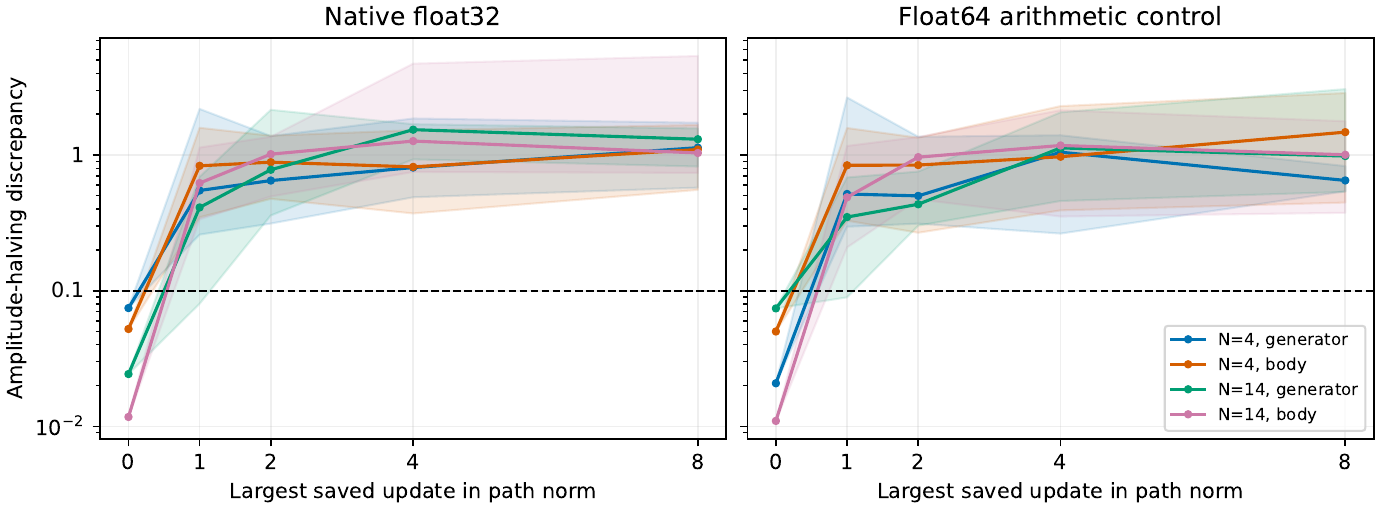}
\caption{Response domain at the selected small radius. Lines are
pointwise medians and shaded bands span all four source values in each
state/direction cell. The dashed line is the fixed $0.10$ discrepancy
criterion. Each ordinate uses all retained times up to the displayed
update. Pointwise medians are descriptive, not additional trajectories.}
\label{model:fig:numerical-response-domain}
\end{figure}

The even-to-odd full-amplitude path ratios span $0.680$--$8.580$ in
float32 and $0.186$--$5.960$ in the arithmetic control. Finite pulse
reconstruction holds to $4.45\times10^{-16}$ in the retained coordinates.
This is an algebraic check of measured sectors, not a forecast of an
unobserved response. The zero-arm float32/control difference grows
from $1.65\times10^{-7}$ and $2.78\times10^{-7}$ RMS at the two
incoming states to endpoint values ranging from $0.00156$ to $0.03694$.
Such differences do not bound either implementation's error relative
to a real-arithmetic map.

All 1,184 executed updates, including replays and arithmetic controls,
have pre-clipping gradient norm above one, with range
$19.53$--$246.74$. No on/off clipping change is observed.
The native power activation remains finite; its minimum before the
positive offset is zero. Between 52,224 and 182,784 stored second-moment
coordinates are zero in the inspected states. Neither those counts
nor active clipping identify the cause of response amplification:
constant zero-moment coordinates can be harmless, and the relevant
smoothness condition concerns the varying coupled trajectory.
The observations therefore support the explicit implementation and
approximation budgets of Proposition~\ref{model:prop:measured-pulse-error},
without estimating its constants or a native relaxation gap.

An independent reducer rereads all 148 raw archives and reconstructs
risk and entropy from 757,760,000 vocabulary coordinates. The maximum
reduction discrepancy is $1.60\times10^{-14}$. Common and transverse
energies are checked through their stored summaries; the full metric
tensors are not independently reconstructed. Narrow and wide workers
took 235 and 793 seconds, respectively, including both arithmetic
modes and their replays. Maximum allocated GPU memory was 20.13 GiB
for the wide float64 control. Input preparation and independent
reduction are separate costs. These two paired widths and four source
draws do not estimate corpus-population variability or a forcing spectrum.

\subsection{A predicted slow mode with zero observed inference response}
\label{model:sec:invisible-sector-results}
A separate finite intervention tests the source-support qualification
in Proposition~\ref{model:prop:invisible-decay}. The input embedding of token
zero has zero incoming Adam moments and is untied from the output
embedding. Before any perturbed trajectory is run, token zero is
verified absent from every selected training input and evaluation
prefix. An independent scan also verifies its absence from all
33,554,432 input positions of the 65,536-document materialized corpus
and all 2,048 evaluation-prefix positions. The support condition therefore
holds for every available source block in this finite corpus.
At each of the same two incoming widths, two new eight-update
source sequences are frozen. The pulse multiplies that input-embedding
row by $1.01$; every other parameter, moment and source index is held
fixed initially. Baseline and pulse arms give 64 native updates, and
a full baseline replay at each width adds sixteen updates.

The full native row forecast is computed before its trajectory, using
the stored row and the body-group multiplier
$1-\eta_{\rm body}\lambda_d$ at each update. All nine saved row
values, including the incoming row, agree bitwise with this forecast.
Every saved vocabulary logit, twelve-coordinate observation and training
loss is bitwise identical between the baseline and perturbed paths.
Digests of all other model coordinates and every optimizer coordinate
also agree at each saved update. The nonzero initial pulse norms and
exact recipe multipliers are given in Table~\ref{model:tab:invisible-sector}.
The independent reduction checks all saved arrays and row forecasts,
including 92,160,000 vocabulary coordinates; equality of other full
state tensors is a producer-digest check.

\begin{table}[htbp]\centering\small
\caption{Native source-support null sector. Each row summarizes both
fresh source sequences at that width; both have identical pulse norms
and zero observed logit difference. The multiplier is specified by the
learning recipe, not estimated from a relaxation fit.}
\label{model:tab:invisible-sector}
\begin{tabular}{rrrr}
\toprule $N$ & Decay multiplier & Initial pulse norm & Maximum logit difference\\
\midrule
4 & 0.999998500000 & 0.154268 & 0\\
14 & 0.999999571429 & 0.304081 & 0\\
\bottomrule
\end{tabular}

\end{table}

The result supplies a direct training-to-inference prediction: a
parameter coordinate evolves slowly while its projection onto the
specified predictive observations is exactly zero. It establishes
neither a thermodynamic critical mode nor invisibility for prompts
containing that token. This mechanism gives empirical content to the
nonvanishing-emission requirement in the critical transfer theory.
A parameter time scale imposed by learning rate and decay cannot
substitute for a connected, observable critical response.

\FloatBarrier
\par\medskip\noindent
Chapter~\ref{ch:adaptation-evidence} follows optimizer memory and
operator changes into native adaptation and frozen inference. Its
interventions examine when the state transport observed here becomes
predictively visible.

\chapter{Inference interventions and native adaptation}
\label{ch:adaptation-evidence}
This chapter presents inference interventions and native fine-tuning
experiments. It connects complete incoming optimizer states, coupled source
sectors, learned-operator replacement and task observations, retaining the
conditioning and finite-response limits of each comparison.

\section{Optimizer memory and emerging inference response}
\label{model:sec:optimizer-results}

\subsection{A finite prediction on complete incoming states}
The finite source calculation in Proposition~\ref{model:prop:finite-optimizer}
was tested at the update-2,048 states with $N=4,14$, corpus zero and
complete initialization identity zero from Section~\ref{model:sec:outer-results}.
The first intervention multiplies every stored first moment by $1+h$.
The second multiplies every positive second moment by $e^h$ and preserves
its zero coordinates. The directions are radial in the first-moment
field and constant in log second-moment coordinates. Equal values of
$h$ do not mean equal Euclidean displacement in those different fields.
Parameters, counters, schedule and remaining source resource are fixed
initially. These are admissible optimizer-state interventions; they are
not asserted to be naturally sampled alternative training histories.

The amplitude $h=0.1$ and its half amplitude were fixed before the
assessment. At each width, two fresh eight-update source sequences
sample batch-32 blocks uniformly without replacement from the remaining
458,752 blocks. Every path excludes the 65,536 consumed incoming blocks.
The zero arm and both signs at both amplitudes for each moment field
give nine arms per sequence. Native float32 and paired float64
implementations share the stored incoming state and source indices.
The arithmetic control has the conventions of
Section~\ref{model:sec:numerical-response-results}. The experiment contains
36 native and 36 arithmetic-control paths, with 288 updates in each
implementation. Four complete zero replays add 32 separate updates.

The twelve-coordinate observation and equal saved-time weighting use
$\mathcal T_8=\{0,1,2,4,8\}$. The first two coordinates are proper-prefix
external-target risk and entropy; the other ten are the common and
transverse metric RMS coordinates at every decoder. The full path is
the specified finite observation. All widths, sources, signs, amplitudes
and implementations remain in Table~\ref{model:tab:optimizer-transport-all}
and the compact outcome record.

\subsection{Predicted state transport and measured visibility}
Before each first optimizer step, the finite formula computes the
outgoing parameters and both moments in float64 from the stored pulsed
state and clipped gradient. No perturbed outgoing tensor is used to fit
that prediction. The first gradient is identical across all nine arms
at a given source and implementation, checked by its complete tensor
digest. Every incoming vocabulary logit is also identical across arms.
The numerical prediction is compared with every outgoing parameter and
moment coordinate. This tests the one-step optimizer calculation; it is
not a forecast without observing the first batch gradient.

The comparison retains cancellation-sensitive summands. For machine
epsilon $\epsilon_{\rm mach}$, put
\[
 A_i=\beta_1|m_i|+(1-\beta_1)|g_i|,\qquad
 B_i=\beta_2v_i+(1-\beta_2)g_i^2.
\]
The respective coordinate denominators for parameter, first-moment and
second-moment error are
\[
 \epsilon_{\rm mach}\left(|\theta_i|+
          \frac{\eta_i A_i}{a_1D_i}\right)+\tau_{\rm mach},\qquad
 \epsilon_{\rm mach}A_i+\tau_{\rm mach},\qquad
 \epsilon_{\rm mach}B_i+\tau_{\rm mach},
\]
where $\tau_{\rm mach}$ is the smallest positive normal value of the
native dtype. The fixed implementation criterion is at most 32 in each
normalized coordinate. It is an engineering tolerance, not a proved
roundoff enclosure or an estimate of a derivative remainder.

All 72 scientific first-step predictions and four replay checks meet the fixed criterion. The largest normalized parameter, first-moment and second-moment errors are 2.580, 1.250 and 1.993, respectively. These are measured implementation residuals of the finite formula. All four zero replays match bitwise in their saved arrays and complete terminal-state digests.

The incoming predictive response is exactly zero. At the first update, the native absolute antisymmetric risk response ranges from $2.857\times10^{-5}$ to $8.026\times10^{-4}$ nats across the eight width/source/moment cells. Thus the effect is visible in the predictive loss as well as the internal metric observations. All sixteen complete-path half-amplitude signals, including arithmetic controls, exceed the specified resolution floor. The whole-path even/odd RMS ratios range from 0.370 to 4.166; halving discrepancies range from 0.529 to 3.022. The measured response therefore retains both finite pulse sectors and its amplitude dependence. The one-step identity supplies no uniform tangent approximation over the later trajectory.

An independent reducer reconstructs 389,120,000 vocabulary coordinates and checks 1,678,080 saved first-step parameter coordinates, including their two moments. The largest risk/entropy/coordinate reconstruction error is $1.599\times10^{-14}$. The complete model prediction residuals and full-gradient identities are separately bound producer reductions or digests; full metric tensors are not independently reconstructed. The narrow and wide workers take 188.1 and 611.5 seconds, including arithmetic controls and replay; maximum allocated GPU memory is 18.47 GiB. Preparation and independent reduction are separate costs. The two conditional source draws at each paired width do not quantify variability over new corpora or initializations.

\begin{table}[htbp]\centering\small
\setlength{\tabcolsep}{4pt}
\caption{Every optimizer-response cell. $O_t$ is the full-amplitude
antisymmetric observation at update $t$, with RMS over twelve coordinates.
The even/odd ratio and halving discrepancy $d_h$ use the whole five-time
path. Each source label denotes a conditional draw shared across signs,
moment fields and implementations. Halving discrepancy is descriptive;
the primary test is the finite state prediction.}
\label{model:tab:optimizer-transport-all}
\input{content/model/generated/optimizer-transport-complete.tex}
\end{table}

\begin{figure}[htbp]\centering
\includegraphics[width=.98\textwidth]{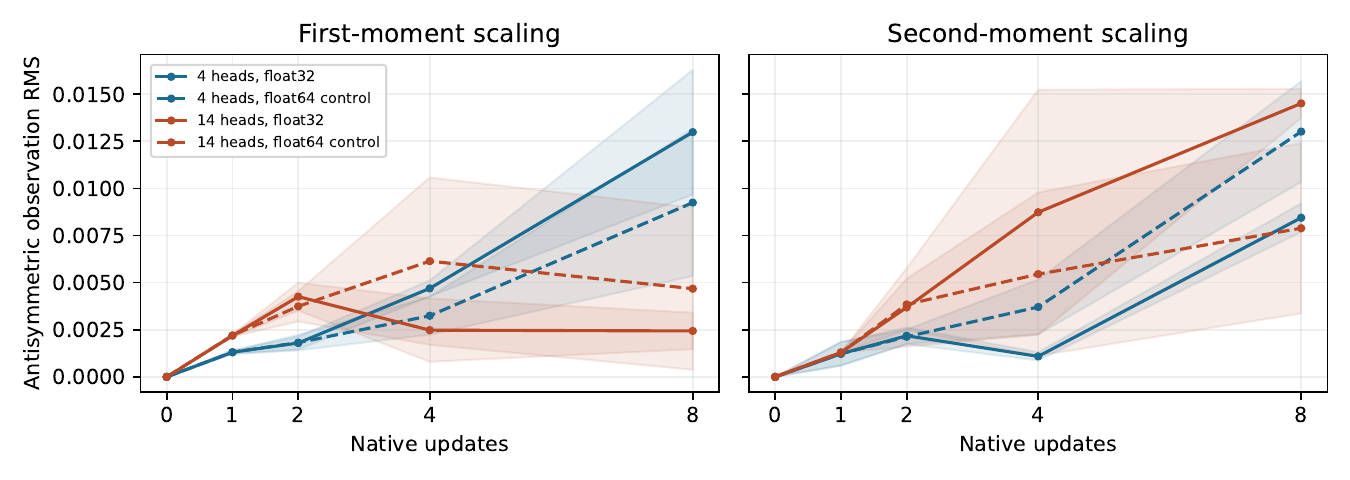}
\caption{Inference response emerges from moment-only interventions.
Lines average the two conditional source draws; bands span their full
range. Every incoming response is zero. The later response concerns
the fixed proper-prefix observation law and finite moment pulse. It
does not estimate a population spectrum or a thermodynamic exponent.}
\label{model:fig:optimizer-transport-visibility}
\end{figure}

The result supplies a concrete training-to-inference mechanism for the
augmented RG state. An incoming model alone has identical predictions
under these interventions, while its stored optimizer history affects
the successor and becomes visible in inference. The complete-state law
retains that history and composes over the eight updates. Its finite
pulse sectors commute with linear observation maps, but their algebraic
reconstruction does not predict a new incoming state or unmeasured pulse.
Together with the support-null experiment, this distinguishes observable
memory effects from parameter motion with zero predictive projection.

\section{Native fine-tuning and source-response sectors}
\label{model:sec:finetuning-results}

This section separates controlled source-response experiments from
adaptation of a qualified released model. The controlled incoming states
identify finite transition mechanisms under their stated short pretraining
law. Section~\ref{model:sec:released-adaptation} compares long-pretrained released
bases and reports their held-out adaptation scores. These training
histories and statistical units are kept distinct.

\subsection{A controlled adapted family}

The fine-tuning study conditions on twelve complete single-pass RefinedWeb
states: $N=4,8,14$ heads per layer and four body initializations at each
width. All models have five decoder layers, head dimension 64, and the
native shared metric learner with eight residual units and intermediate
width 170. Their parameter counts are 23,423,332, 52,032,274 and
109,689,362. They share the pretraining corpus, block permutation and
initial metric-generator draw. The four body draws define a conditional
initialization ensemble; the shared quantities are not additional
replicates. Learned metric generators can subsequently differ because
they interact with the bodies.

Each incoming path has consumed 40,960 batches of 32 distinct 64-target
blocks, or 83,886,080 supervised source positions. The incoming schedule
has completed its 2,000-update warm-up and cosine horizon of 32,768,
followed by 8,192 floor-rate updates. Both parameter groups continue at
the same applied rate $1.2\times10^{-4}$ at every width. AdamW has
$(\beta_1,\beta_2)=(0.9,0.95)$, $\epsilon=10^{-5}$, weight decay $0.1$
and value clipping at 1. The primary arms preserve both moments, their
counters and the scheduler phase. These are controlled incoming states,
distinct from the public model weights and their longer pretraining
histories. The bounded family does not establish a limit for those
longer histories.

Lexical selection uses fixed word lists applied to the first 128 stored
tokens of each document. A technical or narrative label requires at
least two words from that list and a score advantage of at least two
over the other list. Unassigned documents meet neither rule. These are
reproducible lexical strata, not human semantic annotations. Among
524,288 materialized documents, the counts are 19,443 technical,
271,512 narrative and 233,333 unassigned. Of 26,281 documents with no
previously consumed block, 160 from each stratum are reserved for
evaluation. Every block of each reserved document is excluded from
fine-tuning. Each stratum has 80 calibration and 80 held-out documents.

After these exclusions, the remaining pools contain 105,659 technical,
1,492,420 narrative and 2,879,744 general-source blocks. The general
pool includes all lexical labels. For the primary source-mixture arms,
the technical probability is
\[
 \rho\in\{0,1/8,1/4,1/2,3/4,7/8,1\}.
\]
Every block position makes its own Bernoulli selection; each reservoir
is consumed without replacement as in
Proposition~\ref{model:prop:finite-source-polynomial}. The same uniforms and
reservoir permutations couple widths, body draws and counterfactual
arms. They supply one realized source experiment, rather than independent
corpora. A further general-source continuation supplies the paired
reference. Each of the 96 primary paths has 512 updates and 1,048,576
supervised positions. Pure technical continuation consumes about $15.5\%$
of its eligible technical pool, so remaining-resource effects are explicit.

Nine optimizer controls reset both moments and their counters while
retaining the scheduler phase. Twelve component controls apply the rate
only to the generator or only to the body; the inactive group's moment
recursion is retained. These controls use one body initialization per
width and the endpoint sources, with an additional general-source reset.
The complete study contains 117 scientific paths and 59,904 updates.
Qualification and replay are counted separately.

\subsection{Observation law and finite predictions}
\label{model:sec:finetuning-observation-results}

Native training retains the all-nonpadding-target objective on 64-token
inputs and the native global Gram computation. Evaluation instead uses
prefixes of lengths 32 and 64 to predict the next token outside each
prefix. Thus reported proper-prefix risk is not identified with the
parallel training objective. The evaluation law assigns equal weight to
the three reserved lexical strata; it is not their frequency-weighted
mixture in the training pool.

At incoming and terminal times every model is evaluated on all 480
documents. Intermediate times $16,64,128,256$ use the same 96-document
subset. Time comparisons use that subset also at the endpoints. Full
vocabulary logits are saved for both prefix lengths. Common fields and centered row energies use native $A$; paired
discrepancies use $A$ and $G_{\mathrm{LM}}$ at every layer and head.
Raw tensors for six fixed
documents permit independent reconstruction of their reductions.

The predictive response matrix is the Gram matrix of narrative and
technical logit changes relative to the general arm in
\eqref{model:eq:finetuning-fisher}, with the incoming distribution supplying
the fixed weights. The common-field Gram uses the corresponding native
residual means, divided by the number of layer, head and feature
coordinates. Endpoint and midpoint observations determine the affine
and quadratic approximations in~\eqref{model:eq:source-quadratic}. Their errors
are measured at the four unused mixture values on held-out documents.
Three anchor-model forward evaluations on each evaluated document are
part of this prediction's acquisition cost. The four unused mixture
outputs are its prediction targets. The documents are excluded from all
training, while their anchor outputs remain available to the interpolant.
These predictions condition on the realized source permutations and
uniforms. They do not estimate the word-averaged source derivative in
Proposition~\ref{model:prop:finite-source-polynomial}.

\input{content/model/generated/finetuning-risk.tex}
\begin{table}[htbp]\centering\small
\caption{Ranges over four body initializations of the two-source response geometry on 240 untouched documents at 512 updates. The metric column uses the common residual field. These are finite effect directions.}
\label{model:tab:finetuning-sectors}
\begin{tabular}{@{}rrrr@{}}
\toprule
$N$ & Predictive $\lambda_2/\lambda_1$ & Metric $\lambda_2/\lambda_1$ & $\cos\angle(D_N,D_T)$\\\midrule
4 & 0.264--0.458 & 0.228--0.612 & 0.291--0.555\\
8 & 0.327--0.582 & 0.210--0.517 & 0.256--0.507\\
14 & 0.377--0.539 & 0.236--0.401 & 0.283--0.451\\
\bottomrule
\end{tabular}
\end{table}

\begin{table}[htbp]\centering\small
\caption{Fisher RMS prediction error relative to the source-response RMS, pooled over four body initializations and four untouched mixture values at 512 updates. Quadratic coefficients use only mixture values 0, 1/2 and 1. Each evaluated document supplies its three anchor logits; its four target-mixture outputs remain unused by the interpolation rule.}
\label{model:tab:finetuning-mixtures}
\begin{tabular}{@{}rrrr@{}}
\toprule
$N$ & Affine error & Quadratic error & Improved cells\\\midrule
4 & 0.7916 & 0.7926 & 10/16\\
8 & 0.8324 & 0.8188 & 9/16\\
14 & 0.8531 & 0.8403 & 13/16\\
\bottomrule
\end{tabular}
\end{table}

\begin{table}[htbp]\centering\small
\caption{Proper-prefix risk in the component and optimizer controls on the balanced held-out panel. These controls use one body initialization per width. Inactive parameter groups have zero applied rate while their moments continue to update.}
\label{model:tab:finetuning-controls}
\begin{tabular}{@{}rlrrrr@{}}
\toprule
$N$ & Source & Full & Reset memory & Generator only & Body only\\\midrule
4 & Narrative & 6.2205 & 6.2186 & 6.1522 & 6.2133\\
4 & Technical & 6.1510 & 6.1474 & 6.1536 & 6.1579\\
8 & Narrative & 5.9883 & 5.9785 & 5.9847 & 6.0162\\
8 & Technical & 5.9469 & 5.9424 & 5.9989 & 5.9692\\
14 & Narrative & 5.9847 & 6.0790 & 5.9653 & 6.0557\\
14 & Technical & 5.9574 & 5.9435 & 5.9857 & 5.9961\\
\bottomrule
\end{tabular}
\end{table}

\begin{table}[htbp]\centering\small
\caption{Mean paired-discrepancy order parameters for residual metric $A$ and energy-curvature tensor $G_{\mathrm{LM}}$, under the prefix-32/prefix-64 coupling on held-out documents. The row fraction is the centered residual-metric energy divided by its total energy. Numerators, denominators and individual cells are retained separately.}
\label{model:tab:finetuning-order}
\begin{tabular}{@{}rlrrr@{}}
\toprule
$N$ & Source & $\mathcal O_A$ & $\mathcal O_G$ & Row fraction\\\midrule
4 & Incoming & 570 & 1.1 & 0.404\\
4 & General & 207 & 1.36 & 0.408\\
4 & Narrative & 53.6 & 1.12 & 0.397\\
4 & Technical & 299 & 1.01 & 0.403\\
8 & Incoming & 13.7 & 10 & 0.43\\
8 & General & 14.4 & 10.7 & 0.435\\
8 & Narrative & 14.5 & 10.7 & 0.426\\
8 & Technical & 14.2 & 7.88 & 0.422\\
14 & Incoming & 13.9 & 0.826 & 0.453\\
14 & General & 15.3 & 0.801 & 0.459\\
14 & Narrative & 14.6 & 0.831 & 0.462\\
14 & Technical & 14.4 & 0.881 & 0.452\\
\bottomrule
\end{tabular}
\end{table}

\subsection{Conditional size behavior}

For common residual coordinate $\mu_{s\ell hj}(x)$, average heads first:
$\bar\mu_{s\ell j}=N^{-1}\sum_h\mu_{s\ell hj}$. The susceptibility
convention is
\begin{equation}
 \chi_C(N,\rho,t)=\frac{N}{5\cdot64}
   \sum_{\ell,j}\E_{x\sim\nu}
      \widehat{\Var}_{s=1,\ldots,4}\bar\mu_{s\ell j}(x),
 \label{model:eq:finetuning-susceptibility}
\end{equation}
where the sample variance uses divisor three at each fixed input.
Analogous diagnostics use head-averaged normalized attention entropy
and proper-prefix risk. The risk diagnostic has the same declared
factor $N$; that convention does not identify a spatial correlation
length. The layer covariance retains cross-layer terms before taking
its spectrum. Variation between document means is excluded from all
these initialization covariances. No fluctuation--dissipation relation is
assumed between this conditional initialization variance and the response
to the source probability $\rho$.

For a positive size statistic $\chi$, the adjacent secant
$k_{N_1,N_2}=\log(\chi(N_2)/\chi(N_1))/\log(N_2/N_1)$ is descriptive.
The complete reduction also predicts the middle size from the two
outer sizes under a pure power law, records its error, and retains the
time and source dependence of every secant. This separates an observed
finite size trend from the common control, time and fluctuation-scaling
hypotheses needed for native critical exponents.

\begin{table}[htbp]\centering\small
\caption{Conditional common-field susceptibility and adjacent logarithmic size secants at 512 fine-tuning updates, on the same held-out inputs across widths. Covariance is centered across four body initializations at each input before averaging. The secants are descriptive.}
\label{model:tab:finetuning-size}
\begin{tabular}{@{}lrrrrr@{}}
\toprule
Source & $\chi_C(4)$ & $\chi_C(8)$ & $\chi_C(14)$ & $k_{4,8}$ & $k_{8,14}$\\\midrule
General & 0.03565 & 0.03789 & 0.04952 & 0.088 & 0.478\\
Narrative & 0.03551 & 0.03763 & 0.04937 & 0.084 & 0.485\\
$
ho=1/4$ & 0.03554 & 0.03783 & 0.0489 & 0.090 & 0.459\\
$
ho=1/2$ & 0.03541 & 0.03785 & 0.05073 & 0.096 & 0.523\\
$
ho=3/4$ & 0.03507 & 0.03883 & 0.04922 & 0.147 & 0.423\\
Technical & 0.03548 & 0.03763 & 0.0501 & 0.085 & 0.512\\
\bottomrule
\end{tabular}
\end{table}

\subsection{Persistence after returning to a common source}
\label{model:sec:finetuning-return-results}

Each narrative, technical and general endpoint is continued for 256
updates on an identical fresh general-source permutation. The return
pool excludes every block used by any of those parents as well as all
evaluation documents and all pretraining blocks. Each complete optimizer
state and its incoming emission is reproduced before the return begins.
The resulting 36 scientific paths contain 9,216 additional updates.
All selected source-return outcomes are retained.

At every saved return time, the two source contrasts are measured
relative to the continuing general parent, using the original incoming
Fisher weights. A $2\times2$ transport matrix is fitted on calibration
documents and assessed on held-out documents through
\eqref{model:eq:source-span-residual}. Its residual is compared with the weaker
incoming response singular value. This tests a finite persistence
condition from Corollary~\ref{model:cor:finite-source-span}. The coefficients
are reacquired at each state, so the measured benefit is conditional
observation transport. Since a separate map is fitted for each body draw,
its finite certificate is a per-state statement. It does not establish
one deterministic observation map for the initialization ensemble in
\eqref{model:eq:finetuning-susceptibility}.

A nested description additionally retains the five interior-mixture
contrasts at the incoming adapted states. Together with the two endpoint
contrasts they form a seven-column dictionary. The same least-squares
normal equations fit its $7\times2$ map to the two return targets.
Equation~\eqref{model:eq:source-span-residual} holds for this rectangular map
with a $7\times7$ incoming Gram. A larger dictionary can only decrease
calibration error because it contains the smaller span; improvement on
the separate test law is an empirical question. The seven contrasts require eight incoming checkpoint forward evaluations
per document, including the shared general reference; two contrasts require
three. The original incoming checkpoint is also evaluated once per document
to fix Fisher weights. At every return time, the three successor checkpoints
are evaluated on calibration documents to acquire the transport coefficients. Terminal full-panel
tables use 240 calibration and 240 held-out documents. Every time-comparison
curve instead uses the same 48 calibration and 48 held-out documents,
including its terminal point. The dictionary
comparison was specified before collecting return outcomes.

\begin{table}[htbp]\centering\small
\caption{Two-source transport after 256 common general-source updates. Fit error is residual Fisher RMS divided by current target RMS. The transport matrix is calibrated separately on 240 calibration documents at that time; errors use the 240 untouched documents. The weak scale is the smaller incoming source singular value. A positive finite lower bound certifies retention of two effect directions on this panel.}
\label{model:tab:finetuning-return}
\begin{tabular}{@{}rrrrr@{}}
\toprule
$N$ & Fit error & Error/weak & Bound positive & Fit improves\\\midrule
4 & 0.842--0.974 & 0.812--1.460 & 0/4 & 4/4\\
8 & 0.891--0.978 & 0.984--1.494 & 0/4 & 4/4\\
14 & 0.946--0.974 & 1.221--1.444 & 0/4 & 4/4\\
\bottomrule
\end{tabular}
\end{table}

\begin{table}[htbp]\centering\small
\caption{Nested source dictionaries after 256 common-source updates. Both descriptions predict the same two endpoint effects; errors are relative to their current Fisher RMS. The seven-contrast dictionary adds all five interior-mixture contrasts observed before the return. Coefficients are fitted on calibration documents; these error ranges use the held-out half.}
\label{model:tab:finetuning-dictionary}
\begin{tabular}{@{}rrrr@{}}
\toprule
$N$ & Two-contrast error & Seven-contrast error & Improved cases\\\midrule
4 & 0.842--0.974 & 0.802--0.972 & 4/4\\
8 & 0.891--0.978 & 0.883--0.954 & 4/4\\
14 & 0.946--0.974 & 0.936--0.950 & 4/4\\
\bottomrule
\end{tabular}
\end{table}

The terminal dictionary comparison therefore uses $8\times480$ incoming
checkpoint evaluations for seven contrasts or $3\times480$ for two,
plus 480 original-checkpoint evaluations for the shared Fisher geometry.
Both fits acquire $3\times240$ successor checkpoint evaluations on
calibration documents. These counts exclude the separate held-out target
measurements used only for scoring and the native training of the anchors.
The small held-out improvement in Table~\ref{model:tab:finetuning-dictionary}
is an observation-transport result with this target-state acquisition.

\subsection{Complete fixed-dictionary projection diagnostic}
\label{model:sec:projection-results}

We evaluate Proposition~\ref{model:prop:projection-obstruction} on every terminal
return state, at update 256, with both prespecified dictionaries. The
geometry is fixed by the original incoming probabilities, the same 240
held-out documents and full 32,000-token vocabulary. The projection
coefficients use that assessment panel only to measure the attainable
finite-panel lower bound. They are not used as a forecast and supply no
population guarantee. The calibration-fitted coefficients retain their
separate, held-out prediction score.

\begin{table}[htbp]\centering\small
\caption{Fixed-Fisher projection of common-source return effects on all terminal held-out panels. Ranges span four body initializations at each width. The oracle uses the assessment panel only as a descriptive lower bound; it is not a forecast. The fraction is orthogonal residual energy divided by calibration-fit residual energy.}
\label{model:tab:projection-summary}
\begin{tabular}{@{}rrcc@{}}
\toprule Heads & Columns & Oracle relative RMS & Orthogonal energy fraction\\\midrule
4 & 2 & 0.8412--0.9739 & 99.72--99.97\%\\
8 & 2 & 0.8859--0.9745 & 98.87--99.92\%\\
14 & 2 & 0.9426--0.9735 & 99.21--99.86\%\\
4 & 7 & 0.7978--0.9653 & 98.44--99.77\%\\
8 & 7 & 0.8721--0.9481 & 97.63--99.66\%\\
14 & 7 & 0.9286--0.9464 & 98.40--99.27\%\\
\bottomrule
\end{tabular}
\end{table}

For all 24 dictionary/state cells, the normal-equation projection gives
the same residual decomposition as independent reconstruction from the
full vocabulary arrays. At least 97.63\% of the calibrated residual
energy is orthogonal to the respective fixed dictionary. Even the oracle
relative RMS residual ranges from 0.7978 to 0.9745. Thus the dominant
component on these panels is representation error. Adjusting coefficients
within these dictionaries cannot yield a small residual. The two-column
and seven-column comparisons retain all states and neither dictionary is
selected from these diagnostic outcomes.

The complete 24-cell vector is in Appendix~\ref{model:app:projection-details}.
The raw reconstruction also checks the existing source Grams, off-knot
mixture errors, calibration-only return fits and proper-prefix risks at
all twelve states. This is a new reduction of completed single-pass paths,
with zero additional scientific training updates. It supports the
representation lower bound for these finite emissions and does not rule
out a richer or state-dependent representation. In particular, failure
of a sufficient weak-scale inheritance certificate does not imply
absence of every native dynamical critical mode.

\begin{figure}[htbp]
 \centering\includegraphics[width=\textwidth]{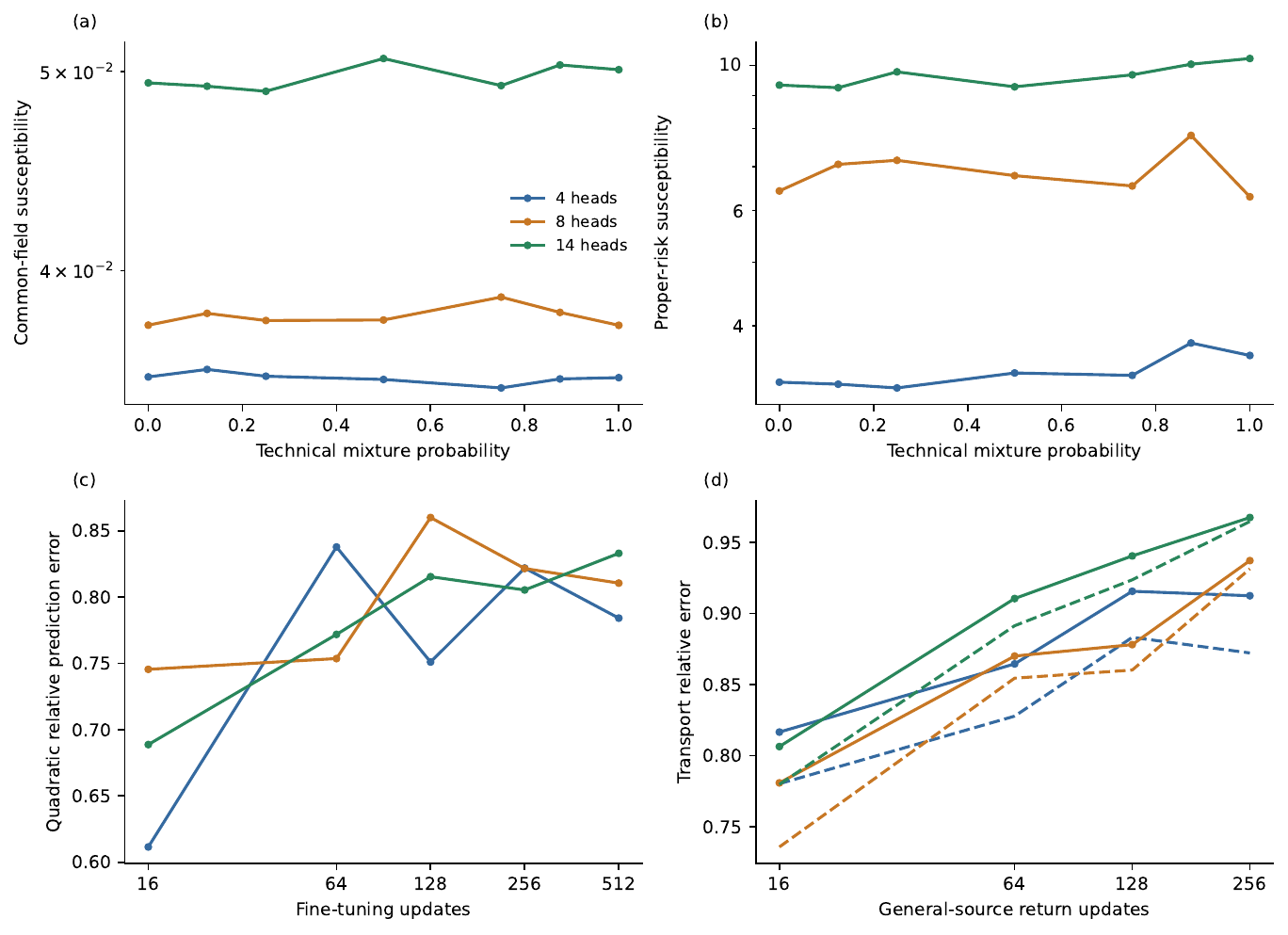}
 \caption{Source-conditioned fine-tuning and common-source return.
 The upper panels show the complete mixture grid at 512 updates on
 held-out documents. The lower panels use exactly the same 48 held-out
 documents at every time; return fits also retain the same 48 calibration
 documents. Mixture errors pool all four untouched mixture
 values and four body initializations. Return errors average the four
 initialization-specific, separately calibrated transports. Solid return
 curves use two source contrasts; dashed return curves use all seven.}
 \label{model:fig:finetuning-summary}
\end{figure}

\subsection{A released reference in the concentrated regime}

The controlled incoming family has a much shorter pretraining history than
released reference model PLDR-LLM-v51-SOC-110M-1. We therefore separately
fine-tune that released five-layer, fourteen-head model. Its reported
pretraining exposure is eight billion tokens on RefinedWeb documents
$[16{,}000{,}000,32{,}000{,}000)$~\cite{gokden2026soc}. The original
optimizer state is unavailable, so all four paths begin with zero AdamW
moments and counters and constant rate $1.2\times10^{-4}$. This is an
explicit new fine-tuning origin, not a recovered continuation of its
pretraining optimizer.

The four paths use the narrative, equal-mixture, technical and general
source streams for 512 updates each. They add 2,048 scientific updates.
Each path uses distinct blocks and the same excluded evaluation documents.
Cached shard lengths and document offsets place every materialized
document at global positions between 46,369,549 and 951,732,640, outside
the reported pretraining interval. This separation is conditional on the
cached document ordering and released source description; it does not
recover an unavailable original training manifest. The protocol binds the
weights, cached ordering and every selected position. A separate native
qualification and exact replay precede this assessment.

\input{content/model/generated/finetuning-reference.tex}

The final predictive source Gram has eigenvalues $0.7803$ and $0.2849$
on held-out documents, with ratio $0.3651$ and source cosine $0.4649$.
Thus two distinct finite source effects coexist with a row-fluctuation
fraction between $5.65\times10^{-8}$ and $2.25\times10^{-7}$ after
adaptation. Strong within-model row concentration does not remove
between-source predictive response.

The incoming proper-prefix risk is $3.6362$ nats and the four terminal
risks range from $3.7086$ to $3.8285$. Meanwhile $\mathcal O_G$ decreases
from $0.01599$ to $0.00320$--$0.00922$. The common-source and equal-mixture
arms also decrease $\mathcal O_A$. These measurements separate operator
invariance from the direction of predictive performance change. In the
narrative arm the raw $A$ discrepancy decreases from
$1.88\times10^{-5}$ to $1.80\times10^{-5}$, while its normalized order
increases from $0.00895$ to $0.01950$ because the reference mean magnitude
falls from $0.002104$ to $0.0009233$. Retaining both parts of the
normalization is therefore necessary even for this finite comparison.

This single released checkpoint is a reference for a different training
regime. It contributes neither a fourth width nor extra initialization
replicates to~\eqref{model:eq:finetuning-susceptibility}. Its proper-prefix
coupling also differs from the stochastic-generation coupling in the
self-criticality study. All source-arm, time and normalization components
are retained separately in the accompanying evidence.

\subsection{What the source experiment establishes}

Every controlled incoming state develops two resolved endpoint source
effects in prediction. At 512 updates the held-out predictive Gram ratio
$\lambda_2/\lambda_1$ ranges from $0.264$ to $0.582$ across the twelve
states. The common-field ratio ranges from $0.210$ to $0.612$. These
measurements resolve more than one finite source direction in both
observations, without identifying relaxation eigenmodes. The released
reference shows the same distinction in a much more concentrated regime.

Source selection also changes prediction on untouched documents. Relative
to general-source continuation, equal-mixture adaptation lowers the
balanced held-out risk in all twelve controlled states, by
$0.00084$--$0.12334$ nats, with equal-state mean difference $-0.06305$.
The pure technical arm improves ten of twelve states and the narrative
arm seven. Each width's mean technical-stratum risk is lower under the
technical endpoint than under the general arm; its mean narrative-stratum
risk is lower under the narrative endpoint. These are paired finite
effects under the fixed lexical and initialization conditioning.

For the realized source coupling, endpoint directions and the midpoint
leave substantial mixture-prediction error. At the terminal time, relative Fisher RMS
errors of the three-knot quadratic rule are $0.7926$, $0.8188$ and
$0.8403$ at increasing widths. It improves 32 of 48 individual
mixture/initialization cells, but its pooled error at four heads is
slightly larger than the affine rule's $0.7916$. These complete errors
quantify the approximation in~\eqref{model:eq:source-quadratic}; the exact
finite-law polynomial does not imply a useful quadratic truncation.

The source span also depends on adaptation age. On the same fixed
held-out documents, its smaller canonical cosine between updates 256
and 512 ranges from $0.019$ to $0.282$. Thus a terminal rank-two
description is not an approximately fixed pair of directions throughout
the observed path. Source-return transport explicitly retains its
state-dependent coefficients and the measured residual rather than
identifying this response span with an autonomous pair of native modes.

For the common-field susceptibility, the terminal $4$-to-$8$ secants
over the seven mixtures and general source are $0.084$--$0.147$; the
$8$-to-$14$ secants are $0.423$--$0.523$. Predicting the middle width
from the two endpoints under one pure power has relative errors
$0.089$--$0.141$. These finite trends retain shared initialization,
corpus and clock conditioning. They do not identify a common native
critical exponent or a critical source probability.

After 256 common-source updates, recalibration improves over leaving
both incoming contrasts unchanged in all twelve states. The two-column
relative residual nevertheless remains $0.842$--$0.978$. Adding the five
interior-source directions improves the terminal full-panel prediction
in all twelve states, with residuals $0.802$--$0.972$. The remaining
error is substantial. With the fixed budget of 48 calibration and 48
held-out documents used for time comparisons, the larger dictionary
improves nine of twelve terminal states. Neither dictionary has
established accurate autonomous source dynamics.

The two-column residual is $0.812$--$1.494$ times the weaker incoming
source singular scale. No terminal state satisfies the positive lower
bound in Corollary~\ref{model:cor:finite-source-span}. This failure of a
sufficient condition does not prove loss of every inherited direction:
the directly observed smaller current Gram eigenvalues are still
$0.0391$--$0.1876$. It establishes the narrower conclusion that these
calibrated transports do not certify inheritance at the weaker source
scale. The theory consequently retains the residual and distinguishes
finite response rank from preserved critical fluctuation exponents.

The supported description therefore retains finite source responses,
their changing predictive projections, optimizer history, remaining
corpus and observation-transport error. Several critical surfaces are
compatible with the conditional theory, and fine-tuning is a valid way
to probe their visibility. The executed observations establish distinct
finite effects in two training regimes. They do not establish that those
effects are inherited critical modes, that training self-organizes to
their intersection, or that a fitted head-count secant is a native
thermodynamic critical exponent.

\subsection{A qualified released base and finite-corpus adaptation}
\label{model:sec:released-adaptation}

The public PLDR-LLM-v51-SOC-110M models 1, 4 and 5 provide a matched
architecture comparison: five decoder layers, fourteen heads of dimension
64, hidden dimension 896 and a 32,000-token vocabulary. The reported
pretraining exposures are approximately eight, eight and forty-one billion
tokens, respectively~\cite{gokden2026soc}. The same paper reports model 5
as the strongest of these three on each of its aggregate benchmark scores.
Independent proper-prefix validation on the same 768 documents also
selects model 5 by cross entropy. The selected immutable release is
\path{de8e539c0ba1829072f4b8c2c5fae3bde0a3a2d2}. Native full-output and
restricted-output computations agree under the specified forward and
backward arithmetic checks at all three bases.

The new lexical cohort contains 4,096 training, 256 validation and 512 test
documents in each of the technical, narrative and unassigned strata.
All 14,592 document identities are distinct across these roles. They exclude
the source-return observation documents and have cached global positions
at least 160 million. The released pretraining document manifest is
unavailable, so this is a declared position-based separation under the
cached ordering; it is not a verified text-deduplication claim against the
released pretraining corpus. Fine-tuning and held-out assessment are
strictly separated at document level.

Each language arm uses 4,096 training documents and eight proper-prefix
examples per document. A 64-token input predicts the external next token;
the global Gram has no access to the supervised target. An epoch is one
random permutation of these 32,768 examples, or 1,024 batches of 32.
Three epochs revisit the same finite corpus explicitly. The technical and
narrative arms use their respective strata; the mixture uses 2,048 documents
from each, and the general arm uses 151 technical, 2,122 narrative and
1,823 unassigned documents. Source-arm randomization and every repeated
example identity are retained.

Model and hyperparameter choice use validation only. Full-parameter AdamW
and rank-four constrained adaptation~\cite{hu2021lora} have separate optimizer origins and
validation-selected checkpoints. The constrained parameterization changes
the query and value matrices in all five layers and the native vocabulary
output matrix, with 203,264 trainable factor coordinates. Its factors are
merged into the existing native matrices for assessment. There are no new
inference layers. Full adaptation uses peak rate $10^{-5}$, selected over
$3\times10^{-5}$; constrained adaptation uses $10^{-4}$, selected over
$3\times10^{-4}$. Each choice uses a separate fixed 256-update validation
pilot. Both main parameterizations use fresh AdamW moments $(0.9,0.95)$, numerical constant
$10^{-8}$, weight decay $0.01$, norm clipping at one and a 64-update
warmup followed by cosine decay to one fifth of the peak. The low-rank
factor $B$ starts at zero and entries of $A$ have standard deviation
$0.01$, so the incoming emission equals the released model's emission.
Proposition~\ref{model:prop:adaptation-factor-state} describes
the exact-real correspondence and the difference between their training
state laws. The recorded native export comparisons additionally check
floating-point agreement.

\begin{table}[htbp]\centering\small
\caption{Matched released-base selection. Exposure is in reported billions of pretraining tokens; Avg0 is the aggregate score reported in the source paper. The last two columns are measured proper-prefix validation scores.}
\label{model:tab:released-base}
\begin{tabular}{@{}lrrrr@{}}
\toprule Model & Exposure & Avg0 & NLL (nats) & Accuracy (\%)\\\midrule
1 & 8 & 40.95 & 3.8037 & 33.51\\
4 & 8 & 41.46 & 3.7063 & 34.64\\
5 & 41 & 42.62 & 3.6239 & 34.96\\
\bottomrule
\end{tabular}
\end{table}

The paired document interval for model 5 minus model 4 validation NLL is $[-0.1093, -0.0563]$ nats.

\begin{table}[htbp]\centering\small
\caption{Held-out language scores for validation-selected constrained adaptation. Each row averages its declared evaluation strata equally, including the general-source row. Prefix length is 64; an update-zero selection is the unchanged released base.}
\label{model:tab:released-language}
\begin{tabular}{@{}lrrrrr@{}}
\toprule Source & Update & Base NLL & Selected NLL & Change & Acc. (\%)\\\midrule
Technical & 1024 & 3.5557 & 3.5507 & -0.0050 & 36.52\\
Narrative & 0 & 3.5772 & 3.5772 & +0.0000 & 33.89\\
Mixture & 1024 & 3.5665 & 3.5656 & -0.0009 & 35.25\\
General & 1024 & 3.5975 & 3.5967 & -0.0008 & 34.96\\
\bottomrule
\end{tabular}
\end{table}

Paired 95\% document intervals for the NLL changes are technical $[-0.0086, -0.0011]$, narrative $[0.0000, 0.0000]$, mixture $[-0.0021, 0.0004]$, general $[-0.0017, 0.0000]$. All context lengths and selected full-parameter controls are retained in the bound outcome table. The validation rule selects the incoming state in 4 of four full-parameter source arms. Those selections supply unchanged-base controls, rather than additional learned source states.

The held-out language panel contains two token offsets and prefix lengths
32, 64 and 128 for every test document. Paired uncertainty estimates
resample complete documents and retain all their token positions together.
The 64-token subset matches the fine-tuning context; the remaining lengths
measure context transfer. These source-conditioned language scores qualify
the released base and its adapted emissions. They do not assign a physical
universality class to RefinedWeb or identify a new native-width ensemble.

\subsection{A matched single-pass coordinate comparison}
\label{model:sec:singlepass-released}
Two further paths isolate the finite resource law from corpus repetition.
Both start from model 5 and use the same technical documents, the same
32,768 proper-prefix examples and the same frozen permutation, with seed
220020. Each completes exactly 1,024 batch-32 updates in one pass.
The full-parameter and rank-four paths use the fixed peak rates $10^{-5}$
and $10^{-4}$, respectively, inherited from validation choices rather
than refitted on this test panel. Consequently this is a comparison of
two specified optimizer laws, not a causal effect of factorization alone.
The complete source sequence, native traces, validation observations,
selected states and merged-factor checks are retained.

Full-parameter validation selects update zero. Factor validation selects
update 1,024, with technical validation NLL changing from 3.68125 to
3.68069. The full test panel contains 512 documents in each of three
domains, two offsets and prefixes 32, 64 and 128. All twelve
parameterization/domain/prefix comparisons are retained in
Table~\ref{model:tab:singlepass-adaptation}. Each of 16,000 paired bootstrap
resamples retains a document's complete position panel. Intervals are
conditional on this corpus and the selected checkpoints. The technical
factor NLL change across all prefixes is $-0.00227$ nats, with an
adjusted interval $[-0.00381,-0.00076]$. The matched-64 change is
$-0.00262$ nats; its nominal interval excludes zero but its adjusted
interval includes zero. Other domains do not establish an improvement
under this uncertainty analysis. These modest finite gains coexist
with the unchanged full-parameter comparator.

\begin{table}[htbp]\centering\small
\caption{Single-pass technical-source adaptation: held-out NLL changes from the incoming base, in nats. Every full-parameter entry is exactly zero because validation selects the incoming checkpoint. Factor intervals use paired document resampling, conditional on the fixed models and corpus. The final column allocates nominal 95\% coverage over all twelve source/parameterization/prefix comparisons.}
\label{model:tab:singlepass-adaptation}
\begin{tabular}{@{}llrrrr@{}}
\toprule Domain & Prefix & Full & Factor & Factor 95\% interval & Adjusted interval\\\midrule
Technical & 64 & 0 & -0.00262 & $[-0.00453,-0.00075]$ & $[-0.00535,0.00012]$\\
Technical & 32,64,128 & 0 & -0.00227 & $[-0.00330,-0.00121]$ & $[-0.00381,-0.00076]$\\
Narrative & 64 & 0 & -0.00020 & $[-0.00167,0.00122]$ & $[-0.00230,0.00183]$\\
Narrative & 32,64,128 & 0 & -0.00047 & $[-0.00127,0.00030]$ & $[-0.00163,0.00068]$\\
Unassigned & 64 & 0 & -0.00044 & $[-0.00195,0.00108]$ & $[-0.00261,0.00179]$\\
Unassigned & 32,64,128 & 0 & 0.00002 & $[-0.00091,0.00095]$ & $[-0.00132,0.00139]$\\
\bottomrule\end{tabular}\end{table}

The paths take 125.1 and 100.1 seconds, including their recorded validation
and checkpoint operations, with peak allocated GPU memory 9.15 and
6.02 GiB. These are whole-path costs, not a matched-accuracy speedup.
The factors and their Adam moments remain in the training state; the
native inference export satisfies its measured logit tolerance.
The single-pass law therefore supports source-specific finite adaptation
without invoking repeated exposure. The experiment conditions on one
released base and one realized adaptation corpus. It neither recovers
that base's original pretraining distribution nor identifies a native
critical mode, an economical autonomous RG surrogate or a universal
improvement from adaptation.

\section{Observed operator stabilization and predictive transport}
\label{model:sec:inference-results}

\subsection{Native prefix interventions and residual row transport}
We measured both released PLDR-LLM models and three declared
$N=14$ controlled checkpoints: $g=1$ and $g=0$ at 32,768 updates,
and $g=1$ at 131,072 updates, all with initialization identity
640101. These five states are a descriptive mechanism panel.
Their different pretraining exposures are not randomized treatments.
The reference designations describe the released models; the
controlled $g=0$ condition is the frozen-generator boundary of our
constant-rate family.

At each state, eight fixed held-out RefinedWeb crops supplied common
prefixes of lengths 16, 32, 48 and 64. We compared the original
64-token forward with a proper prefix forward, an identical-input
replay, replacement of only the future token occupying the target's
input position, and replacement of the entire later suffix. Replacements
used a fixed cyclic donor among the eight crops. The scored target
and prefix remained fixed. All vocabulary logits and native $A,G$
outputs were retained. Every identical-input replay agreed bytewise.
The measurements use native CPU float32 forwards and float64 scalar
reductions. Prefix shortening changes sequence length; suffix
replacement holds it fixed and directly tests later-token dependence.

We also passively recorded the normalized input Gram and each of
eight residual row units in all five decoder layers. Endpoints
reproduced the corresponding native prefix experiment bytewise.
Let $E_j$ be the mean centered row energy at stage $j$, and $S_j$
the mean squared difference between original and replaced-suffix
row clouds. The finite observed transport factors are
$K_R=\sqrt{E_8/E_0}$ and $K_S=\sqrt{S_8/S_0}$, with undefined
ratios recorded separately if an input energy is zero. These are
sampled transport factors, not estimates of the global supremum
Lipschitz constants in Proposition~\ref{model:prop:shared-row-input-contraction}.

\begin{table}[htbp]
\centering\small
\caption{Observed row and suffix transport across each complete
residual stack. Extrema cover the five decoder layers. Suffix KL
averages the eight crops and the three prefix lengths below 64.
An exact zero is a zero in this float32 observation program.}
\label{model:tab:inference-prefix-transport}
\input{content/model/generated/inference-prefix-transport.tex}
\end{table}

Both released models suppress row variation and later-token
variation across the residual stack. Their mean predictive KL under
suffix replacement is $5.45\times10^{-10}$ and
$3.85\times10^{-13}$, respectively. The latter scale is reported
as a finite arithmetic observation, without assigning a physical
exponent to it. The joint suppression supports the shared-map
mechanism in Section~\ref{model:sec:inference-mechanism}. It does not
replace the common-row field by the centered row norm.

To measure infinitesimal transport directly, we selected two of the
eight contexts, heads 1, 7 and 14, and rows 1, 32 and 64 in each
of the five decoder layers. This gives 90 input points per state
and 450 complete $64\times64$ Jacobians. Selection preceded the
Jacobian measurements and followed the secant panel, so this is a
descriptive extension of that panel. The native float32 row stacks
first replayed all recorded stages bytewise. Parameters and normalized
Gram inputs were then lifted exactly from float32 to float64 for
differentiation. An independent calculation using the affine,
SiLU, multiplication, residual and layer-normalization formulas
reconstructed every retained Jacobian and singular spectrum.

\begin{table}[htbp]
\centering\small
\caption{Local spectral norms of the complete eight-unit shared
row map. These values concern the declared lifted float64
computation at 90 points per state, not a supremum over the input
domain or an eigenvalue of the training process.}
\label{model:tab:inference-row-jacobians}
\input{content/model/generated/inference-row-jacobians.tex}
\end{table}

Every sampled local row map in the two released models contracts
infinitesimal perturbations. The largest complete-stack gains are
0.0671 and $2.42\times10^{-4}$. In some layers the product of the
separate-unit spectral norms exceeds one while the complete-stack
norm remains below one. This demonstrates why retaining matrix
orientation can give a sharper transport description than
multiplying scalar gains. Equation~\eqref{model:eq:row-input-training-differential}
then distinguishes this input contraction from the response of the
learned operator to training. The measured contraction directions
are not themselves identified as critical training modes.

\subsection{Fixed learned operators and task-margin preservation}
For each state we calibrated fixed native operators on 64 distinct
RefinedWeb contexts, using float64 averaging followed by conversion
to float32. We compared native inference, these fixed operators,
a cyclic permutation of their heads with each recipient head's
RMS norm preserved, and projection of $A$ onto its row centroid
before recomputing the native metric and downstream graph.
No intervention changed the trained weights. The native external
operator branch reproduced each input's original forward bytewise
when supplied with that input's own operators. With the same fixed
operators, equal prefixes under later-token replacement also
reproduced their logits bytewise. Restoring the native computation
after each intervention reproduced its original outputs.

The prediction-loss panel uses 64 fixed held-out external targets.
The task panel uses 64 ARC-Easy and 64 ARC-Challenge questions
selected by a fixed seed from the published test sets
\cite{Clark2018ARC}, subject to a prefix-preserving 64-input-token
limit. It also includes 192 generated location questions in
counterfactual pairs at composition depths one, two and three.
The paired instances change supplied premises and the specified
correct answer. A separate graph traversal checks those answers.
All answer candidates are scored through their proper autoregressive
prefixes using the reference SentencePiece tokenizer. The primary
score sums candidate-token log probabilities; token-mean scores
are retained as a sensitivity analysis. This short-context panel
is a specified task distribution, not the full reference benchmark
evaluation or its length-normalized scoring protocol.

For every item and intervention, retained vocabulary logits determine
all candidate scores, their signed margins and their maximum paired
score error $e$. Table~\ref{model:tab:inference-task-margins} counts the
strictly correct native examples satisfying $m>2e$. Every such
example retained its correct answer, as required by
Proposition~\ref{model:prop:task-margin}. The denominator is the number
of strictly correct native items, not the total panel size.
The ``Changes'' column counts all decision changes among the 320
items, including decisions that were initially incorrect.

\begin{table}[htbp]
\centering\footnotesize
\caption{Completed native operator interventions. NLL concerns the
64 held-out external targets; score error and decision counts
concern the separate 320-item task panel. Certified counts use the
observed sufficient margin bound and do not estimate a uniform
transport constant.}
\label{model:tab:inference-task-margins}
\input{content/model/generated/inference-task-margins.tex}
\end{table}

In both released models, calibration freezing and row projection
preserve all 320 answer decisions. Their native external-target
losses are 3.77717 and 3.64575 nats; head permutation raises them
to 6.10964 and 6.17360 nats. The norms of the fixed reference heads
are preserved in that intervention, so these differences concern
the learned assignment of operators to contextual attention paths.
The stable operators carry useful predictive content on this law.
Stability also preserves incorrect decisions. Consequently these
experiments establish retention of learned predictions and task
margins, rather than an independent establishment of general
multi-fact reasoning.

\begin{figure}[htbp]
\centering
\includegraphics[width=\linewidth]{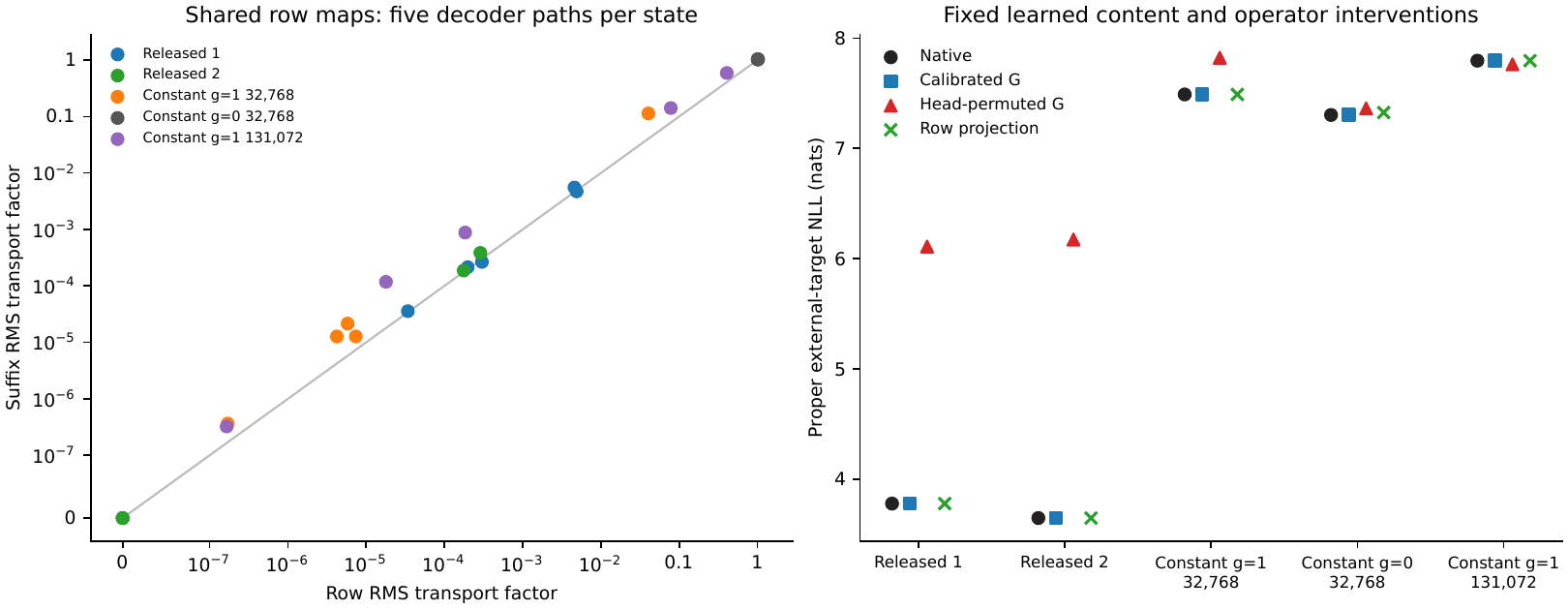}
\caption{Finite observed transport through the residual row stacks
and predictions under native operator interventions. The left panel
has five decoder paths per state and uses a linear neighborhood
of zero on otherwise logarithmic axes. Overlapping zero values
remain zeros. The right panel separates preservation by calibrated
operators from sensitivity to their learned head assignment.}
\label{model:fig:inference-mechanism}
\end{figure}

These completed measurements support a training-selected inference
description that retains learned common operators, their input
variation, predictive transport and task margins. They do not
identify a thermodynamic critical surface from a small inference
variation statistic. The critical scaling and endogenous feedback
hypotheses remain those of the joint training law, as specified in
the analytical sections.

\subsection{Demonstrations and controlled premise changes}
To test whether worked examples elicit the selected compositional
behavior, we evaluated the same 192 location questions with zero,
two and four demonstrations. This yields 576 questions at each
of the same five checkpoints. Twelve demonstrations provide four
examples at each composition depth. Demonstration people and object
names are disjoint from those in the scored questions; answer roles
are balanced in both the first two and all four demonstrations.
The target facts, candidate choices and counterfactual pairs remain
unchanged across demonstration counts. A separate graph traversal
checks every demonstration and target answer. The selection was
fixed after inspecting the initial task panel and before scoring
any of these additional prompts, so it is a descriptive test of
prompt conditioning on those questions.

The longest actual proper prefix has 188 tokens. Native inference
with each of the two longest prefixes first reproduced its own
captured-operator replay and restored computation bytewise. All
four operator modes then scored every candidate through its proper
prefix, in CPU batches of eight. Independent reductions reconstructed
the vocabulary log probabilities and all 11,520 candidate-margin
comparisons. The zero-demonstration prompts exactly repeat the
initial composition questions, whose forward batch size was 32.
Score differences from that arithmetic change are retained in
Table~\ref{model:tab:prompt-repeat}; no answer decision changed in the
3,840 repeated item/mode comparisons.

\begin{table}[htbp]
\centering\footnotesize
\caption{Zero-demonstration repeat under the declared batch-size
change. Maximum score difference covers the 192 questions and all
four modes at each state. Longest prefixes refer to the full
zero/two/four demonstration panel.}
\label{model:tab:prompt-repeat}
\begin{tabular}{lrrr}
\toprule
State & Max. prefix & Max. score change & Changed decisions\\
\midrule
Released 1 & 188 & $6.05\times10^{-6}$ & 0\\
Released 2 & 188 & $7.34\times10^{-6}$ & 0\\
$g=1$, 32,768 & 188 & $1.11\times10^{-4}$ & 0\\
$g=0$, 32,768 & 188 & $5.58\times10^{-6}$ & 0\\
$g=1$, 131,072 & 188 & $4.81\times10^{-6}$ & 0\\
\bottomrule
\end{tabular}

\end{table}

\begin{table}[htbp]
\centering\footnotesize
\caption{Native answers under every demonstration count. ``Correct''
counts individual questions; each depth column counts pairs for
which both opposite premise variants receive the specified answer.
The signed response $\Delta$ is defined in
Proposition~\ref{model:prop:premise-routing-margin} and is averaged across
all 96 pairs.}
\label{model:tab:prompt-native}
\input{content/model/generated/prompt-native-composition.tex}
\end{table}

The released models answer both variants correctly on only zero
to two of the 96 pairs at each demonstration count. Individual
accuracy is close to one half, while the paired result directly
tests whether the answer follows the changed premises. Two or four
worked examples do not establish reliable composition on this
finite task law. This statement does not evaluate every possible
prompt format or assign a population reasoning score.
Equation~\eqref{model:eq:premise-routing-margin} identifies the missing
property precisely: useful directional premise response must
overcome the pair's average answer preference.

\begin{figure}[htbp]
\centering
\includegraphics[width=\linewidth]{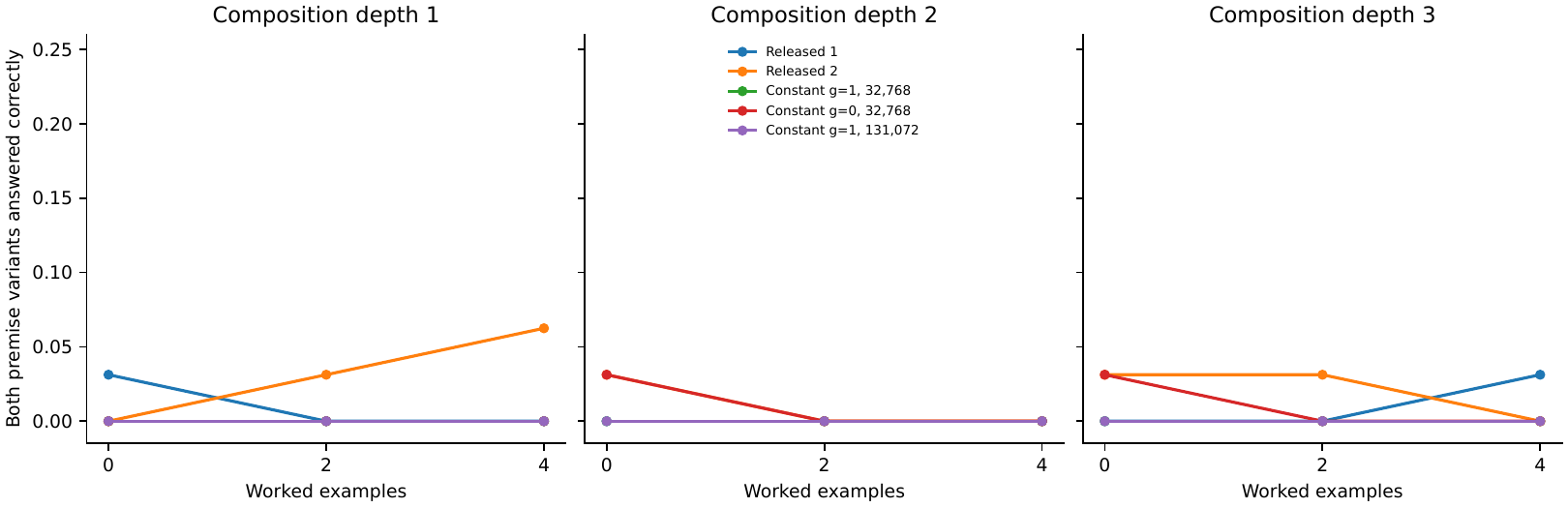}
\caption{Completed native premise-pair outcomes at every selected
composition depth and demonstration count. Each point contains
32 complete counterfactual pairs; overlapping zero outcomes remain
zero. These are fixed-panel counts rather than independent model
replications or a finite-size critical scaling curve.}
\label{model:fig:prompt-conditioning}
\end{figure}

\begin{table}[htbp]
\centering\footnotesize
\caption{Operator interventions with zero demonstrations. ``Both''
counts correctly answered counterfactual pairs out of 96; ``Changes''
counts decisions changed from native inference among 192 questions.
``Certified'' gives the sufficient margin count over strictly
correct native questions. The last column counts pairs with
positive signed premise response, which alone is insufficient for
both answers to be correct.}
\label{model:tab:prompt-operators-zero}
\input{content/model/generated/prompt-operators-shots0.tex}
\end{table}

\begin{table}[htbp]
\centering\footnotesize
\caption{Operator interventions with two demonstrations, using the
same counts and denominators as Table~\ref{model:tab:prompt-operators-zero}.}
\label{model:tab:prompt-operators-two}
\input{content/model/generated/prompt-operators-shots2.tex}
\end{table}

\begin{table}[htbp]
\centering\footnotesize
\caption{Operator interventions with four demonstrations, using the
same counts and denominators as Table~\ref{model:tab:prompt-operators-zero}.}
\label{model:tab:prompt-operators-four}
\input{content/model/generated/prompt-operators-shots4.tex}
\end{table}

Fixed calibration and row projection again preserve every answer
decision in both released models across all demonstration counts.
The controlled checkpoints and the head-permutation outcomes are
retained alongside them. Preservation of these weak paired outcomes
makes the separation between operator stability and task competence
observable. A theory relating a training regime to reasoning must
therefore explain the selection of the contextual computation and
readout, its premise-response margins, and the preservation of those
margins at inference. The row-contraction mechanism addresses the
last part when the required task knowledge has been learned.

\subsection{Full ARC test sets and the released score convention}
\label{model:subsec:full-arc}

Both released checkpoints are evaluated on all 2,376 ARC-Easy and
1,172 ARC-Challenge test questions. The 14,188 candidate answers
require 55,940 unique proper prefixes per model and operator mode.
No question is excluded and no prefix is truncated; the longest
prefix has 180 tokens. The template is \texttt{Question:} followed
by the question and a new line beginning \texttt{Answer:}. Neither
BOS nor EOS is inserted. Every candidate-token score uses its own
proper prefix, with CPU float32 native forwards and float64
log-softmax scoring.

The preserved public evaluation harness defines its normalized
accuracy by dividing the answer log-likelihood by the number of
Unicode characters in the answer text, excluding the added leading
separator. This differs from normalization by tokenizer tokens or
UTF-8 bytes. The latter two are retained as explicit sensitivity
observations; 55 candidate answers have different character and
byte counts. All four conventions are reported in
Tables~\ref{model:tab:arc-characters}--\ref{model:tab:arc-bytes}.

\begin{table}[htbp]
\centering\footnotesize
\caption{Full ARC evaluation with answer-character normalization,
matching the preserved public harness's normalization rule.
The two accuracy columns use 2,376 and 1,172 questions, respectively.
``Changes'' counts changed decisions among all 3,548 questions.
``Certified'' gives the sufficient margin certificate count over
strictly correct native questions under this same score convention.}
\label{model:tab:arc-characters}
\input{content/model/generated/benchmark-answer-characters.tex}
\end{table}

Native character-normalized accuracies are 36.15\% and 22.95\%
for released model 1, and 36.87\% and 20.99\% for released model 2.
These match the corresponding rounded entries reported in
\cite{gokden2026soc}. Agreement of these benchmark aggregates does
not assert identity of every implementation detail of the original
pretraining or evaluation program.

Fixed calibration operators and row projection preserve every
native decision in both models under every score convention.
For the character convention, all 1,128 and 1,122 strictly correct
native answers, respectively, satisfy the sufficient preservation
certificate. Head permutation changes 1,297 and 1,319 decisions.
It reduces ARC-Easy accuracy to 31.73\% and 30.77\%, respectively,
while ARC-Challenge accuracy becomes 21.25\% and 21.76\%.
The latter changes have different signs in the two models; a
wrong-operator perturbation need not lower every benchmark score.

\begin{table}[htbp]
\centering\footnotesize
\caption{Full ARC evaluation with the summed proper-prefix answer
log-likelihood. Counts and certificates use this convention's
native decisions, with the denominators described in
Table~\ref{model:tab:arc-characters}.}
\label{model:tab:arc-sum}
\input{content/model/generated/benchmark-sum.tex}
\end{table}

\begin{table}[htbp]
\centering\footnotesize
\caption{Full ARC evaluation with answer-token normalization.
This is a separate score convention from the public harness's
character normalization.}
\label{model:tab:arc-tokens}
\input{content/model/generated/benchmark-answer-tokens.tex}
\end{table}

\begin{table}[htbp]
\centering\footnotesize
\caption{Full ARC evaluation with answer-byte normalization.
UTF-8 byte counts and Unicode character counts are kept distinct.}
\label{model:tab:arc-bytes}
\input{content/model/generated/benchmark-answer-utf8-bytes.tex}
\end{table}

The independent reconstruction covers 113,536 normalized
question-margin comparisons across both checkpoints, all modes
and all four conventions. Native own-operator replay and restoration
also agree bytewise on the two longest-prefix qualification inputs
for each model. These results validate preservation of measured
benchmark decisions by the learned operator reduction. Benchmark
knowledge, premise-sensitive composition and thermodynamic
criticality remain distinct properties. The two pretrained
checkpoints do not supply independent replications of a training
transition or establish absence of benchmark contamination.

\FloatBarrier
\par\medskip\noindent
Chapter~\ref{ch:conditional-criticality} states the limiting laws that
would justify a scaling classification beyond these finite observations.
It specifies renormalized families, conditional head limits and the role of
shared environments.

\addtocontents{toc}{\protect\newpage}
\part{Conditional scaling and inference visibility}
\chapter{Renormalized laws and conditional criticality}
\label{ch:conditional-criticality}
This chapter develops conditional scaling and criticality for specified
renormalized laws. It treats cumulant flow, Gaussian and signed-head limits,
collective susceptibility and stability, and the additional assumptions
needed to interpret a width family as a thermodynamic comparison.

\section{RG flow and independent-context universality}
\label{model:sec:flow}

Here the random units are independent contexts drawn from the declared law
$\D$ at one fixed model state. The observation is the centered finite
full-graph field; its dimension and the model stay fixed while the number
$b$ of contexts in a block grows. Head and layer coordinates within one
context retain their joint dependence. This context limit does not vary
training initialization or head count.

\subsection{Tensor cumulants and scaling fields}
Let $\kappa_n=\nabla^n W(\bm0)$ be the connected cumulant tensor of order $n$.
For finite calls all these derivatives exist by
Proposition~\ref{model:prop:regularity}.

\begin{theorem}[Complete cumulant flow]\label{model:thm:cumulants}
Under independent-context blocking,
\begin{equation}
 \kappa_n' = b^{1-nH} C^{\otimes n}\kappa_n.\label{model:eq:cumulant-flow}
\end{equation}
In particular the covariance transforms as
$\Sigma'=b^{1-2H}C\Sigma C^{\mathsf T}$, including all cross-head and
cross-layer entries. With $C=I$, dyadic iteration is
\begin{equation}
 \kappa_n^{(k+1)}=2^{1-nH}\kappa_n^{(k)},\qquad
 \beta_n^{\mathrm{disc}}=
 \frac{2^{1-nH}-1}{\log 2}\kappa_n.\label{model:eq:discrete-beta}
\end{equation}
At $H=1/2$, a mean displacement is relevant with RG eigenvalue exponent $1/2$,
the covariance is marginal, and every cumulant of order $n>2$ has negative
RG eigenvalue exponent $1-n/2$.
\end{theorem}
\begin{proof}
Differentiate~\eqref{model:eq:rg-cgf} $n$ times at $j=\bm0$. Each derivative supplies
one factor $b^{-H}C$, while the exterior factor is $b$. This proves the
tensor formula, the covariance formula, and the dyadic recursion.
Subtracting successive iterates and dividing by $\log2$ gives the stated
discrete beta function. The signs follow directly from $1-n/2$.
\end{proof}

For a linear observation without document blocking, the covariance law is
$C\Sigma C^{\mathsf T}$. Thus head elimination and document averaging
commute in the sense of Theorem~\ref{model:thm:law-rg}, but discarding off-diagonal
covariances does not commute with either operation.

The familiar differential form can be recorded precisely. The analytic
interpolation $W_s(j)=e^s W(e^{-Hs}j)$ satisfies
\begin{equation}
 \partial_s W_s=W_s-Hj^{\mathsf T}\nabla W_s,
 \qquad \partial_s\kappa_n=(1-nH)\kappa_n.\label{model:eq:continuous-beta}
\end{equation}
This follows by the chain rule. For a general finite law this interpolation
need not be a log moment generating function at noninteger $e^s$;
probability powers require divisibility. Consequently
Equation~\eqref{model:eq:discrete-beta} is the exact probability flow used here.
Equation~\eqref{model:eq:continuous-beta} is its analytic scaling interpolation,
or a probability flow when the required infinitely divisible law exists,
as it does at the Gaussian fixed point. No continuous probability
semigroup for an arbitrary discrete empirical law is assumed.

\subsection{The joint Gaussian fixed point}
The following is the multivariate independent-sum central limit theorem,
specialized to the bounded full-graph field \cite{kallenberg2021,jona2001}.
\begin{theorem}[Independent-context fluctuation class]\label{model:thm:clt}
Let $\varphi$ be any centered finite-dimensional observation of a fixed
finite PLDR call, sampled independently from $\D$, with covariance
$\Sigma$. Then
\begin{equation}
 b^{-1/2}\sum_{i=1}^b\varphi_i\ \Rightarrow\ N(\bm0,\Sigma)
       \quad\text{as }b\to\infty.\label{model:eq:clt}
\end{equation}
The Gaussian law is fixed by $\RG_{b,I,1/2}$ for every positive integer
$b$. On $\range\Sigma$, whitening identifies its nondegenerate part with
$N(\bm0,I_r)$, $r=\rank\Sigma$. Joint observations from two models belong to
this same independent-context Gaussian fluctuation class after their own
centering and covariance normalization. Their covariances and predictive
response coefficients need not agree.
\end{theorem}
\begin{proof}
Write $\chi(t)=\E e^{it^{\mathsf T}\varphi}$. Boundedness and centering
allow Taylor expansion with a uniform local remainder:
\[
 \chi(t)=1-\tfrac12t^{\mathsf T}\Sigma t+O(\norm{t}^3).
\]
For fixed $t$, independence gives
\[
 \chi(t/\sqrt b)^b
 =\left[1-\frac{t^{\mathsf T}\Sigma t}{2b}+O(b^{-3/2})\right]^b
 \longrightarrow e^{-t^{\mathsf T}\Sigma t/2}.
\]
The limit is continuous at the origin and is the characteristic function
of the possibly degenerate Gaussian with covariance $\Sigma$. The
characteristic-function continuity theorem proves~\eqref{model:eq:clt}. A sum
of $b$ independent such Gaussians has covariance $b\Sigma$, so division by
$\sqrt b$ proves fixedness. Diagonalize $\Sigma$ orthogonally and divide
each positive-eigenvalue coordinate by its standard deviation. Coordinates
in its kernel have zero variance and are almost surely zero after
centering; quotienting them gives $N(\bm0,I_r)$. Applying the same argument to
each model proves the final assertion.
\end{proof}

This is a model-wide statement about all chosen fields simultaneously,
not separate central limit claims for each head. It remains true for the
complete finite trace. Its hypothesis is the independently sampled
\emph{context}; tokens within a context and tensors within a model may
have arbitrary dependence. The Gaussian attractor is a theorem of this
specified blocking. It does not imply that one unblocked model output is
Gaussian, or that the model was trained at a critical point.

\subsection{The Gaussian effective action under head elimination}
On a nondegenerate retained fluctuation space, the Gaussian fixed point has
quadratic action
\begin{equation}
 \Gamma(y)=\tfrac12 y^{\mathsf T}Qy+\text{constant},\qquad Q=\Sigma^{-1}.
\end{equation}
This action describes a probability law of observations. It is not the
architecture tensor $\G$.

\begin{theorem}[Gaussian marginalization]\label{model:thm:schur}
Partition a positive-definite precision matrix as
\[
 Q=\begin{pmatrix}Q_{rr}&Q_{ru}\\Q_{ur}&Q_{uu}\end{pmatrix}.
\]
Integrating the unresolved field $u$ gives the exact retained precision
\begin{equation}
 Q_{\mathrm{eff}}=Q_{rr}-Q_{ru}Q_{uu}^{-1}Q_{ur}.\label{model:eq:schur}
\end{equation}
It equals $(\Sigma_{rr})^{-1}$. Sequential and simultaneous elimination
of the same coordinates agree.
\end{theorem}
\begin{proof}
Complete the square:
\[
 \begin{pmatrix}r\\u\end{pmatrix}^{\mathsf T}Q
 \begin{pmatrix}r\\u\end{pmatrix}
 =r^{\mathsf T}Q_{\mathrm{eff}}r+
 (u+Q_{uu}^{-1}Q_{ur}r)^{\mathsf T}Q_{uu}
 (u+Q_{uu}^{-1}Q_{ur}r).
\]
The Gaussian integral over the shifted $u$ contributes only a factor
independent of $r$. The marginal covariance of $r$ is, by definition,
$\Sigma_{rr}$, which proves the inverse-covariance identity. Gaussian
densities are integrable and nonnegative, so iterated integration agrees
with joint integration. Uniqueness of the retained quadratic form proves
the last statement.
\end{proof}

Replacing $Q_{\mathrm{eff}}$ by $Q_{rr}$ is conditioning $u=\bm0$, a different
operation. Non-Gaussian finite-scale laws produce higher interactions on
elimination; the covariance rule alone is then an approximation to their
full action. The exact state continues to be the joint law of
Section~\ref{model:sec:maps}.

\section{A criticality model with an explicit scale}
\label{model:sec:criticality}

\subsection{Conjugate fields and observable susceptibility}
For a centered observable law $\mu$, introduce a statistical source $j$ by
\begin{equation}
 d\mu_j(y)=e^{j^{\mathsf T}y-W(j)}d\mu(y).
\end{equation}
Differentiating the finite moment generating function gives
$\nabla W(j)=\E_j y$ and
$\nabla^2 W(j)=\Cov_j(y)$.
In particular $\partial_j\E_j y|_{j=\bm0}=\Sigma$. This susceptibility
measures a response to reweighting contexts. It has a different source
from the head intervention $\eta$ in Section~\ref{model:sec:response}.

Let $Y_1,\ldots,Y_b$ be vector observations at ordered positions or
segments, and center each using its own population mean. Define
\begin{equation}
 M_b=\frac1b\sum_{i=1}^bY_i,
 \qquad \chi_b=b\Cov(M_b)
       =\Cov\left(b^{-1/2}\sum_{i=1}^bY_i\right).\label{model:eq:chi}
\end{equation}
The factor $b$ is fixed before measurement. A common deterministic
position-dependent mean is not counted as a fluctuating correlation.

\begin{theorem}[Exact model-wide susceptibility RG]\label{model:thm:chi}
For arbitrary dependence and without stationarity,
\begin{equation}
 \chi_b=\frac1b\sum_{i,j=1}^b\Cov(Y_i,Y_j).\label{model:eq:chi-general}
\end{equation}
For a second-order stationary sequence, writing
$C(r)=\Cov(Y_0,Y_r)$,
\begin{equation}
 \chi_b=C(0)+\sum_{r=1}^{b-1}(1-r/b)
                       \bigl[C(r)+C(r)^{\mathsf T}\bigr].\label{model:eq:chi-lag}
\end{equation}
For independent identically distributed fields this is $\chi_b=C(0)$.
If $\sum_{r\geq1}\norm{C(r)}<\infty$, then
\begin{equation}
 \chi_b\longrightarrow C(0)+\sum_{r\geq1}
                     [C(r)+C(r)^{\mathsf T}],\label{model:eq:summable}
\end{equation}
a finite positive-semidefinite matrix.
\end{theorem}
\begin{proof}
Expand the covariance of the sum bilinearly, proving
\eqref{model:eq:chi-general}. Under stationarity there are $b$ terms at zero
lag, and $b-r$ terms of each orientation at positive lag $r$; counting them
gives~\eqref{model:eq:chi-lag}. Independence removes all off-diagonal terms.
Under absolute norm summability, each coefficient $1-r/b$ tends to one
and has absolute value at most one on its range. Dominated convergence
for the matrix series proves~\eqref{model:eq:summable}. Each $\chi_b$ is a
covariance, and the positive-semidefinite cone is closed.
\end{proof}

\begin{definition}[Fluctuation criticality for a specified observation]
\label{model:def:fluctuation-criticality}
Fix a nondegenerate fine-scale covariance $C_0$ on the retained field
space. The dimensionless largest susceptibility is
\begin{equation}
 \Lambda_b=\sup_{v\ne\bm0}\frac{v^{\mathsf T}\chi_bv}{v^{\mathsf T}C_0v},
 \qquad m_b=\Lambda_b^{-1}.\label{model:eq:criticality}
\end{equation}
We set $m_b=+\infty$ if $\Lambda_b=0$. A regular regime has $\sup_b\Lambda_b<\infty$. Fluctuation criticality
requires $\Lambda_b\to\infty$ along the specified scale limit, or a
divergence on approaching a separately specified control-parameter limit.
The mass $m_b$ then tends to zero. This divergence identifies a
fluctuation sector; its origin and the nature of any thermodynamic
transition require additional analysis. A universality class additionally
specifies a limiting joint law after field normalization and its relevant
scaling fields, rather than only a fitted variance exponent.
\end{definition}

This criterion is invariant under invertible linear changes of retained
coordinates: substitute $C_0'=TC_0T^{\mathsf T}$ and
$\chi_b'=T\chi_bT^{\mathsf T}$, then set $w=T^{\mathsf T}v$ in the
Rayleigh quotient. Zero-variance coordinates must first be quotiented
out. Their vanishing is not divergent susceptibility. With independent
contexts and $C_0=\Sigma$, Theorem~\ref{model:thm:chi} gives
$\Lambda_b=1$ at every scale, although the \emph{normalized law} flows to
a Gaussian fixed point. A fixed point of an RG and a critical point of a
physical family are distinct notions.

\subsection{Conditional long-range scaling}
\begin{proposition}[A correlated critical sector]\label{model:prop:longrange}
Suppose a stationary sequence satisfies, in matrix norm,
\begin{equation}
 \tfrac12[C(r)+C(r)^{\mathsf T}]=Mr^{-\gamma}+o(r^{-\gamma}),
 \qquad 0<\gamma<1,\qquad M\succeq\bm0_{\mathrm{mat}},\ M\ne\bm0_{\mathrm{mat}}.
\end{equation}
Then
\begin{equation}
 \chi_b\sim\frac{2M}{(1-\gamma)(2-\gamma)}b^{1-\gamma}
 \label{model:eq:longrange}
\end{equation}
in the sense that the difference divided by $b^{1-\gamma}$ tends to zero.
The variance-normalizing field exponent is $H=1-\gamma/2$ in the
nonzero sector of $M$. This covariance condition alone does not specify
the limiting distribution.
\end{proposition}
\begin{proof}
Insert the assumed symmetric covariance into~\eqref{model:eq:chi-lag}. The
leading scalar sum obeys
\[
 b^{\gamma-1}\sum_{r=1}^{b-1}(1-r/b)r^{-\gamma}
 \longrightarrow\int_0^1(1-u)u^{-\gamma}\,du
 =\frac1{(1-\gamma)(2-\gamma)}.
\]
The integrable singularity at zero is controlled by splitting at a fixed
small positive fraction of $b$ and bounding the initial sum by a constant
times that fraction to power $1-\gamma$. For the remainder, choose a
fixed large lag beyond which its norm is at most
$\varepsilon r^{-\gamma}$. The finite initial remainder vanishes after
division by $b^{1-\gamma}$ and the tail is bounded by $\varepsilon$ times
the leading sum. Let $\varepsilon\downarrow0$. This proves
\eqref{model:eq:longrange}. Since
$\Cov(\sum_iY_i)=b\chi_b$, its nonzero leading sector grows as
$b^{2-\gamma}$, giving $2H=2-\gamma$. Covariances determine only second
moments; no limiting characteristic function follows from the hypothesis.
\end{proof}

Likewise, a document-level latent field can produce growing susceptibility
without a critical scale limit being identified. Let $b\geq1$ and let
$Y_i=M_{\mathrm{doc}}+\varepsilon_i$ be square-integrable vector observations.
Assume the residuals are centered, mutually independent, and independent
of $M_{\mathrm{doc}}$. Their covariance matrices may depend on position. Then
\begin{equation}
 \chi_b=b\Cov(M_{\mathrm{doc}})
       +\frac1b\sum_{i=1}^b\Cov(\varepsilon_i).\label{model:eq:latent}
\end{equation}
Indeed, $\Cov(Y_i,Y_j)=\Cov(M_{\mathrm{doc}})$ for $i\ne j$, whereas
$\Cov(Y_i)=\Cov(M_{\mathrm{doc}})+\Cov(\varepsilon_i)$. Summing these
covariances and dividing by $b$ proves the formula. Pairwise zero
cross-covariances suffice for this calculation. If, in addition,
$\Cov(\varepsilon_i)=\Sigma_\varepsilon$ for every $i$, it specializes to
$\chi_b=b\Cov(M_{\mathrm{doc}})+\Sigma_\varepsilon$.

A finite stretch of enhanced susceptibility can therefore reflect document
heterogeneity, short-range correlations, or a long-range sector. Resolving
these possibilities requires the specified limit and joint-law information.
The segment measurements use the nonstationary exact
Equation~\eqref{model:eq:chi-general}. Finite covariance-shape fits are predictive
diagnostics and do not estimate a thermodynamic critical exponent.

More generally, let $\mathcal H$ record the document or corpus environment,
$m_i=\E[Y_i\mid\mathcal H]$, and
$C_{ij}=\Cov(Y_i,Y_j\mid\mathcal H)$, with all $Y_i$ square integrable.
The law of total covariance gives the source-conditioned identity
\begin{equation}
 \chi_b=\frac1b\Cov\!\left(\sum_{i=1}^b m_i\right)
       +\frac1b\E\!\left[\sum_{i,j=1}^b C_{ij}\right].
 \label{model:eq:latent-conditional}
\end{equation}
To prove it, apply total covariance to $Z=\sum_iY_i$:
$\Cov Z=\Cov(\E[Z\mid\mathcal H])+\E[\Cov(Z\mid\mathcal H)]$.
Conditional expectation is linear, and bilinearity expands the conditional
covariance into the double sum. Division by $b$ gives the displayed formula.
This retains position-dependent means, heterogeneous residual covariances
and conditional cross-position dependence. In particular, discarding the
second sum's off-diagonal terms needs its own zero-covariance hypothesis.
Source heterogeneity is therefore an explicit part of the declared RG law,
not evidence of a critical limit by itself.

Finally, a finite fixed network on a finite context space has no divergent
finite-source partition function by Proposition~\ref{model:prop:regularity}.
A thermodynamic transition requires a controlled sequence of models,
context extents, or correlated ensembles. The training construction in
Section~\ref{model:sec:training} supplies the evolving law on which such a
limit can be defined; it does not by itself assert that this limit is
critical.

\subsection{Thermodynamic families and exponents from an RG}
\label{model:sec:thermodynamic}
A marginal temporal point occurs at $q=1$ in the homogeneous
source-free comparison law $E_{t+1}=qE_t$: blocking $b$ updates sends
$q$ to $q^b$, with logarithmic-scale beta function $q\log q$ and
temporal decay scale $-1/\log q$ for $0<q<1$.
Indeed, $q(\ell)=q_0^{e^\ell}$ differentiates to
$dq/d\ell=q\log q$, and $q^t=\exp(t\log q)$ fixes that decay scale.
Its divergence as $q\to1$ concerns a specified temporal comparison
model. It does not identify the endogenous PLDR training law, a
singularity across model sizes, or attraction of training to $q=1$.
The temporal cocycle and the thermodynamic construction consequently
retain separate scales and hypotheses.

A model-wide thermodynamic family specifies architectures indexed by
$N$, their parameter normalization, data laws, retained intensive fields
$q_N$, and a control parameter $g$. It also specifies the training law:
either a suitable invariant law, if one exists for the declared update,
or a controlled sequence of finite training horizons $t_N$.
A scheduled optimizer with an increasing counter is not automatically a
stationary physical process. Its finite-time distributions remain valid
objects in Equation~\eqref{model:eq:emission}.
The declared constant-rate, clipped and decayed native family has
invariant laws for its limiting bias-correction kernel at each fixed width
by Proposition~\ref{model:prop:finite-stationarity}. That existence result
neither selects a unique stationary family across widths nor establishes
relaxation of the measured finite trajectories.

For a scalar collective field one candidate thermodynamic functional is
\begin{equation}
 f_N(g,j)=\frac1N\log\E_{\mu_{N,g}}
                    e^{Njq_N},\qquad
 \partial_j^2 f_N(g,0)=N\Var_{\mu_{N,g}}(q_N).
 \label{model:eq:thermodynamic-functional}
\end{equation}
The derivative identity follows by differentiating the moment generating
function. The observation units and the meaning of $N$ must remain fixed
by the family definition. A thermodynamic assertion additionally
establishes an appropriate limiting functional, its singular behavior,
and the associated collective scale. Neither a large finite curvature
nor a change of observable normalization proves that limit. Training
mixtures and fixed-checkpoint laws are different choices of $\mu_{N,g}$;
Equation~\eqref{model:eq:training-mixture} resolves their covariance difference.
No equilibrium interpretation of a learned PLGA matrix is required by
\eqref{model:eq:thermodynamic-functional}.

\begin{proposition}[A singular limit induced by a finite training mixture]
\label{model:prop:mixture-potential}
Let $\pi_s>0$ be a probability distribution on finitely many checkpoints,
let $A_s(j)=\log\E_\D e^{j\Phi_{\theta(s)}}$ for a bounded scalar
observation, and draw independent contexts sharing one checkpoint. The
block-average potential satisfies
\begin{equation}
 f_b(j)=\frac1b\log\sum_s\pi_s e^{bA_s(j)}
       \longrightarrow \max_s A_s(j).\label{model:eq:mixture-potential}
\end{equation}
If the conditional means $m_s=A_s'(0)$ are not all equal, the limiting
function has right derivative $\max_s m_s$ and left derivative
$\min_s m_s$ at zero, despite each conditional $A_s$ being analytic.
\end{proposition}
\begin{proof}
Conditional independence gives the displayed finite-$b$ expression.
Let $M(j)=\max_s A_s(j)$ and $\pi_{\min}=\min_s\pi_s$. The weighted sum
lies between $\pi_{\min}e^{bM(j)}$ and $e^{bM(j)}$. Taking its logarithm
and dividing by $b$ bounds $f_b(j)-M(j)$ between
$b^{-1}\log\pi_{\min}$ and zero, proving the limit. Since the collection
is finite, $A_s(j)=m_sj+O(j^2)$ with a uniform local bound. Maximizing
this expansion gives the stated one-sided derivatives.
\end{proof}

This nonanalyticity comes from the shared training-state mixture and its
unequal inference means. Its finite-$b$ curvature is exactly $A+bB$ from
Theorem~\ref{model:thm:training-mixture}. Here the growing size is a number of
conditionally independent document draws, not the number of interacting
model degrees of freedom. Even a singular averaged generating function
therefore requires its ensemble and physical scale to be identified
before it can establish an intrinsic continuous critical transition of
PLDR-LLM. The entropy mixture evaluated in Section~\ref{model:sec:training-results},
with its finite-law methods in Appendix~\ref{model:sec:auxiliary-methods},
supplies a concrete finite law to which this proposition applies.

\begin{proposition}[Critical exponents on a relevant RG orbit]
\label{model:prop:rg-exponent}
Suppose a physically specified RG changes a length scale by $s>1$ and
has a relevant scalar coordinate $r$ with $r'=\lambda r$, $\lambda>1$.
Assume positive correlation length and susceptibility functions satisfy
$\xi(\lambda r)=\xi(r)/s$ and
$\chi(\lambda r)=s^{-y_\chi}\chi(r)$ on this orbit. Then for $r_0>0$ along
$r_n=r_0\lambda^{-n}$,
\begin{equation}
 \nu=\lim_{n\to\infty}\frac{\log\xi(r_n)}{-\log r_n}
      =\frac{\log s}{\log\lambda},\qquad
 \gamma_\chi=\lim_{n\to\infty}\frac{\log\chi(r_n)}{-\log r_n}
      =y_\chi\nu.\label{model:eq:rg-exponents}
\end{equation}
\end{proposition}
\begin{proof}
Iteration gives $\xi(r_n)=s^n\xi(r_0)$ and
$\chi(r_n)=s^{ny_\chi}\chi(r_0)$. Their logarithms divided by
$n\log\lambda-\log r_0$ have the displayed limits. Values between this
geometric orbit, nonlinear scaling coordinates, and irrelevant-field
corrections require additional control to infer a full asymptotic scaling
function. They are not supplied by the orbit identities alone.
\end{proof}

A usual finite-size hypothesis would then have the form
\begin{equation}
 \chi_N(g)=N^{\kappa}
 \left[F((g-g_c)N^\phi)+O(N^{-\omega})\right],\qquad \omega>0.
 \label{model:eq:fss}
\end{equation}
If $N=L^d$ for an established geometry, the conventional identifications
are $\kappa=\gamma_\chi/(d\nu)$ and $\phi=1/(d\nu)$. Dense attention
and parameter count do not automatically define such a $d$. Equation
\eqref{model:eq:fss} is a conditional model, not a fit to the width measurements in
Sections~\ref{model:sec:criticality-results} and \ref{model:sec:scaling-results}. A critical universality assignment involves the fixed joint law,
its relevant fields, and scaling relations. Finite-size order-parameter
distributions can provide information beyond one fitted slope
\cite{binder1981}. The Gaussian document-block eigenvalues in
Section~\ref{model:sec:flow} belong to another RG and cannot be substituted for
$\lambda$ in Equation~\eqref{model:eq:rg-exponents} without identifying the
same physical scale and relevant coordinate.

There is also an explicit ordering-of-limits issue. If the average
within-checkpoint covariance stays bounded and the between-state
covariance behaves as $B_N\sim N^{-\alpha}B$, Equation
\eqref{model:eq:training-mixture} gives, for shared-checkpoint document blocks
$b_N\sim N^p$,
\begin{equation}
 \chi_{N,b_N}=A_N+b_NB_N
             =A_N+N^{p-\alpha}(B+o(1)).\label{model:eq:double-scale}
\end{equation}
This follows by substitution. Its exponent depends on the chosen joint
limit; it is not automatically the exponent of an interacting critical
sector. The finite training-ensemble calculation in
Section~\ref{model:sec:training-results} instantiates the exact
finite-$N$ identity that precedes this conditional limit; its arithmetic
agreement is not an independent validation of that identity.

\subsection{A testable mathematical condition for self-organization}
\label{model:sec:selforganization}
A self-organized training transition requires both an independently
identified critical surface and training dynamics that approach it.
The effective control can be a function of the complete augmented state;
it need not equal an externally scheduled learning rate. A return toward
an optimizer stability boundary alone does not establish the singular
fluctuation law in Equation~\eqref{model:eq:thermodynamic-functional}. Full-batch
gradient descent can approach an edge of stability under ordinary
optimization dynamics \cite{cohen2021}. The separation of a critical
surface from the feedback selecting it is likewise central in analyses
of self-organized criticality \cite{dickman1998,dickman2000}.

The following normal form gives a quantitative sufficient criterion once
such an effective recursion has actually been derived. It is not assumed
to be the AdamW equation for PLDR-LLM.
\begin{proposition}[Affine feedback and a shrinking critical window]
\label{model:prop:feedback}
Let the signed distance to a known critical surface evolve as
\begin{equation}
 r_{t+1}=(1-a_N)r_t+d_N+\sigma_N\zeta_t,
 \qquad 0<a_N<2,\label{model:eq:feedback}
\end{equation}
where the innovations are independent and identically distributed,
centered, of unit variance, and
independent of the past. Its stationary finite-variance law has
\begin{equation}
 \E r=\frac{d_N}{a_N},\qquad
 \Var r=\frac{\sigma_N^2}{a_N(2-a_N)}.\label{model:eq:feedback-variance}
\end{equation}
For a critical window of width $N^{-\phi}$, the condition
\begin{equation}
 N^{2\phi}\left[\left(\frac{d_N}{a_N}\right)^2+
            \frac{\sigma_N^2}{a_N(2-a_N)}\right]\longrightarrow0
 \label{model:eq:feedback-window}
\end{equation}
is sufficient for $N^\phi r\to0$ in probability in stationarity.
\end{proposition}
\begin{proof}
Set $u_t=r_t-d_N/a_N$ and $c=1-a_N$, so $|c|<1$ and
$u_{t+1}=cu_t+\sigma_N\zeta_t$. The infinite past series
$u_t=\sigma_N\sum_{k\geq0}c^k\zeta_{t-1-k}$ converges in mean square
because $\sum_kc^{2k}<\infty$. It defines a stationary solution with
zero mean and variance $\sigma_N^2/(1-c^2)$, which is the displayed
formula. For any finite-variance initial state, the homogeneous transient
$c^tu_0$ vanishes in mean square, giving the same limiting moments.
Finally, Markov's inequality gives
$\Pr(|N^\phi r|>\varepsilon)\leq
N^{2\phi}\E r^2/\varepsilon^2$, and the stated condition sends this bound
to zero.
\end{proof}

A finite training horizon must additionally control the scaled transient
$N^\phi(1-a_N)^{t_N}u_0$. The result exposes distinct requirements:
restoring feedback, small bias from the critical surface, sufficiently
small fluctuations relative to its shrinking window, and enough time to
relax. It does not assert that slow drive, vanishing noise, or a particular
conservation law is a universal mechanism for all forms of
self-organization. A PLDR training transition would need its own derived
control and dynamics satisfying the appropriate conditions. The completed trajectories measure finite inference emissions and
optimizer interventions. They do not identify this critical feedback mechanism.

\subsection{Closed covariance flows with retained relaxation modes}
A finite effective description of a correlated field can retain its
lag-zero covariance separately from its relaxation poles. This distinction
is essential because averaging an autoregressive field produces memory.
The following is a conditional covariance RG; it does not assume that all
PLDR fields have geometric correlations.

\begin{proposition}[Geometric covariance blocking]\label{model:prop:geometric-blocking}
Let a centered stationary vector field have covariance
\[
 C(0)=C_0,\qquad C(r)=B+\sum_{a=1}^k T_a\rho_a^{r-1},\quad r\ge1,
 \qquad C(-r)=C(r)^{\mathsf T},
\]
where $B=B^{\mathsf T}$, $0<\rho_a<1$, and the entire covariance sequence
is positive semidefinite. For nonoverlapping blocks
$Z_j=b^{-H}\sum_{i=0}^{b-1}Y_{bj+i}$, the same representation holds with
\begin{gather}
 B'=b^{2-2H}B,\quad\rho'_a=\rho_a^b,\quad
 T'_a=b^{-2H}\left(\frac{1-\rho_a^b}{1-\rho_a}\right)^2T_a,
 \label{model:eq:pole-flow}\\
 C'_0=b^{-2H}\left[bC_0+b(b-1)B+
 \sum_a\sum_{r=1}^{b-1}(b-r)\rho_a^{r-1}(T_a+T_a^{\mathsf T})\right].
 \label{model:eq:pole-zero}
\end{gather}
These maps compose on aligned blocks. A stationary Gaussian field with
this covariance remains Gaussian, so the retained covariance state then
specifies its complete coarse law.
\end{proposition}
\begin{proof}
For blocks separated by $n\ge1$ all fine lags $bn+j-i$ are positive.
Their constant component sums to $b^2B$. For a relaxation mode,
\[
 \sum_{i,j=0}^{b-1}\rho^{bn+j-i-1}
 =\rho^{b(n-1)}\left(\sum_{i=0}^{b-1}\rho^i\right)^2.
\]
Multiply by $b^{-2H}$ to obtain the lag-positive formulas. Within one
block, the diagonal gives $bC_0$ and there are $b-r$ pairs of each
orientation at lag $r$, proving the lag-zero formula. Successive blocking
is the same linear sum as direct aligned blocking, so the covariance maps
compose. Linear images preserve Gaussianity and positive semidefiniteness.
\end{proof}

In particular, the residual length $\xi_a=-1/\log\rho_a$ obeys
$\xi'_a=\xi_a/b$. For one symmetric mode $T=\rho A$ and $C_0=B+A$,
the fine field is the covariance of a shared Gaussian component plus an
AR(1) residual. Its blocked lag-zero term must still be retained; imposing
$T'=\rho' (C'_0-B')$ would generally change the law. This is an explicit
finite collective state with compatible maps and residual memory.
The general hierarchy in Equation~\eqref{model:eq:three-sectors} includes
nongeometric and nonstationary residuals. Observing a large shared sector
does not remove that freedom.

\subsection{A normalized width family and a regular thermodynamic reference}
The finite family keeps $L=5$, head dimension 64, generator width 170,
and $w=64h$, with FFN width $\lfloor8w/3\rfloor$. Common observations
average over heads and use the same vocabulary projections at every width.
In a second finite parameterization, wide linear weights are scaled at
initialization by the square root of reference fan-in divided by actual
fan-in, with reference width 128 and FFN width 341. Wide-model parameters
outside the fixed-width generator use learning rate $\eta_0\,128/w$;
generator parameters retain $\eta_0$. This specifies a reproducible
finite family, without asserting that every generator derivative has a
uniform width bound.

An inverse-width learning rate controls a coherent direct output update
under bounded Adam ratios. This learning-rate calculation is separate from
the initial variance convention. Proposition~\ref{model:prop:shape-initialization}
derives a shape-aware law that keeps the initial per-head units
nondegenerate across widths.

\begin{proposition}[Bounded Adam ratio and direct output displacement]
\label{model:prop:adam-normalization}
Let $m_0=v_0=0$, $m_t=\beta_1m_{t-1}+(1-\beta_1)g_t$ and
$v_t=\beta_2v_{t-1}+(1-\beta_2)g_t^2$, with
$0<\beta_1^2<\beta_2<1$. Write the bias-corrected quantities as
$\widehat m_t=m_t/(1-\beta_1^t)$ and
$\widehat v_t=v_t/(1-\beta_2^t)$. For $\epsilon>0$,
\begin{equation}
 \frac{|\widehat m_t|}{\sqrt{\widehat v_t}+\epsilon}
 \le c_t:=\frac{1-\beta_1}{\sqrt{1-\beta_2}}
 \frac{\sqrt{1-\beta_2^t}}{1-\beta_1^t}
 \sqrt{\sum_{k=0}^{t-1}(\beta_1^2/\beta_2)^k}.\label{model:eq:adam-bound}
\end{equation}
For a direct vocabulary-weight update at fixed hidden vector $x\in\R^w$
with learning rate $\eta_0\,128/w$, the adaptive part of each logit
increment is bounded by $128\eta_0c_t w^{-1}\sum_i|x_i|$.
\end{proposition}
\begin{proof}
Expand $m_t=(1-\beta_1)\sum_{k=0}^{t-1}\beta_1^k g_{t-k}$ and
$v_t=(1-\beta_2)\sum_{k=0}^{t-1}\beta_2^k g_{t-k}^2$. Weighted
Cauchy--Schwarz bounds the square of the first sum by
$(\sum_k(\beta_1^2/\beta_2)^k)(\sum_k\beta_2^k g_{t-k}^2)$.
Apply the bias corrections. If $v_t=0$ every contributing gradient and
$m_t$ are zero; otherwise divide and use the positive denominator offset.
For the second claim sum the coordinatewise update bounds multiplied by
$|x_i|$. Decoupled weight decay adds its explicit deterministic linear
term; changes in $x$ are governed by the full graph transport and are not
part of this direct-output bound.
\end{proof}

The ratio bound holds pathwise, including clipped gradients. It does not
make generator dynamics width-independent. A regular thermodynamic
reference can be stated precisely for any intensive head field
$q_h=h^{-1}\sum_{a=1}^h u_{h,a}$ in common units.

\begin{proposition}[Regular cumulant-density limit]\label{model:prop:regular-limit}
Let $U_h=\sum_a u_{h,a}$ be bounded for each $h$. Suppose for all $n\ge1$
that $h^{-1}\kappa_n(U_h)\to c_n$ and
$|h^{-1}\kappa_n(U_h)|\le n!C^n$ uniformly in $h$. Then near zero,
\begin{equation}
 f_h(j)=h^{-1}\log\E e^{jU_h}\longrightarrow
 f(j)=\sum_{n\ge1}\frac{c_nj^n}{n!},\qquad |j|<C^{-1},
\end{equation}
with local uniform convergence of derivatives. In particular the limit is
analytic there and $h\Var(q_h)\to c_2$ is finite.
\end{proposition}
\begin{proof}
The uniform coefficient bound dominates the cumulant series on any disk
$|j|\le r<C^{-1}$ by $\sum_n(Cr)^n$. Its power series equals the finite
law's log generating function near zero and extends it along the real
interval by analyticity. Dominated convergence gives the displayed limit.
On a smaller disk the derivative series are dominated by the corresponding
convergent differentiated geometric series, proving derivative convergence.
Finally $f_h''(0)=\kappa_2(U_h)/h=h\Var(q_h)$.
\end{proof}

\begin{corollary}[Gaussian fluctuations under the extensive-cumulant bound]
\label{model:cor:head-gaussian}
Under Proposition~\ref{model:prop:regular-limit}, let
$Z_h=(U_h-\E U_h)/\sqrt h$. Then $Z_h$ converges in distribution to
a centered Gaussian of variance $c_2$, with a point mass when $c_2=0$.
For a vector field the corresponding conclusion holds if the same
hypotheses apply to every fixed scalar projection and the normalized
covariance matrices converge.
\end{corollary}
\begin{proof}
Centering removes the first cumulant. For fixed real $t$, the absolute
contribution of orders $n\ge3$ to the log moment generating function
of $Z_h$ is bounded, once $C|t|<\sqrt h$, by
\[
 h\sum_{n\ge3}(C|t|/\sqrt h)^n
 =\frac{C^3|t|^3}{\sqrt h(1-C|t|/\sqrt h)}.
\]
This tends to zero, while the quadratic term tends to $c_2t^2/2$.
The cumulant series is valid in this neighborhood by the same analytic
argument as in Proposition~\ref{model:prop:regular-limit}. Convergence of
moment generating functions near zero gives the scalar distributional
limit. Applying this result to every scalar projection and using
Cram\'er--Wold gives the vector result.
\end{proof}

These are conditional universality statements when the hypotheses hold
at fixed $(x,\mathcal C)$. Averaging over contexts can give a mixture
of Gaussians with different covariance matrices. Neither bounded
susceptibility nor the influence bound alone proves the higher-cumulant
hypotheses, and the finite initialization ensembles do not determine a limiting
order-parameter distribution.

This is a sufficient regular limit, not an assumption established by a finite
collection of widths and initializations. Shared training or document fields may
violate its extensive-cumulant hypothesis. The measured families therefore
supply finite coefficients, intervention effects and approximation errors;
no thermodynamic singularity is inferred by multiplying their variance by
parameter count.

\section{Conditional Gaussian limits of signed operators}
\label{model:sec:native-sign-limit}
The signed limit requires complete conditional state-law invariance under
head signs, a fixed observation dimension, a uniform head bound, and joint
convergence of the invariant observation and covariance at the chosen times
$T_N$. It yields a Gaussian mixture with possibly random or singular
covariance. These premises are stronger than a finite sign-orbit check;
they do not assert independence of trained heads or critical dynamics.

The orbit average uses independent uniform head signs under the assumed
invariance of the complete conditional state law. The observation dimension
and context panel remain fixed as $N$ grows, and the sign-invariant observation
$H_N$ and covariance $C_N$ must converge jointly at the chosen times $T_N$.
Conditional on the limiting covariance environment, the signed observation
is Gaussian. Averaging over that environment gives a covariance mixture,
which need not be a single Gaussian. These sign averages do not assert
independence of trained heads or supply additional independent training
initializations. See the guide to
\hyperref[sec:physical-terminology]{physical terminology and renormalization objects}.

\begin{theorem}[Native signed-operator Gaussian mixture]
\label{model:thm:native-sign-gaussian-mixture}
At a selected training time $T_N$, let
$v_{N,a}=d^{-1}\operatorname{vec}G_{\ell a,T_N}\in\R^q$, where
$q=d^2$ is fixed. More generally use a fixed linear observation of a
finite context panel that receives the same sign on head $a$.
Assume the complete conditional state law is invariant under independent
head signs, and $\|v_{N,a}\|\le B$ uniformly in $N$ and $a$.
Define
\[
 Z_N=\frac1{\sqrt N}\sum_{a=1}^N v_{N,a},\qquad
 C_N=\frac1N\sum_{a=1}^Nv_{N,a}v_{N,a}^{\mathsf T}.
\]
Let $H_N$ be any sign-invariant observation taking values in a fixed
Polish space. If $(H_N,C_N)$ converges weakly to $(H,C)$, then
\begin{equation}
 (H_N,C_N,Z_N)\ \Longrightarrow\ (H,C,C^{1/2}W),
 \label{model:eq:native-sign-mixture-limit}
\end{equation}
where $W$ is a standard $q$-dimensional Gaussian independent of $(H,C)$.
The covariance $C$ may be singular or random. At finite $N$, averaging
over the head-sign orbit gives the exact characteristic function
\begin{equation}
 \varphi_N(t)=\prod_{a=1}^N
           \cos\!\left(\frac{t^{\mathsf T}v_{N,a}}{\sqrt N}\right).
 \label{model:eq:sign-orbit-characteristic}
\end{equation}
If $\max_a|t^{\mathsf T}v_{N,a}|/\sqrt N\le1/2$, then
\begin{equation}
 \left|\varphi_N(t)-e^{-t^{\mathsf T}C_Nt/2}\right|
 \le\frac1{3N^2}\sum_a(t^{\mathsf T}v_{N,a})^4
 \le\frac{\|t\|^4B^4}{3N}.
 \label{model:eq:sign-characteristic-bound}
\end{equation}
Under Corollary~\ref{model:cor:head-sign-uniform-concentration}, the required
operator bound is native and uniform along the chosen horizons.
\end{theorem}
\begin{proof}
For a fixed state, independent uniform signs $\zeta_a$ give
\[
 2^{-N}\sum_{\zeta\in\{-1,1\}^N}
 e^{it^{\mathsf T}N^{-1/2}\sum_a\zeta_av_{N,a}}
 =\prod_a\cos(t^{\mathsf T}v_{N,a}/\sqrt N).
\]
Both $H_N$ and $C_N$ are unchanged by this action. Invariance of the
state law therefore gives the same formula inside expectations
multiplied by any bounded continuous $f(H_N,C_N)$.

For $|u|\le1/2$, positivity of cosine permits its real logarithm.
The inequalities $\sin u\le u$ and
$\cos u\ge1-u^2/2\ge7/8$ for $0\le u\le1/2$ give
$\tan u\le2u$. Since
$(\tan u-u)'=\tan^2u$, integration yields
$0\le\tan u-u\le4u^3/3$. A second integration and evenness give
\[
 0\le-\log\cos u-u^2/2\le u^4/3.
\]
Summing over heads writes the product as
$e^{-t^{\mathsf T}C_Nt/2-r_N}$, with
$0\le r_N\le(3N^2)^{-1}\sum_a(t^{\mathsf T}v_{N,a})^4$.
Using $1-e^{-r}\le r$ proves the first bound; Cauchy--Schwarz and
$\|v_{N,a}\|\le B$ prove the second.

For every fixed $t$, its domain condition holds for all sufficiently
large $N$. Multiplication by $f$ and averaging thus replaces the
characteristic function by
$\E[f(H_N,C_N)e^{-t^{\mathsf T}C_Nt/2}]$ with error tending to zero.
Weak convergence of $(H_N,C_N)$ gives the limit
$\E[f(H,C)e^{-t^{\mathsf T}Ct/2}]$.
The sign identity also gives
$\E\|Z_N\|^2=\E\tr C_N\le B^2$, proving tightness of $Z_N$.
Together with tightness of $(H_N,C_N)$, these mixed characteristic
functions identify every subsequential joint limit as
$(H,C,C^{1/2}W)$. This proves the stated convergence.
\end{proof}

The fixed context, row/common observations and predictive probability
vectors can be included in $H_N$, since the sign action preserves them.
Their compact observation spaces and the bounded positive semidefinite
matrices $C_N$ provide convergent subsequences. The conclusion applies
to those subsequences; it does not prove a unique selected covariance
law. Independent decoder sign groups similarly give a conditionally
block-diagonal Gaussian covariance across decoders: apply the theorem
to $LN$ vectors, each multiplied by $\sqrt L$ and embedded in its own
decoder block. Their covariance blocks can remain jointly dependent
through the invariant state.

Conditional sign blocking has an exact composition rule. For disjoint
blocks of $b$ heads, normalize each block sum by $b^{-1/2}$. Given the
complete orbit, the signs in disjoint blocks are independent, so their
characteristic functions multiply. A further block of $c$ such sums has
normalization $(bc)^{-1/2}$, independent of the order of blocking.
Covariances average under this operation and the fourth-cumulant
correction below decreases on the declared fluctuation scale when no
single head dominates. The retained orbit still evolves by the native
quotient successor. A finite covariance matrix by itself is not asserted
to close that training evolution.

\begin{proposition}[Finite sign-orbit fourth-moment budget]
\label{model:prop:sign-orbit-fourth-budget}
Fix scalar head projections $x_a$ and let
$Y=N^{-1/2}\sum_a\zeta_ax_a$ with independent uniform signs.
For $\sigma_N^2=N^{-1}\sum_ax_a^2$,
\begin{equation}
 \E_\zeta Y^2=\sigma_N^2,\qquad
 \E_\zeta Y^4=3\sigma_N^4-\frac2{N^2}\sum_ax_a^4.
 \label{model:eq:sign-fourth-budget}
\end{equation}
When the second moment is positive, its orbit excess kurtosis is
$-2/N_{\rm eff}$, where
\[
 N_{\rm eff}=\frac{(\sum_ax_a^2)^2}{\sum_ax_a^4}\in[1,N].
\]
If the invariant state is also random, its unconditional scalar fourth
cumulant is
\begin{equation}
 \kappa_4(Y)=3\Var(\sigma_N^2)
               -\frac2{N^2}\E\sum_ax_a^4.
 \label{model:eq:sign-mixture-fourth-budget}
\end{equation}
These formulas also hold exactly for a finite uniform mixture of saved
sign orbits, using population moments for that mixture.
\end{proposition}
\begin{proof}
In the square, only two identical sign indices survive expectation.
In the fourth power, only four identical indices or two pairs survive,
giving $N^{-2}(\sum_ax_a^4+6\sum_{a<b}x_a^2x_b^2)$.
Expanding $3(\sum_ax_a^2)^2$ gives the stated fourth moment.
Division by $\sigma_N^4$ gives the excess kurtosis.
Nonnegative cross terms prove $N_{\rm eff}\ge1$;
Cauchy--Schwarz gives $(\sum_ax_a^2)^2\le N\sum_ax_a^4$.
Finally average the fourth-moment identity over the invariant state
and subtract $3(\E\sigma_N^2)^2$.
\end{proof}

If $|x_a|\le B_x$ and $\sigma_N^2\ge c>0$, then
$\sum_ax_a^4\le B_x^2\sum_ax_a^2$ gives
$N_{\rm eff}\ge Nc/B_x^2$. The conditional standardized fourth
correction is then $O(N^{-1})$. A large head count alone does not give
this conclusion: if one projection is nonzero and all others vanish,
$N_{\rm eff}=1$ and the standardized sign remains non-Gaussian, even
though its unstandardized $N^{-1/2}$ sum converges to zero.

\begin{corollary}[A nondegenerate regular rate at zero generator control]
\label{model:cor:zero-control-regular-rate}
Use the native normalized real initialization and zero generator control,
so all metric-learner and PLGA parameters stay fixed while the body may
train. The output PLGA biases start at zero. Suppose the initial output
coupling matrices have one bounded, width-independent law with
$\E\sigma_{\min}(a_{\ell h,0})^2>0$, and retain the conditional head-sign
invariance above. Then constants $0<c\le C<\infty$, independent of width,
training horizon and valid source stream, satisfy
\begin{equation}
 c\le\chi_{N,T}^G\le C,\qquad
 \frac cN\le\frac1{d^2}\E\left\|\frac1N\sum_hG_{\ell h,T}\right\|_F^2
            \le\frac CN.
 \label{model:eq:zero-control-regular-rate}
\end{equation}
Thus the RMS signed-operator average has the regular rate
$\Theta(N^{-1/2})$ throughout this frozen-generator family.
\end{corollary}
\begin{proof}
The bounded terminal LayerNorm and frozen power-layer parameters put
every positive base in one compact interval bounded away from zero,
and bound every exponent. Hence some constant $m_->0$ bounds every
entry of the powered potential from below, uniformly in the incoming
query and training time. The unchanged zero output bias gives
$G_{\ell h,T}=a_{\ell h,0}P^{\rm pot}_{\ell h,T}$. Therefore
\[
 \frac{\|G_{\ell h,T}\|_F^2}{d^2}
 \ge\frac{\sigma_{\min}(a_{\ell h,0})^2}{d^2}
             \|P^{\rm pot}_{\ell h,T}\|_F^2
 \ge m_-^2\sigma_{\min}(a_{\ell h,0})^2.
\]
No independence of the current potential and initial coupling is used.
Averaging this inequality in the exact sign-covariance identity gives
$c=m_-^2\E\sigma_{\min}(a_{\ell h,0})^2>0$.
The uniform upper envelope gives $C$ and the zero-mean identity gives
the second pair of bounds. The normalized real Xavier matrix has a
width-independent continuous density and bounded support. Its
singular matrices form the zero set of the nonzero determinant
polynomial, so it is invertible almost surely and satisfies the stated
strictly positive finite expectation.
\end{proof}

The $N^{-1/2}$ amplitude is a regular averaging normalization, not a
critical exponent. A non-Gaussian unconditional operator distribution
can arise from the random covariance in
Equation~\eqref{model:eq:native-sign-mixture-limit}. Neither such a mixture
nor a nonzero fourth-moment ratio identifies a critical singularity.
Sign-orbit integration supplies no additional independent training
realizations, and numerical sign equivariance retains its separate
native execution qualification.

\section{Conditional head limits and collective stability}
\label{model:sec:conditional-heads}

The native architecture has a finite shared subsystem: the residual
metric learner applies the same eight residual units, with input
dimension $d$ and hidden width 170, to every head within a decoder.
The head dimension and the shared subsystem stay fixed as $N$ grows;
the recorded experiments use $d=64$. The native rotary implementation
uses even $d\ge2$.
The minibatch at each update also drives every head. These shared
variables require an explicit conditioning convention in a head-count
limit. Write $\mathcal C$ for the initial shared metric parameters and the
complete realized sequence of training batches, and $\xi_N$ for the
remaining initialization. A conditional family fixes $\mathcal C$ and
varies $\xi_N$; averaging over $\mathcal C$ is a separate operation.
Learned shared parameters and optimizer moments remain endogenous
coordinates of $S_t$. Conditioning on their final values could remove
the collective fluctuation under investigation and is not the convention
used here.

At a selected finite horizon, the random unit is a complete draw of
$\xi_N$, transported by that fixed training history. The observations
are the head mean $Q_N$ and its covariance at fixed $(x,\mathcal C)$.
A width limit must additionally specify the horizon at each $N$ and
the hypotheses that control this covariance; head exchangeability
alone supplies no independence or Gaussian limit.

\subsection{Head exchangeability and the uniform collective mode}
For fixed evaluation context $x$, let $U_{N,a}$ be a vector of bounded
head observations, and let $Q_N=N^{-1}\sum_{a=1}^N U_{N,a}$. The two
fields in decoder $\ell$, for $s\ge2$ allowed keys at the final query, are
\begin{alignat}{2}
 H_{\ell a}&=-\frac{\sum_{i=1}^{s}p_{\ell a,i}\log p_{\ell a,i}}{\log s},
 &\quad P_d&=I_d-\frac{\mathbf1\mathbf1^{\mathsf T}}{d},\notag\\
 R_{\ell a}&=\frac{\langle(P_dA_{\ell a})^{\odot2}\rangle}
 {\max\{\langle A_{\ell a}^{\odot2}\rangle,\epsilon_R\}}.
 \label{model:eq:conditional-head-fields}
\end{alignat}
Here $p$ is last-query attention on its $s$ allowed keys, and
$A\in\R^{d\times d}$ is the residual metric learner output in
Equation~\eqref{model:eq:gram}. The angle brackets average its $d^2$
matrix entries; $I_d$ is the identity and $\mathbf1\in\R^d$ is the
vector of ones. The key count $s$ and head dimension $d$ are distinct;
the reported 64-token head observations use $s=d=64$. For a singleton
allowed support, set $H_{\ell a}=0$. The fixed denominator floor is
$\epsilon_R=10^{-30}$. Thus the row observation concerns $A$, before
the positive PLGA transform $\Alm$. Unnormalized entropy is between zero and $\log s$;
$P_d$ is an orthogonal row-centering projection, which cannot increase
the Frobenius norm. Consequently both displayed fields lie in $[0,1]$
in real arithmetic, at every width and training time.
The conditional susceptibility is
\begin{equation}
 \chi_N^{\rm q}(x,\mathcal C)
   =N\Cov_{\xi_N}(Q_N\mid x,\mathcal C).
 \label{model:eq:quenched-head-chi}
\end{equation}
The superscript records conditioning on the exogenous shared environment;
it does not assert an equilibrium Gibbs law. For the $L$-component
decoder vector, reported scalar susceptibilities are the context-averaged
trace divided by $L$. The corresponding sample statistic, including the factor $N$, is
Equation~\eqref{model:eq:sample-chi-method}.

\begin{lemma}[Conditional head-permutation symmetry]
\label{model:lem:head-symmetry}
In the real-arithmetic, dropout-free native architecture, independently
permuting the attention-head labels within each layer leaves the output
unchanged when the query, key and value output blocks, power-layer head
parameters, and outgoing attention projection columns are permuted
consistently. A permutation-invariant initialization, the common
generator-rate convention, and AdamW with global gradient clipping
preserve this symmetry under any fixed sequence of training batches.
Consequently the conditional law of head observations is exchangeable.
\end{lemma}
\begin{proof}
The query, key and value row-block permutations only relabel their heads.
Gram formation, the shared metric learner, the per-head power layer,
rotary maps, and attention softmax commute with this relabeling. The
inverse column permutation in the outgoing projection restores exactly
the original decoder output. Residual paths and later layers therefore
receive the original boundary state. The parameter transformation is an
orthogonal permutation of coordinates. Differentiation transforms the
loss gradient by the same permutation because the specified forward activations
are smooth. Coordinatewise Adam moment
updates, their bias corrections, and weight decay commute with it; the
global clipping factor is unchanged because the gradient norm is
unchanged. The rates are constant within each permuted parameter group.
Induction over updates and invariance of the initialization law prove
the last assertion. This is a symmetry of the mathematical program;
floating-point reduction order can introduce rounding differences.
\end{proof}

The exchangeable variance identity \cite{kallenberg2021} takes the following fixed-head-unit form.
\begin{proposition}[Conditional head susceptibility]
\label{model:prop:head-susceptibility}
For $N\ge2$ exchangeable square-integrable head vectors, condition on
$(x,\mathcal C)$ and write
$V_N=\Cov(U_{N,1})$ and $C_N=\Cov(U_{N,1},U_{N,2})$. Then $C_N$ is
symmetric and
\begin{equation}
 \chi_N^{\rm q}=V_N+(N-1)C_N,\qquad
 \bm0_{\mathrm{mat}}\preceq\chi_N^{\rm q}\preceq NV_N,\qquad V_N-C_N\succeq\bm0_{\mathrm{mat}}.
 \label{model:eq:head-susceptibility}
\end{equation}
In particular, uniformly bounded $V_N$ and $N\|C_N\|$ imply bounded
conditional susceptibility. Divergence of the ratio to $V_N$ alone is
not divergence in fixed physical units if $V_N$ itself vanishes.
\end{proposition}
\begin{proof}
Exchanging labels 1 and 2 gives $C_N=C_N^{\mathsf T}$. The covariance
of the head sum contains $N$ diagonal terms $V_N$ and $N(N-1)$
off-diagonal terms $C_N$, which gives the identity. The covariance of
$U_{N,1}-U_{N,2}$ is $2(V_N-C_N)$ and is positive semidefinite.
Also $\chi_N^{\rm q}$ is a covariance. Finally,
$NV_N-\chi_N^{\rm q}=(N-1)(V_N-C_N)\succeq\bm0_{\mathrm{mat}}$. The norm bound follows
from the identity and triangle inequality.
\end{proof}

The finite-data estimator uses all off-diagonal head pairs, so its
covariance identity remains valid without assuming that an individual
realization is invariant under head permutation. The experimental
replicates are independently initialized models under fixed $\mathcal C$.
Their heads and repeated evaluations are not independent realizations.
For a vector with one component per decoder, independent head relabeling
in each layer is also a symmetry. The finite one-head reference is averaged
over these relabelings. Its diagonal entries are the per-layer single-head
variances; its off-diagonal entries are $\chi_{N,\ell m}^{\rm q}/N$.
This removes arbitrary cross-layer head-index alignment. Both the
susceptibility and this symmetrized reference retain all cross-layer
collective covariance. The reported trace enhancement uses the ratio
$\tr\chi_N^{\rm q}/\tr V_N$ alongside the absolute trace in common units.

\begin{proposition}[Compatible exchangeable covariance blocking]
\label{model:prop:head-blocking}
For disjoint groups of $b$ exchangeable heads and
$Z_j=b^{-H}\sum_{a\in j}U_a$, the diagonal and off-diagonal block
covariances are
\begin{equation}
 V'=b^{1-2H}[V+(b-1)C],\qquad C'=b^{2-2H}C.
 \label{model:eq:head-blocking}
\end{equation}
These maps compose on aligned groups. They are covariance maps within
a declared head law; they do not identify a separately trained
smaller-width architecture with the blocked larger model.
\end{proposition}
\begin{proof}
Within one group there are $b$ diagonal covariance terms and $b(b-1)$
off-diagonal terms. Between two groups all $b^2$ terms are off diagonal.
Multiplication by $b^{-2H}$ gives the formulas. Applying block sizes
$b$ and then $a$ yields the same weighted sum as one block of size $ab$,
which proves composition. No Gaussian assumption is needed for these
second-moment identities.
\end{proof}

At Gaussian block normalization $H=1/2$, a subset of $b\le N$
heads in the same trained architecture has susceptibility
$\chi_{N,b}^{\rm q}=V_N+(b-1)C_N$. Growth with $b$ at fixed $N$
therefore measures that architecture's exchangeable common sector.
The thermodynamic endpoint instead uses $b=N$ while the law itself
changes with $N$. Its behavior depends on the width dependence of
both $V_N$ and $C_N$. These are different limits even though their
covariance maps are compatible.

\subsection{Initialization units for a nondegenerate head family}
For head dimension $d$ fixed independently of $N$, the head count fixes
residual width $w=dN$ and FFN width $\lfloor8w/3\rfloor$, with fixed
decoder depth $L$ and the complete metric learner.
Parameter count grows quadratically in $N$ at large width. The
thermodynamic normalization below counts head units, each with fixed
internal dimension, rather than treating every dense coupling as another
observation. Extensivity of a limiting law is an additional property of
this convention, not a consequence of the parameter count.

The native program uses Xavier uniform initialization
\cite{glorot2010}: a linear matrix with input and output dimensions
$n_i,n_o$ has entry variance $2/(n_i+n_o)$. Its three-dimensional
PLGA tensors of shape $(N,d,d)$ have fan dimensions $d^2$ and $dN$,
so their entry variance is $2/[d(d+N)]$. Applying another
inverse-square-root fan-in factor to an already initialized square
wide matrix would give variance proportional to $w^{-2}$.
That finite parameterization has different initial units from a
variance-preserving head family.

\begin{proposition}[Shape-aware initial variance]
\label{model:prop:shape-initialization}
Let a centered linear weight have native entry variance $2/(n_i+n_o)$.
For reference dimensions $r_i,r_o>0$, multiply it by
\begin{equation}
 c(n_i,n_o;r_i,r_o)=
 \left[\frac{r_i(n_i+n_o)}{n_i(r_i+r_o)}\right]^{1/2}.
 \label{model:eq:shape-normalization}
\end{equation}
The resulting entry variance is
$[2r_i/(r_i+r_o)]/n_i$. For a deterministic input $x$ independent
of the independent weight entries, the variance of one output coordinate
is $[2r_i/(r_i+r_o)]\|x\|^2/n_i$.
Multiplying the native PLGA head tensors by $\sqrt{(d+N)/(d+2)}$ gives
the identical per-entry uniform law at every $N$, equal to the law at
$N=2$ at the same fixed head dimension $d$.
\end{proposition}
\begin{proof}
Multiplication by $c$ multiplies the weight variance by $c^2$;
cancelling $n_i+n_o$ gives the first formula. Independence removes
cross terms from the variance of $\sum_i W_i x_i$, proving the second.
The same calculation for a head tensor gives $2/[d(d+2)]$.
Scaling a centered uniform distribution scales its interval, so equality
of these variances also gives equality of its uniform one-entry laws.
\end{proof}

Write $r_w=2d$, $r_f=\lfloor8r_w/3\rfloor$, and $r_v$ for the fixed
vocabulary size of the reference architecture. The reference head count
is two; it specifies the initialization units and is held fixed as $N$
varies. In the recorded family, $d=64$, $L=5$, and
$(r_w,r_f,r_v)=(128,341,32000)$. The correction applies to wide linear matrices
and to the three randomly initialized PLGA tensors. Embeddings,
normalization parameters, zero biases and the fixed shared metric learner
retain their native laws. At $N=2$ it is exactly the identity.
The generator learning rate is $3\times10^{-4}g$; other parameters use
$3\times10^{-4}(2/N)$. These choices define the measured training
family. They do not prove an infinite-width Adam law: gradient scaling,
the fixed denominator offset, global clipping and learned shared-state
transport remain part of that problem. The distinction between an
initialization law and a full training limit is essential
\cite{yang2021,yang2023adaptive}.
Sections~\ref{model:sec:criticality-results} and \ref{model:sec:optimizer-scale-results} give the finite family and its moment and clipping diagnostics.

\subsection{A native initialization limit}
\label{model:sec:initialization-limit}
The corrected initial law permits a more specific limit before any
optimizer update. Many-head Gaussian limits provide a useful context for
attention architectures \cite{hron2020}. Here the finite-dimensional
PLGA head remains in the limiting kernel as an explicit nonlinear random
operator; it is not replaced by dot-product attention.

Fix finitely many nonpadding contexts of fixed finite lengths and
concatenate their token positions into an index set of size $T$. All attention and metric
computations still use the appropriate individual context. For a
positive-semidefinite residual kernel $K\in\R^{T\times T}$, draw
independent query, key and value arrays with $d$ independent Gaussian
columns of covariance $K$. Apply the native rotary maps to queries and
keys, and draw independent PLGA head entries $W,P,a$ from the corrected
uniform law. Their biases vanish. Apply the fixed initial shared metric
learner in each context, including its Gram-input normalization and all
eight residual units. The resulting attention matrix $\Pi_\ell$ is
block diagonal across contexts and causal within each context. It
depends on the queries, keys and head parameters, but not on the values.
Define
\begin{equation}
 \mathcal A_\ell(K)=\E[\Pi_\ell K\Pi_\ell^{\mathsf T}],\qquad
 \mathcal L_\epsilon(K)=D_\epsilon(K)^{-1/2}K D_\epsilon(K)^{-1/2},
 \quad D_\epsilon(K)=\diag(K_{ii}+\epsilon).
 \label{model:eq:initial-attention-kernel}
\end{equation}
The expectation includes the finite head law and conditions on the
shared initialization. If $O=\Pi_\ell V$ is its value output, independence
of $V$ gives $d^{-1}\E[OO^{\mathsf T}]=\mathcal A_\ell(K)$.
This integrates values exactly at the covariance level.

Put
\[
 c_{\rm in}=\frac{2r_w}{r_w+r_f},\qquad
 c_{\rm out}=\frac{2r_f}{r_w+r_f},\qquad
 c_{\rm voc}=\frac{2r_w}{r_w+r_v},
\]
and let $\mathcal S(K)_{ij}=\E[\operatorname{SiLU}(Z_i)
\operatorname{SiLU}(Z_j)]$ for $Z\sim N(\bm0,K)$. These constants follow
from the reference dimensions $r_w,r_f,r_v$ specified after
Proposition~\ref{model:prop:shape-initialization}. They are held fixed
as $N$ grows.
The initial residual kernels obey
\begin{align}
 K_0(i,j)&=\mathbf1\{\text{token}_i=\text{token}_j\},\notag\\
 K_{\ell,a}&=\mathcal L_{\epsilon_{\rm LN}}
                  (K_\ell+\mathcal A_\ell(K_\ell)),\notag\\
 K_{\ell+1}&=\mathcal L_{\epsilon_{\rm LN}}\left(
 K_{\ell,a}+c_{\rm out}c_{\rm in}K_{\ell,a}
               \odot\mathcal S(c_{\rm in}K_{\ell,a})\right).
 \label{model:eq:initial-kernel-recursion}
\end{align}
The token-indicator kernel includes repeated words. Native embeddings
have independent standard-normal entries for distinct nonpadding tokens,
are multiplied by $\sqrt w$, and then undergo the input LayerNorm.
Its normalized covariance tends to $K_0$; the initial denominator offset
becomes negligible under this embedding scaling.

\begin{theorem}[Conditional native initial-law limit]
\label{model:thm:initialization-limit}
Use the uncached real-arithmetic native architecture at initialization, with
head dimension $d$ and decoder depth $L$ fixed independently of $N$, the
complete fixed shared metric learner, zero biases, unit normalization
gains, no dropout, and an untied readout. Let $w=dN\to\infty$, use the
shape-aware independent uniform
wide weights and head coefficients, and retain the declared FFN width.
For each fixed finite context collection, the empirical residual kernels
converge in probability to Equation~\eqref{model:eq:initial-kernel-recursion}.
Any fixed finite collection of residual coordinates converges jointly
to centered Gaussian fields with these kernels. The vocabulary logits
converge jointly to a centered Gaussian field with covariance
$c_{\rm voc}K_L$ for each vocabulary coordinate and zero covariance
between distinct vocabulary coordinates.

At each layer, finitely many distinct heads converge to independent
copies of the finite head law used in
Equation~\eqref{model:eq:initial-attention-kernel}, conditional on the shared
initialization. Consequently each bounded head-mean entropy and row
field converges in mean square to the corresponding head-law
expectation. This statement does not give a rate for that convergence
or a limit for $N$ times its variance.
\end{theorem}
\begin{proof}
Write the residual token matrix as $X\in\R^{T\times w}$. The induction
maintains convergence of $XX^{\mathsf T}/w$, vanishing empirical feature
means, and uniformly bounded moments of every fixed coordinate, to any
fixed finite order. The Gaussian embedding law initializes these
properties. Normalized Gaussian coordinates have uniformly bounded
moments: on the event that the empirical embedding variance is at least
one half, the normalized coordinate is bounded by a constant times its
centered Gaussian input. On the complementary event its absolute value
is at most $\sqrt w$, while the chi-square lower-tail bound obtained
from $\E e^{-t\chi_k^2}=(1+2t)^{-k/2}$ decreases exponentially in $w$.
The input denominator offset is nonnegative and does not weaken these
bounds. The ordinary law of large numbers gives the token-indicator
kernel and the Gaussian finite-coordinate limit.

Consider a fresh wide projection with independent centered uniform
entries of variance $c/w$. Conditional on $X$, its output coordinates
are independent weighted sums and have covariance $cXX^{\mathsf T}/w$.
The fourth-moment bound implies
\[
 \max_{i\le w}\|X_{:,i}\|/\sqrt w\longrightarrow0
 \quad\hbox{in probability},
\]
because its fourth power is at most
$w^{-2}\sum_i\|X_{:,i}\|^4$, whose expectation is $O(w^{-1})$.
The sum of the third-order characteristic-function remainders is bounded
by a constant times
\[
 \frac{\max_i\|X_{:,i}\|}{\sqrt w}
       \frac1w\sum_i\|X_{:,i}\|^2.
\]
It vanishes in probability. The product of the conditional characteristic
functions therefore converges to that of the stated Gaussian projection.
Its limiting covariance is deterministic after conditioning on the
shared initialization, so this conditional limit also gives independence
from any previously retained finite collection of variables.

Query, key and value head blocks use independent weights. Conditional
on $X$, the heads are independent, with the same initial head law.
Rotary maps are fixed linear maps. Gram formation, the positive-base
power map and the normalized residual metric learner are continuous.
The head observations $H$ and $R$ are bounded and continuous, including
their fixed row denominator floor. The conditional projection limit
therefore gives their limiting head law. Attention probabilities are
bounded and the value output is a convex combination of finitely many
value coordinates, so all needed output moments remain uniformly
bounded. Conditional averaging over heads consequently gives
\[
 \frac1{dN}\sum_{a=1}^N O_aO_a^{\mathsf T}
       \longrightarrow\mathcal A_\ell(K_\ell)
       \quad\hbox{in probability}.
\]
Uniform moment bounds justify convergence of the corresponding
conditional expectations, in addition to bounded test functions.

The outgoing attention projection has fresh independent weights of
variance $1/w$. Applying the same conditional characteristic-function
argument yields a Gaussian residual increment with covariance
$\mathcal A_\ell(K_\ell)$, independent in the limit of the retained
input coordinates. Its empirical covariance converges to that same
matrix. Conditional independence of its wide output coordinates bounds
the empirical covariance error and the cross term with $X$ by quantities
that vanish in mean square. The residual sum thus has kernel
$K_\ell+\mathcal A_\ell(K_\ell)$ and empirical mean tending to zero.
The positive LayerNorm denominator permits continuous passage to its
normalized kernel, giving $K_{\ell,a}$.

For the two independent GLU input projections, the limiting covariance
is $c_{\rm in}K_{\ell,a}$. Independence gives covariance
$c_{\rm in}K_{\ell,a}\odot\mathcal S(c_{\rm in}K_{\ell,a})$ for
their gated product. The output projection has variance
$c_{\rm out}/m$ at FFN width $m$, giving the displayed FFN contribution.
The same conditional law of large numbers, fresh output projection,
residual addition and normalization prove the next line of the recursion.
Uniform higher moments are preserved: centered bounded-weight sums
have moment bounds controlled by their normalized input second moment;
$|\operatorname{SiLU}(u)|\le|u|$ controls the gate, and each subsequent
LayerNorm divides by at least $\sqrt{\epsilon_{\rm LN}}$.
The number of layers is fixed. This completes the induction.
The independent readout projection with variance $c_{\rm voc}/w$
gives the final Gaussian logit law by the same argument.

Finally, for either bounded head field at a fixed layer and context,
let $m_N(X)$ be its conditional one-head expectation. Conditional
independence gives
\[
 \E|Q_N-m|^2
 =\E\Var(Q_N\mid X)+\E|m_N(X)-m|^2
 \le\frac1{4N}+\E|m_N(X)-m|^2.
\]
The projection limit and boundedness give $m_N(X)\to m$ in mean square.
Both terms vanish. No rate for the second term has been used, which
proves the stated coverage boundary.
\end{proof}

The head operator and all token dependence remain inside
$\mathcal A_\ell$. Contexts are not assumed independent in this
construction. The numerical comparison in Section~\ref{model:sec:initialization-kernel-results}
evaluates its diagonal
context blocks, which suffice for per-context head means and logit
variances. Gaussian residual and logit limits do not make the individual
nonlinear PLGA head Gaussian. Nor does the head-mean law of large numbers
establish an analytic thermodynamic potential, convergence of its
curvature, or the extensive-cumulant hypothesis. Training reuses the
weights and introduces gradient-dependent correlations, so the fresh
projection argument is confined to initialization.

If the shared initialization is also random, boundedness permits
integration of the head-mean mean-square limit over that environment.
Thus $Q_N\to m(\mathcal C)$ in mean square, and
$\Var(Q_N)\to\Var(m(\mathcal C))$. A nonconstant limiting conditional
mean gives an extensive unconditional susceptibility already at
initialization. This instantiates the shared-environment decomposition
without identifying such variation with a trained critical transition.

\subsection{A native generator envelope}
The fixed head dimension and the final normalization in the residual
metric learner give a bound on the actual PLGA operator, independently
of the size of its incoming Gram matrix. Write $\|M\|_{\max}$ for the
largest absolute matrix entry.

\begin{proposition}[Uniform native generator envelope]
\label{model:prop:generator-envelope}
Use the shape-aware initialization, head dimension $d\ge2$ fixed
independently of head count, zero initial Adam moments, generator rate
$\eta_g\ge0$, and weight decay $\lambda_d>0$
with $0\le\eta_g\lambda_d\le1$. Suppose
$0<\beta_1<1$ and $\beta_1^2<\beta_2<1$. Let $c_t$ be the bound in
Equation~\eqref{model:eq:adam-bound}, and set
\[
 K_0=1,\qquad
 K_{t+1}=(1-\eta_g\lambda_d)K_t+\eta_g c_{t+1}.
\]
Then the absolute entries of $W,P,a,b,b_a$ in the power layer and the
affine parameters of the last metric-learner LayerNorm are at most $K_t$.
Define
\begin{align}
 M_t&=d(\sqrt d+1)K_t^2+K_t,\notag\\
 L_t&=\max\{|\log\epsilon_A|,
                  |\log(M_t^2+\epsilon_A)|\}.
\end{align}
For every head, layer and finite input in real arithmetic,
\begin{equation}
 \|G_{\ell a,t}\|_{\max}
 \le dK_t e^{K_tL_t}+K_t.
 \label{model:eq:generator-envelope}
\end{equation}
Moreover $\overline c=\sup_{t\ge1}c_t<\infty$ and
$K_t\le\max\{1,\overline c/\lambda_d\}$, so the envelope is uniform
in head count and training time.
\end{proposition}
\begin{proof}
The corrected PLGA initial entries are centered uniform variables of
variance $2/[d(d+2)]$. Their support has endpoints
$\pm\sqrt{6/[d(d+2)]}$ and lies in $[-1,1]$ for $d\ge2$.
The displayed biases start at zero and the LayerNorm affine parameters
at one and zero. Each AdamW coordinate update is
$(1-\eta_g\lambda_d)\theta-\eta_g r$, with $|r|\le c_{t+1}$.
Induction proves the $K_t$ bound. The geometric sum in
Equation~\eqref{model:eq:adam-bound} is uniformly bounded, and
$1-\beta_1^t\ge1-\beta_1>0$, proving finiteness of $\overline c$.
The recursion for $K_t$ preserves the stated constant upper bound,
including the frozen case $\eta_g=0$.

For one row entering LayerNorm, the centered, variance-normalized
vector has Euclidean norm at most $\sqrt d$. The affine output therefore
has every absolute entry at most $(\sqrt d+1)K_t$. This applies to
$A$ after the final metric-learner residual unit, whatever the earlier
Gram matrix or shared hidden activations. Matrix multiplication and
the bias give $\|WA+b\|_{\max}\le M_t$.
The native positive activation is $f(z)=z^2\sigma(z)$, so
$0\le f(z)\le z^2$. Every entry of $\Alm=f(WA+b)+\epsilon_A$ thus lies
between $\epsilon_A$ and $M_t^2+\epsilon_A$.
Since $|P_{ij}|\le K_t$, each entry of
$\Alm^{\odot P}=\exp(P\odot\log\Alm)$ is at most $e^{K_tL_t}$.
The final multiplication by $a$ and addition of $b_a$ give
Equation~\eqref{model:eq:generator-envelope}.
\end{proof}

The bound is deliberately conservative, especially at large power
coefficients. It establishes a native finite-operator sector under the
declared update, rather than a useful numerical forecast of its magnitude.
Collective susceptibility can still grow through correlations among
bounded heads. Uniform control of derivatives, learned shared-state
influence, and the full training limit also requires more than this
operator envelope.

\subsection{Stationary laws of the limiting finite-width update}
For the declared constant-rate, clipped AdamW family, existence of a
finite-width invariant law can be established separately from relaxation
and uniqueness. This statement averages over the iid minibatch law;
the conditional experiment fixes one complete realization of that law.

\begin{proposition}[Finite-width limiting stationary laws]
\label{model:prop:finite-stationarity}
In real arithmetic, fix a finite native PLDR architecture and a finite iid
minibatch law.
Use global gradient clipping by a continuous map with norm bound $C_g$,
Adam offset $\epsilon_o>0$, $0<\beta_1,\beta_2<1$, and constant
coordinate rates $\eta_i\ge0$ with
$0\le\eta_i\lambda_d\le1$ and $\lambda_d>0$.
Assume the initial parameter law has finite second moment and the
initial moments vanish. Replace the integer bias-correction counter by
$u_{1,t}=\beta_1^t$, $u_{2,t}=\beta_2^t$, and extend their state space
to $[0,1]^2$.
The resulting continuous-state Markov kernel has an invariant probability
measure with finite parameter second moment. Every invariant measure
is supported on $u_1=u_2=0$. Its projection onto weights and Adam moments
is consequently invariant for the limiting kernel with bias corrections
equal to one.
\end{proposition}
\begin{proof}
Every clipped gradient coordinate has absolute value at most $C_g$.
The moment recursions preserve $|m_i|\le C_g$ and $0\le v_i\le C_g^2$.
At the next update, the first bias-correction denominator is
$1-\beta_1u_1\ge1-\beta_1$. The adaptive coordinate ratio is therefore
at most
\[
 B=\frac{C_g}{(1-\beta_1)\epsilon_o}
\]
in absolute value, uniformly over this extended state space. Hence
\[
 |\theta_{i,t+1}|
 \le(1-\eta_i\lambda_d)|\theta_{i,t}|+\eta_iB,
 \qquad
 |\theta_{i,t}|\le\max\{|\theta_{i,0}|,B/\lambda_d\}.
\]
The last inequality includes a frozen coordinate. It gives a uniform
second-moment bound on parameters at this fixed architecture size.
The moment coordinates and $u_1,u_2$ lie in compact boxes, so the
full family of state laws is tight.

The finite native graph has a continuous loss gradient: its LayerNorm
denominators are positive, its power bases are bounded below by
$\epsilon_A>0$, and its SiLU and softmax maps are smooth. Continuous
clipping and the positive Adam denominator preserve continuity of each
minibatch update. Averaging the finitely many updates therefore maps
bounded continuous functions to bounded continuous functions, the Feller
property.

Let $\nu_T=T^{-1}\sum_{t=0}^{T-1}\mathcal L(S_t)$. Tightness gives a
weakly convergent subsequence with limit $\nu$. For every bounded
continuous $f$, telescoping gives
\[
 \left|\int Pf\,d\nu_T-\int f\,d\nu_T\right|
 =\frac{|\E f(S_T)-\E f(S_0)|}{T}
 \le\frac{2\|f\|_\infty}{T}.
\]
The Feller property permits passage to the weak limit, proving
$\nu P=\nu$. The uniform second-moment bound passes to the limit by
lower semicontinuity. This is the occupation-measure existence argument
for Feller kernels \cite{hairer2021}.
Finally, invariance gives $\E_\nu u_i=\beta_i\E_\nu u_i$.
Since $u_i\ge0$ and $\beta_i<1$, both variables vanish almost surely.
The update restricted to this boundary has bias corrections equal to
one, proving the projection statement.
\end{proof}

A head-exchangeable initial law remains exchangeable under each update
by Lemma~\ref{model:lem:head-symmetry}. Averaging in time and taking a weak
limit preserve this invariance under the finite head-permutation group.
The invariant law obtained above can therefore be chosen exchangeable,
consistently with the head-covariance construction.

The boundary $u_1=u_2=0$ represents the limiting update; it is not
an invariant law of an increasing integer counter. The proof supplies
subsequential limits of time-averaged state laws and does not prove
uniqueness or convergence of the individual-time laws. It also gives
no uniform relaxation rate as $N$ grows. A fixed realized minibatch
history, the averaged stationary ensemble, and a joint width--time
limit remain different objects. The measured horizon dependence is
therefore retained explicitly when interpreting conditional head
fluctuations.

\subsection{Paired dynamics and stationary marginal laws}
The paired experiment has its own exact training kernel. With an update
map $T(S,B)$ and minibatch law $\mathcal B$, define
\begin{equation}
 (P_{\rm pair}F)(S,S')
 =\int F(T(S,B),T(S',B))\,d\mathcal B(B).
 \label{model:eq:paired-kernel}
\end{equation}
Both copies use the same batch. The weights and optimizer coordinates of
each copy are retained, so this definition also covers finite state pulses.

The product-kernel marginal identity is classical Markov-kernel algebra \cite{kemeny1976}; here the coupling retains the same PLDR training batch.
\begin{proposition}[Compatible paired training laws]
\label{model:prop:paired-kernel}
For either coordinate projection $\pi_i$, the paired kernel satisfies
\[
 P_{\rm pair}(f\circ\pi_i)=(Pf)\circ\pi_i,
 \qquad
 P_{\rm pair}^{\,b}(f\circ\pi_i)=(P^bf)\circ\pi_i.
\]
Its temporal blocks compose and retain the common driving sequence.
If $\nu$ is invariant for $P$, its diagonal image under $S\mapsto(S,S)$
is invariant for $P_{\rm pair}$. These identities do not assert attraction
of a displaced pair to the diagonal.
\end{proposition}
\begin{proof}
A function of one coordinate in Equation~\eqref{model:eq:paired-kernel}
ignores the other coordinate, giving the first identity. Induction gives
the second. Kernel iteration proves temporal composition with the same
batch used in both copies at every fine update. On the diagonal both
copies remain identical; their common state has law $\nu P=\nu$.
This proves diagonal invariance.
\end{proof}

Stationary marginals and paired contraction are different requirements.
For a direct example, doubling on the unit circle preserves the uniform
law: splitting $\int_0^1 f(2x\bmod1)\,dx$ into two half intervals gives
$\int_0^1f(x)\,dx$. Two initially uniform copies separated by
$1/(3\cdot2^k)$ have identical stationary marginals, but after $k$ steps
their circular distance is $1/3$ forever. Arbitrarily small initial
separations therefore need not contract even with stationary marginals.
This example establishes the logical distinction, without identifying
the native training map with circle doubling.

For differentiable native updates, an infinitesimal displacement of the
full augmented state instead obeys the product of its update Jacobians
along the realized history. Replacing that product by a fixed collective
matrix $F$ requires a local amplitude regime and a validated reduction.
Finite pulse separation under Equation~\eqref{model:eq:paired-kernel} can measure
nonlinear propagation even when the marginal stationary-law theorem
applies. Attraction to a critical surface concerns its independently
identified relevant coordinate, rather than every component of a paired
state difference.

\subsection{The conditional thermodynamic functional}
For the $L$-component head-mean entropy or row-energy field, every
component of $Q_N$ lies in $[0,1]$. Define
\begin{equation}
 f_N^{\rm q}(j\mid x,\mathcal C)
 =\frac1N\log\E_{\xi_N}
   e^{N j^{\mathsf T}Q_N(x,\mathcal C,\xi_N)},
 \qquad
 \overline f_N^{\rm q}(j)=\E_{x\sim\D_e}
                      f_N^{\rm q}(j\mid x,\mathcal C),
 \label{model:eq:conditional-potential}
\end{equation}
where $\D_e$ is the declared evaluation-context law. Differentiation
at zero gives
$\nabla^2\overline f_N^{\rm q}(0)=\E_x\chi_N^{\rm q}(x,\mathcal C)$.
Averaging the exponential over contexts before taking its logarithm
would instead include a document-mean sector. Here $j$ reweights the
conditional initialization ensemble at fixed $(x,\mathcal C)$. The
physical operator gain changes model weights, whereas $g$ changes the
training update. No fluctuation--dissipation identity between these
different sources is assumed.

\begin{proposition}[Compactness and compatible observation potentials]
\label{model:prop:potential-compactness}
Suppose $\|Q_N\|_2\le B$ uniformly in size and evaluation context.
At fixed $(x,\mathcal C)$, the functions $f_N^{\rm q}$ are convex,
vanish at zero, and have Lipschitz constant at most $B$. Every sequence
of sizes tending to infinity has a subsequence on which they converge
locally uniformly to a finite convex $B$-Lipschitz function. The
context-averaged family $\overline f_N^{\rm q}$ has the same properties
and its own subsequential conclusion.

For a fixed linear observation $T$, the potential of $TQ_N$ is
$f_N^{\rm q}(T^{\mathsf T}j)$, and its limit along any such subsequence
is $f^{\rm q}(T^{\mathsf T}j)$. The analogous identity holds for the
context-averaged potential. Successive fixed linear observations remain
compatible with these limits.
\end{proposition}
\begin{proof}
Boundedness permits differentiation under the expectation at every
finite size and real source. The gradient is the exponentially tilted
mean of $Q_N$, so its norm is at most $B$. The Hessian is $N$ times
the tilted covariance and is positive semidefinite. Thus each potential
is convex and $B$-Lipschitz. Its value at zero is zero, giving
$|f_N^{\rm q}(j)|\le B\|j\|_2$.

Choose a countable dense subset of source space. The displayed bound
and a diagonal subsequence argument give convergence at all of its
points. The common Lipschitz bound extends the limit uniquely to a
$B$-Lipschitz function on the source space. On any compact set, a finite
net from the dense subset bounds the uniform difference by the maximum
difference at its net points plus twice $B$ times the net radius.
Taking the subsequence limit and then the radius to zero proves local
uniform convergence. Convexity and the value at zero pass to the limit.
Averaging over contexts preserves the bounds and convexity, so the same
argument applies to $\overline f_N^{\rm q}$. This does not assert a
single subsequence resolving every individual-context limit.

Finally $j^{\mathsf T}TQ_N=(T^{\mathsf T}j)^{\mathsf T}Q_N$ proves
the finite observation identity, also after context averaging. A fixed
linear transpose takes compact sets to compact sets, proving the
asserted convergence and composition.
\end{proof}

For the normalized decoder head-mean vector one can take $B=\sqrt L$,
including at any chosen horizon sequence $t_N$. Fixed vocabulary
probabilities can be retained jointly as additional bounded observations.
The observation identity keeps the original size factor $N$; it does
not identify a separately trained smaller architecture. A finite seed
panel also does not control the exponentially weighted tails needed to
estimate the potential at a fixed nonzero source.

This compactness result supplies possible limiting functionals without
identifying a unique limit, its differentiability, or convergence of
its second derivatives. The regular cumulant-density hypothesis of
Proposition~\ref{model:prop:regular-limit} is much stronger. A critical
interpretation requires a singular collective sector in these fixed
units and the appropriate order of width and training-time limits.

\subsection{Shared environments and the order of averaging}
Let $m_N(x,\mathcal C)=\E[Q_N\mid x,\mathcal C]$. The total covariance
formula gives the exact decomposition
\begin{equation}
 N\Cov(Q_N\mid x)
 =\E_{\mathcal C}\chi_N^{\rm q}(x,\mathcal C)
   +N\Cov_{\mathcal C}(m_N(x,\mathcal C)).
 \label{model:eq:environment-mixture}
\end{equation}
Indeed, subtract the conditional mean, expand the centered outer product,
and condition on $\mathcal C$; the mixed terms have zero expectation.
If $m_N$ converges in mean square to a nonconstant $m(x,\mathcal C)$,
the second term has an extensive leading coefficient even when the first
term remains bounded. A further average over evaluation contexts has
its own conditional-mean contribution. Shared initialization, common
training drive, and shared document structure are therefore separate
sectors in a model-wide limit.

This decomposition does not declare every shared mode noncritical.
An instability of the learned shared state belongs to the endogenous
dynamics and must be retained. Fixing its initial condition and the
external driving sequence removes two specified sources of mixture
variation while allowing that instability to occur. The averaged environmental
sector can also have singular control dependence. Its explicit factor
of $N$ alone does not identify an intrinsic transition among head units;
the conditional and environmentally averaged ensembles answer different
questions.

Fixing the specified environment does not assert that every remaining
initial collective self-averages. For example, the shape-aware readout
retains finite initial logit variance in its declared units. Such an
initial field can transmit initialization dependence to learned shared
coordinates. Equation~\eqref{model:eq:environment-mixture} can be applied again
to an additional specified initial collective. The measured ensemble
integrates over the remaining wide initialization, and its finite-seed
covariance does not identify every possible surviving common mode.

\subsection{A regular limit from bounded initialization influence}
The classical independent-replacement variance bound \cite{efron1981} gives a sufficient
condition directly in terms of a trained observable.
\begin{proposition}[Conditional influence bound]\label{model:prop:head-influence}
Given $(x,\mathcal C)$, suppose the remaining initialization is a product
of $N$ independent blocks $Z_1,\ldots,Z_N$. Let $Q_N^{(a)}$ replace
block $a$ by an independent copy and rerun the same training history.
If
\[
 \E\|Q_N-Q_N^{(a)}\|^2\le J_N^2/N^2\quad\hbox{for every }a,
\]
then $\tr\chi_N^{\rm q}\le J_N^2/2$. A uniformly bounded $J_N$
therefore excludes divergent conditional susceptibility for this field.
\end{proposition}
\begin{proof}
Reveal the independent blocks successively, and decompose
$Q_N-\E Q_N$ into its conditional-expectation martingale differences
$D_a$. Orthogonality gives
$\E\|Q_N-\E Q_N\|^2=\sum_a\E\|D_a\|^2$. At fixed earlier blocks,
$D_a$ is the centered conditional expectation over the later blocks.
The independent-copy identity for variance and conditional Jensen's
inequality imply
$\E\|D_a\|^2\le\E\|Q_N-Q_N^{(a)}\|^2/2$. Sum these inequalities,
use the assumed bound for each replacement, and multiply by $N$.
\end{proof}

The replacement blocks need not be physical heads, but their independence
and normalization must be specified. This is a sufficient bound; it is
not a uniform certificate obtained by inspecting a few initializations.
For PLDR, establishing its hypothesis requires controlling the full
training and emission transport, including the learned shared generator.
The finite direct-output Adam bound alone does not supply that control.

\subsection{A collective stability mode and its temporal RG}
Let $N\geq1$ and let $u_{a,0}\in\mathbb R^d$ be centered random vectors
with finite second moments. Fix deterministic matrices $A,B$ and put
$F=A+B$. Suppose the centered innovations $\xi_{a,t}$ have finite
second moments, covariance $Q$, are mutually independent across heads
and times, and their entire collection is independent of
$(u_{1,0},\ldots,u_{N,0})$. Initial heads need not be independent.
Consider
\[
 u_{a,t+1}=Au_{a,t}+B\bar u_t+\xi_{a,t},\qquad
 \bar u_t=\frac1N\sum_{a=1}^N u_{a,t},\qquad
 \chi_t=N\operatorname{Cov}(\bar u_t).
\]
All assumptions and covariances may instead be imposed conditionally
on one fixed external environment, with $A,B,Q$ fixed in that environment.

\begin{proposition}[Collective covariance and relaxation]
\label{model:prop:collective-stability}
Under the stated hypotheses,
\begin{equation}
 \chi_{t+1}=F\chi_tF^{\mathsf T}+Q,\qquad
 F_b=F^b,\qquad Q_b=\sum_{k=0}^{b-1}F^kQ(F^k)^{\mathsf T}.
 \label{model:eq:collective-temporal-rg}
\end{equation}
For integer blocks $a,b\geq0$, $F_{a+b}=F_bF_a$ and
$Q_{a+b}=F_bQ_aF_b^{\mathsf T}+Q_b$.
For positive integers $a,b$, the blocking transformations
$\mathcal R_b(F,Q)=(F_b,Q_b)$ also obey
$\mathcal R_a\circ\mathcal R_b=\mathcal R_{ab}$.
If $\rho(F)<1$, the limiting covariance, which is the unique fixed
covariance of this recursion, is
$\chi_\infty=\sum_{k\geq0}F^kQ(F^k)^{\mathsf T}$.
For a real left eigenvector $v^{\mathsf T}F=\lambda v^{\mathsf T}$,
$|\lambda|<1$,
\begin{equation}
 v^{\mathsf T}\chi_\infty v=\frac{v^{\mathsf T}Qv}{1-\lambda^2},
 \qquad \tau=-\frac1{\log|\lambda|}\quad(\lambda\ne0).
 \label{model:eq:collective-mode}
\end{equation}
Here $\tau$ is the e-folding time of the conditional mean response
envelope. The stationary covariance statement does not assert
stationarity of a native consuming-corpus training process.
\end{proposition}
\begin{proof}
Let $\mathcal F_t$ contain the initial head vector and all innovations
at times strictly below $t$. The state $\bar u_t$ is
$\mathcal F_t$-measurable. The independence and centering assumptions
give $\mathbb E[\bar\xi_t\mid\mathcal F_t]=\bm0$ and
$\mathbb E[\bar\xi_t\bar\xi_t^{\mathsf T}\mid\mathcal F_t]=Q/N$,
where $\bar\xi_t=N^{-1}\sum_a\xi_{a,t}$. In particular the
cross covariance of $\bar u_t$ and $\bar\xi_t$ is zero.
Average the head recursion, expand the covariance of
$F\bar u_t+\bar\xi_t$, and multiply by $N$ to obtain the
one-step identity. Iteration gives
\[
 \chi_{t+b}=F^b\chi_t(F^b)^{\mathsf T}+Q_b.
\]
Splitting the finite sum defining $Q_{a+b}$ into $k<b$ and $k\geq b$
gives the concatenation formula. Blocking an already $b$-blocked
recursion for $a$ steps gives noise covariance
\[
 \sum_{j=0}^{a-1}\sum_{k=0}^{b-1}
 F^{bj+k}Q(F^{bj+k})^{\mathsf T}=Q_{ab},
\]
and drift $(F^b)^a=F^{ab}$, proving the RG composition law.
When $\rho(F)<1$, powers of $F$
decay geometrically up to a polynomial factor. Consequently the
covariance series converges and $F^t\chi_0(F^t)^{\mathsf T}\to\bm0_{d\times d}$.
The same decay applied to the difference of two fixed covariances
proves uniqueness. Contracting the series with $v$ yields a scalar
geometric series with ratio $\lambda^2$. Iterating the conditional
mean equation multiplies the mode by $\lambda$ at each step,
which gives its stated decay time.
\end{proof}

The probabilistic requirements can be weakened to an adapted
filtration with $\mathbb E[\xi_{a,t}\mid\mathcal F_t]=\bm0$ and
$\mathbb E[\xi_{a,t}\xi_{b,t}^{\mathsf T}\mid\mathcal F_t]
=\delta_{ab}Q$. These conditional moment assumptions give the same
second-order proof. Without them, set
$Q_t=N\operatorname{Cov}(\bar\xi_t)$ and
$C_t=N\operatorname{Cov}(\bar u_t,\bar\xi_t)$; the identity is
\[
 \chi_{t+1}=F\chi_tF^{\mathsf T}+Q_t+FC_t+C_t^{\mathsf T}F^{\mathsf T}.
\]
Thus independence across innovation times alone does not remove
correlations with a random incoming state. Empirical risk increments
are permitted to be correlated and are not assumed to be martingale
differences. A deterministic incoming state removes its initial
cross terms, while cross-interval terms can remain.

Approaching $\lambda=1$ can link increased collective variance to
slow restoration when forcing remains in that mode. Nonlinear
terms control finite-size rounding and the behavior beyond loss of
stability. Application to PLDR requires measured perturbation
transport and a justified conditional noise model. Neither a large
covariance nor moment-reset sensitivity determines $F$ or proves
that unforced training approaches a critical surface.

\subsection{A finite saturation mechanism and a physical training pulse}
An attention head can lose sensitivity through ordinary softmax
saturation. This mechanism is distinct from divergent connected
fluctuations and can be tested within the native graph.

The categorical covariance is the softmax Fisher Hessian \cite{martens2020}; its concentration bound specifies the native attention application.
\begin{proposition}[Local saturation bound]\label{model:prop:saturation}
For $p=\softmax(s)$ on $S$ allowed keys, let $J_p=\diag(p)-pp^{\mathsf T}$
and $\delta=1-\max_i p_i$. Then
\begin{equation}
 \bm0_{S\times S}\preceq J_p,\qquad
 \|J_p\|_2\le\tr J_p=1-\sum_i p_i^2\le2\delta.
 \label{model:eq:saturation-bound}
\end{equation}
If the largest score exceeds every other score by at least $\Delta\ge0$,
then $\delta\le(S-1)e^{-\Delta}$. For the head output $O=V^{\mathsf T}p$
and a local uniform score gain $s\mapsto(1+\alpha)s$,
\begin{equation}
 \left\|\frac{dO}{d\alpha}\bigg|_{0}\right\|
 \le2\delta\,\|V_c\|_2\,\|s_c\|,
 \label{model:eq:saturation-gain}
\end{equation}
where $V_c=V-\mathbf1 v_0^{\mathsf T}$ and $s_c=s-c\mathbf1$
for arbitrary fixed $v_0,c$.
\end{proposition}
\begin{proof}
For any vector $u$, $u^{\mathsf T}J_pu$ is the categorical variance
of its entries. This proves positive semidefiniteness, and the largest
eigenvalue is at most the trace. Also
$\sum_i p_i^2\ge(1-\delta)^2$, which gives the upper bound.
Dividing the softmax denominator by its largest exponential gives
$\delta=\sum_{i\ne i_*}e^{s_i-s_{i_*}}/
(1+\sum_{i\ne i_*}e^{s_i-s_{i_*}})$ and the gap bound follows.
Differentiation gives $dO/d\alpha=V^{\mathsf T}J_ps$.
Since $J_p\mathbf1=\bm0$ and $\mathbf1^{\mathsf T}J_p=\bm0^{\mathsf T}$,
the centering choices leave this expression unchanged. Apply the
operator-norm inequality and Equation~\eqref{model:eq:saturation-bound}.
\end{proof}

\begin{corollary}[Hard selection at fixed head size]
\label{model:cor:hard-selection}
Fix a score vector $s$ with a unique largest entry at $i_*$, gap
$\Delta>0$, and a value matrix $V$. For
$O_\kappa=V^{\mathsf T}\softmax(\kappa s)$,
\[
 O_\kappa\longrightarrow V_{i_*,:}^{\mathsf T},\qquad
 \left.\partial_\alpha
 V^{\mathsf T}\softmax((1+\alpha)\kappa s)\right|_{\alpha=0}
 \longrightarrow\bm0\quad(\kappa\to\infty).
\]
\end{corollary}
\begin{proof}
The total probability on other keys is at most
$(S-1)e^{-\kappa\Delta}$, which proves the output limit.
Proposition~\ref{model:prop:saturation} bounds the gain derivative by
$2(S-1)\kappa e^{-\kappa\Delta}\|V_c\|_2\|s_c\|$, which tends to zero.
\end{proof}

Across contexts, the selected key and its value can both vary, so this
limit retains context dependence even when local gain sensitivity
vanishes. It is a source-amplitude limit at fixed head size.
An average concentration statistic does not establish its uniform-gap
hypothesis throughout a model.

The corresponding full-model source response sums local injections
transported through the remaining graph, including every affected token
position. Vanishing local factors imply a small full response only with
control of those transport factors and of the centered scores and values.
A measured last-query Jacobian is therefore a local diagnostic.
It does not by itself bound the whole vocabulary response or imply that
predictions are independent of context.

The native PLGA operator has affine form $G=a\,\mathcal V+b_a$, where
$\mathcal V$ depends on the metric and power parameters but not on the
two displayed affine coefficients. Thus replacing both $a$ and $b_a$
by $(1+\alpha)$ times themselves realizes $G\mapsto(1+\alpha)G$ at
fixed layer input. This is an exact algebraic equality. Applying it in
every layer reproduces the inference head-gain source when all downstream
states are recomputed, by induction over decoder layers.

For a training pulse the optimizer moments and counters are left at the
same pre-pulse values. Subsequent matched updates then measure the
response of the augmented training state. This is a state perturbation,
not a coordinate reparameterization of Adam. Freezing generator learning
rates in paired branches provides a causal intervention on one proposed
restoring mechanism, while retaining the input dependence of the generated
operators. Recovery of these finite observables can establish a restoring
regime; attraction to criticality additionally requires that the recovered
state approach the independently identified singular surface.
Section~\ref{model:sec:paired-pulse-results} details the paired pulses, generator controls and finite response measurements.

\FloatBarrier
\par\medskip\noindent
Chapter~\ref{ch:joint-clocks} adds colored sources and joint size--time
limits. It examines how relaxation, observation scaling and a moving
selection window modify the conditional classifications developed here.

\chapter{Colored sources, joint clocks, and selection windows}
\label{ch:joint-clocks}
This chapter studies conditional fluctuations when forcing is temporally
correlated and training is nonstationary. Chronological covariance, coupled
width and time limits, predictive visibility and shrinking selection windows
specify the scale comparisons required before assigning an exponent.

\section{Colored forcing, response gaps and conditional exponents}
\label{model:sec:colored-collectives}
The collective covariance in Proposition~\ref{model:prop:collective-stability}
uses adapted orthogonal innovations. A law with temporally correlated
forcing instead retains a covariance kernel or its spectrum. The
following conditional class shows why the relaxation gap alone cannot
determine a susceptibility exponent.

The observation is the stationary response to one declared centered
stationary source with its specified low-frequency spectrum. The limit
closes the response gap within that family; its exponent depends on the
forcing spectrum as well as on the gap. No finite consuming-corpus
training path is identified with this stationary source law. See the guide to
\hyperref[sec:physical-terminology]{physical terminology and renormalization objects}.

\begin{proposition}[A colored collective mode]
\label{model:prop:colored-mode}
Let $0<\lambda<1$ and let $(\eta_t)_{t\in\mathbb Z}$ be a centered
second-order stationary real process with spectral density $s$, in the
convention
$\Cov(\eta_0,\eta_k)=(2\pi)^{-1}\int_{-\pi}^{\pi}
e^{ik\omega}s(\omega)\,d\omega$.
Then the stationary causal solution
$y_t=\sum_{j\ge0}\lambda^j\eta_{t-1-j}$ exists in $L^2$ and satisfies
\begin{equation}
 \Var(y_t)=\frac1{2\pi}\int_{-\pi}^{\pi}
 \frac{s(\omega)}{(1-\lambda)^2+2\lambda(1-\cos\omega)}\,d\omega.
 \label{model:eq:colored-variance}
\end{equation}
If $s(\omega)\sim s_0|\omega|^\alpha$ as $\omega\to0$,
where $s_0>0$ and $-1<\alpha<1$, then, with $\delta=1-\lambda$,
\begin{equation}
 \Var(y_t)\sim
 \frac{s_0}{2\cos(\pi\alpha/2)}\delta^{\alpha-1}.
 \label{model:eq:colored-asymptotic}
\end{equation}
The difference of two solutions with identical subsequent forcing
is multiplied by $\lambda$ at each update. Its e-folding response
time is $\tau=-1/\log\lambda\sim\delta^{-1}$.
\end{proposition}
\begin{proof}
Minkowski's inequality bounds the $L^2$ norm of the series tail by
$\norm{\eta_0}_{L^2}\sum_{j\ge J}\lambda^j$, proving convergence.
Substitution gives the recurrence $y_{t+1}=\lambda y_t+\eta_t$.
For finite partial sums, the spectral covariance formula gives the
integral of the squared finite transfer polynomial. The polynomials
converge uniformly to $e^{-i\omega}/(1-\lambda e^{-i\omega})$;
since $s$ is integrable, passage to the limit yields
Equation~\eqref{model:eq:colored-variance}.

For any fixed small $\epsilon>0$, the integral over
$|\omega|\ge\epsilon$ remains bounded as $\delta\to0$ and is
negligible after multiplication by $\delta^{1-\alpha}$. On the
remaining interval set $\omega=\delta x$. The denominator divided
by $\delta^2$ tends to $1+x^2$. For $\lambda\ge1/2$ it is bounded
below by a positive constant times $1+x^2$; the spectral assumption
bounds the scaled numerator by a constant times $|x|^\alpha$.
The function $|x|^\alpha/(1+x^2)$ is integrable exactly for the
stated range of $\alpha$. Dominated convergence therefore gives
\[
 \lim_{\delta\downarrow0}\delta^{1-\alpha}\Var(y_t)
 =\frac{s_0}{\pi}\int_0^\infty\frac{x^\alpha}{1+x^2}\,dx
 =\frac{s_0}{2\cos(\pi\alpha/2)}.
\]
The last integral follows by $u=x^2$ and the beta integral
$\int_0^\infty u^{p-1}/(1+u)\,du=\pi/\sin(\pi p)$,
$0<p<1$, with $p=(\alpha+1)/2$.
Subtracting the two forced recurrences and iterating proves the
response formula; $-\log(1-\delta)\sim\delta$ gives its asymptotic.
\end{proof}

\begin{corollary}[Gap, forcing and inference scaling]
\label{model:cor:colored-exponents}
Suppose a normalized collective coordinate has
$\chi_N=\Var(y_N)=N\Var(\bar u_N)$ and obeys the preceding
stationary model with
\[
 \lambda_N=1-cN^{-\zeta}+o(N^{-\zeta}),\qquad
 s_N(\omega)=a_Ns(\omega),\qquad
 a_N\sim a_0N^{-\upsilon},
\]
where $c,a_0,\zeta>0$ and $s$ has the stated low-frequency
asymptotic. Then
\[
 \tau_N\sim c^{-1}N^\zeta,\qquad
 \chi_N\sim\frac{a_0s_0c^{\alpha-1}}{2\cos(\pi\alpha/2)}
 N^{\zeta(1-\alpha)-\upsilon}.
\]
For an inference observable $X_N=h_N\bar u_N+e_N$ with
$h_N\to h\ne0$ and
$\norm{e_N}_{L^2}=o(\sqrt{\chi_N/N})$, one also has
$N\Var(X_N)\sim h^2\chi_N$.
\end{corollary}
\begin{proof}
Apply Equation~\eqref{model:eq:colored-asymptotic} to the fixed spectral
shape and multiply by $a_N$. Insert the gap asymptotic. For the
inference statement, apply Lemma~\ref{model:lem:master-perturbation}
with $A=h_N$ and use contraction of $L^2$ under centering:
\[
 |N\Var(X_N)-h_N^2\chi_N|
 \le 2|h_N|\sqrt{N\chi_N}\,\norm{e_N}_{L^2}
       +N\norm{e_N}_{L^2}^2=o(\chi_N).
\]
The limit of $h_N$ gives the claim.
\end{proof}

The flat-spectrum case $\alpha=0$ recovers the white-forcing
relation $\kappa=\zeta-\upsilon$. A depleted low-frequency spectrum
with $\alpha>0$ weakens the divergence; an enhanced spectrum with
$\alpha<0$ strengthens it. Neither an exponent for response time
nor a signed finite-interval covariance sum identifies $\alpha$.
The response time above refers to matched-forcing perturbations,
not a conditional-mean or autocorrelation decay assertion for
colored noise. Nonnormal matrix modes, size-dependent spectral shapes
and nonlinear saturation require separate limits.

Single-pass native training is nonstationary on its augmented state.
Application of this stationary class to a local training window
requires both a justified effective mode and control of corpus,
optimizer and drive changes throughout its response history. At
fixed consumed fraction $q<1$ and batch size $B$, a sufficient
sampling coupling condition for a window of length $\tau_N$ is
$(B\tau_N)^2/((1-q)M_N)\to0$, as in
Proposition~\ref{model:prop:data-path-coupling}. For a postulated response exponent
$\zeta$, $M_N$ growing faster than $N^{2\zeta}$ meets this resource
condition. It does not establish stationarity or a native value of
$\zeta$. The finite conditional studies use their full covariance
without fitting a low-frequency exponent or imposing this limit.

\section{Chronological covariance before a stationary limit}
\label{model:sec:nonstationary-response}

The exact temporal map remains available when a stationary scalar
mode has not been identified. It retains the same initial-state and
forcing sectors as the finite-resource transport, now with a general
deterministic chronological response.

\begin{corollary}[Nonstationary response specialization]
\label{model:prop:nonstationary-response-covariance}
For $y_{k+1}=J_ky_k+\eta_k$ with deterministic $J_k$ and jointly
square-integrable initial state and forcing, let $\Phi_{t,k}$,
$C_0,H_k,Q_{k\ell}$ have the meanings in
Proposition~\ref{model:prop:master-chronological}, with $q=y$ and $b_k=\bm0$.
Then
\begin{equation}
 \Cov(y_t)=L_t\Cov(U)L_t^{\mathsf T}.
 \label{model:eq:nonstationary-response-covariance}
\end{equation}
Its initial, initial/forcing and forcing/forcing blocks are exactly
Equation~\eqref{model:eq:law-innovation-covariance}. Consecutive regrouping preserves them.
\end{corollary}
\begin{proof}
Take the conditioning sigma-field to be trivial and all offsets to be
zero in Proposition~\ref{model:prop:master-chronological}. Its jointly retained
initial and forcing law gives the displayed covariance and regrouping.
\end{proof}

Choosing a deterministic fitted $J_k$ and defining the residual
$\eta_k=y_{k+1}-J_ky_k$ makes the identity exact, but does not supply
a predictive law for those residuals. Native Jacobians generally
depend on the random path and cannot be moved outside these
expectations. Conditioning on their entire history requires the
corresponding conditional covariance, cross terms and the outer
covariance of the conditional mean in Equation~\eqref{model:eq:master-outer}.
Conditioning on a future response history is not a causal predictor. A native
linear approximation additionally retains its nonlinear remainder.
Thus neither temporal covariance enhancement nor a four-interval fit
identifies a response gap or a low-frequency forcing exponent.

Under the stationary scalar assumptions of
Corollary~\ref{model:cor:colored-exponents}, the distinct limit is
$\chi_N\asymp N^{\zeta(1-\alpha)-\upsilon}$ and
$\tau_N\asymp N^\zeta$. At fixed batch and a consumed fraction
bounded away from one, the sufficient whole-window sampling condition
$(B\tau_N)^2/[(1-q_N)M_N]\to0$ requires
$M_N/N^{2\zeta}\to\infty$ when $\tau_N\asymp N^\zeta$.
This only controls the sampling approximation. Coefficient drift,
optimizer memory, observation overlap and the response remainder
remain separate assumptions. The finite conditional path description
and the stationary critical limit consequently have different,
explicit domains.

\section{Joint training clocks and critical scaling}
\label{model:sec:joint-scaling}

\subsection{The optimizer belongs to the scale map}
The head family fixes $d$ independently of $N$ and varies $N$, so $N$ is a count of
interacting components, not an independently specified spatial length.
A susceptibility law $\chi_N\asymp N^\kappa$ and a control window
$|r|\asymp N^{-\phi}$ therefore define count exponents. Identifying
$\kappa$ with $\gamma/\nu$ requires a separate length convention and its
relation to $N$. An exponent of a finite training clock has another
meaning again.

For the controlled constant-rate family the two rates are
\begin{equation}
 \eta_{\rm gen}=\eta_0g,\qquad
 \eta_{\rm body}=2\eta_0/N,\qquad \eta_0=3\times10^{-4}.
 \label{model:eq:two-clocks}
\end{equation}
Consequently fixing $g$ while increasing $N$ changes the ratio of
generator and body rates. A width comparison at fixed update count
changes the nominal body clock. These rate factors specify parameter
update conventions. A clock for a fixed emitted coordinate additionally
depends on its Jacobian and on the conditional preconditioned velocity,
both of which can vary with $N$. Global clipping, Adam normalization
and the growing body parameter dimension therefore remain in the
source and emission laws. If a measured relaxation time in
physical time is $N^z$ and the declared time increment per update is
$\eta_N\asymp N^{-a}$, its update count is $N^{z+a}$.
This follows by dividing the relaxation time by the time increment.
The apparent update exponent contains the optimizer convention.
Fixed Adam coefficients also fix memory in update units; exact temporal
blocking must retain the moments and counters, as in
Section~\ref{model:sec:row-optimizer-theory}.

A deterministic learning-rate schedule is an additional drive coordinate.
Let the two peak rates be $\eta_{\rm body}^{\rm pk}$ and
$\eta_{\rm gen}^{\rm pk}$. If a common nonnegative factor $s_k$
multiplies them at update $k$, the nominal accumulated body,
generator-drift and generator-noise clocks are respectively
\begin{equation}
 \eta_{\rm body}^{\rm pk}\sum_{k<t}s_k,\qquad
 \eta_{\rm gen}^{\rm pk}\sum_{k<t}s_k,\qquad
 (\eta_{\rm gen}^{\rm pk})^2\sum_{k<t}s_k^2.
 \label{model:eq:scheduled-nominal-clocks}
\end{equation}
For the schedule-only controlled family, Equation~\eqref{model:eq:two-clocks}
specifies the peak rates. For each reproduced reference recipe,
both parameter groups instead use its common peak rate. Thus their
nominal first-order clocks coincide; the controlled body factor
$1/N$ must not be inserted into those reference trajectories.
The first two expressions sum rate factors, whereas the last sums
squared rate factors for a fixed conditional velocity covariance.
They do not replace the state-dependent transported drift and covariance.
The exact transition over an interval is the ordered composition
$K_{s_0}K_{s_1}\cdots K_{s_{t-1}}$, with the native optimizer state
retained. Including the schedule phase in the augmented state makes
this a homogeneous transition on the enlarged state space.
An arbitrary schedule need not admit an invariant probability law on
that enlarged space. In particular, the schedule phase is not an
endogenous feedback variable. A vanishing rate can suppress motion
without demonstrating attraction toward a critical surface.
Warmup and annealing therefore define a different conditioned process.
The clocks in Equation~\eqref{model:eq:scheduled-nominal-clocks} state its
explicit dependence on the drive; empirical transport between schedules
additionally concerns the evolving optimizer and emission laws.

The reference pretraining schedule~\cite{gokden2026soc} has a linear
warmup of $W$ updates, a cosine interval ending at update $T$, and a
positive floor fraction $\alpha$. With
$a=(1+\alpha)/2$ and $b=(1-\alpha)/2$, its multiplier is
\begin{equation}
 s_k=\begin{cases}
 k/W,&0\le k\le W,\\
 a+b\cos\!\bigl(\pi(k-W)/(T-W)\bigr),&W<k<T,\\
 \alpha,&k\ge T.
 \end{cases}
 \label{model:eq:reference-schedule}
\end{equation}
The update numbered $k+1$ uses $s_k$. Thus the first update has zero
parameter rate while its gradients still update the Adam moments and
counter. The stored scheduler phase after $t$ updates specifies the
rate for the next update.

\begin{proposition}[Exact nominal clocks for the reference drive]
\label{model:prop:reference-schedule-clocks}
Let $W,T,L$ be integers with $W\ge1$, $T-W\ge2$, and $L\ge0$.
For the schedule in Equation~\eqref{model:eq:reference-schedule},
\begin{align}
 \sum_{k=0}^{T+L-1}s_k
 &=\frac W2+(T-W)a-\frac\alpha2+L\alpha,\\
 \sum_{k=0}^{T+L-1}s_k^2
 &=\frac W3+(T-W)\left(a^2+\frac{b^2}{2}\right)
   -\frac{\alpha^2}{2}+\frac1{6W}+L\alpha^2.
 \label{model:eq:reference-clock-sums}
\end{align}
\end{proposition}
\begin{proof}
Split the sum into $0\le k<W$, $W\le k<T$, and the final $L$
updates. The warmup contributions are $(W-1)/2$ and
$(W-1)(2W-1)/(6W)$. Put $M=T-W$. The real part of the finite
geometric sum of $\exp(i\pi j/M)$ gives
$\sum_{j=0}^{M-1}\cos(\pi j/M)=1$. Likewise
$\cos^2 x=(1+\cos(2x))/2$ and the sum over $M$ roots of unity give
$\sum_{j=0}^{M-1}\cos^2(\pi j/M)=M/2$.
The cosine contributions are consequently $Ma+b$ and
$M(a^2+b^2/2)+2ab$. Substituting $a,b$ and adding the floor
contributions $L\alpha,L\alpha^2$ proves both formulas.
\end{proof}

These two sums differ even before the transported native velocity law
is considered. For example, $T=250000$, $W=2000$, and $\alpha=0.1$
give $137399.95$ and $100796.66175$ over the annealing horizon.
An identification of the entire process with constant-rate training
at one effective update count cannot simultaneously preserve both
nominal sums. Equality of one emitted observable under such a time
change therefore requires a separate, observable-specific argument.

\begin{proposition}[Invariant laws of a clamped schedule phase]
\label{model:prop:clamped-phase-invariant}
Let a Markov state include a phase $j\in\{0,\ldots,T\}$ whose next
value is deterministically $\min(j+1,T)$. Every invariant probability
law of the augmented process is supported on $j=T$.
\end{proposition}
\begin{proof}
The phase is bounded, so its expectation is finite. Its expected
one-step increment under any probability law is exactly
$\Pr(j<T)$. Invariance makes that increment zero, giving
$\Pr(j<T)=0$.
\end{proof}

This statement does not assert existence of an invariant law for the
learned parameters at the floor. It distinguishes transient schedule
phases from a possible invariant training law after the imposed drive
has stopped changing. In particular, a fluctuation time series pooled
across warmup, annealing and the floor mixes different conditioned
transition laws. An integer Adam counter that increases at every update
also excludes an invariant probability law on the exact counter state.
A limiting stationary description can instead compactify the two bias
clocks as $q_i=\beta_i^t$, with update $q_i'=\beta_iq_i$ and the
boundary $q_i=0$ included. Any use of that boundary requires the
corresponding limiting parameter, moment and emission laws. Stationarity
of a projected observable is a weaker statement than stationarity of
the complete native state.

A candidate second-order generator clock is $g^2t$. The simultaneous
transformation
\begin{equation}
 (N,t,g)\longmapsto (bN,bt,b^{-1/2}g)
 \label{model:eq:joint-clock-map}
\end{equation}
preserves both $t/N$ and $g^2t$ and composes multiplicatively in $b$.
It is an experimental coordinate map. Its interpretation as a kinetic
RG requires control of the native drift, noise, memory and emission.
The first-order generator clock is $gt$. If $\mu$ is the conditional
mean complete generator velocity in fixed resolved coordinates,
a frozen-coefficient expected displacement is $\eta_0g\mu t$.
Under Equation~\eqref{model:eq:joint-clock-map}, preserving this displacement
also requires
\begin{equation}
 \mu'=b^{-1/2}\mu.
 \label{model:eq:drift-coupling-map}
\end{equation}
Indeed $(g/\sqrt b)(\mu/\sqrt b)(bt)=g\mu t$; leaving $\mu$
unchanged instead multiplies the displacement by $\sqrt b$.
A fixed conditional velocity covariance preserves its accumulated
second-order contribution because $(g/\sqrt b)^2(bt)=g^2t$.
The drift factor composes multiplicatively. Thus nonzero drift must
be retained as a transformed coupling, centered in a justified moving
frame, or shown to have the requisite smallness. The learned native
velocity law is not a free hyperparameter that can be rescaled by
declaration. Neither a norm of its complete drift nor a passage-time
fit identifies the drift of a particular putative critical coordinate.
The nominal weight-decay clocks are
\[
 D_{\rm gen}=\eta_0g\lambda_{\rm gen}t,\qquad
 D_{\rm body}=2\eta_0\lambda_{\rm body}t/N.
\]
To preserve these clocks as well, Equation~\eqref{model:eq:joint-clock-map}
requires
\begin{equation}
 \lambda_{\rm gen}'=b^{-1/2}\lambda_{\rm gen},\qquad
 \lambda_{\rm body}'=\lambda_{\rm body}.
 \label{model:eq:decay-scale-map}
\end{equation}
Direct substitution proves the invariance, and the factors compose
under successive scale changes. The native choice
$\lambda_{\rm gen}=\lambda_{\rm body}=0.01$ is consequently not
preserved by this joint transformation.
A comparison that holds native decay fixed tests a finite two-clock
approximation, while retaining the additional decay coordinate.
These nominal clocks do not assert pure decay of a coordinate that
has nonzero learned force.

Likewise, preserving a fixed physical Adam-memory duration while
replacing $t$ updates by $bt$ requires $\beta'=\beta^{1/b}$, since
$(\beta')^b=\beta$. Fixed native coefficients instead give a fast
memory sector on this rescaled clock. A correct limiting generator
must treat that sector or justify its elimination.
The following statement makes one sufficient local requirement precise.

\begin{proposition}[A two-clock infinitesimal generator]
\label{model:prop:two-clock-generator}
Let $h\downarrow0$ and let a retained state $x=(u,v)$ have one-step
increment $(\sqrt h\,V_h,hW_h)$. Expectations below condition on the
complete incoming state. Suppose, uniformly on the compact region in
question,
\begin{alignat*}{2}
 \E V_h&=\sqrt h\,b(x)+o(\sqrt h),&\quad \E W_h&=c(x)+o(1),\\
 \E[V_hV_h^{\mathsf T}]&=S(x)+o(1),&\quad \E(\norm{V_h}+\norm{W_h})^3&\le C.
\end{alignat*}
For any $C^3$ test function with bounded derivatives through order three,
\begin{equation}
 \frac{\E[f(x+\Delta x)-f(x)]}{h}
 \longrightarrow b\cdot\nabla_u f+c\cdot\nabla_v f
       +\tfrac12\tr(S\nabla^2_{uu}f).
 \label{model:eq:two-clock-generator}
\end{equation}
\end{proposition}
\begin{proof}
Taylor expansion to second order gives the two first-derivative terms
and $\frac12\E[\Delta x^{\mathsf T}D^2f\Delta x]$.
After division by $h$, the $uu$ term is the displayed trace. The mixed
term is $O(\sqrt h)$ and the $vv$ term is $O(h)$ by the third-moment
bound and H\"older's inequality. The Taylor remainder divided by $h$
is $O(\sqrt h)$ because
$\E\norm{\Delta x}^3=O(h^{3/2})$. Substitution of the conditional
mean assumptions proves the limit.
\end{proof}

With $h=1/N$ and $g=c_0/\sqrt N$, Equation~\eqref{model:eq:two-clocks}
has these two step scales. It does not supply the required centering
of $V_h$. In particular, an incoming Adam first moment can give a
conditional drift of order one. Then a $\sqrt h$ generator step has
drift of order $h^{-1/2}$ on the proposed clock. Averaging a fast
optimizer process can require a corrector and a separate mixing
argument. Neither a fitted $g^2t$ collapse nor the local generator
limit alone proves that averaging, tightness of paths, uniqueness of
a limiting process, or a native critical point.

\subsection{Innovation covariance and the conditioning law}
Let $K_N(s,\cdot)$ be the native augmented update kernel, with fresh
batch randomness averaged conditional on its complete incoming state
$s$. Write $\Delta\theta=\theta_{t+1}-\theta_t$ and define
\[
 b_N(s)=\E[\Delta\theta\mid S_t=s],\qquad
 Q_N(s)=\Cov(\Delta\theta\mid S_t=s).
\]
When the next batch follows this conditional kernel,
$\varepsilon_{t+1}=\Delta\theta-b_N(S_t)$ is a martingale difference
with respect to the training filtration containing the complete incoming
state and all revealed batches. Indeed its conditional expectation is
zero by the definition of $b_N$. The remaining corpus belongs to $S_t$
for a single pass; batches need not be independent across time. This is
a statement under the conditional next-batch law. After conditioning
on an entire realized batch history $\mathcal C$, that next batch is fixed. The same innovation
interpretation no longer applies to the remaining initialization
ensemble. Resampling batches at a saved state measures the first law;
whole-initialization susceptibility at fixed $\mathcal C$ measures the
second. Neither is a substitute for the other.
These two measurements are reported in Sections~\ref{model:sec:conditional-noise-results} and \ref{model:sec:conditional-fluctuation-results}, respectively.

\begin{proposition}[Drift and innovation contributions to accumulated motion]
\label{model:prop:innovation-motion}
For a square-integrable adapted vector process with increments
$\Delta\theta_t=b_t+\varepsilon_{t+1}$, suppose $b_t$ is measurable at
time $t$ and $\E[\varepsilon_{t+1}\mid\mathcal F_t]=\bm0$.
Set $B_T=\sum_{t<T}b_t$ and $M_T=\sum_{t<T}\varepsilon_{t+1}$.
Then
\begin{align}
 \E\norm{M_T}^2
 &=\sum_{t<T}\E\tr Q_t,\label{model:eq:innovation-isometry}\\
 \E\norm{\theta_T-\theta_0}^2
 &=\E\norm{B_T}^2+\sum_{t<T}\E\tr Q_t
                  +2\E\ip{B_T}{M_T},\label{model:eq:drift-innovation-motion}
\end{align}
where $Q_t=\E[\varepsilon_{t+1}\varepsilon_{t+1}^{\mathsf T}
\mid\mathcal F_t]$. In general the final cross term is nonzero.
\end{proposition}
\begin{proof}
If $i<j$, $\varepsilon_{i+1}$ is $\mathcal F_j$-measurable.
Conditional expectation therefore gives
$\E\ip{\varepsilon_{i+1}}{\varepsilon_{j+1}}=0$.
Expanding the squared norm of the sum proves
Equation~\eqref{model:eq:innovation-isometry}. Summing the increment equation
and expanding $\norm{B_T+M_T}^2$ proves
Equation~\eqref{model:eq:drift-innovation-motion}.
A future drift can depend on earlier innovations, so its covariance
with $M_T$ need not vanish. For example, take two scalar increments,
$b_0=0$, $\varepsilon_1$ centered with unit variance,
$b_1=\varepsilon_1$, and $\varepsilon_2=0$. Then
$\E[B_2M_2]=1$.
\end{proof}

Adam's incoming moments, clipping and the learned state make $b_t$
state dependent. Consequently a measured conditional noise trace does
not alone predict an accumulated displacement proportional to $g^2T$.
The drift, its feedback through the state, and their cross contribution
remain in Equation~\eqref{model:eq:drift-innovation-motion}. Passive records
of realized updates also contain all three contributions. Their
within-window temporal covariance is not itself an innovation
covariance. A diffusion approximation must justify the centering or
averaging assumptions in Proposition~\ref{model:prop:two-clock-generator}
for the appropriate probability law.

\subsection{Exact finite steps and coupled parameter sources}
A local generator response need not approximate the complete native
optimizer step. Fix an incoming state $z$, including parameters,
optimizer moments and counters. Partition the parameters as $(a,b)$,
where $a$ contains the declared generator parameter group of
Section~\ref{model:sec:model} and $b$ its complementary body group. A fresh
batch $\xi$ produces the realized displacements
$(d_a(z,\xi),d_b(z,\xi))$ of one complete optimizer step. Both
displacements use the same incoming state, batch and full-model
clipping rule. The following identity holds for any fixed observation
$F$ taking values in a finite-dimensional inner-product space.

\begin{proposition}[Finite source corners and their coupled covariance]
\label{model:prop:finite-source-corners}
Define, conditional on $z$,
\begin{gather}
 Y_a=F(a+d_a,b)-F(a,b),\quad
 Y_b=F(a,b+d_b)-F(a,b),\nonumber\\
 Y_{ab}=F(a+d_a,b+d_b)-F(a+d_a,b)
              -F(a,b+d_b)+F(a,b).
 \label{model:eq:finite-source-corners}
\end{gather}
Then the complete observation increment is
$Y=Y_a+Y_b+Y_{ab}$. If these variables have finite second moments,
write $\mu_i=\E[Y_i\mid z]$ and
$Q_{ij}=\E[(Y_i-\mu_i)\otimes(Y_j-\mu_j)\mid z]$
for $i,j\in\{a,b,ab\}$. The exact conditional identities are
\begin{equation}
 \E[Y\mid z]=\sum_i\mu_i,\qquad
 \Cov(Y\mid z)=\sum_{i,j}Q_{ij},\qquad
 \E[\norm{Y}^2\mid z]=
 \sum_{i,j}\bigl(\ip{\mu_i}{\mu_j}+\tr Q_{ij}\bigr).
 \label{model:eq:finite-source-covariance}
\end{equation}
For a twice continuously differentiable real-arithmetic observation,
\begin{equation}
 Y_{ab}=\int_0^1\!\!\int_0^1
 D_aD_bF(a+s d_a,b+u d_b)[d_a,d_b]\,du\,ds.
 \label{model:eq:finite-mixed-transport}
\end{equation}
If the operator norm of this mixed derivative on the entire rectangle
is bounded by $K$, then
$\norm{Y_{ab}}\le K\norm{d_a}\norm{d_b}$.
\end{proposition}
\begin{proof}
Adding the three differences cancels the intermediate corners.
Subtract their conditional means, expand the tensor product and
take conditional expectations to obtain the covariance identity.
The squared-norm identity follows by taking traces and adding
the squared conditional mean. Applying the fundamental theorem of
calculus first in the $a$ direction and then in the $b$ direction
gives Equation~\eqref{model:eq:finite-mixed-transport}. Integrating the
operator-norm bound on the unit square gives the last assertion.
\end{proof}

The algebraic corner and covariance identities require no
differentiability and apply to the recorded deterministic
floating-point forward program. The mixed-derivative representation
requires its stated smooth real-arithmetic extension. Partial corners
are interventions using displacements from the same coupled step;
they are not separately trained models. Global clipping and a common
batch can correlate $Y_a$ and $Y_b$ even when $Y_{ab}=\bm0$.
For example, $F(a,b)=a+b$, $d_a=\xi$ and $d_b=-\xi$ give zero
complete variance despite two nonzero diagonal source variances.
For $F(a,b)=a+b+ab$ at $(0,0)$, the same displacements instead give
$Y=-\xi^2$, entirely through the mixed corner.

Thus the two source rates do not justify adding independent
generator and body diffusion coefficients without their cross terms.
A generator-only approximation must control both $Y_b+Y_{ab}$
and its conditional covariance, followed by accumulated error over
the relevant training window. A small infinitesimal emission
residual for $Y_a$ supplies neither of those controls. A measured
conditional restoring mean is also a function of the complete
incoming state $z$; identifying it with a drift depending only on
one row statistic requires a separate closure argument.

\subsection{Local predictive source coordinates}
At a fixed incoming state and fixed evaluation cohort, let $J$ map a
parameter displacement to its full-vocabulary logit tangent. For each
context let $p$ be the base prediction. Define the linear Fisher image
by centering the logit tangent under $p$, multiplying its entries by
$\sqrt p$, and including the square root of the fixed context weight.
Write the resulting Euclidean linear map as $L$. It includes the
complete downstream graph after the chosen parameter source.
Zero-probability numerical coordinates, when present, contribute zero
in this measured seminorm. A generator-only source and a full-optimizer
source are different domains for $L$.

\begin{proposition}[Visible source coordinates and finite projection error]
\label{model:prop:visible-source-coordinates}
Let $L:\R^D\to\R^m$ be linear, let $v_1,\ldots,v_k$ be orthonormal
in the emitted coordinates, and let $P$ project onto their span.
The frozen source coordinates
$a_i=L^{\mathsf T}v_i$ predict the projected response to every source
$u$ exactly:
\begin{equation}
 PLu=\sum_{i=1}^k v_i\ip{a_i}{u}.
 \label{model:eq:visible-source-pullback}
\end{equation}
For any square-integrable random response $Y$ under a law for which
$P$ is fixed, including a law conditional on the calibration that chose it,
\begin{align}
 \E\norm{Y}^2&=\E\norm{PY}^2+\E\norm{(I-P)Y}^2,\label{model:eq:visible-second}\\
 \tr\Cov(Y)&=\tr\Cov(PY)+\tr\Cov((I-P)Y).\label{model:eq:visible-covariance}
\end{align}
If $M=\E[YY^{\mathsf T}]$ has decreasing eigenvalues
$\lambda_1,\ldots,\lambda_m$, the smallest second-moment residual
among rank-$k$ orthogonal projections is
$\sum_{j>k}\lambda_j$. This last optimality statement holds for the
law defining $M$, including a finite calibration law, and is not a
generalization guarantee for another law.
\end{proposition}
\begin{proof}
The adjoint identity gives
$\ip{v_i}{Lu}=\ip{L^{\mathsf T}v_i}{u}$, proving
Equation~\eqref{model:eq:visible-source-pullback}. Orthogonality gives
$\norm{y}^2=\norm{Py}^2+\norm{(I-P)y}^2$ for each $y$.
Take expectations to obtain Equation~\eqref{model:eq:visible-second}; apply
the same identity to $Y-\E Y$ to obtain
Equation~\eqref{model:eq:visible-covariance}. Finally,
$\E\norm{(I-P)Y}^2=\tr M-\tr(PM)$. In an eigenbasis of $M$,
$\tr(PM)=\sum_j\lambda_j\alpha_j$, where
$0\le\alpha_j\le1$ and $\sum_j\alpha_j=k$.
This sum is at most $\sum_{j\le k}\lambda_j$, attained by projection
onto the leading eigenspace. Subtracting from $\tr M$ proves the bound.
\end{proof}

\begin{corollary}[Drift and innovation require different relative errors]
\label{model:cor:source-noise-priority}
Let $Y$ have mean $\mu$ and covariance $Q$, with $\tr Q>0$, and let
$P$ be a fixed orthogonal projection. Write $e_{\rm total}$ and
$e_{\rm noise}$ for the residual fractions defined using the population
second moment and covariance, respectively. Then
\begin{align}
 \E\norm{(I-P)Y}^2
 &=\norm{(I-P)\mu}^2+\tr((I-P)Q),\notag\\
 e_{\rm noise}&\leq e_{\rm total}
 \left(1+\frac{\norm{\mu}^2}{\tr Q}\right).
 \label{model:eq:source-noise-relative-bound}
\end{align}
A rank-$k$ leading eigenspace of $Q$ minimizes the covariance residual;
its minimum is the sum of the remaining eigenvalues of $Q$. Small
relative total-response error alone does not give small relative
innovation error uniformly over the drift-to-noise ratio.
For a sample of size $s$ with unbiased covariance $\widehat Q$, the
same bound holds with the factor
$1+s\norm{\bar Y}^2/((s-1)\tr\widehat Q)$.
\end{corollary}
\begin{proof}
Write $Y=\mu+Z$ with $\E Z=\bm0$. The cross term in the residual squared
norm has zero expectation, proving the decomposition. Dropping its
nonnegative mean term and dividing by $\tr Q$ proves the bound, since
$\E\norm Y^2=\norm\mu^2+\tr Q$.
Apply the eigenspace argument of
Proposition~\ref{model:prop:visible-source-coordinates} to $Y-\mu$ for the
covariance optimum. For a counterexample to uniform relative transfer,
take $Y=(M,Z)$ with $\E Z=0$, $\Var Z=1$, and project onto the first
coordinate. Then $e_{\rm total}=1/(M^2+1)\to0$ but
$e_{\rm noise}=1$. Finally, in the empirical uniform law,
$\widehat\E\norm{Y-\bar Y}^2=(s-1)\tr\widehat Q/s$.
Substitution gives the stated sample factor.
\end{proof}

An uncentered response basis optimizes a second moment that includes
conditional drift. It need not optimize conditional covariance. The
trace identity also does not remove cross blocks of the complete
covariance matrix or identify an independent-component law.
Moreover, participation rank is not monotone under a linear emission:
$\diag(100,1)$ has participation rank $101^2/10001$, while applying
$\diag(1/10,1)$ gives the identity covariance, whose participation rank
is $2$. A comparison of parameter and predictive concentration must
therefore use matched source samples and the declared observation
metric. The retained basis and its adjoints are state dependent; a
future training law requires their transport or the appropriate memory,
in addition to the fixed-state result above.

Instantaneous noise fidelity also differs from long-time covariance
fidelity. An explicit resolved linear process is
$X_{t+1}=A_\epsilon X_t+\xi_t$, with independent centered Gaussian
innovations and
\[
 A_\epsilon=\diag(1-\epsilon^2,0),\quad
 \Cov(\xi_t)=\diag(\epsilon,1),\quad
 P=\diag(0,1),\qquad 0<\epsilon<1.
\]
Keeping $P\xi_t$ loses only the fraction $\epsilon/(1+\epsilon)$
of instantaneous noise trace. Yet the stationary covariance is
\[
 \Sigma_\epsilon=
 \diag\!\left(\frac1{\epsilon(2-\epsilon^2)},1\right).
\]
To verify this formula, sum the convergent coordinatewise geometric
series $\sum_{j\ge0}A_\epsilon^j\Cov(\xi_t)(A_\epsilon^j)^{\mathsf T}$;
the first entry is $\epsilon/[1-(1-\epsilon^2)^2]$.
The discarded fraction of stationary covariance is therefore
$[1+\epsilon(2-\epsilon^2)]^{-1}\to1$. Weakly forced slow directions
can dominate a long-time law. This example is not a native PLDR
criticality claim. It shows why fixed-state source approximation must
be combined with memory, response and accumulated-closure control
before transferring a scaling class.

\subsection{Slow parameter modes invisible to the declared emission}
\begin{proposition}[An exact input-support decay sector]
\label{model:prop:invisible-decay}
Consider the native architecture with untied input and output
embeddings. Suppose a token index $j$ has zero probability of appearing
in any training input or in an input of the declared observation law.
Start its input-embedding row with zero Adam moments, as in the native
initialization. With constant input-embedding rate $\eta_N$ and decay
$\lambda_d$, that row obeys, in real arithmetic,
\begin{equation}
 e_{j,t}=(1-\eta_N\lambda_d)^t e_{j,0}.
 \label{model:eq:invisible-decay}
\end{equation}
Perturbing only $e_{j,0}$ changes neither any other parameter or moment
trajectory nor any full-vocabulary predictive emission on that input
law. Nevertheless, for $0<\eta_N\lambda_d<1$, the parameter relaxation
time is $-1/\log(1-\eta_N\lambda_d)$. Under
Equation~\eqref{model:eq:two-clocks} it is asymptotic to
$N/(2\eta_0\lambda_d)$.
\end{proposition}
\begin{proof}
An input-embedding row is read only when its token occurs in the input.
Untying ensures that the same row is not used by the full-vocabulary
output projection. The loss and the declared emission are therefore
independent of this row. Its gradient is zero at every update, and
zero moments stay zero. Global clipping preserves the zero gradient.
The only update to this row is multiplication by
$1-\eta_N\lambda_d$, proving Equation~\eqref{model:eq:invisible-decay} by
induction. The entire gradient of every other parameter is unchanged,
so induction also preserves their moments, clipping factors and
updates under the coupled perturbation. The exponential relaxation
time follows by taking the logarithm of the decay factor. Finally
$-\log(1-x)\sim x$ at zero and
$\eta_N\lambda_d=2\eta_0\lambda_d/N$ give the stated asymptotic.
\end{proof}

Thus an eigenvalue approaching one in the full parameter update can
come from the declared learning-rate and decay convention, with zero
response in the retained predictive field. The conclusion is relative
to the specified input law: another prompt containing $j$ can observe
that embedding row. A model-wide critical theory must retain source
visibility and distinguish such a data-dependent null sector from a
soft interacting collective field. Merely extending every parameter
relaxation time would not make this distinction.

\subsection{Bounded fields constrain count exponents}
\begin{proposition}[An admissible exponent for a bounded intensive field]
\label{model:prop:bounded-count-exponent}
Suppose $0\le q_N\le1$ under the declared ensemble and let
$\chi_N=N\Var(q_N)$. Then
\begin{equation}
 \chi_N\le N\E q_N(1-\E q_N)\le N/4.
 \label{model:eq:bounded-count-susceptibility}
\end{equation}
In particular, $\chi_N\sim C N^\kappa$ with $C>0$ requires
$\kappa\le1$. For $s>1$ observed values in $[0,1]$, the susceptibility
with the unbiased variance divisor is at most $Ns/[4(s-1)]$.
These bounds also hold after averaging variances at fixed contexts and
coordinates with nonnegative weights of total one.
\end{proposition}
\begin{proof}
Since $q_N^2\le q_N$, writing $\mu_N=\E q_N$ gives
$\Var(q_N)\le\mu_N-\mu_N^2\le1/4$.
If $\kappa>1$, dividing the asserted asymptotic by $N$ contradicts
this bound. Apply the same inequality to the uniform empirical law
on the $s$ observations and multiply its population variance by
$s/(s-1)$ to obtain the finite-sample bound. Weighted averaging
preserves each inequality.
\end{proof}

The native normalized row field and normalized attention entropy
satisfy this range condition. A finite log--log slope exceeding one
can arise when the smaller widths have already concentrated while
larger widths are still relaxing. It cannot be continued as the
asymptotic susceptibility exponent of that unchanged bounded
observation. It may instead indicate a transient or a correction to
scaling. Conversely, a slope near one is allowed but need not be
intrinsic criticality: a finite common random coordinate has precisely
that susceptibility normalization.

\begin{proposition}[Radial and directional common fluctuations]
\label{model:prop:common-radial-directional}
Let $C_1,\ldots,C_s$ be vectors in an inner-product space, with
$s\ge2$, and let $\bar C=s^{-1}\sum_iC_i$.
Write $r_i=\norm{C_i}$ and $u_i=C_i/r_i$ when $r_i>0$;
set $u_i=\bm0$ when $r_i=0$. For any $N>0$,
\begin{align}
 \widehat\chi_C
 &=\frac{N}{s-1}\sum_i\norm{C_i-\bar C}^2
   =\widehat\chi_{\rm rad}+\widehat\chi_{\rm dir},\nonumber\\
 \widehat\chi_{\rm rad}
 &=\frac{N}{s(s-1)}\sum_{i<j}(r_i-r_j)^2,\qquad
 \widehat\chi_{\rm dir}
 =\frac{N}{s(s-1)}\sum_{i<j}r_i r_j\norm{u_i-u_j}^2.
 \label{model:eq:common-radial-directional}
\end{align}
Both contributions are nonnegative. The identity remains valid after
averaging over a fixed context and decoder cohort, with radii and
directions defined separately at each cohort coordinate.
\end{proposition}
\begin{proof}
Expanding squared norms gives
$\sum_{i<j}\norm{C_i-C_j}^2
=s\sum_i\norm{C_i-\bar C}^2$.
For positive radii, expand the inner product and use
$\norm{u_i}=\norm{u_j}=1$ to obtain
\[
 \norm{C_i-C_j}^2=(r_i-r_j)^2+
                  r_i r_j\norm{u_i-u_j}^2.
\]
If a radius is zero, the same formula follows by direct substitution.
Summing and multiplying by $N/[s(s-1)]$ proves the decomposition.
Every summand is nonnegative, and cohort averaging is linear.
\end{proof}

In the native centroid measurement the inner product is the mean over
the $d$ column coordinates ($d=64$ in the recorded measurements).
Directional variation across independent
training identities differs from head misalignment within one identity.
A large directional fraction can describe a learned orientation
sector without establishing whether it is a physical order parameter,
an emission redundancy or a critical coordinate. Those interpretations
require the corresponding predictive source and response information.
The decomposition does not condition away any learned direction.
Section~\ref{model:sec:common-coordinate-results} reports the complete common-coordinate and radial/directional comparisons.

\begin{proposition}[A finite common sector and a source corner]
\label{model:prop:finite-common-corner}
Let $Z$ take values $-1$ and $1$ with equal probability, and for each
$N$ let all $N$ observed components equal $mZ$, where $m>0$ is fixed.
The intensive head mean is $q_N=mZ$. Then
\[
 \chi_N=Nm^2,\qquad
 p_N(j)=\frac1N\log\E e^{Njq_N}
       =\frac1N\log\cosh(Njm)\longrightarrow m|j|.
\]
These laws can be generated by one random bit at every $N$, with a
constant-in-time process and no attraction or feedback.
\end{proposition}
\begin{proof}
The mean has variance $m^2$. The two-point expectation gives the
hyperbolic cosine. For real $x$,
$|x|-\log2\le\log\cosh x\le|x|$; substitute $x=Njm$ and divide
by $N$. Sampling $Z$ once and keeping it fixed realizes the last
assertion.
\end{proof}

Thus the mathematical existence of a source corner does not by itself
supply an extensive interacting critical mechanism or endogenous
selection. In an interacting model, order-one common fluctuations may
instead describe coexistence of distinct phases. The appropriate
within-phase source response and the order of source, size and time
limits then matter. If a common coordinate is a gauge redundancy, its
physical emission must be quotiented appropriately. If it is a learned
order parameter, conditioning on its final value merely to suppress
its variance can remove the physical phenomenon of interest. The
conditional construction fixes the exogenous environment, not an
outcome-dependent phase or learned common state. The measured
common/contrast and predictive decompositions are used to determine
what this finite evidence resolves, without imposing any of those
interpretations in advance.

\subsection{Scaling dimensions belong to observations}
The linear row contrast and its squared energy are different
observations. Their scaling dimensions need not agree even when a
single collective mode controls both.

\begin{proposition}[Nonlinear visibility of a fluctuation scale]
\label{model:prop:nonlinear-scaling}
Let $p\ge1$ be an integer, $\Delta\in\R$, and
$q_N=N^{-\Delta}X_N$. Suppose $X_N$ converges in distribution to $X$,
the family $|X_N|^{2p}$ is uniformly integrable, and
$0<\Var(X^p)<\infty$. Let
$Y_N=c_Nq_N^p+e_N$, where $c_N\to c\ne0$ and
\[
 N^{p\Delta}\norm{e_N-\E e_N}_{L^2}\longrightarrow0.
\]
Then
\begin{equation}
 N\Var(Y_N)\sim c^2\Var(X^p)N^{1-2p\Delta}.
 \label{model:eq:nonlinear-scaling}
\end{equation}
If $0<\Var(X)<\infty$, writing $\kappa_q=1-2\Delta$ gives
$\kappa_Y=p\kappa_q-(p-1)$.
\end{proposition}
\begin{proof}
Distributional convergence and uniform integrability imply convergence
of the moments of orders $p$ and $2p$, hence
$\Var(X_N^p)\to\Var(X^p)$. Center $Y_N$ and multiply by
$N^{p\Delta}$. Its main term is $c_N(X_N^p-\E X_N^p)$ and its
remaining term tends to zero in $L^2$. Cauchy--Schwarz bounds the
cross term by the product of their $L^2$ norms. Expanding the square
therefore gives
$N^{2p\Delta}\Var(Y_N)\to c^2\Var(X^p)$, proving the result.
The exponent identity follows by substitution.
\end{proof}

For example, a linear mode with $\kappa_q=1/2$ can have a quadratic
energy with $\kappa_Y=0$. Thus failure of the normalized row-energy
susceptibility to diverge does not exclude every possible linear
critical mode. Conversely, a surviving common coordinate with
order-one variance gives an extensive susceptibility. Identifying its
physical origin still requires the ensemble decomposition and
gauge-invariant emission. Application to the native ratio $R=E/T$
also requires control of $T$ and of the remainder of any proposed
single-mode expansion. The exact row/common decomposition does not
assume that expansion.

\subsection{Dynamic closure near a slow sector}
Small one-step or fixed-state prediction error can accumulate on a
long collective relaxation time. A sufficient observation-accuracy condition
for susceptibility transfer depends on the exponent and response gap.

\begin{proposition}[Signed susceptibility transfer]
\label{model:prop:signed-transfer}
Let $X_N,e_N$ be square-integrable scalar variables on the same
specified probability space, with $\chi_N=N\Var(X_N)>0$. Write
$\widetilde\chi_N=N\Var(X_N+e_N)$ and
$a_N=\norm{e_N-\E e_N}_{L^2}$. Then
\begin{align}
 \widetilde\chi_N-\chi_N
 &=N\{2\Cov(X_N,e_N)+\Var(e_N)\},\label{model:eq:signed-transfer}\\
 |\widetilde\chi_N-\chi_N|
 &\le2\sqrt{N\chi_N}\,a_N+Na_N^2.\label{model:eq:centered-transfer}
\end{align}
If $\chi_N\sim C N^\kappa$, $C>0$, the condition
$a_N=o(N^{(\kappa-1)/2})$ is sufficient to preserve this leading
asymptotic. For the specified coupling, preservation of the same
leading asymptotic is equivalent to
\[
 2\Cov(X_N,e_N)+\Var(e_N)=o(\Var(X_N)).
\]
\end{proposition}
\begin{proof}
Apply Lemma~\ref{model:lem:master-perturbation} under the specified coupling
with $A=1$, $X=X_N$ and $e=e_N$. Its scalar identity, centered-error
bound and normalized equivalence give all the assertions.
\end{proof}

Centering contracts $L^2$, so the stronger condition with
$\norm{e_N}_{L^2}$ is also sufficient. It is not necessary for
a particular coupled pair. For example, if $Z=\pm1$ equiprobably,
$0<a<1/2$ and $0\le\kappa\le1$, the bounded observations
\[
 X_N=\tfrac12+aN^{(\kappa-1)/2}Z,\qquad
 \widetilde X_N=\tfrac12-aN^{(\kappa-1)/2}Z
\]
both have susceptibility $a^2N^\kappa$. Their difference has
$L^2$ norm $2aN^{(\kappa-1)/2}$: its variance cancels twice its
signed covariance with $X_N$. A deterministic shift likewise
has no effect on susceptibility. Preservation of only an exponent,
allowing a different leading constant, is weaker than the
equivalence in Proposition~\ref{model:prop:signed-transfer}.

\begin{proposition}[Accumulated closure and susceptibility transfer]
\label{model:prop:critical-closure}
Couple two processes on the same declared probability space and let
$\varepsilon_t$ bound the $L^2$ distance of their scalar observations.
Suppose a closed discrepancy estimate gives
\[
 \varepsilon_{t+1}\le\rho_N\varepsilon_t+\delta_N,
 \qquad 0\le\rho_N<1,\quad \delta_N\ge0.
\]
Then
\begin{equation}
 \varepsilon_t\le\rho_N^t\varepsilon_0+
             \frac{\delta_N(1-\rho_N^t)}{1-\rho_N}.
 \label{model:eq:accumulated-closure}
\end{equation}
If their susceptibilities are $\chi_{N,t}$ and $\widetilde\chi_{N,t}$,
with normalization $N$, then
\begin{equation}
 |\widetilde\chi_{N,t}-\chi_{N,t}|
 \le2\sqrt{N\chi_{N,t}}\,\varepsilon_t+N\varepsilon_t^2.
 \label{model:eq:dynamic-chi-transfer}
\end{equation}
If $\chi_{N,t_N}\sim C N^\kappa$, $C>0$, the condition
$N^{(1-\kappa)/2}\varepsilon_{t_N}\to0$ transfers this asymptotic
with the same leading constant. In particular, when
$1-\rho_N\asymp N^{-\zeta}$, a sufficient forcing condition is
\begin{equation}
 \delta_N=o\bigl(N^{(\kappa-1)/2-\zeta}\bigr),
 \label{model:eq:critical-closure-window}
\end{equation}
together with the corresponding vanishing initial transient.
\end{proposition}
\begin{proof}
Iterate the scalar recursion and sum its finite geometric series to
obtain Equation~\eqref{model:eq:accumulated-closure}. If $e$ is the difference
of the observations, then $\Var(e)\le\E e^2\le\varepsilon_t^2$.
The scalar bound in Lemma~\ref{model:lem:master-perturbation} gives
Equation~\eqref{model:eq:dynamic-chi-transfer}. Divide it by $N^\kappa$.
Both terms vanish under the stated observation condition. Finally
substitute the geometric bound and the assumed order of
$1-\rho_N$.
\end{proof}

The discrepancy recursion is an additional dynamic hypothesis; it
does not follow from predictive fidelity at one frozen state.
The full clipped-Adam tangent and the measured shared-force transport
identify which terms a reduced evolution must retain. A closure that
is accurate on ordinary time intervals can consequently fail in a
critical relaxation window.
The tangent is derived in Section~\ref{model:sec:row-optimizer-theory}; Section~\ref{model:sec:adjoint-source-results} gives the source-transport measurements.

\subsection{Moving peaks and a limiting source potential}
An enhanced finite-time fluctuation sector need not be a singular
thermodynamic transition. This distinction persists even when its
variance has an exact power of size.

\begin{proposition}[An exact moving-peak family]
\label{model:prop:moving-peak}
Fix $0<\kappa<1$ and $a>0$. Let $Z$ be equally likely to be $-1$ or
$1$, set $F(u)=u e^{-u}$, and define
\[
 q_{N,g,t}=\tfrac12F(gt/N^a)
             \bigl(1+N^{(\kappa-1)/2}Z\bigr),\qquad N\ge1,\quad g,t\ge0.
\]
This is a bounded nonnegative intensive field, with
\begin{equation}
 N\Var(q_{N,g,t})=\tfrac14N^\kappa F(gt/N^a)^2.
 \label{model:eq:moving-peak}
\end{equation}
At fixed $N,t>0$ its susceptibility peaks at $g=N^a/t$. At fixed
$N,g>0$, the law converges to a point mass at zero as $t\to\infty$.
Along $gt/N^a=u>0$, its source potential converges locally uniformly
in $j$ to the analytic function $jF(u)/2$, even though its curvature
at zero diverges as $N^\kappa$.
\end{proposition}
\begin{proof}
Since $0\le F(u)\le1/e$ for $u\ge0$ and
$0<N^{(\kappa-1)/2}\le1$, the field lies in $[0,1/e]$.
The variance of $Z$ gives Equation~\eqref{model:eq:moving-peak}.
The derivative $F'(u)=(1-u)e^{-u}$ locates the unique positive peak,
and $F(u)\to0$ at infinity proves the fixed-size long-time limit.
For fixed $u$, put $m=F(u)/2$ and
$d_N=mN^{(\kappa-1)/2}$. Directly,
\[
 \frac1N\log\E e^{Njq_{N,g,t}}
   =mj+\frac1N\log\cosh(Njd_N).
\]
The second term lies between zero and $|j|d_N$, which tends to zero
uniformly on compact source intervals. Its second derivative at zero
is $Nd_N^2=m^2N^\kappa$. Thus convergence of these convex potentials
does not license interchange with their second derivatives.
\end{proof}

This construction has a fluctuation scale in the sense of
Section~\ref{model:sec:criticality}; it has no singular limiting potential
on the stated orbit. A fitted moving peak must therefore be accompanied
by the declared joint limit, a critical control and a limiting joint law.
A related finite mechanism is heterogeneous collapse time. If a field
has two conditional branch means differing by $d$ and branch
probability $p$, the law of total variance contributes $Nd^2p(1-p)$
to its susceptibility, in addition to the within-branch term. A smooth
motion of $p$ through $1/2$ can produce a large transient peak. This
decomposition retains both branches; it does not discard a learned
common fluctuation by conditioning on its final value.

\subsection{Conditional histories and stationary laws}
The fixed-width invariant-law existence result concerns the kernel
that averages fresh data batches. The head susceptibility measurements
instead condition on an exogenous shared initialization and a complete
batch history. These laws must be distinguished when judging a long
training horizon. A stationary law after averaging the environment
can have conditional means that continue to move with that environment.

For an explicit example, let $|a|<1$, let $(B_t)_{t\in\mathbb Z}$ be
IID bounded variables, and consider
\[
 X_{t+1}=aX_t+B_t,\qquad
 X_t^*=\sum_{k=0}^{\infty}a^k B_{t-1-k}.
\]
Absolute convergence proves that $X_t^*$ solves the recursion.
The process is stationary and
$\Var(X_t^*)=\Var(B_0)/(1-a^2)$ by independence. Two solutions sharing
the batches differ by $a^t$ times their initial difference, so initial
conditions are forgotten. Conditional on the entire batch sequence,
$X_t^*$ is a point mass at a generally changing value. Thus neither
pointwise constancy of a conditional trajectory nor agreement of two
individual checkpoints is the definition of stationarity.

For native training, convergence of the evolving conditional law,
existence of an averaged invariant law, and criticality of a chosen
family of either law remain separate statements. The matched
initialization and batch-history measurements identify their finite
conditioning sensitivity. Longer trajectories test the specified
finite-horizon laws and temporal response; their lengths alone do not
turn these statements into an asymptotic theorem. Likewise, an
endogenous return mechanism establishes self-organized criticality
only when its attracting set is independently located inside a
critical scaling window of the same physical family.
Section~\ref{model:sec:extended-environment-results} gives the crossed environments and extended-horizon comparisons.

\subsection{Feedback with a retained correlation time}
A feedback estimate based only on fresh-batch variance can miss a
correlated forcing sector. The complete incoming Adam state and the
shared-update time series distinguish conditional batch noise from
forcing correlations along training. The following extension of
Proposition~\ref{model:prop:feedback} quantifies this distinction in an
explicit normal form.

\begin{proposition}[Correlated forcing and a critical window]
\label{model:prop:colored-feedback}
Let $0<a_N<1$, $0\le\rho_N<1$, and let $(\xi_t)$ be a centered
stationary process with
$\Cov(\xi_t,\xi_{t+k})=\rho_N^{|k|}$. Suppose the signed distance to
an independently known critical surface obeys
\[
 r_{t+1}=(1-a_N)r_t+d_N+\sigma_N\xi_t.
\]
Its stationary solution with finite second moment has
\begin{equation}
 \E r=\frac{d_N}{a_N},\qquad
 \Var r=\frac{\sigma_N^2}{a_N(2-a_N)}
       \frac{1+(1-a_N)\rho_N}{1-(1-a_N)\rho_N}.
 \label{model:eq:colored-feedback}
\end{equation}
For a window $N^{-\phi}$, the condition
$N^{2\phi}[(\E r)^2+\Var r]\to0$ is sufficient for
$N^\phi r\to0$ in probability in this stationary law. If
$a_N\asymp N^{-\zeta}$, $\zeta>0$, and
$1-\rho_N\asymp N^{-m}$, $m\ge0$, then
\begin{equation}
 \Var r\asymp\sigma_N^2 N^{\zeta+\min(\zeta,m)}.
 \label{model:eq:colored-feedback-order}
\end{equation}
In this normal form sufficient bias and noise conditions are therefore
\[
 \frac{d_N}{a_N}=o(N^{-\phi}),\qquad
 \sigma_N^2=o\bigl(N^{-2\phi-\zeta-\min(\zeta,m)}\bigr).
\]
\end{proposition}
\begin{proof}
Put $c=1-a_N$. The series
$r_t=d_N/a_N+\sigma_N\sum_{k\ge0}c^k\xi_{t-1-k}$ converges in
$L^2$, since its summands have $L^2$ norms bounded by
$|\sigma_N|c^k$. It defines a stationary solution. Summing the
diagonal and the two off-diagonal parts of its covariance gives
\[
 \sum_{k,l\ge0}c^{k+l}\rho_N^{|k-l|}
 =\frac{1}{1-c^2}\left(1+2\sum_{j\ge1}(c\rho_N)^j\right)
 =\frac{1+c\rho_N}{(1-c^2)(1-c\rho_N)}.
\]
All series converge absolutely. This proves
Equation~\eqref{model:eq:colored-feedback}; Markov's inequality gives the
window statement. For the order estimate,
$1-c\rho_N=a_N+(1-\rho_N)-a_N(1-\rho_N)$ is comparable to
$a_N+(1-\rho_N)$, and $1+c\rho_N$ is bounded above and below by
positive constants. Substitution proves
Equation~\eqref{model:eq:colored-feedback-order}.
\end{proof}

Two solutions coupled to the same forcing differ by $c^t$ times their
initial difference. A finite horizon consequently also needs the
corresponding scaled transient to vanish. Neither geometric forcing
covariance nor the signed critical distance in this proposition is
assumed to be the native PLDR training law. In particular, a measured
Adam coefficient is not an estimate of $\rho_N$ for the evolving
nonlinear shared force. If a noise correlation time grows as fast as
or faster than the restoring time, its additional factor changes the
window condition. Temporal covariance and instantaneous conditional
variance cannot then be substituted for one another.

An adaptive optimizer's numerical stability boundary is also distinct
from a thermodynamic critical surface. Adaptive edge-of-stability
measurements concern the preconditioned loss Hessian in specified
full-batch or sufficiently large-batch settings
\cite{cohen2022adaptive}. Importing their threshold into a clipped,
stochastic two-rate update would require the appropriate augmented
linearization, including preconditioner and clipping transport. The
native tangent derived in Section~\ref{model:sec:row-optimizer-theory}
retains these terms. Even a validated
numerical stability threshold would additionally need a size-dependent
collective law and a visible limiting response before it supplied
critical exponents for this architecture.

\begin{proposition}[Visibility at the fluctuation scale]
\label{model:prop:scalar-fluctuation-visibility}
Let $X_N,r_N$ have finite second moments under the same specified
ensemble, and let $Y_N=b_NX_N+r_N$, where $b_N$ is deterministic under
that ensemble. Define $\chi_X=N\Var X_N$ and $\chi_Y=N\Var Y_N$.
Then
\begin{equation}
 |\chi_Y-b_N^2\chi_X|\le
 N\left(2|b_N|\sqrt{\Var X_N\Var r_N}+\Var r_N\right).
 \label{model:eq:scalar-fluctuation-visibility}
\end{equation}
If $\chi_X\sim C N^\kappa$ with $C>0$, $b_N\to b\ne0$, and
$\|r_N-\E r_N\|_{L^2}=o(N^{(\kappa-1)/2})$, then
$\chi_Y\sim b^2 C N^\kappa$.
\end{proposition}
\begin{proof}
Center the equality defining $Y_N$ under the stated ensemble. Expansion
gives $\chi_Y-b_N^2\chi_X=N[2b_N\Cov(X_N,r_N)+\Var r_N]$.
Cauchy--Schwarz bounds the absolute covariance by
$\sqrt{\Var X_N\Var r_N}$, proving the inequality. The residual
assumption makes its two right-hand terms $o(N^\kappa)$; the asserted
asymptotic follows from $b_N^2\chi_X\sim b^2CN^\kappa$.
\end{proof}

This sufficient condition explicitly joins a training fluctuation scale
to an inference observable. It is not necessary for a specified coupled
residual, because the signed covariance terms can cancel. Random
projections require their conditional and outer covariance sectors.
The finite support-null experiment and the optimizer-state intervention
test the two possibilities of an absent and an emergent predictive
projection; neither identifies a thermodynamic critical mode.
Sections~\ref{model:sec:invisible-sector-results} and \ref{model:sec:optimizer-results} give the respective designs and full outcomes.

\begin{proposition}[Vector transport of a fluctuation scale]
\label{model:prop:vector-fluctuation-visibility}
Let $X_N,Y_N,r_N$ be centered square-integrable vectors under the same
conditional law, with $Y_N=A_NX_N+r_N$ and deterministic linear $A_N$.
Write $v_X=\E\|X_N\|^2$, $v_r=\E\|r_N\|^2$ and
$v_A=\E\|A_NX_N\|^2$. Then
\begin{equation}
 \bigl|\E\|Y_N\|^2-v_A\bigr|\le2\sqrt{v_Av_r}+v_r.
 \label{model:eq:vector-fluctuation-visibility}
\end{equation}
If $\sup_N\|A_N\|<\infty$, $v_A/v_X\to a>0$ and
$v_r/v_X\to0$, then $\E\|Y_N\|^2/v_X\to a$.
All finite identities and bounds hold for centered empirical vectors
using a common covariance divisor and fixed context weights.
\end{proposition}
\begin{proof}
Expand $\|A_NX_N+r_N\|^2$ and take expectations. The cross inner
product has absolute value at most $\sqrt{v_Av_r}$ by Cauchy--Schwarz
in the space of square-integrable vectors. This proves the inequality.
Divide by $v_X$ and use $v_A\le\|A_N\|^2v_X$ to prove the limit.
The same expansion and Cauchy--Schwarz apply to a finite weighted sum.
\end{proof}

A small absolute predictive error does not imply the relative remainder
condition. This proposition applies after centering under the specified
training ensemble. Freezing weights stops optimizer-time evolution;
persistence in inference refers to this ensemble of predictive laws or
to a separately specified source-response or generated process.

\begin{proposition}[Categorical elimination with a fixed tail law]
\label{model:prop:categorical-tail-visibility}
Fix a proper subset $I$ of a finite vocabulary and a strictly positive
probability vector $r$ on its complement $J$. Let $Kp$ retain each
$p_i$, $i\in I$, and the single tail mass $q_J=\sum_{j\in J}p_j$.
The lift $R$ restores the retained coordinates and assigns $q_Jr_j$
to each $j\in J$. Both $K$ and $R$ are stochastic channels, $KR$ is
the identity and $RK$ is idempotent. For positive $p$,
\begin{equation}
 \KL(p\|RKp)=q_J\KL(p(\cdot\mid J)\|r).
 \label{model:eq:categorical-tail-kl}
\end{equation}
Let $\phi(p)=2\sqrt p$. There is a fixed linear isometry $A$ with
$\phi(RKp)=A\phi(Kp)$, and hence, for the corresponding population
or unbiased empirical variance traces,
\begin{gather}
 v_{RKp}=v_{Kp}\le v_p,\notag\\
 0\le v_p-v_{Kp}\le2\sqrt{v_{Kp}v_\rho}+v_\rho,
 \qquad \rho=\phi(p)-\phi(RKp),
 \label{model:eq:categorical-tail-variance}
\end{gather}
where $v_\rho$ is the centered variance trace of $\rho$. Fixed context
averaging preserves the statements, with a fixed $r$ for each context.
Nested retained-token sets give compatible elimination channels.
\end{proposition}
\begin{proof}
Channel columns sum to one. Applying $K$ to a lifted distribution
recovers every retained coordinate and the tail mass, proving $KR=1$
and $(RK)^2=RK$. Retained-coordinate terms in the KL sum vanish. Each
tail term equals $p_j\log[p_j/(q_Jr_j)]$, giving
Equation~\eqref{model:eq:categorical-tail-kl} by summation.
The columns of $A$ are the retained coordinate unit vectors and the
vector with entries $\sqrt{r_j}$ on $J$ and zero elsewhere. They have
disjoint supports and unit norms, so $A^{\mathsf T}A=1$ and direct
substitution proves the embedding identity. Isometry gives equal
variance traces; categorical Hellinger contraction in
Proposition~\ref{model:prop:predictive-hellinger} gives $v_{Kp}\le v_p$.
Center the embedding identity and apply
Proposition~\ref{model:prop:vector-fluctuation-visibility}. Finally, deleting
more retained coordinates and adding their masses to the tail produces
the same $Kp$ as deleting them in one step, which proves compatibility.
\end{proof}

The maps act on a predictive law already acquired from the model.
A tail dictionary and a small representation error do not provide a
successor law for the reduced state. A context-dependent tail dictionary
also has storage and acquisition costs separate from the coarse vector.

\section{Training selection relative to a shrinking critical window}
\label{model:sec:selection-window}

Fix a family under the same pretraining law $\D$, document and
token-block construction, and conditioning level. Head count $N$
determines residual width $dN$; head dimension $d$, decoder depth and
metric-generator width stay fixed independently of $N$. The executed
families use $d=64$, depth five and metric-generator width 170.
The specification also includes
initialization, optimizer scaling, batch $B_N$, horizon $t_N$, corpus
size $M_N$, consumed fraction $B_Nt_N/M_N$, schedule phase and
arithmetic policy. Conditional-on-corpus and corpus-averaged limits
have different fluctuation sectors. A physical input length $L$ and
an adaptation or sampling clock are additional variables, not substitutes
for $N$.

A shrinking critical window is meaningful only after a critical
surface and distance $r_{N,t}$ have been identified independently of
the trajectory being tested. The following benchmark specifies a
training-selection mechanism and the noise control it needs.

\begin{proposition}[Restoring drift with a finite-time noise budget]
\label{model:prop:selection-window}
For each $N$, let $0<a_N<1$, $w_N>0$, and
\[
 r_{t+1}=(1-a_N)r_t+b_N+\xi_t,
\]
with deterministic $r_0$ and square-integrable centered forcing.
Set $g=1-a_N$. Then
\begin{align*}
 \E r_t&=g^tr_0+\frac{b_N}{a_N}(1-g^t),\\
 \Var(r_t)&=\sum_{j,k<t}g^{t-1-j}g^{t-1-k}\Cov(\xi_j,\xi_k).
\end{align*}
If the forcing is independent with common variance $\sigma_N^2$,
the second expression is
$\sigma_N^2(1-g^{2t})/(2a_N-a_N^2)$.
At times $t_N$, sufficient conditions for
$r_{N,t_N}/w_N\to0$ in probability are
\[
 \frac{g^{t_N}|r_0|}{w_N}\to0,\qquad
 \frac{|b_N|}{a_Nw_N}\to0,\qquad
 \frac{\Var(r_{N,t_N})}{w_N^2}\to0.
\]
In the independent-forcing case the last condition follows from
$\sigma_N/(\sqrt{a_N}w_N)\to0$.
\end{proposition}
\begin{proof}
Induction gives
$r_t=g^tr_0+b_N\sum_{j<t}g^j+\sum_{j<t}g^{t-1-j}\xi_j$.
Take expectations and sum the geometric series. Centering and expanding
the square of the last sum gives the full covariance expression.
Independence deletes its off-diagonal terms, and summing the remaining
geometric series gives the displayed variance. Since
$2a_N-a_N^2\ge a_N$, the stated independent-forcing condition is
sufficient. The first two conditions make the mean divided by $w_N$
vanish. Chebyshev's inequality applied to the centered variable proves
convergence in probability.
\end{proof}

This proposition is a conditional mechanism, not an identified AdamW
law. In single-pass training the remaining corpus and optimizer memory
can color the forcing and alter its drift. Their measured finite
effects motivate retaining the covariance sum and the augmented state.
A prescribed drive selected at a peak cannot establish self-organization;
nor do restoring drift and a finite response identify a critical surface.
The empirical results establish finite training-selected emissions and
conditional memory requirements. Native attraction into a shrinking
critical window is not established by those measurements.

\FloatBarrier
\par\medskip\noindent
Chapter~\ref{ch:metric-reduction} applies these scale questions to the
native optimizer and metric-row variables. Its tangent, projection and
observation analysis identifies which fluctuations survive a proposed reduction.

\chapter{Optimizer scaling and metric-row reduction}
\label{ch:metric-reduction}
This chapter examines optimizer scaling and reductions of the learned
metric-row sector. It combines augmented Adam tangents, clipping and offset
scales, shared-state interventions, nonlinear covariance transport and
numerical observation bounds with explicit fluctuation-fidelity requirements.

\section{The augmented Adam tangent and its temporal composition}
\label{model:sec:row-optimizer-theory}
\label{model:sec:adam-tangent}
A physical training perturbation changes an augmented state, so its
local transport must retain the optimizer moments. The loss Hessian
alone does not give that transport. We now differentiate the declared
native update at fixed batch and counter in real arithmetic.

Let $g=\nabla_\theta\ell(\theta;B)$, $G=\|g\|_2$, and write the
implemented global clipping map as
\[
 \widetilde g=cg,\qquad c=\min\{1,C_g/(G+\delta_c)\},
 \quad C_g=1,\quad\delta_c=10^{-6}.
\]
The positive offset here belongs to gradient clipping and is distinct
from the Adam denominator offset $\epsilon_o$. For an incoming weight
direction $v_\theta$, set $h=\nabla^2\ell\,v_\theta$. Away from the
clipping boundary $G+\delta_c=C_g$, its directional derivative is
\begin{equation}
 \delta\widetilde g=ch+\delta c\,g,\qquad
 \delta c=\begin{cases}
 0,&G+\delta_c<C_g,\\
 -\displaystyle\frac{C_g\langle g,h\rangle}
 {G(G+\delta_c)^2},&G+\delta_c>C_g.
 \end{cases}
 \label{model:eq:clip-tangent}
\end{equation}
The un-clipped case includes $G=0$.

At an update with counter $k+1$, put
$a_i=1-\beta_i^{k+1}$, and use componentwise products in
\begin{alignat}{2}
 m^+&=\beta_1m+(1-\beta_1)\widetilde g,&\quad v^+&=\beta_2v+(1-\beta_2)\widetilde g^{\odot2},\notag\\
 s&=\sqrt{v^+/a_2},&\quad d&=s+\epsilon_o.\notag
\end{alignat}
The coordinate rates $\eta_i$ are held fixed in the physical state
perturbation. Its three components are denoted
$(v_\theta,v_m,v_v)$.

\begin{proposition}[Local augmented update transport]
\label{model:prop:adam-tangent}
Suppose the batch loss is twice continuously differentiable near the
state, the clipping boundary is avoided, and each coordinate satisfies
either $v_i^+>0$, or $v_i^+=m_i^+=0$ with admissible zero first-order
variation of $v_i^+$. Then the update direction is
\begin{align}
 v_m^+&=\beta_1v_m+(1-\beta_1)\delta\widetilde g,\notag\\
 v_v^+&=\beta_2v_v+2(1-\beta_2)
                           \widetilde g\odot\delta\widetilde g,\notag\\
 v_{\theta,i}^+&=(1-\eta_i\lambda_d)v_{\theta,i}
 -\eta_i\left[\frac{v_{m,i}^+}{a_1d_i}
       -\frac{m_i^+v_{v,i}^+}{2a_1a_2s_i d_i^2}\right].
 \label{model:eq:adam-tangent}
\end{align}
The last bracket's second term is zero at the admitted zero-moment
coordinates. The tangent includes the derivative of the global clipping
factor as well as both moment directions.
\end{proposition}
\begin{proof}
For $G>0$, differentiating its squared norm gives
$\delta G=\langle g,h\rangle/G$. Differentiating the two smooth branches
of the clipping map gives Equation~\eqref{model:eq:clip-tangent}. In a
neighborhood of $G=0$ clipping is inactive. The first two lines of
Equation~\eqref{model:eq:adam-tangent} then follow from the affine first-moment
recursion and the product rule for the second moment. At $v_i^+>0$,
$\delta s_i=v_{v,i}^+/(2a_2s_i)$. Apply the quotient rule to
$m_i^+/(a_1d_i)$ and differentiate weight decay to obtain the last line.

At a zero-moment coordinate, the map
$F(m,v)=m/(\sqrt{v/a_2}+\epsilon_o)$ on $v\ge0$ has relative derivative
$\delta m/\epsilon_o$ at $(0,0)$. Indeed,
\[
 \left|F(m,v)-m/\epsilon_o\right|
 \le |m|\sqrt{v/a_2}/\epsilon_o^2,
\]
which, divided by $\sqrt{m^2+v^2}$, tends to zero. This proves the
specified zero denominator contribution along admissible directions.
A zero second moment with nonzero first moment is excluded from this
argument. In exact reachable Adam histories with positive moment decay,
zero second moment forces all contributing gradients, and hence the
first moment, to vanish. Native underflow need not preserve that exact
implication and requires a separate arithmetic check.
\end{proof}

For this derivative, $S=(\theta,m,v)$ denotes the continuous state on
a fixed realized counter sequence. The physical sources leave that
sequence unchanged. For an initial source $\alpha\mapsto S_t(\alpha)$
and its direction $V_t$, let $J_j=DT_{B_j,k_j}(S_j)$ denote the augmented
update derivative on these fixed-counter fibers.
Whenever the displayed differentiability conditions hold along the
finite path, the chain rule gives
\begin{equation}
 V_{t+b}=J_{t+b-1}\cdots J_tV_t,\qquad
 \partial_\alpha\Phi(S_{t+b}(\alpha))\big|_0
 =D\Phi(S_{t+b})J_{t+b-1}\cdots J_tV_t.
 \label{model:eq:adam-emission-tangent}
\end{equation}
Splitting the ordered product at any intermediate time proves temporal
composition and compatibility with the inference emission. Optimizer
memory is retained by the first two lines of
Equation~\eqref{model:eq:adam-tangent}; discarding those coordinates generally
changes the product. The batch sequence and the evaluation contexts
remain distinct sources of variation.

These identities define a local response of the specified mathematical
program. A finite product can amplify a displacement through transient
or nonlinear-state effects without identifying a stationary critical
mode. Eigenvalues of one local Jacobian, a finite-time growth rate,
and a thermodynamic susceptibility are different quantities. Relating
them requires the collective reduction, its noise law and the limiting
regime stated in Proposition~\ref{model:prop:collective-stability}.
The numerical controls compare the full-graph Hessian-vector calculation,
this augmented tangent, and decreasing-amplitude physical-weight pulses
with identical incoming moments. Their arithmetic and amplitude windows
are measured independently of the instantaneous inference windows.

\section{Clipping, denominator offsets and the width limit}
\label{model:sec:optimizer-scales}
The native family keeps the Adam offset fixed at
$\epsilon_o=10^{-8}$. This is a physical convention of the declared
update, including when the head count tends to infinity. A comparison
of finite learning rates alone does not determine its asymptotic
adaptive dynamics.

\begin{proposition}[A coordinatewise decay-dominated update sector]
\label{model:prop:adam-decay-sector}
In real arithmetic, use Adam moments with $0\le\beta_1,\beta_2<1$,
start with zero moments, and consider a specified set of coordinates
with constant rate $\eta_N\ge0$, decay $\lambda_d>0$,
$0\le\eta_N\lambda_d\le1$, and offset $\epsilon_N>0$.
If every clipped gradient in these coordinates has absolute value at
most $b_N$ during the first $T_N$ updates, then
\begin{align}
 \left|\theta_{i,t}-(1-\eta_N\lambda_d)^t\theta_{i,0}\right|
 &\le\frac{b_N}{\lambda_d\epsilon_N}
             \left[1-(1-\eta_N\lambda_d)^t\right]\notag\\
 &\le\eta_N t\,b_N/\epsilon_N,
 \qquad t\le T_N.\label{model:eq:adam-decay-sector}
\end{align}
Thus $\eta_N T_N b_N/\epsilon_N\to0$ is sufficient for uniform
coordinatewise agreement with the pure weight-decay path on that
horizon. If the unclipped coordinates are bounded by $B_N$ and the
full gradient norm is at least $c_N\ge0$, the stated native clipping map
permits
\[
 b_N=B_N\min\{1,C_g/(c_N+\delta_c)\}.
\]
\end{proposition}
\begin{proof}
The bias-corrected first moment is a convex combination of the preceding
clipped gradients, so its absolute value is at most $b_N$.
The adaptive ratio is therefore at most $b_N/\epsilon_N$.
Subtract the pure decay recursion from the AdamW update and iterate.
The resulting geometric sum gives the first bound, including a zero
rate. Bounding each factor in the sum by one gives the second.
Finally, multiply the unclipped coordinate bound by the upper bound
on the clipping factor supplied by the lower bound on the full norm.
\end{proof}

The gradient bounds in this proposition are hypotheses about the
whole training path. A finite set of saved states does not certify
them uniformly in width. Moreover, a coordinatewise bound need not
control an emission that combines a growing number of coordinates;
the full graph transport remains necessary. The pure decay path
itself can change a fixed shared generator when its rate remains
nonzero.

Already one Adam step shows why the offset limit matters. If a clipped
coordinate gradient at the first step is $u/\sqrt N$, with $u\ne0$,
its bias-corrected adaptive ratio is exactly
\begin{equation}
 \frac{u/\sqrt N}{|u|/\sqrt N+\epsilon_N}
       =\frac{u}{|u|+\epsilon_N\sqrt N}.
 \label{model:eq:adam-offset-limit}
\end{equation}
The ratio tends to zero for fixed positive offset, to
$u/(|u|+e)$ if $\epsilon_N\sqrt N\to e\in(0,\infty)$, and to
$\operatorname{sign}(u)$ if that product tends to zero.
These limits follow directly by division. They are distinct update
families; none is inferred from the initial variance correction alone.
In particular this conditional example does not assert the
$N^{-1/2}$ gradient scaling for native trained PLDR coordinates.
The recorded moment and clipping diagnostics locate the measured
finite family relative to its actual denominator scales.
Section~\ref{model:sec:optimizer-scale-results} reports those diagnostics at the recorded widths and training ages.

\section{Interventions on learned shared collectives}
\label{model:sec:learned-collectives}
The metric learner is shared across the heads of one decoder. Conditioning
on its initial value leaves its subsequently trained state inside the
endogenous law. A component intervention can identify its influence
without declaring that state an external nuisance variable.

Split a trained parameter state into recipient coordinates $r$ and shared
metric coordinates $a$. At a fixed evaluation context let
$q(r,a)\in\R^k$ be the full inference emission after combining the two
components. A balanced crossing of finitely many recipients and donors
defines a counterfactual product law $\rho\otimes\alpha$. This law is
different from the factual joint training law, which retains correlations
between the components.

\begin{proposition}[Exact crossed collective decomposition]
\label{model:prop:crossed-collective}
For any square-integrable $q$ under this product law, set
\[
 \overline q=\E q,\quad q_r=\E[q\mid r]-\overline q,\quad
 q_a=\E[q\mid a]-\overline q,\quad
 q_{ra}=q-\overline q-q_r-q_a.
\]
Then
\begin{equation}
 \Cov(q)=\E q_rq_r^{\mathsf T}+\E q_aq_a^{\mathsf T}
                      +\E q_{ra}q_{ra}^{\mathsf T}.
 \label{model:eq:crossed-collective}
\end{equation}
The decomposition commutes with every deterministic linear observation
map. It does not, in general, equal the covariance decomposition under
the factual joint law.
\end{proposition}
\begin{proof}
The first two components have mean zero and are independent under the
product law. Direct conditional expectation gives
$\E[q_{ra}\mid r]=\E[q_{ra}\mid a]=\bm0$. Conditioning each cross product
on its relevant coordinate therefore makes every mixed term vanish.
Expanding the covariance proves Equation~\eqref{model:eq:crossed-collective}.
Linearity of conditional expectation shows that each component of $Tq$
is the corresponding component of $q$ multiplied by $T$, which proves
compatibility. A factual joint law need not make $r$ and $a$ independent,
so the orthogonality calculation does not apply to that law.
\end{proof}

Thus a dominant donor contribution identifies an intervention-sensitive
shared coordinate of the measured field. A large interaction contribution
identifies recipient--donor compatibility. Neither conclusion implies
that the shared coordinate is noncritical. The full nonlinear native
graph remains in the emission used for this intervention; an additive
approximation to its input is unnecessary.
Section~\ref{model:sec:shared-crossing-results} specifies the recipient--donor experiment and its complete variance decomposition.

A repeated component constraint changes the training kernel as well as
its initial state. For two declared updates $T,T'$ driven by the same
batch, the paired kernel is
$P_{T,T'}F(S,S')=\E_B F(T(S,B),T'(S',B))$.
Conditioning on either coordinate gives
$P_{T,T'}(f\circ\pi_1)=(Pf)\circ\pi_1$ and
$P_{T,T'}(f\circ\pi_2)=(P'f)\circ\pi_2$.
Iterating these identities proves compatibility at every temporal block.
Restoring a fixed component after each optimizer step is such a declared
$T'$, with the original moment recursion retained; it need not have the
same stationary marginal law as $T$.

\section{Row concentration and the order of width and time limits}
For the native normalized row energy, $0\le q_N\le1$ at every context
and decoder: subtracting the row mean is an orthogonal projection, and
the denominator is at least the uncentered squared norm. This elementary
bound gives an absolute diagnostic even when a microscopic variance
reference becomes small.

\begin{proposition}[Concentration bound for a bounded row field]
\label{model:prop:row-concentration}
At fixed context and training conditions, let
$p_N=\E_{\xi_N}q_N$ and $\chi_N=N\Var_{\xi_N}(q_N)$. Then
\begin{equation}
 0\le\chi_N\le Np_N(1-p_N)\le Np_N.
 \label{model:eq:row-concentration}
\end{equation}
Consequently $Np_N\to0$ implies vanishing absolute susceptibility.
For a $k$-component bounded row field, the trace of its susceptibility
is at most $N\sum_{\ell=1}^k p_{N,\ell}$, including after averaging
over the evaluation-context law. These statements apply along the
specified sequence of training horizons as well as at stationarity.
\end{proposition}
\begin{proof}
The pointwise inequality $q_N^2\le q_N$ gives
$\Var(q_N)=\E q_N^2-p_N^2\le p_N(1-p_N)$. Multiplication by $N$
proves the scalar bound. Summing the diagonal scalar inequalities gives
the trace bound; expectation over contexts preserves it. Positive
semidefiniteness bounds every eigenvalue by the trace.
\end{proof}

Finite small row energies alone do not establish the rate $p_N=o(N^{-1})$.
They also do not constrain another observable's limiting sector. For
example, attention or predictive response can remain active while the
row field approaches a point mass. Dividing by a simultaneously vanishing
one-head reference must not replace the absolute bound.

An exactly soluble transient law demonstrates the importance of the
time coordinate. Let all heads share
$q_N(t)=q_*\mathbf1_{\{\tau>t\}}$, where $0<q_*\le1$ and the collapse
time $\tau$ is exponential with rate $a>0$, independent of $N$. Then
\begin{equation}
 \chi_N(t)=Nq_*^2e^{-at}(1-e^{-at}),\qquad
 \chi_N(c\log N)\sim q_*^2N^{1-ac}\quad(c>0).
 \label{model:eq:collapse-time-example}
\end{equation}
Indeed, the indicator is Bernoulli with success probability $e^{-at}$;
its mean and variance give both identities. For each fixed width the
limiting law is the point mass at zero. Its maximal finite-time
susceptibility is nevertheless $Nq_*^2/4$, and different joint time--width
paths produce different apparent powers. This example is not a fitted
PLDR collapse-time law. It supplies an explicit alternative mechanism
that a persistent critical interpretation must distinguish using the
measured temporal and distributional information.

The same example also separates weak concentration from the limiting
tilted functional. Along $t_N=c\log N$, write $p_N=N^{-ac}$. Then
\[
 \frac1N\log\bigl(1-p_N+p_N e^{Njq_*}\bigr)
 \longrightarrow\max\{0,jq_*\}.
\]
For $j>0$, the logarithm divided by $N$ equals
$jq_*+N^{-1}\log p_N+o(1)$; for $j<0$ the term $1-p_N$ dominates,
and at zero the expression vanishes exactly. This proves the limit.
It has a cusp even when $ac>1$ makes the zero-source curvature tend
to zero and the un-tilted law concentrate at zero. Subextensive rare-state
weights explain that distinction. Controlling the limiting tilted law
and the order of limits therefore requires more than weak concentration
or a finite susceptibility curve.

\section{Physical row energy and downstream transfer}
\label{model:sec:row-correspondence}
Fix a context, decoder and head along a training trajectory, with
fixed head dimension $d$ and $P_d=I_d-d^{-1}\mathbf1\mathbf1^{\mathsf T}$.
The recorded experiments use $d=64$. The physical
row quotient and its unnormalized energy are
\[
 Z_t=P_dA_t,\qquad E_t=\|Z_t\|_{\mathrm F}^2.
\]
The field in Equation~\eqref{model:eq:conditional-head-fields} instead uses
\begin{equation}
 \mathcal N_t=\max\{\|A_t\|_{\mathrm F}^2,d^2\epsilon_R\},
 \qquad R_t=E_t/\mathcal N_t.
 \label{model:eq:row-normalization-correspondence}
\end{equation}
Both are observations of the same metric, but their numerical scales
and their fluctuations differ. In particular, $E_t$ is not the
conditional susceptibility over independently initialized models.

For the increment $D_t=Z_{t+1}-Z_t$, direct expansion gives
\[
 E_{t+1}=E_t+2\langle Z_t,D_t\rangle+\|D_t\|_{\mathrm F}^2.
\]
The associated canonical affine coordinates are
$(q_t^E,\zeta_t^E)=(E_{t+1}/E_t,0)$ when $E_t>0$ and
$(q_t^E,\zeta_t^E)=(0,E_{t+1})$ when $E_t=0$.
These coordinates satisfy $E_{t+1}=q_t^EE_t+\zeta_t^E$ by direct
substitution in each of the cases $E_t>0$ and $E_t=0$.
For two chronological steps the affine composition is
$(q_2,\zeta_2)\circ(q_1,\zeta_1)=(q_2q_1,q_2\zeta_1+\zeta_2)$;
substitution proves associativity and iteration gives the block law.
Normalization changes the one-step coefficients to
\begin{equation}
 q_t^R=q_t^E\frac{\mathcal N_t}{\mathcal N_{t+1}},\qquad
 \zeta_t^R=\frac{\zeta_t^E}{\mathcal N_{t+1}},\qquad
 R_{t+1}=q_t^RR_t+\zeta_t^R.
 \label{model:eq:normalized-row-cocycle}
\end{equation}
Indeed, divide the affine energy identity by $\mathcal N_{t+1}>0$
and substitute $E_t=\mathcal N_tR_t$. If chronological blocking on
$[k,n)$ gives $E_n=Q^E(n,k)E_k+Z^E(n,k)$, the same calculation gives
\[
 Q^R(n,k)=Q^E(n,k)\frac{\mathcal N_k}{\mathcal N_n},\qquad
 Z^R(n,k)=\frac{Z^E(n,k)}{\mathcal N_n}.
\]
Thus normalization is compatible with blocking when its endpoint
factors and transported sources are retained. Neither coordinate
system is autonomous merely because this pathwise identity is exact.
An observer floor or a floating-point zero also cannot be substituted
for the analytical condition $E_t=0$.

Under Proposition~\ref{model:prop:generator-envelope}, let
$\overline K=\max\{1,\overline c/\lambda_d\}$ and
\[
 B_R=\max\{d^2(\sqrt d+1)^2\overline K^2,d^2\epsilon_R\}.
\]
The final affine LayerNorm gives $\mathcal N_t\le B_R$ uniformly
in head count and time. Consequently
\[
 d^2\epsilon_R R_t\le E_t\le B_RR_t.
\]
Thus raw and normalized row collapse are equivalent along sequences
within this bounded native family. Their covariance laws still need
not be proportional, because $\mathcal N_t$ is a learned random
quantity. The common row itself remains a separate retained
coordinate of the physical quotient construction.

The physical quotient has a second boundary at the learned PLGA map.
Write $\Psi(A)$ for the fixed-parameter transformation from $A$ to
$\G$ in Equation~\eqref{model:eq:plga}, and decompose
$A=\overline A+Z$, where
$\overline A=\mathbf1\mu^{\mathsf T}$ and $Z=P_dA$.
Define the constant-input defect
\begin{equation}
 \delta_\Psi(\mu)=P_d\Psi(\mathbf1\mu^{\mathsf T}).
 \label{model:eq:plga-row-defect}
\end{equation}
Integrating the derivative on the segment gives the exact
response-plus-defect identity:
\begin{align}
 P_d\Psi(A)
 &=\delta_\Psi(\mu)+
   P_d\int_0^1D\Psi(\overline A+sZ)[Z]\,ds,\notag\\
 \bigl|\|P_d\Psi(A)\|_{\mathrm F}
              -\|\delta_\Psi(\mu)\|_{\mathrm F}\bigr|
 &\le L\|Z\|_{\mathrm F},
 \label{model:eq:plga-row-transfer}
\end{align}
where $L$ bounds the derivative operator norm on that segment.
The positive power-base offset makes $\Psi$ smooth, so such a finite
$L$ exists on every fixed finite segment. The bound uses
$\|P_d\|_{\rm op}=1$ and the reverse triangle inequality.
At an exactly row-constant input, the output quotient is exactly
$\delta_\Psi(\mu)$. It vanishes only when the learned transform
preserves that input. For a sequence with uniformly bounded $L$,
vanishing input quotient and vanishing defect imply vanishing output
quotient; a defect bounded away from zero prevents that conclusion.
The empirical tests here retain the complete transform and all later
decoders when measuring predictive fidelity.
Sections~\ref{model:sec:row-projection-results} and \ref{model:sec:inference-results} specify those complete-emission interventions.

\section{The surviving common-row collective sector}
\label{model:sec:metric-collective-theory}
Row collapse removes differences between matrix rows and leaves their
common row unrestricted. To retain its
endogenous fluctuations, fix a context and the exogenous environment
of the head family, and set
\[
 \overline A_{\ell,N}=\frac1N\sum_{a=1}^N A_{\ell a},\qquad
 \overline\mu_{\ell,N}=\frac1d\overline A_{\ell,N}^{\mathsf T}\mathbf1.
\]
Define $Q_A$ by vectorizing all $\overline A_{\ell,N}$ and dividing
by $\sqrt{Ld^2}$, define $Q_\perp$ similarly using
$P_d\overline A_{\ell,N}$, and define $Q_\mu$ by vectorizing all
$\overline\mu_{\ell,N}$ and dividing by $\sqrt{Ld}$.
These are fixed mean-square units across the head-count family.
Common refers to the rows within a head matrix; equality between
different heads is a separate property.
The linear map $I_\mu$ that broadcasts the row vector into the full
matrix coordinates is an isometry under these normalizations.
One has $Q_A=I_\mu Q_\mu+Q_\perp$ with orthogonal components.

\begin{proposition}[Orthogonal metric collective decomposition]
\label{model:prop:metric-collective-sectors}
Assume the fields are square integrable under the specified conditional
initialization law, and write
$\Sigma_A=N\Cov(Q_A)$, $\Sigma_\mu=N\Cov(Q_\mu)$,
$\Sigma_\perp=N\Cov(Q_\perp)$ and $\chi_i=\tr\Sigma_i$.
Then
\begin{align}
 \chi_A&=\chi_\mu+\chi_\perp,\label{model:eq:metric-collective-trace}\\
 \|\Sigma_A-I_\mu\Sigma_\mu I_\mu^{\mathsf T}\|_{\rm op}
 &\le2\sqrt{\chi_\mu\chi_\perp}+\chi_\perp.
 \label{model:eq:metric-collective-transfer}
\end{align}
The identities and bound also hold after averaging the conditional
covariances over a fixed context law. If $\chi_\mu>0$ and
$\chi_\perp/\chi_\mu\to0$, the covariance of the complete head-averaged metric,
normalized by $\chi_\mu$, differs in operator norm by a vanishing
amount from the embedded common-row covariance.
\end{proposition}
\begin{proof}
Center each field over the same initialization law. Orthogonal
projections commute with this centering, so the common and contrast
vectors remain orthogonal in every realization. Expanding their
squared norm and taking expectation proves
Equation~\eqref{model:eq:metric-collective-trace}.
The covariance expansion contains the embedded common covariance,
the contrast covariance, and two transposed cross terms.
For unit vectors $u,v$, Cauchy--Schwarz bounds each cross term by
$\sqrt{\chi_\mu\chi_\perp}$. A positive semidefinite covariance has
operator norm at most its trace. The triangle inequality therefore
gives Equation~\eqref{model:eq:metric-collective-transfer}.
Context averaging preserves the trace identity; Cauchy--Schwarz
for that average gives the same bound with averaged traces.
Dividing by $\chi_\mu$ proves the final assertion.
\end{proof}

For empirical covariances with the same unbiased seed denominator,
all covariance terms acquire the same factor relative to the
uniform empirical law. The trace identity and the homogeneous norm
bound therefore hold with that denominator as well.

Orthogonality gives a trace decomposition, not statistical independence:
the common and contrast sectors can have nonzero cross covariance.
Their linear projections commute with head averaging and enter the
compatible observation-map composition of Section~\ref{model:sec:maps}.
Two arbitrary linear observations need not commute with one another. The nonlinear row-energy
ratio $R$ is a different field. A small susceptibility of $R$ therefore
does not show that the complete metric has small collective covariance.
When the transfer condition holds, the common row supplies effective
variables for the head-averaged metric covariance sector. Establishing
singularity or a universality class for that sector still requires the
specified width and training-law limits.

The common-row vector is representation dependent. In each decoder,
simultaneously replacing the terminal LayerNorm affine parameters
$(\gamma,\beta)$ by $(-\gamma,-\beta)$ and every head's input power
matrix $W_a$ by $-W_a$ sends $A_a$ and $\mu_a$ to their negatives
while preserving $W_aA_a$. All subsequent PLGA tensors and predictions
are unchanged. This substitution establishes the layer-sign symmetry directly.
Co-transforming first Adam moments by
the same signs and keeping second moments fixed also preserves
the corresponding mathematical update: loss gradients change by
the orthogonal sign action, the global clipping norm is invariant,
and the moment and decay recursions commute with that action.
The initial conditioning convention fixes a parameter representative.
The common-row covariance reported in that frame is nevertheless
not itself a gauge-invariant predictive susceptibility. Its role is
to retain a nonvanishing endogenous coordinate for the downstream
map; a physical critical interpretation additionally needs an
invariant emission or a justified gauge quotient.

\section{Metric-row covariance transport with nonlinear sources}
\label{model:sec:metric-transport}
Each native residual metric unit applies the same smooth map
$F_j:\R^d\to\R^d$ to every matrix row, with fixed head dimension $d$
($d=64$ in the recorded experiments). The index $j$
counts the eight internal units of a fixed decoder at a fixed training
state. It is distinct from the thermodynamic head count $N$ and the
training counter. Write the rows as $x_i^{\mathsf T}$, and define
\[
 \mu=\frac1d\sum_i x_i,\qquad \delta_i=x_i-\mu,\qquad
 C^{\rm row}=\frac1d\sum_i\delta_i\delta_i^{\mathsf T}.
\]
The native row energy can then be written exactly as
\begin{equation}
 R=\frac{\tr C^{\rm row}}
 {\max\{\tr C^{\rm row}+\|\mu\|^2,d\epsilon_R\}},
 \qquad\epsilon_R=10^{-30}.
 \label{model:eq:row-energy-moments}
\end{equation}
Indeed, the squared Frobenius norm decomposes into its centered part
and $d\|\mu\|^2$; division by the number of matrix entries gives the
implemented mean-square normalization. This internal row covariance
is different from the five-component covariance over initialization
identities used for head susceptibility.

\begin{proposition}[Row moment map with retained curvature defects]
\label{model:prop:metric-row-transport}
Let $F$ be twice continuously differentiable on a neighborhood of the
segments from $\mu$ to the rows. Set $J=DF(\mu)$ and
\[
 r_i=F(\mu+\delta_i)-F(\mu)-J\delta_i,\quad
 b=\frac1d\sum_i r_i,\quad e_i=r_i-b.
\]
Writing $\mu',C^{\rm row\prime}$ for the output row moments, one has
\begin{alignat}{2}
 \mu'&=F(\mu)+b, &\quad \delta_i'&=J\delta_i+e_i,\notag\\
 C^{\rm row\prime}&=JC^{\rm row}J^{\mathsf T}
                          +B+B^{\mathsf T}+C_e,\label{model:eq:metric-row-covariance}
\end{alignat}
where $B=d^{-1}\sum_i(J\delta_i)e_i^{\mathsf T}$ and
$C_e=d^{-1}\sum_i e_ie_i^{\mathsf T}$.
If the bilinear Hessian norm of $F$ is at most $M$ on those segments,
then
\begin{align}
 \tr C_e&\le\frac{M^2}{4d}\sum_i\|\delta_i\|^4,\notag\\
 \|C^{\rm row\prime}-JC^{\rm row}J^{\mathsf T}\|_{
 \rm op}
 &\le 2\|J\|_{\rm op}\sqrt{\tr C^{\rm row}\tr C_e}+\tr C_e.
 \label{model:eq:metric-row-defect}
\end{align}
The centered maps compose by ordered Jacobian products with their
transported residual sources retained.
\end{proposition}
\begin{proof}
The mean of $\delta_i$ is zero. Averaging the definition of $r_i$
therefore proves the mean identity and, on subtracting it, the
centered identity. Expanding the outer product of $J\delta_i+e_i$
and averaging gives Equation~\eqref{model:eq:metric-row-covariance}, including
both signed cross terms. Taylor's integral remainder bounds
$\|r_i\|\le M\|\delta_i\|^2/2$. Centering is an orthogonal projection
on the array of row residuals, so
$\sum_i\|e_i\|^2\le\sum_i\|r_i\|^2$, giving the first bound.
For any unit vectors $u,v$, Cauchy--Schwarz gives
\[
 |u^{\mathsf T}Bv|
 \le\left(\frac1d\sum_i\|J\delta_i\|^2\right)^{1/2}
      \left(\frac1d\sum_i\|e_i\|^2\right)^{1/2}
 \le\|J\|_{\rm op}\sqrt{\tr C^{\rm row}\tr C_e}.
\]
Use this bound for $B$ and $B^{\mathsf T}$, and
$\|C_e\|_{\rm op}\le\tr C_e$, to obtain the second bound.
Finally, substitution at two units gives
$\delta_i^{(2)}=J_1J_0\delta_i^{(0)}+J_1e_i^{(0)}+e_i^{(1)}$.
Induction yields, for any finite block,
\[
 \delta_i^{(b)}=J_{b-1}\cdots J_0\delta_i^{(0)}
       +\sum_{a=0}^{b-1}J_{b-1}\cdots J_{a+1}e_i^{(a)}.
\]
Empty products are identities. Splitting this product and sum at an intermediate unit proves the
composition statement. The means and residuals belong to the same
actual path; no closure in a scalar norm has been assumed.
\end{proof}

\begin{corollary}[A local row-concentrating sector]
\label{model:cor:metric-row-contraction}
If $F$ is $L$-Lipschitz on a convex region containing the rows, then
$\tr C^{\rm row\prime}\le L^2\tr C^{\rm row}$.
If $\|DF(\mu)\|_{\rm op}<1$, continuity of the derivative supplies
an input neighborhood with such an $L<1$.
Identical rows remain identical under every metric unit.
\end{corollary}
\begin{proof}
Expanding pair differences gives
$\tr C^{\rm row}=(2d^2)^{-1}\sum_{i,k}\|x_i-x_k\|^2$.
Apply the Lipschitz inequality to each pair and use the same identity
for the outputs. If the derivative norm at $\mu$ is below one, choose
a larger number still below one. Continuity bounds the derivative by
that number on a sufficiently small ball, and integration along line
segments gives the Lipschitz property there. Equal inputs have equal
outputs, proving the last assertion.
\end{proof}

The measured Jacobian norm is a local diagnostic. Its value at a centroid
does not bound a widely dispersed row cloud. The signed residual terms
in Equation~\eqref{model:eq:metric-row-covariance} quantify that distinction;
their omission is justified only within a validated local reduction.
Even its linear part depends on the orientation of $C^{\rm row}$:
$\tr(JC^{\rm row}J^{\mathsf T})/\tr C^{\rm row}$ can be much smaller
than $\|J\|_{\rm op}^2$.
Native arithmetic adds a separately measured row residual to the
smooth-map identity. It is transported by the same finite product
when composing recorded native steps. The inference effect of the
resulting metric still travels through the learned power layer and
all later decoders. Neither the finite unit-depth contraction nor a
centroid Jacobian crossing one establishes a critical head-count
limit or a stationary training eigenmode.
Section~\ref{model:sec:row-transport-results} reports the centroid Jacobians, row residuals and retained nonlinear transport.

\begin{proposition}[Loss adjoints with retained exceptional rows]
\label{model:prop:metric-row-adjoint}
Let a differentiable scalar loss depend on a rowwise unit
$y_i=F_\vartheta(x_i)$ through its outputs, with all later graph
operations retained. The index $i$ can include the batch and head
indices as well as the matrix row. Write
$\lambda_i'=\nabla_{y_i}\mathcal L$ and
$J_i=D_xF_\vartheta(x_i)$. Then
\begin{equation}
 \lambda_i=J_i^{\mathsf T}\lambda_i',\qquad
 \nabla_\vartheta\mathcal L
       =\sum_i(D_\vartheta F_\vartheta(x_i))^{\mathsf T}\lambda_i'.
 \label{model:eq:metric-row-adjoint}
\end{equation}
If $\|J_i\|_{\rm op}\le\rho$ on a retained core $C$ of rows, then
\begin{equation}
 \sum_i\|\lambda_i\|^2
 \le\rho^2\sum_{i\in C}\|\lambda_i'\|^2
          +\sum_{i\notin C}\|J_i^{\mathsf T}\lambda_i'\|^2.
 \label{model:eq:adjoint-core-tail}
\end{equation}
Across successive units the adjoints compose by the transposed ordered
Jacobian product. Uniform bounds $\rho_j$ on all their actual row
Jacobians consequently bound the backward norm gain by $\prod_j\rho_j$.
\end{proposition}
\begin{proof}
The chain rule gives the input gradient of each independently applied
row map. The parameter is shared across rows, so its chain-rule
contributions add, proving Equation~\eqref{model:eq:metric-row-adjoint}.
Apply the operator-norm inequality on $C$ and keep the complementary
terms without approximation to obtain
Equation~\eqref{model:eq:adjoint-core-tail}. Applying the same chain rule at
each preceding unit yields the transposed product. Submultiplicativity
of its norm gives the final bound.
\end{proof}

This is the adjoint of the forward row transport, with the source set
by the complete loss graph. It provides a mechanism for attenuating
upstream gradients when the actual source-bearing rows are contracting.
A centroid derivative does not control the exceptional-row term, and
forward row concentration alone does not determine the loss adjoint.
The unit's own parameter gradient additionally contains its parameter
Jacobian. Relating these quantities to a decay-dominated Adam sector
requires the clipping and moment-history conditions of
Proposition~\ref{model:prop:adam-decay-sector}.

\begin{corollary}[Shared-gradient closure requires a transported source]
\label{model:cor:shared-gradient-closure}
For a shared metric unit, define the stacked parameter Jacobian $B$ by
$(Bv)_i=D_\vartheta F_\vartheta(x_i)v$, and let $\lambda'$ be its
stacked output loss adjoint. Its full shared-coordinate gradient is
$\Gamma=B^{\mathsf T}\lambda'$. For any coupled approximations
$\widetilde B,\widetilde\lambda'$ in the same spaces,
\begin{equation}
 \|\Gamma-\widetilde B^{\mathsf T}\widetilde\lambda'\|
 \le\|B\|_{\rm op}\|\lambda'-\widetilde\lambda'\|
       +\|B-\widetilde B\|_{\rm op}\|\widetilde\lambda'\|.
 \label{model:eq:shared-gradient-closure}
\end{equation}
In particular, $\|\Gamma\|\le\|B\|_{\rm op}\|\lambda'\|$.
\end{corollary}
\begin{proof}
Stacking Equation~\eqref{model:eq:metric-row-adjoint} gives the formula
for $\Gamma$. Subtract the approximate gradient and write the difference
as $B^{\mathsf T}(\lambda'-\widetilde\lambda')+
(B-\widetilde B)^{\mathsf T}\widetilde\lambda'$.
The triangle and operator-norm inequalities prove the bound.
Applying the operator-norm inequality to the exact formula gives the
last assertion.
\end{proof}

The shared parameter dimension is fixed along the declared head-count
family, but its gradient aggregates all heads and rows with the full
loss source. Substituting $c\Gamma$ into the shared first- and
second-moment recursions supplies its exact native training flow.
The common clipping factor still depends on the complete model gradient.
If the exact and reduced global factors are $c,\widetilde c\in[0,1]$,
then adding and subtracting $\widetilde c\Gamma$ gives
\begin{equation}
 \|\widetilde c\widetilde\Gamma-c\Gamma\|
 \le\widetilde c\|\widetilde\Gamma-\Gamma\|
       +|\widetilde c-c|\,\|\Gamma\|.
 \label{model:eq:clipped-shared-gradient-error}
\end{equation}
Thus controlling the shared force before clipping and controlling its
full-model clipping factor are separate requirements.
A reduction of that flow must control the parameter Jacobian and the
transported loss source as well as the forward row variables.
A small predictive KL or a small final-unit input-adjoint gain supplies
neither error term in Equation~\eqref{model:eq:shared-gradient-closure} by
itself. These terms state a sufficient quantitative closure condition;
their uniform vanishing in a native trained width limit is not assumed.

\section{Compatible row reductions and fluctuation fidelity}
\label{model:sec:row-projection}
The curvature sources in Equation~\eqref{model:eq:metric-row-covariance}
need not be spread uniformly over rows. A reduction can therefore retain
exceptional rows together with the centroid of their complement.
For an $m\times d$ row matrix $X$ and a retained index set
$I\subseteq\{1,\ldots,m\}$, let $P_I X$ keep the rows indexed by $I$
and replace all other rows by their own centroid. Set $P_I X=X$ if
the complement is empty. The retained state contains the indices, the
retained row vectors, the complement centroid and its multiplicity.
It can be represented by at most $|I|+1$ row vectors.

\begin{proposition}[Nested row maps and their covariance deficit]
\label{model:prop:exception-projection}
For fixed index sets $I\subseteq J$, the maps satisfy
\[
 P_I^2=P_I,\qquad P_I P_J=P_J P_I=P_I.
\]
They preserve the global row centroid. With
$E=X-P_I X$ and row covariances normalized by $m$,
\begin{equation}
 C_X^{\rm row}=C_{P_I X}^{\rm row}+\frac1m E^{\mathsf T}E.
 \label{model:eq:exception-covariance}
\end{equation}
In particular the covariance deficit is positive semidefinite, and its
trace is the mean squared residual of the discarded row deviations.
For the native square metric, row energy cannot increase under this
single projection at a fixed input.
\end{proposition}
\begin{proof}
Let $V_I$ be the linear subspace of matrices whose rows outside $I$
are equal. The Frobenius least-squares projection onto $V_I$ leaves
the free rows unchanged and sets the remaining rows to their centroid,
so it is exactly $P_I$. If $I\subseteq J$, then $V_I\subseteq V_J$.
Orthogonal projections onto nested subspaces have the stated
composition: projection onto $V_I$ removes both orthogonal complements,
whereas projection onto $V_J$ fixes every element of $V_I$.
The sum of the rows is preserved by replacing any group by its
centroid. Moreover, $E$ vanishes on $I$ and its rows sum to zero on
the complement. The centered rows of $P_I X$ are constant on that
complement. Both matrix cross products of these centered rows with
$E$ are therefore zero. Expanding the centered covariance proves
Equation~\eqref{model:eq:exception-covariance}. The deficit has the form
$E^{\mathsf T}E/m$, which is positive semidefinite. The row-energy
formula~\eqref{model:eq:row-energy-moments} is nondecreasing in
$\tr C^{\rm row}$ at fixed centroid, proving the last claim.
\end{proof}

The discarded covariance also determines the row-energy change at that
fixed matrix. Put $V=\tr C_X^{\rm row}$,
$M=\|\mu\|^2$ and $D=m^{-1}\|X-P_I X\|_F^2$. If the denominator
floor is inactive before and after the projection, then
\begin{equation}
 R(X)-R(P_I X)=\frac{MD}{(V+M)(V-D+M)}.
 \label{model:eq:exception-row-energy-error}
\end{equation}
This follows by subtracting $V/(V+M)$ and
$(V-D)/(V-D+M)$. For $M>0$ the difference is at most $D/M$.
The formula links a retained-row variance budget to the error in the
chosen intensive field. Downstream changes in the metric input remain
a separate full-graph transport effect.

Selecting a nested list once from a fine matrix gives a compatible
family of these maps. The numerical implementation orders rows by
their squared distance from the global centroid, resolves ties by
index, and stores the retained indices. Composition uses that retained
information. Independently recomputing a ranking at every scale is a
different procedure and is not covered by the proposition. Nor does
the proposition bound the error after the rest of the decoder graph:
the later context and metric inputs can change. Those predictions are
measured through the complete native emission.
Section~\ref{model:sec:row-projection-results} gives the nested retained-row experiment and full-vocabulary errors.

At zero exceptions the reduced metric has identical rows. Rowwise
application of any metric unit preserves this property, so a centroid
inserted before the first unit can be propagated as one row and then
broadcast after the last. This is an exact identity for the
\emph{projected} smooth graph. Its discrepancy from the original
graph includes the centroid defects $b$ in
Proposition~\ref{model:prop:metric-row-transport}. The discrepancy between
the one-row and broadcast implementations is an additional native
arithmetic question.

This equal-row identity stops at the metric learner. The subsequent
power transform has head-specific matrix multiplications, biases and
entrywise powers. Its constant-input quotient defect in
Equation~\eqref{model:eq:plga-row-defect} is retained in every complete-graph
projection measurement. Thus a centroid metric reduction does not
impose a row-constant generated operator.

Predictive equivalence alone does not imply thermodynamic equivalence
for a chosen internal observable. In particular, the zero-exception
projection sets the internal row susceptibility to zero even if the
predictive law is well approximated. The following quantitative
condition states what fluctuation fidelity additionally requires.

\begin{proposition}[Susceptibility transfer under a coupled reduction]
\label{model:prop:susceptibility-transfer}
Let $Q_N,\widetilde Q_N$ be square-integrable scalar observables
coupled under the same conditional law, and put
\[
 \chi_N=N\Var Q_N,\qquad
 \widetilde\chi_N=N\Var\widetilde Q_N,\qquad
 \varepsilon_N^2=\E(\widetilde Q_N-Q_N)^2.
\]
Then
\begin{equation}
 |\widetilde\chi_N-\chi_N|
 \le 2\sqrt{N\chi_N}\,\varepsilon_N+N\varepsilon_N^2.
 \label{model:eq:susceptibility-transfer}
\end{equation}
Consequently, if $\chi_N>0$ and
$\varepsilon_N=o(\sqrt{\chi_N/N})$, the ratio
$\widetilde\chi_N/\chi_N$ tends to one. For vector observables with
$\varepsilon_N^2=\E\|\widetilde Q_N-Q_N\|^2$, their susceptibility
matrices satisfy
\[
 \|\widetilde\chi_N-\chi_N\|_{\rm op}
 \le 2\sqrt{N\tr\chi_N}\,\varepsilon_N+N\varepsilon_N^2.
\]
The scalar bound also holds for the trace averaged over components
and contexts, using the correspondingly averaged squared error.
\end{proposition}
\begin{proof}
Use Lemma~\ref{model:lem:master-perturbation} with $A=I$ and
$e=\widetilde Q_N-Q_N$. Centering contracts $L^2$, so its centered
error is at most $\varepsilon_N$. Multiply the scalar or matrix bound
by $N$ to obtain the two inequalities. In the scalar case, divide by
$\chi_N$ and apply the stated small-error condition. Averaging the
scalar inequalities over the declared components and contexts and
applying Cauchy--Schwarz to their mixed bounds proves the final assertion.
\end{proof}

This bound transfers an already specified fluctuation scaling; it
does not establish that scaling or the singularity of a limiting
thermodynamic functional. For an empirical ensemble of $s$ seeds
using unbiased covariances, the same argument replaces
$N\varepsilon_N^2$ by $Ns\widehat\varepsilon_N^2/(s-1)$.
The factor accounts for the denominator of the sample covariance.
Contexts remain paired observations under every reduction.

\section{Observation error and inference visibility}
\label{model:sec:observation-theory}

An arithmetic observation is a second map applied to the mathematical
model. Its errors propagate through the chosen collective law, with
units fixed before a susceptibility is compared. A small individual
row field and a small ensemble susceptibility are different statements.

\subsection{Energy enclosures and a sample susceptibility bound}
\begin{proposition}[Energy enclosure from a matrix error bound]
\label{model:prop:observation-energy}
Let $P$ be an orthogonal row projection on $d\times d$ real matrices.
Suppose $\|\widehat A-A\|_{\rm F}/d\le\delta$ and put
$\widehat e=\|P\widehat A\|_{\rm F}^2/d^2$ and
$e=\|PA\|_{\rm F}^2/d^2$. Then
\begin{equation}
 (\max\{0,\sqrt{\widehat e}-\delta\})^2
 \le e\le(\sqrt{\widehat e}+\delta)^2.
 \label{model:eq:observation-energy}
\end{equation}
Write $e_-,e_+$ for these endpoints, and $t_-,t_+$ for the
corresponding total-energy endpoints obtained with $P=I$.
For the normalized row field with positive floor $\epsilon_R$,
\begin{equation}
 \frac{e_-}{\max\{t_+,\epsilon_R\}}
 \le R(A)\le
 \min\left\{1,\frac{e_+}{\max\{t_-,\epsilon_R\}}\right\}.
 \label{model:eq:observation-ratio}
\end{equation}
\end{proposition}
\begin{proof}
Orthogonal projection contracts the Frobenius norm. The triangle and
reverse triangle inequalities bound $\|PA\|_{\rm F}/d$ between
$\max\{0,\sqrt{\widehat e}-\delta\}$ and
$\sqrt{\widehat e}+\delta$. Squaring these nonnegative bounds proves
the energy enclosure. The same reasoning applies with $P=I$.
The denominator floor is positive and monotone; divide the lower
numerator bound by the upper denominator bound and conversely.
Finally $\|PA\|_{\rm F}\le\|A\|_{\rm F}$ gives $R(A)\le1$.
\end{proof}

A certified matrix error gives a certified enclosure. A float32 versus
float64 difference supplies an empirical arithmetic sensitivity unless
the reference error is itself bounded. In particular, a threshold on
$R$ is neither a universal rounding floor nor a threshold on
$N\Var(Q_N)$. Equal stored rows do not certify the real-arithmetic
row-constant face.

\begin{proposition}[Susceptibility error of observed seed fields]
\label{model:prop:sample-observation}
Let $q_i,\widehat q_i$ be elements of a real Hilbert space, for
$i=1,\ldots,s$ with $s>1$. The norm includes the declared average
over contexts and decoders. For $N>0$ set
\[
 \chi=\frac{N}{s-1}\sum_i\|q_i-\bar q\|^2,\qquad
 \widehat\chi=\frac{N}{s-1}\sum_i
             \|\widehat q_i-\overline{\widehat q}\|^2,
 \qquad \Delta_i=\widehat q_i-q_i,
\]
and $\chi_\Delta=N(s-1)^{-1}\sum_i\|\Delta_i-\bar\Delta\|^2$.
Then
\begin{align}
 |\widehat\chi-\chi|
 &\le2\sqrt{\chi\chi_\Delta}+\chi_\Delta,
 \label{model:eq:sample-observation}\\
 \chi_\Delta&\le\frac{Ns}{s-1}\,
                   \frac1s\sum_i\|\Delta_i\|^2.
 \label{model:eq:sample-observation-mse}
\end{align}
\end{proposition}
\begin{proof}
Corollary~\ref{model:cor:master-empirical} applies in the declared observation
Hilbert space with $\Delta_i=\widehat q_i-q_i$. Its product-space norm
uses the same divisor $s-1$ and gives both displayed inequalities.
\end{proof}

The factor $s/(s-1)$ is required by the unbiased sample convention.
This deterministic inequality does not require independent seeds;
independence is needed for interpreting the sample variance as an
estimator of the initialization population. A conservative bound can
be much larger than the observed change when covariance cross terms
cancel. Both quantities are retained in the arithmetic results.

For a diagnostic threshold $u$, split each field into
$R_{<u}=R\mathbf1_{R<u}$ and $R_{\ge u}=R-R_{<u}$.
After head averaging, the unscreened susceptibility is exactly
\[
 \chi_R=\chi_{<u}+\chi_{\ge u}
             +2N\Cov(Q_{<u},Q_{\ge u}),
\]
with the same sample normalization when estimated. Small entries can
be numerous while a minority of larger entries carries most of this
second moment. Removing entries changes the estimand and is used
only as a sensitivity diagnostic.

\subsection{Which training fluctuations survive freezing}
Freezing weights changes the observation of a trained-state ensemble;
it does not continue its optimizer dynamics. The following sufficient
visibility criterion applies to one declared ensemble throughout.

\begin{proposition}[Visibility of a collective fluctuation in inference]
\label{model:prop:inference-visibility}
At fixed conditioning, let $r_N$ be a scalar training collective with
$\chi_r=N\Var(r_N)>0$. Suppose an inference observable satisfies
$Y_N=a_Nr_N+e_N$, where $a_N$ is deterministic under that
conditioning. With $\chi_e=N\Var(e_N)$,
\begin{equation}
 \left|\frac{N\Var(Y_N)}{\chi_r}-a_N^2\right|
 \le2|a_N|\sqrt{\chi_e/\chi_r}+\chi_e/\chi_r.
 \label{model:eq:inference-visibility}
\end{equation}
If $a_N\to a\ne0$ and $\chi_e/\chi_r\to0$, then
$N\Var(Y_N)/\chi_r\to a^2$.
\end{proposition}
\begin{proof}
Apply the scalar covariance identity of
Lemma~\ref{model:lem:master-perturbation} with $A=a_N$ and $X=r_N$.
After multiplication by $N$, its two mixed terms have absolute sum
at most $2|a_N|\sqrt{\chi_r\chi_e}$. Divide by $\chi_r$ to obtain
the bound and take the stated limits. The same calculation applies
to paired sample covariances with a common divisor.
\end{proof}

This transfers a critical scaling property only if such a property
has first been established for $r_N$. Vanishing predictive overlap,
a material residual, or changing the ensemble prevents that inference.
The layer-sign symmetry and the saturated score directions give
concrete reasons why internal covariance need not have a comparable
predictive response. Across-initialization variation at fixed context
and across-context variation at one released checkpoint remain distinct.

\FloatBarrier
\par\medskip\noindent
Chapter~\ref{ch:model-wide-evidence} measures the corresponding
model-wide fluctuations and size--time transport. The experiments test finite
response, shared-state influence and observation limits under the protocols
needed to interpret the theory.

\chapter{Measured model-wide fluctuations and size--time transport}
\label{ch:model-wide-evidence}
This chapter assembles the model-wide fluctuation and transport evidence.
Pretrained observations, controlled training, conditional head scans,
width--time continuations and native source interventions resolve finite
patterns while retaining replication units, arithmetic qualifications and
limits on scaling inference.

\section{Completed experiments on pretrained models}
\label{model:sec:methods}

\subsection{Checkpoints, dataset, and numerical execution}
We used the public checkpoints
\texttt{PLDR-LLM-v51-SOC-110M-1} and
\texttt{PLDR-LLM-v51-SOC-110M-2}, denoted checkpoints 1 and 2
\cite{model1,model2}. Both contain five decoders, fourteen heads per decoder,
head dimension 64, residual width 896, feed-forward width 2389, and a
32,000-token SentencePiece vocabulary. Their model cards report
approximately eight billion RefinedWeb pretraining tokens. Checkpoint 1
used 2,000 warm-up steps and maximum learning rate $1.2\times10^{-3}$;
checkpoint 2 used 1,000 steps and $10^{-3}$. These are two distinct released
models, not a controlled training intervention. Their immutable revisions
and file hashes are recorded in Appendix~\ref{model:app:reproduction}.

The local RefinedWeb Arrow dataset was read without modification. Its
5,518 original shards were divided into sixteen contiguous index strata;
one shard was selected uniformly within each stratum using seed 20260905.
Rows of each selected shard were randomly permuted. We retained 288
distinct documents per shard with at least 513 tokens and drew a uniform
513-token crop from each. Duplicate content hashes were rejected. The
4,608 resulting documents were shuffled, and the first 512 were reserved
for observation calibration. The other 4,096 supplied evaluation
observations. Document content hashes, original shard and row, crop
offset, tokenizer hash, and complete token arrays are retained. No BOS,
EOS, or padding tokens were added to these fixed-length crops.

This is a distribution of sufficiently long RefinedWeb documents with
equal selected-shard contributions, rather than token-weighted sampling of
the complete corpus. Calibration and evaluation document sets are
disjoint, but overlap with the historical pretraining stream is unknown.
The measured NLL is therefore an in-source-distribution diagnostic, not a
held-out-pretraining generalization benchmark.

The calculations used the author's pinned model modules and released
safetensors. A narrow compatibility adapter supplied the explicit
triangular additive mask for uncached, unpadded inference under the
installed Transformers API. It changed no PLGA, RoPE, normalization,
feed-forward, or vocabulary-projection formula. Direct decoder-loop
evaluation matched the public wrapper's final logits exactly in the
native validation. The retained native neural operations used float32, with TF32 disabled;
statistical accumulation, KL evaluation, and Fisher products used float64.
The released weights were held fixed for these measurements. One checkpoint ran on each of two NVIDIA
GeForce RTX 4090 GPUs with 24 GB memory. Complete commands, package
versions, peak memory, source hashes, and measured run times accompany the
artifacts. The main joint-feature measurements took approximately eight seconds per
checkpoint after loading. Eight-segment collection took 47--48 seconds.
The 256-context response runs took 204 and 212 seconds including array
serialization, with peak allocated GPU memory below 1 GB. These recorded
times describe this implementation and hardware, rather than a general
performance benchmark. All measurements reported here have completed.

\subsection{Joint observations and independent-document flow}
\label{model:sec:joint-observation-methods}
At context length 128 we collected a fixed 174-vector per document:
70 final-row attention entropies, 70 RMS final-row head outputs before
their concatenation and output projection, four fixed random projections
of each of the scaled embedding and five decoder states (24 total), eight
fixed centered-logit projections, predictive entropy, and the NLL of the
external next token. Every layer and head appears in this vector. We also
recorded deductive-tensor norms and row energies as precision-sensitive
diagnostics, which were not used to standardize the active fluctuation
field.

Each coordinate was centered and scaled using only the 512 calibration
documents. Thirty-two Gaussian random directions, drawn with seed 773,
were then fixed and normalized to unit calibration variance. They are
projections of the joint 174-vector and generally mix heads and layers.
All projection arrays are saved. They provide tractable joint-law
diagnostics, not proof of isotropy or completeness of the empirical field.

\begin{figure}[ht]
\centering\includegraphics[width=\textwidth]{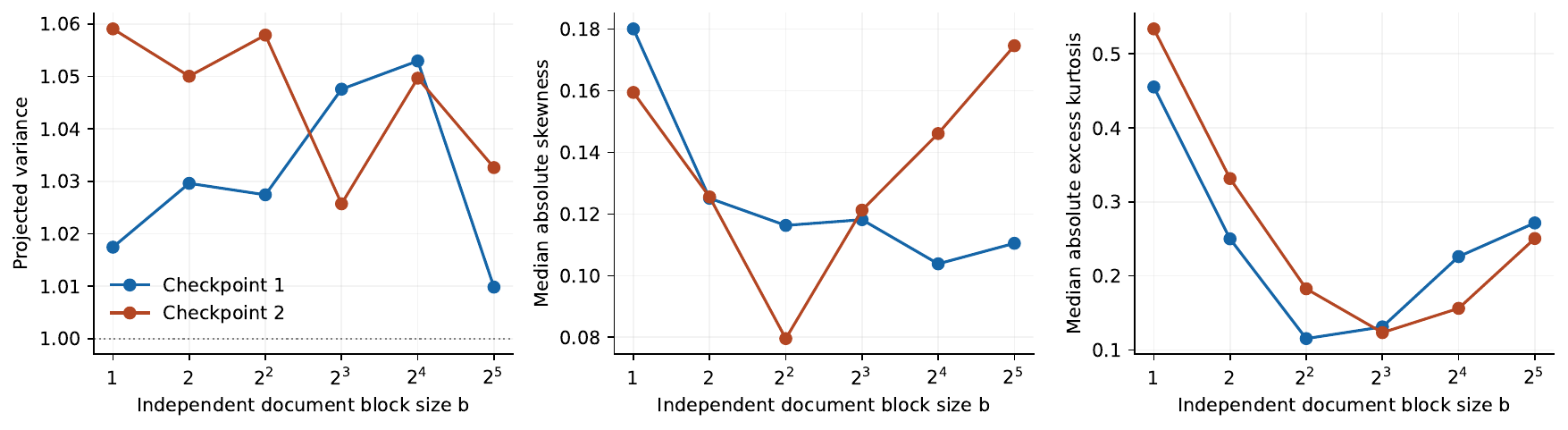}
\caption{Document-block fluctuation flow. Covariance stays approximately
constant under $b^{-1/2}$ normalization. Higher moments decrease on the
well populated early scales; coarse-scale moment estimates reach a
sampling floor as the number of blocks falls.}
\label{model:fig:flow}
\end{figure}

The population RG of Theorem~\ref{model:thm:clt} is defined for fresh independent
draws from the declared context law. The executed document blocks use a
random permutation without replacement of one finite stratified sample;
they therefore retain finite-population dependence. Their finite-scale
statistics approximate the independent-draw construction and do not prove
its sampling hypothesis. The 4,096 evaluation documents were partitioned into aligned blocks of
$b=1,2,4,8,16,32$. For each projection we formed the sum divided by
$\sqrt b$, then measured its central variance, skewness, and excess
kurtosis. Table~\ref{model:tab:flow} and Figure~\ref{model:fig:flow} give medians
across the fixed projections. The plotted reference variance is one in
calibration units; its small displacement on evaluation data includes
calibration sampling error.

\begin{table}[htbp]
\centering\small
\caption{Independent-document RG on 4,096 evaluation contexts. Columns
are medians over 32 fixed projections; the last two use absolute central
standardized moments. Projection coordinates are correlated and are not
counted as independent experimental replicates.}
\label{model:tab:flow}
\input{content/model/generated/flow-table.tex}

\end{table}

To summarize the second-moment flow, we fit
$\log\Var(b^{-1/2}\sum_iY_i)$ against $\log b$ separately in each fixed
projection and define the descriptive value $H=(1+\text{slope})/2$.
The median is $\ModelSourceHOne$ for checkpoint 1, with document-bootstrap interval
$[\ModelSourceHLoOne,\ModelSourceHHiOne]$, and $\ModelSourceHTwo$ for checkpoint 2, with interval
$[\ModelSourceHLoTwo,\ModelSourceHHiTwo]$. These are consistent with the $H=1/2$ normalization derived in
Theorem~\ref{model:thm:clt}. The class assignment follows from the specified
independent-context law and boundedness; the finite observations measure
how its flow appears on these models and this sample.

At $b=32$ there are only 128 blocks. Gaussian reference standard errors
for individual skewness and excess-kurtosis estimates are approximately
$\sqrt{6/128}=0.217$ and $\sqrt{24/128}=0.433$. Consequently the remaining
absolute moments at these scales do not determine separate correction
exponents. No full 174-dimensional Gaussian goodness-of-fit conclusion is
drawn from thirty-two projections. Mean next-token NLL on evaluation
contexts is $\ModelSourceNativeNLLOne$ and $\ModelSourceNativeNLLTwo$ nats, respectively; this
is a descriptive outcome rather than a reasoning benchmark.

\subsection{Whole-model response and the importance of cross terms}
\label{model:sec:whole-model-response-results}
The response experiment used a separate 256-document crop subset at
length 128. The first 128 contexts supplied ensemble diagnostics and the
last 128 supplied the reported perturbation-validation statistics.
For \emph{each} context we measured the 70 source columns of the logit
Jacobian using symmetric differences at step 0.01 in
Equation~\eqref{model:eq:measured-source}. We then formed the complete
$70\times70$ matrix $\Hess_x$ using the baseline vocabulary probabilities.
The test directions were eight fixed Gaussian vectors with RMS component
one, and the test sources were $\eta=\epsilon v$ at
$\epsilon=0.003,0.01,0.03,0.1$. These retained observations are descriptive evidence; context reuse and
the two response sample sizes are recorded in the complete run inventory.

This tests prediction of simultaneous perturbations from the source
basis derivatives at the same context. It does not predict a context's response matrix from
another context. Such a replacement is not part of
Theorem~\ref{model:thm:kl}. The full field $x\mapsto\Hess_x$ remains in the
dataset-conditioned state.

Table~\ref{model:tab:response} compares actual output KL with
$\epsilon^2v^{\mathsf T}\Hess_xv/2$ on the 128 validation contexts.
Uncertainty intervals resample complete contexts, retaining all eight
directions together. At RMS head perturbation 0.003, the median relative
errors are $\ModelSourceResponseOne\%$ and $\ModelSourceResponseTwo\%$; maxima are about
6.90\% and 4.81\%, respectively. Table~\ref{model:tab:response} includes q90
and maximum errors at every amplitude. They grow approximately
linearly with perturbation amplitude across the measured range, consistent with a local expansion whose cubic and numerical
terms vary across contexts (Proposition~\ref{model:prop:response-window}). The small-perturbation result is thus a direct test of the
full-model derivative construction and the predictive source metric.

\begin{table}[htbp]
\centering\small
\caption{Finite whole-model perturbations. Error is
$|\KL_{\mathrm{observed}}/\KL_{\mathrm{quadratic}}-1|$, summarized over
128 contexts and eight simultaneous directions. The diagonal comparison
retains the same measured diagonal coefficients and removes all
cross-source terms.}
\label{model:tab:response}
\input{content/model/generated/response-table.tex}

\end{table}

The off-diagonal Frobenius fractions
$\norm{\Hess-\diag\Hess}_F/\norm{\Hess}_F$ of the first-half mean
matrices are $\ModelSourceOffDiagonalOne$ and $\ModelSourceOffDiagonalTwo$. Removing these
cross terms increases the smallest-amplitude median errors to
$\ModelSourceDiagonalOne\%$ and $\ModelSourceDiagonalTwo\%$. This supplies a concrete reason to
retain a model-wide response matrix: separate scalar head contributions
do not reproduce the response to simultaneous perturbations in the fixed
head-gain coordinates.

Independent automatic-differentiation directional checks on four
contexts and two directions per context agreed with the finite-difference
Jacobian to maximum relative predictive-Fisher discrepancies
$\ModelSourceADMaxOne$ and $\ModelSourceADMaxTwo$. A second central-difference check at step
0.005 was retained for every context and direction, with its median, q90
and maximum reported in the numerical evidence table in Appendix~\ref{model:app:reproduction}. These checks
measure the numerical derivative approximation; they do not certify a
uniform remainder over all contexts or arbitrary perturbation sizes.

\subsection{Adjacent text segments and a correlated sector}
\label{model:sec:segment-results}
For every document we additionally evaluated eight adjacent, nonoverlapping
64-token contexts, each as a separate native model call with rotary
positions reset and its next token excluded. The next-token target of one
segment can be an input token of the following segment; the resulting
cross-segment dependence is deliberately retained. Each segment supplies
the same 174 observation types. A common projection map was calibrated
from the first 512 complete documents, and analysis used the remaining
4,096 documents. Position-specific means were removed before covariance
calculation, so deterministic changes with segment index do not inflate
the fluctuation statistic.

For $b=1,2,4,8$ we formed aligned sums within each document. For each
projection, its susceptibility was the average covariance of these
aligned block sums divided by $b$, relative to its average fine-position
variance. Table~\ref{model:tab:segments} reports median directional ratios.
Table~\ref{model:tab:lambda} additionally reports the defined maximum generalized
eigenvalue $\Lambda_b$ and an independently evaluated selected direction. A control independently permuted document
labels at each segment position, preserving marginal laws and within-call
head dependence while removing paired cross-segment dependence.

\begin{table}[htbp]
\centering\small
\caption{Finite text-segment susceptibility. The interval resamples
complete documents and preserves all eight segment fields within each
sample. The control uses a fixed independent permutation at each segment
position. No stationarity across segment positions is assumed.}
\label{model:tab:segments}
\input{content/model/generated/segment-table.tex}

\end{table}

At eight segments the enhancements are $\ModelSourceSegmentOne$
$[\ModelSourceSegmentLoOne,\ModelSourceSegmentHiOne]$ and $\ModelSourceSegmentTwo$
$[\ModelSourceSegmentLoTwo,\ModelSourceSegmentHiTwo]$. The control medians are
$\ModelSourcePermutedOne$ and $\ModelSourcePermutedTwo$. Direct construction of the
cross-position covariance matrix and its contraction according to
Equation~\eqref{model:eq:chi-general} reproduce the blocked covariance with
relative residual below $4\times10^{-16}$. This is a numerical algebra
check. The independent empirical content is the reproducible enhancement
and its removal by the permutation control.

\begin{figure}[ht]
\centering\includegraphics[width=.96\textwidth]{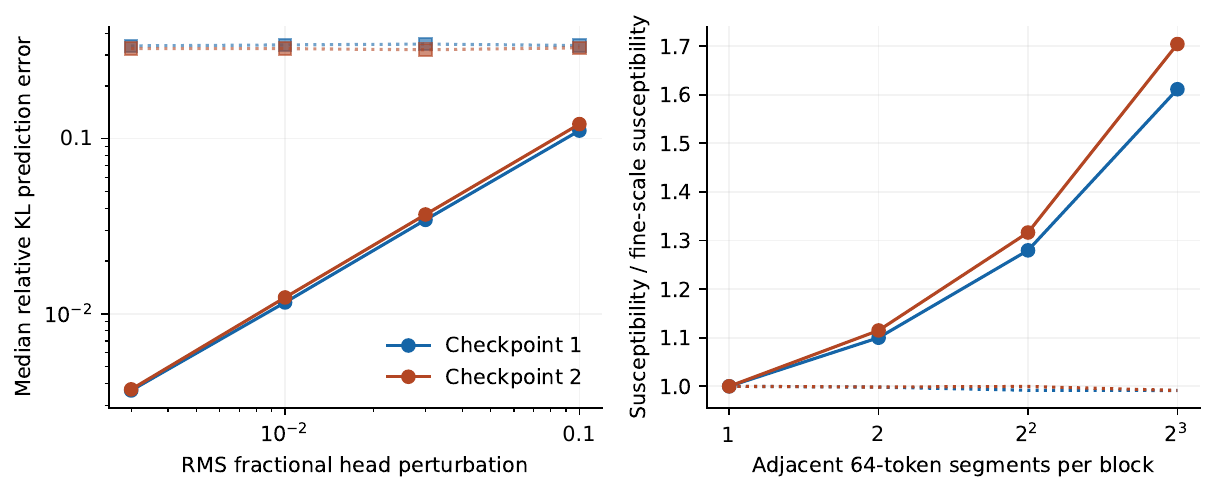}
\caption{Left: predictive response errors with the complete Fisher matrix
(solid circles) and its diagonal approximation (dotted squares). Right:
susceptibility from adjacent text segments (solid circles) and independently
permuted segment controls (dotted). Each color denotes one checkpoint.}
\label{model:fig:response}
\end{figure}

The finite correlated sector is therefore resolved by the joint-law map.
It is not represented by the independent-document product law. Eight
segments span 512 tokens and do not identify a limiting correlation
exponent, a diverging correlation length, or a critical universality class
of text. We accordingly apply Theorem~\ref{model:thm:chi}, whose finite
nonstationary hypotheses match the measurement, rather than
Proposition~\ref{model:prop:longrange}.

\begin{table}[htbp]\centering\small
\caption{Generalized susceptibility in the retained 32-dimensional projection.
No ridge is added. The selected direction is estimated on rows $[512,2560)$
and evaluated on rows $[2560,4608)$; it is a Rayleigh quotient on other
documents, not an unbiased estimate of a population supremum. The minimum
and maximum eigenvalues of $C_0$ are included to specify numerical support.}
\label{model:tab:lambda}
\input{content/model/generated/lambda-table.tex}

\end{table}

\subsection{Decoder regrouping and the nearly frozen operator background}
\label{model:sec:pretrained-freezing-results}
We independently composed the native decoder at boundaries after layers
2 and 4, checking four contexts per checkpoint. Direct primal evaluation
matched the wrapper's logits exactly. Automatic differentiation through
the entire loop and the transported two-block derivative in
Lemma~\ref{model:thm:jet} also matched exactly in the recorded float32
calculations. These checks validate the source routing and decoder
adapter, rather than provide an independent empirical test of the chain
rule.

Finally, one calibration context supplied all seventy generated operators
for a joint replacement of their generating networks. We compared this
fully frozen operator background with uncached native evaluation on 256
evaluation contexts. Table~\ref{model:tab:freeze} gives the output errors.
The maximum predictive KL is $\ModelSourceFrozenKLOne$ for checkpoint 1 and
$\ModelSourceFrozenKLTwo$ for checkpoint 2, with no changed argmax prediction in
either sample. The underlying logit differences are retained, so these
results are not presented as exact operator invariance. The smallest
computed KL values reach floating-point roundoff, including a value of
order $-10^{-15}$ for checkpoint 2; the raw signed values are preserved.

\begin{table}[htbp]
\centering\small
\caption{Replacement of all generated $\G$ tensors by one calibration
context's tensors. Errors concern final-position vocabulary predictions;
NLL uses the external next token. Each model is evaluated on 256 contexts.}
\label{model:tab:freeze}
\input{content/model/generated/freeze-table.tex}

\end{table}

This controlled approximation is compatible with the finite-variance
active fluctuation field measured above. A nearly constant deductive
background can coexist with variable attention, hidden representations,
and predictions. Corollary~\ref{model:cor:freezing} describes the exact limiting
case, while the table states the precision of the observed approximation.

\begin{table}[htbp]\centering\small
\caption{Direct matrix-level operator dispersion on 64 contexts at length
128, rows $[4096,4160)$, relative to anchor row 0. Fluctuation is the RMS
matrix deviation from its sample mean divided by that mean's Frobenius
norm. Norm CVs use the separate 4,608-context field cohort. Native float32
operators are compared with float64 diagnostic accumulation.}
\label{model:tab:operators}
\begin{tabular}{rrrrrr}
\toprule
Model & Median distance & Max distance & Max fluctuation & Median norm CV & Max norm CV\\
\midrule
1 & $6.86\times10^{-7}$ & $1.59\times10^{-3}$ & $4.41\times10^{-4}$ & $5.57\times10^{-8}$ & $1.19\times10^{-4}$\\
2 & 0 & $1.37\times10^{-6}$ & $3.04\times10^{-7}$ & $7.38\times10^{-14}$ & $1.20\times10^{-8}$\\
\bottomrule
\end{tabular}

\end{table}

The direct matrix comparison distinguishes near constancy from invariant
norms alone. The maximum relative matrix distance for checkpoint 1 is
about $1.59\times10^{-3}$, while its retained predictive freezing test has
maximum logit change $4.10\times10^{-3}$. Small KL and matrix dispersion
support a finite approximation, not equality of real operators. This is
the measured boundary of Proposition~\ref{model:prop:operator-approximation}.

\subsection{Uncertainty and evidence scope}
All statistical intervals are percentile intervals from 1,000 resamples.
The sampling unit is a complete document for response and segment
experiments, and a complete independent-document block for the displayed
block-moment intervals. Exponent intervals recompute every block scale
from the same resampled document sequence. Projection directions and
calibration normalizations stay fixed. The intervals are conditional on
the selected shards, calibration map, checkpoints, and observation family;
they do not quantify uncertainty over shard choice, training seeds, or
all possible model sources. No independent-replicate count is assigned
to the 70 heads, 32 projections, or eight perturbation directions.

\section{Initialization trajectories and finite response windows}
\label{model:sec:training-results}

Twelve full-parameter initialization trajectories use four head counts
$h=2,4,8,14$, widths $128,256,512,896$, five decoders, head dimension 64,
eight generator residual units, $A_{\rm dff}=170$, and FFN width
$\lfloor8w/3\rfloor$. Each uses AdamW at learning rate $3\times10^{-4}$,
$(\beta_1,\beta_2)=(0.9,0.95)$, $\epsilon=10^{-8}$, weight decay 0.01 and
global gradient clipping at norm 1. All parameters are trainable. Seeds
12001--12003 provide three initializations and associated sampler streams;
a stream is paired across widths. These are the first 256 updates, with
32 external targets per update, totaling 8,192 supervised targets per run.
They do not reconstruct historical pretraining of the released models.

Training samples come from rows $[0,3072)$ of the 4,608-document cohort.
Each update samples documents with replacement and uniform offsets
$0,\ldots,448$ inside its stored 513-token crop. The input is 64 tokens and
the target is the next excluded token. The 512 quality probes use rows
$[3584,4096)$ at offset zero. This probe crop rule differs from random-offset
training; the consistently cropped study in Section~\ref{model:sec:controlled-training-methods}
has a separate sampling boundary.

\begin{table}[htbp]\centering\small
\caption{Every step-256 checkpoint on the same 512 probe targets. NLL,
entropy and entropy SD are in nats. The matched unigram uses add-$1/2$
smoothing on the exact 8,192 supervised targets of each run. The full-pool
unigram uses all 1,575,936 tokens in the training pool. Neither comparison
establishes mature contextual language modeling.}
\label{model:tab:initialization-quality}
\input{content/model/generated/initialization-quality.tex}

\end{table}

Eleven model point estimates are below the matched-target unigram; every
model is above the full-pool unigram. These estimators use different
information budgets. Entropy differs substantially between initializations,
whereas within-checkpoint loss fluctuations are dominated by documents.
For the finite law uniform over the three checkpoints and 512 documents,
Table~\ref{model:tab:training-fractions} reports all-coordinate variance fractions.
The identity $\chi_b/\chi_1=1+(b-1)f_{\rm between}$ explains the effect of a
shared state, but recomputing it from the same array is arithmetic
consistency, not validation of a predictive training reduction.

\begin{table}[htbp]\centering\small
\caption{Between-checkpoint fractions at step 256. Median and maximum use
every coordinate; the final column counts fractions above one half.
Population denominators define the declared finite three-state law.}
\label{model:tab:training-fractions}
\begin{tabular}{rrrrrr}
\toprule
Width & Entropy & NLL & Median & Max & Count $>1/2$\\
\midrule
128 & 0.681 & $5.15\times10^{-5}$ & 0.385 & 1.000 & 24\\
256 & 0.883 & $6.12\times10^{-5}$ & 0.257 & 1.000 & 30\\
512 & 0.987 & $2.19\times10^{-4}$ & 0.296 & 1.000 & 45\\
896 & 0.998 & $7.38\times10^{-4}$ & 0.281 & 1.000 & 56\\
\bottomrule
\end{tabular}

\end{table}

A saved next minibatch and optimizer state determine an actual full-parameter
float32 AdamW displacement $\Delta\theta$. Sixteen external probes, rows
$[4096,4112)$ at offset zero, measure the predictive Fisher curvature in
that displacement. Forward/reverse differentiation agreement checks the
same graph's derivatives; the central secant at $h=0.05$ also includes
finite-step error. Table~\ref{model:tab:training-windows} keeps these diagnostics
separate and reports each run's error distribution.

\begin{table}[htbp]\centering\scriptsize
\caption{Per-run native response errors at $e=0.03$ and $e=1$, in percent;
$d_{.05}$ is the maximum relative Fisher norm of the central secant-minus-JVP
difference. The final column is the upper end of a contiguous tested grid
starting at $10^{-4}$ for which all 16 KL errors are at most 5\% in the
float64-parameter, dtype-preserving-softmax control. It is not a certified interval or a population
Taylor radius.}
\label{model:tab:training-windows}
\input{content/model/generated/initialization-response.tex}

\end{table}

The precision scans hold each float32 displacement fixed, use float64
parameters, and compare native float32 attention softmax with the same
formula preserving dtype. Both arms retain the native float32 rotary input
cast as well as the rotary cache coefficients, so they are mixed-precision
controls. The smooth control in Section~\ref{model:sec:augmented-response-results}
additionally removes that input cast. A centered
logit-increment identity evaluates KL with a small-increment polynomial to
reduce cancellation. The complete amplitude arrays and both precision
arms are retained. The local window depends on state: the tested upper
limits for width 128 at seeds 12002 and 12003 are $10^{-3}$ and
$3\times10^{-4}$, respectively. Mathematical small-displacement accuracy
and a numerically resolved amplitude window are therefore separate claims.
Proposition~\ref{model:prop:response-window} explains both limits explicitly.

\begin{figure}[ht]\centering
\includegraphics[width=.98\textwidth]{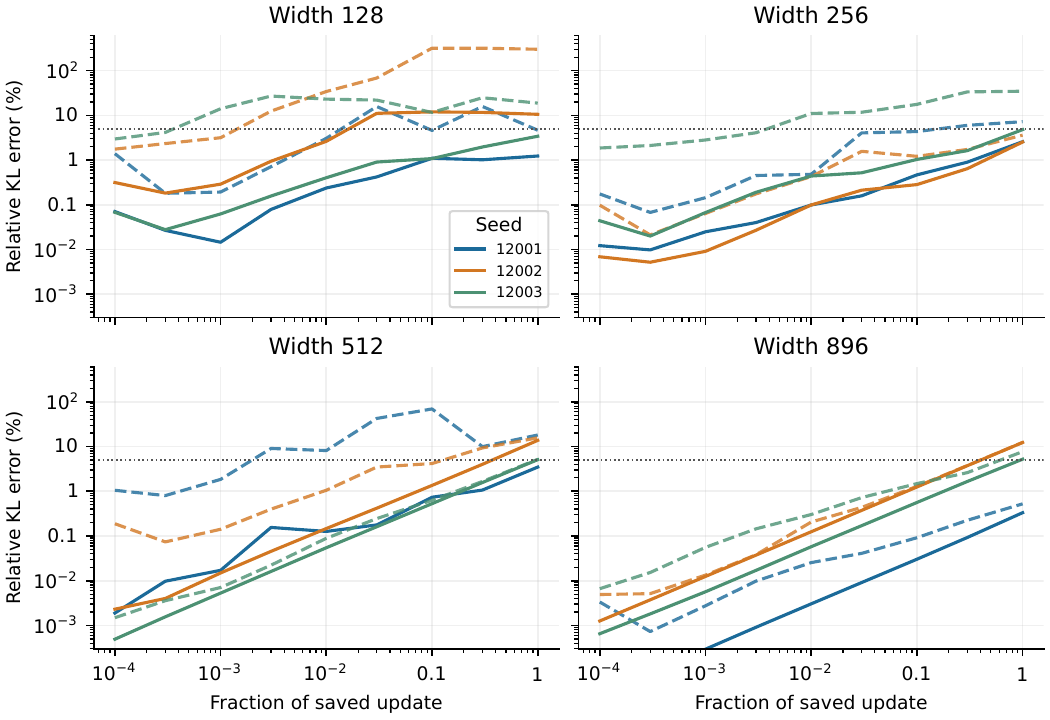}
\caption{Float64-parameter precision-control response scans at the twelve stored
initialization states. Solid curves are medians over 16 contexts; dashed
curves are maxima. The horizontal line marks a 5\% grid diagnostic.}
\label{model:fig:precision-windows}
\end{figure}

Four-step deterministic checkpoint replay, restoring parameters and
optimizer state at the regrouping boundary, gives zero recorded parameter
and moment discrepancies. Finite covariance decompositions agree to
roundoff, and sampling from the declared empirical mixture agrees with
its computed moments. These are numerical implementation and sampling
consistency checks. They do not establish autonomous coarse dynamics or a
critical training transition.

\section{Controlled training, conditional prediction and extended inference}
\label{model:sec:controlled-results}

\subsection{Finite parameterizations and training quality}
\label{model:sec:controlled-training-methods}
The controlled study contains 24 complete 2,048-update trajectories, each
with 65,536 supervised targets. Twelve use the native parameterization
at four widths and seeds 630101--630103. Nine pair the three larger widths
with the normalized parameterization of
Proposition~\ref{model:prop:adam-normalization}, using identical initial random
draws and sampler streams. Width 128 is their common reference.
Three further width-896 runs use fresh initialization seeds 630111--630113
and independent streams with the normalized design fixed. Generator width,
depth, head dimension, optimizer coefficients and target-safe objective
are otherwise the specified common architecture.

Training still samples the finite 3,072-document pool. The fresh
2,048-document short cohort supplies calibration and evaluation under the
same random-offset crop rule. Fixed 64-document calibration probes are
recorded every 32 updates in 36 common coordinates: five head-averaged
attention entropies, five head-averaged output RMS values, six boundary
RMS values, eight fixed vocabulary projections, predictive entropy and
NLL, and five per-layer normalized row energies and operator RMS values.
The vocabulary projection uses a separate fixed random generator so that
changing residual width does not change its coordinates. At five
milestones the 1,536 evaluation documents additionally measure loss,
entropy, within-context position permutations and baselines. Context
replacement by another document is measured at the final endpoint. Direct operator dispersion uses
64 fixed calibration documents. On the two RTX 4090 GPUs, a complete
training worker, including milestone evaluations and paired branches,
takes 118--334 seconds and uses at most 9.98 GiB of allocated device
memory. The 24 workers total 83.5 worker-minutes; CPU preprocessing and
separate response and inference measurements are additional.

\begin{table}[htbp]\centering\scriptsize
\caption{All 24 controlled endpoints. N denotes the native parameterization,
R the paired normalization, and S the three fresh-seed checks. Width 128
is printed once as the common reference. Matched unigram uses the exact
65,536 supervised targets. $\Delta_{\rm ctx}$ is the mean NLL increase
when each target's context is replaced by another document;
$\widehat I_{\rm ctx}$ is the corresponding mean full-vocabulary KL.
These context-derangement values are finite-cohort point estimates.
G dispersion is the median relative RMS matrix fluctuation across
heads and layers on the 64 calibration probes.}
\label{model:tab:controlled-quality}
\input{content/model/generated/controlled-quality.tex}

\end{table}

The two parameterizations select different finite output regimes.
The predictive context-dispersion coordinate distinguishes active
context-dependent predictions from nearly context-insensitive predictions,
even where operator dispersion is small in both. The joint initialization
and update normalization yields more comparable learning across the
measured widths. This intervention does not isolate its initialization
and learning-rate contributions. The matched-target and full-pool
unigram comparisons specify the information budget; lower probe loss
alone is not a criticality diagnostic.

\begin{figure}[ht]\centering
\includegraphics[width=.99\textwidth]{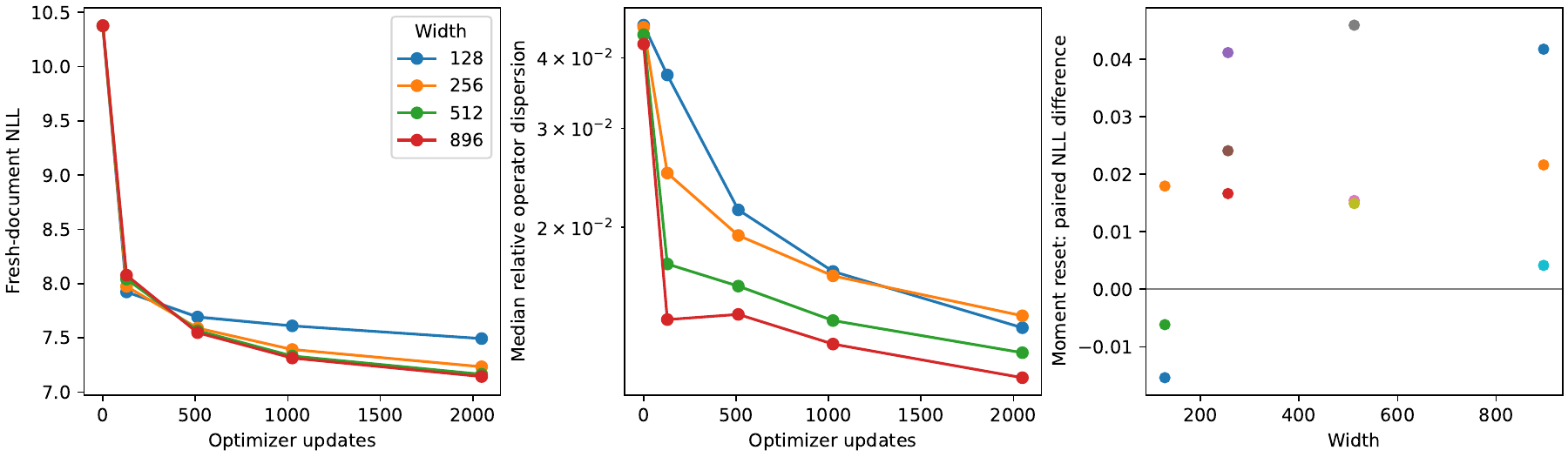}
\caption{Normalized-family learning, operator dispersion and paired moment
intervention. Width 128 uses the common reference trajectory. Shading
shows the range across three initializations, not a confidence interval
for a seed population. The right panel displays individual paired
16-update NLL differences after resetting moments.}
\label{model:fig:normalized-training}
\end{figure}

The three fresh-seed width-896 runs finish at NLL 7.144, 7.149 and 7.146
nats, compared with matched-target unigram losses 7.578, 7.560 and 7.564.
Replacing their full contexts increases NLL by 0.852, 0.789 and 0.798 nats;
preserving the final context token leaves increases of 0.078, 0.071 and
0.082 nats. Thus the fixed normalized design retains predictive use of
context on other training realizations while reusing the evaluation cohort. They do not provide an
additional model size or an independent corpus population. The results
support finite normalization and regime comparisons; the uniform
cumulant-density assumptions of Proposition~\ref{model:prop:regular-limit}
remain analytical hypotheses.

\clearpage
\subsection{Optimizer and generator interventions}
At each endpoint four branches begin with identical weights and the same
sixteen minibatches. They retain the optimizer, zero both moment tensors
while retaining their counters, suppress generator parameter updates,
or combine the two interventions. Global clipping is computed before
suppressing generator gradients, keeping that clipping factor common
at the first paired step. Suppression also disables weight decay for the
frozen parameters. The frozen parameter set comprises the residual metric learner and the
power layer, including their internal normalizations. Query/key maps and
the Gram-input LayerNorm remain in the other parameter group.
Generator parameter freezing does not replace the
input-dependent generated operator by a constant.

\begin{table}[htbp]\centering\small
\caption{Paired mean NLL differences after sixteen updates, relative to
ordinary continuation, on the same 256 evaluation documents. Positive
values increase loss. The full per-document arrays, intermediate steps
$1,2,4,8$, and document-bootstrap intervals are retained. Each row is a
training realization; probe documents are not additional realizations.}
\label{model:tab:controlled-branches}
\input{content/model/generated/controlled-branches.tex}

\end{table}

The interventions expose optimizer dependence at fixed initial weights
and quantify the effects of suppressing generator updates. They do not
force those effects to have one sign. Equations~\eqref{model:eq:inference-drift}
and \eqref{model:eq:covariance-drift} retain precisely these state- and
source-dependent contributions. An inference field alone cannot determine
a unique next transition for all these augmented states. We accordingly
retain optimizer coordinates in the training law, with reduced-state
errors governed by Propositions~\ref{model:prop:closure-error} and
\ref{model:prop:metric-closure}; no uniformly closed 36-coordinate kernel is
claimed from these finite branches.

\clearpage
\subsection{Prediction for independent minibatch ensembles}
At each of the twelve native step-2048 states, sixteen independent
minibatches draw from the fixed training pool. Each restores the complete
saved parameter and optimizer state before generating one native float32
AdamW displacement. Eight displacements supply calibration coefficients;
the remaining eight test a prediction for the unseen update ensemble.
All use the same sixteen fresh-cohort probe documents, rows $[768,784)$.

For each displacement the smooth arithmetic control computes
$z'(0)$ and $z''(0)$ by nested directional differentiation, then the
predictive quadratic and cubic coefficients in
Equation~\eqref{model:eq:response-cubic}. It also predicts NLL drift through
second order. The tested amplitudes range from $10^{-6}$ to one full
update. Softmax and rotary input arithmetic preserve float64; the native
rotary cache coefficients are retained. Explicit SiLU and LayerNorm
formulas support second derivatives. A native-cast primal comparison and
a central difference of first derivatives at $10^{-5}$ are separate
numerical checks. The full amplitude arrays remain available.

\begin{table}[htbp]\centering\scriptsize
\setlength{\tabcolsep}{3pt}
\caption{Local response and conditional drift at $e=10^{-5}$ for each
saved native state. Response errors use 16 minibatches and 16 probes;
percentiles describe spread. Cubic error uses the same quadratic
normalization, with the cubic correction subtracted. The standardized
ensemble discrepancy compares the first eight batches' second-order
prediction with the other eight batches' observed mean NLL increment.
Its SE combines the two independent batch-mean variance estimates.
Taylor bias is the mean observed-minus-predicted increment on the test
batches using their own directions, separating approximation from
calibration sampling error.}
\label{model:tab:drift}
\input{content/model/generated/drift-table.tex}

\end{table}

Across the twelve states, the largest smooth-versus-native-cast logit difference is $6.43\times10^{-6}$. The first-direction second-derivative secant check has median relative Fisher error $8.59\times10^{-8}$, 90th percentile $3.18\times10^{-5}$, and maximum $3.68\times10^{-3}$. At $e=10^{-5}$, the maximum quadratic response error across all 3,072 batch--probe pairs is 0.534\%, compared with 0.0091\% with the cubic correction. The largest calibration-versus-test discrepancy is 2.04 estimated standard errors; this is a descriptive ensemble check, not a simultaneous coverage guarantee.

This calculation tests two different predictions: a directional local
expansion and a conditional expectation estimated from other minibatches.
The latter retains sampling uncertainty even when the Taylor remainder
is small. At larger displacements the local expansion has state-dependent
accuracy; adding a cubic term does not make an arbitrary full update local.
The observed numerical window is consistent with
Proposition~\ref{model:prop:response-window} and is not a measured uniform
third-derivative bound. The conditional ensemble does not identify a
low-dimensional autonomous training process.

\clearpage
\subsection{Sixty-four adjacent segments on fresh documents}
A separate length-qualified cohort contains 1,024 documents with at least
4,097 tokens and a uniform crop of that length. Each released checkpoint
provides 64 adjacent length-64 calls with targets excluded. This gives
65,536 calls per checkpoint. The first 256 documents calibrate sixteen
fixed projections of the 174-dimensional field; the next 256 fit covariance
models; the final 512 evaluate sums, generalized susceptibility and disjoint
contrasts. Whole documents are the statistical units. Position-specific
means are removed within each split.

\begin{table}[htbp]\centering\small
\caption{Fresh long-document susceptibility in a fixed 16-dimensional
projection, with no covariance ridge. Sum and contrast columns are trace
ratios to the fine-position covariance. Contrasts assign $+1/\sqrt b$ to
one half of an aligned block and $-1/\sqrt b$ to the other half. The
selected-direction Rayleigh quotient uses a direction chosen on the fit
documents and evaluated on the other 512 documents. The permutation control
independently reassigns document identities at each position.}
\label{model:tab:long}
\input{content/model/generated/long-table.tex}

\end{table}

At $b=64$, sum trace ratios are 8.308 and 7.189; shard-stratified
document-bootstrap intervals are $[7.038,9.826]$ and $[6.409,8.070]$.
The corresponding contrast ratios are 1.237 and 1.223, with intervals
$[1.145,1.323]$ and $[1.147,1.310]$. Maximum generalized eigenvalues
are 32.935 and 25.599, and independently selected directions give
30.867 and 24.588. The independently permuted controls have trace ratios
1.000 and 1.007. These figures identify a strong persistent sum sector
and a much smaller but nonzero residual contrast sector on the measured
extent.

The shared-component-plus-residual hierarchy accounts for this separation
while retaining residual dependence. Coherent Gaussian covariance models
with a shared component and geometric residual, a geometric residual
alone, and a completely monotone power mixture were fitted on the first
sixteen positions of fit documents and evaluated on other documents up to
64 positions. Their complete residual errors are retained; none is used
to certify exact stationary closure or to identify a unique latent
mechanism. The general residual law is retained in the analytical state.
Proposition~\ref{model:prop:geometric-blocking} specifies the compatible RG of
the geometric family, including its lag-zero covariance, when that
conditional model applies. A finite covariance-shape exponent is not a
thermodynamic critical exponent.

\subsection{Properties retained by frozen pretrained inference}
\label{model:sec:controlled-inference-results}
The two released models are also evaluated on the 1,536 fresh short-cohort
documents. Context replacement increases NLL substantially, including
when the final context token is preserved. Replacing all seventy generated
operators by one calibration context's operators produces a separate
predictive approximation measured in the last two columns below.

\begin{table}[htbp]\centering\small
\caption{Fresh-cohort released-model inference. $\Delta_{\rm ctx}$ replaces
the entire input by another document, while $\Delta_{\rm earlier}$ restores
the true final context token. Targets are unchanged. Operator replacement
uses calibration row 0 and the same 1,536 evaluation targets. These
measurements concern frozen weights and do not reconstruct historical
training trajectories.}
\label{model:tab:fresh-inference}
\begin{tabular}{rrrrrr}
\toprule
Model & NLL & $\Delta_{\rm ctx}$ & $\Delta_{\rm earlier}$ & Max frozen KL & Max logit\\
\midrule
1 & 3.781 & 6.849 & 3.412 & $2.24\times10^{-7}$ & $4.51\times10^{-3}$\\
2 & 3.712 & 7.268 & 3.593 & $3.28\times10^{-12}$ & $1.34\times10^{-5}$\\
\bottomrule
\end{tabular}

\end{table}

The retained properties are the conditional emission, operator dispersion,
context-dependent vocabulary predictions and connected segment statistics.
Training-time optimizer autocorrelation does not persist as an inference
process after the weights are frozen. The measured active prediction and
small operator-replacement error can coexist, as the complete PLGA graph
and its source transport allow. No critical training surface has been
identified here, so no critical exponent is claimed to persist in these
inference measurements.

\section{Conditional head fluctuations and paired training responses}
\label{model:sec:criticality-results}

\subsection{Executed families and statistical units}
The conditional head experiment uses five decoders, 64-dimensional heads,
the complete eight-unit metric learner, and the 32,000-token vocabulary.
Its training law samples uniformly from 3,072 recorded RefinedWeb documents
and the 449 possible length-65 crops within each stored 513-token prefix.
Each batch contains 32 crops and supervises the next token after 64 input
tokens. The held-out 512-document cohort and crop offsets are fixed across
all conditions. Two hundred and fifty-six disjoint pairs supply the
finite-cohort predictive context statistic.

The scan varies $N\in\{2,4,8,14\}$ and generator-rate multiplier
$g\in\{0,1/4,1,4,16\}$, with four wide initializations per condition.
All runs use the same initial shared metric learner and the same complete
minibatch history. The learned shared parameters remain trainable when
$g>0$. The two initial laws are the extra fan-in scaling and the
shape-aware scaling of Equation~\eqref{model:eq:shape-normalization}.
They coincide exactly at $N=2$, giving \ModelSourceCriticalityScanRuns{} distinct
2,048-update training runs. The observed row-energy change between
$g=1$ and $g=4$ motivates a denser scan at $g=1.25,1.5,2$ in the
shape-aware family, adding \ModelSourceCriticalityFineRuns{} trajectories with
the same four seeds at each base width. These measurements sample the
finite transition bracket; their selection is exploratory. A withheld width $w=1536$, with $N=24$ and
$g=1$, adds \ModelSourceCriticalityWiderRuns{} initializations.
Eight of the original trajectories, at $N=2,14$ and $g=1$, continue to
8,192 updates with complete Adam states and sampler prefixes restored.
These continuations do not add independent initializations.

The reference AdamW update uses $(\beta_1,\beta_2)=(0.9,0.95)$,
denominator offset $10^{-8}$, weight decay $0.01$, and global gradient
clipping at norm one. Generator parameters have rate $3\times10^{-4}g$;
other parameters have rate $3\times10^{-4}(2/N)$. The two RTX 4090 GPUs
each have 24\,GB memory. The withheld wider runs have a maximum recorded
per-process allocation of \ModelSourceCriticalityMaximumMemory{}\,GiB.

For each context, covariance is centered across the four training
initializations before averaging over contexts. Let
$\chi_H=5^{-1}\tr\E_x\chi_N^{\rm q}$ for normalized attention entropy,
and define $\chi_R$ analogously for the row field in
Equation~\eqref{model:eq:conditional-head-fields}. The associated
trace enhancements are $A_H=\tr\E_x\chi_N^{\rm q}/\tr\E_x V_N$ and
$A_R$ for the row field. The reference is symmetrized over independent
head relabelings in each decoder. Heads, documents and time checkpoints
are not additional independent training realizations.
Seed ranges and leave-one-seed jackknife errors are descriptive;
four seeds do not supply calibrated population intervals or reliable
order-parameter tail distributions.

\subsection{Initialization units and finite width comparisons}
The initialization diagnostic evaluates four seeds at each of five widths
under three initial laws, for \ModelSourceCriticalityInitializationModels{} model
initializations. Table~\ref{model:tab:criticality-initialization} verifies the
variance calculation directly in the native architecture. The corrected
query gain and PLGA coefficient variance retain their reference units,
while the logits have nearly constant mean square on the fixed 16-context
diagnostic cohort. These are initial-law checks, with the full-training
limit qualifications of Section~\ref{model:sec:conditional-heads}.

Separately, the initial states of the four-seed training ensemble,
measured on its 512-context cohort, have $\chi_H$ between
\ModelSourceCriticalityInitialEntropyMinimum{} and \ModelSourceCriticalityInitialEntropyMaximum{}
across the four base widths; $\chi_R$ ranges from
\ModelSourceCriticalityInitialRowMinimum{} to \ModelSourceCriticalityInitialRowMaximum{}.
This finite stability of the initial conditional fluctuations is
consistent with the common-unit construction. It supplies no
higher-cumulant or infinite-width training certificate.

\begin{table}[ht]
\centering\small
\caption{Measured initialization units, averaged over four seeds.
Query gains use the first decoder. The final two columns use the
shape-aware family. The PLGA coefficient is the affine matrix $a$.
The reported logit mean square is a finite-cohort average.}
\label{model:tab:criticality-initialization}
\input{content/model/generated/criticality-initialization.tex}

\end{table}

\subsection{Native initialization-kernel comparison}
\label{model:sec:initialization-kernel-results}
Theorem~\ref{model:thm:initialization-limit} gives a prediction from the
initial law without fitting the measured head means. Its numerical
evaluation uses the native finite-dimensional PLGA head, analytically
integrates its independent values, and evaluates the Gaussian GLU gate
by quadrature. Each of \ModelSourceCriticalityKernelReplicas{} independent Monte
Carlo replicas uses \ModelSourceCriticalityKernelSamples{} heads per context and
layer on the sixteen-context initialization cohort. Comparing quadrature
orders 20 and 28 on the first replica gives a maximum entry difference
of \ModelSourceCriticalityKernelQuadratureError{}. This is a numerical comparison,
not a rigorous integration-error bound.

\begin{table}[ht]
\centering\small
\caption{Initial-law predictions compared with four independent wide
initializations per width. RMS H and R average squared head-mean errors
over seeds, contexts and decoders before taking the square root. Kernel
errors use every within-context token pair at the normalized decoder
input ($K_0$) and final decoder output ($K_5$). The twenty models repeat
members of the sixty-model diagnostic and do not add training seeds.}
\label{model:tab:criticality-initialization-limit}
\begin{tabular}{rrrrr}
\toprule
Width & RMS H & RMS R & RMS $K_0$ & RMS $K_5$\\
\midrule
128 & 0.1267 & $3.71\times10^{-3}$ & 0.0872 & 0.1078\\
256 & 0.0836 & $2.73\times10^{-3}$ & 0.0617 & 0.0789\\
512 & 0.0577 & $1.65\times10^{-3}$ & 0.0436 & 0.0564\\
896 & 0.0481 & $1.40\times10^{-3}$ & 0.0329 & 0.0433\\
1536 & 0.0358 & $1.07\times10^{-3}$ & 0.0253 & 0.0303\\
\bottomrule
\end{tabular}

\end{table}

From $N=2$ to $N=24$, the entropy RMS difference decreases from
\ModelSourceCriticalityKernelEntropySmall{} to \ModelSourceCriticalityKernelEntropyLarge{},
and the row-field difference from \ModelSourceCriticalityKernelRowSmall{} to
\ModelSourceCriticalityKernelRowLarge{}. The final residual-kernel RMS difference
falls from \ModelSourceCriticalityKernelCovarianceSmall{} to
\ModelSourceCriticalityKernelCovarianceLarge{}. All six measured residual stages
show decreasing discrepancy across the five widths. The repeated native
head fields are bitwise identical to their original initialization
diagnostics. The logit mean-square prediction is
\ModelSourceCriticalityKernelLogitPrediction{}, in the units of
Table~\ref{model:tab:criticality-initialization}.

\begin{figure}[ht]
\centering
\includegraphics[width=\textwidth]{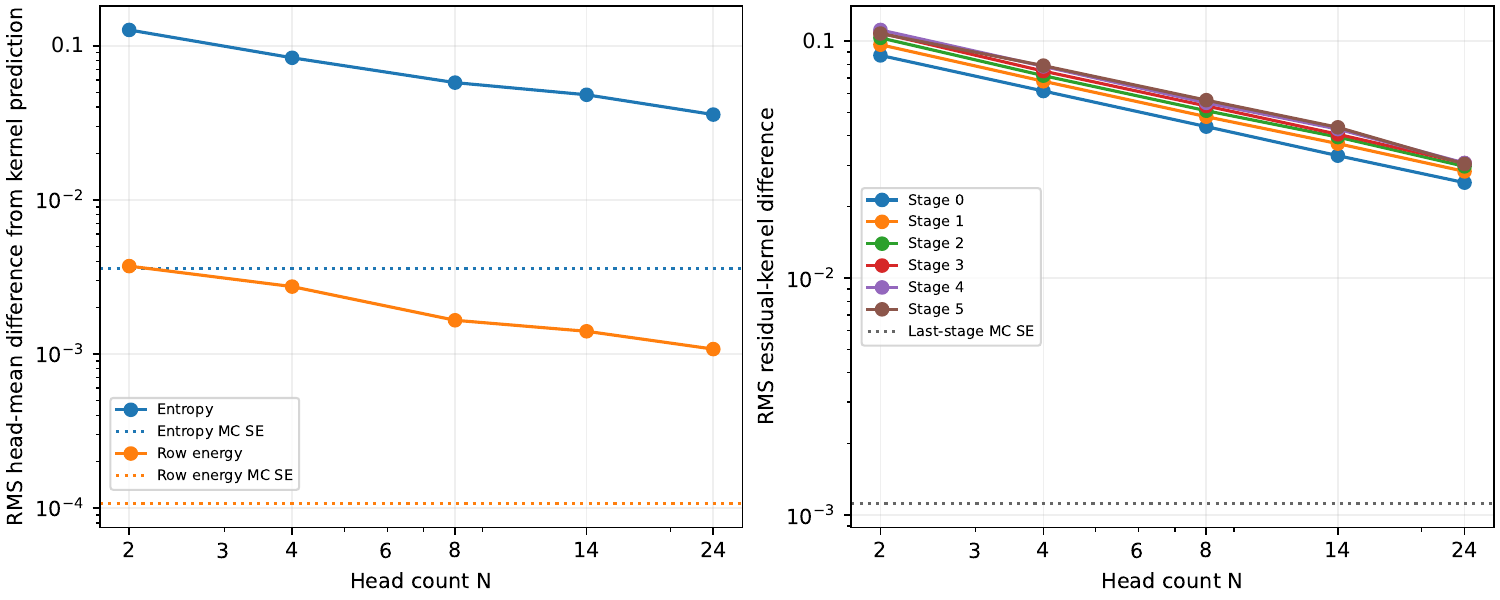}
\caption{Finite-width approach to the native initial-law predictions.
Dotted lines show RMS standard errors of the Monte Carlo mean, computed
across four numerical replicas. They describe numerical integration
variance, not its bias or population uncertainty. Stage 0 is the normalized
input and stages 1--5 are decoder outputs. No convergence exponent is
fitted.}
\label{model:fig:criticality-initialization-limit}
\end{figure}

The RMS Monte Carlo standard errors for the entropy and row predictions
are \ModelSourceCriticalityKernelMonteEntropy{} and \ModelSourceCriticalityKernelMonteRow{}.
The finite-model discrepancies remain larger than these numerical
standard errors at the measured widths. The comparison supports the
native initialization recursion and its observation boundary. It does
not establish a trained many-head law or the higher-cumulant bounds
needed to identify a thermodynamic critical point.

\subsection{Conditional fluctuations after training}
\label{model:sec:conditional-fluctuation-results}
At $g=1$, the row enhancement $A_R$ changes from
\ModelSourceCriticalityIntermediateEnhancementSmall{} at $N=2$ to
\ModelSourceCriticalityIntermediateEnhancementLarge{} at $N=14$, whereas the
absolute susceptibilities are respectively
\ModelSourceCriticalityIntermediateRowSmall{} and \ModelSourceCriticalityIntermediateRowLarge{}.
Thus the increased ratio occurs with a smaller absolute row fluctuation
at the larger endpoint. Reporting both quantities is essential to the
interpretation of Equation~\eqref{model:eq:head-susceptibility}; the ratio
alone would suggest a different width trend.

\begin{table}[p]
\centering\small
\caption{Shape-aware family after 2,048 updates. Each row contains four
training initializations under the same exogenous environment. NLL and
predictive context KL are seed means; susceptibilities are conditional
covariance estimates in the fixed bounded observation units. At $g=16$,
superscripts A and Q identify arithmetic agreement and qualification
under the paired bound criterion of Section~\ref{model:sec:observation-results}.}
\label{model:tab:criticality-variance}
\input{content/model/generated/criticality-variance-table.tex}

\end{table}

\begin{figure}[p]
\centering
\includegraphics[width=\textwidth]{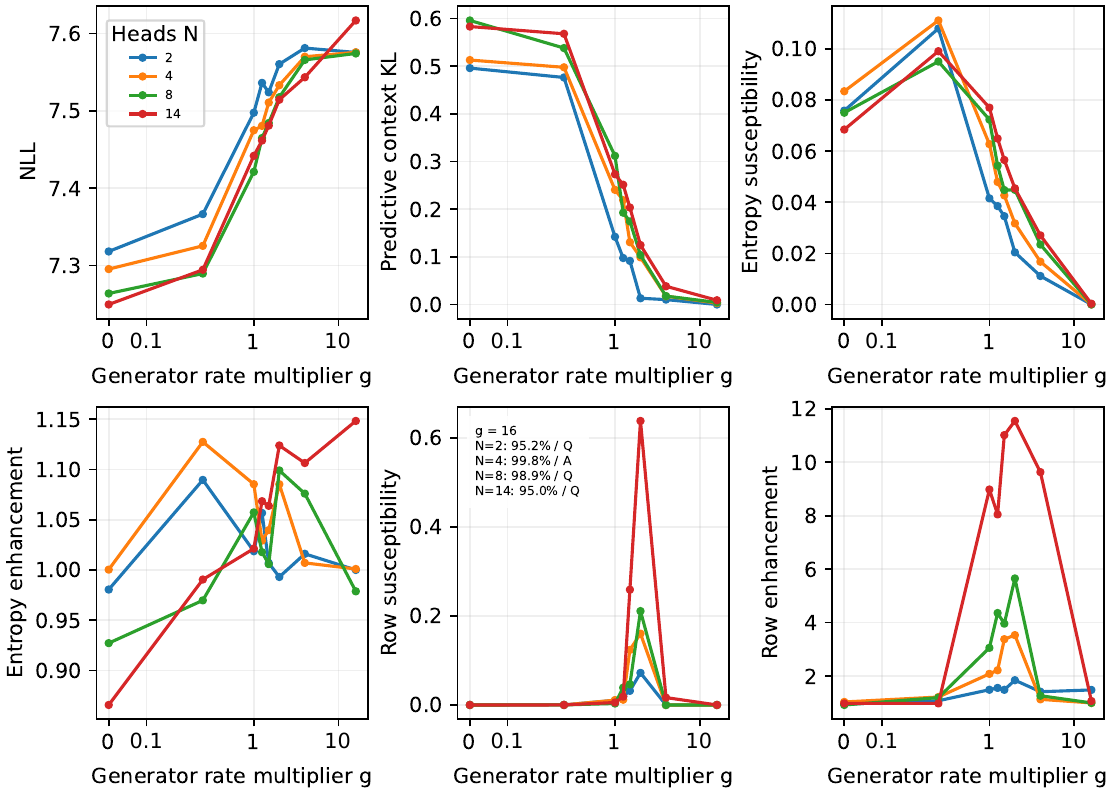}
\caption{Completed shape-aware width and generator-rate scan. The bottom
panels separate absolute row fluctuations from correlation enhancement.
Lines connect the measured controls and do not represent an inferred
critical scaling function. The row panel reports the $R<10^{-13}$
entry-screen fraction and arithmetic status of each $g=16$ condition;
the screen is not a susceptibility floor.}
\label{model:fig:criticality-variance}
\end{figure}

\begin{table}[ht]
\centering\small
\setlength{\tabcolsep}{4pt}
\caption{Largest measured row susceptibility at each base width in the
combined shape-aware control grid. These selected rates are not estimates
of a thermodynamic critical rate. The head-mean variance is $\chi_R/N$.
The jackknife and omission range hold the displayed rate fixed; they
do not include the selection of the largest point.}
\label{model:tab:criticality-sampled-peaks}
\input{content/model/generated/criticality-sampled-peaks.tex}

\end{table}

The largest measured row susceptibility occurs at $g=2$ at each of the
four base widths. These are selected grid maxima. Their values retain an intensive variance $\chi_R/N$
between \ModelSourceCriticalityPeakMeanVarianceMinimum{} and
\ModelSourceCriticalityPeakMeanVarianceMaximum{} across the base widths.
Equation~\eqref{model:eq:head-susceptibility} separates the one-head variance
and the shared covariance responsible for this normalization. A growing
finite-grid maximum consequently requires its covariance sector and
training-time dependence to be identified before it can determine a
critical scaling law.

\begin{table}[ht]
\centering\small
\caption{Seed-level spread at $g=1$ in the shape-aware family.
The covariance jackknife omits one training initialization at a time.
It describes sensitivity to the four measured seeds and is not a
calibrated uncertainty interval for a thermodynamic limit.}
\label{model:tab:criticality-spread}
\input{content/model/generated/criticality-seed-spread.tex}

\end{table}

The coarse $N=14$, $g=4$ row susceptibility illustrates the remaining
seed sensitivity. Its value is \ModelSourceCriticalityCoarseTailRow{}, but omitting
seed 640101 gives \ModelSourceCriticalityCoarseTailOmitted{}. That seed has mean
row field \ModelSourceCriticalityCoarseTailMean{}, while the other three means are
at most \ModelSourceCriticalityCoarseOtherMean{}. A large contribution from this
finite seed ensemble is therefore retained explicitly in interpreting
the generator-rate profile. It does not resolve the frequency of that outcome
under the initialization law.

The extra fan-in family is retained as a finite comparison in
Table~\ref{model:tab:criticality-fan-in}. Both families use the same optimizer
rates and generator interventions. Their different initial units are
therefore part of the family definition. Width trends across these two
laws cannot be combined into one thermodynamic extrapolation.

\subsection{Microscopic saturation and inference response}
All 140 scan checkpoints have an additional fixed-cohort last-query
attention diagnostic. Table~\ref{model:tab:criticality-saturation} compares
the microscopic concentration and categorical Jacobian trace with
the separately measured full predictive gain secants. Their Fisher-weighted squared norms use
centered source secants at amplitudes $0.01$ and $0.005$; the analysis records their
disagreement. The complete per-context controls are retained with the
numerical evidence.

\begin{table}[ht]
\centering\small
\caption{Selected controls from the complete saturation diagnostic.
The concentration column is the fraction of context--layer--head
observations whose largest attention probability exceeds 0.99.
The Jacobian concerns the last query; $q_\eta$ is the vocabulary Fisher
squared norm of the uniform head-gain secant at amplitude 0.005. These quantities have different
transport factors. Float64 categorical trace calculations are interpreted
at absolute numerical resolution, not as exact zero tests.}
\label{model:tab:criticality-saturation}
\input{content/model/generated/criticality-saturation.tex}

\end{table}

\begin{figure}[p]
\centering
\includegraphics[width=\textwidth]{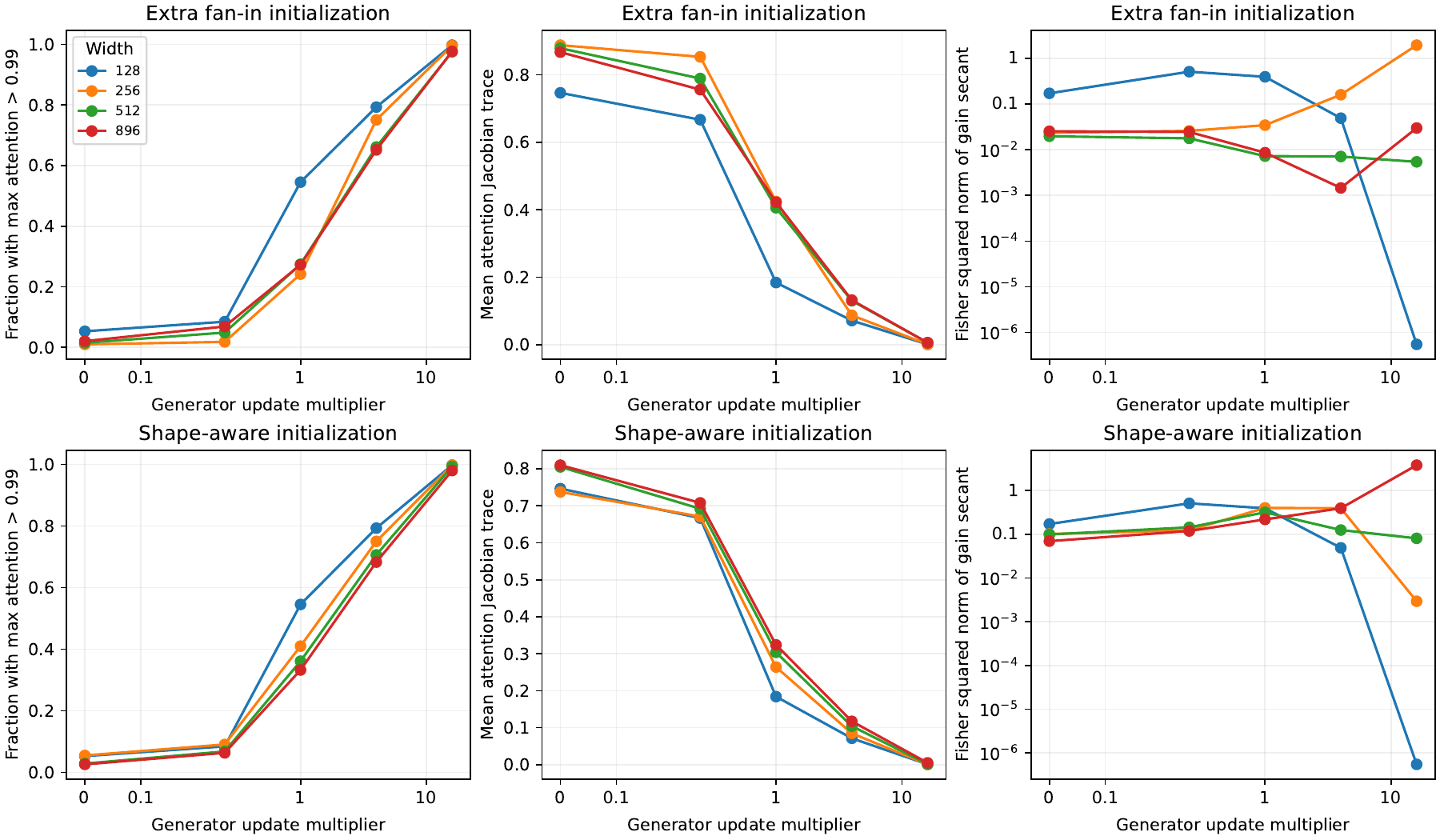}
\caption{Microscopic saturation and full predictive gain response in both
finite initialization families. The same $N=2$ trajectories supply the
common reference. A small local Jacobian tests the mechanism in
Proposition~\ref{model:prop:saturation}; the full-model secant response is measured
independently.}
\label{model:fig:criticality-saturation}
\end{figure}

The finite comparison family also separates concentration from context
independence. At width 896 and $g=16$, \ModelSourceCriticalitySaturatedComparisonPercent{}\%
of its measured last-query attention observations have $p_{\max}>0.99$,
while the predictive context KL is
\ModelSourceCriticalitySaturatedComparisonContextKL{} nats. Selected values can
retain context information in the local limit of
Corollary~\ref{model:cor:hard-selection}. The observed average concentration
is not a uniform-gap certificate or a bound on the full predictive
source response.

\begin{table}[ht]
\centering\small
\caption{Finite source-resolution controls for the shape-aware family.
The relative discrepancy is the Fisher norm of the difference between
the 0.01 and 0.005 source secants, divided by the Fisher norm of the
0.005 secant with denominator floor $10^{-14}$. The first two discrepancy columns bracket the four seed
medians; the maxima cover all recorded seed--context pairs. The absolute
column retains the unnormalized Fisher discrepancy when the source norm
is small. A finite secant norm is not an independently certified
infinitesimal derivative.}
\label{model:tab:criticality-source-controls}
\input{content/model/generated/criticality-source-controls.tex}

\end{table}

\subsection{Resolved source windows in selected larger-model states}
The widest-model source controls motivate a targeted numerical comparison
at $g=1,16$ for seeds 640101 and 640104. For each of these
\ModelSourceCriticalityPrecisionCases{} saved states, we select the four largest
recorded amplitude-0.005 squared secants and the fixed relative source
rows $0,16,32,48$, removing overlaps. The resulting
\ModelSourceCriticalityPrecisionContexts{} context--state observations are selected
controls, not a population tail survey or extra training realizations.

The float64 control preserves the dtype of softmax and rotary inputs,
retains native rotary-cache constants, and expands SiLU and LayerNorm
into their equivalent smooth formulas. At the same dtype its primal
logits agree with the unexpanded formula control within
\ModelSourceCriticalityPrecisionPrimalError{}. A full-graph directional JVP is then
compared with centered secants at ten amplitudes from $10^{-7}$ to
$10^{-2}$. The reported radius is a finite-grid result for resolved
source norms, with no bound asserted between grid points.

\begin{table}[ht]
\centering\small
\setlength{\tabcolsep}{4pt}
\caption{Selected full-graph source-resolution controls at width 896.
The count $n_r/n$ reports source norms above $10^{-14}$ out of all
selected contexts. $q^{64}_0$ is the float64 JVP Fisher squared norm;
$E_{.005}$ is the maximum relative secant error at amplitude 0.005.
The radius $e_{5\%}$ passes a 5\% error threshold for every selected
resolved context and every smaller tested amplitude. The final column
is the maximum KL from native CPU float32 predictions to the smooth
float64 predictions at zero source.}
\label{model:tab:criticality-gain-precision}
\input{content/model/generated/criticality-gain-precision.tex}

\end{table}

The tested radii are $5\times10^{-6}$ or $10^{-5}$ in these cases.
Thus an amplitude-0.005 finite response can be far outside the local
response window even when the zero-source prediction is well resolved.
The selected high-rate states also expose a separate arithmetic effect:
the KL from recorded GPU float32 predictions to native CPU float32
predictions reaches \ModelSourceCriticalityPrecisionNativeDeviceKL{}, and native CPU float32
versus smooth float64 predictions differ by as much as
\ModelSourceCriticalityPrecisionArithmeticKL{} nats. The smooth coefficient is
therefore reported with its primal precision change. It is not
substituted for a certified derivative of native finite arithmetic.
These observations instantiate the distinction between a finite source
measurement, its local expansion, and its arithmetic control.

\begin{figure}[p]
\centering
\includegraphics[width=\textwidth]{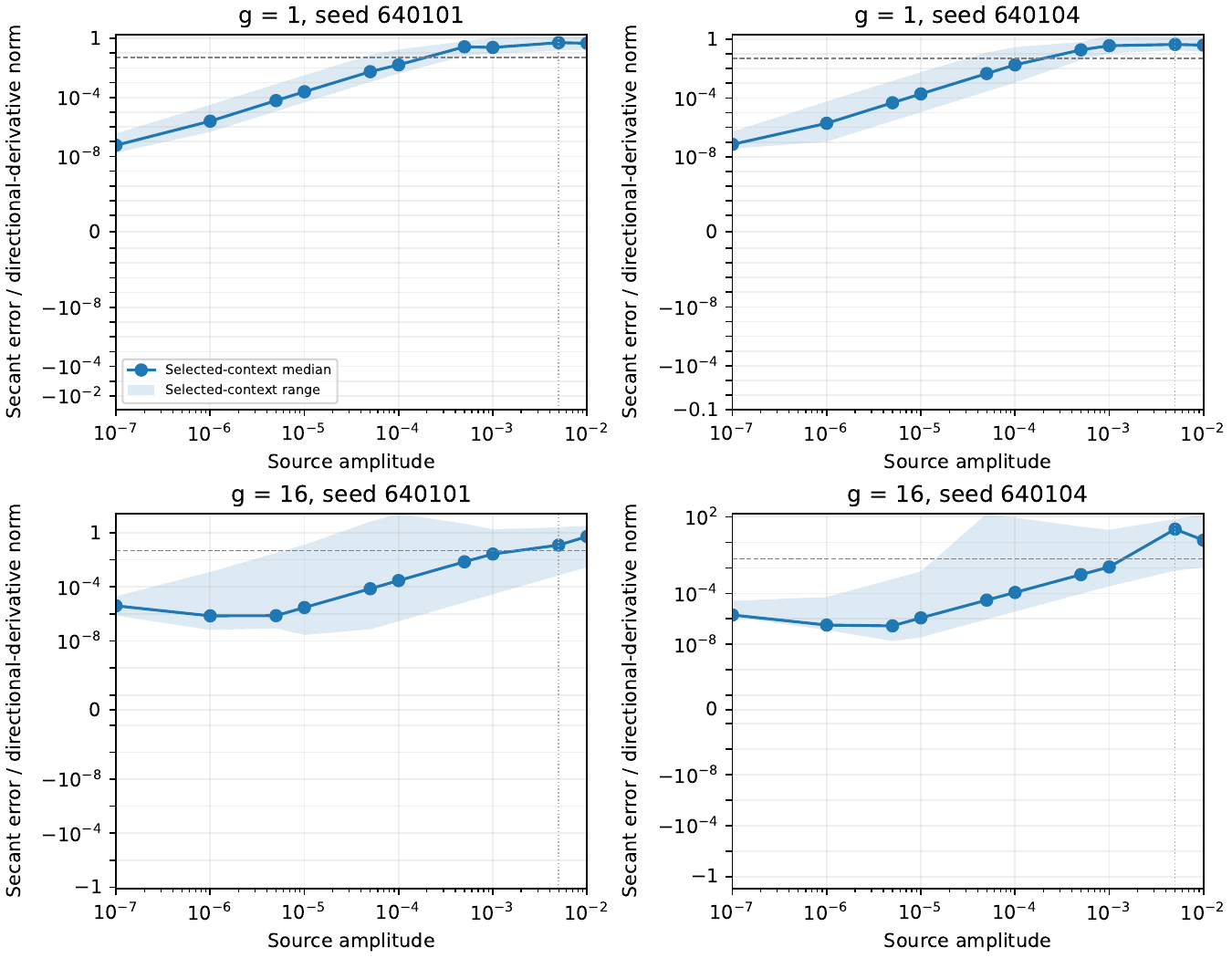}
\caption{Finite source secants converge to the float64 full-graph
JVP on the selected resolved contexts. Curves and bands show their
median and range; selection includes inspected tails and the bands
are not confidence intervals. The horizontal line is 5\% error and
the vertical line marks amplitude 0.005.}
\label{model:fig:criticality-gain-precision}
\end{figure}

\subsection{Horizon dependence and the withheld larger width}
Table~\ref{model:tab:criticality-horizon} follows the same initializations to
four times the scan horizon. Every continuation restores weights, moments,
counters and supervised-target counts, and uses the same subsequent
batch sequence. These measurements distinguish a finite training transient
from a statement about a stationary law.
The residual bias factor is $\beta_2^{2048}<3\times10^{-46}$,
so both Adam bias corrections already equal one at native floating-point
resolution. The continuation tests evolution of the weights and moments,
whose distribution need not have reached an invariant law merely because
the limiting update kernel admits one.

\begin{table}[ht]
\centering\small
\caption{Continued shape-aware trajectories at $g=1$. The smallest width
uses the exactly identical reference parameterization. Covariances at
different times use the same four seeds and are statistically dependent.}
\label{model:tab:criticality-horizon}
\input{content/model/generated/criticality-horizon.tex}

\end{table}

At $g=1$, the mean row field changes from
\ModelSourceCriticalityHorizonRowSmallStart{} to \ModelSourceCriticalityHorizonRowSmallEnd{}
at $N=2$, and from \ModelSourceCriticalityHorizonRowLargeStart{} to
\ModelSourceCriticalityHorizonRowLargeEnd{} at $N=14$. The endpoint ratio
$\chi_R(N=14)/\chi_R(N=2)$ changes from
\ModelSourceCriticalityHorizonWidthRatioStart{} at 2,048 updates to
\ModelSourceCriticalityHorizonWidthRatioEnd{} at 8,192 updates. At the larger width,
$\chi_R$ increases from \ModelSourceCriticalityHorizonChiLargeStart{} to
\ModelSourceCriticalityHorizonChiLargeEnd{}, while mean held-out NLL decreases
from \ModelSourceCriticalityHorizonNLLLargeStart{} to
\ModelSourceCriticalityHorizonNLLLargeEnd{}. These repeated trajectories show
continued regime selection and collective spread at fixed external
rate. Their strong horizon dependence is retained in the width
comparison; the measured endpoints do not identify a stationary
critical scaling law.

\begin{figure}[p]
\centering
\includegraphics[width=\textwidth]{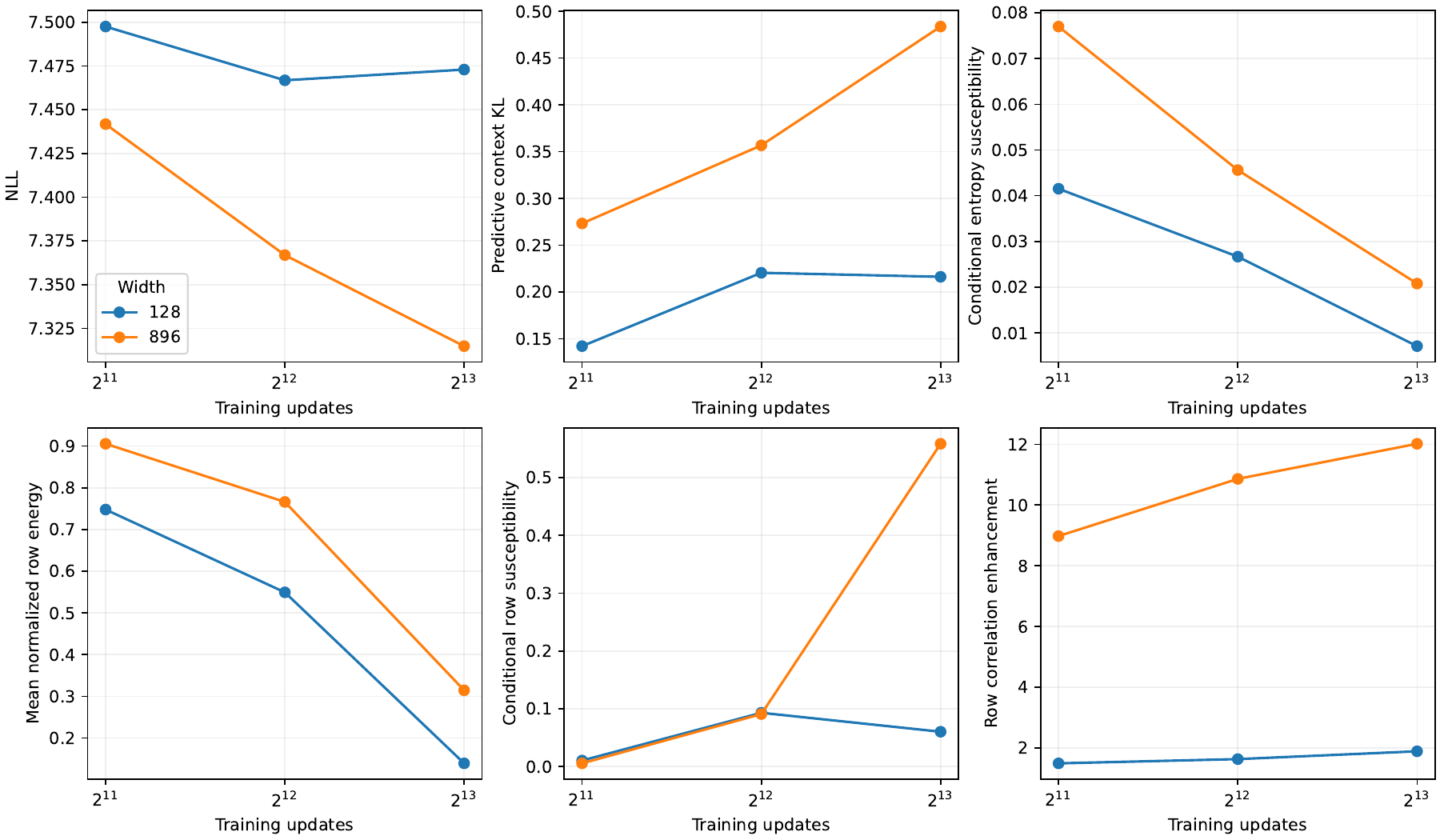}
\caption{Horizon dependence at fixed generator rate and exogenous
environment. Time is not pooled with width or counted as another
independent realization.}
\label{model:fig:criticality-horizon}
\end{figure}

At $g=1$, the four base widths fit two finite predictive descriptions,
$a+b/N$ with $a\ge0$, and $aN^p$ with $a\ge0$.
The predictions at $N=24$ are saved before the wider training runs begin
and are bound as inputs to their producer. Initial-law diagnostics at
that width are separate from this withheld trained endpoint. Table~\ref{model:tab:criticality-holdout}
scores them without refitting. The second form is a finite competing fit;
its power is not a derived critical exponent.

\begin{table}[ht]
\centering\small
\caption{Prediction at withheld width 1,536 using the four base widths
at $g=1$. Relative error is absolute prediction error divided by the
measured susceptibility; it is undefined when that susceptibility is zero.
The two fields retain their fixed units. Jackknife SE omits one of the
four wider-model initialization seeds at a time.}
\label{model:tab:criticality-holdout}
\input{content/model/generated/criticality-holdout.tex}

\end{table}

The regular prediction has \ModelSourceCriticalityWiderEntropyRegularPercent{}\%
relative error for entropy, compared with
\ModelSourceCriticalityWiderEntropyPowerPercent{}\% for the power prediction.
For row fluctuations the power prediction has
\ModelSourceCriticalityWiderRowPowerPercent{}\% error and the regular prediction
\ModelSourceCriticalityWiderRowRegularPercent{}\%. This mixed preference and the
large row-field jackknife error do not select a common scaling law.
A single withheld width at a fixed rate and horizon cannot establish
universality or identify a critical exponent.

\subsection{Paired operator pulses and finite response propagation}
\label{model:sec:paired-pulse-results}
Eight parent models at $N=2,14$ and $g=1$ each generate eight matched
512-update branches. The ordinary branches use physical operator pulses
$\alpha=0,\pm0.05,\pm0.1$. Three further branches use
$\alpha=0,\pm0.1$ with generator learning rates set to zero.
All branches retain their pre-pulse Adam moments and counters and share
the subsequent batches. Direct prediction comparisons check equivalence
between coefficient scaling and the inference gain source.

For a head-mean field $Q$, the signed response is
$d_\alpha(t)=[Q_{+\alpha}(t)-Q_{-\alpha}(t)]/(2\alpha)$.
Its norm uses the same 64 context and five decoder coordinates at every
time and is divided by $\|d_\alpha(0)\|$. An initial RMS below $10^{-7}$
is unresolved. Both the projection on the initial displacement and
the full response norm are retained. Comparing the two pulse amplitudes
tests the finite linear-response interpretation. The gain pulse is a
defined physical direction; no identification with the relevant coordinate
$r$ in Proposition~\ref{model:prop:feedback} is assumed. A decreasing paired
response measures contraction along this perturbation direction; it does
not require the unperturbed trajectory to be stationary. Baseline NLL
and the full observation trajectories are retained to distinguish this
contraction from the drift of the state itself.

\begin{table}[ht]
\centering\small
\caption{Paired response norms after 512 updates, divided by the initial
signed response norm at pulse amplitude 0.1. H and R denote entropy and
row energy. The final column, $\Delta_H$, is the relative discrepancy between the
entropy secants at amplitudes 0.05 and 0.1, with denominator floor
$10^{-14}$. An unresolved initial response
is displayed as a dash. These are eight parent realizations, not 64
independent training seeds.}
\label{model:tab:criticality-restoration}
\input{content/model/generated/criticality-restoration.tex}

\end{table}

After 512 updates the ordinary entropy response norms range from
\ModelSourceCriticalityPulseEntropyOrdinaryMinimum{} to
\ModelSourceCriticalityPulseEntropyOrdinaryMaximum{} times their initial values;
the row response norms range from \ModelSourceCriticalityPulseRowOrdinaryMinimum{}
to \ModelSourceCriticalityPulseRowOrdinaryMaximum{}. Generator freezing gives
entropy factors \ModelSourceCriticalityPulseEntropyFrozenMinimum{} to
\ModelSourceCriticalityPulseEntropyFrozenMaximum{} and row factors
\ModelSourceCriticalityPulseRowFrozenMinimum{} to \ModelSourceCriticalityPulseRowFrozenMaximum{}.
These finite pulses propagate into separated trajectories in both
update conditions. The entropy secants at the two pulse amplitudes
differ by relative factors \ModelSourceCriticalityPulseAmplitudeMinimum{} to
\ModelSourceCriticalityPulseAmplitudeMaximum{} at the endpoint. They consequently
do not identify a common linear relaxation coefficient. The source-window
measurements above concern instantaneous inference and do not themselves
bound the amplitude window of a 512-update training response.

\begin{figure}[p]
\centering
\includegraphics[width=\textwidth]{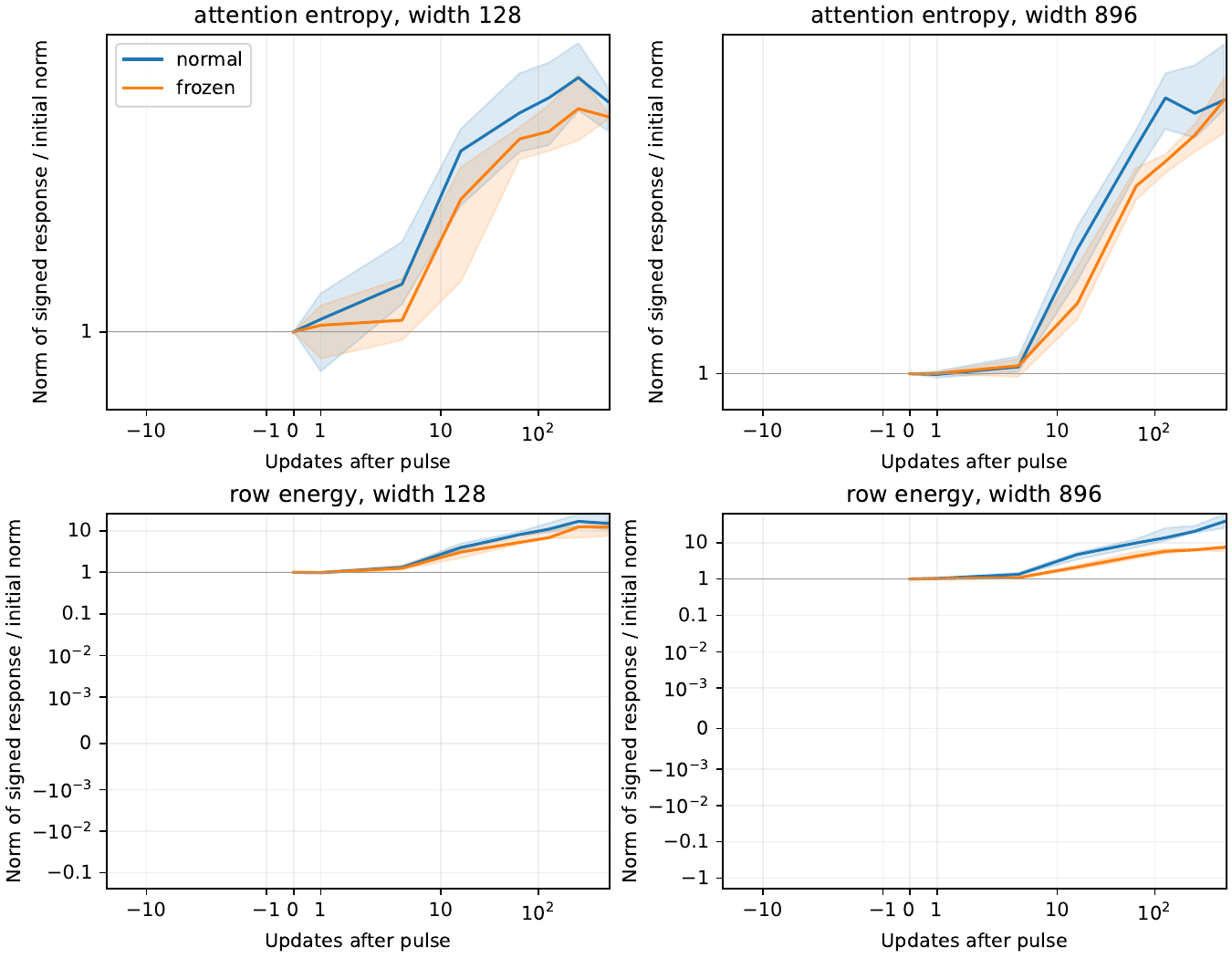}
\caption{Signed-pulse response norms during matched continuation.
Curves show seed medians and bands the observed seed range for ordinary
updates and frozen generator weights. Bands are descriptive, not
confidence intervals. The curves describe finite displacement propagation
under the paired kernel of Equation~\eqref{model:eq:paired-kernel}.}
\label{model:fig:criticality-restoration}
\end{figure}

\subsection{Numerical replay of the paired dynamics}
Four of the pulse parents, at $N=2,14$ with seeds 640101 and 640104,
have two additional unperturbed 512-update branches each. Both copies
restore the same full training state, use identical subsequent batches,
and receive zero physical pulse. Their head-field separation measures
native numerical repeatability. A comparison with the originally recorded
unperturbed branch also measures repeatability across executions.
These \ModelSourceCriticalityReplayBranches{} branches reuse
\ModelSourceCriticalityReplayParents{} initializations.

\begin{table}[ht]
\centering\small
\setlength{\tabcolsep}{4pt}
\caption{Unperturbed native replay controls after 512 updates. RMS values
use the 64 context and five decoder head-mean coordinates. The H and R
ratios divide this separation by the corresponding original $+0.1$
versus $-0.1$ pulse separation. The final column instead uses the
difference between the new and original unperturbed row fields in the
numerator. These are numerical controls on existing parents.}
\label{model:tab:criticality-training-replay}
\begin{tabular}{rrrrrrr}
\toprule
Width & Seed & $\mathrm{RMS}(\Delta H)$ & $\mathrm{RMS}(\Delta R)$ & H ratio & R ratio & R original ratio\\
\midrule
128 & 640101 & 0 & 0 & 0.000 & 0.000 & 0.000\\
128 & 640104 & 0 & 0 & 0.000 & 0.000 & 0.000\\
896 & 640101 & 0 & 0 & 0.000 & 0.000 & 0.000\\
896 & 640104 & 0 & 0 & 0.000 & 0.000 & 0.000\\
\bottomrule
\end{tabular}

\end{table}

For all four parents, the two new branches and the original unperturbed
branch have bitwise identical recorded losses and head fields. Their
measured replay separations, replay-to-pulse ratios and predictive replay
KL are zero. Thus these controls resolve no unperturbed replay floor
for the measured finite-pulse separation. They certify repeatability
only for the recorded native executions and observations.

The stationary-law result and the paired-kernel result have different
conclusions. Proposition~\ref{model:prop:finite-stationarity} supplies an
invariant marginal law at fixed width, whereas
Proposition~\ref{model:prop:paired-kernel} does not assert attraction of a
perturbed pair to the diagonal. Neither a finite pulse amplification
nor numerical trajectory separation by itself determines the stationary
population law, a Lyapunov exponent, or feedback toward a critical
surface.

\clearpage
\subsection{Complete finite comparison family}
\begin{figure}[ht]
\centering
\includegraphics[width=\textwidth]{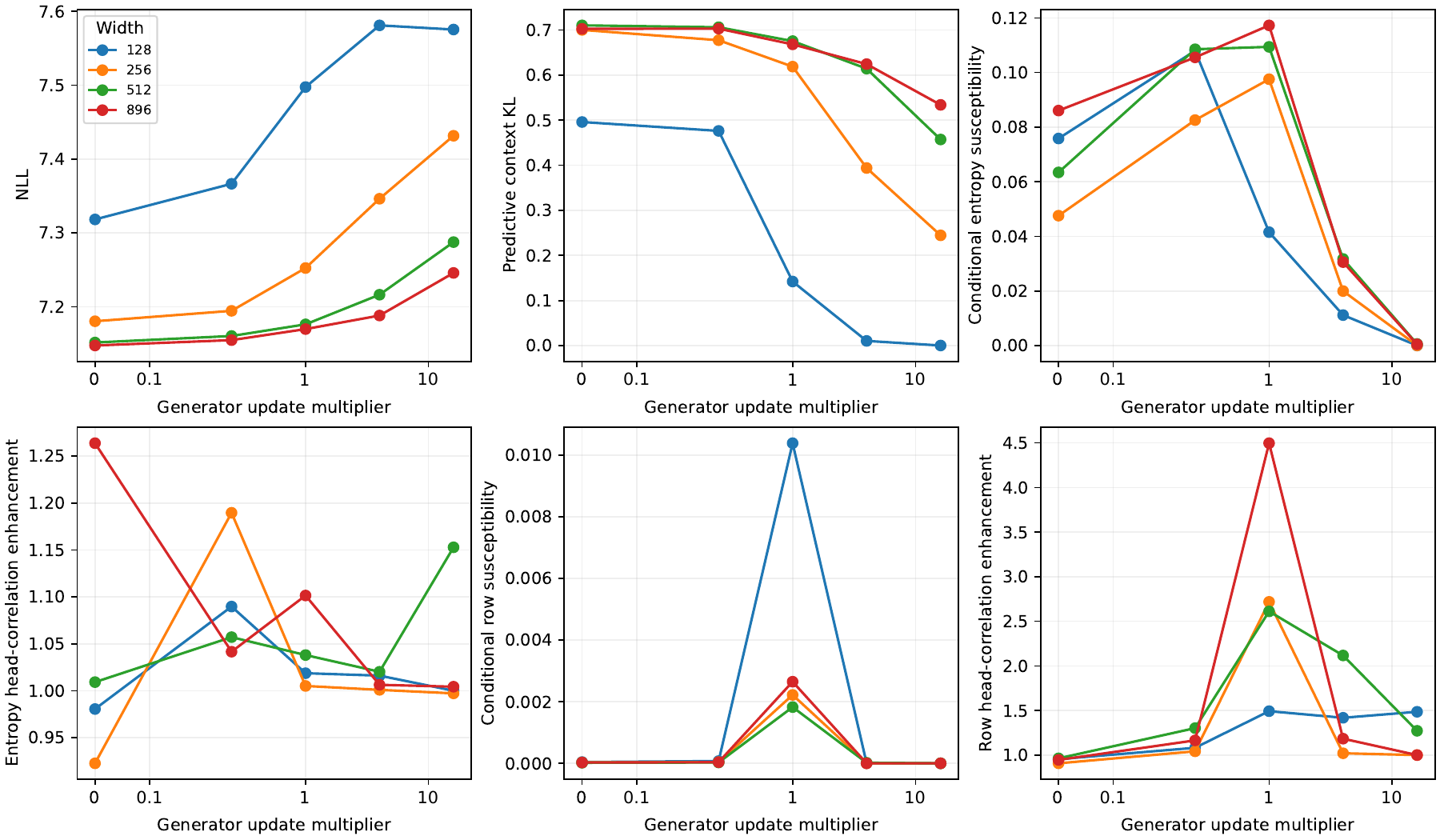}
\caption{Completed finite comparison under extra fan-in scaling.
The initial per-head units differ from those of the shape-aware family;
the two scans are analyzed separately.}
\label{model:fig:criticality-fan-in}
\end{figure}

Table~\ref{model:tab:criticality-fan-in} and
Figure~\ref{model:fig:criticality-fan-in} report the extra fan-in family
after 2,048 updates. The runs use the same coarse generator-rate grid,
four initializations per condition and conditional covariance convention
as the shape-aware family in
Section~\ref{model:sec:conditional-fluctuation-results}.
At every displayed width, the largest sampled row susceptibility $\chi_R$
occurs at $g=1$. This is a finite-grid comparison within the extra fan-in
family: its different initial per-head units preclude pooling the two scans
into a single width extrapolation.

\begin{table}[ht]
\centering\small
\caption{Extra fan-in family with the same generator-rate scan and
conditional covariance convention as Table~\ref{model:tab:criticality-variance}.
The width-128 reference is counted once in the unique run inventory.}
\label{model:tab:criticality-fan-in}
\input{content/model/generated/criticality-scan-table.tex}

\end{table}

\clearpage

\section{Width, training time and learned collective dynamics}
\label{model:sec:dynamics-results}

\subsection{Executed ensembles and temporal measurements}
The width--time study retains the shape-aware initial law, shared
initialization 640011, batch-history seed 640001, and native clipped AdamW
specified above. A rate grid uses $g\in\{1.5,2,3,4\}$ at
$N\in\{4,8,14\}$, with four initialization identities per condition
and a horizon of 8,192 updates. These are 36 exact continuations and
12 fresh trajectories. At $g=1$, 48 fresh trajectories and twelve
continuations complete a sixteen-seed ensemble at each of
$N=2,4,8,14$ and a four-seed ensemble at $N=24$, including eight
bound trajectories already present in the horizon evidence. The wide
initialization identities are 640101--640116 at the four smaller widths
and 640101--640104 at the largest. Twenty further continuations double
the horizon of the first four identities at all five widths to 16,384.
Thus these measurements add 128 trajectory artifacts, comprising
60 fresh trainings and 68 continuations. Continuing one trajectory
never increases the number of independent initializations.

The full population probe uses the same 512 held-out contexts at each
reported horizon (Table~\ref{model:tab:statistical-methods}). Dense observations reuse sixteen of them every 32
updates. For $g=1$, covariance uncertainty resamples whole initialization
identities 5,000 times, with fixed contexts and paired time points.
The sixteen-seed covariance divisor is fifteen. At $N=24$ and in the
rate grid, the four-seed divisor is three. The $N=2$ reference reuses
its identity normalization exactly, including the recorded convention
of its continuation inputs. All runs retain complete Adam moments,
counters and the original sampler prefix. Intermediate checkpoints and
per-tensor moment, update and denominator-offset diagnostics are bound
with the observations.

The rate and mechanism choices are developmental. Shared initialization,
the complete batch history and the evaluation contexts are held fixed;
the intervals therefore quantify initialization uncertainty within this
conditional ensemble. They do not include another common training drive
or an alternative shared initial state.

\subsection{Temporal resolution of the rate-dependent row sector}
At $g=2$, the conditional row susceptibility changes from
\ModelSourceDynamicsPeakSmallStart{} to \ModelSourceDynamicsPeakSmallEnd{} at $N=4$,
from \ModelSourceDynamicsPeakMiddleStart{} to \ModelSourceDynamicsPeakMiddleEnd{} at $N=8$,
and from \ModelSourceDynamicsPeakLargeStart{} to \ModelSourceDynamicsPeakLargeEnd{} at $N=14$
between 2,048 and 8,192 updates. The row means decrease over the same
interval. This is a measured evolution of the original ensemble, with
its optimizer state retained exactly. Its early rate-grid maximum
therefore cannot serve as an estimate of a stationary critical control.

\begin{table}[htbp]\centering\small
\caption{Four-seed rate conditions at matched horizons. Susceptibilities
are conditional seed covariances averaged over 512 fixed contexts and
five decoders. The endpoint standard error is a delete-one-seed jackknife.
The final column gives the percentage of entries with $R<10^{-13}$
and the arithmetic status: A meets the paired bound criterion, Q is
qualified, and M has no paired arithmetic control. The screen does not
change the estimator. Complete fields remain in the bound analysis.}
\label{model:tab:dynamics-rate-time}
\input{content/model/generated/dynamics-rate-time.tex}

\end{table}

\begin{figure}[htbp]\centering
\includegraphics[width=\textwidth]{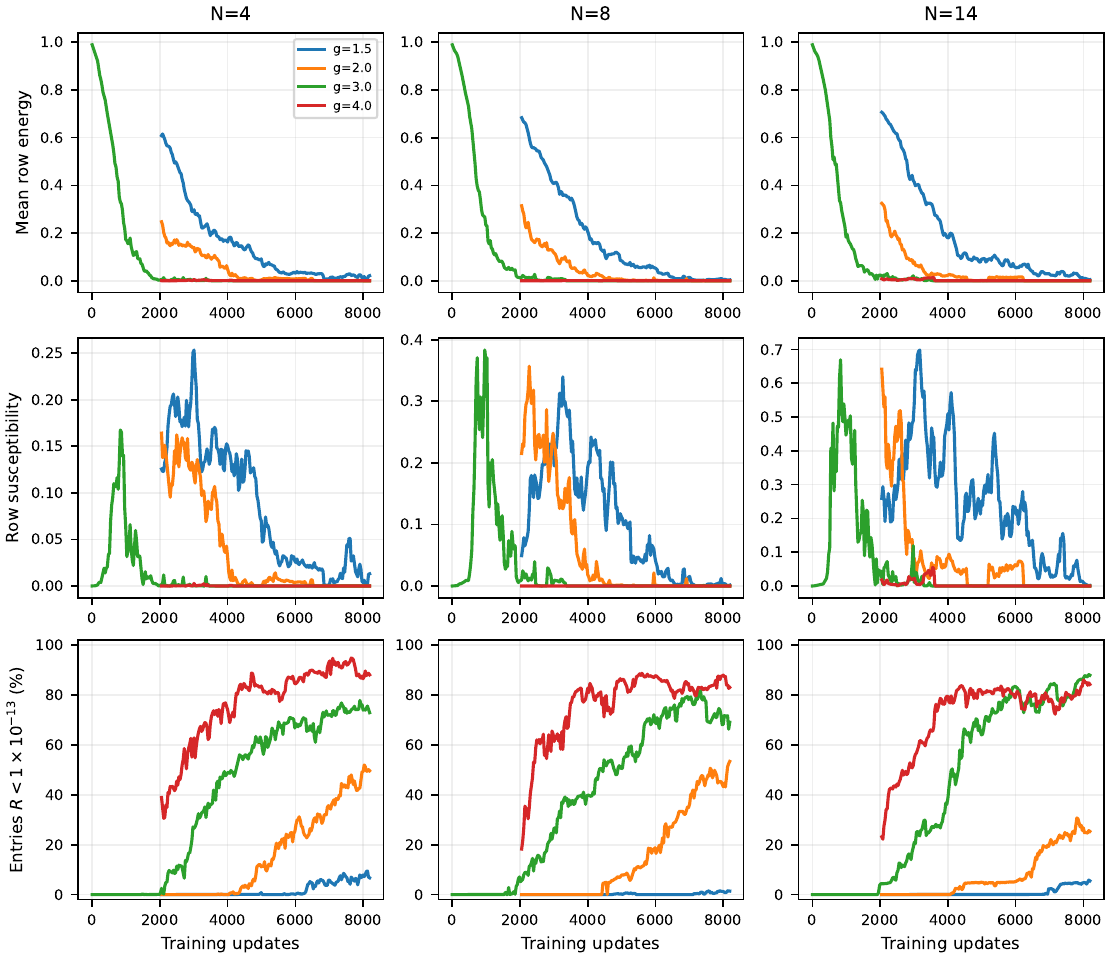}
\caption{Row means and absolute susceptibility along the rate-grid
trajectories. Every curve uses the same sixteen-context dense cohort
and four initialization identities. Time samples are repeated
observations of those identities. The bottom panels report the
entry-screen fraction on this same dense cohort; the threshold acts
on $R$, not on the susceptibility. Complete-cohort arithmetic
comparisons are reported separately.}
\label{model:fig:dynamics-rate-time}
\end{figure}

The archived point estimates are lower at 8,192 than at 2,048 updates
in each rate--width condition. Their four-seed uncertainty and
condition-specific arithmetic qualifications remain separate
(Section~\ref{model:sec:observation-results}); the magnitude of the change
differs substantially.
At $g=1.5$, the final mean row energies range from about 0.0037 to
0.0211; the faster-rate conditions attain much smaller values.
The endpoint attention susceptibilities remain nonzero and do not
follow the same degree of concentration.
The absolute decrease is consistent with the bounded row-concentration
sector in Proposition~\ref{model:prop:row-concentration}. The finite data do
not establish its asymptotic $o(N^{-1})$ mean condition. Nor does a
concentrating row field imply a constant predictive distribution:
attention and the full vocabulary response retain their own measured
coordinates. Equation~\eqref{model:eq:collapse-time-example} shows explicitly
how a width-growing transient peak can coexist with a noncritical
long-time point mass; identifying the native limiting law requires the
actual joint width--time limit.

\subsection{Independent replication and horizon doubling}
\begin{figure}[htbp]\centering
\includegraphics[width=\textwidth]{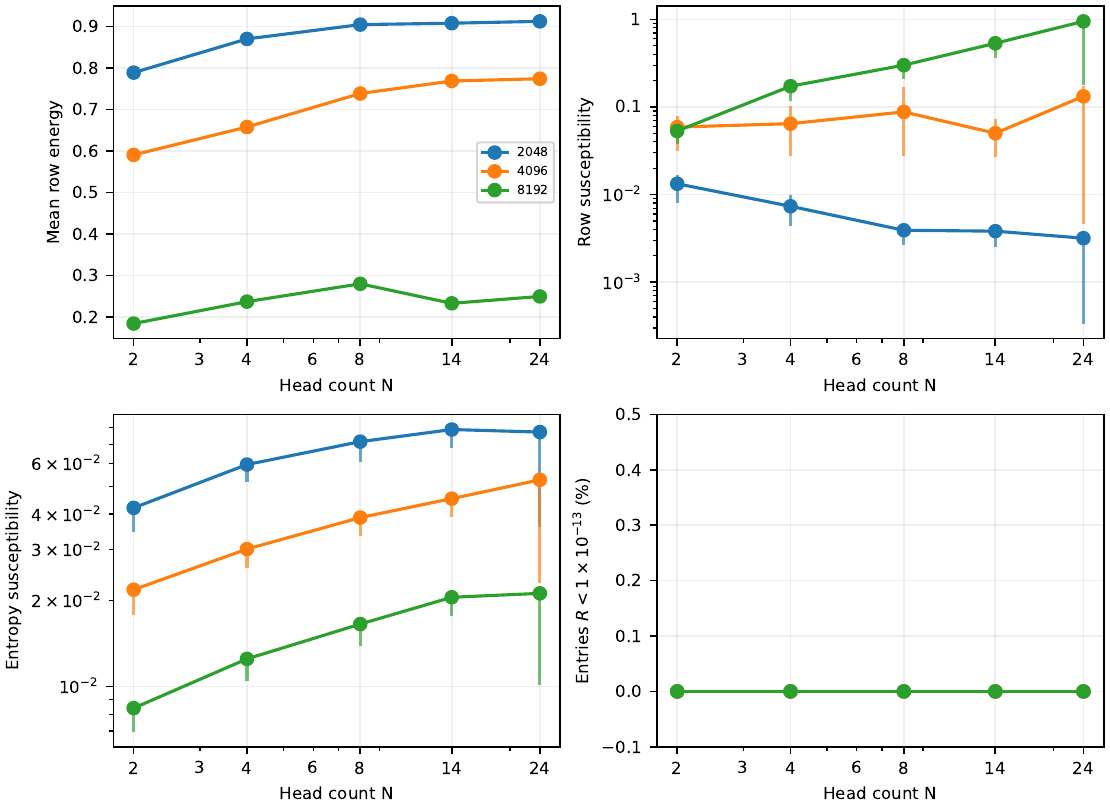}
\caption{The slower-rate width family at three matched horizons.
Vertical intervals show 95\% whole-seed bootstrap percentiles for
absolute susceptibility. Means, entropy and row fluctuations retain
separate observation units. The fourth panel gives the $R<10^{-13}$
entry-screen fraction on the same 512-context cohort. It does not
provide an arithmetic certificate for the susceptibility.}
\label{model:fig:dynamics-replication}
\end{figure}

Table~\ref{model:tab:dynamics-replication} resolves the $g=1$ ensemble with
sixteen independent initializations at four widths. The fixed-context
centering is unchanged as replication increases. The five-width
comparison and its whole-seed bootstrap intervals appear in
Figure~\ref{model:fig:dynamics-replication}; the largest width retains four
initializations and correspondingly different statistical precision.
Table~\ref{model:tab:dynamics-long} uses the first four identities at every
width to compare 8,192 and 16,384 updates without changing the ensemble
between its two columns.

\begin{table}[htbp]\centering\small
\caption{Replicated conditional means and susceptibilities at $g=1$.
There are sixteen independent seeds at $N=2,4,8,14$ and four at $N=24$.
Intervals resample whole seeds; contexts and repeated horizons do not
increase that count. These finite-sample percentile intervals are not
a thermodynamic confidence statement.}
\label{model:tab:dynamics-replication}
\input{content/model/generated/dynamics-replication.tex}

\end{table}

\begin{table}[htbp]\centering\small
\caption{Matched four-seed horizon doubling at $g=1$. Each width uses
the same four trajectories in both horizon columns. The quoted
standard error is over initialization identities, not time samples.}
\label{model:tab:dynamics-long}
\input{content/model/generated/dynamics-long.tex}

\end{table}

At update 8,192, the sixteen-seed row susceptibilities are
$0.0532$, $0.1726$, $0.3002$ and $0.5336$ at $N=2,4,8,14$,
respectively. Their whole-seed 95\% percentile intervals are
$[0.0382,0.0612]$, $[0.1166,0.2091]$, $[0.2076,0.3502]$ and
$[0.3585,0.6439]$. The four-seed $N=24$ estimate is $0.9490$,
with the much wider interval $[0.1794,0.9490]$.
The measured width dependence is therefore substantial at this
horizon. However, the paired changes in mean row energy from
4,096 to 8,192 updates range from $-0.406$ to $-0.536$ across
the four sixteen-seed widths, and every corresponding interval is
strictly below zero. The fluctuation increase occurs during ongoing
row concentration.

Horizon doubling reduces the row mean in every one of the twenty
continued trajectories. The five final ensemble means are
$0.00794$, $0.0101$, $0.0159$, $0.0608$ and $0.0305$ in increasing
head-count order. Their susceptibilities are $0.00119$, $0.00234$,
$0.0189$, $0.2686$ and $0.1420$. Relative to the matched four-seed
8,192-update ensembles, these retain $1.97\%$, $0.849\%$, $13.3\%$,
$48.1\%$ and $15.0\%$ of the earlier absolute susceptibility.
The point estimates at the longer horizon are also no longer
monotone at the two largest widths; the four-seed sample does not
establish their population ordering. Entropy susceptibility point estimates are also lower at each width.
For row susceptibility, the paired $N=14$ difference is $-0.2901$,
with jackknife standard error $0.3670$ and exact empirical percentiles
$[-0.5678,0.0541]$. The four-seed uncertainty therefore includes zero
at this width. Mean row energy separately decreases in all twenty
continued trajectories. Section~\ref{model:sec:observation-results} reports
the completed sixteen-seed extension of this selected condition.

Thus replication resolves finite width-dependent fluctuations while
the paired continuations establish a substantial change of the law
with training time. These observations are compatible with the
conditional row-concentrating sector and do not identify a stationary
singularity. In particular, fitting a width power to the 8,192-update
row susceptibilities would assign an exponent to a measured transient.
Neither the observed temporal decrease nor the later width dependence
determines the joint infinite-width and long-time limit.

\subsection{Measured optimizer scales in the conditional family}
\label{model:sec:optimizer-scale-results}
The saved Adam states provide a direct denominator-scale diagnostic.
The 64-state inventory contains all four $g=2$ identities at
$N=4,8,14$ at updates 2,048 and 8,192, and all four matched $g=1$
identities at $N=2,4,8,14,24$ at 8,192 and 16,384 updates.
Coordinates are grouped into the shared metric learner, the remaining
per-head power parameters, and all other parameters. The input
LayerNorm belongs to the last group, consistently with the component
intervention. A coordinate is called active here precisely when its
stored second moment is positive. Zero second moments are counted
separately, including unused embedding coordinates.

\begin{table}[htbp]\centering\small
\caption{Native Adam denominator scales at the measured endpoints.
Percentages are active coordinates with
$\sqrt{v/(1-\beta_2^t)}<10^{-8}$, pooled over four seeds within each
parameter group. The norm column is the average over seeds of the
median total gradient norm divided by $\sqrt N$ in the final 256
updates. The final column averages their fractions of clipped updates.
Coordinates and updates do not add independent seed realizations.}
\label{model:tab:dynamics-optimizer-scales}
\input{content/model/generated/dynamics-optimizer-scales.tex}

\end{table}

At $g=2$, the active shared-metric fraction below the offset is at most
0.0118\% in the 2,048-update states and ranges from 61.7\% to 88.1\%
at update 8,192 across the twelve width--seed cases. The offset thus
contributes more than half of the denominator in many of the late
shared coordinates. The corresponding late power-parameter fractions
are only about 2.0--5.5\%, demonstrating distinct optimizer scales
inside the same model.

The analyzer also retains the norms of the stored adaptive coefficient
$\widehat m/(\sqrt{\widehat v}+\epsilon_o)$ and the decay coefficient
$\lambda_d\theta$, together with their coordinatewise comparisons.
At $g=2,t=8,192$, the shared adaptive coefficient norm is still
44.4--115.6 times its decay coefficient norm across the twelve cases.
Only 24.7--48.6\% of active shared coordinates have smaller adaptive
than decay magnitude; the corresponding power-group norm ratios are
345--421. The late shared group therefore retains substantial adaptive
updates within a heterogeneous coordinate distribution. These diagnose
stored scales; the next update additionally depends on its new gradient.
A small denominator alone does not imply that a complete parameter
group follows pure decay. Proposition~\ref{model:prop:adam-decay-sector}
instead supplies a pathwise sufficient condition, including its
dependence on horizon and offset. The finite endpoint fractions and
clipping norms do not verify its uniform gradient hypotheses as
$N\to\infty$. Equation~\eqref{model:eq:adam-offset-limit} identifies why
the offset convention must remain part of any claimed limiting
training law, including one inferred from finite-width fluctuations.

At $g=1$, the fraction of active shared coordinates below the
Adam offset increases in every one of the twenty paired continuations.
The range across width--seed cases is $0$--$16.2\%$ at update 8,192
and $16.4$--$53.7\%$ at update 16,384. The later power-group fractions
are $1.24$--$6.60\%$. Thus the denominator scales continue to evolve
through the horizon at which the row mean and susceptibility decrease.

The corresponding late shared adaptive-to-decay norm ratios remain
$84.1$--$223$, while only $0.993$--$23.8\%$ of active shared
coordinates have smaller adaptive than decay magnitude.
The late power ratios are $426$--$561$ and those of the remaining
parameters are $39.3$--$64.4$. The slower-rate family therefore also
retains substantial adaptive coefficients after horizon doubling.
These simultaneous changes in row observations and optimizer scales
resolve the measured state of the native update; they do not isolate
a single causal term or establish its uniform head-count limit.

\subsection{Causal influence and compatibility of the shared metric learner}
\label{model:sec:shared-crossing-results}
We cross the trained shared metric learner with the remaining recipient
parameters in all $4\times4$ combinations at $N=4,g=2$, separately at
2,048 and 8,192 updates. These are full native float32 inference calls
on 64 fixed contexts. The shared component comprises the eight residual
metric units in each decoder. The recipient contains the remaining
parameters, including the input LayerNorm. Every diagonal combination reproduces its
recipient's fields and logits exactly. The sixteen combinations define
the finite product law of Proposition~\ref{model:prop:crossed-collective};
they are not sixteen independent training outcomes.

\begin{table}[htbp]\centering\small
\caption{Finite product-law covariance decomposition under shared-metric
transplantation. Recipient, donor and interaction components are orthogonal at each
fixed context. Displayed fractions divide each averaged component by
the averaged crossed variance.
The factual diagonal variance is retained separately.}
\label{model:tab:dynamics-crossing}
\input{content/model/generated/dynamics-crossing.tex}

\end{table}

At the early checkpoint, the donor contribution accounts for about
93\% of crossed row-energy variance, whereas the recipient contribution
accounts for about 93\% of crossed entropy variance. These fields
therefore select different components of the joint trained state.
At the later checkpoint, the row decomposition contains about 42\%
donor and 44\% interaction variance. The factual diagonal row variance
is much smaller than the crossed variance. This identifies learned
compatibility between components of the concentrated regime.
Off-diagonal predictive KL remains measurable at both horizons, so
small factual row energy does not establish interchangeability of the
shared learner in the full predictive graph. Conditioning on its initial
value did not remove its learned causal role.

\subsection{Response to finite shared-state displacement}
Four late $N=4,g=2$ parents support four branches each. The base branch
continues the recorded state. A second branch replaces all shared metric
weights by those of the same trajectory at update 2,048, retaining the
late Adam state. A third also replaces their first and second moments,
while retaining the current counters. A fourth keeps the transplanted
shared weights fixed by restoring them after each ordinary optimizer
step. Its moment updates and full-gradient clipping remain active.
All branches use the same subsequent 4,096 minibatches. Complete 32,000-token logits are retained at the initial and final
observations. Each branch is compared to the ordinary continuation
at the same context, and the saved weights and both moment arrays
provide separate endpoint state distances. The interventions
are finite augmented-state changes under the paired-kernel construction,
not infinitesimal sources. The frozen branch uses its stated controlled
update kernel; its marginal dynamics differ from the ordinary branch.

\begin{table}[htbp]\centering\small
\setlength{\tabcolsep}{2pt}
\caption{Return of the row field after finite shared-state interventions.
Each reported displacement is the RMS paired difference from the base
trajectory over 64 fixed contexts and five decoders, divided by its own
initial RMS. Four parent identities are retained separately in the
complete analysis.}
\label{model:tab:dynamics-return}
\input{content/model/generated/dynamics-return.tex}

\end{table}

\begin{figure}[htbp]\centering
\includegraphics[width=\textwidth]{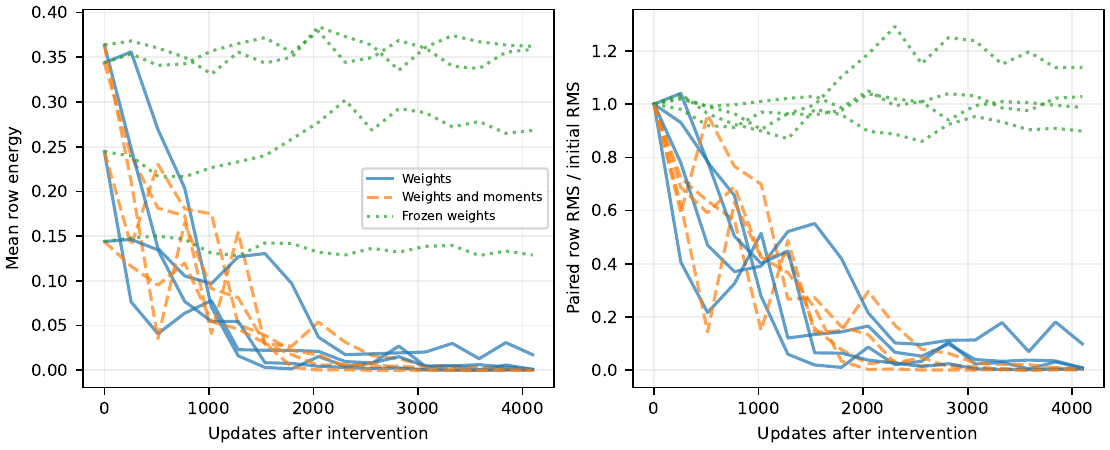}
\caption{Native finite component-return trajectories. Solid, dashed
and dotted curves distinguish weight replacement, weight-and-moment
replacement, and the frozen shared-weight intervention. Different
curves within a branch type represent the four parent initializations.}
\label{model:fig:dynamics-return}
\end{figure}

\begin{table}[htbp]\centering\small
\caption{Predictive and shared-state displacement in the same return
branches. Each entry is the maximum over four parent identities.
$K_0$ and $K_T$ are context-mean predictive KL from the paired base to
the intervention, before and after 4,096 updates; $K_T^{\max}$ is the
largest endpoint context KL. The three $d$ columns are Euclidean
endpoint displacements of the shared weights and moments, divided by
the norm of the corresponding base coordinate. Complete per-parent
values and the other parameter groups remain in the bound analysis.}
\label{model:tab:dynamics-return-emission}
\input{content/model/generated/dynamics-return-emission.tex}

\end{table}

After 4,096 updates, the relative row RMS displacement is
$0.00502$--$0.0993$ for shared-weight replacement and
$0.00119$--$0.00725$ when the early moments are replaced as well.
It remains $0.898$--$1.137$ in the frozen-weight controls.
Both adapting branches therefore approach the concentrated row
regime in every parent, whereas the fixed shared weights largely
preserve the displacement despite continued adaptation of the
remaining model. Replacing the moments produces the smaller endpoint
row displacement in all four matched parents. These finite paired
comparisons identify a role for the evolving shared learner and its
optimizer state. They do not separately identify the contributions
of its adaptive and decay terms.

Recovery of this internal field does not restore the paired predictive
law. Every intervened branch has larger context-mean predictive KL
at the endpoint than at its initial displacement: the initial means
range from $0.00671$ to $0.0149$, whereas the final means over all
three intervention types range from $0.106$ to $0.242$.
The entropy RMS displacement also exceeds its initial value in
every intervened branch. In the two adapting branch types, the
endpoint shared-weight distance is $0.375$--$0.428$ of the base
weight norm; both Adam moment coordinates remain distinct as well.
The paired NLL changes have both signs. Thus row concentration selects
an observed regime containing distinct predictive and augmented
states, rather than demonstrating return to one complete trajectory.

The measured adaptation occurs at the unchanged external training
rate and provides finite basin evidence for this row regime.
An endogenous critical attraction would additionally require the
independently identified relevant coordinate and critical-window
concentration of Proposition~\ref{model:prop:feedback}. The observed
row recovery and its frozen control supply neither that critical
surface nor a thermodynamic scaling law.

\subsection{Local row contraction and retained nonlinear transport}
\label{model:sec:row-transport-results}
Thirty-two checkpoint diagnostics measure all eight metric units in
every decoder on four fixed contexts. At $N=4$, all four initialization
identities are observed at updates 0, 2,048, 4,096 and 8,192. At
$N=8,14$, the four identities are observed at 2,048 and 8,192 updates.
The training rate is $g=2$. Native float32 execution supplies each actual
input row cloud; the individual unit is then evaluated and differentiated
in float64. The recorded quantities include its full $64\times64$
centroid Jacobian, centered linear image, nonlinear residual, centroid
defect and signed covariance cross term. Five radial contractions of
the input deviations, from $0.1$ to $0.001$, check the local derivative
against finite changes.

\begin{table}[htbp]\centering\small
\caption{Transport through the final internal metric unit, pooled over
five decoders and four fixed contexts at each of four initializations.
The norm columns summarize the centroid Jacobian. The empirical gain
is the square root of output divided by input row variance. The last
two columns bound the range across seeds of total nonlinear residual
energy divided by total output row variance. This ratio is a variance
budget with signed cross terms, and can exceed one.}
\label{model:tab:dynamics-row-transport}
\input{content/model/generated/dynamics-row-transport.tex}

\end{table}

At initialization the $N=4$ median centroid norm is about 10.2, while
the empirical row-variance gain is about one. At update 8,192 the
maximum centroid norms are below 0.397 at all three widths; every
sampled final-unit centroid therefore belongs to the local sector in
Corollary~\ref{model:cor:metric-row-contraction}. The median empirical gains
are approximately 0.015--0.016. This identifies a learned local
row-concentrating mechanism inside the metric graph, with its effect
measured at frozen inference states.

The full covariance budget is essential. At the late checkpoint the
ratio of aggregate nonlinear residual energy to aggregate output
variance ranges from 0.00137 to 0.909 at $N=4$, from 0.377 to 1.07 at
$N=8$, and from $6.63\times10^{-5}$ to 0.695 at $N=14$.
Small local gains and small typical residuals can coexist with rare row
excursions that carry much of the variance. The linear term alone is
therefore not a uniform covariance closure, even in the late regime.
Equation~\eqref{model:eq:metric-row-covariance} retains exactly the terms that
resolve this distinction. These finite unit-depth measurements are
separate from a head-count limit and from an eigenmode of training.

\subsection{Full predictive reduction with exceptional rows retained}
\label{model:sec:row-projection-results}
The row maps of Proposition~\ref{model:prop:exception-projection} are tested
in 34 complete-graph cases: the same 32 controlled checkpoints and
both released models, each on 64 fixed contexts. Six variants per case
compare the original graph; terminal metric projections retaining
zero, one or four exceptional rows; and two initial-centroid variants.
The latter either broadcast the centroid through all eight units or
propagate it as a single row and broadcast only at the output. Every
variant recomputes all later attention, residual and decoder operations
and the full vocabulary distribution. Thus a projection in one decoder
can change the inputs of later metric learners. Its retained indices
and matrix error are stored at the actual projected input.

\begin{table}[htbp]\centering\small
\caption{Full-graph four-exception reduction of the controlled models.
The first two error columns are predictive KL over four seeds and
64 contexts. The susceptibility ratio uses the same four initialization
identities before and after reduction, centered at fixed context.
The last column is the relative sample-error bound from
Proposition~\ref{model:prop:susceptibility-transfer}. It is a finite-ensemble
bound, not uncertainty for a population limit.}
\label{model:tab:dynamics-row-projection}
\input{content/model/generated/dynamics-row-projection.tex}

\end{table}

At 8,192 updates, retaining four exceptional rows and the complement
centroid gives row-susceptibility ratios of approximately 0.910, 1.003
and 0.994 at $N=4,8,14$, respectively. The corresponding relative
error bounds are 14.7\%, 3.88\% and 2.46\%. The mean predictive KL
is below $1.83\times10^{-7}$ at each width. These observations support
a compact inference state that preserves the measured late row
fluctuations as well as predictions. Its validity is state dependent:
at 2,048 updates the same four-exception reduction retains only about
6\% of the row susceptibility, and at initialization its redistribution
of an almost uniformly large row field amplifies a small seed variance.
The complete controlled range in Table~\ref{model:tab:dynamics-row-projection}
therefore identifies the approximation's scope across training.

The centroid-only projection has small late predictive error but removes
the row susceptibility down to native arithmetic residuals. One retained
exception recovers about 42\%, 63\% and 86\% of late susceptibility
at the three widths. The additional retained rows improve fluctuation
fidelity; decreasing the matrix projection error does not require
monotone predictive KL after nonlinear graph transport.

\begin{table}[htbp]\centering\small
\caption{Centroid reductions of the two released models on 64 contexts.
Terminal projection averages the completed metric; initial compression
propagates a single centroid through its eight internal units.
The unprojected row field is summarized over $64\times5\times14$
context--decoder--head entries; KL refers to the full native prediction.
The two checkpoints are individual frozen laws, not an ensemble for
estimating a trained-model susceptibility.}
\label{model:tab:dynamics-row-released}
\input{content/model/generated/dynamics-row-released.tex}

\end{table}

The unprojected normalized row field has mean
$2.05\times10^{-7}$, median $1.31\times10^{-7}$ and maximum
$2.11\times10^{-6}$ in the first model. The corresponding values
in the second are $2.24\times10^{-14}$, $1.39\times10^{-14}$ and
$1.91\times10^{-13}$. Each cohort contains 64 contexts, five decoders
and fourteen heads. Thus these released-checkpoint reductions test
inference consistency in an already concentrated regime. Controlled
initial and intermediate states supply the separate nontrivial
row-contrast tests. For both released models the terminal-centroid maximum KL is below
$2.77\times10^{-11}$. Initial-centroid compression has maximum KL
$6.00\times10^{-8}$ in the first model and
$2.05\times10^{-12}$ in the second. The single-row and broadcast
implementations give identical recorded logits in 26 of the 34 cases.
The eight $N=14$ cases have native arithmetic differences, with maximum
paired KL $1.80\times10^{-10}$ across their contexts. Thus their exact
smooth-graph identity and measured native implementation accuracy remain
separate statements. The original-to-reduced comparisons measure the
predictive effect of replacing the nonlinear centroid flow. They do not make
the original and reduced inference laws exactly equal, transfer a
training-time critical mode, or identify a critical thermodynamic law.

\subsection{Common-row fluctuations after contrast concentration}
\label{model:sec:metric-collective-results}
The unprojected matrices from the matched training-input study resolve
the two sectors of Proposition~\ref{model:prop:metric-collective-sectors}.
At each width and checkpoint, both 32-example minibatches are
combined into one fixed 64-context observation mixture. Covariances
use the same four initialization identities at each context. The
native float32 matrices are centered and reduced in float64; these
are observations of stored tensors, not exact-face certificates for
the smooth underlying graph.

\begin{table}[htbp]\centering\small
\caption{Absolute row energy and orthogonal metric collective sectors
at the two fixed training minibatches. The physical per-head energy
$E=\|P_dA\|_{\mathrm F}^2$ is averaged over heads, decoders, contexts
and four initializations. The centroid norm is in per-coordinate
mean-square units. Both susceptibilities use the matching units in
Proposition~\ref{model:prop:metric-collective-sectors}, with
$\chi_A=\chi_\mu+\chi_\perp$. Common reports
$100\chi_\mu/\chi_A$. Contexts are conditioned observations rather
than additional model replicas.}
\label{model:tab:dynamics-metric-collectives}
\input{content/model/generated/dynamics-metric-collectives.tex}

\end{table}

Mean physical row energies fall from approximately
1,002, 1,289 and 1,325 to 0.370, 0.813 and 0.0702 at
$N=4,8,14$, respectively. The mean squared centroid coordinate
instead remains close to one at the late states. Thus the declining
normalized row field reflects a large reduction of the absolute
contrast energy on this cohort, while the common row remains active.

The common-row susceptibility increases from 2.92 to 4.03 at $N=4$,
from 4.78 to 8.21 at $N=8$, and from 8.79 to 13.5 at $N=14$.
Its share of the head-averaged metric covariance rises from 88.4--90.6\%
to more than 99.997\% at every measured late width.
The relative operator-norm bound in
Equation~\eqref{model:eq:metric-collective-transfer} is consequently below
0.0100 at all three late widths. This is a quantitative covariance
reduction on the measured finite ensemble; it does not identify a
width-limit exponent.

Four centered initialization vectors have rank at most three.
The measured effective ranks 2.95--2.99 approach this sample ceiling;
they do not identify an intrinsic rank-three collective manifold or
a physical count of critical modes. The common field remains
endogenous under the fixed shared initialization and batch history,
and its numerical susceptibility is stated in the declared native
parameter frame. The layer-sign symmetry and the learned downstream
PLGA coefficients must be retained before assigning a predictive
or gauge-invariant physical meaning to its fluctuations.

\subsection{Loss-adjoint transport at matched training inputs}
\label{model:sec:adjoint-results}
The backward measurement retains the complete next-token loss graph.
At all four $g=2$ identities and $N=4,8,14$, it crosses the states at
updates 2,048 and 8,192 with the same two training minibatches. Each
batch has 32 examples. These are 48 gradient measurements at 24
checkpoint states, with twelve width--initialization identities.
The native CPU float32 forward and backward passes record the input
and output adjoints of all forty metric units. A separate smooth64
vector--Jacobian product at the actual native rows holds the native
output adjoint fixed.

For internal unit $j$, define the measured gain
\[
 a_j=\left(
 \frac{\sum_{\ell,b,h,i}\|\lambda_{\ell,b,h,i}^{(j)}\|^2}
 {\sum_{\ell,b,h,i}\|\lambda_{\ell,b,h,i}^{(j+1)}\|^2}
 \right)^{1/2},\qquad A=\prod_{j=0}^7a_j.
\]
The sum includes every decoder, example, head and row. The product
telescopes to the full metric-block input/output adjoint norm ratio.
It is a source-weighted gain, not a Jacobian operator norm. The unit
parameter-gradient norms and their fractions of the full unclipped
gradient are retained separately.

\begin{table}[htbp]\centering\small
\caption{Native loss-adjoint transport with the minibatch fixed across
checkpoint times. Batch identifies the zero-based sampler index.
The gain columns report medians over four initializations. The final
column is the largest paired $A_{8192}/A_{2048}$ across those identities,
computed before aggregation. Each row uses the same token crops at both
times; different batches are repeated measurements of the same states.}
\label{model:tab:dynamics-row-adjoint}
\input{content/model/generated/dynamics-row-adjoint.tex}

\end{table}

\begin{figure}[htbp]\centering
\includegraphics[width=\textwidth]{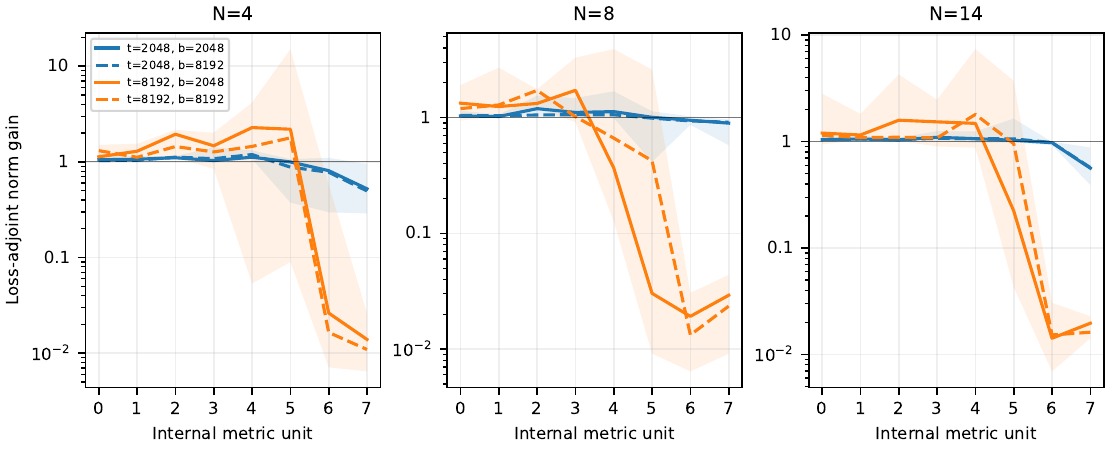}
\caption{Internal loss-adjoint gains across the eight metric units.
Colors identify checkpoint times; solid and dashed curves distinguish
the two fixed training minibatches. Each curve is the median of four
initializations, after aggregating the complete source-weighted norm
within each model. Shaded ranges cover all four identities and both
fixed minibatches at each checkpoint time. The horizontal line marks
unit gain.}
\label{model:fig:dynamics-row-adjoint}
\end{figure}

The final-unit gain decreases in all 24 paired state comparisons.
It ranges from 0.288--0.962 at the early states and
0.00645--0.0437 at the late states; the largest paired late/early ratio
is 0.0630. The complete block retains stronger source dependence.
Twenty-three late block gains are below one, while the remaining gain
is 2.04 at $N=4$, identity 640104, on minibatch 2,048. Its early-state
gain on the same batch is 0.306, giving a paired ratio of 6.66.
In that late case, internal unit five has gain 15.1 while unit seven
has gain 0.0103. The ordered block therefore amplifies that source
despite strong attenuation in the final unit. The late block medians
remain much smaller than the early medians at every width and batch.
These are finite measured distributions of gains, with the complete
input-dependent range retained.

The largest local smooth64/native adjoint relative discrepancy is
$8.06\times10^{-5}$ on the resolved coordinates, using the native output
adjoint as the fixed source. The unit's parameter gradient has its own
scale. For example, the late final-unit parameter gradients across all
five decoders account for 98.1\% of the complete gradient norm at
$N=4$, identity 640101, on minibatch 8,192, despite a final-unit input
adjoint gain of 0.0147. Attenuating the input adjoint consequently does
not eliminate adaptive learning of the unit itself.

Proposition~\ref{model:prop:metric-row-adjoint} relates these measured gains
to the actual loss source. Forward row concentration and a centroid
Jacobian alone do not determine them. The repeated minibatches isolate
a checkpoint-state comparison while retaining each input's own loss
adjoint; they do not identify a stationary training spectrum or a
population rate of critical attraction.

\subsection{Training-force fidelity of the inference reduction}
\label{model:sec:gradient-projection-results}
The same 24 checkpoint states and two matched minibatches also support
a direct backward comparison of the terminal row reductions.
At every decoder, the four-exception projection or the centroid
projection is inserted before the complete learned PLGA transform.
Each branch retains its actual exception indices. The full next-token
loss is differentiated through the reduced graph and the native
global norm clipping includes every model parameter. This gives 48
matched sources and 96 nonidentity reduction comparisons.
The archived base loss and logits are reproduced, and a 64-row
identity projection preserves the full stored shared-gradient vectors
before and after clipping exactly.

Write $\Gamma,\widetilde\Gamma_k$ for the complete shared metric
gradient in the base and $k$-exception branches, and
$c,\widetilde c_k$ for their native global clipping factors. We report
\[
 e_k=\frac{\|\widetilde\Gamma_k-\Gamma\|}{\|\Gamma\|},\qquad
 e_k^{\rm clip}=
 \frac{\|\widetilde c_k\widetilde\Gamma_k-c\Gamma\|}
      {\|c\Gamma\|}.
\]
The same raw records retain absolute gradient norms, differences and
cosines. Predictive KL and force error use the same checkpoint and
tokens, so their comparison does not confound a change of minibatch.

\begin{table}[htbp]\centering\small
\caption{Predictive and shared-force fidelity under terminal metric
reduction. Each row covers four initializations and both fixed
minibatches. $K_4$ is the largest mean predictive KL for four retained
exceptions. Error columns are maxima of the paired relative norms;
the cosine column is the smallest clipped-gradient cosine.
The zero-exception column uses the same eight source conditions.}
\label{model:tab:dynamics-row-gradient-projection}
\input{content/model/generated/dynamics-row-gradient-projection.tex}

\end{table}

At the early states, four-exception clipped-gradient errors can
exceed one and the gradient cosine can be negative. At the late
states the largest four-exception errors are 0.403, 0.0172 and 0.247
at $N=4,8,14$, respectively. In the $N=8$ group, the retained
four rows therefore approximate the shared clipped force within
1.72\% on all eight measured sources. The same statement does not
hold uniformly over the other widths.

A concrete source separates the two fidelities. At the late
$N=4$ state with identity 640104 and minibatch 2,048, the mean
predictive KL is $2.29\times10^{-8}$ but the clipped shared-force
relative error is 0.403. Its absolute force difference is 0.00283, against a base shared-force
norm of 0.00701. The full gradient norm changes from 106 to 73.3;
the resulting clipping change takes a raw shared-force error of
0.308 to the clipped error of 0.403.
This is the same state and batch whose complete backward metric
block has gain 2.04. The two measurements are consistent with
input-dependent transport, without isolating that block gain as
the sole cause of the force discrepancy.

Centroid-only clipping errors reach 1.51, 0.887 and 0.275 at the
three late widths. Thus a successful prediction-level centroid
approximation cannot be promoted automatically to an equivalent
shared update. Equation~\eqref{model:eq:shared-gradient-closure} retains
the parameter Jacobian and loss source, and
Equation~\eqref{model:eq:clipped-shared-gradient-error} additionally
retains the full-model clipping factor. The measured reduced
forces test these requirements locally. Exception indices are
fixed within the executed backward branch; a tied native selection
does not establish differentiability of a different continuously
reselected rule. None of these finite comparisons establishes
autonomous reduced training closure.

\subsection{Augmented response and its arithmetic window}
\label{model:sec:augmented-response-results}
The full-graph control differentiates the next batch loss and applies
Equation~\eqref{model:eq:adam-tangent} with both moment directions and the
derivative of the clipping factor. The two physical weight directions
are a common operator gain, implemented through the final PLGA affine
coefficients, and radial scaling of all shared metric weights. Incoming
moment directions are zero in these physical pulses; their outgoing
directions are computed and checked. The initial and final emissions
include the full 32,000-token logits on four fixed evaluation contexts.

There are 24 checkpoint--direction controls: four initialization
identities at $N=4$, one at each of $N=8,14$, two matched horizons and
both directions. Each uses the original next minibatch of 32 training
examples. Independent symmetric pulses test a decreasing amplitude
grid. A local window requires relative errors below 5\% for weights,
both moments and predictive Fisher response at two neighboring tested
amplitudes. The complete grids, including further resolution of five
early shared-metric directions, are retained. Reported windows concern
the sampled amplitudes; no uniform interpolation bound between them
is measured.

\begin{table}[htbp]\centering\small
\caption{Full-state and predictive response resolution for the expanded
smooth64 program. A window requires two adjacent amplitudes passing
all four error criteria. The two amplitude columns give the minimum
and maximum largest qualifying upper amplitude across the stated
initializations. Gain columns are the ratio of the summed predictive
source curvatures after and before one update. Numerical controls at a single trained
initialization do not estimate a population response distribution.}
\label{model:tab:dynamics-tangent}
\input{content/model/generated/dynamics-tangent.tex}

\end{table}

\begin{table}[htbp]\centering\small
\caption{Independent native CPU float32 and expanded float64
comparisons before and after one actual Adam update, using identical
initial stored states and minibatch identities. Maxima are over the
four fixed evaluation contexts and the stated initialization identities.}
\label{model:tab:dynamics-arithmetic}
\input{content/model/generated/dynamics-arithmetic.tex}

\end{table}

The declared update formula also agrees with an independent native
\texttt{torch.optim.AdamW} step on the same double-precision graph,
with maximum parameter discrepancy \ModelSourceDynamicsAdamPrimalError{} and
maximum moment discrepancy \ModelSourceDynamicsAdamMomentError{}.
The twelve native-arithmetic comparisons separately measure the effect
of float32 execution. Their initial predictive differences can increase
after an update, particularly at the early checkpoint. A resolved
derivative window for the smooth64 reference therefore does not certify
the same window for native float32 perturbations.
The paired native finite interventions above have their own arithmetic
and amplitude regime. Neither a large one-step curvature gain nor an
arithmetic amplification identifies a stationary critical mode.

All \ModelSourceDynamicsResolvedTangentCases{} primary controls have two adjacent
passing amplitudes after the recorded resolution extensions. Each
requires agreement of both moment directions and the parameter
direction as well as the predictive Fisher response. Their joint
validation checks the complete local update transport on the sampled
states; it does not identify a stationary response spectrum.

The largest qualifying upper amplitudes are at or below $10^{-7}$ in
15 of 24 controls and at or below $10^{-8}$ in nine. The narrow windows
reach from $10^{-7}$ down to $3\times10^{-10}$. A source amplitude $e$
is the relative Euclidean displacement of its active weight group:
all final PLGA affine coefficients for the gain source, or all shared
residual metric weights for the radial source. Incoming moments are
held fixed. These are local derivative controls in that metric; their
small amplitudes are not a model of a complete finite optimizer update.

\section{Paired horizon and arithmetic observations}
\label{model:sec:observation-results}

The observation bounds distinguish three claims: concentration of a
row field, resolution of its conditional second moment, and visibility
in predictions. The following completed measurements keep those claims
in their declared units and retain every selected condition.

\subsection{Whole-seed uncertainty after horizon doubling}
Table~\ref{model:tab:observation-horizon} reconstructs the paired susceptibility
difference from all endpoint fields, preserving each seed's two
horizons. The first five rows use the four original identities at each
width. The $N=14$ extension continues the remaining twelve existing
initializations with their complete optimizer state, shared generator,
and common batch realization, giving sixteen matched identities.
It adds 98,304 updates across the twelve continuations and no new
initialization identities. The additional-twelve result is also shown
separately because the choice of width followed the unresolved
four-seed comparison.

\begin{table}[htbp]\centering\small
\caption{Paired row susceptibility at 8,192 and 16,384 updates, averaged
over 512 fixed contexts and five decoders. The change is late minus
early. SE is the delete-one-seed jackknife standard error of that
change. Percentiles use all 256 ordered empirical resamples for four
seeds and 20,000 paired resamples for the larger panels. The last two
rows are the full sixteen-seed panel and its additional twelve seeds.
These are finite empirical uncertainty diagnostics.}
\label{model:tab:observation-horizon}
\input{content/model/generated/observation-horizon.tex}

\end{table}

For all sixteen $N=14$ identities, the susceptibility changes from
\ModelSourceObservationHorizonBefore\ to \ModelSourceObservationHorizonAfter, a difference
of \ModelSourceObservationHorizonDifference\ with jackknife standard error
\ModelSourceObservationHorizonSE\ and empirical percentiles
$[\ModelSourceObservationHorizonLow,\ModelSourceObservationHorizonHigh]$.
The additional twelve identities give difference
\ModelSourceObservationAdditionalDifference\ and percentiles
$[\ModelSourceObservationAdditionalLow,\ModelSourceObservationAdditionalHigh]$.
The ensemble mean row field changes from
\ModelSourceObservationHorizonMeanBefore\ to \ModelSourceObservationHorizonMeanAfter;
\ModelSourceObservationMeanDecreases\ of the sixteen individual mean fields
decrease. The complete analysis retains each seed's row change,
leave-one-out susceptibility, NLL and predictive context dispersion.

\begin{table}[htbp]\centering\small
\caption{Predictive emissions and mean row fields in the same paired
horizon panels. NLL averages 512 held-out targets per seed;
$I$ is full-vocabulary context dispersion over 256 fixed context pairs
per seed. Columns are initialization means. The row observation and
predictive observables retain separate units.}
\label{model:tab:observation-prediction}
\input{content/model/generated/observation-prediction.tex}

\end{table}

For the sixteen-seed panel, mean held-out NLL changes from
\ModelSourceObservationHorizonNllBefore\ to \ModelSourceObservationHorizonNllAfter,
and predictive context dispersion changes from
\ModelSourceObservationHorizonContextBefore\ to \ModelSourceObservationHorizonContextAfter.
These emissions are observed on the same trained states as the row
statistics; they are not determined by a row-concentration threshold.

Both enlarged panels have negative empirical percentile intervals and
negative leave-one-out changes. They support a finite conditional
horizon effect beyond the unresolved four-seed comparison. All sixteen
mean row fields decrease, while predictive context dispersion stays
near $0.5$. Thus substantial row concentration coexists with an active
context-dependent predictive law in this lower-rate family.

These paired observations do not establish monotonic population
susceptibility at every time, relaxation to an invariant law, or a
critical size--time scaling function. The conditional covariance flow
retains nonlinear sources and changing learned common coordinates;
it permits the strong movement of the measured second moment without
assigning a stationary critical exponent.

\begin{figure}[htbp]\centering
\includegraphics[width=\textwidth]{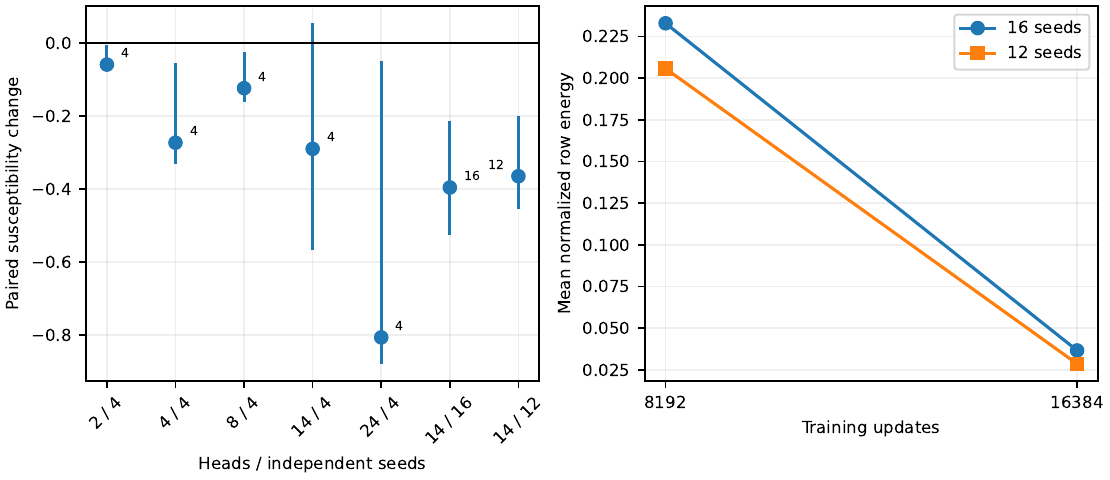}
\caption{Paired whole-seed susceptibility changes with empirical
percentile intervals, and mean row energy for the sixteen-seed and
additional-twelve panels. The horizontal zero line concerns a
susceptibility difference, not the value of the row field. The larger
panels share twelve trajectories and are not independent studies.}
\label{model:fig:observation-horizon}
\end{figure}

\subsection{Complete-cohort arithmetic comparison}
Thirteen frozen-weight conditions were evaluated on the same four
seeds and 512 contexts per condition. Nine cover the complete
$N=4,8,14$, $g=2,3,4$ grid at update 8,192; four cover
$N=2,4,8,14$, $g=16$ at update 2,048. Every matrix retains both its
absolute centered energy $E$ and total energy $T$, in fixed
mean-square-entry units:
\[
 E=\|P_dA\|_{\rm F}^2/d^2,\qquad T=\|A\|_{\rm F}^2/d^2,
 \qquad d=64.
\]
The unnormalized physical energy in
Section~\ref{model:sec:row-correspondence} is $d^2E$.
Native CPU float32
and CPU float64 forward calls both accumulate these statistics in
float64. The $N=4,g=4$ endpoint is included as a repeated numerical
reference; these evaluations add no independent training identities.
The archived CUDA float32 row field supplies the paired
reference for the published native susceptibility. Consequently this
comparison measures the joint sensitivity to forward arithmetic,
platform and statistic accumulation. The CPU float32 control and its
paired full-vocabulary comparison with CPU float64 are retained
separately; the latter does not certify the archived CUDA prediction.

\begin{table}[htbp]\centering\small
\caption{All paired arithmetic conditions. $\chi_{32}$ is the archived
native row susceptibility and $\chi_{64}$ is its frozen-weight CPU
float64 counterpart. The signed relative change and the relative
bound from Equation~\eqref{model:eq:sample-observation} are percentages.
Screen is the percentage of native row entries below $10^{-13}$.
A meets both declared bound tolerances; Q requires qualification.
Neither flag is a thermodynamic classification or an exact-arithmetic
certificate.}
\label{model:tab:observation-arithmetic}
\input{content/model/generated/observation-arithmetic.tex}

\end{table}

\ModelSourceObservationStableCells\ of thirteen conditions meet the declared
bound criterion. At $N=4,g=4,t=8192$, 88.2\% of native entries lie
below the primary screen, yet the susceptibility changes by only
$-0.0302\%$, with a relative bound of $0.0305\%$.
The discrepancy susceptibility is $7.6065\times10^{-17}$ and the
absolute bound is $9.9823\times10^{-13}$. Removing all entries below
the screen changes the native susceptibility by a relative
$3.08\times10^{-12}$. Thus entry concentration and aggregate
resolution coexist in this condition, as allowed by the covariance
decomposition in Section~\ref{model:sec:observation-theory}.

A conservative discrepancy bound can exceed the observed change.
At $N=8,g=2,t=8192$, the actual relative change is $0.0946\%$,
whereas the relative bound is $9.54\%$; this cell is Q under the
fixed rule. At $N=14,g=3,t=8192$, the corresponding magnitudes are
$0.353\%$ and $6.15\%$. These flags do not imply that either point
estimate is made solely of rounding residuals.

The high-rate conditions illustrate a different limitation.
At $N=8,g=16$, the native susceptibility
$3.6087\times10^{-16}$ becomes $7.6642\times10^{-16}$,
a relative change of $112.38\%$. At $N=14,g=16$, its relative
change is $-2.50\%$ and the conservative bound is $81.9\%$.
The $N=2,g=16$ cell also falls outside the relative criterion;
the $N=4,g=16$ cell meets it despite 99.8\% screened entries.
All remain in the tables and evidence object. Their tiny row
susceptibilities do not support quantitative critical scaling fits.

\begin{table}[htbp]\centering\small
\caption{Absolute CPU float64 centered and total metric energies,
and mean row ratios before and after the paired arithmetic change.
Energy and ratio means average the same seed--context--decoder--head
entries; the mean ratio is not the ratio of the energy means.
Maximum KL is over the 2,048 matched seed--context CPU predictions
per condition, comparing CPU float32 with CPU float64.}
\label{model:tab:observation-energies}
\input{content/model/generated/observation-energies.tex}

\end{table}

\begin{figure}[htbp]\centering
\includegraphics[width=\textwidth]{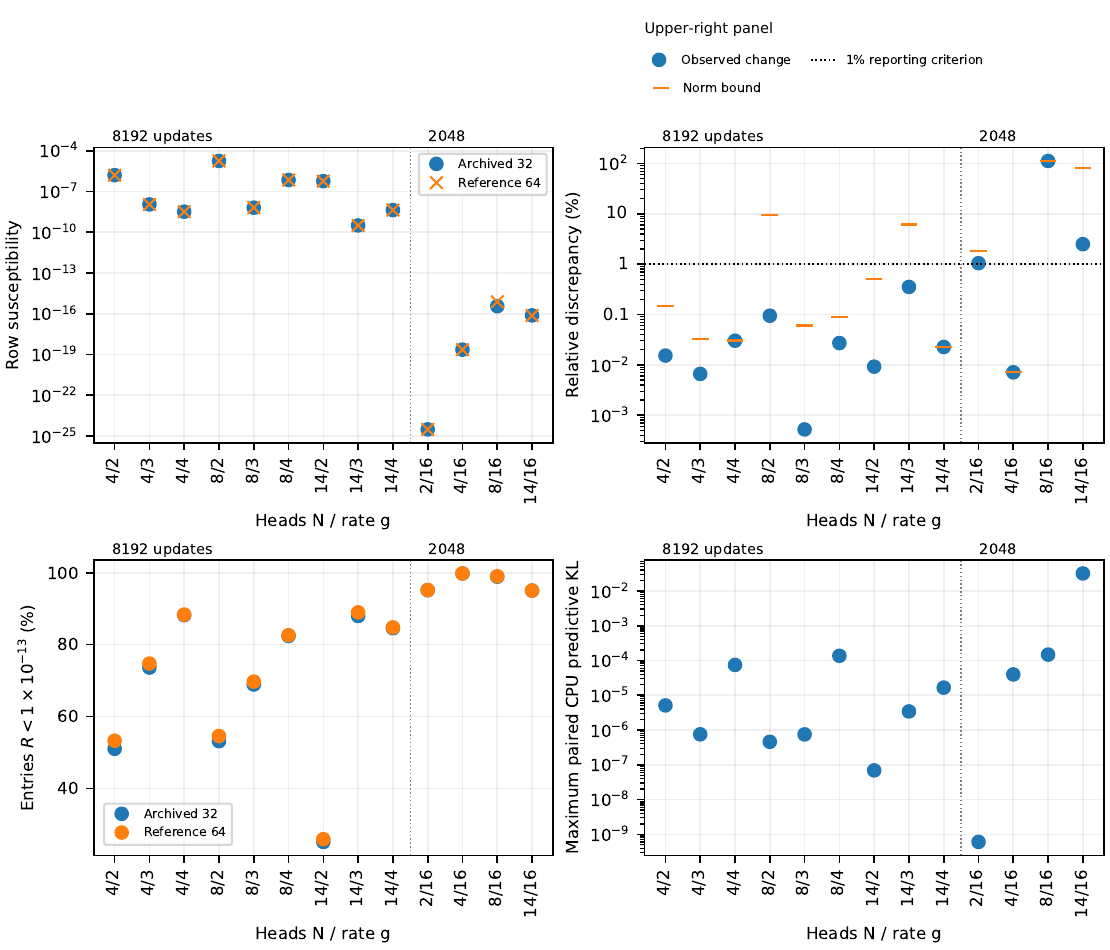}
\caption{The thirteen complete arithmetic cells in table order.
The first nine cells use 8,192 updates and the last four use 2,048.
The panels show paired susceptibility, relative discrepancy and its
bound, entry-screen fractions, and maximum paired CPU predictive KL.
The 1\% line is a declared relative reporting criterion; the absolute
criterion is applied as well. The row screen acts on $R$ and is
never drawn as a threshold on susceptibility.}
\label{model:fig:observation-arithmetic}
\end{figure}

Across all 1,085,440 measured context--decoder--head matrices, the
minimum CPU float64 total energy is $0.8687$. The fixed denominator
floor is therefore inactive throughout this panel; the small ratios
come with small absolute centered energies.

Across the nine late rate-grid conditions, maximum paired CPU
predictive KL is at most $1.361\times10^{-4}$. The high-rate
$N=14,g=16$ condition reaches $0.0318$. This difference in scale
is another reason to observe emission directly rather than infer
its accuracy from the magnitude of an internal row field.
The arithmetic observations support the finite norm-transfer theory
and delimit its quantitative use. They supply no certified matrix
error bound, stationary trained limit or universal rounding floor.

\section{Native size--time laws and visible stochastic sources}
\label{model:sec:scaling-results}

\subsection{The executed family and its random units}
The size--time study contains \ModelSourceScalingTrainingArtifacts\ additional native
training artifacts and \ModelSourceScalingAddedUpdates\ additional optimizer updates,
with a maximum completed horizon of \ModelSourceScalingMaximumHorizon.
An artifact can continue an existing initialization identity; it is not
necessarily an independent training replicate. Head counts are
$N=2,4,8,14,24$, with five decoders, head dimension 64 and shared
metric-network width 170. The variance-corrected initialization fixes
the reference at $N=2$. At that reference the two initialization labels
give the same complete initial parameter digest. Generator and body
learning rates are respectively $3\times10^{-4}g$ and
$6\times10^{-4}/N$. They remain constant throughout each trajectory,
without warmup or annealing. Native AdamW has
$(\beta_1,\beta_2)=(0.9,0.95)$, denominator offset $10^{-8}$,
weight decay 0.01 and global gradient clipping at one.
All training uses full batches of 32 in float32 with TF32 disabled.
Every continuation retains the full incoming optimizer state, absolute
step, sampler prefix and supervised-target counts.

The specified data law is the uniform crop law on the first 3,072
documents of the bound 4,608-document cohort. Each batch samples
documents and offsets independently with replacement, with offsets
$0,\ldots,448$. A crop has 64 input tokens and one supervised external
target. Consequently one update supplies 32 target observations.
This is an empirical distribution derived from RefinedWeb, not the
entire RefinedWeb population or a loss averaged over every token in
the input. Unless an environment factor is varied explicitly, shared
initialization 640011 and complete batch history 640001 are held fixed.
Initialization labels 640101--640104 define the matched four-identity
panels. The additional labels 640105--640108 test frozen variance
predictions at $N=2,4,8,24$. The sixteen-identity panels at their
available shorter horizons retain their original identities.

There are 224 frozen-state CPU observations. The primary context
cohort is always the same 512 short-cohort inputs; calibration
matrices use a separate fixed set of 64 inputs. Selected later states
also retain complete mean matrices on all 512 evaluation inputs.
Float32 forward programs accumulate matrix energies and covariances
in float64. Paired float64 forward observations use the same saved
float32 parameters. Native dense training observations use sixteen
fixed inputs every 64 updates in the added trajectories. These fixed
contexts, heads, decoders, repeated times and arithmetic realizations
do not increase the number of independent initializations.

\subsection{A finite rate--time clock and its predictions}
For a threshold $R_*$, the first observed passage of the sixteen-context
mean row field below $R_*$ determines a time interval. Uncrossed paths
are right censored. The working accelerated-time model is
\begin{equation}
 \log T_{N,g,i}(R_*)=a_N+c_i-\psi\log g+\sigma Z,
 \qquad Z\sim\mathcal N(0,1).
 \label{model:eq:empirical-rate-clock}
\end{equation}
The likelihood integrates over the observed passage interval; it does
not replace an interval by its midpoint. Width and initialization
effects are retained. The 76 calibration paths use $N=2,4,8,14$ and
the available $g=1,1.5,2,3,4$ states, with four common initialization
labels. Fixed powers one and two and a fitted power are all retained.
The primary threshold is $R_*=0.5$; thresholds 0.8, 0.2 and 0.1
describe sensitivity to the measured stage of concentration.

\begin{table}[htbp]
\centering\small
\caption{Interval-censored finite rate--time fits. NLL denotes negative
working log likelihood. The primary empirical range refits all 256
ordered whole-initialization resamples, with the same labels paired
across widths and rates. It is not a population confidence interval.}
\label{model:tab:scaling-clock-fit}
\begin{tabular}{rrrrrr}
\toprule
$R_*$ & $\widehat\psi$ & Empirical range & NLL$_{1}$ & NLL$_{2}$ & NLL$_{\rm free}$\\
\midrule
0.8 & 1.9996 & -- & 123.22 & 60.27 & 60.27\\
0.5 & 1.9467 & [1.8159, 2.0552] & 199.67 & 118.33 & 117.69\\
0.2 & 1.9610 & -- & 272.09 & 190.10 & 189.30\\
0.1 & 1.8863 & -- & 288.89 & 218.77 & 215.33\\
\bottomrule
\end{tabular}

\end{table}

The primary fitted power is 1.9467, with empirical percentiles
$[1.8159,2.0552]$. The other threshold fits range from 1.8863 to
1.9996. At the primary threshold, the working negative log likelihoods
are 199.67, 118.33 and 117.69 for powers one, two and fitted,
respectively. These values support a near-quadratic finite time clock
on the calibration family. They are not critical exponents. In
particular, the $N=2$ width-specific fit reaches the imposed residual
scale floor 0.02; its point optimum cannot establish a distinct
universality class. Exact interval-feasibility calculations retain
that identification limitation.

At ordinary milestones the clock uses the first sixteen inputs of a
batch-32 observation, replacing a simultaneous dense observation where
both exist. A sensitivity calculation instead prefers available dense
batch-16 observations of those same inputs. None of the 304 threshold
intervals changes; the maximum simultaneous row-cohort-mean difference
is $3.082\times10^{-6}$. Historical states lacking a dense observation
keep their available observation in both conventions. This result
concerns the stated clock functional and does not identify the two
floating-point programs in every field.

Before the target runs began, all 54 path forecasts and three joint
size--time forecasts were frozen. Forecasts and target fields use
the native GPU float32 forward program. The path forecasts interpolate the
complete 512-context $g=1$ field linearly in $\log(1+t)$ and evaluate
it at $g^\psi t$, separately for powers one, two and fitted. Four
identities at each of $N=4,14$ are measured at $g=0.5$ through 32,768
updates. The joint family uses
$g=\sqrt{8/N}$ and $t=1024N$, so $g^2t=8192$ and $2t/N=2048$.
Its inverse-$N$ forecasts use only the $N=2,g=2,t=2048$ and
$N=8,g=1,t=8192$ calibration fields. These forecasts extrapolate
each mean and variance in $1/N$; they do not construct a full
transported joint law. Native decay and Adam memory
are held fixed in these training runs, so the joint family tests a
finite approximation to Equation~\eqref{model:eq:joint-clock-map}; it does
not implement the complete scale transformation of those additional
optimizer variables.

Field RMS errors use the natural coupling with the same nonshared
initialization label and batch history. They do not minimize over
couplings between the conditional laws. A large paired field error
alone therefore does not exclude agreement of marginal laws; mean
and covariance comparisons are retained separately.

The completed half-rate controls give an observable-specific finite
clock. At 32,768 updates the mean row fields are 0.2313 and 0.3867
for $N=4,14$. Their frozen linear predictions are 0.01008 and
0.06083, the quadratic predictions are 0.2552 and 0.3145, and the
fitted-clock predictions are 0.2422 and 0.3010. Across all five
frozen horizons at both widths, the largest absolute quadratic
mean error is 0.07228; the corresponding linear and fitted-clock
errors are 0.4890 and 0.09417. These are finite point-prediction
errors for the mean observation.

The endpoint row susceptibilities are 0.14590 and 0.63277. Quadratic
predictions give 0.27562 and 0.55866, while fitted-clock predictions
give 0.24784 and 0.50845. The paired empirical ranges retain the
uncertainty of these four-identity estimates. Mean accuracy also
differs from paired field accuracy: the endpoint linear and
quadratic field RMS errors are respectively 0.2868 and 0.3541 at
$N=4$, and 0.4073 and 0.2635 at $N=14$. There is no uniform ranking
under those distinct criteria.

Predictive observations retain another time dependence. At 8,192
updates the measured NLL means are 7.3154 and 7.2031, compared with
quadratic-clock predictions 7.4748 and 7.4417. At 32,768 the measured
predictive entropy means are 7.1777 and 6.8347, compared with
7.4338 and 7.3870. Thus the mean row clock supplies one measured
component of the scale flow, while the covariance and full-graph
predictive law require their own transported coordinates and source
couplings. It does not supply a model-wide clock or a thermodynamic
critical exponent.

The joint targets give a further finite mean-clock check. At
$N=4,14,24$, their endpoint mean row fields are
$0.22351,0.36094,0.24035$, respectively. The same-width quadratic
clock predicts $0.25522,0.31446,0.24951$. Across all eight frozen
width--horizon groups in this target family, the largest absolute
quadratic mean error is 0.04648, compared with 0.41462 for the
linear clock and 0.05443 for the fitted clock. Each alternative
uses its original calibration. These point errors concern the mean
row observation and do not bound all fields or their fluctuations.

\begin{figure}[htbp]
\centering
\includegraphics[width=\textwidth]{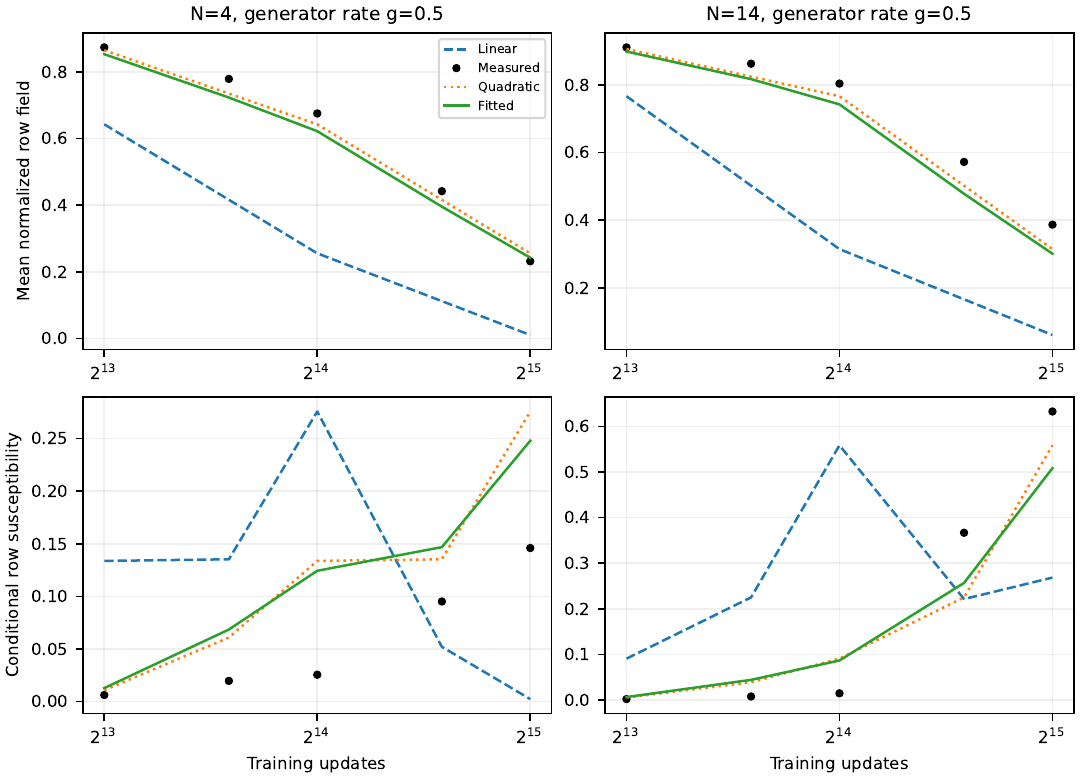}
\caption{Frozen clock predictions and measured $g=0.5$ fields. Every
displayed value uses the same four initialization labels and 512 fixed
contexts. Susceptibility is computed after transporting each seed
field, rather than by interpolating its already aggregated variance.
Lines and markers show point values. Tables~\ref{model:tab:scaling-clock-predictions}
and~\ref{model:tab:scaling-clock-variances} display paired endpoint uncertainty;
every horizon is retained in the numerical evidence.}
\label{model:fig:scaling-clock-predictions}
\end{figure}

\begin{table}[htbp]
\centering\scriptsize
\setlength{\tabcolsep}{3pt}
\caption{Endpoint row-mean prediction errors for all three frozen
clocks. The 32,768-update targets have $g=0.5$; the other endpoints
have $g=\sqrt{8/N}$. Errors are observed minus predicted. Paired
empirical ranges retain each seed's complete reference and target
fields. All intermediate horizons and predictive fields are retained
in the numerical evidence.}
\label{model:tab:scaling-clock-predictions}
\input{content/model/generated/scaling-clock-predictions.tex}

\end{table}

\begin{table}[htbp]
\centering\scriptsize
\setlength{\tabcolsep}{3pt}
\caption{Endpoint row susceptibility forecasts and errors for the
same targets and natural initialization couplings as
Table~\ref{model:tab:scaling-clock-predictions}. Ranges enumerate the
256 ordered whole-initialization empirical resamples and have no
guaranteed population coverage. A small mean error does not imply
a small susceptibility error.}
\label{model:tab:scaling-clock-variances}
\input{content/model/generated/scaling-clock-variances.tex}

\end{table}

\begin{table}[htbp]
\centering\scriptsize
\setlength{\tabcolsep}{3pt}
\caption{The separate frozen inverse-$N$ joint-clock predictions.
Their reference uses two widths; no target outcome is refitted.
The prediction and target retain the same four initialization identities
and fixed observation contexts; the displayed endpoints are finite
conditional comparisons.}
\label{model:tab:scaling-joint-predictions}
\input{content/model/generated/scaling-joint-predictions.tex}

\end{table}

The separate two-width moment extrapolation predicts mean row fields
$0.22329,0.21872,0.21796$ at the three targets. Its row susceptibility
predictions are $0.09568,0.21327,0.33087$, whereas the measured
values are $0.19549,0.67342,1.56191$. The susceptibility point
estimates are thus larger by factors $2.04,3.16,4.72$.
The complete paired four-initialization empirical ranges accompany
these comparisons. Approximate mean-clock agreement does not
identify a common covariance scaling function across widths.

This distinction also appears in predictive units. For the same
two-width extrapolation, the NLL mean errors are
$+0.04424,-0.02613,-0.07483$, while the predictive entropy mean
errors are $+0.49489,+0.63521,+0.39911$. Hence a small mean error
in one predictive observation need not transfer to another.
The resulting finite description retains separate covariance,
predictive and optimizer coordinates. It does not assign a critical
count exponent to these three targets or identify the tested
fixed-memory, fixed-decay family with the complete joint RG orbit.

\subsection{Common coordinates, nonlinear observations and new identities}
\label{model:sec:common-coordinate-results}
Let $C$ be the mean row centroid averaged over heads, in fixed native
column coordinates. Let $V_1$ be the average conditional seed variance
of one head centroid, with the same fixed context and coordinate
weights. The measured effective head count is
\begin{equation}
 N_{\rm eff}=\frac{V_1}{\Var(C)}=\frac{NV_1}{\chi_C}.
 \label{model:eq:measured-effective-heads}
\end{equation}
This ratio does not assume conditional independence or exact
exchangeability. The corresponding mean off-diagonal head covariance
is $(\chi_C-V_1)/(N-1)$. At 16,384 updates the first-four effective
counts are approximately 1.0001, 1.0062, 1.0019, 1.0098 and 1.0098
at $N=2,4,8,14,24$. Thus increasing the nominal head count retains
a strongly coherent common coordinate in this panel. The calibration
mean-matrix susceptibility decomposes exactly into common-row and
row-contrast contributions in mean-square-entry units. No shared
coordinate is removed by conditioning on its learned value.

The susceptibility of $C$ is further split by
Proposition~\ref{model:prop:common-radial-directional}. Radii use the
mean-square norm over the 64 native column coordinates, separately
at every context and decoder. Pair distances give nonnegative radial
and directional terms without subtracting two large second moments.
At 16,384 updates, the directional fractions in the first-four panels
are approximately 0.99964, 0.99927, 0.99877, 0.99015 and 0.99500
for $N=2,4,8,14,24$. Their mean radii are approximately
0.976, 0.969, 0.961, 0.930 and 0.950.
The corresponding mean cosines between directions from different
initializations are $-0.026,-0.017,0.008,-0.009,-0.007$.
Thus heads are coherent within a trained model while the common
direction varies across models. This is a statement in the recorded
coordinate frame, not an identification of a gauge or critical mode.

\begin{table}[htbp]
\centering\scriptsize
\setlength{\tabcolsep}{3pt}
\caption{Radial and directional contributions to the native
common-centroid susceptibility in matched four-identity panels.
Radii and directions are defined within each fixed context and
decoder before averaging. Complete condition, arithmetic and
leave-one-initialization-out results remain in the evidence object.}
\label{model:tab:scaling-common-geometry}
\input{content/model/generated/scaling-common-geometry.tex}

\end{table}

The paired CPU observations at 32,768 updates sharpen this
separation. Their common-centroid susceptibilities are
$1.8404,3.7153,7.2499,12.5523,21.7173$ for
$N=2,4,8,14,24$, respectively. Relative to the same first-four
16,384-update panels, these point estimates change by
$-5.78\%,-2.67\%,-1.35\%,+1.57\%,-1.33\%$.
All five effective head counts lie within $4.8\times10^{-6}$ of one.
The common component accounts for at least 99.9977\% of the
head-averaged metric covariance trace on the full 512-context
evaluation cohort. Thus the measured concentration strongly
suppresses the contrast sector while preserving a coherent common
field across initialization identities. Agreement of these moments
does not establish stationarity of the complete conditional law.

At this horizon the paired float32/float64 bounds on row susceptibility
are, in the same width order, approximately
$0.000139\%,0.01199\%,0.04537\%,0.3397\%,0.04010\%$
of the float64 susceptibilities. Each is below 1\% relatively.
These are finite observation-program comparisons at fixed weights,
with the statistical bound derived in Section~\ref{model:sec:observation-theory}
and its normalization specified in Appendix~\ref{model:sec:statistics};
they are not certificates relative to exact real arithmetic.

\begin{figure}[htbp]
\centering
\includegraphics[width=\textwidth]{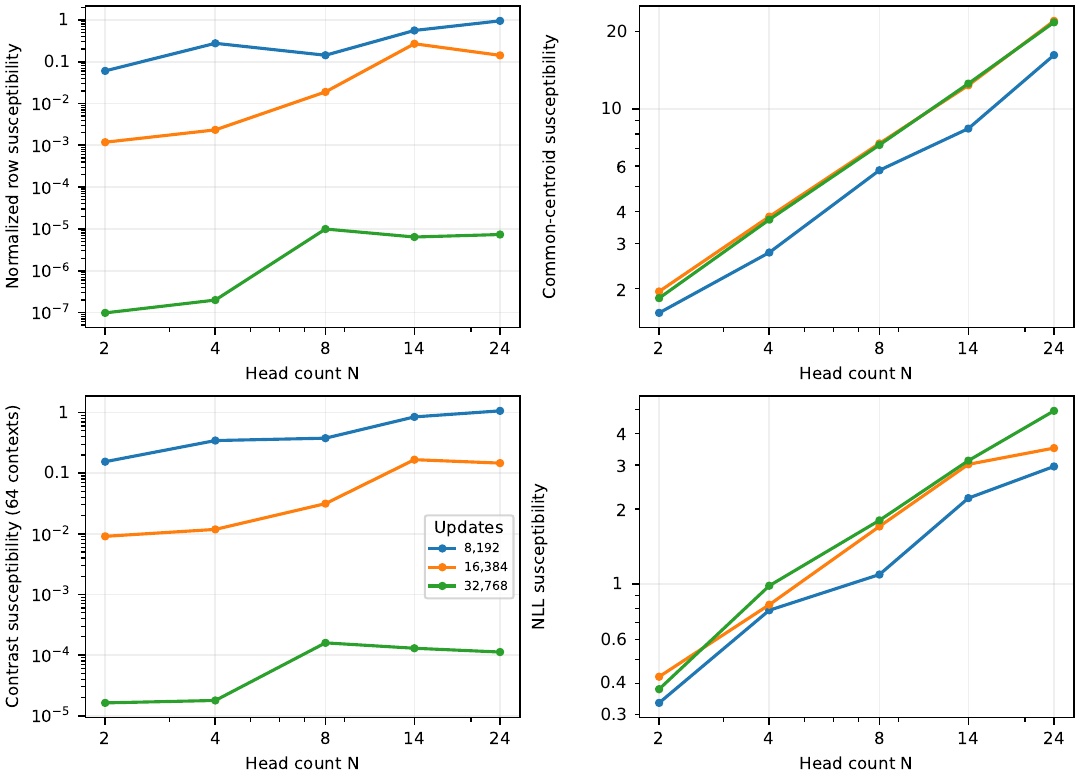}
\caption{Matched four-initialization width dependence at three training
horizons. The contrast field uses the separate 64-context calibration
cohort; other fields use 512 evaluation contexts. The horizontal
variable is a head count. Connecting points does not assert a spatial
length, stationary critical surface or asymptotic exponent.}
\label{model:fig:scaling-collective-widths}
\end{figure}

\begin{table}[htbp]
\centering\scriptsize
\setlength{\tabcolsep}{3pt}
\caption{Four-identity fixed-unit collective and predictive observations
at $g=1$. $C$ denotes the common centroid, and $I_{\rm ctx}$ is the
paired predictive context KL. Complete eight- and sixteen-identity
groups at their available horizons are included in the evidence object.}
\label{model:tab:scaling-collectives}
\input{content/model/generated/scaling-collectives.tex}

\end{table}

In the complete four-identity native GPU panel, extending 16,384 to
32,768 updates lowers the row-susceptibility point estimate at every
width. The decrease ranges from 3.28 to 4.62 orders of magnitude.
The 32,768-update estimates are approximately
$9.80\times10^{-8}$, $1.99\times10^{-7}$,
$9.95\times10^{-6}$, $6.40\times10^{-6}$ and
$7.34\times10^{-6}$ for $N=2,4,8,14,24$.
This temporal change does not eliminate initialization variation in
prediction: the corresponding NLL susceptibilities are 0.380,0.985,
1.800,3.122 and 4.941. Thus concentrating the row field does not
establish stationarity or concentration of every observable in the
full model. The CPU comparison and the paired-arithmetic observations
retain their own explicitly declared numerical program.

Nine observation fields have four frozen susceptibility alternatives:
the finite common-coordinate law $AN+B$, the independent-head correction
$A+B/N$, a free count power $AN^\kappa$, and the first-four estimate
at the same width. The first two fits constrain both coefficients to
be nonnegative and minimize squared relative residuals. The power fit
uses log residuals. Calibration uses the first four identities at
$N=2,4,8,14$ and 16,384 updates. Confirmation uses the next four
identities at $N=2,4,8,24$. The first-four $N=24$ outcome was available
when the design was specified, but was excluded from the three
cross-width fits. The 8,192-update parents of the $N=2,4,8$ targets
were also available. The frozen test concerns their unobserved
16,384-update continuations and four freshly initialized $N=24$
models. Confirmation labels are distinct from the calibration labels;
the selection of the model family remains developmental.

All 144 predictions are scored without refitting. Calibration and
confirmation have distinct initialization identities, so uncertainty
in their difference uses the product of the two finite empirical
resampling laws. Labels remain paired across widths within each law.
Four zero-variance calibration resamples are excluded from the fitted
models and counted explicitly; the same-width alternative retains
all 256 calibration resamples. These empirical percentiles have no
population or simultaneous-coverage guarantee.

The frozen finite-common form predicts the additional-panel
common-centroid susceptibility with relative RMS error 0.02885
and maximum absolute relative error 0.05000. Its four predictions
are $1.9723,3.7416,7.2800,21.4339$; the measured values are
$1.9162,3.7385,6.9160,21.3392$. The count-power and same-width
alternatives have relative RMS errors 0.03406 and 0.03600, whereas
the independent-head alternative gives 3.420. These are comparisons
of finite predictive errors, with the empirical uncertainty reported
separately. The coefficients of $AN+B$ are finite fit parameters;
this comparison alone does not identify them with individual
microscopic covariance sectors.

Prediction quality depends on the field. For NLL susceptibility,
the same-width reference has relative RMS error 0.07948, compared
with 0.1773 for the finite-common form. The row susceptibility
varies more strongly between the two four-identity panels: at $N=4$
the additional-panel value is 5.77 times the calibration-panel value.
The common-centroid result therefore supports a reproducible coherent
sector while leaving nonlinear contrast and predictive observations
with their own finite corrections and uncertainty.

\begin{figure}[htbp]
\centering
\includegraphics[width=\textwidth]{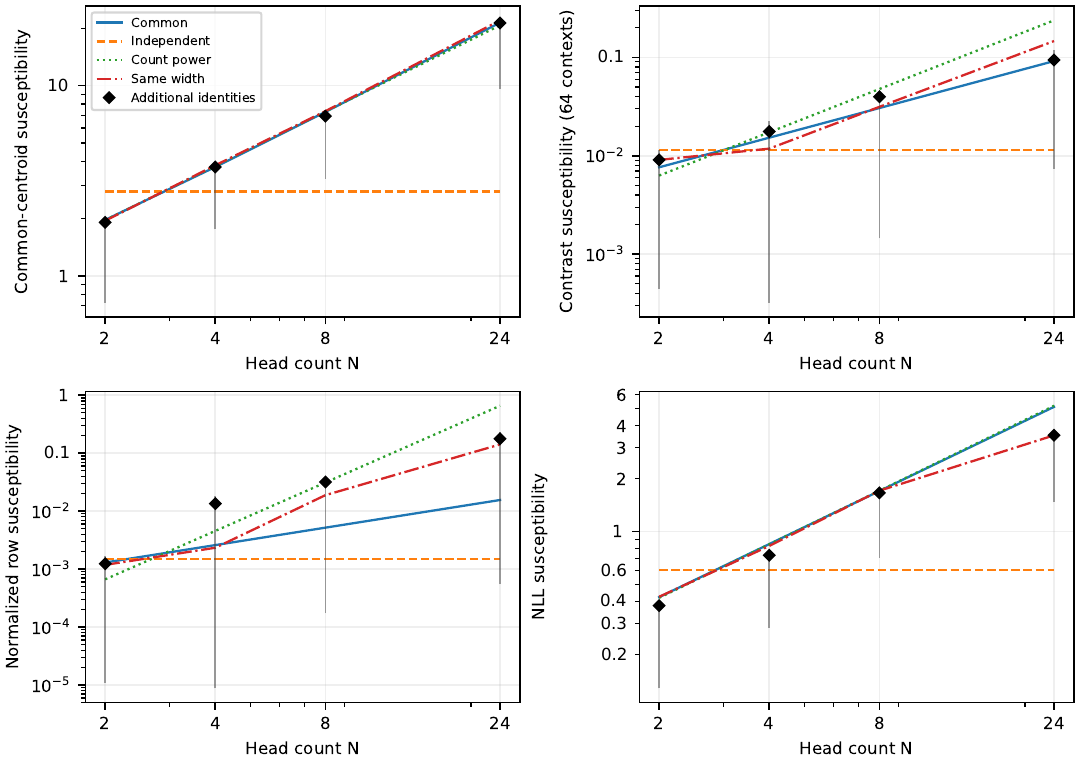}
\caption{Frozen predictions and additional-identity observations at
16,384 updates. Lines connect the four declared target widths and
do not add interpolated predictions to the test. Vertical bars show
the target panel's finite empirical percentile range, without a
population-coverage interpretation. The complete table retains all
nine fields and every alternative.}
\label{model:fig:scaling-variance-predictions}
\end{figure}

\begin{table}[htbp]
\centering\scriptsize
\setlength{\tabcolsep}{4pt}
\caption{All frozen variance alternatives, scored on initialization
identities excluded from calibration. Entries are signed relative errors $(\chi_{\rm obs}
-\chi_{\rm pred})/\chi_{\rm pred}$. RMS averages their squares across
the four target widths. The complete evidence also gives every
unrounded prediction and product-resampling error range.}
\label{model:tab:scaling-variance-predictions}
\input{content/model/generated/scaling-variance-predictions.tex}

\end{table}

The calibrated common-centroid power is 0.9480, whereas the normalized
row-field power is 2.7763. The former is consistent with a finite
coherent common coordinate over this range. The latter cannot be an
asymptotic susceptibility exponent of the unchanged bounded row
observation, by Proposition~\ref{model:prop:bounded-count-exponent}.
Nonlinear observation and evolving training time remain explicit
parts of the finite scaling law. Even a persistent $\chi_C\propto N$
would require the additional intrinsic and predictive identification
in Proposition~\ref{model:prop:finite-common-corner}.

\subsection{Extended horizons, boundary dynamics and environments}
\label{model:sec:extended-environment-results}
The matched $g=1$ continuations at $N=4,14$ retain four identities
through \ModelSourceScalingMaximumHorizon\ updates, with intermediate states at
65,536 and 98,304. Their paired changes include all adjacent available
balanced horizons, regardless of the sign of the change. Dense
observations also retain absolute row energy, common centroids,
predictive entropy and NLL. This separates disappearance of normalized
row contrast from stationarity of the remaining predictive state.

The completed CPU float32 panel is not monotonically concentrating.
At $N=4$, the mean normalized row field changes from
$8.0500\times10^{-6}$ at 32,768 updates to $0.014587$ at
131,072; at $N=14$ it changes from $1.4350\times10^{-4}$ to
$0.015841$. The corresponding endpoint row susceptibilities are
$0.0059791$ and $0.038212$, compared with
$1.9866\times10^{-7}$ and $6.4039\times10^{-6}$ at the earlier
horizon. Every selected horizon and initialization remains in the
paired calculation. The transient concentration therefore does not
identify an absorbing collapsed state under constant-rate training.

The common coordinate remains coherent while its amplitude law
changes. From 32,768 to 131,072 updates, $\chi_C$ decreases from
$3.7153$ to $2.2397$ at $N=4$ and from $12.5523$ to $8.1970$
at $N=14$. Endpoint effective head counts are nevertheless
$1.0023$ and $1.0036$. Thus a head-coherent common mode can persist
through substantial evolution of both its covariance and the row
contrast. This behavior requires a coupled, time-dependent effective
state; it does not establish a stationary critical ensemble.

\begin{figure}[htbp]
\centering
\includegraphics[width=\textwidth]{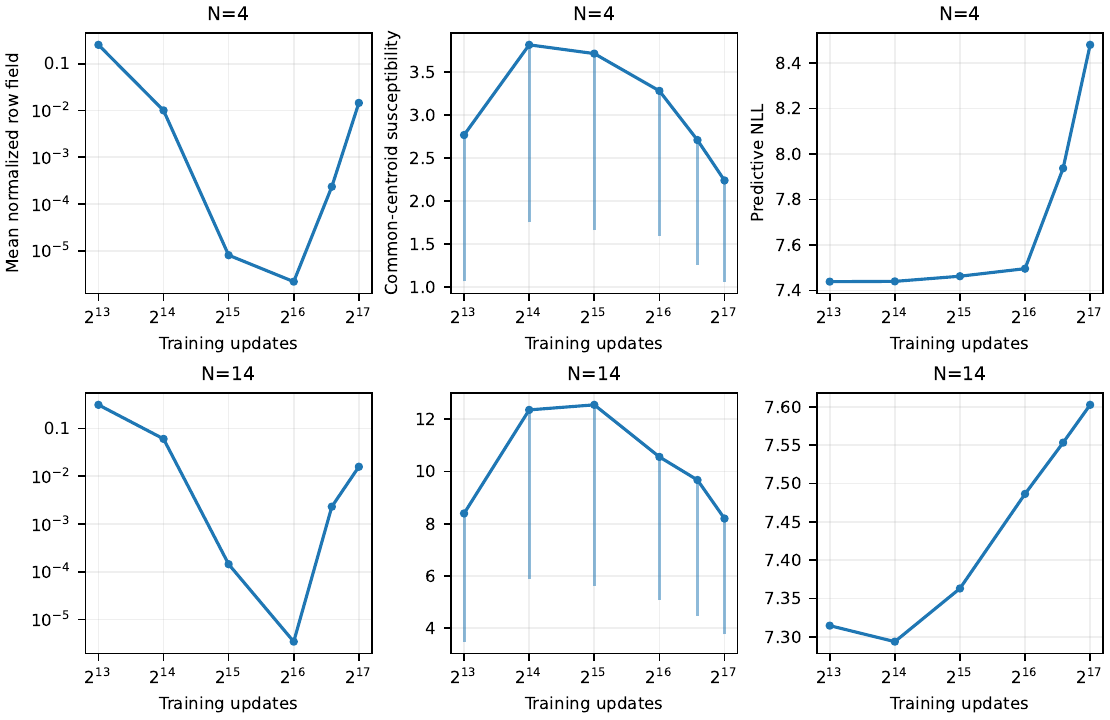}
\caption{Extended matched training horizons at $N=4,14$. The same
four identities are retained throughout. Susceptibility intervals
are whole-initialization empirical percentiles conditional on the fixed
contexts and environment. They are not temporal error bars.
The row-field axis is logarithmic above $10^{-8}$ and linear below
that value, retaining both concentration and subsequent excursions.}
\label{model:fig:scaling-long-horizons}
\end{figure}

\begin{table}[htbp]
\centering\scriptsize
\setlength{\tabcolsep}{3pt}
\caption{Paired long-horizon changes in normalized row, common-centroid
and predictive fields. Every listed difference retains the same
initialization multiplicities at both endpoints.}
\label{model:tab:scaling-long-paired}
\input{content/model/generated/scaling-long-paired.tex}

\end{table}

The paired arithmetic diagnostic for these added observations uses
the explicitly stated rule
$2\sqrt{\chi_{64}\chi_{32-64}}+\chi_{32-64}
\leq\max(0.01\chi_{64},10^{-8})$.
This disjunctive reporting rule differs from the conjunctive relative
and absolute criterion in the thirteen-condition arithmetic study.
An absolute-only agreement at a tiny susceptibility does not authorize
transfer of a relative scaling exponent. Neither comparison certifies
agreement with exact real arithmetic.

\begin{table}[htbp]
\centering\scriptsize
\setlength{\tabcolsep}{3pt}
\caption{Paired forward-arithmetic observations at later matched states.
The rule column applies the disjunctive reporting bound stated in the
text. All unscreened fields remain in the calculation.}
\label{model:tab:scaling-arithmetic}
\input{content/model/generated/scaling-arithmetic.tex}

\end{table}

Eight boundary continuations extend paths whose generator rate has
been zero from initialization. They retain their complete 2,048-update
states and train through 32,768 updates.
All declared generator parameters remain bitwise unchanged, while
body parameters, gradients, clipping and optimizer moments follow
their ordinary rules. The Gram-input normalization is a body
parameter, so freezing generator weights does not freeze the
input-dependent head or its full-graph emission.
At 32,768 updates the CPU float32 mean normalized row contrasts are
$0.98736$ and $0.98779$ for $N=4,14$, respectively. Their
susceptibilities are $4.36\times10^{-5}$ and $4.99\times10^{-5}$.
Thus a small seed susceptibility also occurs in a state whose rows
retain large normalized contrast; it does not by itself diagnose
row concentration. The common-centroid susceptibilities are $0.01460$
and $0.01300$, with effective head counts $3.31$ and $12.53$.
The common component accounts for $1.97\%$ and $1.79\%$ of the
full-cohort head-averaged metric covariance trace, rather than the
near-unit common fraction of the learned concentrated states.

Between 8,192 and 32,768 updates the mean NLL falls from $7.0665$
to $6.9708$ at $N=4$ and from $7.0100$ to $6.9076$ at $N=14$;
every one of the eight matched initialization identities has lower
endpoint NLL. These finite controls demonstrate body-driven
predictive learning without the measured row concentration. Together
with the nonzero-rate paths, they identify generator adaptation as
part of the observed concentration mechanism. They do not determine
an infinite-time law at zero rate or justify exchanging the limits
$g\to0$ and $t\to\infty$.

\begin{table}[htbp]
\centering\small
\caption{The zero-generator boundary with four matched identities per
width. Fixed generator weights are distinguished from a fixed
input-dependent generator output.}
\label{model:tab:scaling-zero-generator}
\input{content/model/generated/scaling-zero-generator.tex}

\end{table}

The environment study crosses two shared initializations, two entire
batch histories and four nonshared initialization identities at each
of $N=4,14$. Its 32 trained paths use 8,192 updates. The seven
orthogonal nonconstant functional-ANOVA sectors are defined on the
uniform finite $2\times2\times4$ product law. Total population
variance uses that finite law's divisor. The average conditional
initialization susceptibility uses divisor three and equals $4N/3$
times the sum of the sectors containing the initialization index.
Interactions are retained; four crossed environment choices are not
four independent draws from an unspecified environment population.

\begin{table}[htbp]
\centering\scriptsize
\setlength{\tabcolsep}{3pt}
\caption{Finite environment decomposition. $C,B,I$ are the shared
initialization, batch-history and nonshared-initialization main-effect
fractions. The interaction column sums the four remaining fractions;
each is retained individually in the evidence object. The decomposition
uses the finite uniform $2\times2\times4$ crossed law at each width;
its environment choices are not draws from an unspecified population.}
\label{model:tab:scaling-environments}
\input{content/model/generated/scaling-environments.tex}

\end{table}

The interaction sectors account for 64.06\% and 70.78\% of the
row-field variance at $N=4,14$, respectively. For the common
centroid their fractions are 65.83\% and 67.44\%. Thus an additive
output model containing only the three main effects omits most of
these finite-design variances. The factors are independent under the
declared product law; it is the trained output that depends on their
interactions.

The conditional row susceptibilities range from 0.07734 to 0.27562
over the four environment cells at $N=4$, and from 0.52212 to
0.61443 at $N=14$. Their averages are 0.17864 and 0.55889.
These differences compare the specified 8,192-update conditional
laws. They do not arise from adding environment cells as extra
initialization replicas to a fixed conditional susceptibility.

The effect of batch history also depends on the inference observation.
Its main-effect fractions are 22.53\% and 28.74\% for NLL,
compared with 58.28\% and 75.06\% for the fixed projected logits.
The pretraining-data distribution is unchanged in these contrasts,
but its realized batch history remains visible in the trained law.
Consequently the conditional law and the law averaged over the
declared environments retain different covariance predictions.
The finite crossed design supplies no assumption of a
history-independent stationary law.

\subsection{Frozen training-law risk and online evolution}
\label{model:sec:frozen-risk-results}
A separate observation of the same long trajectories distinguishes
risk under the empirical training law from risk on the fixed held-out
cohort. It uses 1,024 fresh crops from the exact native row and offset
sampler, with a disjoint seed. The selected incoming states cross
$N=4,14$, four initialization identities and updates
32,768, 65,536, 98,304 and 131,072. The same training-law crops
and the same 512 held-out inputs are evaluated at every state in
CPU float32 batches of 32. Each crop supplies one next-token target.
The state is frozen throughout each risk observation.

The selection and fresh cohort were fixed before these evaluations.
Five long held-out paths and their online training losses were already
known, so this is a developmental diagnostic. All 32 selected states,
both context cohorts, and every adjacent matched horizon remain in
the analysis. Whole-initialization empirical ranges condition on the
two context cohorts; they do not integrate context-sampling uncertainty
or guarantee population coverage.

For each checkpoint, the preceding 8,192 native preupdate minibatch
losses give a separate online average. That average follows an evolving
trajectory, whereas the two frozen-state risks evaluate one fixed model.
Their time conventions and probability laws are consequently kept
distinct. The held-out fields are also compared with the original
CPU checkpoint observation. Full first-batch logits and targets in
both cohorts support independent softmax reconstruction, and all
online losses retain their original native record.

\begin{figure}[htbp]
\centering
\includegraphics[width=\textwidth]{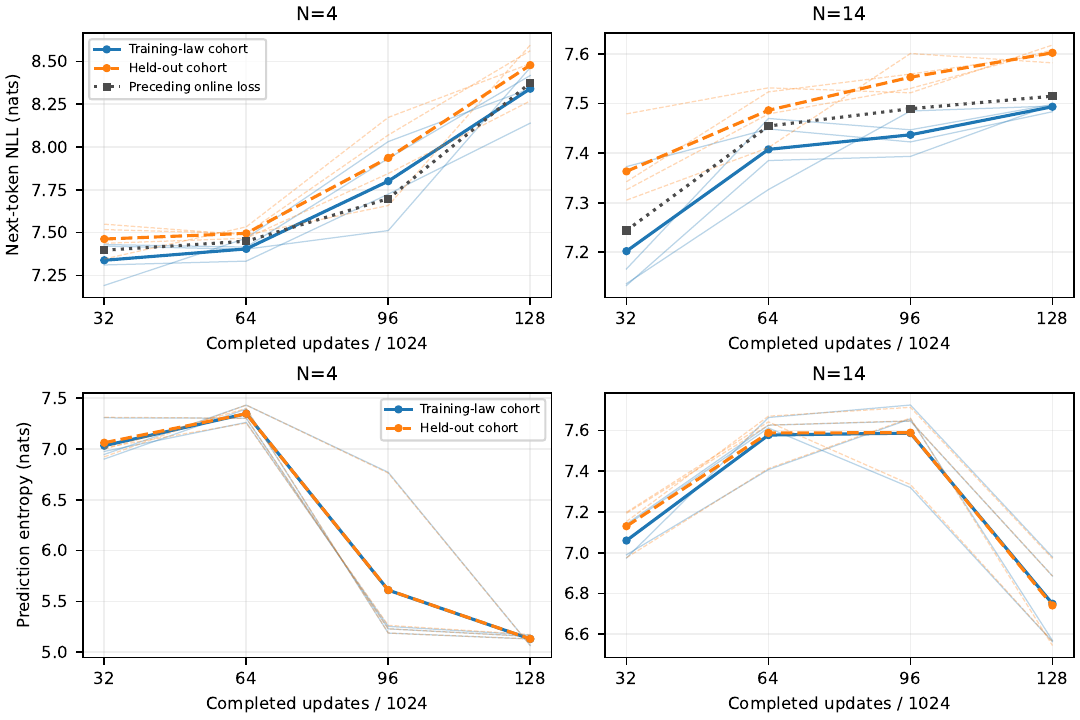}
\caption{Frozen training-law and held-out risk across the matched
long trajectories. Thin lines retain each initialization identity;
thick lines are their arithmetic means. The online loss averages
the preceding 8,192 updates of the evolving model. Every entropy
and loss is in nats per external next-token target.}
\label{model:fig:scaling-training-risk}
\end{figure}

\begin{table}[htbp]
\centering\scriptsize
\setlength{\tabcolsep}{3pt}
\caption{All eight matched width--time risk conditions. The risk gap
is held-out minus training-law NLL at the same frozen states.
Its empirical range pairs the same four initialization
multiplicities across both context cohorts.}
\label{model:tab:scaling-training-risk}
\input{content/model/generated/scaling-training-risk.tex}

\end{table}

\begin{table}[htbp]
\centering\scriptsize
\setlength{\tabcolsep}{3pt}
\caption{All adjacent matched horizon changes in frozen-state NLL.
The empirical ranges preserve initialization identity across
horizons and condition on the fixed contexts.}
\label{model:tab:scaling-training-risk-paired}
\input{content/model/generated/scaling-training-risk-paired.tex}

\end{table}

The complete fresh-crop observations show deterioration under both
laws over the longest matched interval. At $N=4$, the four-identity
mean training-law NLL increases from $7.3380$ to $8.3413$ between
32,768 and 131,072 updates, while held-out NLL increases from
$7.4623$ to $8.4793$. At $N=14$, the corresponding changes are
$7.2023$ to $7.4937$ and $7.3632$ to $7.6025$.
The held-out minus training-law gaps change from $0.1243$ to
$0.1380$ and from $0.1609$ to $0.1088$, respectively. The worsening
held-out loss is therefore not explained solely by an increasing
empirical generalization gap. The trained conditional predictor
also worsens on fresh samples from its own specified training law.

All 32 held-out loss and entropy fields agree bytewise with their
corresponding original checkpoint observations. At the same endpoints,
mean held-out entropy decreases from $7.0626$ to $5.1311$ at $N=4$
and from $7.1305$ to $6.7415$ at $N=14$. Lower predictive entropy
and earlier row concentration do not entail lower proper prediction
risk. These observations constrain the finite model-wide flow to
retain learned predictive competence alongside metric concentration
and its stochastic source variables.

\subsection{Conditional native noise and predictive transport}
\label{model:sec:conditional-noise-results}
At each of 24 fixed incoming states, 32 fresh IID minibatches determine
the conditional generator Adam velocity. The state panel crosses
$N=4,14$, updates 2,048, 8,192 and 16,384, and the first four
initialization identities. The full loss gradient is globally clipped
before restricting its source coordinates. The velocity formula is evaluated in float64 from clipped CPU
float32 gradients. It includes the incoming first and second moments, next-step
bias correction, denominator offset and decoupled decay. With velocities
$v_j$, the squared-drift estimator is
\[
 \widehat{\|b\|^2}=\|\bar v\|^2-\frac{1}{32}\tr\widehat Q,
 \qquad
 \widehat Q=\frac1{31}\sum_j(v_j-\bar v)(v_j-\bar v)^\top.
\]
It is not truncated at zero. For the shared metric network, the ratio
$\widehat{\|b\|^2}/\tr\widehat Q$ ranges from 2.73 to 10.34 in
the selected states. Thus the observed near-quadratic passage clock
does not by itself identify centered diffusion as its microscopic
mechanism. Proposition~\ref{model:prop:innovation-motion} retains both the
drift and its accumulated cross term with the innovations.

Eight of the same directions per state are transported through the
complete inference graph on sixteen fixed inputs. The nominal native
generator displacement includes its learning rate; body parameters
are held fixed in this local source experiment. Predictive tangents
use the full-vocabulary Fisher metric. Comparing parameter and
predictive covariance uses exactly the same eight samples and the
same centering, so both participation ranks are bounded by seven.
The 32-batch parameter spectrum has a different sample-rank bound and
is retained separately. In all 24 matched comparisons the predictive
participation rank is lower. This is an empirical transport property,
not monotonicity of participation rank under arbitrary linear maps.

\begin{table}[htbp]
\centering\small
\caption{Conditional forcing and matched source--emission geometry.
Each interval is the range over four incoming initialization states.
The drift ratio concerns the shared metric network. Participation
ranks concern the complete declared generator parameter group and its
full-graph predictive tangent on the same eight minibatches.}
\label{model:tab:scaling-noise-geometry}
\input{content/model/generated/scaling-noise-geometry.tex}

\end{table}

\begin{figure}[htbp]
\centering
\includegraphics[width=\textwidth]{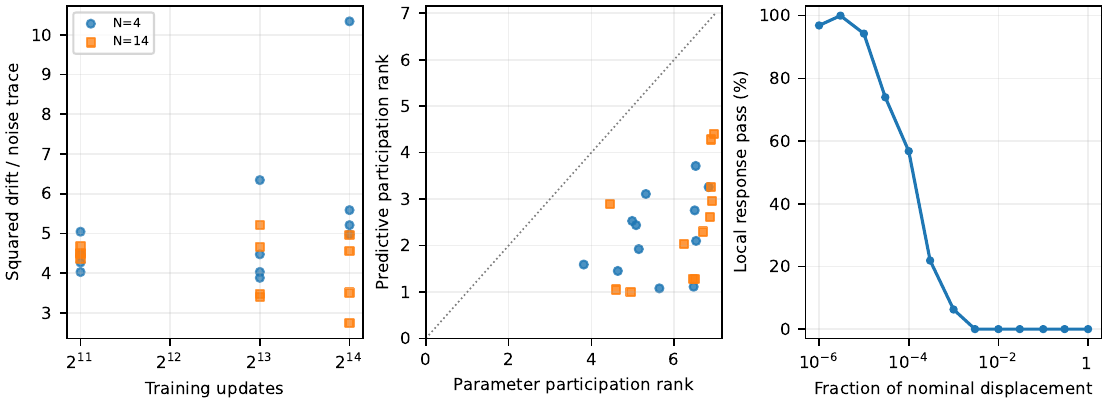}
\caption{Directed conditional forcing, matched finite-sample predictive
transport and the local-response window. The right panel includes all
192 state--direction pairs at every declared positive amplitude.}
\label{model:fig:scaling-noise-geometry}
\end{figure}

Both signs of thirteen amplitudes from $10^{-6}$ to one test the
first field derivative, half the second derivative, and predictive KL
curvature against automatic derivatives of the smooth float64
realization. The joint diagnostic requires relative errors at most
1\% in each of the first and half-second \emph{global field norms},
and at most 5\% in Fisher curvature. It is not a 1\% relative
guarantee for every separate field component. All 192 directions meet
this diagnostic at $\alpha=3\times10^{-6}$; none does at unit
amplitude. Consequently local transport is used within its measured
range and is not substituted for a complete native optimizer step.
All individual field errors and native/smooth primal comparisons are
retained.

\begin{table}[htbp]
\centering\small
\caption{Complete amplitude sensitivity of the joint local-response
diagnostic. Refining the observation scale changes the numerical
validity window; it does not create additional independent states.}
\label{model:tab:scaling-locality}
\input{content/model/generated/scaling-locality.tex}

\end{table}

A projected isotropic common-row force can have very large held-out
residuals in deeper decoders. The reduced description therefore
retains the full transported source rather than identifying a fitted
passage clock with that restricted forcing law. The source coordinates
in Proposition~\ref{model:prop:visible-source-coordinates} provide an exact
local representation on a specified predictive subspace.

\subsection{Frozen adjoint coordinates and fresh-batch validation}
\label{model:sec:adjoint-source-results}
At sixteen incoming states, $N=4,14$ crossed with updates 2,048 and
16,384 and the first four identities, the original eight predictive
tangents calibrate an uncentered Fisher-response basis. Full-graph
adjoints pull its at most eight directions back to generator parameter
space. The basis, adjoints, reference probabilities and their hashes
are frozen before sixteen fresh validation minibatches are generated.
The fresh batch law has a separate seed and is paired across all
incoming states. These measurements test source generalization at the
same fixed context cohort; they do not test new-context generalization.

For each fresh displacement, adjoint source coefficients predict the
projection of a directly recomputed full-graph tangent. Adjoint duality
is checked separately from subspace approximation. Dimensions
$k=0,1,2,4,8$ all remain in the result. The two residual fractions are
\begin{equation}
 e_{\rm total}(k)=
 \frac{\widehat\E\|(I-P_k)Y\|^2}{\widehat\E\|Y\|^2},
 \qquad
 e_{\rm noise}(k)=
 \frac{\tr\widehat\Cov((I-P_k)Y)}{\tr\widehat\Cov(Y)}.
 \label{model:eq:fresh-source-errors}
\end{equation}
The uncentered calibration basis minimizes its calibration total
response residual. It need not minimize fresh conditional noise
error, and preserving a mean response can coexist with appreciable
innovation loss. This separation is required by the two identities
in Proposition~\ref{model:prop:visible-source-coordinates}.

\begin{figure}[htbp]
\centering
\includegraphics[width=\textwidth]{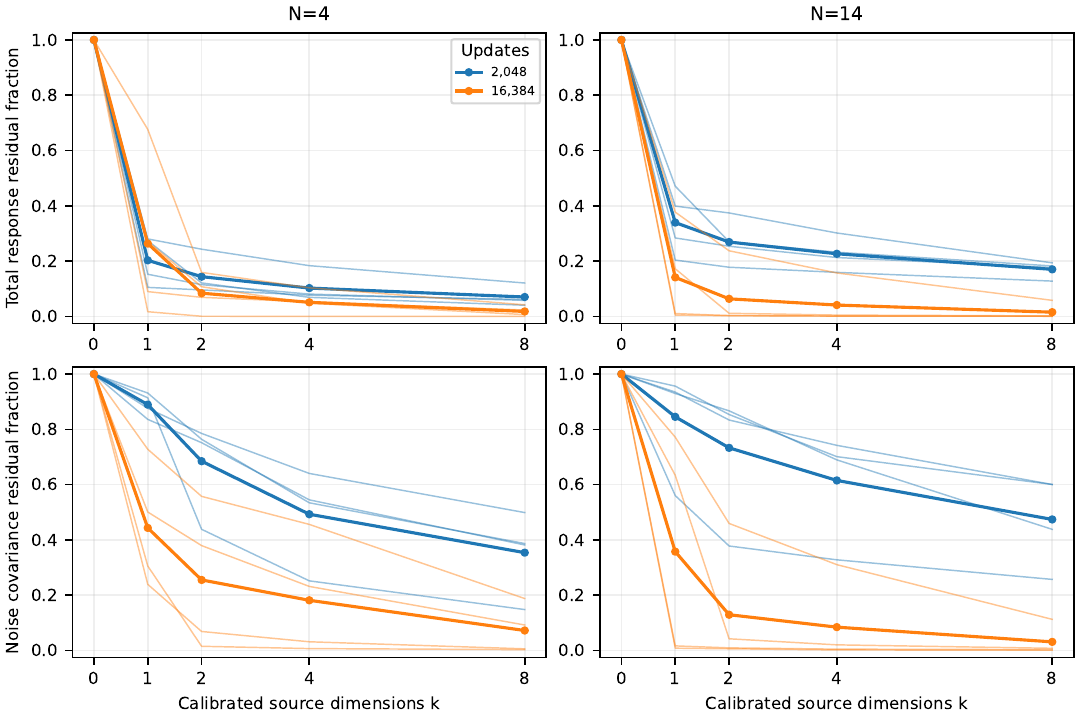}
\caption{Fresh-batch validation of frozen adjoint source coordinates.
Thin curves retain all four incoming initialization states; thick
curves are their arithmetic means. Total response and innovation
covariance have distinct denominators.}
\label{model:fig:scaling-visible-noise}
\end{figure}

\begin{table}[htbp]
\centering\scriptsize
\setlength{\tabcolsep}{3pt}
\caption{All five source dimensions in all four width--time conditions.
Ranges retain the four individual states. Pooled fractions divide
summed residual energy by summed total energy, and need not equal
the arithmetic mean of the four fractions.}
\label{model:tab:scaling-visible-noise}
\input{content/model/generated/scaling-visible-noise.tex}

\end{table}

With eight retained directions, the four-state covariance residual
ranges at 2,048 updates are $[0.148,0.499]$ for $N=4$ and
$[0.257,0.601]$ for $N=14$. At 16,384 they are $[0.00345,0.188]$
and $[0.00123,0.113]$. The mean paired changes are respectively
$-0.2815$ and $-0.4428$, with empirical ranges
$[-0.4454,-0.0640]$ and $[-0.5959,-0.2897]$. Thus this measured source
representation improves during row concentration while retaining
substantial state-to-state variation. The late pooled covariance
residuals, 0.01095 and 0.007883, weight states by their conditional
noise trace. They do not imply the same error at every state and do
not include covariance across the states' conditional means.
Corollary~\ref{model:cor:source-noise-priority} explains why the smaller
total-response residuals alone do not control these noise fractions.

Fresh finite responses at $10^{-6},3\times10^{-6},10^{-5}$ also test
the tangent and Fisher curvature. Every one of the 256 fresh directions
passes the stated first-derivative and Fisher criterion at each of
these three magnitudes. The source law remains conditional
on the complete incoming state and its chosen generator source group.
A local predictive subspace is not an autonomous reduced training
state: body sources, their cross covariance, optimizer memory and
accumulated closure errors remain relevant to such a dynamical claim.

\subsection{Paired expansion of the predictive source state}
The expanded source study retains all sixteen incoming states and
uses all 32 original calibration minibatches. The remaining 24
full-graph tangents extend the eight archived calibration tangents.
A total-response basis is ordered by the uncentered second moment;
a second basis orders the same complete span by centered calibration
noise covariance. Both use frozen full-graph adjoints. The final
covariance coordinate can retain calibration drift outside the
centered-noise span, so retaining the full basis recovers the same
calibration span under either ordering.

Thirty-two fresh minibatches per state come from a second disjoint
validation sample. The already frozen eight-calibration predictor is
also evaluated on these exact new displacements. This reference
comparison was specified before any expanded scientific validation
began. It isolates calibration size and coordinate ordering from a
change of validation-batch realization. All dimensions
$k=0,1,2,4,8,16,32$ are retained for the expanded bases; the reference
retains its available dimensions through eight. Conditional mean,
total response and innovation covariance remain distinct statistics.

Increasing calibration from eight to 32 minibatches improves the
fresh covariance residual at fixed $k=8$ in all sixteen states when
coordinates retain the total-response ordering. At 2,048 updates,
the mean paired changes are $-0.1118$ for $N=4$ and $-0.1535$ for
$N=14$, with empirical ranges $[-0.1272,-0.0889]$ and
$[-0.1986,-0.1077]$. At 16,384 updates the corresponding changes
are $-0.0572$ and $-0.0228$, with ranges $[-0.1246,-0.00231]$
and $[-0.0624,-0.00167]$. These paired comparisons keep both the
validation sample and retained dimension fixed.

With all 32 coordinates, the early statewise noise residuals span
$[0.0239,0.0817]$ for $N=4$ and $[0.1201,0.1907]$ for $N=14$.
The late ranges become $[3.70\times10^{-5},0.001229]$ and
$[7.19\times10^{-5},0.002301]$. Thus every measured late state
retains at least 99.769\% of its conditional predictive tangent
noise trace. The late pooled residual fractions are
$7.39\times10^{-5}$ and $1.78\times10^{-4}$; their weighting again
differs from a uniform average over states. At sixteen coordinates,
one of the eight late states still has a residual above 1\%.
This dimension sensitivity prevents interpreting the eight-coordinate
description as a uniform quantitative closure.

The centered-noise ordering changes the tradeoff between mean response
and innovation covariance. It is not uniformly superior on fresh
batches at a truncated dimension; the two orderings agree at the full
32-dimensional span by construction. All 512 fresh directions satisfy
the declared tangent and Fisher diagnostic at each of the three
tested local magnitudes. These results validate conditional predictive
source compression on the fixed probe law. They do not determine the
population Jacobian rank, a universal collective dimension, or the
memory error of an autonomous reduced training process.

\begin{figure}[htbp]
\centering
\includegraphics[width=\textwidth]{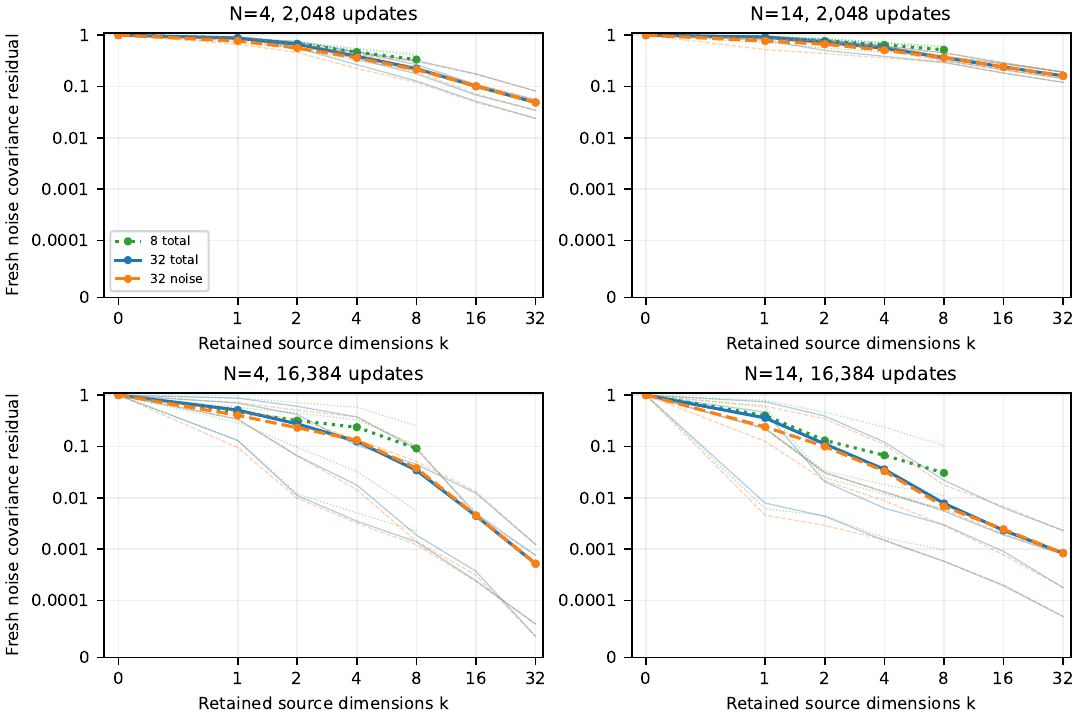}
\caption{Predictive noise under paired source-state expansion. Every
curve uses the same 32 new batches at its incoming state. The labels
specify calibration sample size and basis objective. Thin curves
retain individual states and thick curves their arithmetic mean.
The vertical scale is logarithmic above $10^{-4}$ and linear below it.}
\label{model:fig:scaling-source-basis}
\end{figure}

\begin{table}[htbp]
\centering\scriptsize
\setlength{\tabcolsep}{3pt}
\caption{The eight-coordinate reference and expanded source states on
one common fresh validation sample. Every earlier retained dimension
is included in the numerical evidence. Ranges prevent a small pooled
error from being interpreted as uniform statewise accuracy.}
\label{model:tab:scaling-source-basis}
\input{content/model/generated/scaling-source-basis.tex}

\end{table}

\subsection{The complete native step and its source interactions}
\label{model:sec:native-source-step-results}
The native-step study uses the same sixteen incoming width--time
states, with 32 fresh minibatches per state from a separate batch law.
Before every draw it restores the entire saved model and Adam state.
The full batch-32 gradient is computed in CPU float32, globally
clipped and passed through the actual native AdamW step. Four
batch-16 probe emissions give the parent, generator-only, body-only
and complete weight corners in
Proposition~\ref{model:prop:finite-source-corners}. Partial corners retain
the displacements realized by that same coupled step. No
infinitesimal extrapolation is used.

The primary row and head fields first average over heads and retain
each fixed context and decoder. The common centroid retains its
native column coordinates. Prediction retains all vocabulary
coordinates. Energies and covariance reductions use float64.
For each incoming state, the Fisher logit distance uses its fixed
parent probabilities,
\[
 D_F(\Delta z)=\frac1{16}\sum_{x,v}p_z(v\mid x)
 \left(\Delta z_v(x)-\sum_w p_z(w\mid x)\Delta z_w(x)\right)^2.
\]
This is a fixed-base observation norm. Its relation to finite
predictive KL is measured separately, rather than assumed at the
native step size.

Signed conditional mean increments are retained separately from
their squared norms. In particular, the row-field mean satisfies
$\widehat\mu_R=\widehat\mu_{R,a}+
\widehat\mu_{R,b}+\widehat\mu_{R,ab}$ at each incoming state.
The corresponding absolute-energy and NLL means determine whether a
particular source increases or decreases the measured observable.
A negative mean row increment at selected full states does not
identify a closed scalar drift depending only on $R$.

\begin{table}[htbp]
\centering\scriptsize
\setlength{\tabcolsep}{3pt}
\caption{Signed native mean increments, averaged over four incoming
states and their 32 fresh batch draws. The final column counts the
four states with a negative estimated conditional mean complete increment.
Batch draws are nested within each fixed complete incoming state; the
source corners are paired on each draw. All individual means and conditional cohort-mean covariances are
retained. Source additivity holds for the unrounded values.}
\label{model:tab:scaling-native-step-drift}
\input{content/model/generated/scaling-native-step-drift.tex}

\end{table}

Let $Y$ be the complete increment and $Y_a$ the generator corner.
The two generator-only errors are
\begin{equation}
 e_{\rm gen,total}=
 \frac{\widehat\E\norm{Y-Y_a}^2}{\widehat\E\norm Y^2},
 \qquad
 e_{\rm gen,noise}=
 \frac{\tr\widehat\Cov(Y-Y_a)}{\tr\widehat\Cov(Y)}.
 \label{model:eq:native-source-errors}
\end{equation}
Unlike an orthogonal projection residual, neither fraction is bounded
by one. Define also
$M_{ab}/M=\widehat\E\norm{Y_{ab}}^2/\widehat\E\norm Y^2$.
A mixed contribution larger than the complete response is compatible
with destructive cross terms. All nine signed entries of the
conditional source covariance are retained. Pooled ratios sum
statewise numerators and denominators; they do not include variance
across the incoming states' conditional means.

\begin{table}[htbp]
\centering\scriptsize
\setlength{\tabcolsep}{3pt}
\caption{Actual conditional native-step source errors. The table
retains the row field, common centroid, NLL and full-vocabulary
Fisher logit field. Every measured field and individual incoming
state is retained in the numerical evidence. The mixed column
includes the full finite source interaction.}
\label{model:tab:scaling-native-step}
\input{content/model/generated/scaling-native-step.tex}

\end{table}

\begin{figure}[htbp]
\centering
\includegraphics[width=\textwidth]{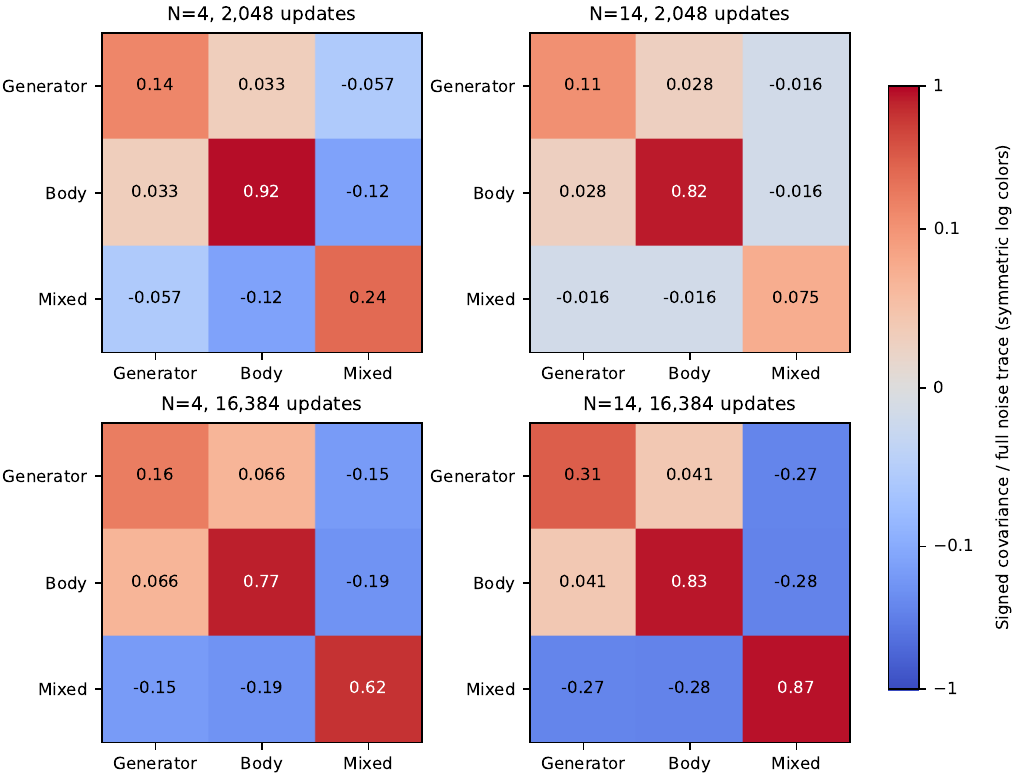}
\caption{Signed conditional source covariance of the actual native
predictive increment. Each matrix is the four-state average of the
within-state covariance divided by the average complete noise trace.
All nine entries sum to one, including both copies of each
off-diagonal cross term. Colors use a symmetric logarithmic scale;
printed entries retain their signed values. The 32 fresh batches per
state are the within-state random units. Signed covariance shares,
including negative values, are neither probabilities nor correlation
coefficients.}
\label{model:fig:scaling-native-step}
\end{figure}

\begin{table}[htbp]
\centering\small
\caption{Finite predictive KL and fixed-base Fisher distance for
every native source corner. The ratio range covers all 128
state--batch pairs in a width--time condition. A ratio near one is
a measured local agreement, not an identity imposed on the data.}
\label{model:tab:scaling-native-step-kl}
\input{content/model/generated/scaling-native-step-kl.tex}

\end{table}

The complete predictive increment is not well represented by the
generator corner alone. At 16,384 updates, its pooled total-response
errors are 0.9797 and 1.0034 for $N=4,14$, and its pooled noise
errors are 1.0129 and 1.1543. The corresponding mixed second-moment
fractions are 0.1714 and 0.6067. By contrast, the internal common
centroid at $N=14$ has generator-only total and noise errors of
0.01711 and 0.02083. Thus a useful approximation to one internal
collective need not close the predictive training increment.

The estimated mean generator-corner row increment is negative in
each of the four width--time conditions after averaging incoming
states. The complete row mean has both signs across states: eight
of sixteen point estimates are negative. At $N=4,t=16{,}384$, the
mean source contributions are approximately
$-5.337\times10^{-5}$, $4.984\times10^{-5}$ and
$2.003\times10^{-5}$, giving a positive complete mean
$1.650\times10^{-5}$. At $N=14$ at the same time they are
$-2.190\times10^{-4}$, $4.039\times10^{-6}$ and
$4.076\times10^{-5}$, giving $-1.742\times10^{-4}$.
These are conditional finite-batch estimates at the stated complete
incoming states. They support retaining the coupled sources and do
not supply a uniform contracting scalar drift or a stationary
critical normal coordinate.

At 16,384 updates, the mean full-step predictive KL is 0.005772
for $N=4$ and 0.002647 for $N=14$. The ratio of twice KL to the
fixed-base Fisher square of the \emph{actual logit increment} ranges
over $[0.874,1.108]$ and $[0.903,1.054]$ respectively.
This comparison differs from replacing a parameter displacement by
its infinitesimal logit tangent. A small predictive KL can therefore
coexist with appreciable nonlinearity and source interaction in the
parameter-to-prediction map.

Independent first-batch replay at each selected state checks every
named parameter, every Adam tensor and every saved scalar in all
four emissions bitwise. The remaining draws are retained raw and
their moments are reconstructed independently. These 512 conditional
steps repeatedly start from sixteen fixed states; they do not add
512 sequential training updates or independent pretraining identities.
Their exact covariance decomposition constrains a native reduced
model, while a long-time transition closure still requires transport
of state, cross terms and memory.

\subsection{Observable time scales and a decoupled slow sector}
Sixty selected trajectories passively record sixteen fixed projections
of the actual native shared-metric parameter update and the clipped
gradient at every update. They also record mean squared update, mean
squared incoming weight and mean update--weight product. The
instrument's native CPU qualification gives equal loss values and
matching byte hashes of weights and the full optimizer state under
paired recorded and unrecorded execution. Independent projection and squared-norm telescoping
identities check the executed paths. This qualification does not
assert unmeasured bitwise GPU equivalence.

A correlation with zero temporal variance is undefined and recorded
as such. If a displayed range has fewer defined values than incoming
states, its count is shown explicitly; undefined values are not zeros.

Temporal summaries retain every complete absolute-time-aligned block
of 2,048, 8,192 and 32,768 updates. Mean-centered and linearly
detrended covariances are both reported. The temporal variance of a
single path is not substituted for conditional initialization
susceptibility. Row-threshold excursions at 0.1, 0.01 and 0.001
are measured on the fixed 64-update mesh with both boundary-censoring
flags. They are observations of threshold crossings, not identified
avalanches with a separated drive and relaxation event law.

\begin{table}[htbp]
\centering\scriptsize
\setlength{\tabcolsep}{3pt}
\caption{The final complete 32,768-update block in each matched long
trajectory. Each range is over four initializations. Trend fraction
is the linear-trend contribution divided by mean-centered temporal
variance. Correlations at the first recorded lag are displayed before
and after detrending; the evidence retains all declared lags and blocks.
Undefined zero-variance ratios and correlations are marked n/a;
a partial range states its number of defined values.}
\label{model:tab:scaling-temporal-windows}
\input{content/model/generated/scaling-temporal-windows.tex}

\end{table}

The final matched interval, updates 98,304--131,072, retains
substantial evolution of the common coordinate. Linear trends account
for $40.6\%$--$55.2\%$ of its temporal variance across the four
$N=4$ paths and $42.3\%$--$51.5\%$ at $N=14$.
For NLL the corresponding fractions are $45.1\%$--$79.1\%$ and
$13.3\%$--$33.9\%$. Thus treating the raw finite-window covariance
as a stationary fluctuation law would mix those trends with the
remaining fluctuations.

At lag 8,192 in that interval, common-coordinate correlations average
$0.0623$ and $0.0589$ over the four identities at $N=4,14$.
After linear detrending they average $-0.2388$ and $-0.2625$.
These are finite-window normalized covariances under two distinct
centering conventions. Their difference prevents interpreting a
single fitted correlation time as an established critical relaxation
time. Complete lag and window records accompany the shorter-lag
values in Table~\ref{model:tab:scaling-temporal-windows}.

An independent support control supplies a useful exact distinction.
There are 1,387 vocabulary indices absent from every possible training
input under the empirical crop law and from all 1,024 probe inputs.
At one fixed index, 6481, a pulse of 0.1 in the untied input embedding
is coupled to an unpulsed path for sixteen native CPU updates at each
of the five head counts. All absent-input gradients are exactly zero;
the pulsed row follows repeated float32 decay exactly in value.
Predictions and losses agree exactly between the paired branches,
and all other parameters and optimizer states have matching byte hashes. Proposition~\ref{model:prop:invisible-decay}
therefore has an executed native instance with relaxation time
\[
 \tau_N=\frac{-1}{\log(1-6\times10^{-6}/N)}
       \sim \frac{N}{6\times10^{-6}}.
\]
Its growing parameter time scale has zero overlap with the specified
input-law emission. This invisibility is limited to that declared
input support; it is not a statement about arbitrary prompts or the
full RefinedWeb distribution.

The finite count laws, conditional source coordinates and paired
time observations together specify the measured scaling regime.
An intrinsic critical exponent would additionally require a
controlled limiting law at an independently identified critical
surface. Self-organized criticality would further require native
restoring feedback toward that surface, the appropriate shrinking
window and forcing-memory control, and nonvanishing predictive
overlap for an inference claim. These requirements are distinct from
row concentration, common-head coherence, a rate--time power or a
slow parameter coordinate.

\enlargethispage{2\baselineskip}
\FloatBarrier
\par\medskip\noindent
Chapter~\ref{ch:physical-calibration} connects model observations to
calibrated physical laws. It asks which error budgets are required for an
internal or predictive readout to retain fluctuations at the physical scale.

\chapter{Physical calibration and the fluctuation readout budget}
\label{ch:physical-calibration}
This chapter studies conditional calibration against physical fluctuation
laws and the accuracy needed to observe them. It derives transport and
readout budgets, then presents finite calibration and size-transfer
experiments that separate physical source behavior from model observation error.

\section{Physical universality as a conditional calibration law}
\label{model:sec:physical-calibration}

Known universality classes provide a controlled distribution on which to
test the observation and scale maps. They do not identify the universality
class of an optimizer. Let $P_{q,L,T}$ be the equilibrium law on a periodic
$L\times L$ square lattice, with spins $s_i\in\{0,\ldots,q-1\}$ and
\begin{equation}
 H(s)=-\sum_{\langle i,j\rangle}\mathbf1_{\{s_i=s_j\}},\qquad
 P_{q,L,T}(s)=Z_{q,L,T}^{-1}\exp[-H(s)/T].
 \label{model:eq:physical-gibbs}
\end{equation}
Each right and down bond is counted once. The cases $q=2$ and $q=3$
have continuous transitions at $T_c(q)=1/\log(1+\sqrt q)$. The $q=2$
Hamiltonian is equivalent to the spin-product Ising Hamiltonian after a
factor-two temperature conversion and an additive energy constant. At
fixed $q$, dimensionality, interaction range and boundary convention,
the exact critical ratios are
\begin{equation}
 (\beta/\nu,\gamma/\nu,\nu)=
 \begin{cases}(1/8,7/4,1),&q=2,\\
 (2/15,26/15,5/6),&q=3.
 \end{cases}
 \label{model:eq:physical-targets}
\end{equation}
Finite-size corrections remain part of an empirical comparison with
these limiting targets~\cite{xu2025corrections}.

The pretraining distribution remains the specified RefinedWeb law
$\D$. Physical adaptation introduces a separate, stated distribution
$\D_{\mathrm{phys}}$ on $(q,L,T,s)$. For a native model of width $N$
at adaptation update $t$, write $Q_{N,t;q,L,T}$ for the generated
configuration law. Its conditional probabilities are the native
proper-prefix logits normalized on a fixed $q$-token spin alphabet.
The prefix contains the physical metadata and preceding spins in the
declared serialization. This defines a positive normalized law even
though the unrestricted language-model vocabulary is larger. The
parameter $L$, the native width $N$, the adaptation clock $t$ and the
Monte Carlo sampling clock label different limits.

A joint asymptotic family specifies $N=N(L)$, adaptation age $t=t(L)$,
remaining-corpus size and a context capacity sufficient for every
$L^2$-site proper prefix. The RefinedWeb pretraining law remains fixed
at the declared conditioning level, and physical adaptation uses
specified corpus-resource and refill laws. The error-rate hypotheses below apply to
such a family. Varying a physical input size at one finite checkpoint
alone does not define a native thermodynamic limit.

\subsection{From proper-prefix approximation to physical moments}

\begin{proposition}[Finite prefix-law control]
\label{model:prop:physical-prefix-control}
Let $q\ge2$ and $V\ge1$ be integers, and let $P$ and $Q$ be strictly positive laws on $\{0,\ldots,q-1\}^V$.
For a fixed ordering define
\[
 d_j=\E_{s_{<j}\sim P}
       \KL\!\left(P(s_j\mid s_{<j})\,\Vert\,
                    Q(s_j\mid s_{<j})\right),\qquad
 \bar d=V^{-1}\sum_{j=1}^V d_j.
\]
Then $\KL(P\Vert Q)=V\bar d$. For every $F$ with $0\le F\le1$,
\begin{equation}
 |\E_PF-\E_QF|\le\operatorname{TV}(P,Q)
       \le\min\{1,\sqrt{V\bar d/2}\}.
 \label{model:eq:physical-prefix-bound}
\end{equation}
\end{proposition}
\begin{proof}
Factor each joint probability into its ordered conditional
probabilities. Taking the logarithm of their ratio, averaging with
respect to $P$ and summing the finite terms gives the chain identity.
Let $A=\{s:P(s)\ge Q(s)\}$. The sum of $P-Q$ on $A$ equals
$\delta=\operatorname{TV}(P,Q)$, and its sum on the complement is
$-\delta$. Multiplication by any number in $[0,1]$ therefore bounds the
expectation difference by $\delta$. Grouping the relative entropy into
$A$ and its complement and applying convexity of $-\log$ bounds it
below by the Bernoulli relative entropy with parameters
$p=P(A)$ and $a=Q(A)$. As a function of $p$ this entropy is zero with
zero first derivative at $p=a$, and its second derivative is
$1/[p(1-p)]\ge4$. Twice integrating gives a lower bound
$2(p-a)^2=2\delta^2$. The endpoint cases follow by continuity.
Together with $\delta\le1$ this proves the claim.
\end{proof}

For color fractions $f_a=V^{-1}\sum_i\mathbf1_{\{s_i=a\}}$, define
\begin{equation}
 m^2=\frac{q\sum_a f_a^2-1}{q-1},\qquad
 \chi_L=L^2\E[m^2],\qquad
 B_L=\frac{\E[m^4]}{\E[m^2]^2}.
 \label{model:eq:physical-observables}
\end{equation}
Here $m\in[0,1]$. For a color-symmetric zero-field law, $\chi_L$ is the
total order-parameter covariance up to a fixed convention-dependent
factor. If generated colors are biased, the uncentered second moment
and the connected covariance must be reported separately. A fixed-size
histogram moment is not by itself a spatial correlation measurement.
Writing $v=\sqrt{q/(q-1)}(f-q^{-1}{\bf1})$, the exact decomposition is
\[
 \chi_L=L^2\operatorname{tr}\Cov(v)+L^2\|\E v\|^2,
\]
obtained by expanding $v=(v-\E v)+\E v$ and averaging the squared
norm. A persistent mean-color sector can therefore dominate an
uncentered susceptibility. If its norm is asymptotic to a positive
constant times $L^{-y}$ and the connected term has exponent $\kappa$,
the uncentered exponent is $\max\{\kappa,2-2y\}$ when both leading
coefficients are positive. Finite generated-sample color imbalance
also contains Monte Carlo error and is not by itself evidence of a
nonzero population field.

\begin{proposition}[Spatial covariance and position-dependent color means]
\label{model:prop:physical-spatial-centering}
For any configuration law on $V=L^2$ sites with $q\ge2$, set
$\pi_i(a)=\Pr(s_i=a)$, $I_i^a=\mathbf1_{\{s_i=a\}}$ and
\begin{alignat}{2}
 C_{ij}&=\frac{q\Pr(s_i=s_j)-1}{q-1},&\quad K_{ij}&=\frac{q}{q-1}\sum_a\Cov(I_i^a,I_j^a),\\
 H_{ij}&=\frac{q}{q-1}\sum_a
       \left(\pi_i(a)-\frac1q\right)
       \left(\pi_j(a)-\frac1q\right).
 \label{model:eq:physical-spatial-centering}
\end{alignat}
Then $C_{ij}=K_{ij}+H_{ij}$. The same identity holds after averaging
pairs at any fixed periodic displacement. Moreover,
\[
 \chi_L=\frac1V\sum_{ij}K_{ij}
       +V\frac{q}{q-1}\left\|\frac1V\sum_i\pi_i-\frac1q\mathbf1\right\|^2.
\]
The first term is the connected susceptibility. If $\pi_i$ depends on
position, subtracting only the squared global color mean generally does
not center a correlation at a fixed displacement.
\end{proposition}
\begin{proof}
The identity $\sum_a I_i^a I_j^a=\mathbf1_{\{s_i=s_j\}}$ and the
definition of covariance give
$\sum_a\Cov(I_i^a,I_j^a)=\Pr(s_i=s_j)-\sum_a\pi_i(a)\pi_j(a)$.
Since both probability vectors sum to one, expanding the centered
inner product gives
$\sum_a(\pi_i(a)-1/q)(\pi_j(a)-1/q)=\sum_a\pi_i(a)\pi_j(a)-1/q$.
Combining these equalities proves $C=K+H$ entrywise. Pair averaging
preserves this equality. Summing $H$ over all ordered pairs gives the
squared norm of the summed color-mean contrast. Dividing the sum of $K$ by $V$ gives
$Vq/(q-1)\sum_a\Var(f_a)$, the stated connected susceptibility.
A fixed-displacement sum retains products of the two position-specific
means and need not equal their global squared mean.
\end{proof}

The identity also holds exactly for an empirical configuration law.
It separates spatial covariance from color-mean profiles without
assuming translation invariance of the native decoder. A small raw
correlation error can conceal cancellation between covariance error
and a learned mean-field contribution. Both coordinates therefore enter
a physical fluctuation comparison; their sum alone does not determine
connected critical behavior.

A finite thermal response requires control of a ratio of moments at
\emph{both} temperatures. Let $m$ be the normalized color-order magnitude,
$0\le m\le1$. For a reference law $P$ and a generated law $Q$, put
\[
 a=\E_Pm^2,\quad b=\E_Pm^4,\quad
 c=\E_Qm^2,\quad d=\E_Qm^4,\qquad
 R_P=b/a^2,\quad R_Q=d/c^2.
\]
The Binder convention here is the moment ratio $R$, without an additive
constant or a reversed sign.

\begin{proposition}[Moment-ratio and finite thermal-contrast budgets]
\label{model:prop:thermal-transport}
Suppose $a,c>0$. Then
\begin{align}
 R_Q-R_P
 &=\frac{a^2(d-b)-b(c-a)(c+a)}{a^2c^2},\label{model:eq:thermal-signed}\\
 |R_Q-R_P|
 &\le D(P,Q):=\frac{|d-b|}{c^2}
       +\frac{b|c-a|(c+a)}{a^2c^2}.\label{model:eq:thermal-ratio}
\end{align}
If $|c-a|\le\epsilon_2a$ and $|d-b|\le\epsilon_4b$, where
$0\le\epsilon_2<1$ and $\epsilon_4\ge0$, then also
\[
 |R_Q-R_P|\le
 R_P\frac{\epsilon_4+2\epsilon_2+\epsilon_2^2}{(1-\epsilon_2)^2}.
 \label{model:eq:thermal-relative}
\]
At temperature ratios $1-h$ and $1+h$, $h>0$, define
$T_h(P)=(R_{P,+}-R_{P,-})/(2h)$, and similarly for $Q$.
Any valid endpoint bounds $|R_{Q,\pm}-R_{P,\pm}|\le D_\pm$ imply
\[
 |T_h(Q)-T_h(P)|\le\frac{D_++D_-}{2h}.\label{model:eq:thermal-contrast}
\]
\end{proposition}
\begin{proof}
Subtract the two ratios over the positive common denominator $a^2c^2$.
The numerator is $a^2(d-b)-b(c^2-a^2)$, giving
\eqref{model:eq:thermal-signed}. The triangle inequality and
$c^2-a^2=(c-a)(c+a)$ give \eqref{model:eq:thermal-ratio}.
For the relative bound, use $c\ge(1-\epsilon_2)a>0$,
$c+a\le(2+\epsilon_2)a$, and the two assumed moment errors in that
inequality. Its right side is at most
$R_P[\epsilon_4+\epsilon_2(2+\epsilon_2)]/(1-\epsilon_2)^2$.
Finally, the difference of the two thermal contrasts is the difference
of the two endpoint ratio errors divided by $2h$. The triangle
inequality and positivity of $2h$ prove the last statement.
\end{proof}

The signed identity retains cancellation between second- and fourth-moment
errors. The absolute bound is a sufficient error budget and can be loose.
It is a useful prediction only when its right side is small at the scale
of the observable being compared. If $h=h_L$ shrinks, vanishing endpoint
errors alone do not ensure a vanishing contrast error: a sufficient
condition is $D_{L,+}+D_{L,-}=o(h_L)$. For relative transport of a nonzero
reference contrast the corresponding condition is
$D_{L,+}+D_{L,-}=o(h_L|T_{h_L}(P_L)|)$.

These statements concern finite differences. If $R_P$ is three times
continuously differentiable in temperature ratio on $[1-h,1+h]$ and its
third derivative has absolute value at most $M$, Taylor expansion at one
with the two signed remainders gives
$|T_h(P)-R_P'(1)|\le Mh^2/6$. A size-dependent bound on $M$ is needed in
a simultaneous size and temperature-step limit. A thermal exponent
requires additional limiting scaling hypotheses and their empirical
tests.

For comparing an adapted checkpoint with its base, the estimand is
$T_h(Q_{\rm adapted})-T_h(Q_{\rm base})$. Its uncertainty is computed
from this direct difference. Separate confidence intervals for the two
responses do not provide an interval for their difference. Configuration
resampling conditions on selected checkpoints; variation across independently
trained checkpoints is a separate random component.

\begin{proposition}[Relative moment and two-size slope control]
\label{model:prop:physical-slope-control}
Let $u_L>0$ and $\widehat u_L>0$ be a reference moment and a learned
moment. If
$|\widehat u_s/u_s-1|\le\epsilon$, with $0\le\epsilon<1$, for $s=L,bL$, where $b>1$, then
\begin{equation}
 \left|\frac{\log(\widehat u_{bL}/\widehat u_L)}{\log b}
       -\frac{\log(u_{bL}/u_L)}{\log b}\right|
 \le\frac{\log[(1+\epsilon)/(1-\epsilon)]}{\log b}.
 \label{model:eq:physical-slope-bound}
\end{equation}
Consequently, vanishing relative moment error preserves an existing
two-size limiting slope. A uniformly bounded multiplicative factor
bounded away from zero is already sufficient to preserve an existing
logarithmic exponent $\lim_{L\to\infty}\log u_L/\log L$.
\end{proposition}
\begin{proof}
The assumption gives
$(1-\epsilon)u_s\le\widehat u_s\le(1+\epsilon)u_s$ and strict
positivity of the lower bound. Divide the upper bound at $bL$ by the
lower bound at $L$, and conversely. The ratio
$(\widehat u_{bL}/\widehat u_L)/(u_{bL}/u_L)$ lies between
$(1-\epsilon)/(1+\epsilon)$ and its reciprocal. Taking logarithms
proves the inequality. For the last assertion, the difference
$\log\widehat u_L-\log u_L$ is bounded, hence becomes zero after
division by $\log L$ in the limit.
\end{proof}

If $\E_Pm^2\asymp L^{-2x}$, where $x=\beta/\nu$, a sufficient
condition for vanishing relative error in $\chi_L$ is
$\operatorname{TV}(P_L,Q_L)=o(L^{-2x})$. By
Proposition~\ref{model:prop:physical-prefix-control}, the stronger condition
$\bar d_L=o(L^{-2-4x})$ suffices in two dimensions. These are sufficient
conditions, not necessary ones. Empirical cross entropy alone supplies
neither the unknown conditional entropy nor a bound on $\bar d_L$.
Direct finite-moment comparisons therefore remain necessary in the
reported calibration. Finite-size bias, Monte Carlo uncertainty and
learned-law error are separately identifiable contributions to a fitted
exponent.

\subsection{An architectural route for physical input fluctuations}

The first native query Gram already contains a possible magnetic
observation before physical adaptation. Its input is the tokenwise
normalized embedding, followed by the query projection and orthogonal
rotary position maps. Metadata have a fixed contribution for a fixed
physical setting and prefix length.

\begin{proposition}[First-Gram histogram visibility]
\label{model:prop:physical-gram}
For head $h$, let $a_{ha}$ be the first-layer query vector for color
$a$ before an orthogonal positional rotation. A prefix has $j\ge1$
spin tokens with fractions $f_a$ and fixed metadata. Write $D_h$ for
its unnormalized query Gram, $d_h$ for the metadata contribution to
its trace and $C_{ha}=\|a_{ha}\|^2$. Then
\begin{equation}
 t_h:=j^{-1}(\tr D_h-d_h)=\sum_a C_{ha}f_a.
 \label{model:eq:physical-gram-histogram}
\end{equation}
If $C$ is injective on the zero-sum color subspace, these head traces
and the known coefficients recover the entire fraction vector.
For any law of the prefix fractions,
$\Cov(t)=C\Cov(f)C^{\mathsf T}$. Uniform positive lower and finite
upper singular-value bounds on that subspace preserve the covariance
scaling exponent of an existing family of color-symmetric input laws.
\end{proposition}
\begin{proof}
The trace of a Gram matrix is the sum of squared query norms.
Orthogonal rotation preserves each norm, so grouping the spin tokens
by color and separating metadata gives
$\tr D_h=d_h+j\sum_a C_{ha}f_a$. Put
$u=q^{-1}{\bf1}$. Since $f-u$ has zero sum,
$t-Cu=C(f-u)$, and a left inverse of $C$ on that subspace recovers
$f-u$. Centering this identity and multiplying its two factors gives
the covariance formula. If the restricted singular values lie between
$c_-$ and $c_+$, then
\[
 c_-^2\E\|f-\E f\|^2
 \le\E\|t-\E t\|^2
 \le c_+^2\E\|f-\E f\|^2.
\]
Constants bounded away from zero and infinity preserve a logarithmic
scaling exponent, as in
Proposition~\ref{model:prop:physical-slope-control}.
\end{proof}

At the last proper prefix, $j=L^2-1$. Its fraction vector differs
from the full configuration's vector by at most $2/L^2$ in $\ell^1$,
since $f_{\rm full}=[(L^2-1)f_{\rm prefix}+e_{s_{L^2}}]/L^2$.
Consequently the squared magnetic invariant differs by $O(L^{-2})$.
The sharper fluctuation-scale argument in
Proposition~\ref{model:prop:prefix-readout-budget} controls relative moments
when $x<2$ and includes an explicit Gram-inversion arithmetic term.
The identity is before Gram normalization: neither LayerNorm nor the
nonlinear metric generator is assumed injective. Their retained
common-row visibility is an empirical question. This mechanism can
expose an externally critical input law even at random initialization;
it does not imply a critical generated law or a training transition.

\subsection{Bounded magnetic coordinates from internal observations}

A linear readout of an invariant need not preserve the underlying
order-parameter geometry. Let $\Delta_q$ be the probability simplex,
$\phi$ a retained native observation and $\widehat f(\phi)$ a fitted
color-fraction vector projected in Euclidean distance onto $\Delta_q$.
Define
\[
 v=\sqrt{q/(q-1)}(f-q^{-1}{\bf1}),\qquad
 \widehat v=\sqrt{q/(q-1)}(\widehat f-q^{-1}{\bf1}).
\]
Both vectors belong to the same $(q-1)$-dimensional zero-sum space,
and $m^2=\|v\|^2$. The reconstructed invariant
$\widehat m^2=\|\widehat v\|^2$ respects the color-fraction geometry.
This construction does not assume that the fitted native readout itself
is equivariant under relabeling of the input tokens.

\begin{proposition}[Bounded invariant reconstruction]
\label{model:prop:physical-invariant}
For any $f,\widehat f\in\Delta_q$, $\|v\|,\|\widehat v\|\le1$ and
\begin{equation}
 |\widehat m^2-m^2|\le2\|\widehat v-v\|,\qquad
 |\E\widehat m^2-\E m^2|
 \le2\bigl(\E\|\widehat v-v\|^2\bigr)^{1/2}.
 \label{model:eq:physical-invariant-bound}
\end{equation}
Thus a sufficient condition for this internal readout to preserve an
existing magnetic second-moment exponent with
$\E m_L^2\asymp L^{-2x}$ is a vector root-mean-square error
$o(L^{-2x})$.
\end{proposition}
\begin{proof}
For a probability vector, $\sum_a f_a^2\le(\sum_a f_a)^2=1$.
Consequently $\|v\|^2=(q\sum_a f_a^2-1)/(q-1)\le1$, and the same
holds for $\widehat v$. The difference of squared norms is
$\langle\widehat v-v,\widehat v+v\rangle$. Cauchy--Schwarz and the
triangle inequality bound its absolute value by $2\|\widehat v-v\|$.
Averaging and applying Cauchy--Schwarz again proves the second
inequality. Division by the reference moment gives vanishing relative
error under the stated rate, so
Proposition~\ref{model:prop:physical-slope-control} applies.
\end{proof}

Readout error under a reference configuration law and error of the
model's generated law are separate quantities. The first tests whether
native variables retain a physical direction; the second tests whether
the trained conditional probabilities reproduce its population law.
Both error budgets are required before transferring a physical exponent
through a learned generative representation.

\subsection{A joint error budget for physical scale transport}

\begin{proposition}[Bounded physical-law and observation transport]
\label{model:prop:physical-transport}
Let $P,Q$ be laws on a finite configuration set, and let $B$ be a
blocking kernel with $P'=B_*P$. Let $v,\widehat v$ on the fine set
and $v'$ on the coarse set take values in a subset $K$ of a normed
space's closed unit ball. Let $S:K\to K$ be $k$-Lipschitz. Define
\[
 \delta=\operatorname{TV}(P,Q),\qquad
 e=\E_P\|\widehat v-v\|,\qquad
 c=\E_{(x,x')\sim P(dx)B(x,dx')}\|S(v(x))-v'(x')\|.
\]
For the Wasserstein distance with this norm cost,
\begin{equation}
 W_1(S_*\widehat v_*Q,v'_*P')\le k(2\delta+e)+c.
 \label{model:eq:physical-transport-budget}
\end{equation}
For two successive physical blocking kernels and two such retained
maps, with constants $k_1,k_2$ and defects $c_1,c_2$ under their actual
successive physical laws, the error is at most
$k_2k_1(2\delta+e)+k_2c_1+c_2$.
\end{proposition}
\begin{proof}
Couple the common mass $\min\{P(x),Q(x)\}$ at each configuration
identically and couple the remaining disjoint masses arbitrarily. This
finite coupling agrees with probability $1-\delta$.
Applying the same map $\widehat v$ to its two coordinates gives equal
vectors when they agree and distance at most two otherwise. Hence
$W_1(\widehat v_*Q,\widehat v_*P)\le2\delta$. Coupling $v$ and
$\widehat v$ on the same draw from $P$ adds at most $e$. Transport
of a coupling by a $k$-Lipschitz map multiplies its expected cost by
at most $k$. Finally, the stated fine/coarse coupling gives cost at
most $c$ between $S_*v_*P$ and $v'_*P'$. The triangle inequality
proves the first result. Apply that result's coupling argument at the
second scale to multiply the incoming error by $k_2$ and add $c_2$.
\end{proof}

This budget keeps generated-law error, native readout error and
physical block closure separate. Its blocked reference is $B_*P$;
replacing that law by a nearest-neighbor Gibbs family at a named
temperature needs an additional coupling-closure bound. On a finite
paired empirical reference, root-mean-square vector errors give valid
upper bounds by Cauchy--Schwarz and Minkowski. The executed transport
check uses an affine map of color fractions followed by Euclidean
simplex projection. Projection is nonexpansive: applying the two
projection variational inequalities to one another and adding them
gives $\|\Pi x-\Pi y\|^2\le\langle\Pi x-\Pi y,x-y\rangle$.
Thus the affine matrix norm bounds the fitted map's Lipschitz constant.
The empirical check does not supply the unknown large-lattice
population value of $\delta$.
Section~\ref{model:sec:physical-transport-results} reports the finite physical calibration and held-out blocking comparisons.

\subsection{Compatible internal variables and RG eigenvalues}

A physical blocking kernel, such as color-symmetric majority blocking,
pushes forward the whole configuration law. Its repeated applications
compose exactly as kernels, but a short list of effective couplings
requires a closure argument. The same distinction applies to a PLDR
representation of that law.

\begin{proposition}[Visibility under a compatible RG representation]
\label{model:prop:physical-intertwining}
Let $U,W$ be finite-dimensional real normed vector spaces. Let
$R:U\to U$ and $S:W\to W$ be differentiable scale maps, with
$R(u_*)=u_*$, and let $\phi:U\to W$ be differentiable near $u_*$.
Suppose $S\circ\phi=\phi\circ R$ there. Write
$A=DR(u_*)$, $C=D\phi(u_*)$ and $B=DS(\phi(u_*))$. Then
$BC=CA$. If $Av=\lambda v$ and $Cv\ne\bm0$, the represented direction
$Cv$ is an eigenvector of $B$ with eigenvalue $\lambda$.
More generally, if $\|BC-CA\|\le\delta$ and
$\|Cv\|\ge a\|v\|>0$, then
\begin{equation}
 \frac{\|B(Cv)-\lambda Cv\|}{\|Cv\|}\le\frac{\delta}{a}.
 \label{model:eq:physical-intertwining-residual}
\end{equation}
\end{proposition}
\begin{proof}
The chain rule at $u_*$ gives $BC=CA$. Substitution of $Av=\lambda v$
then gives $B(Cv)=C(Av)=\lambda Cv$. For the approximate assertion,
$B(Cv)-\lambda Cv=(BC-CA)v$; the operator-norm bound and the stated
lower bound on $\|Cv\|$ give the result.
\end{proof}

For a nonnormal $B$, a small eigenvector residual need not imply a
nearby eigenvalue without an additional conditioning bound. Moreover,
the correspondence must be tested on perturbations and held-out
configurations. Prediction of an order parameter alone does not
establish $S\circ\phi=\phi\circ R$. In a physical thermal and magnetic
scaling basis, $\lambda_t=b^{y_t}$ and $\lambda_h=b^{y_h}$ imply
$\nu=1/y_t$, $\beta=(2-y_h)/y_t$ and
$\gamma=(2y_h-2)/y_t$, under the usual homogeneous free-energy
scaling assumptions. Attention eigenvalues and temporal optimizer
relaxation factors are different quantities.

\subsection{A spatial control that preserves bulk critical moments}

\begin{proposition}[Histogram preservation under spatial shuffling]
\label{model:prop:physical-shuffle}
For any configuration law and any possibly configuration-dependent
random permutation of its sites, the complete distribution of
$(f_0,\ldots,f_{q-1})$ is unchanged. Hence every moment of $m$, the
quantity $\chi_L$ and the Binder ratio in
Equation~\eqref{model:eq:physical-observables} are unchanged whenever defined.
\end{proposition}
\begin{proof}
A permutation reorders the finite summands defining each $f_a$ and
preserves their sum for every configuration and every realized
permutation. Every function of that histogram is therefore preserved
pointwise, which proves equality of the corresponding distributions
and moments after averaging.
\end{proof}

Spatially separated spin correlations need not be preserved. This
control tests whether apparent exponent recovery can be explained by
learning a distribution of global histograms while losing the physical
geometry. It also separates a learned inference law from a training
transition. Providing examples at a known $T_c$ imposes a critical
source distribution; an endogenous training transition additionally
requires an independently identified critical surface and a mechanism
that drives training toward it.

\section{Magnetic readout accuracy at the fluctuation scale}
\label{model:sec:readout-budget}

Fix a model checkpoint, a configuration law $P_L$, and a fitted readout.
All expectations in this section condition on that checkpoint and fit.
For color fractions $f_L$ and their Euclidean-simplex-projected estimate
$\widehat f_L$, put
\[
 v_L=\sqrt{q/(q-1)}(f_L-q^{-1}{\bf1}),\qquad
 \widehat v_L=\sqrt{q/(q-1)}(\widehat f_L-q^{-1}{\bf1}).
\]
Both vectors have norm at most one. Define
$\mu_L=\E_{P_L}\|v_L\|^2>0$,
$e_L=\widehat v_L-v_L$, and
$\delta_L=(\E_{P_L}\|e_L\|^2)^{1/2}$.
The ratio $r_L=\delta_L/\sqrt{\mu_L}$ compares reconstruction error
with the amplitude of the reference order parameter. The sufficient
second-moment condition is $r_L\to0$, which preserves an existing slope.
Connected variance, generated-law transport, and fourth-moment or Binder
ratios require their own centered, distributional, or higher-moment controls.

\begin{proposition}[Second-moment reconstruction budget]
\label{model:prop:readout-refinement}
For these vectors,
\begin{align}
 \E\|\widehat v_L\|^2-\mu_L
   &=2\E\langle v_L,e_L\rangle+\delta_L^2,\label{model:eq:readout-signed}\\
 \left|\E\|\widehat v_L\|^2-\mu_L\right|
   &\le\min\{2\delta_L,\,2\sqrt{\mu_L}\delta_L+\delta_L^2\},
   \label{model:eq:readout-min}\\
 \left|\frac{\E\|\widehat v_L\|^2}{\mu_L}-1\right|
   &\le 2r_L+r_L^2.\label{model:eq:readout-relative}
\end{align}
If $\mu_L\asymp L^{-2x}$, then $\delta_L=o(L^{-x})$ suffices
for a vanishing relative second-moment error. It preserves an existing
limiting two-size slope of $\mu_L$, and hence that of $L^2\mu_L$.
It does not assert existence of either limiting slope.
\end{proposition}
\begin{proof}
Expand $\|v_L+e_L\|^2$ and average to obtain
\eqref{model:eq:readout-signed}. Cauchy--Schwarz on the joint probability
space gives
$|\E\langle v_L,e_L\rangle|\le\sqrt{\mu_L}\delta_L$.
Separately,
$|\|\widehat v_L\|^2-\|v_L\|^2|
\le\|\widehat v_L-v_L\|\,\|\widehat v_L+v_L\|
\le2\|e_L\|$; averaging and applying Cauchy--Schwarz bounds it by
$2\delta_L$. Both inequalities hold, so their minimum holds.
Division by $\mu_L$ proves \eqref{model:eq:readout-relative}.
The stated rate implies $r_L\to0$. Write the reconstructed moment
as $\mu_L(1+\eta_L)$, with $\eta_L\to0$.
For any fixed $b>1$ its two-size logarithmic slope differs from
that of $\mu_L$ by
$[\log(1+\eta_{bL})-\log(1+\eta_L)]/\log b$, which tends to zero.
\end{proof}

The scale-dependent bound need not be smaller than $2\delta_L$ at
every finite size. Taking the minimum avoids an unnecessary finite-size
loss while retaining the weaker asymptotic rate. Cancellation of the
signed cross term in \eqref{model:eq:readout-signed} can make the observed
moment error much smaller than either sufficient bound.
Section~\ref{model:sec:readout-results} evaluates both bounds and the signed error on finite native readouts and fresh physical sizes.

\begin{proposition}[Connected fluctuations]
\label{model:prop:connected-refinement}
Let $\kappa_L=\E\|v_L-\E v_L\|^2$ and
$\delta_{c,L}=\|e_L-\E e_L\|_{L^2}$. Then
\[
 \left|\operatorname{tr}\operatorname{Cov}(\widehat v_L)
       -\operatorname{tr}\operatorname{Cov}(v_L)\right|
 \le2\sqrt{\kappa_L}\delta_{c,L}+\delta_{c,L}^2,
 \qquad \delta_{c,L}\le\delta_L.
\]
For $\kappa_L>0$, the sufficient relative-error condition is
$\delta_{c,L}=o(\sqrt{\kappa_L})$.
\end{proposition}
\begin{proof}
Center both vectors. Their difference is $e_L-\E e_L$.
The expansion and Cauchy--Schwarz argument above applies to these
centered vectors. Finally,
$\delta_{c,L}^2=\delta_L^2-\|\E e_L\|^2\le\delta_L^2$.
Dividing by $\kappa_L$ gives the relative statement.
\end{proof}

If a mean-color sector dominates $\mu_L$, controlling error relative
to $\sqrt{\mu_L}$ alone does not control the smaller connected
fluctuation. The two budgets must be assessed separately.

\begin{proposition}[Generated-law and readout errors]
\label{model:prop:joint-refinement}
Let $Q_L$ be a second law on the same configurations and let
$\Delta_L=\operatorname{TV}(P_L,Q_L)$, with total variation defined
as one half the sum of absolute probability differences. For the
same bounded readout under both laws,
\[
 \left|\E_{Q_L}\|\widehat v_L\|^2-\mu_L\right|
 \le\Delta_L+\min\{2\delta_L,
                    2\sqrt{\mu_L}\delta_L+\delta_L^2\}.
\]
Thus $\Delta_L=o(\mu_L)$ and $\delta_L=o(\sqrt{\mu_L})$
jointly suffice for relative moment transport.
\end{proposition}
\begin{proof}
Since $0\le\|\widehat v_L\|^2\le1$, its expectation changes by
at most $\Delta_L$ between $P_L$ and $Q_L$.
Add and subtract $\E_{P_L}\|\widehat v_L\|^2$ and apply
\eqref{model:eq:readout-min}. Division by $\mu_L$ proves the last assertion.
\end{proof}

This separates internal visibility from generative fidelity. In two
dimensions the proper-prefix KL identity
$\operatorname{KL}(P_L\Vert Q_L)=L^2\bar d_L$ and Pinsker's
inequality give the sufficient law condition
$\bar d_L=o(L^{-2-4x})$ when $\mu_L\asymp L^{-2x}$.
The weaker readout condition does not weaken that separate TV-based
condition. Cross entropy alone does not measure $\bar d_L$ without
the source conditional entropy.

Second-moment transport also does not by itself transport a Binder
ratio. An adequate additional condition is
$\|e_L\|_{L^4}=o(L^{-x})$ and
$\E\|v_L\|^4\asymp L^{-4x}$, together with the second-moment
condition. Indeed, the reverse triangle inequality in $L^4$ bounds
the change of the fourth-moment fourth root by $\|e_L\|_{L^4}$.
The resulting vanishing relative fourth-moment error, divided by the
squared second moment, preserves an existing Binder limit.

For a generated-law Binder ratio, the separate sufficient condition is
$\operatorname{TV}(P_L,Q_L)=o(L^{-4x})$, since the fourth moment is
also a bounded observable. Second-moment accuracy under $P_L$ does not
imply that stronger generated-law condition.

\subsection{A constructive proper-prefix readout}

The first-Gram identity gives a readout whose acquisition consists of
known token-query coefficients, rather than a fit to magnetic targets.
This exposes a concrete source of size dependence in the observation
map and a quantitative arithmetic error.

\begin{proposition}[Prefix omission and Gram inversion]
\label{model:prop:prefix-readout-budget}
Let $V=L^2>1$, $j=V-1$, and let $U$ have orthonormal columns spanning
the zero-sum subspace of $\R^q$. In the setting of
Proposition~\ref{model:prop:physical-gram}, suppose $CU$ has smallest singular
value $\sigma_L>0$. The measured normalized trace is
$\widetilde t=Cf_{\rm prefix}+\varepsilon$. Put $u=q^{-1}{\bf1}$ and
\[
 \widehat f=\Pi_{\Delta_q}\bigl[u+U(CU)^+
                      (\widetilde t-Cu)\bigr],
\]
where $(CU)^+$ is the left pseudoinverse and $\Pi_{\Delta_q}$ is
Euclidean projection onto the probability simplex. For every configuration,
\begin{equation}
 \|\widehat v-v_{\rm full}\|
 \le\sqrt{\frac q{q-1}}\left(\frac{\sqrt2}{L^2}
                               +\frac{\|\varepsilon\|}{\sigma_L}\right).
 \label{model:eq:prefix-readout-budget}
\end{equation}
Consequently, under any configuration law,
\[
 \delta_L\le\sqrt{\frac q{q-1}}
 \left(\frac{\sqrt2}{L^2}+\frac{\|\varepsilon\|_{L^2}}{\sigma_L}\right).
\]
For a context-compatible family with $\mu_L\asymp L^{-2x}$ and
$x<2$, exact arithmetic gives vanishing relative second-moment error.
The same conclusion with arithmetic error requires
$\|\varepsilon\|_{L^2}/\sigma_L=o(L^{-x})$.
\end{proposition}
\begin{proof}
Write $f_{\rm prefix}=u+Uz$. The unprojected reconstruction is
$f_{\rm prefix}+U(CU)^+\varepsilon$. The pseudoinverse has norm
$1/\sigma_L$, and Euclidean projection onto a closed convex set is
nonexpansive. Since $f_{\rm prefix}\in\Delta_q$,
$\|\widehat f-f_{\rm prefix}\|\le\|\varepsilon\|/\sigma_L$.
Also
$f_{\rm full}=[(V-1)f_{\rm prefix}+e_{s_V}]/V$.
The Euclidean diameter of the simplex is $\sqrt2$, so
$\|f_{\rm prefix}-f_{\rm full}\|\le\sqrt2/V$.
The triangle inequality and the definition of $v$ prove the
pointwise bound. Minkowski's inequality gives its $L^2$ form.
Dividing by $\sqrt{\mu_L}\asymp L^{-x}$ and using
Proposition~\ref{model:prop:readout-refinement} proves the last claims.
\end{proof}

For the two specified magnetic source classes, $x=1/8$ and $2/15$,
the omission condition holds. It is conditional on the existence of
the source scaling law and an available context at each size. The
finite checkpoints have a declared context capacity and do not realize
that infinite family. The inverse coefficients and metadata offset
are checkpoint- and setting-dependent. Injectivity before Gram
normalization makes no claim about injectivity of a normalized Gram,
common metric rows, spatial blocking, or generated configurations.

\section{Executed calibration with two physical universality classes}
\label{model:sec:physical-results}

Controlled-width experiments and adaptation of a qualified released
base answer complementary questions here. The controlled paths isolate
architecture and finite optimizer response. Section~\ref{model:sec:released-physical}
uses the long-pretrained model selected by independent validation and tests
next-spin competence together with native generated laws. The two
adaptation seeds share one released base and do not constitute a native
width or independent-pretraining ensemble.

The Ising and three-state Potts laws in
Section~\ref{model:sec:physical-calibration} supply physical targets with known
limiting exponents. The experiment separates equilibrium-source
calibration, learned configuration laws, internal physical readouts and
finite optimizer response. Conditioning a language model on a physical
critical law changes its adaptation distribution. It does not assert
that RefinedWeb or the optimizer has the same critical class.

\subsection{Physical source and native adaptation}

The periodic square-lattice sampler uses fixed numbers of Wolff cluster
updates~\cite{wolff1989collective}. Independent random-site heat-bath
chains and exact enumeration qualify the equilibrium moments at three
temperatures for $q=2,L=4$ and $q=3,L=3$. All 72 specified checks pass.
Every source cohort has distinct random seeds. Half the chains start
ordered and half disordered, with 4,096 burn-in updates and twelve
cluster updates between retained configurations. The stopping rule is
a fixed update count, not a state-dependent amount of flipped volume.
The Monte Carlo clock is an algorithmic sampling clock; its relaxation
exponent is not a physical Glauber exponent or an optimizer
exponent~\cite{swendsen1987nonuniversal}.

At $T_c$, the precision reference contains sixteen independent chains
of 8,192 retained configurations for each of
$L\in\{4,6,8,12,16,24,32,48,64\}$ and each class. The thermal reference
derivative is evaluated using
$\partial_T\E F=\operatorname{Cov}(F,H)/T^2$. Applied to the Binder
ratio, this retains both the fourth-moment derivative and its
second-moment denominator derivative. Logarithmic finite-size fits use
all sizes above each stated lower cutoff. Independent-chain bootstrap
intervals quantify sampling uncertainty; cutoff dependence measures a
different, systematic finite-size contribution. The limiting values in
Equation~\eqref{model:eq:physical-targets} are reference targets, not fitted
constraints.

The native family retains five decoders, 64 dimensions per head and
the full eight-unit metric generator. Four- and eight-head models
each have three complete initialization replicas. Each enters physical
adaptation after 40,960 single-pass RefinedWeb updates, corresponding to
83,886,080 supervised positions. The physical adaptation origin retains
the complete model weights and explicitly resets Adam moments. It is
conditioned on a separate corpus of both classes, sizes $4,8,16$ and
temperature ratios $0.85,0.925,1,1.075,1.15$.

Each of the eight paths executes 16,384 updates, with one uniformly
selected physical cell and one uniformly selected target site per
batch of 32 fresh configuration identities. A path never revisits
a $(\text{cell},\text{chain},\text{sample})$ identity. Equal physical
spin configurations can recur under the equilibrium law and are not
deduplicated. The six primary replicas and two paired four-head
controls share the frozen stream. One control holds the residual
metric-network parameters fixed; the other independently shuffles
sites within each configuration. The first still allows metric changes
through other trainable parameters. The second preserves the source
histogram law exactly, as in Proposition~\ref{model:prop:physical-shuffle}.

AdamW uses peak learning rate $3\times10^{-4}$, moments $(0.9,0.95)$,
numerical constant $10^{-8}$, weight decay $0.01$ and gradient-norm
clipping at one. A 128-update warmup precedes cosine decay to one fifth
of the peak. The supervised objective is native cross entropy on
the existing two- or three-token spin alphabet. Every call contains
only metadata and spins strictly before its target. There is no
padding or cache. This proper-prefix construction prevents the global
Gram computation from accessing the target or future spins. Selected
output-row evaluation agrees with full-vocabulary gathering within
the stated arithmetic qualification, including gradients.

The independent assessment uses sizes $4,6,8,12,16$ and temperature
ratios $0.94,0.97,1,1.03,1.06$. Sizes six and twelve are absent from
adaptation. Checkpoints at 512, 2,048 and 16,384 updates retain common
held-out native observations. At the terminal checkpoint, each critical
cell has 512 generated configurations and each off-critical cell has
128; each earlier critical cell has 128. All 560 selected checkpoint
cells are retained. Initialization variation is reported across the
three models at each width, separately from conditional generation
Monte Carlo error. Model temperature contrasts use the discrete
metadata settings. They are not analytic derivatives of tokenization.

\begin{table}[htbp]\centering\small
\caption{Unconstrained physical reference slopes through $L=64$. Brackets are 95\% independent-chain bootstrap intervals and do not include finite-size systematic error.}
\label{model:tab:physical-reference}
\begin{tabular}{@{}rrccc@{}}
\toprule $q$ & $L_{\min}$ & $\beta/\nu$ & $\gamma/\nu$ & $\nu$\\\midrule
2 & 4 & 0.1236 [0.1232, 0.1240] & 1.7540 [1.7533, 1.7547] & 0.9821 [0.9775, 0.9869]\\
2 & 8 & 0.1242 [0.1235, 0.1250] & 1.7521 [1.7509, 1.7533] & 0.9918 [0.9845, 0.9999]\\
2 & 16 & 0.1245 [0.1233, 0.1259] & 1.7513 [1.7493, 1.7531] & 0.9918 [0.9778, 1.0061]\\
3 & 4 & 0.1303 [0.1296, 0.1312] & 1.7447 [1.7434, 1.7460] & 0.7928 [0.7881, 0.7978]\\
3 & 8 & 0.1313 [0.1302, 0.1325] & 1.7414 [1.7395, 1.7433] & 0.8051 [0.7971, 0.8127]\\
3 & 16 & 0.1328 [0.1308, 0.1347] & 1.7381 [1.7349, 1.7414] & 0.8041 [0.7918, 0.8183]\\
\bottomrule
\end{tabular}
\end{table}

The physical reference shows distinct sampling and finite-size errors.
In particular, the thermal Potts estimate remains cutoff dependent;
its sampling interval alone is not a certificate of asymptotic exponent
recovery. The reference therefore calibrates finite slopes and their
drift, while the exact class exponents specify the limiting targets.

\begin{table}[htbp]\centering\small
\caption{Native generated-law finite slopes. Entries are means and ranges across all three initialization replicas at each width. The ranges are not confidence intervals; conditional generation-bootstrap intervals for every replica are retained in the compact data. Sizes extend only through sixteen.}
\label{model:tab:physical-native-slopes}
\begin{tabular}{@{}rrrcc@{}}
\toprule Heads & $q$ & $L_{\min}$ & $\beta/\nu$ & $\gamma/\nu$\\\midrule
4 & 2 & 4 & 0.102 [0.066, 0.144] & 1.805 [1.740, 1.856]\\
4 & 2 & 8 & 0.077 [-0.033, 0.165] & 1.846 [1.702, 2.020]\\
4 & 3 & 4 & 0.176 [0.154, 0.191] & 1.674 [1.644, 1.715]\\
4 & 3 & 8 & 0.176 [0.105, 0.219] & 1.667 [1.597, 1.798]\\
8 & 2 & 4 & 0.129 [0.066, 0.177] & 1.762 [1.682, 1.867]\\
8 & 2 & 8 & 0.097 [-0.005, 0.155] & 1.812 [1.719, 1.990]\\
8 & 3 & 4 & 0.244 [0.177, 0.289] & 1.569 [1.489, 1.680]\\
8 & 3 & 8 & 0.201 [0.044, 0.322] & 1.638 [1.447, 1.894]\\
\bottomrule
\end{tabular}
\end{table}

\begin{table}[htbp]\centering\small
\caption{Absolute generated-moment relative error over all five critical sizes and three replicas: median / maximum. Color bias is the maximum deviation of a mean color fraction from $1/q$.}
\label{model:tab:physical-native-error}
\begin{tabular}{@{}rrcccc@{}}
\toprule Heads & $q$ & $\E m$ & $\chi_L$ & $B_L$ & Color bias\\\midrule
4 & 2 & 3.10\% / 10.67\% & 4.61\% / 19.18\% & 2.52\% / 6.20\% & 0.344\\
4 & 3 & 3.13\% / 18.27\% & 5.27\% / 28.61\% & 2.85\% / 16.80\% & 0.419\\
8 & 2 & 2.56\% / 12.56\% & 4.70\% / 23.45\% & 2.22\% / 7.81\% & 0.241\\
8 & 3 & 5.63\% / 29.78\% & 9.54\% / 43.74\% & 5.50\% / 33.69\% & 0.234\\
\bottomrule
\end{tabular}
\end{table}

On the fixed development holdout, native proper-prefix risk decreases in 8 of eight paths. The largest primary critical-cell color bias is 0.4194; the largest relative difference between connected covariance and the uncentered $\chi_L$ is 74.16\%. The uncentered quantity is retained for comparison with its specified physical moment; color imbalance is reported separately. Complete off-critical outcomes, discrete-temperature contrasts and both finite-size windows are included in the compact evidence.

\begin{table}[htbp]\centering\small
\caption{Paired four-head spatial controls. Susceptibility is divided by the physical reference. Near and half-lattice correlations retain the declared color normalization; the complete data include their reference values. Spatial shuffling preserves source histogram moments but can change spatial correlations.}
\label{model:tab:physical-spatial}
\begin{tabular}{@{}rrlrrr@{}}
\toprule $q$ & $L$ & Arm & $\chi/\chi_P$ & $C(1)$ & $C(L/2)$\\\midrule
2 & 8 & full & 0.990 & 0.719 & 0.624\\
2 & 8 & fixed residual net & 1.034 & 0.746 & 0.652\\
2 & 8 & shuffled & 1.067 & 0.686 & 0.688\\
2 & 16 & full & 1.192 & 0.742 & 0.647\\
2 & 16 & fixed residual net & 0.937 & 0.677 & 0.507\\
2 & 16 & shuffled & 0.964 & 0.526 & 0.523\\
3 & 8 & full & 0.961 & 0.694 & 0.602\\
3 & 8 & fixed residual net & 0.899 & 0.674 & 0.556\\
3 & 8 & shuffled & 1.134 & 0.728 & 0.725\\
3 & 16 & full & 1.005 & 0.663 & 0.544\\
3 & 16 & fixed residual net & 0.625 & 0.547 & 0.332\\
3 & 16 & shuffled & 1.027 & 0.557 & 0.555\\
\bottomrule
\end{tabular}
\end{table}

\begin{table}[htbp]\centering\small
\caption{Fresh-chain reconstruction of magnetic second moments from common metric rows. Skill is one minus held-out squared error divided by the calibration-mean predictor error. Both readouts use the same eight PCA coordinates. The invariant first reconstructs all color fractions and then forms their bounded quadratic invariant. The final column is relative error in the mean magnetic second moment.}
\label{model:tab:physical-invariant}
\begin{tabular}{@{}rrrrrr@{}}
\toprule Heads & $q$ & $L$ & Linear skill & Invariant skill & Mean rel. error\\\midrule
4 & 2 & 8 & 0.155 & 0.793 & -1.66\%\\
4 & 2 & 16 & 0.718 & 0.746 & +6.71\%\\
4 & 3 & 8 & 0.296 & 0.512 & -8.98\%\\
4 & 3 & 16 & 0.353 & 0.621 & -3.51\%\\
8 & 2 & 8 & 0.665 & 0.973 & -0.72\%\\
8 & 2 & 16 & 0.809 & 0.903 & -3.05\%\\
8 & 3 & 8 & 0.674 & 0.920 & -1.94\%\\
8 & 3 & 16 & 0.403 & 0.823 & -8.14\%\\
\bottomrule
\end{tabular}
\end{table}

The invariant improves all 8 of 8 unblocked common-row comparisons; 6 paired chain-bootstrap intervals for its squared-error difference exclude zero. This assessment uses fresh simulation seeds after the invariant construction was fixed. All hidden-state, common-row and effective-metric controls, including both blocking depths, remain in the compact record. These are readout errors under the reference law, separate from the generative errors in Table~\ref{model:tab:physical-native-error}.

\begin{table}[htbp]\centering\small
\caption{Matched origin controls for the same common-row invariant and physical configurations. The entries are held-out skill. These additional descriptive controls reuse the fresh cohort and its unchanged calibration split. High magnetic readout skill can precede physical adaptation.}
\label{model:tab:physical-origins}
\begin{tabular}{@{}rrrrrr@{}}
\toprule Heads & $q$ & $L$ & Random & Pretrained & Adapted\\\midrule
4 & 2 & 8 & 0.964 & 0.967 & 0.793\\
4 & 2 & 16 & 0.971 & 0.844 & 0.746\\
4 & 3 & 8 & 0.923 & 0.594 & 0.512\\
4 & 3 & 16 & 0.935 & 0.543 & 0.621\\
8 & 2 & 8 & 0.976 & 0.902 & 0.973\\
8 & 2 & 16 & 0.971 & 0.873 & 0.903\\
8 & 3 & 8 & 0.933 & 0.755 & 0.920\\
8 & 3 & 16 & 0.926 & 0.823 & 0.823\\
\bottomrule
\end{tabular}
\end{table}

The native first-Gram check covers all 72 selected origin, class, size and prefix cases. The largest relative trace discrepancy is \ensuremath{3.71\times10^{-7}}, and the largest recovered color-fraction error is \ensuremath{3.8\times10^{-6}}. The color-tangent matrix has rank $q-1$ in every case. This directly checks Proposition~\ref{model:prop:physical-gram} before Gram normalization. The origin controls in Table~\ref{model:tab:physical-origins} separate architectural visibility from the learning of a generated configuration law.

\begin{table}[htbp]\centering\small
\caption{Bounded magnetic transport on the fresh fine/coarse pairs. All entries use normalized color-vector RMS error. The prediction bound is the fitted map Lipschitz constant times fine readout error plus physical closure. Route discrepancy compares transported fine readout and direct coarse readout. The reference is the actual majority-blocked law.}
\label{model:tab:physical-transport}
\begin{tabular}{@{}rrrrrrr@{}}
\toprule Heads & $q$ & $L$ & Closure & Prediction & Bound & Routes\\\midrule
4 & 2 & 8 & 0.086 & 0.106 & 0.179 & 0.133\\
4 & 2 & 16 & 0.040 & 0.083 & 0.123 & 0.101\\
4 & 3 & 8 & 0.076 & 0.162 & 0.237 & 0.159\\
4 & 3 & 16 & 0.049 & 0.164 & 0.208 & 0.219\\
8 & 2 & 8 & 0.086 & 0.086 & 0.122 & 0.103\\
8 & 2 & 16 & 0.040 & 0.064 & 0.086 & 0.067\\
8 & 3 & 8 & 0.076 & 0.090 & 0.141 & 0.096\\
8 & 3 & 16 & 0.049 & 0.113 & 0.152 & 0.123\\
\bottomrule
\end{tabular}
\end{table}

All 24 observation-family transport checks satisfy the coupled empirical bounds of Proposition~\ref{model:prop:physical-transport}. The common-row predictions have the errors in Table~\ref{model:tab:physical-transport}; these checks leave generated-law population error as a separate term.

\begin{table}[htbp]\centering\small
\caption{Held-out relative mean-square defects of the first fitted physical block map. Ranges include both classes and $L=8,16$. The denominator is the calibration-mean predictor error on held-out chains. These finite fitted maps have measured closure defects, not identified RG eigenvalues.}
\label{model:tab:physical-closure}
\begin{tabular}{@{}rlrr@{}}
\toprule Heads & Observation & Minimum & Maximum\\\midrule
4 & Hidden & 0.078 & 0.286\\
4 & Common rows & 0.288 & 0.590\\
4 & Effective metric & 0.279 & 3.063\\
8 & Hidden & 0.103 & 0.341\\
8 & Common rows & 0.069 & 0.136\\
8 & Effective metric & 0.195 & 0.684\\
\bottomrule
\end{tabular}
\end{table}

\begin{table}[htbp]\centering\small
\caption{Finite residual-generator pulses at adapted endpoints. Gain divides the small-amplitude centered logit-response norm by its own initial value. Halving discrepancy compares the two symmetric secants. Even remainder is normalized by the large-amplitude linear-response scale. These quantities do not identify a uniform tangent or a critical relaxation exponent.}
\label{model:tab:physical-response}
\begin{tabular}{@{}rrrrr@{}}
\toprule Heads & Updates & Gain & Halving discrepancy & Even remainder\\\midrule
4 & 0 & 1.000 & 0.319 & 0.320\\
4 & 8 & 34.937 & 0.933 & 7.580\\
4 & 128 & 2160.344 & 1.313 & 0.679\\
8 & 0 & 1.000 & 0.165 & 0.409\\
8 & 8 & 11.458 & 0.590 & 0.683\\
8 & 128 & 95.764 & 1.030 & 1.405\\
\bottomrule
\end{tabular}
\end{table}

\begin{table}[htbp]\centering\small
\caption{Native inference interventions across both classes and $L=8,16$. Entries give the largest panel-mean forward KL and largest prefix-logit discrepancy among the four cells. Generated intervention moments and paired native controls are retained separately.}
\label{model:tab:physical-interventions}
\begin{tabular}{@{}rlrr@{}}
\toprule Heads & Intervention & Mean KL maximum & Logit maximum\\\midrule
4 & row-projected & 0.00015 & 0.4804\\
4 & G0.9 & 0.000165 & 0.1409\\
4 & G1.1 & 0.000143 & 0.1304\\
8 & row-projected & \ensuremath{1.95\times10^{-6}} & 0.0104\\
8 & G0.9 & 0.000378 & 0.1485\\
8 & G1.1 & 0.000258 & 0.1143\\
\bottomrule
\end{tabular}
\end{table}

\begin{figure}[htbp]\centering\includegraphics[width=.98\textwidth]{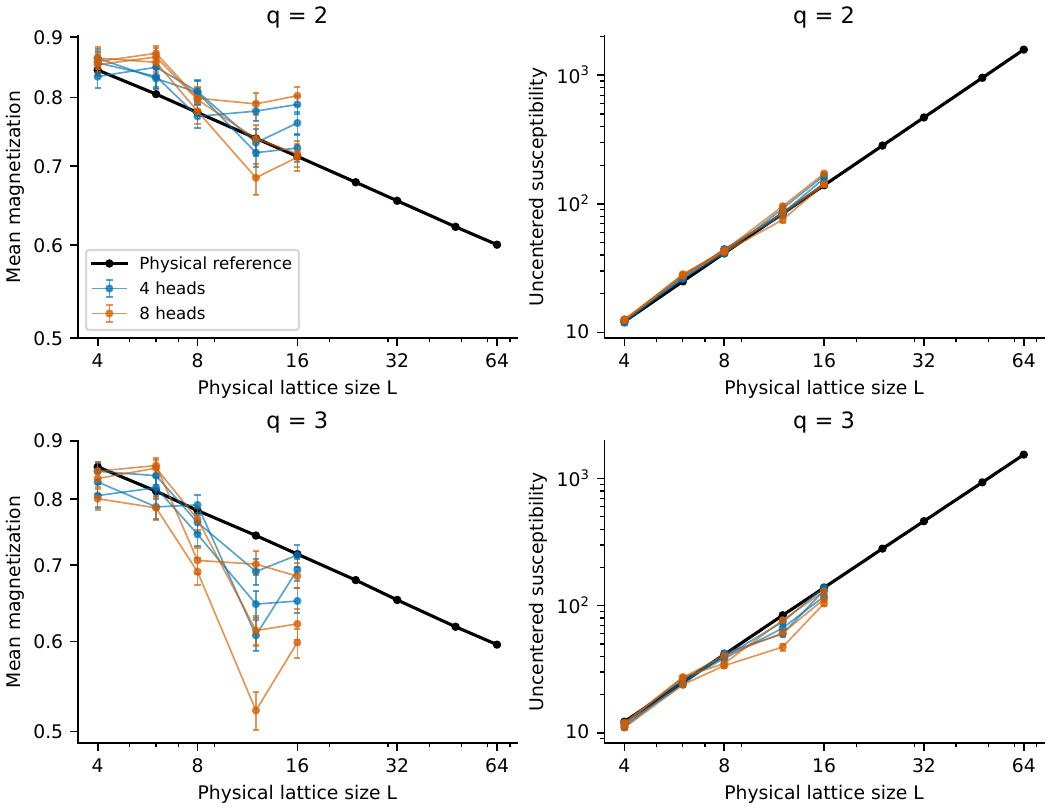}\caption{Critical physical moments and native generated-law moments. Thin curves show every primary model replica. Vertical bars are pointwise 95\% generation-bootstrap intervals conditional on each trained model; the physical reference spans sizes through 64. Native generation is assessed through size sixteen.}\label{model:fig:physical-scaling}\end{figure}

\begin{figure}[htbp]\centering\includegraphics[width=.95\textwidth]{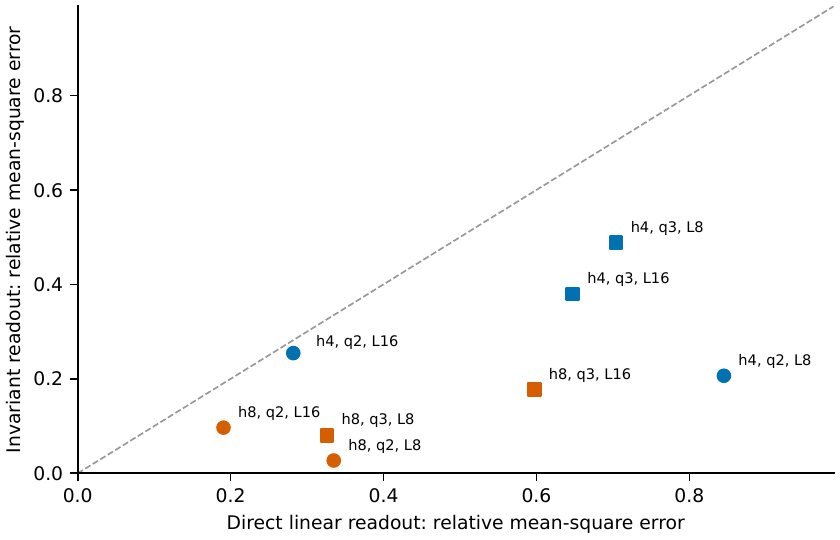}\caption{Fresh-chain magnetic-invariant prediction from common metric rows. Each point is one class, size and width cell. The diagonal denotes equal squared prediction error; lower values favor the invariant.}\label{model:fig:physical-collectives}\end{figure}

\subsection{What the calibration identifies}
\label{model:sec:physical-transport-results}

The generated configuration law is an explicit product of native
proper-prefix conditionals, as in autoregressive statistical-mechanics
models~\cite{wu2019autoregressive}. Complete enumeration of all sixteen
binary and 81 ternary configurations on a periodic $2\times2$ lattice
checks the normalization, KL chain rule and bounded-moment inequalities
in 48 class, temperature and model cells. These are exact finite-law
checks, including extrapolation to an untrained size. They establish
neither a thermodynamic limit nor a small large-lattice KL error.

The physical source fits, generated-law errors and internal magnetic
readouts answer distinct questions. The common rows retain information
about the physical color-order vector, including in matched random and
RefinedWeb-pretrained origin controls. The first-Gram identity in
Proposition~\ref{model:prop:physical-gram} supplies an architectural route for
this information; physical adaptation is not required to create it. Forming its bounded quadratic
invariant uses that geometry and has the explicit error bound in
Proposition~\ref{model:prop:physical-invariant}. Fitted physical blocking maps
still carry their measured held-out closure and composition defects.
No native scale eigenvalue is assigned to a representation merely
because it predicts magnetization. A low-dimensional physical readout
can itself display finite-size scaling~\cite{kim2021finite}.

Paired finite optimizer continuations retain the complete incoming
moments and consume 128 additional fresh updates on each of ten
branches. Symmetric perturbations multiply residual metric-network
parameters by $1\pm10^{-3}$ and $1\pm5\times10^{-4}$; the unperturbed
branch has the same future stream and terminal learning rate.
Amplitude-halving comparisons determine whether a measured response
can be identified with a tangent. When those comparisons are not
uniformly small, the nonlinear conditional transition law and its
finite response remain the appropriate objects. A large ratio of
response norms to a small initial response is not a measured critical
relaxation time.

The experiment therefore supports a finite architectural mechanism for
observing externally imposed magnetic structure in native common fields, with
separate controls on generated moments and empirical scale transport.
Exponent inheritance additionally requires the asymptotic error and
visibility hypotheses in Section~\ref{model:sec:physical-calibration}.
The two native widths and physical sizes through sixteen do not
establish a native thermodynamic exponent or an endogenous mechanism
for self-organized criticality.

\subsection{Physical fine-tuning from the qualified released model}
\label{model:sec:released-physical}

We also adapt the selected public model 5 from
Section~\ref{model:sec:released-adaptation} directly to native next-spin prediction.
Two independent adaptation seeds share its released pretrained weights.
They are adaptation replicas, not independently pretrained models. Every
native parameter is trainable, and both Adam moments start at zero because
the released pretraining optimizer is unavailable. AdamW uses moments
$(0.9,0.95)$, numerical constant $10^{-8}$, weight decay $0.01$ and
unit gradient-norm clipping. A 64-update warmup reaches $3\times10^{-5}$,
followed by cosine decay to one fifth of that rate. A fixed 256-update
validation pilot selects this rate over $10^{-4}$.

The adaptation corpus contains 61,440 configurations: sixteen independent
chains with 128 retained configurations in every combination of
$q=2,3$, $L=4,8,16$ and $T/T_c=0.85,0.925,1,1.075,1.15$.
The fixed Wolff sampler uses 4,096 burn-in cluster updates and 96 updates
between retained configurations. Validation and test use separate chain
seeds. Each epoch visits every training configuration identity exactly once
in a randomized order. Every visit draws a uniform color permutation and a uniform square-lattice
symmetry; a uniform target site is shared by the 32 configurations in
each batch. Color
permutation and lattice symmetry preserve the Hamiltonian and hence the
population Gibbs law. They change the presentation of the finite corpus;
they do not count as new independent configurations.

Each run completes 5,760 updates, or three epochs of 1,920 batches of 32.
The selected checkpoint minimizes native proper-prefix validation cross
entropy. The assessment then uses untouched chains at
$L=4,6,8,12,16,20,24$ and $T/T_c=0.94,0.97,1,1.03,1.06$.
The sizes six, twelve, twenty and twenty-four are absent from adaptation.
Sixteen test chains contribute eight configurations each. Eight target
sites are sampled in disjoint quantile strata, with stratum-width weights
when $L^2$ is not divisible by eight. All intervals retain the complete
within-chain panel; their sampling scope conditions on the frozen target
sites. A Dirichlet-half prefix-histogram predictor provides a fixed
nonspatial baseline in addition to the incoming released model and the
uniform $q$-color law.

\begin{table}[htbp]\centering\small
\caption{Native next-spin test performance, equally weighted over the 35 physical cells of each class and weighted by target-site stratum width. The fixed prefix-histogram baseline uses Dirichlet parameter one half.}
\label{model:tab:released-spin}
\begin{tabular}{@{}llrrr@{}}
\toprule $q$ & Model & NLL (nats) & Accuracy (\%) & Histogram NLL\\\midrule
2 & Released base & 0.4019 & 84.46 & 0.3631\\
2 & Adaptation 0 & 0.3284 & 86.97 & 0.3631\\
2 & Adaptation 1 & 0.3348 & 86.98 & 0.3631\\
3 & Released base & 0.6391 & 77.52 & 0.5847\\
3 & Adaptation 0 & 0.5309 & 80.52 & 0.5847\\
3 & Adaptation 1 & 0.5311 & 80.36 & 0.5847\\
\bottomrule
\end{tabular}
\end{table}

The selected physical checkpoints are update 3,328 (1.733 epochs), update 5,632 (2.933 epochs). Both runs complete three epochs; validation, rather than final-test scores, determines which checkpoint is exported.
For $q=2$, paired chain intervals for the NLL changes are $[-0.0780, -0.0690]$ and $[-0.0719, -0.0624]$ nats.
For $q=3$, paired chain intervals for the NLL changes are $[-0.1132, -0.1034]$ and $[-0.1135, -0.1023]$ nats.

\begin{table}[htbp]\centering\small
\caption{Task transfer to sizes absent from adaptation: $L=6,12,20,24$. These are the same untouched physical test chains, reported separately from the trained-size cells.}
\label{model:tab:released-spin-transfer}
\begin{tabular}{@{}llrr@{}}
\toprule $q$ & Model & NLL (nats) & Accuracy (\%)\\\midrule
2 & Released base & 0.3842 & 85.37\\
2 & Adaptation 0 & 0.3520 & 85.75\\
2 & Adaptation 1 & 0.3652 & 85.69\\
3 & Released base & 0.6216 & 78.79\\
3 & Adaptation 0 & 0.5675 & 78.73\\
3 & Adaptation 1 & 0.5819 & 78.07\\
\bottomrule
\end{tabular}
\end{table}

\begin{table}[htbp]\centering\small
\caption{Exhaustive source-to-model law comparison on the untrained $2\times2$ lattice at critical metadata. All configurations and proper-prefix probabilities are retained. The conditional KL sum equals the joint KL, and the bounded-moment and total-variation inequalities hold in all 18 temperature/model cells.}
\label{model:tab:released-exact}
\begin{tabular}{@{}llrrr@{}}
\toprule $q$ & Model & KL (nats) & TV & $|\Delta\mathbb E m^2|$\\\midrule
2 & Released base & 2.6818 & 0.8742 & 0.6375\\
2 & Adaptation 0 & 0.0977 & 0.1613 & 0.1153\\
2 & Adaptation 1 & 0.1017 & 0.1453 & 0.0986\\
3 & Released base & 3.6649 & 0.8794 & 0.5220\\
3 & Adaptation 0 & 0.2355 & 0.2976 & 0.1873\\
3 & Adaptation 1 & 0.1894 & 0.2129 & 0.1238\\
\bottomrule
\end{tabular}
\end{table}

Both adapted checkpoints substantially reduce the source-to-model joint KL for each physical class. This is a measured improvement of the complete finite generative law at a size excluded from adaptation. These small-lattice identities validate the finite probability-law accounting. They do not estimate conditional excess entropy or certify the corresponding law-error budget at larger sizes.
Across all physical-source cells, the largest classical split-chain variance ratios for energy density, magnetic second moment and first-color fraction are 1.0072 (train), 1.0120 (validation), 1.0050 (test). These finite diagnostics do not certify equilibration or independent configurations. Every uncertainty interval retains the declared complete-chain sampling unit.

Both adaptations improve aggregate held-out accuracy and NLL for each
class. On sizes absent from adaptation, both improve NLL, while accuracy
transfer depends on the class and seed. For three-state Potts the accuracy
changes are $-0.062$ and $-0.720$ percentage points. Whole-chain paired
95\% intervals are $[-0.446,0.323]$ and $[-1.186,-0.243]$ percentage
points, respectively, using 4,000 resamples with a fresh fixed seed.
These intervals condition on the released base, fitted checkpoint and
fixed target sites. Improved proper log score need not improve top-one
accuracy: the two functionals order predictive laws differently.

Native ancestral generation uses 512 configurations per critical cell and
128 per thermal cell, with fresh declared seeds, model temperature one,
and complete proper-prefix recomputation. Thermal generation covers
$L=4,8,16,24$ at each of the four noncritical test ratios. Every selected
cell is retained. The comparison reports connected and uncentered
susceptibilities, Binder ratios, color balance and correlations at fixed
fractions of the lattice size. Independent-chain bootstrap intervals for
the source and conditional generation intervals describe different random
units. Finite slopes use every size above each stated cutoff and report
the two adaptation seeds separately.

\begin{table}[htbp]\centering\scriptsize
\caption{Finite critical-metadata log slopes for $q=2$ through $L=24$. Brackets give 95\% conditional generation or independent source-chain intervals. The lower cutoff is explicit. $x_m$ is minus the slope of mean magnetic magnitude; the susceptibility slopes are uncentered and connected, respectively. Finite fits do not establish limiting critical exponents.}
\label{model:tab:released-slopes-q2}
\begin{tabular}{@{}lrlll@{}}
\toprule Model & $L_{\min}$ & $x_m$ [95\%] & $x_\chi$ [95\%] & $x_{\chi,c}$ [95\%]\\\midrule
Physical source & 4 & 0.12 [0.12, 0.13] & 1.76 [1.75, 1.76] & \\
Released base & 4 & -0.22 [-0.25, -0.20] & 2.39 [2.35, 2.43] & 2.81 [2.74, 2.89]\\
Physical source & 8 & 0.13 [0.11, 0.14] & 1.75 [1.73, 1.77] & \\
Released base & 8 & -0.35 [-0.41, -0.30] & 2.58 [2.50, 2.67] & 2.51 [2.42, 2.61]\\
Physical source & 12 & 0.12 [0.10, 0.15] & 1.75 [1.72, 1.79] & \\
Released base & 12 & -0.29 [-0.38, -0.21] & 2.49 [2.38, 2.61] & 2.48 [2.31, 2.64]\\
Adaptation 0 & 4 & 0.09 [0.07, 0.11] & 1.84 [1.81, 1.87] & 1.48 [1.43, 1.54]\\
Adaptation 0 & 8 & 0.03 [-0.01, 0.06] & 1.96 [1.91, 2.02] & 1.52 [1.41, 1.63]\\
Adaptation 0 & 12 & -0.00 [-0.07, 0.06] & 2.00 [1.90, 2.09] & 1.77 [1.58, 1.97]\\
Adaptation 1 & 4 & -0.01 [-0.02, 0.00] & 2.00 [1.98, 2.01] & 1.93 [1.90, 1.97]\\
Adaptation 1 & 8 & -0.05 [-0.07, -0.04] & 2.09 [2.06, 2.11] & 1.96 [1.90, 2.01]\\
Adaptation 1 & 12 & -0.14 [-0.17, -0.12] & 2.24 [2.20, 2.29] & 2.07 [1.98, 2.15]\\
\bottomrule
\end{tabular}
\end{table}

\begin{table}[htbp]\centering\scriptsize
\caption{Finite critical-metadata log slopes for $q=3$ through $L=24$. Brackets give 95\% conditional generation or independent source-chain intervals. The lower cutoff is explicit. $x_m$ is minus the slope of mean magnetic magnitude; the susceptibility slopes are uncentered and connected, respectively. Finite fits do not establish limiting critical exponents.}
\label{model:tab:released-slopes-q3}
\begin{tabular}{@{}lrlll@{}}
\toprule Model & $L_{\min}$ & $x_m$ [95\%] & $x_\chi$ [95\%] & $x_{\chi,c}$ [95\%]\\\midrule
Physical source & 4 & 0.14 [0.13, 0.14] & 1.74 [1.73, 1.75] & \\
Released base & 4 & 0.05 [0.02, 0.08] & 1.98 [1.94, 2.03] & 2.65 [2.60, 2.71]\\
Physical source & 8 & 0.14 [0.13, 0.16] & 1.72 [1.70, 1.74] & \\
Released base & 8 & -0.11 [-0.16, -0.05] & 2.25 [2.14, 2.35] & 2.45 [2.34, 2.55]\\
Physical source & 12 & 0.14 [0.12, 0.15] & 1.73 [1.70, 1.76] & \\
Released base & 12 & -0.04 [-0.14, 0.06] & 2.14 [1.96, 2.31] & 2.11 [1.93, 2.29]\\
Adaptation 0 & 4 & 0.13 [0.11, 0.15] & 1.76 [1.73, 1.79] & 1.15 [1.08, 1.20]\\
Adaptation 0 & 8 & -0.02 [-0.06, 0.02] & 2.03 [1.97, 2.09] & 1.23 [1.11, 1.34]\\
Adaptation 0 & 12 & -0.11 [-0.17, -0.04] & 2.18 [2.07, 2.30] & 1.60 [1.38, 1.83]\\
Adaptation 1 & 4 & 0.07 [0.06, 0.09] & 1.86 [1.84, 1.88] & 1.81 [1.79, 1.83]\\
Adaptation 1 & 8 & -0.00 [-0.03, 0.02] & 1.98 [1.95, 2.02] & 1.98 [1.93, 2.04]\\
Adaptation 1 & 12 & -0.10 [-0.14, -0.05] & 2.15 [2.07, 2.22] & 2.43 [2.31, 2.57]\\
\bottomrule
\end{tabular}
\end{table}

\begin{table}[htbp]\centering\small
\caption{Direct critical-metadata discrepancies at the largest tested size, $L=24$. The susceptibility difference is relative to the physical source; Binder and distance-$L/2$ correlation differences are absolute. Color bias is $\max_a|\mathbb E f_a-1/q|$. The last column separates connected fluctuation from the uncentered second moment.}
\label{model:tab:released-law-errors}
\begin{tabular}{@{}llrrrrr@{}}
\toprule $q$ & Model & $\Delta\chi$ (\%) & $\Delta B$ & $\Delta C_{12}$ & Bias & $\chi_c/\chi$\\\midrule
2 & Released base & +12.5 & +0.174 & +0.123 & 0.119 & 0.897\\
2 & Adaptation 0 & +4.8 & +0.141 & +0.056 & 0.243 & 0.540\\
2 & Adaptation 1 & +61.4 & -0.139 & +0.320 & 0.103 & 0.947\\
3 & Released base & -46.1 & +1.065 & -0.085 & 0.097 & 0.913\\
3 & Adaptation 0 & -18.7 & +0.200 & -0.063 & 0.326 & 0.388\\
3 & Adaptation 1 & +27.9 & -0.030 & +0.152 & 0.157 & 0.908\\
\bottomrule
\end{tabular}
\end{table}

\begin{table}[htbp]\centering\small
\caption{Spatial covariance decomposition at $L=24$ and critical metadata. The first two differences compare the native law with its physical reference at distance 12. The mean sector uses each position-specific color mean. The last column is $q/(q-1)$ times the spatial mean square of the color-mean profile around its global mean. These are finite empirical point diagnostics.}
\label{model:tab:released-spatial-centering}
\begin{tabular}{@{}llrrrr@{}}
\toprule $q$ & Model & $\Delta C_{12}$ & $\Delta K_{12}$ & $H_{12}$ & Profile variance\\\midrule
2 & Released base & +0.123 & +0.058 & 0.066 & 0.0555\\
2 & Adaptation 0 & +0.056 & -0.188 & 0.244 & 0.0228\\
2 & Adaptation 1 & +0.320 & +0.278 & 0.042 & 0.0004\\
3 & Released base & -0.085 & -0.110 & 0.025 & 0.0209\\
3 & Adaptation 0 & -0.063 & -0.307 & 0.245 & 0.0151\\
3 & Adaptation 1 & +0.152 & +0.095 & 0.057 & 0.0022\\
\bottomrule
\end{tabular}
\end{table}

Proposition~\ref{model:prop:physical-spatial-centering} is applied to every generated cell. Its empirical decomposition is checked both from direct categorical covariance and from raw correlation minus the mean sector. The position-dependent means are retained explicitly, so a near match of uncentered correlations is not identified with connected fluctuation fidelity.

\begin{table}[htbp]\centering\small
\caption{Finite temperature response of the Binder ratio at $L=24$. The contrast is $[B(1+w)-B(1-w)]/(2w)$ in reduced temperature. Brackets are 95\% intervals from independent generated samples or source chains. Both frozen widths are retained; this is a finite contrast, without a fitted thermal exponent.}
\label{model:tab:released-thermal}
\begin{tabular}{@{}llrl@{}}
\toprule $q$ & Model & $w$ & Binder contrast [95\%]\\\midrule
2 & Physical source & 0.03 & 6.13 [5.57, 6.67]\\
2 & Released base & 0.03 & -0.18 [-1.93, 1.83]\\
2 & Adaptation 0 & 0.03 & 4.04 [2.47, 5.84]\\
2 & Adaptation 1 & 0.03 & 1.10 [0.58, 1.74]\\
2 & Physical source & 0.06 & 6.44 [6.08, 6.75]\\
2 & Released base & 0.06 & -0.18 [-1.26, 0.92]\\
2 & Adaptation 0 & 0.06 & 2.03 [1.28, 2.91]\\
2 & Adaptation 1 & 0.06 & 0.27 [0.12, 0.47]\\
3 & Physical source & 0.03 & 10.44 [9.76, 11.17]\\
3 & Released base & 0.03 & 1.79 [-4.25, 7.22]\\
3 & Adaptation 0 & 0.03 & 4.11 [2.99, 5.38]\\
3 & Adaptation 1 & 0.03 & 2.53 [1.69, 3.41]\\
3 & Physical source & 0.06 & 8.36 [7.88, 8.79]\\
3 & Released base & 0.06 & -1.78 [-4.78, 0.83]\\
3 & Adaptation 0 & 0.06 & 2.29 [1.58, 3.02]\\
3 & Adaptation 1 & 0.06 & 1.14 [0.72, 1.63]\\
\bottomrule
\end{tabular}
\end{table}

\begin{table}[htbp]\centering\small
\caption{Operator geometry on critical test configurations at $L=8,16,24$. Maxima cover both physical classes, all samples, layers and heads for the row fraction, and the six complete input panels for the common-field fraction and paired $G$ order. The common-field fraction is $\mathbb E\|\mu-\mathbb E\mu\|^2/\mathbb E\|\mu\|^2$. The last column couples the half and final proper prefixes of each configuration.}
\label{model:tab:released-geometry}
\begin{tabular}{@{}lrrr@{}}
\toprule Model & Max row fraction & Max common fraction & Max $\mathcal O_G$\\\midrule
Released base & \ensuremath{1.05\times10^{-16}} & \ensuremath{2.44\times10^{-26}} & \ensuremath{0.00}\\
Adaptation 0 & \ensuremath{2.36\times10^{-11}} & \ensuremath{1.44\times10^{-16}} & \ensuremath{3.74\times10^{-6}}\\
Adaptation 1 & \ensuremath{2.64\times10^{-12}} & \ensuremath{4.26\times10^{-15}} & \ensuremath{3.87\times10^{-5}}\\
\bottomrule
\end{tabular}
\end{table}

The incoming released base has zero observed paired $G$ discrepancy on
these panels, while its common-row input variation is numerically negligible.
Fine-tuning improves held-out spin prediction while retaining strong row
concentration and small paired operator discrepancy. Thus task learning can
coexist with an almost input-invariant metric generator. The native query,
key, value and body computations remain input dependent. These observations
do not make the common-row magnetic readout a test of task competence or
establish equality of every input pair outside the measured panels.
\begin{figure}[htbp]\centering
\includegraphics[width=\textwidth]{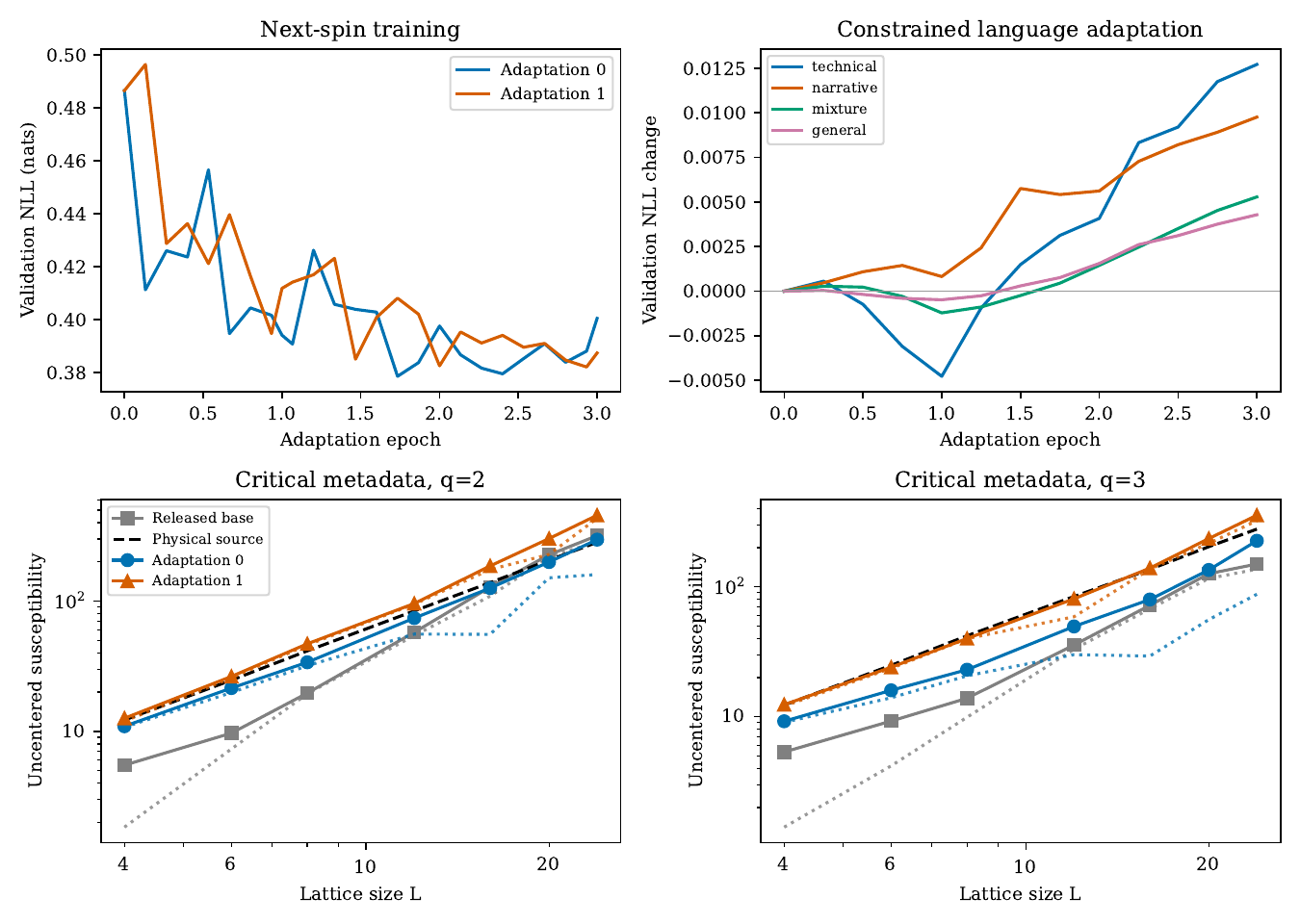}
\caption{Released-base adaptation and physical inference. The upper panels
show validation histories under the actual epoch clock. Lower panels compare
critical-metadata generated susceptibilities with independent physical-source
measurements. Dotted curves give the corresponding connected susceptibilities. Each adapted model is a separate fine-tuning seed of the same
released base. All displayed sizes and models are retained.}
\label{model:fig:released-adaptation}
\end{figure}

The direct adapted-minus-base thermal response is reported in
Table~\ref{model:tab:thermal-direct}. Every $L=24$ endpoint cell contributes
128 independently generated configurations. The 16,000 configuration
resamples reuse a cell's draws wherever that cell enters a comparison.
All eight point estimates are positive; four nominal intervals exclude
zero and three do so with the eight-comparison quantile allocation.
This exploratory reanalysis uses fixed selected checkpoints and no
additional fitting. It supports some finite response increases, with
class-, seed- and step-dependent uncertainty. It supplies no thermal
exponent. Comparing significance against zero in separate checkpoints
would not answer the adapted-minus-base question.

\input{content/model/generated/thermal-transport.tex}

Proposition~\ref{model:prop:thermal-transport} passes all eighteen exact-law
ratio checks and all six $h=0.06$ contrast checks on the complete
$2\times2$ spaces. These use normalized full laws and their second and
fourth moments, rather than substituting ordinary NLL for excess KL.
The signed identity is retained alongside its conservative absolute
budget. A bound that is large relative to the source contrast certifies
an error envelope, not an accurate reduced thermal predictor.

The magnetic size fits require a different conclusion. For example,
adaptation 0 at $q=3$ has an uncentered susceptibility slope of 1.759
when all seven sizes are included, compared with 1.738 for the source.
Raising the lower cutoff to twelve gives 2.177 and 1.728, respectively.
Adaptation 1 gives 1.859 and 2.145 at those two cutoffs. Retaining all
cutoffs and both seeds prevents a single close finite fit from being
assigned the source universality class. The reported bootstrap intervals
condition on each selected model; they do not absorb variation between
adaptation seeds or extrapolation error beyond the tested sizes.

The spatial decomposition explains another limitation of a raw moment
comparison. At $L=24$, adaptation 0 has distance-twelve raw-correlation
errors of $+0.056$ and $-0.063$ for $q=2,3$, while its connected errors
are $-0.188$ and $-0.307$. The corresponding position-mean contributions
are both approximately $0.245$. Thus mean and covariance errors can partly
cancel in a raw correlation. The second adaptation has different mean
and covariance errors, which are retained in
Table~\ref{model:tab:released-spatial-centering}. The exact identity in
Proposition~\ref{model:prop:physical-spatial-centering} therefore supplies a
necessary refinement of the observable state: spatial color means and
connected covariance are measured separately, even after symmetry-augmented
training. An empirical color imbalance is a finite-sample diagnostic;
these observations alone do not prove spontaneous symmetry breaking.

The task scores and generated-law measurements test two different parts
of Proposition~\ref{model:prop:physical-prefix-control}. Accuracy is not a
joint-law divergence, and cross entropy includes the unknown physical
conditional entropy. Even a small per-site excess can accumulate over
$L^2$ sites. Physical exponent transfer therefore retains the explicit
relative-moment, spatial-law and visibility conditions of
Section~\ref{model:sec:physical-calibration}. The actual finite comparisons
below those hypotheses determine the supported inference scope; a known
critical adaptation temperature is not an endogenous training transition.

\section{Magnetic readout errors and size transfer}
\label{model:sec:readout-results}
\subsection{Finite native assessment of the budget}

An independent float64 reduction evaluated fixed simplex readouts from
24 saved native observation arrays. It crossed two widths, three
origins, two alphabets, two physical sizes, two blocking depths, and
three feature families, giving 144 cells. Each cell used eight
held-out source chains with eight retained configurations per chain;
the other eight chains supplied the fitted readout. The reduction
checked the signed identity, both second-moment bounds, and the
connected-variance bound. The maximum signed-identity residual was
$2.35\times10^{-16}$. The scale-dependent bound was smaller than
the unit-ball bound in 142 cells. These are identities and bounds
under finite empirical laws; the checks do not certify a population
error rate.

\begin{table}[htbp]
\centering\small
\caption{Adapted common-row readout at blocking depth zero. The
columns give empirical $r=\delta/\sqrt\mu$, absolute relative
moment error, and the two relative upper bounds. Each row describes
one fixed model and eight test chains; the displayed numbers are
point estimates, not simultaneous confidence bounds.}
\label{model:tab:new-readout}
\input{content/model/generated/readout-table.tex}
\end{table}

\begin{figure}[htbp]
\centering
\includegraphics[width=.91\textwidth]{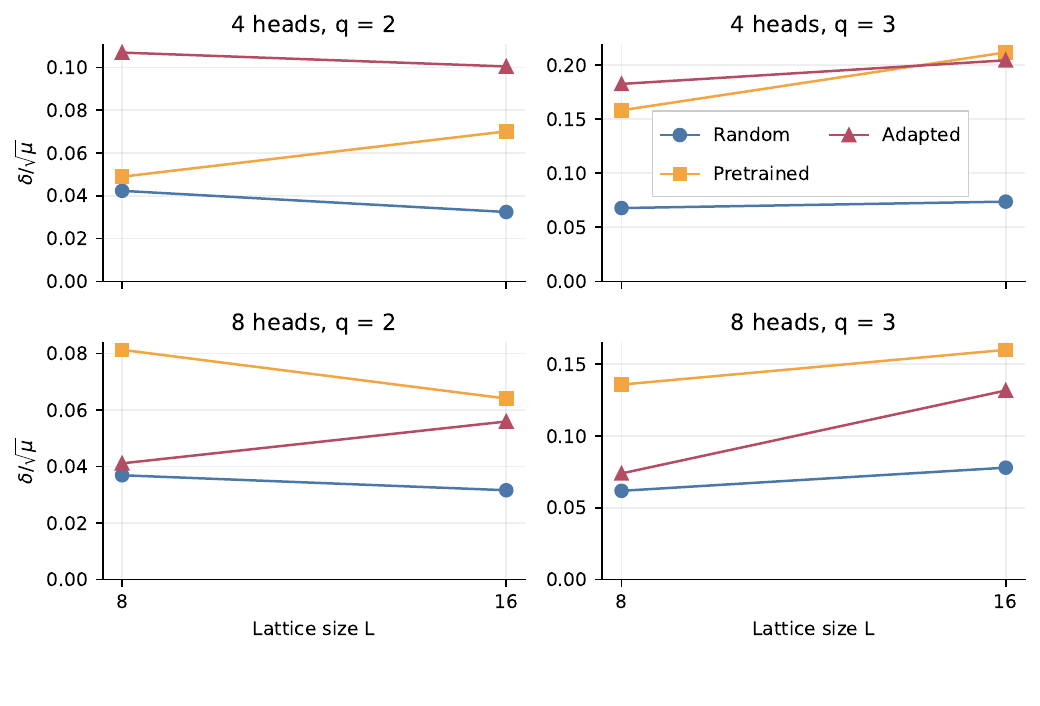}
\caption{Relative vector error of the fixed common-row readout on
the same source cohort, by model origin. Lines connect two empirical
sizes only. These descriptive comparisons condition on the calibration
fit and share the test configurations across origins. They are not
independent size replications or estimates of an asymptotic exponent.}
\label{model:fig:new-readout}
\end{figure}

The adapted common-row cells have $r$ between $0.0411$ and $0.2044$
and relative moment errors between $0.0072$ and $0.0898$.
The two available lattice sizes do not determine whether $r_L\to0$.
The origin comparison and the first-query-Gram histogram identity
motivate treating the existence of magnetic information as an
architectural observation. Physical adaptation, predictive quality,
and compatibility with spatial blocking require their own tests.

For the eight adapted common-row comparisons at sizes eight and
sixteen, all invariant gains are positive in point estimates. Six
fixed-fit paired-chain 95\% intervals exclude zero. This does not
provide a simultaneous all-cell guarantee.

\subsection{Fresh chains and unused physical sizes}

A separate frozen study drew 32 independent Wolff chains in each
$(q,L)$ cell, with $q\in\{2,3\}$ and $L\in\{8,12,16,20\}$.
Each chain used 4,096 burn-in cluster updates and retained eight
configurations separated by 96 cluster updates. Alternating ordered
and random starts gave matched start-type diagnostics; the largest
absolute standardized difference between start-type mean magnetic
moments was 1.68. This finite diagnostic does not prove equilibration.
Chain resampling retains within-chain dependence.

The first sixteen chains in each cell supplied calibration, and the
last sixteen supplied assessment. Six native checkpoints crossed four
and eight heads with random, single-pass RefinedWeb-pretrained and
physically adapted origins at body seed 640101. Every checkpoint saw
the same configurations and the same two successive majority-blocked
versions. There were 2,048 source configurations, 256 independent
chains and 36,864 model--configuration evaluations, with no optimizer
updates. The complete 648 readout cells and 144 fitted block-map cells
are retained in the accompanying arrays.

Each observation used the final proper prefix, including metadata and
excluding the target spin. The largest call had 424 tokens, within the
native 1,024-token context. Float32 selected/full output checks at
unused sizes passed with maximum absolute logit discrepancy
$4.08\times10^{-7}$. The three origin jobs took 50.5 seconds in total
at four heads and 59.0 seconds at eight heads, on separate RTX 4090
GPUs; peak allocated memory was 0.184 and 0.372 GiB. These times cover
native observation and output writing, not source simulation or CPU
bootstrap analysis.

Within each size and blocking depth, eight calibration principal
components and ridge penalty $10^{-3}$ times the mean Gram diagonal
were used for both the color-fraction readout and the direct magnetic
second-moment readout. A distinct transfer fit used only calibration
chains at $L=8,16$ and was evaluated without refitting at $L=12,20$.
Thus size twelve tests interpolation to an unused size and size twenty
tests extrapolation beyond the fitted sizes. Metadata and native
coordinates were not renormalized using assessment targets.

The 1,000-resample pointwise intervals independently resampled
calibration and assessment chains and refitted the ridge coefficients.
They condition on the acquired calibration PCA basis and on one
checkpoint per origin and width. They do not include PCA acquisition,
initialization or pretraining-corpus uncertainty. Fifteen of the sixteen
adapted, unblocked common-row cells improved the direct readout in point
estimates and in their pointwise paired intervals. This is a statement
about these cells, not uniform improvement over a model population.

\input{content/model/generated/fresh-readout-results.tex}

\begin{figure}[htbp]\centering
\includegraphics[width=.94\textwidth]{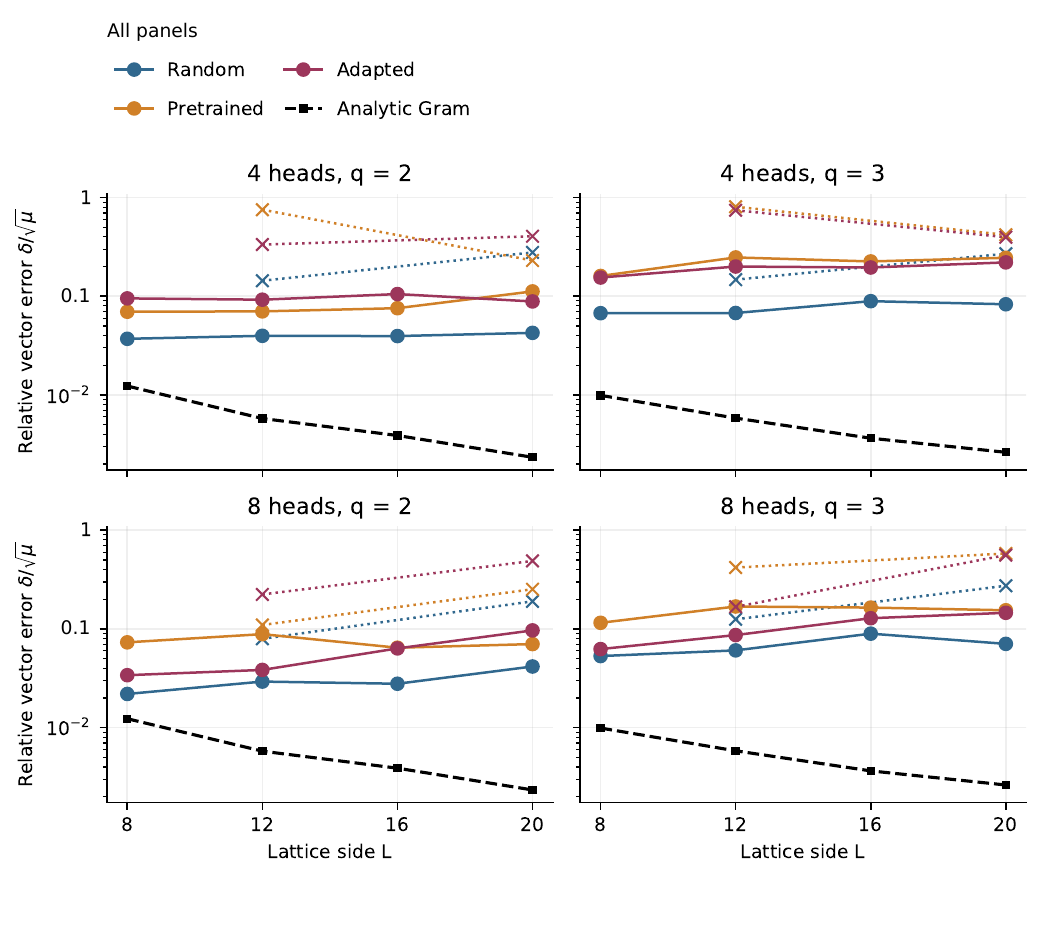}
\caption{Fresh-source relative vector error. Solid colored lines use a
separate calibration at each size. Dotted lines connect the two unused
sizes evaluated by a fit frozen at sizes eight and sixteen. The dashed
black line shows analytic Gram inversion for the pretrained origin;
the random and adapted Gram values agree to the displayed precision.
Lines guide comparison of finite points and are not fitted critical
powers. All estimates condition on the stated checkpoint and fit.}
\label{model:fig:fresh-readout}
\end{figure}

\subsection{A finite mechanism for scale-sensitive observation}

Across all 144 analytic Gram cells, the smallest restricted singular
value was 0.619 and the maximum prefix-fraction error was
$1.23\times10^{-6}$. Independent reduction verified
Equation~\eqref{model:eq:prefix-readout-budget} for all 36,864 measured
configuration/readout pairs. In the 48 unblocked Gram cells,
$r$ ranged from 0.00233 to 0.01231. Their empirical sufficient relative
second-moment budget $2r+r^2$ was below 0.025, and their largest actual
relative second-moment error was 0.00149. These are bounds under the
finite assessment laws, not confidence bounds on an unobserved
population or an asymptotic arithmetic certificate.

The origin comparison identifies the first-Gram observation as an
architectural mechanism that already operates before training. In
contrast, fitted common-row coefficients carry appreciable size
conditioning. Table~\ref{model:tab:fresh-readout} separates their within-size
accuracy from actual transfer to unused sizes. Their two-block versus
iterated one-block discrepancies ranged from 0.0068 to 0.3638 in units
of the held-out coarse-code RMS about its calibration mean, across the three fitted
feature families. Histogram recovery therefore does not supply an
autonomous spatial block map. A scale-dependent observation must retain
its metadata, normalization, coefficient acquisition and closure error.
The constructive Gram readout and the measured finite common-row
readouts occupy different levels of this effective-state hierarchy.

\subsection{Finite-size corrections and uncertainty in the reference law}

To separate model-form sensitivity from Monte Carlo uncertainty, a
fixed reanalysis used all available reference-chain sufficient statistics
for sizes through 64. At each lower cutoff $8$, $16$ or $24$, it compared
$\log|\partial_T B_L|=a+p\log L$ with
$\log|\partial_T B_L|=a+p\log L+cL^{-\omega}$ for each
$\omega\in\{0.8,1,2\}$. The estimator was $\nu=1/p$;
no exponent was fixed to its known value. All 24 fits and their
leave-one-size-out predictions are retained. These correction terms
are sensitivity models; their amplitudes and applicability to this
Binder derivative are not inferred from corrections in other
observables~\cite{xu2025corrections}.

For $q=3$ and $L_{\min}=8$, the leading estimate was 0.8051 and
its chain-resampling interval excludes $5/6$. At $L_{\min}=16$,
correction-family point estimates ranged from 0.8166 to 0.8383.
However, each correction fit had larger leave-one-size-out prediction
error than the corresponding leading fit. Closeness to a known exponent
therefore cannot select a correction as validated. The wider sensitivity
at the largest cutoff reflects fewer sizes and weakly constrained
correction coefficients. The reference-law values calibrate finite
observables; this analysis supplies no native thermodynamic exponent.

\FloatBarrier
\par\medskip\noindent
Chapter~\ref{ch:synthesis} brings these results together. It separates
established finite conclusions from conditional limiting statements and
identifies the remaining requirements for a broader theory of training and
inference.

\chapter{Synthesis and the boundary of the established theory}
\label{ch:synthesis}
This chapter synthesizes the complete-law, geometric and predictive
descriptions developed in the book. It states what their combination
establishes, summarizes the hypotheses behind the limiting classes, and
locates the unresolved mechanisms at the boundary of the finite evidence.

\section{The closed law and its conditional classification}
\label{model:sec:theory-synthesis}

The compatible state is the training path law together with its
full-graph emission and declared sources. Decoder elimination retains
internal source insertions; temporal elimination composes the augmented
kernels chronologically; linear observation and aligned document or
head blocking push forward the same joint dependence. These operations
describe one computation. Identifying their result with a separately
trained smaller architecture requires a further matching assumption.
A state-dependent exception map retains its indices and tie rule;
it need not commute with a different adaptive selection.

Table~\ref{model:tab:claim-index-main} states the central claims and their
empirical scope. Conditional memory in Proposition~\ref{model:prop:conditional-memory} supplies
exact closure when finite coordinates are insufficient.

\subsection{Scale laws and their hypotheses}
The thermodynamic family fixes the data law, decoder depth $L$, head
dimension $d$ and shared generator width independently of $N$, with
residual width $dN$. The recorded family uses $L=5$, $d=64$ and shared
generator width 170. The specification fixes initialization, both
learning-rate groups, clipping, decay, moment coefficients and offsets,
and the horizon sequence $t_N$. For a single pass it also specifies
the corpus-generation and blocking rule, resource sequence $M_N$,
batch sequence $B_N$, consumed fraction $q_N=B_Nt_N/M_N$ and
numerical policy $\nu_N$. The latter belongs to the implemented
conditional law; Proposition~\ref{model:prop:implementation-error} states
a sufficient comparison with an idealized arithmetic law.
Section~\ref{model:sec:data-resource} gives their compatible scaling
conditions. Conditioning on each realized corpus and ordering differs
from averaging over the corpus-generation law; the latter retains the
corpus-mean sector of Proposition~\ref{model:prop:corpus-conditioning}.
An $N$-dependent Adam offset defines another family. A conditional source potential averages over
$\xi_N$ at fixed $(x,\mathcal C)$ before averaging over contexts.
Taking a logarithm after mixing those environments produces another
potential and can introduce a mixture contribution. Bounded intensive
fields have compatible subsequential source potentials by
Proposition~\ref{model:prop:potential-compactness}; uniqueness and derivative
limits require further hypotheses.

\small
\begin{longtable}{>{\raggedright\arraybackslash}p{.22\textwidth}>{\raggedright\arraybackslash}p{.34\textwidth}>{\raggedright\arraybackslash}p{.35\textwidth}}
\caption{Conditional classification under specified scale maps.}
\label{model:tab:conditional-classes}\\
\toprule Sector & Scale law and hypotheses & Mathematical and empirical scope\\
\midrule\endfirsthead
\toprule Sector & Scale law and hypotheses & Mathematical and empirical scope\\
\midrule\endhead
Independent documents & $H=1/2$; centered iid finite-variance blocks &
Gaussian joint limit; cumulant multipliers $b^{1-n/2}$. Completed
document measurements give finite-scale covariance and cumulants.\\
Consuming finite population & $M_Nh_N\to m$, fixed batch,
frozen centered population and deterministic chronological transport &
Exact negative cross-time covariance and a physical-clock second-moment
limit with a finite-resource correction; no automatic Gaussian or adaptive
gradient limit (Proposition~\ref{model:prop:consuming-bridge}).\\
Persistent shared component & $H=1$ for a block mean; a nonzero shared
field and a mean-square vanishing residual average &
The shared law survives. Mixture or document coherence alone is not
intrinsic connected criticality.\\
Correlated text or time & $C(r)\sim Mr^{-\gamma}$,
$0<\gamma<1$; $H=1-\gamma/2$ &
Conditional second-moment scaling. A covariance tail does not by itself
identify the full limiting distribution.\\
Native initialization & Shape-aware initial law; conditional wide
projection central-limit and moment hypotheses &
The native kernel and concentrating head means, with completed
finite-width comparisons; no automatic trained-state limit.\\
Regular trained heads & Uniform extensive cumulant bounds and
convergent cumulant densities &
Analytic limiting source potential and Gaussian fluctuations under
the stated hypotheses. Finite-time training data do not certify them.\\
Critical trained heads & Singular conditional limit, compatible
scaling fields, and a specified size or length convention &
Critical exponents are conditional properties of that limit. No
trained critical class is assigned by these finite measurements.\\
Endogenous critical selection & Identified critical surface, restoring
feedback, sufficient relaxation, and fluctuations below its shrinking
window &
Conditional attraction criterion. Measured recovery of row
concentration establishes a finite regime, without identifying such
a critical surface.\\
\bottomrule
\end{longtable}
\normalsize

For a length-based relevant coordinate $r'=\lambda r$,
$\xi'=\xi/b$ implies $\nu=\log b/\log\lambda$ under the
homogeneity hypotheses already stated. Head-count exponents in a form
$\chi_N=N^\kappa F((g-g_c)N^\phi,t/N^z)$ require their own
size--time correspondence and corrections to scaling. Parameter count,
head count and geometric length are not interchangeable. Neither a
finite rate peak nor a local tangent amplification supplies that
correspondence. The completed experiments identify finite
conditional regimes and effective inference variables within this
classification; they do not fit a thermodynamic critical exponent.

Holding Adam coefficients fixed preserves update-count memory.
Maintaining finite memory under a rescaled step $h_N\to0$ requires an
explicitly changed family, such as $\beta_i(N)=\exp(-\gamma_i h_N)$.
Proposition~\ref{model:prop:matched-memory-force} gives a uniform adaptive-force
bound when $2\gamma_1>\gamma_2>0$, without asserting convergence of
that changed native family.
The thermodynamic, long-time and vanishing-drive limits must be ordered
or jointly specified. A consuming reservoir has no nonterminal stationary
law without a refill rule. A local stationary approximation additionally
requires relaxation faster than resource and schedule drift, including
the source-path coupling conditions of Section~\ref{model:sec:data-resource}.

Fitting adds another conditioning sector. Independent future paths
centered using the same $m$-path estimated mean have cross covariance $C/m$
when fitting is averaged over, although their cross covariance conditional on
that fitted mean is zero (Proposition~\ref{model:prop:acquisition-noise}).
This shared uncertainty belongs to the observation procedure. Its
scaling depends on the acquisition sequence $m_N$ and must be distinguished
from a persistent field generated by the trained model. The two fresh
fit cohorts in Section~\ref{model:sec:refresh-results} expose this additional
sampling axis within the finite experiment.

The finite categorical envelope in Proposition~\ref{model:prop:metric-budget}
controls a further observation approximation at each temporal scale.
It uses acquired logits and does not close their successor law or certify
a reduction in acquisition cost. The matched width/schedule experiment
fixes the initialization, optimizer, remaining resource and observation
cohort explicitly, so its finite contrasts have a declared ensemble.

The fluctuation criterion is stricter than finite risk prediction.
For a coupled observation error $e$ and $\chi=N\Var(X)$,
Proposition~\ref{model:prop:critical-closure} gives
\[
 |N\Var(X+e)-\chi|\le
 2\sqrt{N\chi}\,\norm e_{L^2}+N\norm e_{L^2}^2.
\]
Thus a count law $\chi\sim C N^\kappa$, $C>0$, is guaranteed to
transfer with the same leading constant if
$\norm e_{L^2}=o(N^{(\kappa-1)/2})$. This is sufficient, not necessary
for a specified coupling: the exact difference is
$N[2\Cov(X,e)+\Var(e)]$, whose signed terms can cancel
(Proposition~\ref{model:prop:signed-transfer}). If its relaxation gap
is of order $N^{-\zeta}$, a sufficient one-step discrepancy is
$o(N^{(\kappa-1)/2-\zeta})$, with a negligible initial transient.
The inference emission must also retain a nonzero transported
limiting covariance (Proposition~\ref{model:prop:inference-fluctuation-transfer});
constant-logit directions have zero categorical Fisher weight.
These conditions join dynamical closure to inference visibility.
The law-averaged $W_1$ risk bound supplies neither the stronger
$L^2$ coupling nor a measurement of these native exponents.

The forcing spectrum supplies another scaling field. Under the
stationary scalar-mode hypotheses of Corollary~\ref{model:cor:colored-exponents},
a response gap of order $N^{-\zeta}$ and forcing spectral amplitude
of order $N^{-\upsilon}$ for $y_N=\sqrt N\,\bar u_N$ and low-frequency spectral shape
$|\omega|^\alpha$, $-1<\alpha<1$, give
$\kappa=\zeta(1-\alpha)-\upsilon$ and response time of order
$N^\zeta$. A nonzero predictive projection and a smaller coupled
emission error preserve that susceptibility exponent in inference.
Equivalently, if the forcing spectral amplitude for the intensive
$\bar u_N$ is $s_{0,N}\asymp N^{-a}$, then $\upsilon=a-1$ and
\[
 \Var(\bar u_N)\sim
 \frac{s_{0,N}}{2\cos(\pi\alpha/2)}\delta_N^{\alpha-1},
 \qquad \kappa=1-a+\zeta(1-\alpha).
\]
This uses the corollary's fixed spectral shape and uniform limiting
hypotheses. For $\bar u_N\in[0,1]$, $\chi_N\le N/4$ forces
$\kappa\le1$ when the leading coefficient is positive. These are derived
conditional exponent relations, not fits to the paired interventions. The native finite
covariance forecasts test retained dependence without estimating
its thermodynamic parameters.

\section{The resulting picture}
\label{sec:conclusion}
The row map, the complete training law, and the frozen predictive
emission form a connected theory with different retained information
at each level. Absolute row energy gives a physically interpretable
quotient and an exact finite work ledger. Its radial-face cocycle
composes across zero without inventing a relative coordinate there.
The augmented optimizer cocycle identifies memory, forcing, and
observation cancellation. These results establish exact identities
and explicit sufficient conditions for attraction. They do not
establish attraction for every admissible training process.

Renormalization preserves chronology and conditioning. On realized
energies it is positive affine composition; on predictive states it
is composition of complete transition kernels; on inference it is
pushforward and elimination with a specified emission. A reduced
autonomous kernel requires closure on its declared fibers. The
executed counterexamples make that requirement operational. Retaining
conditional memory or accepting a quantified finite prediction error
is a mathematically supported response when a proposed smaller state
does not close.
Sections~\ref{rg:sec:matched-fiber-results} and \ref{rg:sec:staged-closure-results} give the counterexamples, and Sections~\ref{model:sec:conditional-memory-law} and \ref{model:sec:evolving-law-closure} derive the retained-memory and finite-error alternatives.

Single-pass training adds a resource coordinate that cannot be hidden
inside stationary noise. Remaining source positions, consumed fraction,
optimizer memory, and the learning-rate schedule specify the family.
The physical-clock and finite-flux formulas keep these choices visible.
The measured matched-clock paths still have width-dependent row motion.
Signed temporal energy records both reinforcement and cancellation;
neither sign alone identifies a connected critical covariance.
Sections~\ref{model:sec:matched-clock-results} and \ref{model:sec:equal-time-flux} give the measured clock and finite-duration energy comparisons.

The inference results show why row concentration is scientifically
useful without making it a universal explanation of prediction. A
common row can carry a learned operator; PLGA can introduce a
constant-input defect; full-graph propagation can amplify or suppress
remaining contrast. State-specific caches and categorical projections
can preserve finite predictive behavior within their measured domains.
Their error and acquisition budgets provide the relevant quantitative
claims. Prefix consistency is evaluated on the proper supplied prefix,
because a globally generated operator can depend on every query
provided to its generator.
Sections~\ref{model:sec:cache-state-results}, \ref{model:sec:categorical-scale-results} and \ref{model:sec:inference-results} provide the corresponding designs and predictive measurements.

\section{Conditional limits and unresolved mechanisms}
The conditional scaling theory identifies regular signed-operator
limits, shared-environment mixtures, connected covariance sectors,
colored forcing, and possible critical modes. These are different
classes of laws. A common random component or a moving finite-width
peak is insufficient to select a thermodynamic universality class.
The native single-pass observations support finite kinetic transport
and training-selected fluctuations, while the shrinking-window,
limiting-potential, and inference-visibility conditions remain separate
requirements.

The signed-operator mixture requires complete-law head-sign invariance,
a fixed observation dimension, a uniform bound and joint convergence of
invariant observation and covariance. Its independent Gaussian factor does
not imply independent trained heads; its covariance may be random or singular.
Physical second-moment readout conditions on the checkpoint, fitted map and
configuration law and requires $\delta_L=o(\sqrt{\mu_L})$ for positive
reference moment $\mu_L$. This transfers an existing slope. Connected
variance, generated-law transport and higher moments have separate budgets.

There are three principal unresolved mechanisms. The first is a
model-specific all-future attraction certificate that controls
transported forcing and reopenings. The second is a predictive reduced
training state whose retained dependence and acquisition cost are both
controlled across size and time. The third is a native limiting
critical law with a specified size convention and a nonvanishing
predictive projection. These questions can be pursued within the
same mathematical framework without changing the interpretation of
the exact identities or the finite evidence already established.

Further confirmation should compare complete incoming states on
disjoint remaining RefinedWeb blocks, reserve independent contexts
for assessment, and retain all prespecified outcomes. Matched time,
memory, and consumed fraction are necessary design coordinates.
The existing two-GPU producers provide those execution mechanisms;
this synthesis introduces no additional empirical hypothesis requiring
a new training campaign. The present result is a unified set of exact
and conditional laws, together with the finite observations that
identify their current domain of application.

\section{Finite conclusions and their experimental units}
The evidence follows the same chain as the theory: finite row mechanics,
chronological description, closure and visibility, finite predictive
reduction, and conditional scaling. Three source trajectories give nine
trajectory-layer units in the long row study. Its 864 maps per time
come from three layers, four heads, and 24 fixed contexts in each path;
they are not 864 independent training replicas. The row-RG study has 12
trajectory-layer cells, ten with complete positive-energy records and two
interrupted at the face. All ten complete cells fail the registered nominal
precision gate. Neither a phase direction nor marginality follows.

The single-pass refinement has six fresh initialization identities per
width/gain cell, conditional on the shared generator, corpus and observation
panels. Its 216 paths contain 884,736 updates.
Together with its coarse family, it gives 384 paths and 1,228,800 scientific
updates on two disjoint partitions of one master corpus. Intensive peak
variances at head counts $4,8,14,24$ are $0.0416,0.0374,0.0520,0.0317$;
the half-height widths $0.617,0.558,0.593,0.705$ do not resolve systematic
narrowing. Growing size times variance alone does not establish a critical
class, and a finite clock fit does not determine a universal exponent.
Section~\ref{model:sec:critical-independent-results} details this family, and Appendix~\ref{model:app:critical-independent-details} retains its complete outcomes.

Two outer initialization identities supply eight frozen incoming states,
each with 16 development, eight calibration and 16 validation continuations
on a fixed risk panel. Tubes cover 117 of 128 validation continuations, with
half-widths $0.0431$--$0.3472$ nats. The $8/9$ guarantee averages over
calibration and one future path conditional on state and fit; $117/128$ is
descriptive validation. Six paired initialization identities at each of two
widths supply twelve initial states and 24 trained endpoints for the
disjoint cache panel, with 16 calibration and 64 assessment documents. There,
all 18 recalibrated aggregate cells satisfy relative centered RMS at most
$0.25$ and mean KL at most $0.03$ nats. Observed RMS ranges from $0.0325020$
to $0.1438165$, and maximum cell mean KL is $0.0011533895$ nats, with
individual-context exceptions reaching six of 64. The six zero-control
initial-cache cells meet both targets; the six positive-control
initial-cache cells fail. These are state- and panel-specific successes,
with retained exceptions, rather than uniform accuracy across histories.

All-position training risk and proper-prefix next-token risk remain
different estimands because the native generator uses the supplied query
window before the later attention mask. In the cache measurements only
tokens before position $L$ enter the call; the target at $L$ is external.
A displaced cache can improve target NLL while reducing fidelity to the
native predictive law. This distinction is essential when translating a
geometric concentration result into an inference claim.

\FloatBarrier
\par\medskip\noindent
Appendix~\ref{app:statistics} supplies the statistical methods and
complete claim scopes supporting these conclusions. The later appendices
provide aligned outcomes, reproduction guidance, formal correspondence,
notation and the table and figure lists.

\part{Appendices}
\appendix
\chapter{Statistical methods and complete claim scopes}
\label{app:statistics}
This appendix records the statistical methods and observation boundaries
used in the experimental chapters. It identifies conditioning assumptions,
replication units and acquisition costs so that each reported conclusion can
be read at the level supported by its design.

\section{Statistical methods and the observation boundary}
\label{model:sec:statistics}

\input{content/model/generated/study-lineage.tex}

\begin{proposition}[Count-weighted replica variance]
\label{model:prop:replica-count-variance}
Let $x_1,\ldots,x_S\in\R^D$, with $S\ge2$, $D\ge1$, and let
nonnegative integer counts $n_i$ sum to $S$. Put
$\bar x_*=S^{-1}\sum_i n_ix_i$. Then
\begin{equation}
 \widehat V_* =\frac{\sum_i n_i\|x_i-\bar x_*\|^2}{(S-1)D}
 =\frac{\sum_{i<j}n_in_j\|x_i-x_j\|^2}{S(S-1)D}\ge0.
 \label{model:eq:replica-count-variance}
\end{equation}
It vanishes when all supported vectors coincide, including every
single-support count vector. The scalar identity summed over the $D$
coordinates gives the vector formula.
\end{proposition}
\begin{proof}
The centered numerator expands as
$\sum_i n_i\|x_i\|^2-S\|\bar x_*\|^2$.
Expansion of the unordered pair sum gives
$S\sum_i n_i\|x_i\|^2-\|\sum_i n_ix_i\|^2$, which is $S$ times
the centered numerator. Division proves the equality. Every pair term
is nonnegative, and every supported pair difference is zero when the
supported vectors coincide. Normalized weights $n_i/S$ sum to one;
the ordered pair version has divisor $2S(S-1)D$.
\end{proof}

The computation uses direct coordinate differences to form squared pair
distances. It never subtracts nearly equal Gram entries. Counts are
nonnegative integers with the declared total, and a zero susceptibility
profile has no resolved peak width. One shared count matrix preserves
pairing across controls, widths, times and fields. Predictive traces retain
the vocabulary sum in Table~\ref{model:tab:singlepass-observation-units}.
This real-arithmetic identity is separate from a floating-point error
certificate or a population coverage theorem.
For six seeds, $\binom{11}{5}=462$ count vectors with multiplicities
$6!/\prod_i n_i!$ represent all $6^6=46,656$ ordered resamples.
The single-support mass is $6/6^6=1/7776$. Exact empirical percentiles
use the weighted inverse CDF; the 10,000-draw summaries use interpolated
percentiles with the fixed seed 9152592. Neither convention adds training
replicas or supplies population coverage.

Training initialization, training-batch randomness and evaluation-context
sampling define different random units. Unless stated otherwise, head
susceptibilities condition on the initial shared metric learner and the
complete batch realization, center across independent initializations
at each fixed context, and then average the covariance over contexts.
Continuations, heads, decoder layers and repeated observations never
increase the number of training replicates.

For the single-pass family, the ordering is a fixed recorded permutation
and no source target position repeats within a path. Conditioning on
that entire stream differs from marginalizing its unrevealed future.
The native initialization comparisons use the former convention;
Proposition~\ref{model:prop:finite-population-innovation} explicitly uses the
latter at a frozen incoming state. Their covariances are not equated. Averaging newly sampled corpora
introduces the separate corpus-mean term in
Proposition~\ref{model:prop:corpus-conditioning}; no training
seed is counted as a corpus replication. The separate outer-state
study in Section~\ref{model:sec:outer-results} samples two corpora and four
complete initialization identities nested within them, paired across
two widths. Its 32 assessment branches per state quantify conditional
Monte Carlo variation, not eight independent outer replications.
Groups retain the exact drive horizon, source recipe, head count,
saved update, shared initialization and batch-stream identity. A
constant-rate continuation and a shorter path can enter the same
saved-time group only when their applied histories agree.

\begin{table}[htbp]
\centering\small
\caption{Conditioning conventions. Variation in one row is not a
replication of another random unit. The native initialization panel
fixes its corpus; the outer-state pilot separately samples two corpora.
Distribution-averaged limits retain their stated analytical hypotheses.}
\label{model:tab:conditioning-conventions}
\begin{tabular}{@{}>{\raggedright\arraybackslash}p{.27\textwidth}
>{\raggedright\arraybackslash}p{.29\textwidth}
>{\raggedright\arraybackslash}p{.38\textwidth}@{}}
\toprule
Objects that vary & Objects held fixed & Quantity described\\
\midrule
Nonshared model initialization &
Corpus, complete ordering, shared initialization and recipe &
Across-initialization collective covariance at each fixed evaluation
context.\\[4pt]
Next batch from remaining blocks &
Complete incoming weights, Adam state, drive and remaining corpus &
Conditional update drift and covariance; the local CPU probes restore
this same incoming state for every draw.\\[4pt]
Complete 64-update continuation paths &
Incoming model, Adam state, drive, consumed block set and fixed
evaluation cohort &
Joint future-risk law under fresh ordering of remaining blocks;
calibration and validation randomize whole reset paths.\\[4pt]
Paired 128-update pulse continuations &
Complete incoming state, remaining corpus, direction panel and pulse &
Four fresh source sequences per state, shared across nine arms;
covariance uses divisor three and does not estimate outer-model
uncertainty.\\[4pt]
Evaluation contexts or tasks &
One trained checkpoint and its inference rule &
Context dispersion or task behavior under the specified finite
observation law.\\[4pt]
Newly sampled corpus and its ordering &
Corpus-generation and initialization laws, training recipe &
Distribution-averaged training law, including variation of the
corpus-conditioned mean.\\
\bottomrule
\end{tabular}
\end{table}

One-initialization data-law comparisons report paired means and
checkpoint observations without a covariance or uncertainty interval.
For two or four initializations, every $s^s$ ordered whole-identity
empirical resample is retained. Removing one of two identities cannot
estimate a variance; the corresponding jackknife is unavailable.
An endpoint field observed only at preselected final checkpoints names
its selected and unselected initialization identities explicitly.
An undefined measured ratio is retained as undefined and prevents the
selected-identity mean. An unselected endpoint is not treated as a
measured zero or omitted silently from a four-identity estimand.

The selection records preserve information already available when each
measurement was specified. The conditional native-risk probes follow
the observed source-recipe geometry, decoder interventions and late loss
rise of the first single-pass $N=4$ constant-rate path. They test local
mechanisms at selected incoming states. The paired constant-versus-cosine
analysis was fixed before its compact controlled trajectories started,
with some constant-rate outcomes already available. Complete paired
identity coverage is retained; neither adaptive checkpoint choice nor
reuse of a trajectory is interpreted as untouched population confirmation.
The source-recipe comparison between cosine horizons is a secondary
comparison of the already selected paths. Its pairings retain the same
initial parameters, recorded sampling prefix and common observation
cohorts, while the post-warmup drive history differs.

The conditional path study fixes its incoming-state selection before
fresh calibration and validation. Development branches determine the
risk center and score scales within each state. Calibration and
validation branches then use separate saved random identities and
uniform sampling without replacement from the remaining blocks.
Each branch has a distinct source prefix internally; overlap between
counterfactual branches does not constitute reuse along a training
trajectory. Its eight prediction strata contain only two outer
initialization identities. The 128 validation paths quantify the
conditional continuation law, not population variation across
pretrained models. Whole-path coverage and the calibration-averaged
rank guarantee are reported separately in
Section~\ref{model:sec:law-closure-results}.

The fresh covariance and memory test in Section~\ref{model:sec:memory-results}
uses those states again, with 40 available paths per state assigned
to development and 32 further paths to validation. Its covariance
forecasts use divisor 39 and its validation covariances divisor 31;
MSE uses divisor 32. Each of the eight states has equal weight in
aggregate errors. The two paired initialization identities remain
the outer units. All fixed predictors and per-state errors are
reported without a population confidence interval. History forecasts
use only their current branch's observations through update 16.

The primary predictive comparison uses identical held-out contexts.
Risk on consumed training blocks characterizes fit under the recorded
seen-data law, whose cohort can change with time. All-position
parallel risk and proper-prefix risk remain separate because the native
global metric can depend on later input tokens. Task accuracy uses its
fixed question law and correct-answer margins, rather than a language
entropy or operator-dispersion proxy. Detailed methods for the
auxiliary finite-law studies are in Section~\ref{model:sec:auxiliary-methods}.

For $s$ seeds, $M$ contexts and $L$ decoders the scalar row estimator is
\begin{equation}
 \widehat\chi_N=\frac{N}{(s-1)ML}
    \sum_{i,x,\ell}(q_{i,x,\ell}-\bar q_{x,\ell})^2,
 \qquad q_{i,x,\ell}=N^{-1}\sum_a R_{i,x,\ell,a}.
 \label{model:eq:sample-chi-method}
\end{equation}
For a horizon difference $\widehat\Delta$, the same resampled seed
indices are applied to both horizons before subtraction. The jackknife
standard error is
\[
 \left[\frac{s-1}{s}\sum_i
 (\widehat\Delta_{(-i)}-\overline{\widehat\Delta}_{(-\cdot)})^2
 \right]^{1/2}.
\]
At four seeds the 256 ordered resamples are equiprobable; the 35
distinct count vectors have their multinomial weights. Reported
percentiles describe this discrete empirical distribution, with no
guarantee of population or simultaneous coverage. Resampling can
produce variances above the observed estimate: $(0,0,0,1)$ has
unbiased variance $1/4$, whereas $(0,0,1,1)$ has variance $1/3$.
An empirical sign fraction is neither a posterior probability nor a
significance-test $p$-value.

For paired float32/float64 collective observations, stored training
weights remain fixed. Native row fields, absolute centered energies,
common metric coordinates and predictive fields are retained separately.
The analysis reports the observed susceptibility discrepancy and its
paired-error bound. Its descriptive threshold is the larger of
$10^{-8}$ in the stated absolute susceptibility units and one percent
of the float64 value. This comparison concerns two implemented
arithmetic programs; neither program is an exact-real oracle. No
threshold defines a physical phase or deletes a small native field.

Retained native float32 risk and entropy reductions are checked by
bytewise replay on their saved vocabulary logits. Their differences
from float64 mathematical softmax reductions are reported separately;
a fixed absolute screen is not a universal rounding-error bound.
The arithmetic qualification also checks the float64 references
against extended-precision reductions. All 3,840 retained scalar
reductions in its six states replay bytewise, and the largest
float64--extended-precision reference difference is
$1.17\times10^{-14}$. These checks concern the retained logits and
their reductions. They do not certify a full native forward at every
checkpoint or exact-real arithmetic for the underlying model.

Temporal diagnostics use both centered and linearly detrended windows,
with every-update native increments separated from fixed-context
emissions on the regular 64-update grid. A window never crosses an
imposed drive phase. Constant external rates still coexist with a
changing remaining corpus in a single pass. The threshold intervals
retain observed onset and offset bounds, including censoring. A
finite temporal trend, small variance or threshold crossing is not
assigned a relaxation exponent or avalanche interpretation.

\section{Methods of the auxiliary finite-law studies}
\label{model:sec:auxiliary-methods}

These measurements test additional exact identities and finite reductions
under their separately recorded laws. The controlled training and
conditional fresh-batch probes here use the small resampled corpus;
their empirical coefficients are restricted to that law. Released-model
and long-document observations retain their own specified input laws.
They supply auxiliary checks, while the primary training interpretation
uses the single-pass results in Section~\ref{model:sec:onepass-results}.
The released-model, short controlled-training and extended size--time results are in Sections~\ref{model:sec:methods}, \ref{model:sec:controlled-results} and \ref{model:sec:scaling-results}.

\small
\begin{longtable}{>{\raggedright\arraybackslash}p{.20\textwidth}>{\raggedright\arraybackslash}p{.28\textwidth}>{\raggedright\arraybackslash}p{.23\textwidth}>{\raggedright\arraybackslash}p{.20\textwidth}}
\caption{Estimands, independent units and adaptive choices. Exact
cohort ranges are in Table~\ref{model:tab:rows}.}
\label{model:tab:statistical-methods}\\
\toprule Study & Estimand and divisor & Random unit and uncertainty & Pairing and selection\\
\midrule\endfirsthead
\toprule Study & Estimand and divisor & Random unit and uncertainty & Pairing and selection\\
\midrule\endhead
Conditional single-pass risk paths & Joint coverage at horizons
$1,4,16,64$; interval covariance uses divisor sixteen & Whole reset
path conditional on one complete incoming state; eight strata from
two paired initializations & Sixteen development, eight fresh
calibration and sixteen fresh validation paths per displayed state;
all selected validation paths retained.\\
Corpus and complete-initialization transfer & Risk and joint increment
scores; covariance fit uses divisor fifteen & 32 whole assessment paths
per state; paired sample Monte Carlo errors conditional on adaptation
and the fixed document panel & Two corpora, two nested initialization
identities each, paired at $N=4,14$. Sixteen adaptation and eight
calibration paths precede assessment; all 48 state/target/scale cells
and all three covariance comparisons retained.\\
Outer empirical hierarchy & Endpoint-risk variance with population
divisors for the specified finite law & Uniform corpus label, uniform
nested initialization label and uniform assessment branch & Three
exact sectors; branches are not independent corpora or initializations.\\
Small-radius single-pass response & Full-path odd-pulse discrepancy;
equal weight on five times and twelve coordinates & Four source draws
conditional on each of two incoming states; source labels repeat the
same time-zero observation & Radius selected by a separate instantaneous
diagnostic, then fixed before fresh paths. All signs, controls and
32 primary cells retained.\\
Input-support null sector & Precomputed input-row decay and zero paired
emission difference & Two conditional source draws per width; no added
corpus or complete initialization & Paired source indices and complete
states, with intervention fixed before execution. All four comparisons
and all saved update times retained.\\
Frozen inference fields & Joint context covariance, document sums,
contrasts and predictive response & Documents; covariance and response
use their declared calibration and evaluation splits & Head and decoder
coordinates observed jointly; response formulation follows a pilot.\\
Long documents & Covariance of block sums and contrasts in a fixed
16-dimensional projection & Whole documents resampled within shard
strata; retained projection held fixed & Adjacent segments stay together;
projection and fit documents precede evaluation.\\
Short controlled training & Finite three-checkpoint mixture; population
divisor for that finite law & Three initializations describe a finite
ensemble; quality intervals resample documents within strata & Branches
share parent states and contexts; normalization is developmental.\\
Conditional rate and width scans & $\widehat\chi_N=N(s-1)^{-1}
\sum_i\|q_i-\bar q\|^2$, $s=4$ & Independent initialization; jackknife
or whole-seed empirical resampling & Common environment and fixed
contexts; targeted rate choices are developmental.\\
Width replication & Same conditional susceptibility, $s=16$ at
$N=2,4,8,14$; $s=4$ at $N=24$ & Divisor fifteen or three;
5,000 whole-seed resamples & Shared seed identities across widths
and times; fixed observation law.\\
Paired horizon comparison & Difference of the two conditional
susceptibilities & Exact $4^4$ resamples for $s=4$; 20,000 paired
resamples for the $N=14$ extension, with jackknife and leave-one-out
diagnostics & Each seed retains both horizons. $N=14$ selected because
the four-seed difference was unresolved.\\
Arithmetic observation & Unscreened row susceptibility; paired
discrepancy and its bound & Same four trained seeds and 512 contexts;
no new training replicates & All thirteen selected conditions and
all seeds retained; tolerances fixed before these calls.\\
Shared crossing and return & Finite product-law ANOVA or paired
trajectory difference & Donors and recipients are four parent
identities; combinations are not independent pretrainings & Same
contexts and batches; all stated interventions retained.\\
Row projection and adjoints & Predictive KL, sample susceptibility,
and full-loss shared gradient & Four initializations per width;
matched context or minibatch arrays & All reductions, both horizons
and both matched batches; indices and tie rule retained.\\
Tangent and initialization kernel & Local derivative error or numerical
kernel-integration error & Checkpoint--direction controls or four
Monte Carlo replicas, respectively & Amplitude refinement and repeated
initializations are numerical controls, not new trained seeds.\\
Joint size--time and longer horizons & Fixed-unit conditional covariance,
$s=4,8,16$ at each complete balanced group; $N$ is a count & Exact 256
resamples for four identities; 20,000 for eight or sixteen & Continued
paths retain identities, full moments and batch prefixes. All adjacent
balanced horizon differences are retained.\\
Common-centroid geometry & Exact pairwise split of the same
conditional susceptibility into radial and directional contributions &
The existing initialization identities and fixed contexts; jackknife
sensitivity across identities & Native coordinates are held fixed;
no alignment or conditioning on the learned direction is applied.\\
Frozen rate--time predictions & Interval-censored first passage; transported
seed fields and their covariance & Four paired initialization labels;
empirical refits and paired target errors & All 54 path alternatives and
three inverse-$N$ predictions frozen before target execution.\\
Frozen variance predictions & Four finite susceptibility alternatives
for nine fields & Distinct four-identity calibration and confirmation
laws; product resampling & All 144 scalar predictions retained; fitted
calibration excludes explicitly counted zero-variance resamples.\\
Crossed training environments & Seven orthogonal population-variance
sectors of a finite $2\times2\times4$ law & Four initialization identities,
paired across two shared states and two entire batch histories & Main
and interaction sectors retained; crossed environments are not IID
pretrainings.\\
Fresh conditional optimizer noise & Conditional drift and unbiased noise
trace at a complete incoming state & 32 fresh minibatches per state;
eight matched source--emission directions & 24 states, all amplitudes;
this fresh-batch law differs from the fixed-history seed law.\\
Frozen visible sources & Total predictive second moment and innovation
covariance residuals, with separate denominators & Sixteen fresh batches
per incoming state; four paired state identities per condition & Basis
and full-graph adjoints frozen from eight calibration directions;
all five dimensions and all sixteen states retained.\\
Expanded source basis & Conditional mean response and noise covariance
on a common fresh law & 32 disjoint validation batches per state;
four matched identities per width--time condition & Both 32-calibration
orderings and the frozen eight-calibration reference share the same targets;
all dimensions and all sixteen states retained.\\
Native source corners & Conditional complete increments, finite
mixed differences and all signed covariance cross terms & 32 fresh
batches at each of sixteen full incoming states & Every draw restores
the complete model and Adam state. Partial weight corners share the
actual globally clipped displacement; they are not separately trained
models.\\
Passive temporal observations & Finite-window moments and lagged
covariances, with and without linear detrending & One path is one
correlated temporal record & All selected windows and mesh-censored
threshold excursions retained; no avalanche law is assumed.\\
\bottomrule
\end{longtable}
\normalsize

The arithmetic comparison holds stored weights fixed and measures
archived CUDA float32 fields, CPU float32 forward fields accumulated
in float64, and CPU float64 forward and statistic fields. It retains
absolute centered and total matrix energies and paired vocabulary KL.
The primary entry screen is $R<10^{-13}$, with fixed diagnostic levels
$10^{-15},10^{-14},10^{-13},10^{-12}$. These dimensionless screens
are not subtraction rules or certified arithmetic floors. A cell
meets the declared quantitative reporting tolerance only if the
paired susceptibility bound is at most 1\% relatively and $10^{-8}$
absolutely. The actual difference, the conservative bound, and failure
to meet that criterion remain separate quantities. These tolerances
specify a finite reporting use and do not define a physical phase.

A 32-context developmental pilot for each released model gave
mean-matrix transfer ratios 0.277--0.569 for the first model and
0.329--1.013 for the second. The context-conditioned response formulation was specified
after that diagnostic; the pilot arrays are retained in the evidence
object and are not counted as independent confirmation data.
This selection fact applies to the response formulation, not to an
additional training-seed ensemble.

\section{Claim assumptions, replication units and acquisition costs}
\label{model:app:evidence-scope}
This view and Table~\ref{model:tab:claim-index-main} are generated from the same
maintained claim records. Every proof retains its assumptions at the point
of use. A reference to complete evidence identifies the full outcome grid,
including controls, rather than a selected favorable subset. The source
bundle's \path{docs/RELEASE_INDEX.json} binds each named reconstruction
route to commands, dependencies, availability, toolchains and qualification.
It distinguishes a typesetting package from access to local raw data.
\subsection*{C1: Compatible complete laws}

\phantomsection\label{model:claim:C1}

\textbf{Assumptions and written basis.} One conditioned law; full parameters, optimizer memory, remaining identities, clocks and shared observation measure. See \ref{model:sec:maps}, \ref{model:sec:data-resource}.

\textbf{Complete evidence.} Recorded single-pass histories and native replay. See \ref{model:sec:matched-clock-results}.

\textbf{Replication and cost.} A complete initialization, conditional on the corpus and order. Native path acquisition; exact composition itself requires no fitted coefficients.

\textbf{Remaining obligation.} An economical retained state needs an additional successor law.

\textbf{Reconstruction route.} \texttt{law} in \path{docs/RELEASE_INDEX.json}.

\subsection*{C2: Native signed class}

\phantomsection\label{model:claim:C2}

\textbf{Assumptions and written basis.} Head-sign invariant initialization and optimizer, uniformly bounded signed operators, convergent orbit-covariance law. See \ref{model:prop:native-head-sign-symmetry}, \ref{model:thm:native-sign-gaussian-mixture}.

\textbf{Complete evidence.} Complete operator observations distinguish signed and invariant fluctuations. See \ref{model:sec:critical-independent-results}.

\textbf{Replication and cost.} Whole initialization; heads and decoders are inner observations. Retained all-head fields from the completed single-pass family.

\textbf{Remaining obligation.} Trained covariance convergence; deterministic Gaussian behavior additionally needs covariance concentration.

\textbf{Reconstruction route.} \texttt{singlepass} in \path{docs/RELEASE_INDEX.json}.

\subsection*{C3: Conditional training reduction}

\phantomsection\label{model:claim:C3}

\textbf{Assumptions and written basis.} Evolving-law successor defect, coarse stability, initial-law discrepancy and emission error. See \ref{model:prop:law-closure}, \ref{model:prop:finite-optimizer}.

\textbf{Complete evidence.} Moment-only interventions and source-return observations retain memory and representation errors. See \ref{model:app:finite-native-transport}.

\textbf{Replication and cost.} Paired continuations conditional on complete incoming states. Native continuation and coefficient acquisition are included in the study-specific budgets.

\textbf{Remaining obligation.} Held-out multistep successor accuracy and total cost advantage for an autonomous reduction.

\textbf{Reconstruction route.} \texttt{closure} in \path{docs/RELEASE_INDEX.json}.

\subsection*{C4: Finite collective row flow}

\phantomsection\label{model:claim:C4}

\textbf{Assumptions and written basis.} Nonzero endpoint energies, fixed averaging, actual chronological increments; derivative bounds additionally need small relative increments. See \ref{model:prop:collective-clock}, \ref{model:lem:finite-defect}, \ref{model:cor:row-path-error}, \ref{model:prop:finite-flux-forecast}, \ref{model:prop:energy-cocycle}, \ref{model:prop:temporal-energy}.

\textbf{Complete evidence.} Eight single updates, sixteen fixed-count continuations and sixteen equal-duration continuations resolve finite transport and energy-coordinate blocking. Every derivative diagnostic is retained. Complete signed temporal energy is reconstructed independently by ordered pair products on all 64 equal-duration cells. See \ref{model:sec:collective-flux-results}, \ref{model:sec:row-path-results}, \ref{model:app:row-quadrature}, \ref{model:sec:equal-time-flux}.

\textbf{Replication and cost.} One initialization for single updates; two other initialization identities reused by both continuation studies. 8 single-update steps and 40 forwards; fixed-count paths: 128 steps and 272 forwards. Equal-duration paths: 1,600 executed updates (128 prefix replays and 1,472 additional updates), 3,216 forwards. Separate quadrature: 8 replays and 272 forwards.

\textbf{Remaining obligation.} A flux successor law and uniform accumulated generator-remainder control.

\textbf{Reconstruction route.} \texttt{row / equal-time} in \path{docs/RELEASE_INDEX.json}.

\subsection*{C5: Source and physical clocks}

\phantomsection\label{model:claim:C5}

\textbf{Assumptions and written basis.} Specified width, resource and time family and matched Adam memory; reference covariance uses a frozen population and deterministic transports. See \ref{model:prop:matched-memory-force}, \ref{model:prop:consuming-bridge}, \ref{model:cor:consuming-clock}.

\textbf{Complete evidence.} 48 paths at four widths and four controls; complete 64 field and 26 cache cells. See \ref{model:sec:matched-clock-results}.

\textbf{Replication and cost.} Three whole initialization identities, paired across width/control/age; one fixed corpus/order. 76,800 scientific updates, 3.391 aggregate GPU worker-hours; qualification and replay separate.

\textbf{Remaining obligation.} A trained limit and native reference-remainder estimates; independent corpus variability.

\textbf{Reconstruction route.} \texttt{clock} in \path{docs/RELEASE_INDEX.json}.

\subsection*{C6: Predictive and cache reduction}

\phantomsection\label{model:claim:C6}

\textbf{Assumptions and written basis.} Common reference for nested observation; state-specific calibration, complete context-risk and operator-displacement budgets. See \ref{model:prop:finite-density}, \ref{model:prop:tail-risk}.

\textbf{Complete evidence.} All stated state-calibrated aggregate targets pass on the two disjoint state panels; unchanged-initial-cache controls and context exceptions remain visible. See \ref{model:sec:cache-state-results}, \ref{model:app:cache-state-details}.

\textbf{Replication and cost.} Whole initialization with paired contexts; context counts do not add trained models. Inference-only acquisition, calibration and timing costs remain separated in the complete study records.

\textbf{Remaining obligation.} Population guarantees and justified downstream sensitivity; workload-specific timing does not establish general throughput.

\textbf{Reconstruction route.} \texttt{cache} in \path{docs/RELEASE_INDEX.json}.

\subsection*{C7: Conditional critical scaling}

\phantomsection\label{model:claim:C7}

\textbf{Assumptions and written basis.} Declared resource limit, physical response and forcing scales, and nonvanishing predictive coupling. See \ref{model:cor:colored-exponents}, \ref{model:prop:selection-window}.

\textbf{Complete evidence.} Finite kinetic observations and covariance/response controls delimit the applicable conditional theory. See \ref{model:sec:critical-independent-results}.

\textbf{Replication and cost.} Whole paths and initialization identities at fixed corpus law. Completed native measurements and finite reference controls only.

\textbf{Remaining obligation.} Native critical exponents, a critical surface, restoring feedback and inference persistence remain unestablished.

\textbf{Reconstruction route.} \texttt{critical} in \path{docs/RELEASE_INDEX.json}.

\subsection*{S1: Moment precision and acquisition}

\phantomsection\label{model:claim:S1}

\textbf{Assumptions and written basis.} Deterministic linear blocking, finite moments, fixed regularization, relative mean and covariance control. See \ref{model:prop:moment-blocking}, \ref{model:prop:moment-precision}, \ref{model:prop:correlation-transfer}.

\textbf{Complete evidence.} Primary outer risk favors local prediction in 4/8 states and transfer in 6/8; both secondary joint scores favor 8/8 at all three scales. See \ref{model:app:evidence-scope}.

\textbf{Replication and cost.} Incoming state; the local conditional paths are its inner sample. 16 paths of 64 updates acquire each local mean/scale (1,024 updates); 32 assessment paths add 2,048 updates. Zero-adaptation means miss the 0.01-nat maximum target in all eight states.

\textbf{Remaining obligation.} Population-relative precision and an autonomous moment successor.

\textbf{Reconstruction route.} \texttt{moments} in \path{docs/RELEASE_INDEX.json}.

\subsection*{S2: Calibration and state adaptation}

\phantomsection\label{model:claim:S2}

\textbf{Assumptions and written basis.} Conditional exchangeability; adaptation-fixed center/scales and explicit source-edit coupling. See \ref{model:prop:calibrated-scale-images}, \ref{model:prop:state-moment-reuse}, \ref{model:prop:resource-edit-coupling}.

\textbf{Complete evidence.} 16 parent/successor states, two fit cohorts and 512 assessment paths; refitting lowers primary scores in all 16 successor/fit cells per candidate. See \ref{model:app:evidence-scope}.

\textbf{Replication and cost.} Conditional path at its declared incoming state. Eight further calibration paths add 512 updates per state; 237/256 assessment paths are covered, with endpoint half-widths 0.120--0.275 nats.

\textbf{Remaining obligation.} Uniform mean accuracy, population drift control and economical reuse.

\textbf{Reconstruction route.} \texttt{moments} in \path{docs/RELEASE_INDEX.json}.

\subsection*{S3: Finite pulses and optimizer memory}

\phantomsection\label{model:claim:S3}

\textbf{Assumptions and written basis.} Complete-state coupling and symmetric/antisymmetric response; fixed-gradient finite moment identities. See \ref{model:prop:finite-pulse}, \ref{model:prop:finite-optimizer}.

\textbf{Complete evidence.} 288 long and 144 short paired continuations retain all amplitudes, arithmetic controls and signed covariance sectors. See \ref{model:app:finite-native-transport}.

\textbf{Replication and cost.} Paired continuation from a complete incoming state. All branch, replay and qualification roles are retained in the execution ledger.

\textbf{Remaining obligation.} Uniform derivative control, native gap/forcing law and naturally sampled memory generalization.

\textbf{Reconstruction route.} \texttt{closure} in \path{docs/RELEASE_INDEX.json}.

\subsection*{S4: Inference-null mode}

\phantomsection\label{model:claim:S4}

\textbf{Assumptions and written basis.} Excluded input support and untied input embeddings. See \ref{model:prop:invisible-decay}.

\textbf{Complete evidence.} Four fresh paired comparisons match row decay while preserving every saved logit. See \ref{model:prop:invisible-decay}.

\textbf{Replication and cost.} Fixed incoming state and paired intervention. Four recorded paired comparisons; no new pretraining sample.

\textbf{Remaining obligation.} Visibility under a changed input law.

\textbf{Reconstruction route.} \texttt{visibility} in \path{docs/RELEASE_INDEX.json}.

\subsection*{S5: Finite training selection}

\phantomsection\label{model:claim:S5}

\textbf{Assumptions and written basis.} Matched objectives, ordered distinct blocks, initialization and complete optimizer/schedule. See \ref{model:sec:training-selection-main}.

\textbf{Complete evidence.} A four-width, three-seed matched family favors cosine risk in all twelve pairs; row-contrast changes have both signs. See \ref{model:sec:training-selection-main}.

\textbf{Replication and cost.} Three whole initializations paired across conditions. 24 paths and 24,576 scientific updates; 64 qualification replay updates counted separately.

\textbf{Remaining obligation.} Corpus-law replication and a selection mechanism; row suppression alone does not establish lower target NLL.

\textbf{Reconstruction route.} \texttt{singlepass} in \path{docs/RELEASE_INDEX.json}.

\subsection*{S6: Inference task fidelity}

\phantomsection\label{model:claim:S6}

\textbf{Assumptions and written basis.} Proper-prefix observations, intervention bounds and adequate correct-answer margins. See \ref{model:prop:task-margin}.

\textbf{Complete evidence.} Full-vocabulary interventions and premise-pair evaluations. See \ref{model:prop:task-margin}.

\textbf{Replication and cost.} Checkpoint and paired proper-prefix task context. Native evaluation and intervention calls, with retained task-level outcomes.

\textbf{Remaining obligation.} Acquisition of premise-sensitive competence and broader task transfer.

\textbf{Reconstruction route.} \texttt{visibility} in \path{docs/RELEASE_INDEX.json}.

\FloatBarrier

\chapter{Complete aligned experimental outcomes}
\label{app:outcomes}
This appendix collects the complete aligned experimental outcomes behind
the selected results in the main chapters. It includes width and time
profiles, cache and continuation studies, finite transport checks and matched
intervention grids with their stated observation conventions.

\section{Complete conditional critical-region observations}
\label{model:app:critical-onepass-details}
The following table contains every coarse endpoint cell. The empirical
fourth-moment statistic $U_4$ centers across initializations separately
at each context and decoder before pooling the central moments; it is
not a thermodynamic Binder limit. NLL averages the 128 fixed held-out
documents. All six initializations are included in every cell.
\begingroup\footnotesize
\input{content/model/generated/critical-coarse-publication-complete.tex}
\endgroup

\begin{figure}[htbp]
\centering
\includegraphics[width=\textwidth]{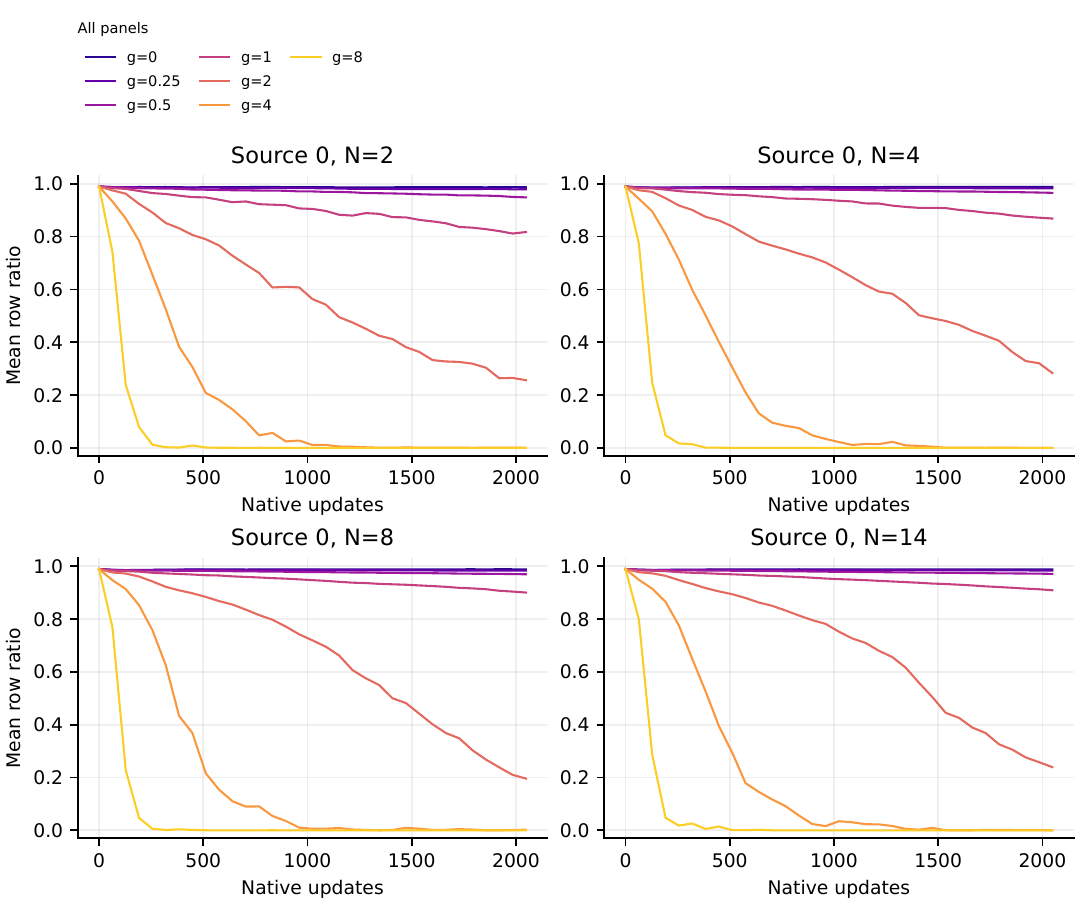}
\caption{All coarse mean row trajectories on the declared 64-update
observation cadence. Each point averages six complete initialization
realizations and the fixed context, decoder and head panel. The
conditional source stream contains distinct source block identities
within each path.}
\end{figure}

\begin{figure}[htbp]
\centering
\includegraphics[width=\textwidth]{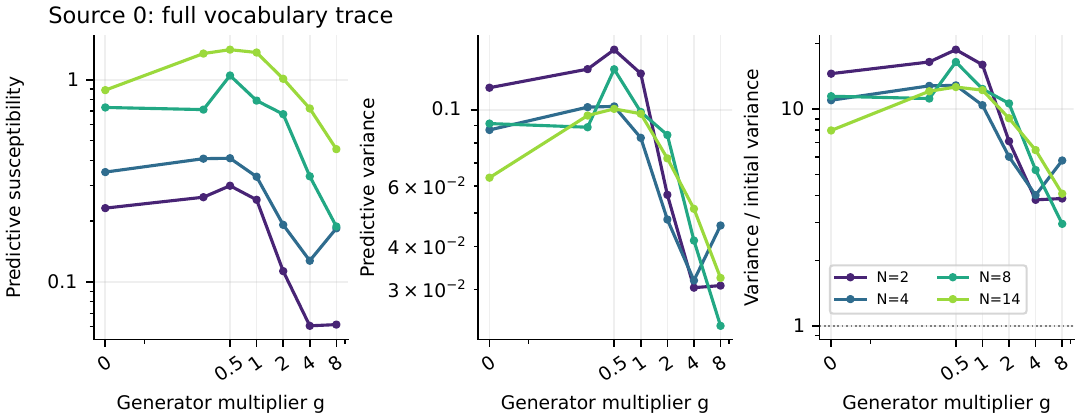}
\caption{Complete predictive-law fluctuations at the trained coarse
endpoints. The vocabulary-summed covariance trace of $2\sqrt p$ is
reported with and without its factor of $N$, and relative to the
identical-unit initialization baseline. Independent unordered Hellinger
pairs reconstruct the trace. These are differences among trained
predictive laws, not responses to an applied parameter perturbation.}
\end{figure}

\subsection{Complete conditional sign-orbit budgets}
The normalized fourth-cumulant terms retain each fixed context,
decoder and projection before averaging. The signs and the finite
mixture of saved orbits use population moment divisors. They supply
conditional symmetry calculations for the original training states.
\begingroup\small
\input{content/model/generated/critical-coarse-sign-limit.tex}
\endgroup

\section{Complete independent single-pass observations}
\label{model:app:critical-independent-details}
All 36 endpoint cells retain the six declared initializations. The
fourth-moment statistic has the same conditional centering and finite
scope as in Appendix~\ref{model:app:critical-onepass-details}.
\begingroup\footnotesize
\input{content/model/generated/critical-refinement-publication-complete.tex}
\endgroup

\begin{figure}[htbp]\centering
\includegraphics[width=\textwidth]{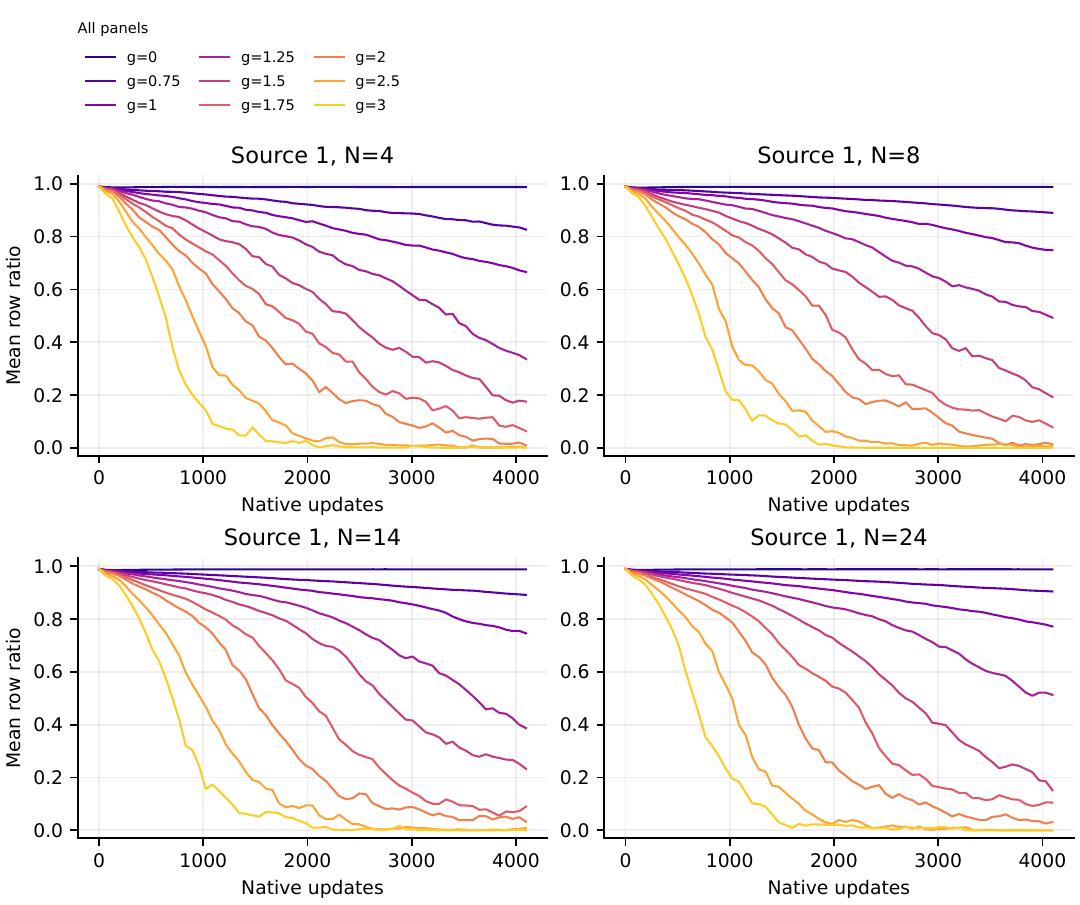}
\caption{Complete independent mean-row trajectories on the recorded
64-update cadence, including every control and width.}
\end{figure}

\begin{figure}[htbp]\centering
\includegraphics[width=\textwidth]{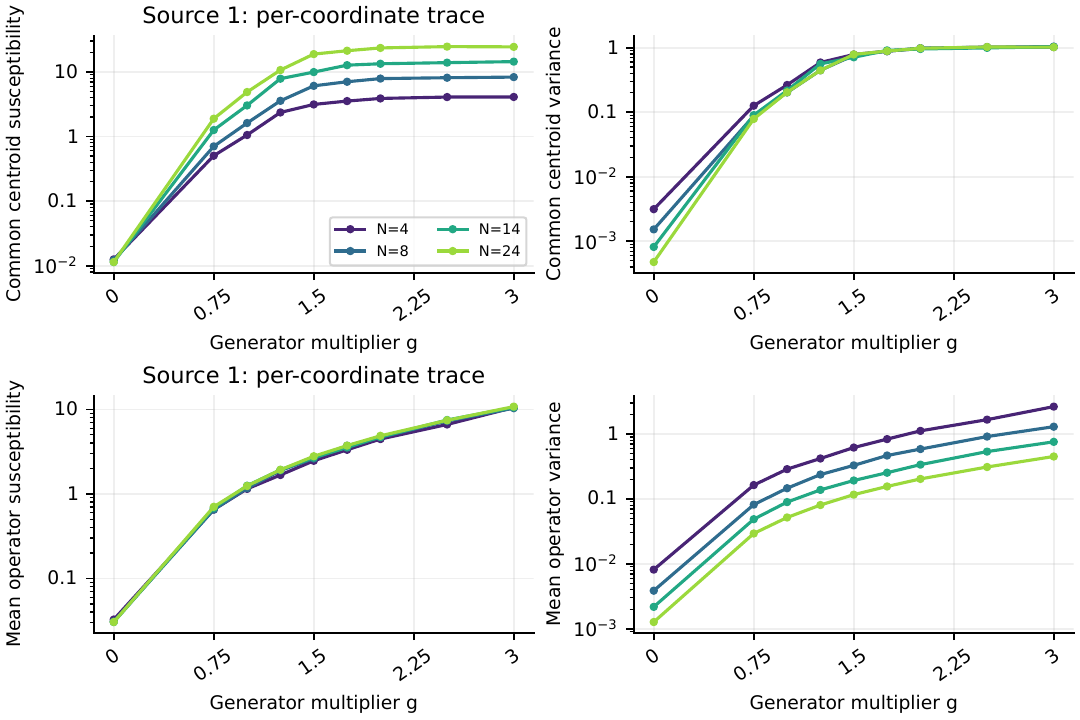}
\caption{Independent complete common-centroid and head-mean operator
observations. Each covariance trace uses its stated per-coordinate
normalization; the susceptibility multiplies it by head count.}
\label{model:fig:critical-independent-fields}
\end{figure}

\begin{figure}[htbp]\centering
\includegraphics[width=\textwidth]{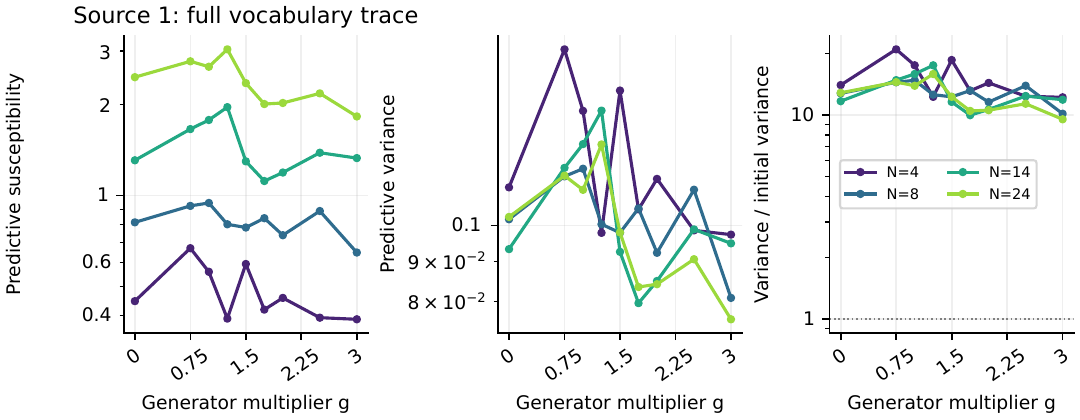}
\caption{Independent full-vocabulary predictive variation in the
$2\sqrt p$ embedding, with the same units at initialization and after
training. Initialization variation is retained as a separate reference.}
\label{model:fig:critical-independent-predictions}
\end{figure}

\subsection{Complete shared-parameter budgets}
The diagonal and signed cross-block contributions use the same
256-update partition. Their sum reconstructs the endpoint variance.
The finite-sample squared-mean correction is retained even if negative.
\begingroup\small
\input{content/model/generated/critical-refinement-detail-shared-budget.tex}
\endgroup
\begin{figure}[htbp]\centering
\includegraphics[width=\textwidth]{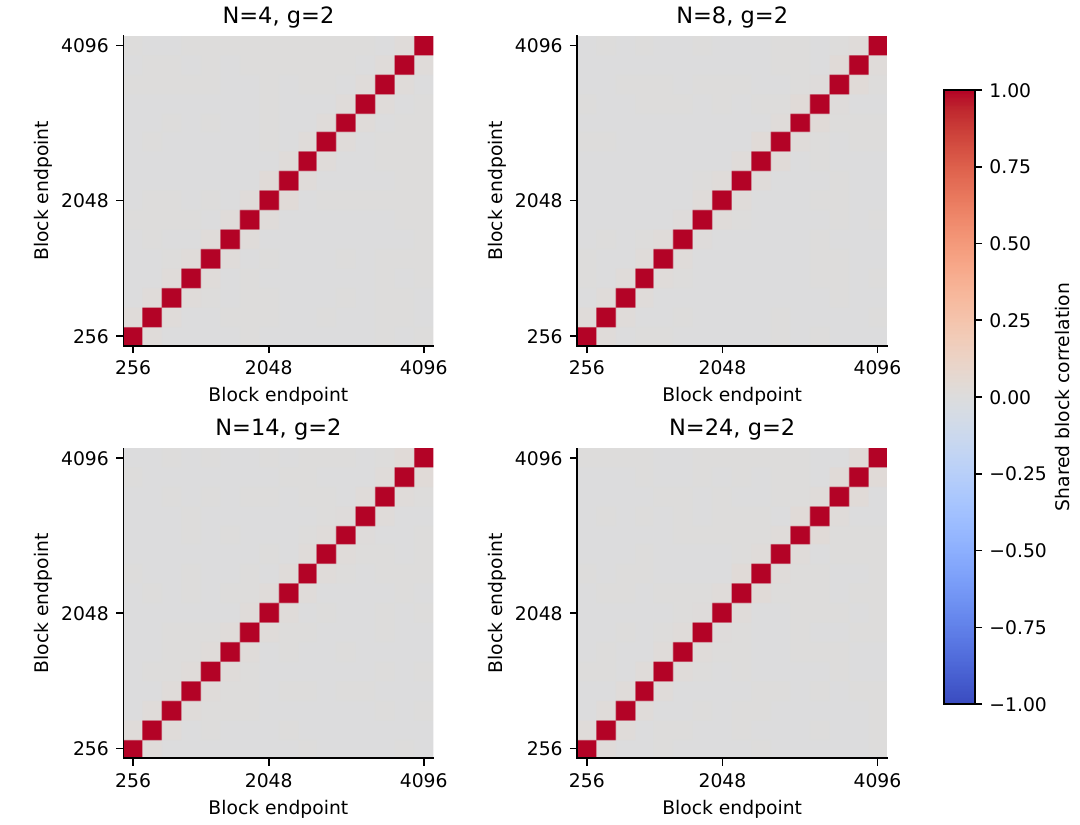}
\caption{Complete shared-parameter block-correlation matrices at the
registered reference $g=2$. These are initialization covariances under
one source order; they are not fresh-batch innovation covariances.}
\label{model:fig:critical-independent-blocks}
\end{figure}

\subsection{Complete conditional sign-orbit budgets}
The same fixed projection panel and exact conditional sign formulas
are used at every independent endpoint. Population orbit moments do
not add independent training realizations.
\begin{figure}[htbp]\centering
\includegraphics[width=\textwidth]{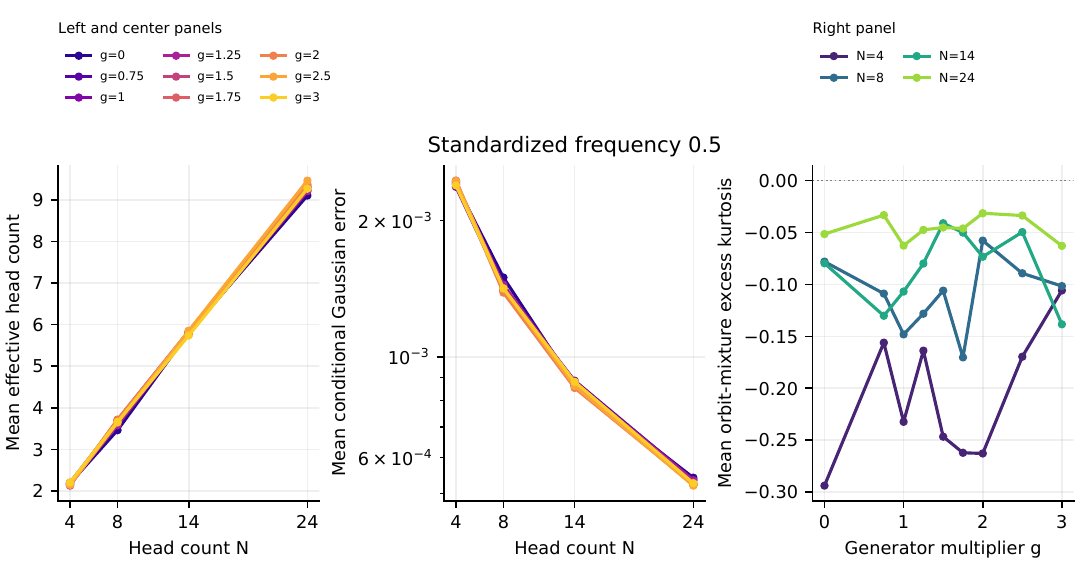}
\caption{Independent conditional Gaussian comparisons and finite
fourth-cumulant budgets. Effective-head counts and bound-domain
fractions retain finite head domination explicitly. The 24-head
product formula is evaluated directly without enumerating all signs.}
\label{model:fig:critical-independent-sign-limit}
\end{figure}
\begingroup\small
\input{content/model/generated/critical-refinement-sign-limit.tex}
\endgroup

\subsection{Coupled-seed uncertainty and all time windows}
All reported intervals below use one joint seed resampling across the
control, width and time coordinates. The fixed-grid interpolation
segments remain separate from these finite-seed percentiles.
\begingroup\small
\input{content/model/generated/critical-refinement-joint-profiles.tex}
\input{content/model/generated/critical-refinement-joint-drift-windows.tex}
\endgroup

\section{Complete predictive resolution and width profiles}
\label{model:app:predictive-resolution-details}
The main results use fixed dictionaries, fixed context laws and whole
initialization replicas. This appendix retains every resolution and
width profile, including all cells outside the finite accuracy target.
Table~\ref{model:tab:context-risk-main} reports context-resolved summaries at
the fixed 8,192-token scale. Tables below retain the other scales and
absolute variances; their ranges describe cells rather than confidence
intervals. The complete per-context arrays are in the bound observation
package and compact numerical records.
Sections~\ref{model:sec:categorical-scale-results} and \ref{model:sec:categorical-visibility-results} define the corresponding measurement designs and observation laws.

\subsection{Predictive resolution across training ages}
Tables~\ref{model:tab:categorical-visibility} and
\ref{model:tab:nested-categorical} collect the fixed-resolution summaries.
The training-age profiles reuse the same initialization replicas and fixed
contexts.

\input{content/model/generated/categorical-visibility.tex}
\input{content/model/generated/nested-categorical.tex}
\clearpage
\subsection{Disjoint context panels}
Tables~\ref{model:tab:context-categorical},
\ref{model:tab:context-transfer} and \ref{model:tab:context-risk-scales}
report all fixed resolutions on the disjoint-context panels at the retained
endpoint.

\begin{table}[htbp]\centering\small
\caption{Complete disjoint-context panel at the retained endpoint. Each of four width/control cells uses six complete initialization replicas and the same 32 fixed contexts. A single context-independent reference is fixed from development data.}
\label{model:tab:context-categorical}
\begin{tabular}{@{}rrrrrr@{}}
\toprule $T$ & Tokens & Target cells & Centered RMS & Variance retained & Mean KL\\\midrule
4096 & 128 & 0/4 & 0.458--0.510 & 76.1--80.6\% & 0.129--0.152\\
4096 & 512 & 0/4 & 0.351--0.384 & 85.6--88.9\% & 0.094--0.101\\
4096 & 2,048 & 0/4 & 0.256--0.285 & 91.9--94.3\% & 0.056--0.061\\
4096 & 8,192 & 4/4 & 0.176--0.197 & 96.1--97.2\% & 0.023--0.025\\
4096 & 16,384 & 4/4 & 0.129--0.144 & 97.9--98.5\% & 0.011--0.012\\
\bottomrule\end{tabular}
\end{table}

\begin{table}[htbp]\centering\small
\caption{All fixed resolutions on 64 additional disjoint documents at $T=4096$. Each of four width/control cells has six complete initialization replicas. Variances sum vocabulary coordinates and average fixed contexts, with unbiased replica divisor five. Ranges span the four cells and are not confidence intervals. The reference, ordering and 25\% centered-RMS target are fixed before acquisition.}
\label{model:tab:context-transfer}
\begin{tabular}{@{}rrrrrr@{}}
\toprule Tokens & Target cells & Centered RMS & Uncentered RMS & Retained (\%) & Mean KL\\\midrule
128 & 0/4 & 0.494--0.525 & 1.624--1.935 & 74.972--78.489 & 0.133--0.164\\
512 & 0/4 & 0.357--0.390 & 1.403--1.599 & 86.019--88.883 & 0.098--0.110\\
2,048 & 0/4 & 0.256--0.281 & 1.082--1.217 & 92.557--94.146 & 0.061--0.067\\
8,192 & 4/4 & 0.175--0.193 & 0.673--0.759 & 96.353--97.140 & 0.024--0.028\\
16,384 & 4/4 & 0.128--0.142 & 0.455--0.515 & 97.981--98.423 & 0.011--0.013\\
\bottomrule\end{tabular}\end{table}

\input{content/model/generated/context-risk-scales.tex}
\clearpage
{\footnotesize\input{content/model/generated/context-transfer-full.tex}}
{\footnotesize\input{content/model/generated/context-risk-full.tex}}
\clearpage
\subsection{Independent width and time profiles}
\begingroup\small
\input{content/model/generated/critical-refinement-detail-peaks.tex}
\input{content/model/generated/critical-refinement-detail-crossing-predictions.tex}
\input{content/model/generated/exact-replica-widths.tex}
\endgroup

\input{content/model/generated/operator-cache-details.tex}
\input{content/model/generated/cache-risk-details.tex}
\input{content/model/generated/cache-state-details.tex}
\input{content/model/generated/cache-confirmation-details.tex}
\input{content/model/generated/matched-clock-details.tex}
\section{Numerical scope of finite row transport}
\label{model:app:row-quadrature}

The finite transport measurements in Section~\ref{model:sec:collective-flux-results}
and the matrix identities require no parameter differentiation.
An independent finite-arithmetic check evaluates
Lemma~\ref{model:lem:finite-defect} on 1,000 generated matrix transitions
and all 4,000 retained native matrix pairs. The largest identity
discrepancy is $6.23\times10^{-16}$; every native pair satisfies
$\delta<1$, with maximum $0.43328$, and its observed differential
remainder satisfies \eqref{model:eq:row-derivative-bound}.
These inner measurements are conditional on eight states. The bound is
conservative and does not establish a width-uniform small-step regime
or control the generator's parameter-linearization remainder.

Table~\ref{model:tab:row-matrix-errors} gives the complete descriptive
matrix-displacement and derivative-error budget for these same eight
states. These saved-data reductions add no new native forward or
training replicate.
\input{content/model/generated/row-matrix-errors.tex}

We also reconstruct a completed derivative integration diagnostic along
each of the same eight realized parameter displacements. The full native
update is known before constructing this path. The retained incoming
parameter directional derivative estimate $D_0$ and the composite trapezoidal
integrals therefore measure retrospective approximation error, not a
forecast of the next optimizer state. The protocol fixes the absolute
increment target $10^{-5}$ and the nested interval counts
$m\in\{1,2,4,8,16,32\}$.

\input{content/model/generated/row-quadrature-diagnostics.tex}
\input{content/model/generated/row-quadrature-levels.tex}

At 32 intervals, three of eight cases meet the fixed absolute target.
The $N=8,g=1.5$ incoming and integrated estimates have the opposite
sign to the observed increment. At $N=4,g=2$, the finest integral
also reverses sign; its absolute target is met because the actual
increment is extremely small. Its relative increment error is about
$3.36$. Absolute tolerance success in this case does not imply relative
accuracy. Refinement is nonmonotone; all 48 resolution outcomes are
retained, including favorable intermediate levels that do not persist
at the finest grid.

Independent endpoint-energy and adjacent-interval reconstruction agrees
within $2.34\times10^{-15}$. This checks faithful reduction of the recorded
arrays. It does not certify real-arithmetic derivatives or convergence
of their quadrature. The records do not distinguish unresolved integrand
variation, arithmetic effects and other path-dependent numerical errors;
they establish no failure of the real-arithmetic fundamental theorem.
The recorded diagnostic workload is eight optimizer replays and 272
native forwards. It adds no independent training trajectory. The parent
finite transport measurements retain their separate eight scientific
updates and 40 native forwards.

\begin{proposition}[Sufficient derivative-quadrature error budget]
\label{model:prop:row-quadrature-budget}
Let $f\in C^3([0,1])$ be a real row observable along a specified
parameter path, and $M_3=\sup_{s\in[0,1]}|f'''(s)|$.
On a uniform $m$-interval grid, suppose the recorded derivatives obey
$|\widehat d_j-f'(j/m)|\leq\epsilon_d$ and the recorded endpoints
obey $|\widehat f_i-f(i)|\leq\epsilon_e$ for $i=0,1$.
The composite trapezoidal integral $\widehat T_m$ then satisfies
\begin{equation}
 \left|\widehat T_m-(\widehat f_1-\widehat f_0)\right|
 \leq\frac{M_3}{12m^2}+\epsilon_d+2\epsilon_e.
 \label{model:eq:row-quadrature-budget}
\end{equation}
\end{proposition}
\begin{proof}
Write $h=1/m$. On an interval $[a,a+h]$, the linear-interpolation
remainder of the integrand $f'$ is bounded by
$M_3(s-a)(a+h-s)/2$. Integration gives $M_3h^3/12$ on that
interval, hence $M_3/(12m^2)$ after summation. Trapezoidal weights
are nonnegative and sum to one, so their derivative perturbation
is at most $\epsilon_d$. The endpoint perturbations contribute at
most $2\epsilon_e$. The fundamental theorem and the triangle inequality
give the stated bound.
\end{proof}

Applying this proposition to the PLDR path requires a smooth real
generator and nonzero row energy along the entire segment, together
with justified arithmetic enclosures. The saved observations provide
neither a certified $M_3$ nor uniform derivative-error enclosures.
Endpoint replay, a positive sampled power base and nearby quadrature
values do not establish those hypotheses. The supported collective
description is the exact finite transport identity, with the complete
measured approximation error kept separate from successor closure.

\section{Complete chronological row-path observations}
\label{model:app:row-path-details}
Section~\ref{model:sec:row-path-results} defines the entire acquisition grid.
The two initialization identities are paired across conditions. Contexts,
heads, decoders, ages and aligned blocks are inner observations. No
confidence interval or additional independent model count is assigned
to them. All sixteen endpoint budgets and all 64 path/scale cells are
reported below. The block derivative is evaluated on a realized matrix
displacement; its error is distinct from any generator linearization.
\input{content/model/generated/row-path-errors.tex}
\input{content/model/generated/row-path-blocks.tex}
The source protocol, checkpoint identities, exact unused source suffixes,
package versions and acquisition snapshots are retained with the raw
arrays in the experiment workdir. The bounded source package contains
the maintained acquisition, reduction, independent verification and
rendering scripts and their compact records. The commands and external
local dependencies are listed under route \texttt{row} in
\path{docs/RELEASE_INDEX.json}; none of the native matrix payloads is
bundled in the manuscript source.

\subsection{Equal physical-duration grid}
\input{content/model/generated/equal-time-flux-blocks.tex}

\input{content/model/generated/equal-time-temporal.tex}

\input{content/model/generated/projection-complete.tex}
\section{Complete outer-state scale and inference comparisons}
\label{model:app:outer-details}

These tables retain all selected coarse moment and proper-prefix
intervention cells of Section~\ref{model:sec:outer-results}. Statistical units,
conditioning, estimation budgets and arithmetic are those specified
there. They are completed finite observations.

\input{content/model/generated/outer-coarse-scores.tex}
\input{content/model/generated/outer-inference.tex}

\begin{table}[htbp]\centering\small
\caption{All five decoder row ratios before training (upper panel) and
at the incoming 2,048-update state (lower panel). The statistic and its
native arithmetic convention are defined in Section~\ref{model:sec:outer-results}.}
\label{model:tab:outer-rows}
\input{content/model/generated/outer-rows-initial.tex}
\medskip
\input{content/model/generated/outer-rows-incoming.tex}

\end{table}

\section{Complete acquisition and state-change comparisons}
\label{model:app:refresh-details}

The primary table preserves all eight lineages, both incoming ages,
two independent fit cohorts and three nested adaptation budgets.
$N,C,I$ give head count and one-based corpus and initialization labels.
The primary covariance comparisons have common fitted means; their
mean-error contribution to a score difference remains present.
All uncertainty is conditional on the recorded fit and evaluation panel.
Successors retain their parent lineage, and both fit cohorts use the
same independent assessment paths at a given state.

\input{content/model/generated/refresh-primary-complete.tex}

The next table gives every risk and joint risk/entropy resolution,
without comparing score magnitudes across dimensions. The complete
compact result object also retains parent-reuse controls, all
budget-versus-reference and reuse-versus-refit paired differences,
conditional bootstrap intervals, acquisition times and calibrated
linear-image enclosures. These are full prespecified outcomes;
no cell is removed because of its sign or uncertainty.

\input{content/model/generated/refresh-secondary-complete.tex}

\begin{table}[htbp]\centering\small
\caption{Zero-adaptation controls at every state. Both retain the same
regularized archived physical covariance. Mean errors are maximum empirical
risk-path errors in nats. The primary score is an equal-state mean across
eight states, using each state's 32 assessment paths. These controls have
one frozen fit, so they are not counted again for the two local fit cohorts.
Threshold counts use unrounded errors.}
\label{model:tab:refresh-zero-controls}
\begin{tabular}{@{}llccc@{}}
\toprule
State & Mean control & Mean-error range & Mean target & Primary score\\
\midrule
Parent & No change & 0.016--0.128 & 0/8 & $-19.308$\\
Parent & Archived & 0.015--0.159 & 0/8 & $-15.130$\\
Successor & No change & 0.017--0.071 & 0/8 & $-17.954$\\
Successor & Archived & 0.010--0.096 & 0/8 & $-15.809$\\
\bottomrule
\end{tabular}

\end{table}

\section{Complete finite-pulse amplitude outcomes}
\label{model:app:directional-details}
Table~\ref{model:tab:directional-all} reports every discrepancy in
Equation~\eqref{model:eq:native-pulse-domain}. State names encode head count,
corpus and nested initialization identity. The four source indices
are paired across arms within a state; they are not extra pretrained
model identities. Each four-digit gate string uses one for a comparison
that meets both the relative-discrepancy and absolute-signal conditions.
The machine-readable record also retains every signal norm, every
recorded-time response, both radii's even/odd decompositions and all
signed temporal and pulse covariance terms.

\begin{table}[htbp]\centering\footnotesize
\setlength{\tabcolsep}{4pt}
\caption{All 64 primary full-path amplitude comparisons, without
selection by sign, source, family or incoming state.}
\label{model:tab:directional-all}
\input{content/model/generated/directional-primary-complete.tex}
\end{table}

\section{Complete paired native intervention outcomes}
\label{model:app:factorial-corners}
Suffixes A and B are disjoint block sequences at the same four incoming
states, with definitions and source reservations in
Section~\ref{model:sec:potential-factorial-results}. Each table contains every
scientific corner. Long and early states have distinct training objectives
and schedules; none of these rows is an independent pretraining seed.
NLL averages the same fixed 64-target cohort, and $K$ uses the first
of the two fixed probes. The comparison supplies conditional path effects,
without population confidence intervals or fitted critical exponents.
\input{content/model/generated/potential-factorial.tex}
\input{content/model/generated/potential-factorial-effects.tex}
\input{content/model/generated/factorial-disjoint-corners.tex}

\section{Complete matched single-pass outcomes}
\label{model:app:matched-complete}
The following table contains every path of the family in
Section~\ref{model:sec:matched-onepass}. All comparisons share their specified
observation cohort and preserve the initialization/order pairing.
\input{content/model/generated/matched-onepass-complete.tex}

\FloatBarrier

\chapter{Reproducibility, evidence access, and numerical methods}
This appendix describes the companion code and evidence, the supported
levels of reproduction, and the numerical methods used to qualify the
results. Sampling roles, reused observations and a worked reconstruction
make the practical scope of the deposited material explicit.

\label{model:app:matched-reproduction}
\label{app:reproducibility}
\label{row:sec:interfaces}
\label{rg:app:reproducibility}
\label{model:app:reproduction}
\section{Code and compact evidence}
The code companion provides selected Lean developments, native experiment
producers, numerical reducers and scientific validation tools. Its subject
index follows the monograph's chapters. Stable mathematical namespaces are
\path{PldrLlmCurvatureSandpile}, \path{RowRGMap}, \path{ModelRG}, and
\path{PldrTrainingDynamics}. The last contains the energy-comparison,
finite second-moment, sequence-limit, path-error, cancellation, subpower
crossover and scalar face-memory kernels.

The code is available at
\url{https://github.com/burcgokden/PLDR-LLM-Training-Dynamics}.
The reported numerical evidence is available at
\url{https://huggingface.co/datasets/fromthesky/pldr-llm-training-dynamics-data}.
Both companions are self-contained for their stated purposes.

\paragraph{Immutable public references.}
The code and data snapshots below identify the public companions for the
commands in this appendix. The code payload is defined by
\path{scientific-manifest.json}; the data binding is recorded in
\path{release-gate.json} of that code snapshot.

\addtocounter{table}{-1}
\begin{longtable}{@{}p{.24\textwidth}p{.68\textwidth}@{}}
\toprule Identity & Public reference\\\midrule\endhead
Code commit & \path{36ef5e640966759734eeac0e9dacf9801e5311be}.\\[4pt]
Code payload SHA-256 &
\hashpair{22447f2cd550c165c41eaa7c506928fa}{dda51bc3a93e3177751048b25b055927}.\\[4pt]
Code manifest SHA-256 &
\hashpair{5b00345c64b204424feaec1b55850632}{3efecc0285e87067d164fb885808348d}.\\[4pt]
Dataset commit & \path{f03c292a227e58a03d145d478103f587d98a8c8a}.\\[4pt]
Dataset manifest SHA-256 &
\hashpair{225427ff888f9d4ecd4506ae34cc39e6}{7b028aa7f684bd96de3d4398aeb49c90}.\\[4pt]
Dataset payload SHA-256 &
\hashpair{41020c31446a02d27caaa73a3a317c3}{0c231f5884ab7a74d010b80d8c216a947}.\\
\bottomrule
\end{longtable}

These public references identify source and evidence availability. They do
not reassign the software identities of the executed studies or establish
new scientific executions. Historical acquisition and analysis identities
remain attached to their records. The dataset's published-PDF destinations
refer to the arXiv artifact identified in its metadata; the labels and page
numbers printed here identify this monograph.

The data companion stores numerical records and complete reported outcome
grids in a content-addressed compressed store. Its index specifies logical
identities, uncompressed and compressed SHA-256 hashes, and byte counts.
Its coverage index maps monograph displays and claims to numerical evidence
or identifies their mathematical or explanatory role. Display-value records
preserve printed precision where a table is the available numerical source;
figure entries distinguish deposited numerical summaries from unavailable
raw plotting observations. Data dictionaries record conditioning, centering,
denominators, independent units and nested observations.

Portable metadata replaces local acquisition locations with logical asset
identities and omits machine identifiers and acquisition dates. Exported records have their own hashes; these are distinct from
hashes of the original acquisitions. Numerical values, sample counts,
seeds, source positions, signs, uncertainties and outcomes are preserved.
Null controls, rejected candidates, failed gates and context exceptions remain
part of the reported evidence. Code for reading, extracting and checking the
data is supplied by the GitHub companion.

The compact deposit does not contain large raw observation arrays, model or
optimizer checkpoints, corpus shards, or token arrays. Raw-array reanalysis
and native replay require the separately identified inputs. The deposit
supports examination of the reported quantities; integrity checking alone
is not an independent scientific replication.

\section{Reproduction levels}
The following levels distinguish checks of reported evidence and mathematical
kernels from reanalysis and native replay, which require separately supplied
inputs.

\begingroup
\small
\setlength{\tabcolsep}{3pt}
\begin{longtable}{@{}>{\raggedright\arraybackslash}p{.20\textwidth}>{\raggedright\arraybackslash}p{.36\textwidth}>{\raggedright\arraybackslash}p{.38\textwidth}@{}}
\toprule Level & Inputs and activity & Meaning and limit of success\\\midrule\endhead
Integrity & Pinned code manifest; regular dataset files, index and
compressed/decompressed records. & Authenticates supplied bytes and
deposited quantities; does not reconstruct absent raw observations.\\[6pt]
Formal and finite checks & Pinned Lean/Mathlib, owned modules, mapped
types, axiom checks and CPU scientific tests. & Checks stated formal
clauses and exercised finite identities; does not certify all written
proofs or empirical assumptions.\\[6pt]
Bounded interfaces & Declared dispatch fixtures, argument handling,
selected negative controls and resource/accounting checks. & Establishes
only the exercised boundaries; does not establish complete native
qualification or repair incomplete admission.\\[6pt]
Raw reanalysis & Complete raw arrays, protocols and hash-bound external
input graph; reducer and independent verifier. & Reconstructs retained
results from those inputs; compact summaries alone are insufficient.\\[6pt]
Specified-state replay & Exact parameters, optimizer and RNG state,
source position, clocks, arithmetic policy, native code and platform. &
Compares the declared finite states/observations; does not imply replay
under another incoming state, source law or platform.\\[6pt]
New acquisition & Frozen design, complete native assets and qualified
producer-specific execution, reduction and verification. & A separate
activity requiring its own evidence; the retained support inventory
determines the scope of each route.\\
\bottomrule
\end{longtable}
\endgroup

\paragraph{Prerequisites and working directory.}
Run the commands below at the root of the code companion, using the
immutable code commit stated above. The documented scientific-check stack
is Python 3.14.6 with PyTorch 2.12.1 and the pinned packages in
\path{requirements-checks.txt}, including its referenced
\path{requirements-publication.txt}. In a fresh Python 3.14.6 environment:
\begin{verbatim}
python3 -m pip install torch==2.12.1+cpu \
  --index-url https://download.pytorch.org/whl/cpu
python3 -m pip install -r requirements-checks.txt
\end{verbatim}
The standard-library evidence reader separately requires Python 3.11 or
later. A documented test environment does not establish arbitrary-version
or native GPU compatibility. Native acquisition additionally requires
producer-specific dependencies and supplied model and corpus assets.
The Lean toolchain is \path{leanprover/lean4:v4.33.0-rc1}; Mathlib is pinned to
\hashpair{9c0c555bde5a8277cd36}{dc4dc6dfe2a5a77a2b11}.

\paragraph{Integrity and code/data correspondence.}
Replace \path{DATA-REPO} and other uppercase placeholders with caller paths.
The dataset must contain actual regular evidence files, not Git LFS
pointers or a Hub-cache layout. Keep extraction and validation outputs
outside its root.
\begin{verbatim}
python3 scripts/verify_scientific_manifest.py
python3 scripts/check_formal_manifest.py
python3 scripts/verify_evidence.py --data-repo DATA-REPO
python3 scripts/read_evidence.py --data-repo DATA-REPO --list
python3 scripts/verify_evidence.py --data-repo DATA-REPO \
  --extract /tmp/pldr-monograph-evidence
\end{verbatim}
The optional \verb|--data-repo DATA-REPO| on
\path{scripts/check_formal_manifest.py} checks the code/data mapping when the
complete dataset is supplied. The logical families \path{row}, \path{rg}
and \path{model} name scientific subjects, not additional replication units.
The dataset's \path{coverage.json}, \path{data-dictionary.json} and
\path{raw-inputs.json} state the interpretation and external-input boundary.

\paragraph{Finite scientific checks.}
\begin{verbatim}
python3 scripts/run_scientific_checks.py
\end{verbatim}
The CPU suites exercise finite work, face completion, chronological
composition and normalized-flux interfaces. They qualify the exercised
calculations, not native training laws. Missing raw-fixture skips and
explicit publication-fixture exclusions remain separate from passes.

\paragraph{Bounded process and resource checks.}
\begin{verbatim}
python3 scripts/check_process_boundaries.py
python3 scripts/check_campaign_budget.py
python3 scripts/check_resource_execution.py
\end{verbatim}
These commands check their declared fixtures and boundaries. Successful
dispatch or accounting does not establish a complete native acquisition.

\paragraph{Pinned formal checks.}
\begin{verbatim}
lake exe cache get
lake build
python3 scripts/check_lean.py
\end{verbatim}
\path{scripts/check_lean.py} rebuilds every owned module, exports referenced
declaration types, and checks transitive axiom use, including generated
and unused declarations. Its optional \verb|--dependency-cache PATH| reads
matching compiled third-party dependencies while rebuilding owned modules.
The statement manifest resolves selected formal clauses without reading
manuscript sources; Appendix~\ref{app:formal} states the axiom boundary.
New outputs stay in the chosen checkout. These command groups describe
activities, not additional scientific execution records.

\paragraph{Raw reanalysis and native replay.}
Raw reanalysis needs the complete arrays, protocols and hash-bound external
input graph. Native replay additionally needs exact model and optimizer
checkpoints, RNG state, token arrays, corpus assets, source cursor, clocks
and the declared numerical law and platform. The compact deposit does not
supply these large inputs. The pinned
\href{https://github.com/burcgokden/PLDR-LLM-Training-Dynamics/blob/36ef5e640966759734eeac0e9dacf9801e5311be/docs/EXECUTION.md}{execution guide}
describes the thirty-cell cache reconstruction and its independent relocation
check. \path{scripts/run_source.py} supplies family imports and explicit
input roots. GPU replay retains the finite scope stated with each experiment.

\paragraph{Support and accounting boundaries.}
The public support inventory distinguishes 25 interfaces with their
stated bounded validation levels from 950 archival source utilities;
four documentation entries form a separate category. Bounded dispatch
does not establish full acquisition. Durable cumulative campaign
accounting applies only to the block-normal, source-resolved and
orbitwise shared interfaces. Those campaign launchers remain archival,
with bounded accounting validation. Generic process supervision does
not provide this contract to every launcher, and historical acquisition
records are not upgraded by a current check. Sampled resource peaks
are observations rather than continuous hard ceilings.

At the stated code commit, the
\href{https://github.com/burcgokden/PLDR-LLM-Training-Dynamics/blob/36ef5e640966759734eeac0e9dacf9801e5311be/provenance/supported-programs.json}{support inventory},
\href{https://github.com/burcgokden/PLDR-LLM-Training-Dynamics/blob/36ef5e640966759734eeac0e9dacf9801e5311be/provenance/campaign-accounting.json}{campaign-accounting inventory}
and
\href{https://github.com/burcgokden/PLDR-LLM-Training-Dynamics/blob/36ef5e640966759734eeac0e9dacf9801e5311be/docs/RESOURCE_EXECUTION.md}{resource-execution guide}
specify the corresponding boundaries. The historical row inventory contains
fourteen launchers; thirteen are shipped. The excluded publication launcher
is not a missing supported acquisition route.

\section{Single-pass and retained-source boundaries}
The primary corpus law consumes distinct registered target blocks in
their declared ordering. Evaluation panels and calibration panels have
explicit roles and are not silently reused as independent confirmation.
Further confirmation can use the retained two-GPU
single-pass producers with a fresh, disjoint source interval and a
finite resource budget. The present monograph reports the already
executed studies and performs no new model training.

The retained row-mechanism and long-run closure records answer finite
geometric and state-sufficiency questions under their recorded data
protocols. They are not pooled with the single-pass families to
estimate a common critical law. Deliberately repeated-corpus comparisons
in the model-wide studies are auxiliary changes of the source law.
The reported negative and insufficient outcomes constrain which
conditional theory may be applied.

\section{Numerical differentiation and experimental scope}
The predictive-response calculations compare central finite differences at
steps 0.01 and 0.005 and use automatic differentiation as an independent
local check. Table~\ref{model:tab:derivative-evidence} reports relative
predictive-Fisher discrepancies for the two released models. These checks
assess derivative approximation on the measured contexts; they do not
establish a uniform Taylor remainder or an asymptotic training law.

\begin{table}[htbp]\centering\small
\caption{Numerical derivative checks. Median, q90 and maximum compare the
central-difference steps across the retained contexts and directions.
The AD maximum compares automatic differentiation with finite differences
on four contexts and two directions per context. All entries are relative
discrepancies in the predictive-Fisher norm.}
\label{model:tab:derivative-evidence}
\begin{tabular}{rrrrr}
\toprule
Model & Median & q90 & Maximum & AD maximum\\
\midrule
1 & $9.83\times10^{-4}$ & $1.78\times10^{-3}$ & $4.43\times10^{-2}$ & $2.25\times10^{-3}$\\
2 & $1.01\times10^{-3}$ & $1.74\times10^{-3}$ & $8.98\times10^{-3}$ & $1.35\times10^{-3}$\\
\bottomrule
\end{tabular}

\end{table}

The experiment inventory distinguishes training, conditional continuations,
paired interventions and observation-only panels. Counts of nested contexts,
heads and evaluation times do not add independent training replications.
The common study registry gives a classified native subtotal of 4,253,160 updates and 4,254,024 including 864 arithmetic-control updates. The scope is the named RefinedWeb-source inventory under its retained classification. Physical-source adaptation and separate released-base adaptation have different source and parameter-law scopes and are not added to this subtotal. The designs and their sampling scopes are listed in Table~\ref{model:tab:execution-ledger}. The 384 central single-pass trajectories and 1,228,800 updates are a subset of this inventory.

\input{content/model/generated/execution-ledger.tex}

\section{Sampling roles and reuse across studies}
All row ranges are half-open in the declared source resource.
\begin{longtable}{p{.27\textwidth}p{.23\textwidth}p{.40\textwidth}}
\caption{Exact sampling roles and cross-experiment reuse.}\label{model:tab:rows}\\
\toprule Cohort and use & Rows & Calls and overlap\\\midrule\endfirsthead
\toprule Cohort and use & Rows & Calls and overlap\\\midrule\endhead
Single-pass training: 524,288 documents & Prefix of one permutation of
4,194,304 blocks & Eight disjoint 64-target source blocks per 513-token
crop; 32 blocks per update. Complete training-document hashes are disjoint
from all three retained text-observation cohorts.\\
Source-selection adaptation and return & Twelve controlled parents at
40,960 updates; seven source mixtures and a general arm & All adaptation
paths exclude consumed supervised blocks. The common return additionally
excludes every parent block. Paired counterfactual arms share their
permutations and uniforms.\\
Fine-tuning text observations & 480 wholly unconsumed materialized
documents, 160 per lexical stratum & Each stratum has 80 calibration and
80 held-out documents. Intermediate times use the first sixteen of each
80-document half; the same subset is used for endpoint time comparisons.
Every block of every evaluation document is excluded from adaptation.\\
Released-model adaptation reference & One released fourteen-head state;
four source paths & The same streams and observation documents, with
cached global positions outside the reported pretraining interval.
A fresh optimizer origin is explicit. This model is not a width or
initialization replicate of the controlled family.\\
Corpus-transfer training: two 65,536-document draws & Eight block indices
per crop; first 65,536 of 524,288 shuffled indices per primary path &
Two initialization identities per corpus paired across two widths;
56 continuations per state draw distinct blocks from the 458,752
remaining indices, with possible counterfactual overlaps.\\
State-change continuations & All eight corpus-transfer parents at
2,048 updates and their 64-update successors & Two disjoint fitting
cohorts per state, each with 16 adaptation and 8 calibration paths;
32 fresh assessment paths shared by those fits. All successor paths
exclude both the original prefix and the state-advance blocks.\\
Native finite-pulse continuations & The eight corpus-transfer parents at
2,048 updates; four fresh source sequences per state & Nine paired arms
per source, each consuming 4,096 distinct remaining blocks; all 35
observations use the same evaluation panel and pulse directions use
the same separate donor panel.\\
Small-radius response paths & Two corpus-zero, identity-zero parents;
four fresh sequences per width & Nine arms and two numerical laws share
256 distinct remaining blocks per sequence. Five times and twelve
coordinates have equal observation weight; all use the fixed evaluation
and donor panels.\\
Source-support intervention & The same two parents; two further source
sequences per width & Baseline and input-row pulse share 256 distinct
remaining blocks per sequence. Every update is observed on the fixed
evaluation panel; the full corpus input support is checked separately.\\
Corpus-transfer text observations & Separate 32-document evaluation and
32-document donor panels & Selected before either new corpus; both panels
are disjoint from both new corpora and the 524,288-document corpus.
All new states and horizons reuse the same evaluation panel.\\
Single-pass seen-training risk & 1,024 already consumed blocks per saved state &
Sampled without replacement from the consumed prefix. Membership changes
with the checkpoint; this is a conditional fit diagnostic.\\
Single-pass prediction and metric observations & $[512,1024)$ of the short cohort &
The identical held-out contexts under every training law. Whole training
initializations supply replication; contexts and checkpoints remain paired.\\
Complete proper-prefix risk and decoder interventions & $[512,544)$ of the short cohort &
All 64 positions per context, with complete vocabulary logits retained.
The 32 contexts are a subset of the common held-out prediction cohort.\\
Conditional risk-path calibration and validation & $[512,544)$ of the short cohort &
Four future horizons share the same 32 external-target contexts.
Each reset branch uses 2,048 distinct remaining training blocks in
64 batch-32 updates; branches may overlap counterfactually. The
fresh memory study uses the same contexts at eight recorded horizons
and retains 32 further paths per incoming state.\\
Single-pass row input-domain and Jacobians & $[512,520)$ of the short cohort &
Lengths 16, 32, 48 and 64 at four saved states. Eighteen fixed local
points per decoder and length; these are input derivatives.\\
4,608 documents: field map & $[0,512)$ calibration; $[512,4608)$ field & Length 128 at offset zero; field and eight-segment analyses reuse these documents.\\
Retained response samples & $[512,544)$ and $[768,1024)$ & Length 128; each sample split into equal calibration and evaluation halves; both overlap the field cohort.\\
Retained operator replacement & anchor 0; $[512,768)$ & Length 128; these evaluation documents overlap the field cohort.\\
Direct operator matrices & anchor 0; $[4096,4160)$ & Length 128; separate 64-context matrix probe.\\
Initialization training & $[0,3072)$ & 64-token inputs, uniform offsets $0,\ldots,448$; documents sampled with replacement.\\
Initialization quality and response & $[3584,4096)$; $[4096,4112)$ & Length 64 at offset zero; response probes reused by precision controls.\\
Fresh short cohort: 2,048 documents & $[0,512)$ calibration; $[512,2048)$ evaluation & Uniform 513-token crop, then one saved uniform offset $0,\ldots,448$; matches the controlled training crop rule.\\
Short-cohort trajectory, branch and drift probes & $[0,64)$; $[512,768)$; $[768,784)$ & Repeated states reuse these fixed document sets. Branch and drift probes are subsets of the evaluation cohort.\\
Conditional head scans & $[512,1024)$ of the fresh short cohort &
Four training seeds per condition, common shared-generator initialization
and complete batch history; 256 disjoint context pairs.\\
Initial-law and source/pulse probes & $[512,528)$; $[512,576)$ of the fresh short cohort &
Sixteen contexts for initial laws; 64 for gain secants, local attention
Jacobians and paired training pulses. These subsets are reused.\\
Selected precision and replay controls & Subsets of $[512,576)$ of the fresh short cohort &
Thirty selected context--state pairs in four source controls; four
existing pulse parents with two unperturbed numerical branches each.\\
Width--time replication and horizon doubling & $[512,1024)$ of the fresh short cohort &
Sixteen seeds at $N=2,4,8,14$ and four at $N=24$ for $g=1$; the rate grid reuses four identities; doubled horizons use sixteen
at $N=14$ and four at the other widths.\\
Dense width--time paths & $[512,528)$ of the fresh short cohort &
Every 32 updates in the initial width--time study and every 64 in the
joint size--time continuations; contexts and times are paired observations.\\
Shared crossing, component return and row projections & $[512,576)$ of the fresh short cohort &
The same 64 contexts under every component or row intervention; no added training identities.\\
Augmented tangent and metric-row transport & $[512,516)$ of the fresh short cohort &
Four contexts; amplitude and arithmetic controls reuse the same checkpoint states.\\
Metric-unit loss adjoints & Two 32-example minibatches from the training cohort &
Sampler indices 2,048 and 8,192, crossed with both checkpoint horizons
at every width and seed; the same token crops are held fixed.\\
Joint size--time and extended paths & $[512,1024)$ evaluation;
$[0,64)$ calibration matrices & Four matched identities through 32,768
at all five widths; eight through 16,384 at $N=2,4,8,24$; four long
identities at each of $N=4,14$. Selected states retain evaluation
matrices and paired arithmetic.\\
Conditional source laws & $[512,528)$ for the full-vocabulary tangent &
32 fresh training batches per incoming state; eight initially
transported directions, followed by 32-direction calibration and
separate 16- and 32-batch validation laws. The reference predictor is
paired on the latter law.\\
Complete native source corners & $[512,528)$ for all four emissions &
32 fresh batches at each of sixteen incoming states. Every draw
restores model, moments and counters before the coupled update.\\
Fresh long cohort: 1,024 documents & $[0,256)$ projection; $[256,512)$ fit; $[512,1024)$ evaluation & One uniform 4,097-token crop; 64 adjacent length-64 calls at offsets $64i$, target $64(i+1)$.\\
\bottomrule
\end{longtable}

\section{Numerical reanalysis}
A CPU reconstruction of the disjoint-context cache panel uses the retained
raw observations, paired input identities and content hashes. All 30
scientific cell dictionaries agree exactly with the retained analysis;
the independent raw-moment and pair-distance checks also pass. The 3,024
forward calls belong to the acquisition of this panel. Recomputing its
summaries requires no additional model forward or training update.
The compact deposit contains the reconstruction and verification records.
Raw arrays and model checkpoints remain separately declared inputs.

\section{A worked bounded reconstruction}
\label{sec:worked-reproduction}
The route reconstructs the disjoint-context cache confirmation under protocol
\path{rg-cache-state-transfer-v1} and implementation admission
\path{cache-state-contract-v2}. Supply its raw study and complete external
input graph separately. A \path{pldr-acquisition-locations-v1} manifest binds
every immutable acquisition identity to an actual regular file and expected
SHA-256. Missing identities, altered bytes and incompatible protocols are
errors. The compact evidence deposit supplies reported results, not these
raw arrays or native checkpoints.

From the combined code checkout, the following is one shell command; replace
the three argument names by admitted locations:
\begin{quote}\small\ttfamily
sh \path{vendor/model/scripts/workspace-wrappers/analyze-cache-state-transfer.sh}\par
\hspace*{1em}RAW-STUDY FRESH-OUTPUT INPUT-LOCATIONS.json
\end{quote}
It writes \path{analysis.json} and \path{verification.json} in the fresh output
directory. The expected result is exact equality of all thirty retained
scientific cell dictionaries and a passing independent verifier. No native
forward or training update is performed. The guarded route
\path{scripts/check_cache_relocation.py}, with \path{--source-study} and
\path{--workspace}, assembles the locations manifest, checks original and
relocated hashes before and after reduction, and blocks reads from original
locations. Allow about 3 GB of temporary input space and at most fifteen
minutes per bounded CPU subprocess. Current details are in
\path{docs/EXECUTION.md}.

\FloatBarrier

\chapter{Correspondence with selected formal statements}
This appendix records the correspondence between selected statements in
the monograph and their formal implementations. It explains the scope of
the checks across row geometry, chronological renormalization and model-wide
laws; the mathematical arguments remain stated independently in the text.

\label{app:formal}
\label{rg:app:formal-correspondence}
\label{model:app:formal}
\section{Scope of the independent checks}
The proofs in this monograph are ordinary mathematical arguments.
No result takes a Lean implementation as a premise. The tables below
map selected statements to independent formal checks and identify
which analytical, probabilistic, or empirical clauses remain outside
those checks. A verified scalar identity does not by itself establish
the corresponding matrix, limiting, or physical interpretation.

The common toolchain is Lean 4.33.0-rc1 with Mathlib pinned to
\hashpair{9c0c555bde5a8277cd36}{dc4dc6dfe2a5a77a2b11}.
The combined axiom check admits only \path{propext},
\path{Classical.choice}, and \path{Quot.sound} as transitive axiom
dependencies. Compilation, the axiom boundary, and semantic
correspondence to a written theorem are separate checks. The first
two are mechanical; the stated coverage and the hypotheses in the
written proof delimit the third.

\section{Row geometry and optimizer dynamics}
Paths in this table are relative to \path{vendor/row}.
The qualifiers \emph{faithful}, \emph{partial}, and
\emph{not formalized} describe the named kernel, not an experimental claim.
\newcommand{\RowSourceManuscriptLeanDeclarationCount}{125}
\newcommand{\RowSourceManuscriptLeanProofDeclarationCount}{88}
\newcommand{\RowSourceManuscriptLeanDefinitionCount}{37}
\newcommand{\RowSourceManuscriptLeanActiveProofDeclarationCount}{417}
\newcommand{\RowSourceManuscriptLeanActiveDefinitionCount}{145}
\newcommand{\RowSourceManuscriptLeanAPIAccessorCount}{2}
\newcommand{\RowSourceManuscriptLeanUnmappedProofCount}{329}
\newcommand{\RowSourceManuscriptLeanArchivedProofCount}{765}
\newcommand{\RowSourceManuscriptLeanCorrespondenceRowCount}{40}

Each result retains its module, status and declarations. The complete
\path{provenance/statement-manifest.json} at the code commit in
Appendix~\ref{app:reproducibility} resolves these names to the shipped modules.
\begingroup
\small
\setlength{\tabcolsep}{3pt}
\begin{longtable}{@{}>{\raggedright\arraybackslash}p{.34\textwidth}>{\raggedright\arraybackslash}p{.30\textwidth}>{\raggedright\arraybackslash}p{.30\textwidth}@{}}
\caption{Exhaustive correspondence between stand-alone written results and selected Lean kernels.}\label{row:tab:formal-correspondence}\\
\toprule Written result and formal locator & Checked scope & Remaining obligations\\\midrule\endfirsthead
\toprule Written result and formal locator & Checked scope & Remaining obligations\\\midrule\endhead
Result~\ref{row:lem:row-quotient-geometry}: Physical quotient and row diameter\par {\footnotesize\ttfamily Vector\allowbreak{}Centered\allowbreak{}Row\allowbreak{}Geometry}\par {\textsc{partial}}\par {\footnotesize\ttfamily vector\allowbreak{}Centered\allowbreak{}Energy}, {\footnotesize\ttfamily vector\allowbreak{}Pair\allowbreak{}Distance\allowbreak{}Sq}, {\footnotesize\ttfamily vector\allowbreak{}Ordered\allowbreak{}Pair\allowbreak{}Energy}, {\footnotesize\ttfamily vector\_\allowbreak{}pairwise\_\allowbreak{}variance\_\allowbreak{}identity}, {\footnotesize\ttfamily vector\_\allowbreak{}pair\_\allowbreak{}distance\_\allowbreak{}le\_\allowbreak{}twice\_\allowbreak{}centered\_\allowbreak{}energy}, {\footnotesize\ttfamily vector\_\allowbreak{}pair\_\allowbreak{}collapse\_\allowbreak{}of\_\allowbreak{}centered\_\allowbreak{}energy} & Finite-dimensional centered energy, the vector pairwise-variance identity, the direct pair-distance bound, and the centered-energy collapse implication. & Attainment of the finite maximum, the reverse diameter-to-energy bound, and the three-way sequence equivalence.\\[6pt]
Result~\ref{row:thm:layer-resolved-gate-shape}: Layer-resolved mixed gate-shape collapse\par {\footnotesize\ttfamily Gate\allowbreak{}Shape\allowbreak{}Factorization}\par {\textsc{partial}}\par {\footnotesize\ttfamily coordinate\allowbreak{}Energy}, {\footnotesize\ttfamily gate\allowbreak{}Shape\allowbreak{}Energy}, {\footnotesize\ttfamily coordinate\_\allowbreak{}energy\_\allowbreak{}nonnegative}, {\footnotesize\ttfamily gate\_\allowbreak{}shape\_\allowbreak{}energy\_\allowbreak{}zero\_\allowbreak{}iff}, {\footnotesize\ttfamily coordinate\_\allowbreak{}energy\_\allowbreak{}collapse\_\allowbreak{}iff} & The finite nonnegative coordinate factorization and equivalence between total-energy convergence and convergence of every mixed coordinate energy. & Matrix row centering, Frobenius column assembly, and transfer to maximum physical row diameter.\\[6pt]
Result~\ref{row:thm:observer-stratum-partition}: Exact observer-stratum partition\par {\footnotesize\ttfamily Observer\allowbreak{}Resolved\allowbreak{}Energy}\par {\textsc{partial}}\par {\footnotesize\ttfamily Exact\allowbreak{}Face}, {\footnotesize\ttfamily Floor\allowbreak{}Censored\allowbreak{}Positive}, {\footnotesize\ttfamily Resolved\allowbreak{}Positive}, {\footnotesize\ttfamily observer\_\allowbreak{}stratum\_\allowbreak{}partition}, {\footnotesize\ttfamily observer\_\allowbreak{}strata\_\allowbreak{}disjoint}, {\footnotesize\ttfamily upper\_\allowbreak{}zero\_\allowbreak{}certifies\_\allowbreak{}exact\_\allowbreak{}face}, {\footnotesize\ttfamily positive\_\allowbreak{}lower\_\allowbreak{}certifies\_\allowbreak{}strict\_\allowbreak{}positivity}, {\footnotesize\ttfamily enclosure\_\allowbreak{}certifies\_\allowbreak{}floor\_\allowbreak{}censored}, {\footnotesize\ttfamily lower\_\allowbreak{}above\_\allowbreak{}floor\_\allowbreak{}certifies\_\allowbreak{}resolved}, {\footnotesize\ttfamily energy\_\allowbreak{}eq\_\allowbreak{}zero\_\allowbreak{}iff\_\allowbreak{}state\_\allowbreak{}zero} & The exhaustive and pairwise-disjoint trichotomy for nonnegative stored energy at a positive observer floor, exact energy zero if and only if every finite state coordinate is zero, and the exact enclosure implications for face, strict-positive, censored, and resolved certification. & Identification of the finite coordinate state with a centered Frobenius tensor, construction of numerical enclosures, and the protocol-specific choice of observer floor.\\[6pt]
Result~\ref{row:thm:block-orbit-criterion}: Necessary and sufficient block orbit criterion\par {\footnotesize\ttfamily Block\allowbreak{}Excursion}\par {\textsc{partial}}\par {\footnotesize\ttfamily block\allowbreak{}Offsets}, {\footnotesize\ttfamily block\allowbreak{}Excursion}, {\footnotesize\ttfamily block\allowbreak{}Maximum}, {\footnotesize\ttfamily energy\_\allowbreak{}le\_\allowbreak{}anchor\_\allowbreak{}add\_\allowbreak{}excursion}, {\footnotesize\ttfamily block\_\allowbreak{}excursion\_\allowbreak{}eq\_\allowbreak{}maximum}, {\footnotesize\ttfamily block\_\allowbreak{}excursion\_\allowbreak{}collapse\_\allowbreak{}iff}, {\footnotesize\ttfamily block\_\allowbreak{}excursion\_\allowbreak{}le\_\allowbreak{}positive\_\allowbreak{}variation} & Attained finite block maxima, exact anchor-plus-excursion identity, the nonnegative block collapse equivalence, and domination by positive variation. & Passage from successive unbounded blocks to the complete natural-number orbit and the finite-registry maximum.\\[6pt]
Result~\ref{row:thm:exact-direct-work-charge}: Exact physical work-charge identity\par {\footnotesize\ttfamily Direct\allowbreak{}Observable\allowbreak{}Energy}\par {\textsc{partial}}\par {\footnotesize\ttfamily energy}, {\footnotesize\ttfamily pairing}, {\footnotesize\ttfamily exact\_\allowbreak{}work\_\allowbreak{}charge}, {\footnotesize\ttfamily exact\_\allowbreak{}three\_\allowbreak{}component\_\allowbreak{}energy\_\allowbreak{}ledger}, {\footnotesize\ttfamily exact\_\allowbreak{}four\_\allowbreak{}component\_\allowbreak{}energy\_\allowbreak{}ledger} & The finite squared-energy work-charge identity and complete three- and four-component cross-Gram expansions after flattening the observable tensor. & The arbitrary finite source-index expansion, implemented tensor capture, and floating-point residual policy.\\[6pt]
Result~\ref{row:thm:observer-safe-work-charge}: Branch-free work-charge classification\par {\footnotesize\ttfamily Observer\allowbreak{}Resolved\allowbreak{}Energy}\par {\textsc{faithful}}\par {\footnotesize\ttfamily work}, {\footnotesize\ttfamily charge}, {\footnotesize\ttfamily observer\_\allowbreak{}safe\_\allowbreak{}work\_\allowbreak{}charge}, {\footnotesize\ttfamily strict\_\allowbreak{}contraction\_\allowbreak{}iff\_\allowbreak{}work\_\allowbreak{}gt\_\allowbreak{}charge}, {\footnotesize\ttfamily preservation\_\allowbreak{}iff\_\allowbreak{}work\_\allowbreak{}eq\_\allowbreak{}charge}, {\footnotesize\ttfamily strict\_\allowbreak{}reopening\_\allowbreak{}iff\_\allowbreak{}work\_\allowbreak{}lt\_\allowbreak{}charge}, {\footnotesize\ttfamily exact\_\allowbreak{}face\_\allowbreak{}work\_\allowbreak{}charge} & The finite-vector branch-free energy balance, exact strict contraction, preservation, and reopening comparisons, and pure-charge specialization at exact energy zero. & Only the notational identification of a finite coordinate vector with the stacked Frobenius registry.\\[6pt]
Result~\ref{row:thm:interval-certified-work-charge}: Interval-certified work-charge decision\par {\footnotesize\ttfamily Observer\allowbreak{}Resolved\allowbreak{}Energy}\par {\textsc{partial}}\par {\footnotesize\ttfamily interval\_\allowbreak{}certifies\_\allowbreak{}contraction}, {\footnotesize\ttfamily interval\_\allowbreak{}certifies\_\allowbreak{}reopening} & The two strict interval-separation implications that certify the true ordering of work and charge. & The endpoint perturbation bounds from Cauchy--Schwarz, the energy-enclosure classification, and the statement that overlapping intervals alone leave the sign unresolved.\\[6pt]
Result~\ref{row:prop:exact-gate-shape-secant}: Exact gate-shape secant\par {\footnotesize\ttfamily Direct\allowbreak{}Observable\allowbreak{}Energy}\par {\textsc{partial}}\par {\footnotesize\ttfamily shape\allowbreak{}Secant}, {\footnotesize\ttfamily gate\allowbreak{}Secant}, {\footnotesize\ttfamily interaction\allowbreak{}Secant}, {\footnotesize\ttfamily exact\_\allowbreak{}gate\_\allowbreak{}shape\_\allowbreak{}secant}, {\footnotesize\ttfamily implementation\allowbreak{}Defect\allowbreak{}Secant}, {\footnotesize\ttfamily exact\_\allowbreak{}precision\_\allowbreak{}resolved\_\allowbreak{}gate\_\allowbreak{}shape\_\allowbreak{}secant} & The exact scalar two-factor secant and the native scalar secant with old-gate shape, old-shape gate, bilinear interaction, and endpoint implementation-defect terms. & Matrix diagonal assembly and the source-component energy Gram expansion, which is checked separately.\\[6pt]
Result~\ref{row:prop:precision-resolved-collapse}: Precision-resolved collapse criterion\par \emph{None}\par {\textsc{not formalized}}\par \emph{None} & No formal coverage is claimed. & The norm convergence equivalence, reverse triangle bound, and cancellation counterexample for native and factorized endpoint sequences.\\[6pt]
Result~\ref{row:thm:integrated-parameter-secant}: Integrated implemented-parameter secant\par \emph{None}\par {\textsc{not formalized}}\par \emph{None} & No formal coverage is claimed. & Piecewise Fréchet differentiability, the fundamental theorem on a partitioned segment, and the implemented AdamW parameter-source partition.\\[6pt]
Result~\ref{row:thm:exact-paired-energy-contrast}: Exact paired-contrast mechanism\par {\footnotesize\ttfamily Direct\allowbreak{}Observable\allowbreak{}Energy}\par {\textsc{partial}}\par {\footnotesize\ttfamily exact\_\allowbreak{}paired\_\allowbreak{}energy\_\allowbreak{}contrast} & The complete finite-coordinate energy contrast between natural and controlled increments from one source state, including state projection, natural-increment interaction, and paired charge. & Implemented intervention construction, floating-point source decomposition, and empirical sign classification.\\[6pt]
Result~\ref{row:thm:exact-radial-tangential-gain}: Exact radial and tangential gain and closing criterion\par {\footnotesize\ttfamily Radial\allowbreak{}Tangential\allowbreak{}Energy}\par {\textsc{partial}}\par {\footnotesize\ttfamily radial\allowbreak{}Coefficient}, {\footnotesize\ttfamily tangent\allowbreak{}Residual}, {\footnotesize\ttfamily tangent\allowbreak{}Charge}, {\footnotesize\ttfamily gain\allowbreak{}Squared}, {\footnotesize\ttfamily tangent\_\allowbreak{}residual\_\allowbreak{}orthogonal}, {\footnotesize\ttfamily radial\_\allowbreak{}tangent\_\allowbreak{}split}, {\footnotesize\ttfamily exact\_\allowbreak{}radial\_\allowbreak{}tangential\_\allowbreak{}energy}, {\footnotesize\ttfamily exact\_\allowbreak{}radial\_\allowbreak{}tangential\_\allowbreak{}gain}, {\footnotesize\ttfamily scalar\_\allowbreak{}closing\_\allowbreak{}criterion}, {\footnotesize\ttfamily strict\_\allowbreak{}energy\_\allowbreak{}decrease\_\allowbreak{}iff} & The finite-coordinate orthogonal residual, exact unnormalized and normalized gain, and scalar strict-closing equivalence. & Identification with the stacked Frobenius registry and the empirical failure-class terminology.\\[6pt]
Result~\ref{row:prop:source-resolved-radial-ledger}: Complete source-resolved radial ledger\par {\footnotesize\ttfamily Radial\allowbreak{}Tangential\allowbreak{}Energy}\par {\textsc{partial}}\par {\footnotesize\ttfamily radial\_\allowbreak{}coefficient\_\allowbreak{}sum}, {\footnotesize\ttfamily tangent\_\allowbreak{}residual\_\allowbreak{}sum}, {\footnotesize\ttfamily exact\_\allowbreak{}four\_\allowbreak{}tangent\_\allowbreak{}gram} & Finite-source radial-coefficient and tangent-residual recomposition, together with the complete four-source tangent Gram identity. & The arbitrary source-index Gram assembly and identification of the four coordinates with implemented named secants.\\[6pt]
Result~\ref{row:thm:radial-tangential-paired-contrast}: Exact radial and tangential paired contrast\par {\footnotesize\ttfamily Radial\allowbreak{}Tangential\allowbreak{}Energy}\par {\textsc{faithful}}\par {\footnotesize\ttfamily increment\_\allowbreak{}energy\_\allowbreak{}radial\_\allowbreak{}tangent}, {\footnotesize\ttfamily pairing\_\allowbreak{}radial\_\allowbreak{}tangent\_\allowbreak{}increments}, {\footnotesize\ttfamily exact\_\allowbreak{}paired\_\allowbreak{}radial\_\allowbreak{}tangential\_\allowbreak{}vector}, {\footnotesize\ttfamily paired\_\allowbreak{}radial\_\allowbreak{}tangential\_\allowbreak{}normalize} & The full finite-vector radial and tangential paired contrast derived from the source, natural increment, and paired increment, followed by its exact normalization. & Identification with the implemented intervention tensors and empirical sign classification.\\[6pt]
Result~\ref{row:cor:exact-positive-excursion-product}: Exact positive-excursion product\par {\footnotesize\ttfamily Radial\allowbreak{}Tangential\allowbreak{}Energy}\par {\textsc{partial}}\par {\footnotesize\ttfamily exact\_\allowbreak{}positive\_\allowbreak{}excursion\_\allowbreak{}product} & The exact chronological product for a scalar zero-force gain recurrence from its initial index. & Reindexing to an arbitrary positive-excursion start and the infinite-horizon equivalence wording.\\[6pt]
Result~\ref{row:thm:exact-radial-face-cocycle}: Exact radial-face affine cocycle\par {\footnotesize\ttfamily Radial\allowbreak{}Tangential\allowbreak{}Energy}\par {\textsc{partial}}\par {\footnotesize\ttfamily face\allowbreak{}Gain}, {\footnotesize\ttfamily face\allowbreak{}Reopening}, {\footnotesize\ttfamily exact\_\allowbreak{}face\_\allowbreak{}affine\_\allowbreak{}step}, {\footnotesize\ttfamily exact\_\allowbreak{}face\_\allowbreak{}affine\_\allowbreak{}cocycle}, {\footnotesize\ttfamily canonical\_\allowbreak{}face\_\allowbreak{}affine\_\allowbreak{}cocycle} & The canonical scalar face branches, exact edge identity, and inductive affine product-convolution across every zero-face restart. & Energy nonnegativity, the physical Frobenius identification, and identification of the positive branch with radial gain.\\[6pt]
Result~\ref{row:thm:intermittent-block-radial-face-closure}: Intermittent block radial-face closure\par {\footnotesize\ttfamily Radial\allowbreak{}Face\allowbreak{}Block\allowbreak{}Closure}\par {\textsc{partial}}\par {\footnotesize\ttfamily shifted\allowbreak{}Face\allowbreak{}Envelope}, {\footnotesize\ttfamily exact\_\allowbreak{}shifted\_\allowbreak{}face\_\allowbreak{}cocycle}, {\footnotesize\ttfamily uniform\_\allowbreak{}anchor\_\allowbreak{}convolution}, {\footnotesize\ttfamily anchor\_\allowbreak{}tendsto\_\allowbreak{}zero\_\allowbreak{}of\_\allowbreak{}uniform\_\allowbreak{}gain}, {\footnotesize\ttfamily geometric\allowbreak{}Convolution}, {\footnotesize\ttfamily geometric\_\allowbreak{}convolution\_\allowbreak{}nonnegative}, {\footnotesize\ttfamily geometric\_\allowbreak{}convolution\_\allowbreak{}tendsto\_\allowbreak{}zero}, {\footnotesize\ttfamily anchor\_\allowbreak{}excursion\_\allowbreak{}maximum\_\allowbreak{}collapse}, {\footnotesize\ttfamily intermittent\_\allowbreak{}block\_\allowbreak{}maximum\_\allowbreak{}tendsto\_\allowbreak{}zero} & Exact canonical face-cocycle composition from arbitrary finite anchors, the finite geometric anchor convolution, the explicit vanishing-forcing geometric-tail limit, and closure from vanishing anchors and excursions to vanishing attained block maxima. & Construction of manuscript block coefficients from arbitrary anchor endpoints, passage from a covering block family to every natural-number time, and transfer to a finite row-map registry.\\[6pt]
Result~\ref{row:thm:direct-energy-envelope}: Direct nonautonomous energy envelope\par {\footnotesize\ttfamily Observable\allowbreak{}Energy\allowbreak{}Dissipation}\par {\textsc{partial}}\par {\footnotesize\ttfamily ordered\allowbreak{}Envelope}, {\footnotesize\ttfamily ordered\_\allowbreak{}affine\_\allowbreak{}recursion} & Inductive chronological domination of a nonnegative affine energy recurrence with nonnegative time-varying gains. & The explicit finite product notation and transfer from stacked energy to maximum row diameter.\\[6pt]
Result~\ref{row:thm:summable-reopening-closure}: Summable reopening closure\par {\footnotesize\ttfamily Reopening\allowbreak{}Budget}\par {\textsc{partial}}\par {\footnotesize\ttfamily reopening}, {\footnotesize\ttfamily finite\_\allowbreak{}reopening\_\allowbreak{}envelope}, {\footnotesize\ttfamily summable\_\allowbreak{}reopening\_\allowbreak{}collapse} & Absolute positive reopening, its finite telescoping envelope, and convergence from arbitrarily late small nonnegative energy plus summability. & Identification of energy with a specific PLDR row-map registry.\\[6pt]
Result~\ref{row:prop:direct-face-preservation}: Exact face-preservation criterion\par {\footnotesize\ttfamily Direct\allowbreak{}Observable\allowbreak{}Energy}\par {\textsc{partial}}\par {\footnotesize\ttfamily face\_\allowbreak{}endpoint\_\allowbreak{}energy}, {\footnotesize\ttfamily face\_\allowbreak{}preserved\_\allowbreak{}iff} & At zero source observable, endpoint energy is the increment charge and vanishes exactly when every finite increment coordinate vanishes. & Identification of increment coordinates with implemented gate, shape, or optimizer sources.\\[6pt]
Result~\ref{row:thm:comprehensive-direct-collapse}: Exact direct characterization\par \emph{None}\par {\textsc{not formalized}}\par \emph{None} & No formal coverage is claimed for the finite-registry assembly. & The equivalence between stacked native energy convergence and every registered maximum row-diameter convergence.\\[6pt]
Result~\ref{row:prop:cumulative-direct-ledger}: Cumulative ledger telescope\par {\footnotesize\ttfamily Direct\allowbreak{}Energy\allowbreak{}Cumulative}\par {\textsc{faithful}}\par {\footnotesize\ttfamily cumulative\allowbreak{}Ledger}, {\footnotesize\ttfamily cumulative\_\allowbreak{}ledger\_\allowbreak{}eq\_\allowbreak{}energy}, {\footnotesize\ttfamily cumulative\_\allowbreak{}ledger\_\allowbreak{}tendsto\_\allowbreak{}zero\_\allowbreak{}iff} & The exact finite work-charge telescope and the literal equivalence between cumulative-ledger convergence to zero and scalar energy convergence to zero. & Identification of the scalar sequences with captured PLDR row-map tensors.\\[6pt]
Result~\ref{row:thm:sufficient-direct-closures}: Sufficient direct closures\par \emph{None}\par {\textsc{not formalized}}\par \emph{None} & No formal coverage is claimed for the manuscript assembly. & Composition of the separately mapped nonautonomous-envelope and summable-reopening results with finite-registry row-diameter equivalence.\\[6pt]
Result~\ref{row:thm:implemented-adamw-lift}: Implemented lifted AdamW derivative\par {\footnotesize\ttfamily Lifted\allowbreak{}Adam\allowbreak{}W\allowbreak{}Matrix}\par {\textsc{partial}}\par {\footnotesize\ttfamily full\allowbreak{}Linear\allowbreak{}Step}, {\footnotesize\ttfamily full\_\allowbreak{}adamw\_\allowbreak{}block\_\allowbreak{}expansion}, {\footnotesize\ttfamily state\_\allowbreak{}independent\_\allowbreak{}intervention} & The abstract three-block lifted state action and its block expansion. The same additive forcing cancels in the difference of two affine responses; this is a ring identity. & Construction of live Hessian, clipping, bias-correction, denominator, and decay blocks from the implemented optimizer.\\[6pt]
Result~\ref{row:prop:gate-zero-invariance}: Gate-zero invariance criterion\par {\footnotesize\ttfamily Gate\allowbreak{}Zero\allowbreak{}Stratum}\par {\textsc{partial}}\par {\footnotesize\ttfamily zero\_\allowbreak{}gate\_\allowbreak{}successor\_\allowbreak{}iff\_\allowbreak{}zero\_\allowbreak{}moment}, {\footnotesize\ttfamily zero\_\allowbreak{}gate\_\allowbreak{}zero\_\allowbreak{}moment\_\allowbreak{}invariant\_\allowbreak{}iff} & The scalar adaptive-write zero criterion and its reduction to a zero clipped gradient when the incoming first moment vanishes. & The coordinatewise AdamW denominator construction and the final physical gate-shape alternative.\\[6pt]
Result~\ref{row:thm:observable-complete-state}: Observable complete-state variation of constants\par {\footnotesize\ttfamily Observable\allowbreak{}Complete\allowbreak{}State\allowbreak{}Cocycle}\par {\textsc{partial}}\par {\footnotesize\ttfamily homogeneous}, {\footnotesize\ttfamily transported}, {\footnotesize\ttfamily complete\_\allowbreak{}state\_\allowbreak{}variation\_\allowbreak{}of\_\allowbreak{}constants}, {\footnotesize\ttfamily observable\_\allowbreak{}variation\_\allowbreak{}of\_\allowbreak{}constants}, {\footnotesize\ttfamily componentwise\_\allowbreak{}observable\_\allowbreak{}vanishing} & Exact chronological state and observable decompositions in additive groups, with separate transported force and remainder and a componentwise vanishing sufficient condition. & Finite-registry row geometry, differentiable observation remainders, the norm envelope, the Gram cancellation identity, and the necessary-and-sufficient limit wording.\\[6pt]
Result~\ref{row:thm:finite-increment-observable-balance}: Exact finite-increment observable balance\par {\footnotesize\ttfamily Finite\allowbreak{}Increment\allowbreak{}Observable\allowbreak{}Balance}\par {\textsc{partial}}\par {\footnotesize\ttfamily homogeneous}, {\footnotesize\ttfamily accumulated\allowbreak{}Correction}, {\footnotesize\ttfamily finite\_\allowbreak{}increment\_\allowbreak{}state\_\allowbreak{}unroll}, {\footnotesize\ttfamily exact\_\allowbreak{}finite\_\allowbreak{}increment\_\allowbreak{}observable\_\allowbreak{}balance} & The chronological finite-state correction unroll is the inductive mathematical kernel; the final three-term observable equality is algebraic and API-adjacent. & Route labels, differentiability interpretations, numerical effect floors, cross-Gram diagnostics, empirical cancellation classification, and implementation of the observation API.\\[6pt]
Result~\ref{row:prop:stratified-itinerary}: Stratified itinerary identity\par \emph{None}\par {\textsc{not formalized}}\par \emph{None} & No formal coverage is claimed. & Route-dependent chart spaces, nonsmooth compatible linear maps, and separation of exact algebra from routewise remainder regularity.\\[6pt]
Result~\ref{row:thm:lower-metric-transfer}: Lower-metric fixed-norm transfer\par {\footnotesize\ttfamily Observable\allowbreak{}Complete\allowbreak{}State\allowbreak{}Cocycle}\par {\textsc{partial}}\par {\footnotesize\ttfamily lower\_\allowbreak{}metric\_\allowbreak{}transfer} & A uniform positive lower comparison transfers convergence of a nonnegative scheduled quantity to convergence of the fixed quantity without an upper comparison. & Matrix-valued metrics, extraction of square-root norm inequalities, and variable chart dimensions.\\[6pt]
Result~\ref{row:thm:scheduled-tube-collapse}: Scheduled-norm tube collapse\par {\footnotesize\ttfamily Complete\allowbreak{}State\allowbreak{}Cocycle}\par {\textsc{partial}}\par {\footnotesize\ttfamily scheduled\_\allowbreak{}affine\_\allowbreak{}collapse} & A nonnegative scheduled state below a vanishing chronological affine envelope converges to zero without a fixed-metric comparison. & Variable-radius tube induction, scheduled induced norms, quadratic remainder construction, and separation of the two chronological limits.\\[6pt]
Result~\ref{row:cor:scheduled-cover-collapse}: Scheduled-cover physical collapse\par {\footnotesize\ttfamily Complete\allowbreak{}State\allowbreak{}Cocycle}\par {\textsc{partial}}\par {\footnotesize\ttfamily physical\_\allowbreak{}cover\_\allowbreak{}collapse} & A vanishing nonnegative scheduled state and baseline transfer through a uniform linear physical cover. & Finite-registry row-diameter geometry and a uniform context-class supremum.\\[6pt]
Result~\ref{row:cor:fixed-norm-collapse-transfer}: Fixed-norm complete-state transfer\par {\footnotesize\ttfamily Observable\allowbreak{}Complete\allowbreak{}State\allowbreak{}Cocycle}\par {\textsc{partial}}\par {\footnotesize\ttfamily fixed\_\allowbreak{}norm\_\allowbreak{}cover\_\allowbreak{}collapse} & Lower-metric transfer followed by a fixed-norm physical cover, with no upper scheduled-metric comparison. & Matrix-valued metrics, square-root extraction, and finite-registry row-diameter geometry.\\[6pt]
Result~\ref{row:thm:finite-cocycle-envelope}: Finite-horizon cocycle envelope\par {\footnotesize\ttfamily Complete\allowbreak{}State\allowbreak{}Cocycle}\par {\textsc{partial}}\par {\footnotesize\ttfamily complete\_\allowbreak{}state\_\allowbreak{}affine\_\allowbreak{}envelope} & Chronological finite iteration of the nonnegative affine scalar estimate used after the nonlinear tube bound. & The tube induction, scheduled norms, and finite physical-cover maximum.\\[6pt]
Result~\ref{row:cor:continuous-observable}: Observable continuous-time specialization\par \emph{None}\par {\textsc{not formalized}}\par \emph{None} & No formal coverage is claimed. & Fundamental propagators, the Duhamel formula, observation transport, and any discrete-to-continuous scaling limit.\\[6pt]
Result~\ref{row:cor:linear-stochastic-forcing}: Transported stochastic forcing\par \emph{None}\par {\textsc{not formalized}}\par \emph{None} & No formal coverage is claimed. & Martingale orthogonality, the transported second-moment bound, and its mean-square consequence.\\[6pt]
Result~\ref{row:thm:collapse-obstructions}: Dynamical and asymptotic collapse obstructions\par {\footnotesize\ttfamily Complete\allowbreak{}State\allowbreak{}Cocycle}\par {\textsc{partial}}\par {\footnotesize\ttfamily vanishing\_\allowbreak{}invariance\_\allowbreak{}force\_\allowbreak{}necessary}, {\footnotesize\ttfamily nonvanishing\_\allowbreak{}force\_\allowbreak{}obstructs\_\allowbreak{}collapse}, {\footnotesize\ttfamily delayed\allowbreak{}Reopening}, {\footnotesize\ttfamily delayed\_\allowbreak{}reopening\_\allowbreak{}matches\_\allowbreak{}finite\_\allowbreak{}prefix}, {\footnotesize\ttfamily delayed\_\allowbreak{}reopening\_\allowbreak{}not\_\allowbreak{}tendsto\_\allowbreak{}zero} & Necessity of vanishing realized force, its contrapositive obstruction, and an explicit finite-prefix reopening counterexample. & The degenerating scheduled-metric counterexample.\\[6pt]
Result~\ref{row:thm:linear-observation-kernel-obstruction}: Linear observation-kernel obstruction\par {\footnotesize\ttfamily Observation\allowbreak{}Zero\allowbreak{}Fiber}\par {\textsc{partial}}\par {\footnotesize\ttfamily constant\_\allowbreak{}nonzero\_\allowbreak{}not\_\allowbreak{}tendsto\_\allowbreak{}zero}, {\footnotesize\ttfamily linear\_\allowbreak{}common\_\allowbreak{}kernel\_\allowbreak{}obstruction} & A persistent nonzero common-kernel vector gives identically zero linear observations and a nonconvergent constant state path. & Time-varying observation codomains and embedding into a particular optimizer recurrence.\\[6pt]
Result~\ref{row:prop:nonlinear-zero-fiber-obstruction}: Nonlinear zero-fiber obstruction and derivative caveat\par {\footnotesize\ttfamily Observation\allowbreak{}Zero\allowbreak{}Fiber}\par {\textsc{partial}}\par {\footnotesize\ttfamily nonlinear\_\allowbreak{}common\_\allowbreak{}zero\_\allowbreak{}fiber\_\allowbreak{}obstruction}, {\footnotesize\ttfamily square\allowbreak{}Observation}, {\footnotesize\ttfamily square\_\allowbreak{}observation\_\allowbreak{}zero\_\allowbreak{}derivative}, {\footnotesize\ttfamily square\_\allowbreak{}observation\_\allowbreak{}nonzero\_\allowbreak{}at\_\allowbreak{}one}, {\footnotesize\ttfamily derivative\_\allowbreak{}kernel\_\allowbreak{}does\_\allowbreak{}not\_\allowbreak{}imply\_\allowbreak{}zero\_\allowbreak{}fiber} & The constant-path obstruction for a common nonlinear zero fiber and the scalar-square counterexample separating a zero derivative from a nonlinear zero fiber. & Time-varying codomains, Fréchet derivatives in general normed spaces, and the physical row-map observation formula.\\[6pt]
Result~\ref{row:thm:plga-quotient-defect}: PLGA quotient-defect decomposition\par {\footnotesize\ttfamily P\allowbreak{}L\allowbreak{}G\allowbreak{}A\allowbreak{}Row\allowbreak{}Constant\allowbreak{}Defect}\par {\textsc{partial}}\par {\footnotesize\ttfamily constant\_\allowbreak{}input\_\allowbreak{}output\_\allowbreak{}is\_\allowbreak{}defect}, {\footnotesize\ttfamily nonzero\_\allowbreak{}defect\_\allowbreak{}obstructs\_\allowbreak{}row\_\allowbreak{}preservation}, {\footnotesize\ttfamily quotient\_\allowbreak{}defect\_\allowbreak{}bound} & The abstract response-plus-defect bound, exact reduction at zero upstream quotient, and obstruction from a nonzero constant-input defect. & The implemented PLGA formula, segment integral construction of the response bound, and sequential transfer.\\[6pt]
Result~\ref{row:prop:plga-row-preservation}: Sufficient PLGA row preservation\par \emph{None}\par {\textsc{not formalized}}\par \emph{None} & No formal coverage is claimed. & Propagation of row constancy through the five implemented PLGA parameter conditions.\\[6pt]
\midrule
\multicolumn{3}{@{}l}{Mapped-result subtotal: faithful 3; partial 29; not-formalized 8.}\\
\bottomrule
\end{longtable}
\endgroup

\section{Chronological row renormalization}
Paths in this table are relative to \path{vendor/rg}.
The correspondence distinguishes scalar and finite algebra, asymptotic
sequence statements, and conventional probability arguments.
\begingroup
\scriptsize
\hbadness=10000
\setlength{\LTleft}{0pt}
\setlength{\LTright}{0pt}
\begin{longtable}{@{}
 >{\raggedright\arraybackslash}p{0.18\textwidth}
 >{\raggedright\arraybackslash}p{0.18\textwidth}
 >{\raggedright\arraybackslash}p{0.10\textwidth}
 >{\raggedright\arraybackslash}p{0.44\textwidth}@{}}
\caption{Clause-level correspondence between manuscript results and Lean declarations. Every listed Lean declaration and helper has a digest-validated elaborated type under the pinned toolchain; proof terms are outside that statement-only fingerprint.}
\label{rg:tab:formal-map}\\
\toprule
Manuscript result & Lean module & Coverage & Checked clauses and boundary \\
\midrule
\endfirsthead
\toprule
Manuscript result & Lean module & Coverage & Checked clauses and boundary \\
\midrule
\endhead
Result~\ref{rg:thm:observer-energy-enclosure} &
\codepath{RowRGMap.ObserverEnvelope} & faithful &
energy-enclosure: \codepath{lowerEnergy}, \codepath{upperEnergy}, \codepath{energy_enclosed}; exact-decisions: \codepath{positive_of_positive_lower}, \codepath{exact_face_of_zero_upper}, \codepath{unresolved_interval_has_zero_and_positive}; asymptotic-boundary: \codepath{convergence_from_upper_envelope}, \codepath{fixed_tolerance_only_bounds}, \codepath{fixed_tolerance_does_not_imply_convergence}. Lean checks the nonnegative enclosure, exact zero and positivity decisions, the unresolved stratum, upper-envelope convergence, and the fixed-tolerance counterexample. \\

Result~\ref{rg:thm:radial-closing} &
\codepath{RowRGMap.RadialEnergy} & kernel &
checked-kernel: \codepath{gain_nonnegative}, \codepath{relative_increment}, \codepath{strict_closing_iff}; remaining-manuscript-argument: manuscript proof. Lean checks the scalar algebra after orthogonality; projection and uniqueness remain in the manuscript. \\

Result~\ref{rg:def:canonical-edge} &
\codepath{RowRGMap.AffineCocycle} & faithful &
checked-kernel: \codepath{canonical}, \codepath{canonical_act}, \codepath{canonical_nonnegative}. Lean checks the exact positive and face branches, their action, and nonnegativity. \\

Result~\ref{rg:thm:canonical-cone-closure} &
\codepath{RowRGMap.AffineCocycle} & faithful &
checked-kernel: \codepath{canonical_iff_gain_zero_or_source_zero}, \codepath{compose_canonical}, \codepath{identity_canonical}, \codepath{block_canonical}, \codepath{canonical_is_canonical}. Lean checks branch membership, identity membership, and closure under arbitrary finite chronological blocking. \\

Result~\ref{rg:thm:affine-rg-semigroup} &
\codepath{RowRGMap.AffineCocycle} & kernel &
checked-kernel: \codepath{compose_identity_left}, \codepath{compose_identity_right}, \codepath{compose_assoc}, \codepath{act_compose}, \codepath{compose_nonnegative}, \codepath{block_append}, \codepath{block_act}, \codepath{block_nonnegative}; remaining-manuscript-argument: manuscript proof. Lean checks the semigroup, action, positivity, and ordered list blocking. The indexed product-convolution is the manuscript induction. \\

Result~\ref{rg:thm:interval-affine-rg} &
\codepath{RowRGMap.IntervalAffine} & faithful &
objects-and-validity: \codepath{Interval}, \codepath{Valid}, \codepath{Contains}, \codepath{singleton}, \codepath{Edge}, \codepath{EdgeValid}, \codepath{ExactEdge}, \codepath{identity}, \codepath{singleton_valid}, \codepath{identity_valid}; endpoint-arithmetic: \codepath{multiply}, \codepath{add}, \codepath{multiply_lower}, \codepath{multiply_upper}, \codepath{add_lower}, \codepath{add_upper}, \codepath{multiply_valid}, \codepath{add_valid}, \codepath{contains_nonnegative}, \codepath{multiply_contains}, \codepath{add_contains}; exact-semigroup: \codepath{compose}, \codepath{act}, \codepath{compose_assoc}, \codepath{compose_identity_left}, \codepath{compose_identity_right}, \codepath{compose_valid}, \codepath{compose_exact}, \codepath{act_exact}; monotone-enclosure: \codepath{Includes}, \codepath{includes_refl}, \codepath{multiply_includes}, \codepath{add_includes}, \codepath{compose_includes}, \codepath{Encloses}, \codepath{intervalBlock}, \codepath{exactBlock}, \codepath{intervalBlock_valid}, \codepath{intervalBlock_encloses_exact}, \codepath{compose_contains_exact}, \codepath{act_contains_exact}; zero-endpoint-branch: \codepath{gain_bracket}, \codepath{no_finite_gain_bound_with_positive_successor}, \codepath{zero_successor_iff_zero_gain}. Lean checks exact nonnegative endpoint arithmetic, identities, inclusion monotonicity, finite chronological blocking, action enclosure, the positive gain bracket, and both zero-endpoint cases. Directed floating-point rounding remains an implementation obligation and is only claimed to preserve enclosure under a left fold. \\

Result~\ref{rg:thm:gauge-covariance} &
\codepath{RowRGMap.AffineCocycle} & faithful &
two-edge-covariance: \codepath{normalize_compose}; finite-block-covariance: \codepath{normalizeChain}, \codepath{block_normalizeChain}. Lean checks both the two-edge identity and its arbitrary finite-block consequence for nonzero intermediate and target gauges. \\

Result~\ref{rg:prop:matrix-comparison-rg} &
None & not formalized &
manuscript-proof: manuscript proof. The componentwise matrix comparison and associativity proof is supplied only in the manuscript. \\

Result~\ref{rg:thm:two-sector-collapse} &
\codepath{RowRGMap.CollapseCriterion} & kernel &
checked-kernel: \codepath{orbit_decomposition}, \codepath{sectors_suffice}, \codepath{sectors_necessary}, \codepath{orbit_tendsto_zero_iff_sectors}, \codepath{source_le_successor}, \codepath{zero_orbit_iff_zero_sectors}; remaining-manuscript-argument: manuscript proof. Lean checks the scalar nonnegative limit equivalence and pointwise zero criterion. The finite registry and diameter transfer remain in the manuscript. \\

Result~\ref{rg:thm:clean-fixed-points} &
\codepath{RowRGMap.CriticalFlow} & kernel &
checked-kernel: \codepath{binary_gain_fixed_iff}, \codepath{nonnegative_binary_fixed_points}, \codepath{binary_fixed_iff}, \codepath{critical_edge_fixed}, \codepath{absorbing_edge_fixed}; remaining-manuscript-argument: manuscript proof. Lean checks the complete fixed set for binary blocking. The manuscript proves the arbitrary integer block-factor extension. \\

Result~\ref{rg:thm:clean-critical-scaling} &
\codepath{RowRGMap.CriticalFlow}; \codepath{RowRGMap.CriticalSource} & kernel &
binary-scaling: \codepath{gain_perturbation_scaling}, \codepath{critical_source_scaling}, \codepath{subcritical_distance_scaling}, \codepath{stationary_energy_fixed}, \codepath{stationary_energy_nonnegative}; arbitrary-block-source: \codepath{blockRG}, \codepath{blockRG_zero}, \codepath{blockRG_succ}, \codepath{blockRG_critical}, \codepath{blockIterate}, \codepath{blockIterate_critical}, \codepath{critical_source_tendsto_atTop_arbitrary}, \codepath{binaryIterate}, \codepath{binaryIterate_critical}, \codepath{critical_source_tendsto_atTop}; real-log-asymptotics: manuscript proof. Lean checks exact binary scaling, arbitrary integer block-factor source iteration, source relevance, and the stationary forced value. Real-log asymptotics and differentiation remain in the manuscript. \\

Result~\ref{rg:lem:cumulative-log-criterion} &
\codepath{RowRGMap.PositiveTail} & kernel &
checked-kernel: \codepath{cumulativeLog}, \codepath{energy}, \codepath{cumulativeLog_zero}, \codepath{cumulativeLog_succ}, \codepath{energy_zero}, \codepath{energy_succ}, \codepath{log_energy_ratio}, \codepath{energy_tendsto_zero_iff}; remaining-manuscript-argument: manuscript proof. Lean checks the exact exponential representation, one-step recurrence, log identity, and collapse equivalence. Divergence and finite-positive-limit branches remain in the manuscript. \\

Result~\ref{rg:prop:regular-nonautonomous-drift} &
None & not formalized &
manuscript-proof: manuscript proof. The comparison of cumulative asymptotic scales is supplied only in the manuscript. \\

Result~\ref{rg:thm:positive-excursion-phase} &
None & not formalized &
manuscript-proof: manuscript proof. The nonzero mean-rate specialization is supplied only in the manuscript. \\

Result~\ref{rg:thm:recurrent-face-collapse} &
\codepath{RowRGMap.RecurrentFaces}; \codepath{RowRGMap.CollapseCriterion} & faithful &
checked-kernel: \codepath{RecurrentFaceSchedule}, \codepath{excursionInterval}, \codepath{excursionInterval_nonempty}, \codepath{excursionPeak}, \codepath{excursionRestart}, \codepath{excursionAmplification}, \codepath{face_times_unbounded}, \codepath{interval_cover}, \codepath{energy_le_excursionPeak}, \codepath{excursionPeak_nonnegative}, \codepath{peak_zero_of_restart_zero}, \codepath{peak_eq_restart_mul_amplification}, \codepath{energy_tendsto_zero_iff_excursionPeak}, \codepath{restart_mem_excursionInterval}, \codepath{excursionRestart_nonnegative}, \codepath{excursionRestart_le_peak}, \codepath{excursionRestart_tendsto_zero_of_collapse}, \codepath{restart_times_amplification_tendsto_zero_of_collapse}; restart-only-obstruction: \codepath{recurrent_face_energy_nonnegative}, \codepath{obstructionSchedule}, \codepath{obstruction_restart}, \codepath{obstruction_restart_tendsto_zero}, \codepath{restart_tendsto_zero_not_sufficient}. Lean constructs and covers finite excursions, proves the peak criterion and restart-amplification consequence, and instantiates a nonnegative obstruction showing that vanishing restarts alone do not imply collapse. \\

Result~\ref{rg:prop:restart-amplification-criticality} &
None & not formalized &
manuscript-proof: manuscript proof. The logarithmic ratio comparison is supplied only in the manuscript. \\

Result~\ref{rg:thm:uniform-block-closure} &
None & not formalized &
manuscript-proof: manuscript proof. The geometric-convolution recurrence estimate is supplied only in the manuscript. \\

Result~\ref{rg:cor:block-cover-closure} &
None & not formalized &
manuscript-proof: manuscript proof. The independent unbounded-anchor cover argument is supplied only in the manuscript. \\

Result~\ref{rg:thm:complete-state-kernel-rg} &
None & not formalized &
manuscript-proof: manuscript proof. The standard Markov-kernel associativity and blocking semigroup, scale-indexed lumpability theorem, propagation to scale multiples, deterministic characterization, and three-cycle non-descent counterexample are supplied only in the manuscript. \\

Result~\ref{rg:def:multiplicative-universality} &
None & not formalized &
manuscript-proof: manuscript proof. The equivalence relation on centered and scaled limiting laws is defined only in the manuscript. \\

Result~\ref{rg:thm:gaussian-log-gain-fixed-point} &
None & not formalized &
manuscript-proof: manuscript proof. The manuscript invokes the classical Lindeberg-Levy theorem under explicit hypotheses; no Lean probability formalization is claimed. \\

Result~\ref{rg:prop:universality-taxonomy} &
Combined kernels below & partial &
Log balance, subpower sequence limit, bounded-factor inequalities and stable-index
arithmetic are checked in \path{PldrTrainingDynamics/ScalingBalance.lean} at the
combined root. Stable attraction and the probabilistic tail example remain
written arguments; the inherited row-RG namespace has no probability theorem. \\

Result~\ref{rg:prop:plga-face-preservation} &
None & not formalized &
manuscript-proof: manuscript proof. The exact functional criterion, five jointly sufficient conditions, conditional necessity of the final two conditions, and cancellation witnesses are proved only in the manuscript. \\

Result~\ref{rg:thm:plga-rg-transfer} &
\codepath{RowRGMap.ResponseDefect} & kernel &
checked-kernel: \codepath{response_plus_defect_bound}, \codepath{nonzero_defect_obstructs_zero_output}; remaining-manuscript-argument: manuscript proof. Lean checks the norm inequality and nonzero-defect obstruction. The Frechet segment integral remains in the manuscript. \\

Result~\ref{rg:cor:plga-sequential-collapse} &
None & not formalized &
manuscript-proof: manuscript proof. The sequential two-term limit proof is supplied only in the manuscript. \\

Result~\ref{rg:prop:gauge-orbit-null} &
\codepath{RowRGMap.GaugeFiber} & faithful &
finite-equivariant-orbit: \codepath{iterate_equivariant}, \codepath{gauge_orbit_successor_equal}. Lean checks finite-iterate equivariance and invariant-observable equality in the same abstract deterministic setting; the concrete PLDR sign action is validated experimentally. \\

Result~\ref{rg:cor:paired-gauge-successor} &
\codepath{RowRGMap.GaugeFiber} & faithful &
paired-orbit-identity: \codepath{compose_equivariant}, \codepath{compose_invariant}, \codepath{paired_gauge_successor_equal}. Lean checks closure of equivariance and invariance under composition and the finite-scale paired-orbit identity in the same abstract deterministic setting. \\

Result~\ref{rg:lem:augmented-fiber-witness} &
\codepath{RowRGMap.GaugeFiber} & faithful &
equal-source-unequal-successor: \codepath{augmented_fiber_witness_not_closed}. Lean checks the contradiction from one equal augmented-state pair with unequal successor-energy readouts. \\

\bottomrule
\end{longtable}
\endgroup

\section{Model-wide laws and predictive observations}
Paths in this table are relative to \path{vendor/model}.
Finite conditional-expectation analogues, covariance sums, or
normalization formulas do not formalize general disintegration,
native program smoothness, empirical independence, or a thermodynamic
critical limit. Those hypotheses and proofs are stated in the text.
\input{content/model/generated/formal-correspondence.tex}

\section{The interface between absolute and normalized energy}
\begin{longtable}{@{}p{.21\textwidth}p{.39\textwidth}p{.32\textwidth}@{}}
\toprule Written result & Formal implementation & Checked scope\\\midrule\endhead
Proposition~\ref{prop:absolute-relative} &
\path{PldrTrainingDynamics/RowFraction.lean}:
\path{rowFraction_reconstruct},
\path{contrast_le_scaled_fraction},
\path{fraction_le_scaled_contrast} &
Exact reconstruction and scalar inequalities under the stated positive
denominator and amplitude bounds. Orthogonal projection, convergence,
and counterexamples have the written proof.\\
Proposition~\ref{prop:row-predictive-bridge} &
No additional formalization claimed &
The complete Lipschitz recursion and categorical Hessian argument
are proved in the text.\\
\bottomrule
\end{longtable}
The public \path{provenance/statement-manifest.json} records stable
monograph labels, final numbering, chapter subjects, formal module paths,
checked hypotheses and obligations outside coverage. The checker resolves
these entries against shipped Lean modules without reading TeX. The map
preserves the distinction between a selected formal kernel and a complete
written theorem.

\section{Finite forcing and error-budget kernels}
All paths in this table are relative to \path{PldrTrainingDynamics/}.
These are partial checks of the indicated written results. In the
row correspondence above, the general stochastic forcing argument remains
unformalized; the finite probability-space and scalar limit checks here
supply its explicitly delimited additions.
\begingroup
\small
\setlength{\tabcolsep}{3pt}
\begin{longtable}{@{}>{\raggedright\arraybackslash}p{.34\textwidth}>{\raggedright\arraybackslash}p{.30\textwidth}>{\raggedright\arraybackslash}p{.30\textwidth}@{}}
\toprule Written result and formal locator & Checked scope & Remaining obligations\\\midrule\endhead
Corollary~\ref{row:cor:linear-stochastic-forcing}\par
\path{StochasticTransport.lean}:\par \path{weighted_add_sq},
\path{centered_second_moment}, \path{mean_plus_variance} &
Finite normalized nonnegative weights in a real inner-product space;
zero vector means and pairwise expected orthogonality are hypotheses. &
General martingale conditional expectation is proved in the text.\\[6pt]
Corollary~\ref{row:cor:linear-stochastic-forcing}\par
\path{LimitCriteria.lean}:\par \path{nonnegative_sum_iff},
\path{envelope_suffices}, \path{separate_decay_suffices} &
Nonnegative real sequence decomposition and squeeze; separate scalar decay
is sufficient. &
These kernels assume the relevant identity or inequality.\\[6pt]
Corollary~\ref{model:cor:row-path-error}\par
\path{PathErrorBudget.lean}:\par \path{prefix_budget},
\path{prefix_le_full}, \path{add_remainder},
\path{scaled_budget_suffices}, \path{ratio_rescale} &
Telescoping, assumed one-step bounds, positive scales, and scalar ratio
invariance. &
The derivative bound, matrix rescaling, and generator smoothness
have stand-alone analytic arguments.\\[6pt]
Counterexamples following both corollaries\par
\path{CancellationExamples.lean}:\par \path{cancellation_recurrence},
\path{killed_noise_recurrence}, \path{killed_noise_tail},
\path{centered_sign_variance}, \path{retained_unit_direction},
\path{radial_budget_positive} &
Exact finite recurrences, two-sign moments, and the $7/2$ radial budget. &
The operator-norm envelope and the zero directional derivative are justified
in the written examples.\\
\bottomrule
\end{longtable}
\endgroup

\section{Crossover balance and scalar face memory}
Paths in this table are relative to \path{PldrTrainingDynamics/}. These
checks concern the scalar clauses stated below. The sequence formulation
uses the subpower property along the chosen diverging crossover, with
pointwise hypotheses after any required finite tail restriction. It does
not formalize a stable central limit theorem or infer a training law.
\begin{longtable}{@{}p{.22\textwidth}p{.34\textwidth}p{.36\textwidth}@{}}
\toprule Written clause & Module and declarations & Scope and remaining obligations\\\midrule\endhead
Equation~\eqref{rg:eq:log-balance-identity} &
\path{ScalingBalance.lean}: \path{log_balance_identity} &
The logarithmic identity from positive drift magnitude, scale, factor and
balance ratio, with the exact balance as hypothesis.\\[4pt]
Proposition~\ref{rg:prop:universality-taxonomy},
Equation~\eqref{rg:eq:hurst-crossover} &
\path{ScalingBalance.lean}: \path{bounded_log_ratio},
\path{log_exponent_of_subpower}, \path{log_exponent_of_bounded_balance} &
Real-sequence squeeze and reciprocal limit from a diverging positive scale,
subpower along the sequence, and exact balance with bounded positive ratio.
No fluctuation law or crossover selection is established by these kernels.\\[4pt]
Equation~\eqref{rg:eq:bounded-factor-crossover} and the stable-index substitution &
\path{ScalingBalance.lean}: \path{bounded_factor_power_bounds},
\path{stable_exponent_identity}, \path{stable_three_halves} &
Positive-power bounds in the equivalent quotient form
\((ar/\delta)^{1/(1-H)}\), and
\(1/(1-1/\alpha)=\alpha/(\alpha-1)\) for \(\alpha>1\).
At \(\alpha=3/2\) the power and logarithmic powers are three and two.
Tail regular variation and its probability limit have written arguments.\\[4pt]
Equation~\eqref{rg:eq:scalar-face-memory} &
\path{ScalarFaceMemory.lean}: \path{zero_gain_act},
\path{block_zero_gain}, \path{block_zero_gain_independent} &
A zero gain in a fixed chronological list annihilates the scalar homogeneous
coefficient. The existing affine composition checks support this finite
extension. Coefficient dependence on complete training state is not erased.\\
\bottomrule
\end{longtable}

\section{Trajectory-compatible chart connection}
The following selected algebraic checks are in
\path{PldrTrainingDynamics/TrajectoryChart.lean}. Admitted domain types
encode where the maps act; arbitrary extensions outside a chart do not
establish domain admission. The time-indexed one-step identities allow
varying state and coordinate types. The transport wrappers use a fixed
additive coordinate type.
\begingroup
\small
\setlength{\tabcolsep}{3pt}
\begin{longtable}{@{}>{\raggedright\arraybackslash}p{.34\textwidth}>{\raggedright\arraybackslash}p{.30\textwidth}>{\raggedright\arraybackslash}p{.30\textwidth}@{}}
\toprule Written result and formal locator & Checked scope & Remaining obligations\\\midrule\endhead
Equation~\eqref{row:eq:realized-chart-step}\par
\path{realized_step_of_lift} &
Substitution from state evolution, extraction and realized lift. &
Chart existence, smoothness and the chain rule are written obligations.\\[6pt]
Equation~\eqref{row:eq:realized-chart-observation}\par
\path{realized_observation_of_lift} &
Physical observation composed with the same lifted state. &
The admitted lift and its domain remain hypotheses.\\[6pt]
Equations~\eqref{row:eq:fixed-slice-defect} and
\eqref{row:eq:fixed-slice-observation-defect}\par
\path{fixed_slice_defect}, \path{observation_defect} &
Additive mismatch identities. &
No Lipschitz or smallness conclusions are supplied.\\[6pt]
Necessity example in Section~\ref{row:sec:trajectory-chart}\par
\path{right_inverse_not_enough}, \path{adapted_slice_recovers_step} &
Exact two-dimensional example: a right inverse alone fails, while the
adapted slice recovers the successor. &
This example does not establish general chart existence or regularity.\\[6pt]
Theorem~\ref{row:thm:observable-complete-state} and
Proposition~\ref{row:prop:stratified-itinerary}\par
\path{normal_step_of_lift}, \path{transport_of_lift},
\path{observable_transport_of_lift} &
The derived recurrence supplies the existing finite additive transport
kernel. &
General varying-dimension transport, differentiability, domain
admission and remainder bounds remain written obligations.\\
\bottomrule
\end{longtable}
\endgroup

\FloatBarrier

\chapter{Notation and conventions}
\label{app:notation}
This appendix gathers the notation and conventions used throughout the
monograph. Definitions are grouped by mathematical role and local setting
so that repeated letters do not imply an identification of different
states, energies, operators or probability laws.

\section{Scope, indices and decorations}
\label{sec:notation-scope}
A symbol has the domain and conditioning stated in the result where it is
used. The entries below collect those meanings rather than impose a single
meaning on a letter throughout the book. In particular, row centering,
context centering and ensemble centering are different operations. Auxiliary
indices, test vectors and constants in a proof are locally quantified;
renaming such a bound variable does not change its meaning.

\begin{description}[style=nextline,leftmargin=1.5em,font=\normalfont,itemsep=3pt]
\item[{$\ell,a,i,j,k,m$}]
Decoder layer, attention head, and locally declared token, row, coordinate,
block or registry indices. In the forward program $\ell=1,\ldots,L$ and
$a=1,\ldots,h$. A subscript $m$ on a registered row map identifies a
context--layer--head combination, rather than an optimizer moment.
\item[{$t,n,T,b$}]
Discrete update or observation index, endpoint index, finite horizon and
block length, respectively, when used as time indices. A continuous $t$ or
$\tau$ is a separately declared physical time; $b$ also denotes a bias or
an affine source in its local equation.
\item[{$N,h,d,r,w,S,V,L$}]
$N$ is the head-count size coordinate in a specified family; $h$ is the
head count in a fixed architecture. The head dimension $d$, matrix row
count $r$, model width $w=Nd$, context length $S$, vocabulary size $V$ and
decoder count $L$ are distinct. A finite horizon also denoted $N$ is
explicitly quantified as such. The dimension $d$ is held fixed in the
head-count limits unless another scaling is stated.
\item[{$s,p,q,B,M$}]
Typical local dimensions or counts: $s$ is the allowed-key count in the
initialization law, $p$ a parameter dimension, $q$ a normal-coordinate
dimension, $B$ a minibatch or block size, and $M$ a corpus size. These
letters also have the function, probability and matrix meanings specified
below; their domains distinguish those uses.
\item[{$X^+,X',X_{t+1},\Delta X_t$}]
Successor or comparison value, and the finite difference
$\Delta X_t=X_{t+1}-X_t$. A prime on a scalar function can instead denote
its derivative. Superscripts $+$ and $-$ on paired paths identify signed
interventions; they do not assert entrywise positivity.
\item[{$\widehat X,\widetilde X,\bar X,\overline X$}]
A hat marks a computed value, estimator, fitted quantity or bias-corrected
moment as specified locally. A tilde marks a comparison, approximation or
factorized evaluation. A bar denotes the stated average or projection;
its averaging population is part of the definition.
\item[{$X_*,X_0,X^{(a)},X_{:k},X_{i:}$}]
A selected reference, initial value, indexed component or source,
column $k$, and row $i$. A star can denote a fixed point or a fitted
reference, and $A^*$ denotes an adjoint where an operator is being discussed.
Subscripts $\mathrm{in},\mathrm{out}$ indicate input and output stages.
\item[{$\mathrm{nat},\mathrm{ref},\mathrm{cal},\mathrm{ev},\mathrm{val}$}]
Native, reference, calibration, evaluation and validation quantities.
Subscripts for a corpus, context panel, training state or numerical law
are part of the conditioning and cannot be dropped when comparing studies.
\item[{$a,b,c,C,K,L$ as bounds}]
Locally quantified constants, comparison factors, curvature or Lipschitz
bounds. Their finiteness, sign and uniformity are exactly those stated in
the corresponding hypothesis; repeated use of a letter does not identify
constants from different results.
\end{description}

\section{Algebra, order and limiting conventions}
\label{sec:notation-algebra}
\begin{description}[style=nextline,leftmargin=1.5em,font=\normalfont,itemsep=3pt]
\item[{$\R,\N,\mathbb Z,\R^{m\times n},\R_+$}]
Real numbers, natural-number indices, integers, real matrices of the stated
shape, and nonnegative reals in the positive affine cone. Restrictions such
as a positive integer or a strictly positive scalar are stated explicitly.
\item[{$0,\bm0,\bm0_{m\times n},\bm0_{\mathrm{mat}}$}]
Scalar zero, zero vector, zero matrix of specified shape, and zero matrix
whose shape is fixed by the operands. A norm, energy or individual array
entry has a scalar zero even when the underlying object is a vector or matrix.
\item[{$0_{\mathrm{op}},0_{\mathrm{fun}},I,I_d,\mathbf1,\bm1$}]
Zero linear operator, identically zero scalar function, identity operator
or matrix, dimension-$d$ identity, and compatible all-ones vectors.
$\mathbf1\mathbf1^{\mathsf T}$ is an all-ones matrix; a superscript
$\mathsf T$ on a vector makes it a row vector.
\item[{$\le,\ge,<,>,\preceq,\succeq,\prec,\succ$}]
The first four relations are scalar or entrywise. The last four denote
positive-semidefinite and positive-definite order for symmetric matrices.
An entrywise positive power base is not thereby a positive-definite matrix.
\item[{$A^{\mathsf T},A^\top,\langle x,y\rangle,\langle A,B\rangle_F$}]
Transpose, Euclidean inner product, and Frobenius inner product
$\langle A,B\rangle_F=\tr(A^{\mathsf T}B)$. Stacked registry inner
products sum the corresponding matrix inner products.
\item[{$\|x\|_2,\|A\|_F,\|A\|_{\mathrm{op}},\|A\|_{\mathrm{HS}}$}]
Euclidean, Frobenius, induced operator and Hilbert--Schmidt norms.
$\|x\|_H=(x^{\mathsf T}Hx)^{1/2}$ is a declared positive-definite
metric norm; an arrow between two metrics specifies the domain and range
norms of an operator. The unadorned norm uses the ambient norm fixed locally.
\item[{$\diag,\tr,\det,\rank,\ker,\range,\operatorname{span},\operatorname{spec},\operatorname{vec}$}]
Diagonal matrix construction, trace, determinant, rank, null space,
range, linear span, spectrum and vectorization. The vectorization order is
fixed within each covariance or derivative comparison.
\item[{$\odot,\oslash,A^{\odot P},\otimes,\bigoplus$}]
Entrywise multiplication, entrywise division, entrywise real powers,
tensor or Kronecker product, and direct sum. An ordinary juxtaposition of
matrices is matrix multiplication. Entrywise logarithms, exponentials,
roots and powers require their stated domains.
\item[{$D f,\mathrm Df,\nabla f,\partial_i f,D^2f,\dot x$}]
Derivative, gradient, partial derivative, second derivative or Hessian, and
time derivative. $Df(x)[v]$ is the derivative applied to direction $v$;
a finite secant is not identified with this derivative without a remainder
bound.
\item[{$f\circ g,f_\#,\mathrm{id},A\perp B,(x)_+$}]
Composition with $g$ acting first, pushforward under $f$, identity map,
orthogonality, and positive part $\max\{x,0\}$. Set union, intersection,
difference and inclusion have their usual meanings, written
$\cup,\cap,\setminus,\subseteq,\subsetneq$.
\item[{$\sum_{j=k}^{n-1},\prod_{j=k}^{n-1},[k,n)$}]
An ordered range of updates including $k$ and excluding $n$. Empty sums
are zero and empty products are the scalar or operator identity. In a
chronological matrix product the later update acts on the left.
\item[{$\log,\exp,\inf,\sup,\min,\max,\lfloor x\rfloor,\lceil x\rceil$}]
Natural logarithm, exponential, infimum, supremum, minimum, maximum,
floor and ceiling. Information quantities consequently use nats unless a
separate conversion is stated. Trigonometric and hyperbolic functions
$\sin,\cos,\tan,\sinh,\cosh,\tanh,\coth$ use their ordinary real
or complex extensions on the stated domains.
\item[{$O(a_N),o(a_N),\Theta(a_N),\asymp,\sim,\simeq,\propto$}]
A bounded multiple of $a_N$, a quantity negligible relative to $a_N$,
two-sided order, two-sided comparison, ratio tending to one, a stated
approximation, and proportionality. Constants and the limiting variable
must be specified. A fitted finite-size power does not establish an
asymptotic equivalence.
\item[{$\to,\Rightarrow,\Longrightarrow,\Longleftrightarrow,\infty$}]
Convergence arrows, logical equivalence and an unbounded limit. The
double arrow denotes convergence in distribution in probability-limit
statements and implication in logical statements. Distributional,
mean-square, uniform and almost-sure convergence are distinguished by the
qualifying text or subscript. A fixed numerical tolerance is not a limit.
\end{description}

\section{Probability, conditioning and information}
\label{sec:notation-probability}
The conditioning law is part of each statistical object; see
Chapters~\ref{ch:forward-program}, \ref{ch:retained-law} and
\ref{ch:single-pass-resource}.
\begin{description}[style=nextline,leftmargin=1.5em,font=\normalfont,itemsep=3pt]
\item[{$\Prob,\Pr,\mathbb P,\E,\mathbb E,\Var,\Cov$}]
Probability, expectation, variance and covariance under the declared law.
A subscript specifies that law or the variable being averaged. For vectors,
$\Cov(X,Y)=\E[(X-\E X)(Y-\E Y)^{\mathsf T}]$; cross covariances
are retained when sources or times are coupled.
\item[{$\Omega,\mathcal F,\mathcal F_t,\mathcal H,\sigma(X)$}]
Sample space, sigma-field, filtration, specified conditioning sigma-field,
and the sigma-field generated by $X$. A conditional expectation or covariance
holds the variables in the conditioning sigma-field fixed.
\item[{$\mu,\nu,\rho,\mathcal L(X),f_\#\mu$}]
Locally specified probability laws, the law of $X$, and its pushforward:
$(f_\#\mu)(A)=\mu(f^{-1}(A))$. The same letters can denote means,
normalization maps or contraction coefficients in other sections.
\item[{$K(s,\cdot),K_t,K^b,K_{k:n},\mu K$}]
Transition kernel from a complete state, a time-indexed kernel, repeated
composition of a homogeneous augmented kernel, chronological composition,
and the successor law. The time and remaining-corpus coordinates needed
for the kernel are retained in the state.
\item[{$\mathcal N(m,\Sigma),\operatorname{Bern}(p)$}]
Gaussian law with mean $m$ and covariance $\Sigma$, and Bernoulli law with
success probability $p$. $\Sigma$ and its indexed variants denote the
covariance matrix specified in the corresponding result.
\item[{$L^2(\Omega),\|X\|_{L^2},\|f\|_r$}]
Square-integrable random variables, root mean square norm
$(\E\|X\|^2)^{1/2}$, and reference-weighted norm
$\|f\|_r^2=\sum_i r_i f_i^2$ for a finite probability vector $r$.
\item[{$H(p),D_{\rm KL}(p\|q),\KL(p\|q)$}]
Shannon entropy and Kullback--Leibler divergence. The finite-vocabulary
divergence is $\sum_i p_i\log(p_i/q_i)$ on its stated support. Entropy
normalized by the logarithm of the allowed-key count is specified separately.
\item[{$\operatorname{TV}(p,q),W_1$}]
Total-variation distance and Wasserstein distance of order one where used.
For probability vectors total variation is one half of the $\ell^1$
distance. A Wasserstein comparison uses the metric stated for its state
or observation space.
\item[{$W(j),\kappa_n,\Gamma,\chi$}]
Log moment generating function, its cumulant tensors, a locally defined
effective action or covariance object, and susceptibility. These are
law-dependent objects; the argument, index and source field distinguish
their occurrences below.
\item[{$\delta_x,\delta_{ij},\mathbf1_A$}]
Point mass at $x$, Kronecker delta and indicator of an event or set.
A scalar $\delta$ without these indices is generally a perturbation,
error bound or distance from a reference, rather than a point mass.
\item[{$\mathrm{iid},\mathrm{q},\mathrm{ann},\mathrm{ctx},\mathrm{doc}$}]
Independent and identically distributed; conditional or quenched;
annealed; context; and document qualifiers, where used. Averaging over a
shared environment is different from conditioning on its realized value.
\end{description}

\section{The architecture and its forward observations}
\label{sec:notation-forward}
These definitions follow Chapter~\ref{ch:forward-program}.
\begin{description}[style=nextline,leftmargin=1.5em,font=\normalfont,itemsep=3pt]
\item[{$x,y,X_0,X_\ell,U_\ell$}]
Input token prefix, external next-token target, embedding array, decoder
boundary state and the intermediate normalized residual state. The target
$y$ is used by the loss, not supplied to the forward computation.
\item[{$Q_{\ell a},K_{\ell a},V_{\ell a},Q_{\rm rope}$}]
Query, key and value arrays for layer $\ell$ and head $a$, and the rotary
query array. $W^Q,W^K,W^V$ and $b^Q,b^K,b^V$ are their learned projection
weights and broadcast biases.
\item[{$\operatorname{RoPE},\LN,\operatorname{FFN},F_\ell$}]
Rotary positional embedding, LayerNorm, decoder feed-forward map and the
complete decoder map. $\epsilon_{\rm LN}$ is the positive LayerNorm
variance offset.
\item[{$D_{\ell a},\Phi_\ell,A_{\ell a},A_{\rm dff}$}]
Query Gram matrix $Q_{\ell a}^{\mathsf T}Q_{\ell a}$, shared metric
learner, its generated matrix, and its hidden width. The row program also
writes $U=Q_{\rm rope}^{\mathsf T}Q_{\rm rope}$ for its Gram input.
\item[{$\Alm,\epsilon_A,\epsilon_{\rm P},P,P^{\mathrm{pot}}$}]
Strictly entrywise positive PLGA base, its positive offset in two local
notations, learned exponent matrix and entrywise powered potential.
A learned entry of $P$ is a parameter, not a critical exponent.
\item[{$W,b,a,b^a,\G$}]
PLGA base weights and bias, output affine weights and bias, and the signed
operator $\G=aP^{\mathrm{pot}}+b^a$. The source partition into generator
and body parameters is specified by the experiment and differs from a
row-space projection.
\item[{$\sigma,\operatorname{silu},\operatorname{SiLU},f,\phi$}]
The logistic sigmoid, the activation $z\sigma(z)$, and the iSwiGLU
nonlinearity $z^2\sigma(z)$ when $f$ or $\phi$ names the PLGA base
activation. SwiGLU is the gated product used in the residual metric units.
These activation meanings are separate from an observation map $\phi$.
\item[{$T_{\ell a},O_{\ell a},O_\ell$}]
Causally masked attention probability matrix, head output $T_{\ell a}V_{\ell a}$,
and projected concatenated attention output. The score uses
$Q_{\ell a}\G^{(\ell a)}K_{\ell a}^{\mathsf T}/\sqrt d$.
\item[{$z_\theta,p_\theta,W_{\rm out},b_{\rm out}$}]
Vocabulary logits, softmax predictive distribution, and vocabulary
projection weight and bias. A proper-prefix conditional is evaluated on
that prefix itself; an earlier position in a longer parallel call is a
different observation when the generated operator depends on the full call.
\item[{$\Xi_\theta,\psi,\Phi_\theta,m_{\theta,\mathcal D},\varphi_\theta$}]
Complete forward trace, a measurable observation of that trace, the observed
vector, its mean under the data law, and the centered observation
$\varphi_\theta=\Phi_\theta-m_{\theta,\mathcal D}$.
\item[{$\mu_{\theta,\mathcal D},j,W_{\theta,\mathcal D}(j)$}]
Joint law of the centered trace observation, its conjugate source vector,
and $\log\E_{\mathcal D}\exp(j^{\mathsf T}\varphi_\theta)$.
A product of marginal head laws is a separate approximation to this joint law.
\end{description}

\section{Row geometry and numerical observation}
\label{sec:notation-row}
See Chapters~\ref{ch:row-geometry} and \ref{ch:finite-work} for the
implemented geometry and its finite balances.
\begin{description}[style=nextline,leftmargin=1.5em,font=\normalfont,itemsep=3pt]
\item[{$P_d,c(x),s(x),\nu(x),\mathcal N_{\gamma,\beta}$}]
Coordinate-centering projector $I_d-d^{-1}\mathbf1\mathbf1^{\mathsf T}$,
centered row input, regularized root variance, normalized row and affine
LayerNorm. Here $c(x)=P_dx$, $s(x)^2=\epsilon_{\rm LN}+d^{-1}\|c(x)\|^2$,
and $\mathcal N_{\gamma,\beta}(x)=\gamma\odot\nu(x)+\beta$.
\item[{$G(z;\psi),G_{j,1},G_{j,2},w_j,x_j$}]
Gated row block with its parameter collection $\psi$, the two blocks of
residual unit $j$, the residual sum before normalization, and the output
row after that unit. $G$ here is a function, not the signed PLGA operator.
\item[{$X,Y,Y_c,\gamma,\beta,\Gamma$}]
Implemented physical row map, normalized rows before the final affine
LayerNorm, their row-centered version, shared affine gate and bias, and
$\Gamma=\diag(\gamma)$. The factorization is
$X=Y\diag(\gamma)+\mathbf1\beta^{\mathsf T}$.
\item[{$C_r,\rowq,\Pi,Z,J(X),E(X),D(X)$}]
Row-centering projector $C_r=I_r-r^{-1}\mathbf1\mathbf1^{\mathsf T}$,
its operator notation $\rowq$, the corresponding locally dimensioned
projector $\Pi$, centered map $Z=C_rX$, quotient norm $J(X)=\|Z\|_F$,
absolute quotient energy $E(X)=J(X)^2$, and maximum pairwise row distance
$D(X)$. This $D(X)$ is not a finite increment.
\item[{$\bar X,\bar x,\upsilon_k,e_k,S_{m,t},a_{m,t}$}]
Constant-row projection $(I-C_r)X=\mathbf1\bar x^{\mathsf T}$, its common
row, coordinate shape energy $\upsilon_k=\|Y_{c,:k}\|^2$, physical
coordinate energy $e_k=\gamma_k^2\upsilon_k$, total centered shape energy
$S_{m,t}=\sum_k\upsilon_{m,k,t}$ and effective squared gate
$a_{m,t}=E_{m,t}/S_{m,t}$ when its denominator is positive.
\item[{$\mathcal M,\mathcal Y,E_t^{\mathcal M}$}]
Finite registry of observed row maps, direct-sum Frobenius space over that
registry, and sum of their absolute quotient energies. A finite registry
does not by itself provide uniform control over unseen contexts.
\item[{$\tau_j,M_j,A_j,B_j,\mathfrak r_t$}]
Block endpoints, maximum energy in a block, excursion above its left anchor,
positive energy variation in the block, and one-step positive reopening
$(E_{t+1}-E_t)_+$. These are absolute-energy quantities.
\item[{$\mathcal F_0,\mathcal C_{\epsilon_*},\mathcal R_{\epsilon_*},\epsilon_*$}]
Exact row-constant face, positive energies at or below a declared effect
floor, resolved positive-energy stratum, and the floor. The middle stratum
contains positive states and is not an exact face.
\item[{$\widehat E,\delta,I(\widehat E,\delta),E_-,E_+,L,U$}]
Computed energy, a certified absolute error when one is assumed, and its
nonnegative enclosure. $E_-,E_+$ or $L,U$ are the enclosure endpoints.
An empirical replay discrepancy is not automatically such a certificate.
\item[{$E^{32},E^{64},\delta_{\rm arith}$}]
Float32 and float64 energy evaluations and the specified arithmetic
resolution floor, $8|E^{64}-E^{32}|$ in the row replay protocol. Numerical
precision, bitwise equality and analytical equality have different scopes.
\end{description}

\section{Finite work, source increments and downstream defects}
\label{sec:notation-work}
The work and PLGA definitions are developed in
Chapters~\ref{ch:finite-work} and \ref{ch:plga-emission}.
\begin{description}[style=nextline,leftmargin=1.5em,font=\normalfont,itemsep=3pt]
\item[{$D_t,\mathcal W_t,\mathcal Q_t$}]
Executed increment $Z_{t+1}-Z_t$, inward work
$-2\langle Z_t,D_t\rangle_F$, and finite-step charge $\|D_t\|_F^2$.
The exact energy change is $-\mathcal W_t+\mathcal Q_t$.
\item[{$D_t^{(a)},\mathcal A,\mathcal A_4$}]
Declared source increments, their index set, and the four native source
labels $\{\mathrm{shape},\mathrm{gate},\mathrm{int},\mathrm{def}\}$.
Their cross-Gram inner products contribute to the charge.
\item[{$U_t,\widetilde Z_t,\delta_t$}]
Centered pre-affine shape $C_rY_t$, factorized evaluation $U_t\Gamma_t$,
and native implementation defect $Z_t^{\rm nat}-\widetilde Z_t$.
The defect is an arithmetic source in the native finite ledger.
\item[{$D^{\rm shape},D^{\rm gate},D^{\rm int},D^{\rm def}$}]
Shape increment at the old gate, gate increment at the old shape,
bilinear shape--gate increment, and change of the implementation defect.
Their exact sum is the native increment.
\item[{$\Psi_{m,t},\Delta\theta^{\rm wd},\Delta\theta^{\rm loss}$}]
Centered implemented row observation as a function of parameters at a fixed
probe, decoupled weight-decay displacement, and the remaining AdamW
loss-driven displacement. Integrating $D\Psi$ on the declared parameter
segment assigns endpoint-exact source secants.
\item[{$D^N,D^C,H_t,I_t$}]
Natural and controlled increments, their difference $H_t=D^C-D^N$, and
the paired energy contrast. Its state projection, natural-increment
interaction and intervention charge are all retained.
\item[{$\alpha_t,T_t,\mathcal T_t,\tau_t^2,g_t^2$}]
Inward radial coefficient $-\langle Z_t,D_t\rangle/E_t$, tangential
finite residual $D_t+\alpha_tZ_t$ in two notations, its relative squared
norm, and squared norm gain $E_{t+1}/E_t$. These ratios require $E_t>0$.
\item[{$\alpha_t^{(a)},\mathcal T_t^{(a)},\beta_t,\mathcal S_t$}]
Source-resolved radial coefficient and tangential residual, followed by the
radial coefficient and tangential residual of a paired intervention.
Here $\beta_t$ is not an Adam moment-decay coefficient.
\item[{$\varrho_t,b_t,P_E(n,k)$}]
Nonnegative energy-comparison gain and source, and chronological scalar
product $\prod_{j=k}^{n-1}\varrho_j$. They appear in a proved inequality
for energy, rather than a definition of the complete optimizer state.
\item[{$B(X),P(X),\mathcal P(X),p,b_a$}]
PLGA positive base, powered intermediate, complete downstream map, learned
exponent matrix and final affine bias in the row-transfer notation.
These correspond to the base, potential and signed operator stages of the
forward notation.
\item[{$\delta(\bar X),r(X;\bar X),\kappa(X)$}]
Constant-input PLGA quotient defect, response along the occupied quotient
segment, and supremum of the quotient derivative norm on that segment.
A measured secant ratio is distinct from the independently bounded supremum.
\item[{$\kappa_W,\pi,q_{\rm PLGA},\zeta_{\rm PLGA}$}]
Common row sum of $W$, common exponent row written as a column vector,
and gain and source in the positive affine downstream energy bound.
The vector $\pi$ in the structural PLGA criterion is not a state projection.
\item[{$e_t,T_t,u_t,L_\ell,\varepsilon_\ell,L_{\rm out},B$}]
In the geometric-to-predictive bridge, centered energy, total matrix energy,
their ratio, decoder Lipschitz factor, replacement defect, output Lipschitz
factor and accumulated logit-error bound. Total energy $T_t$ in this bridge
is distinct from the tangential increment also written $T_t$ above.
\end{description}

\section{Complete optimizer states and their response}
\label{sec:notation-optimizer}
Chapter~\ref{ch:complete-adamw} fixes the row-response notation;
Chapters~\ref{ch:retained-law} and \ref{ch:metric-reduction} retain the
additional resource and scaling coordinates.
\begin{description}[style=nextline,leftmargin=1.5em,font=\normalfont,itemsep=3pt]
\item[{$\theta_t,m_t,v_t,s_t,S_t$}]
Parameters, Adam first moment, Adam second moment, continuous optimizer
state $s_t=(\theta_t,m_t,v_t)$, and complete augmented training state.
The latter also retains schedule and bias clocks, remaining data and
stochastic or numerical program state.
\item[{$\omega_t,\mathcal F_t,g,c,C,K,K_c$}]
Declared update environment, deterministic successor conditional on that
environment, raw minibatch gradient, clipped gradient, clipping derivative,
loss-gradient derivative and $K_c=CK$. The componentwise clipping cell of
the row derivative and the global-norm clipping used in a declared training
protocol have their respective derivatives.
\item[{$\eta,\beta_1,\beta_2,\lambda_{\rm wd},\Lambda,d$}]
Learning rate, first- and second-moment decay coefficients, decoupled
weight-decay coefficient, diagonal decay operator and its binary mask in
the optimizer formula. This mask $d$ is distinct from head dimension.
\item[{$\tau,a_1,a_2,\widehat m^+,\widehat v^+,\epsilon_{\rm A}$}]
Live optimizer step, bias corrections $a_i=1-\beta_i^\tau$,
bias-corrected successor moments and positive Adam denominator offset.
The offset $\epsilon_{\rm A}$ is separate from the PLGA base offset.
\item[{$M_\theta,V_\theta,D_m,D_v,\mathcal A_t$}]
Derivatives of the moment updates with respect to parameters, derivatives
of the adaptive ratio with respect to its two moments, and the lifted
Jacobian on $(\theta,m,v)$. The live denominator contributes the $D_v$
blocks.
\item[{$\mathcal X_t,\mathscr S_{\ell,t},U_t,\mathcal V_t,q_t$}]
Complete-state space, row-constant stratum, normal-coordinate domain,
chart neighborhood and its normal dimension. The chart dimension $q_t$
is distinct from the canonical scalar-energy gain.
\item[{$\Psi_t,V_t^{\rm tan},a_t,\iota_t,\pi_t,T_t,z_t$}]
Full local chart, tangential domain and realized tangential coordinate;
normal insertion $\iota_t(z)=\Psi_t(a_t,z)$, normal extraction,
charted successor $\pi_{t+1}\circ\mathcal F_t\circ\iota_t$, and realized
normal coordinate. The admitted trajectory satisfies both
$\pi_t\iota_t=\mathrm{id}$ and $s_t=\iota_t(z_t)$.
The slice is fixed during differentiation, but can depend on complete
history; it does not imply predictive closure
(Section~\ref{row:sec:trajectory-chart}). The scalar $a_t$ in the worked
row example has a separate local meaning.
\item[{$\bar\iota_t,\bar T_t,e_t,d_t$}]
Prescribed slice, its charted successor, realized successor mismatch and
physical observation mismatch. Their exact definitions are
Equations~\eqref{row:eq:fixed-slice-defect} and
\eqref{row:eq:fixed-slice-observation-defect}; neither is automatically a
quadratic remainder.
\item[{$f_t,A_t,R_t(z),\Phi(n,k)$}]
Invariance defect $T_t(\bm0)$, homogeneous derivative or declared compatible
linear map, exact nonlinear remainder, and chronological propagator
$A_{n-1}\cdots A_k$. A derivative interpretation requires the selected
smooth cell; the algebraic remainder identity does not.
\item[{$h_t,y_t,o_t,O_t,Q_t(z),\mathcal H_t$}]
Normal-coordinate observation, its value, value at the chart origin,
linearized observation, remaining observation nonlinearity, and the
complete-state observation map. $Q_t(z)$ here is not a query matrix.
\item[{$s_t^0,s_t^\pm,\Delta_t,\rho_t,Y_n,L_n,N_n,Q_n$}]
Base and paired perturbed trajectories, their half-difference,
departure from declared linear state transport, observed half-difference,
homogeneous observed response, transported state correction and observation
correction. Their exact finite balance is $Y_n=L_n+N_n+Q_n$.
\item[{$\chi_n,c_{a,n}$}]
Cancellation index formed from the sum of component norms divided by the
resolved total norm, and the individual terms of the observable cocycle.
This $\chi_n$ is not a count susceptibility.
\item[{$\mathsf H_t,h_-,h_+,r_t,C_t,F_t,\mathfrak a_t,\mathfrak q_t$}]
Scheduled positive-definite metric, lower and optional upper comparisons,
tube radius, quadratic remainder bound, forcing norm, homogeneous operator
norm between scheduled metrics and effective tube gain
$\mathfrak q_t=\mathfrak a_t+C_tr_t$.
\item[{$P_{\rm tube}(n,k),b_{m,n},g_{m,n},G$}]
Product of effective tube gains, additive physical-cover defect,
cover coefficient and its uniform finite bound. The bound $G$ in this
cover is a scalar constant, not the PLGA operator.
\item[{$B(t),b(t),r(t,z),U(t,s),\xi_t,\mu_n,V_n,B_n$}]
Continuous homogeneous generator, forcing, remainder and fundamental
propagator; then discrete martingale forcing, exact propagated mean,
transported second-moment contribution and its norm-based upper envelope.
The scalar $V_n$ here is not the value-projection array.
\end{description}

\section{Chronological affine maps and closure}
\label{sec:notation-affine}
These objects belong to Chapters~\ref{ch:affine-blocking}--\ref{ch:closed-states}.
\begin{description}[style=nextline,leftmargin=1.5em,font=\normalfont,itemsep=3pt]
\item[{$\edge_t,(q_t,\zeta_t),\idEdge,\mathcal C,\mathcal A_+$}]
Canonical affine edge, its gain and restart source, identity edge $(1,0)$,
realized cone $q\zeta=0$ with nonnegative coordinates, and full nonnegative
comparison cone. On a positive source $q_t=E_{t+1}/E_t$ and $\zeta_t=0$;
on the exact face $q_t=0$ and $\zeta_t=E_{t+1}$.
\item[{$Q(n,k),Z(n,k),\RG_b$}]
Chronological product of scalar gains, convolution of transported sources,
and blocking by scale $b$. The affine action is
$E_n=Q(n,k)E_k+Z(n,k)$; the right factor in a composition acts first.
\item[{$\bar q_t,\bar\zeta_t,\mathbf A=(Q,Z),\odot$}]
Proved comparison gain and source, interval-valued affine edge, and its
interval action or composition. This interval use of $\odot$ is defined
by endpoint arithmetic and is separate from entrywise matrix multiplication.
Underlines and overlines mark lower and upper enclosure endpoints.
\item[{$s_t,\widehat E_t,\mathcal G_{s_t,s_{t+1}}$}]
Positive change of energy units, rescaled energy $E_t/s_t$, and the
induced affine gauge map. This gauge changes energy coordinates and is
separate from a parameter symmetry of the optimizer.
\item[{$\bm E_t,K_t,\bm h_t$}]
Vector of registry energies, entrywise nonnegative comparison matrix and
nonnegative source vector in the coupled registry inequality. These are
comparison coordinates, not a claim that the energy vector closes as a state.
\item[{$\ell,\beta_q,\beta_\zeta,\tau,u,E_*,\chi_\zeta$}]
Log blocking scale $\ell=\log b$, affine beta functions, distance
$\tau=1-q$ from unit gain, invariant coordinate $u=\zeta/(1-q)$,
subcritical stationary comparison energy and response to its source.
Here $\ell$ is a scale coordinate rather than a decoder index.
\item[{$y_\tau,y_\zeta,\phi,\xi_\parallel,\nu_\parallel$}]
Linearized affine scaling dimensions, their crossover ratio,
longitudinal scale $1/|\log q|$ and its declared exponent near $q=1$.
These symbols describe the specified clean comparison model.
\item[{$Y_t,S_{n,k},\lambda_k$}]
Positive-branch log gain $\log q_t$, cumulative log gain and its limiting
mean rate when that limit exists. A zero rate does not determine the
sublinear cumulative behavior.
\item[{$\sigma_j,r_j,P_j,M_j,\rho_j,a_j,\chi_{\rm exc}$}]
Successive face times, restart energy, excursion peak, peak amplification,
negative log restart, log amplification and limiting ratio
$a_j/\rho_j$ when defined. Log marks are used only for positive restarts.
\item[{$\mathsf X,\mathcal X,\mathsf E,\mathcal E,\mathsf A,e,\bar K_b,D_b$}]
Complete and observed standard Borel spaces with their sigma-fields,
proposed summary-state space, observation map, reduced $b$-step kernel
and same-fiber successor defect. A candidate state map
$a:\mathsf X\to\mathsf A$ and readout $h:\mathsf A\to\mathsf E$
make the intermediate summary explicit. A reduced kernel valid for every initial law requires equal
pushed-forward successors throughout each relevant fiber.
\item[{$G,g,C_g,P_g,M_g,[m,v]_G$}]
Parameter symmetry group, a group action, its consistent action on parameters
and first moments, parameter-only action, moment-only action, and a chosen
gauge-canonical optimizer coordinate. Second moments are unchanged by the
sign action considered here.
\item[{$R_P,a_P,\phi$}]
Row maps observed on a fixed probe $P$, candidate reduced state combining
that probe with optimizer information, and the retained exogenous phase.
A fixed finite probe may omit dynamically active model coordinates.
\item[{$\mathcal T_b,\mathcal T_{k,b},\mu,\sigma^2,\sigma_{\rm eff}^2,h$}]
Standardized log-gain law under blocking, its origin-dependent version,
mean log gain, one-step variance, long-run variance and standardized drift
$h=\mu/\sigma$. Here $h$ is neither head count nor a physical time step.
\item[{$m_t,\varepsilon_t,M_{k,b},\Xi_{k,b},a_{k,b},c_{k,b}$}]
Nonautonomous log-gain drift, centered fluctuation, their block sums, and
the declared block centering and positive normalization. The drift $m_t$
is distinct from the Adam first moment.
\item[{$\xi_{\rm dis},\nu_{\rm dis},H,L(b),\alpha$}]
Disorder crossover scale and exponent, fluctuation-sum exponent,
subpower normalization factor, and stable-tail index where a heavy-tailed
class is specified. A logarithmic correction in $L(b)$ is retained;
it cannot be replaced silently by a bounded constant.
\end{description}

\section{Complete laws, emissions and the consuming corpus}
\label{sec:notation-laws}
See Chapters~\ref{ch:retained-law} and \ref{ch:single-pass-resource}.
\begin{description}[style=nextline,leftmargin=1.5em,font=\normalfont,itemsep=3pt]
\item[{$\mathcal D_{\rm tr},\mathcal D_{\rm ev},\mathcal C_M,\mathcal C_N,\mathbf P_N^{\mathcal D}$}]
Training and evaluation laws, a realized corpus of $M$ indexed blocks,
a declared size-family conditioning specification, and the complete joint
path, evaluation and emission law. In $\mathbf P_N^{\mathcal D}$,
$s_{0:T}$ is the complete-state path, $x=(x^-,y_{\rm target})$ the
evaluation example and $u$ the emitted observation; $ds_{0:T},dx,du$
indicate their integration variables. The initial law is
$\rho_{N,0}^{\mathcal D}$, the terminal evaluation kernel is
$\mu_{N,T}(dx\mid s_T)$, and the emission kernel is
$\mathcal E_{N,T}(du\mid s_T,x^-)$, excluding the target. A corpus
object and a conditioning specification are distinguished where the letter
$\mathcal C$ is used.
\item[{$S_t,\sigma_t,R_t,\mathcal R_t,\xi_t,U_t$}]
Augmented state, schedule and bias clocks, remaining block identities in
two notations, sampler or stochastic-program state, and the native complete
update. In an alternative state notation $o_t$ collects optimizer variables
and $r_t$ the sampler state; in resource formulas $r_t=|R_t|$ is a count.
\item[{$\rho_t,\mathcal E,\mu_t,\zeta_t$}]
Training-state law, inference emission kernel, emitted law
$\mu_t=\rho_t\mathcal E$, and a training innovation in the update
$U_t(s_t,\zeta_t)$. This innovation $\zeta_t$ is distinct from the affine
face-restart energy.
\item[{$K_{t:t+m},T_{t,j},a_t$}]
Chronological training kernel product, source-weighted transfer kernel,
and retained training observation. Products written
$\rho_tK_tK_{t+1}\cdots$ propagate a law in time order from the left;
functional or matrix compositions retain their separately stated action order.
\item[{$(F,w),\mathcal T_{F,w},B_{b,C,H},\RG_{b,C,H}$}]
Decoder boundary map with its internal source insertion, weighted transfer
operator, blocked linear observation and its pushforward law. The blocking
map is $b^{-H}C\sum_{i=1}^b\varphi_i$ and retains the joint input law.
\item[{$c,P,\bar K_t,\bar{\mathcal E},Q,d_Q$}]
Retained-state map, its deterministic pushforward kernel, proposed reduced
transition and emission, reduced state space and its metric.
Here $P$ is a kernel, not a learned exponent matrix.
\item[{$\delta_t,\varepsilon_E,L_t,L_E,d_t,e_0,a_m$}]
Uniform successor defect, emission defect, transition and emission
Lipschitz constants, evolving-law averaged successor defect, initial-law
discrepancy and terminal averaged emission defect. The averaging and metric
are part of each closure bound.
\item[{$q_t,\pi_t,\widetilde\pi_{t+1},P_s$}]
Retained coordinate $c(S_t)$, conditional complete-state law given its
observed history, predicted conditional law $\pi_tK_t$, and
state-conditioned evaluation-context kernel. This filtering law $\pi_t$
is distinct from a chart extraction map or a remaining-data law.
\item[{$M,B,T,R_t,r_t,(M)_K,\pi_t^{(B)}$}]
Resource size, batch size, update horizon, remaining identities, their
count, falling factorial $M(M-1)\cdots(M-K+1)$, and uniform ordered
next-batch law. A single pass requires $BT\le M$ and removes consumed
identities from the state.
\item[{$h_i,\bar h,\widehat h_B,\Sigma_R$}]
Frozen-state population vectors, their population mean, mean of a batch
sampled without replacement, and population covariance with divisor $r$.
The batch covariance factor is $(r-B)/(B(r-1))$ when $r>1$.
\item[{$n_i,\bar n,n_*,g_*,a_i,\Sigma_a,g_{\mathcal B}$}]
Valid-target counts, their mean and admitted positive batch-mean lower
bound; population gradient ratio, centered gradient-count residual,
its covariance and masked batch gradient. Unequal valid-target counts make
this a ratio problem rather than an unweighted block mean.
\item[{$\mathcal J,p_j,H_j,H,V$}]
Indexed supervised source positions, probability that a crop supervises
position $j$, its exposure count, total target events and number of distinct
supervised positions. Here $V$ is an occupancy count, not vocabulary size.
\item[{$q_N(t),M_N,B_N,T_N,\mu_D,\beta_B,z_T$}]
Consumed fraction $B_Nt/M_N$, size-dependent resource, batch and horizon,
and their declared resource, batch and horizon scaling exponents. These
are imposed family coordinates, not automatically critical exponents.
\item[{$Y_t,A_t,Z,\Sigma,T_{\rm eff},\mathcal F_q$}]
Frozen-population batch mean, deterministic transport weight, transported
centered sum, population covariance, effective response length
$(\sum_t w_t)^2/\sum_t w_t^2$, and finite-resource scaling function.
These $A_t$ are transport matrices, rather than necessarily native Jacobians.
\item[{$\mu_{\mathcal C},\Sigma_{\mathcal C},\bar\mu,V_C,V_\epsilon,Q$}]
Corpus mean and covariance, distribution-averaged mean, common-document
and block-residual covariance, and the induced covariance between distinct
sampled positions in the stated document model. Corpus and permutation
averaging are separate operations.
\item[{$L,\ell_a,R,g_*,b,V,e_y$}]
Locally specified loss or risk, Bernoulli cross-entropy with target
frequency $a$, observed cohort risk, a reference gradient, conditional mean
and covariance of a native displacement, and the unit vector of target $y$.
Training risk, external-target risk and native-law fidelity are separate
quantities even when all use logarithmic losses.
\end{description}

\section{Potential increments, response and fluctuation sectors}
\label{sec:notation-response}
See Chapter~\ref{ch:predictive-transport}.
\begin{description}[style=nextline,leftmargin=1.5em,font=\normalfont,itemsep=3pt]
\item[{$M_t,P_t,V_t,q_t,b_t$}]
Positive power base, learned exponent, entrywise potential $V_t=M_t^{P_t}$,
log potential $q_t=\log V_t$ and log base $b_t=\log M_t$.
Here $q_t$ is a matrix, distinct from a retained state or consumed fraction.
\item[{$u_t,v_t,u_*,v_*,c_t,Q_I$}]
Exponent and base contributions to a log-potential increment, their
symmetric allocations, cross increment $c_t=\Delta P_t\odot\Delta b_t$,
and accumulated log increment on the time interval $I$.
\item[{$d_t,e_t,\mathcal A_I,\star,U_b,V_p$}]
Exponent and log-base increments, their signed chronological area,
its concatenation product, and the corresponding base/exponent path
readouts. The order of increments is part of these path coordinates.
\item[{$\alpha_t,r_t,A_I$}]
Declared physical step, log-potential speed $\|\Delta q_t\|/\alpha_t$,
and elapsed physical time $\sum_{t\in I}\alpha_t$.
\item[{$j,\eta,J_\ell,B_\ell,\mathcal J_x,\Hess_x$}]
Statistical source, instrumental forward perturbation, boundary Jacobian,
source Jacobian, complete source-to-logit Jacobian and the pulled-back
categorical Fisher form $\Hess_x=\mathcal J_x^{\mathsf T}F(p_x)\mathcal J_x$.
The source $\eta$ is distinct from an optimizer rate.
\item[{$A(z),F(p),v_x,q_x,w_x,k_x(h)$}]
Log-sum-exp, categorical Fisher matrix
$F(p)=\diag(p)-pp^{\mathsf T}$, directional logit derivative,
its Fisher quadratic form, second directional logit derivative, and
finite predictive KL along that parameter direction.
\item[{$\bar F_j,K_j,Q,r,a_\pm(r),\eta(r),T_m$}]
Integrated Fisher curvature, finite KL, half the reference-weighted logit
variance, logit-increment range, upper/lower curvature factors,
a relative quadratic-error bound, and the refined variance quadrature.
The scalar $Q$ here is distinct from a query matrix or forcing covariance.
\item[{$m_s,C_s,\bar m_t,A_t,B_t$}]
Conditional context mean and covariance at one training state, mean
across training states, mean within-state covariance and covariance of
state means. Their sum is the total covariance under the specified joint law.
\item[{$Y_i,m_S,a_S(U),\epsilon_i,B_{\rm tr},B_{\rm doc},A_\epsilon,\Gamma(r)$}]
Observed field, training-state mean, document effect, residual,
training and document covariance sectors, residual variance and lag-$r$
residual covariance. The conditioning determines which sectors remain random.
\item[{$I_{\rm ctx},\Sigma_p$}]
Mean predictive KL between two independent contexts at fixed weights,
and covariance of the vocabulary probability vector under that context law.
\item[{$Y=b+AX+e,\Sigma_X,\Sigma_e,C_{Xe}$}]
Affine observation with residual, input and residual covariances, and their
cross covariance. A covariance error budget retains both signed cross terms.
\item[{$\Phi_{t,j},U,L_t,d_t,\mu_U,C_U,H_j,Q_{ij}$}]
Chronological response product, stacked initial state and innovations,
its observation matrix and deterministic drift, stacked mean and covariance,
initial-state/innovation covariance and cross-time innovation covariance.
\item[{$C_{XX},C_{XY},C_{YY},K,S$}]
Blocks of a joint covariance, linear conditional predictor coefficient
$K=C_{YX}C_{XX}^{-1}$ and Schur residual covariance
$S=C_{YY}-C_{YX}C_{XX}^{-1}C_{XY}$. Gaussianity is a separate assumption.
\end{description}

\section{Shared collectives, sign symmetry and physical clocks}
\label{sec:notation-clocks}
See Chapters~\ref{ch:shared-collectives}--\ref{ch:physical-clocks}.
\begin{description}[style=nextline,leftmargin=1.5em,font=\normalfont,itemsep=3pt]
\item[{$\mu_{N,T,g},g_N,r,\tau_*,a_N,\nu_N$}]
Conditioned trained-state law, width-dependent generator rate multiplier,
fixed relative-rate coordinate, physical horizon, shared parameter path
and empirical head-observation measure. In this family $g_N=2r/N$;
$a_N$ here is a vector path, not a scalar noise amplitude.
\item[{$\zeta_{\ell a},\Gamma_N,\pi_N,\Phi_N,F_N$}]
Native head signs, sign group $(\mathbb Z/2\mathbb Z)^{LN}$,
orbit quotient, equivariant training update and invariant predictive
emission. Sign changes act jointly on the specified query and PLGA factors.
\item[{$\chi_N^G,B_G,\phi(p),H^2(p,p'),\chi_N^\phi$}]
Signed-operator susceptibility, uniform operator envelope,
Hellinger embedding $\phi(p)=2\sqrt p$, squared Hellinger distance
$H^2=\tfrac12\sum_i(\sqrt{p_i}-\sqrt{p_i'})^2$, and predictive susceptibility
$N\tr\Cov(\phi(p))$. The operator statistic uses its declared $d^2$
normalization; the predictive statistic has no vocabulary divisor.
\item[{$c_{\ell a},h_N,u_N,s_N,d_N$}]
Common row centroid, first clipped shared gradient, its normalized Adam
force, coordinate signs and their mean discrepancy in the shared-motion
calculation. Elsewhere $h_N$ is a physical time step.
\item[{$q,b_i,K_{ij},w_i,\widehat V$}]
Decay multiplier, shared-path block residual, its per-coordinate
cross covariance, endpoint decay weight and the resulting variance
reconstruction. Here $q=1-\eta\lambda$, rather than consumed fraction.
\item[{$Z,p,m_z(x),V_z(x),r(u,\xi),u_*,u_\pm,g_*,w_{\rm half}$}]
Whole-realization kinetic label, its probability, conditional means and
variances, scalar kinetic field with random environment, peak and
half-height clock positions, induced rate peak and half-height width.
The kinetic exponent in $g^pT$ is a separate local use of $p$.
\item[{$h_N,\gamma_i,\beta_i(N),C_h,\bar C,C_\beta$}]
Physical optimizer step, positive memory decay rates, matched factors
$\beta_i(N)=e^{-\gamma_i h_N}$, finite matched-memory force envelope,
its uniform bound and the fixed-memory envelope. The two envelopes have
different definitions and hypotheses.
\item[{$\kappa_N,\lambda_N,\sigma_N^2,a_N,\tau_{\rm updates},\tau_{\rm physical}$}]
Restoring rate, multiplier $e^{-\kappa_Nh_N}$, forcing intensity,
one-step innovation variance and relaxation times in updates and
physical units. The physical clock separates a small step from a small
restoring rate.
\item[{$a,b,v,\zeta,\upsilon$}]
Clock, restoring-rate and physical-noise powers, update-gap exponent
$\zeta=a+b$ and innovation exponent $\upsilon=a+v$ in the specified
power-law reference family. These letters have other local uses.
\item[{$D_k,\eta_k,Z_T,C,m,\tau,F(s)$}]
Deterministic transport, without-replacement centered batch mean,
transported sum, finite-population covariance, limiting scaled corpus
size, physical horizon and continuous transport kernel in the consuming
bridge calculation.
\item[{$A,D,\Pi,E,E',u(A),\ell(A,D),q(A,D)$}]
Metric matrix, actual finite matrix increment, row-centering projector,
predecessor and successor total squared norms, centered-energy fraction,
and its exact cross and quadratic fluxes. Both fluxes use $E'$ in their
denominator. The data field \texttt{linear\_flux} stores the stated average
of these individual $\ell$ values, as explained in
Section~\ref{model:sec:collective-clock} and measured in
Section~\ref{model:sec:collective-flux-results}.
\item[{$a,b,c,\widehat f,\varepsilon$}]
Predecessor-normalized cross and quadratic numerators, relative total-energy
change, forecast $(\widehat a+\widehat b)/(1+\widehat c)$ and positive
lower bound for the true denominator $1+c$.
\item[{$C_i,r_{ij},q_{ij},\star$}]
Centered row energy, total-energy ratio, normalized centered-energy
increment and chronological energy-coordinate composition. Here
$r_{ij}=E_j/E_i$, $q_{ij}=(C_j-C_i)/E_i$ and
$u_j=(u_i+q_{ij})/r_{ij}$; this $q$ includes more than the one-step
quadratic flux.
\end{description}

\section{Predictive resolution, operator caches and risk}
\label{sec:notation-resolution}
See Chapters~\ref{ch:predictive-resolution}--\ref{ch:cache-evidence}.
\begin{description}[style=nextline,leftmargin=1.5em,font=\normalfont,itemsep=3pt]
\item[{$r,I_k,J_k,K_k,R_k,\Pi_k,K_{k\leftarrow\ell}$}]
Positive vocabulary reference law, retained and tail token sets,
elimination, reference-conditional lift, reconstruction $\Pi_k=R_kK_k$,
and map between retained resolutions. Here $R_k$ is a lift, not a risk.
\item[{$\mathcal P,\mathcal Q,B(i),f,A_{\mathcal P},\|\cdot\|_r$}]
Vocabulary partitions, the block containing token $i$, density $f_i=p_i/r_i$,
reference-conditional density projection and reference-weighted norm.
The corresponding probability projection is $\Pi_{\mathcal P}$.
\item[{$v_k,v_p,e_k,U_k,B_k,V_{e,k},\overline D_k$}]
Retained and full Hellinger variance traces, reconstruction residual,
its empirical mean squared norm and squared mean, unbiased centered
residual variance and mean KL reconstruction error. The empirical
variance uses the replica divisor $S-1$.
\item[{$V_{p,c},V_{e,c},R_c,R_\mu,w_c$}]
Native and residual variance at context $c$, contextwise and aggregated
relative root variance errors, and variance-weighted context mass
$w_c=\mu_cV_{p,c}/\sum_d\mu_dV_{p,d}$. These ratios require positive
native variance on their stated domain.
\item[{$\overline G_{\ell h},\widehat G,c,a_j,X_j$}]
State-specific cached operator, estimated cache, fixed comparison cache,
calibration weights and proper-prefix calibration inputs. Caches are
formed separately at each trained state unless explicitly pooled.
\item[{$L_\ell,d_\ell,D,A_\ell,b_\ell$}]
Decoder Lipschitz constant, same-input substitution defect, propagated
logit-error bound, operator-to-decoder sensitivity and its downstream
weight. Their validity is restricted to the compared domain.
\item[{$m_\mu,V_\mu,m_\nu,V_\nu,\bar G_w,R_t(c_s),\Delta,a$}]
Evaluation and calibration operator means and scatters, empirical panel
mean, risk of reusing an earlier cache, mean-state displacement and
earlier calibration displacement. The risk budget retains their signed
cross term.
\item[{$p,\widehat p,r,V_x,V_{\widehat x},V_e$}]
Native, approximated and external-target laws, and the variance traces
of their native, retained and residual Hellinger observations.
External-target NLL difference and native-law KL fidelity are different
quantities.
\end{description}

\section{Finite interventions, autoregression and adaptation}
\label{sec:notation-adaptation}
See Chapters~\ref{ch:finite-adaptation}--\ref{ch:adaptation-evidence}.
\begin{description}[style=nextline,leftmargin=1.5em,font=\normalfont,itemsep=3pt]
\item[{$F_t(h,\omega),X,O_h,E_h,B$}]
Native observation after a parameter pulse $hv$ with coupled source order
$\omega$, unperturbed observation, odd and even pulse components, and a
linear path-observation map. The odd component is a half difference,
without division by $h$.
\item[{$M,V,a_j,D_i,u,w,\epsilon_o,\lambda_d$}]
Updated first and second Adam moments, bias corrections
$a_j=1-\beta_j^{k+1}$, corrected denominator, incoming moment displacements,
positive denominator offset and decoupled decay coefficient in the finite
optimizer intervention. Here $V$ is a moment vector.
\item[{$z_j^F,z_j^P,p_j^F,p_j^P,v_j,\omega_j$}]
Full-window and proper-prefix logits and laws, their logit difference
and its coordinate range. Superscripts $F,P$ specify the input available
at the prediction position.
\item[{$R_{\rm parallel},R_{\rm prefix},E_{\rm parallel},E_{\rm prefix},I_P(Y;Z\mid X)$}]
Target risks, excess risks relative to the corresponding true conditional
laws, and conditional information carried by the extra suffix $Z$ about
target $Y$ given prefix $X$.
\item[{$P,Q_\theta,\Delta_t,s_a(x),a_*(x),m(x),e(x)$}]
True continuation law, proper autoregressive model law, excess conditional
log loss, candidate answer score, correct answer, signed task margin and
uniform candidate-score error bound.
\item[{$u_0,u_1,b,\Delta,\mathbf G,V_*,\Lambda$}]
Paired answer-score differences, their common bias and premise contrast,
collection of operators used by the task scores, its squared error from
a fixed collection, and score sensitivity to that collection.
\item[{$\Pi,\bar\mu_V,\mathcal O_\Pi$}]
Coupling of paired tensor observations, mean reference entry and normalized
paired operator discrepancy. This $\Pi$ is a probability coupling, not
a row or vocabulary projector; the discrepancy requires
$\bar\mu_V\ne0$.
\item[{$\rho,c,|c|,f_c,F_T(\rho),\widehat y_\rho$}]
Source-selection probability, binary source word, its number of selected
technical blocks, conditional path expectation, resulting finite source
polynomial and the three-knot quadratic interpolant.
\item[{$D_N,D_T,p_0,\nu,\langle\cdot,\cdot\rangle_{p_0,\nu}$}]
Narrative and technical logit contrasts against the general-source arm,
incoming probabilities, fixed input law and averaged categorical Fisher
inner product. Here subscripts $N,T$ denote source arms, not width or time.
\item[{$D_0,D_t,G_0,G_t,C_{0t},M,E,P_D$}]
Incoming and current source dictionaries, their Gram matrices, cross Gram,
fitted coefficient transport, representation residual and orthogonal
projection onto a fixed dictionary's range. Coefficient error and the
component outside that range have separate squared-norm budgets.
\item[{$A_N,X_N,Y_N,e_N,\lambda_{i,N},\epsilon_N$}]
Sector-to-observation map, incoming and observed fluctuations, coupled
residual, ordered incoming covariance eigenvalues and residual $L^2$ norm.
Uniform singular-value bounds and sufficiently small residuals are the
stated hypotheses for inheriting several fluctuation sectors.
\end{description}

\section{Conditional moments, calibration and numerical observation}
\label{sec:notation-moments}
See Chapters~\ref{ch:conditional-moments}--\ref{ch:forecast-evidence}.
\begin{description}[style=nextline,leftmargin=1.5em,font=\normalfont,itemsep=3pt]
\item[{$V_j,D_j,D,\mu,C,B,\mathcal M_B$}]
Path observations, successive increments, stacked increment vector,
its conditional mean and covariance, linear observation map and moment
map $\mathcal M_B(\mu,C)=(B\mu,BCB^{\mathsf T})$.
\item[{$S(\mu,C;x),\mathcal L,\mu_*,C_*$}]
Dawid--Sebastiani score, its excess expectation, and true mean and positive
definite covariance. The score is $\log\det C+(x-\mu)^{\mathsf T}C^{-1}(x-\mu)$;
Gaussianity is not required for its moment interpretation.
\item[{$C_{\rm path},C_{\rm init},C_{\rm corpus},D_s,R_s$}]
Conditional continuation, initialization and corpus covariance sectors;
state-specific diagonal standard-deviation matrix and correlation matrix
in $C_s=D_sR_sD_s$. The nested conditioning determines the decomposition.
\item[{$T(x),m,a_j,q,\mathcal T_s,n$}]
Maximum standardized coordinate score, adaptation-fitted center and scales,
maximum calibration score, calibrated path set and calibration sample size.
The chapter's coverage statement conditions on adaptation and requires
its specified exchangeability.
\item[{$\mu_0,C_0,\mu_1,C_1,\epsilon,\delta,\rho$}]
Parent and successor conditional moments and the relative covariance,
mean and coupled state-change error bounds used in forecast reuse.
These local error coordinates do not denote source-mixture probabilities.
\item[{$R,A,r,k,\ell,I,J,H,d_H,W,c,d_0$}]
Remaining resource and removed subset, their sizes, sequence length,
coupled source sequences, their Hamming distance and metric, observation
whitening map, sequence sensitivity and program-change bound.
\item[{$A,c_r,c_d,Q,m$}]
Acquisition cost, reduced and direct per-query costs, query count and
number of acquired paths. The cost comparison is made at matched
prediction quality. Here $A$ is a scalar cost, not a metric matrix.
\item[{$\widehat F,F,\epsilon_T,\widehat T_t,T_t,\eta_t,L_t$}]
Numerical and idealized observation maps, accumulated observation error,
numerical and real-arithmetic updates, local execution defect and
propagation Lipschitz constant. The measured finite pulse remains defined
even where a differentiable approximation is not admitted.
\end{description}

\section{Renormalized laws and conditional scaling fields}
\label{sec:notation-criticality}
See Chapters~\ref{ch:conditional-criticality}--\ref{ch:joint-clocks}.
\begin{description}[style=nextline,leftmargin=1.5em,font=\normalfont,itemsep=3pt]
\item[{$\kappa_n,W_s,\beta_n^{\rm disc},H,C$}]
Connected cumulant tensor, analytically rescaled cumulant-generating
function, discrete logarithmic-scale increment, field-normalization
exponent and retained linear observation. Here $C$ acts on a field;
it is not necessarily a covariance matrix.
\item[{$\Gamma(y),Q,\Sigma$}]
Gaussian effective action, precision $Q=\Sigma^{-1}$ and covariance on
its nondegenerate retained space. A singular covariance is first restricted
to its range when an inverse is used.
\item[{$M_b,\chi_b,C(r),C_0,\Lambda_b,m_b$}]
Block mean, susceptibility matrix $b\Cov(M_b)$, lag covariance,
fine-scale covariance reference, largest generalized susceptibility and
its reciprocal mass. The mass is $+\infty$ at zero susceptibility.
\item[{$\gamma,M,B,T_a,\rho_a$}]
Long-range covariance-tail exponent and matrix amplitude, constant common
sector, geometric-pole amplitudes and pole locations. The tail
$Mr^{-\gamma}$ and the geometric representation are different covariance
classes with separately stated assumptions.
\item[{$q_N,g,j,f_N,g_c,\xi,\nu,\gamma_\chi,y_\chi$}]
Intensive field, training control, statistical conjugate source,
normalized log moment-generating functional, critical control,
correlation length, length exponent, susceptibility exponent and
susceptibility scaling dimension. A statistical source does not change
the native optimizer or the forward graph.
\item[{$\kappa,\phi,z,\omega,\mathcal F,u_i,\phi_i$}]
Count-susceptibility, count-window, dynamic and correction exponents,
conditional scaling function, relevant fields and their count-window
exponents. A physical length relation is required to convert count
exponents to length exponents. Here $\phi$ is an exponent, not the
Hellinger embedding or an observation map.
\item[{$U_{N,a},Q_N,V_N,C_N,\chi_N^{\rm q},f_N^{\rm q},\overline f_N^{\rm q}$}]
Head observation, intensive head mean, one-head and distinct-head
covariances, conditional susceptibility, conditional thermodynamic
functional and its context average. The superscript $\rm q$ records the
specified conditioning; the context average is outside the logarithm.
\item[{$H_{\ell a},R_{\ell a},\epsilon_R,A_H,A_R$}]
Allowed-key normalized attention entropy, floored row-energy fraction,
its denominator floor, and the reported entropy and row susceptibility
enhancements relative to their one-head references. For one allowed key,
the normalized entropy is defined to be zero.
\item[{$n_i,n_o,r_i,r_o,c,c_{\rm in},c_{\rm out},c_{\rm voc}$}]
Actual and reference fan dimensions, shape-aware initialization
multiplier, and the induced variance constants for the input, output
and vocabulary projections in the initialization kernel calculation.
\item[{$K_\ell,\mathcal A_\ell,\mathcal L_\epsilon,\mathcal S,\Pi_\ell$}]
Residual covariance kernel, native attention kernel transform,
LayerNorm kernel transform, SiLU covariance transform and causal
attention matrix, including its context blocks. These are initialization
objects, not a closed trained-state kernel.
\item[{$v_{N,a},Z_N,C_N,H_N,W,\varphi_N$}]
Vectorized signed head operator, normalized head sum, invariant empirical
second moment, sign-invariant observation, independent standard Gaussian
and sign-orbit characteristic function. This $C_N$ is distinct from the
distinct-head covariance used in the exchangeable-head formula.
\item[{$K_t,M_t,L_t,\bar c,u_{1,t},u_{2,t},P_{\rm pair}$}]
Parameter, power-base and log-base envelopes, uniform Adam-force bound,
continuous bias-clock coordinates $\beta_1^t,\beta_2^t$, and paired
training kernel driven by the same batch.
\item[{$Q_N^{(a)},J_N,A,B,F,Q,F_b,Q_b$}]
Independent-initialization replacement observation, its influence bound,
individual and shared response matrices, collective response $F=A+B$,
innovation covariance and their temporally blocked counterparts.
\item[{$J_p,\delta,\Delta,V_c,s_c,O_\kappa$}]
Softmax Jacobian, probability mass outside the most likely key,
score gap, centered values and scores, and output at score gain $\kappa$.
This gap is a finite attention margin, not a collective relaxation gap.
\item[{$s(\omega),s_0,\alpha,\delta_N,\zeta,\upsilon$}]
Forcing spectral density at angular frequency $\omega$, its low-frequency
amplitude and exponent, relaxation gap, gap exponent and forcing-amplitude
exponent. The spectral convention includes the factor $1/(2\pi)$;
the resulting scalar susceptibility power is
$\zeta(1-\alpha)-\upsilon$ under the stated hypotheses.
\item[{$\eta_{\rm gen},\eta_{\rm body},\eta_0,s_k,W,T,\alpha$}]
Generator and body rates, reference rate, schedule multiplier, warmup and
cosine-end counters and floor fraction. A superscript $\rm pk$ denotes
a peak rate. Accumulated drift clocks sum rates; the nominal noise clock
sums squared rates.
\item[{$b_N(s),Q_N(s),\varepsilon_{t+1},Y_a,Y_b,Y_{ab},Q_{ij}$}]
Conditional native parameter drift and covariance, centered innovation,
generator, body and mixed finite observation increments, and their
conditional cross covariances. Their sum gives the complete finite-step
covariance, including mixed contributions.
\item[{$L,v_i,P,a_i,e_{\rm total},e_{\rm noise}$}]
Full Fisher-image map, orthonormal emitted directions, their projector,
pulled-back source coordinates $a_i=L^{\mathsf T}v_i$, and residual
fractions for total second moment and centered noise. Here $L$ is a
linear map, not the layer count.
\item[{$\widehat\chi_{\rm rad},\widehat\chi_{\rm dir},r_i,u_i,\Delta$}]
Radial and directional parts of common-field susceptibility, sample
radii and unit directions, and the separately declared field-scaling
dimension in a nonlinear observation $q_N=N^{-\Delta}X_N$.
\item[{$r_{N,t},a_N,b_N,d_N,\sigma_N,\rho_N,w_N$}]
Distance to an independently identified critical surface, restoring
coefficient, deterministic bias in the selection recursion, bias in the
feedback recursion, forcing amplitude, forcing-correlation multiplier
and critical-window width. Selection requires the transported mean and
fluctuation errors to be small relative to that window.
\end{description}

\section{Metric-row transport and observation accuracy}
\label{sec:notation-metric}
See Chapters~\ref{ch:metric-reduction}--\ref{ch:model-wide-evidence}.
\begin{description}[style=nextline,leftmargin=1.5em,font=\normalfont,itemsep=3pt]
\item[{$g,G,\widetilde g,c,C_g,\delta_c,h$}]
Raw gradient, its Euclidean norm, clipped gradient, clipping multiplier,
clipping threshold, clipping denominator offset and loss-Hessian action
in the augmented Adam tangent. This scalar $G$ is distinct from the PLGA
operator; the local direction $h$ is distinct from head count.
\item[{$v_\theta,v_m,v_v,J_j,b_N,\epsilon_N$}]
Parameter and moment components of an augmented tangent, native update
Jacobian, coordinatewise clipped-gradient envelope and size-dependent
Adam offset in the width-limit comparison.
\item[{$\rho\otimes\alpha$}]
Counterfactual product law of recipient and donor components in a
crossed shared-state intervention. It differs from their factual joint
training law, which retains component dependence.
\item[{$Z,\mu,C^{\rm row},Q_A,Q_\perp,Q_\mu,I_\mu$}]
Centered row matrix, centroid, row covariance, normalized full,
transverse and common head-mean fields, and the isometric broadcast of
the common row field. Row covariance uses divisor $d$; it describes a
finite row cloud, not independent training replicas.
\item[{$J,r_i,b,e_i,B,C_e,M$}]
Row-map derivative at the centroid, Taylor residual, residual mean,
centered residual, transported-row/residual cross covariance,
residual covariance and a bilinear Hessian bound. Both $B$ and
$B^{\mathsf T}$ are retained in the exact covariance transport.
\item[{$\vartheta,\lambda_i,\lambda_i',J_i,B,\Gamma$}]
Shared row-map parameters, incoming and outgoing loss adjoints,
row-input Jacobian, stacked parameter Jacobian and full shared-coordinate
gradient $\Gamma=B^{\mathsf T}\lambda'$. The shared parameter gradient
requires the downstream adjoints as well as row moments.
\item[{$P_I,\delta_\Psi(\mu),\chi,\widehat\chi,\chi_\Delta$}]
Retained-row/complement-centroid projection, constant-input PLGA row
defect, exact and observed sample susceptibilities, and susceptibility
of their paired observation error. A common divisor and common units
are required for the finite comparison.
\item[{$S_{ij},B_{ij},X_{ij},V,Q,H,\odot$}]
Normalized sum of squared matrix increments, squared net displacement
and signed cross-time energy; unnormalized net increment, diagonal
energy and cross energy; and their compatible merge operation in
Section~\ref{model:sec:temporal-energy}. This use of $\odot$ denotes the
explicit block-summary product, rather than entrywise multiplication.
\item[{$\widehat e,e,\delta,\epsilon_N,a_N,r_N$}]
Observed and exact projected matrix energies, matrix observation-error
bound, coupled reduction error, emission coefficient and retained
scalar collective in the corresponding energy and visibility bounds.
These are local error-budget variables, with the norm and ensemble
specified in each result.
\end{description}

\section{Physical calibration and magnetic readouts}
\label{sec:notation-physical}
See Chapter~\ref{ch:physical-calibration}.
\begin{description}[style=nextline,leftmargin=1.5em,font=\normalfont,itemsep=3pt]
\item[{$q,L,V,s_i,H(s),T,Z_{q,L,T},P_{q,L,T}$}]
Number of spin colors, square-lattice side, site count $V=L^2$, spin
color, Potts Hamiltonian, temperature, partition function and Gibbs law.
The bond sum counts each right and down nearest-neighbor bond once.
These uses of $L,V,H,T$ differ from decoder depth, vocabulary size,
field-normalization exponent and training horizon.
\item[{$T_c(q),\mathcal D_{\rm phys},Q_{N,t;q,L,T}$}]
Critical temperature, declared physical-data law, and generated
configuration law at model width $N$ and adaptation age $t$.
Each conditional prediction uses a proper prefix and the fixed spin alphabet.
\item[{$f_a,m,v,\Delta_q,\chi_L,B_L$}]
Color fractions, order-parameter magnitude, centered normalized color
vector, probability simplex, susceptibility $L^2\E m^2$ and Binder
moment ratio $\E m^4/(\E m^2)^2$. The ratio has no additive constant or
reversed sign. Connected susceptibility separately subtracts the mean vector.
\item[{$d_j,\bar d,T_h,D_\pm,x,\beta,\nu$}]
True-law conditional prefix KL, its site average, finite thermal contrast,
endpoint Binder-ratio error bounds, magnetic scaling dimension
$x=\beta/\nu$, magnetization exponent and correlation-length exponent.
These $\beta,\nu$ are physical critical exponents, not Adam factors or
calibration distributions.
\item[{$a_{ha},D_h,d_h,C_{ha},t_h,\phi,\widehat f,\widehat v$}]
Color-query vector, first-layer query Gram, metadata contribution to its
trace, squared query coefficient, metadata-corrected normalized trace,
retained native observation, fitted color fractions and reconstructed
magnetic vector.
\item[{$\mu_L,e_L,\delta_L,r_L,\kappa_L,\delta_{c,L},\Delta_L$}]
Reference magnetic second moment, vector reconstruction error, its RMS,
relative amplitude error, connected reference variance, centered error
RMS and total variation between reference and generated configuration laws.
Second-moment and connected-variance budgets use their corresponding scales.
\item[{$B,S,k,\delta,e,c,R,\lambda$}]
Physical blocking kernel, represented scale map, Lipschitz constant,
law, readout and compatibility defects, fine scale map and retained
RG eigenvalue. These are the objects of the physical transport and
intertwining statements, with spaces declared there.
\item[{$U,\sigma_L,(CU)^+,\Pi_{\Delta_q},\varepsilon$}]
Orthonormal basis for the zero-sum color space, smallest singular value
of the restricted query coefficient matrix, its left pseudoinverse,
Euclidean simplex projection and trace-measurement error in the
constructive proper-prefix readout.
\item[{$p,\omega,a,c$ in physical fits}]
Fitted logarithmic slope, finite-size correction exponent, intercept
and correction amplitude. In the Binder thermal-slope fit the estimated
correlation-length exponent is $\nu=1/p$.
\end{description}

\section{Statistical, experimental and presentation conventions}
\label{sec:notation-experiments}
Appendices~\ref{app:statistics}--\ref{app:reproducibility} specify the
statistical units and methods used by the reported studies.
\begin{description}[style=nextline,leftmargin=1.5em,font=\normalfont,itemsep=3pt]
\item[{$\widehat\mu,\widehat C,s,S,m,n$}]
Empirical mean and covariance, and locally declared sample counts.
Across-replica sample covariance uses divisor one less than the number
of replicas. Finite-population covariance and row-cloud covariance use
their stated population divisors. Heads, contexts, repeated measurements
and adaptation replicas are not interchangeable independent units.
\item[{$\chi_R,\chi_H,\chi_G,\chi_\phi,V_R,V_H$}]
Susceptibilities of row, entropy, signed-operator and Hellinger fields,
and microscopic row and entropy variance references. A subscript or
superscript specifying a state, context, decoder, width, conditioning
law or averaging convention is part of the observable.
\item[{NLL, KL, RMS, RMSE, SE, SD, CI}]
Negative log likelihood, Kullback--Leibler divergence, root mean square,
root mean squared error, standard error, standard deviation and confidence
interval. Logarithmic prediction losses use natural logarithms and nats
unless explicitly normalized. A percentage includes its factor of 100.
\item[{$\Delta_H,\widehat\psi$}]
Relative discrepancy between the paired entropy secants at the two
reported amplitudes, and fitted rate--time clock exponent. Finite
contrasts and fitted powers
retain their declared amplitudes, windows and units; they are not
automatically asymptotic critical exponents.
\item[{Native, $+h$, $-h$, $h/2$; paired and crossed arms}]
Unperturbed and signed finite interventions, the smaller comparison
radius, shared-source comparisons and factorial component substitutions.
Their common incoming state, source indices and assessment panels are
specified by each experiment.
\item[{float32, float64, bfloat16; dtype}]
Numerical floating-point formats and data type. The stored dtype,
reduction order, casts and arithmetic policy are part of an executed
observation. A higher-precision observation need not rerun training.
\item[{Intervals, error bars and fitted curves}]
A table range is the stated observed range unless a confidence or
prediction interval is identified. Bootstrap resampling respects the
specified independent unit and pairing. A fitted curve, an empirical
secant and a uniform analytical bound have different meanings.
\item[{$\mathrm{tr},\mathrm{init},\mathrm{doc},\mathrm{path},\mathrm{corpus}$}]
Training-state, initialization, document, continuation and corpus
averaging or covariance sectors. Conditioning removes only the declared
randomness. Reusing a state, document or source order preserves the
corresponding dependence across comparisons.
\item[{Single pass, proper prefix, frozen inference and refill}]
A single pass consumes distinct indexed source blocks or target positions
under the declared rule. A proper prefix excludes its next target.
Frozen inference holds weights fixed while evaluating new inputs.
Repeated-corpus adaptation includes the epoch and refill mechanism in
its complete state; it is a different resource law.
\item[{Inequalities and tolerances}]
Reported finite precision and tolerances qualify the stated numerical
comparison. Strict, non-strict, entrywise and matrix inequalities retain
the distinctions in Section~\ref{sec:notation-algebra}.
\end{description}

\FloatBarrier

\chapter{Lists of tables and figures}
\label{app:lists}
This appendix lists the numbered tables and figures in their order of
appearance. Each entry retains its caption and links to the corresponding
display, allowing the theoretical diagrams and experimental results to be
located directly.

\section{List of tables}
\label{app:tables}
\appendixcaptionlist{lot}

\clearpage
\section{List of figures}
\label{app:figures}
\appendixcaptionlist{lof}

\FloatBarrier

\backmatter
\part*{Back matter}
\chapter*{Acknowledgements}
I am grateful to my parents for their support and patience. This research was
conducted independently without support from a grant or corporation.

\chapter*{Disclosure of the use of AI tools}
The author used (Codex, model GPT-6-Astra, OpenAI) and (Codex, model
GPT-5.6-Sol, OpenAI) in preparing most chapters of this monograph. Their
assistance included organizing the material, drafting and editing exposition
and proof arguments, developing and checking selected Lean formalizations
and supporting code, analyzing experimental results, and checking
mathematical consistency, references, and presentation. (Claude, model
Claude Fable 5, Anthropic) was used in part of the work on the earlier
chapters.

\bibliographystyle{plain}
\bibliography{references}
\end{document}